\def\paperID{5894}
\def\confName{CVPR}
\def\confYear{2024}

\def\paperTitle{DiffusionLight: Light Probes for Free by Painting a Chrome Ball}

\def\authorBlock{
    Pakkapon Phongthawee\footnotemark[1] \hspace{1pt}\vistec \qquad\quad
    Worameth Chinchuthakun\footnotemark[1] \hspace{1pt}\vistectokyo \qquad\quad
    Nontaphat Sinsunthithet\hspace{0.8pt}\vistec \\
    Amit Raj\google \qquad\quad
    Varun Jampani\stabilityai \qquad\quad
    Pramook Khungurn\pixiv \qquad\quad
    Supasorn Suwajanakorn\vistec \\
    \vspace{-0.65em}\\
    \vistec \hspace{-0.4em} VISTEC \qquad
    \tokyotech \hspace{-0.4em} Tokyo Tech \qquad
    \google \hspace{-0.4em} Google Research\qquad
    \stabilityai \hspace{-0.4em} Stability AI \qquad
    \pixiv \hspace{-0.4em} Pixiv \\ 
    {\small \url{https://diffusionlight.github.io/} }
    
}

\newif\ifreview 
\newif\ifarxiv \newcommand{\arxiv}{\arxivtrue}
\newif\ifcamera 
\newif\ifrebuttal 

\newif\ifpurecompactformat 

\newif\ifwithackowledgement

\arxiv
\let\mypdfximage\pdfximage
\def\pdfximage{\immediate\mypdfximage}
\pdfoutput=1
\documentclass[10pt,twocolumn,letterpaper]{article}
\usepackage[accsupp]{axessibility}

\ifreview \usepackage[review]{cvpr} \fi
\ifarxiv \usepackage[pagenumbers]{cvpr} \fi
\ifrebuttal \usepackage[rebuttal]{cvpr} \fi
\ifcamera \usepackage{cvpr} \fi

\usepackage{graphicx}	
\usepackage{amsmath}	
\usepackage{amssymb}	
\usepackage{booktabs}
\usepackage{times}
\usepackage{microtype}
\usepackage{epsfig}
\usepackage[table,xcdraw,dvipsnames]{xcolor}
\usepackage{caption}
\usepackage{float}
\usepackage{placeins}
\usepackage{color, colortbl}
\usepackage{stfloats}
\usepackage{enumitem}
\usepackage{tabularx}
\usepackage{xstring}
\usepackage{multirow}
\usepackage{xspace}
\usepackage{url}
\usepackage{subcaption}
\usepackage{xcolor}
\usepackage{gensymb}
\usepackage[hang,flushmargin]{footmisc}
\usepackage{tabu}
\usepackage[ruled,vlined]{algorithm2e}
\usepackage{float}
\usepackage{algpseudocode}
\usepackage{bbm}
\usepackage{colortbl}
\usepackage{color,soul}

\ifcamera \usepackage[accsupp]{axessibility} \fi

% ...

\ifarxiv  \fi

\newcommand{\todo}[1]{{\textcolor{red}{[TODO: #1]}}}

\newcommand{\pure}[1]{{\textcolor{orange}{#1}}}

\definecolor{tabfirst}{rgb}{0.96, 0.77, 0.77} 
\definecolor{tabsecond}{rgb}{0.98 , 0.93, 0.77} 
\definecolor{tabthird}{rgb}{1, 1, 0.7} 

\newcommand{\myparagraph}[1]{\vspace{0.5em}\noindent\textbf{#1}}

\newcommand{\scfirst}[1]{\colorbox{tabfirst}{\makebox[1.5cm][c]{#1}}}
\newcommand{\scsecond}[1]{\colorbox{tabsecond}{\makebox[1.5cm][c]{#1}}}

\usepackage{mathtools}
\DeclarePairedDelimiter{\parens}{\lparen}{\rparen}

\newcommand{\vect}[1]{\ensuremath{\mathbf{#1}}}

\newcommand{\R}[1]{{%
    \textbf{%
        \ifstrequal{#1}{1}{\textcolor{red}{R#1}}{%
        \ifstrequal{#1}{2}{\textcolor{blue}{R#1}}{%
        \ifstrequal{#1}{3}{\textcolor{magenta}{R#1}}{%
        \ifstrequal{#1}{4}{\textcolor{teal}{R#1}}{%
                           \textcolor{cyan}{R#1}%
        }}}}%
    }%
}}

\newcommand{\vistec}{%
  $^1$
}
\newcommand{\vistectokyo}{%
  $^{1,2}$
}
\newcommand{\tokyotech}{%
  $^2$
}

\newcommand{\google}{%
  $^3$
}

\newcommand{\stabilityai}{%
  $^4$
}

\newcommand{\pixiv}{%
  $^5$
}

\usepackage{xr-hyper}

\makeatletter
\newcommand*{\addFileDependency}[1]{
  \typeout{(#1)}
  \@addtofilelist{#1}
  \IfFileExists{#1}{}{\typeout{No file #1.}}
}

\makeatother

\definecolor{cvprblue}{rgb}{0.21,0.49,0.74}
\usepackage[pagebackref,breaklinks,colorlinks,citecolor=cvprblue]{hyperref}
\usepackage[capitalize]{cleveref}
\crefname{section}{Sec.}{Secs.}
\crefname{table}{Table}{Tables}
\crefname{figure}{Fig.}{Figs.}

\begin{document}
\title{\paperTitle}
\author{\authorBlock}
\twocolumn[{%
\renewcommand\twocolumn[1][]{#1}%
\maketitle
\begin{center}
    \centering
    \captionsetup{type=figure}
    \vspace{-2em}
    \includegraphics[width=1.0\textwidth]{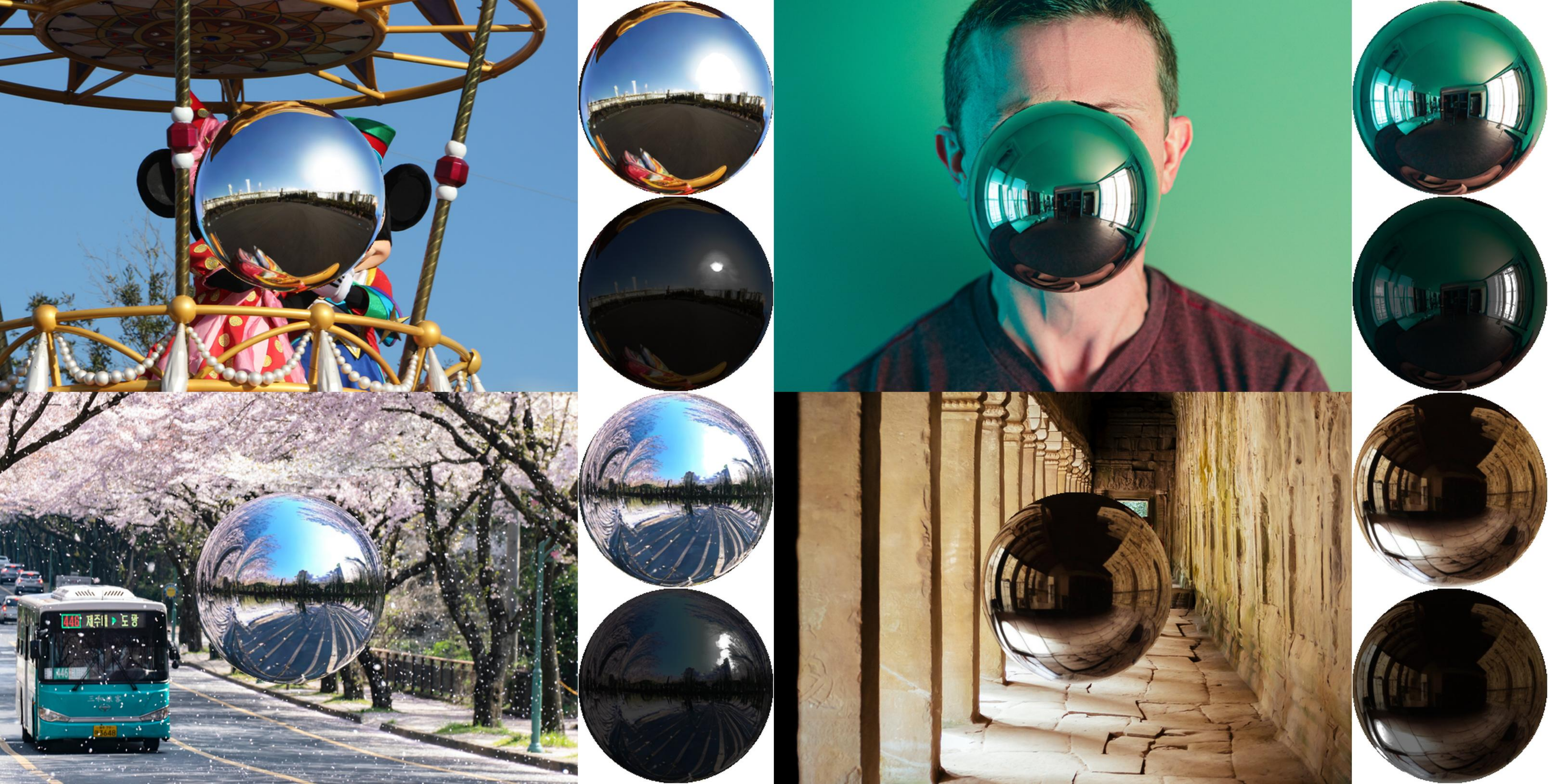}
    \vspace{-1.8em}
    \captionof{figure}{We leverage a pre-trained diffusion model (Stable Diffusion XL) for light estimation by rendering an HDR chrome ball.
    In each scene, we show our normally exposed chrome ball on top and our underexposed version, which reveals bright light sources, on the bottom.
    } \label{fig:teaser}
\end{center}%
}]

\begin{abstract}
\vspace{-0.8em}
\footnotetext[1]{Authors contributed equally to this work.}
We present a simple yet effective technique to estimate lighting in a single input image. Current techniques rely heavily on HDR panorama datasets to train neural networks to regress an input with limited field-of-view to a full environment map. However, these approaches often struggle with real-world, uncontrolled settings due to the limited diversity and size of their datasets. To address this problem, we leverage diffusion models trained on billions of standard images to render a chrome ball into the input image. Despite its simplicity, this task remains challenging: the diffusion models often insert incorrect or inconsistent objects and cannot readily generate chrome balls in HDR format. Our research uncovers a surprising relationship between the appearance of chrome balls and the initial diffusion noise map, which we utilize to consistently generate high-quality chrome balls. We further fine-tune an LDR diffusion model (Stable Diffusion XL) with LoRA, enabling it to perform exposure bracketing for HDR light estimation. Our method produces convincing light estimates across diverse settings and demonstrates superior generalization to in-the-wild scenarios.

\end{abstract}
\vspace{-0.4cm}
\section{Introduction}
\vspace{-0.2em}
\label{sec:intro}

Single-view lighting estimation is the problem of inferring the lighting conditions from an input image. In this work, we represent lighting as an environment map \cite{blinn1976environmentmap}, which facilitates seamless insertion of virtual objects, including highly reflective ones.
This problem is ill-posed because the environment map extends beyond the limited field of view of the input image. Moreover, the output must have a high dynamic range (HDR) to capture the true intensity of the incoming light. These difficulties have spurred numerous attempts to \emph{regress} HDR environment maps from LDR images.

A common strategy used in state-of-the-art techniques is to train a neural regressor with a dataset of HDR panoramas. For example, StyleLight \cite{wang2022stylelight} trains a GAN on thousands of panoramas and, at test time, uses GAN inversion to find a latent code that generates a full panorama whose cropped region matches the input image. Everlight \cite{dastjerdi2023everlight} trains a conditional GAN on 200k panoramas to directly predict an HDR map from an input image. However, these training panoramas offer limited scene variety, often featuring typical viewpoints like those from a room's center due to tripod use. Panoramas featuring elephants from inside a Safari Jeep, would be extremely rare (Figure \ref{fig:qualitative_wild}). 
So, how can we estimate lighting in-the-wild for any image under any scenario?

Our key idea is to use a pre-trained large-scale text-to-image (T2I) diffusion model to render a chrome ball into the input image.\footnote{Note that the practice of \emph{physically} placing a chrome ball into the scene dates back to the early days of computer graphics and photography \cite{devebechistory}.} 
One clear advantage of our approach is the ability to leverage the image prior in the T2I diffusion models. Our method also does not assume a known camera pose, unlike many methods that outpaint panoramas \cite{wang2022stylelight, dastjerdi2023everlight}.



Inpainting a chrome ball into an image, while seemingly simple, poses challenges even for SOTA diffusion models with inpainting capabilities \cite{avrahami2023blendedlatent, avrahami2022blendeddiffusion, yang2023paint, ye2023ip-adapter} (see Figure \ref{fig:inpaint_sota}). These models frequently fail to generate a chrome ball or generate one with undesirable patterns (e.g., a disco ball with tiny square mirrors) that do not convincingly reflect environmental lighting. Another key limitation is that these diffusion models were trained on low dynamic range (LDR) images and cannot produce HDR chrome balls.

To produce consistent, high-quality chrome balls, our solution involves three key ideas. First, we make inserting a ball reliable by using depth map conditioning \cite{zhang2023adding} on top of a standard T2I diffusion model (Stable Diffusion XL \cite{podell2023sdxl}). Second, to better mimic genuine chrome ball appearance, we fine-tune the model using LoRA \cite{hu2021lora} on a small number of synthetically generated chrome balls. Third, we start the diffusion sampling process from a good initial noise map, and we propose an algorithm to find one. The last idea is based on surprising findings we discovered about diffusion model behavior and chrome ball appearance.

To generate HDR chrome balls, our idea is to generate and combine multiple LDR chrome balls with varying exposure values, similar to exposure bracketing. One naive method that requires no additional training is to use two text prompts: one for generating a standard chrome ball, and the other with ``black dark'' added to the text prompt. Alternatively, we propose utilizing our previous LoRA \cite{hu2021lora} to map continuous interpolations of the two text prompts to a target ball image with varying, known exposures. This enables specifying the exposure values of the generated balls at test time. While this fine-tuning requires a small number of panoramas for training, the core task of producing reflective chrome balls still relies on the model initial's capability, which remains generalizable to a broad range of scenes.

We evaluate our method against StyleLight \cite{wang2022stylelight}, EverLight \cite{dastjerdi2023everlight}, Weber et al. \cite{weber2022editableindoor}, and EMLight \cite{zhan2021emlight} on standard benchmarks: Laval Indoor \cite{garder2017lavelindoor} and Poly Haven \cite{polyhaven} datasets.
Our method is competitive with StyleLight and achieves better performance on two out of three metrics across both datasets, while ranking second and third, when tested using a protocol in \cite{dastjerdi2023everlight} on Laval Indoor. Note that the baselines were directly trained on the datasets, with some tailored to indoor scenes. When applied to more challenging, in-the-wild images beyond the benchmarks, our method still produces convincing results while the baselines fail to do so.

To summarize, our contributions are:
\begin{itemize}
\item A novel light estimation technique that generalizes across a wide variety of scenes based on a simple idea of inpainting a chrome ball using a pre-trained diffusion model.
\item An iterative inpainting algorithm that enhances quality and consistency by leveraging our discovered relationship between the initial noise map and chrome ball appearances.
\item A continuous LoRA fine-tuning technique for exposure bracketing to produce HDR chrome balls.
\end{itemize}

\ifpurecompactformat
\vspace{-0.1cm}
\fi
\section{Related Work}
\label{sec:related}
\ifpurecompactformat
\vspace{-0.1cm}
\fi

\tabulinesep=0.1pt
\begin{figure}
    \centering

    \begin{tabu} to \textwidth {
        @{}
        c@{}
        c@{\hspace{0.5pt}}
        c@{\hspace{0.5pt}}
        c@{\hspace{0.5pt}}
        c@{\hspace{0.5pt}}
        c@{\hspace{0.5pt}}
        c@{}
    }
        
        \multicolumn{1}{l}{\rotatebox[origin=c]{90}{\shortstack[l]{\scriptsize Input \\ \scriptsize image}}} &
        \noindent\parbox[c]{0.083\textwidth}{\includegraphics[width=0.083\textwidth]{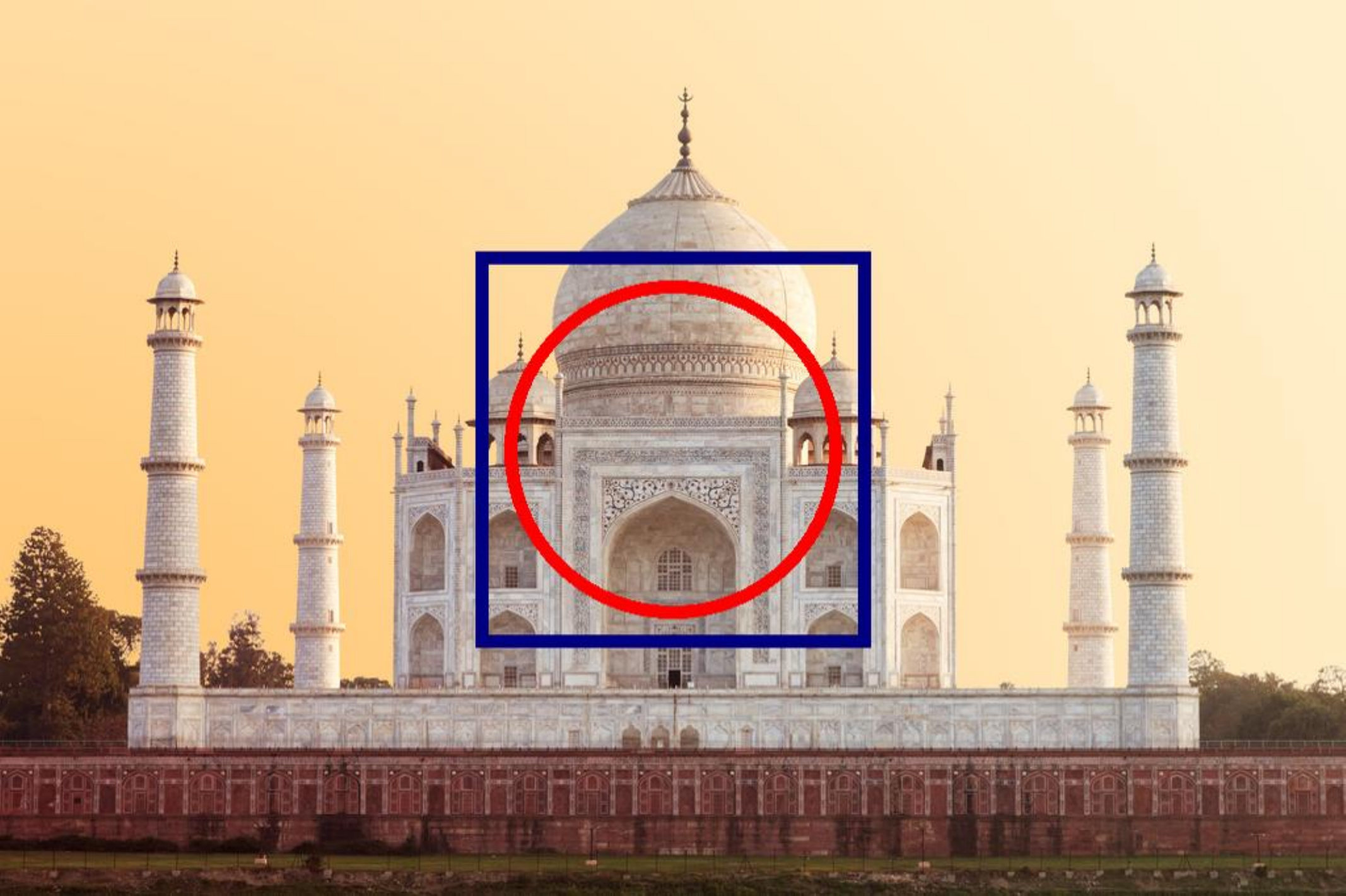}} & 
        \noindent\parbox[c]{0.083\textwidth}{\includegraphics[width=0.083\textwidth]{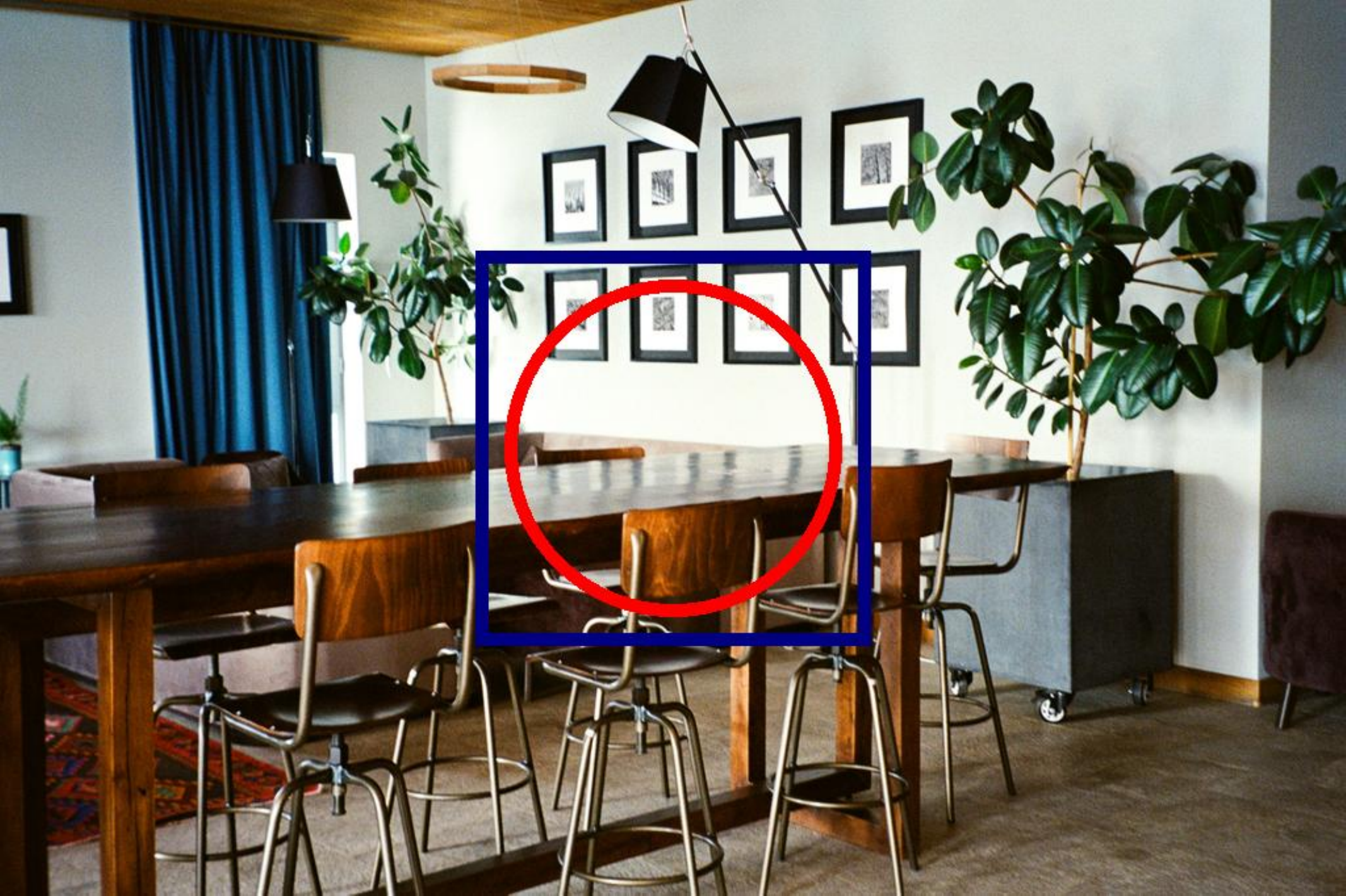}} & 
        \noindent\parbox[c]{0.083\textwidth}{\includegraphics[width=0.083\textwidth]{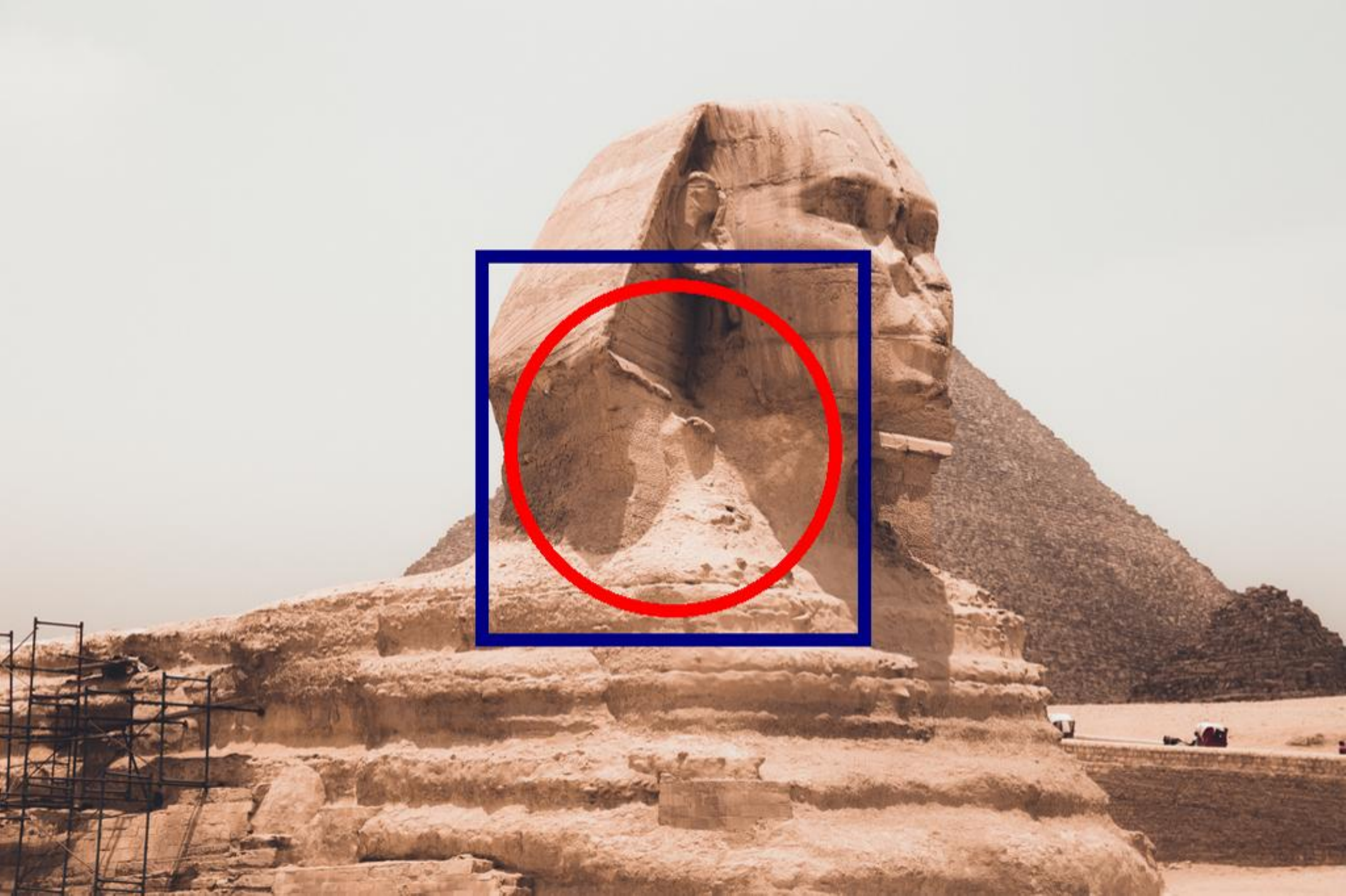}} & 
        \noindent\parbox[c]{0.083\textwidth}{\includegraphics[width=0.083\textwidth]{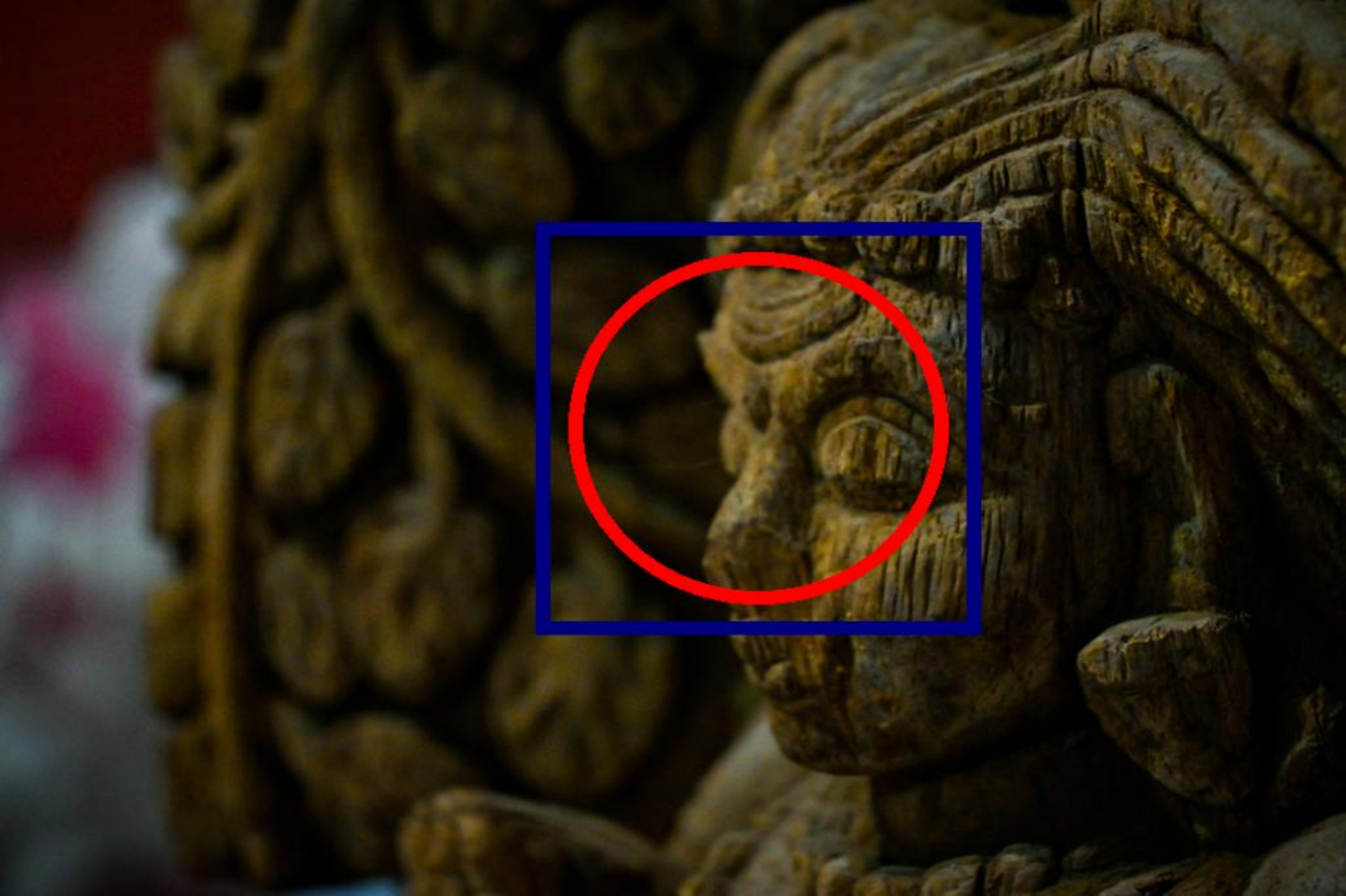}} & 
        \noindent\parbox[c]{0.083\textwidth}{\includegraphics[width=0.083\textwidth]{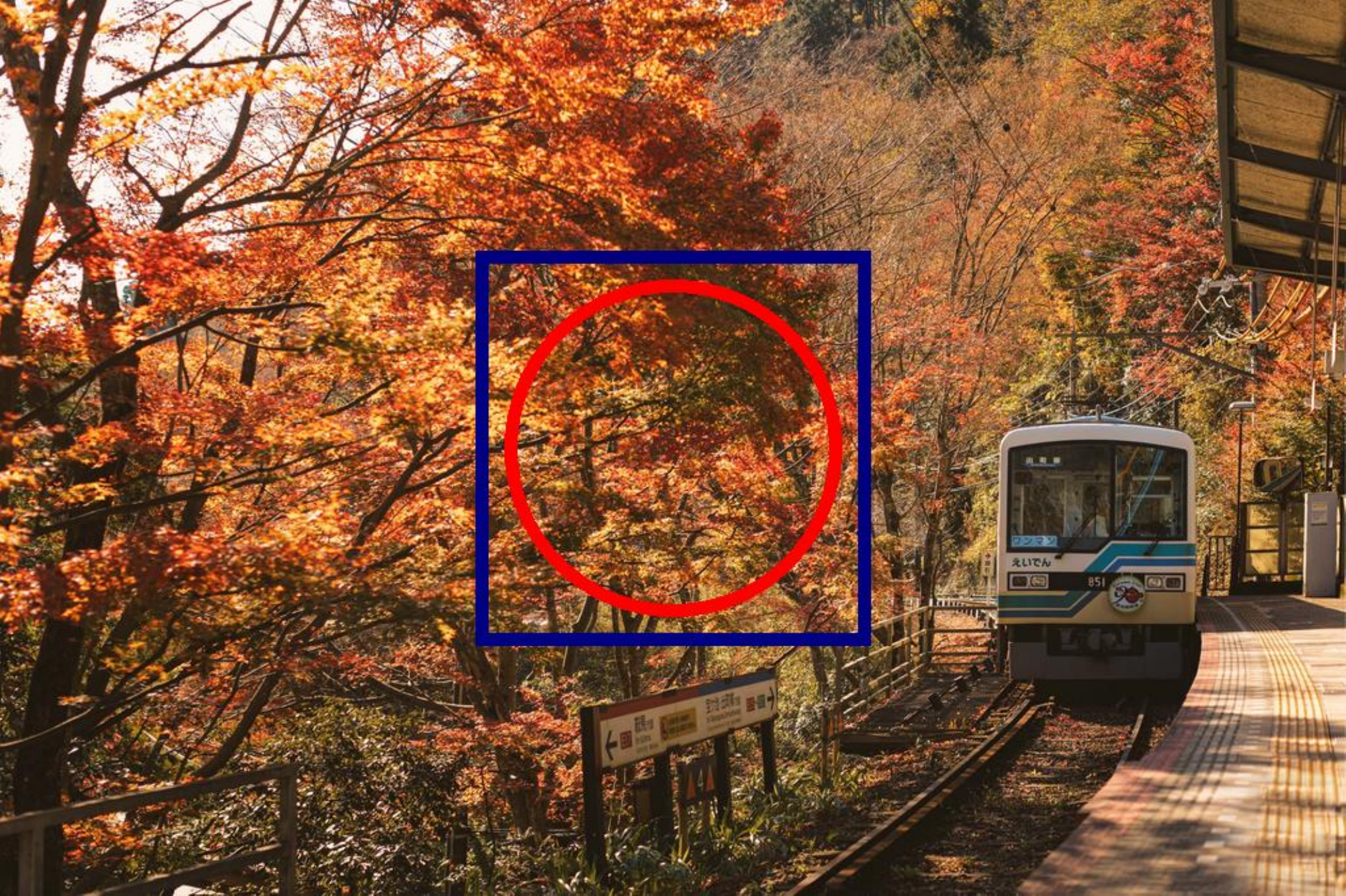}} &
        \\

        \multicolumn{1}{l}{\rotatebox[origin=c]{90}{\shortstack[l]{\scriptsize Blended Dif-\\ \scriptsize fusion \cite{avrahami2023blendedlatent, avrahami2022blendeddiffusion}}}} &
        \noindent\parbox[c]{0.083\textwidth}{\includegraphics[width=0.083\textwidth]{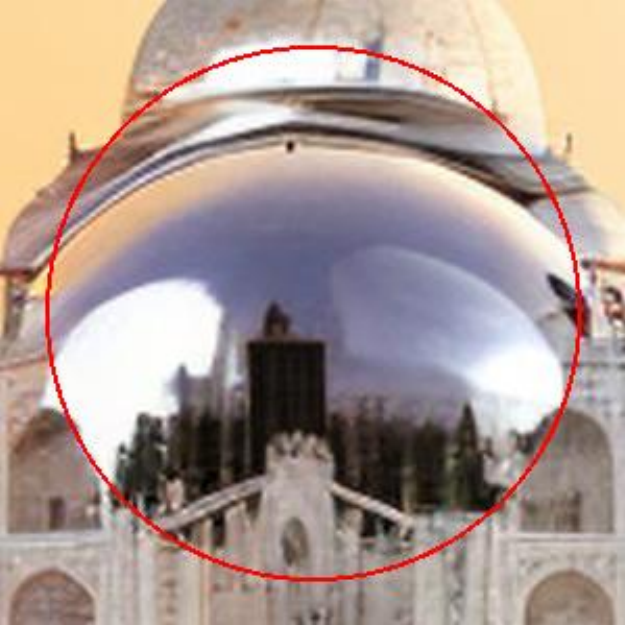}} & 
        \noindent\parbox[c]{0.083\textwidth}{\includegraphics[width=0.083\textwidth]{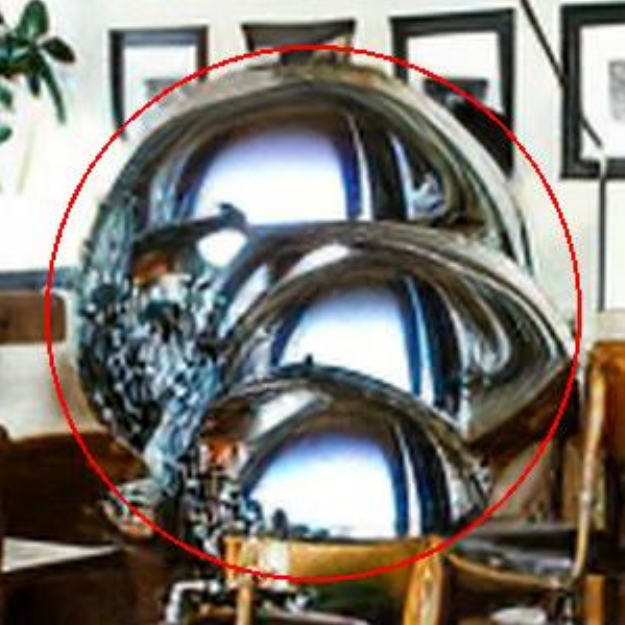}} & 
        \noindent\parbox[c]{0.083\textwidth}{\includegraphics[width=0.083\textwidth]{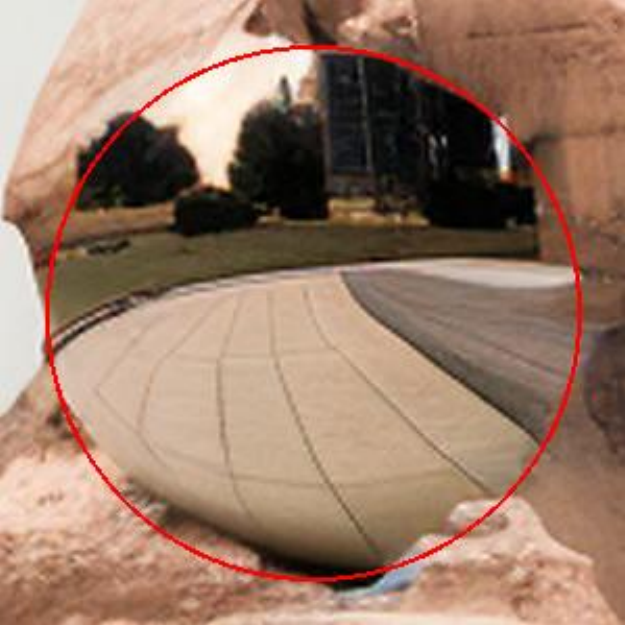}} & 
        \noindent\parbox[c]{0.083\textwidth}{\includegraphics[width=0.083\textwidth]{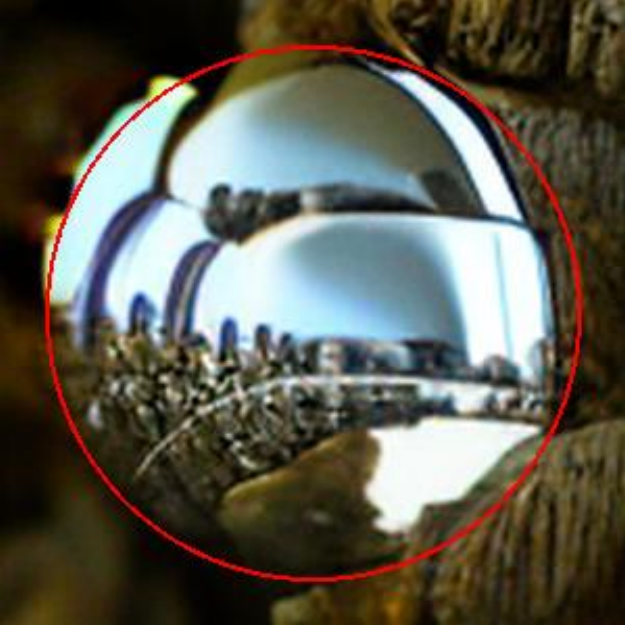}} & 
        \noindent\parbox[c]{0.083\textwidth}{\includegraphics[width=0.083\textwidth]{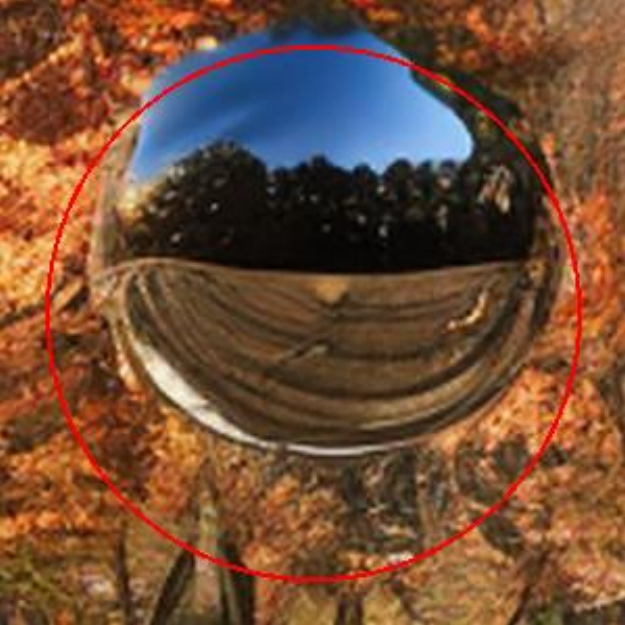}} &
        
        \\

        \multicolumn{1}{l}{\rotatebox[origin=c]{90}{\shortstack[l]{\scriptsize Paint-by-Ex\\ \scriptsize ample \cite{yang2023paint}}}} &
        \noindent\parbox[c]{0.083\textwidth}{\includegraphics[width=0.083\textwidth]{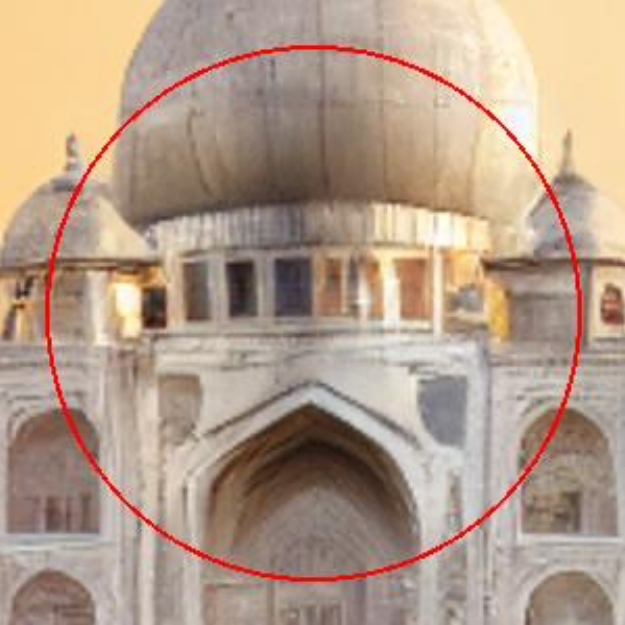}} & 
        \noindent\parbox[c]{0.083\textwidth}{\includegraphics[width=0.083\textwidth]{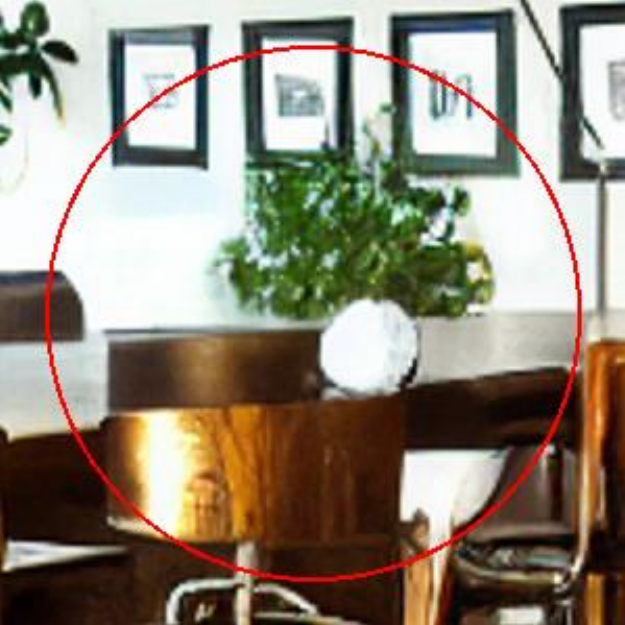}} & 
        \noindent\parbox[c]{0.083\textwidth}{\includegraphics[width=0.083\textwidth]{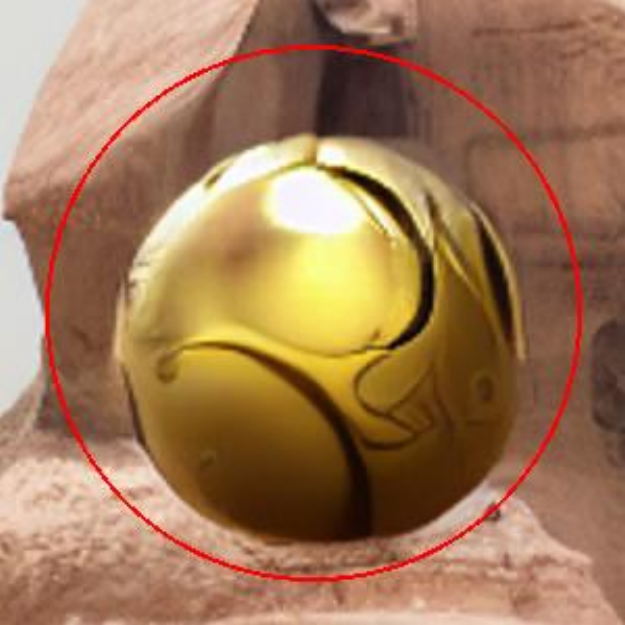}} & 
        \noindent\parbox[c]{0.083\textwidth}{\includegraphics[width=0.083\textwidth]{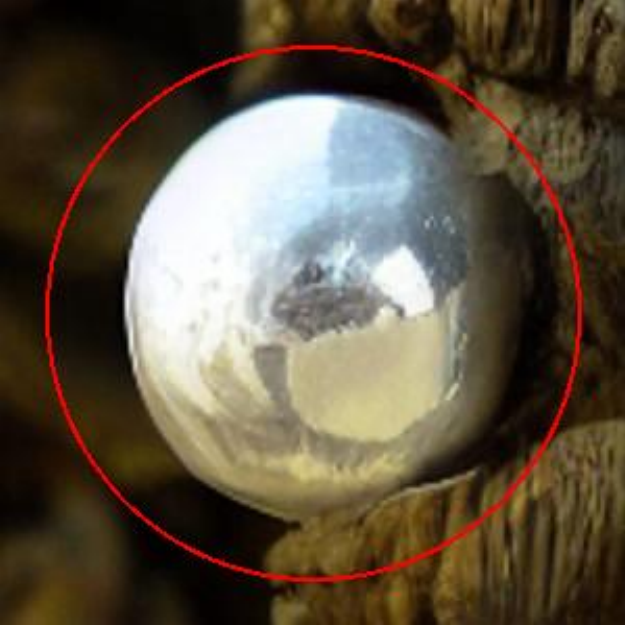}} & 
        \noindent\parbox[c]{0.083\textwidth}{\includegraphics[width=0.083\textwidth]{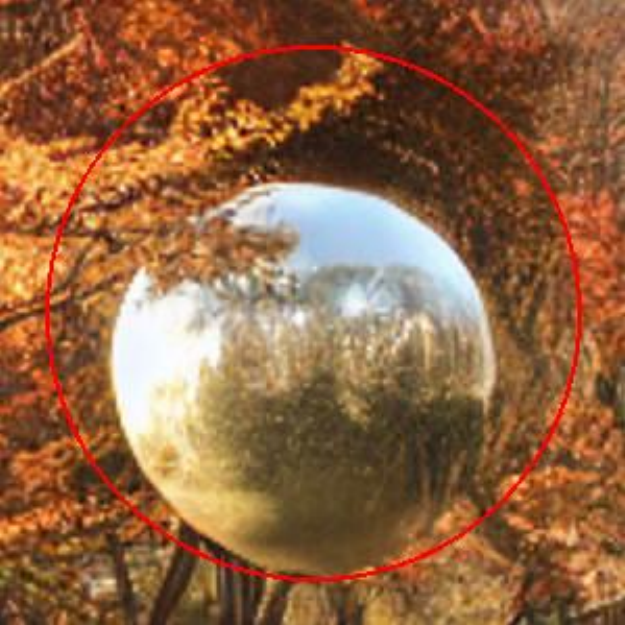}} &
        
        \\

        \multicolumn{1}{l}{\rotatebox[origin=c]{90}{\shortstack[l]{\scriptsize IP-Adapter\\ \scriptsize \cite{ye2023ip-adapter}}}} &
        \noindent\parbox[c]{0.083\textwidth}{\includegraphics[width=0.083\textwidth]{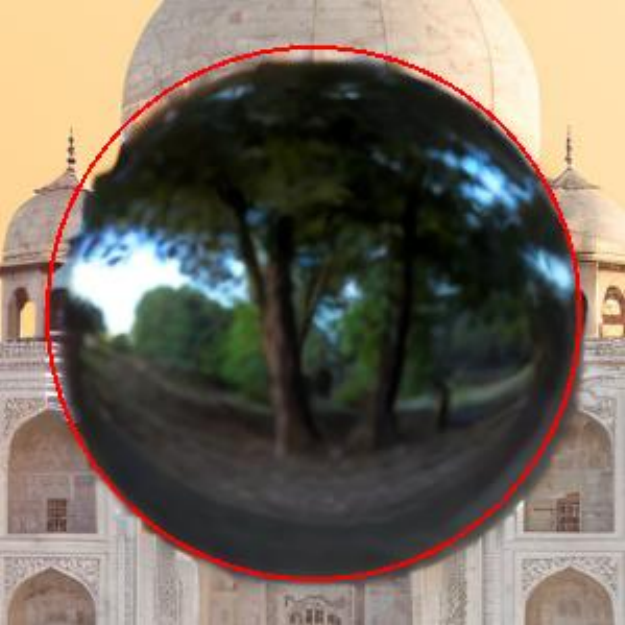}} & 
        \noindent\parbox[c]{0.083\textwidth}{\includegraphics[width=0.083\textwidth]{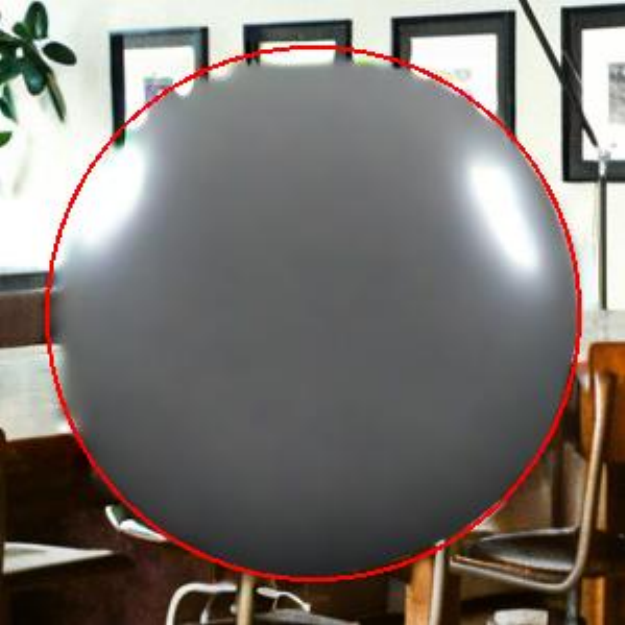}} & 
        \noindent\parbox[c]{0.083\textwidth}{\includegraphics[width=0.083\textwidth]{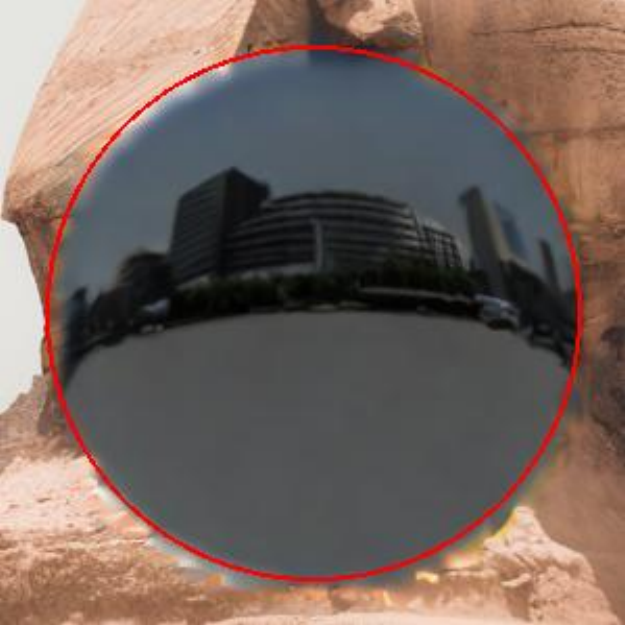}} & 
        \noindent\parbox[c]{0.083\textwidth}{\includegraphics[width=0.083\textwidth]{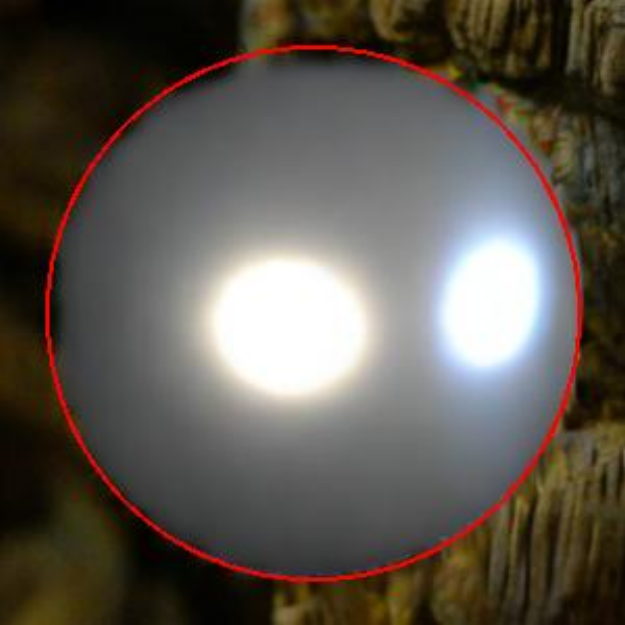}} & 
        \noindent\parbox[c]{0.083\textwidth}{\includegraphics[width=0.083\textwidth]{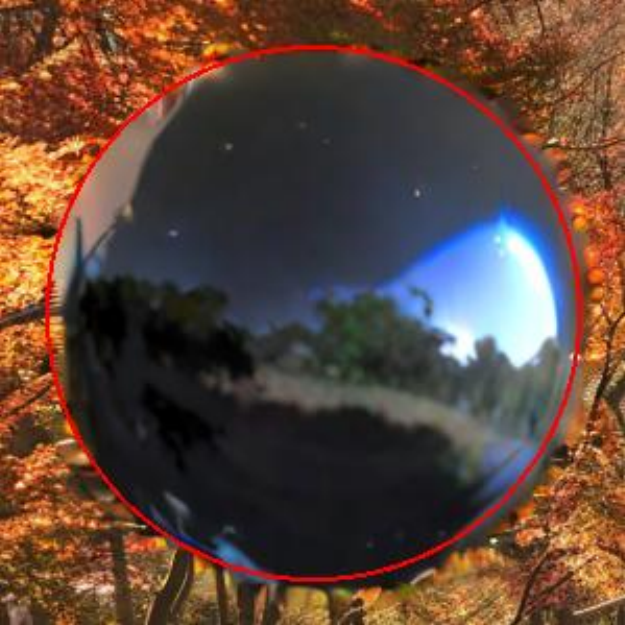}} &
        
        \\

        \multicolumn{1}{l}{\rotatebox[origin=c]{90}{\shortstack[l]{\scriptsize DALL·E2 \cite{dalle2}}}} &
        \noindent\parbox[c]{0.083\textwidth}{\includegraphics[width=0.083\textwidth]{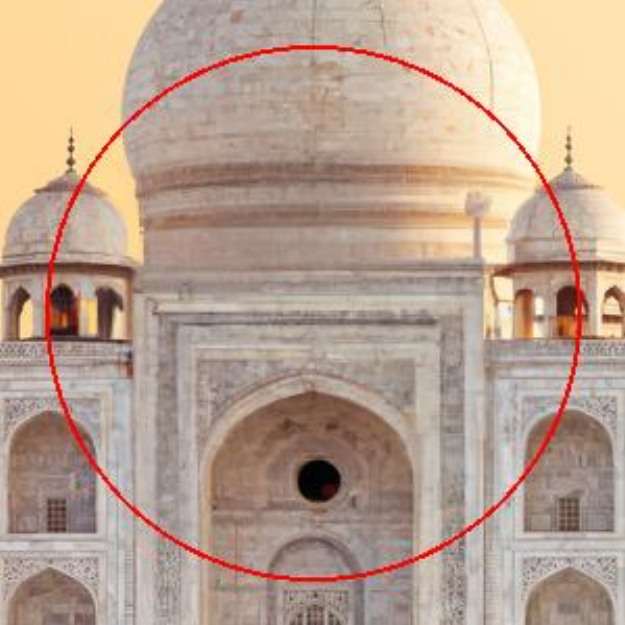}} & 
        \noindent\parbox[c]{0.083\textwidth}{\includegraphics[width=0.083\textwidth]{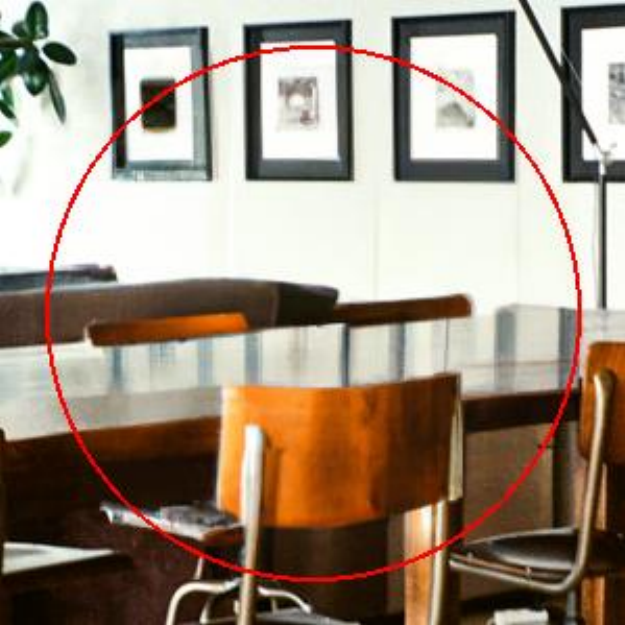}} & 
        \noindent\parbox[c]{0.083\textwidth}{\includegraphics[width=0.083\textwidth]{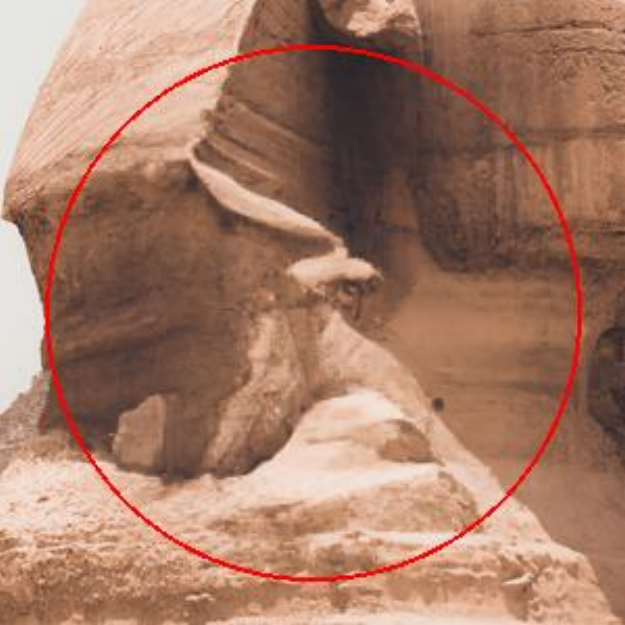}} & 
        \noindent\parbox[c]{0.083\textwidth}{\includegraphics[width=0.083\textwidth]{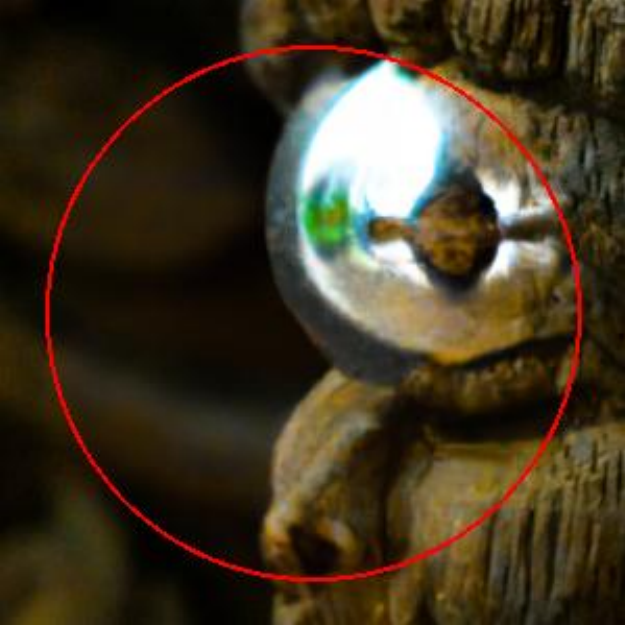}} & 
        \noindent\parbox[c]{0.083\textwidth}{\includegraphics[width=0.083\textwidth]{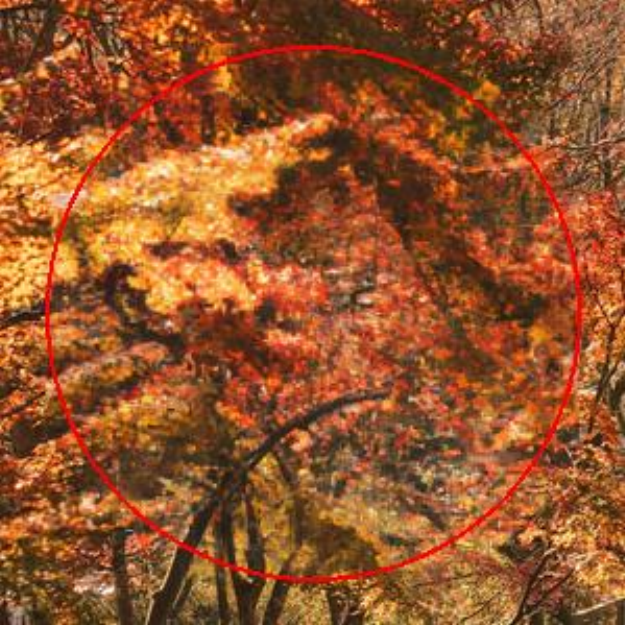}} &
        
        \\

        \multicolumn{1}{l}{\rotatebox[origin=c]{90}{\shortstack[l]{\scriptsize Adobe \\ \scriptsize Firefly \cite{adobefirefly}}}} &
        \noindent\parbox[c]{0.083\textwidth}{\includegraphics[width=0.083\textwidth]{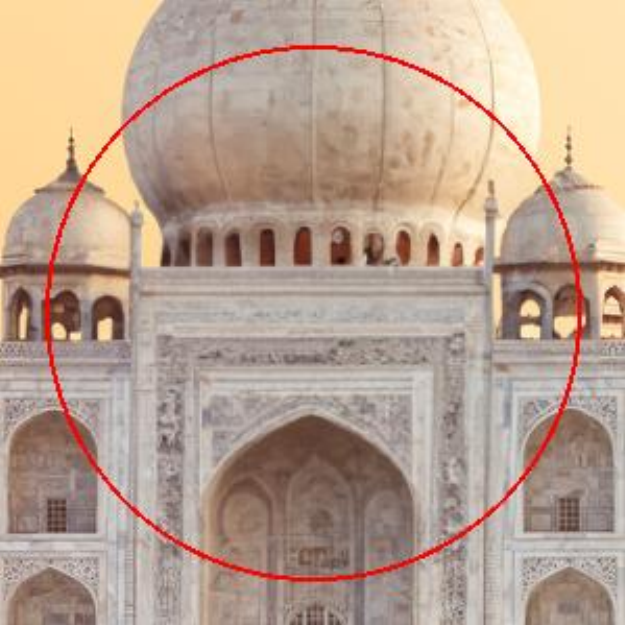}} & 
        \noindent\parbox[c]{0.083\textwidth}{\includegraphics[width=0.083\textwidth]{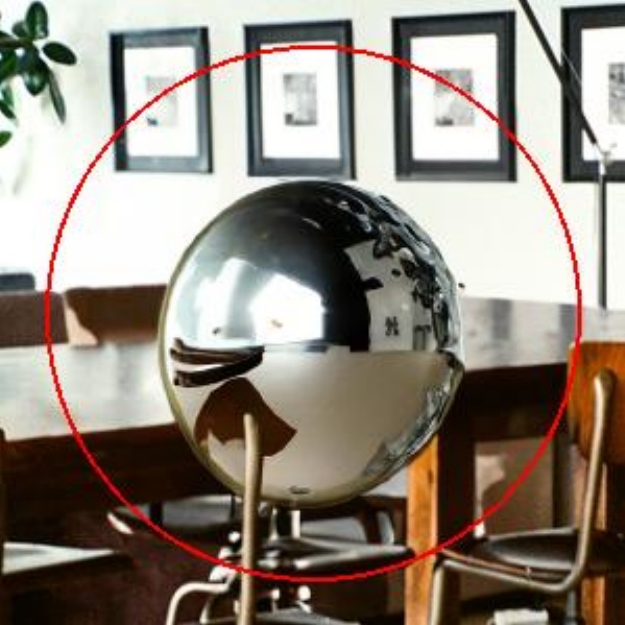}} & 
        \noindent\parbox[c]{0.083\textwidth}{\includegraphics[width=0.083\textwidth]{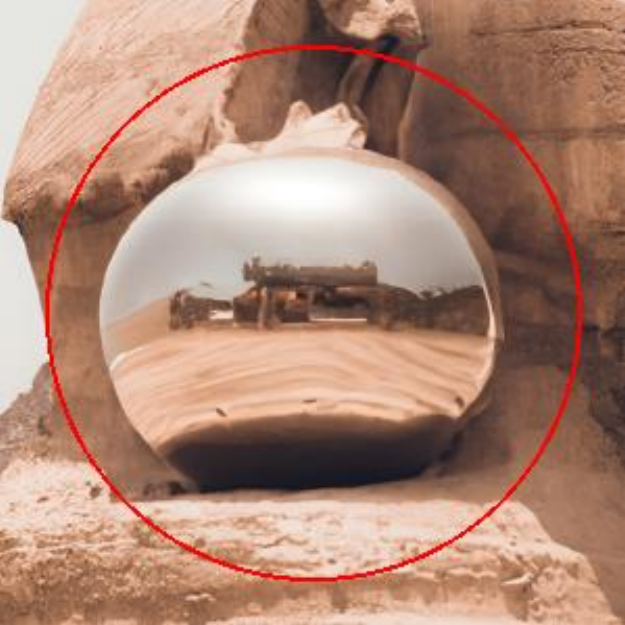}} & 
        \noindent\parbox[c]{0.083\textwidth}{\includegraphics[width=0.083\textwidth]{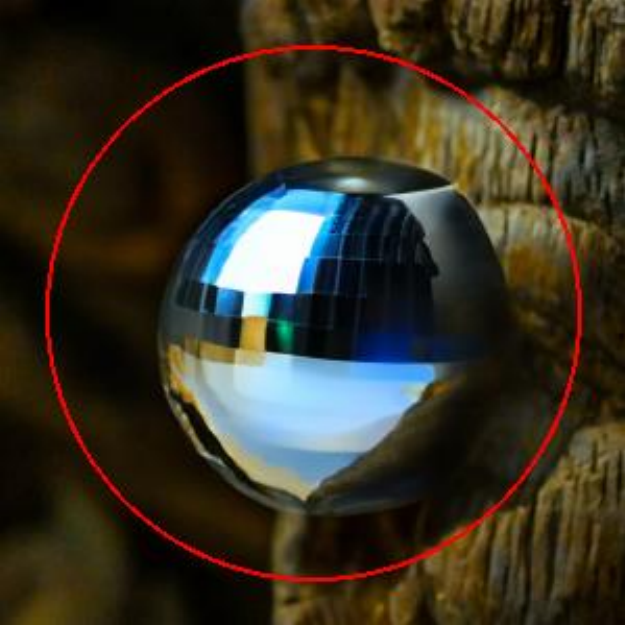}} & 
        \noindent\parbox[c]{0.083\textwidth}{\includegraphics[width=0.083\textwidth]{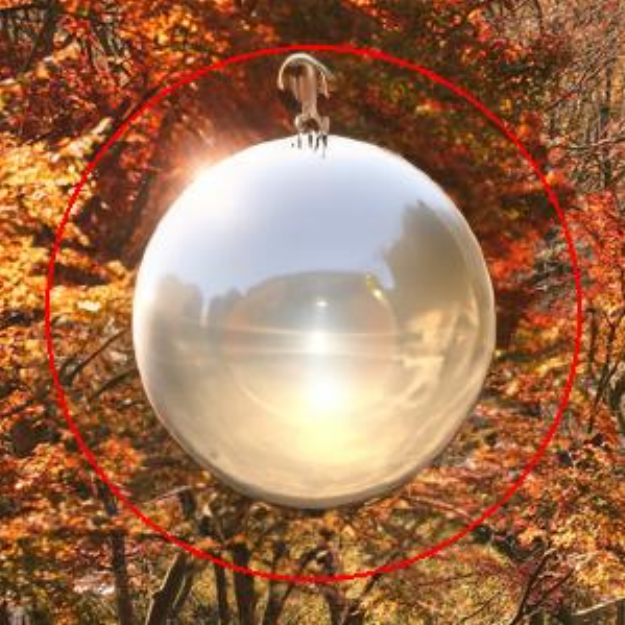}} &

        \\

        \multicolumn{1}{l}{\rotatebox[origin=c]{90}{\shortstack[l]{\scriptsize SDXL \cite{podell2023sdxl}}}} &
        \noindent\parbox[c]{0.083\textwidth}{\includegraphics[width=0.083\textwidth]{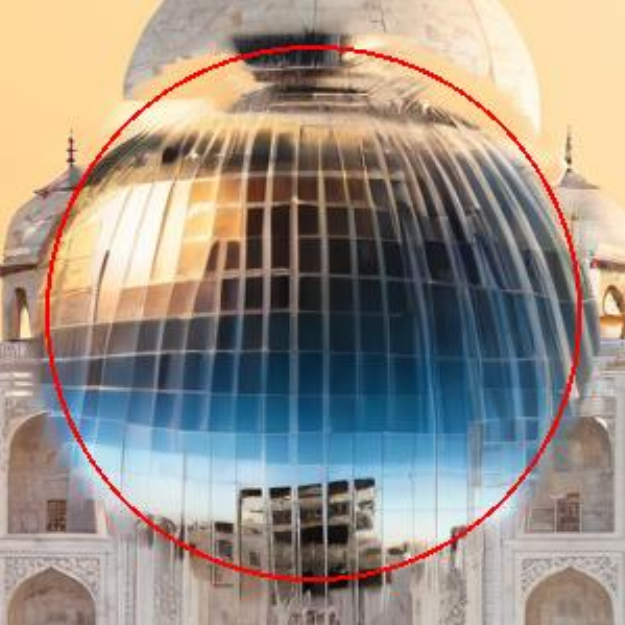}} & 
        \noindent\parbox[c]{0.083\textwidth}{\includegraphics[width=0.083\textwidth]{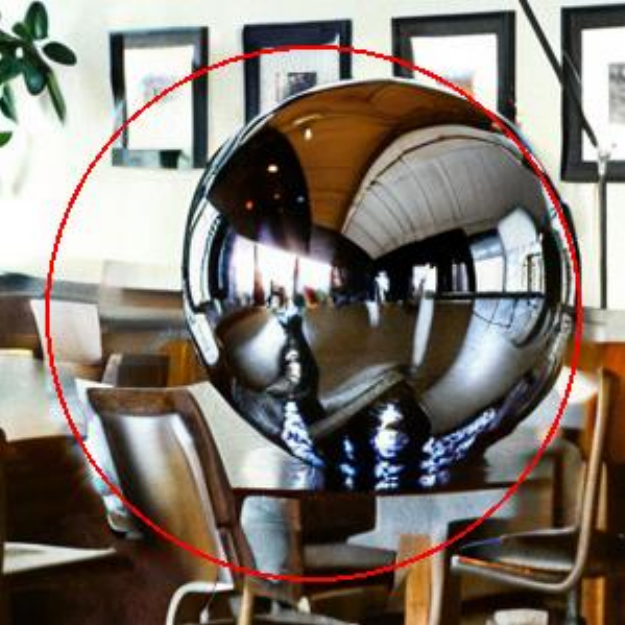}} & 
        \noindent\parbox[c]{0.083\textwidth}{\includegraphics[width=0.083\textwidth]{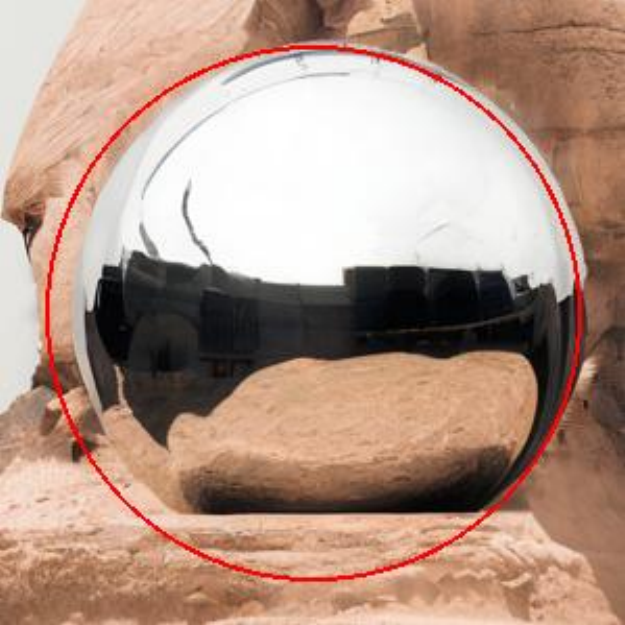}} & 
        \noindent\parbox[c]{0.083\textwidth}{\includegraphics[width=0.083\textwidth]{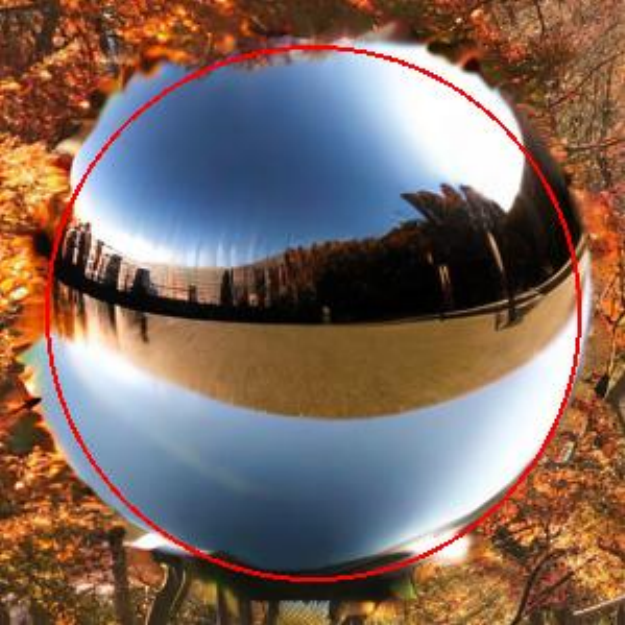}} & 
        \noindent\parbox[c]{0.083\textwidth}{\includegraphics[width=0.083\textwidth]{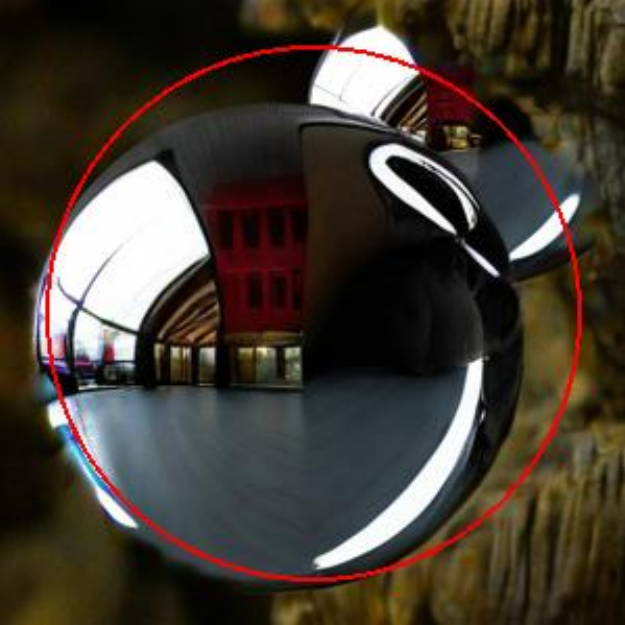}} &
        
        \\ \hline

        \multicolumn{1}{l}{\rotatebox[origin=c]{90}{\shortstack[l]{\scriptsize \textbf{Ours}}}} &
        \noindent\parbox[c]{0.083\textwidth}{\includegraphics[width=0.083\textwidth]{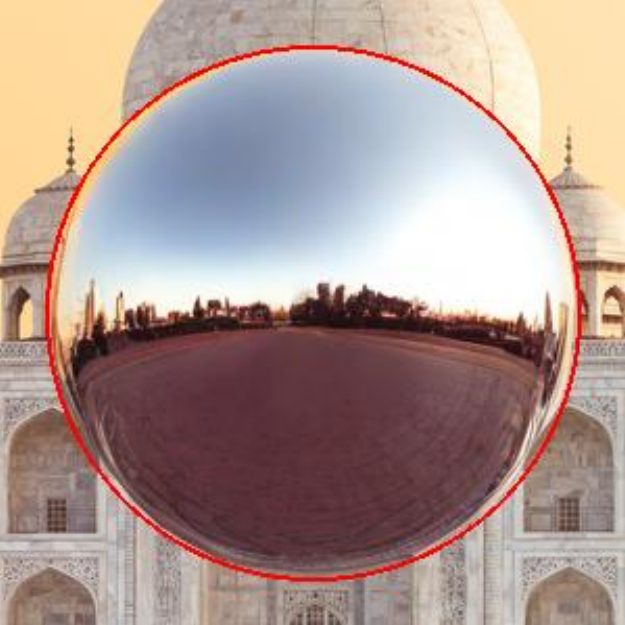}} & 
        \noindent\parbox[c]{0.083\textwidth}{\includegraphics[width=0.083\textwidth]{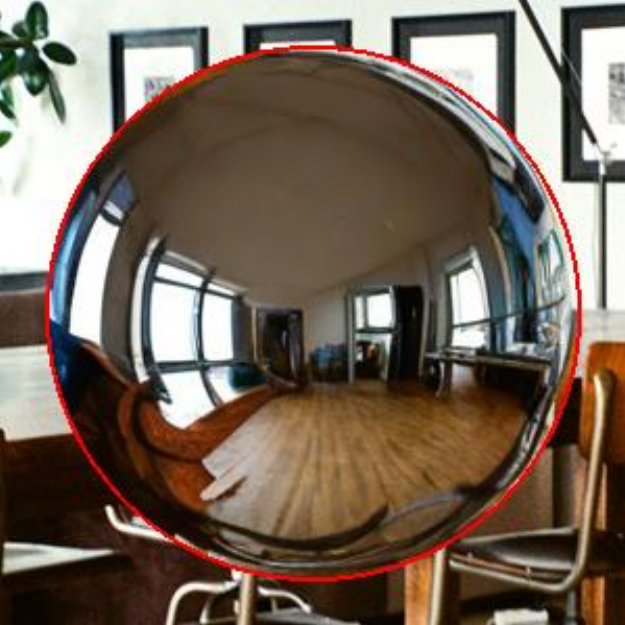}} & 
        \noindent\parbox[c]{0.083\textwidth}{\includegraphics[width=0.083\textwidth]{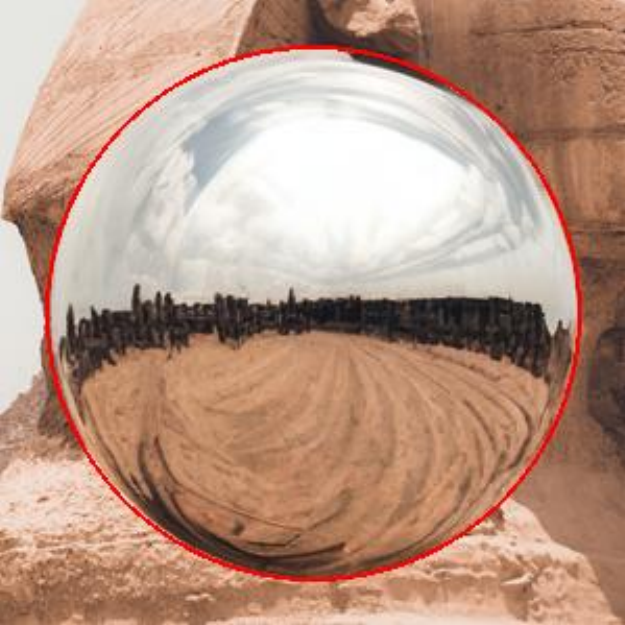}} & 
        \noindent\parbox[c]{0.083\textwidth}{\includegraphics[width=0.083\textwidth]{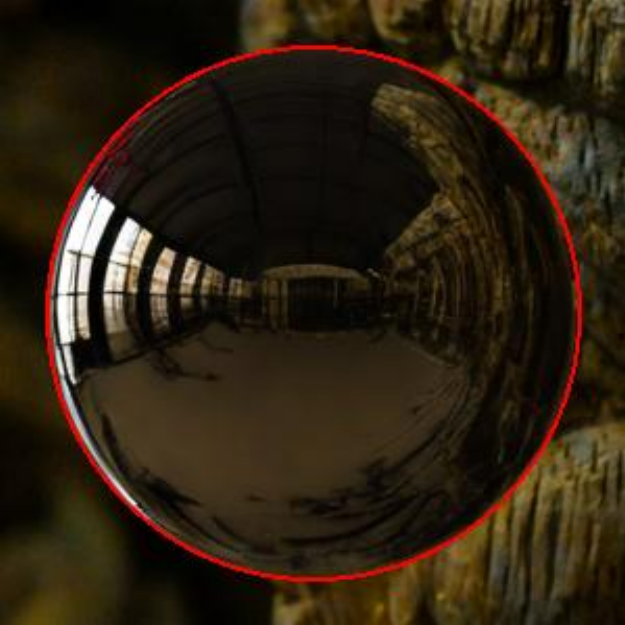}} & 
        \noindent\parbox[c]{0.083\textwidth}{\includegraphics[width=0.083\textwidth]{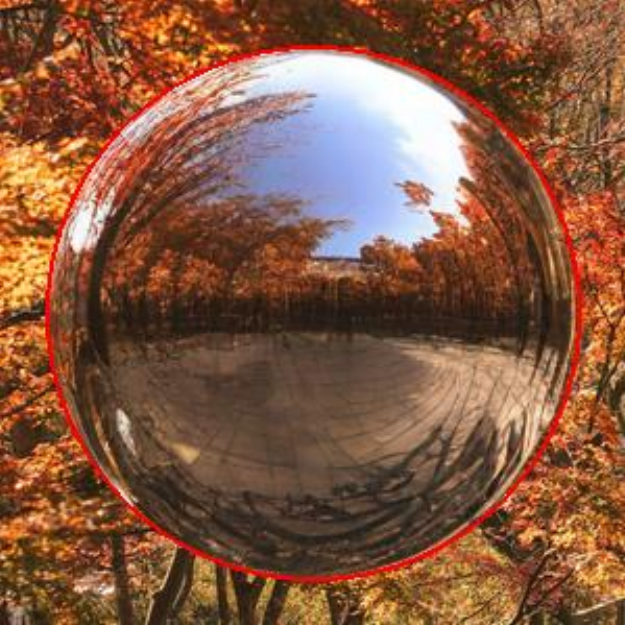}} & 
        
        \\
        
    \end{tabu}
    \caption{Chrome ball inpainting results from various methods. The red circle indicates the inpainted region, and we show a zoomed-in view of the blue crop. These diffusion models tend to produce distorted balls with undesirable textures, or completely fail to produce a ball and instead just reconstruct the original content. Our method addresses all these issues and precisely follows the inpainting mask.}
    \vspace{-0.6cm}
    \label{fig:inpaint_sota}
\end{figure}

\textbf{Lighting estimation.} 
While there exists light estimation methods that require specific light probes \cite{Debevec1998, weber2018learning, lombardi2015reflectance, park2020seeing, yu2023accidental, georgoulis2017around} or naturally occurring elements \cite{calian2018faces, yi2018faces, nishino2004eyes} in the image, we focus our review on approaches that do not require such probes. These methods are designed to handle indoor \cite{gardner2019deepparam, garon2019fastspatialvary, weber2022editableindoor, zhan2021emlight} or outdoor scenes \cite{hold2017outdoor, hold2019laveloutdoor}, or both \cite{dastjerdi2023everlight, legendre2019deeplight}. 


Unlike earlier approaches that rely on limited lighting models \cite{lalonde2012estimating, karsch2011rendering, karsch2014automatic, hosek2012analytic, hold2017outdoor, garon2019fastspatialvary, gardner2019deepparam}, modern techniques have shifted towards predicting 360$^{\circ}$ HDR environment maps, which are essential for tasks such as virtual insertion of highly reflective objects.
A common strategy among these technique involves regressing an LDR input with a limited field of view into an HDR map with neural networks. Gardner et al. \cite{garder2017lavelindoor} use a CNN classifier to locate lights using a large dataset of LDR panoramas and then fine-tune the CNN to predict HDR maps using a smaller HDR dataset. 
Hold-Geoffroy et al. \cite{hold2019laveloutdoor} first train an autoencoder for outdoor sky maps, then use another network to encode and decode an input image. Weber et al. \cite{weber2022editableindoor} predict LDR maps along with parametric light sources \cite{gardner2019deepparam}, also with a CNN. Somanath et al. \cite{somanath2021envmapnet} introduce two loss functions based on randomly masked L1 and cluster classification to enhance estimation accuracy. Zhan et al. \cite{zhan2021emlight, zhan2022gmlight} a two-step process involving a spherical light distribution predictor and an HDR map predictor. 
These methods are typically demonstrated in either indoor or outdoor settings due to the specific lighting models or the lack of sufficiently large and diverse HDR datasets.

To solve both indoor and outdoor settings, LeGendre et al. \cite{legendre2019deeplight} collected a new dataset of natural scenes captured using a mobile device with three reflective probing balls. Their method, DeepLight, regresses HDR lighting from an input image using a loss function that minimizes the difference between ground truth and rendered balls under the predicted lighting. Dastjerdi et al. \cite{dastjerdi2023everlight}'s EverLight combines multiple indoor and outdoor datasets and predicts an editable lighting representation, which then conditions a GAN to generate a full HDR map. Some GAN-based techniques focus on outpainting an input image to a 360$^{\circ}$ panorama \cite{akimoto2019360outpainting2stategan, dastjerdi2022immersegan, akimoto2022outpainting}, but they perform poorly in light estimation due to the LDR prediction \cite{dastjerdi2023everlight, wang2023360}. In contrast, StyleLight by Wang et al. \cite{wang2022stylelight} trains a two-branch StyleGAN network to predict LDR and HDR maps from noise and, at test time, uses GAN inversion to predict an HDR map from an input image.

Despite many attempts to combine indoor and outdoor panoramas, the resulting datasets remain small and limited in diversity.
In contrast, we leverage diffusion models trained on billions of images, leading to better generalization.

\myparagraph{Image inpainting using diffusion models}. Our method relies on text-conditioned diffusion models to synthesize chrome balls. While there are many diffusion models for image editing or inpainting \cite{avrahami2022blendeddiffusion, avrahami2023blendedlatent, rombach2021highresolution, meng2022sdedit, tang2023realfill, yang2023paint, ye2023ip-adapter, preechakul2022diffusion, wallace2023edict, wallace2023end}, we only discuss work related to object insertion in images.
Blended Diffusion \cite{avrahami2022blendeddiffusion, avrahami2023blendedlatent} allows arbitrary object insertion using text prompts for masked regions in an input image. This is done using guided sampling \cite{dhariwal2021diffusion} based on the cosine distance between the CLIP embedding of the inpainted region and the prompt. Paint-by-Example \cite{yang2023paint} uses example images and their CLIP embeddings as prompts to condition diffusion models. ControlNet \cite{zhang2023adding} and IP-Adapter \cite{ye2023ip-adapter} enable conditioning pre-trained diffusion models for image generation using both image and text prompts. Commercial products like DALL·E2 \cite{dalle2} and Adobe Firefly\cite{adobefirefly} also offer similar inpainting capabilities with text prompts.

Unfortunately, these methods fail to consistently produce chrome balls or produce ones that do not convincingly reflect the environmental lighting (see Figure \ref{fig:inpaint_sota}).


\myparagraph{Personalized text-to-image diffusion models.} Our work enhances pre-trained diffusion models for consistent generation of known objects, which is related to \emph{personalized} image generation. 
For this task, models are fine-tuned using single or a few reference object images while preserving their unique appearance. 
DreamBooth \cite{ruiz2022dreambooth} uses a special token during fine-tuning to represent the object while maintaining the pre-trained distribution.
Gal et al. \cite{gal2022image} introduce a learned word in the embedding space for representing referenced objects, and Voynov et al. \cite{voynov2023P+} adopt separate word embeddings per network layer. 
Additionally, there are studies exploring techniques to simplify fine-tuning of diffusion models \cite{hu2021lora, han2023svdiff, oft2023}, with LoRA \cite{hu2021lora} being a popular choice that enforces low-rank weight changes.



These methods can be adapted to our task by providing chrome ball images for fine-tuning. In fact, a portion of our pipeline can be viewed as a variant of DreamBooth with LoRA, albeit without the prior preservation loss.

\ifpurecompactformat
\vspace{-0.1cm}
\fi
\section{Approach}
\ifpurecompactformat
\vspace{-0.1cm}
\fi

\setlength\tabcolsep{0pt}
\begin{figure*}[!t]
    \centering
    \includegraphics[width=1.0\textwidth]{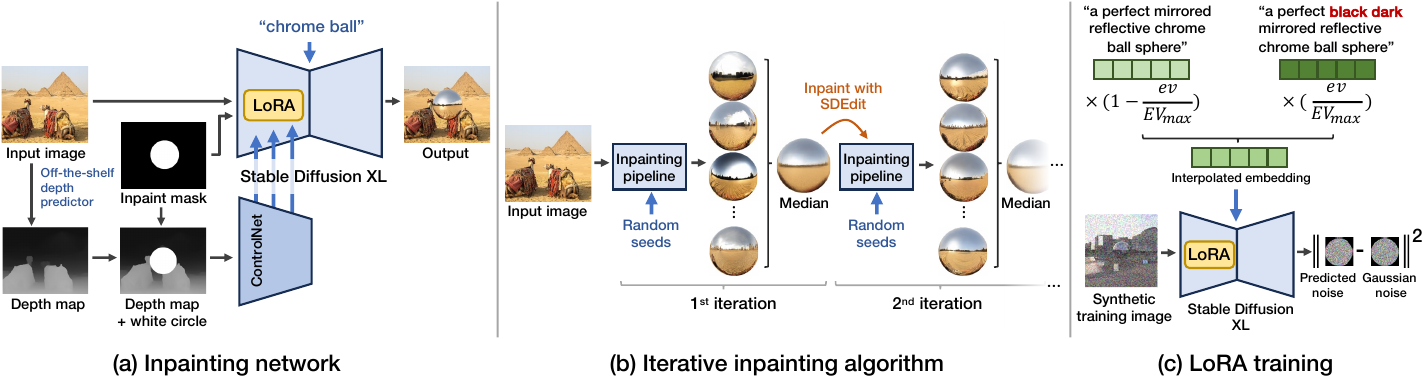}
    \caption{(a) We use Stable Diffusion XL \cite{podell2023sdxl} with depth-conditioned ControlNet \cite{zhang2023adding} to inpaint a chrome ball. (b) Our iterative inpainting algorithm helps improve generation quality and consistency by constraining the initial noise through sample averaging (Section \ref{sec:median_algo}). (c) We train LoRA for exposure bracketing, which produces multiple LDR chrome balls with varying exposures for HDR merging (Section \ref{sec:cont_lora}).
    \vspace{-0.5cm}
 }
    \label{fig:pipeline_overview}
\end{figure*}

Given a standard LDR input image, our goal is to estimate the scene's lighting as an HDR environment map. 
Our solution is based on inserting a chrome ball into the image using a diffusion model, then unwraps it to an environment map. We tackle two key challenges of (1) how to consistently generate chrome balls and (2) how to use an LDR diffusion model to generate HDR chrome balls.


\myparagraph{Overview.} As shown in Figure \ref{fig:pipeline_overview}, our key component is based on Stable Diffusion XL \cite{podell2023sdxl} with depth-conditioned ControlNet \cite{zhang2023adding}. We first predict a depth map from the input image using an off-the-shelf depth prediction network \cite{ranftl2020towards, ranftl2021vision}. Then, we paint a circle both at the depth map's center with the distance closest to the camera (visualized as white) and in an inpaint mask. We feed them along with the input image and the prompt ``\emph{a perfect mirrored reflective chrome ball sphere}'' to the diffusion model. 

We make two improvements to the above base model. First, we propose a technique called `iterative inpainting' to locate a neighborhood of good initial noise maps that lead to consistent and high-quality chrome balls (Section \ref{sec:median_algo}). Second, to further improve the generated appearance and generate multiple LDR images for exposure bracketing, we fine-tune the diffusion model using LoRA \cite{hu2021lora} on a set of synthetically generated chrome balls with varying exposures (Section \ref{sec:cont_lora}). 
To explain our method in detail, we first cover background and standard notations of diffusion models.



\subsection{Preliminaries} \label{sec:prelim}

\textbf{Diffusion models} \cite{ho2020denoising} form a family of generative models that can transform a prior distribution (Gaussian distribution) to a target data distribution $p_{\mathrm{data}}$ by learning to revert a Gaussian diffusion process. Following the convention in \cite{song2020denoising}, it is represented by a discrete-time stochastic process $\{ \vect{x}_t \}_{t=0}^T$ where $\vect{x}_0 \sim p_{\mathrm{data}}$, and $\mathbf{x}_t \sim \mathcal{N}(\vect{x}_{t-1}; \sqrt{\alpha_t / \alpha_{t-1} } \vect{x}_{t-1}, (1 - \alpha_t / \alpha_{t-1})\vect{I})$. The decreasing scalar function $\alpha_t$, with constraints that $\alpha_0 = 1$ and $\alpha_T \approx 0$, controls the noise level through time. It can be shown that
\begin{equation} \label{eq:add_noise}
    \vect{x}_t = \sqrt{\alpha_t} \vect{x}_0 + \sqrt{1 - \alpha_t} \vect{\epsilon}, ~ \text{where} ~ \vect{\epsilon} \sim \mathcal{N}(\vect{0}, \vect{I}).
\end{equation}
A diffusion model is a neural network $\epsilon_{\theta}$ trained to predict from $\vect{x}_t$ the noise $\epsilon$ that was used to generate it according to \eqref{eq:add_noise}. The commonly employed, simplified training loss is
\begin{equation} \label{eq:diffusion_objective}
    \mathcal{L} = \mathbb{E}_{\vect{x}_0, t, \vect{\epsilon}}\lVert \epsilon_{\vect{\theta}} (\sqrt{\alpha_t}\vect{x}_0 + \sqrt{1 - \alpha_t}\epsilon, t, \vect{C}) - \vect{\epsilon} \rVert_2^2,
\end{equation}
where $\vect{C}$ denotes conditioning signals such as text. The trained network can then be used to convert a Gaussian noise sample to a data sample through several sampling methods \cite{ho2020denoising, song2020denoising, zhao2023unipc}. In this paper, we use Stable Diffusion \cite{rombach2021highresolution, podell2023sdxl}, which operates on latent codes of images according to a variational autoencoder (VAE). As such, we use $\vect{x}_t$ to denote images and $\vect{z}_t$ to denote latent codes.


\myparagraph{Lora fine-tuning.} 
Instead of fine-tuning each full weight matrix $\vect{W}_i \in \mathbb{R}^{m \times n}$, LoRA\cite{hu2021lora} optimizes a learnable low-rank residual matrix $\Delta \vect{W}_i = \vect{A}_i\vect{B}_i$, where $\vect{A}_i \in \mathbb{R}^{m \times d}$, $\vect{B}_i \in \mathbb{R}^{d \times n}$, and $d \ll m,n$. The final weight matrix is given by $\vect{W}_i' = \vect{W}_i + \alpha \Delta \vect{W}_i$, where $\alpha$ is the ``LoRA scale.''

\subsection{Iterative inpainting for improving quality } \label{sec:median_algo}

\begin{figure}
    \centering

    \begin{tabu} to \textwidth {
        @{}
        c@{}
        c@{\hspace{1.0pt}}
        c@{\hspace{0.5pt}}
        c@{\hspace{0.5pt}}
        c@{\hspace{0.5pt}}
        c@{\hspace{0.5pt}}
        c@{}
    }
        
        \multicolumn{1}{l}{\rotatebox[origin=c]{90}{\shortstack[l]{\scriptsize Input}}} &
        \noindent\parbox[c]{0.083\textwidth}{\includegraphics[width=0.083\textwidth]{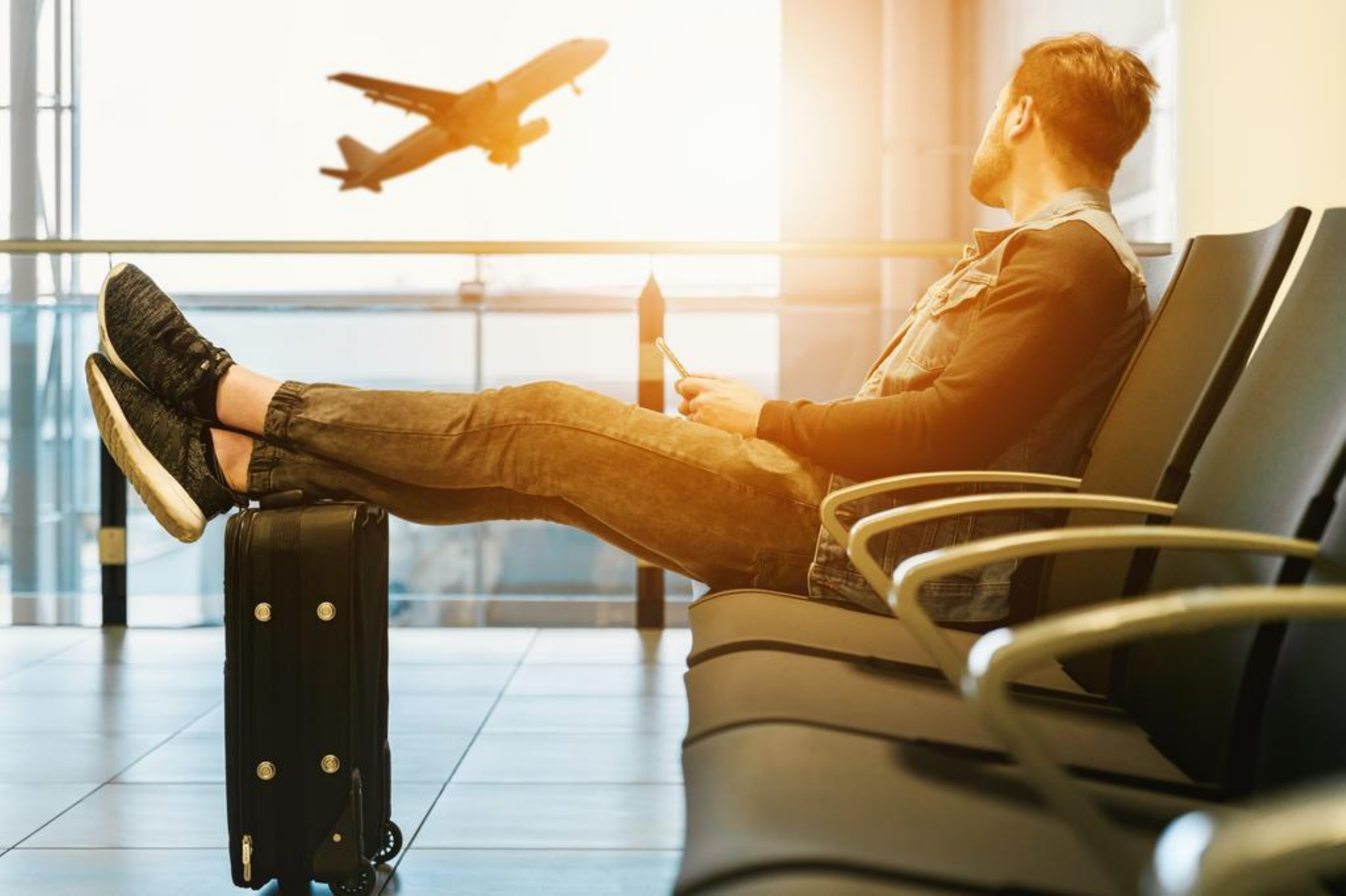}} & 
        \noindent\parbox[c]{0.083\textwidth}{\includegraphics[width=0.083\textwidth]{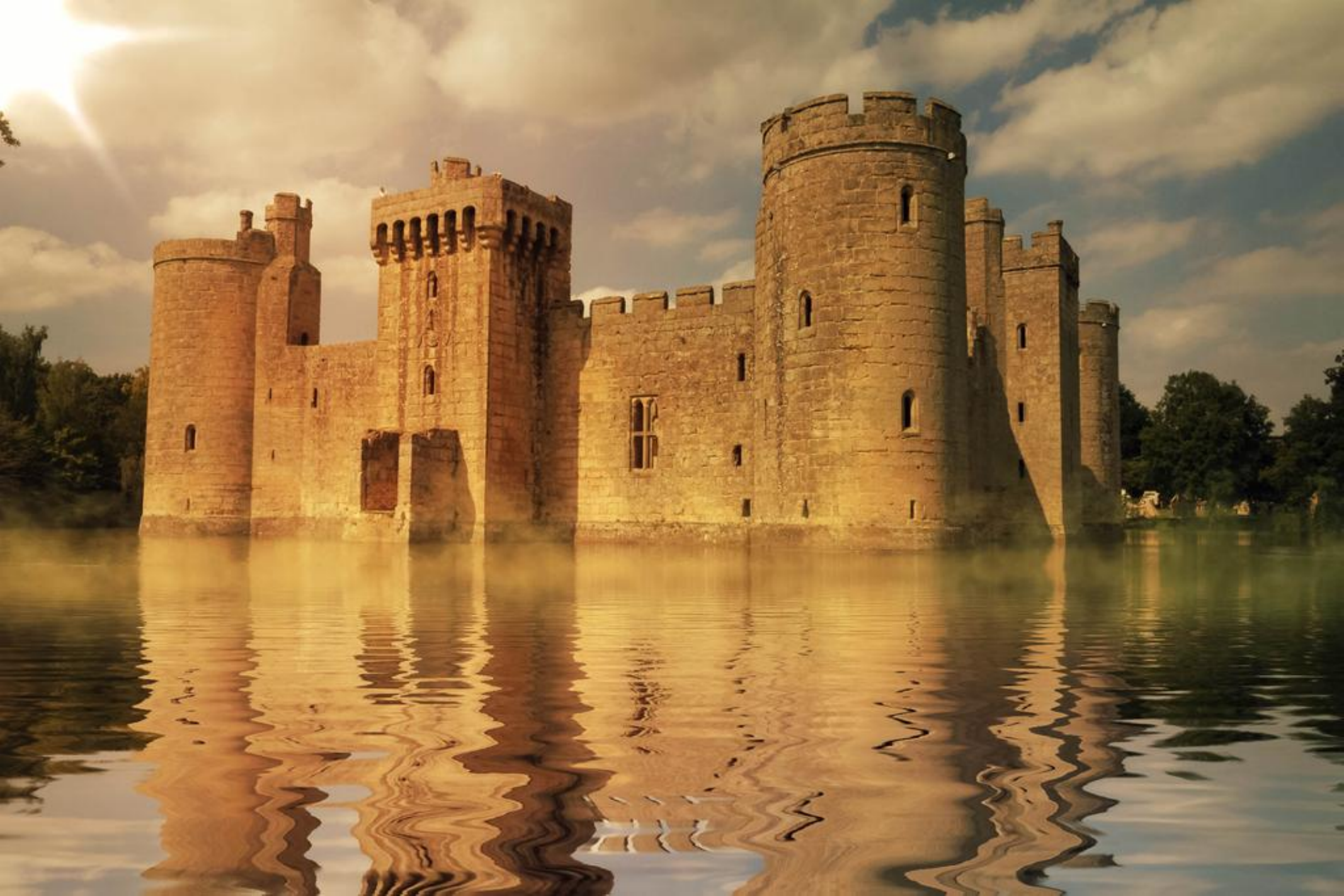}} & 
        \noindent\parbox[c]{0.083\textwidth}{\includegraphics[width=0.083\textwidth]{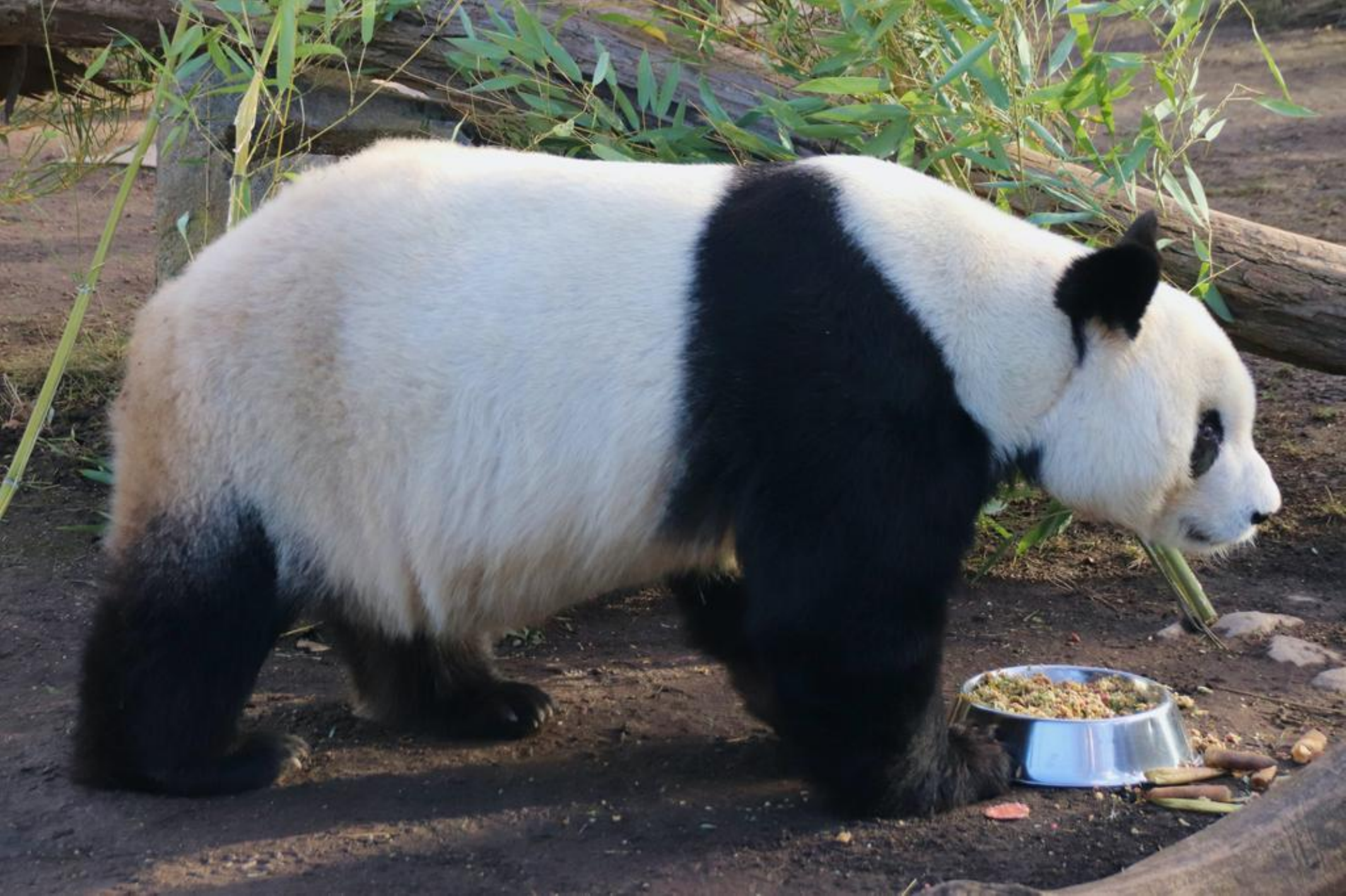}} & 
        \noindent\parbox[c]{0.083\textwidth}{\includegraphics[width=0.083\textwidth]{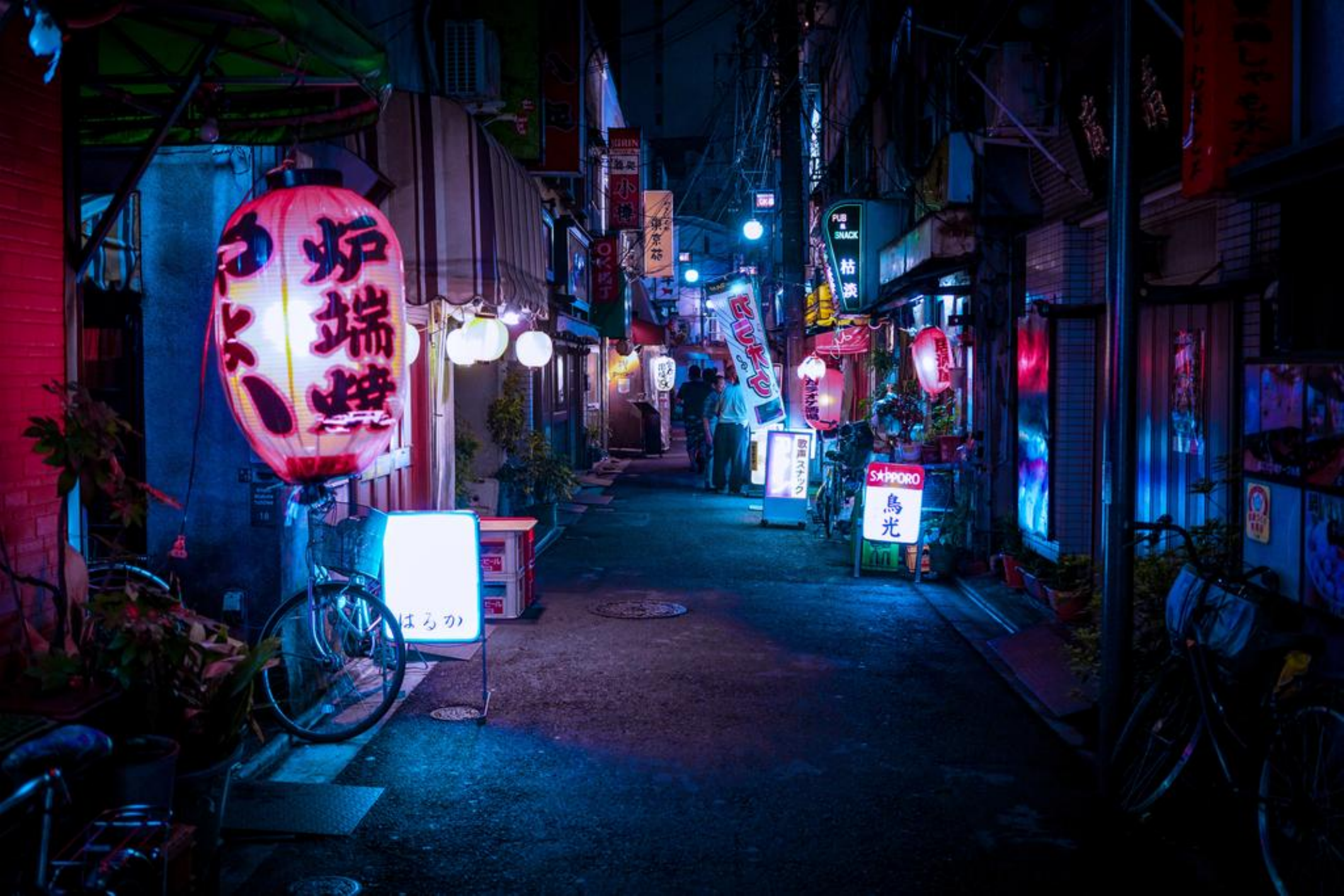}} & 
        \noindent\parbox[c]{0.083\textwidth}{\includegraphics[width=0.083\textwidth]{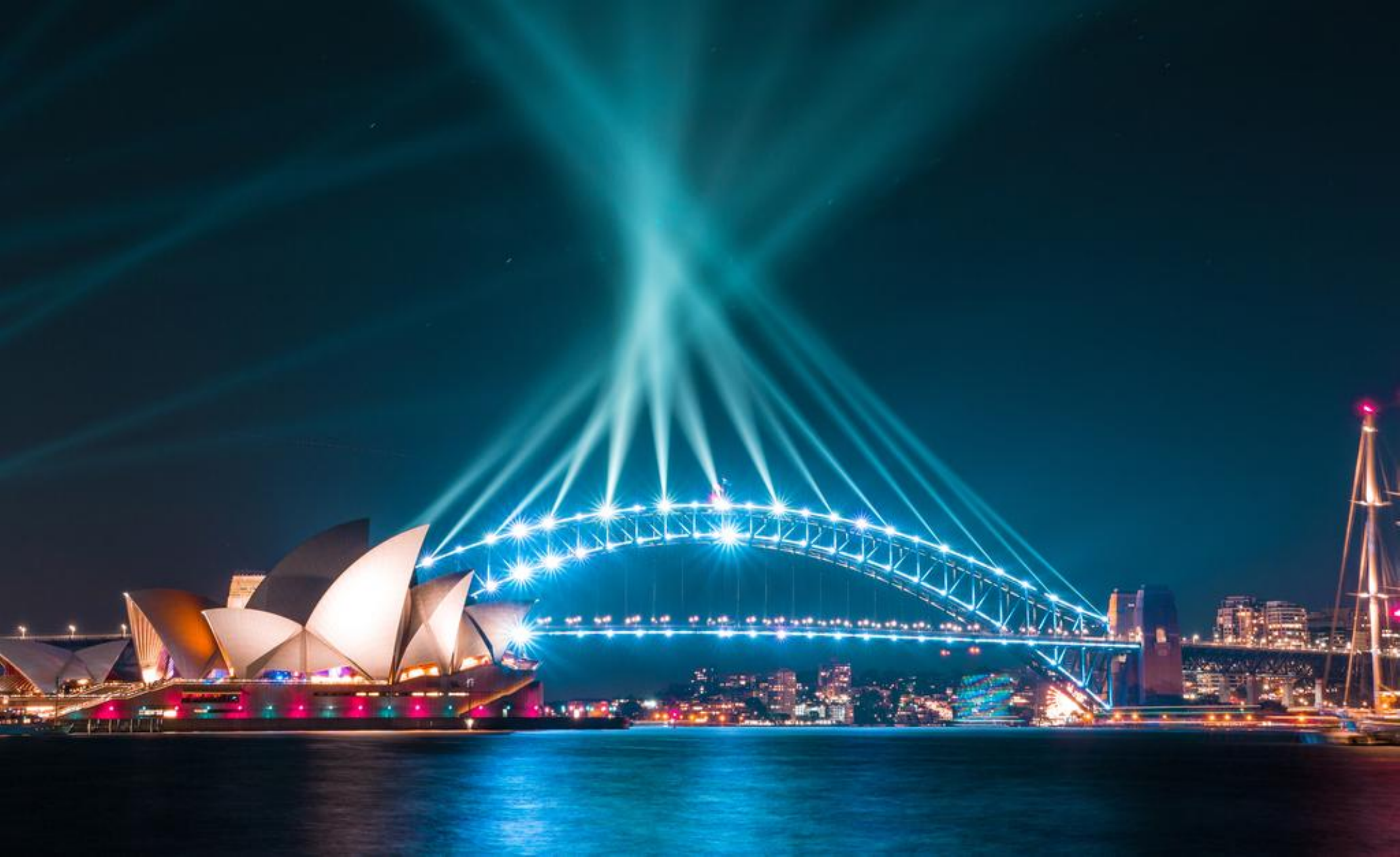}} & 

        \\

        \multicolumn{1}{l}{\rotatebox[origin=c]{90}{\shortstack[l]{\scriptsize Bad noise \#1}}} &
        \noindent\parbox[c]{0.083\textwidth}{\includegraphics[width=0.083\textwidth]{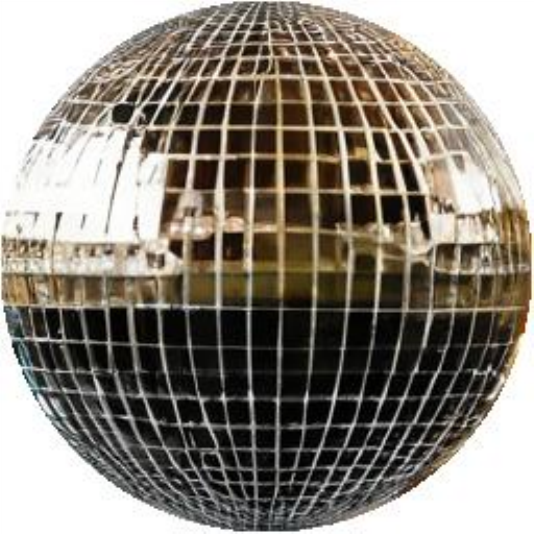}} & 
        \noindent\parbox[c]{0.083\textwidth}{\includegraphics[width=0.083\textwidth]{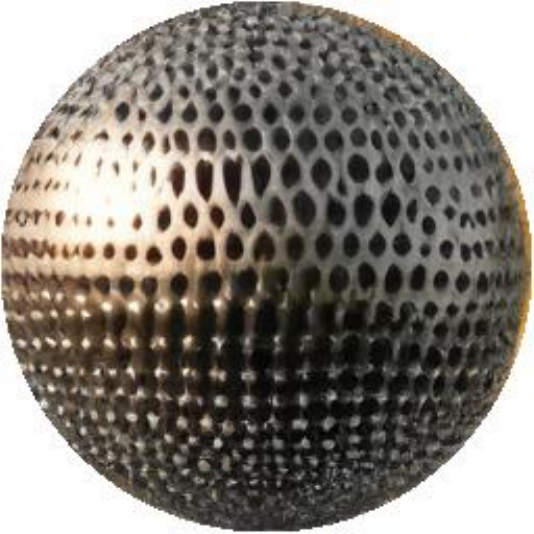}} & 
        \noindent\parbox[c]{0.083\textwidth}{\includegraphics[width=0.083\textwidth]{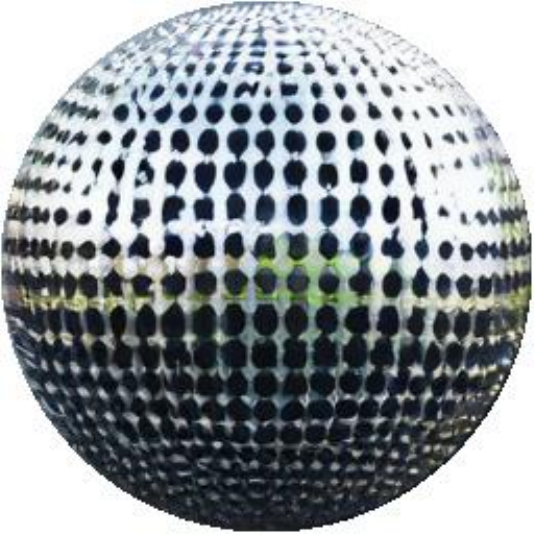}} & 
        \noindent\parbox[c]{0.083\textwidth}{\includegraphics[width=0.083\textwidth]{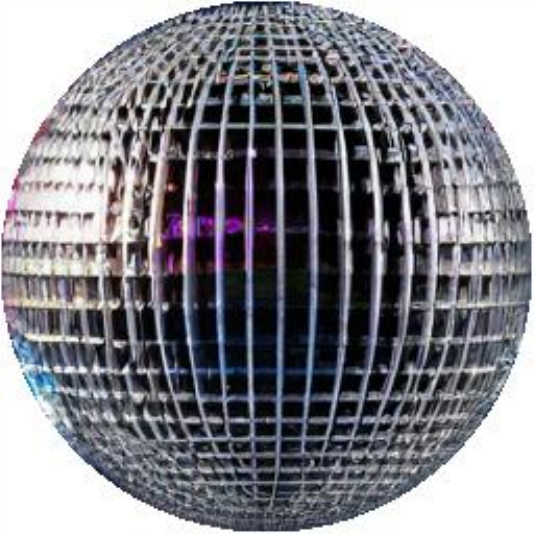}} & 
        \noindent\parbox[c]{0.083\textwidth}{\includegraphics[width=0.083\textwidth]{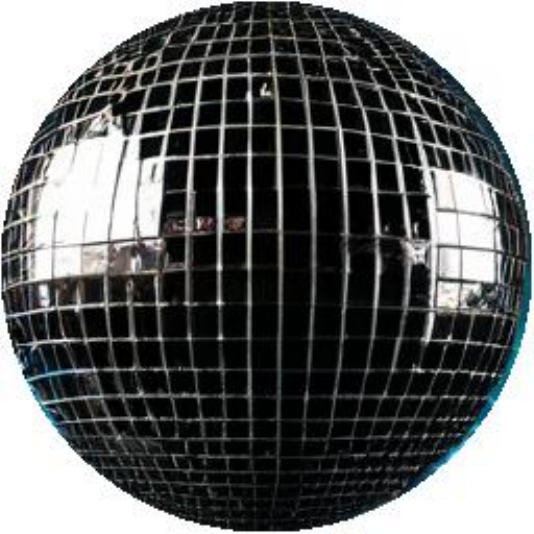}} & 
        
        \\

        \multicolumn{1}{l}{\rotatebox[origin=c]{90}{\shortstack[l]{\scriptsize Bad noise \#2}}} &
        \noindent\parbox[c]{0.083\textwidth}{\includegraphics[width=0.083\textwidth]{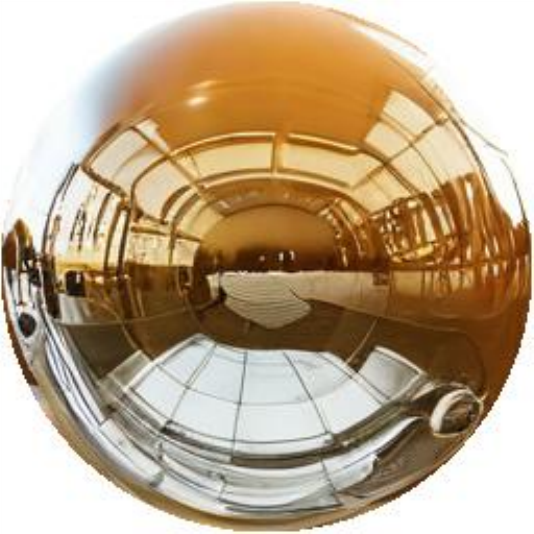}} & 
        \noindent\parbox[c]{0.083\textwidth}{\includegraphics[width=0.083\textwidth]{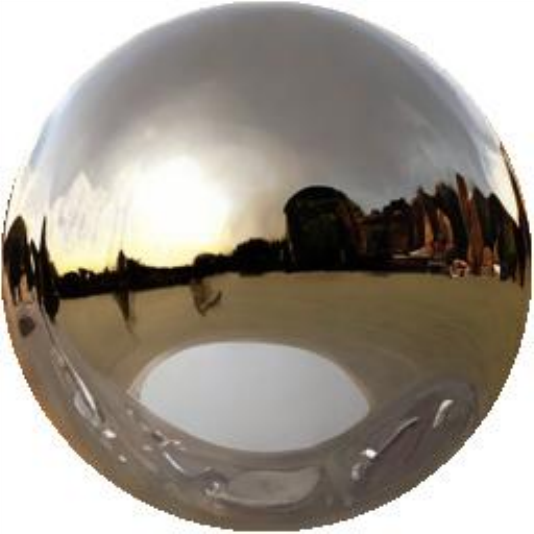}} & 
        \noindent\parbox[c]{0.083\textwidth}{\includegraphics[width=0.083\textwidth]{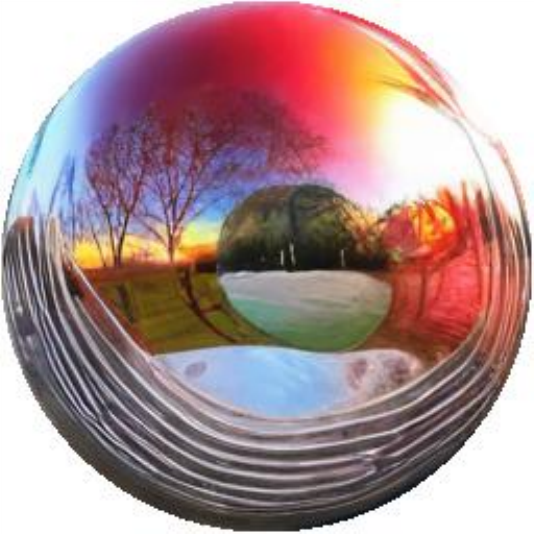}} & 
        \noindent\parbox[c]{0.083\textwidth}{\includegraphics[width=0.083\textwidth]{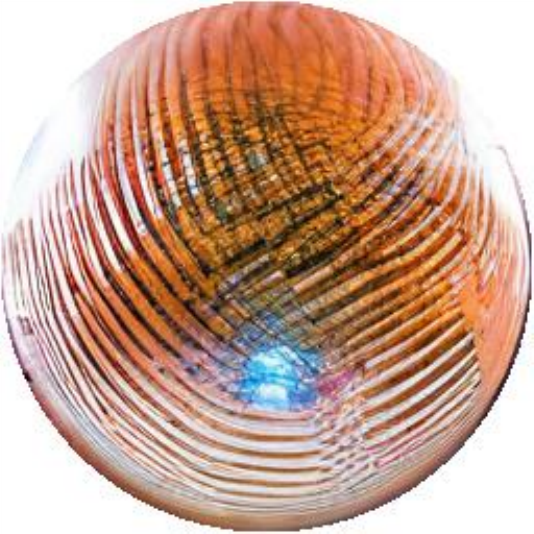}} & 
        \noindent\parbox[c]{0.083\textwidth}{\includegraphics[width=0.083\textwidth]{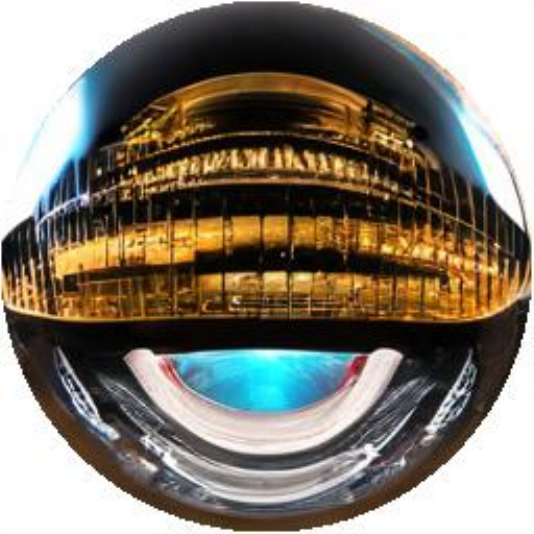}} & 
        
        \\
        
        \arrayrulecolor{red}\cline{2-6}

        \multicolumn{1}{l}{\rotatebox[origin=c]{90}{\shortstack[l]{\scriptsize \textbf{Ours}}}} &
        \multicolumn{1}{|c}{
        \noindent\parbox[c]{0.083\textwidth}{\includegraphics[width=0.083\textwidth]{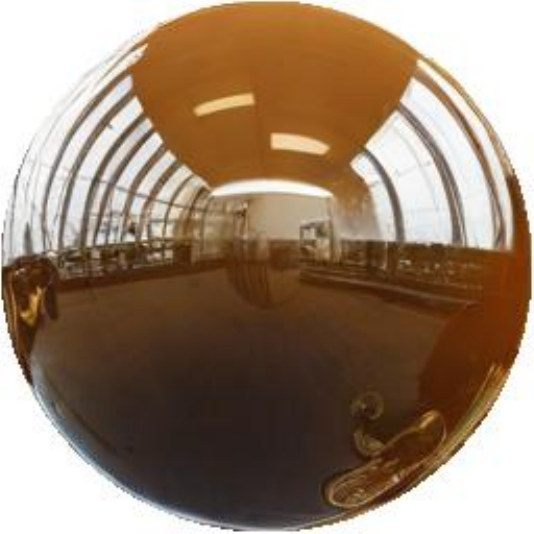}}} & 
        \noindent\parbox[c]{0.083\textwidth}{\includegraphics[width=0.083\textwidth]{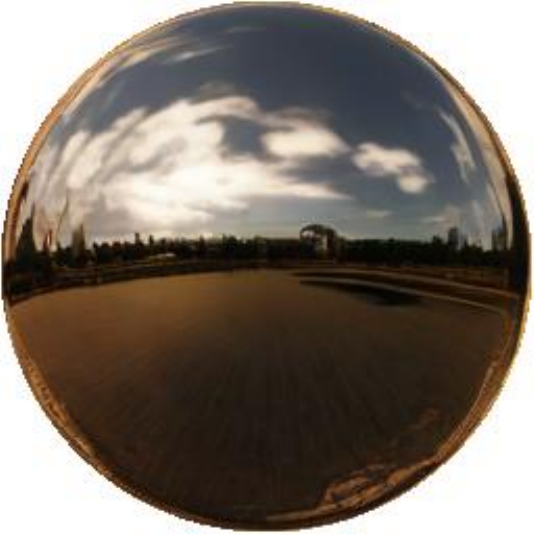}} & 
        \noindent\parbox[c]{0.083\textwidth}{\includegraphics[width=0.083\textwidth]{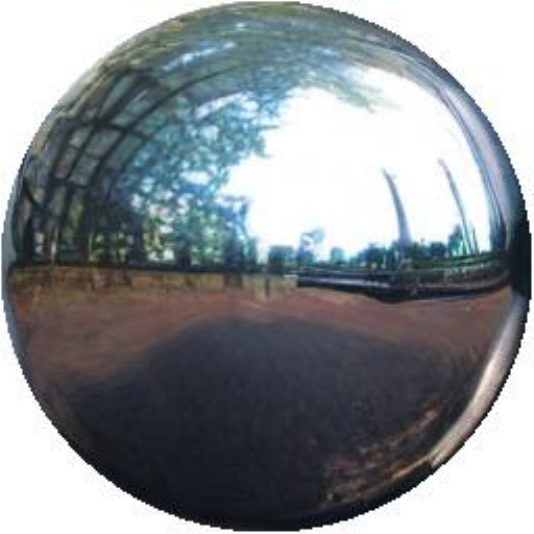}} & 
        \noindent\parbox[c]{0.083\textwidth}{\includegraphics[width=0.083\textwidth]{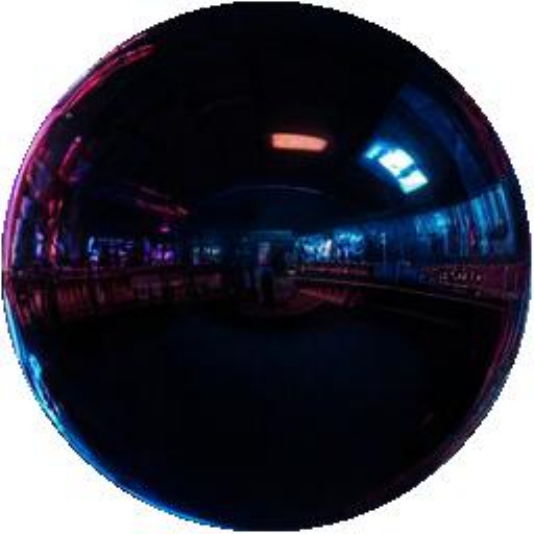}} & 
        \multicolumn{1}{c|}{\noindent\parbox[c]{0.083\textwidth}{\includegraphics[width=0.083\textwidth]{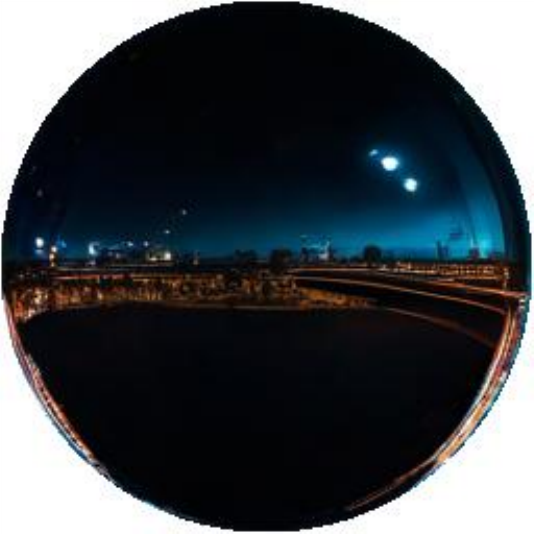}}} & 

        \\ \arrayrulecolor{red}\cline{2-6}
        \arrayrulecolor{black}
        
    \end{tabu}

    \caption{Observations: the initial noise map encodes some semantic patterns. We found certain noise maps to consistently produce a disco chrome ball or bad patterns \emph{across} input images. In contrast, our results produce consistently clean chrome balls.
    }
    \label{fig:noise_appearance_relationship}
\end{figure}

We found that the base depth-conditioned Stable Diffusion model could reliably \emph{insert} a chrome ball as opposed to merely recovering the masked out content. However, the chrome balls often contain undesirable patterns and fail to convincingly reflect environment lighting (Figure \ref{fig:noise_appearance_relationship}).

Our first improvement stems from a few observations: The same initial noise map leads to the generation semantically similar inpainted areas \emph{regardless} of the input image. For instance, there exists a ``disco'' noise map that consistently produces a disco ball \emph{across} different input images,  while a good noise map almost always produces a reflective chrome ball (Figure \ref{fig:noise_appearance_relationship}).   When searching for images of a ``chrome ball'' on the Internet, not all results match the specific reflective chrome ball we want. 
So, the encoded semantics found within the noise map are perhaps understandable, as text prompts alone cannot encode all visual appearances of ``chrome ball.''
Here, adding ``disco'' to the negative prompt may fix this specific instance, but many other failure modes are not as easy to describe and exclude via text prompts. 

\tabulinesep=0.5pt

Another observation is that the average of multiple ball samples tends to approximate the overall lighting reasonably well, but the average ball itself is too blurry and not as useful. 

Based on these insights, we propose a simple algorithm to automatically locate a good noise neighborhood by sample averaging.
Specifically, we first inpaint $N$ chrome balls onto an input image using different random seeds. Then, we calculate a pixel-wise median ball and composite it back to the input image. Let us denote this composited image by $\vect{B}$. To sample a better chrome ball, we apply SDEdit \cite{meng2022sdedit}\footnote{Commonly known as ``image-to-image'' by Stable Diffusion users \cite{HuggingFace:i2i}.} by adding noise to $\vect{B}$ to simulate the diffusion process up to time step $t$: $\vect{B}' = \sqrt{\alpha_t} \vect{B} + \sqrt{1 - \alpha_t}\vect{\epsilon}$ where $\vect{\epsilon} \sim \mathcal{N}(\vect{0}, \vect{I})$ and $t < T$, the maximum timestep. Then, we continue denoising $\vect{B}'$ starting from $t$ to 0. In our implementation, $t = 0.8T$.


We can repeat the process by using SDEdit to generate another set of $N$ chrome balls from the output and recompute another median chrome ball.
This repetition not only minimizes artifacts and spurious patterns from incorrect ball types but also enhances the consistency of light estimation.

In our implementation, we perform two iterations of median computation. This involves generating $N$ chrome balls with standard diffusion sampling to compute the first median, performing $N$ SDEdits from the first median to compute the second, and doing one last SDEdit 
to generate the output.


\tabulinesep=2pt
\begin{figure}
    \centering

    \begin{tabu} to \textwidth {
        @{}
        c@{\hspace{1pt}}
        c@{\hspace{1pt}}
        @{\hspace{8pt}}
        c@{\hspace{1pt}}
        c@{\hspace{1pt}}
        c@{}
    }
        

        
        
        


        \noindent\parbox[c]{0.08\textwidth}{\includegraphics[width=0.08\textwidth]{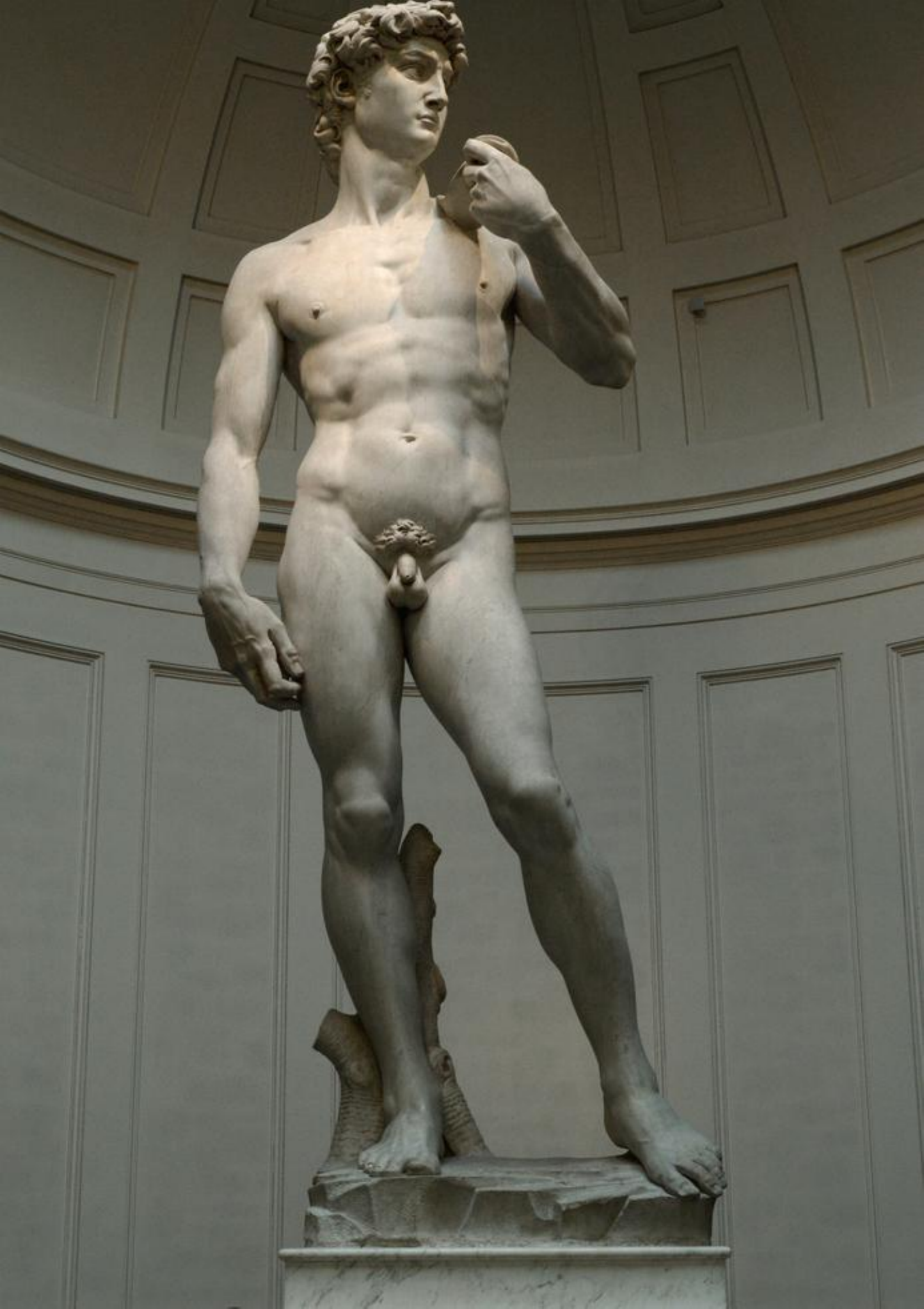}} 
        &
        \noindent\parbox[c]{0.14\textwidth}{\includegraphics[width=0.14\textwidth]{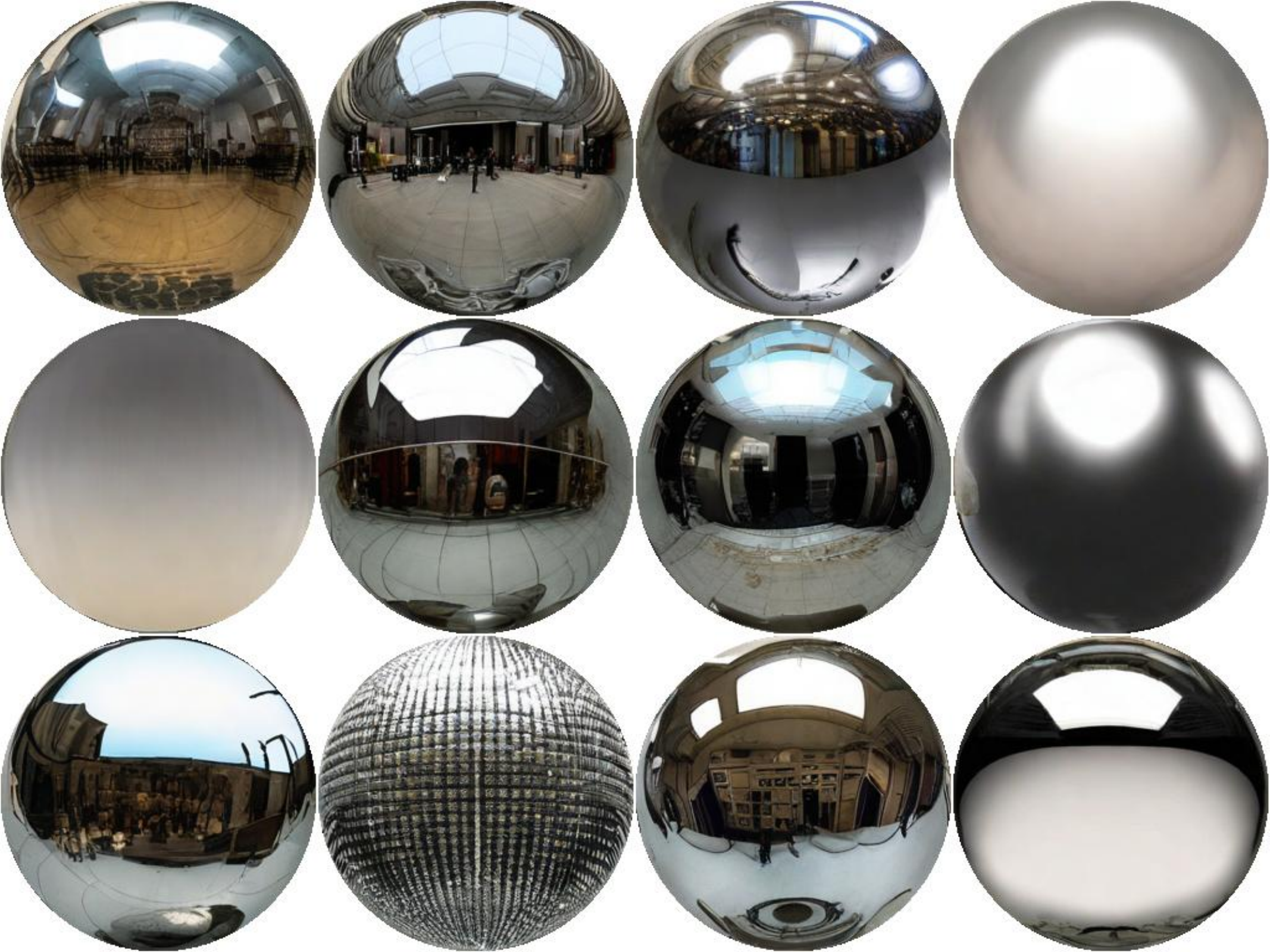}}

        &
        \noindent\parbox[c]{0.08\textwidth}{\shortstack{\tiny Median ball \\ \includegraphics[width=0.08\textwidth]{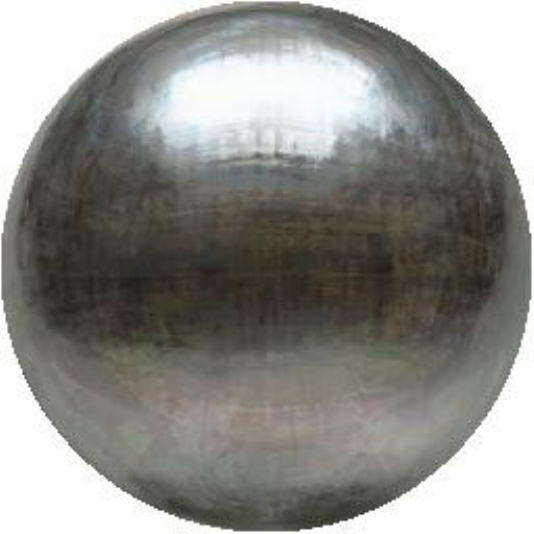}}}
        
        &
        \noindent\parbox[c]{0.14\textwidth}{\includegraphics[width=0.14\textwidth]{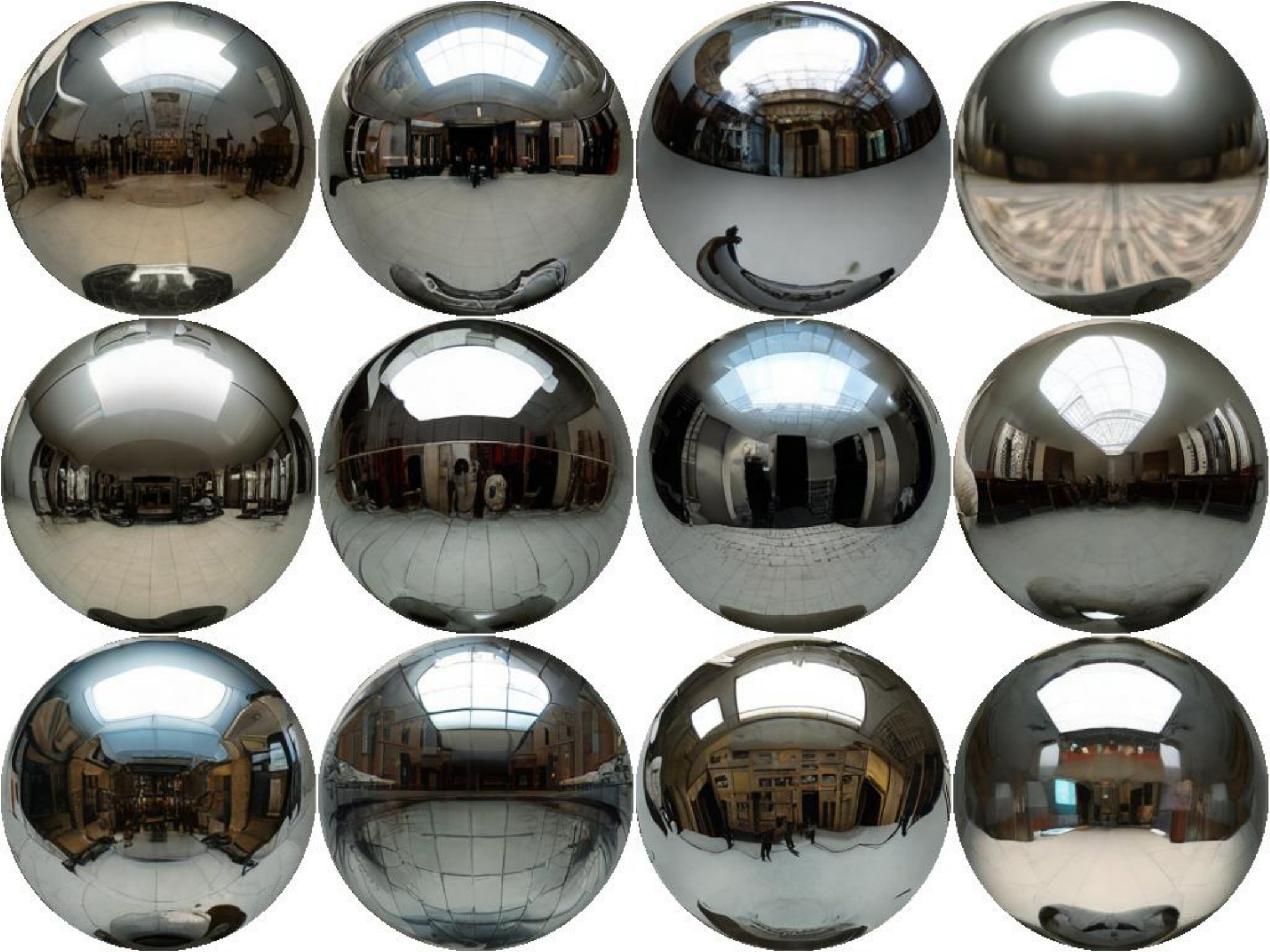}} 
        
        \\
        
    \end{tabu}
    \vspace{-0.5em}
    \caption{Balls before (left) and after (right) iterative inpainting.
    }
    \label{fig:compare_median_distribution}
    \vspace{-1.5em}
\end{figure}

\subsection{Predicting HDR chrome balls} \label{sec:cont_lora}

\tabulinesep=0.5pt
\begin{figure}
    \centering

    \begin{tabu} to \textwidth {
        @{}
        c@{\hspace{1pt}}
        c@{\hspace{1pt}}
        c@{\hspace{1pt}}
        c@{\hspace{1pt}}
        c@{\hspace{1pt}}
        c@{\hspace{1pt}}
        c@{}
    }
        
        \noindent\parbox[c]{0.090\textwidth}{\includegraphics[width=0.090\textwidth]{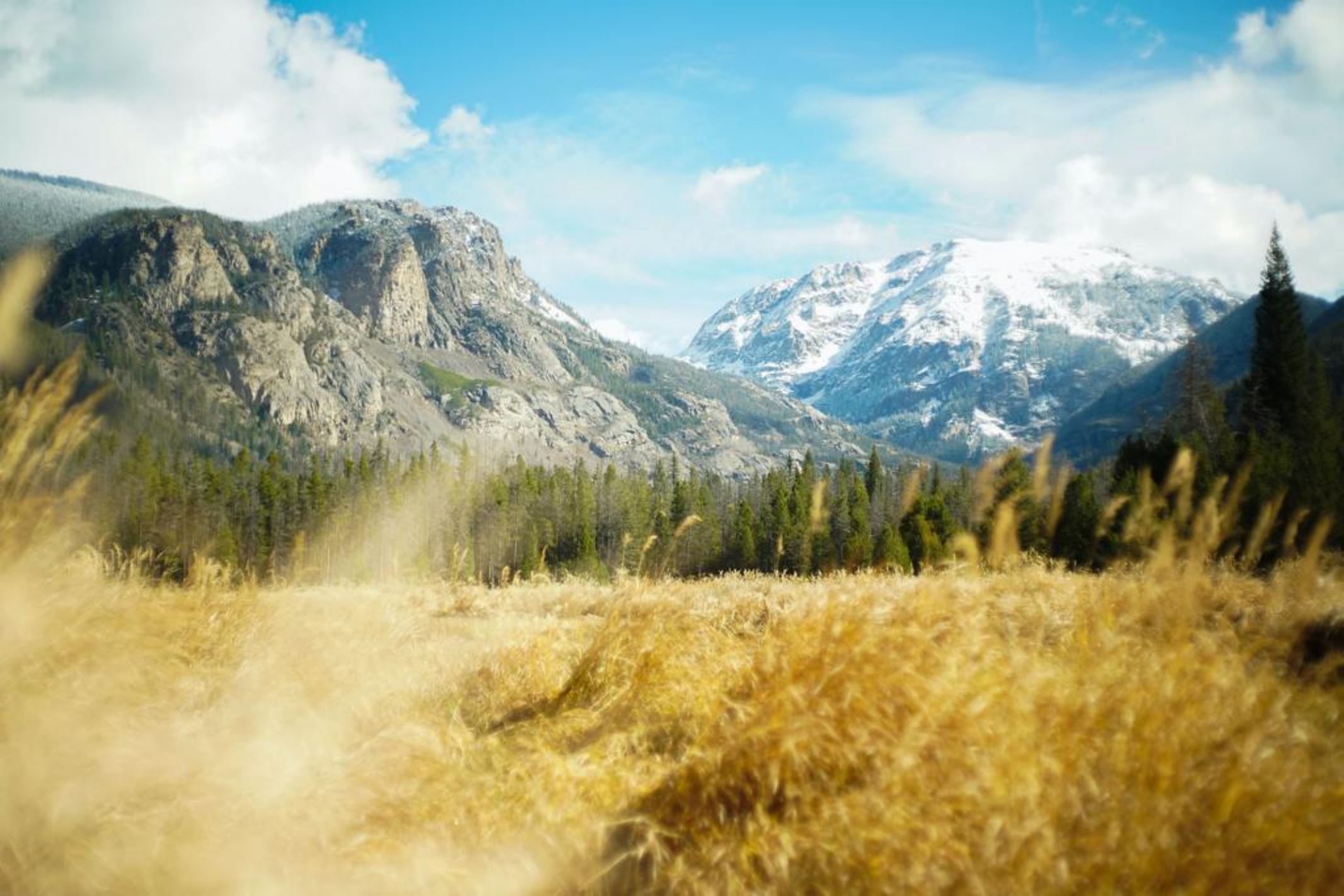}} &
        \noindent\parbox[c]{0.071\textwidth}{\includegraphics[width=0.071\textwidth]{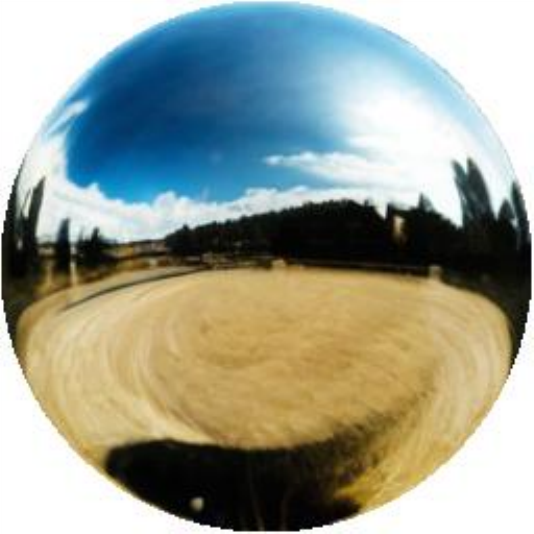}} & 
        \noindent\parbox[c]{0.071\textwidth}{\includegraphics[width=0.071\textwidth]{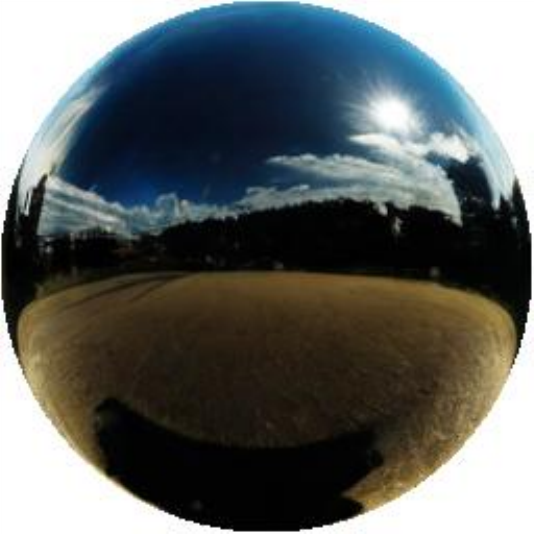}} & 
        \noindent\parbox[c]{0.090\textwidth}{\includegraphics[width=0.090\textwidth]{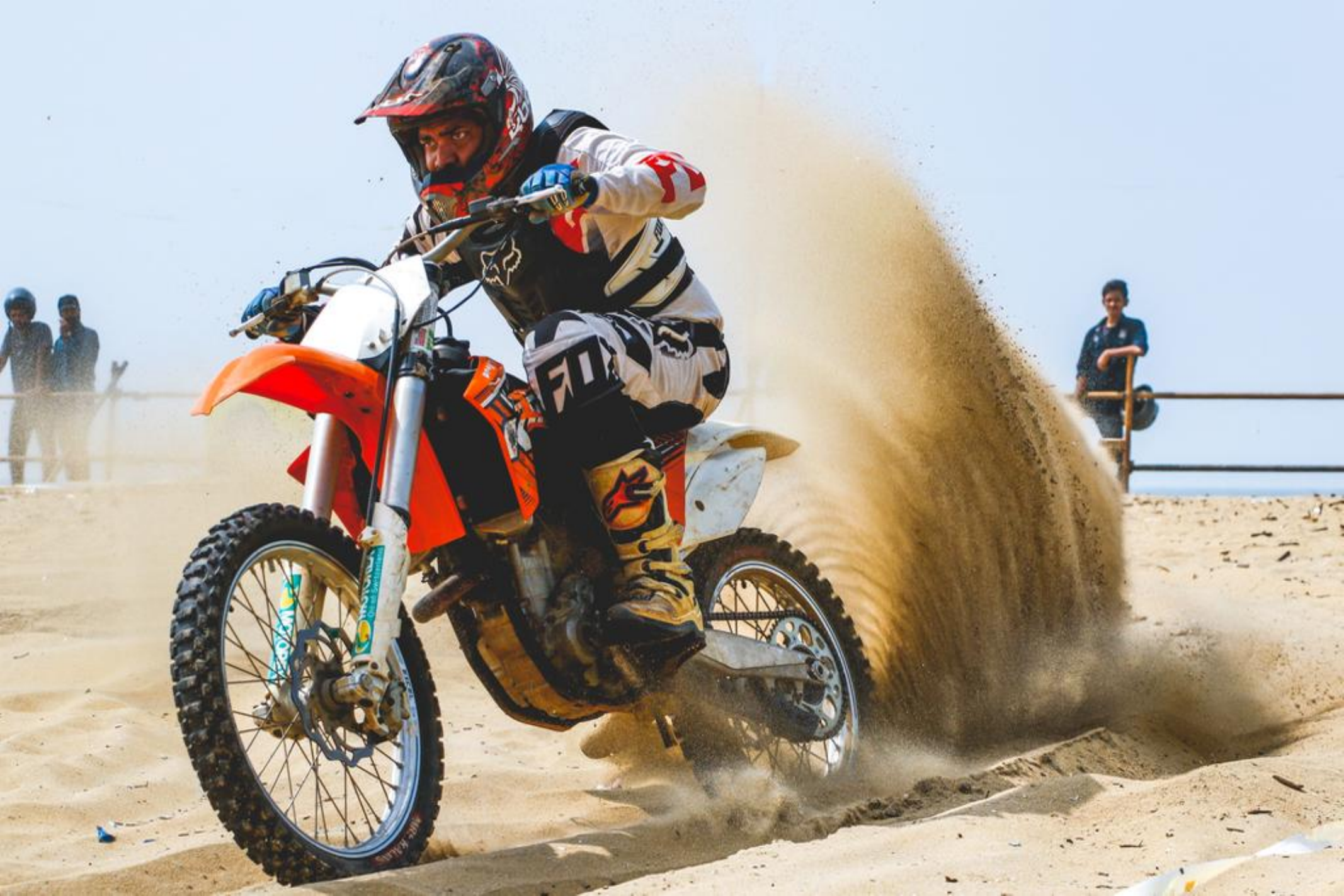}} &
        \noindent\parbox[c]{0.071\textwidth}{\includegraphics[width=0.071\textwidth]{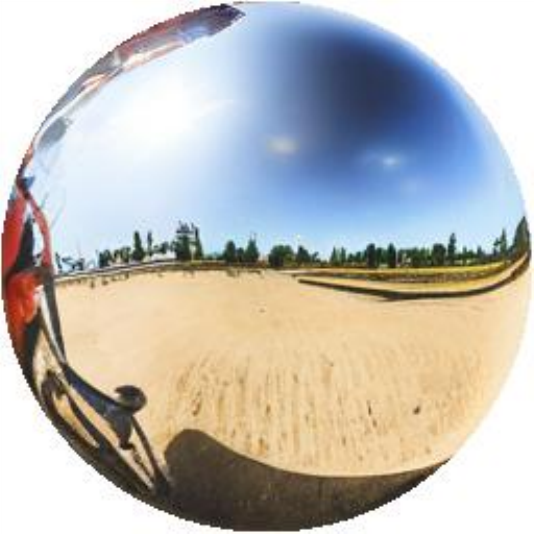}} & 
        \noindent\parbox[c]{0.071\textwidth}{\includegraphics[width=0.071\textwidth]{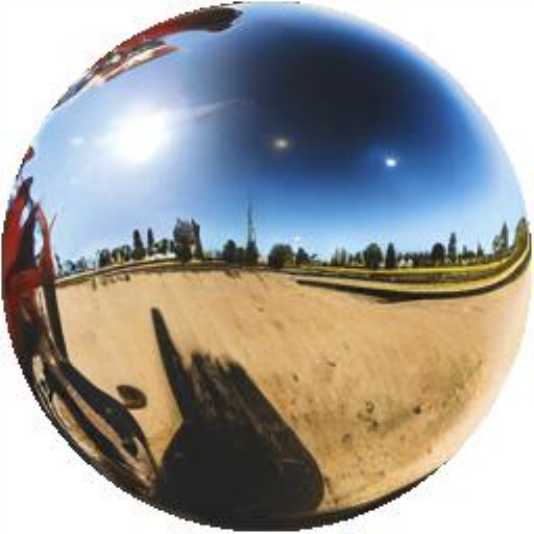}} &


        \\

        



        
    \end{tabu}

    \caption{Simply adding ``\emph{black dark}'' to the prompt allows the overexposed sun to emerge. However, we need a way to specify the target EV, which is addressed by our LoRA (Section \ref{sec:cont_lora}).
    \vspace{-0.5cm}}
    \label{fig:add_black_dark}
    
\end{figure}
The main issue with using pre-trained diffusion models for HDR prediction is that they have never seen HDR images. Nonetheless, these models can still indirectly learn about HDR and the wide range of luminance through examples of under and overexposed images in their training sets. This ability is evident in our experiment where adding `black dark' to our text prompt can reduce the overexposed white sky, allowing the round sun to emerge on outdoor scenes (see Figure \ref{fig:add_black_dark}).
To leverage this ability, we propose to use the exposure bracketing technique by inpainting multiple chrome balls with different exposure values and combining them to produce a linearized HDR output.
Our idea is to train a LoRA to steer the sampling process such that the output conforms to the appearances associated with specific exposure compensation values (EVs).

\ifpurecompactformat
\vspace{-0.1cm}
\fi
\myparagraph{Training set.} We construct our LoRA training set using HDR panoramas synthetically generated from Text2Light \cite{chen2022text2light} to avoid direct access to scenes in benchmark datasets.
As illustrated in Figure \ref{fig:pipeline_overview}, each training pair consists of a random EV, denoted by $ev$, and a panorama crop with a chrome ball rendered with EV=$ev$ at the center.
This crop is constructed by projecting a full HDR panorama to a small field-of-view image and then tone-mapping to LDR \emph{without} exposure compensation (EV0). The chrome ball is rendered using the panorama as the environment map in Blender \cite{blender}, but its luminance is scaled by $2^{ev}$ before being tone-mapped to LDR.
Following \cite{wang2022stylelight}, we use a simple $\gamma$-2.4 tone-mapping function and map the 99$^\text{th}$ percentile intensity to 0.9. 

Here, we assume that the typical output images from the diffusion model have a mean EV of zero. For light estimation purposes, our focus is on recovering high-intensity light sources, which are crucial for relighting and are captured in underexposed or negative EV images. Therefore, we randomly sample the EV values from [$EV_\text{min}$, 0].

\myparagraph{Training.} To generate a chrome ball with a specific EV, we condition our model on an interpolation of two text prompts as a function of $ev$. The two prompts are the original prompt and the original with ``black dark'' added. We denote their text embeddings by $\vect{\xi}_o$ and $\vect{\xi}_d$, respectively. The resulting embedding is given by:

\begin{align}
    \vect{\xi}^{ev} = \vect{\xi}_o + (ev / EV_{\text{min}}) (\vect{\xi}_d - \vect{\xi}_o).
\end{align}
We train our LoRA with a masked version of the standard L$_2$ loss function computed only on the chrome ball pixels given by a mask $\vect{M}$:
\begin{equation} \label{eq:lora_loss_fn}
    \mathcal{L} = \mathbb{E}_{\vect{z}_0, t, \vect{\epsilon}, ev}\left[ \lVert \vect{M} \odot \parens*{\epsilon_{\vect{\theta}} (\vect{z}_t^{ev}, t, \vect{\xi}^{ev}) - \epsilon} \rVert_2^2 \right],
\end{equation}
where $\vect{z}_t^{ev}$ is computed using Equation \eqref{eq:add_noise} from our training image with EV=$ev$. We choose to train a single LoRA as opposed to multiple LoRAs for individual EVs because it helps preserve the overall scene structure across exposures due to weight sharing. Refer to Appendix \ref{appendix:aba_lora} for details.

\ifpurecompactformat
\vspace{-0.1cm}
\fi

\myparagraph{LDR balls generation and HDR merging. } We generate chrome balls with multiple EVs = \{-5, -2.5, 0\}, each using their own median ball computation (Section \ref{sec:median_algo}). While our LoRA can  maintain the overall scene structure across exposures, some details do become altered. As a result, using standard HDR merging algorithms can lead to ghosting artifacts when details in each LDR are not fully aligned. As our primary goal is to gather high-intensity light sources from underexposed images to construct a useful light map, we can merge the luminances while retaining the chroma from the normally exposed EV0 image to reduce ghosting.

In particular, we first identify overexposed regions in each LDR image with a simple threshold of 0.9, assuming the pixel range is between 0 and 1. Then, the luminance values in these regions are replaced by the exposure-corrected luminances from lower EV images. This luminance replacement is performed in pairs, starting from the lowest EV to the normal EV0 image, detailed in Algorithm \ref{algo:ldr2hdr} in Appendix \ref{appendix:implement}.

\ifpurecompactformat
\vspace{-0.3cm}
\fi
\section{Experiments}
\label{sec:experiment}
\ifpurecompactformat
\vspace{-0.2cm}
\fi



\textbf{Implementation details.} We fine-tuned SDXL \cite{podell2023sdxl} for multi-exposure generation using a rank-4 LoRA \cite{hu2021lora}. 
We trained our LoRA on 1,412 HDR panoramas synthetically generated by Text2Light \cite{chen2022text2light} for 2,500 steps with a learning rate of $10^{-5}$ and a batch size of $4$. The process took about 5 hours on an NVIDIA RTX 3090Ti. During training, we sampled timestep $t \sim U(900, 999)$ as we found that the light information is determined at the early stage of the denoising process. 
When applying iterative inpainting (Section~\ref{sec:median_algo}), we generated $N = 30$ chrome balls per each median computation iteration. We use UniPC \cite{zhao2023unipc} sampler with 30 sampling steps, a guidance scale of 5.0, and a LoRA scale of $0.75$.

\vspace{-0.1cm}
\myparagraph{Datasets.} We evaluated our approach on two standard benchmarks: Laval Indoor HDR \cite{garder2017lavelindoor} and Poly Haven \cite{polyhaven}. The latter covers both indoor and outdoor settings.
\vspace{-0.1cm}
\myparagraph{Evaluation metrics.}
Following previous work \cite{wang2022stylelight, zhan2021emlight}, we used three scale-invariant metrics: scale-invariant Root Mean Square Error (si-RMSE) \cite{grosse2009groundtruth}, Angular Error \cite{legendre2019deeplight}, and normalized RMSE. The normalization for the last metric is done by mapping the 0.1$^\text{st}$ and 99.9$^\text{th}$ percentiles to 0 and 1, following \cite{marnerides2019expandnet}. We chose these metrics instead of standard RMSE because each benchmark dataset has its own specific range and statistics of light intensity, but our method was not trained on any of them. 

\tabulinesep=0.5pt
\begin{figure}
    \centering

    \subcaptionbox{Laval Indoor \cite{garder2017lavelindoor} \label{fig:qualitative_indoor}}{
        \begin{tabu} to \textwidth {
            @{}
            c@{\hspace{2pt}}
            c@{\hspace{2pt}}
            c@{\hspace{2pt}}
            c@{\hspace{2pt}}
            c@{\hspace{2pt}}
            c@{}
        }
            \multicolumn{1}{c}{\shortstack{\scriptsize Input image}}
            & \multicolumn{1}{c}{\shortstack{\scriptsize GT}}
            & \multicolumn{1}{c}{\shortstack{\scriptsize StyleLight \cite{wang2022stylelight}}}
            & \multicolumn{1}{c}{\shortstack{\scriptsize Ours (I)}}
            & \multicolumn{1}{c}{\shortstack{\scriptsize Ours (I+LR)}} &
            \\
    
            \noindent\parbox[c]{0.080\textwidth}{\includegraphics[width=0.080\textwidth]{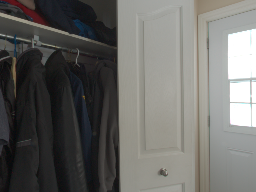}} 
            
            &
            \noindent\parbox[c]{0.080\textwidth}{\includegraphics[width=0.080\textwidth]{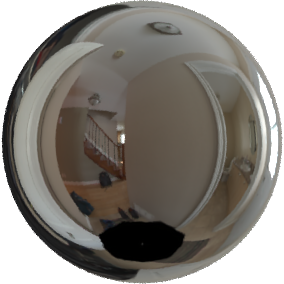}} & 
            \noindent\parbox[c]{0.080\textwidth}{\includegraphics[width=0.080\textwidth]{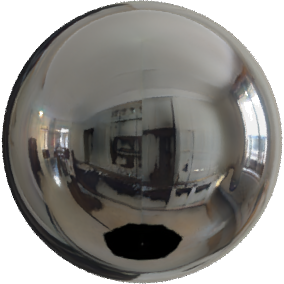}} &  
            \noindent\parbox[c]{0.080\textwidth}{\includegraphics[width=0.080\textwidth]{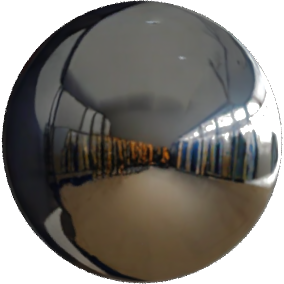}} &
           
            \noindent\parbox[c]{0.080\textwidth}{\includegraphics[width=0.080\textwidth]{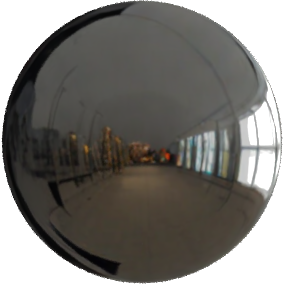}} & 
            
            \\
    
            &
            \noindent\parbox[c]{0.080\textwidth}{\includegraphics[width=0.080\textwidth]{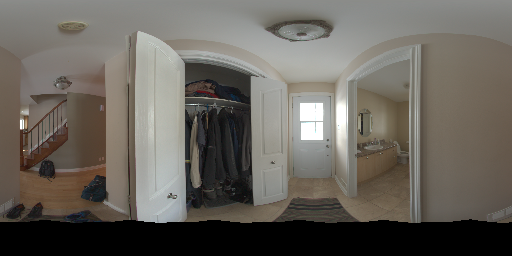}} & 
            \noindent\parbox[c]{0.080\textwidth}{\includegraphics[width=0.080\textwidth]{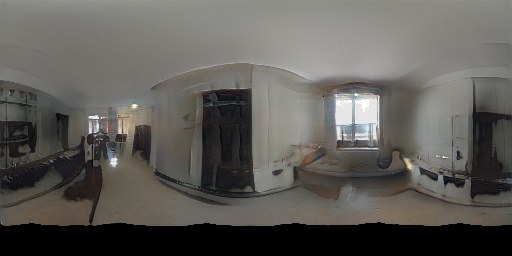}} & 
            \noindent\parbox[c]{0.080\textwidth}{\includegraphics[width=0.080\textwidth]{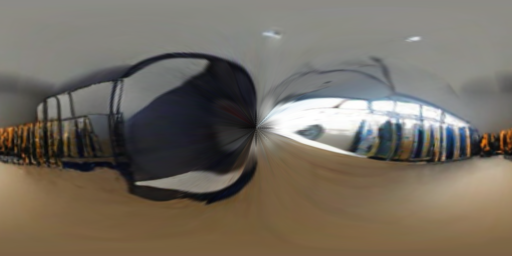}} &
            \noindent\parbox[c]{0.080\textwidth}{\includegraphics[width=0.080\textwidth]{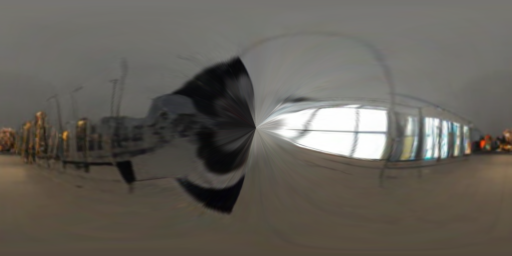}} & 
            \\
    
    
            \noindent\parbox[c]{0.080\textwidth}{\includegraphics[width=0.080\textwidth]{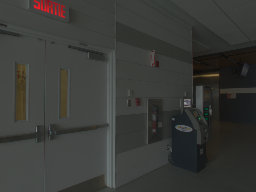}} 
            
            &
            \noindent\parbox[c]{0.080\textwidth}{\includegraphics[width=0.080\textwidth]{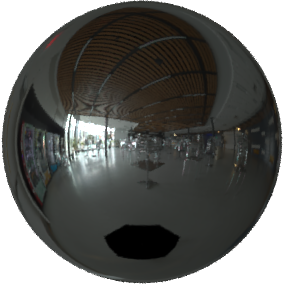}} & 
            \noindent\parbox[c]{0.080\textwidth}{\includegraphics[width=0.080\textwidth]{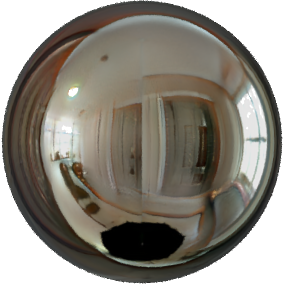}} &  
            \noindent\parbox[c]{0.080\textwidth}{\includegraphics[width=0.080\textwidth]{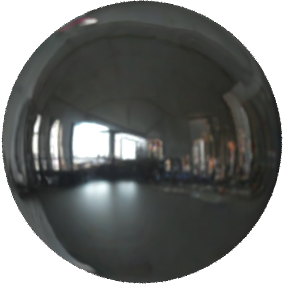}} &
           
            \noindent\parbox[c]{0.080\textwidth}{\includegraphics[width=0.080\textwidth]{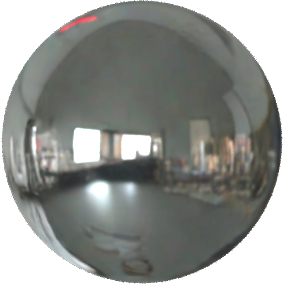}} & 
            
            \\
    
            &
            \noindent\parbox[c]{0.080\textwidth}{\includegraphics[width=0.080\textwidth]{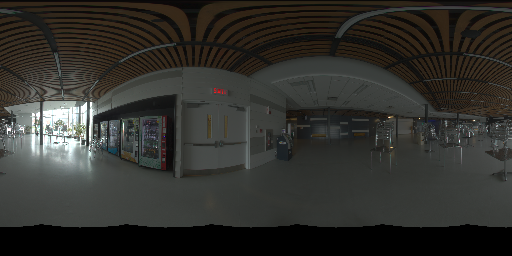}} & 
            \noindent\parbox[c]{0.080\textwidth}{\includegraphics[width=0.080\textwidth]{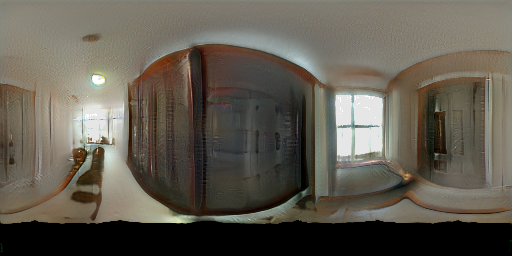}} & 
            \noindent\parbox[c]{0.080\textwidth}{\includegraphics[width=0.080\textwidth]{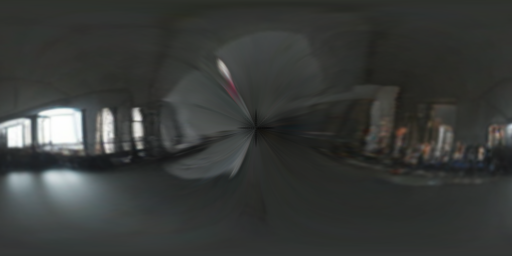}} &
            \noindent\parbox[c]{0.080\textwidth}{\includegraphics[width=0.080\textwidth]{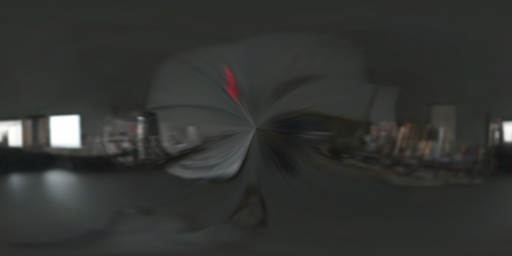}} & 
            \\
    
    
            \noindent\parbox[c]{0.080\textwidth}{\includegraphics[width=0.080\textwidth]{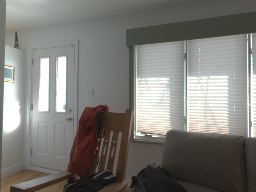}} 
            
            &
            \noindent\parbox[c]{0.080\textwidth}{\includegraphics[width=0.080\textwidth]{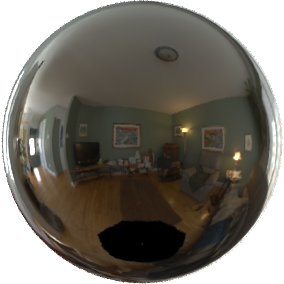}} & 
            \noindent\parbox[c]{0.080\textwidth}{\includegraphics[width=0.080\textwidth]{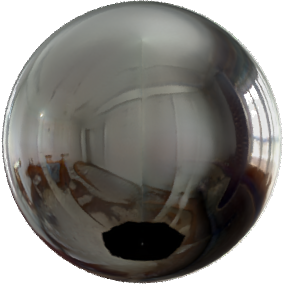}} &  
            \noindent\parbox[c]{0.080\textwidth}{\includegraphics[width=0.080\textwidth]{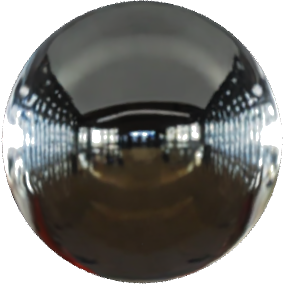}} &
           
            \noindent\parbox[c]{0.080\textwidth}{\includegraphics[width=0.080\textwidth]{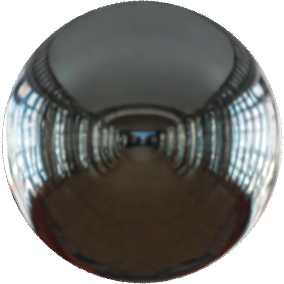}} & 
            
            \\
    
            &
            \noindent\parbox[c]{0.080\textwidth}{\includegraphics[width=0.080\textwidth]{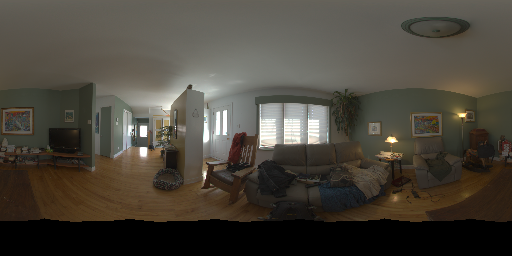}} & 
            \noindent\parbox[c]{0.080\textwidth}{\includegraphics[width=0.080\textwidth]{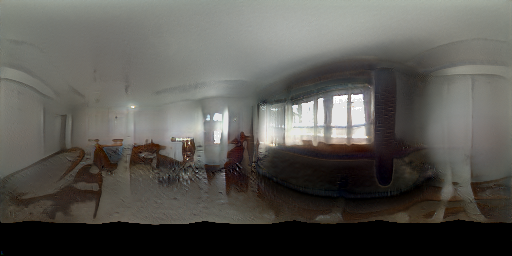}} & 
            \noindent\parbox[c]{0.080\textwidth}{\includegraphics[width=0.080\textwidth]{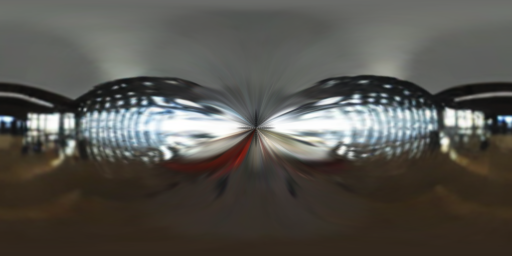}} &
            \noindent\parbox[c]{0.080\textwidth}{\includegraphics[width=0.080\textwidth]{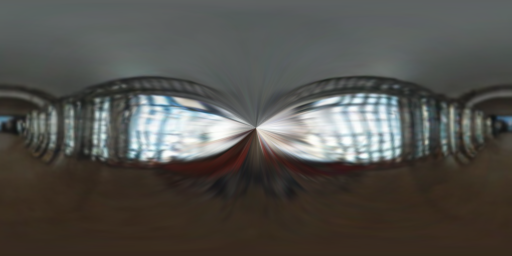}} & 
            \\
    
        \end{tabu}
    }

    \subcaptionbox{Poly Haven \cite{polyhaven} \label{fig:qualitative_polyhaven}}{
        \begin{tabu} to \textwidth {
            @{}
            c@{\hspace{2pt}}
            c@{\hspace{2pt}}
            c@{\hspace{2pt}}
            c@{\hspace{2pt}}
            c@{\hspace{2pt}}
            c@{}
        }

            \hline
    
            \noindent\parbox[c]{0.080\textwidth}{\includegraphics[width=0.080\textwidth]{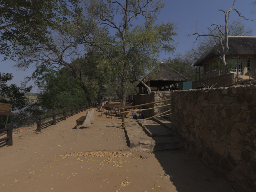}} 
            
            &
            \noindent\parbox[c]{0.080\textwidth}{\includegraphics[width=0.080\textwidth]{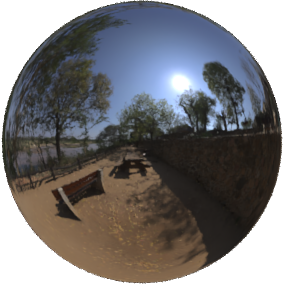}} & 
            \noindent\parbox[c]{0.080\textwidth}{\includegraphics[width=0.080\textwidth]{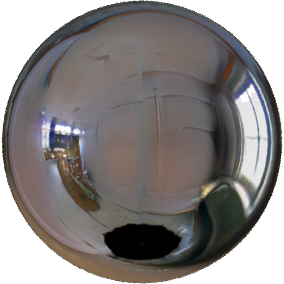}} &  
            \noindent\parbox[c]{0.080\textwidth}{\includegraphics[width=0.080\textwidth]{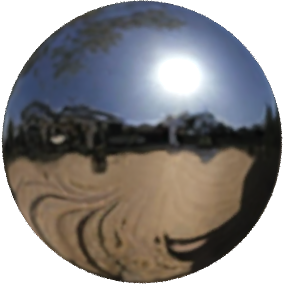}} &
           
            \noindent\parbox[c]{0.080\textwidth}{\includegraphics[width=0.080\textwidth]{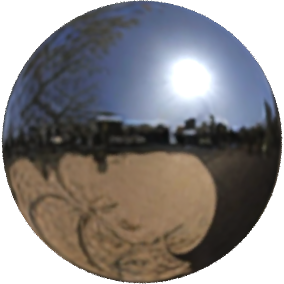}} & 
            
            \\
    
            &
            \noindent\parbox[c]{0.080\textwidth}{\includegraphics[width=0.080\textwidth]{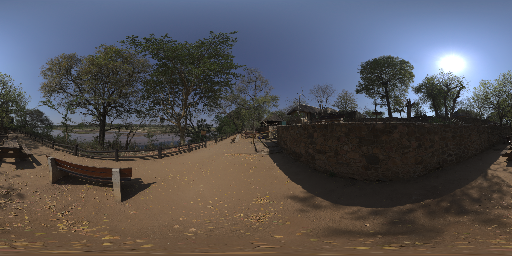}} & 
            \noindent\parbox[c]{0.080\textwidth}{\includegraphics[width=0.080\textwidth]{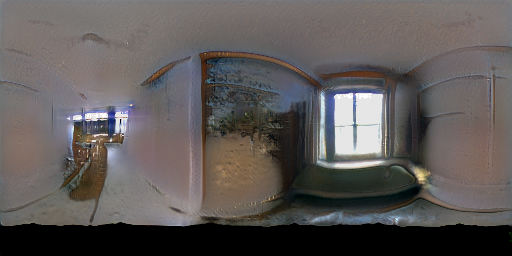}} & 
            \noindent\parbox[c]{0.080\textwidth}{\includegraphics[width=0.080\textwidth]{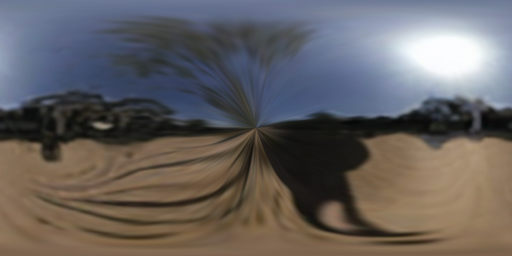}} &
            \noindent\parbox[c]{0.080\textwidth}{\includegraphics[width=0.080\textwidth]{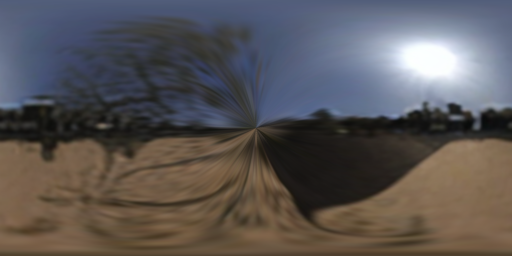}} & 
            \\
    
    
            \noindent\parbox[c]{0.080\textwidth}{\includegraphics[width=0.080\textwidth]{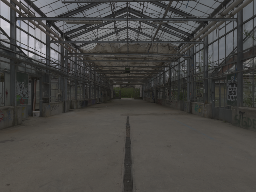}} 
            
            &
            \noindent\parbox[c]{0.080\textwidth}{\includegraphics[width=0.080\textwidth]{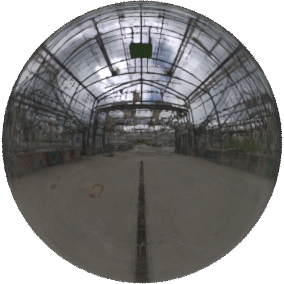}} & 
            \noindent\parbox[c]{0.080\textwidth}{\includegraphics[width=0.080\textwidth]{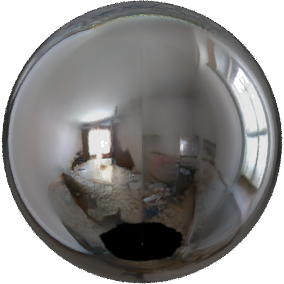}} &  
            \noindent\parbox[c]{0.080\textwidth}{\includegraphics[width=0.080\textwidth]{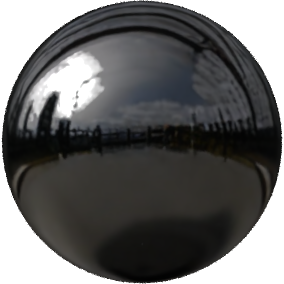}} &
           
            \noindent\parbox[c]{0.080\textwidth}{\includegraphics[width=0.080\textwidth]{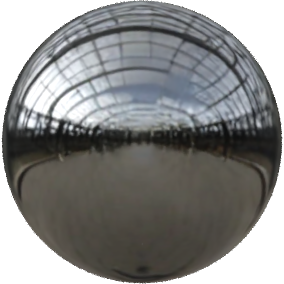}} & 
            
            \\
    
            &
            \noindent\parbox[c]{0.080\textwidth}{\includegraphics[width=0.080\textwidth]{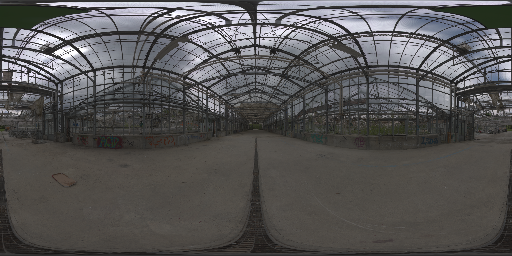}} & 
            \noindent\parbox[c]{0.080\textwidth}{\includegraphics[width=0.080\textwidth]{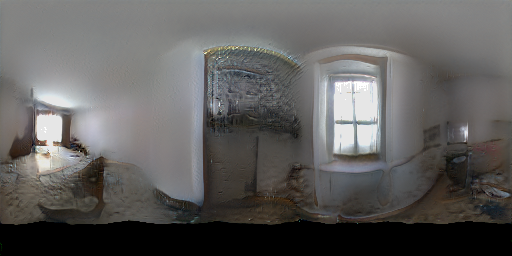}} & 
            \noindent\parbox[c]{0.080\textwidth}{\includegraphics[width=0.080\textwidth]{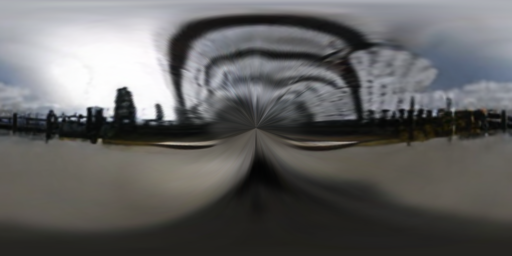}} &
            \noindent\parbox[c]{0.080\textwidth}{\includegraphics[width=0.080\textwidth]{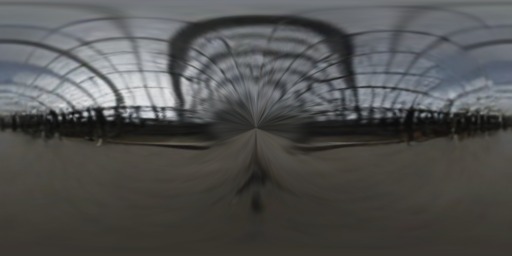}} & 
            \\
    
    
            \noindent\parbox[c]{0.080\textwidth}{\includegraphics[width=0.080\textwidth]{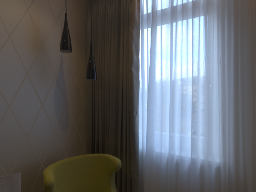}} 
            
            &
            \noindent\parbox[c]{0.080\textwidth}{\includegraphics[width=0.080\textwidth]{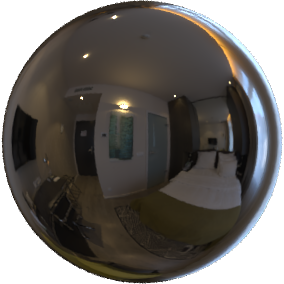}} & 
            \noindent\parbox[c]{0.080\textwidth}{\includegraphics[width=0.080\textwidth]{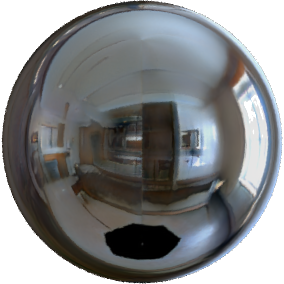}} &  
            \noindent\parbox[c]{0.080\textwidth}{\includegraphics[width=0.080\textwidth]{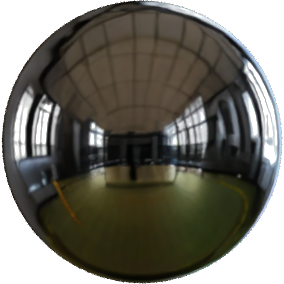}} &
           
            \noindent\parbox[c]{0.080\textwidth}{\includegraphics[width=0.080\textwidth]{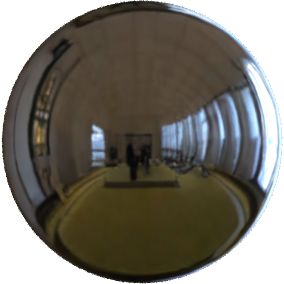}} & 
            
            \\
    
            &
            \noindent\parbox[c]{0.080\textwidth}{\includegraphics[width=0.080\textwidth]{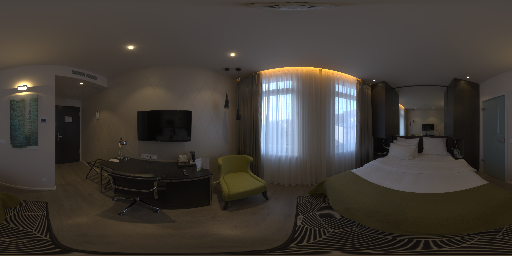}} & 
            \noindent\parbox[c]{0.080\textwidth}{\includegraphics[width=0.080\textwidth]{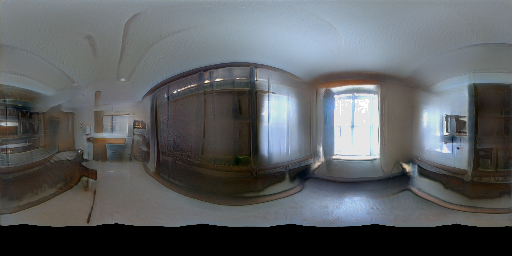}} & 
            \noindent\parbox[c]{0.080\textwidth}{\includegraphics[width=0.080\textwidth]{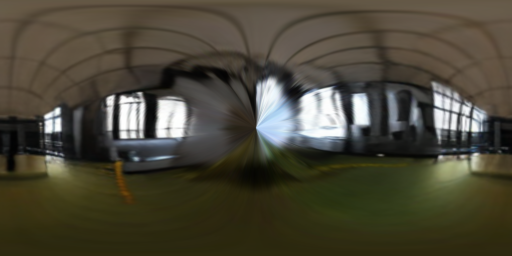}} &
            \noindent\parbox[c]{0.080\textwidth}{\includegraphics[width=0.080\textwidth]{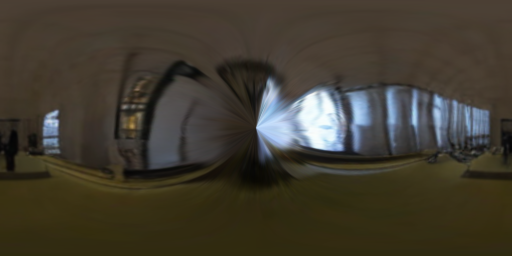}} & 
            \\
    
        \end{tabu}
        }

    \vspace{-0.5em}
    \caption{
    Qualitative results on benchmark datasets. For each input image, we show the rendered chrome ball (1\textsuperscript{st} row) and the corresponding environment map (2\textsuperscript{st} row) from each method. (I: iterative, LR: LoRA).
    }
    \label{fig:qualitative_benchmark}
    \vspace{-1.5em}
\end{figure}
\begin{table*}[!h]
\centering
\small
\resizebox{1\textwidth}{!}{%
\setlength{\tabcolsep}{1pt}
\begin{tabular}{
    l@{\hspace{9pt}}
    l@{\hspace{9pt}}
    c@{\hspace{9pt}}
    c@{\hspace{9pt}}
    c@{\hspace{9pt}}
    c@{\hspace{9pt}}
    c@{\hspace{9pt}}
    c@{\hspace{9pt}}
    c@{\hspace{9pt}}
    c@{\hspace{9pt}}
    c
}
\toprule
\multirow{2}{*}{\textbf{Dataset}} & \multirow{2}{*}{\textbf{Method}} & \multicolumn{3}{c}{\textbf{Scale-invariant RMSE} $\downarrow$} & \multicolumn{3}{c}{\textbf{Angular Error} $\downarrow$}  & \multicolumn{3}{c}{\textbf{Normalized RMSE} $\downarrow$}                                                            \\
 & & Diffuse           & Matte           & Mirror           & Diffuse           & Matte           & Mirror   & Diffuse           & Matte           & Mirror      \\ 
\midrule

Laval Indoor \cite{garder2017lavelindoor} & StyleLight {\footnotesize(reported by the paper)} & 0.11 & 0.29 & 0.55 & 2.41 & 2.96 & 4.30 & - & - & - \\

\hline

\multirow{5}{*}{Laval Indoor \cite{garder2017lavelindoor}}  & StyleLight {\footnotesize (reproduced using official code)} & 0.13 & 0.31 & 0.55 & 4.24 & 4.74 & 6.78 & \colorbox{tabsecond}{0.23} & \colorbox{tabsecond}{0.40} & 0.51 \\
& SDXL$^\dagger$ & 0.20 & 0.45 & 0.72 & 5.87 & 6.20 & 8.28 & 0.32 & 0.47 & 0.51 \\
& SDXL$^\dagger$ + iterative (ours) & 0.15 & 0.39 & 0.67 & 3.58 & 4.55 & 7.05 & 0.25 & 0.41 & 0.47 \\
& SDXL$^\dagger$ + LoRA (ours) & 0.15 & 0.38 & 0.65 & \colorbox{tabsecond}{3.47} & \colorbox{tabsecond}{3.86} & \colorbox{tabsecond}{6.15} & 0.25 & \colorbox{tabsecond}{0.40} & \colorbox{tabsecond}{0.45} \\
& SDXL$^\dagger$ + iterative + LoRA (ours) & \colorbox{tabsecond}{0.14} & \colorbox{tabsecond}{0.33} & \colorbox{tabsecond}{0.60} & \colorbox{tabfirst}{2.14} & \colorbox{tabfirst}{3.42} & \colorbox{tabfirst}{5.94} & \colorbox{tabfirst}{0.20} & \colorbox{tabfirst}{0.36} & \colorbox{tabfirst}{0.43} \\

\hline

\multirow{5}{*}{Poly Haven \cite{polyhaven}} & StyleLight {\footnotesize (reproduced using official code)} & 0.17 & \colorbox{tabfirst}{0.44} & \colorbox{tabfirst}{0.64} & 3.53 & 4.44 & 7.12 & 0.23 & \colorbox{tabsecond}{0.41} & 0.49 \\
& SDXL$^\dagger$ & 0.22 & 0.59 & 0.80 & 2.98 & 4.25 & 5.74 & 0.30 & 0.50 & 0.54 \\
& SDXL$^\dagger$ + iterative (ours) & \colorbox{tabsecond}{0.16} & 0.50 & 0.73 & 2.57 & 4.18 & 5.31 & \colorbox{tabsecond}{0.22} & 0.44 & 0.50 \\
& SDXL$^\dagger$ + LoRA (ours) & \colorbox{tabsecond}{0.16} & 0.50 & 0.72 & \colorbox{tabsecond}{2.35} & \colorbox{tabfirst}{3.56} & \colorbox{tabsecond}{4.43} & 0.24 & 0.43 & \colorbox{tabsecond}{0.47} \\
& SDXL$^\dagger$ + iterative + LoRA (ours) & \colorbox{tabfirst}{0.14} & \colorbox{tabsecond}{0.45} & \colorbox{tabsecond}{0.66} & \colorbox{tabfirst}{2.14} & \colorbox{tabsecond}{3.60} & \colorbox{tabfirst}{4.29} & \colorbox{tabfirst}{0.20} & \colorbox{tabfirst}{0.39} & \colorbox{tabfirst}{0.43} \\

\bottomrule
\end{tabular}}
\vspace{-0.5em}
\caption{Comparison using the three-sphere evaluation protocol between StyleLight \cite{wang2022stylelight}, simple inpainting with SDXL \cite{podell2023sdxl} and depth-conditioned ControlNet \cite{zhang2023adding}  (``SDXL$^\dagger$'' in the table), and ablated versions of our method. The \colorbox{tabfirst}{best} and \colorbox{tabsecond}{second-best}are color coded.}
\label{tab:indoor_stylelight}
\vspace{-1.5em}
\end{table*}
\begin{table}[!h]
\centering
\resizebox{0.95\columnwidth}{!}{%
\setlength{\tabcolsep}{12pt}
\begin{tabular}{lcc
}
\toprule
\textbf{Method} & \textbf{si-RMSE} $\downarrow$ & \textbf{Angular Error} $\downarrow$ \\
\midrule
EverLight \cite{dastjerdi2023everlight} & 0.091 & 6.36 \\
StyleLight \cite{wang2022stylelight} & 0.123 & 7.09 \\
Weber et al. \cite{weber2022editableindoor} & \scfirst{0.081} & \scsecond{4.13} \\
EMLight \cite{zhan2021emlight} & 0.099 & \scfirst{3.99} \\
\textbf{Ours} & \scsecond{0.090} & 5.25 \\

\bottomrule
\end{tabular}}
\vspace{-0.6em}
\caption{Scores on indoor array-of-spheres protocol (Section~\ref{sec:eval_benchmark}) 
}
\label{tab:indoor_everlight}
\vspace{-0.5em}
\end{table}
\begin{table}[]
\centering
\small
\begin{tabular}{
    l@{\hspace{5pt}}
    l@{\hspace{5pt}}
    c@{\hspace{5pt}}
    c@{\hspace{5pt}}
    c
}
\toprule
\textbf{Sphere} & \textbf{Method} & \textbf{si-RMSE} $\downarrow$ & \begin{tabular}[c]{@{}c@{}}\textbf{Angular}\\ \textbf{Error}\end{tabular} $\downarrow$ & \begin{tabular}[c]{@{}c@{}}\textbf{Normalized}\\ \textbf{RMSE}\end{tabular} $\downarrow$ \\
\midrule

Diffuse & StyleLight & 0.143 & 3.741 & 0.236\\
        & \textbf{Ours} & \colorbox{tabfirst}{0.135} & \colorbox{tabfirst}{2.337} & \colorbox{tabfirst}{0.219} \\
\hline
Matte   & StyleLight & \colorbox{tabfirst}{0.347} & 4.492 & 0.429 \\
        & \textbf{Ours} & 0.359 & \colorbox{tabfirst}{3.483} & \colorbox{tabfirst}{0.369} \\
  \hline
Mirror & StyleLight & \colorbox{tabfirst}{0.606} & 7.655 & 0.544\\
 & \textbf{Ours} & 0.644 & \colorbox{tabfirst}{5.988} & \colorbox{tabfirst}{0.438} \\

\bottomrule
\end{tabular}
\vspace{-0.6em}
\caption{Scores on the random-camera protocol (Section~\ref{sec:random-camera-protocol}).
}
\label{tab:randfov}
\vspace{-0.5em}
\end{table}
\begin{table}[!h]
\centering
\small
\begin{tabular}{
    l@{\hspace{5pt}}
    l@{\hspace{5pt}}
    c@{\hspace{5pt}}
    c@{\hspace{5pt}}
    c
}
\toprule
\textbf{Dataset} & \textbf{Method} & \textbf{RMSE} $\downarrow$ & \textbf{si-RMSE} $\downarrow$ & \begin{tabular}[c]{@{}c@{}}\textbf{Angular}\\ \textbf{Error}\end{tabular} $\downarrow$\\
\midrule

Laval Indoor \cite{garder2017lavelindoor} & StyleLight & 0.246 & \colorbox{tabfirst}{0.271} & 5.814\\
 & \textbf{Ours} & \colorbox{tabfirst}{0.187} & 0.303 & \colorbox{tabfirst}{4.412} \\
 \hline
 Poly Haven  \cite{polyhaven}& StyleLight & 0.241 & 0.324 & 6.291 \\
 & \textbf{Ours} & \colorbox{tabfirst}{0.179} & \colorbox{tabfirst}{0.275} & \colorbox{tabfirst}{4.567} \\

\bottomrule
\end{tabular}
\vspace{-0.6em}
\caption{Scores on LDR environment maps (Section~\ref{sec:ldr-panoramas-eval}). 
}
\label{tab:ldr-stylelight}
\vspace{-1.3em}
\end{table}

\ifpurecompactformat
\vspace{-0.2cm}
\fi
\subsection{Evaluation on benchmark datasets} \label{sec:eval_benchmark}
\ifpurecompactformat
\vspace{-0.2cm}
\fi

We adopt two different evaluation protocols used in the literature: from each input LDR image, we generate an HDR panorama of size $128 \times 256$ pixels and use it to render (1) three spheres with different materials (gray-diffuse, silver-matte, and silver-mirror spheres) \cite{wang2022stylelight, gardner2019deepparam, garder2017lavelindoor} or (2) an array of diffuse spheres \cite{dastjerdi2023everlight, weber2022editableindoor}. Then, we computed the evaluation metrics on these renderings. Many studies do not publish their source code and use only one of the protocols, resulting in missing baselines' scores in some experiments.


\ifpurecompactformat
\vspace{-0.1cm}
\fi
\myparagraph{Evaluation on three spheres.} We compared our method to StyleLight \cite{wang2022stylelight} on (1) 289 panoramas from the Laval Indoor dataset and (2) 500 panoramas from Poly Haven dataset. 
It is important to note that StyleLight was trained on the Laval \emph{Indoor} dataset; its scores on Poly Haven are provided solely as a reference to demonstrate how existing methods perform in out-of-distribution scenarios.
Following StyleLight's protocol, we created one input image from each panorama by cropping it to a size of $192 \times 256$ with a vertical FOV of 60\degree and then applying tone-mapping, setting the 99th percentile to 0.9 and using $\gamma = 2.4$. In only our pipeline, we upscale the image while keeping the aspect ratio for SDXL.



Table \ref{tab:indoor_stylelight} shows that our method outperforms StyleLight in terms of Angular Error and Normalized RMSE on Laval indoor dataset, with significantly lower Angular Error: 49.5\% (diffuse), 27.8\% (matte), and 12.4\% (mirror). Our method is also effective in Poly Haven outdoor scenes, while StyleLight's performance drops with a large 39.7\% gap in Angular Error for mirror spheres. Qualitative results are in Figure \ref{fig:qualitative_benchmark} and Appendix \ref{appendix:more_result_benchmark}. 
Note that we used StyleLight's official code to produce these scores; however, discrepancies exist with those reported in the paper. (See Appendix \ref{appendix:stylelight-score-diff} for details and our discussion with StyleLight's authors on this issue).

\ifpurecompactformat
\vspace{-0.1cm}
\fi
\myparagraph{Evaluation on array of spheres.} We compared our approach with StyleLight, Everlight \cite{dastjerdi2023everlight}, EMLight \cite{zhan2021emlight}, and Weber et al. \cite{weber2022editableindoor}. 
We used 224 panoramas (the same ones used to evaluate Everlight) from the Laval Indoor dataset. For each panorama, we generated 10 input LDR images by centering the panorama at certain azimuthal angles and cropping it to $50^\circ$ FOV, following Weber et al. \cite{weber2022editableindoor}. As a result, the metrics were computed from 2240 input-output pairs. 
In Table \ref{tab:indoor_everlight}, our method ranks after Weber et al. and EMLight; however, it outperforms Everlight and StyleLight, despite not being explicitly trained on the dataset.
\ifpurecompactformat
\vspace{-0.2cm}
\fi
\subsection{Evaluation on unknown camera parameters} \label{sec:random-camera-protocol}
\ifpurecompactformat
\vspace{-0.2cm}
\fi

Evaluation protocols in the last section crop HDR panoramas at fixed camera angles and FOVs. In real-world settings, however, we often do not know the camera parameters of a photograph. This evaluation considers more challenging scenarios that reflect this situation better. In particular, to generate an input LDR image, we randomly sample the FOV from the interval $[30^\circ, 150^\circ]$, the elevation from $[-45^\circ, 45^\circ]$, and the azimuth from all $360^\circ$. We then crop an HDR panorama accordingly. We generate one LDR image from 289 HDR panoramas of the Laval Indoor dataset and compare our method with Stylelight using the generated input-output pairs and the three-sphere protocol. Table \ref{tab:randfov} shows that our method outperforms StyleLight in Angular Error and Normalized RMSE and remains competitive in si-RMSE. 

\ifpurecompactformat
\vspace{-0.2cm}
\fi
\subsection{Evaluation on LDR panoramas} \label{sec:ldr-panoramas-eval}
\ifpurecompactformat
\vspace{-0.2cm}
\fi
We demonstrate that our method can generate more plausible panoramas than SOTA approaches. In this evaluation, we generated one LDR input image from each panorama in the test dataset as in the three-sphere protocol. However, we compared output LDR panoramas \emph{directly} to the ground truth. Again, since each dataset uses its own brightness scale unknown to our method, comparison was done in the LDR image domain, where the scale is explicitly defined. Specifically, we compared our panoramas to those from StyleLight at a resolution of $256 \times 512$ after performing tone-mapping as described in \cite{wang2022stylelight}. Table \ref{tab:ldr-stylelight} shows that our method outperforms StyleLight with respect to nearly all metrics and datasets, suggesting that our method can leverage the strong generative prior in pre-trained diffusion models.

\tabulinesep=0.5pt
\begin{figure*}[!t]
    \centering

        \begin{tabu} to \textwidth {
        @{}
        c@{\hspace{7pt}}
        c@{\hspace{7pt}}
        c@{\hspace{7pt}}
        c@{\hspace{1pt}}
        c@{\hspace{1pt}}
        c@{\hspace{7pt}}
        c@{\hspace{1pt}}
        c@{\hspace{1pt}}
        c@{}
    }

        \multicolumn{1}{c}{\shortstack{\scriptsize StyleLight \cite{wang2022stylelight}}}
        & 
        \multicolumn{1}{c}{\shortstack{\scriptsize \hspace{-7pt} Ours (inpainted image)}}
        & 
        \multicolumn{1}{c}{\shortstack{\scriptsize \hspace{-3pt} Prediction \#1}}
        & 
        \multicolumn{1}{c}{\shortstack{\scriptsize \hspace{-1pt} EV-0.5}} &
        \multicolumn{1}{c}{\shortstack{\scriptsize \hspace{-1pt} EV-1.6}} &
        \multicolumn{1}{c}{\shortstack{\scriptsize \hspace{-7pt} EV-3.9}}
        &
        \multicolumn{1}{c}{\shortstack{\scriptsize \hspace{-1pt} Prediction \#2}} &
        \multicolumn{1}{c}{\shortstack{\scriptsize \hspace{-1pt} Prediction \#3}}
        \\

        \noindent\parbox[c]{0.105\textwidth}{\includegraphics[height=0.105\textwidth]{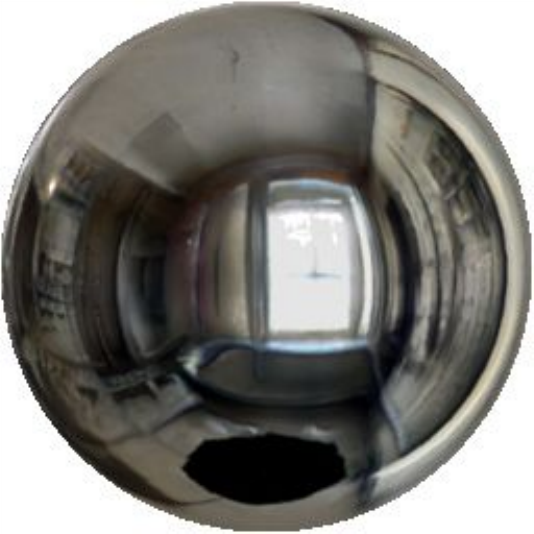}} & 
        \noindent\parbox[l]{0.16\textwidth}{\includegraphics[height=0.105\textwidth]{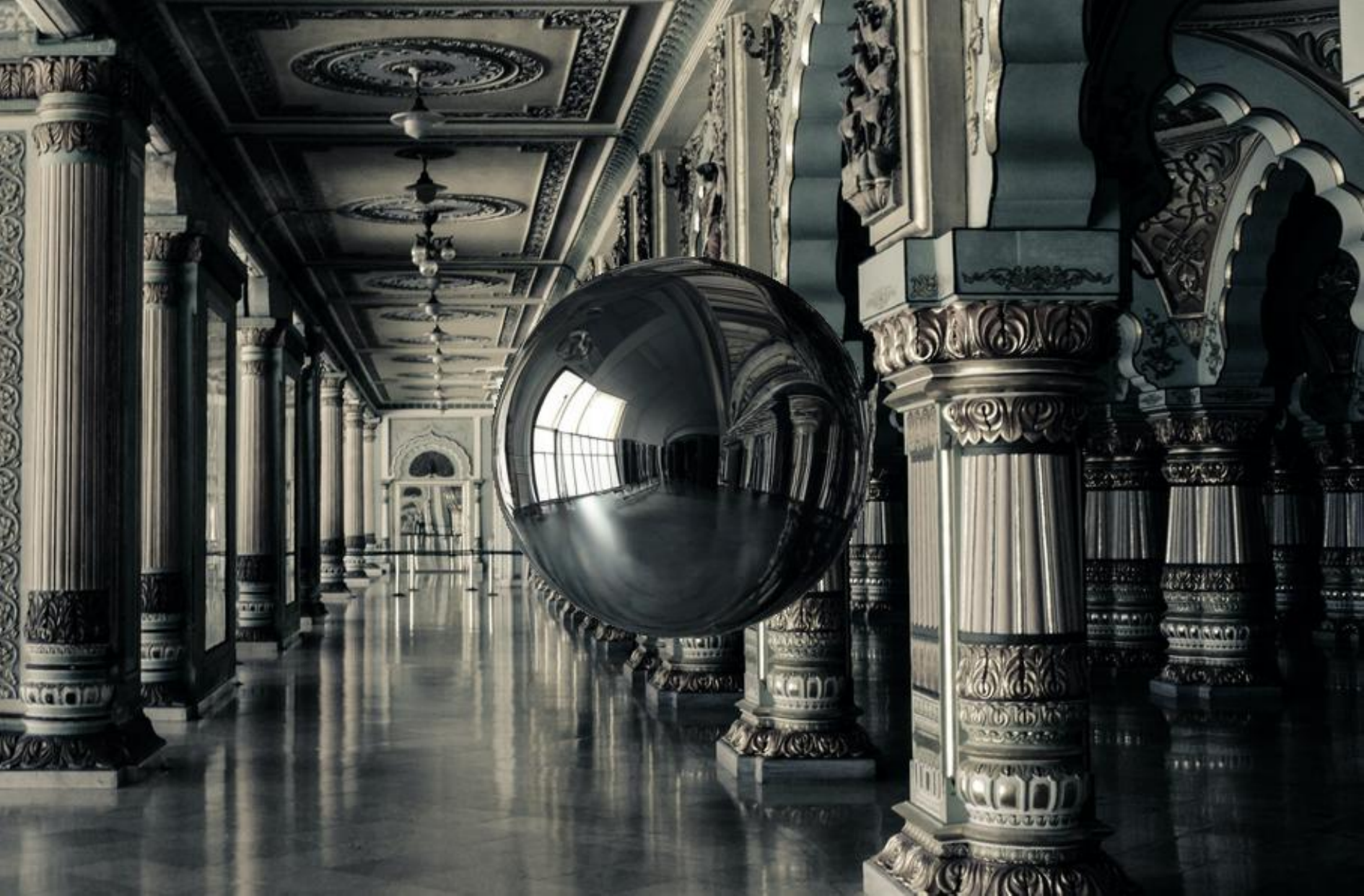}} &  
        \noindent\parbox[c]{0.105\textwidth}{\includegraphics[height=0.105\textwidth]{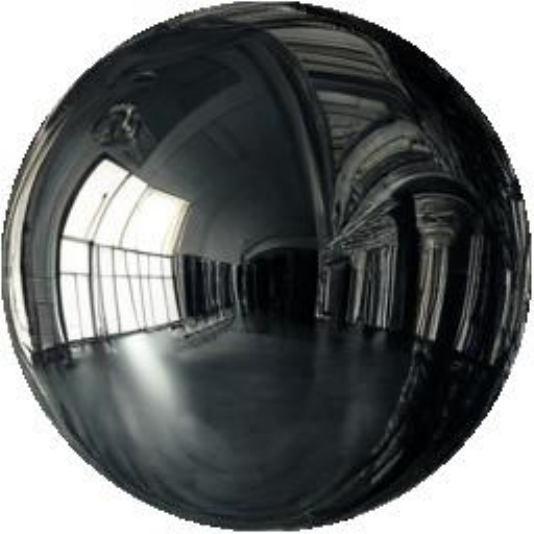}} & 
        
        \noindent\parbox[c]{0.105\textwidth}{\includegraphics[height=0.105\textwidth]{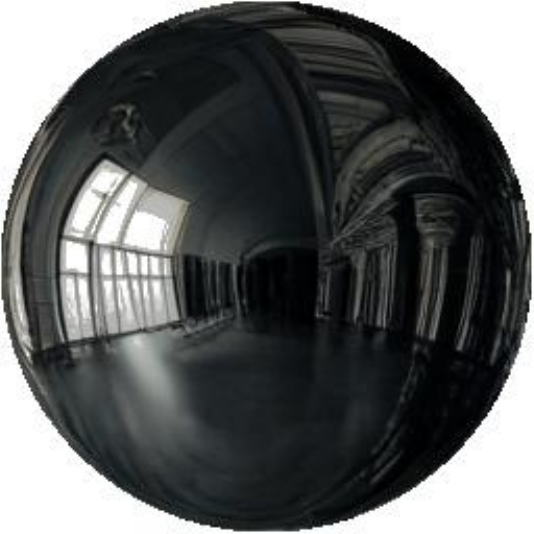}} & 
        \noindent\parbox[c]{0.105\textwidth}{\includegraphics[height=0.105\textwidth]{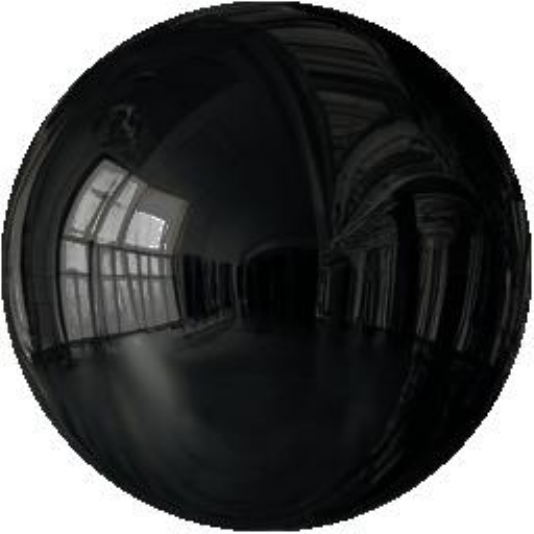}} & 
        \noindent\parbox[c]{0.105\textwidth}{\includegraphics[height=0.105\textwidth]{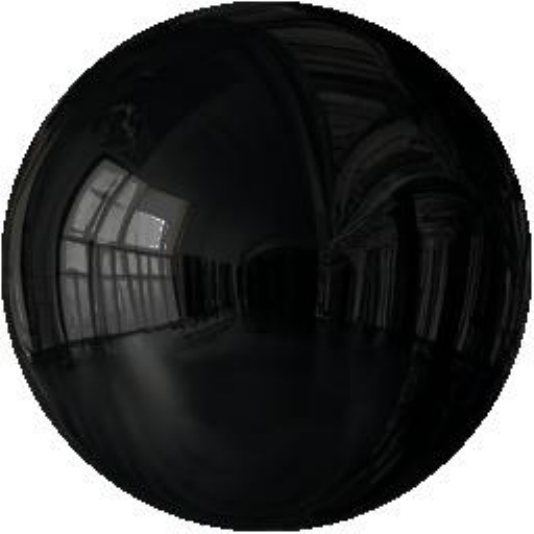}} &

        \noindent\parbox[c]{0.105\textwidth}{\includegraphics[height=0.105\textwidth]{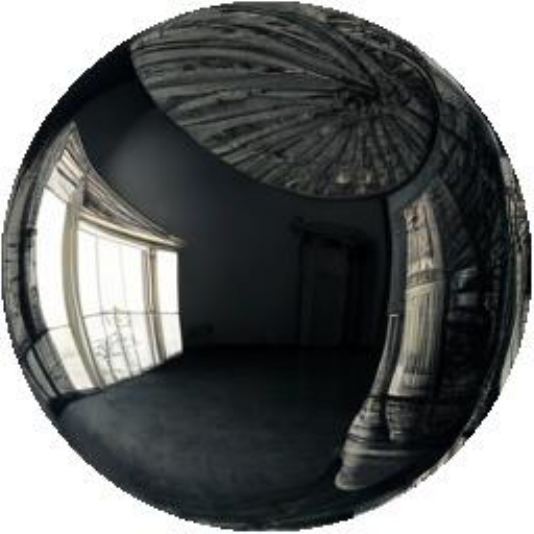}} & 
        \noindent\parbox[c]{0.105\textwidth}{\includegraphics[height=0.105\textwidth]{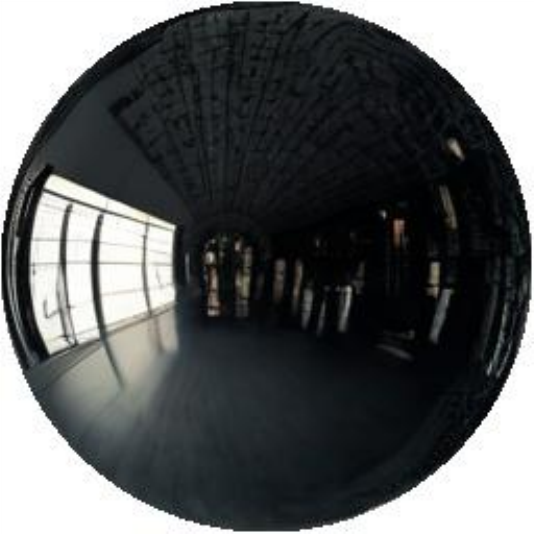}} & 
        
        \\

        \noindent\parbox[c]{0.105\textwidth}{\includegraphics[height=0.105\textwidth]{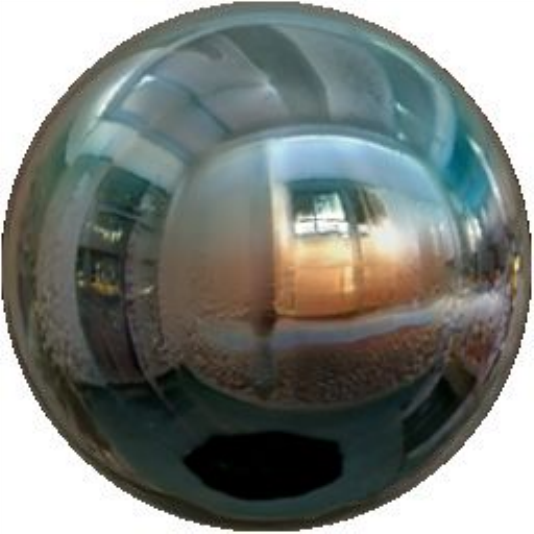}} & 
        \noindent\parbox[l]{0.16\textwidth}{\includegraphics[height=0.105\textwidth]{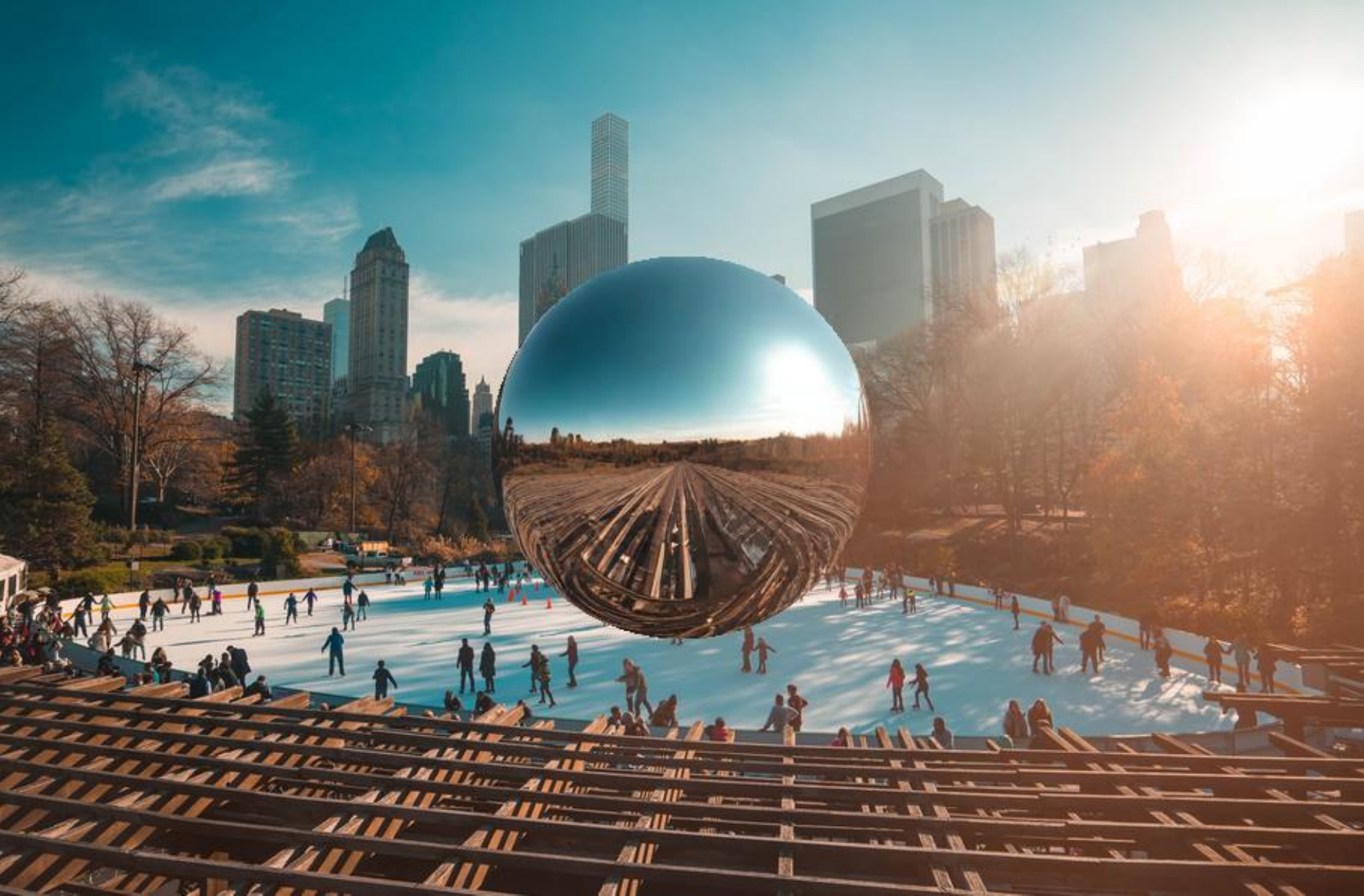}} &  
        \noindent\parbox[c]{0.105\textwidth}{\includegraphics[height=0.105\textwidth]{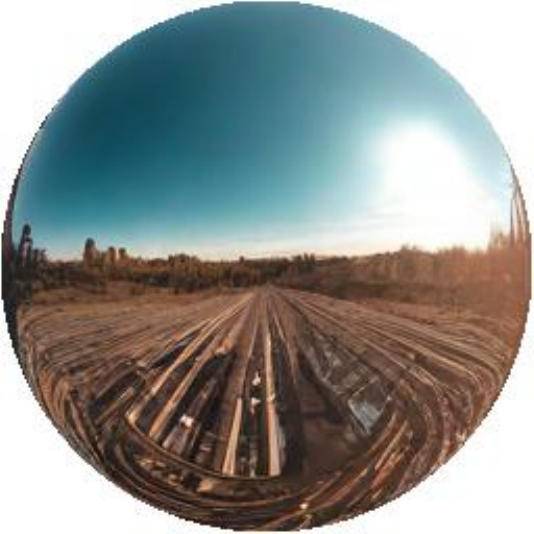}} & 
        
        \noindent\parbox[c]{0.105\textwidth}{\includegraphics[height=0.105\textwidth]{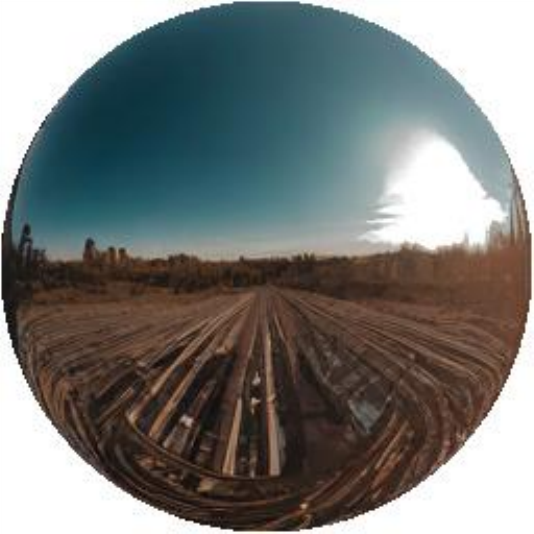}} & 
        \noindent\parbox[c]{0.105\textwidth}{\includegraphics[height=0.105\textwidth]{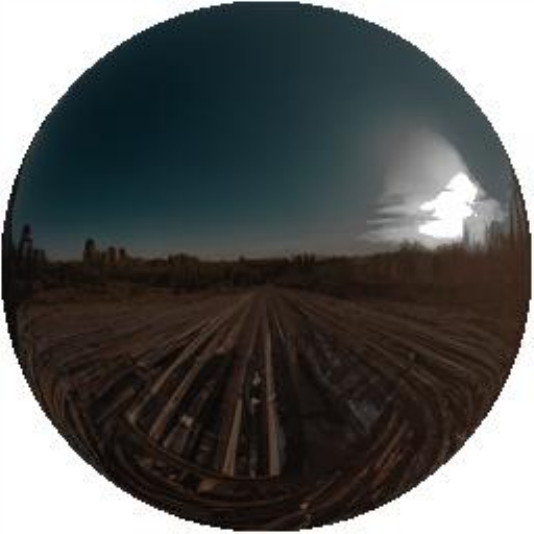}} & 
        \noindent\parbox[c]{0.105\textwidth}{\includegraphics[height=0.105\textwidth]{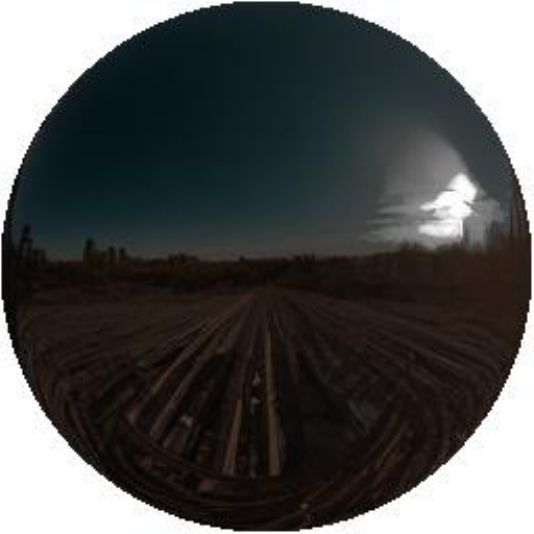}} & 

        \noindent\parbox[c]{0.105\textwidth}{\includegraphics[height=0.105\textwidth]{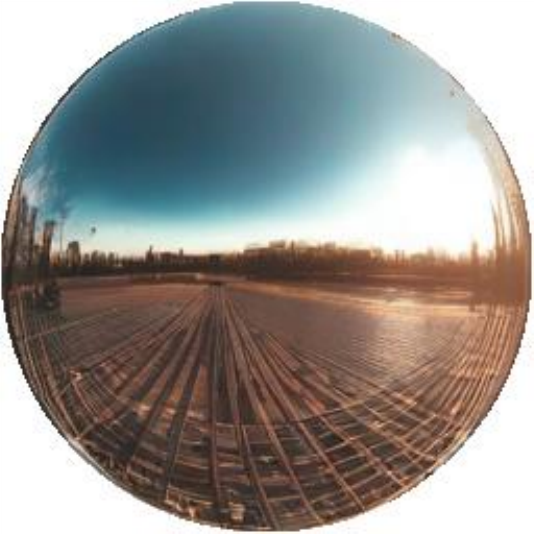}} & 
        \noindent\parbox[c]{0.105\textwidth}{\includegraphics[height=0.105\textwidth]{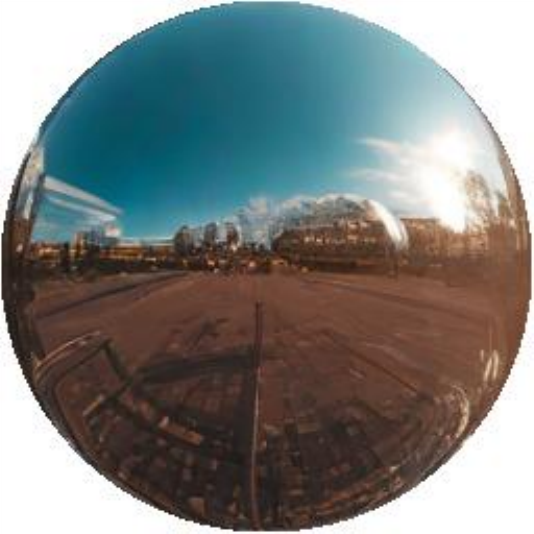}} & 
        \\

        \noindent\parbox[c]{0.105\textwidth}{\includegraphics[height=0.105\textwidth]{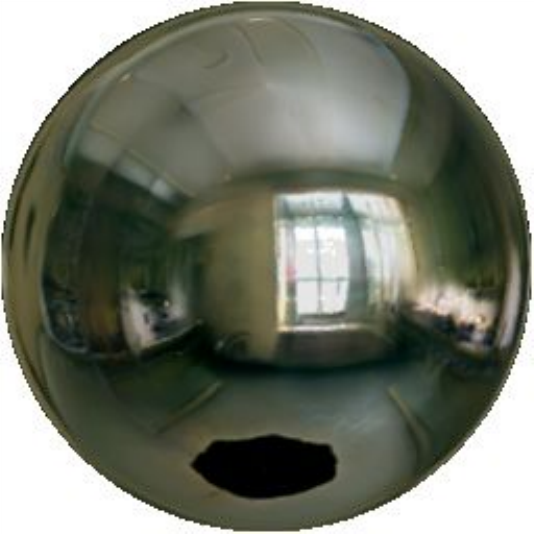}} & 
        \noindent\parbox[l]{0.16\textwidth}{\includegraphics[height=0.105\textwidth]{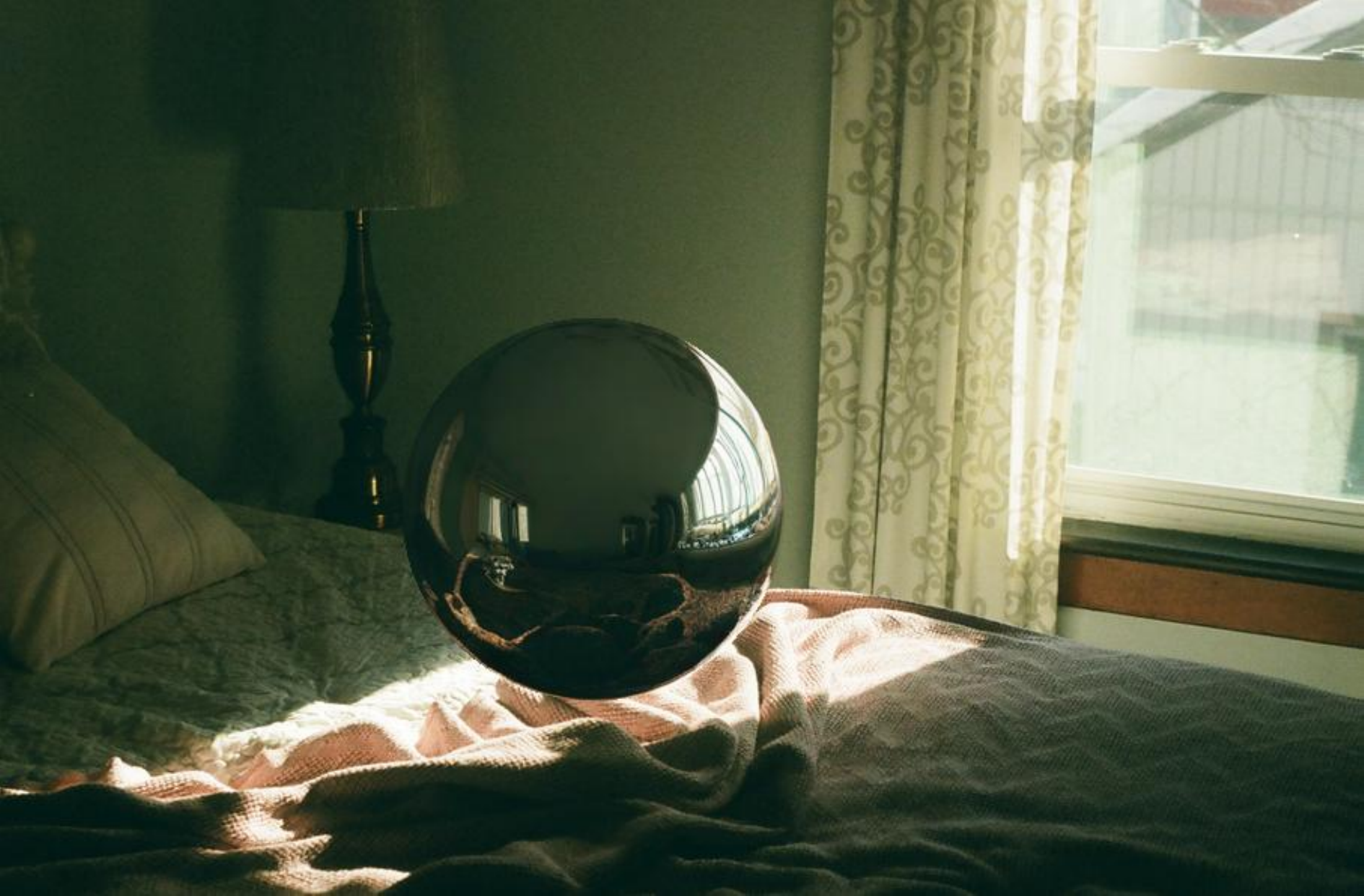}} &  
        \noindent\parbox[c]{0.105\textwidth}{\includegraphics[height=0.105\textwidth]{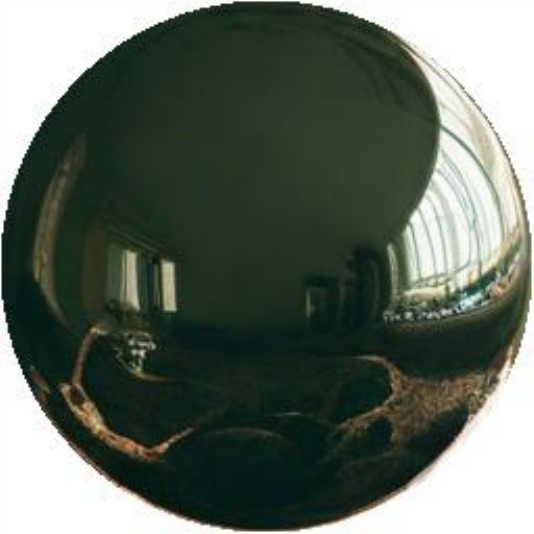}} & 
        
        \noindent\parbox[c]{0.105\textwidth}{\includegraphics[height=0.105\textwidth]{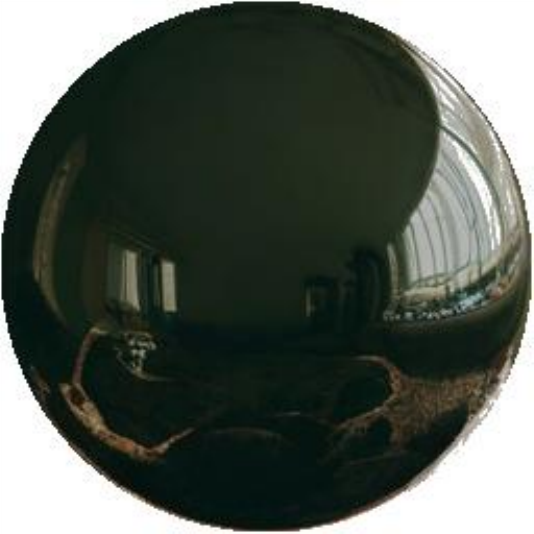}} & 
        \noindent\parbox[c]{0.105\textwidth}{\includegraphics[height=0.105\textwidth]{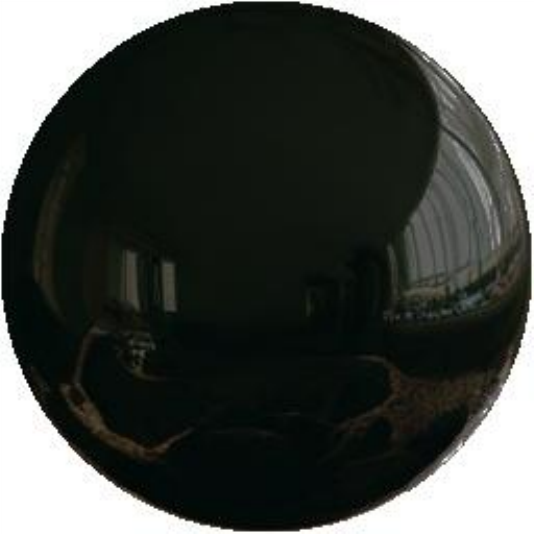}} & 
        \noindent\parbox[c]{0.105\textwidth}{\includegraphics[height=0.105\textwidth]{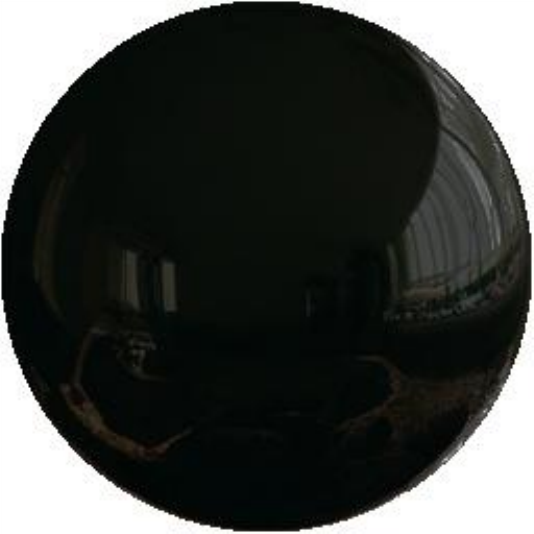}} &

        \noindent\parbox[c]{0.105\textwidth}{\includegraphics[height=0.105\textwidth]{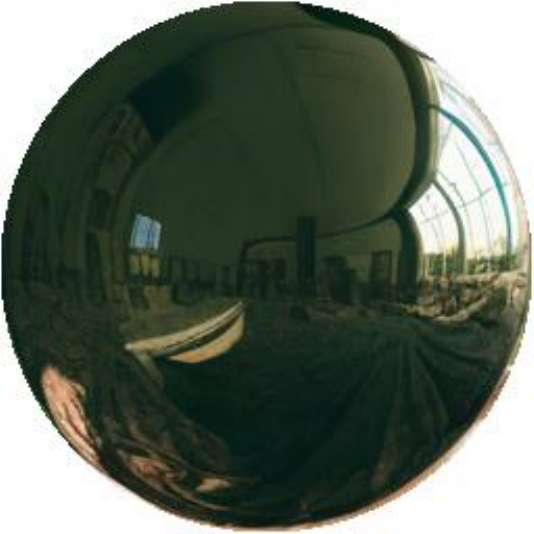}} & 
        \noindent\parbox[c]{0.105\textwidth}{\includegraphics[height=0.105\textwidth]{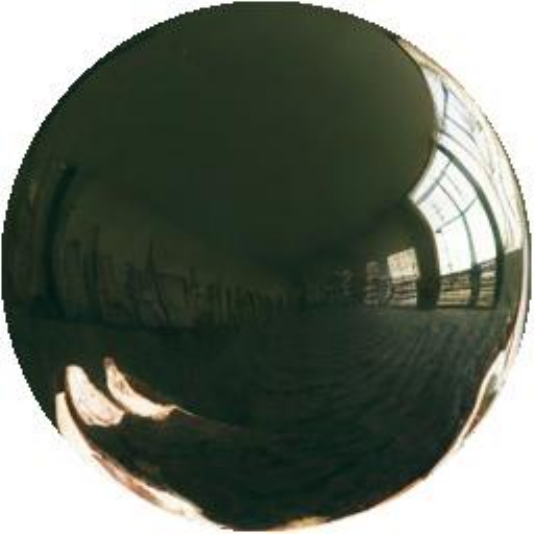}} & 
        \\

        \noindent\parbox[c]{0.105\textwidth}{\includegraphics[height=0.105\textwidth]{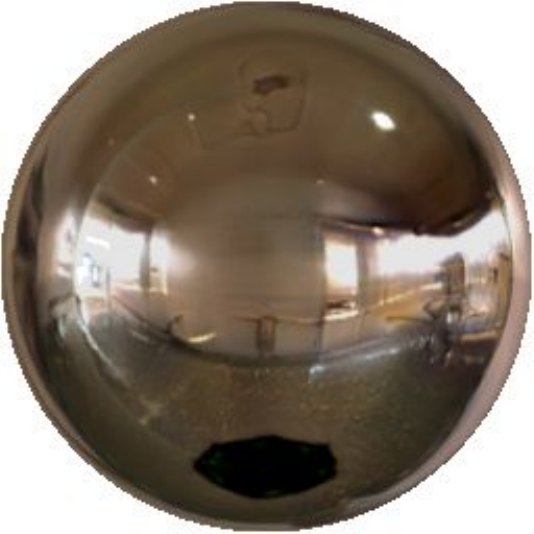}} & 
        \noindent\parbox[l]{0.16\textwidth}{\includegraphics[height=0.105\textwidth]{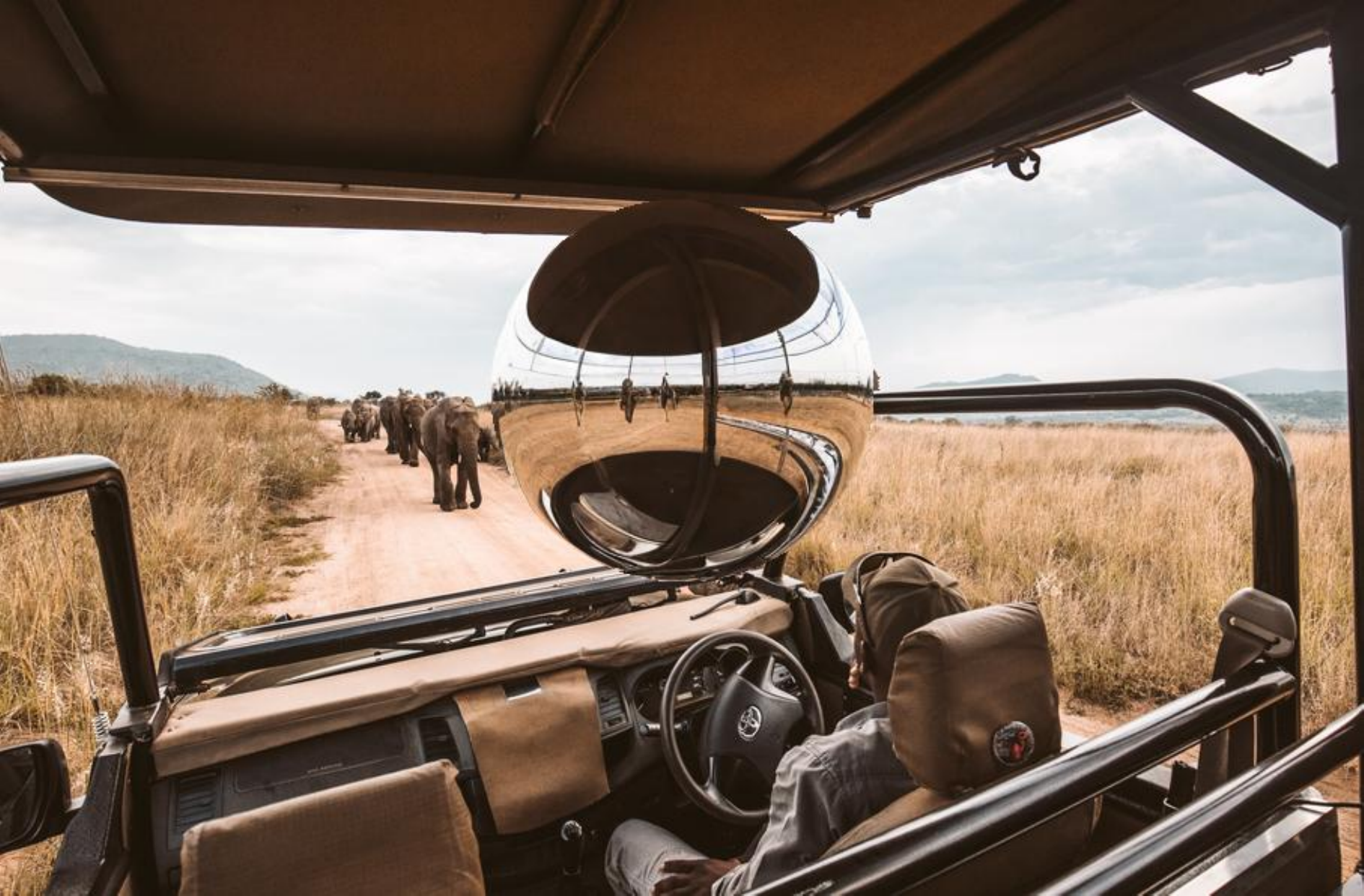}} &  
        \noindent\parbox[c]{0.105\textwidth}{\includegraphics[height=0.105\textwidth]{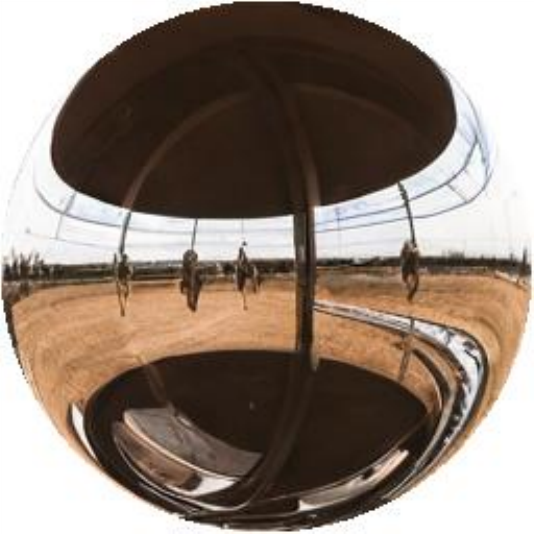}} & 
        
        \noindent\parbox[c]{0.105\textwidth}{\includegraphics[height=0.105\textwidth]{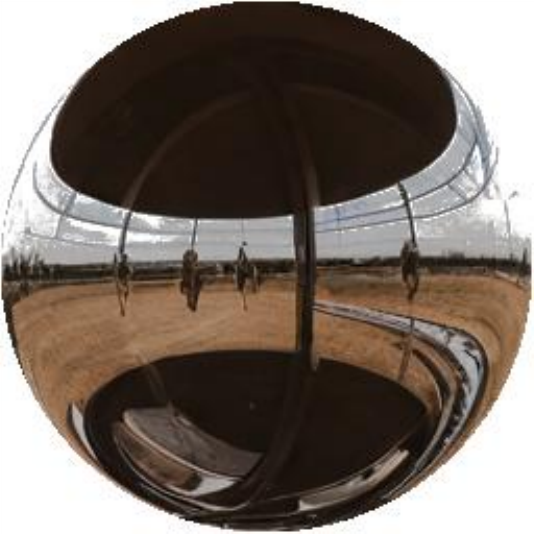}} & 
        \noindent\parbox[c]{0.105\textwidth}{\includegraphics[height=0.105\textwidth]{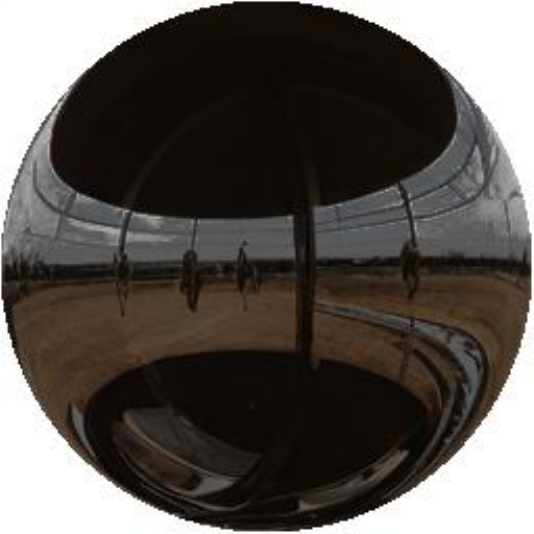}} & 
        \noindent\parbox[c]{0.105\textwidth}{\includegraphics[height=0.105\textwidth]{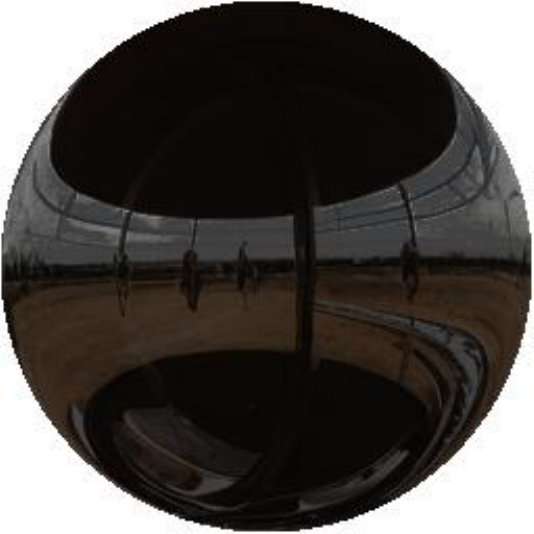}} &

        \noindent\parbox[c]{0.105\textwidth}{\includegraphics[height=0.105\textwidth]{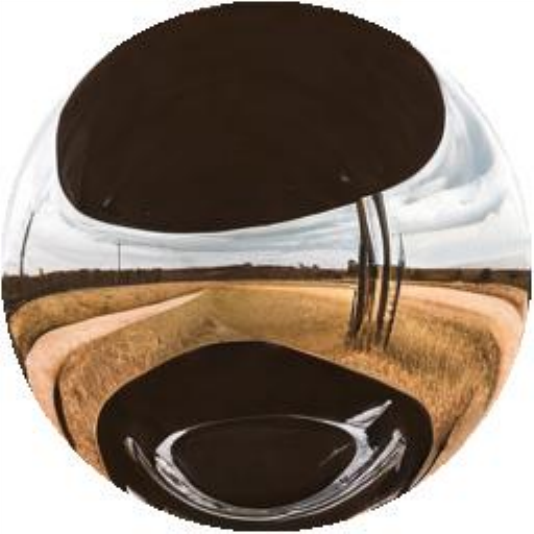}} & 
        \noindent\parbox[c]{0.105\textwidth}{\includegraphics[height=0.105\textwidth]{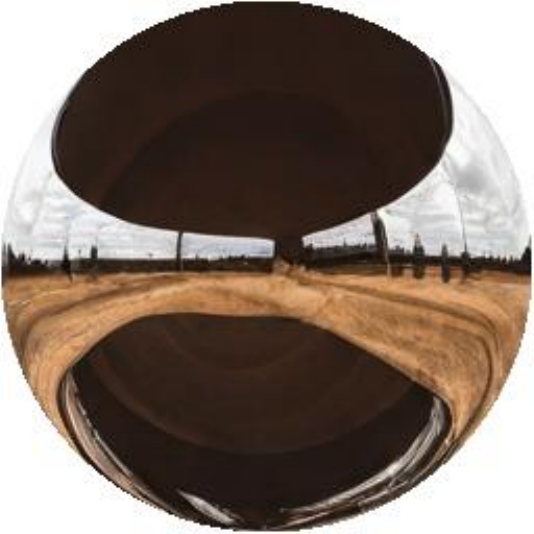}} & 
        \\

        \noindent\parbox[c]{0.105\textwidth}{\includegraphics[height=0.105\textwidth]{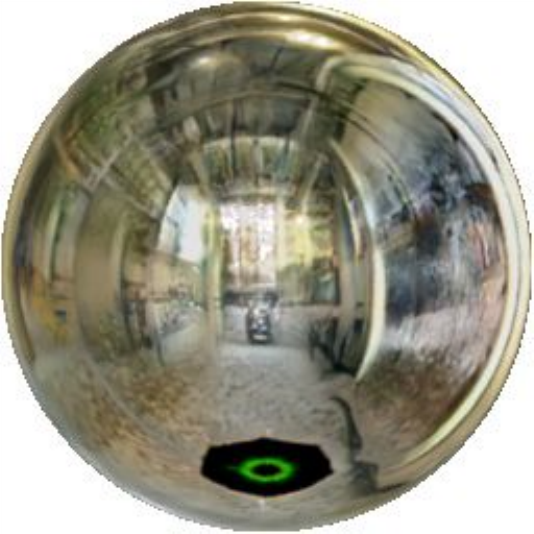}} & 
        \noindent\parbox[l]{0.16\textwidth}{\includegraphics[height=0.105\textwidth]{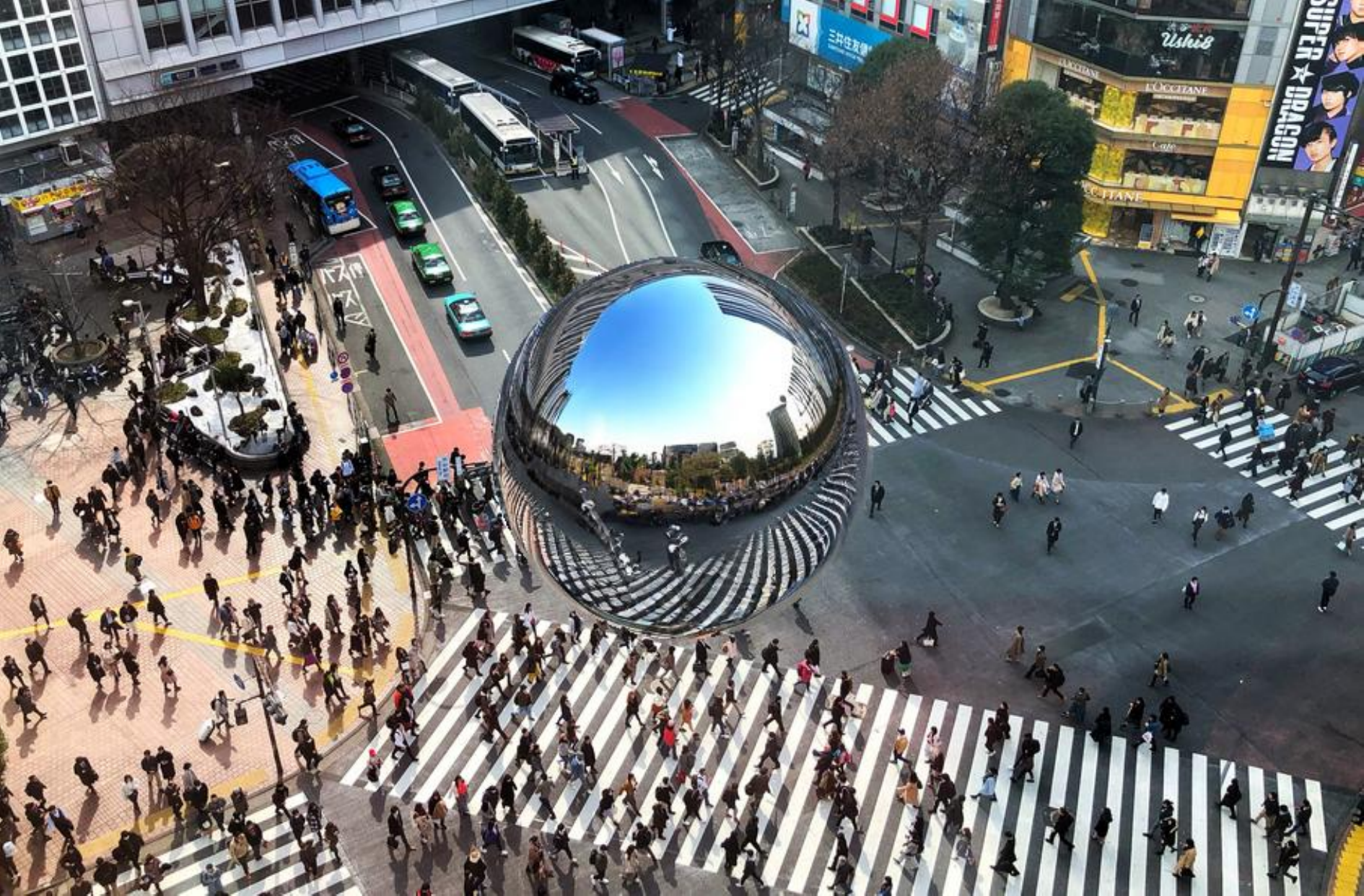}} &  
        \noindent\parbox[c]{0.105\textwidth}{\includegraphics[height=0.105\textwidth]{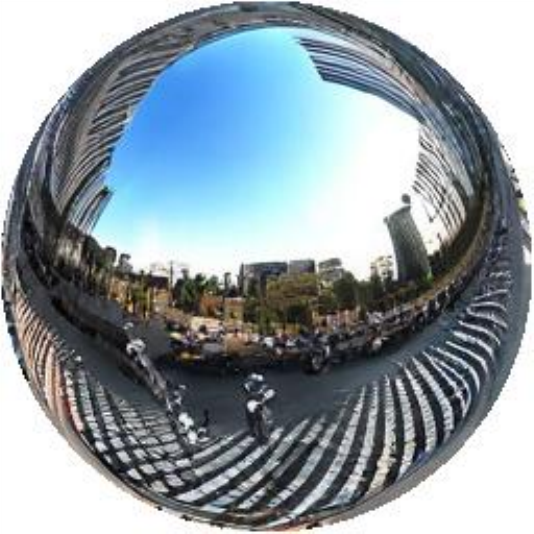}} &

        \noindent\parbox[c]{0.105\textwidth}{\includegraphics[height=0.105\textwidth]{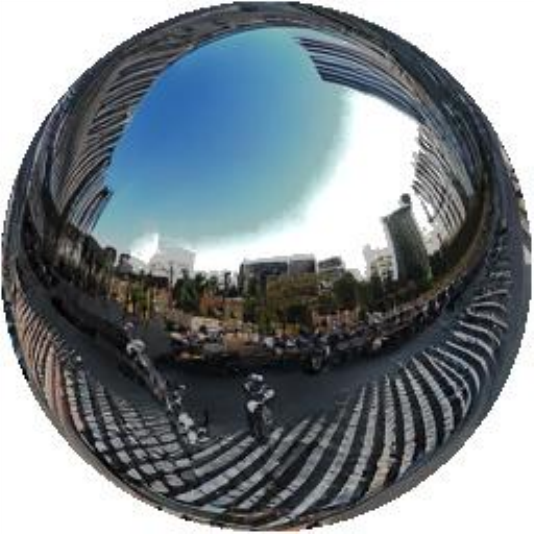}} & 
        \noindent\parbox[c]{0.105\textwidth}{\includegraphics[height=0.105\textwidth]{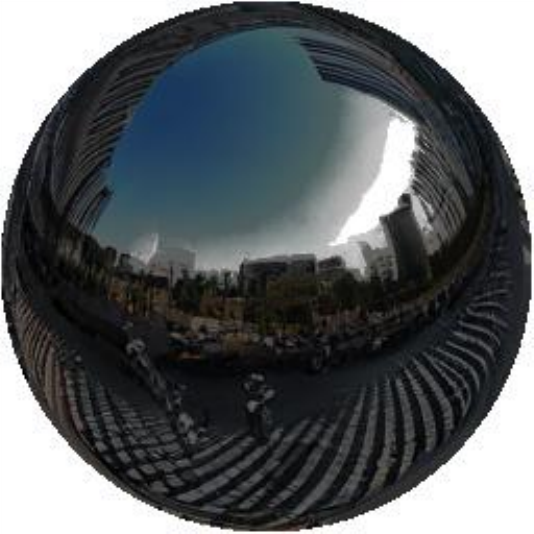}} & 
        \noindent\parbox[c]{0.105\textwidth}{\includegraphics[height=0.105\textwidth]{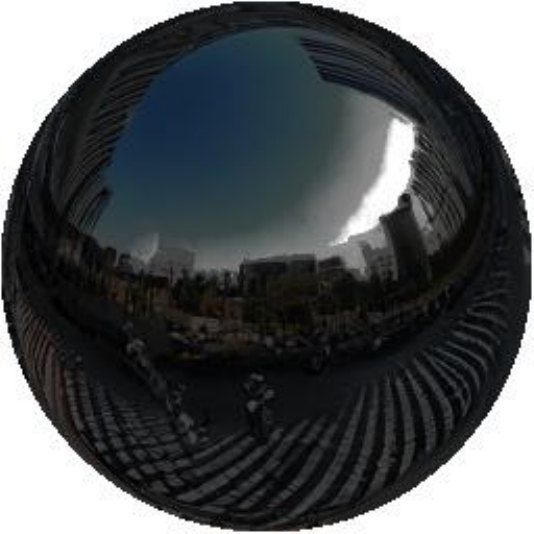}} &

        \noindent\parbox[c]{0.105\textwidth}{\includegraphics[height=0.105\textwidth]{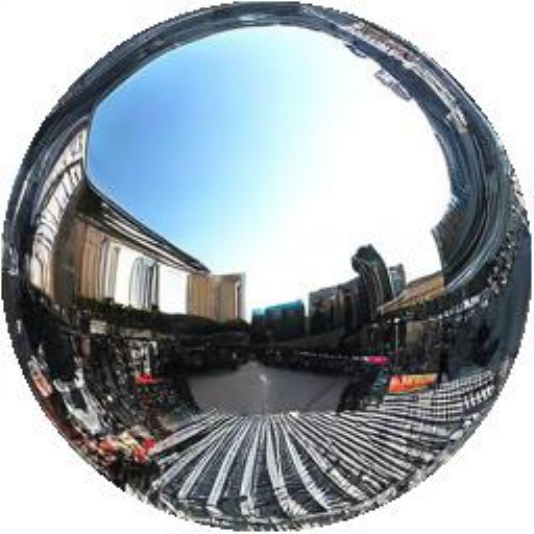}} & 
        \noindent\parbox[c]{0.105\textwidth}{\includegraphics[height=0.105\textwidth]{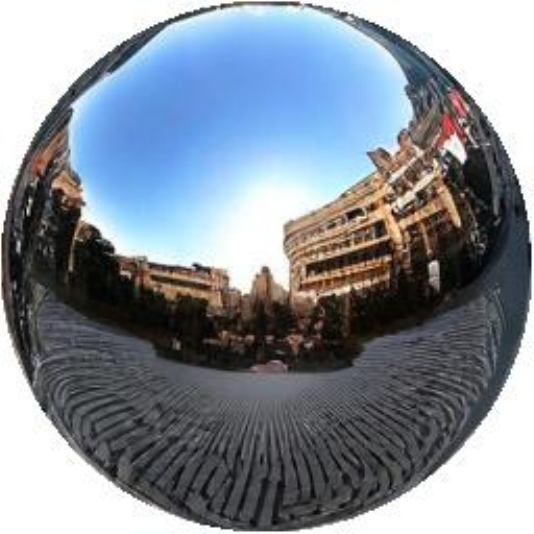}} & 
        \\

        \noindent\parbox[c]{0.105\textwidth}{\includegraphics[height=0.105\textwidth]{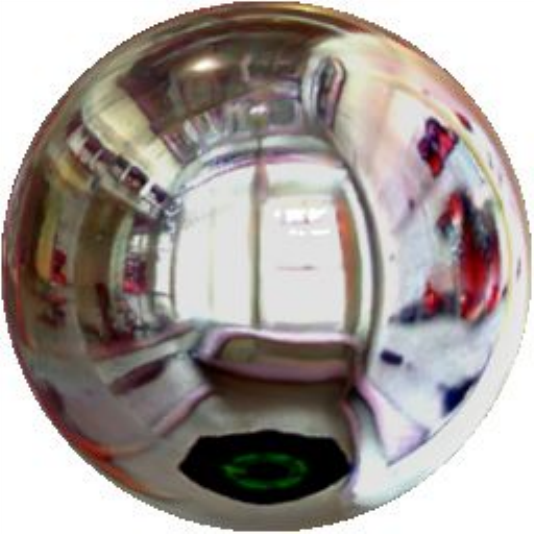}} & 
        \noindent\parbox[l]{0.16\textwidth}{\includegraphics[height=0.105\textwidth]{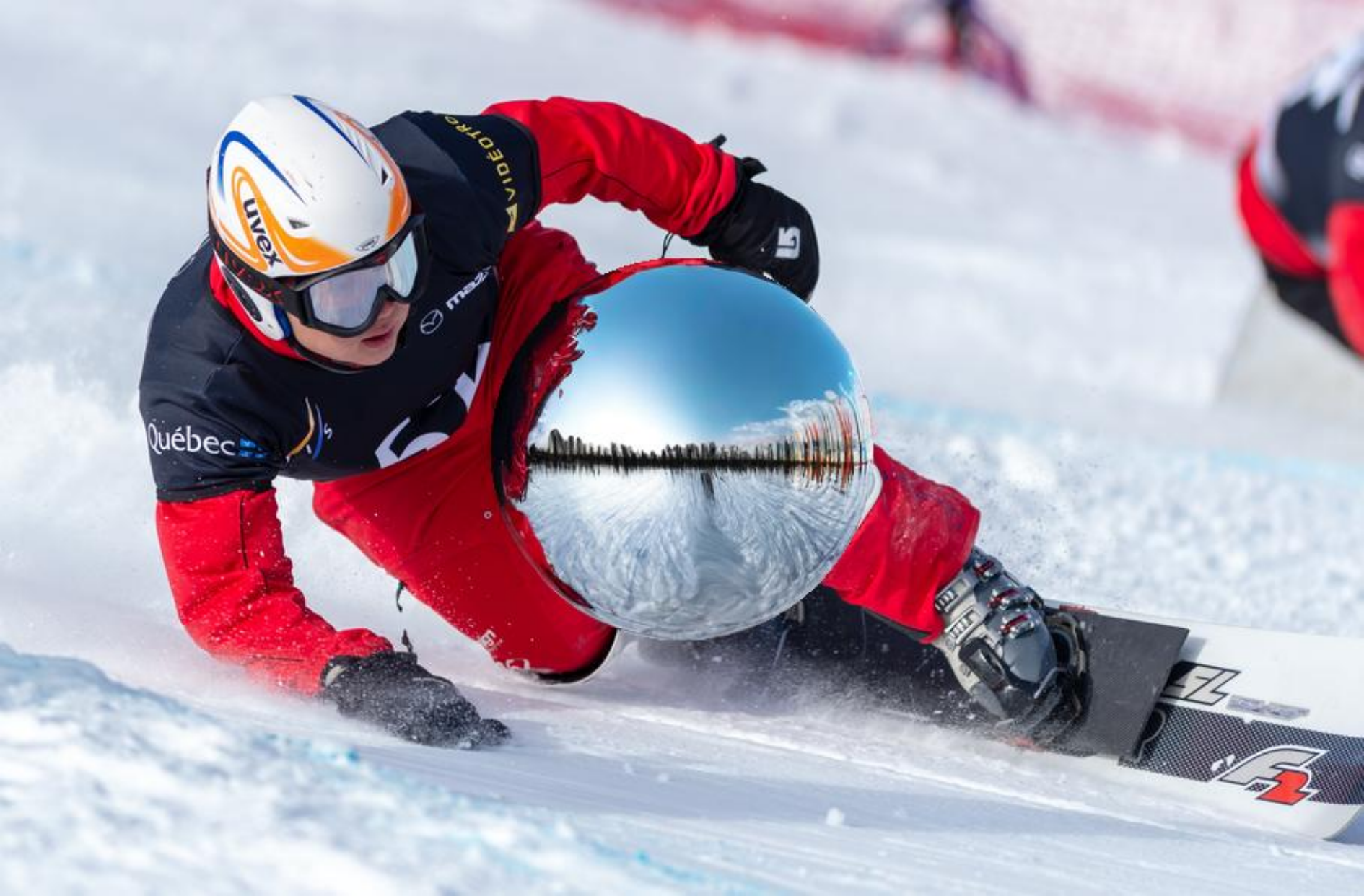}} &  
        \noindent\parbox[c]{0.105\textwidth}{\includegraphics[height=0.105\textwidth]{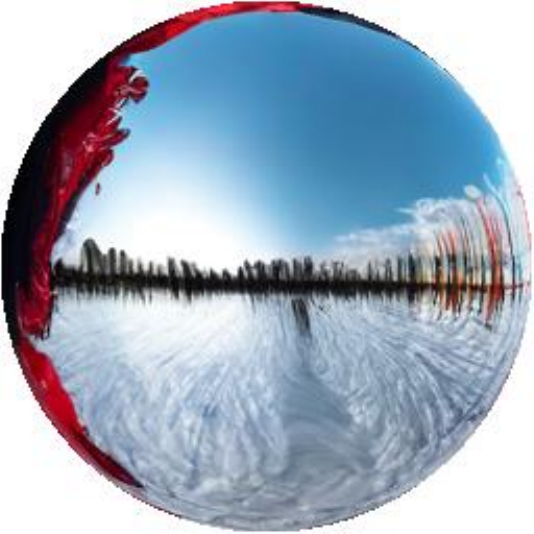}} & 

        \noindent\parbox[c]{0.105\textwidth}{\includegraphics[height=0.105\textwidth]{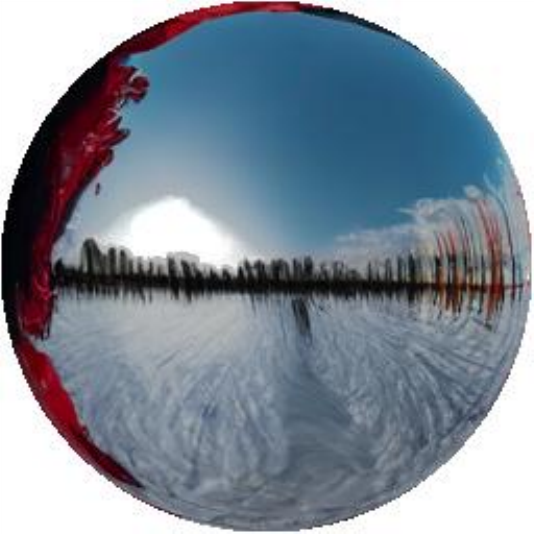}} & 
        \noindent\parbox[c]{0.105\textwidth}{\includegraphics[height=0.105\textwidth]{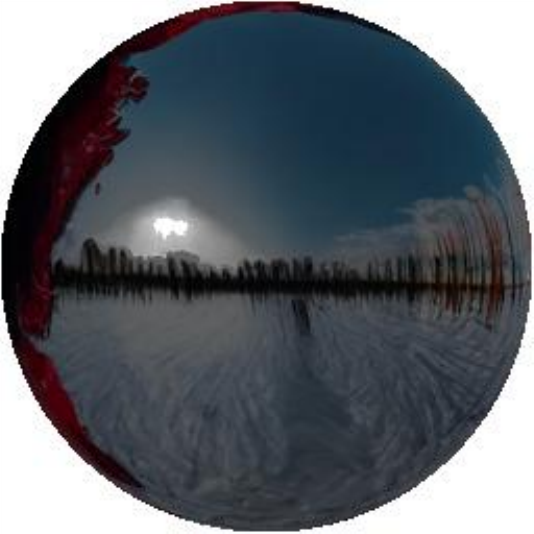}} & 
        \noindent\parbox[c]{0.105\textwidth}{\includegraphics[height=0.105\textwidth]{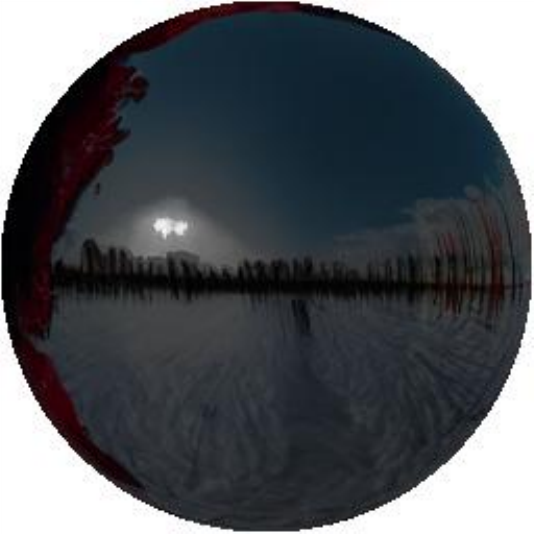}} &

        \noindent\parbox[c]{0.105\textwidth}{\includegraphics[height=0.105\textwidth]{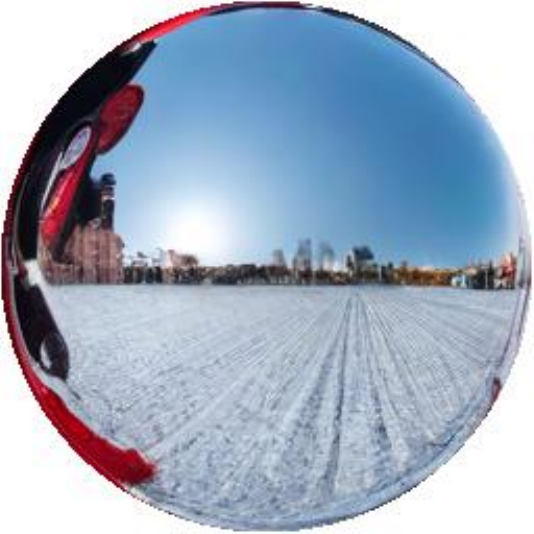}} & 
        \noindent\parbox[c]{0.105\textwidth}{\includegraphics[height=0.105\textwidth]{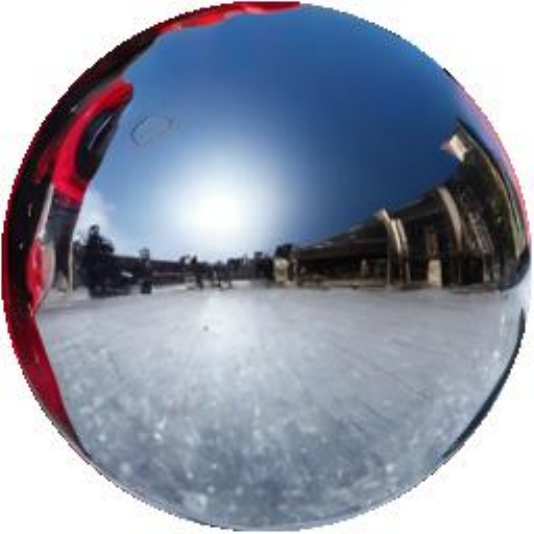}} & 
        \\

        \end{tabu}
    \caption{
    Qualitative results for in-the-wild scenes. We show chrome balls generated from our pipeline along with their HDR outputs, rendered at different EVs in columns 3 to 6. Additional plausible chrome balls are depicted in columns 7 and 8.}
    \label{fig:qualitative_wild}
    \vspace{-1.5em}
\end{figure*}

\ifpurecompactformat
\vspace{-0.1cm}
\fi
\subsection{Qualitative results for in-the-wild scenes}
\ifpurecompactformat
\vspace{-0.2cm}
\fi
We present additional qualitative results on diverse in-the-wild scenes available on Unsplash (www.unsplash.com) and other websites under CC4.0 license in Figure \ref{fig:qualitative_wild} and Appendix \ref{appendix:more_result_wild}. Compared to other existing techniques, our method can produce chrome balls that ``reflect'' what is in the input image: the car's ceiling, the zebra crossing, and the red garment of the snowboarder, as well as reveal overexposed details, such as window frames and the sun (see also Figure \ref{fig:teaser}).


\ifpurecompactformat
\vspace{-0.1cm}
\fi
\subsection{Ablation studies}
\ifpurecompactformat
\vspace{-0.2cm}
\fi
We perform an ablation study on our iterative inpainting and LoRA using Laval Indoor and Poly Haven datasets.
Table \ref{tab:indoor_stylelight} shows that our full method
surpasses all ablated versions on all metrics except Angular Error on matte balls in Poly Haven. Studies on the size and number of balls, as well as the trade-off between running time and quality are in Appendix \ref{appendix:aba_median}. Studies on LoRA scale and timesteps are in Appendix \ref{appendix:aba_lora}.
\ifpurecompactformat
\vspace{-0.2cm}
\fi
\subsection{Limitations}
\ifpurecompactformat
\vspace{-0.1cm}
\fi
 Given the absence of focal length or FOV information, we assume orthographic projection when converting a chrome ball to an environment map, which may not reflect the projection model rendered by the diffusion model. Our chrome balls occasionally fail to reflect surrounding environments in overhead or bird's eye-view images, shown in Appendix \ref{app:fail}. Our method is currently slow with iterative inpainting and diffusion sampling, but utilizing sampling-efficient diffusion models \cite{liu2023insta, luo2023latent, luo2023lcm} can directly improve its speed.
\ifpurecompactformat
\vspace{-0.3cm}
\fi
\section{Conclusion}
\label{sec:conclusion}
\ifpurecompactformat
\vspace{-0.2cm}
\fi
This paper presents a novel HDR light estimation approach by inpainting a chrome ball into the scene using a pretrained LDR diffusion model. To consistently render high-quality chrome balls, we propose an iterative algorithm to locate a suitable initial noise neighborhood and apply continuous LoRA fine-tuning for exposure bracketing and generating HDR chrome balls.
Our method performs competitively with the state of the art in both indoor and outdoor settings and marks the first work that achieves good generalization to in-the-wild images.

{\small
\bibliographystyle{ieeenat_fullname}
\bibliography{11_references}
}

\clearpage \appendix 


\section{Implementation Details} \label{appendix:implement}

\subsection{Inpainting algorithm}
The pseudocode of the iterative inpainting algorithm described in Section~\ref{sec:median_algo} is given in Algorithm \ref{algo:median_algo}. Our implementation uses $\gamma = 0.8$, $k = 2$, and $N = 30$. The algorithm repeatedly invokes the $\textsc{Inpaint}$ function, which stands for an inpainting algorithm based on SDEdit \cite{meng2022sdedit} as implemented in the Diffusers library \cite{diffusers}. For completeness, we include its pseudocode in Algorithm \ref{algo:inpainting_algo}.
This algorithm requires no modification to the diffusion model and resembles standard diffusion sampling except the ``imputing'' step in Line 16. 

\DecMargin{1em}
\SetKwInput{Input}{Input}
\SetKwInput{Output}{Output}
\setcounter{algocf}{-1}
\begin{algorithm}[h!]
  
  \caption{Inpainting using Diffusion Model}\label{algo:inpainting_algo}
  \begin{algorithmic}[1]
    \Function{Update}{$\vect{z}, t, \vect{\epsilon}$} \\
        \quad \Return $\sqrt{\alpha_{t-1}} \parens*{\frac{\vect{z} - \sqrt{1 - \alpha_t} \vect{\epsilon}}{\sqrt{\alpha}}} + \sqrt{1 - \alpha_{t-1}} \vect{\epsilon}$
    \EndFunction \\
    {\color{gray}
    \\
     \small // \textbf{input:} latent code of input image $\vect{z}_I$, \\ // initial latent code $\vect{z}$, // inpainting mask $M$ \\ // conditioning signal (e.g. text, depth map) $\vect{C}$ \\ // timestep to start denoising ($\text{denoising-start}$).\\
    \small // \textbf{output:} Input image with an inpainted chrome ball.
    }
    \Function{Inpaint}{$\vect{z}_I$, $\vect{z}, \vect{M}, \vect{C}, \text{denoising-start} = T$} \\
    \quad $\vect{\epsilon} \sim \mathcal{N}(\vect{0}, \vect{I})$ \\
    \quad \textbf{for} $i \in \{\text{denoising-start}, \ldots, 1\}$ \textbf{do} \\
    \quad \quad $\vect{\epsilon} \leftarrow \epsilon_{\vect{\theta}}(\vect{z}, t, \vect{C})$ \\
    \quad \quad $\vect{z} \leftarrow \Call{Update}{\vect{z}, t, \vect{\epsilon}}$ \\
    
    \quad \quad $\vect{z}_I' \leftarrow \sqrt{\alpha_t}\vect{z}_I + \sqrt{1-\alpha_t} \vect{\epsilon}$ \\
    \quad \quad $\vect{z} \leftarrow (1 - \vect{M}) \odot \vect{z}_I' + \vect{M} \odot \vect{z}$ \\
    \quad \textbf{end for} \\
    \quad $\vect{\epsilon} \leftarrow \epsilon_{\theta}(\vect{z}, 0, \vect{C})$ \\
    \quad $\vect{z} \leftarrow \Call{Update}{\vect{z}, 0, \vect{\epsilon}}$ \\
    \quad \Return $\Call{Decode}{\vect{z}}$
    \EndFunction
  \end{algorithmic}
\end{algorithm}

\SetKwInput{Input}{Input}
\SetKwInput{Output}{Output}
\begin{algorithm}[htbp]
  \caption{Iterative Inpainting Algorithm}\label{algo:median_algo}
  \begin{algorithmic}[1]
    \Function{InpaintBall}{$\vect{I}, \vect{M}, \vect{C}, \gamma, T=1000$} \\
    \quad $\vect{z} \leftarrow \Call{Encode}{\vect{I}}$ \\
    \quad $\vect{\epsilon} \sim \mathcal{N}(\vect{0}, \vect{I})$ \\
    \quad $\vect{z}_{\gamma T} \leftarrow \sqrt{\alpha_{\gamma T}}\vect{z} + \sqrt{1-\alpha_{\gamma T}} \vect{\epsilon}$ \\
    \quad \Return $\Call{Inpaint}{\vect{z}, \vect{z}_{\gamma T}, \vect{M}, \vect{C}, \text{denoising-start}=\gamma T}$
    \EndFunction \\ 
    { \color{gray} \\
    {\small // \textbf{input:} Input image $\vect{I}$, binary inpainting mask $\vect{M}$, \\ // Conditioning signal (e.g. text, depth map) $\vect{C}$ \\ // denoising strength $\gamma$, number of balls for median $N$, \\ // number of median iterations $k$.}\\
    {\small // \textbf{output:} Input image with an inpainted chrome ball.}
    }
    \Function{IterativeInpaint}{$\vect{I}, \vect{M}, \vect{C}, \gamma, k, N$} \\
    \quad \textbf{for} $i \in \{1, \ldots, k\}$ \textbf{do} \\
    \quad \quad \textbf{for} $j \in \{1, \ldots, N\}$ \textbf{do} \\
    \quad \quad \quad $\gamma' \leftarrow \gamma$ \textbf{if} $i > 1$ \textbf{else} $1.0$ \\ 
    \quad \quad \quad $\vect{B}_j \leftarrow \Call{InpaintBall}{\vect{I}, \vect{M}, \vect{C}, \gamma'}$ \\
    \quad \quad \textbf{end for}\\ 
    \quad \quad $\vect{B} \leftarrow \Call{PixelwiseMedian}{\vect{B}_1, \ldots, \vect{B}_N}$ \\
    \quad \quad $\vect{I} \leftarrow (1 - \vect{M}) \odot \vect{I} + \vect{M} \odot \vect{B}$ \\
    \quad \textbf{end for} \\
    \quad \Return $\Call{InpaintBall}{\vect{I}, \vect{M}, \gamma}$
    \EndFunction
  \end{algorithmic}
\end{algorithm}

\subsection{HDR merging algorithm}

Algorithm \ref{algo:ldr2hdr} merges a bracket of LDR images to create an HDR image. As mentioned in the main paper, we merge in the luminance space to avoid ghosting artifacts. Our luminance conversion assumes sRGB \cite{nguyen2017luminance} and gamma value of 2.4, following \cite{wang2022stylelight}. 


\begin{algorithm}[htbp]
  \caption{HDR Merging Algorithm}\label{algo:ldr2hdr}
  \begin{algorithmic}[1]
    \Function{Luminance}{$\vect{I}, ev, \gamma=2.4$} \\
    \quad \Return $\vect{I}^\gamma \cdot [0.21267, 0.71516, 0.07217]^\top (2^{-ev})$
    \EndFunction \\ 
    {\color{gray} \\
    {\small // \textbf{input:} LDR images and a list of exposure values in \\ // descending order, where $ev_0=0$. E.g., [0, -2.5, -5]}\\
    {\small // \textbf{output:} A linearized HDR image.}
    }
    \Function{MergeLDRs}{$\vect{I}_0, ..., \vect{I}_{n-1}, ev_0, ..., ev_{n-1}$} \\
    \quad $\vect{L} \leftarrow  \Call{Luminance}{\vect{I}_{n-1}, ev_{n-1}}$ \\
    \quad \textbf{for} $i \in \{n-2, n-1, ..., 0\}$ \textbf{do} \\ \quad \quad 
        $\vect{L}_{i} \leftarrow \Call{Luminance}{\vect{I}_{i} , ev_i}$ \\ \quad \quad 
        $\vect{M} \leftarrow \Call{clip}{\frac{(2^{ev_i} \vect{L}_i)-0.9}{0.1}, 0, 1} \odot \mathbbm{1}(\vect{L} > \vect{L}_i)$ \\ \quad \quad
        $\vect{L} \leftarrow (1 - \vect{M}) \odot \vect{L}_i + \vect{M} \odot \vect{L}$ \\ \quad
        \textbf{end for}\\ 
    \quad \Return $\vect{I}_0^\gamma \odot \left(\frac{\vect{L}}{\vect{L}_0}\right)$
    \EndFunction
  \end{algorithmic}
\end{algorithm}

\section{Ablation on Iterative Inpainting Algorithm} \label{appendix:aba_median}

\subsection{Inpainting iterations}

\begin{figure}
    \centering

    \begin{tabu} to \textwidth {
        @{}
        c@{\hspace{1pt}}
        c@{\hspace{1pt}}
        @{\hspace{8pt}}
        c@{\hspace{1pt}}
        c@{\hspace{1pt}}
        c@{}
    }
        
        \noindent\parbox[c]{0.08\textwidth}{\includegraphics[width=0.08\textwidth]{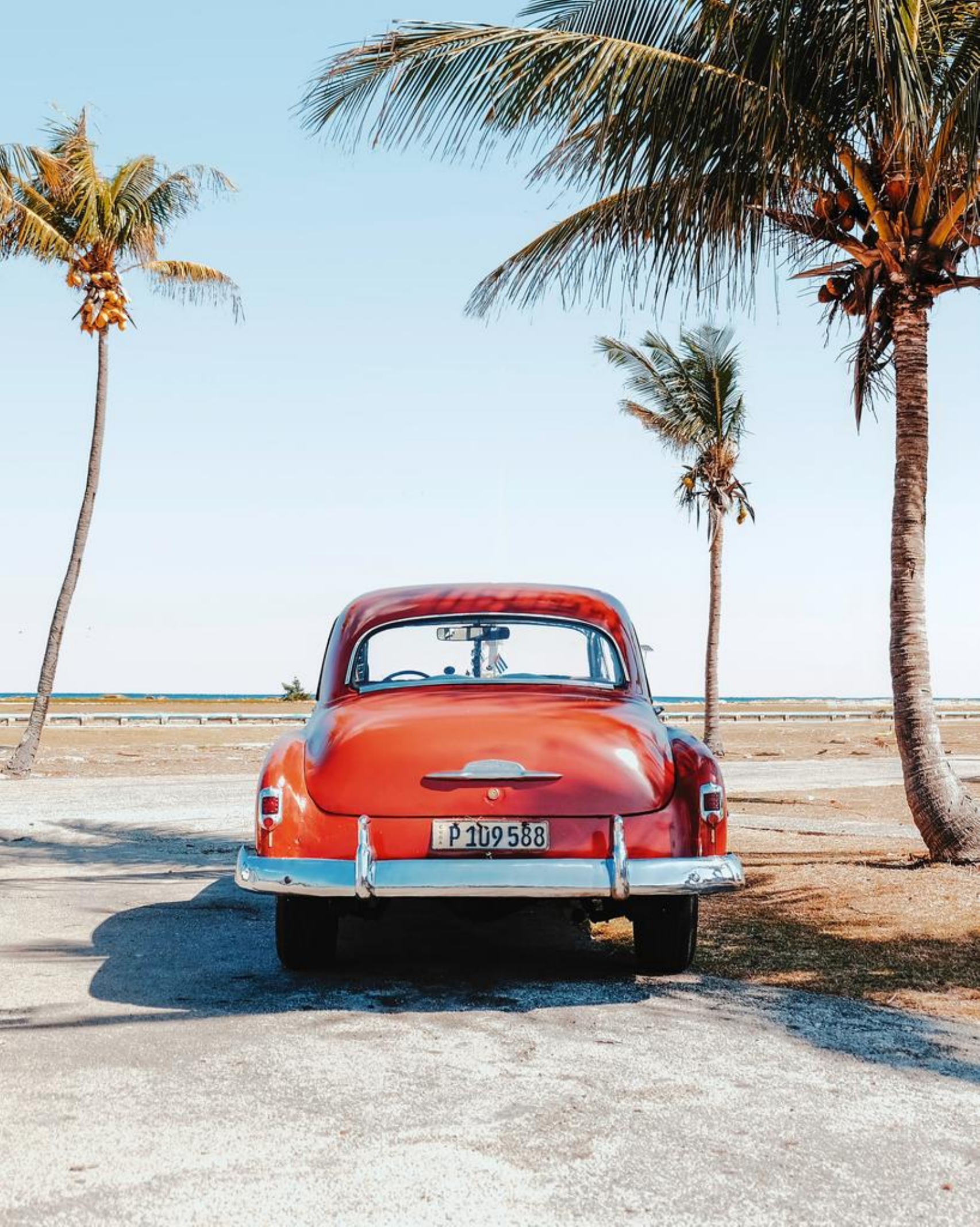}} 
        &
        \noindent\parbox[c]{0.14\textwidth}{\includegraphics[width=0.14\textwidth]{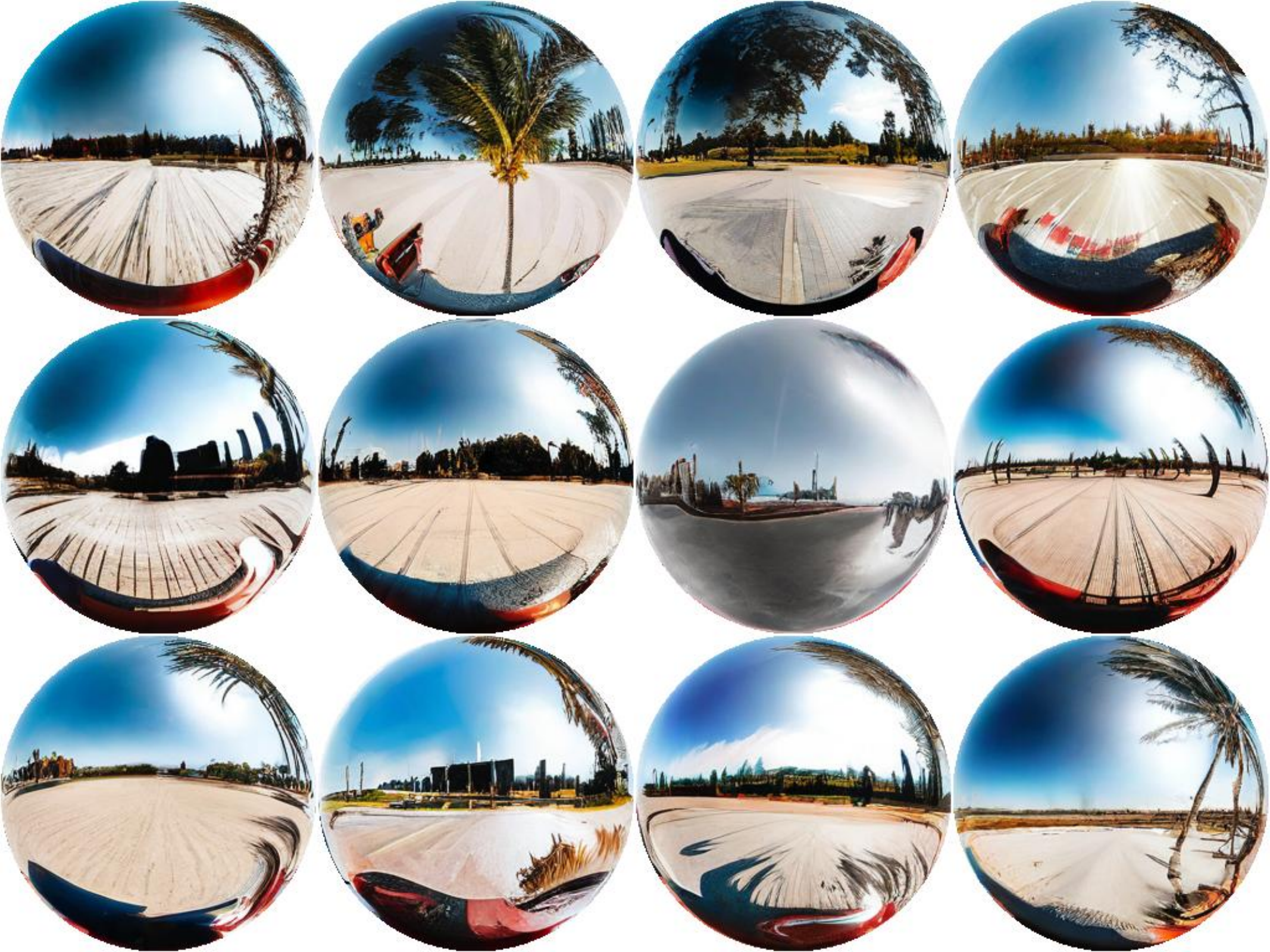}} 

        
        &
        \noindent\parbox[c]{0.08\textwidth}{\shortstack{\tiny Median ball \\ \includegraphics[width=0.08\textwidth]{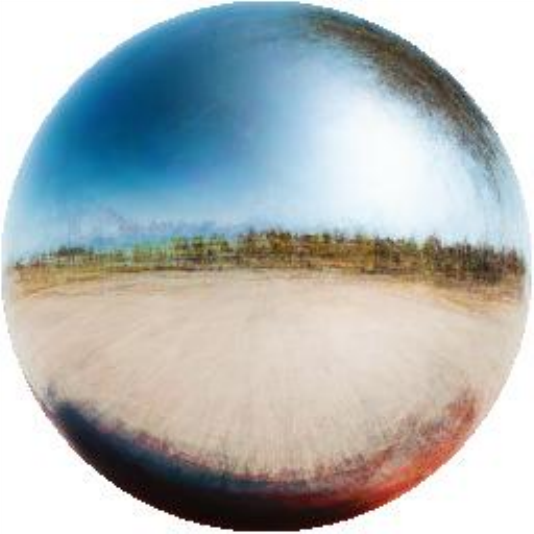}}}
        
        &
        \noindent\parbox[c]{0.14\textwidth}{\includegraphics[width=0.14\textwidth]{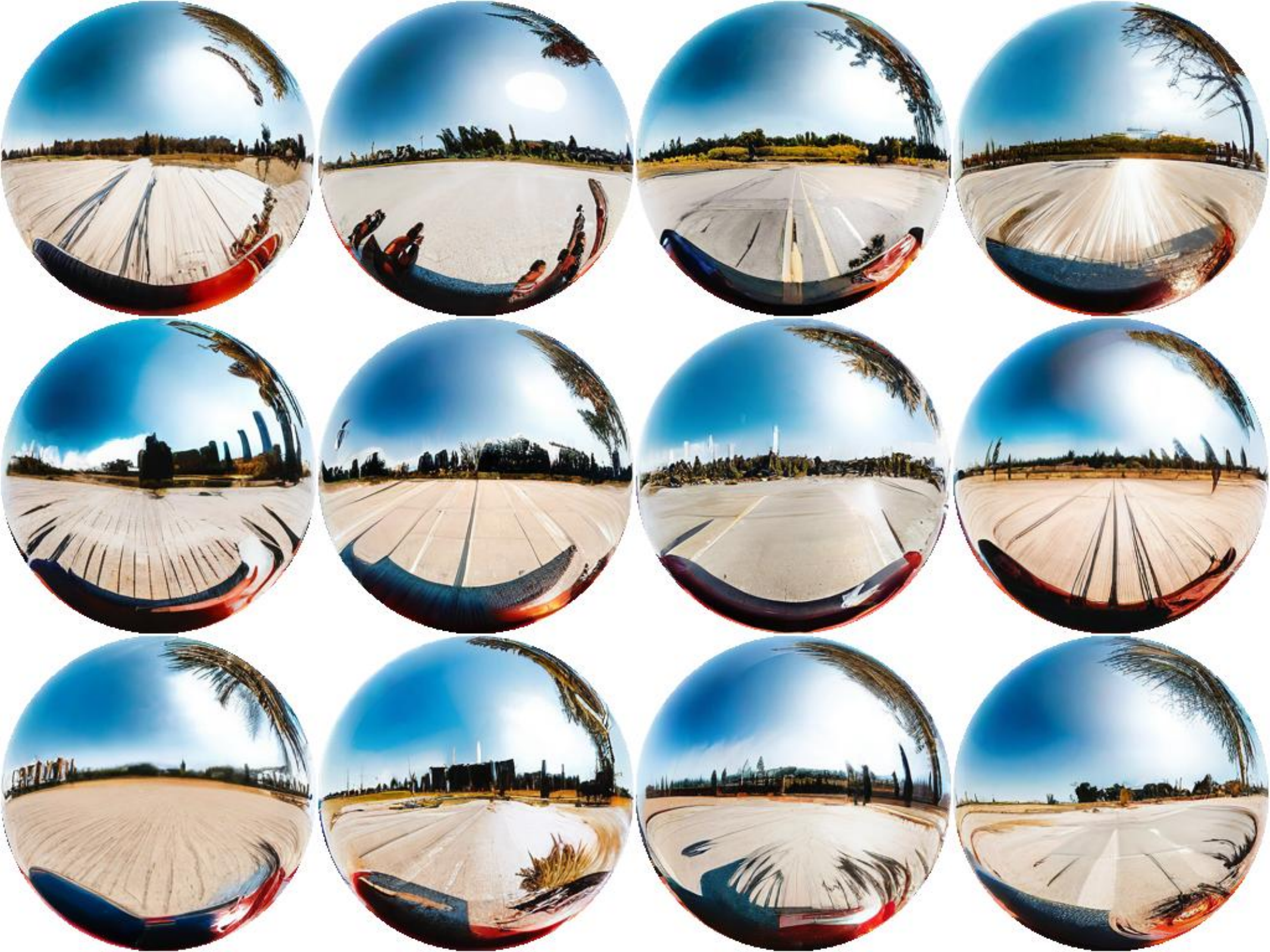}} 
        
        \\

        \hline

        \noindent\parbox[c]{0.08\textwidth}{\includegraphics[width=0.08\textwidth]{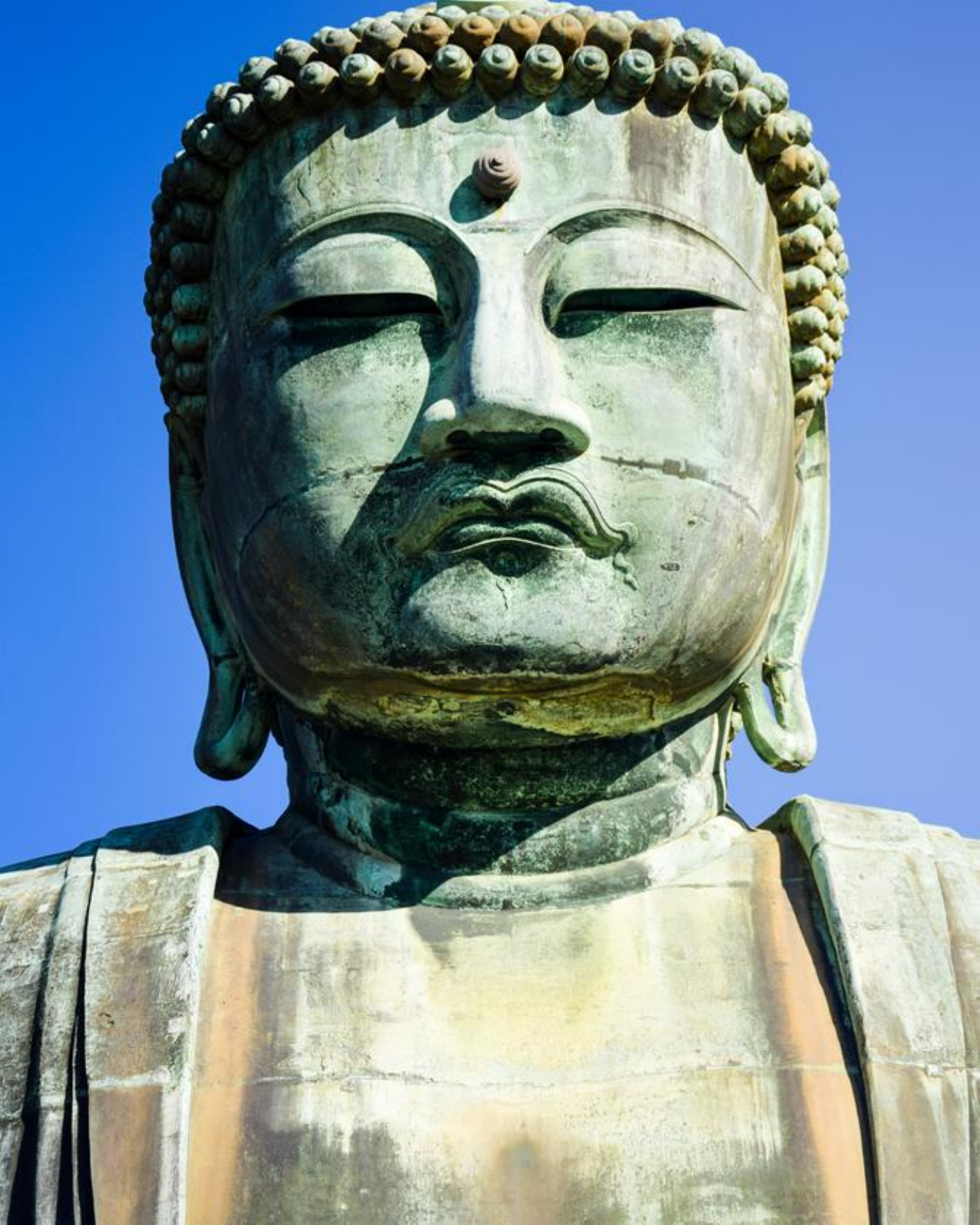}} 
        &
        \noindent\parbox[c]{0.14\textwidth}{\includegraphics[width=0.14\textwidth]{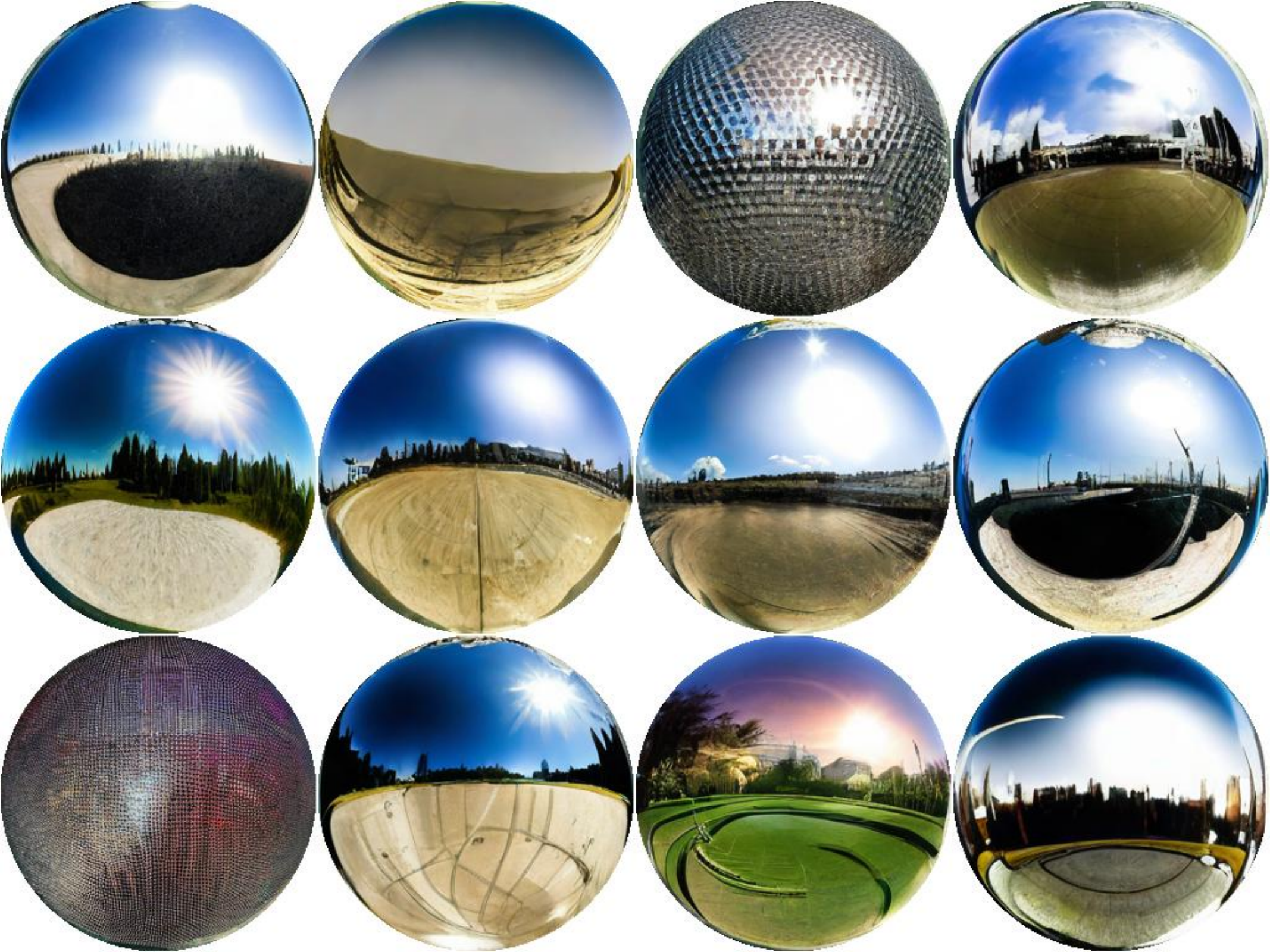}} 

        
        &
        \noindent\parbox[c]{0.08\textwidth}{\shortstack{\tiny Median ball \\ \includegraphics[width=0.08\textwidth]{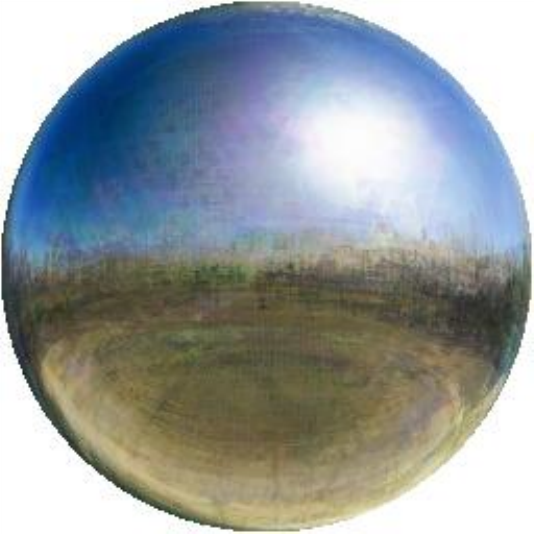}}}
        
        &
        \noindent\parbox[c]{0.14\textwidth}{\includegraphics[width=0.14\textwidth]{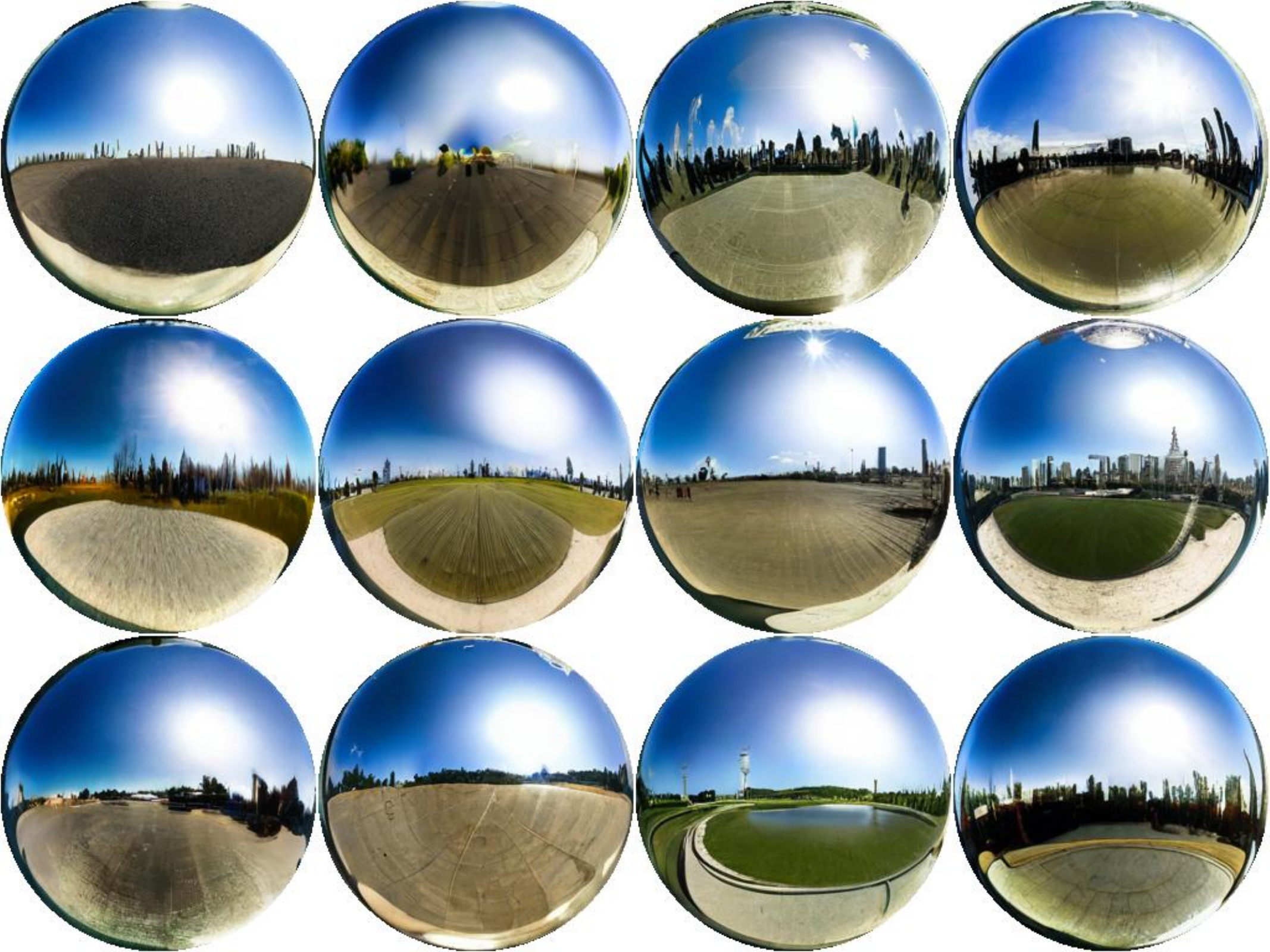}} 
        
        \\

        \hline

        \noindent\parbox[c]{0.08\textwidth}{\includegraphics[width=0.08\textwidth]{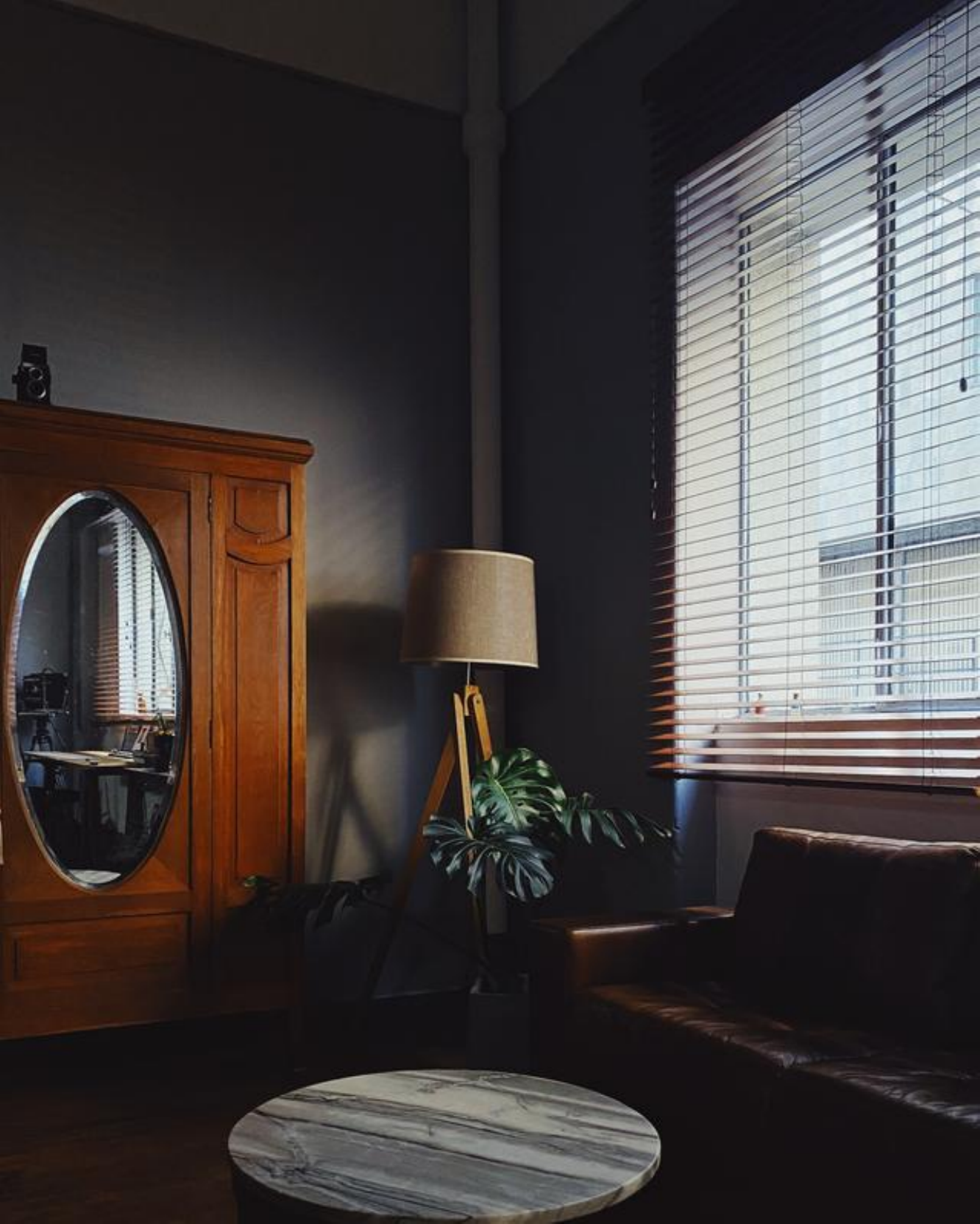}} 
        &
        \noindent\parbox[c]{0.14\textwidth}{\includegraphics[width=0.14\textwidth]{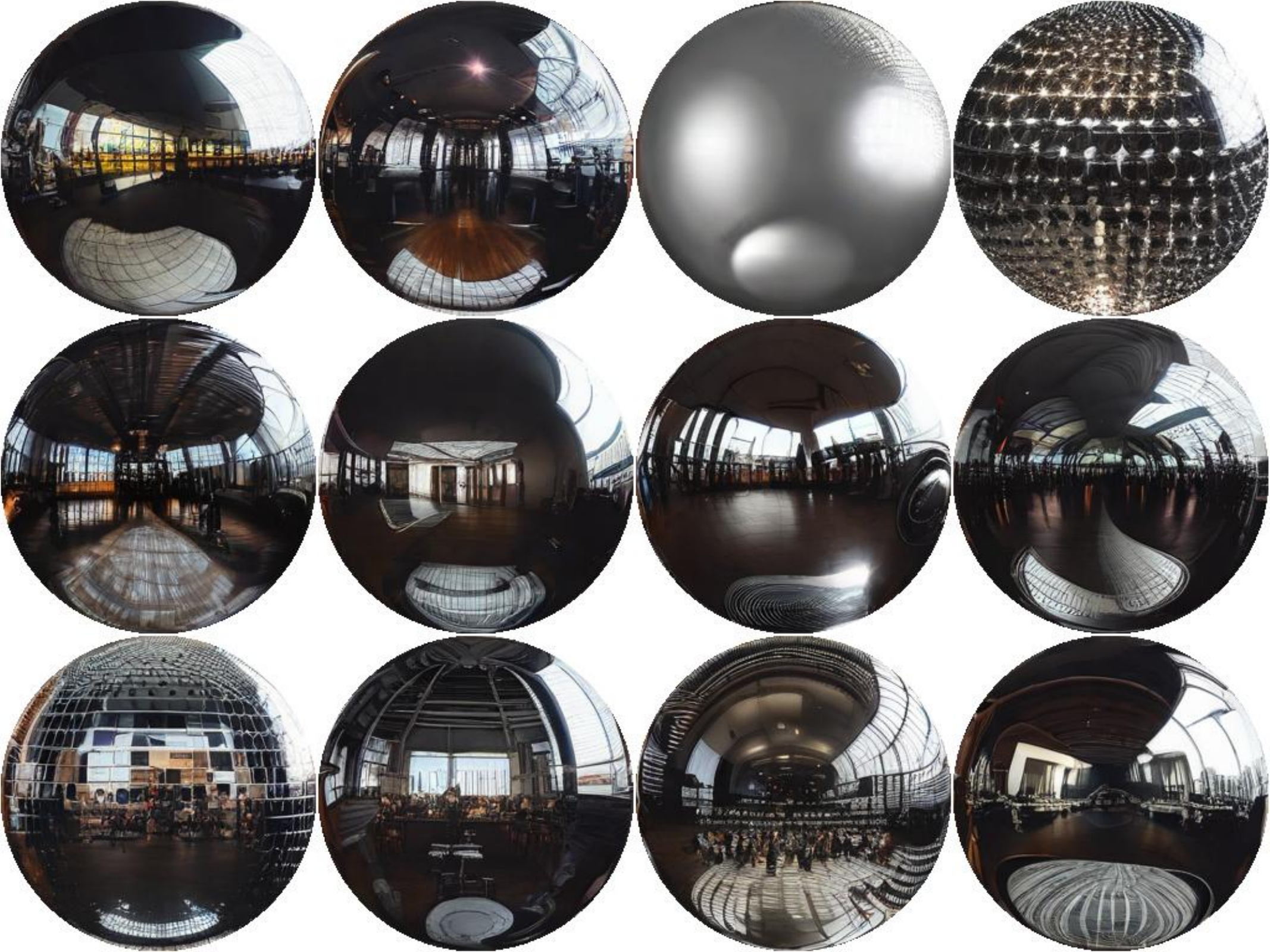}} 

        
        &
        \noindent\parbox[c]{0.08\textwidth}{\shortstack{\tiny Median ball \\ \includegraphics[width=0.08\textwidth]{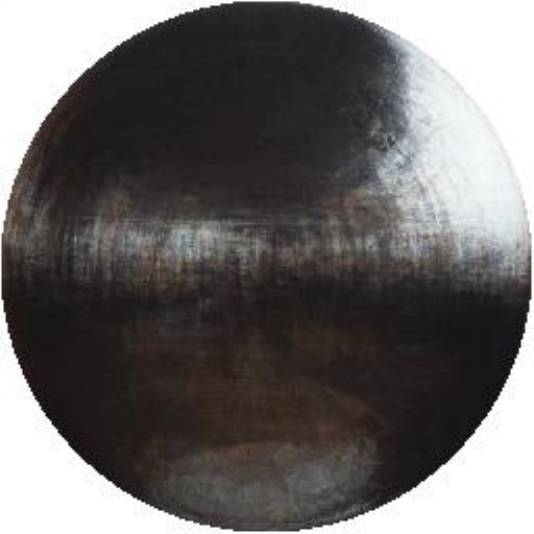}}}
        
        &
        \noindent\parbox[c]{0.14\textwidth}{\includegraphics[width=0.14\textwidth]{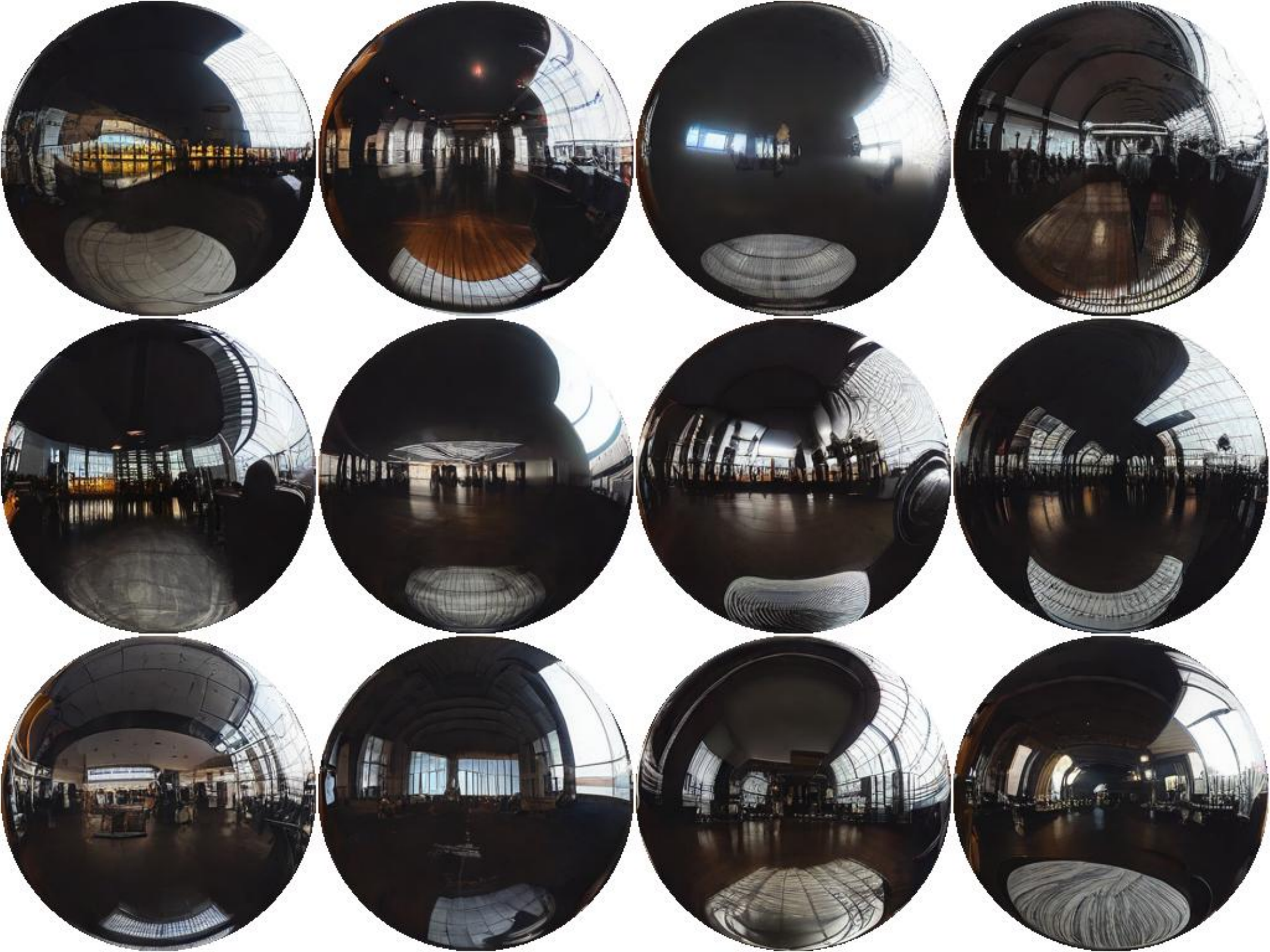}} 
        
        \\
        
    \end{tabu}
    \caption{Chrome balls before (left) and after (right) one iteration of our iterative inpainting algorithm. Notice how poor chrome balls are fixed and the light estimation becomes more consistent.}
    \label{fig:compare_median_distribution_aba}
\end{figure}

\begin{figure*}
    \centering
        \begin{tabu} to \textwidth {
        c@{}|@{\hspace{0pt}}
        c@{}
        c@{\hspace{0.3pt}}
        c@{\hspace{0.3pt}}
        c@{\hspace{0.3pt}}
        c@{\hspace{0.3pt}}
        c@{\hspace{0.3pt}}
        c@{\hspace{0.3pt}}
        c@{\hspace{0.3pt}}
        c@{\hspace{0.3pt}}
        c@{\hspace{0.3pt}}|@{\hspace{0.3pt}}
        c@{}
    }
        \multicolumn{1}{c}{\shortstack{\scriptsize \hspace{0.3pt} Input image}} &
        &
        \multicolumn{1}{c}{\shortstack{\scriptsize Pred\#1}} & 
        \multicolumn{1}{c}{\shortstack{\scriptsize Pred\#2}} & 
        \multicolumn{1}{c}{\shortstack{\scriptsize Pred\#3}} & 
        \multicolumn{1}{c}{\shortstack{\scriptsize Pred\#4}} & 
        \multicolumn{1}{c}{\shortstack{\scriptsize Pred\#5}} & 
        \multicolumn{1}{c}{\shortstack{\scriptsize Pred\#6}} & 
        \multicolumn{1}{c}{\shortstack{\scriptsize Pred\#7}} & 
        \multicolumn{1}{c}{\shortstack{\scriptsize Pred\#8}} & 
        \multicolumn{1}{c}{\shortstack{\scriptsize \hspace{-5pt} Pred\#9}} & 
        \multicolumn{1}{c}{\shortstack{\scriptsize Median Ball}}
        \\
        
        \multirow{2}{*}{\noindent\parbox[c]{0.112\textwidth}{\includegraphics[width=0.112\textwidth]{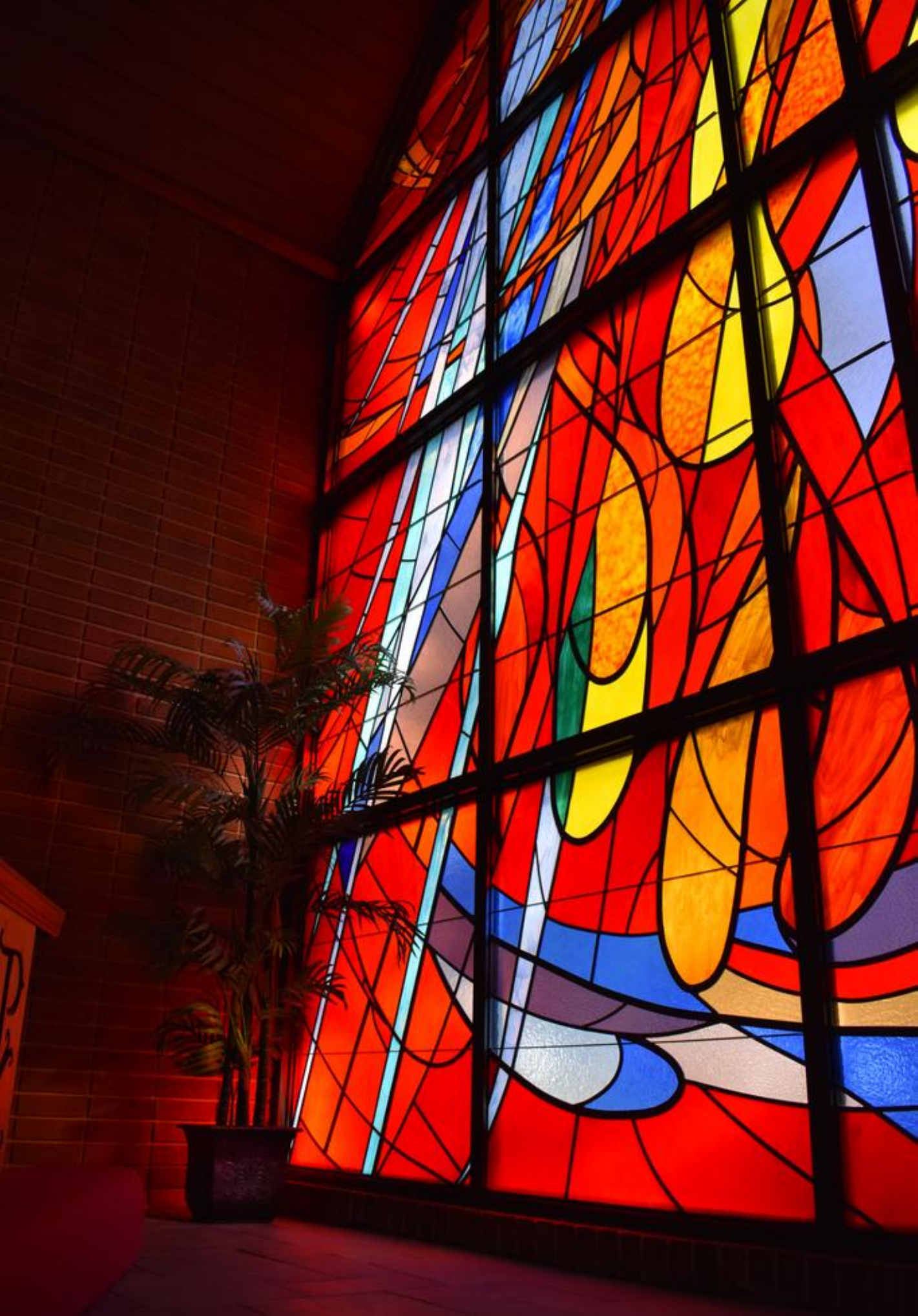}}} &
        \multicolumn{1}{l}{\rotatebox[origin=c]{90}{\shortstack[l]{\tiny 1\textsuperscript{st} iteration}}} &
        \noindent\parbox[c]{0.082\textwidth}{\includegraphics[width=0.082\textwidth]{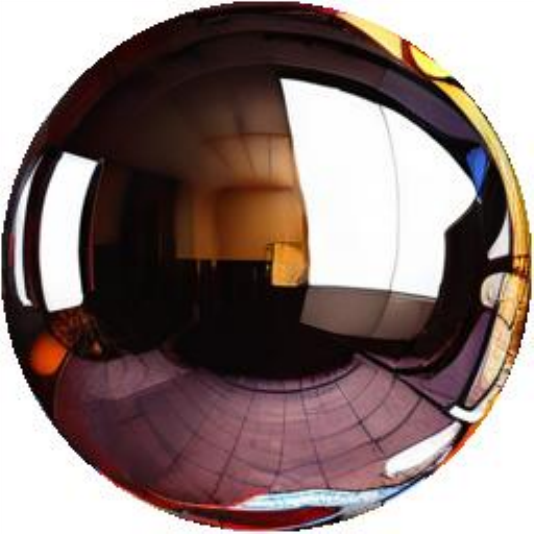}} & 
        \noindent\parbox[c]{0.082\textwidth}{\includegraphics[width=0.082\textwidth]{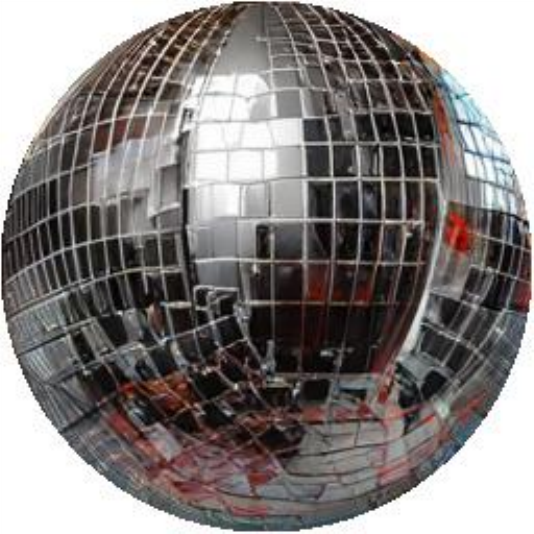}} & 
        \noindent\parbox[c]{0.082\textwidth}{\includegraphics[width=0.082\textwidth]{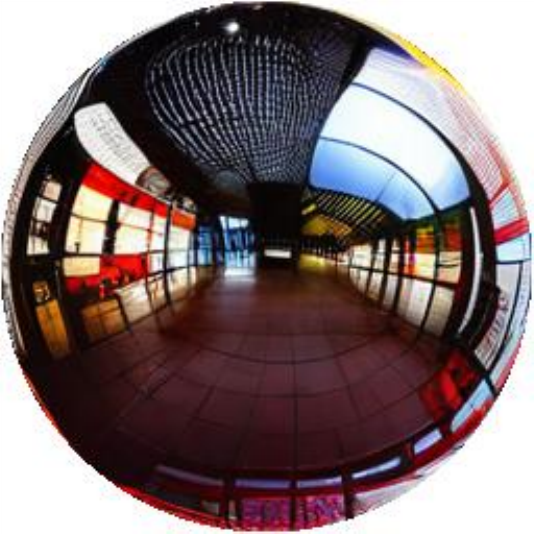}} & 
        \noindent\parbox[c]{0.082\textwidth}{\includegraphics[width=0.082\textwidth]{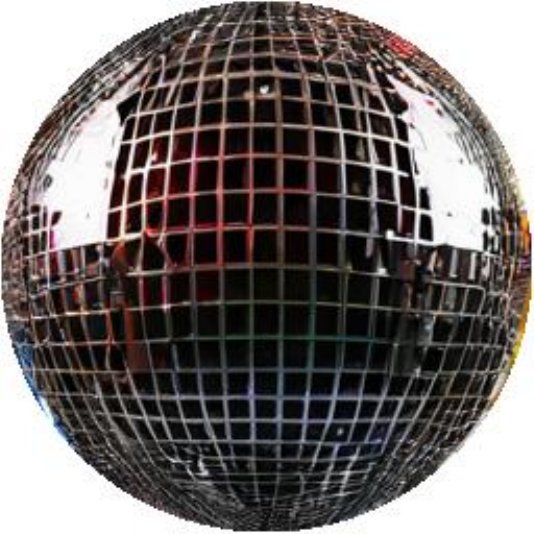}} & 
        \noindent\parbox[c]{0.082\textwidth}{\includegraphics[width=0.082\textwidth]{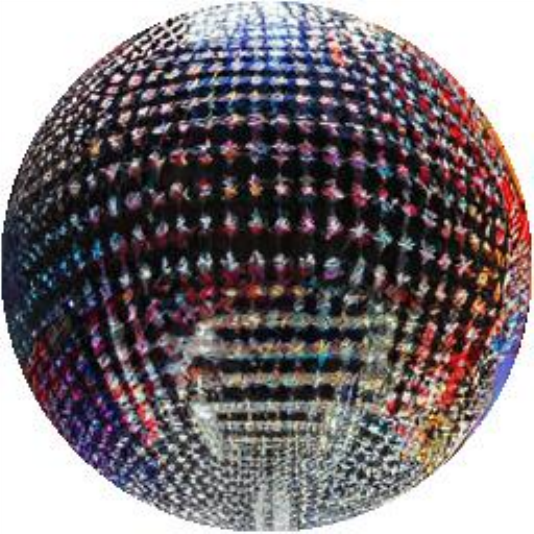}} & 
        \noindent\parbox[c]{0.082\textwidth}{\includegraphics[width=0.082\textwidth]{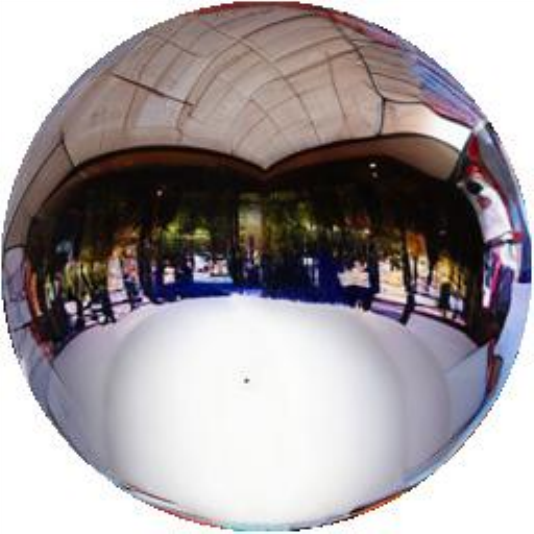}} & 
        \noindent\parbox[c]{0.082\textwidth}{\includegraphics[width=0.082\textwidth]{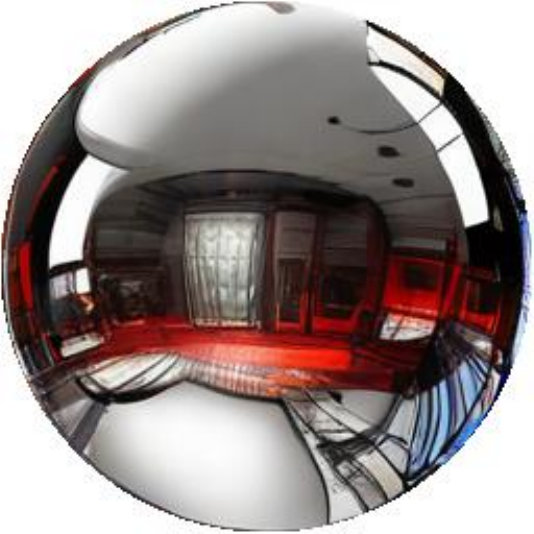}} & 
        \noindent\parbox[c]{0.082\textwidth}{\includegraphics[width=0.082\textwidth]{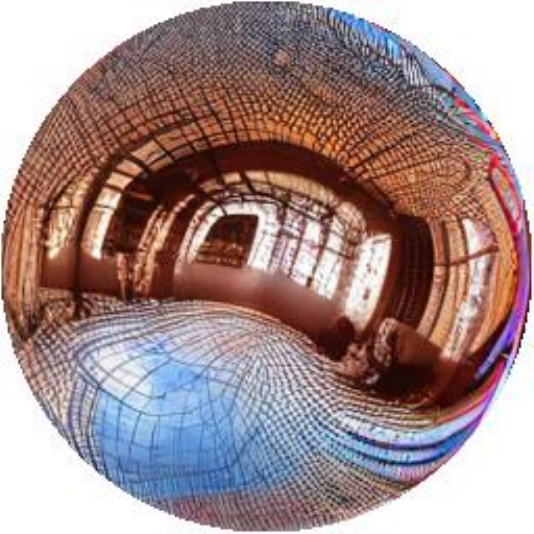}} & 
        \noindent\parbox[c]{0.082\textwidth}{\includegraphics[width=0.082\textwidth]{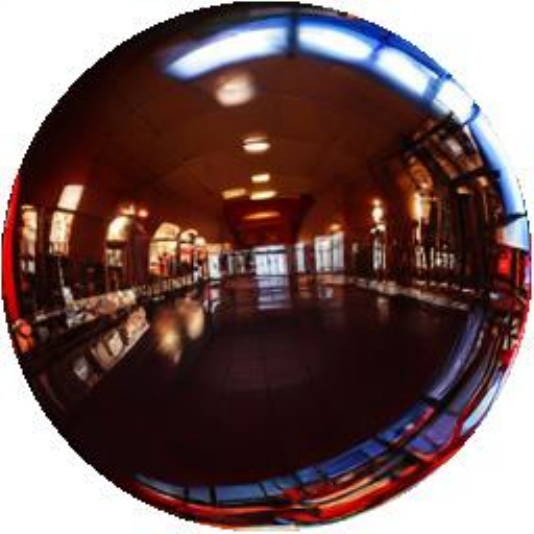}} & 
        \noindent\parbox[c]{0.082\textwidth}{\includegraphics[width=0.082\textwidth]{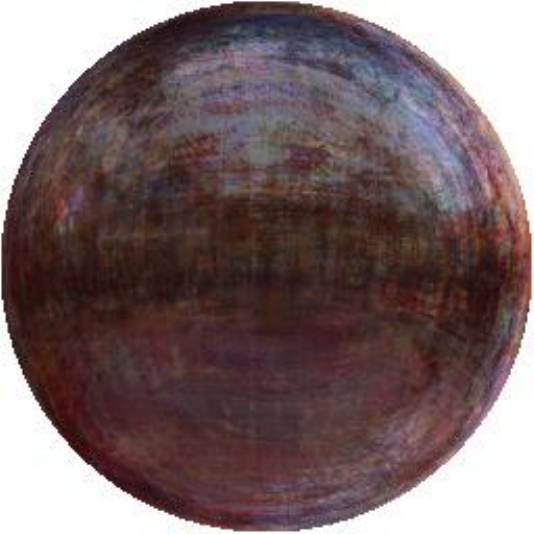}}

        \\ \cline{2-12}

        & 
        \multicolumn{1}{l}{\rotatebox[origin=c]{90}{\shortstack[l]{\tiny 2\textsuperscript{nd} iteration}}} &
        \noindent\parbox[c]{0.082\textwidth}{\includegraphics[width=0.082\textwidth]{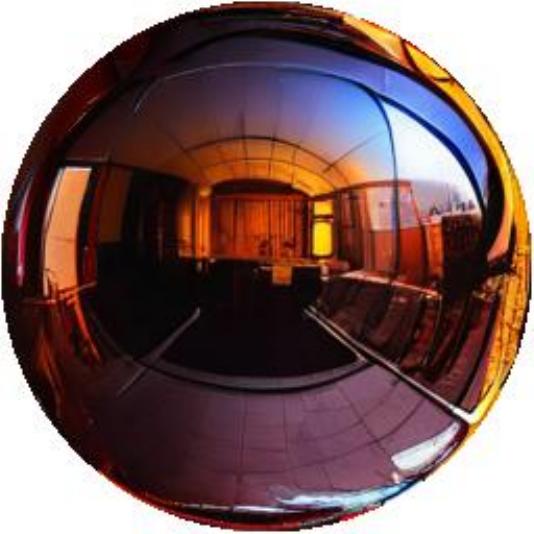}} & 
        \noindent\parbox[c]{0.082\textwidth}{\includegraphics[width=0.082\textwidth]{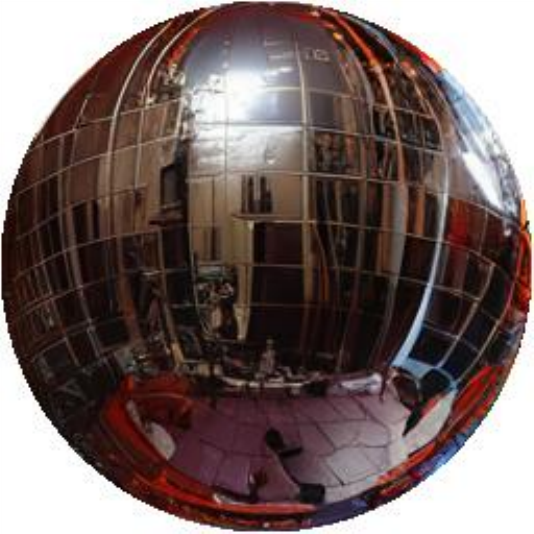}} & 
        \noindent\parbox[c]{0.082\textwidth}{\includegraphics[width=0.082\textwidth]{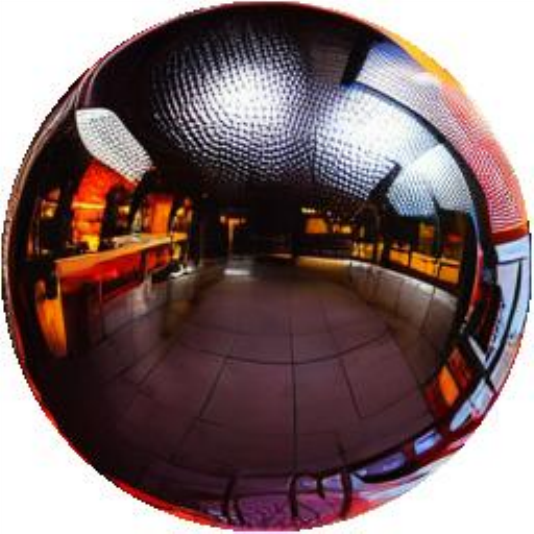}} & 
        \noindent\parbox[c]{0.082\textwidth}{\includegraphics[width=0.082\textwidth]{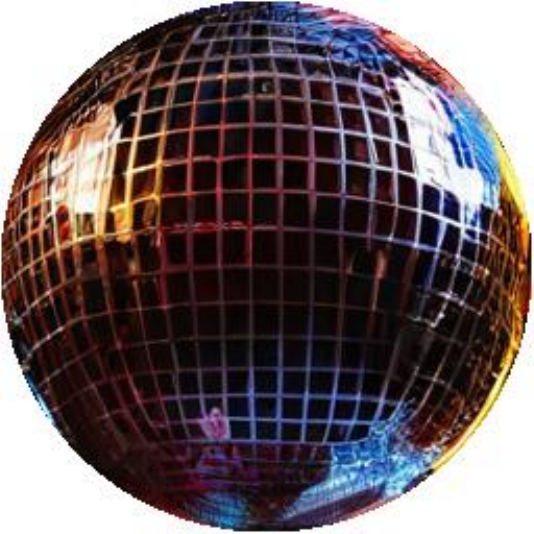}} & 
        \noindent\parbox[c]{0.082\textwidth}{\includegraphics[width=0.082\textwidth]{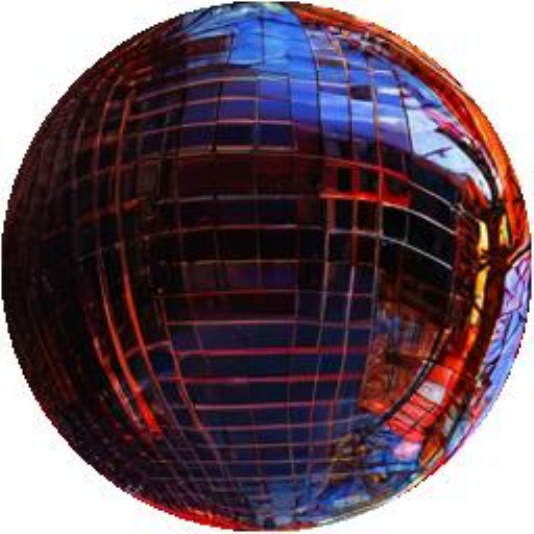}} & 
        \noindent\parbox[c]{0.082\textwidth}{\includegraphics[width=0.082\textwidth]{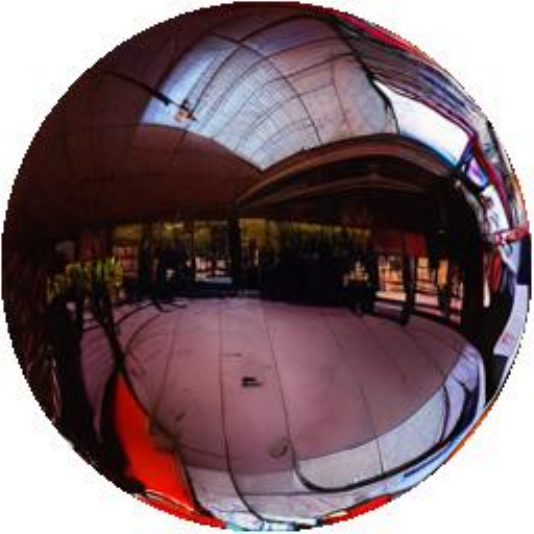}} & 
        \noindent\parbox[c]{0.082\textwidth}{\includegraphics[width=0.082\textwidth]{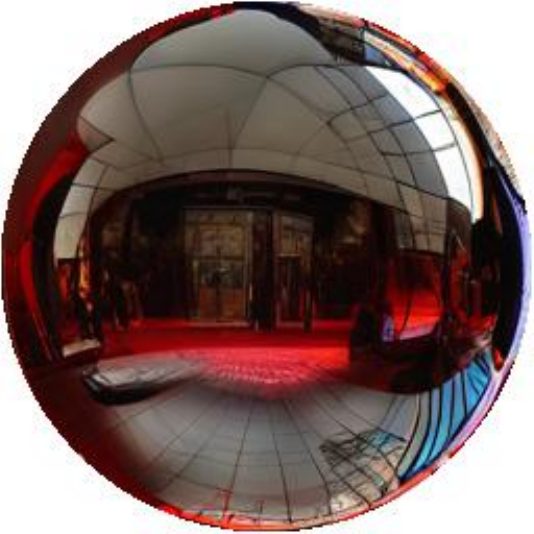}} & 
        \noindent\parbox[c]{0.082\textwidth}{\includegraphics[width=0.082\textwidth]{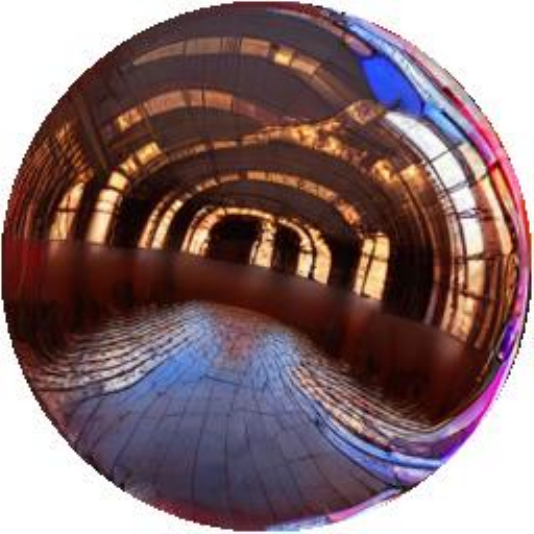}} & 
        \noindent\parbox[c]{0.082\textwidth}{\includegraphics[width=0.082\textwidth]{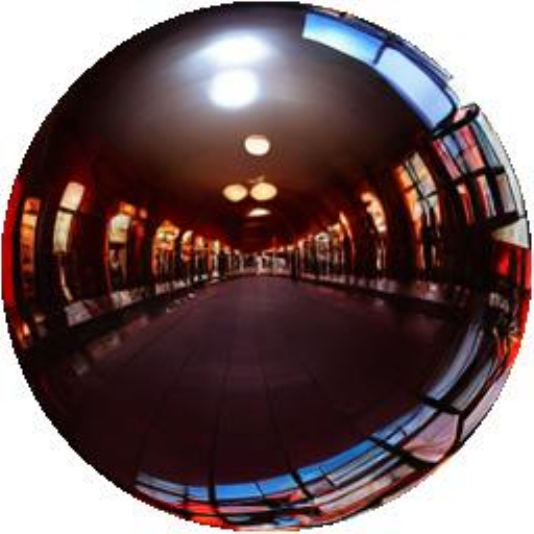}} & 
        \noindent\parbox[c]{0.082\textwidth}{\includegraphics[width=0.082\textwidth]{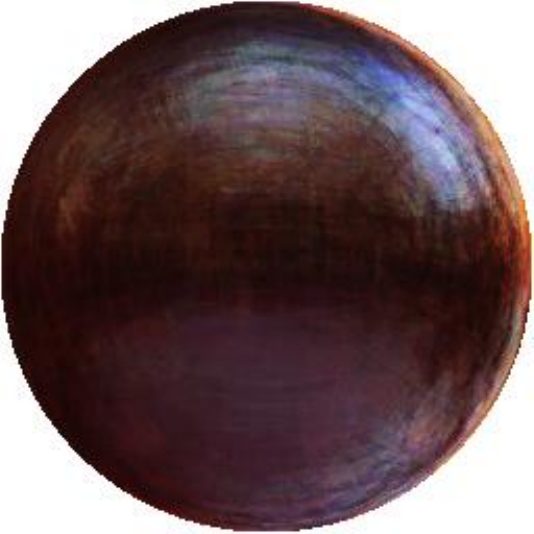}}

        \\

         &
        \multicolumn{1}{l}{\rotatebox[origin=c]{90}{\shortstack[l]{\tiny 3\textsuperscript{rd} iteration}}} &
        \noindent\parbox[c]{0.082\textwidth}{\includegraphics[width=0.082\textwidth]{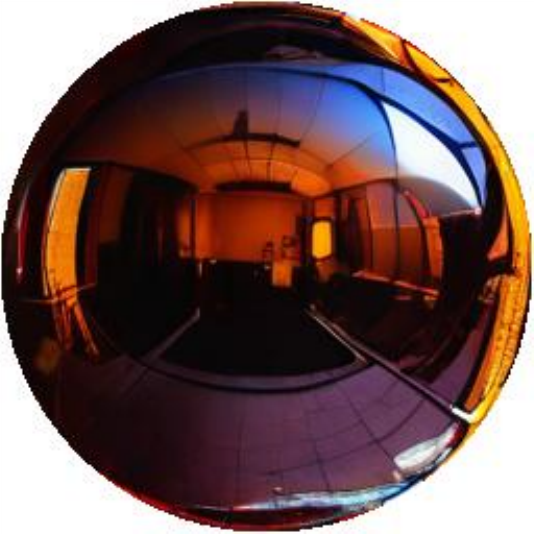}} & 
        \noindent\parbox[c]{0.082\textwidth}{\includegraphics[width=0.082\textwidth]{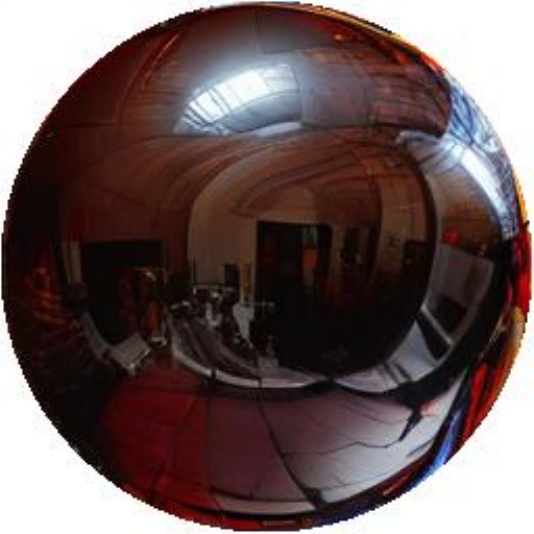}} & 
        \noindent\parbox[c]{0.082\textwidth}{\includegraphics[width=0.082\textwidth]{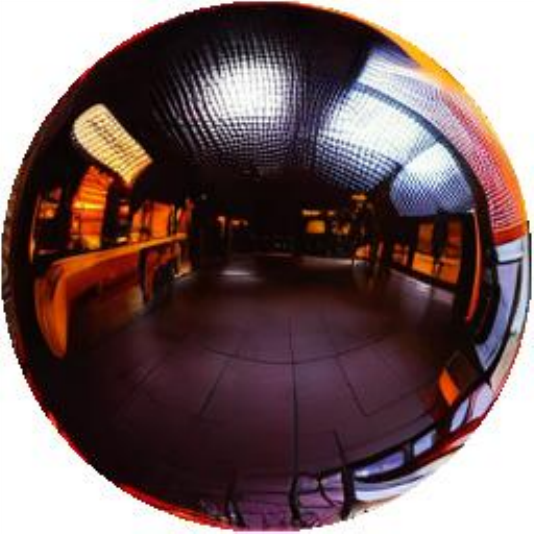}} & 
        \noindent\parbox[c]{0.082\textwidth}{\includegraphics[width=0.082\textwidth]{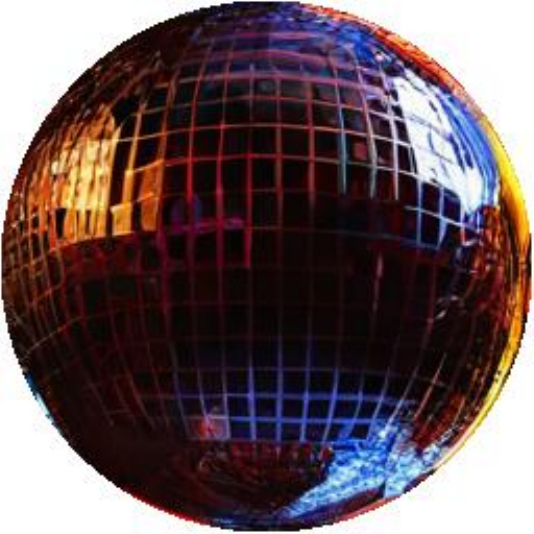}} & 
        \noindent\parbox[c]{0.082\textwidth}{\includegraphics[width=0.082\textwidth]{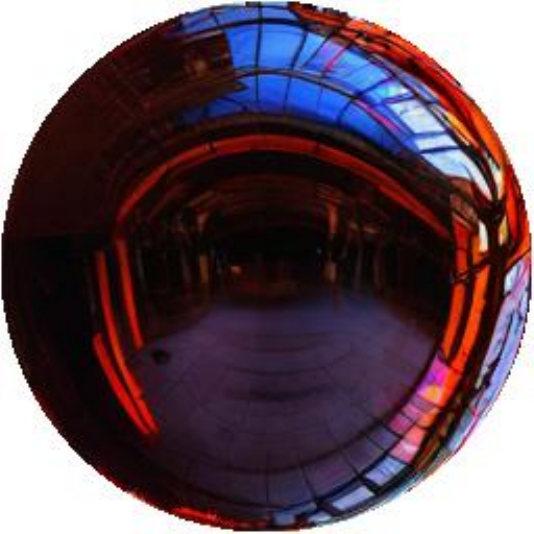}} & 
        \noindent\parbox[c]{0.082\textwidth}{\includegraphics[width=0.082\textwidth]{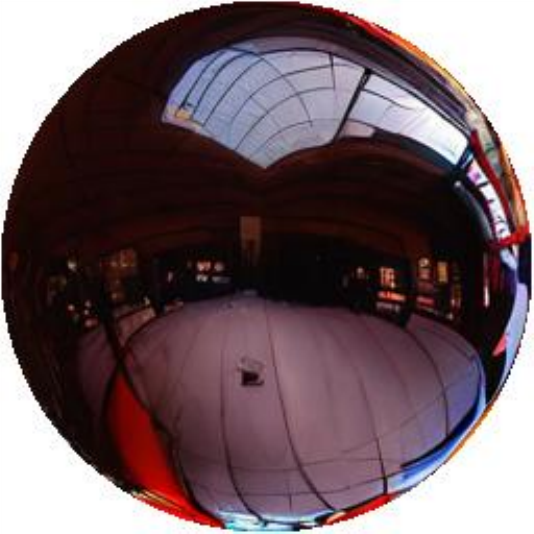}} & 
        \noindent\parbox[c]{0.082\textwidth}{\includegraphics[width=0.082\textwidth]{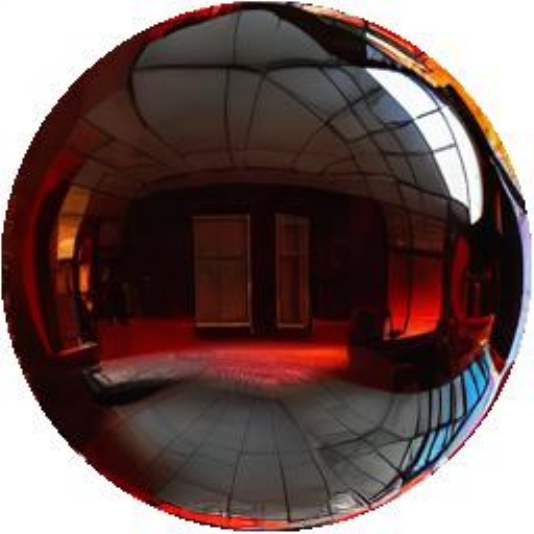}} & 
        \noindent\parbox[c]{0.082\textwidth}{\includegraphics[width=0.082\textwidth]{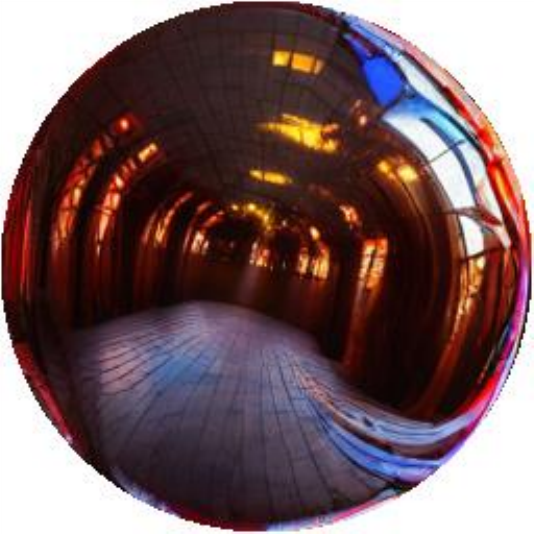}} & 
        \noindent\parbox[c]{0.082\textwidth}{\includegraphics[width=0.082\textwidth]{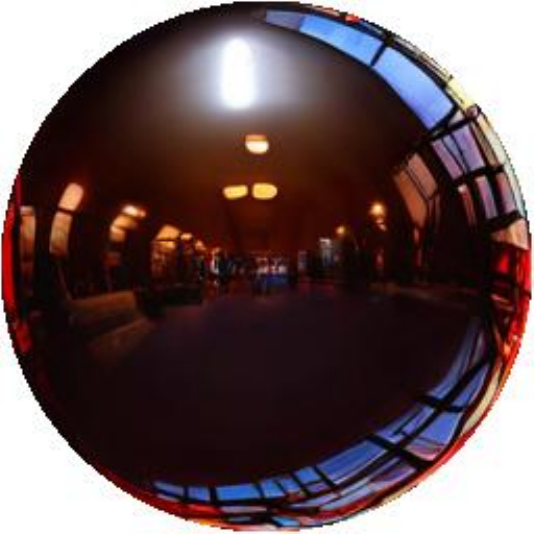}} & 
        \noindent\parbox[c]{0.082\textwidth}{\includegraphics[width=0.082\textwidth]{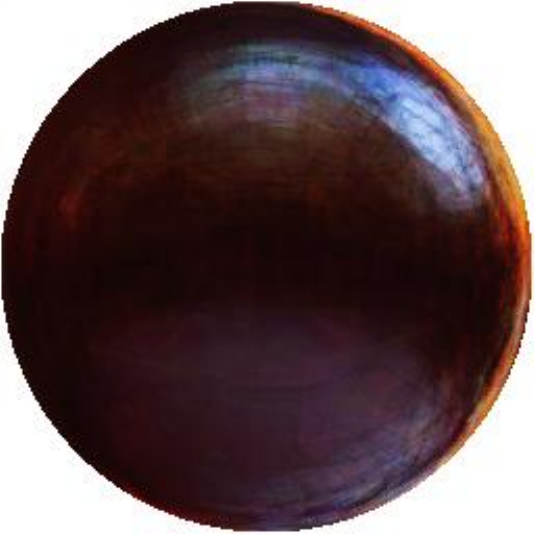}} 

        \\

         &
        \multicolumn{1}{l}{\rotatebox[origin=c]{90}{\shortstack[l]{\tiny 4\textsuperscript{th} iteration}}} &
        \noindent\parbox[c]{0.082\textwidth}{\includegraphics[width=0.082\textwidth]{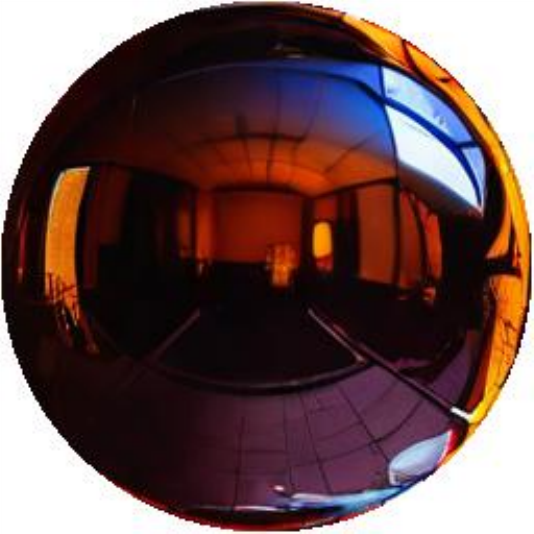}} & 
        \noindent\parbox[c]{0.082\textwidth}{\includegraphics[width=0.082\textwidth]{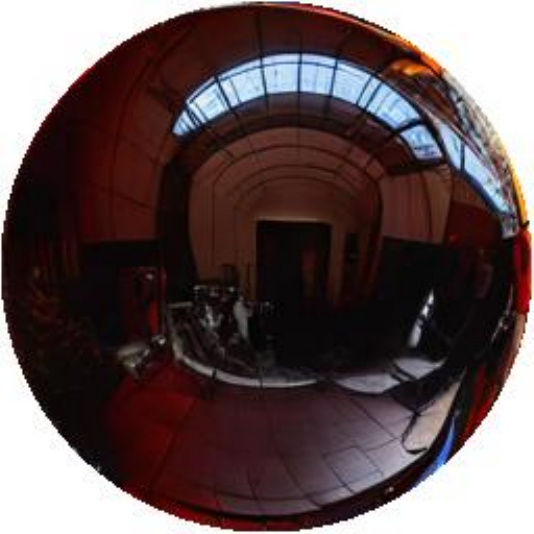}} & 
        \noindent\parbox[c]{0.082\textwidth}{\includegraphics[width=0.082\textwidth]{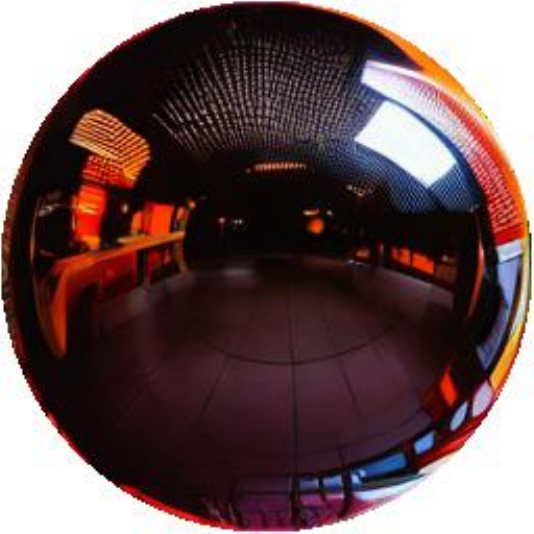}} & 
        \noindent\parbox[c]{0.082\textwidth}{\includegraphics[width=0.082\textwidth]{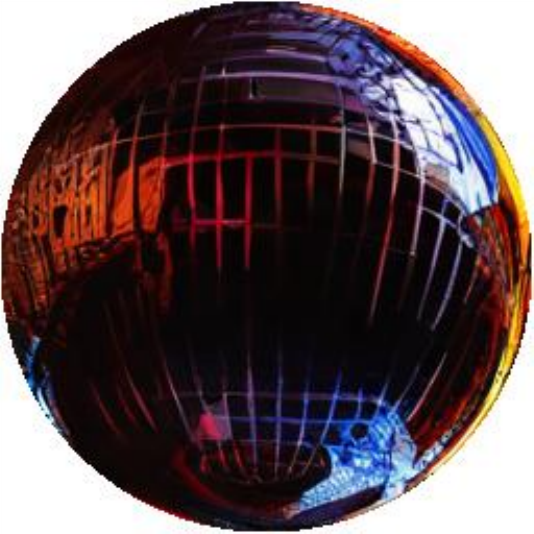}} & 
        \noindent\parbox[c]{0.082\textwidth}{\includegraphics[width=0.082\textwidth]{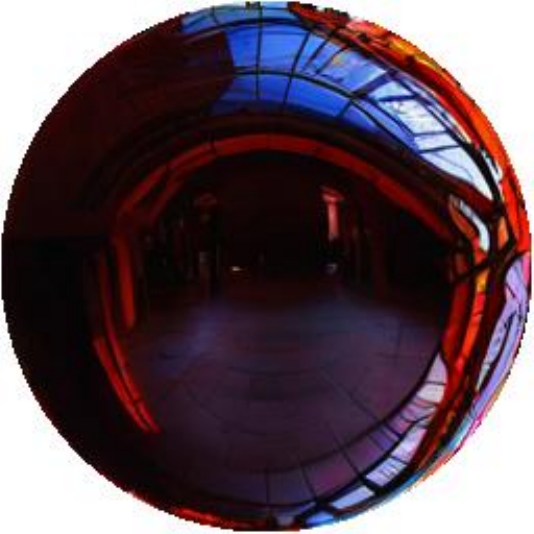}} & 
        \noindent\parbox[c]{0.082\textwidth}{\includegraphics[width=0.082\textwidth]{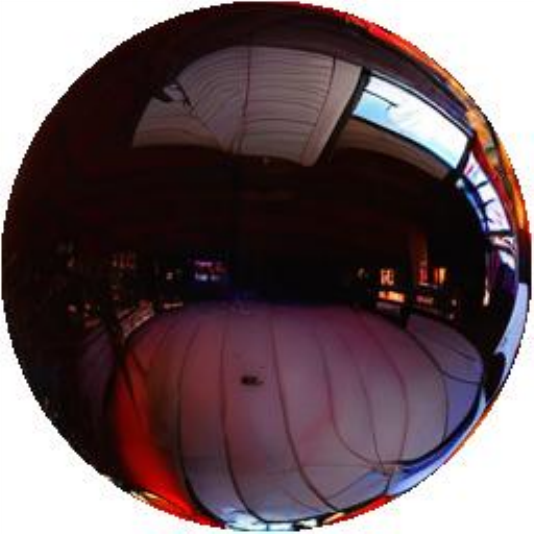}} & 
        \noindent\parbox[c]{0.082\textwidth}{\includegraphics[width=0.082\textwidth]{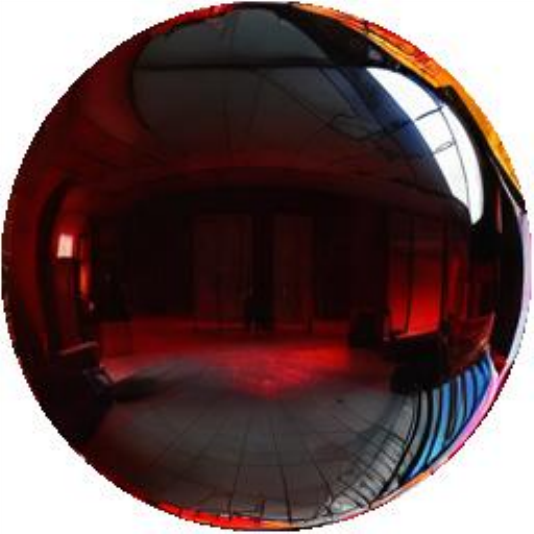}} & 
        \noindent\parbox[c]{0.082\textwidth}{\includegraphics[width=0.082\textwidth]{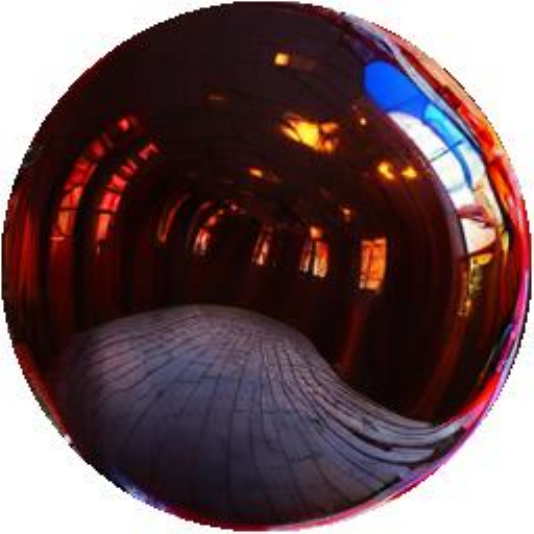}} & 
        \noindent\parbox[c]{0.082\textwidth}{\includegraphics[width=0.082\textwidth]{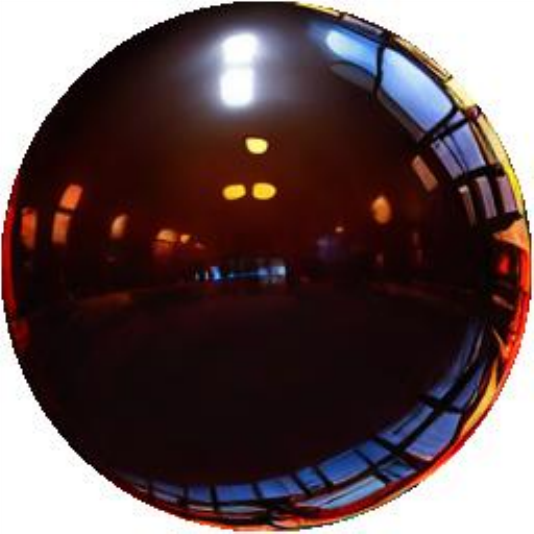}} & 
        \noindent\parbox[c]{0.082\textwidth}{\includegraphics[width=0.082\textwidth]{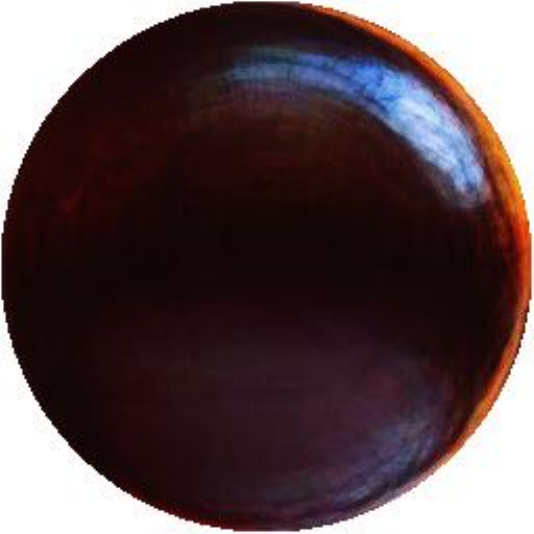}}

        \\

         &
        \multicolumn{1}{l}{\rotatebox[origin=c]{90}{\shortstack[l]{\tiny 5\textsuperscript{th} iteration}}} &
        \noindent\parbox[c]{0.082\textwidth}{\includegraphics[width=0.082\textwidth]{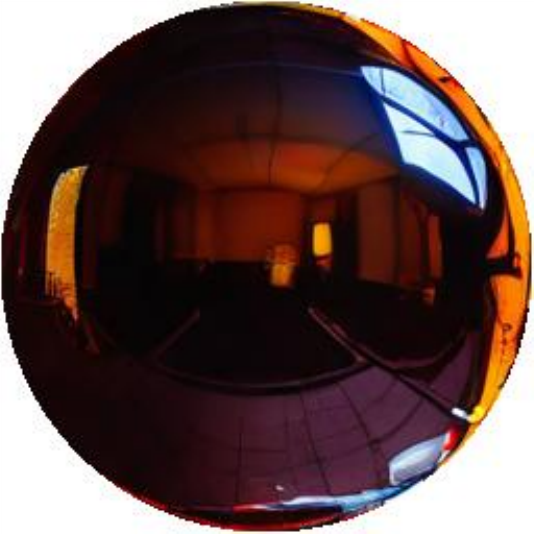}} & 
        \noindent\parbox[c]{0.082\textwidth}{\includegraphics[width=0.082\textwidth]{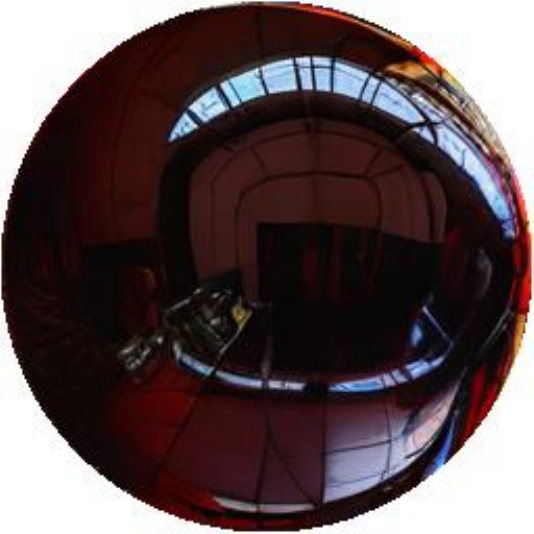}} & 
        \noindent\parbox[c]{0.082\textwidth}{\includegraphics[width=0.082\textwidth]{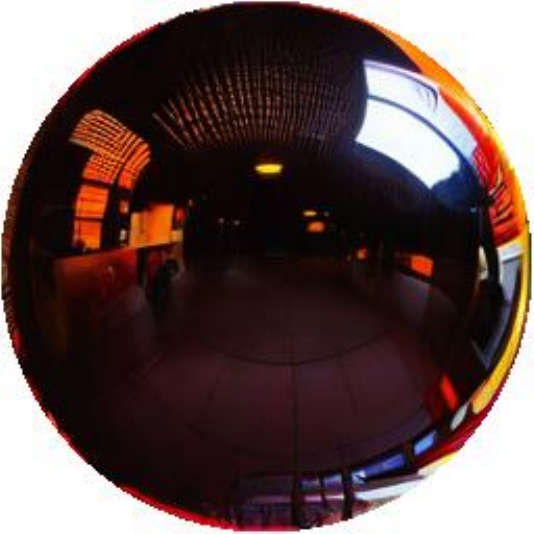}} & 
        \noindent\parbox[c]{0.082\textwidth}{\includegraphics[width=0.082\textwidth]{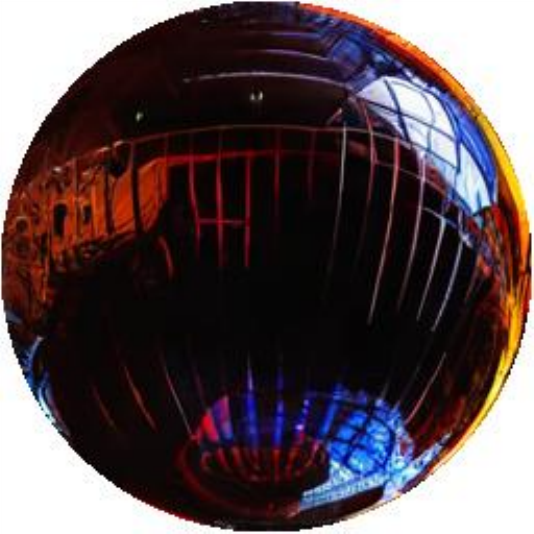}} & 
        \noindent\parbox[c]{0.082\textwidth}{\includegraphics[width=0.082\textwidth]{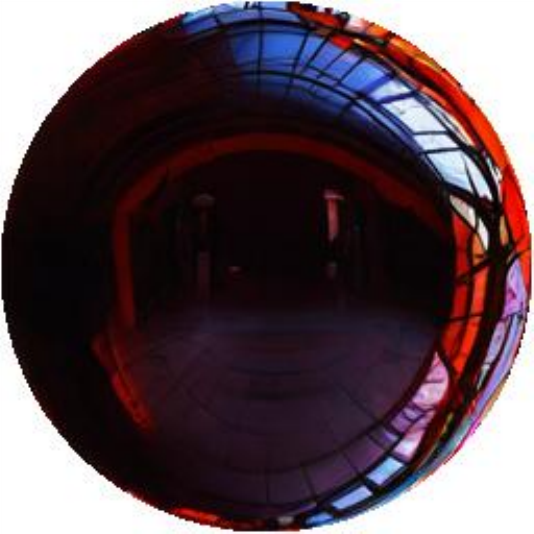}} & 
        \noindent\parbox[c]{0.082\textwidth}{\includegraphics[width=0.082\textwidth]{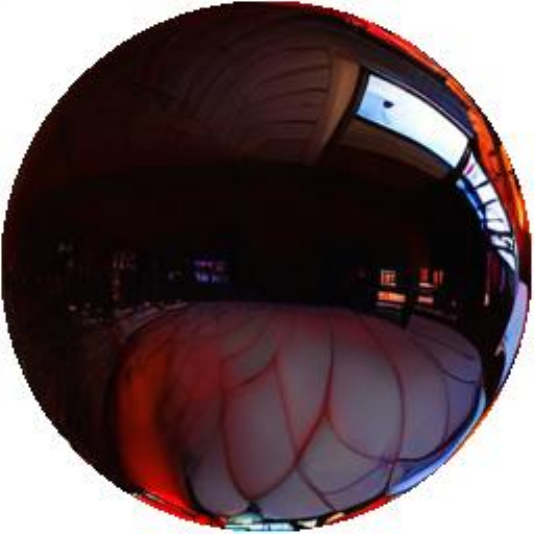}} & 
        \noindent\parbox[c]{0.082\textwidth}{\includegraphics[width=0.082\textwidth]{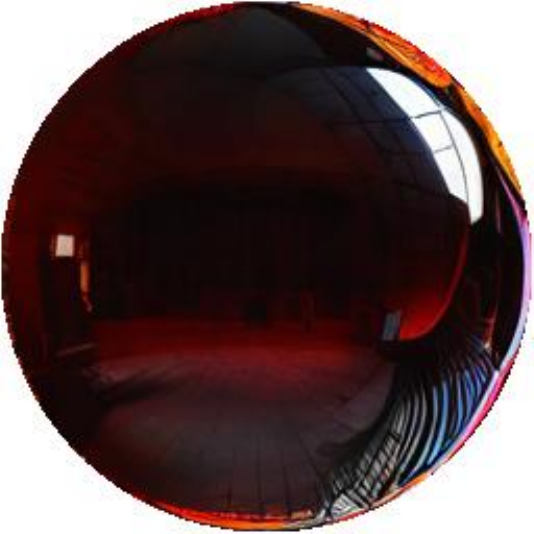}} & 
        \noindent\parbox[c]{0.082\textwidth}{\includegraphics[width=0.082\textwidth]{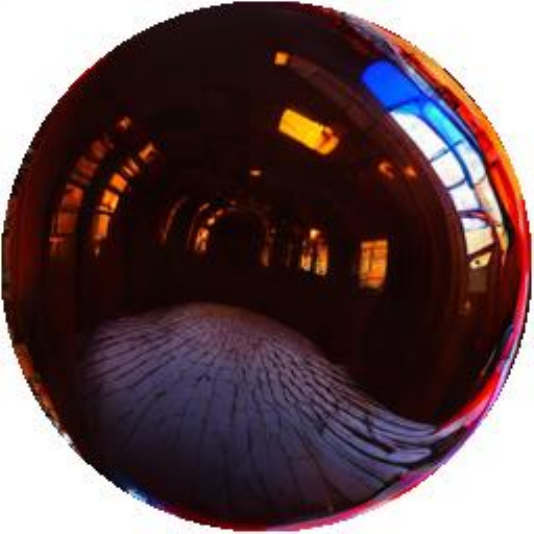}} & 
        \noindent\parbox[c]{0.082\textwidth}{\includegraphics[width=0.082\textwidth]{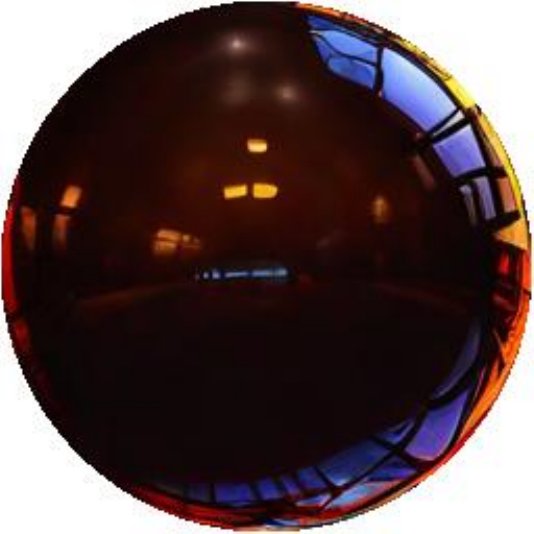}} & 
        \noindent\parbox[c]{0.082\textwidth}{\includegraphics[width=0.082\textwidth]{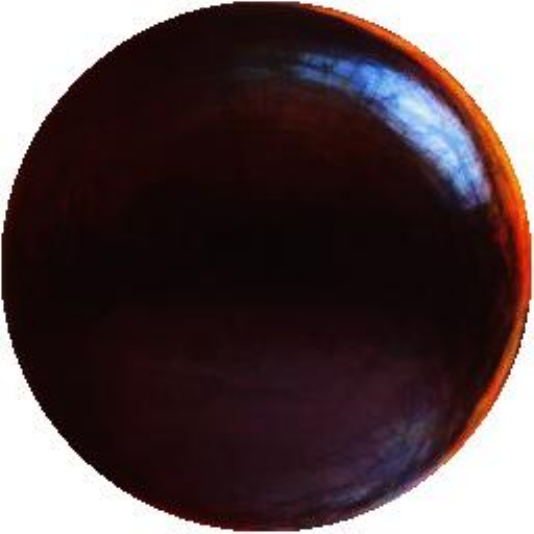}}

        \\

         &
        \multicolumn{1}{l}{\rotatebox[origin=c]{90}{\shortstack[l]{\tiny 6\textsuperscript{th} iteration}}} &
        \noindent\parbox[c]{0.082\textwidth}{\includegraphics[width=0.082\textwidth]{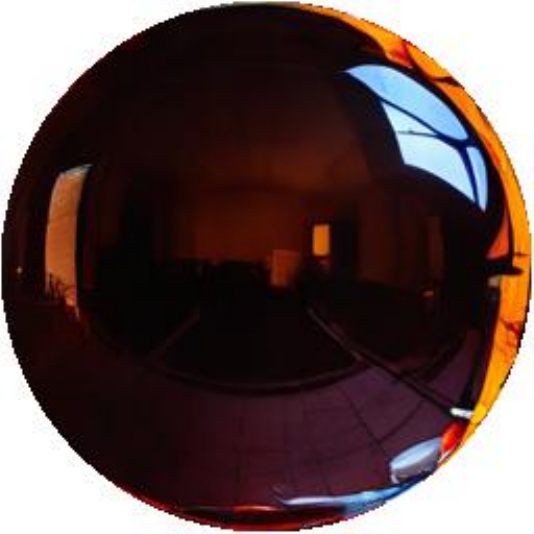}} & 
        \noindent\parbox[c]{0.082\textwidth}{\includegraphics[width=0.082\textwidth]{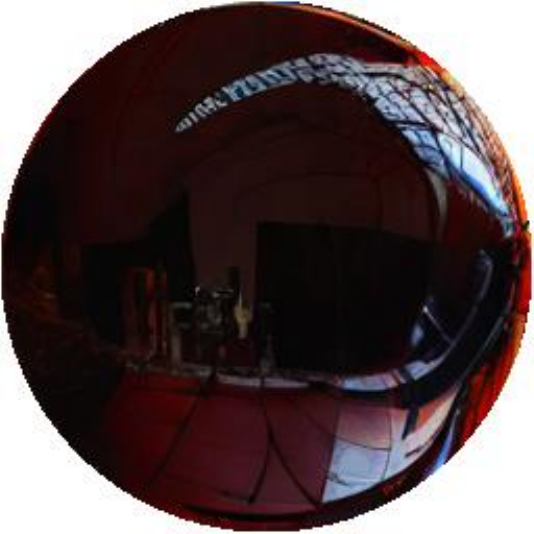}} & 
        \noindent\parbox[c]{0.082\textwidth}{\includegraphics[width=0.082\textwidth]{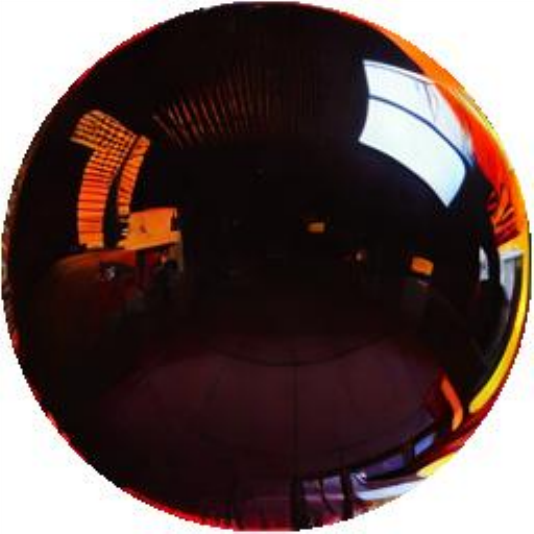}} & 
        \noindent\parbox[c]{0.082\textwidth}{\includegraphics[width=0.082\textwidth]{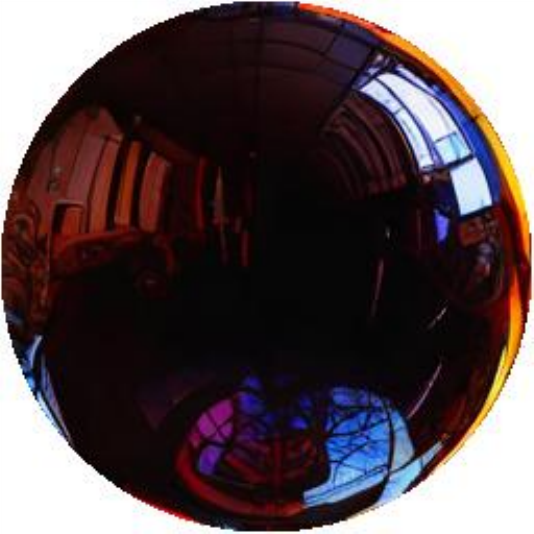}} & 
        \noindent\parbox[c]{0.082\textwidth}{\includegraphics[width=0.082\textwidth]{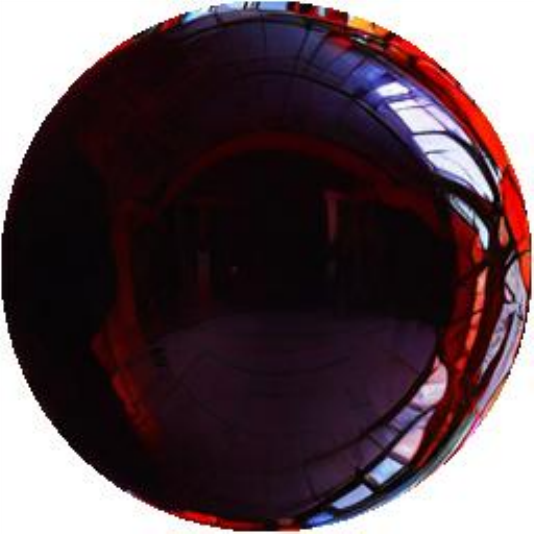}} & 
        \noindent\parbox[c]{0.082\textwidth}{\includegraphics[width=0.082\textwidth]{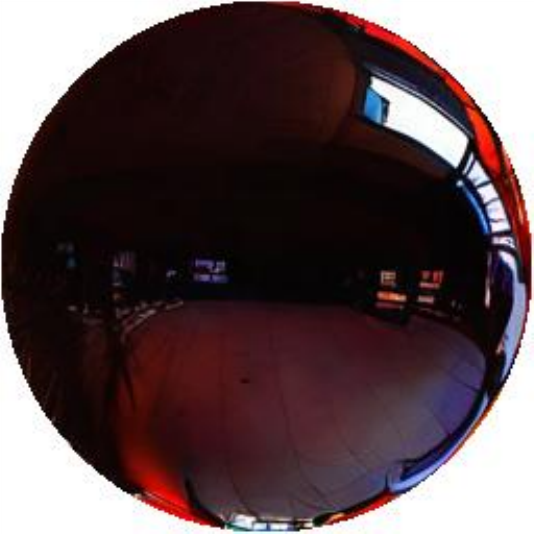}} & 
        \noindent\parbox[c]{0.082\textwidth}{\includegraphics[width=0.082\textwidth]{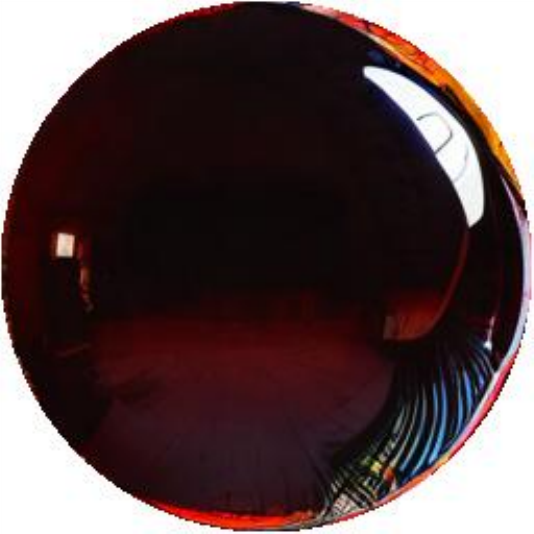}} & 
        \noindent\parbox[c]{0.082\textwidth}{\includegraphics[width=0.082\textwidth]{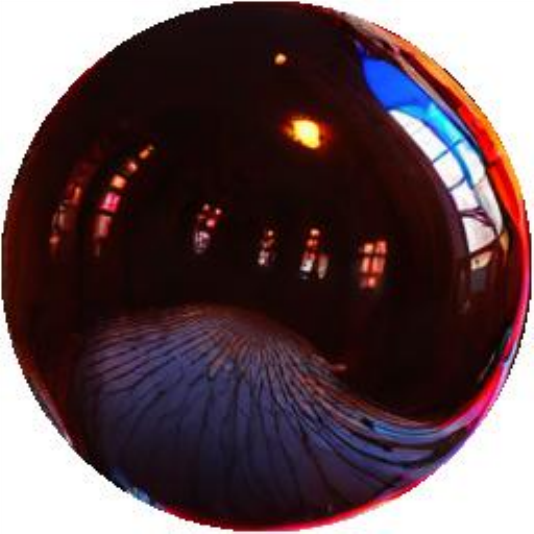}} & 
        \noindent\parbox[c]{0.082\textwidth}{\includegraphics[width=0.082\textwidth]{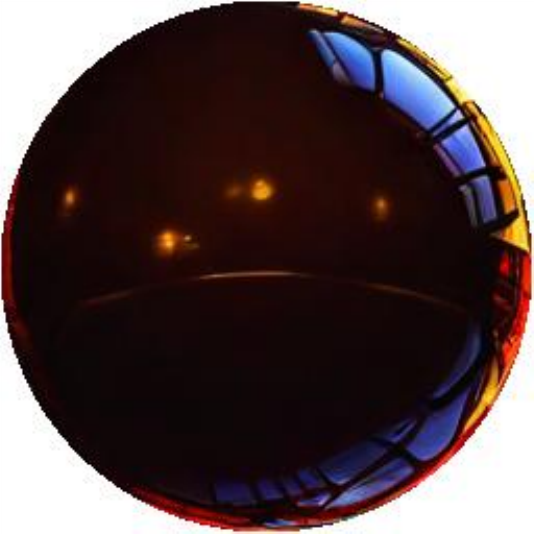}} & 
        \noindent\parbox[c]{0.082\textwidth}{\includegraphics[width=0.082\textwidth]{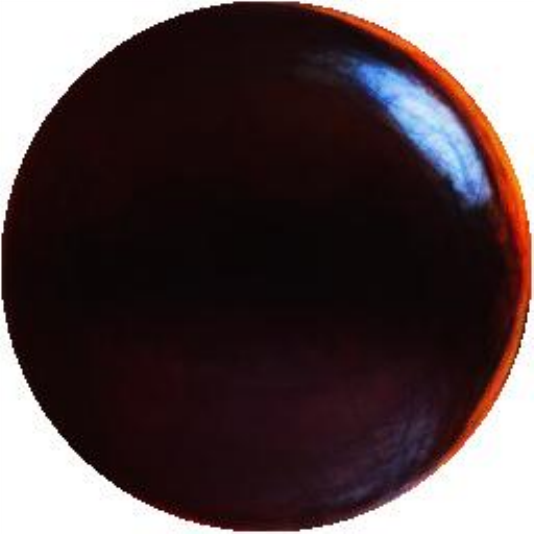}}

        \\

        \end{tabu}
    \caption{
    Repeatedly applying our iterative inpainting algorithm gradually produces chrome balls with better light estimation and fix degenerate balls such as Pred\#2, Pred\#4, and Pred\#5.
    }
    \label{fig:compare_median_distribution_aba_infinite}
\end{figure*}

Figure \ref{fig:compare_median_distribution_aba} presents additional results demonstrating how our iterative inpainting algorithm improves the consistency and quality of the generated chrome balls after one iteration. Figure \ref{fig:compare_median_distribution_aba_infinite} presents results with more iterations. Note that the experiments in our main paper were limited to two iterations due to our resource constraints. 

\subsection{Trade-off between running time and quality}\label{appendix:compute_tradeoff} 

\begin{figure}
    \centering
    \begin{tabu} to \textwidth {
        @{\hspace{3.0pt}}
        c@{\hspace{0.2pt}} 
        c@{\hspace{0.6pt}}
        c@{\hspace{0.2pt}}
        c@{\hspace{0.6pt}}
        c@{\hspace{0.2pt}}
        c@{\hspace{0.2pt}}
        @{}
    }
    \multicolumn{1}{c}{\textcolor{teal}{\footnotesize $k$=1}} &
    \multicolumn{1}{c}{\textcolor{Orchid}{\footnotesize $k$=2}} &
    \multicolumn{1}{c}{\textcolor{teal}{\footnotesize $k$=1}} &
    \multicolumn{1}{c}{\textcolor{Orchid}{\footnotesize $k$=2}} &
    \multicolumn{1}{c}{\textcolor{teal}{\footnotesize $k$=1}} &
    \multicolumn{1}{c}{\textcolor{Orchid}{\footnotesize $k$=2}}
    \\
    
    \noindent\parbox[c]{0.07\textwidth}{\includegraphics[width=0.070\textwidth]{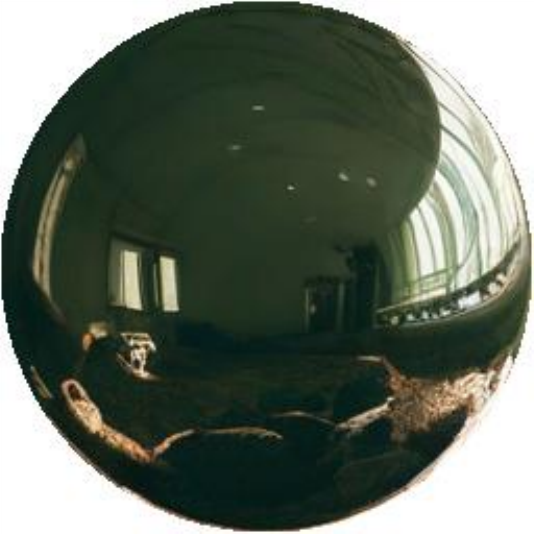}} & 
    \noindent\parbox[c]{0.07\textwidth}{\includegraphics[width=0.070\textwidth]{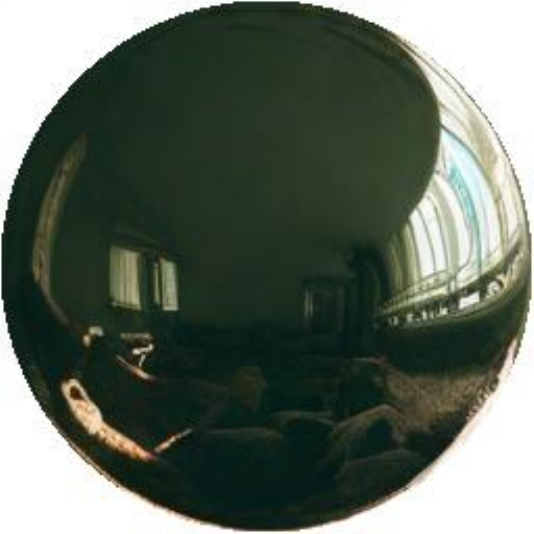}} & 
    \noindent\parbox[c]{0.07\textwidth}{\includegraphics[width=0.070\textwidth]{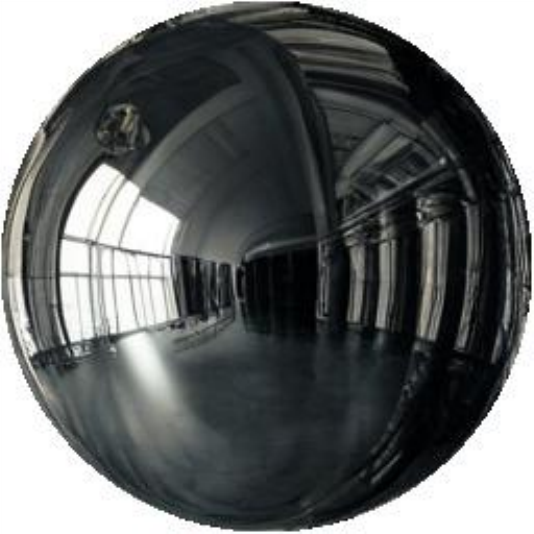}} & 
    \noindent\parbox[c]{0.07\textwidth}{\includegraphics[width=0.070\textwidth]{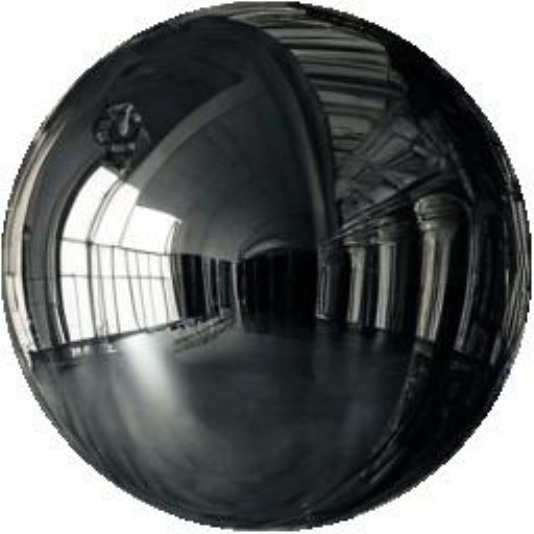}} &
    \noindent\parbox[c]{0.07\textwidth}{\includegraphics[width=0.070\textwidth]{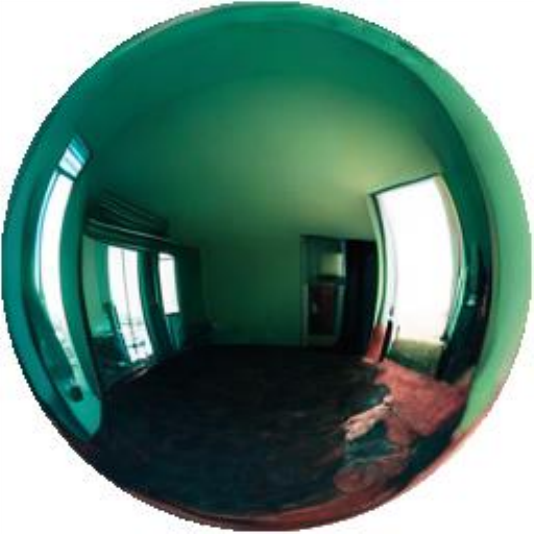}} & 
    \noindent\parbox[c]{0.07\textwidth}{\includegraphics[width=0.070\textwidth]{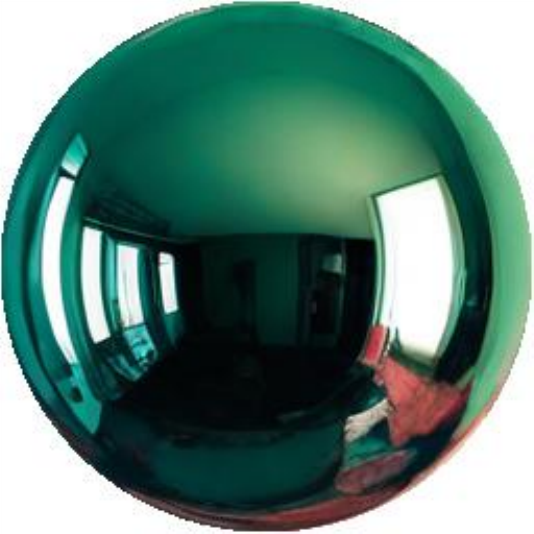}}
    \\
    
    \multicolumn{6}{c}{\footnotesize Input \#1 ($N$=30) \ \ \ \ \ \ \ Input \#2  ($N$=30) \ \ \ \ \ \ \ Input \#3  ($N$=30)} 
    \\
    
    \end{tabu}
    
    \caption{Qualitative results for the two configurations in Table \ref{tab:plot_time_accuracy}, marked with teal and purple colors. We can halve the running time with minimal quality degradation by decreasing the number of iterations $k$ to one.} 
    \label{fig:ablation_inpainting_ball_iteration}
\end{figure}

We experimented with various numbers of balls $N$ and iterations $k$ in the iterative inpainting algorithm on the validation set comprising 100 scenes from the Laval Indoor dataset \cite{garder2017lavelindoor} and 100 scenes from the Poly Haven dataset \cite{polyhaven}. As shown in Table \ref{tab:plot_time_accuracy}, increasing either of them generally leads to more accurate results. Notably, while two iterations of iterative inpainting ($k=2$) deliver the best score, reducing the number of iterations to one ($k=1$) halves the running time with minimal quality degradation as shown in Figure \ref{fig:ablation_inpainting_ball_iteration}. Note that we opted for $N = 30$ and $k = 2$, which takes about 30 minutes per image on an RTX 3090 Ti GPU, because of resource constraints.

\subsection{Inpainting ball size}
\begin{figure}[h]
    \centering
    \vspace{-4pt}
    \begin{tabu} to \textwidth {
        @{}
        c@{}
        c@{\hspace{1.0pt}}
        c@{\hspace{0.5pt}}
        c@{\hspace{0.5pt}}
        c@{\hspace{0.5pt}}
        c@{\hspace{0.5pt}}
        c@{}
    }
    
        &
        \multicolumn{1}{c}{\tiny Ball radius:} &
        \multicolumn{1}{c}{\tiny 128} &
        \multicolumn{1}{c}{\tiny 256} &
        \multicolumn{1}{c}{\tiny 384} &
        \multicolumn{1}{c}{\tiny 512} &
        \\
        
        \multicolumn{1}{l}{\rotatebox[origin=c]{90}{\shortstack[l]{\tiny Input \#1}}} &
        \noindent\parbox[c]{0.098\textwidth}{\includegraphics[width=0.098\textwidth]{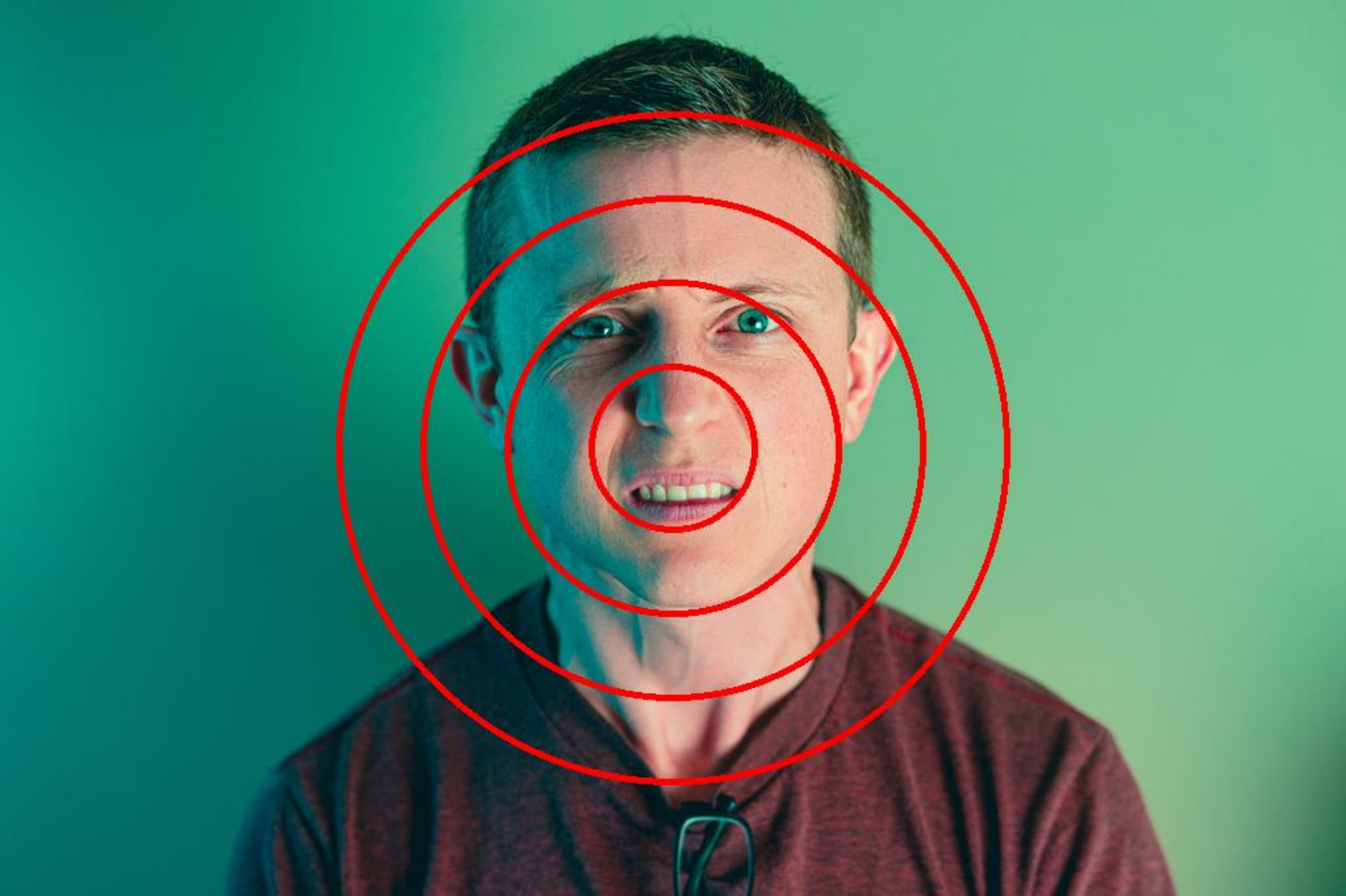}} & 
        \noindent\parbox[c]{0.065\textwidth}{\includegraphics[width=0.065\textwidth]{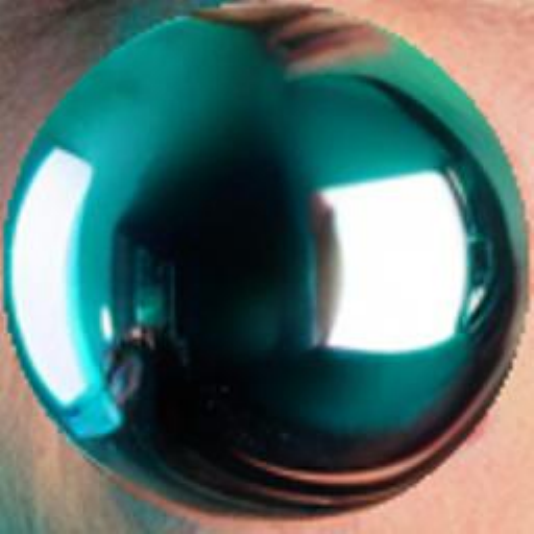}} & 
        \noindent\parbox[c]{0.065\textwidth}{\includegraphics[width=0.065\textwidth]{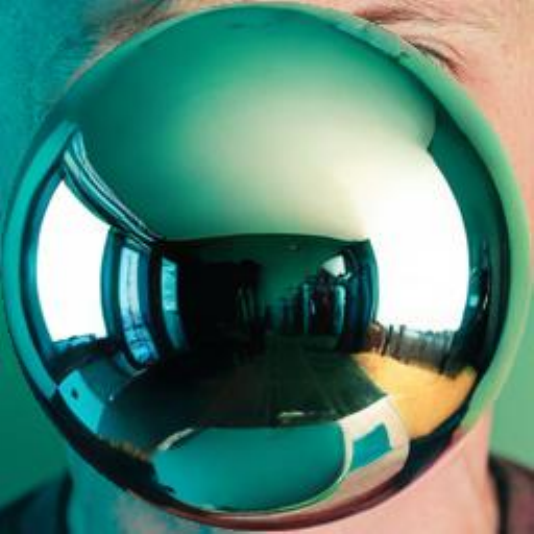}} & 
        \noindent\parbox[c]{0.065\textwidth}{\includegraphics[width=0.065\textwidth]{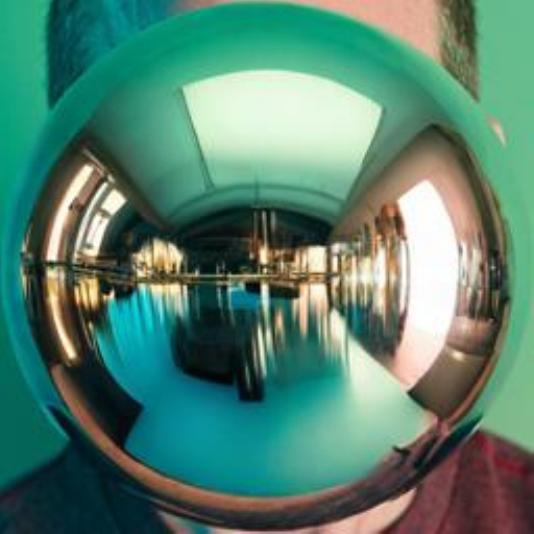}} & 
        \noindent\parbox[c]{0.065\textwidth}{\includegraphics[width=0.065\textwidth]{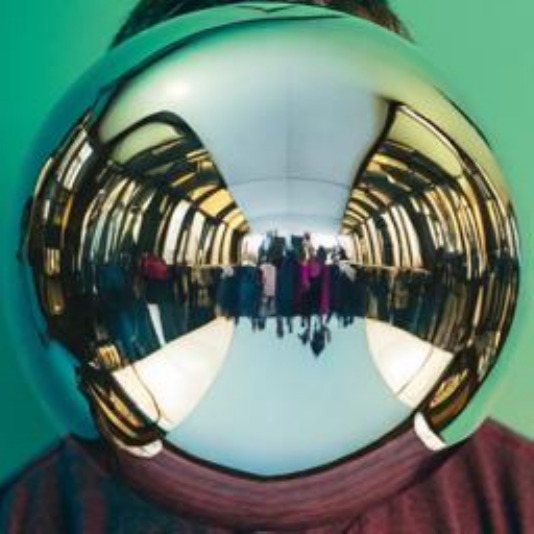}} & 
        \\

        \multicolumn{1}{l}{\rotatebox[origin=c]{90}{\shortstack[l]{\tiny Input \#2}}} &
        \noindent\parbox[c]{0.098\textwidth}{\includegraphics[width=0.098\textwidth]{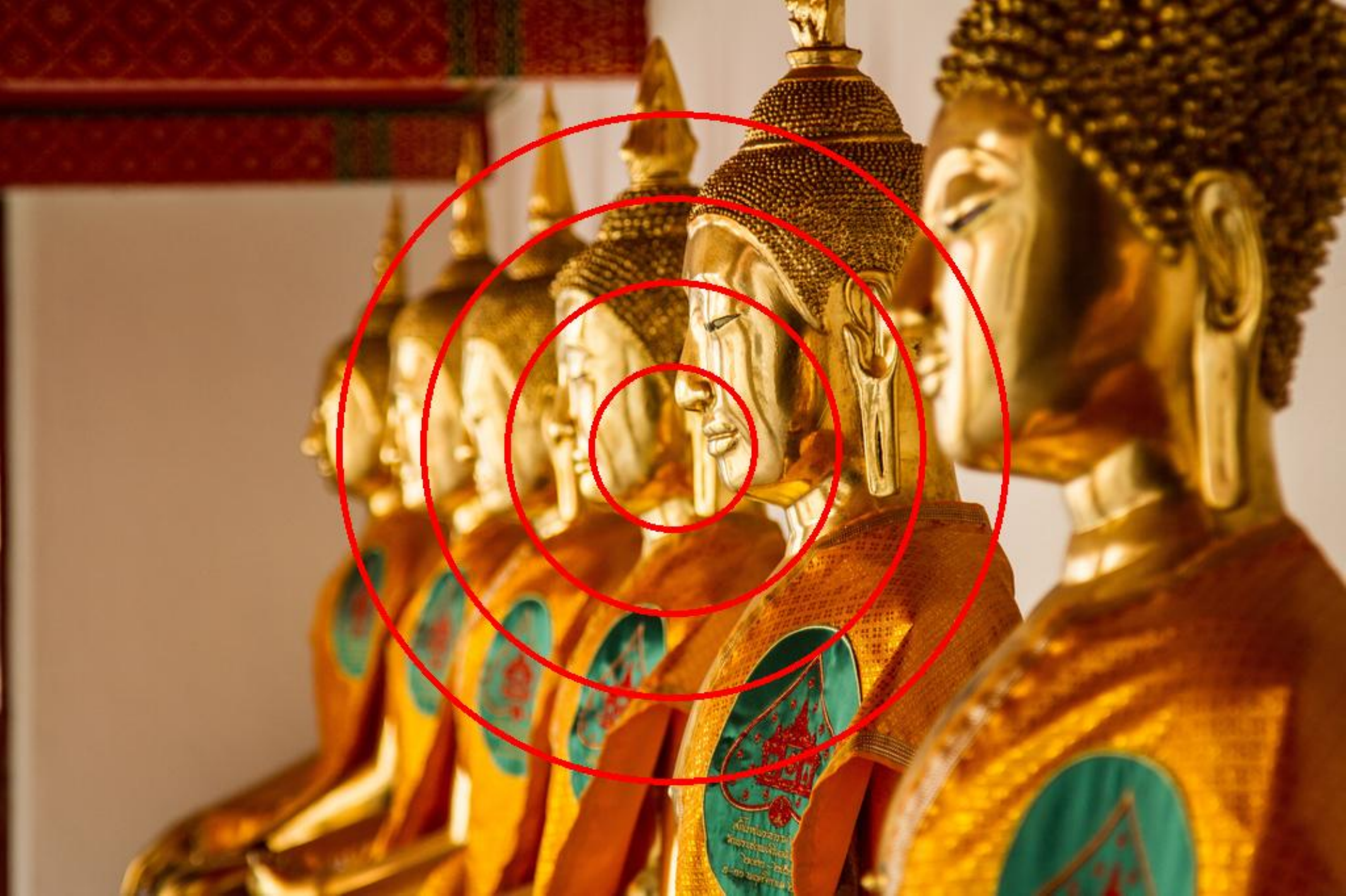}} & 
        \noindent\parbox[c]{0.065\textwidth}{\includegraphics[width=0.065\textwidth]{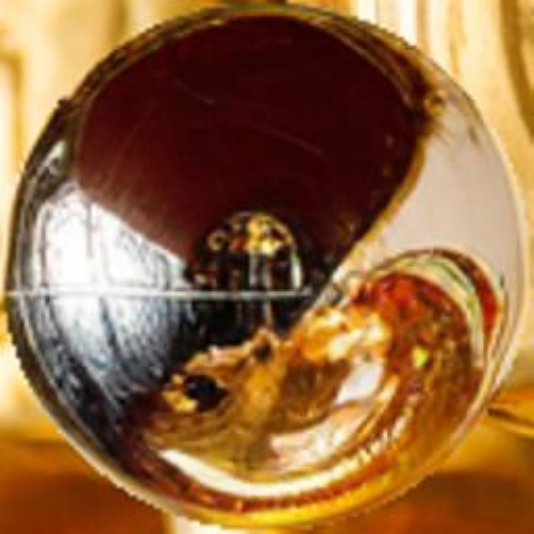}} & 
        \noindent\parbox[c]{0.065\textwidth}{\includegraphics[width=0.065\textwidth]{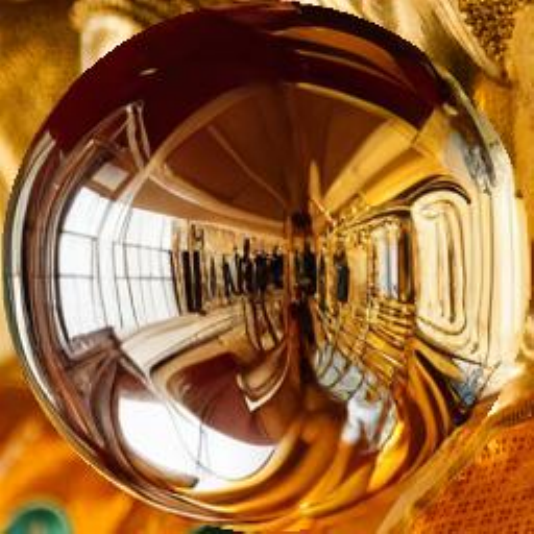}} & 
        \noindent\parbox[c]{0.065\textwidth}{\includegraphics[width=0.065\textwidth]{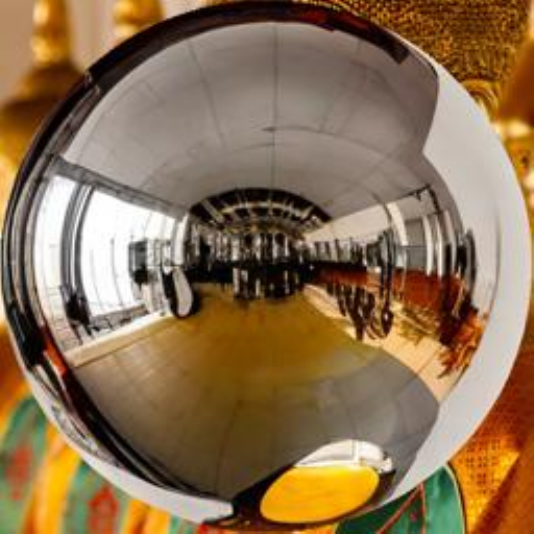}} & 
        \noindent\parbox[c]{0.065\textwidth}{\includegraphics[width=0.065\textwidth]{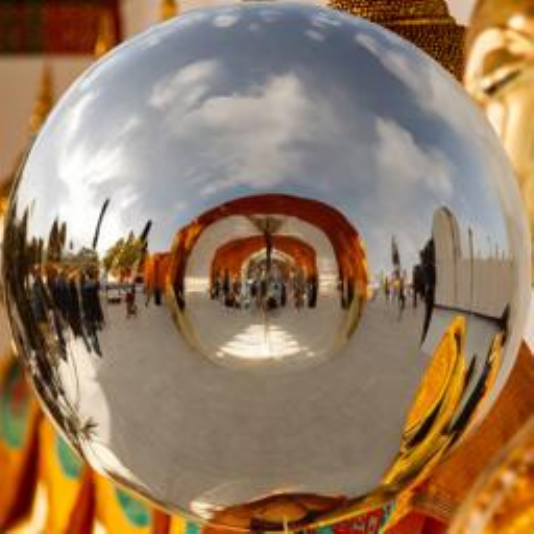}} & 
        \\

        \multicolumn{1}{l}{\rotatebox[origin=c]{90}{\shortstack[l]{\tiny Input \#2}}} &
        \noindent\parbox[c]{0.098\textwidth}{\includegraphics[width=0.098\textwidth]{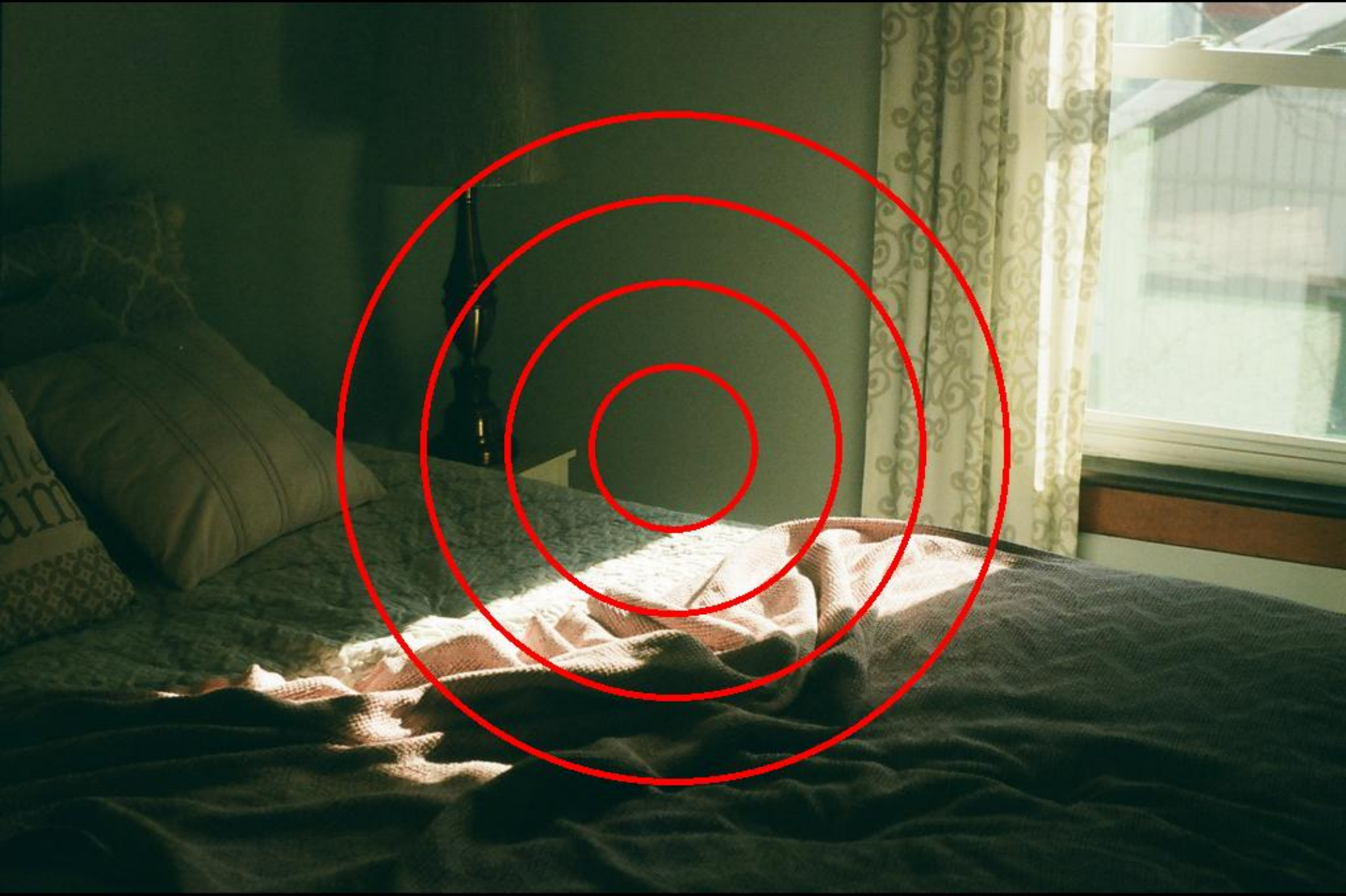}} & 
        \noindent\parbox[c]{0.065\textwidth}{\includegraphics[width=0.065\textwidth]{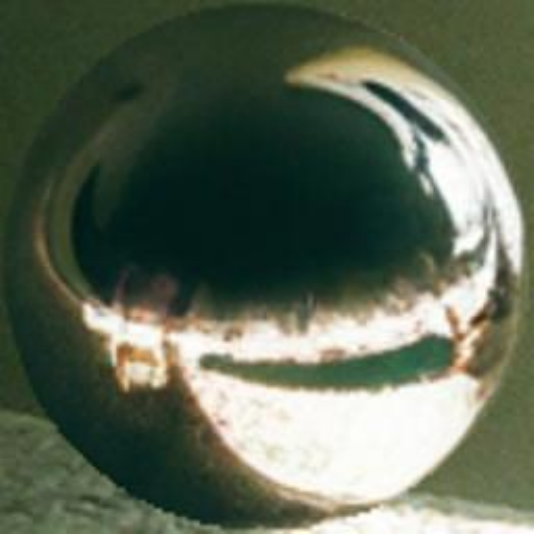}} & 
        \noindent\parbox[c]{0.065\textwidth}{\includegraphics[width=0.065\textwidth]{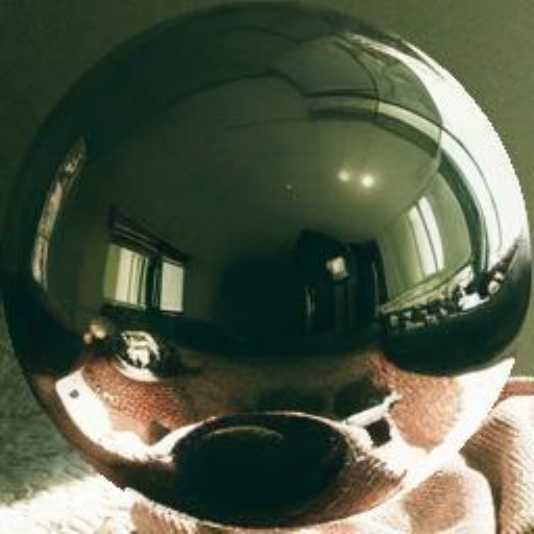}} & 
        \noindent\parbox[c]{0.065\textwidth}{\includegraphics[width=0.065\textwidth]{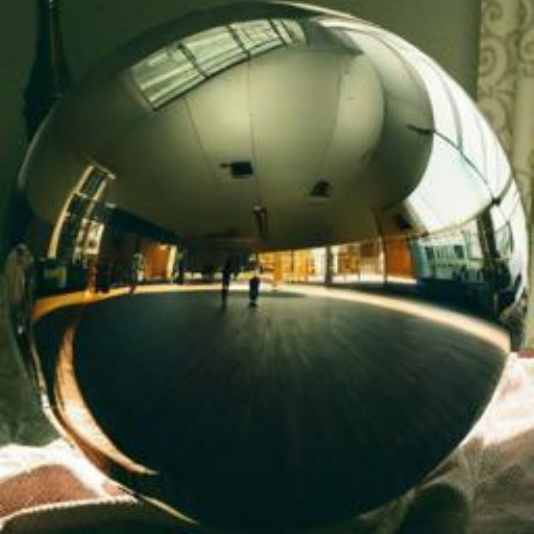}} & 
        \noindent\parbox[c]{0.065\textwidth}{\includegraphics[width=0.065\textwidth]{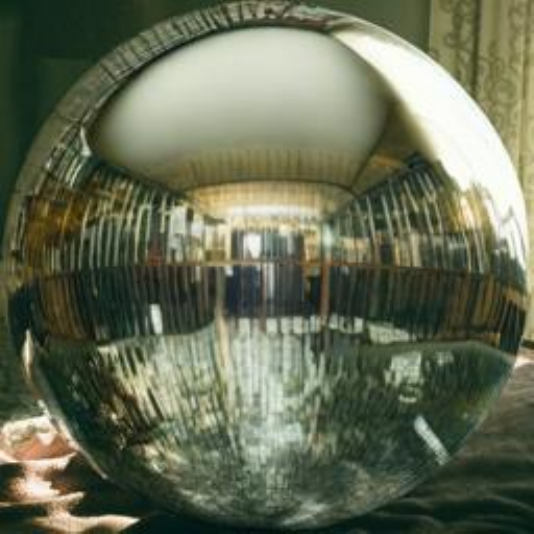}} & 
        \\

        \multicolumn{1}{l}{\rotatebox[origin=c]{90}{\shortstack[l]{\tiny Input \#2}}} &
        \noindent\parbox[c]{0.098\textwidth}{\includegraphics[width=0.098\textwidth]{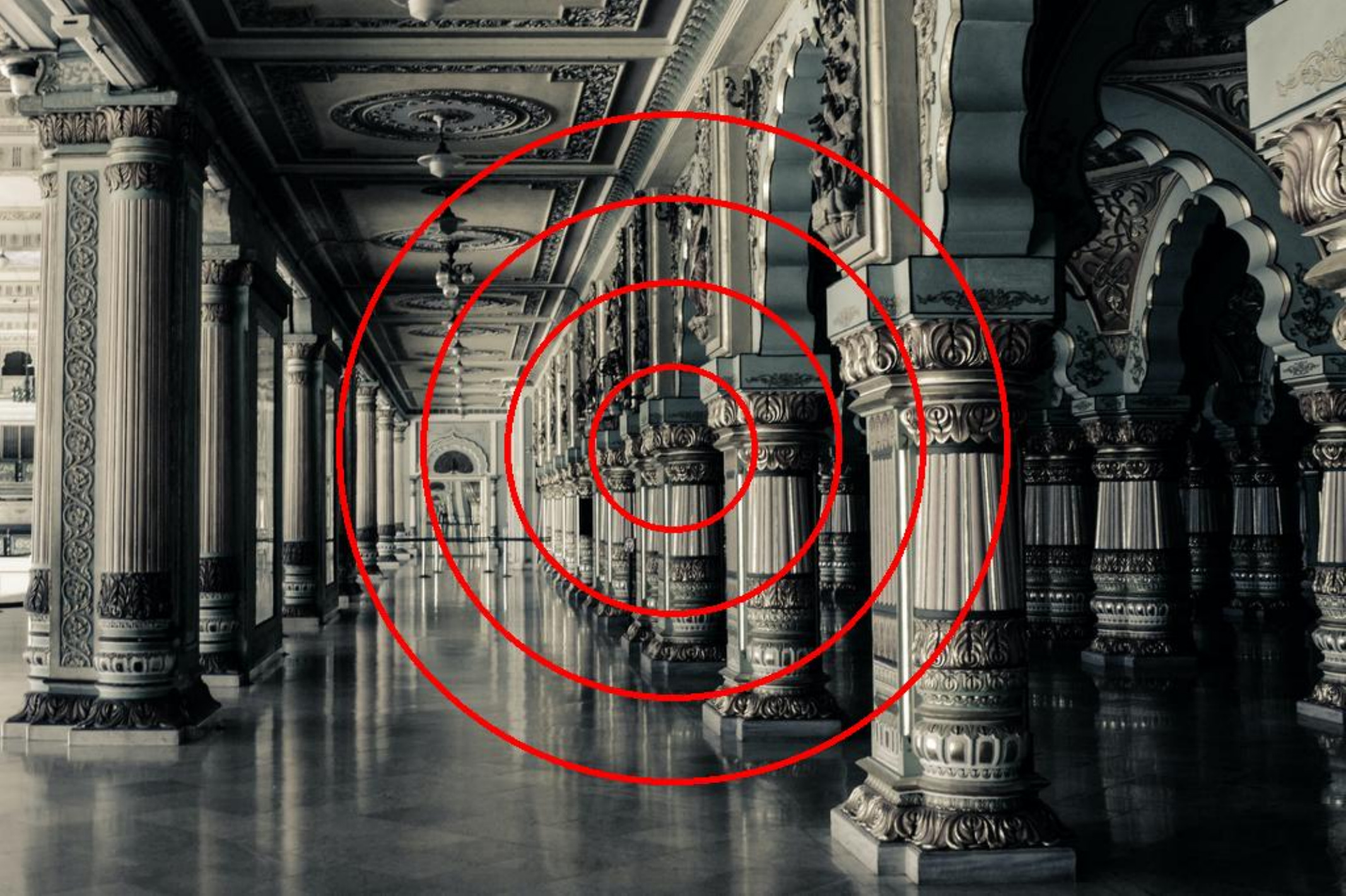}} & 
        \noindent\parbox[c]{0.065\textwidth}{\includegraphics[width=0.065\textwidth]{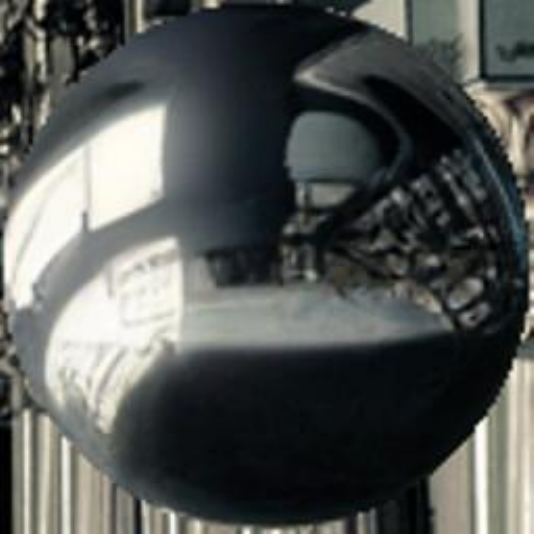}} & 
        \noindent\parbox[c]{0.065\textwidth}{\includegraphics[width=0.065\textwidth]{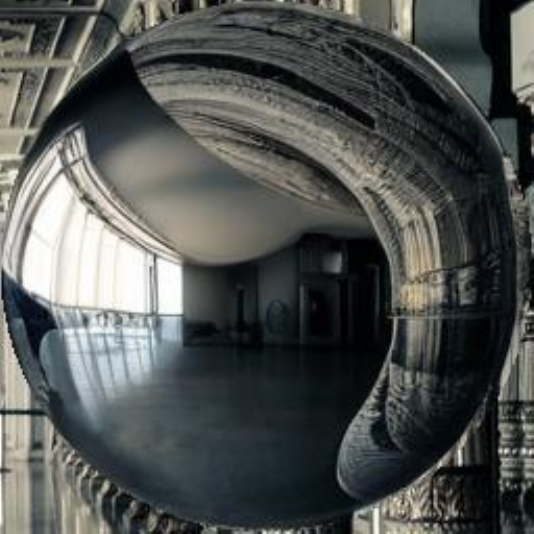}} & 
        \noindent\parbox[c]{0.065\textwidth}{\includegraphics[width=0.065\textwidth]{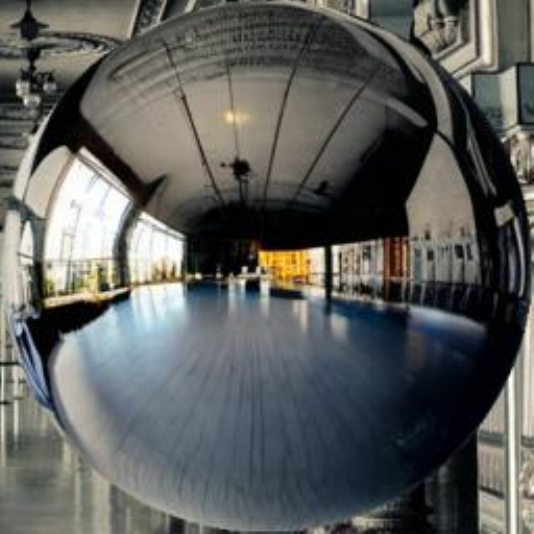}} & 
        \noindent\parbox[c]{0.065\textwidth}{\includegraphics[width=0.065\textwidth]{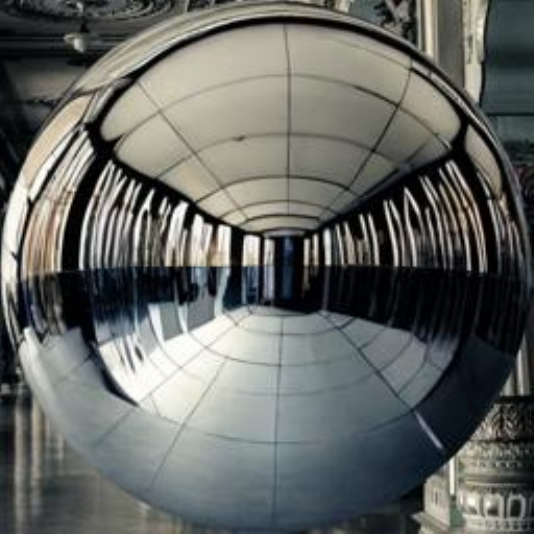}} & 
        \\
        
    \end{tabu}
    \vskip-3pt
    \begin{tabu} to \textwidth {
        @{}
        c@{}
        c@{\hspace{0.2pt}}
        c@{\hspace{0.2pt}}
        c@{\hspace{0.2pt}}
        c@{\hspace{2pt}}
        c@{\hspace{0.2pt}}
        c@{\hspace{0.2pt}}
        c@{\hspace{0.2pt}}
        c@{\hspace{0.2pt}}
        c@{}
    }

        \vspace{-6pt}
        
        &
        \multicolumn{4}{c}{\tiny Median balls for input \#1} &
        \multicolumn{4}{c}{\tiny Median balls for input \#2} &
        \\

        \multicolumn{1}{l}{\vspace{-1pt} \tiny Ball radius:} &
        \multicolumn{1}{c}{\tiny 128} &
        \multicolumn{1}{c}{\tiny 256} &
        \multicolumn{1}{c}{\tiny 384} &
        \multicolumn{1}{c}{\tiny 512} &
        \multicolumn{1}{c}{\tiny 128} &
        \multicolumn{1}{c}{\tiny 256} &
        \multicolumn{1}{c}{\tiny 384} &
        \multicolumn{1}{c}{\tiny 512} &
        \\

        \multicolumn{1}{l}{\shortstack[l]{\tiny Last \\ \tiny iteration}} &
        \noindent\parbox[c]{0.050\textwidth}{\includegraphics[width=0.050\textwidth]{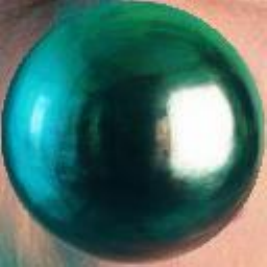}} & 
        \noindent\parbox[c]{0.050\textwidth}{\includegraphics[width=0.050\textwidth]{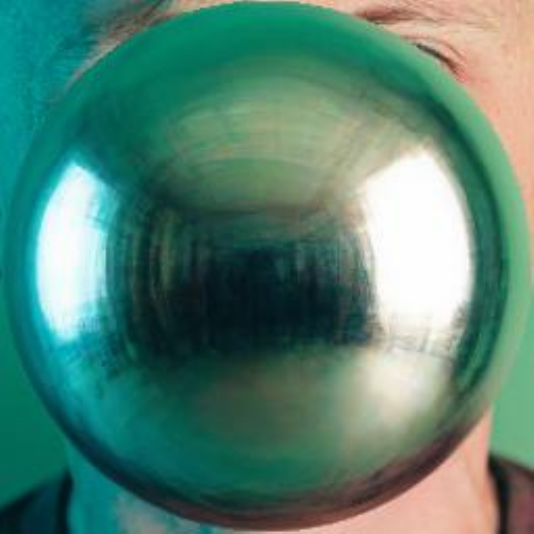}} & 
        \noindent\parbox[c]{0.050\textwidth}{\includegraphics[width=0.050\textwidth]{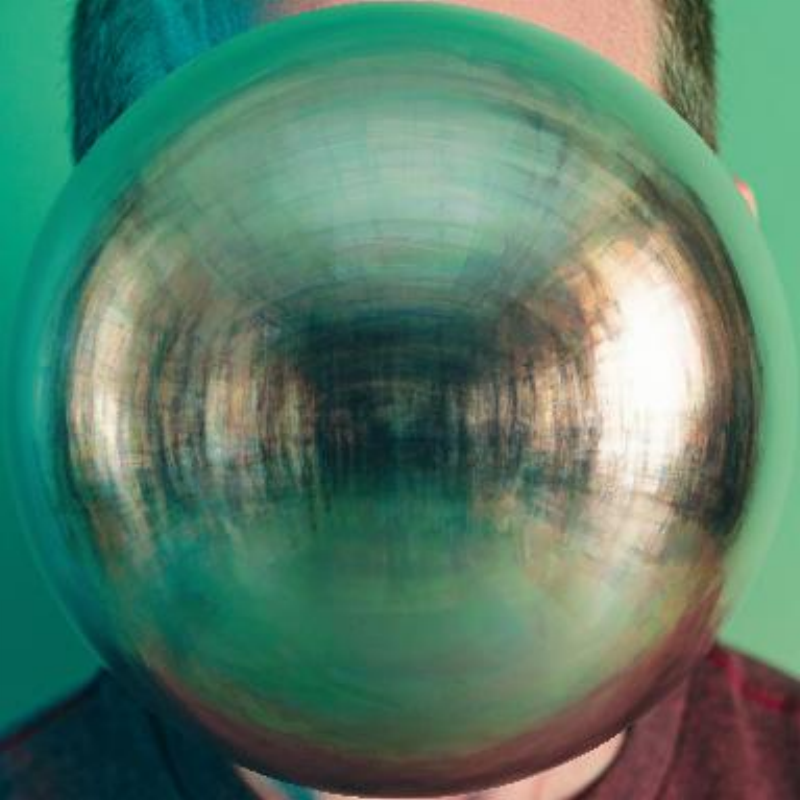}} & 
        \noindent\parbox[c]{0.050\textwidth}{\includegraphics[width=0.050\textwidth]{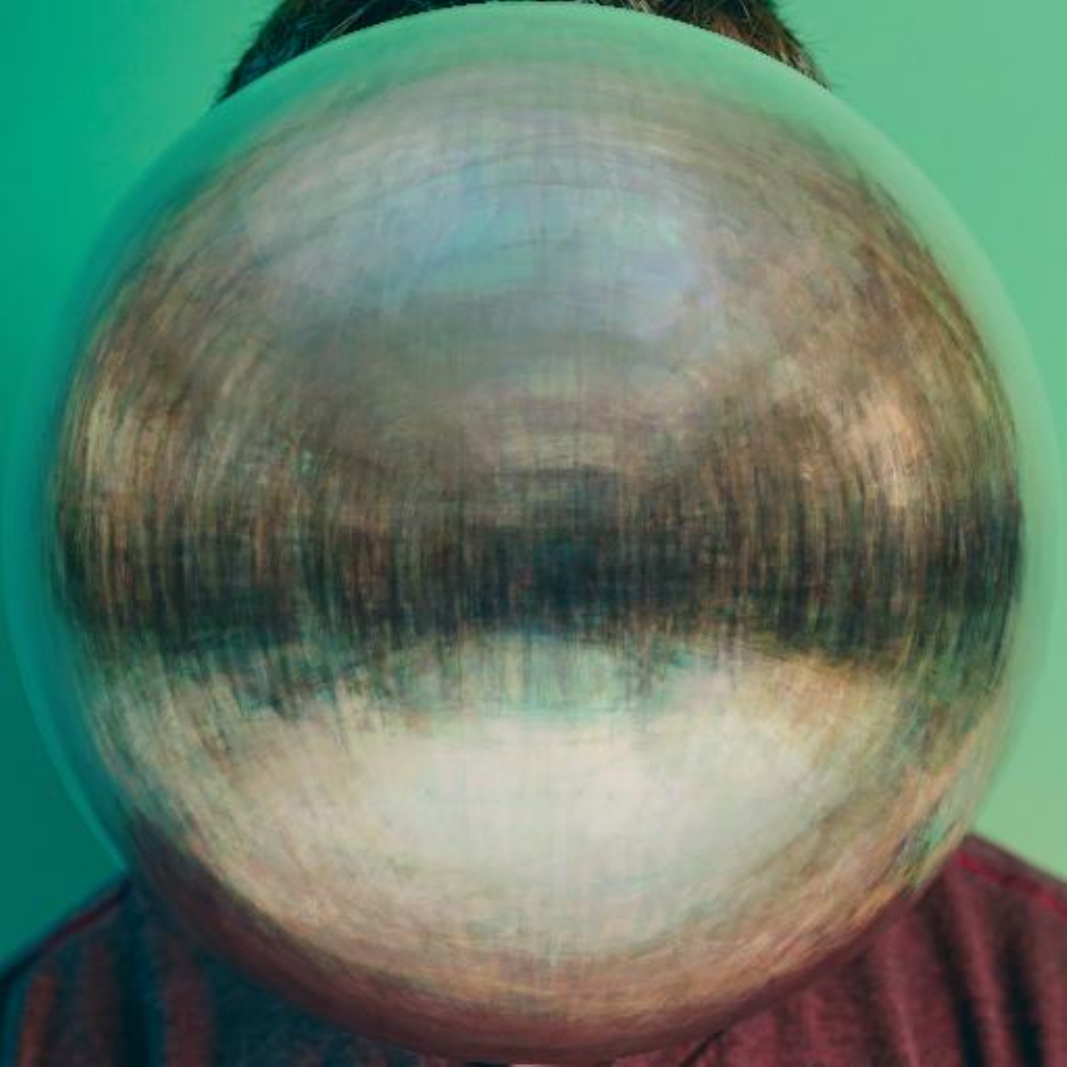}} & 

        \noindent\parbox[c]{0.050\textwidth}{\includegraphics[width=0.050\textwidth]{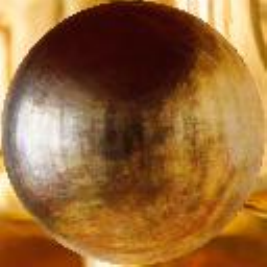}} & 
        \noindent\parbox[c]{0.050\textwidth}{\includegraphics[width=0.050\textwidth]{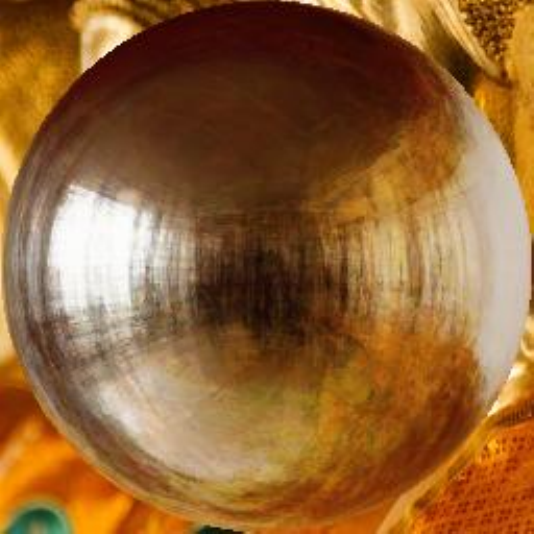}} & 
        \noindent\parbox[c]{0.050\textwidth}{\includegraphics[width=0.050\textwidth]{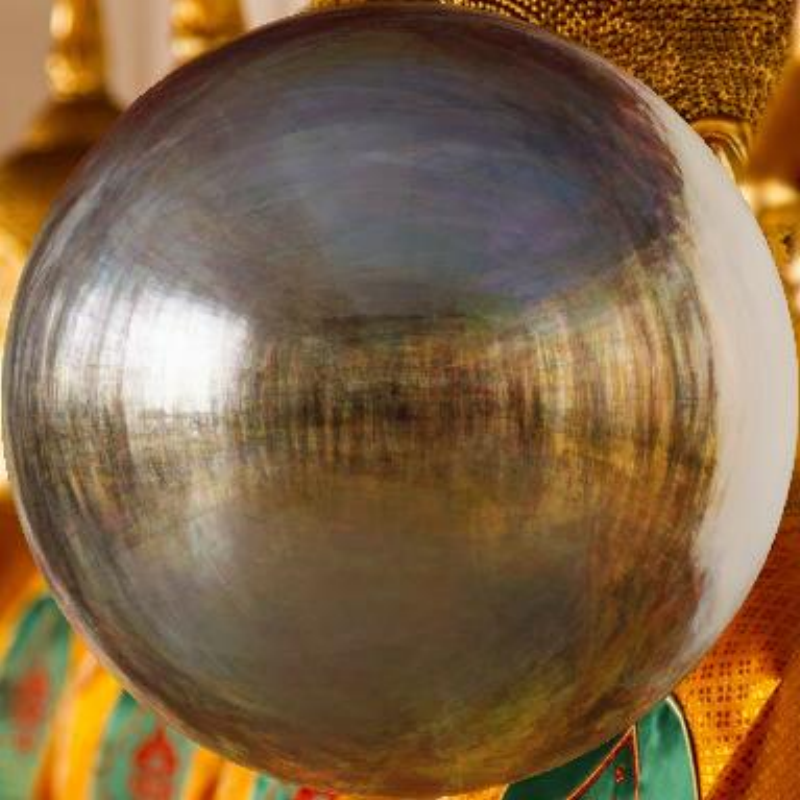}} & 
        \noindent\parbox[c]{0.050\textwidth}{\includegraphics[width=0.050\textwidth]{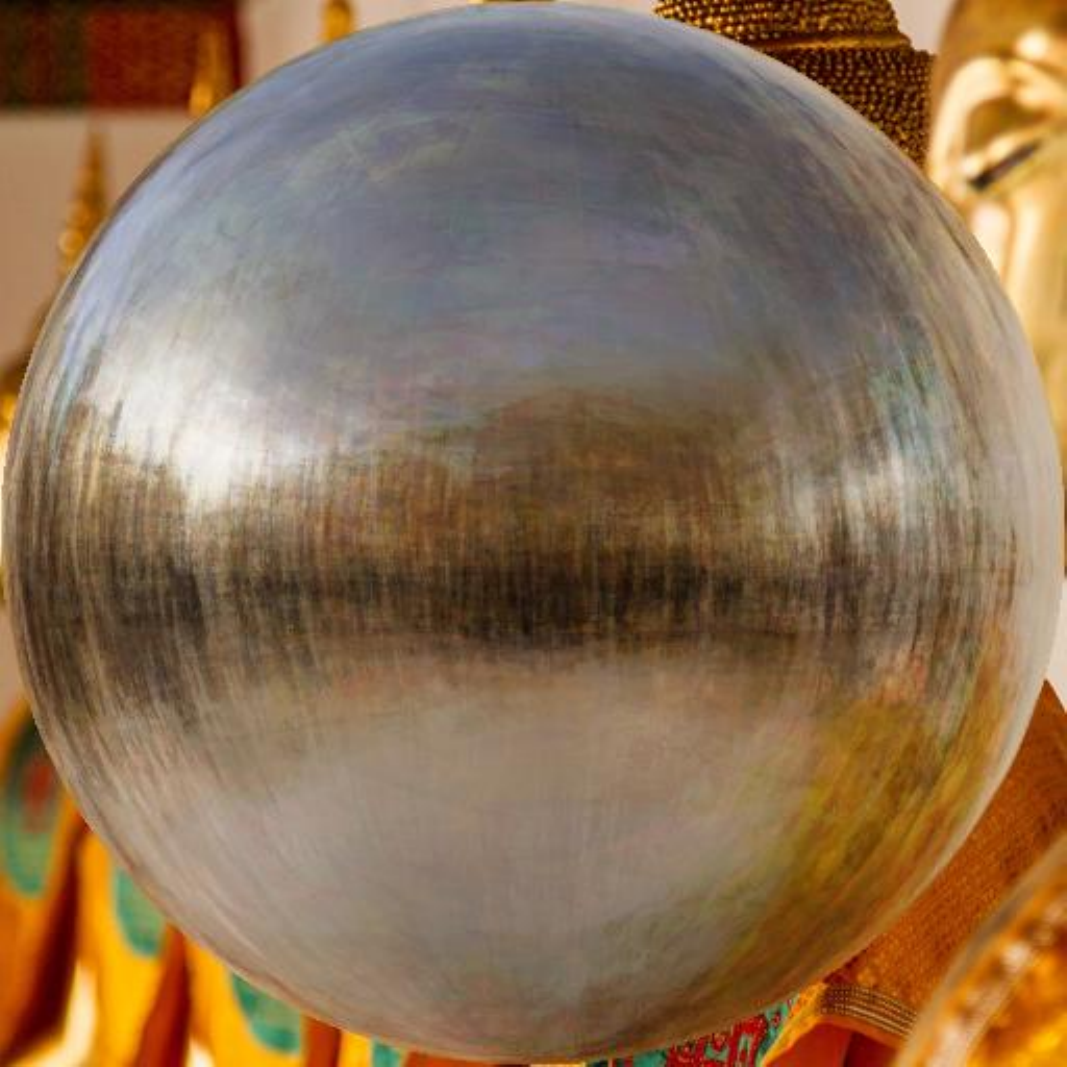}} & 
        \\

        \multicolumn{1}{l}{\shortstack[l]{\tiny Last \\ \tiny iteration}} &
        \noindent\parbox[c]{0.050\textwidth}{\includegraphics[width=0.050\textwidth]{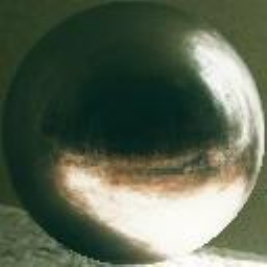}} & 
        \noindent\parbox[c]{0.050\textwidth}{\includegraphics[width=0.050\textwidth]{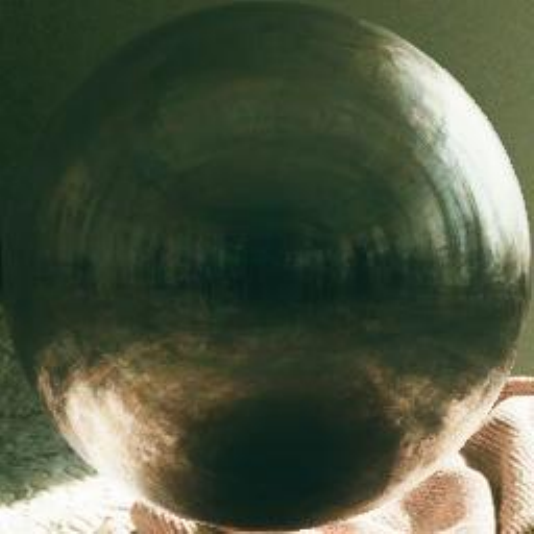}} & 
        \noindent\parbox[c]{0.050\textwidth}{\includegraphics[width=0.050\textwidth]{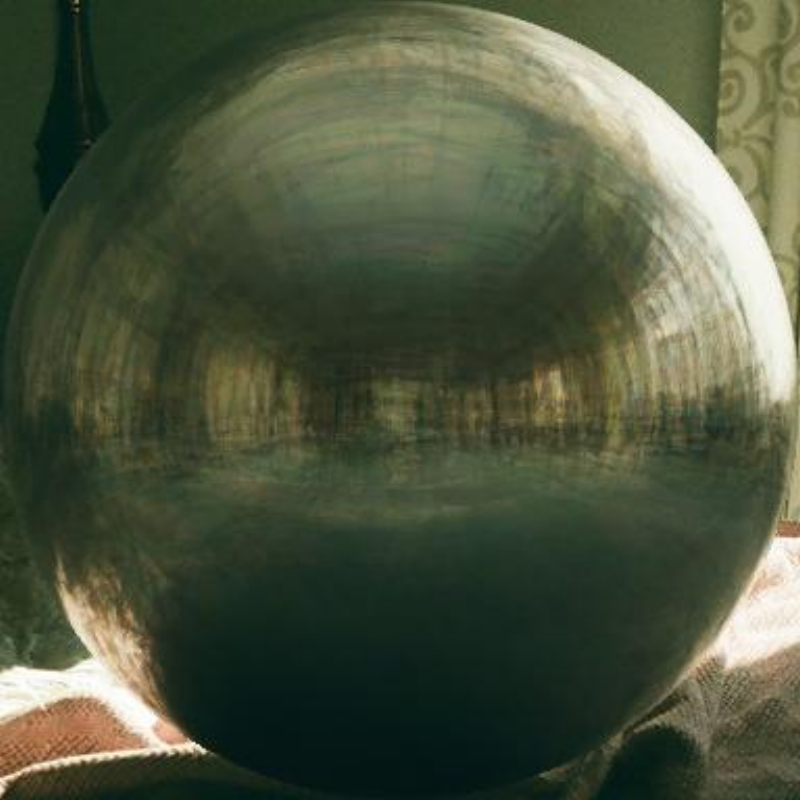}} & 
        \noindent\parbox[c]{0.050\textwidth}{\includegraphics[width=0.050\textwidth]{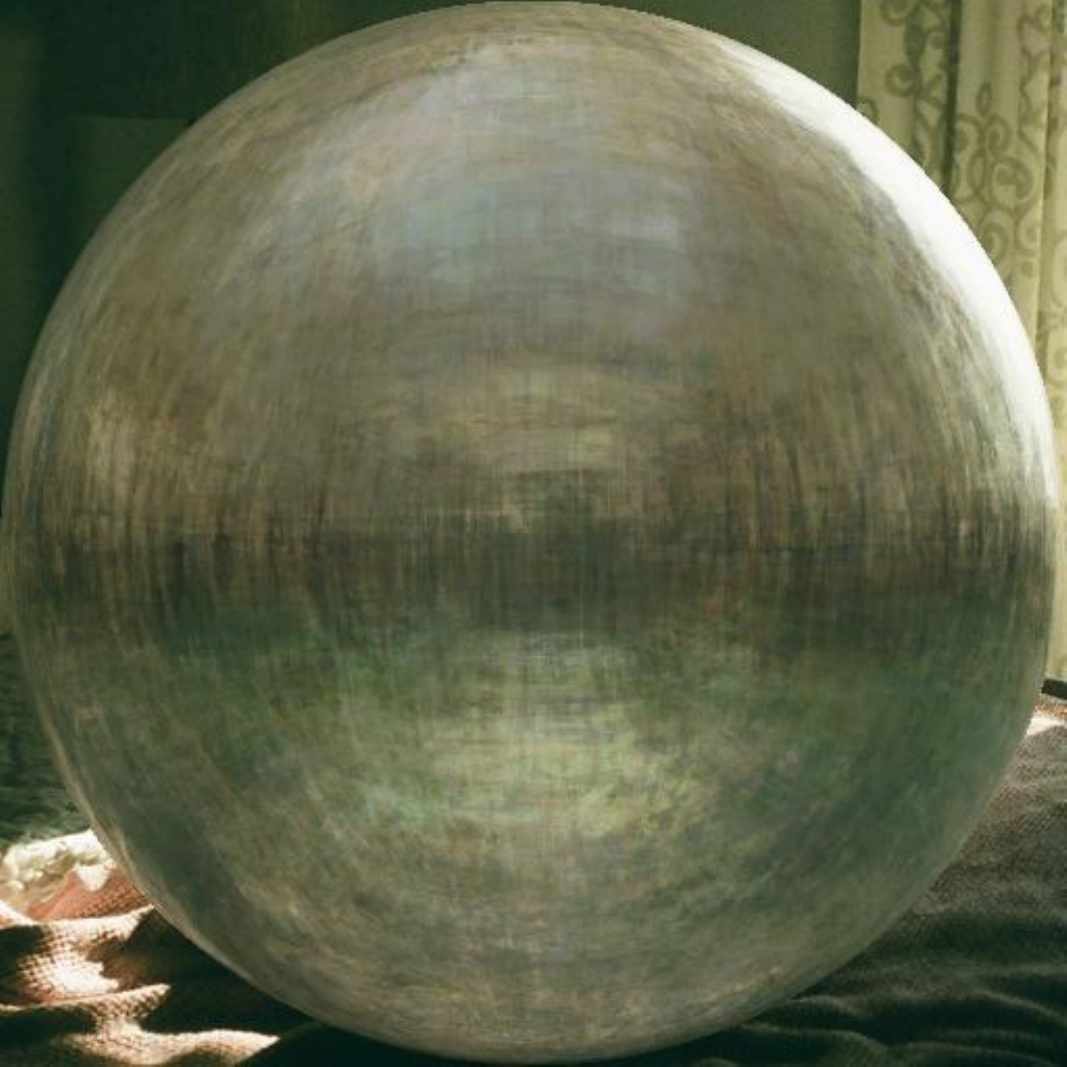}} & 

        \noindent\parbox[c]{0.050\textwidth}{\includegraphics[width=0.050\textwidth]{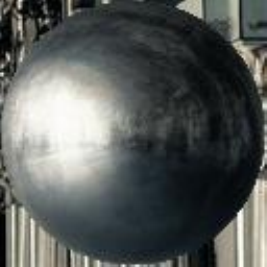}} & 
        \noindent\parbox[c]{0.050\textwidth}{\includegraphics[width=0.050\textwidth]{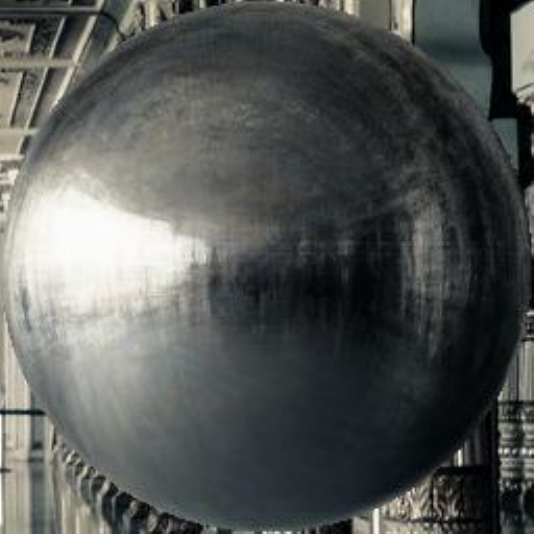}} & 
        \noindent\parbox[c]{0.050\textwidth}{\includegraphics[width=0.050\textwidth]{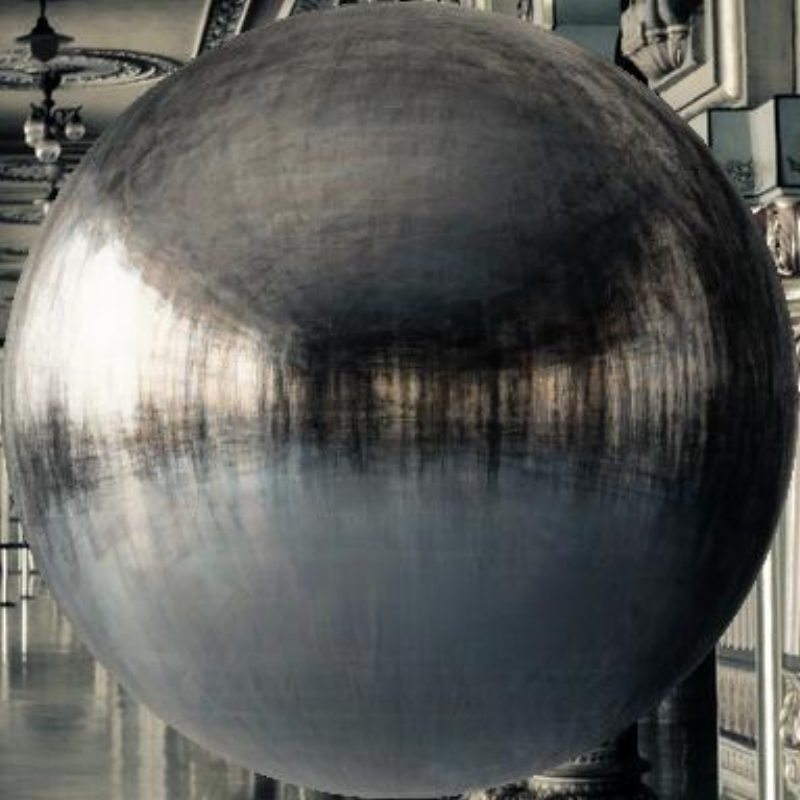}} & 
        \noindent\parbox[c]{0.050\textwidth}{\includegraphics[width=0.050\textwidth]{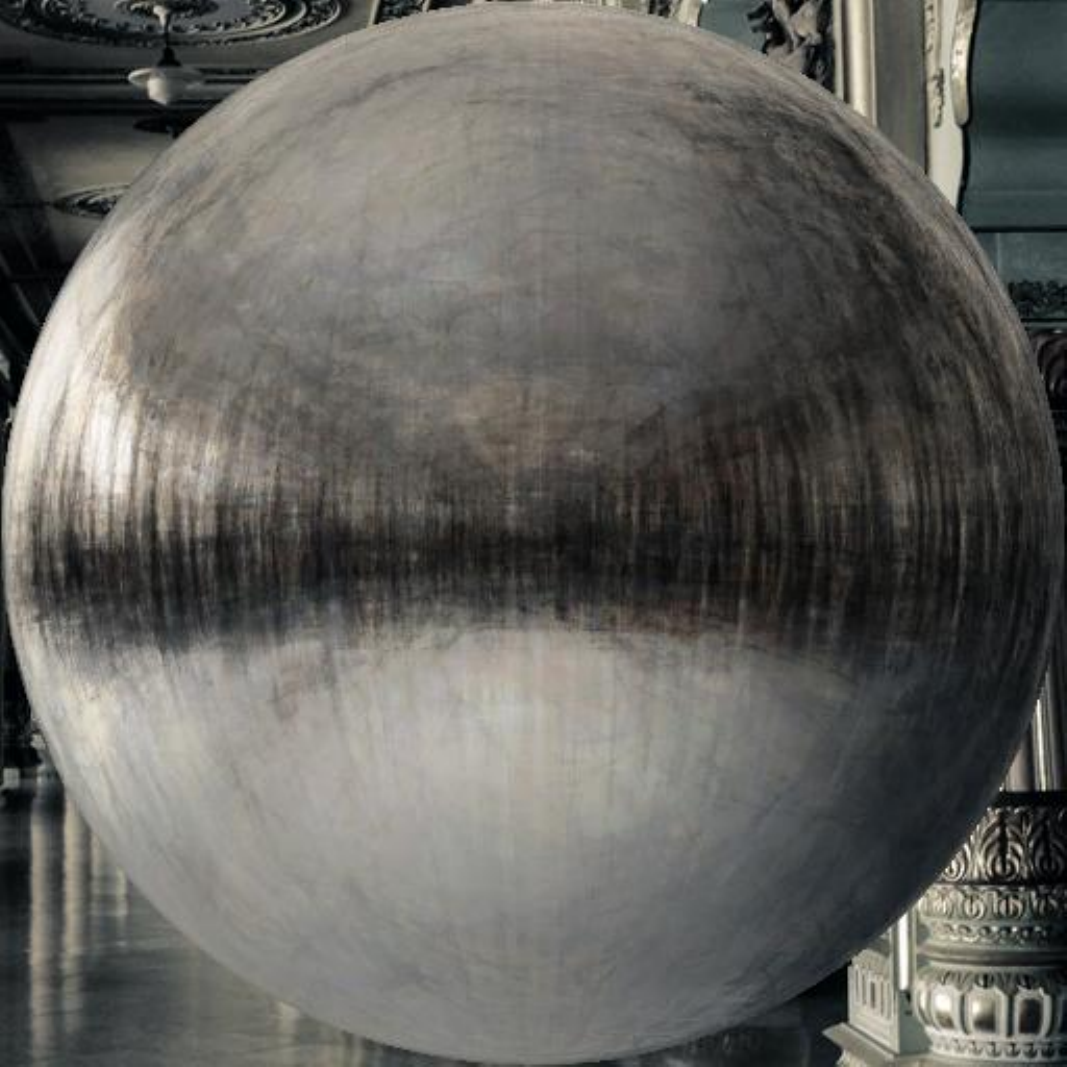}} & 
        \\
        
    \end{tabu}
    \vspace{-8pt}
    \caption{Results when varying the ball size (without LoRA).}
    \label{fig:ball_size}
\end{figure}

We investigated the effects of ball diameter (i.e., white circle) on the depth maps used by ControlNet. Specifically, we analyze and compare the efficacy of various ball sizes, ranging from 128 to 512 pixels in diameter, as illustrated in Figure \ref{fig:ball_size}. The result shows that increasing the ball size from 256 to 384 or 512 still results in realistic balls, but they do not reflect the environment as convincingly. This is likely because the original input content seen by the model is reduced. On the other hand, smaller balls can capture the lighting well but are less detailed and not as useful.

\begin{table}[]
\centering
\small
\begin{tabular}{
    c@{\hspace{5pt}}
    c@{\hspace{5pt}}
    c@{\hspace{5pt}}
    c@{\hspace{5pt}}
    c@{\hspace{5pt}}
    c
}
\toprule
\textbf{\begin{tabular}[c]{@{}c@{}}\textbf{\#Iterations}\\ \textbf{($k$)}\end{tabular}} & \textbf{\begin{tabular}[c]{@{}c@{}}\textbf{\#Balls}\\ \textbf{($N$)}\end{tabular}} & \textbf{\begin{tabular}[c]{@{}c@{}}\textbf{Time}\\ \textbf{(mins)}\end{tabular}} & \textbf{si-RMSE} $\downarrow$ & \begin{tabular}[c]{@{}c@{}}\textbf{Angular}\\ \textbf{Error}\end{tabular} $\downarrow$ & \begin{tabular}[c]{@{}c@{}}\textbf{Normalized}\\ \textbf{RMSE}\end{tabular} $\downarrow$ \\
\midrule
1 & 5  & 2.5 & 0.631 & 5.367 & 0.416\\
  & 15 & 7.5 & 0.615 & 5.229 & 0.410\\
  & \textcolor{teal}{30} & \textcolor{teal}{15}  & \textcolor{teal}{0.609} & \textcolor{teal}{5.210} & \textcolor{teal}{0.405}\\
\hline
2 & 5  & 5  & 0.634 & 5.393 & 0.417\\
  & 15 & 15 & 0.615 & 5.263 & 0.409\\
  & \textcolor{Orchid}{30} & \textcolor{Orchid}{30} & \textcolor{Orchid}{0.607} & \textcolor{Orchid}{5.248} & \textcolor{Orchid}{0.403}\\
\bottomrule
\end{tabular}
\caption{Ablation study on the number of iterations $k$ and balls $N$. We report running time and quality metrics on 200 mirror balls in the validation set. See Figure \ref{fig:ablation_inpainting_ball_iteration} for qualitative results.}
 \label{tab:plot_time_accuracy}
\end{table}

\tabulinesep=2pt

\tabulinesep=2pt

\tabulinesep=0.5pt


\tabulinesep=0.1pt

\section{Ablation on LoRA Training} \label{appendix:aba_lora}

\subsection{Additional details on training set generation}

To generate training panoramas from text prompts as mentioned in Section \ref{sec:cont_lora}, we use Text2Light \cite{chen2022text2light}, using prompts from its official GitHub repository\footnote{\url{https://github.com/FrozenBurning/Text2Light}} and additional prompts generated by Chat-GPT 3.5\footnote{\url{https://chat.openai.com/}} from short instructions and examples.
To eliminate near-duplicate samples, we use the perceptual hashing algorithm implemented in the \texttt{ imagededup} package \footnote{\url{https://github.com/idealo/imagededup}}. This process yielded a dataset of 1,412 unique HDR panoramas at resolution of $2048 \times 4196$ pixels. We used orthographic projection and a 60$^\circ$ field of view.

\subsection{Range of timesteps for LoRA training}

\begin{figure}[!t]
    \centering

        \begin{tabu} to \textwidth {
        @{}
        c@{\hspace{1pt}}
        c@{\hspace{1pt}}
        c@{\hspace{1pt}}
        c@{\hspace{1pt}}
        c@{\hspace{1pt}}
        c@{\hspace{1pt}}
        c@{\hspace{1pt}}
    }

        \multicolumn{1}{c}{\shortstack{\scriptsize Input}} & 
        \multicolumn{1}{c}{\shortstack{\scriptsize 6\textsuperscript{th} step}} &
        \multicolumn{1}{c}{\shortstack{\scriptsize 12\textsuperscript{th} step}} & 
        \multicolumn{1}{c}{\shortstack{\scriptsize 18\textsuperscript{th} step}} & 
        \multicolumn{1}{c}{\shortstack{\scriptsize 24\textsuperscript{th} step}} &
        \multicolumn{1}{c}{\shortstack{\scriptsize 30\textsuperscript{th} step}} &
        \\

        \noindent\parbox[c]{0.107\textwidth}{\includegraphics[width=0.107\textwidth]{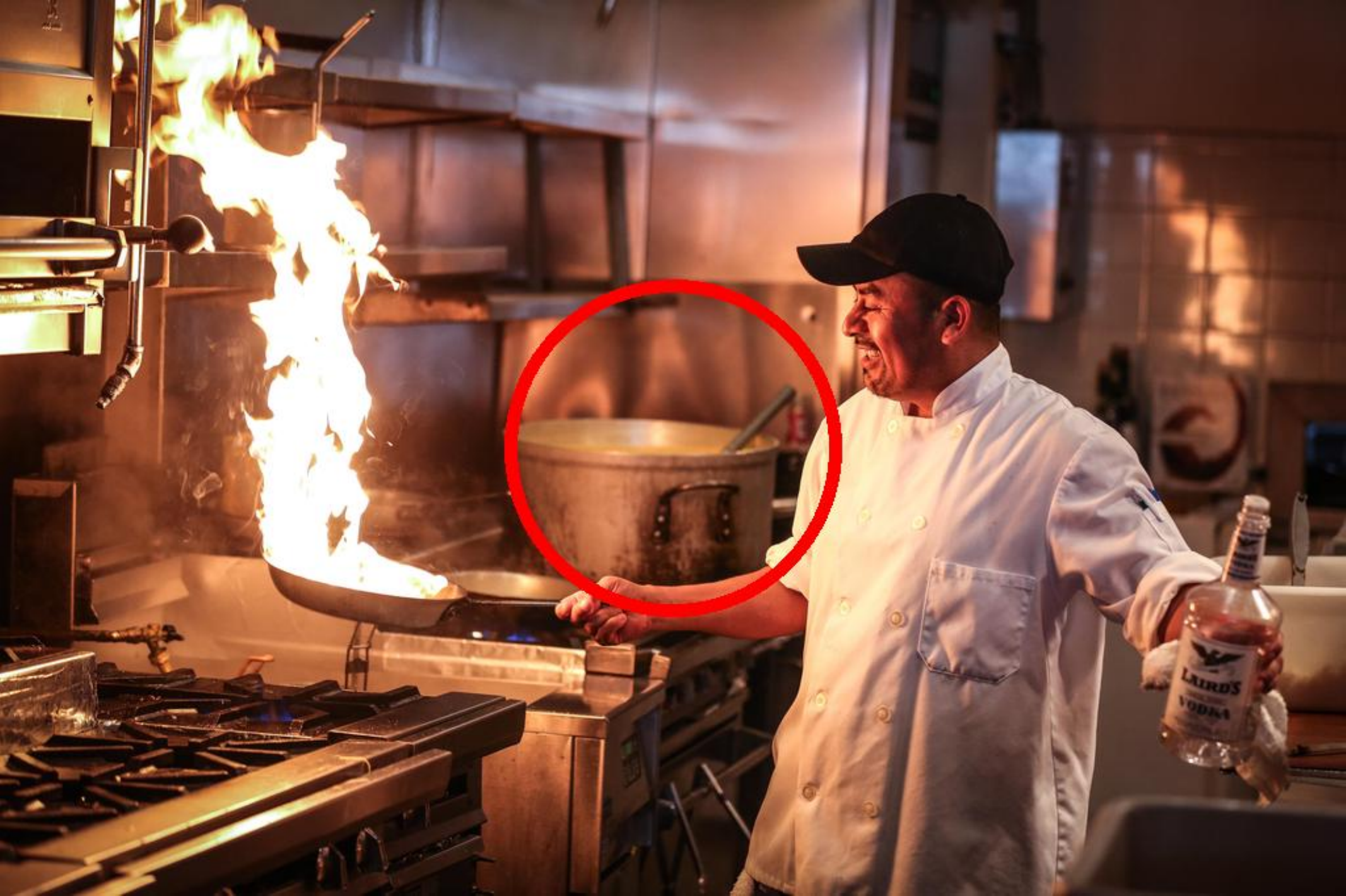}} & 
        \noindent\parbox[c]{0.071\textwidth}{\includegraphics[width=0.071\textwidth]{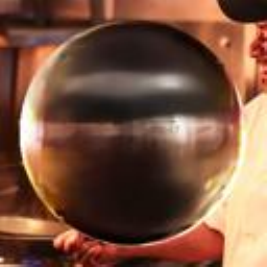}} &  
        \noindent\parbox[c]{0.071\textwidth}{\includegraphics[width=0.071\textwidth]{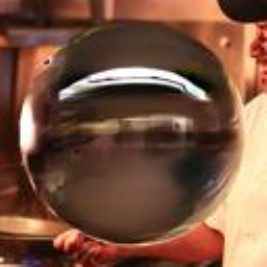}} &
        \noindent\parbox[c]{0.071\textwidth}{\includegraphics[width=0.071\textwidth]{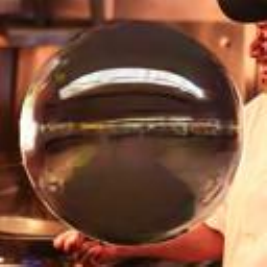}} &
        \noindent\parbox[c]{0.071\textwidth}{\includegraphics[width=0.071\textwidth]{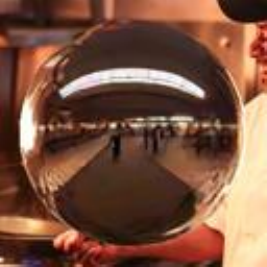}} &
        \noindent\parbox[c]{0.071\textwidth}{\includegraphics[width=0.071\textwidth]{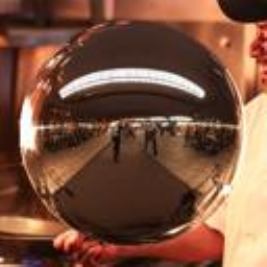}} &
        \\

        & 
        \noindent\parbox[c]{0.071\textwidth}{\includegraphics[width=0.071\textwidth]{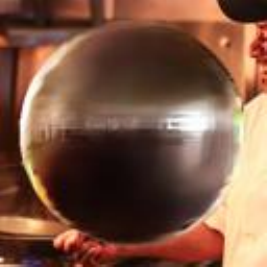}} &  
        \noindent\parbox[c]{0.071\textwidth}{\includegraphics[width=0.071\textwidth]{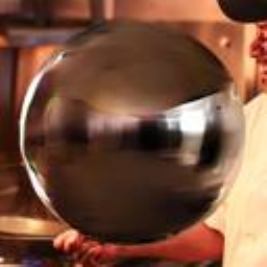}} &
        \noindent\parbox[c]{0.071\textwidth}{\includegraphics[width=0.071\textwidth]{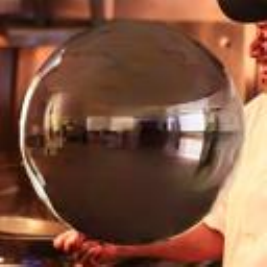}} &
        \noindent\parbox[c]{0.071\textwidth}{\includegraphics[width=0.071\textwidth]{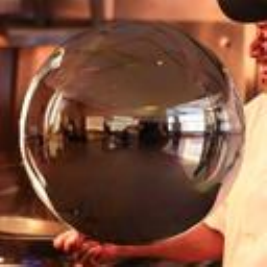}} &
        \noindent\parbox[c]{0.071\textwidth}{\includegraphics[width=0.071\textwidth]{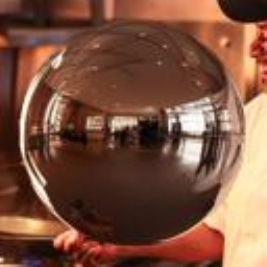}} &
        \\

        & 
        \noindent\parbox[c]{0.071\textwidth}{\includegraphics[width=0.071\textwidth]{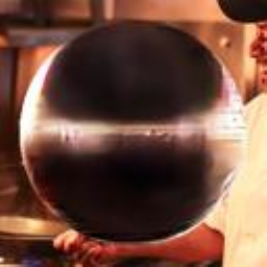}} &  
        \noindent\parbox[c]{0.071\textwidth}{\includegraphics[width=0.071\textwidth]{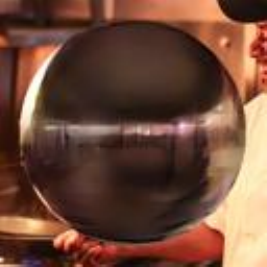}} &
        \noindent\parbox[c]{0.071\textwidth}{\includegraphics[width=0.071\textwidth]{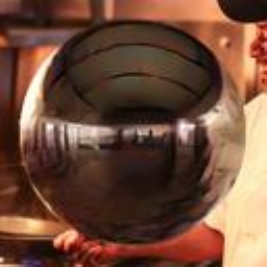}} &
        \noindent\parbox[c]{0.071\textwidth}{\includegraphics[width=0.071\textwidth]{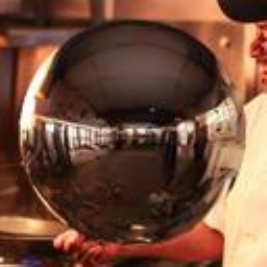}} &
        \noindent\parbox[c]{0.071\textwidth}{\includegraphics[width=0.071\textwidth]{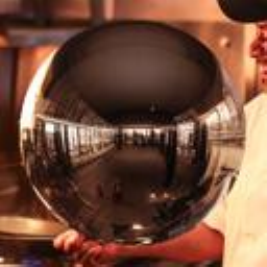}} &
        \\

        \end{tabu}
    \caption{The overall lighting is determined at early sampling steps. Here, we visualize intermediate predictions at various steps during 30-step sampling with UniPC \cite{zhao2023unipc}. These intermediate predictions, or the predicted $\vect{z}_0$, can be computed from $\vect{z}_t$ at any timestep $t$ using Equation \ref{eq:add_noise}. 
    Each row corresponds to a different random seed.
    }
    
    \label{fig:aba_lightinfo_earlystop}
\end{figure}

For LoRA training, we sampled from timesteps 900-999 as we observed that the overall lighting information is determined earlier in the sampling process (see Figure \ref{fig:aba_lightinfo_earlystop}). This choice helped speed up training. 
In Table \ref{tab:ablation_lora_t}, we compare this choice to training from 0-999 and 500-999 given the same number of training iterations and report scores on the same validation set of 200 scenes as in Appendix \ref{appendix:compute_tradeoff}. Our choice of 900-999 yielded the best performance across all three metrics.

\tabulinesep=0.5pt




\begin{table}[]
\centering
\small
\begin{tabular}{
    l@{\hspace{5pt}}
    l@{\hspace{5pt}}
    c@{\hspace{5pt}}
    c@{\hspace{5pt}}
    c
}
\toprule
\textbf{Sphere} & \textbf{Denoising step} & \textbf{si-RMSE} $\downarrow$ & \begin{tabular}[c]{@{}c@{}}\textbf{Angular}\\ \textbf{Error}\end{tabular} $\downarrow$ & \begin{tabular}[c]{@{}c@{}}\textbf{Normalized}\\ \textbf{RMSE}\end{tabular} $\downarrow$ \\
\midrule

Diffuse & $\vect{x}_0:\vect{x}_{999}$ & 0.194 & 3.322 & 0.292\\
        & $\vect{x}_{500}:\vect{x}_{999}$ & 0.188 & 3.260 & 0.284\\
        & $\vect{x}_{900}:\vect{x}_{999}$ & \colorbox{tabfirst}{0.156} & \colorbox{tabfirst}{2.956} & \colorbox{tabfirst}{0.246} \\
\hline
Matte & $\vect{x}_0:\vect{x}_{999}$ & 0.449 & 4.121 & 0.436\\
        & $\vect{x}_{500}:\vect{x}_{999}$ & 0.452 & 4.008 & 0.438\\
        & $\vect{x}_{900}:\vect{x}_{999}$ & \colorbox{tabfirst}{0.385} & \colorbox{tabfirst}{3.575} & \colorbox{tabfirst}{0.371} \\
  \hline
Mirror & $\vect{x}_0:\vect{x}_{999}$ & 0.727 & 6.292 & 0.483\\
        & $\vect{x}_{500}:\vect{x}_{999}$ & 0.730 & 6.277 & 0.479\\
        & $\vect{x}_{900}:\vect{x}_{999}$ & \colorbox{tabfirst}{0.656} & \colorbox{tabfirst}{5.464} & \colorbox{tabfirst}{0.431} \\
\bottomrule
\end{tabular}
\caption{Ablation study on sampled timesteps for LoRA training.
}
\label{tab:ablation_lora_t}
\end{table}

\subsection{LoRA scale}
We conducted an experiment to assess the effect of using different LoRA scales. Here, the LoRA scale refers to the $\alpha$ value in the weight update equation: $\vect{W}^{\prime} = \vect{W} + \alpha \Delta \vect{W}$, where $\vect{W}^{\prime}$ is the new weight for inference, $\vect{W}$ is the original weight of SDXL, $\Delta \vect{W}$ is the weight update from LoRA. (See Section \ref{sec:prelim} for a brief background on LoRA.) In Table \ref{tab:ldr-lorascale}, we report scores computed on EV0 LDR chrome balls evaluated on scenes in Poly Haven, which were never part of Text2Light's training set. We selected the LoRA scale of 0.75, which has the best si-RMSE and angular error scores, for our implementation. 



\begin{table}[!h]
\centering
\small
\begin{tabular}{
    l@{\hspace{20pt}}
    c@{\hspace{5pt}}
    c@{\hspace{5pt}}
    c
}
\toprule
\textbf{LoRA scale} & \textbf{RMSE} $\downarrow$ & \textbf{si-RMSE} $\downarrow$ & \begin{tabular}[c]{@{}c@{}}\textbf{Angular}\\ \textbf{Error}\end{tabular} $\downarrow$\\
\midrule

 0.00 & 0.232 & 0.327 & 6.189\\
 0.25 & 0.220 & 0.307 & 6.287\\
 0.50 & 0.211 & 0.303 & 6.180\\
 \textbf{0.75} & 0.204 & \colorbox{tabfirst}{0.300} & \colorbox{tabfirst}{6.109}\\
 1.00 & \colorbox{tabfirst}{0.199} & 0.303 & 6.267\\

\bottomrule
\end{tabular}
\caption{Ablation study on LoRA scales}
\label{tab:ldr-lorascale}
\vspace{-1.3em}
\end{table}

\subsection{Training a single continuous LoRA v.s. multiple LoRAs for exposure bracketing}

As described in Section \ref{sec:cont_lora}, we train a single \textit{continuous} LoRA for multiple EVs by conditioning it on an interpolated text prompt embedding instead of training multiple LoRAs for individual EVs. This strategy helps preserve the overall scene structure across exposures due to weight sharing. 

\begin{figure*}
    \centering
    \begin{tabu} to \textwidth {
        @{}
        c@{}
        *{10}{c}
        @{}
    }
        
        \multicolumn{1}{l}{\rotatebox[origin=c]{90}{\shortstack[l]{\scriptsize Input}}} &
        \multicolumn{2}{c}{
            \noindent\parbox[c]{0.19\textwidth}{\includegraphics[width=0.19\textwidth]{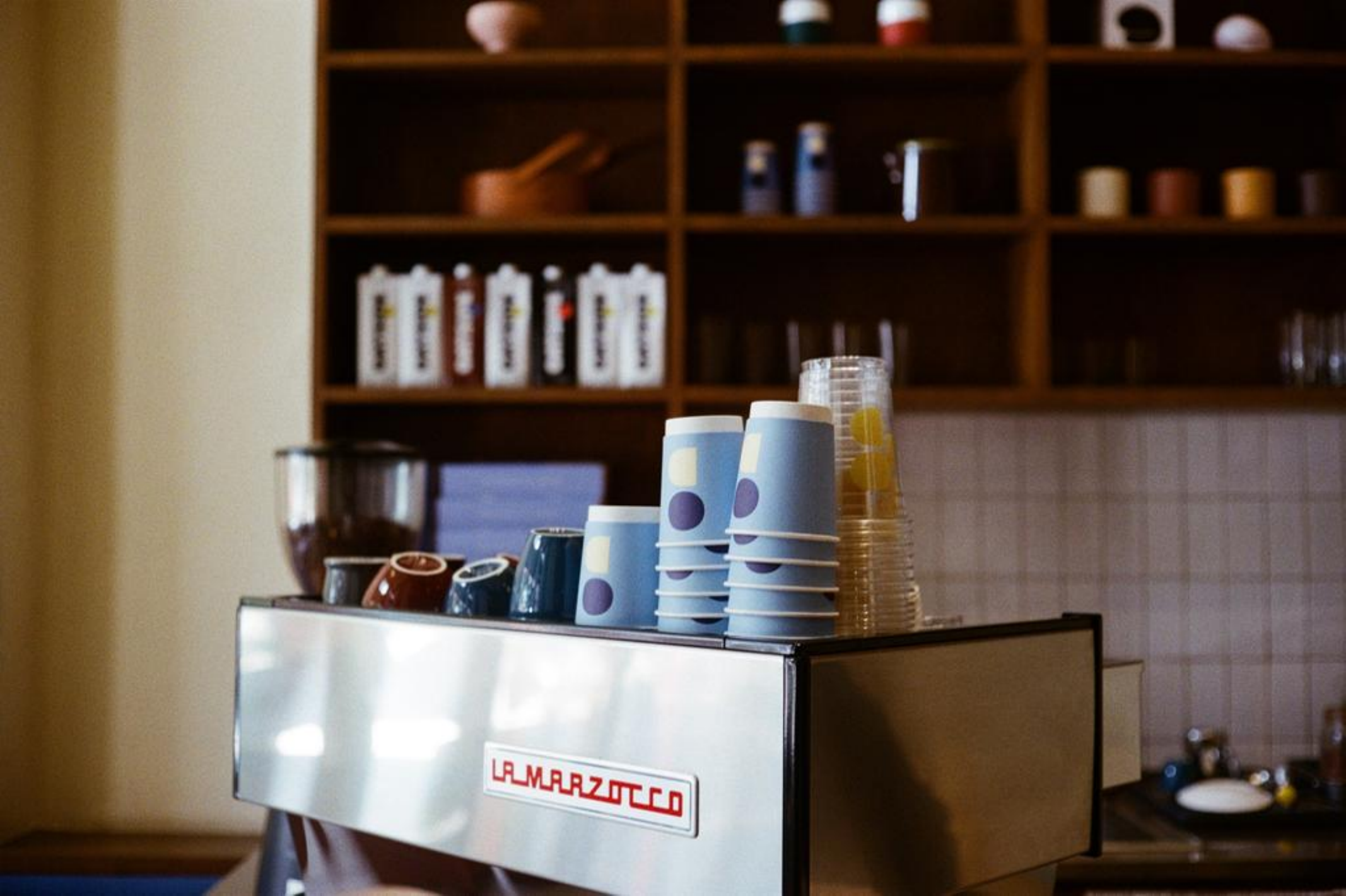}}
        } & 
        \multicolumn{2}{c}{
            \noindent\parbox[c]{0.19\textwidth}{\includegraphics[width=0.19\textwidth]{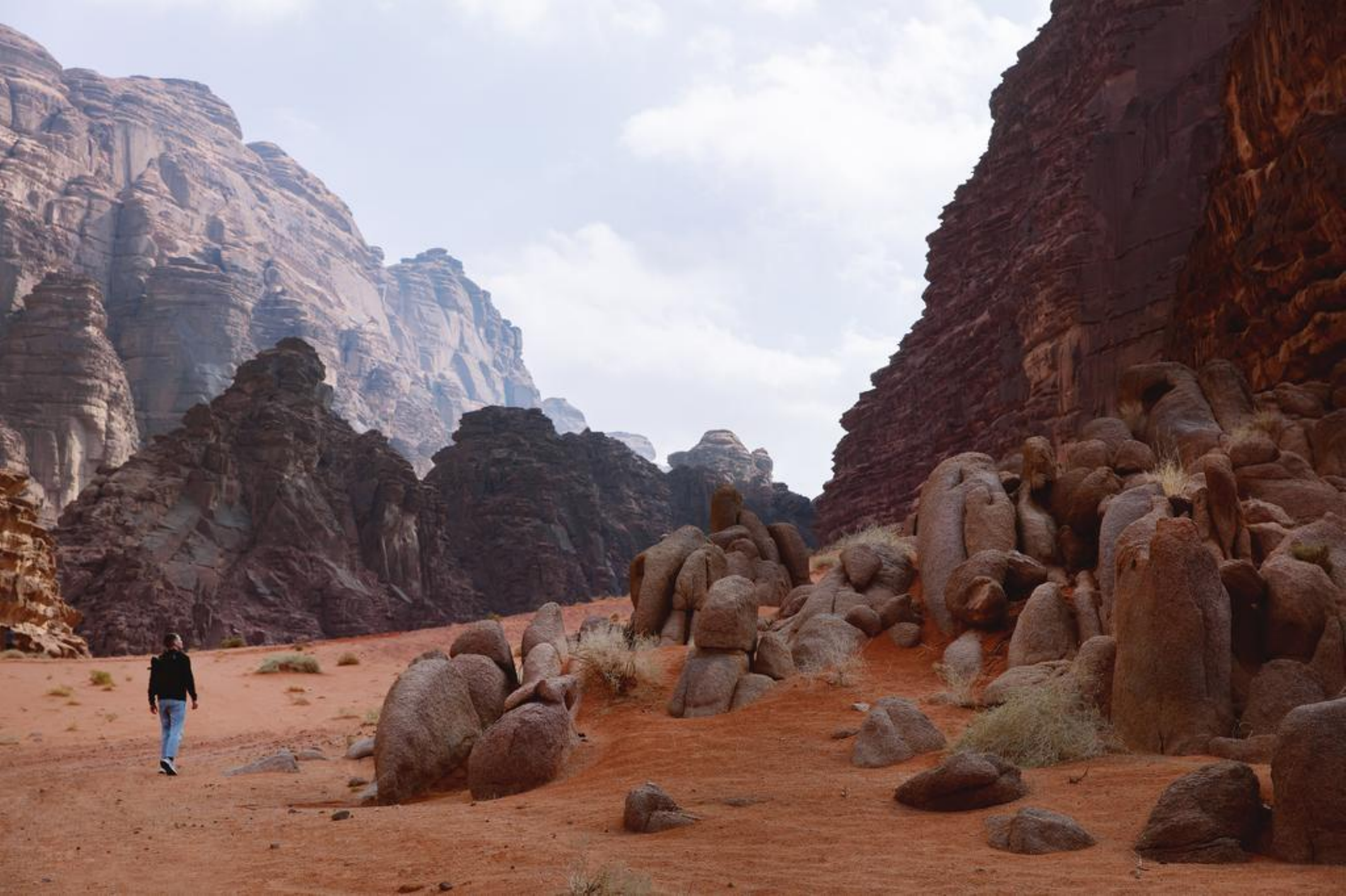}}
        } & 
        \multicolumn{2}{c}{
            \noindent\parbox[c]{0.19\textwidth}{\includegraphics[width=0.19\textwidth]{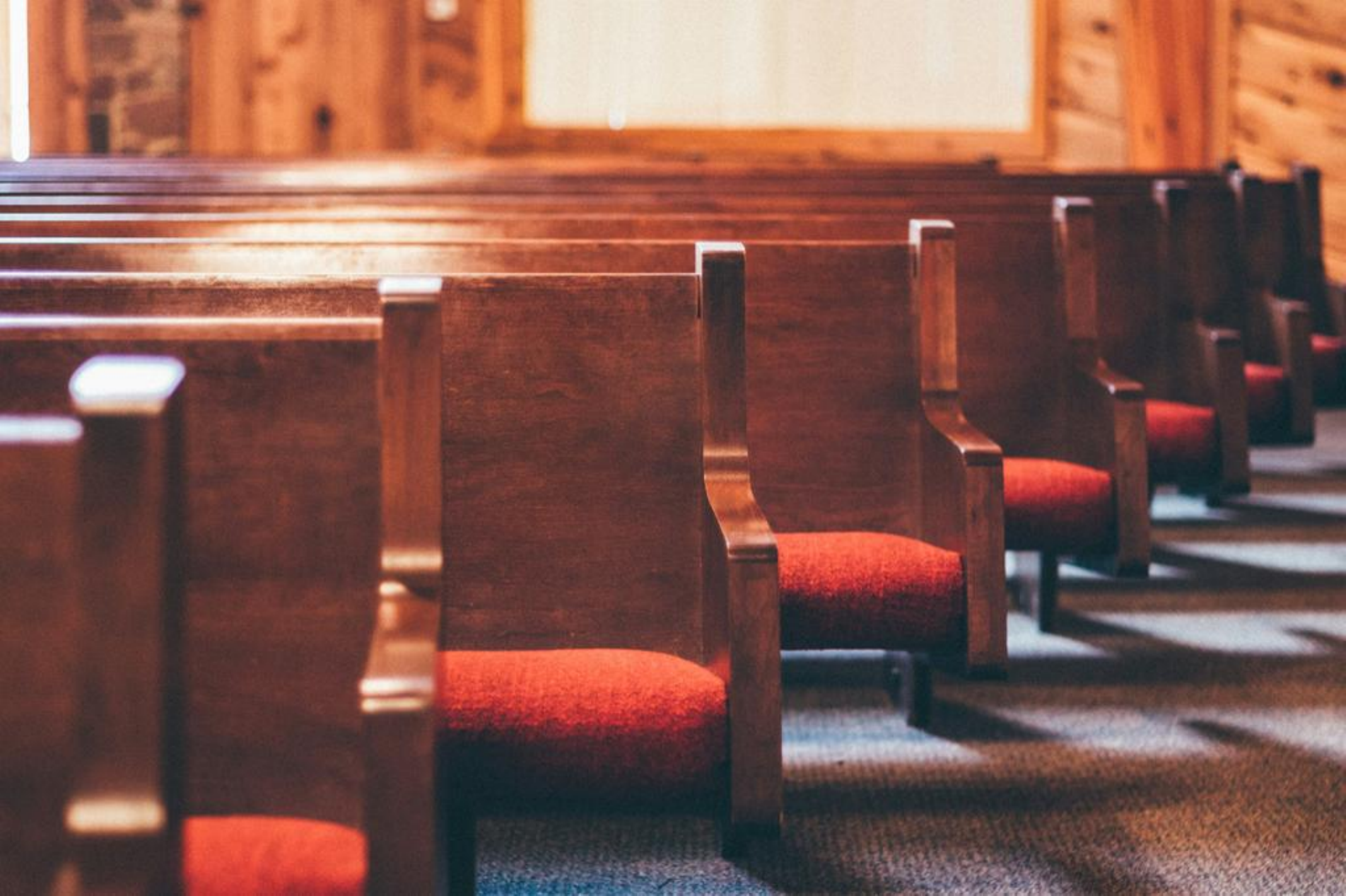}}
        } & 
        \multicolumn{2}{c}{
            \noindent\parbox[c]{0.19\textwidth}{\includegraphics[width=0.19\textwidth]{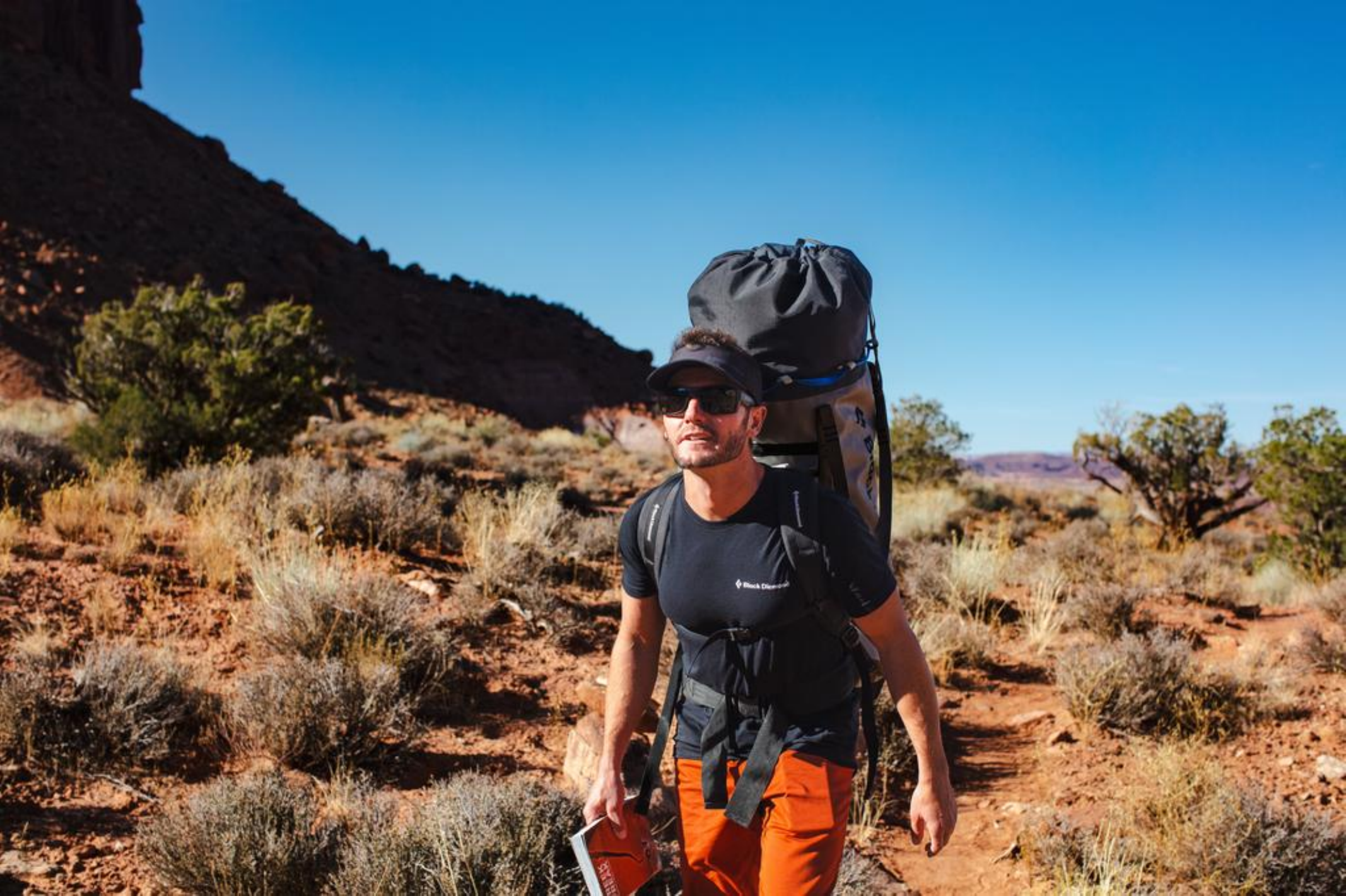}}
        } & 
        \multicolumn{2}{c}{
            \noindent\parbox[c]{0.19\textwidth}{\includegraphics[width=0.19\textwidth]{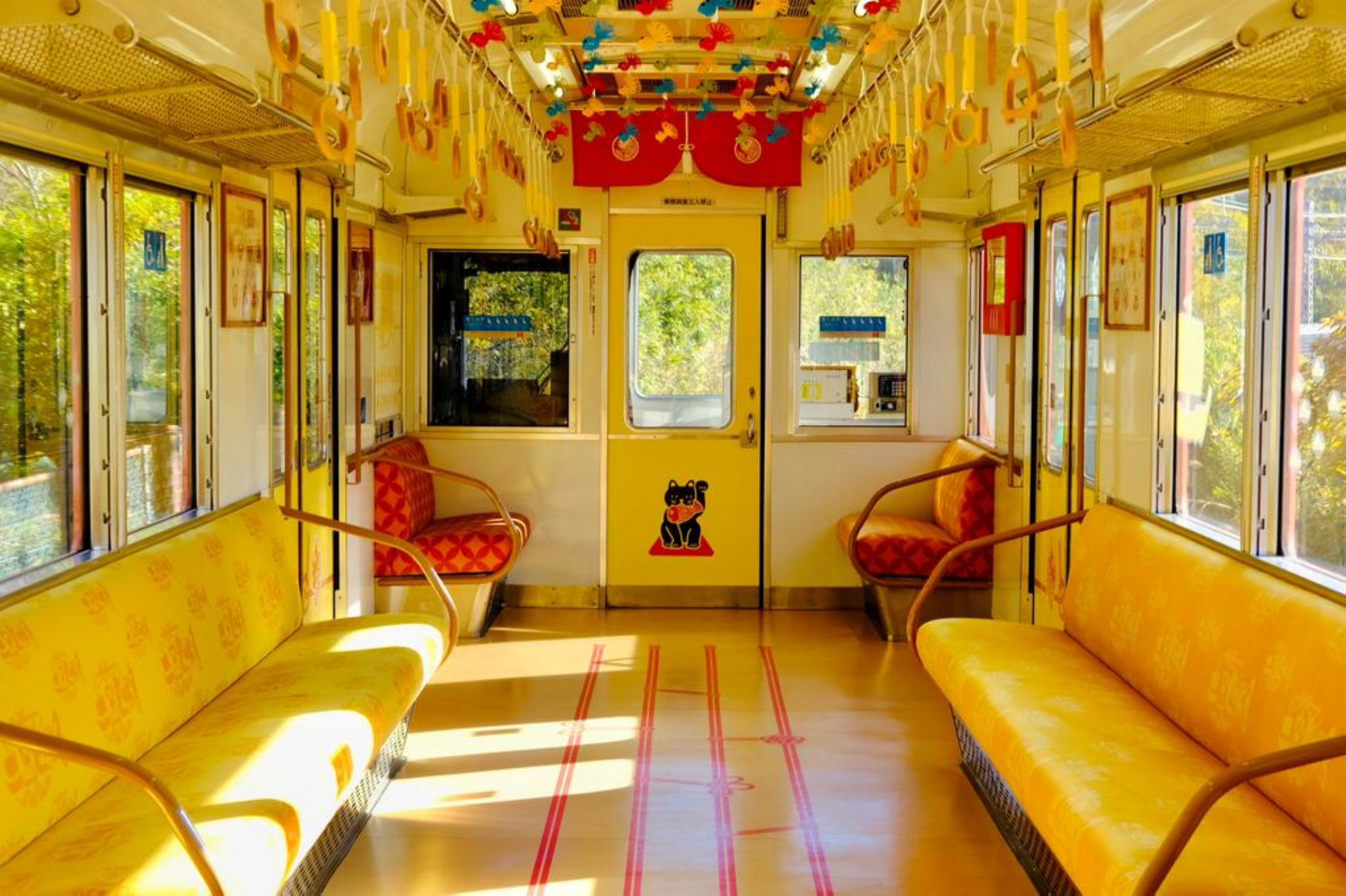}}
        }

        \\

        &
        \multicolumn{1}{c}{\shortstack{\scriptsize 3 LoRAs}} & 
        \multicolumn{1}{c}{\shortstack{\scriptsize Cont. LoRA}} &
        \multicolumn{1}{c}{\shortstack{\scriptsize 3 LoRAs}} & 
        \multicolumn{1}{c}{\shortstack{\scriptsize Cont. LoRA}} &
        \multicolumn{1}{c}{\shortstack{\scriptsize 3 LoRAs}} & 
        \multicolumn{1}{c}{\shortstack{\scriptsize Cont. LoRA}} &
        \multicolumn{1}{c}{\shortstack{\scriptsize 3 LoRAs}} & 
        \multicolumn{1}{c}{\shortstack{\scriptsize Cont. LoRA}} &
        \multicolumn{1}{c}{\shortstack{\scriptsize 3 LoRAs}} & 
        \multicolumn{1}{c}{\shortstack{\scriptsize Cont. LoRA}}
        \\

        \multicolumn{1}{l}{\rotatebox[origin=c]{90}{\shortstack[l]{\scriptsize EV0}}} &
        \noindent\parbox[c]{0.095\textwidth}{\includegraphics[width=0.095\textwidth]{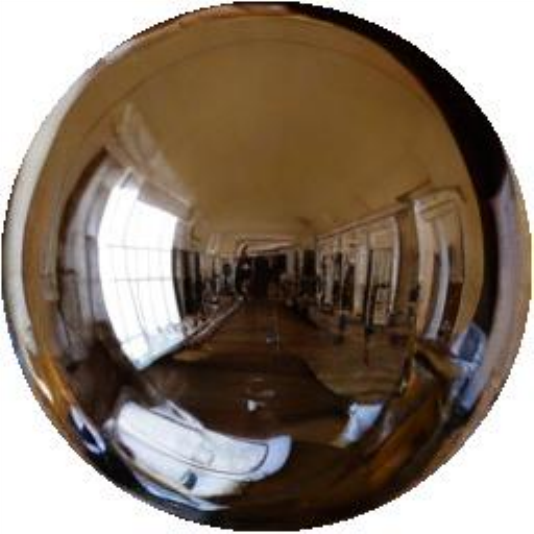}} & 
        \noindent\parbox[c]{0.095\textwidth}{\includegraphics[width=0.095\textwidth]{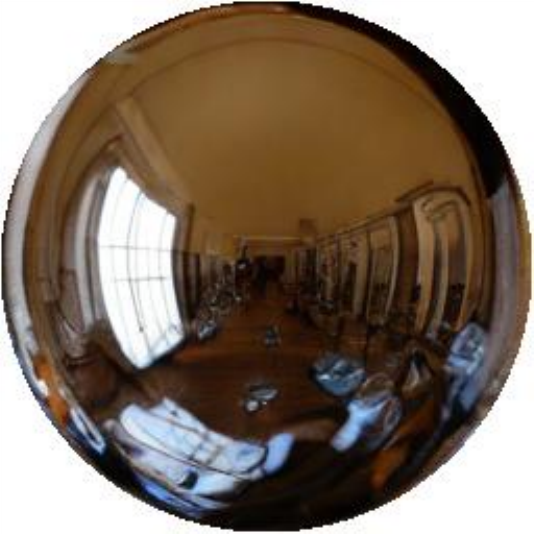}} & 
        \noindent\parbox[c]{0.095\textwidth}{\includegraphics[width=0.095\textwidth]{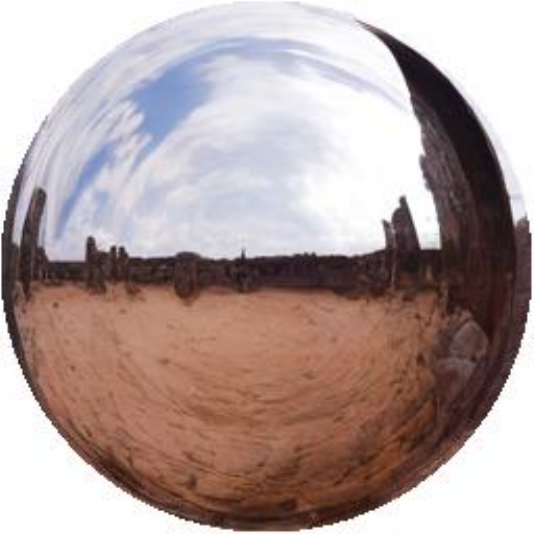}} & 
        \noindent\parbox[c]{0.095\textwidth}{\includegraphics[width=0.095\textwidth]{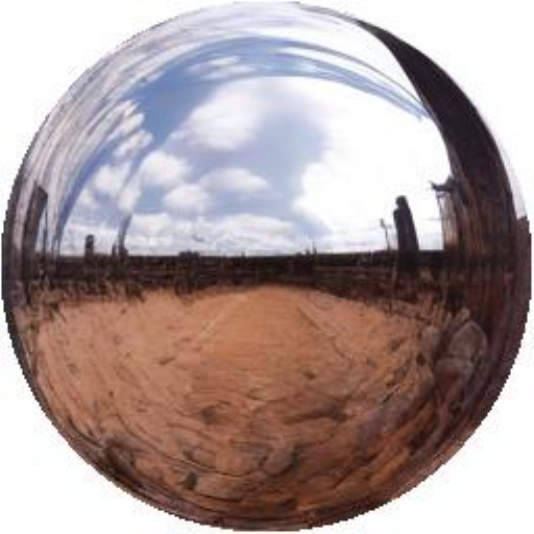}} & 
        \noindent\parbox[c]{0.095\textwidth}{\includegraphics[width=0.095\textwidth]{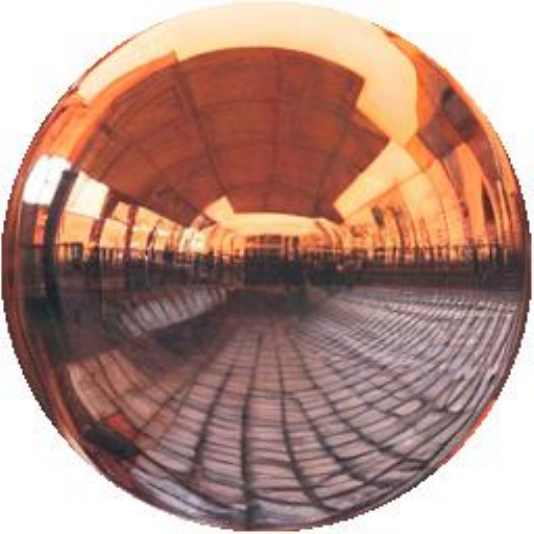}} & 
        \noindent\parbox[c]{0.095\textwidth}{\includegraphics[width=0.095\textwidth]{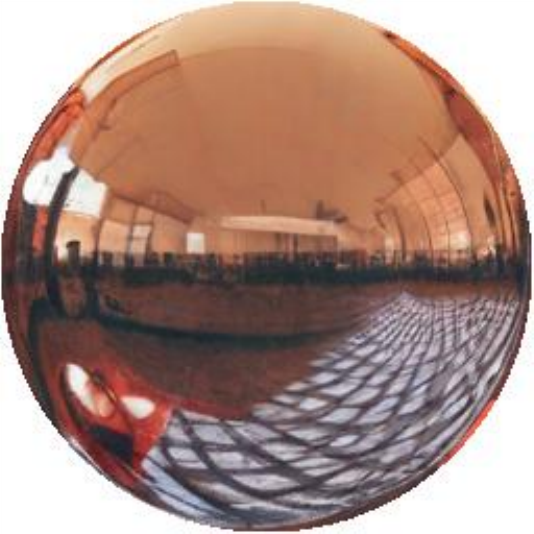}} & 
        \noindent\parbox[c]{0.095\textwidth}{\includegraphics[width=0.095\textwidth]{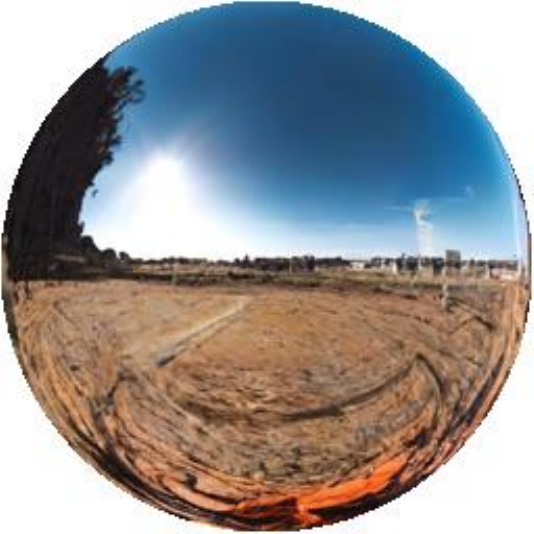}} & 
        \noindent\parbox[c]{0.095\textwidth}{\includegraphics[width=0.095\textwidth]{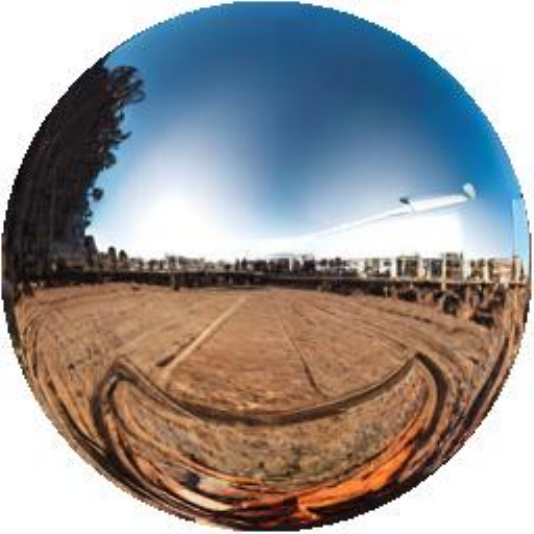}} & 
        \noindent\parbox[c]{0.095\textwidth}{\includegraphics[width=0.095\textwidth]{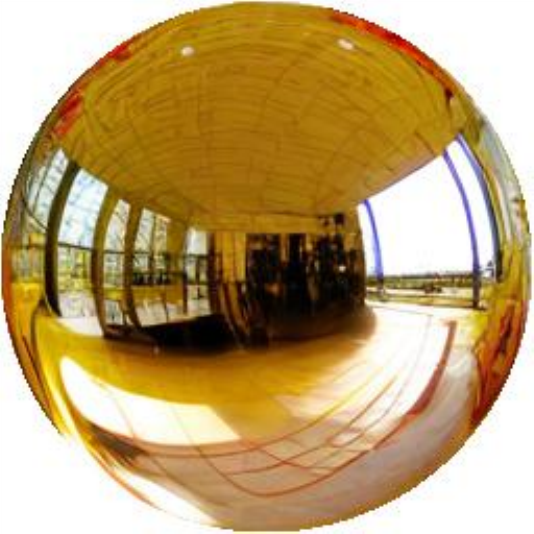}} & 
        \noindent\parbox[c]{0.095\textwidth}{\includegraphics[width=0.095\textwidth]{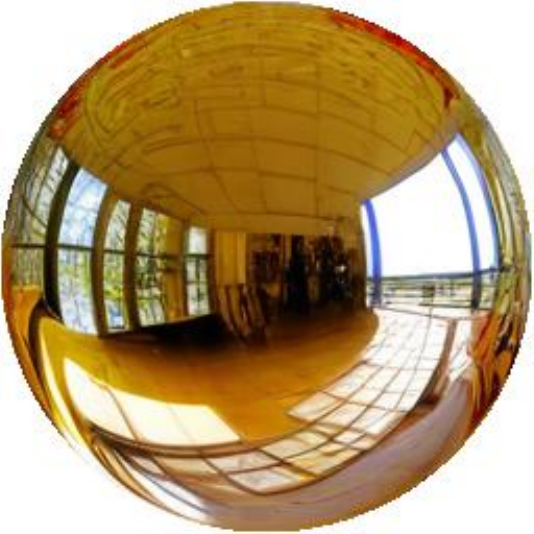}}
        
        \\

        \multicolumn{1}{l}{\rotatebox[origin=c]{90}{\shortstack[l]{\scriptsize EV-2.5}}} &
        \noindent\parbox[c]{0.095\textwidth}{\includegraphics[width=0.095\textwidth]{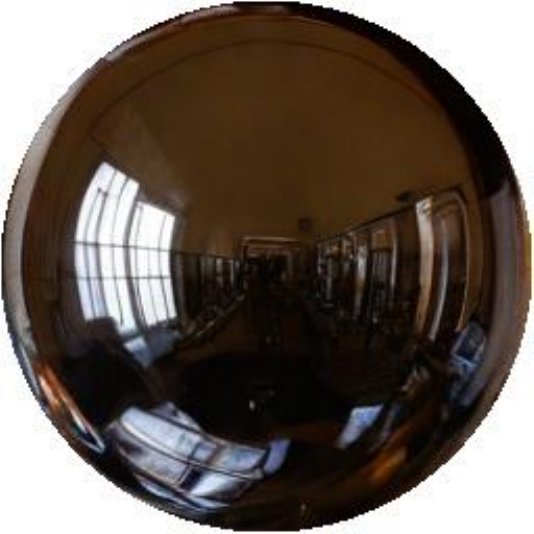}} & 
        \noindent\parbox[c]{0.095\textwidth}{\includegraphics[width=0.095\textwidth]{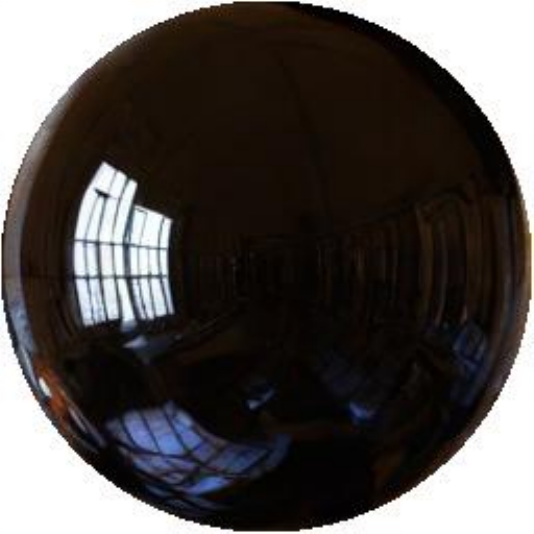}} & 
        \noindent\parbox[c]{0.095\textwidth}{\includegraphics[width=0.095\textwidth]{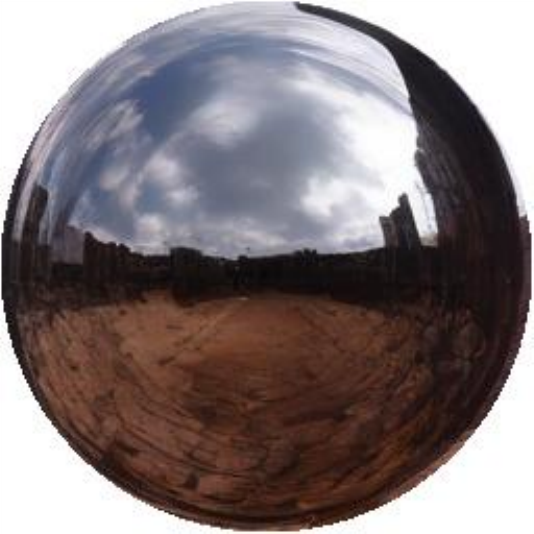}} & 
        \noindent\parbox[c]{0.095\textwidth}{\includegraphics[width=0.095\textwidth]{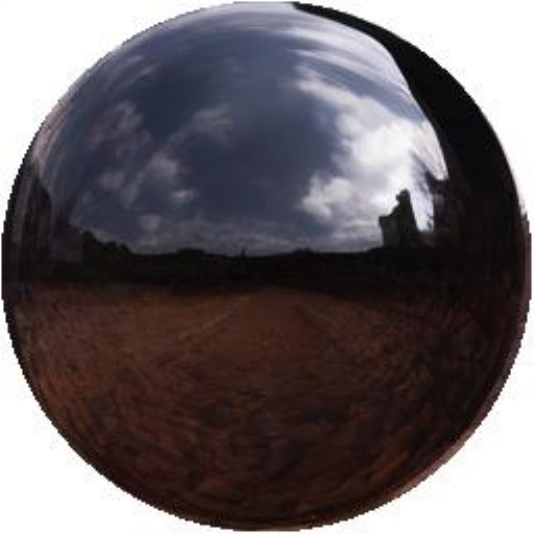}} & 
        \noindent\parbox[c]{0.095\textwidth}{\includegraphics[width=0.095\textwidth]{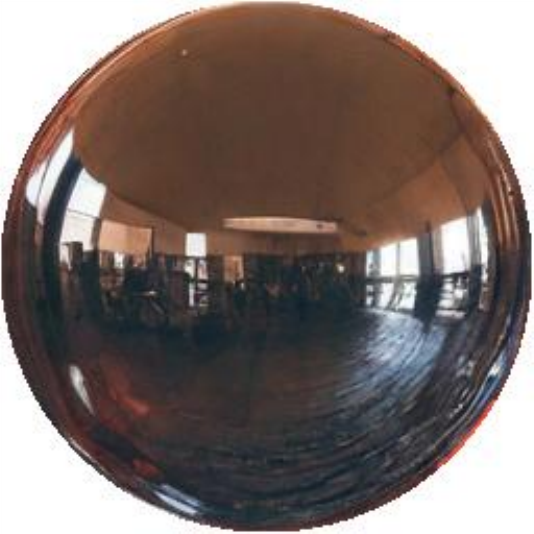}} & 
        \noindent\parbox[c]{0.095\textwidth}{\includegraphics[width=0.095\textwidth]{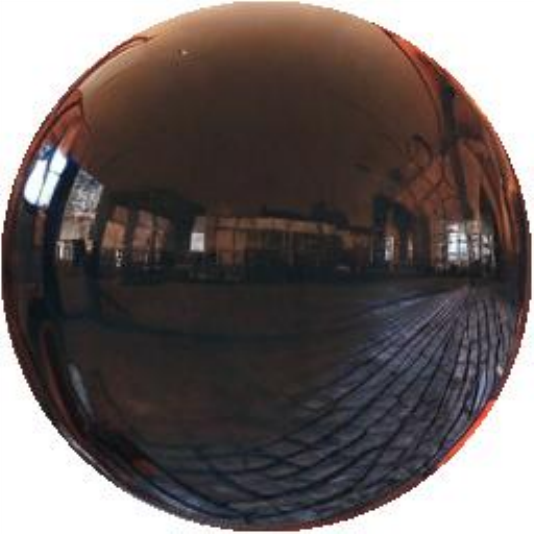}} & 
        \noindent\parbox[c]{0.095\textwidth}{\includegraphics[width=0.095\textwidth]{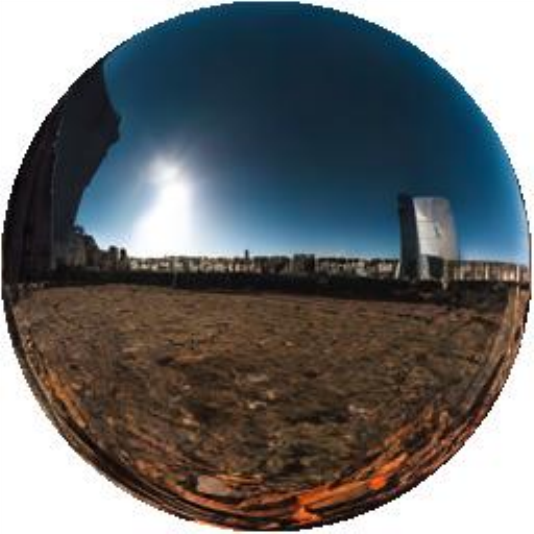}} & 
        \noindent\parbox[c]{0.095\textwidth}{\includegraphics[width=0.095\textwidth]{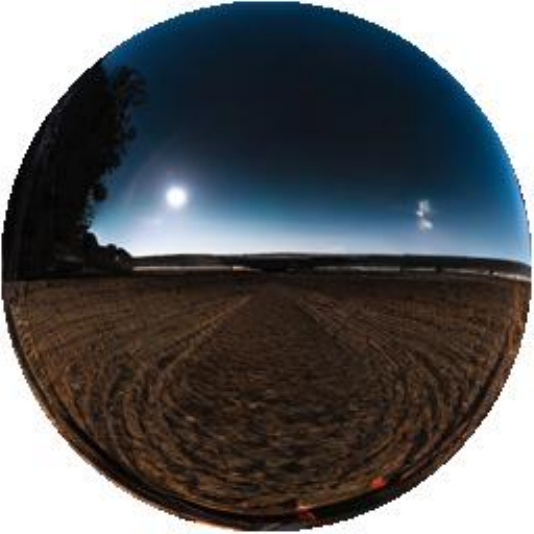}} & 
        \noindent\parbox[c]{0.095\textwidth}{\includegraphics[width=0.095\textwidth]{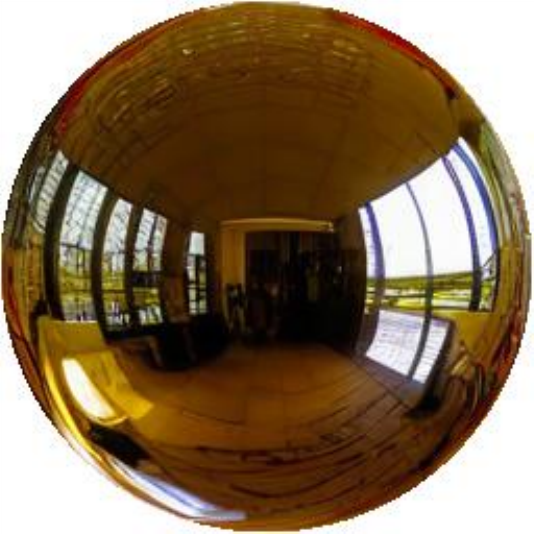}} & 
        \noindent\parbox[c]{0.095\textwidth}{\includegraphics[width=0.095\textwidth]{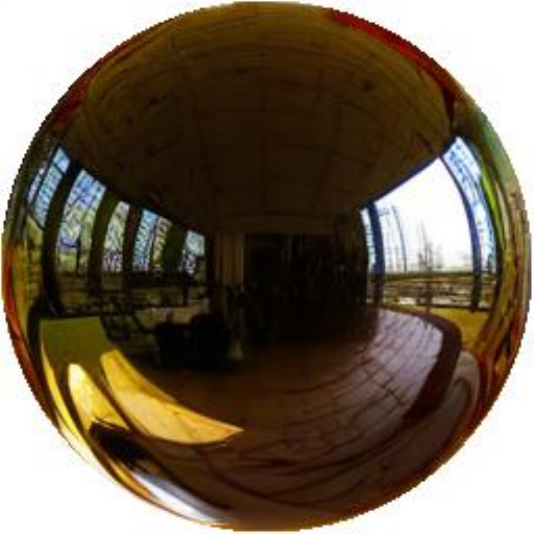}}
        
        \\

        \multicolumn{1}{l}{\rotatebox[origin=c]{90}{\shortstack[l]{\scriptsize EV-5.0}}} &
        \noindent\parbox[c]{0.095\textwidth}{\includegraphics[width=0.095\textwidth]{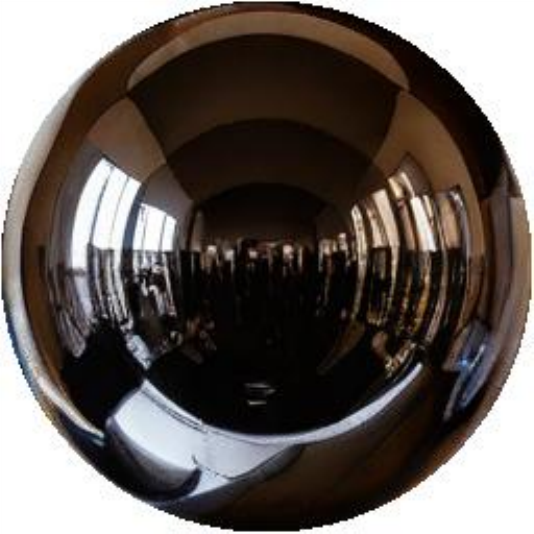}} & 
        \noindent\parbox[c]{0.095\textwidth}{\includegraphics[width=0.095\textwidth]{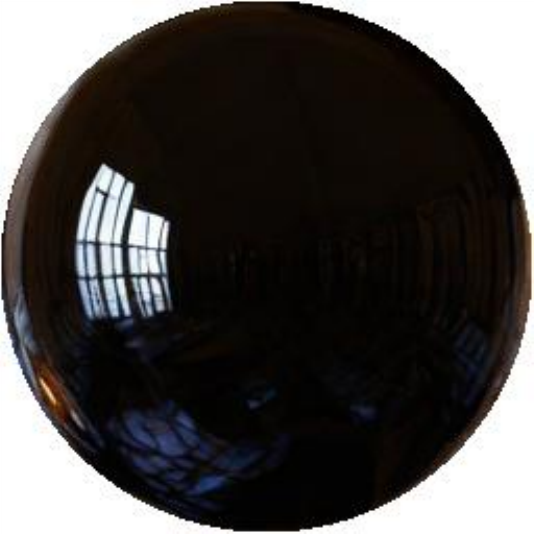}} & 
        \noindent\parbox[c]{0.095\textwidth}{\includegraphics[width=0.095\textwidth]{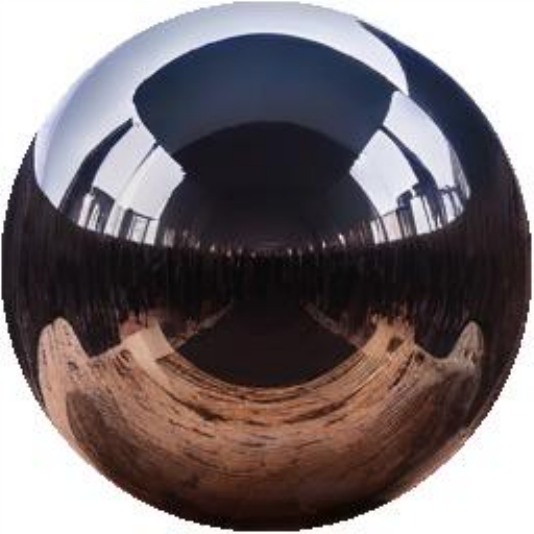}} & 
        \noindent\parbox[c]{0.095\textwidth}{\includegraphics[width=0.095\textwidth]{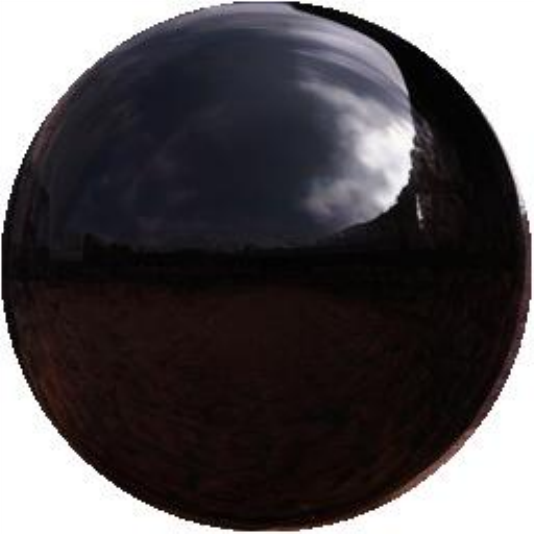}} & 
        \noindent\parbox[c]{0.095\textwidth}{\includegraphics[width=0.095\textwidth]{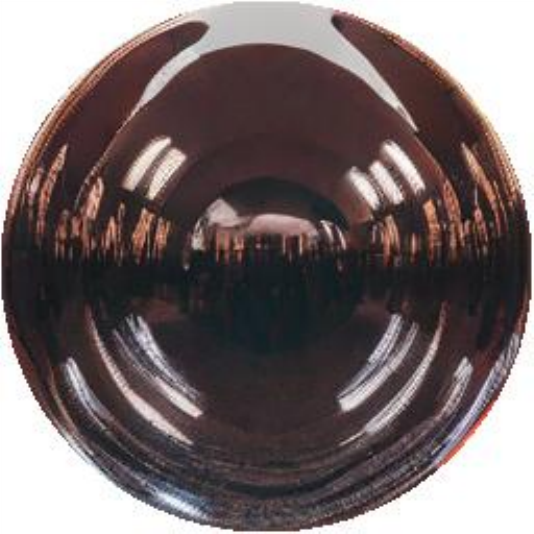}} & 
        \noindent\parbox[c]{0.095\textwidth}{\includegraphics[width=0.095\textwidth]{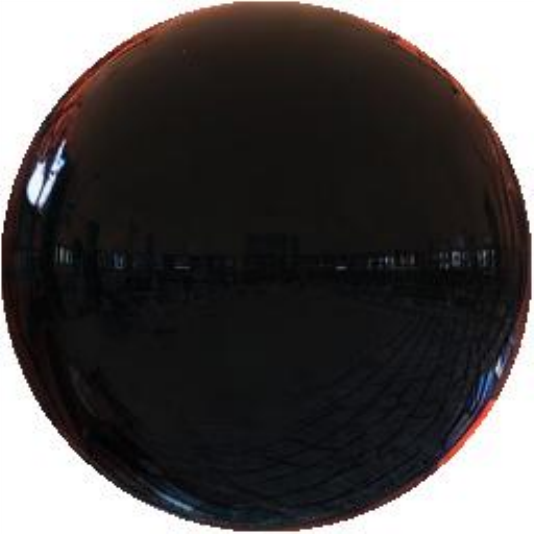}} & 
        \noindent\parbox[c]{0.095\textwidth}{\includegraphics[width=0.095\textwidth]{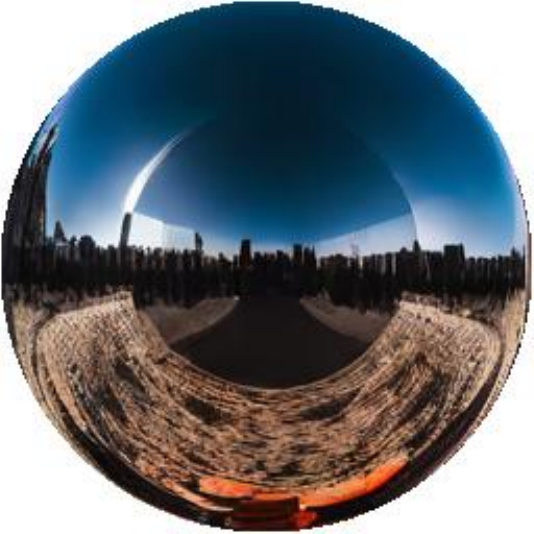}} & 
        \noindent\parbox[c]{0.095\textwidth}{\includegraphics[width=0.095\textwidth]{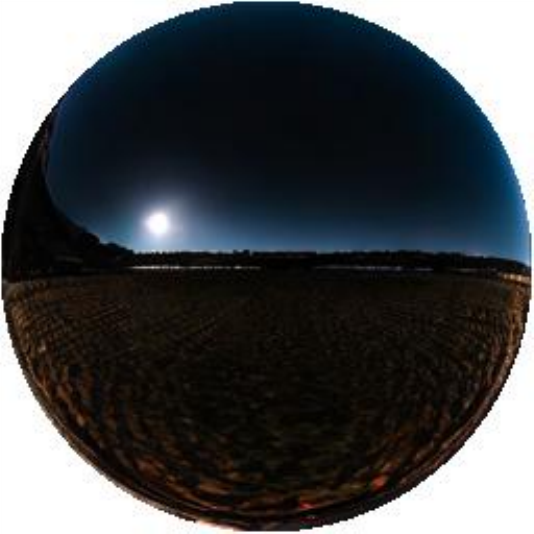}} & 
        \noindent\parbox[c]{0.095\textwidth}{\includegraphics[width=0.095\textwidth]{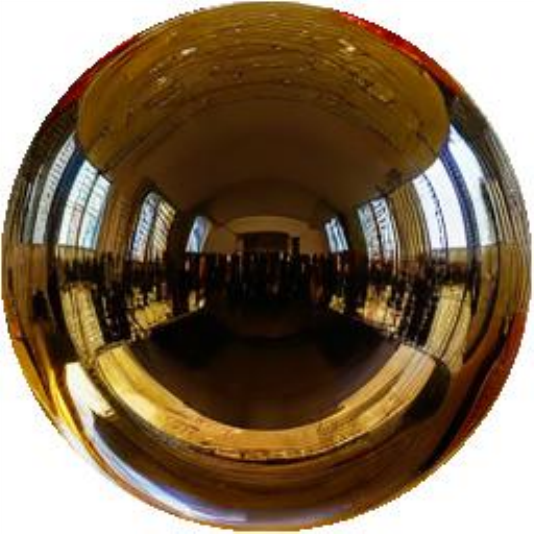}} & 
        \noindent\parbox[c]{0.095\textwidth}{\includegraphics[width=0.095\textwidth]{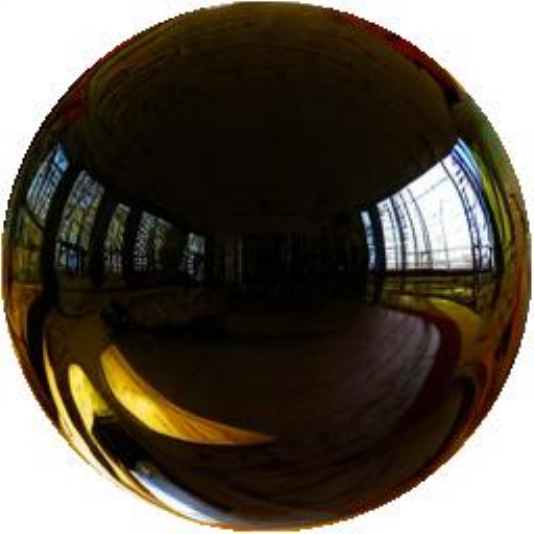}}
        
        \\

        \end{tabu}
    \caption{
    Our proposed continuous LoRA training (Cont. LoRA) yields chrome balls with higher detail consistency across different EVs than results obtained from using three separate LoRAs (3 LoRAs).}
    
    \label{fig:lora_multiple_consistency}
\end{figure*}

To show this, we conducted an experiment comparing results from our LoRA and three separately trained LoRAs at EVs 0, -2.5, and -5.0. Following the commonly adopted training pipeline \cite{ruiz2022dreambooth} implemented in the Diffusers library \cite{diffusers}, our three LoRAs are trained with the prompt containing the `\textit{sks}' token: ``a perfect sks mirrored reflective chrome ball sphere.'' We use the same hyperparameters, random seeds, and HDR panoramas during training. We present results without our iterative algorithm to isolate the effect of LoRA in Figure \ref{fig:lora_multiple_consistency}. Note that we use a lora scale of 1.0 to apply the same weight residual matrices obtained from the training. Our LoRA produces chrome balls with better structure consistency, particularly at EV-5.0.


\tabulinesep=0.5pt
\setlength{\tabcolsep}{0.5pt}


\section{More Comparison with SOTA Inpainting Techniques} \label{appendix:compare_sota}

In Figure \ref{fig:inpaint_sota} in Section \ref{sec:related}, we provide a qualitative comparison between our approach and existing SOTA diffusion-based inpainting methods: Blended Diffusion \cite{avrahami2023blendedlatent, avrahami2022blendeddiffusion}, Paint-by-Example \cite{yang2023paint}, IP-Adapter \cite{ye2023ip-adapter}, DALL·E2 \cite{dalle2}, Adobe Firefly \cite{adobefirefly}, and SDXL \cite{podell2023sdxl}. In this section, we describe the experimental settings for these methods. Additionally, we investigate the behavior of each using different random seeds.

\subsection{Experimental setups}
\begin{figure}
    \centering
    \includegraphics[width=0.49\textwidth]{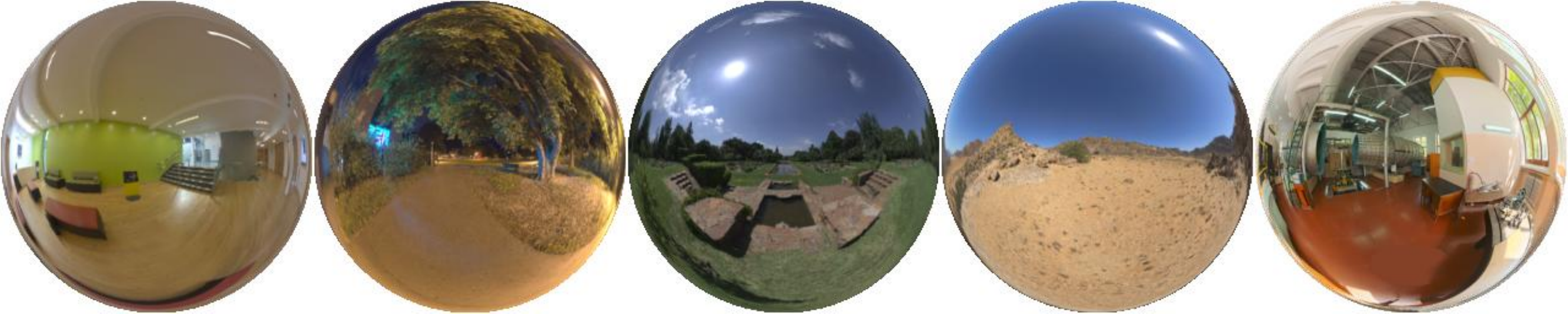}
    \caption{Chrome balls used as inputs for Paint-by-Example \cite{yang2023paint} and IP-Adapter \cite{ye2023ip-adapter}. We generate them from five randomly selected HDR environment maps from Poly Haven dataset \cite{polyhaven}.}
    \label{fig:text2light_reference_ball}
\end{figure}

Blended Diffusion, IP-Adapter, and SDXL shared the same text prompt: ``a perfect mirrored reflective chrome ball sphere.''. We used negative prompt ``matte, diffuse, flat, dull'' when executing methods that can accept one: IP-Adapter and SDXL. We provided Paint-by-Example and IP-Adapter with reference chrome balls from five randomly selected HDR environment maps in Poly Haven dataset \cite{polyhaven} as shown in Figure \ref{fig:text2light_reference_ball}. We used the official OpenAI API \footnote{\url{https://platform.openai.com/docs/guides/images/edits-dall-e-2-only}} for DALL·E2, and we used the Generative Fill feature in Photoshop for Adobe Firefly. We followed the default configurations in the methods' official implementations as described in Table \ref{tab:sota_default_config}.

\setlength{\tabcolsep}{3pt}
\begin{table}
    \centering
    \begin{tabu}{lccc}
    \toprule
      \multicolumn{1}{c}{Method}   &  Sampler & \#step & cfg \\
      \midrule
      Blended Diffusion   &  DDIM \cite{ho2020denoising} & 50 & 7.5 \\
      Paint-by-Example   &  PLMS \cite{liu2022pseudo} & 50 & 5.0 \\
      IP-Adapter   &  UniPC \cite{zhao2023unipc} & 30 & 5.0 \\
      SDXL & UniPC \cite{zhao2023unipc} & 30 & 5.0 \\
      \bottomrule
    \end{tabu}
    \caption{Sampler, number of sampling steps, and classifier-guidance scale (cfg) used in different SOTA methods.}
    \label{tab:sota_default_config}
\end{table}

\subsection{Behavior under different random seeds}
\begin{figure*}
    \centering

    \begin{tabu} to \textwidth {
        @{}
        l@{}
        l@{\hspace{0.5pt}}
    }
        \multicolumn{1}{l}{\rotatebox[origin=c]{90}{\shortstack[l]{\scriptsize Input \\ \scriptsize image}}} &
        \noindent\parbox[c]{0.5\textwidth}{\includegraphics[width=0.5\textwidth]{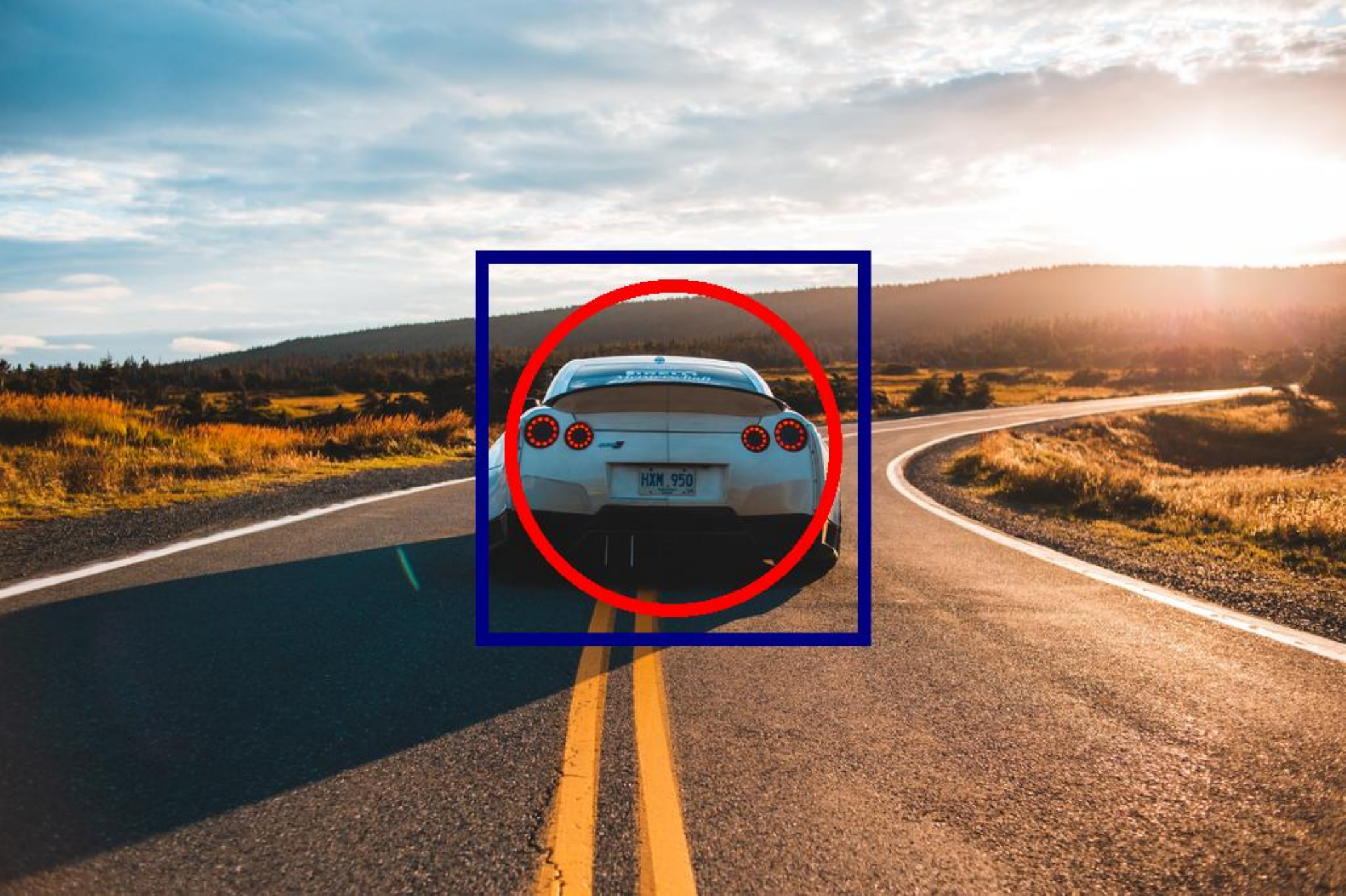}}  \\
    \end{tabu}
    \begin{tabu} to \textwidth {
        @{}
        c@{}
        c@{\hspace{0.5pt}}
        c@{\hspace{0.5pt}}
        c@{\hspace{0.5pt}}
        c@{\hspace{0.5pt}}
        c@{\hspace{0.5pt}}
        c@{\hspace{0.5pt}}
        c@{\hspace{0.5pt}}
        c@{\hspace{0.5pt}}
        c@{\hspace{0.5pt}}
        c@{\hspace{0.5pt}}
        c@{}
    }
        

        \multicolumn{1}{l}{\rotatebox[origin=c]{90}{\shortstack[l]{\scriptsize Blended Dif-\\ \scriptsize fusion \cite{avrahami2023blendedlatent, avrahami2022blendeddiffusion}}}} &
        \noindent\parbox[c]{0.081\textwidth}{\includegraphics[width=0.081\textwidth]{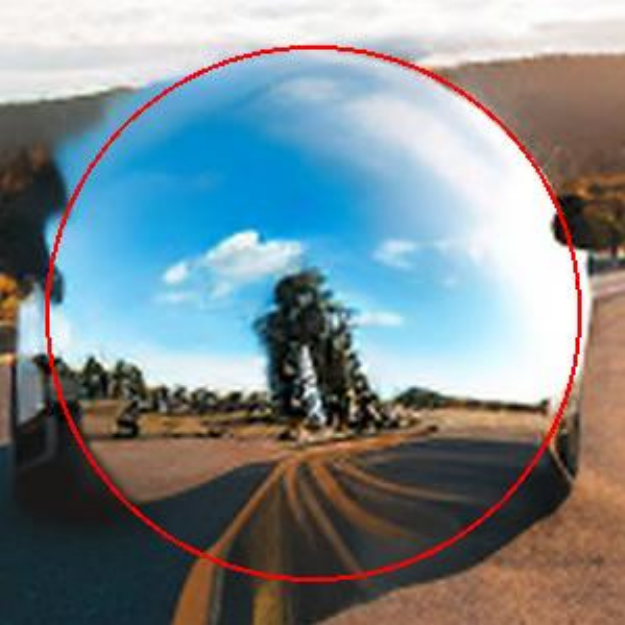}} & 
        \noindent\parbox[c]{0.081\textwidth}{\includegraphics[width=0.081\textwidth]{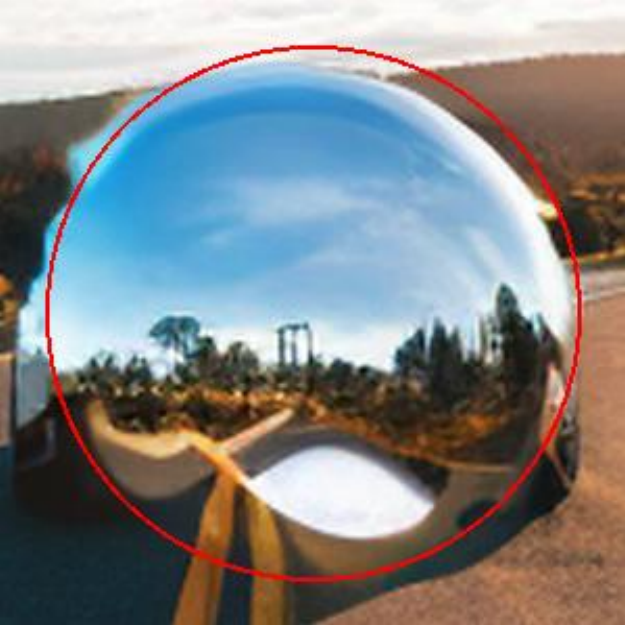}} &  
        \noindent\parbox[c]{0.081\textwidth}{\includegraphics[width=0.081\textwidth]{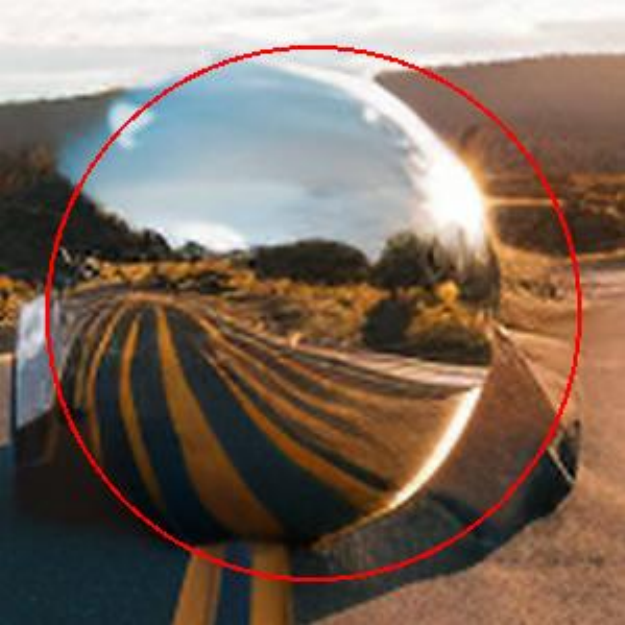}} & 
        \noindent\parbox[c]{0.081\textwidth}{\includegraphics[width=0.081\textwidth]{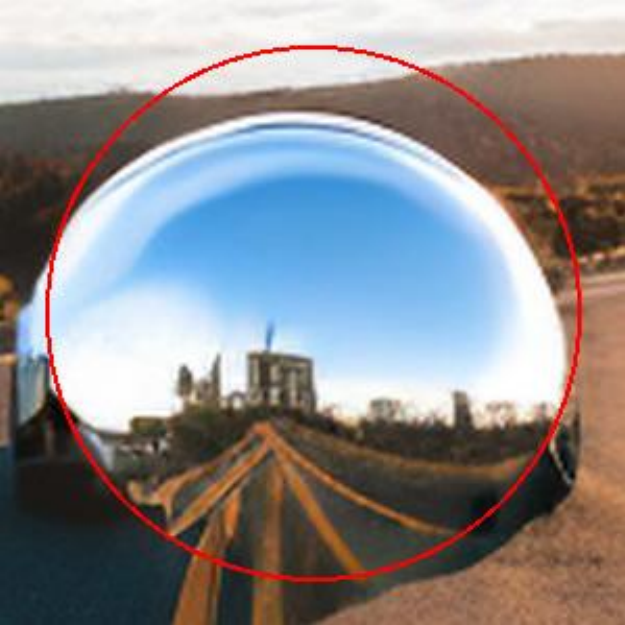}} & 
        \noindent\parbox[c]{0.081\textwidth}{\includegraphics[width=0.081\textwidth]{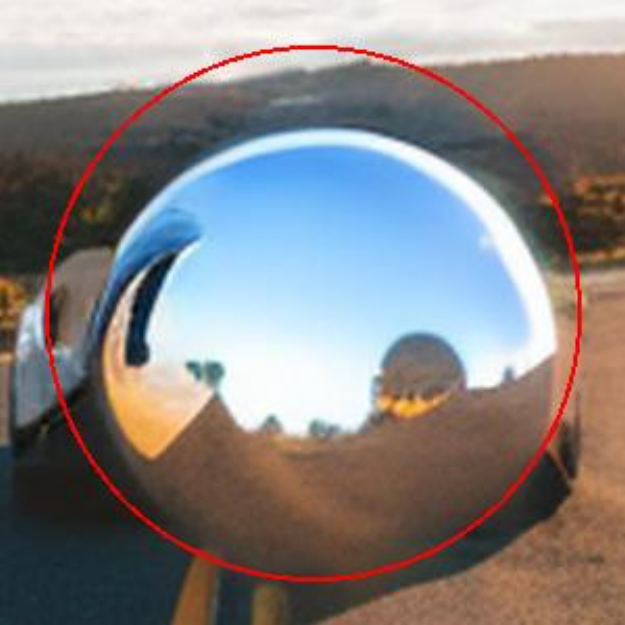}} & 
        \noindent\parbox[c]{0.081\textwidth}{\includegraphics[width=0.081\textwidth]{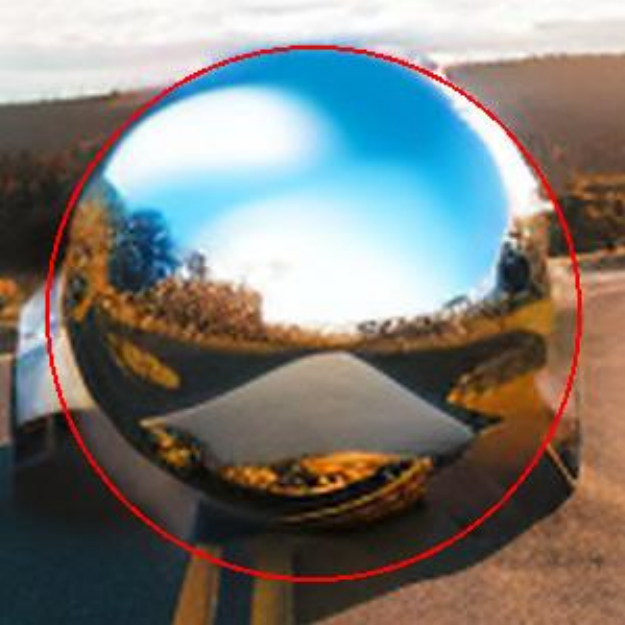}} & 
        \noindent\parbox[c]{0.081\textwidth}{\includegraphics[width=0.081\textwidth]{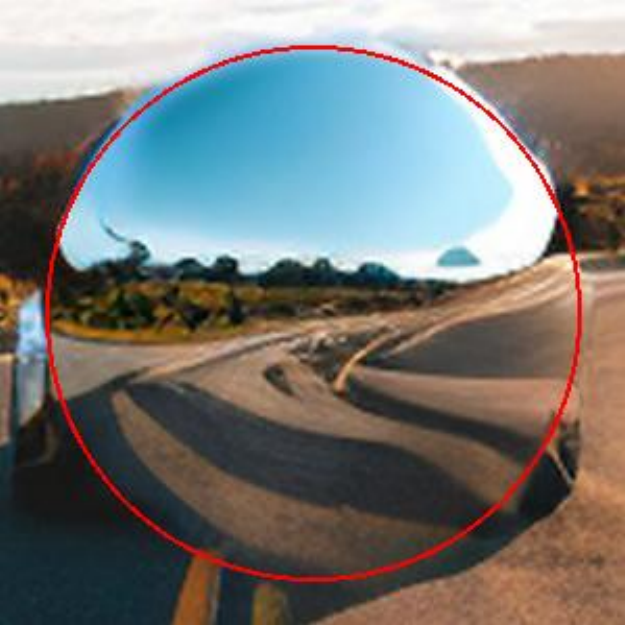}} & 
        \noindent\parbox[c]{0.081\textwidth}{\includegraphics[width=0.081\textwidth]{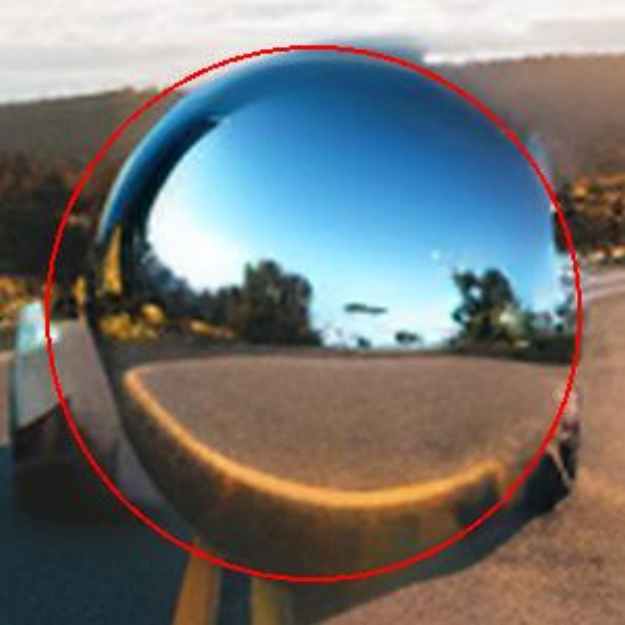}} & 
        \noindent\parbox[c]{0.081\textwidth}{\includegraphics[width=0.081\textwidth]{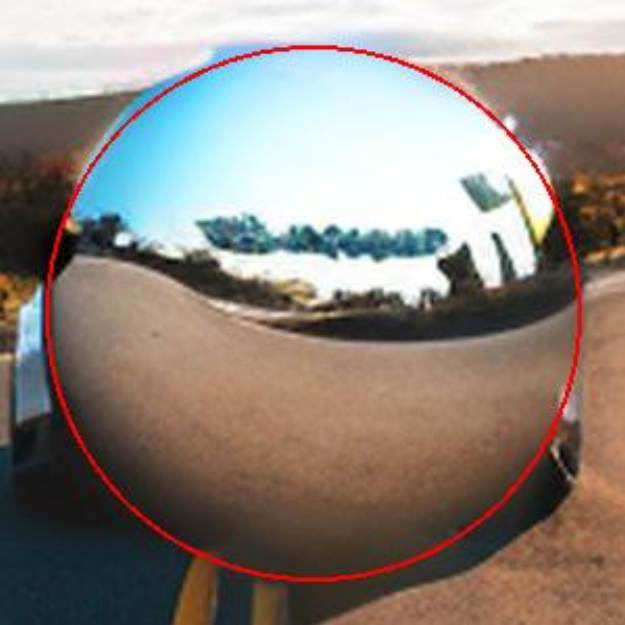}} & 
        \noindent\parbox[c]{0.081\textwidth}{\includegraphics[width=0.081\textwidth]{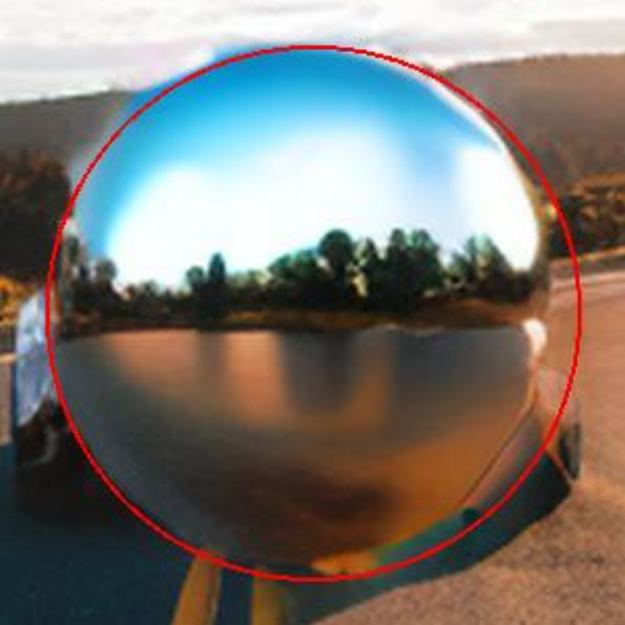}} & 
        
        \\

        \multicolumn{1}{l}{\rotatebox[origin=c]{90}{\shortstack[l]{\scriptsize Paint-by-Ex\\ \scriptsize ample \cite{yang2023paint}}}} &
        \noindent\parbox[c]{0.081\textwidth}{\includegraphics[width=0.081\textwidth]{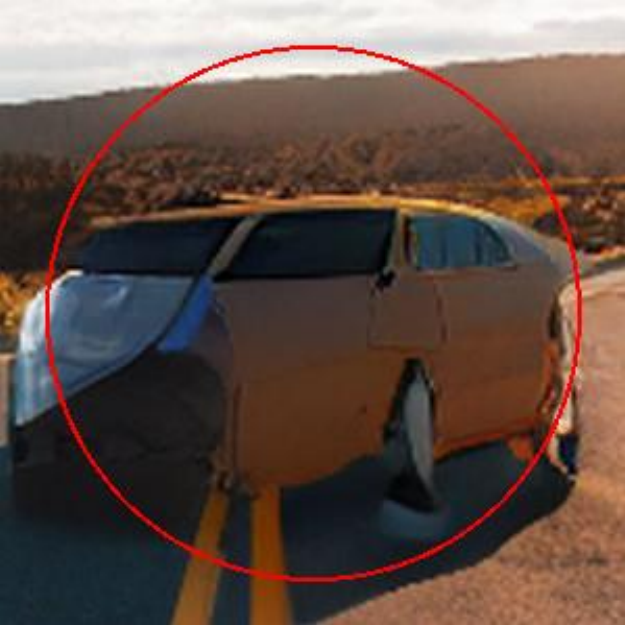}} & 
        \noindent\parbox[c]{0.081\textwidth}{\includegraphics[width=0.081\textwidth]{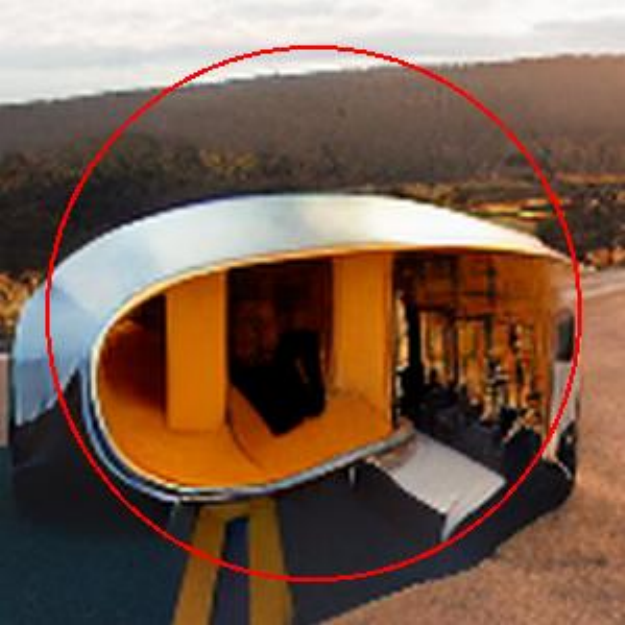}} &  
        \noindent\parbox[c]{0.081\textwidth}{\includegraphics[width=0.081\textwidth]{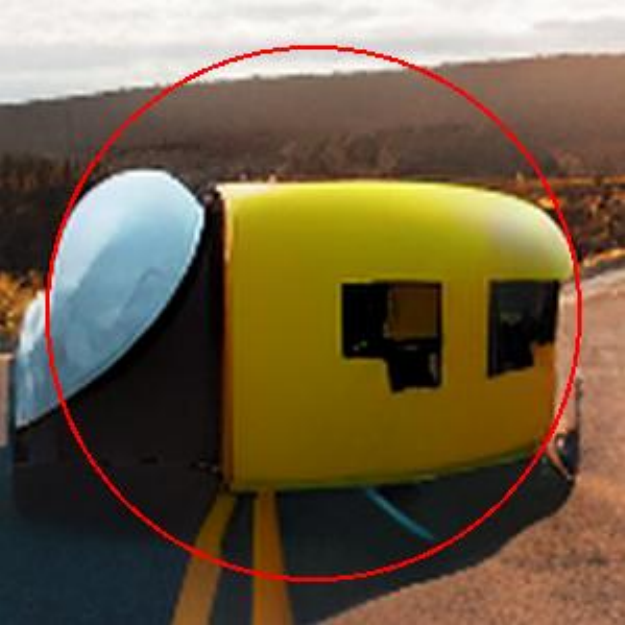}} & 
        \noindent\parbox[c]{0.081\textwidth}{\includegraphics[width=0.081\textwidth]{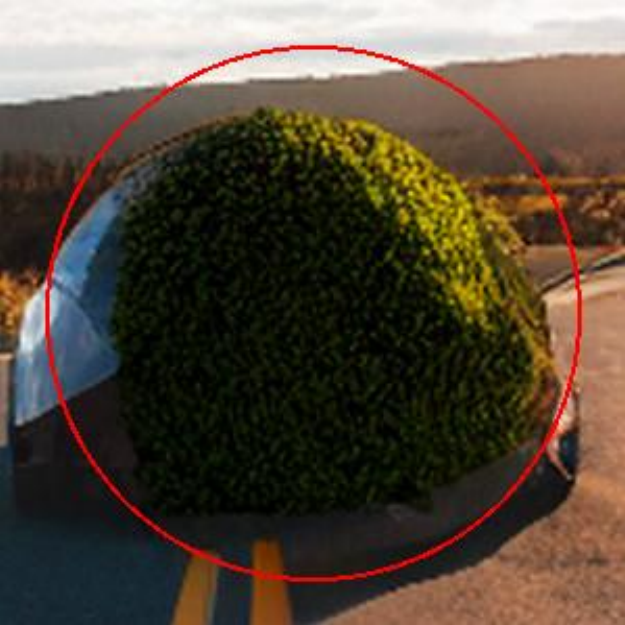}} & 
        \noindent\parbox[c]{0.081\textwidth}{\includegraphics[width=0.081\textwidth]{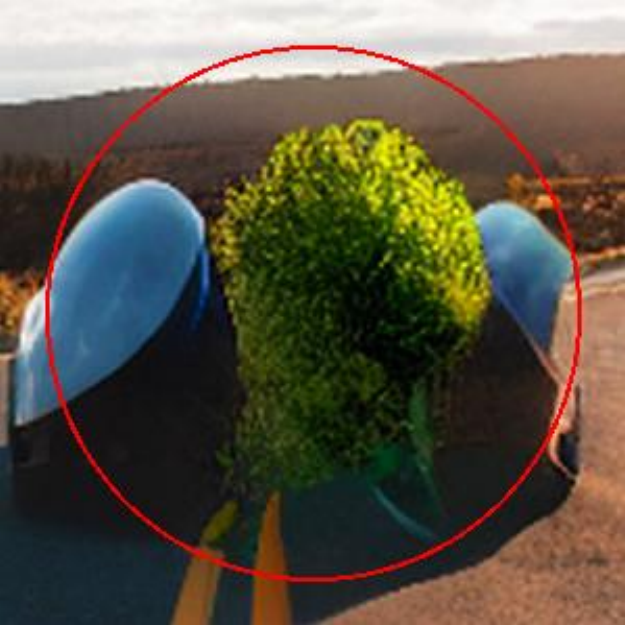}} & 
        \noindent\parbox[c]{0.081\textwidth}{\includegraphics[width=0.081\textwidth]{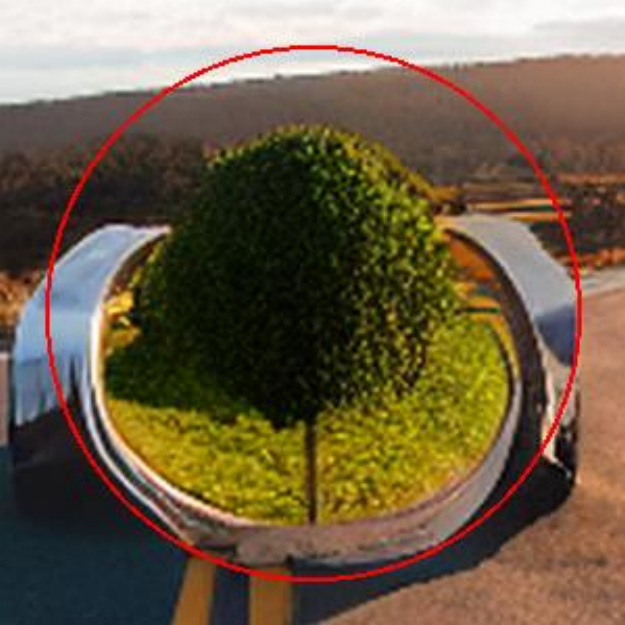}} & 
        \noindent\parbox[c]{0.081\textwidth}{\includegraphics[width=0.081\textwidth]{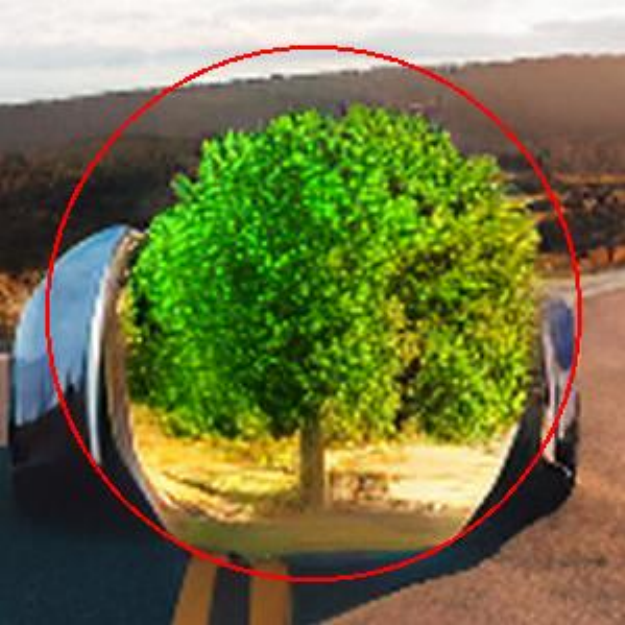}} & 
        \noindent\parbox[c]{0.081\textwidth}{\includegraphics[width=0.081\textwidth]{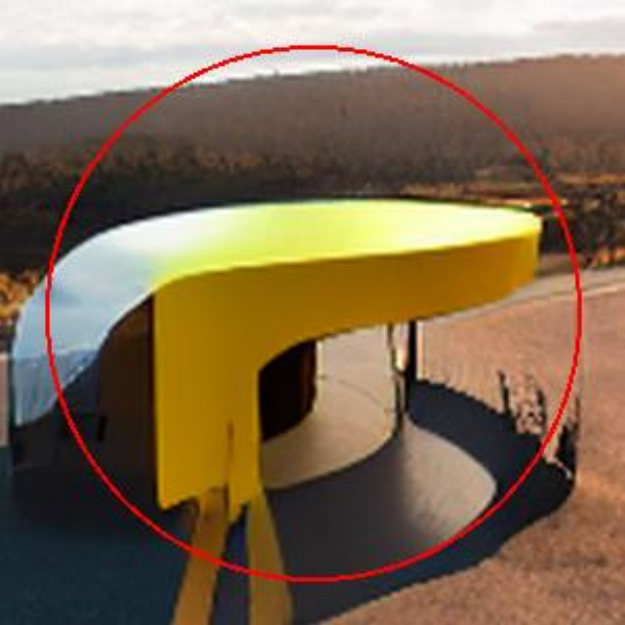}} & 
        \noindent\parbox[c]{0.081\textwidth}{\includegraphics[width=0.081\textwidth]{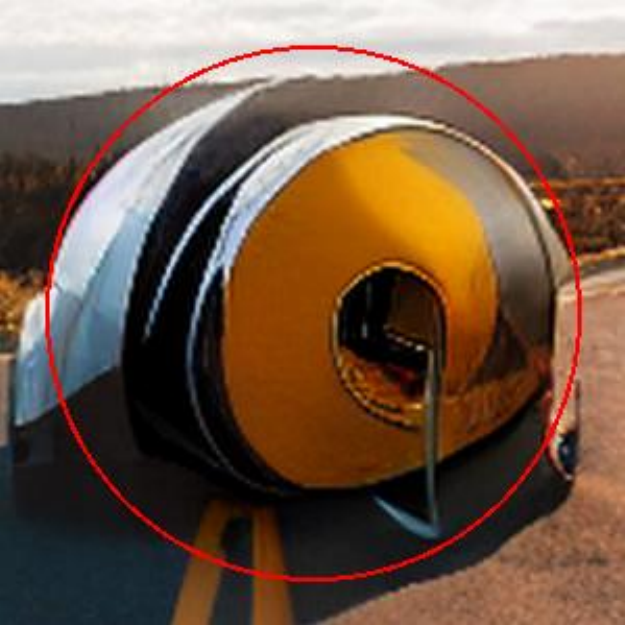}} & 
        \noindent\parbox[c]{0.081\textwidth}{\includegraphics[width=0.081\textwidth]{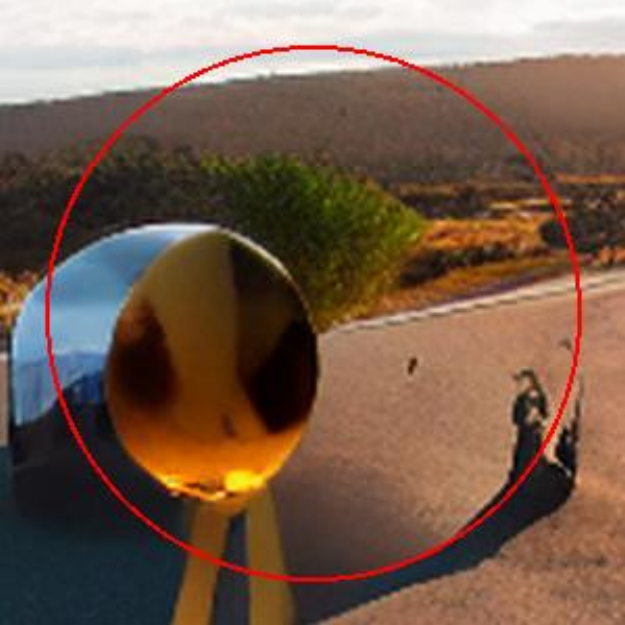}} & 
        
        \\

        \multicolumn{1}{l}{\rotatebox[origin=c]{90}{\shortstack[l]{\scriptsize IP-Adapter\\ \scriptsize \cite{ye2023ip-adapter}}}} &
        \noindent\parbox[c]{0.081\textwidth}{\includegraphics[width=0.081\textwidth]{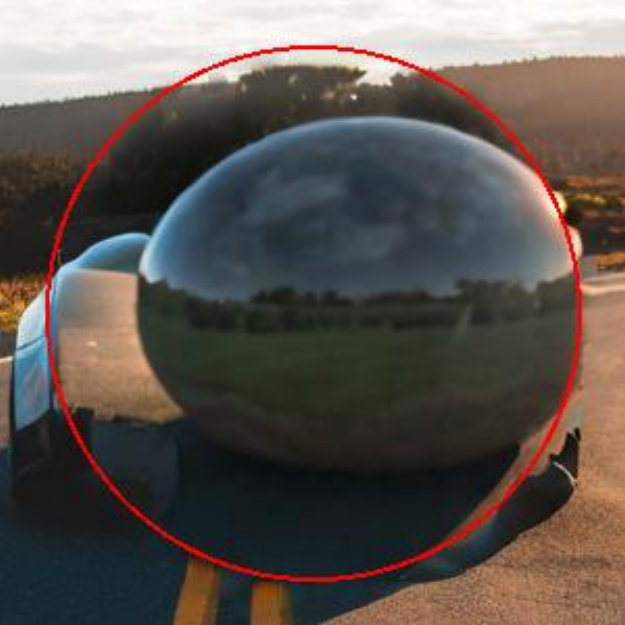}} & 
        \noindent\parbox[c]{0.081\textwidth}{\includegraphics[width=0.081\textwidth]{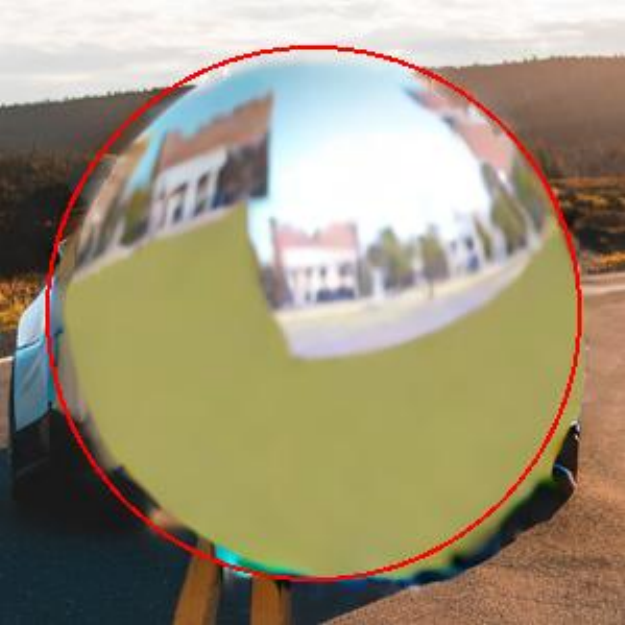}} &  
        \noindent\parbox[c]{0.081\textwidth}{\includegraphics[width=0.081\textwidth]{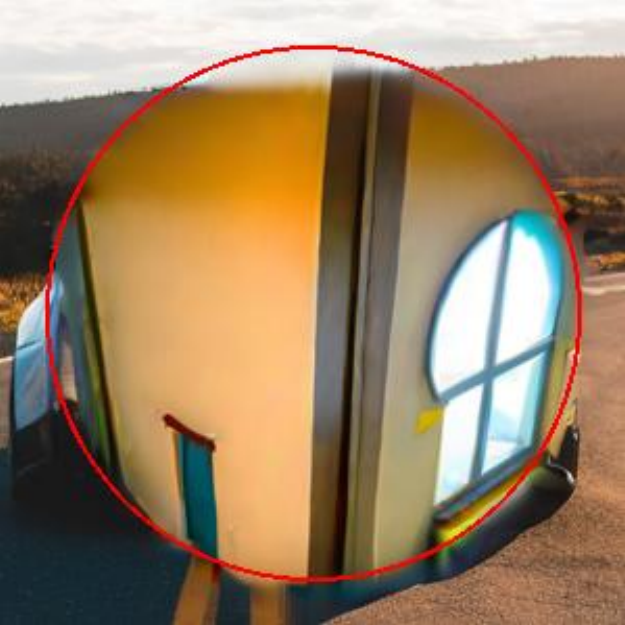}} & 
        \noindent\parbox[c]{0.081\textwidth}{\includegraphics[width=0.081\textwidth]{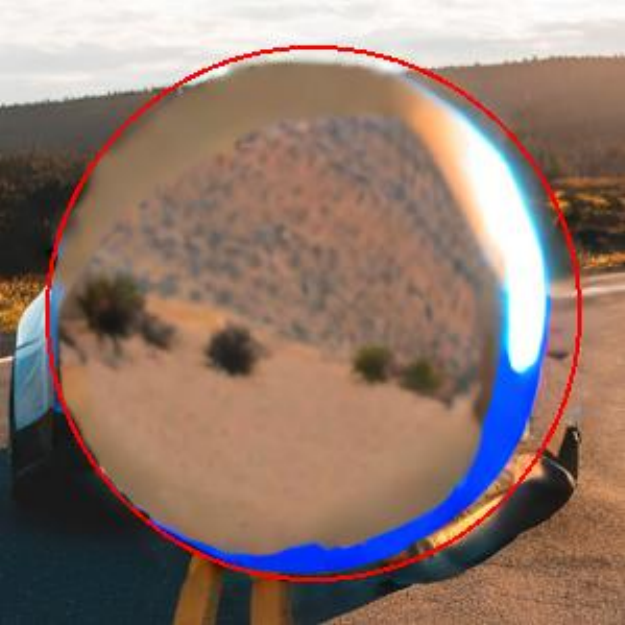}} & 
        \noindent\parbox[c]{0.081\textwidth}{\includegraphics[width=0.081\textwidth]{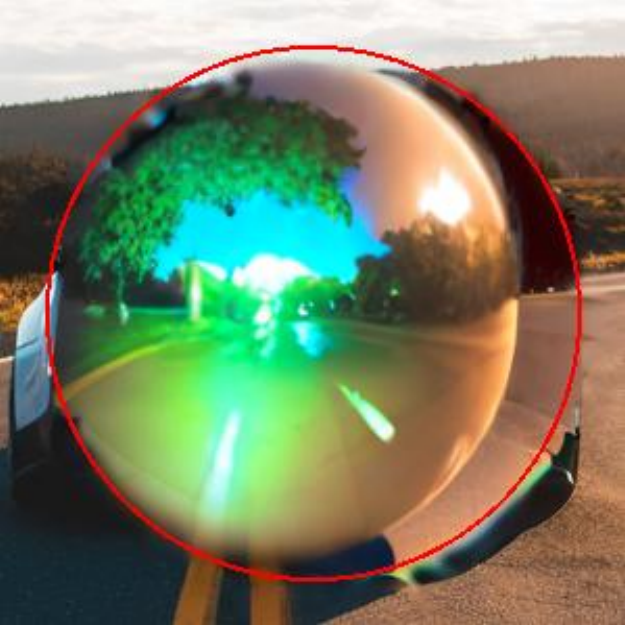}} & 
        \noindent\parbox[c]{0.081\textwidth}{\includegraphics[width=0.081\textwidth]{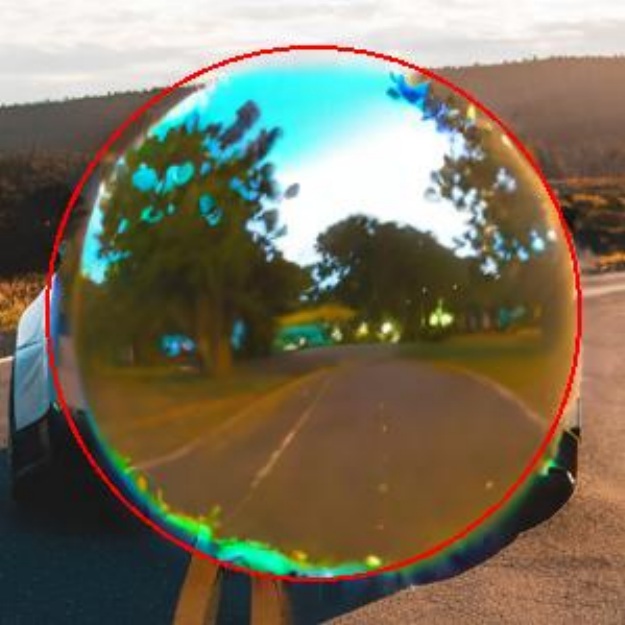}} & 
        \noindent\parbox[c]{0.081\textwidth}{\includegraphics[width=0.081\textwidth]{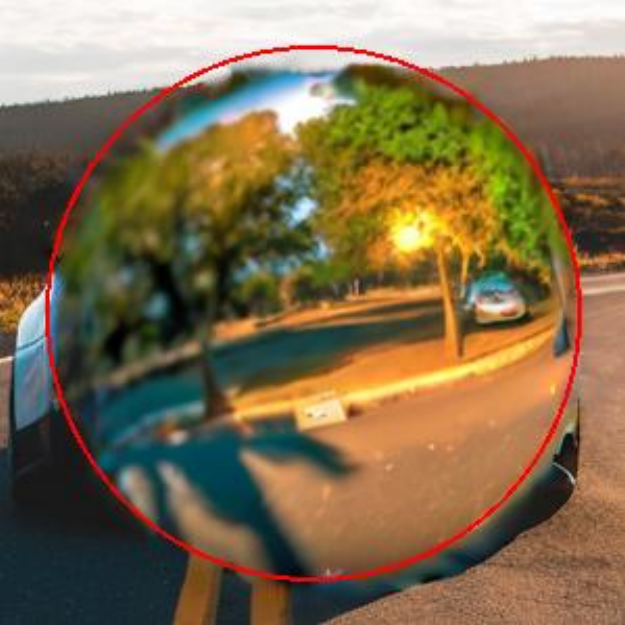}} & 
        \noindent\parbox[c]{0.081\textwidth}{\includegraphics[width=0.081\textwidth]{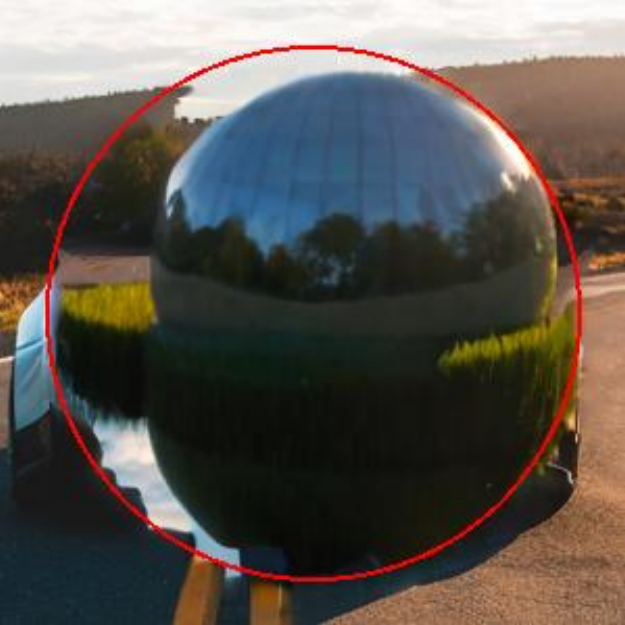}} & 
        \noindent\parbox[c]{0.081\textwidth}{\includegraphics[width=0.081\textwidth]{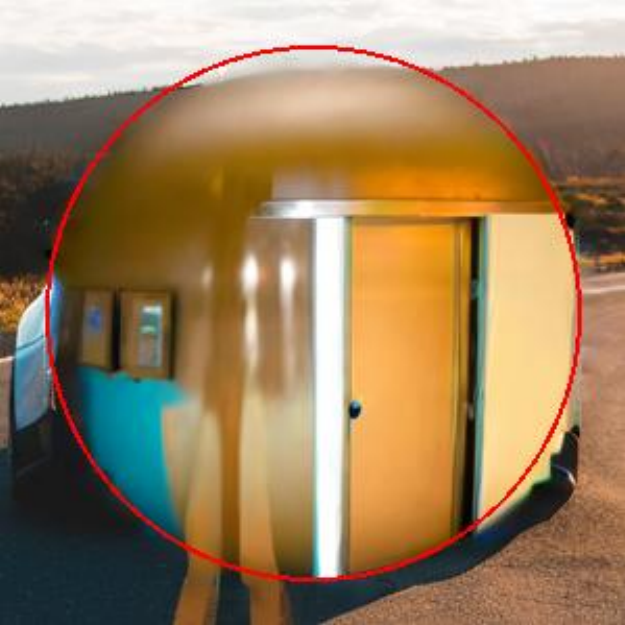}} & 
        \noindent\parbox[c]{0.081\textwidth}{\includegraphics[width=0.081\textwidth]{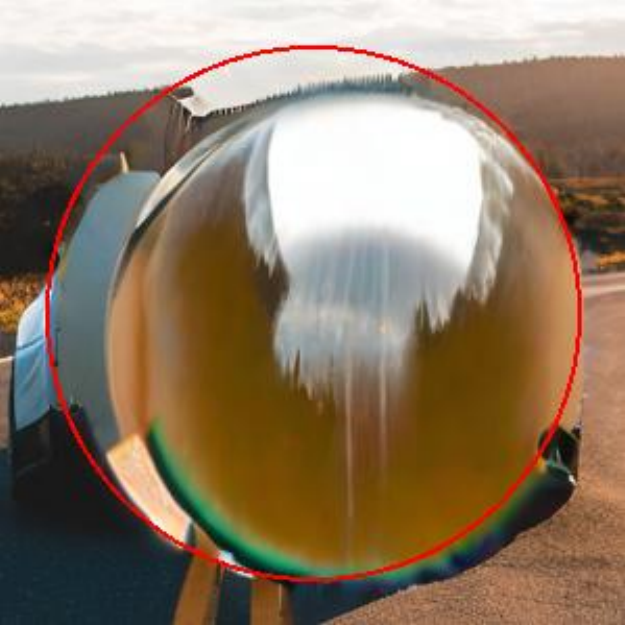}} & 
        
        \\

        \multicolumn{1}{l}{\rotatebox[origin=c]{90}{\shortstack[l]{\scriptsize DALL·E2 \cite{dalle2}}}} &
        \noindent\parbox[c]{0.081\textwidth}{\includegraphics[width=0.081\textwidth]{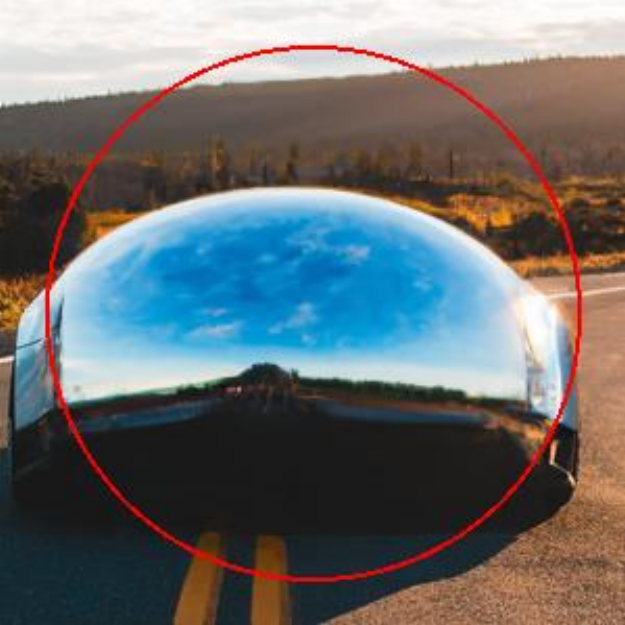}} & 
        \noindent\parbox[c]{0.081\textwidth}{\includegraphics[width=0.081\textwidth]{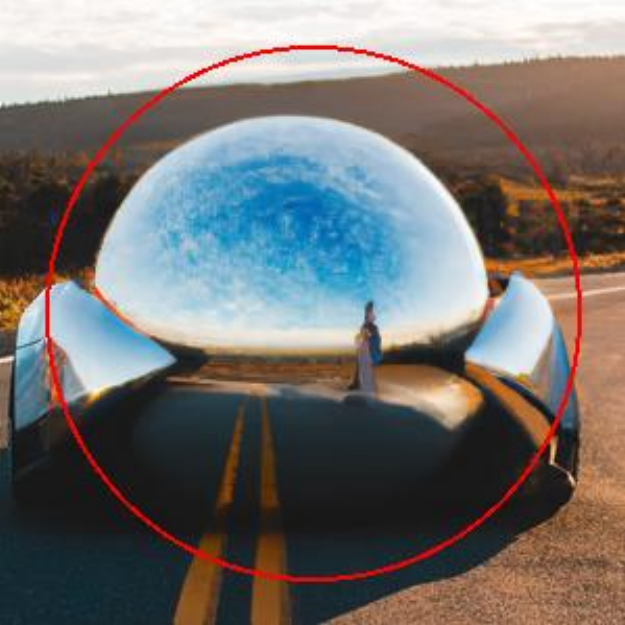}} &  
        \noindent\parbox[c]{0.081\textwidth}{\includegraphics[width=0.081\textwidth]{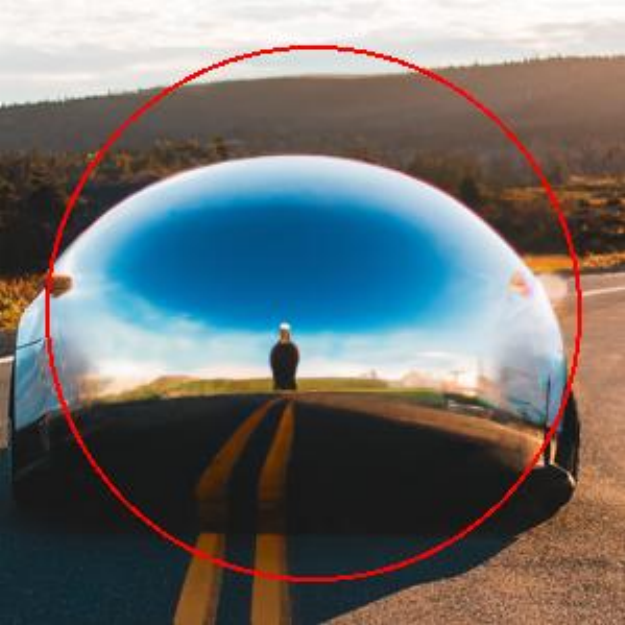}} & 
        \noindent\parbox[c]{0.081\textwidth}{\includegraphics[width=0.081\textwidth]{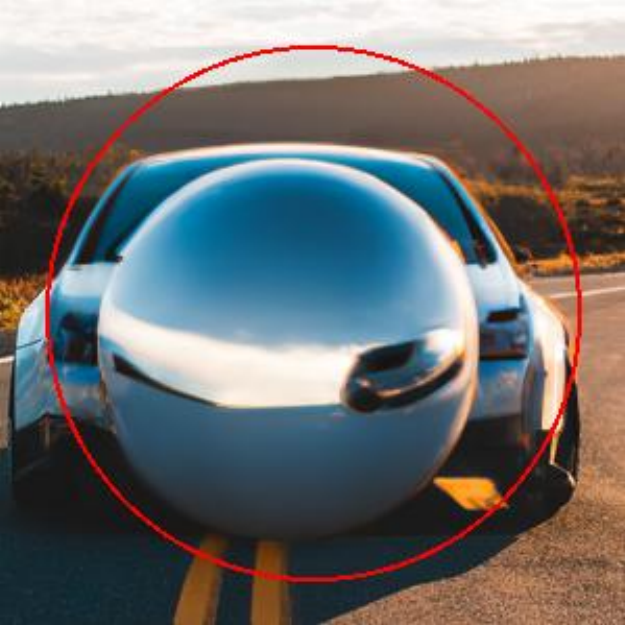}} & 
        \noindent\parbox[c]{0.081\textwidth}{\includegraphics[width=0.081\textwidth]{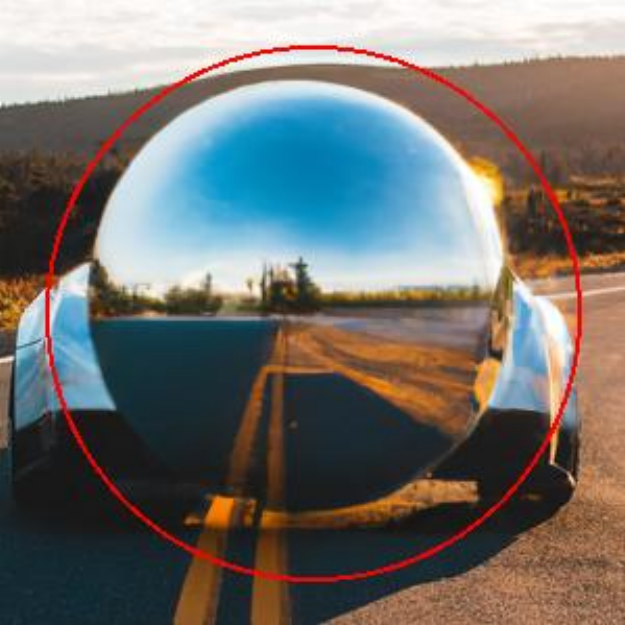}} & 
        \noindent\parbox[c]{0.081\textwidth}{\includegraphics[width=0.081\textwidth]{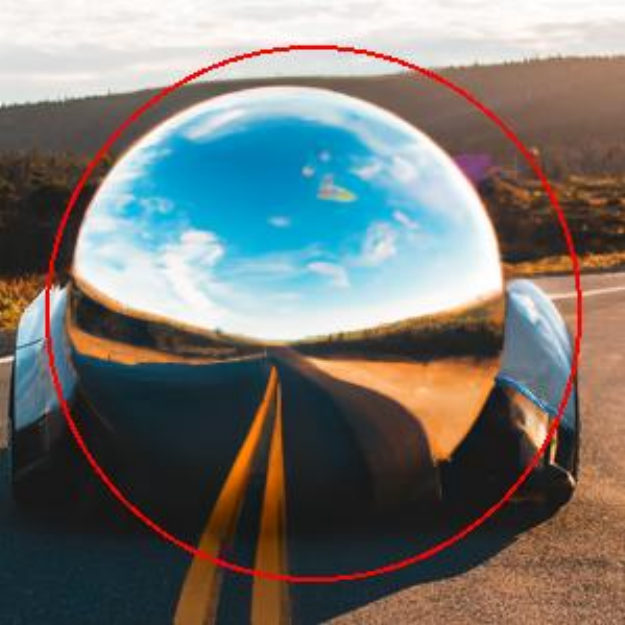}} & 
        \noindent\parbox[c]{0.081\textwidth}{\includegraphics[width=0.081\textwidth]{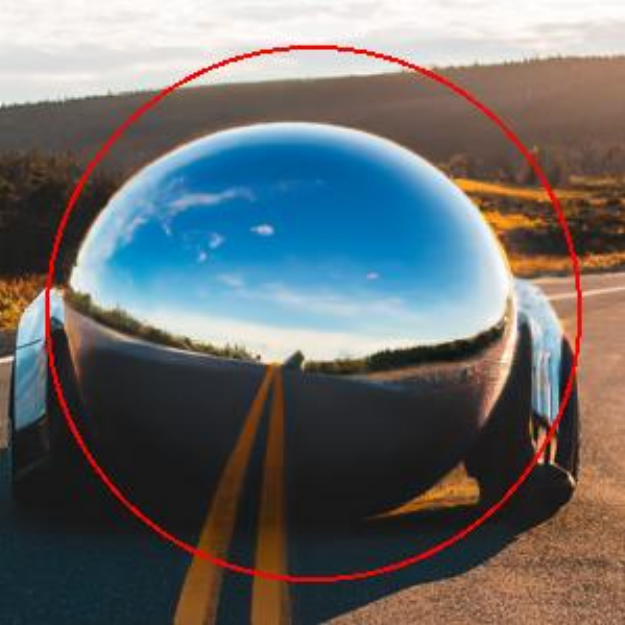}} & 
        \noindent\parbox[c]{0.081\textwidth}{\includegraphics[width=0.081\textwidth]{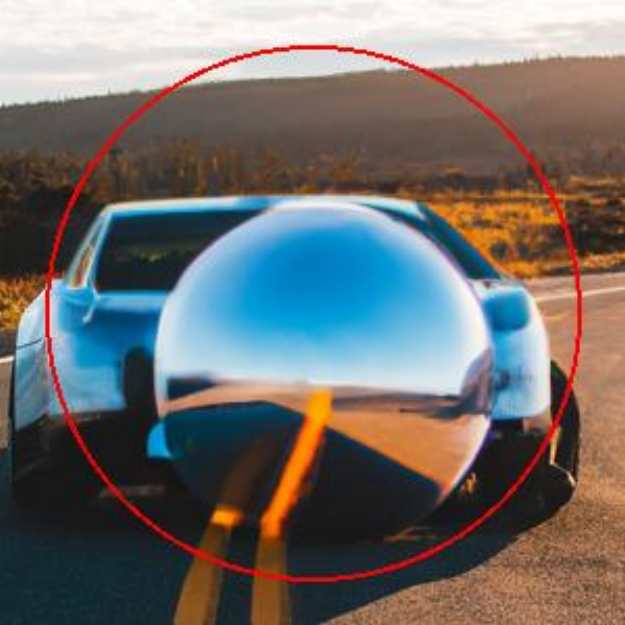}} & 
        \noindent\parbox[c]{0.081\textwidth}{\includegraphics[width=0.081\textwidth]{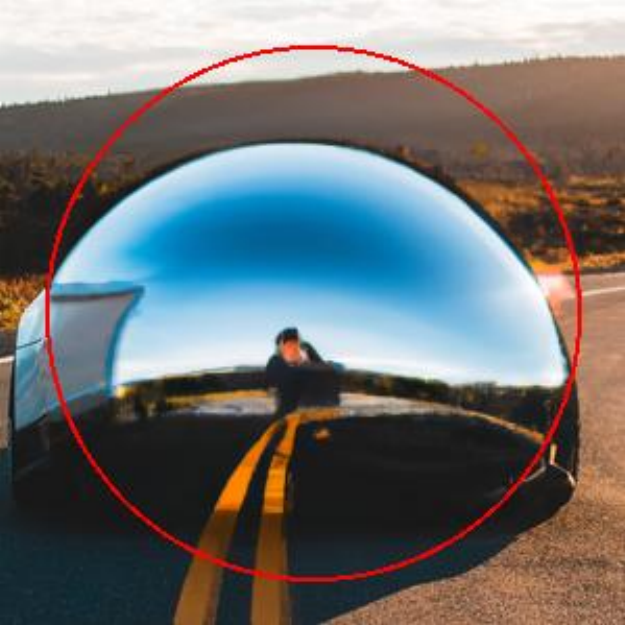}} & 
        \noindent\parbox[c]{0.081\textwidth}{\includegraphics[width=0.081\textwidth]{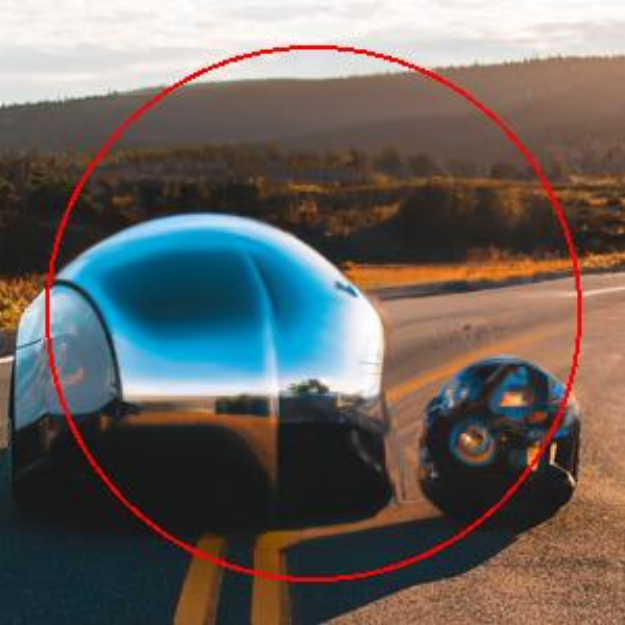}} & 
        
        \\

        \multicolumn{1}{l}{\rotatebox[origin=c]{90}{\shortstack[l]{\scriptsize Adobe \\ \scriptsize Firefly \cite{adobefirefly}}}} &
        \noindent\parbox[c]{0.081\textwidth}{\includegraphics[width=0.081\textwidth]{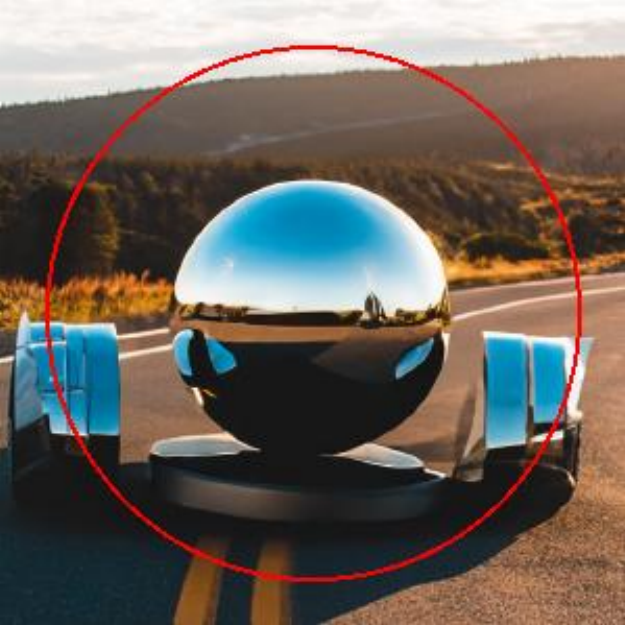}} & 
        \noindent\parbox[c]{0.081\textwidth}{\includegraphics[width=0.081\textwidth]{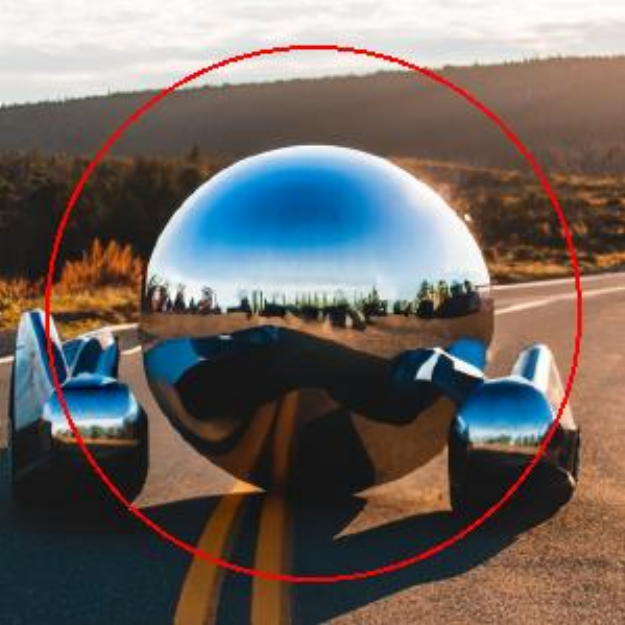}} &  
        \noindent\parbox[c]{0.081\textwidth}{\includegraphics[width=0.081\textwidth]{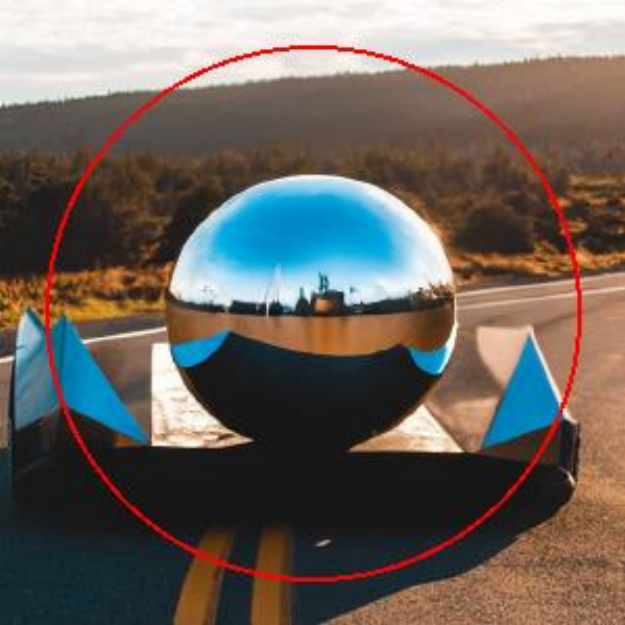}} & 
        \noindent\parbox[c]{0.081\textwidth}{\includegraphics[width=0.081\textwidth]{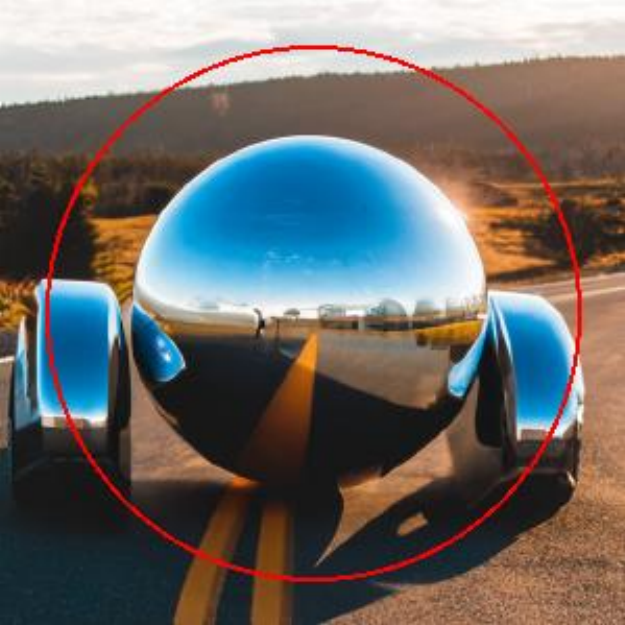}} & 
        \noindent\parbox[c]{0.081\textwidth}{\includegraphics[width=0.081\textwidth]{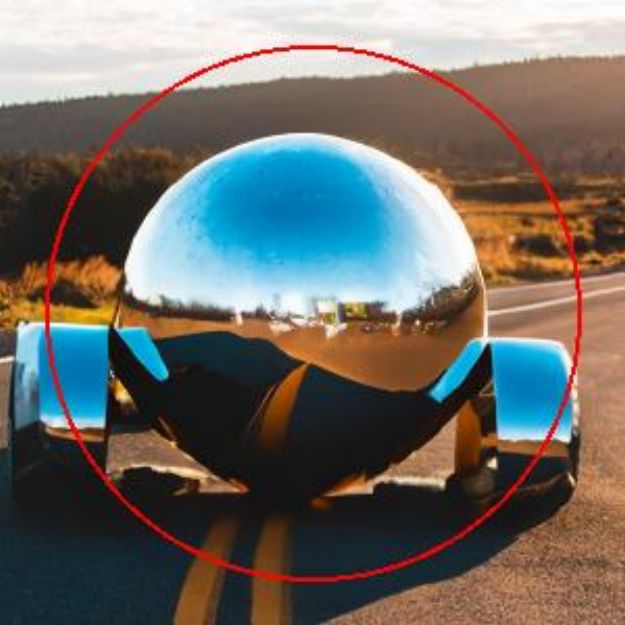}} & 
        \noindent\parbox[c]{0.081\textwidth}{\includegraphics[width=0.081\textwidth]{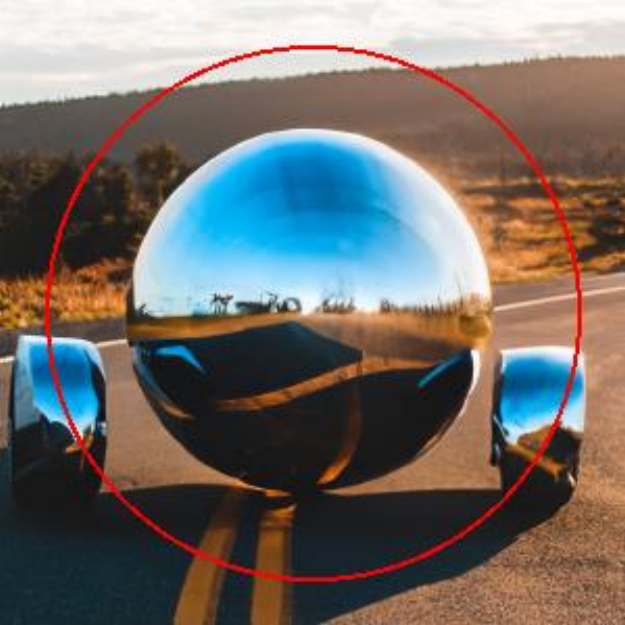}} & 
        \noindent\parbox[c]{0.081\textwidth}{\includegraphics[width=0.081\textwidth]{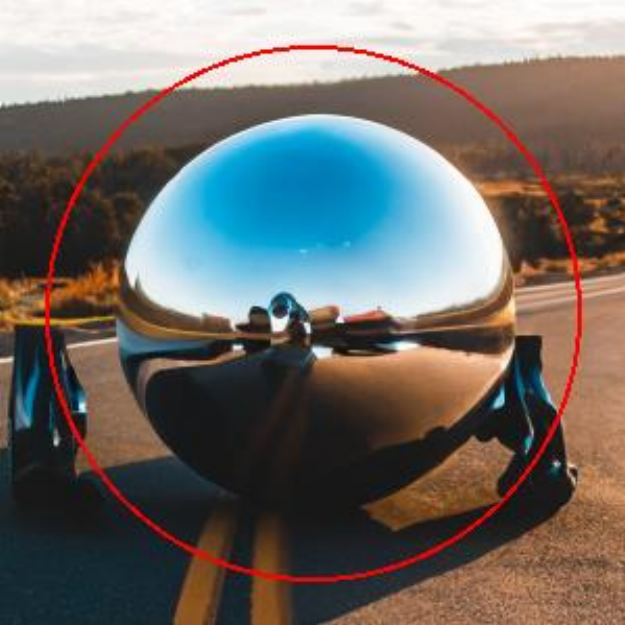}} & 
        \noindent\parbox[c]{0.081\textwidth}{\includegraphics[width=0.081\textwidth]{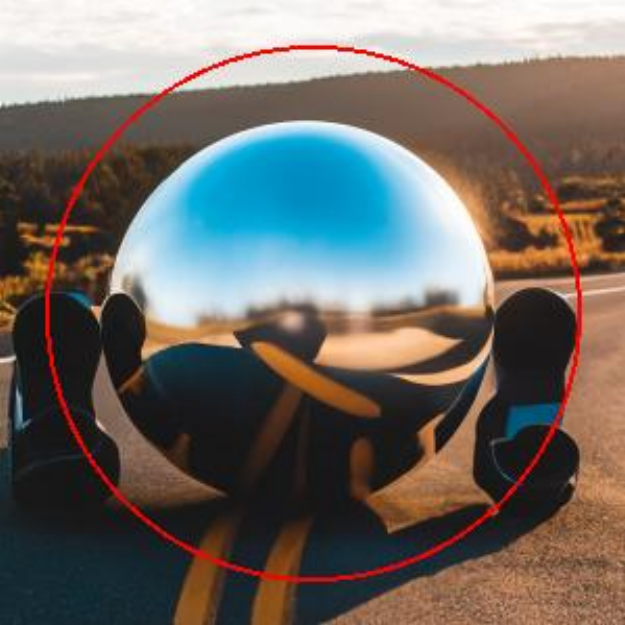}} & 
        \noindent\parbox[c]{0.081\textwidth}{\includegraphics[width=0.081\textwidth]{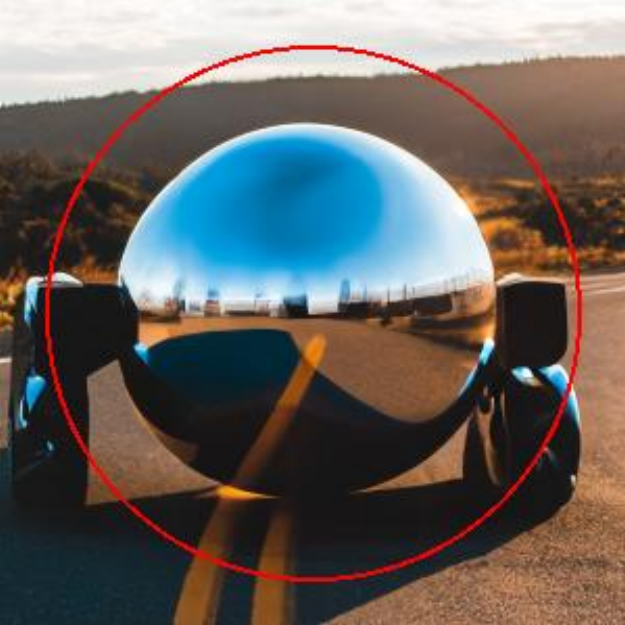}} & 
        \noindent\parbox[c]{0.081\textwidth}{\includegraphics[width=0.081\textwidth]{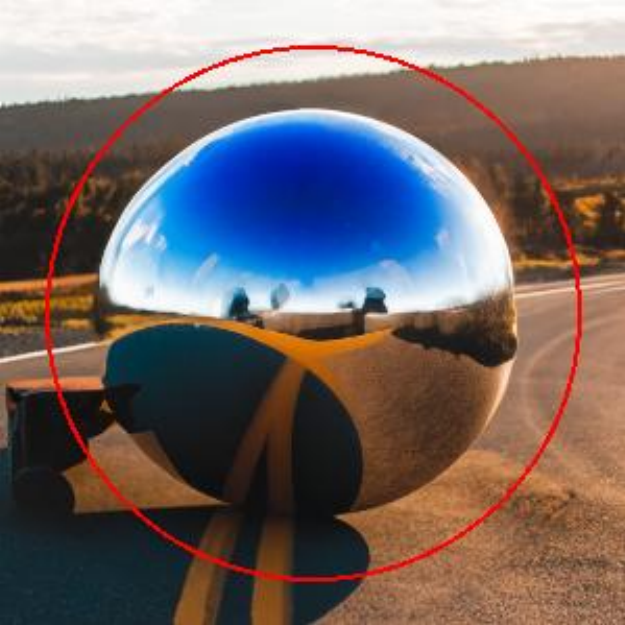}} & 

        \\

        \multicolumn{1}{l}{\rotatebox[origin=c]{90}{\shortstack[l]{\scriptsize SDXL \cite{podell2023sdxl}}}} &
        \noindent\parbox[c]{0.081\textwidth}{\includegraphics[width=0.081\textwidth]{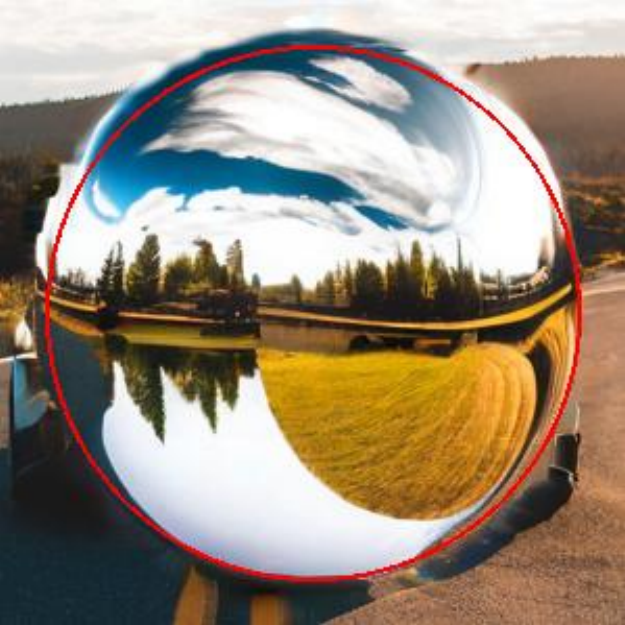}} & 
        \noindent\parbox[c]{0.081\textwidth}{\includegraphics[width=0.081\textwidth]{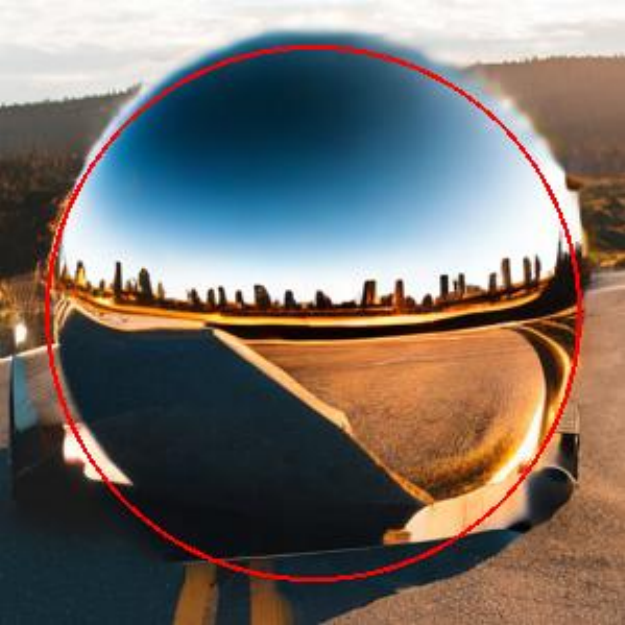}} &  
        \noindent\parbox[c]{0.081\textwidth}{\includegraphics[width=0.081\textwidth]{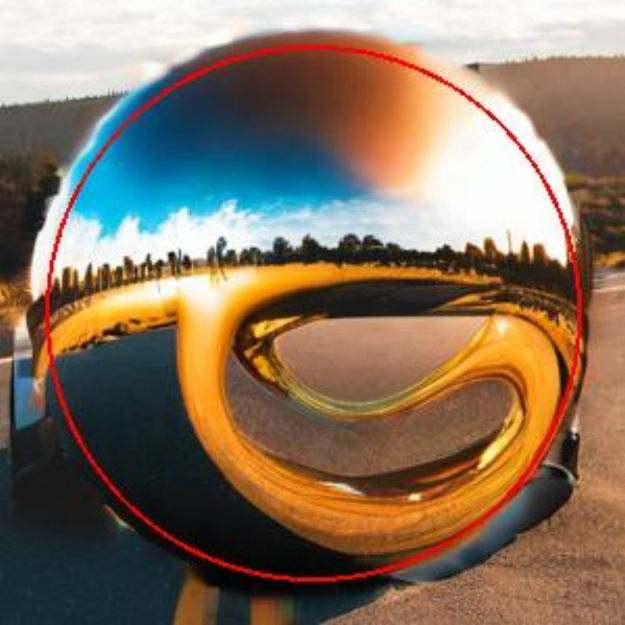}} & 
        \noindent\parbox[c]{0.081\textwidth}{\includegraphics[width=0.081\textwidth]{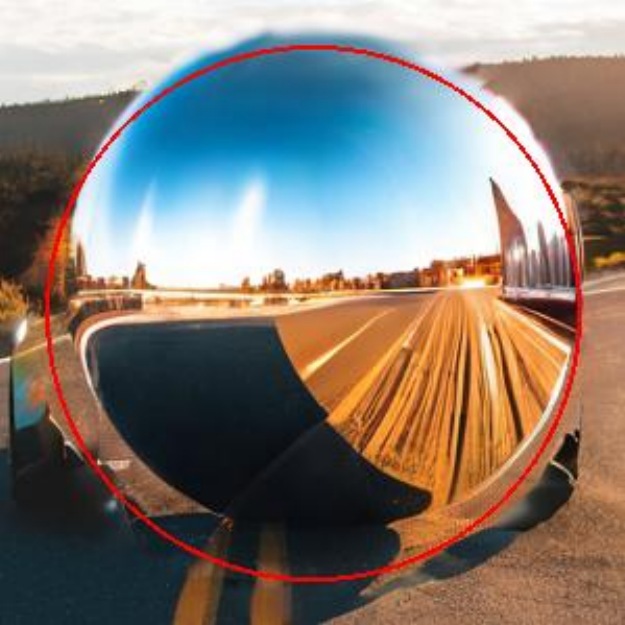}} & 
        \noindent\parbox[c]{0.081\textwidth}{\includegraphics[width=0.081\textwidth]{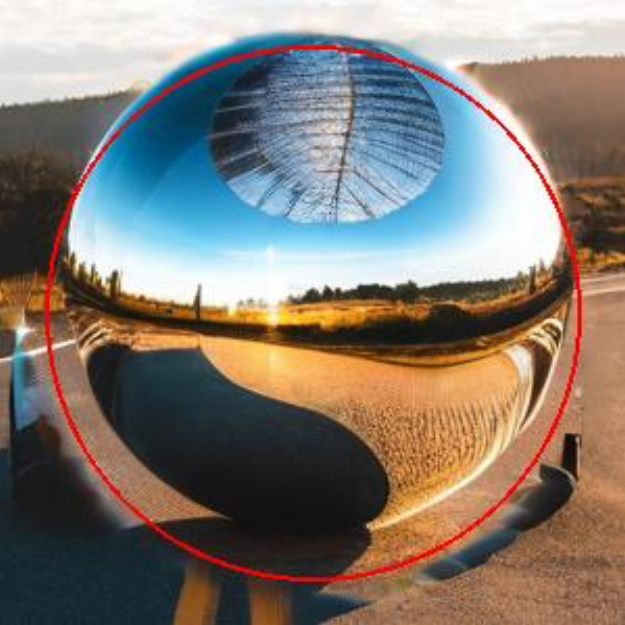}} & 
        \noindent\parbox[c]{0.081\textwidth}{\includegraphics[width=0.081\textwidth]{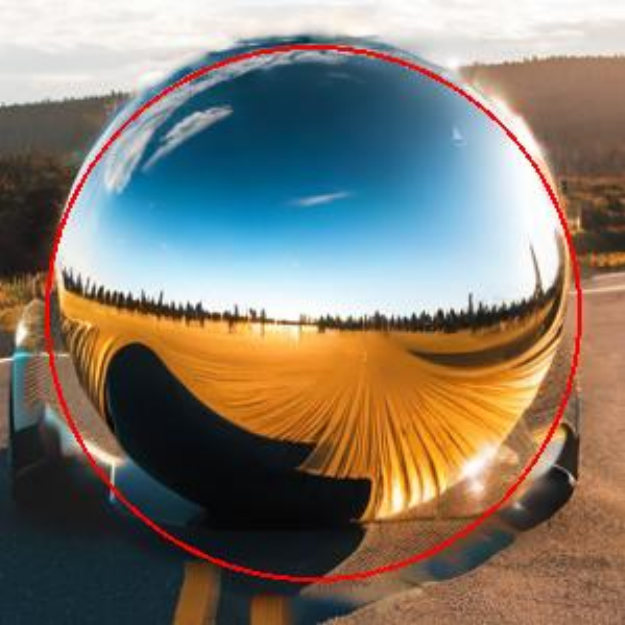}} & 
        \noindent\parbox[c]{0.081\textwidth}{\includegraphics[width=0.081\textwidth]{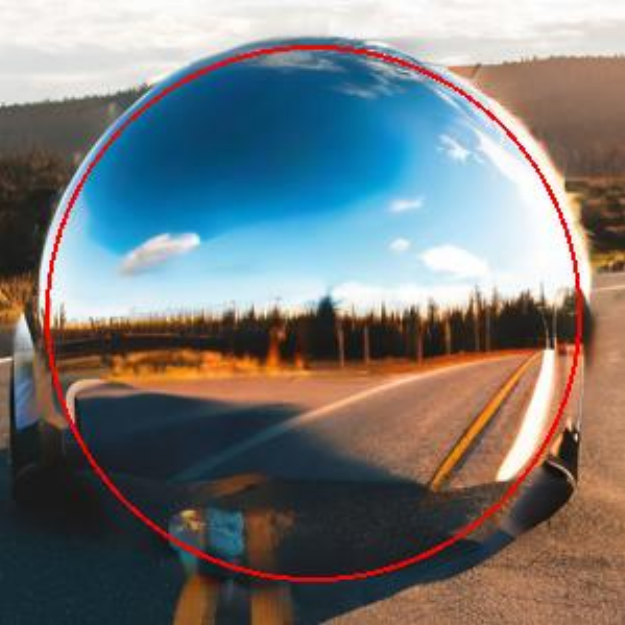}} & 
        \noindent\parbox[c]{0.081\textwidth}{\includegraphics[width=0.081\textwidth]{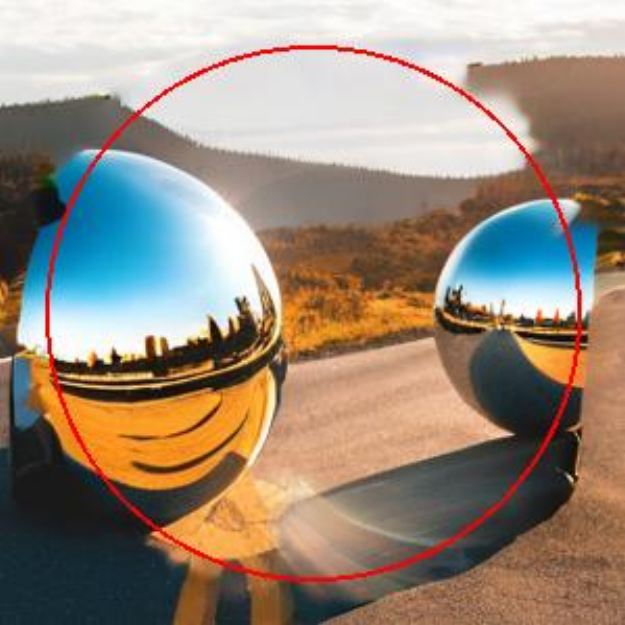}} & 
        \noindent\parbox[c]{0.081\textwidth}{\includegraphics[width=0.081\textwidth]{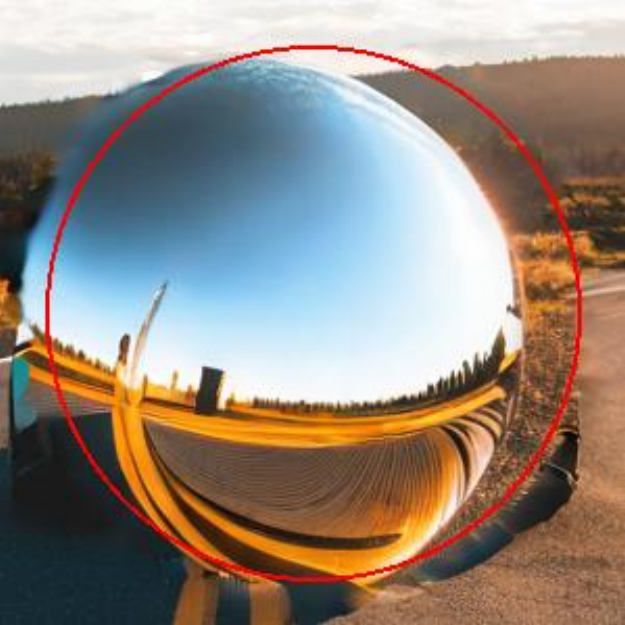}} & 
        \noindent\parbox[c]{0.081\textwidth}{\includegraphics[width=0.081\textwidth]{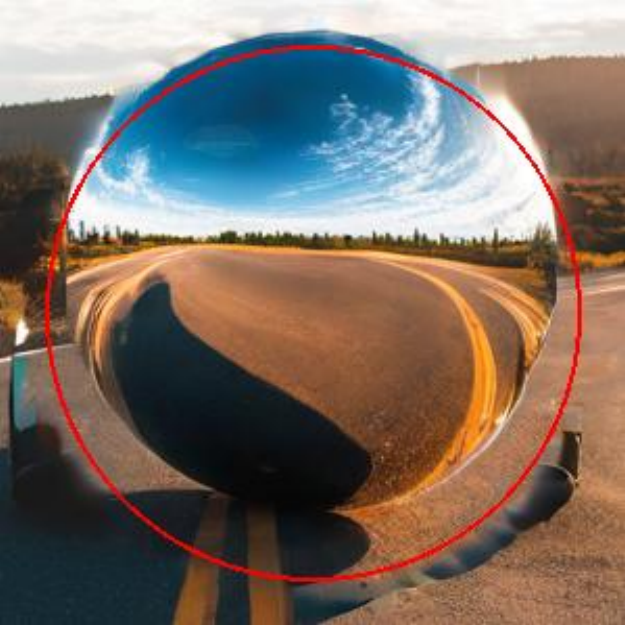}} & 
        
        \\ \hline

        \multicolumn{1}{l}{\rotatebox[origin=c]{90}{\shortstack[l]{\scriptsize \textbf{Ours}}}} &
        \noindent\parbox[c]{0.081\textwidth}{\includegraphics[width=0.081\textwidth]{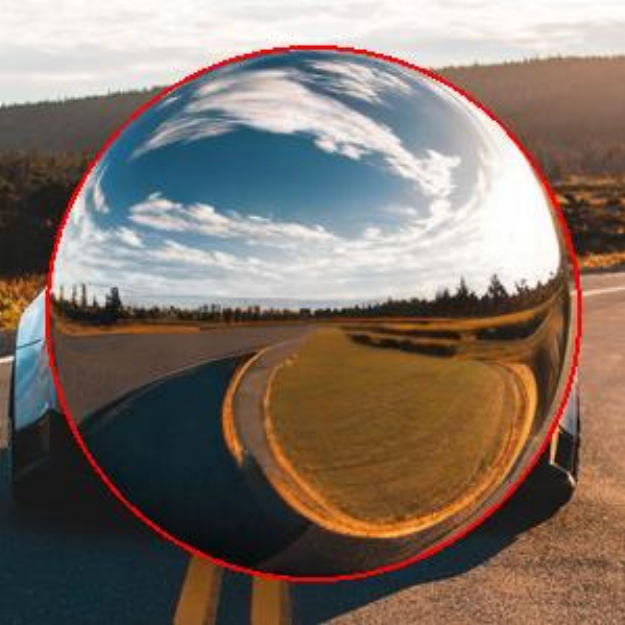}} & 
        \noindent\parbox[c]{0.081\textwidth}{\includegraphics[width=0.081\textwidth]{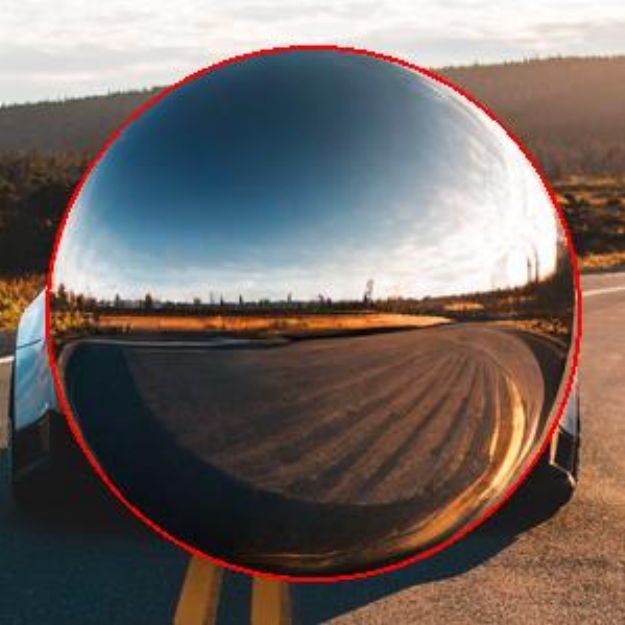}} &  
        \noindent\parbox[c]{0.081\textwidth}{\includegraphics[width=0.081\textwidth]{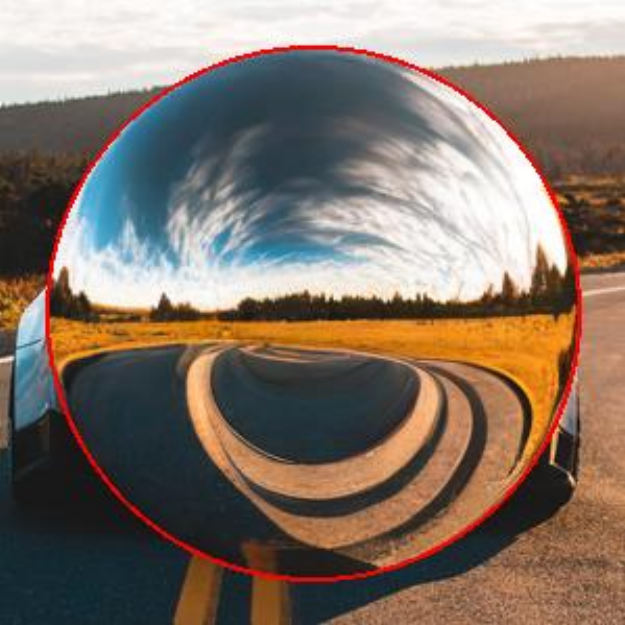}} & 
        \noindent\parbox[c]{0.081\textwidth}{\includegraphics[width=0.081\textwidth]{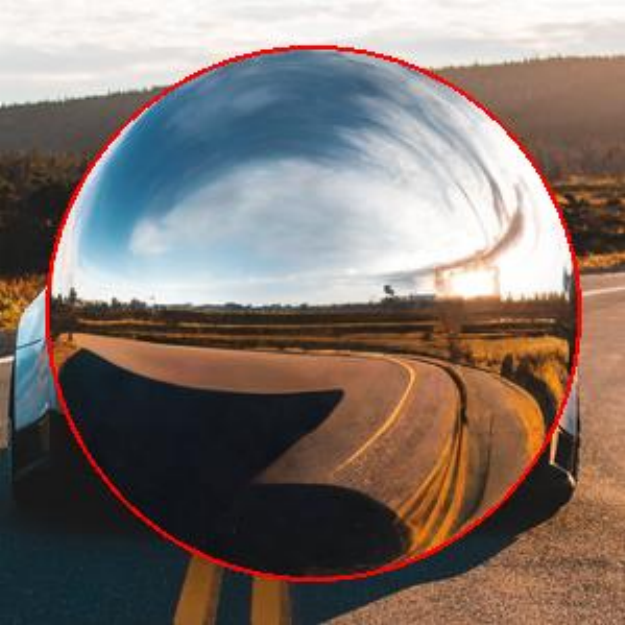}} & 
        \noindent\parbox[c]{0.081\textwidth}{\includegraphics[width=0.081\textwidth]{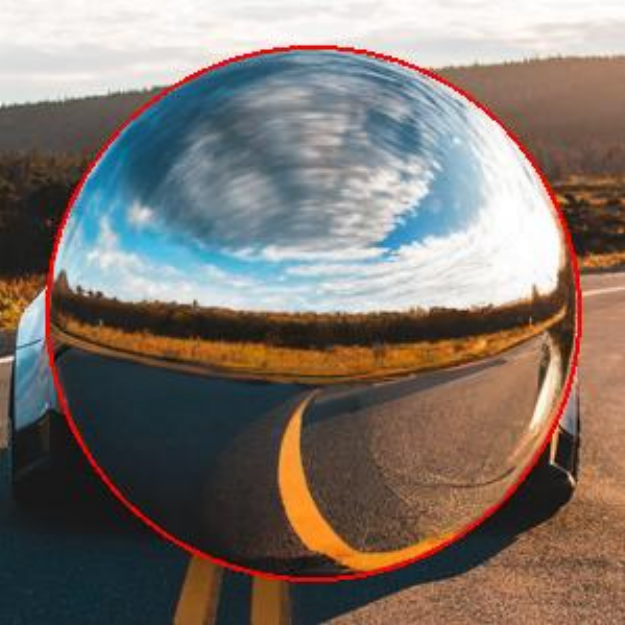}} & 
        \noindent\parbox[c]{0.081\textwidth}{\includegraphics[width=0.081\textwidth]{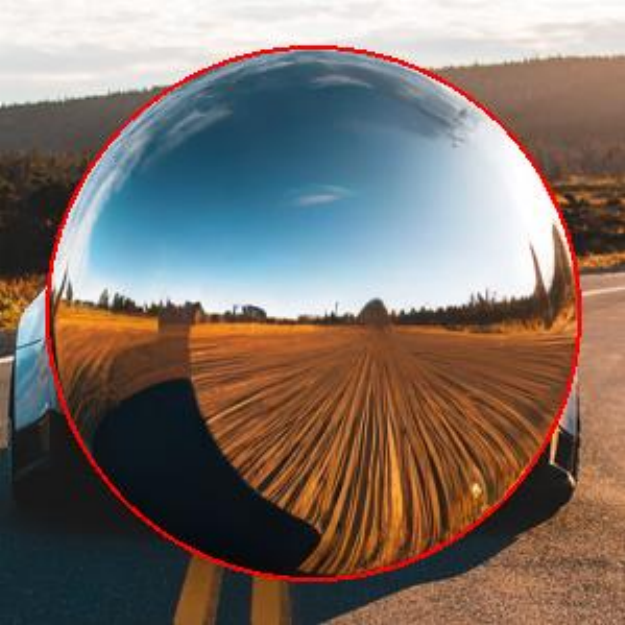}} & 
        \noindent\parbox[c]{0.081\textwidth}{\includegraphics[width=0.081\textwidth]{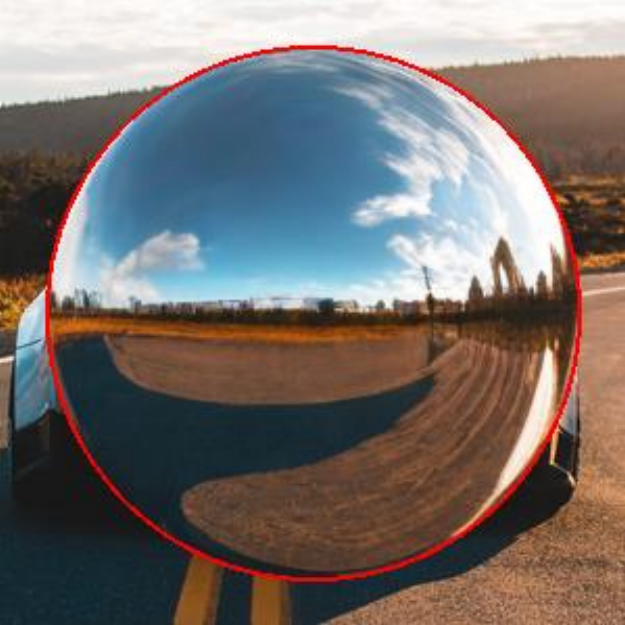}} & 
        \noindent\parbox[c]{0.081\textwidth}{\includegraphics[width=0.081\textwidth]{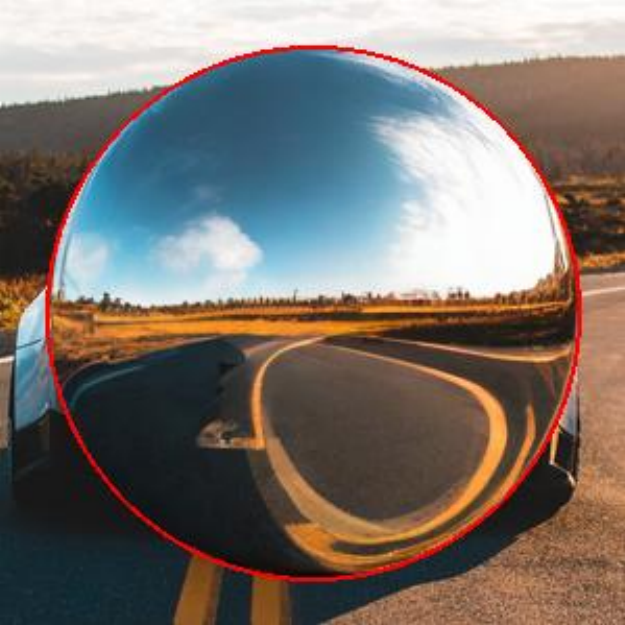}} & 
        \noindent\parbox[c]{0.081\textwidth}{\includegraphics[width=0.081\textwidth]{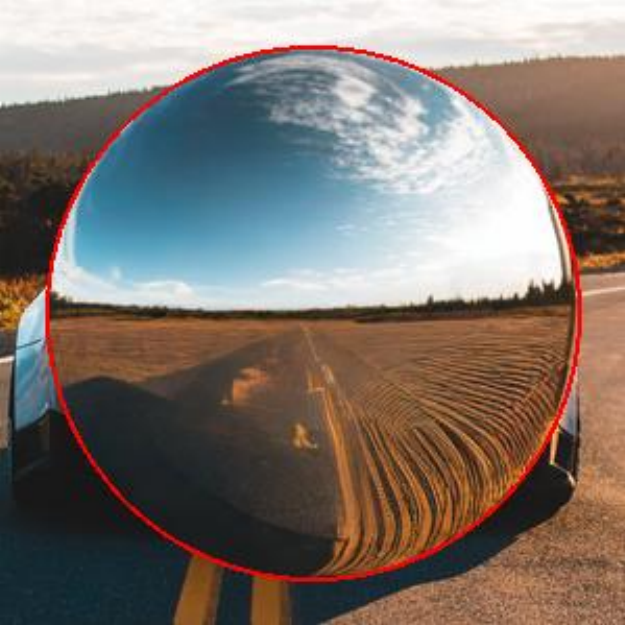}} & 
        \noindent\parbox[c]{0.081\textwidth}{\includegraphics[width=0.081\textwidth]{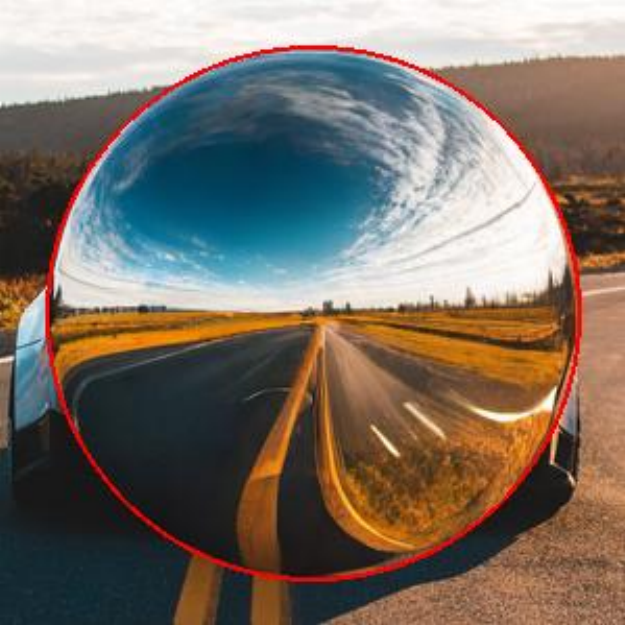}}
        
        \\
        
    \end{tabu}

    \caption{Chrome ball inpainting results from various methods. The
    red circle indicates the inpainted region, and we show a zoomed-in
    view of the blue crop. Each row contains results from ten different random seeds.}
    \label{fig:aba_seed-cherry}
\end{figure*}

\begin{figure*}
    \centering

    \begin{tabu} to \textwidth {
        @{}
        l@{}
        l@{\hspace{0.5pt}}
    }
        \multicolumn{1}{l}{\rotatebox[origin=c]{90}{\shortstack[l]{\scriptsize Input \\ \scriptsize image}}} &
        \noindent\parbox[c]{0.5\textwidth}{\includegraphics[width=0.5\textwidth]{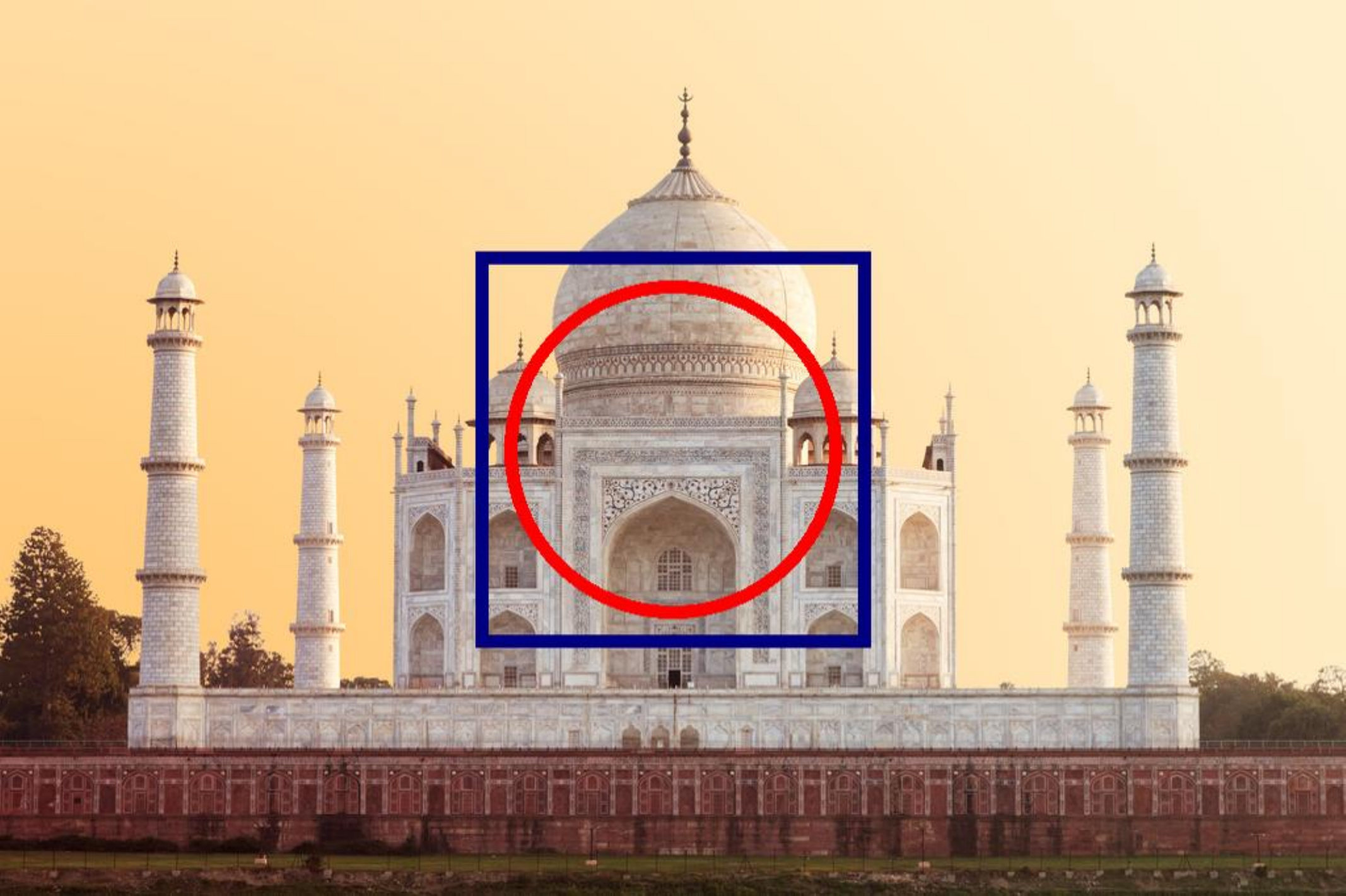}}  \\
    \end{tabu}
    
    \begin{tabu} to \textwidth {
        @{}
        c@{}
        c@{\hspace{0.5pt}}
        c@{\hspace{0.5pt}}
        c@{\hspace{0.5pt}}
        c@{\hspace{0.5pt}}
        c@{\hspace{0.5pt}}
        c@{\hspace{0.5pt}}
        c@{\hspace{0.5pt}}
        c@{\hspace{0.5pt}}
        c@{\hspace{0.5pt}}
        c@{\hspace{0.5pt}}
        c@{}
    }
        

        \multicolumn{1}{l}{\rotatebox[origin=c]{90}{\shortstack[l]{\scriptsize Blended Dif-\\ \scriptsize fusion \cite{avrahami2023blendedlatent, avrahami2022blendeddiffusion}}}} &
        \noindent\parbox[c]{0.081\textwidth}{\includegraphics[width=0.081\textwidth]{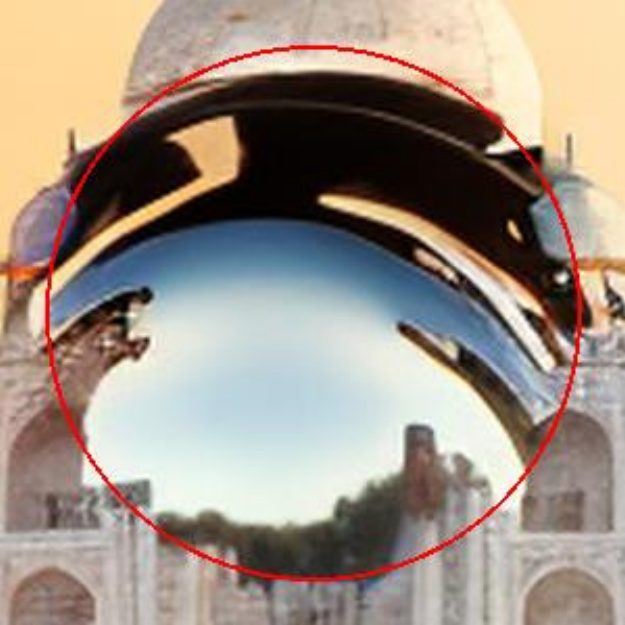}} & 
        \noindent\parbox[c]{0.081\textwidth}{\includegraphics[width=0.081\textwidth]{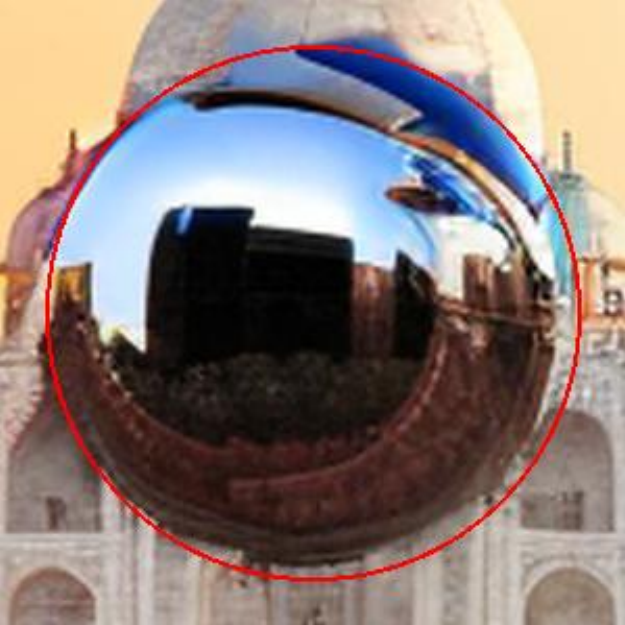}} &  
        \noindent\parbox[c]{0.081\textwidth}{\includegraphics[width=0.081\textwidth]{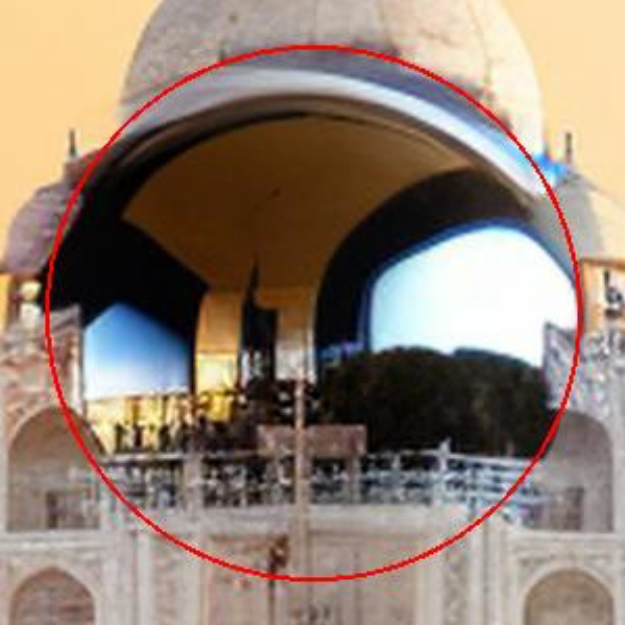}} & 
        \noindent\parbox[c]{0.081\textwidth}{\includegraphics[width=0.081\textwidth]{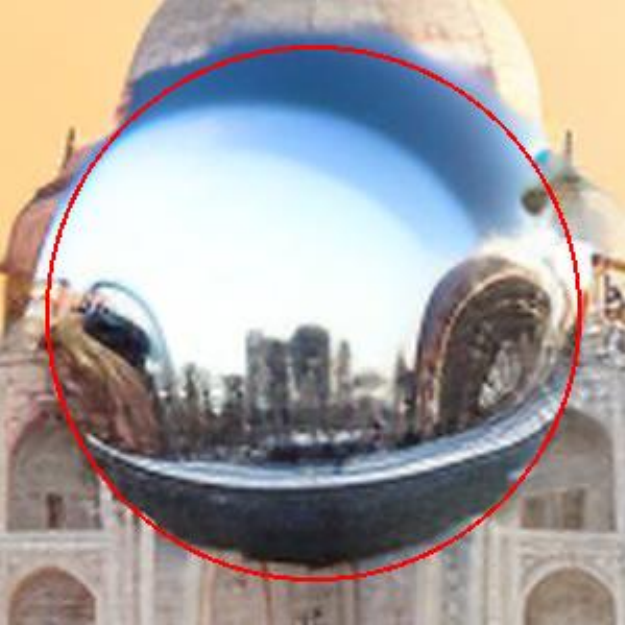}} & 
        \noindent\parbox[c]{0.081\textwidth}{\includegraphics[width=0.081\textwidth]{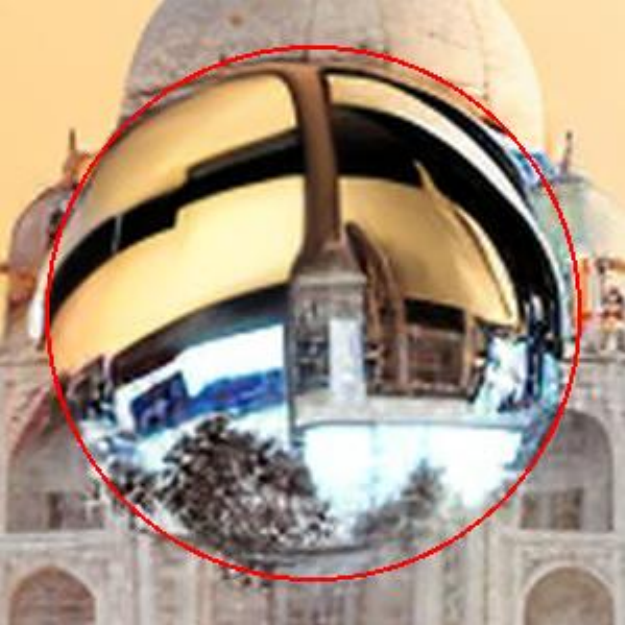}} & 
        \noindent\parbox[c]{0.081\textwidth}{\includegraphics[width=0.081\textwidth]{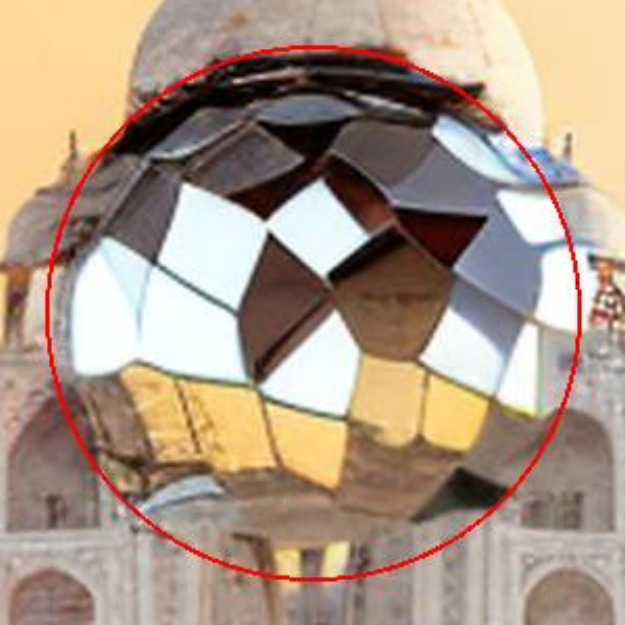}} & 
        \noindent\parbox[c]{0.081\textwidth}{\includegraphics[width=0.081\textwidth]{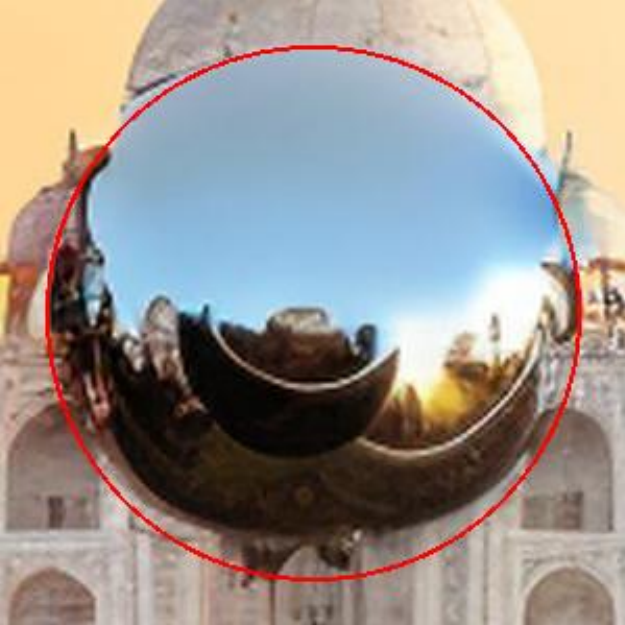}} & 
        \noindent\parbox[c]{0.081\textwidth}{\includegraphics[width=0.081\textwidth]{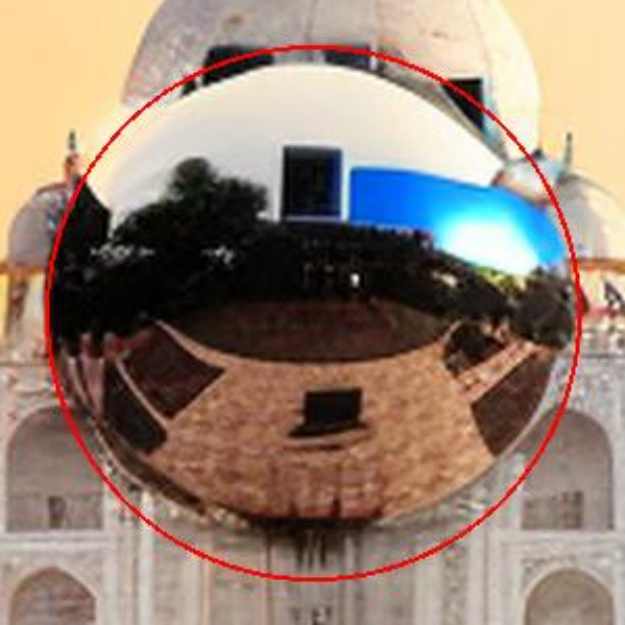}} & 
        \noindent\parbox[c]{0.081\textwidth}{\includegraphics[width=0.081\textwidth]{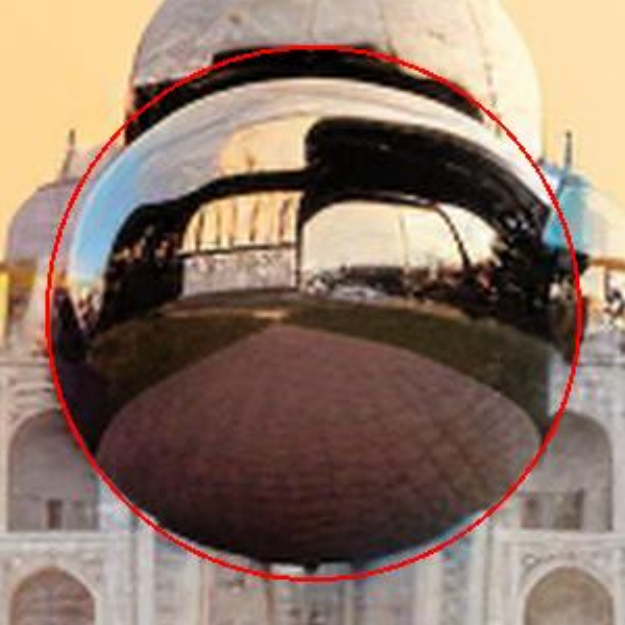}} & 
        \noindent\parbox[c]{0.081\textwidth}{\includegraphics[width=0.081\textwidth]{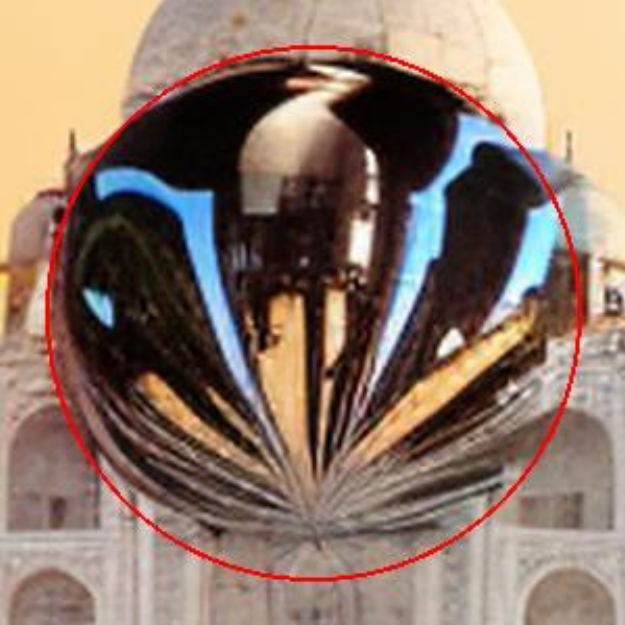}} & 
        
        \\

        \multicolumn{1}{l}{\rotatebox[origin=c]{90}{\shortstack[l]{\scriptsize Paint-by-Ex\\ \scriptsize ample \cite{yang2023paint}}}} &
        \noindent\parbox[c]{0.081\textwidth}{\includegraphics[width=0.081\textwidth]{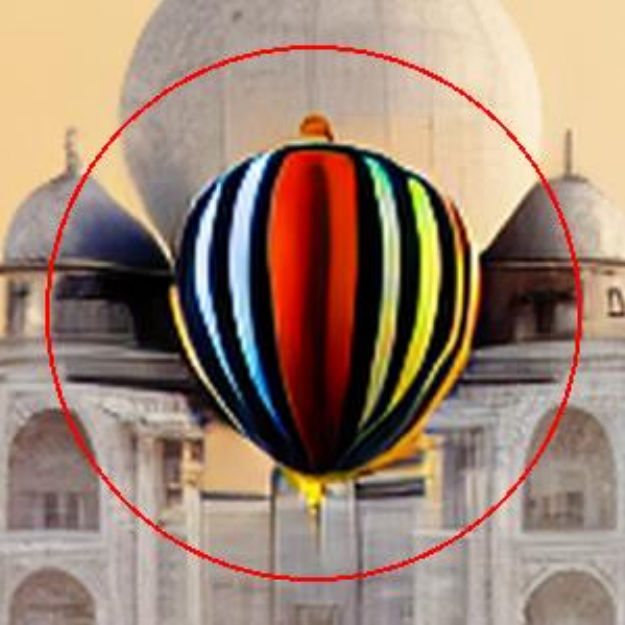}} & 
        \noindent\parbox[c]{0.081\textwidth}{\includegraphics[width=0.081\textwidth]{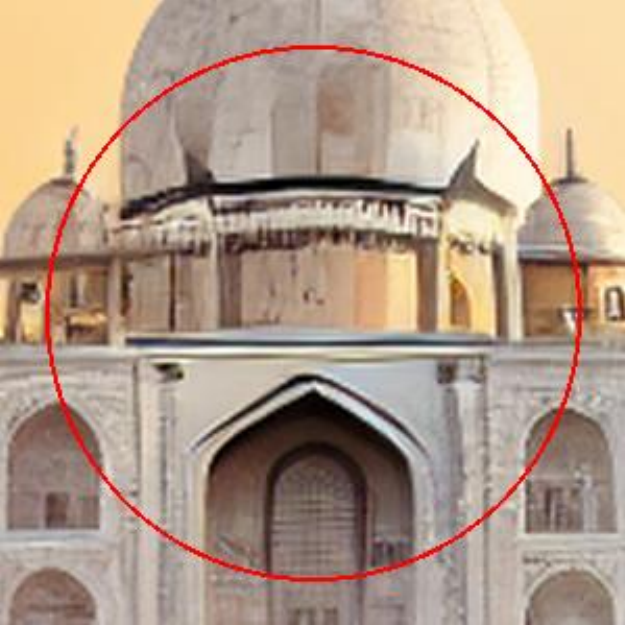}} &  
        \noindent\parbox[c]{0.081\textwidth}{\includegraphics[width=0.081\textwidth]{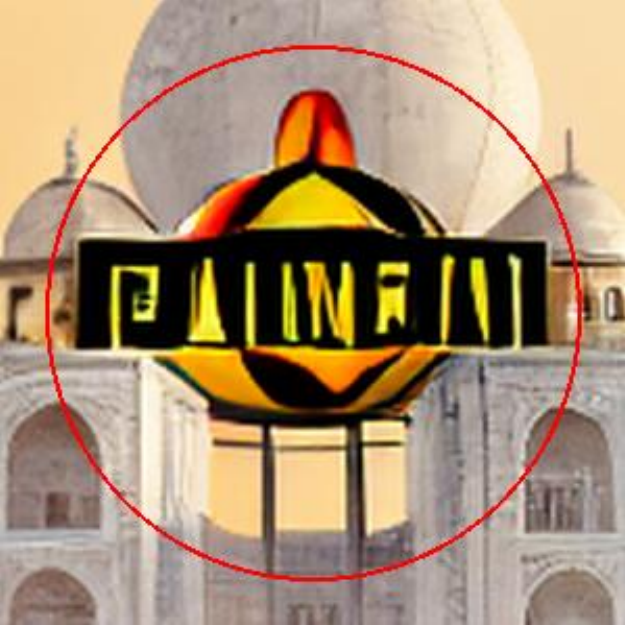}} & 
        \noindent\parbox[c]{0.081\textwidth}{\includegraphics[width=0.081\textwidth]{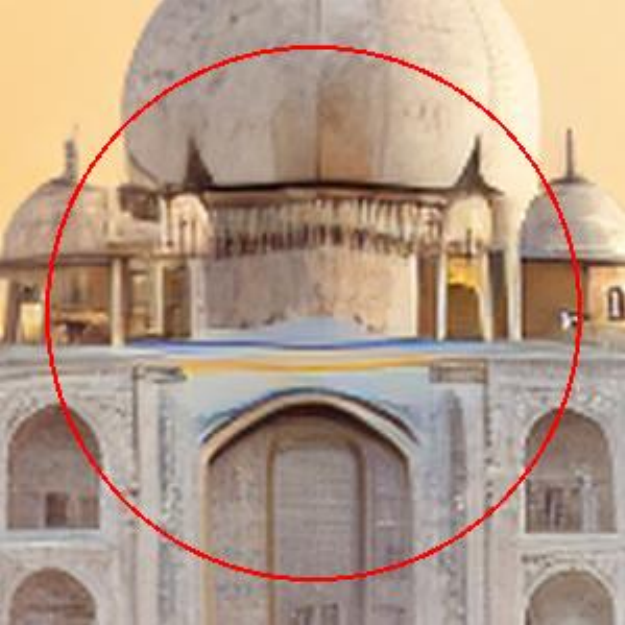}} & 
        \noindent\parbox[c]{0.081\textwidth}{\includegraphics[width=0.081\textwidth]{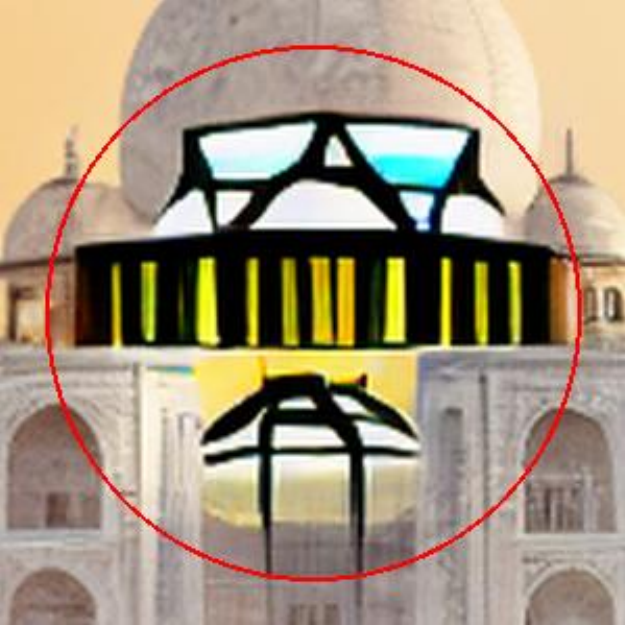}} & 
        \noindent\parbox[c]{0.081\textwidth}{\includegraphics[width=0.081\textwidth]{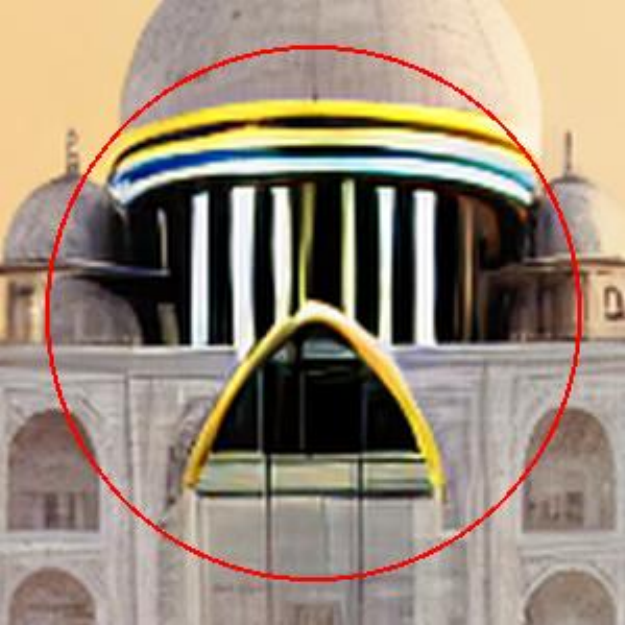}} & 
        \noindent\parbox[c]{0.081\textwidth}{\includegraphics[width=0.081\textwidth]{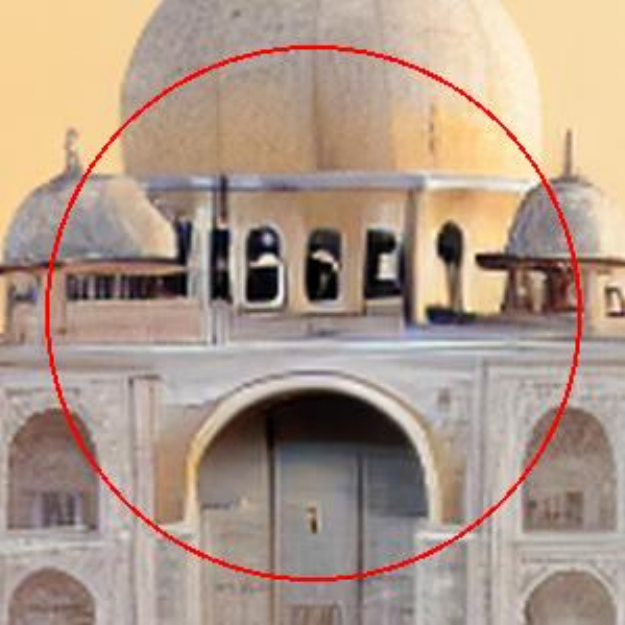}} & 
        \noindent\parbox[c]{0.081\textwidth}{\includegraphics[width=0.081\textwidth]{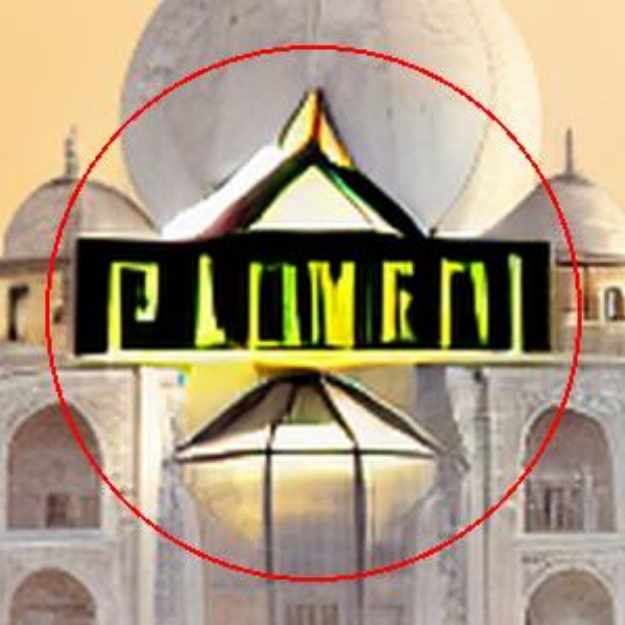}} & 
        \noindent\parbox[c]{0.081\textwidth}{\includegraphics[width=0.081\textwidth]{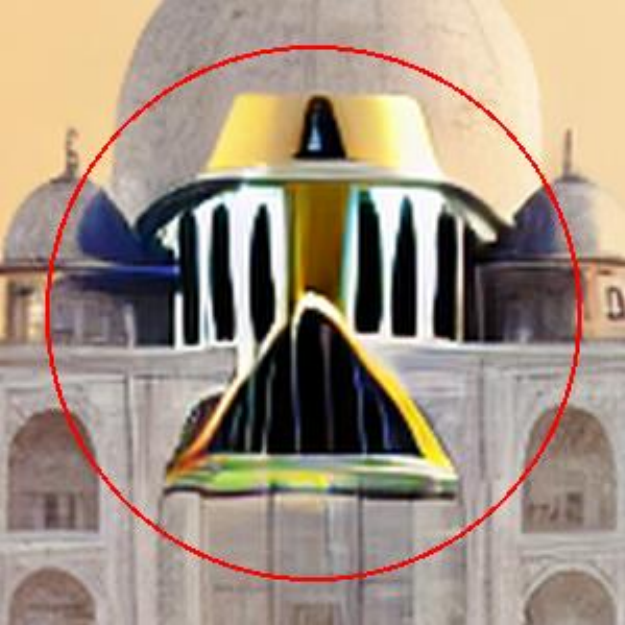}} & 
        \noindent\parbox[c]{0.081\textwidth}{\includegraphics[width=0.081\textwidth]{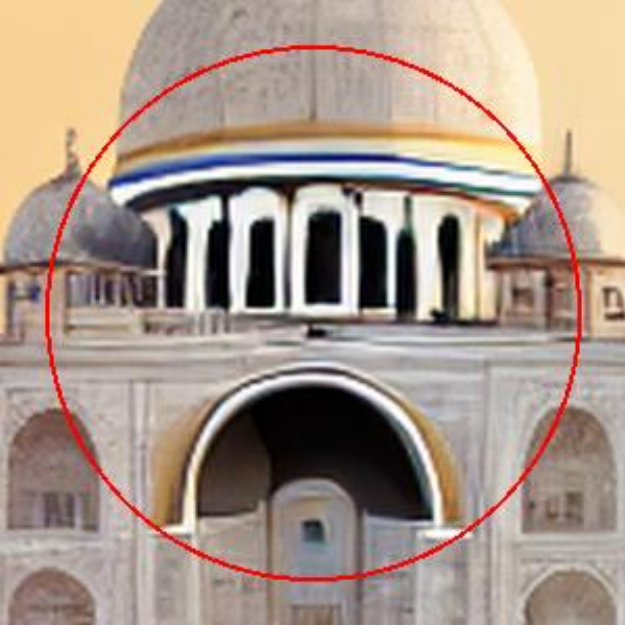}} & 
        
        \\

        \multicolumn{1}{l}{\rotatebox[origin=c]{90}{\shortstack[l]{\scriptsize IP-Adapter\\ \scriptsize \cite{ye2023ip-adapter}}}} &
        \noindent\parbox[c]{0.081\textwidth}{\includegraphics[width=0.081\textwidth]{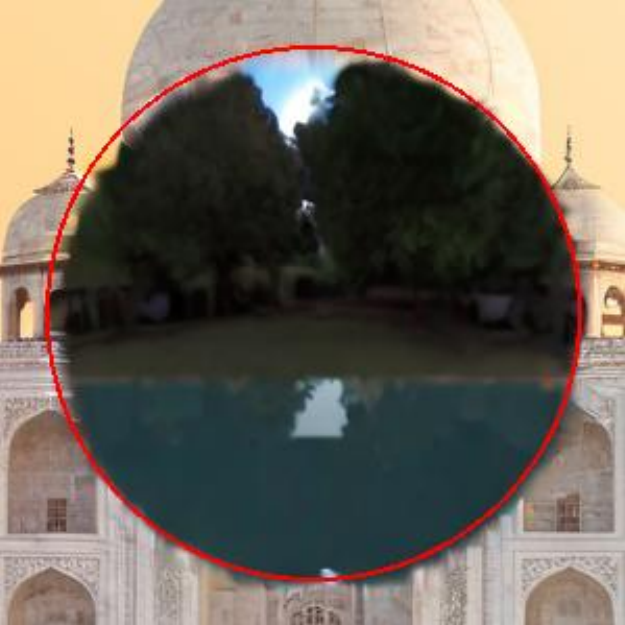}} & 
        \noindent\parbox[c]{0.081\textwidth}{\includegraphics[width=0.081\textwidth]{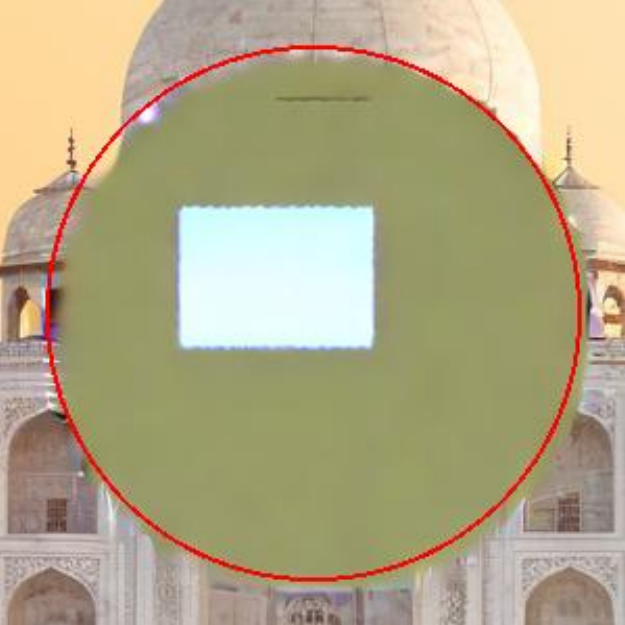}} &  
        \noindent\parbox[c]{0.081\textwidth}{\includegraphics[width=0.081\textwidth]{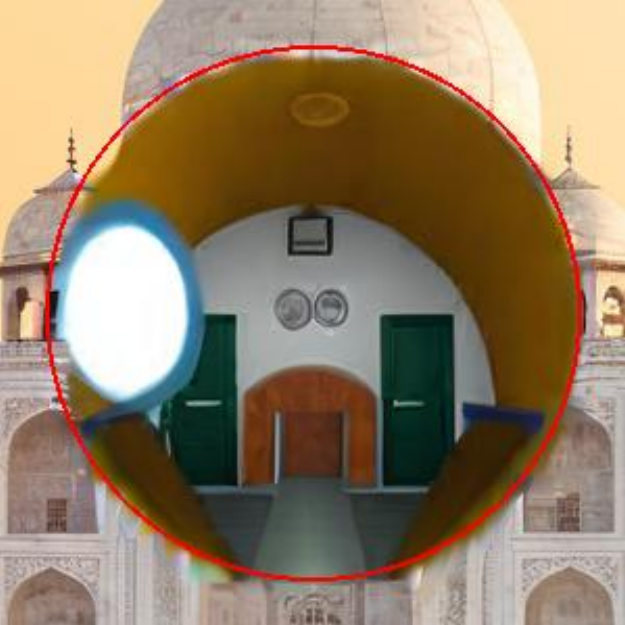}} & 
        \noindent\parbox[c]{0.081\textwidth}{\includegraphics[width=0.081\textwidth]{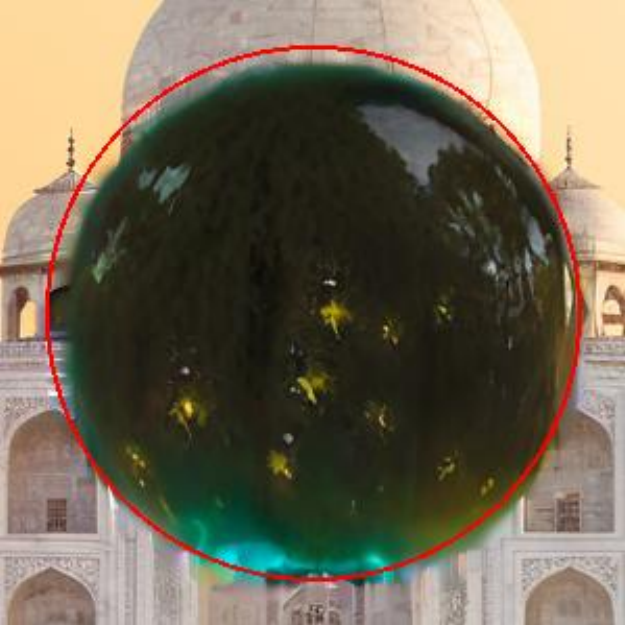}} & 
        \noindent\parbox[c]{0.081\textwidth}{\includegraphics[width=0.081\textwidth]{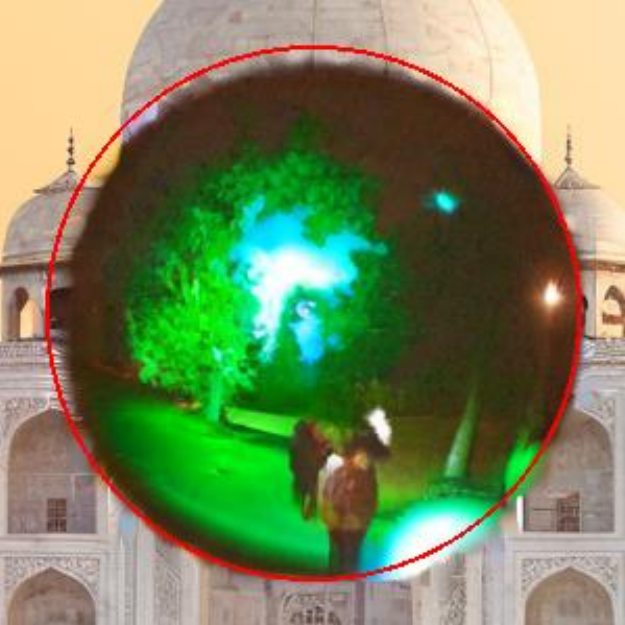}} & 
        \noindent\parbox[c]{0.081\textwidth}{\includegraphics[width=0.081\textwidth]{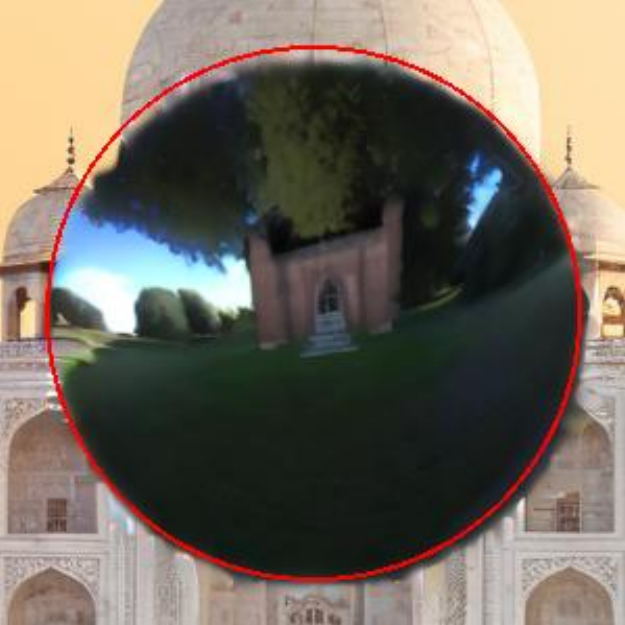}} & 
        \noindent\parbox[c]{0.081\textwidth}{\includegraphics[width=0.081\textwidth]{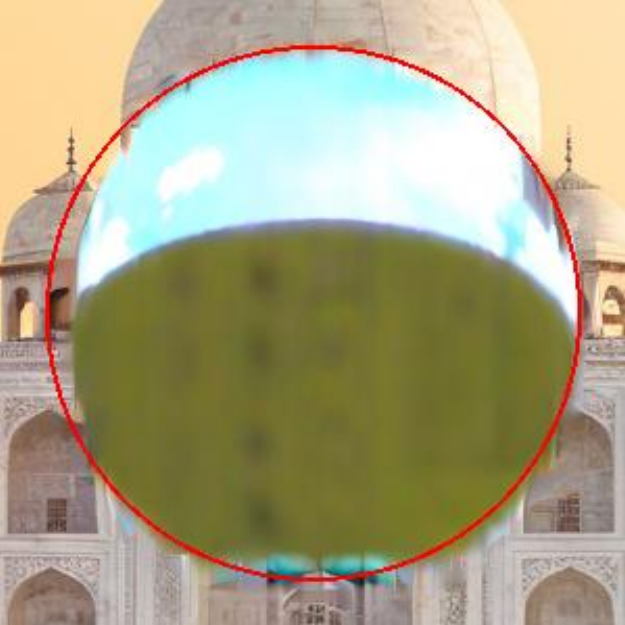}} & 
        \noindent\parbox[c]{0.081\textwidth}{\includegraphics[width=0.081\textwidth]{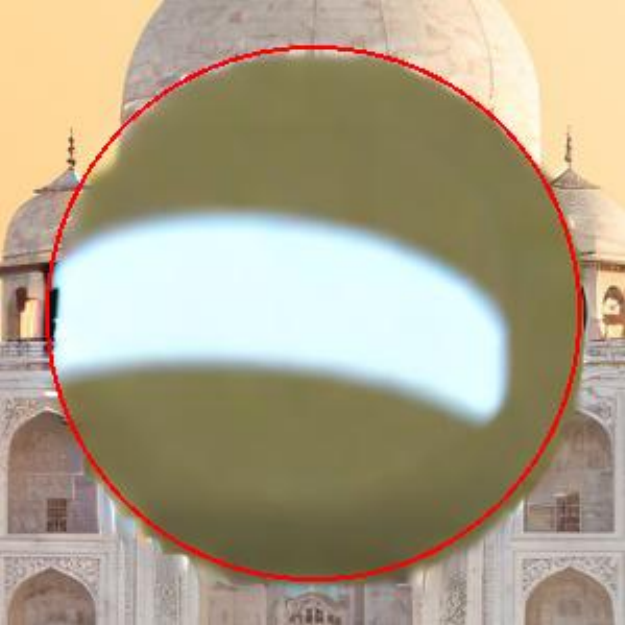}} & 
        \noindent\parbox[c]{0.081\textwidth}{\includegraphics[width=0.081\textwidth]{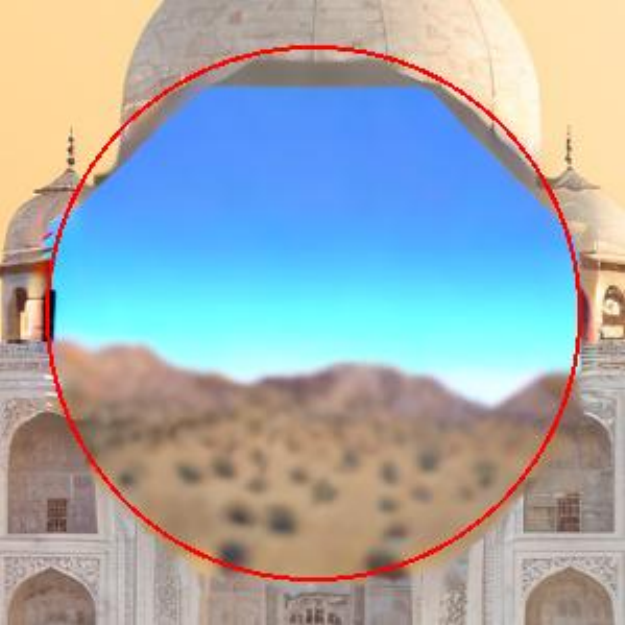}} & 
        \noindent\parbox[c]{0.081\textwidth}{\includegraphics[width=0.081\textwidth]{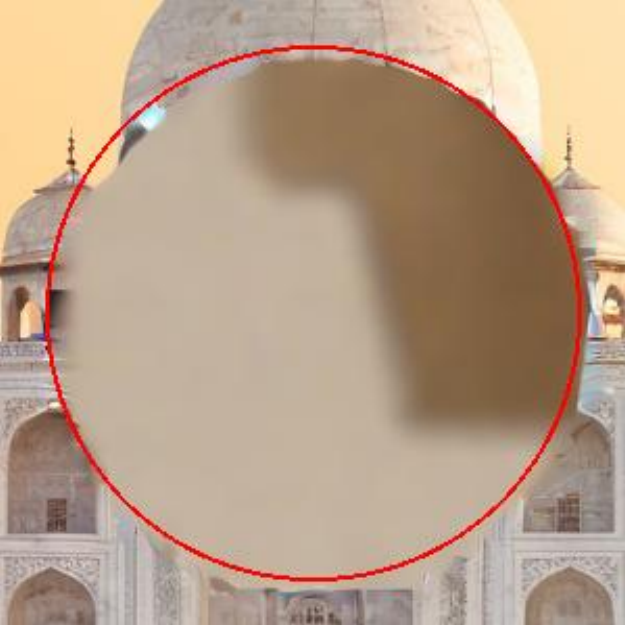}} & 
        
        \\

        \multicolumn{1}{l}{\rotatebox[origin=c]{90}{\shortstack[l]{\scriptsize DALL·E2 \cite{dalle2}}}} &
        \noindent\parbox[c]{0.081\textwidth}{\includegraphics[width=0.081\textwidth]{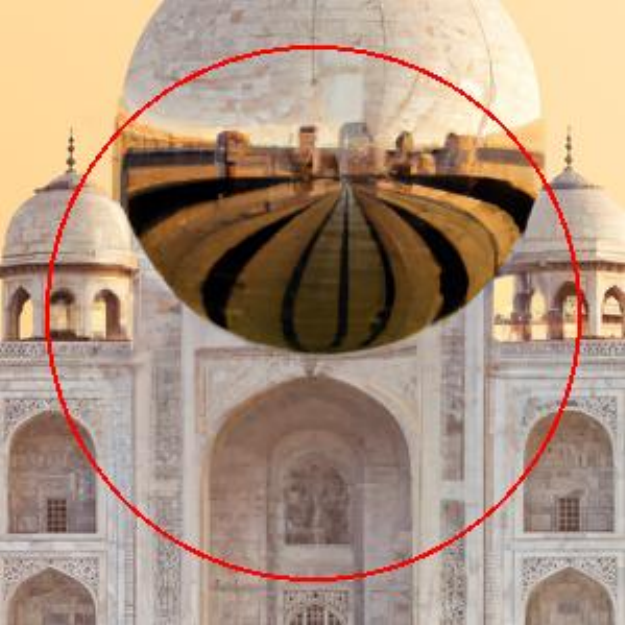}} & 
        \noindent\parbox[c]{0.081\textwidth}{\includegraphics[width=0.081\textwidth]{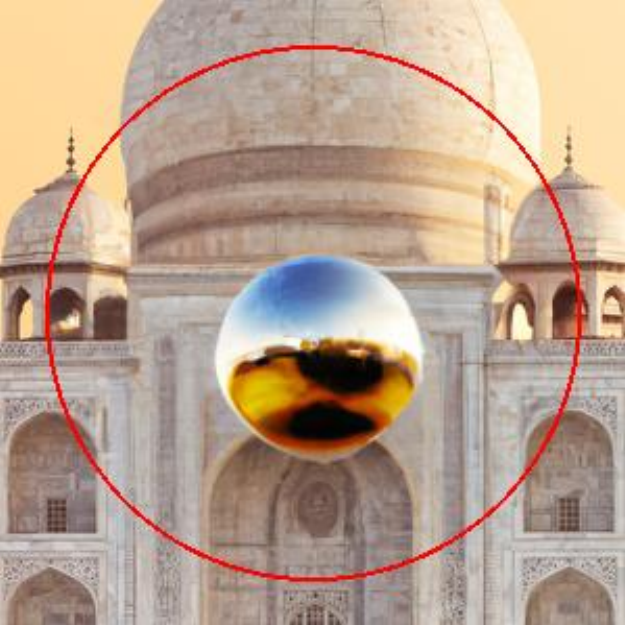}} &  
        \noindent\parbox[c]{0.081\textwidth}{\includegraphics[width=0.081\textwidth]{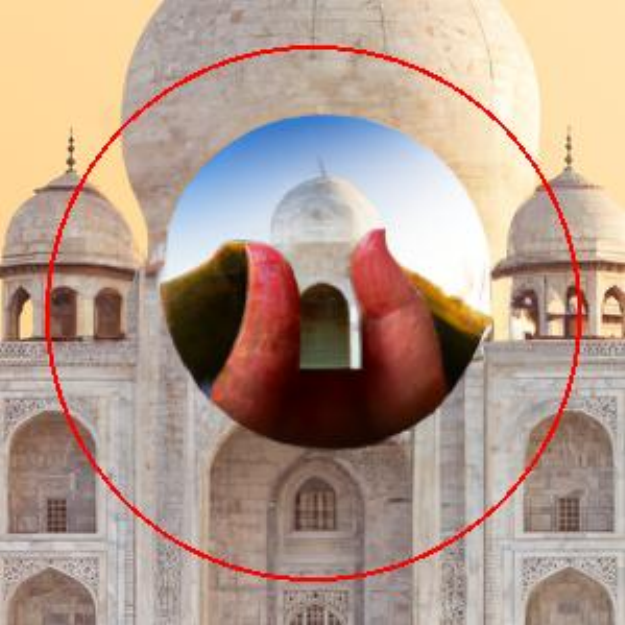}} & 
        \noindent\parbox[c]{0.081\textwidth}{\includegraphics[width=0.081\textwidth]{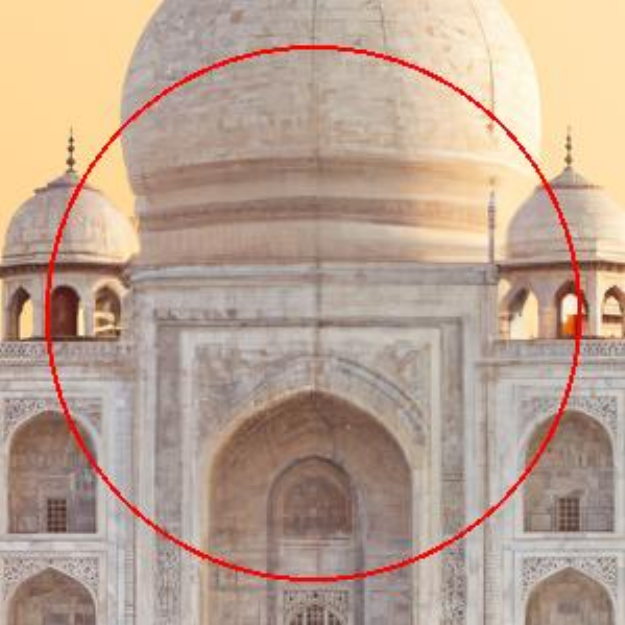}} & 
        \noindent\parbox[c]{0.081\textwidth}{\includegraphics[width=0.081\textwidth]{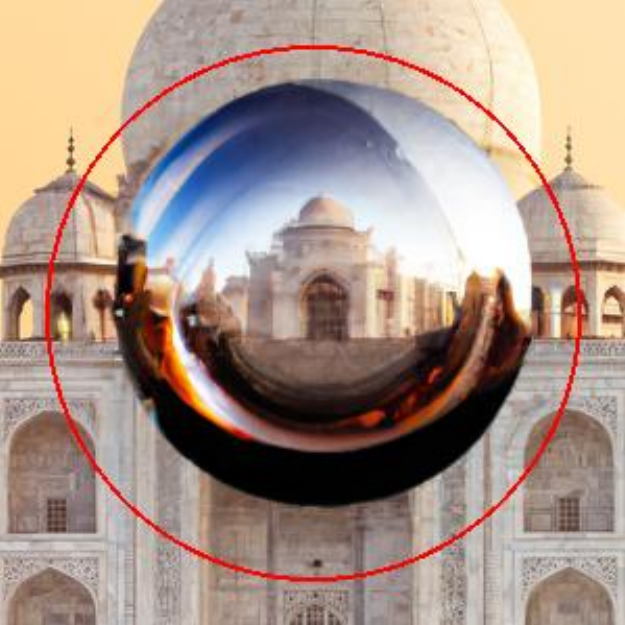}} & 
        \noindent\parbox[c]{0.081\textwidth}{\includegraphics[width=0.081\textwidth]{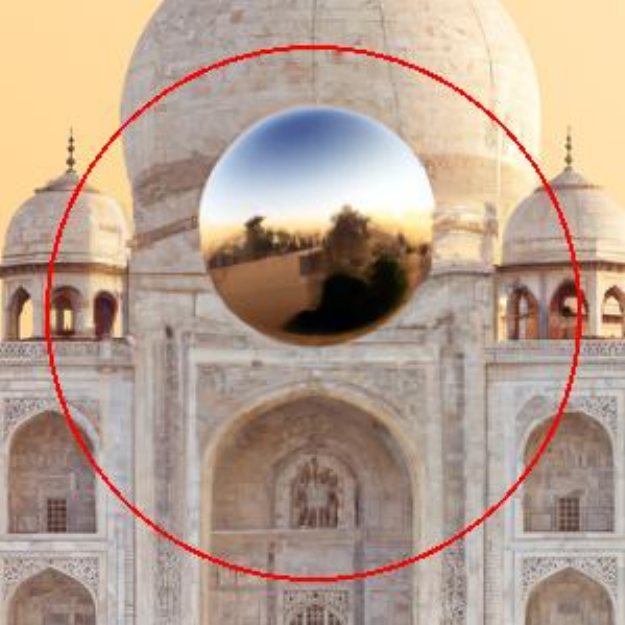}} & 
        \noindent\parbox[c]{0.081\textwidth}{\includegraphics[width=0.081\textwidth]{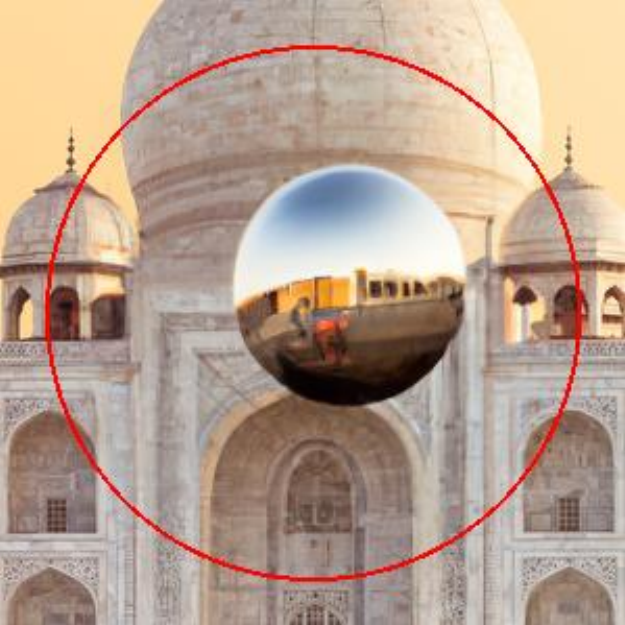}} & 
        \noindent\parbox[c]{0.081\textwidth}{\includegraphics[width=0.081\textwidth]{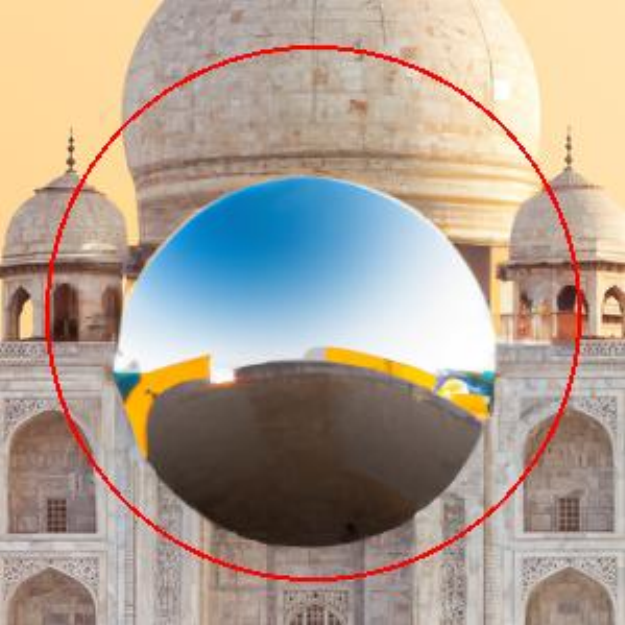}} & 
        \noindent\parbox[c]{0.081\textwidth}{\includegraphics[width=0.081\textwidth]{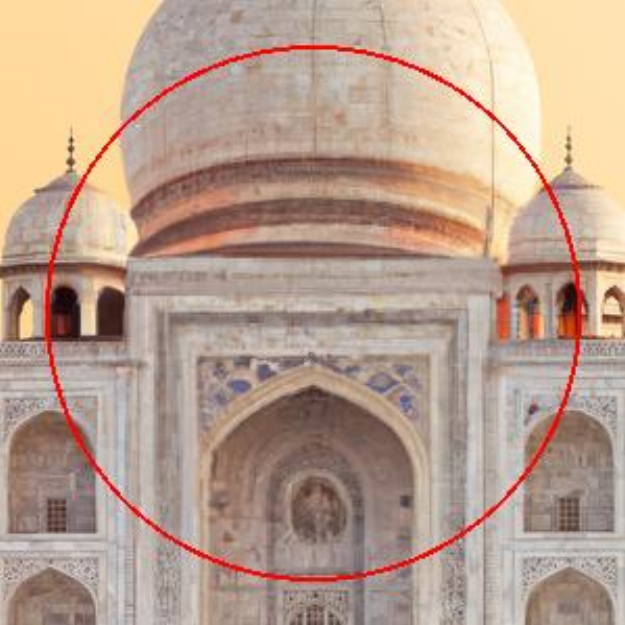}} & 
        \noindent\parbox[c]{0.081\textwidth}{\includegraphics[width=0.081\textwidth]{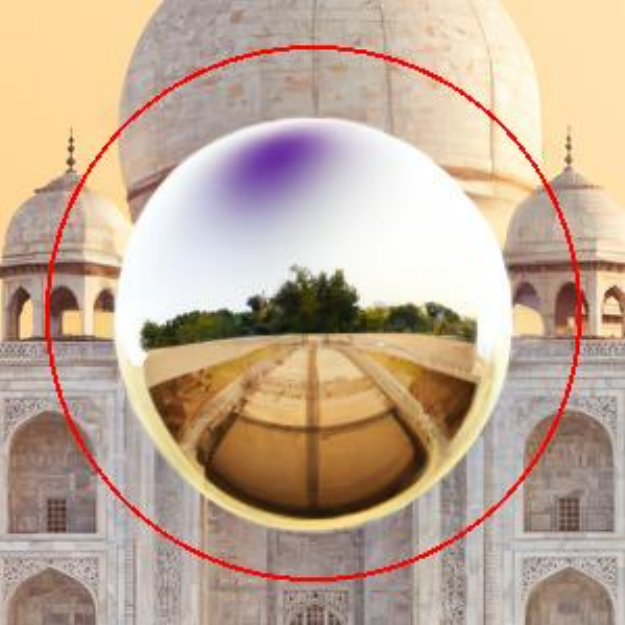}} & 
        
        \\

        \multicolumn{1}{l}{\rotatebox[origin=c]{90}{\shortstack[l]{\scriptsize Adobe \\ \scriptsize Firefly \cite{adobefirefly}}}} &
        \noindent\parbox[c]{0.081\textwidth}{\includegraphics[width=0.081\textwidth]{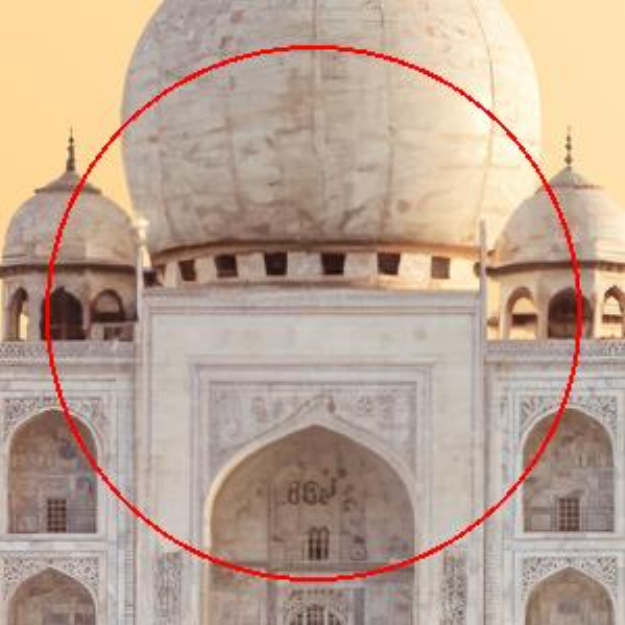}} & 
        \noindent\parbox[c]{0.081\textwidth}{\includegraphics[width=0.081\textwidth]{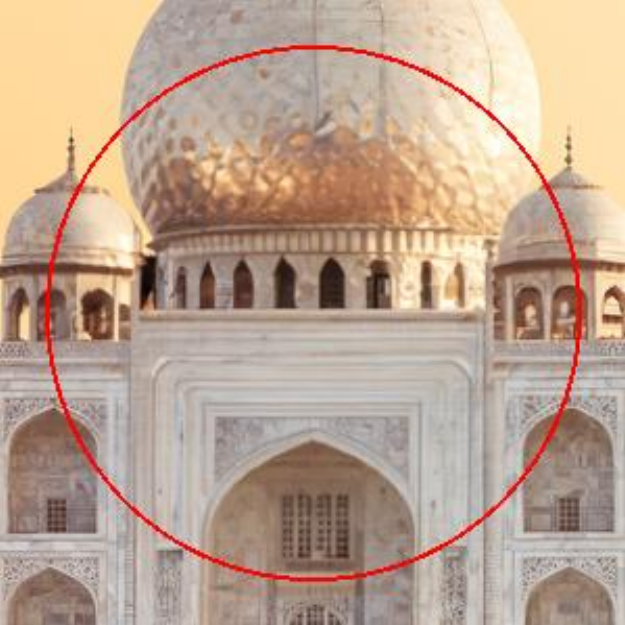}} &  
        \noindent\parbox[c]{0.081\textwidth}{\includegraphics[width=0.081\textwidth]{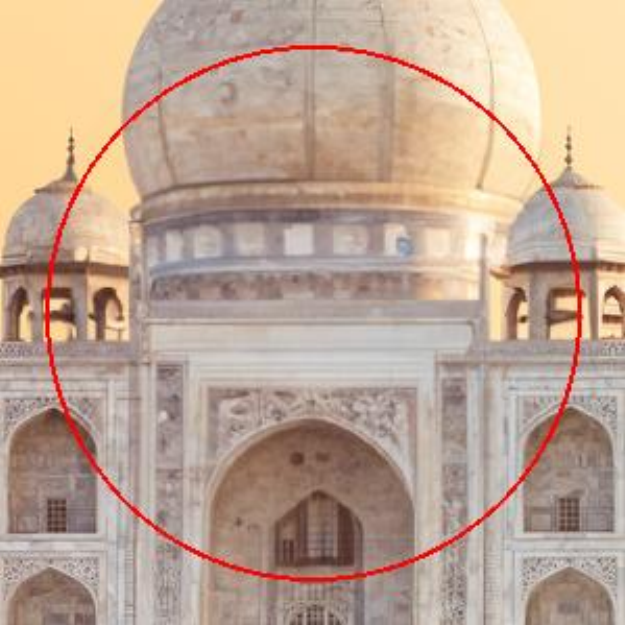}} & 
        \noindent\parbox[c]{0.081\textwidth}{\includegraphics[width=0.081\textwidth]{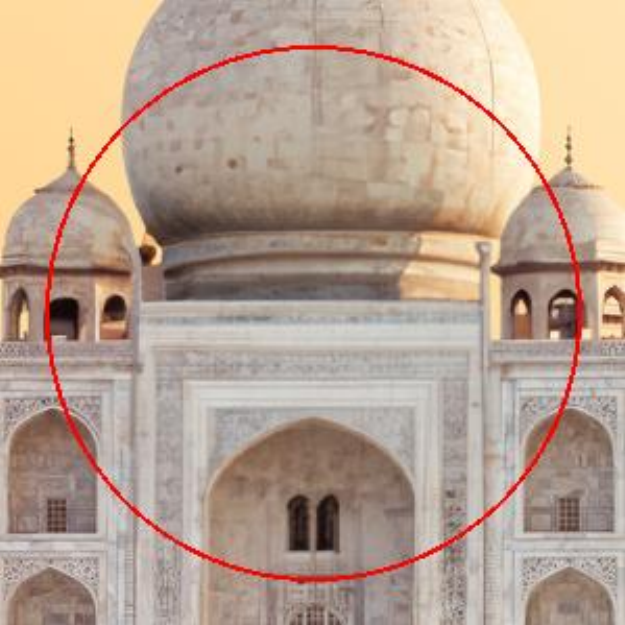}} & 
        \noindent\parbox[c]{0.081\textwidth}{\includegraphics[width=0.081\textwidth]{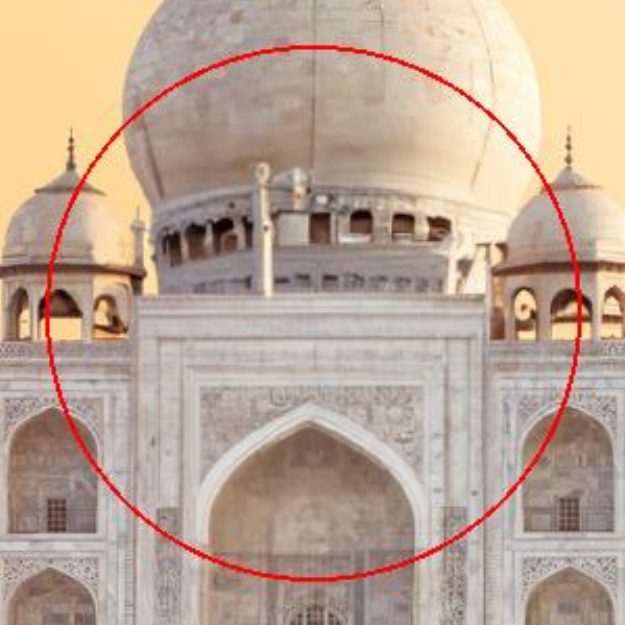}} & 
        \noindent\parbox[c]{0.081\textwidth}{\includegraphics[width=0.081\textwidth]{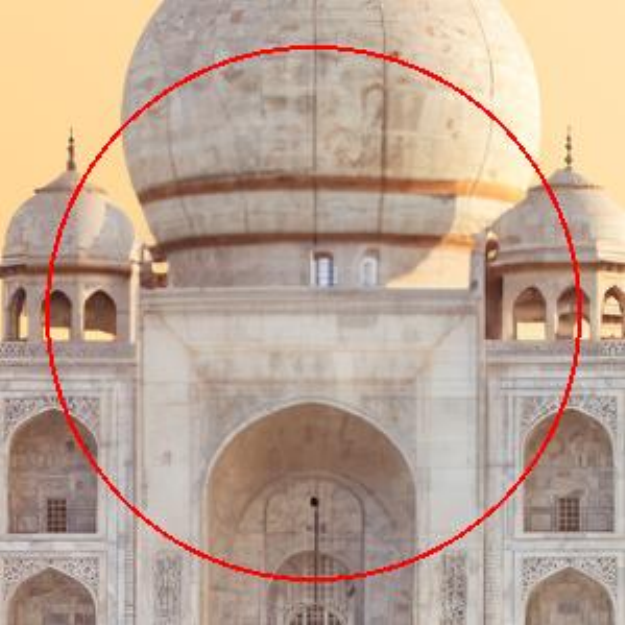}} & 
        \noindent\parbox[c]{0.081\textwidth}{\includegraphics[width=0.081\textwidth]{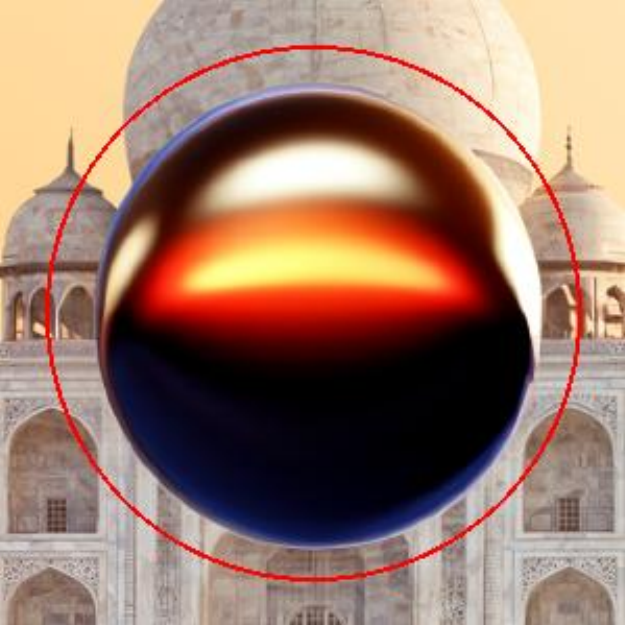}} & 
        \noindent\parbox[c]{0.081\textwidth}{\includegraphics[width=0.081\textwidth]{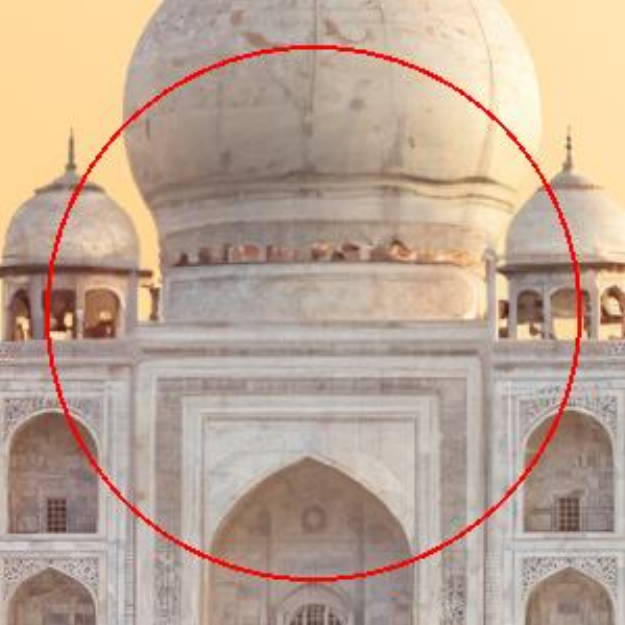}} & 
        \noindent\parbox[c]{0.081\textwidth}{\includegraphics[width=0.081\textwidth]{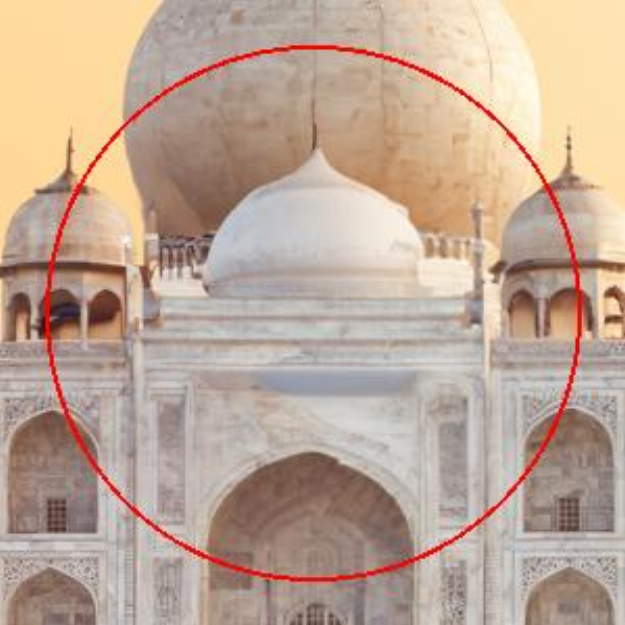}} & 
        \noindent\parbox[c]{0.081\textwidth}{\includegraphics[width=0.081\textwidth]{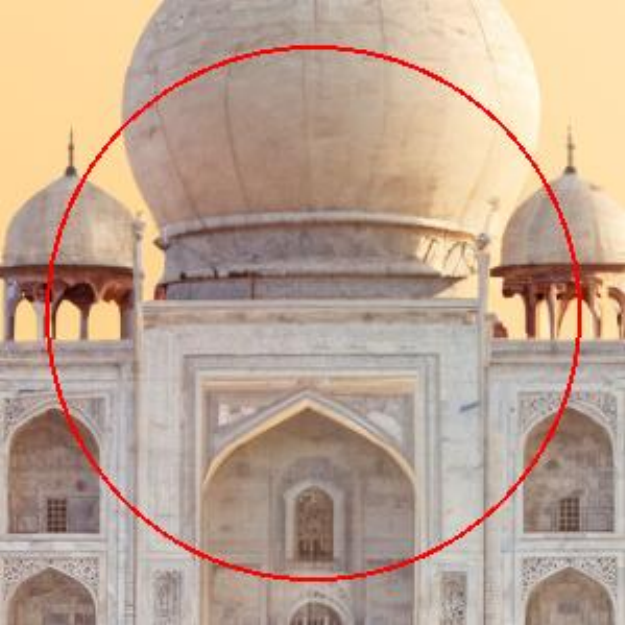}} & 

        \\

        \multicolumn{1}{l}{\rotatebox[origin=c]{90}{\shortstack[l]{\scriptsize SDXL \cite{podell2023sdxl}}}} &
        \noindent\parbox[c]{0.081\textwidth}{\includegraphics[width=0.081\textwidth]{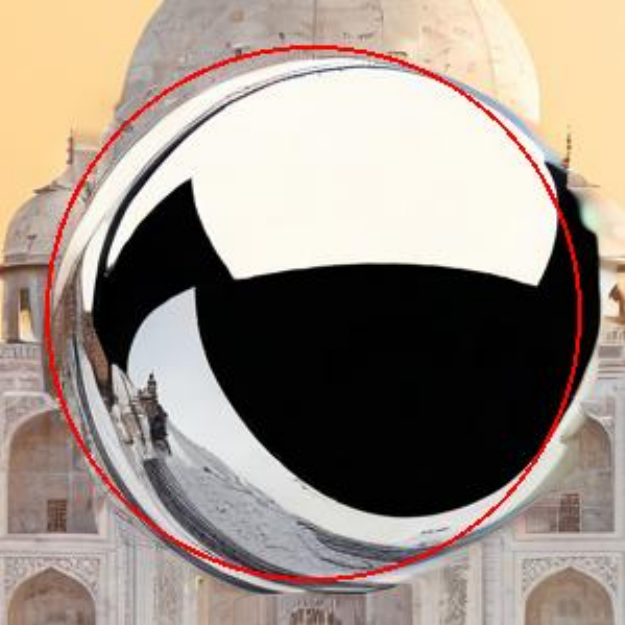}} & 
        \noindent\parbox[c]{0.081\textwidth}{\includegraphics[width=0.081\textwidth]{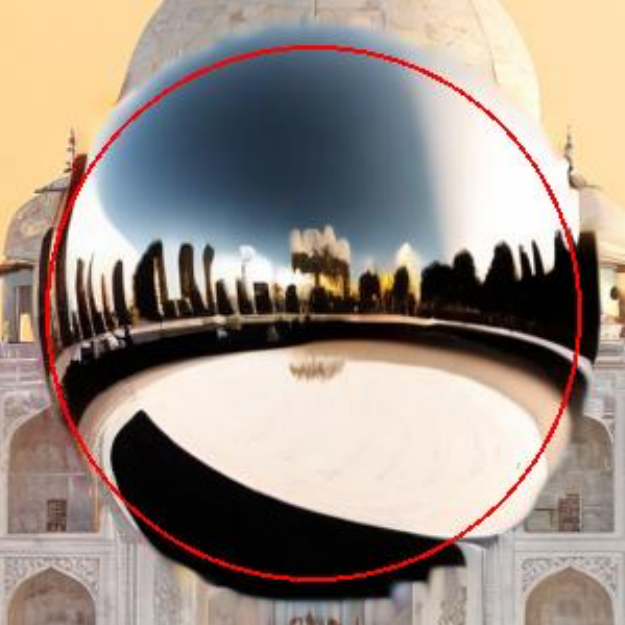}} &  
        \noindent\parbox[c]{0.081\textwidth}{\includegraphics[width=0.081\textwidth]{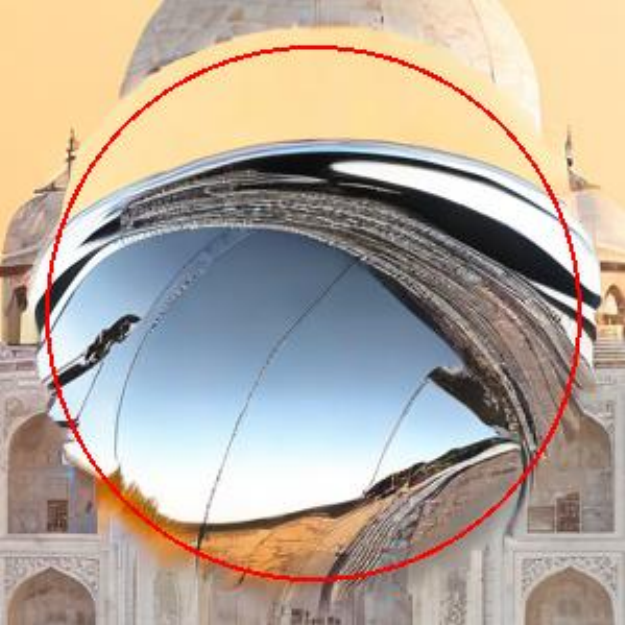}} & 
        \noindent\parbox[c]{0.081\textwidth}{\includegraphics[width=0.081\textwidth]{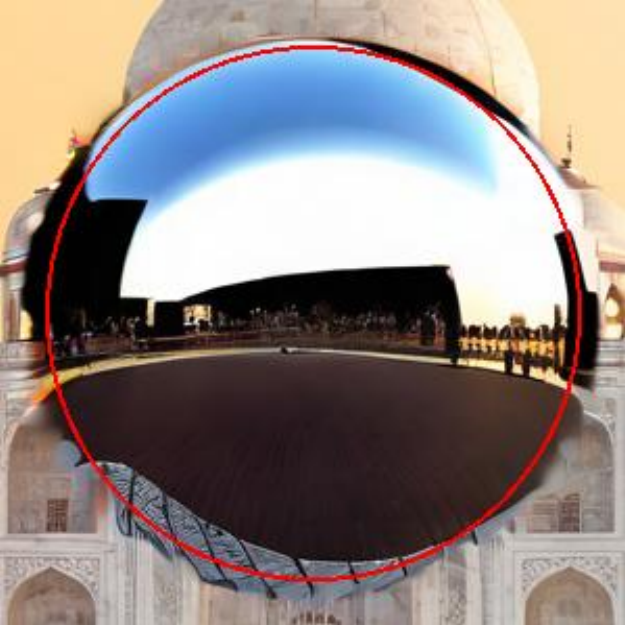}} & 
        \noindent\parbox[c]{0.081\textwidth}{\includegraphics[width=0.081\textwidth]{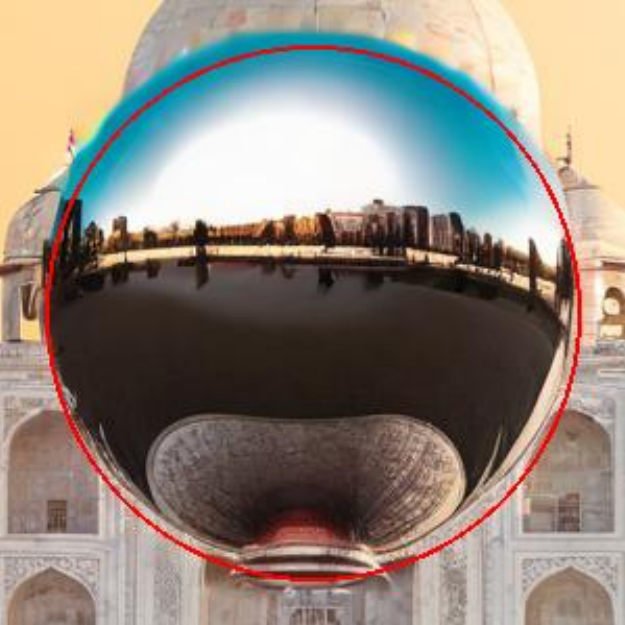}} & 
        \noindent\parbox[c]{0.081\textwidth}{\includegraphics[width=0.081\textwidth]{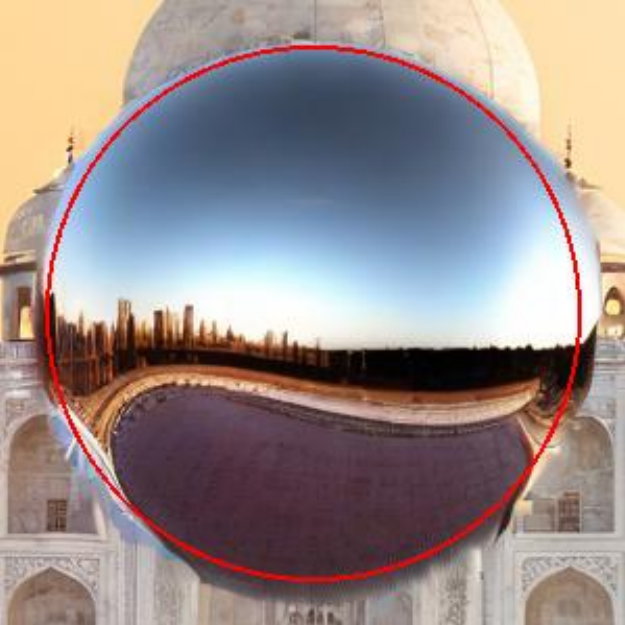}} & 
        \noindent\parbox[c]{0.081\textwidth}{\includegraphics[width=0.081\textwidth]{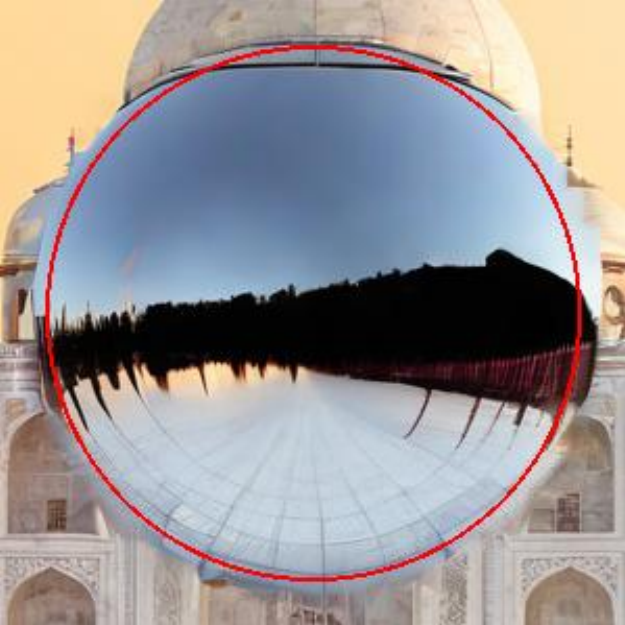}} & 
        \noindent\parbox[c]{0.081\textwidth}{\includegraphics[width=0.081\textwidth]{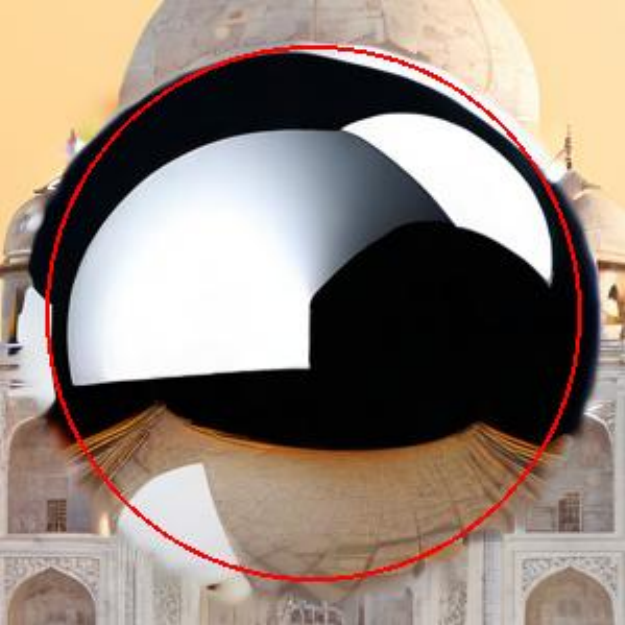}} & 
        \noindent\parbox[c]{0.081\textwidth}{\includegraphics[width=0.081\textwidth]{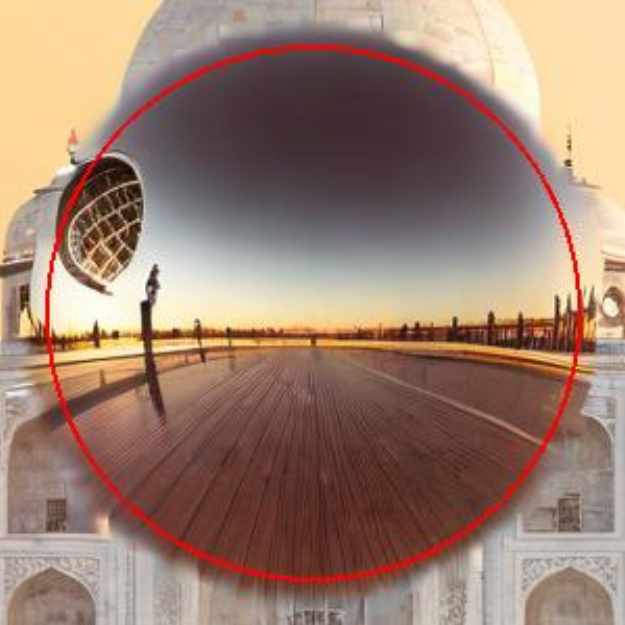}} & 
        \noindent\parbox[c]{0.081\textwidth}{\includegraphics[width=0.081\textwidth]{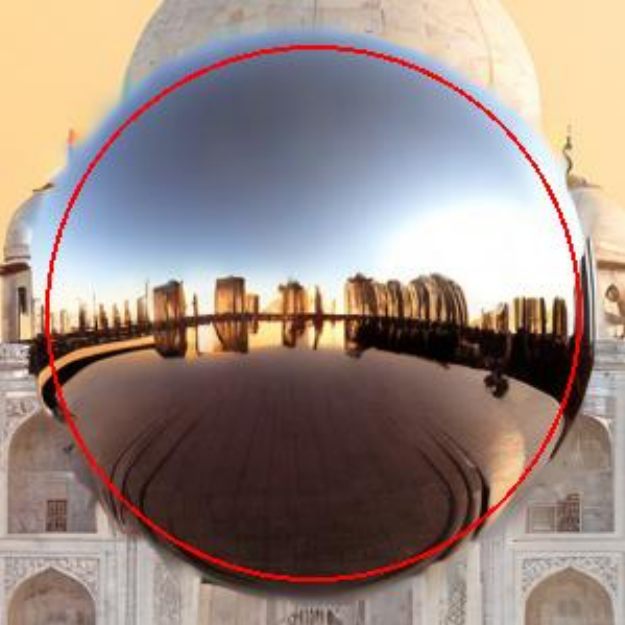}} & 
        
        \\ \hline

        \multicolumn{1}{l}{\rotatebox[origin=c]{90}{\shortstack[l]{\scriptsize \textbf{Ours}}}} &
        \noindent\parbox[c]{0.081\textwidth}{\includegraphics[width=0.081\textwidth]{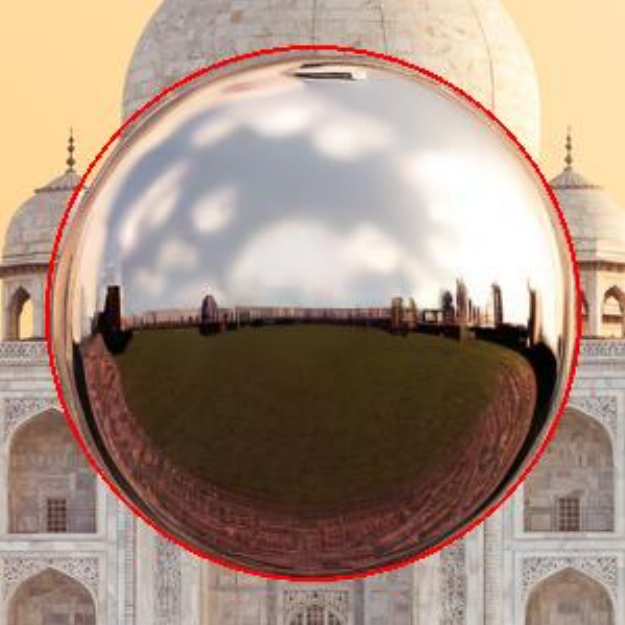}} & 
        \noindent\parbox[c]{0.081\textwidth}{\includegraphics[width=0.081\textwidth]{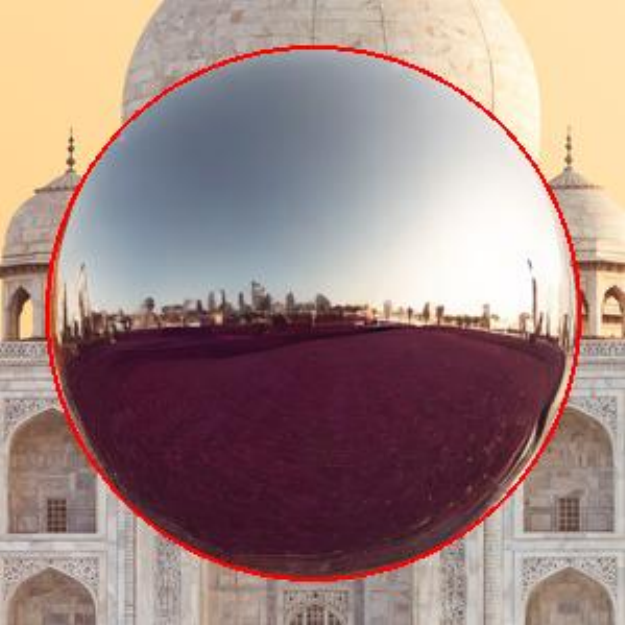}} &  
        \noindent\parbox[c]{0.081\textwidth}{\includegraphics[width=0.081\textwidth]{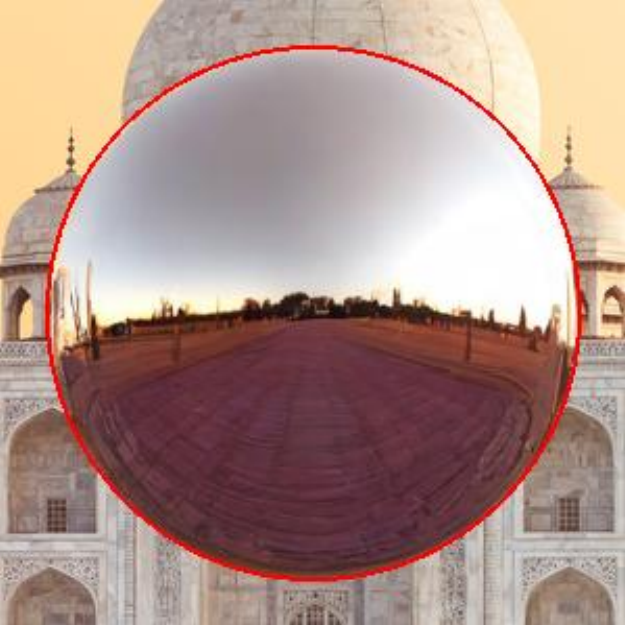}} & 
        \noindent\parbox[c]{0.081\textwidth}{\includegraphics[width=0.081\textwidth]{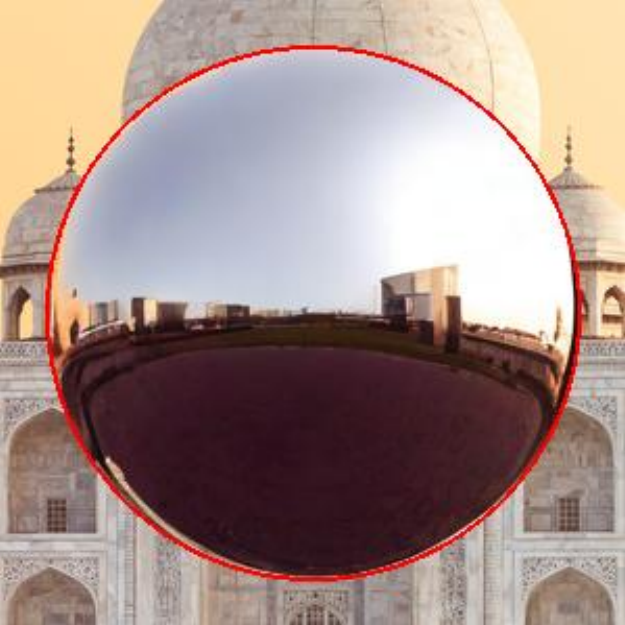}} & 
        \noindent\parbox[c]{0.081\textwidth}{\includegraphics[width=0.081\textwidth]{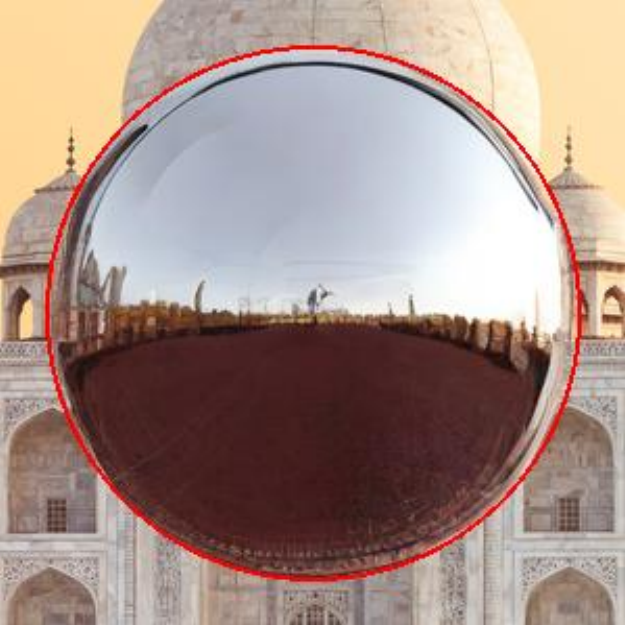}} & 
        \noindent\parbox[c]{0.081\textwidth}{\includegraphics[width=0.081\textwidth]{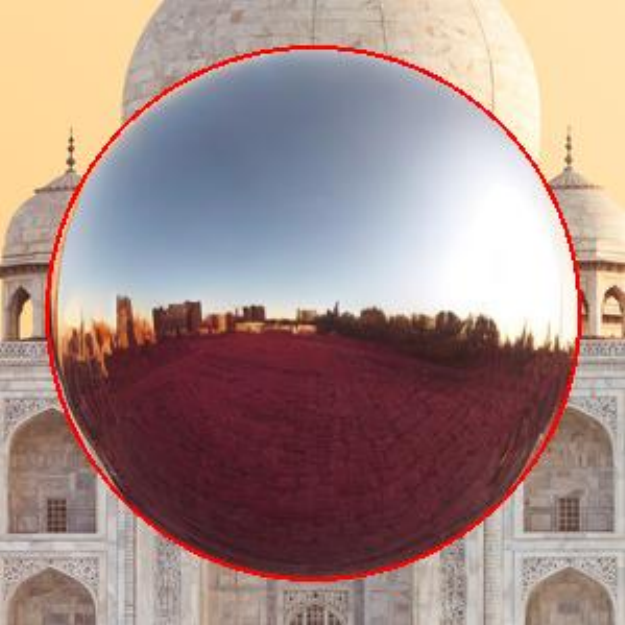}} & 
        \noindent\parbox[c]{0.081\textwidth}{\includegraphics[width=0.081\textwidth]{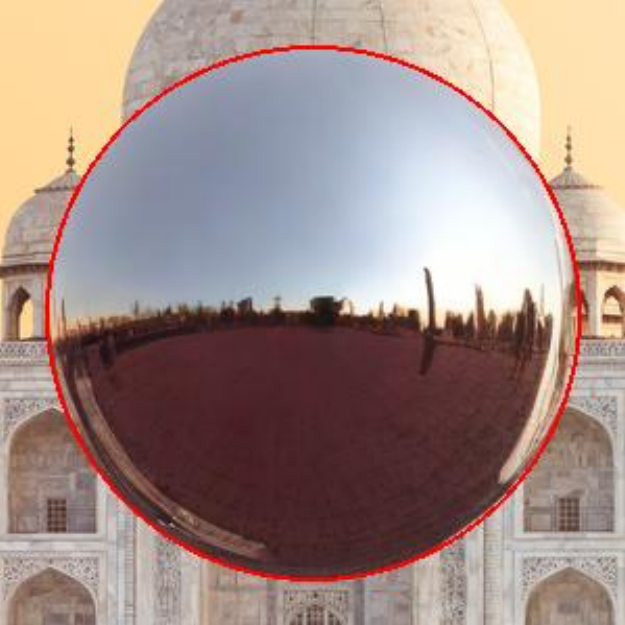}} & 
        \noindent\parbox[c]{0.081\textwidth}{\includegraphics[width=0.081\textwidth]{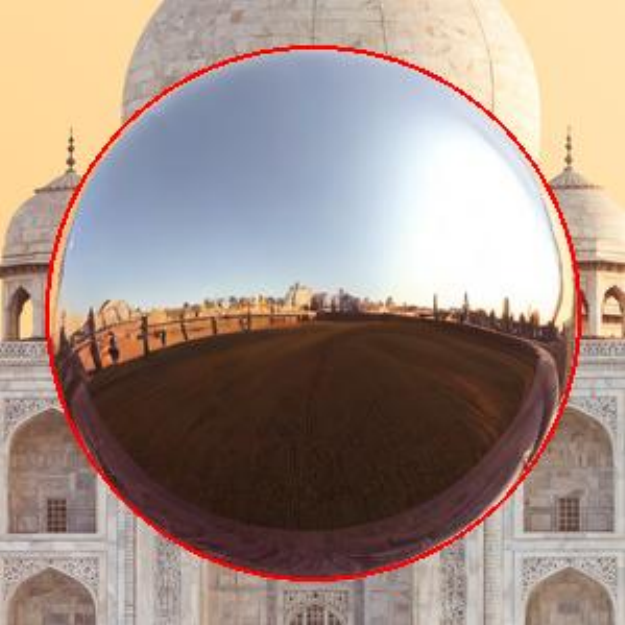}} & 
        \noindent\parbox[c]{0.081\textwidth}{\includegraphics[width=0.081\textwidth]{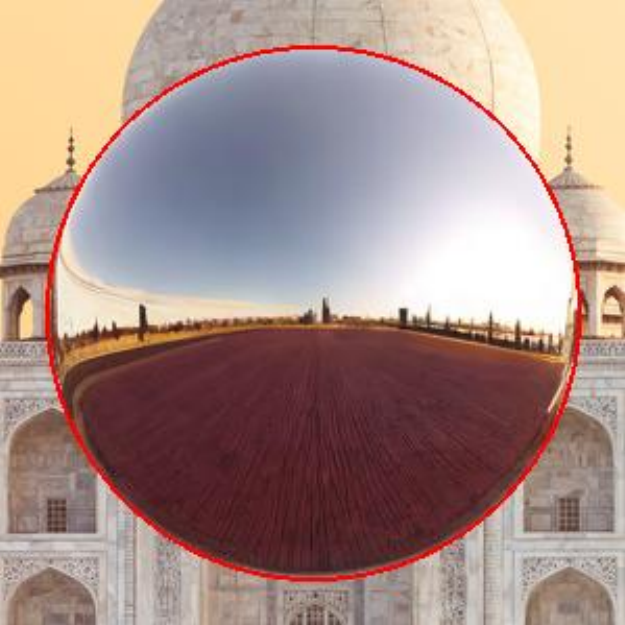}} & 
        \noindent\parbox[c]{0.081\textwidth}{\includegraphics[width=0.081\textwidth]{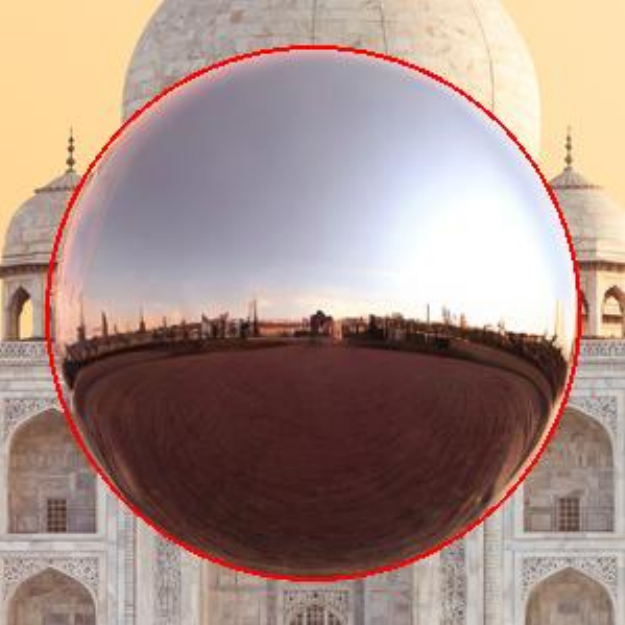}} & 
        
        \\
        
    \end{tabu}

    \caption{Chrome ball inpainting results from various methods. The
    red circle indicates the inpainted region, and we show a zoomed-in
    view of the blue crop. Each row contains results from ten different random seeds.}
    \label{fig:aba_seed-cherry2}
\end{figure*}

We show inpainting results of our method and other baselines using different random seeds in Figure \ref{fig:aba_seed-cherry} and Figure \ref{fig:aba_seed-cherry2}.
What we observed in general was that Blend Diffusion \cite{avrahami2022blendeddiffusion, avrahami2023blendedlatent} produced distorted balls. Paint-by-Example \cite{yang2023paint} failed to reproduce mirrored chrome balls altogether. IP-Adapter \cite{ye2023ip-adapter} replicated textures and details of the example chrome balls, making it unsuitable for light estimation. DALL·E2 \cite{dalle2} often simply reconstructed most of the masked-out content. Adobe Firefly \cite{adobefirefly} had a similar problem with DALL·E2 \cite{dalle2}, albeit more severe (see Figure \ref{fig:aba_seed-cherry2}). Moreover, none of these techniques precisely followed the inpainting mask. Our proposed method can address all these issues and consistently inpaint high-quality chrome balls.





\section{Additional Qualitative Results} \label{appendix:more_result_all}

\subsection{Benchmark datasets} \label{appendix:more_result_benchmark}

This section provides qualitative results for the experiments in Section~\ref{sec:experiment} in the main paper.

\textbf{Evaluation on three spheres.} We show spheres with three material types---mirror, matte, and diffuse---rendered using the inferred environment maps from the following methods: 


1. The ground truth

2. StyleLight \cite{wang2022stylelight}

3. Stable Diffusion XL \cite{podell2023sdxl} with depth-conditioned ControlNet \cite{zhang2023adding} (SDXL$^\dagger$)

4. SDXL$^\dagger$ with our LoRA (SDXL$^\dagger$+LR)

5. SDXL$^\dagger$ with our iterative inpainting (SDXL$^\dagger$+I)

6. SDXL$^\dagger$ with our LoRA and iterative inpainting (SDXL$^\dagger$+LR+I)

\begin{figure*}[!t]
    \centering

        \begin{tabu} to \textwidth {
        @{}
        c@{}
        c@{}
        c@{}
        c@{}
        c@{}
        c@{}
        c@{}
        c@{}
        c@{}
    }

        \multicolumn{1}{c}{\shortstack{\scriptsize Ground truth map}}
        & 
        \multicolumn{1}{c}{\shortstack{\hspace{-6pt} \scriptsize Input}}
        &
        \multicolumn{1}{c}{\shortstack{\scriptsize Ground truth}}
        & 
        \multicolumn{1}{c}{\shortstack{\scriptsize StyleLight \cite{wang2022stylelight}}}
        & 
        \multicolumn{1}{c}{\shortstack{\scriptsize SDXL$^\dagger$}} &
        \multicolumn{1}{c}{\shortstack{\scriptsize \begin{tabular}[c]{@{}c@{}}SDXL$^\dagger$+LR \\ (ours, ablated)\end{tabular}}} &
        \multicolumn{1}{c}{\shortstack{\scriptsize \begin{tabular}[c]{@{}c@{}}SDXL$^\dagger$+I \\ (ours,ablated)\end{tabular}}}
        &
        \multicolumn{1}{c}{\shortstack{\scriptsize \begin{tabular}[c]{@{}c@{}}SDXL$^\dagger$+LR+I \\ (ours)\end{tabular}}} 
        \\

        \noindent\parbox[c]{0.205\textwidth}{\includegraphics[height=0.100\textwidth]{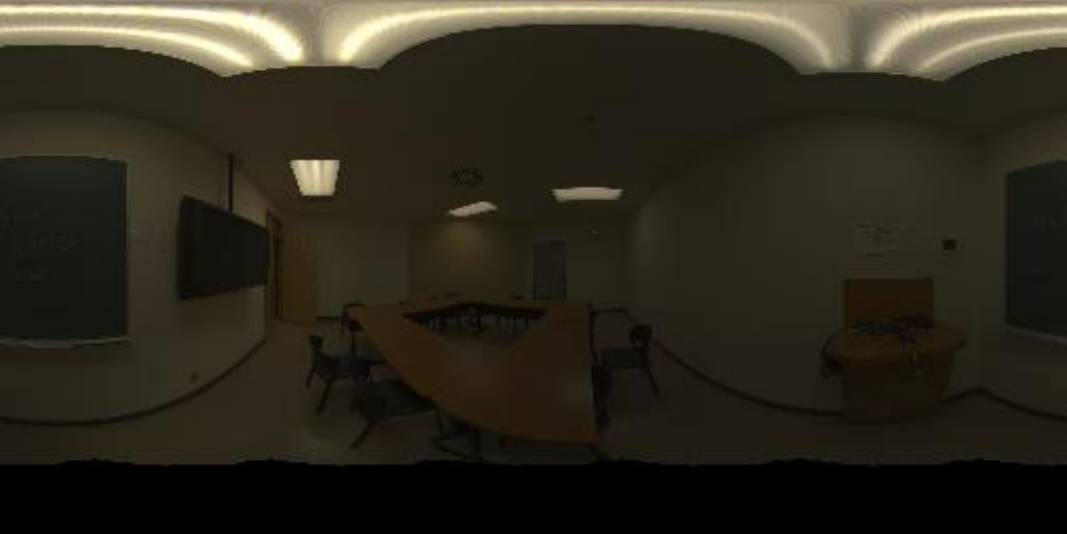}} & 
        \noindent\parbox[c]{0.14\textwidth}{\includegraphics[height=0.100\textwidth]{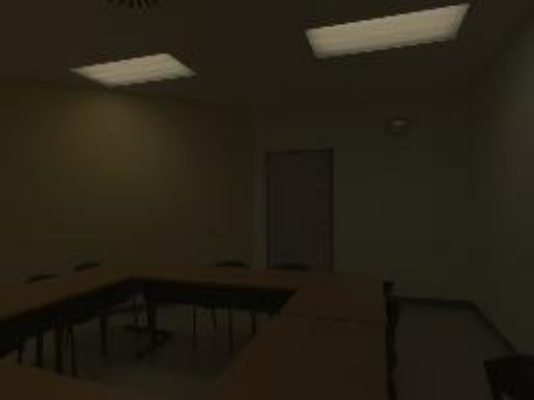}} &  
        
        \noindent\parbox[c]{0.100\textwidth}{\includegraphics[height=0.100\textwidth]{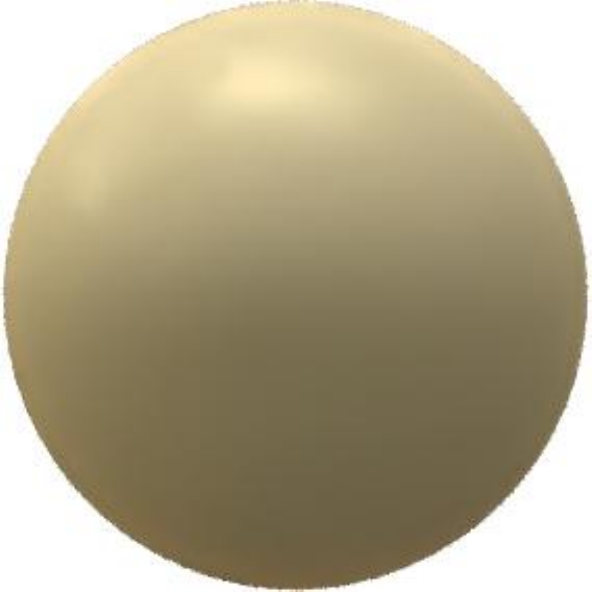}} & 
        \noindent\parbox[c]{0.100\textwidth}{\includegraphics[height=0.100\textwidth]{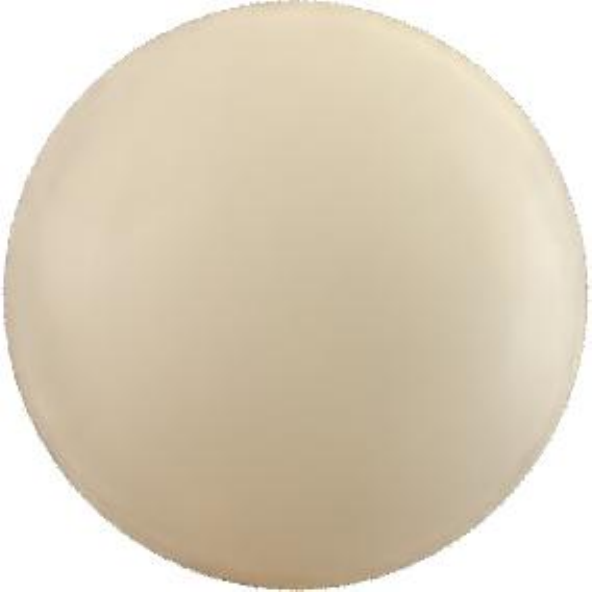}} & 
        
        \noindent\parbox[c]{0.100\textwidth}{\includegraphics[height=0.100\textwidth]{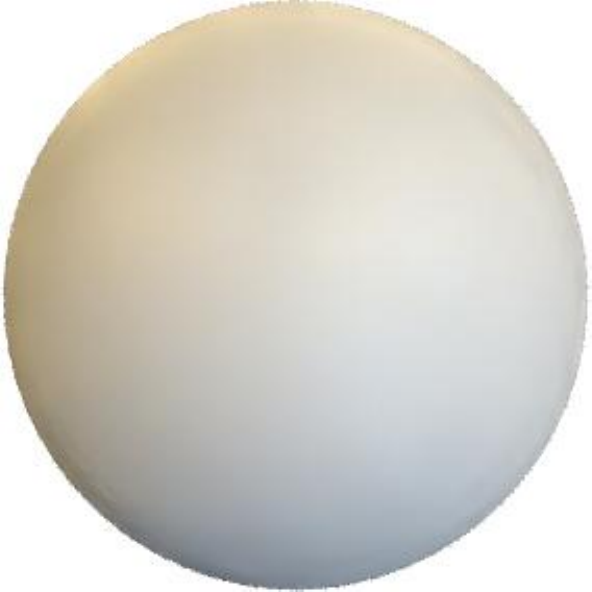}} & 
        \noindent\parbox[c]{0.100\textwidth}{\includegraphics[height=0.100\textwidth]{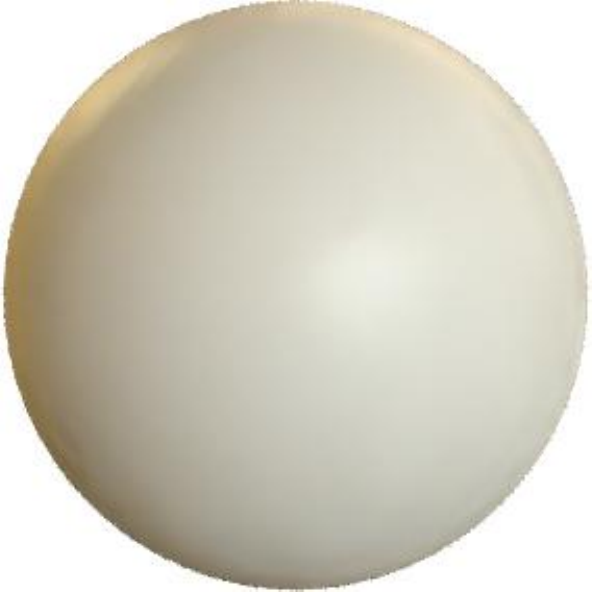}} &
        \noindent\parbox[c]{0.100\textwidth}{\includegraphics[height=0.100\textwidth]{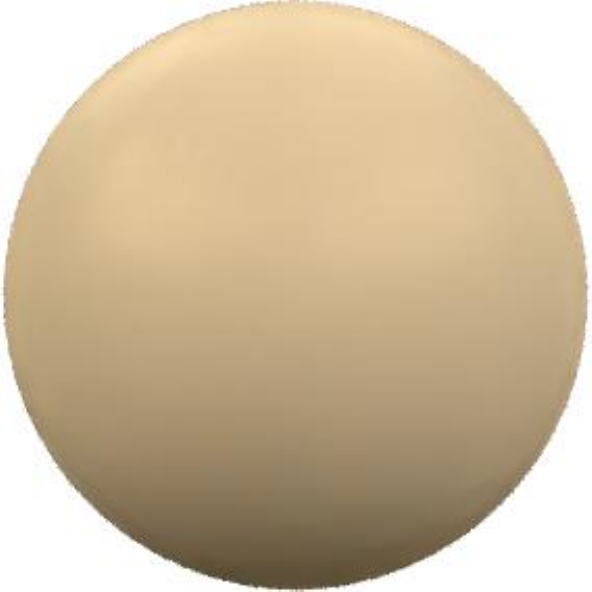}} & 
        \noindent\parbox[c]{0.100\textwidth}{\includegraphics[height=0.100\textwidth]{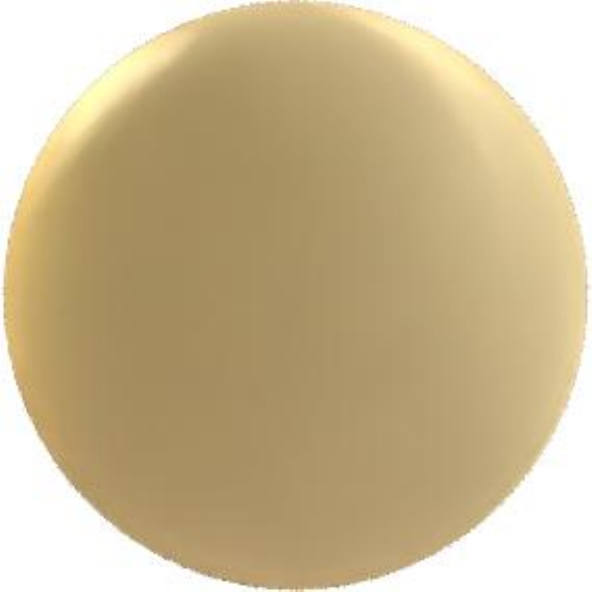}} & 
        \\

        \noindent\parbox[c]{0.205\textwidth}{\includegraphics[height=0.100\textwidth]{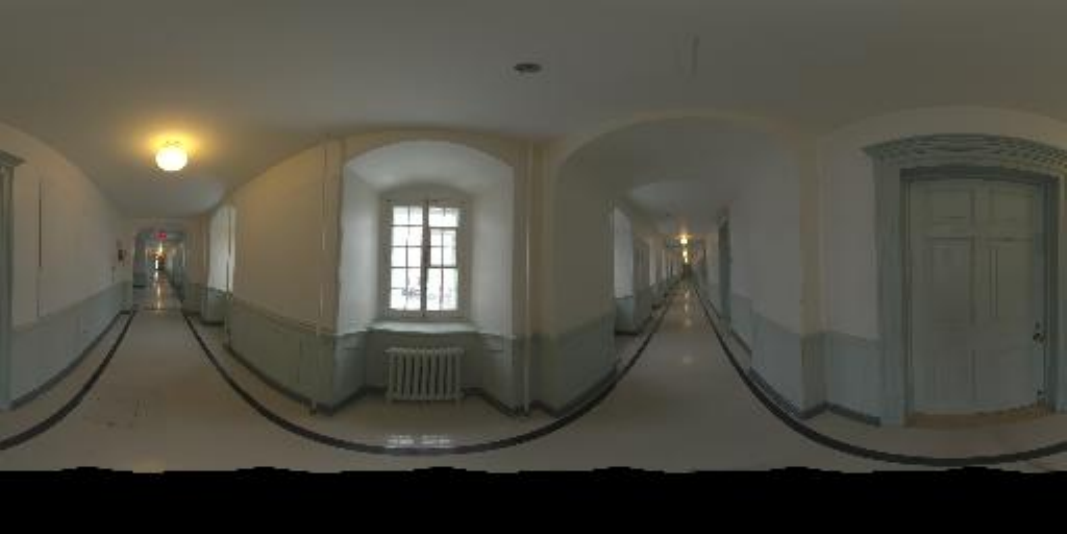}} & 
        \noindent\parbox[c]{0.14\textwidth}{\includegraphics[height=0.100\textwidth]{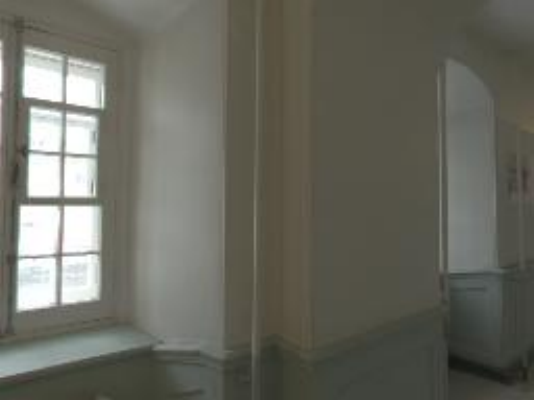}} &  
        
        \noindent\parbox[c]{0.100\textwidth}{\includegraphics[height=0.100\textwidth]{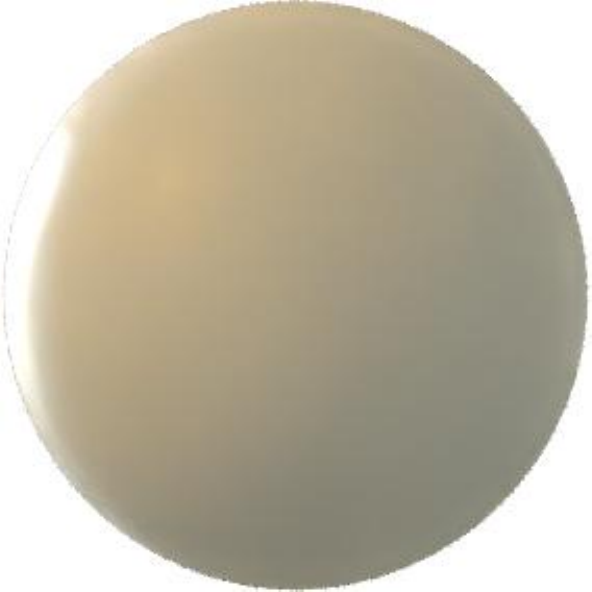}} & 
        \noindent\parbox[c]{0.100\textwidth}{\includegraphics[height=0.100\textwidth]{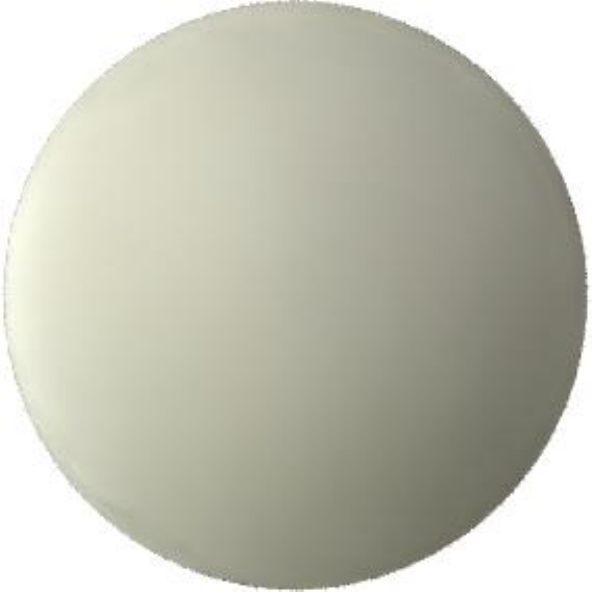}} & 
        
        \noindent\parbox[c]{0.100\textwidth}{\includegraphics[height=0.100\textwidth]{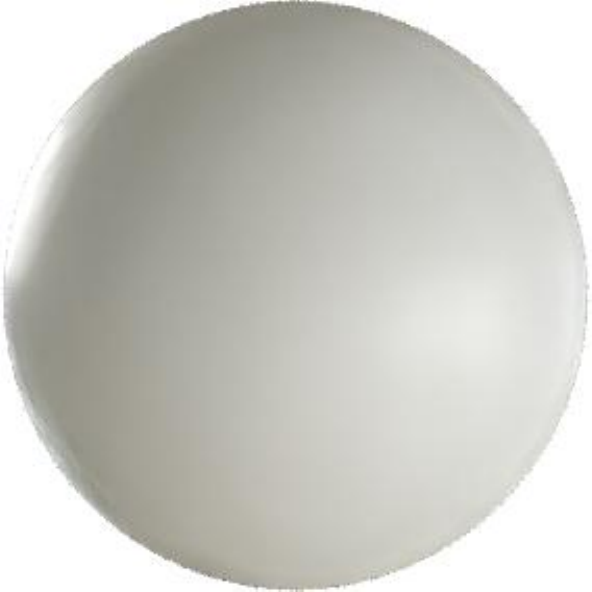}} & 
        \noindent\parbox[c]{0.100\textwidth}{\includegraphics[height=0.100\textwidth]{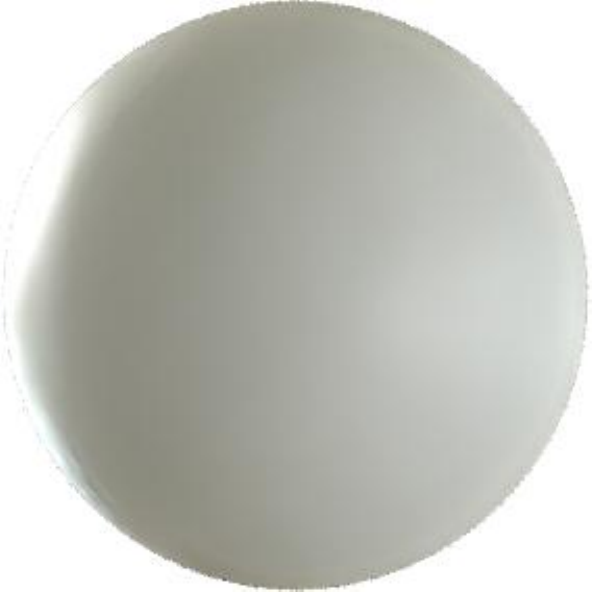}} &
        \noindent\parbox[c]{0.100\textwidth}{\includegraphics[height=0.100\textwidth]{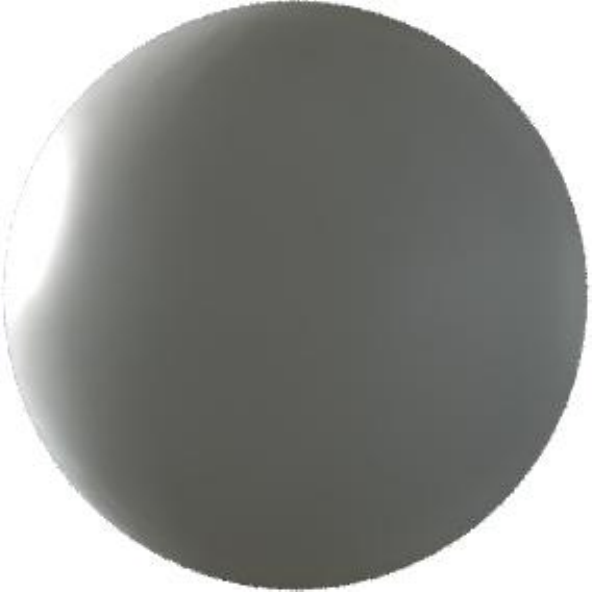}} & 
        \noindent\parbox[c]{0.100\textwidth}{\includegraphics[height=0.100\textwidth]{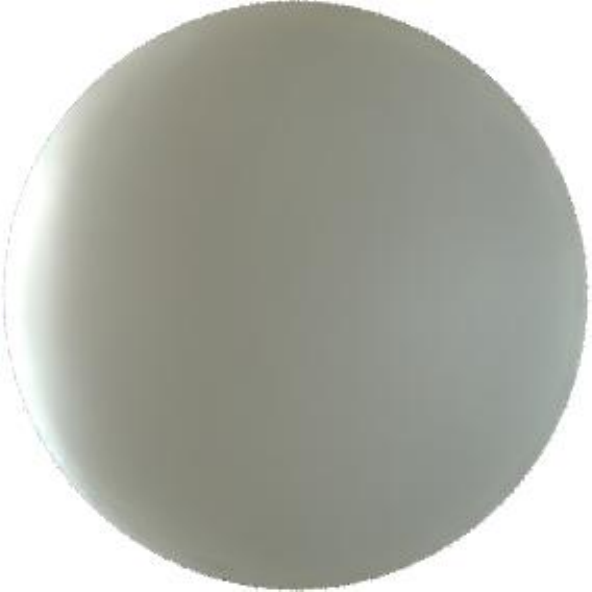}} & 
        \\

        \noindent\parbox[c]{0.205\textwidth}{\includegraphics[height=0.100\textwidth]{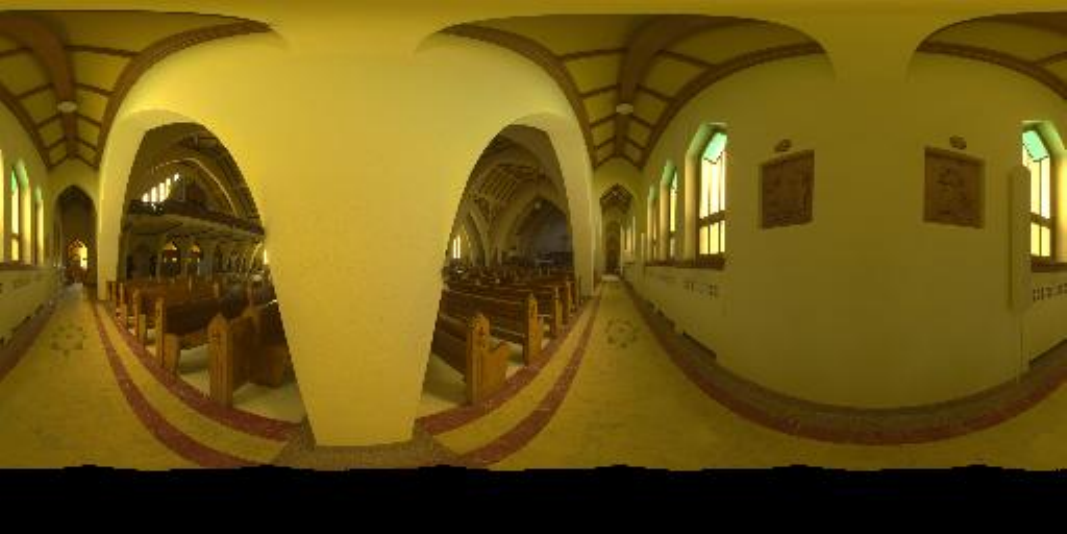}} & 
        \noindent\parbox[c]{0.14\textwidth}{\includegraphics[height=0.100\textwidth]{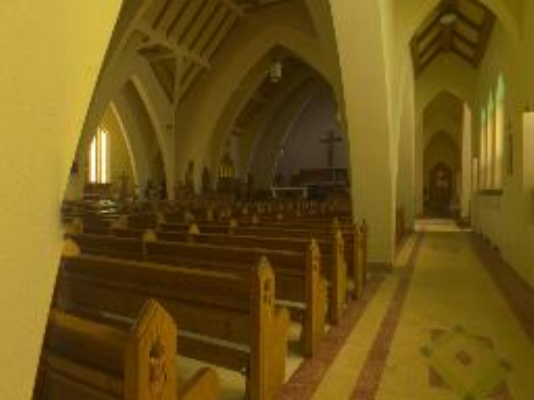}} &  
        
        \noindent\parbox[c]{0.100\textwidth}{\includegraphics[height=0.100\textwidth]{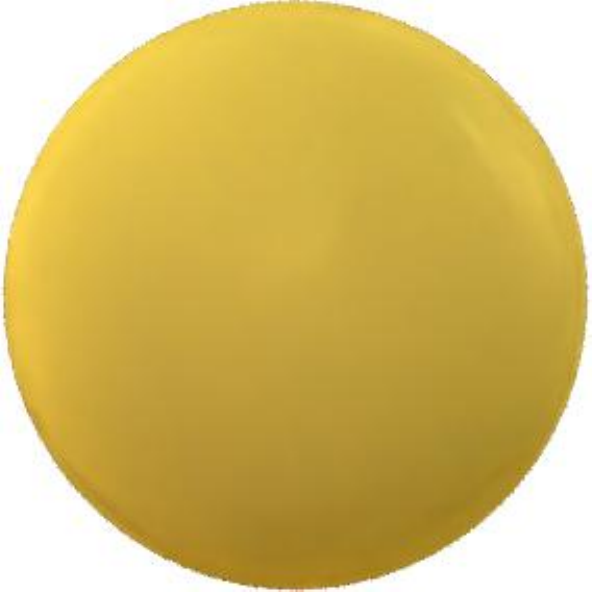}} & 
        \noindent\parbox[c]{0.100\textwidth}{\includegraphics[height=0.100\textwidth]{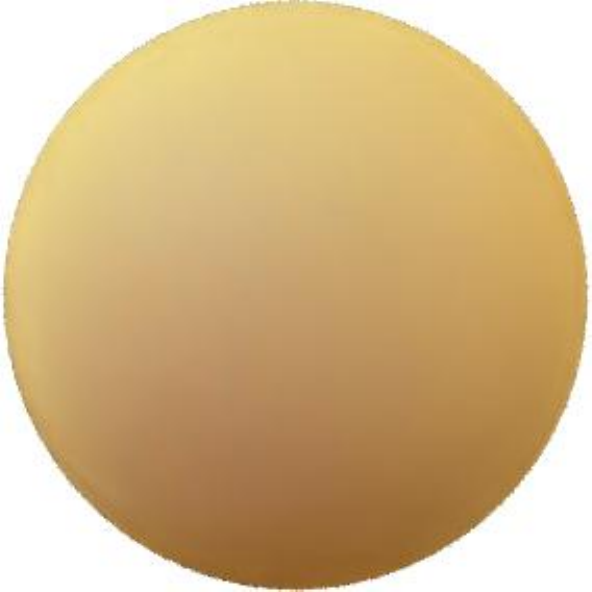}} & 
        
        \noindent\parbox[c]{0.100\textwidth}{\includegraphics[height=0.100\textwidth]{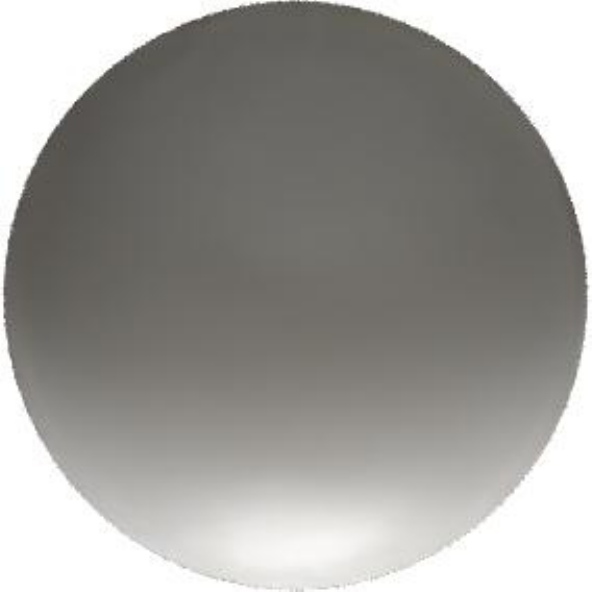}} & 
        \noindent\parbox[c]{0.100\textwidth}{\includegraphics[height=0.100\textwidth]{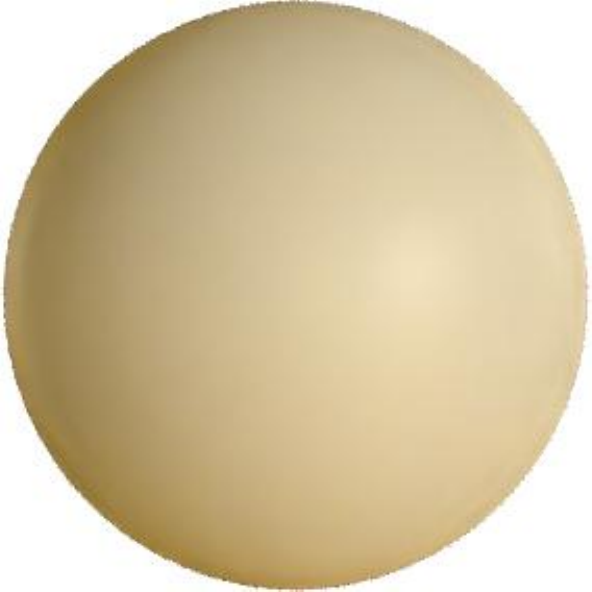}} &
        \noindent\parbox[c]{0.100\textwidth}{\includegraphics[height=0.100\textwidth]{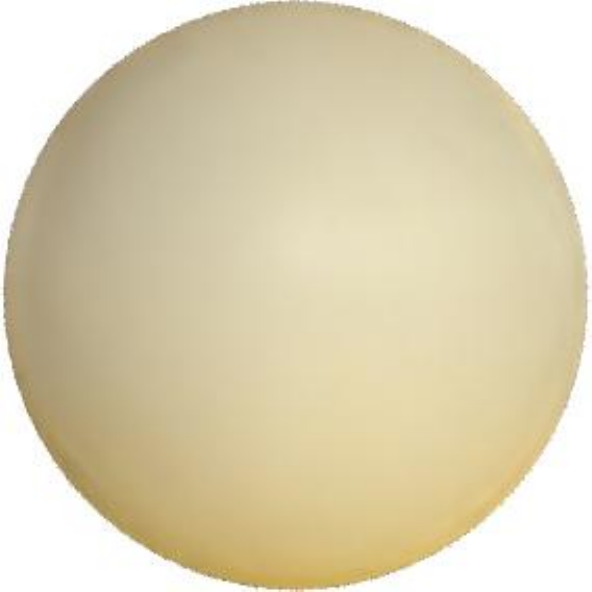}} & 
        \noindent\parbox[c]{0.100\textwidth}{\includegraphics[height=0.100\textwidth]{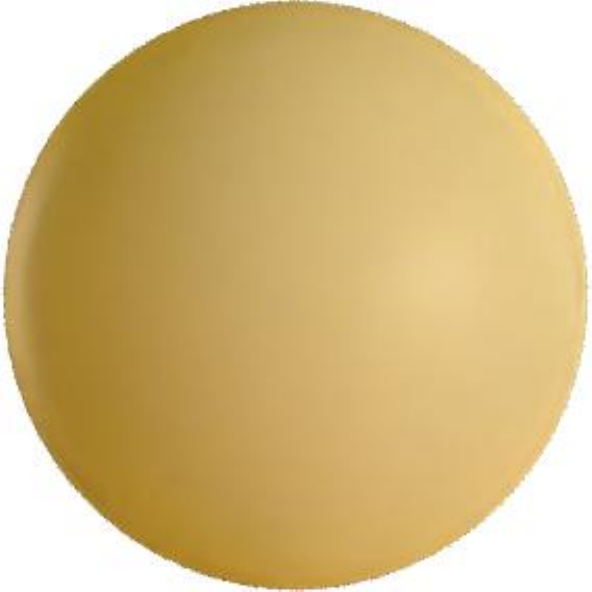}} & 
        \\

        \noindent\parbox[c]{0.205\textwidth}{\includegraphics[height=0.100\textwidth]{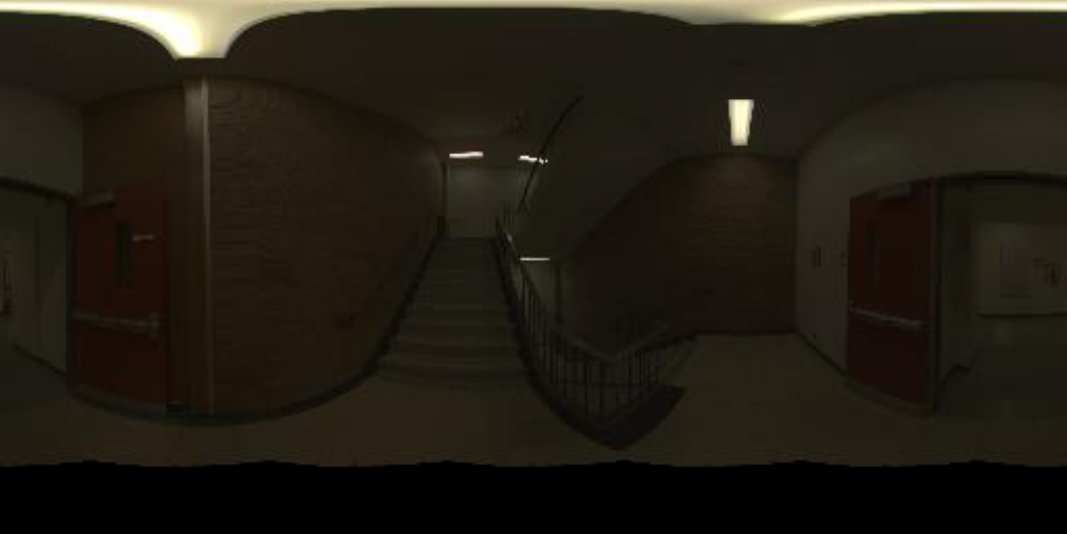}} & 
        \noindent\parbox[c]{0.14\textwidth}{\includegraphics[height=0.100\textwidth]{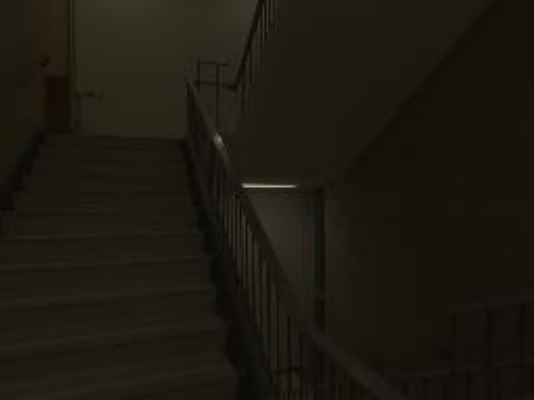}} &  
        
        \noindent\parbox[c]{0.100\textwidth}{\includegraphics[height=0.100\textwidth]{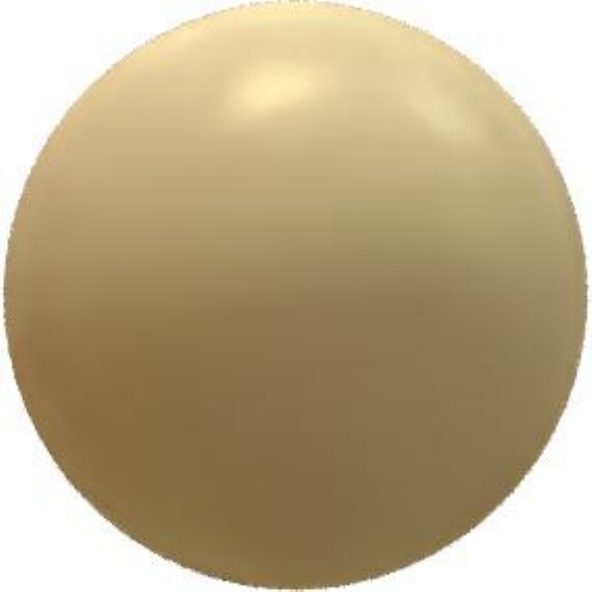}} & 
        \noindent\parbox[c]{0.100\textwidth}{\includegraphics[height=0.100\textwidth]{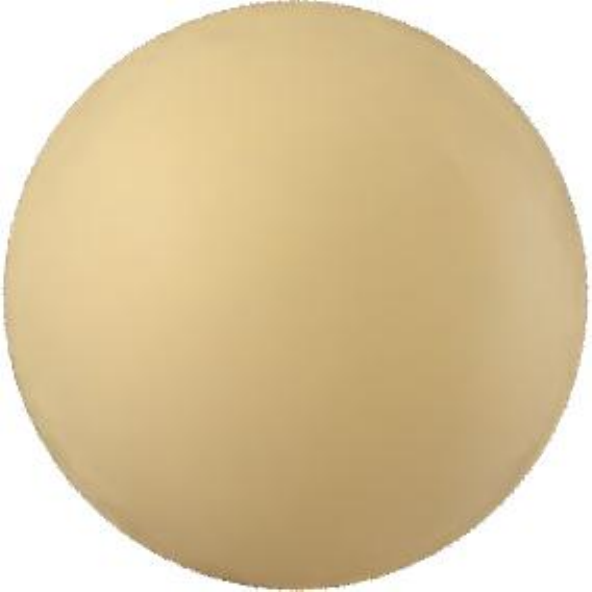}} & 
        
        \noindent\parbox[c]{0.100\textwidth}{\includegraphics[height=0.100\textwidth]{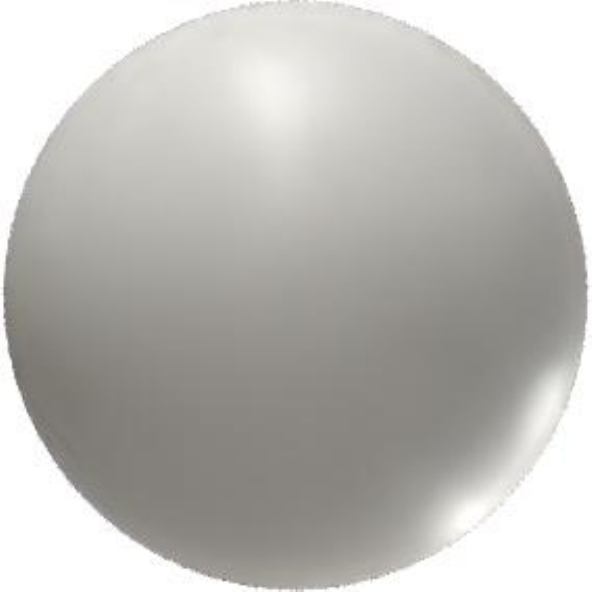}} & 
        \noindent\parbox[c]{0.100\textwidth}{\includegraphics[height=0.100\textwidth]{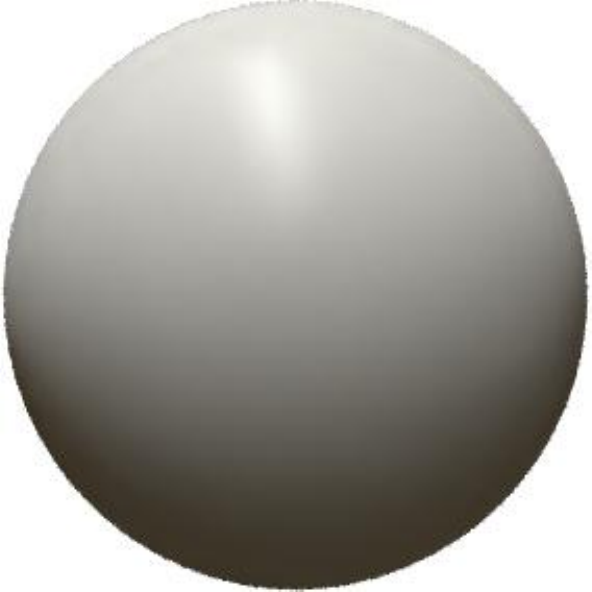}} &
        \noindent\parbox[c]{0.100\textwidth}{\includegraphics[height=0.100\textwidth]{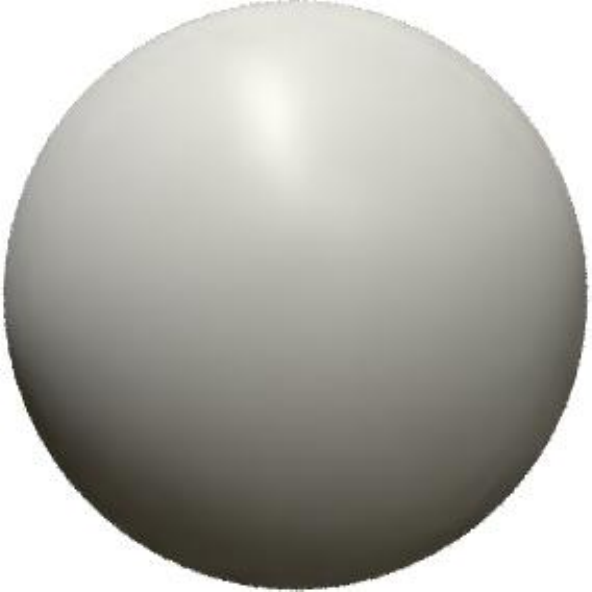}} & 
        \noindent\parbox[c]{0.100\textwidth}{\includegraphics[height=0.100\textwidth]{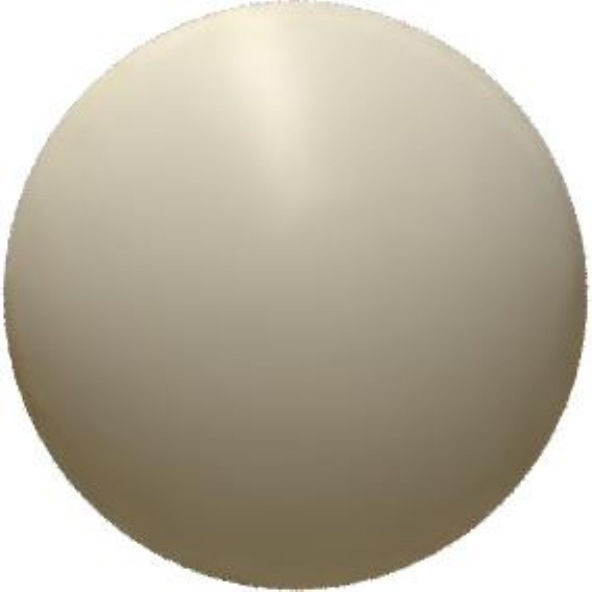}} & 
        \\

        \noindent\parbox[c]{0.205\textwidth}{\includegraphics[height=0.100\textwidth]{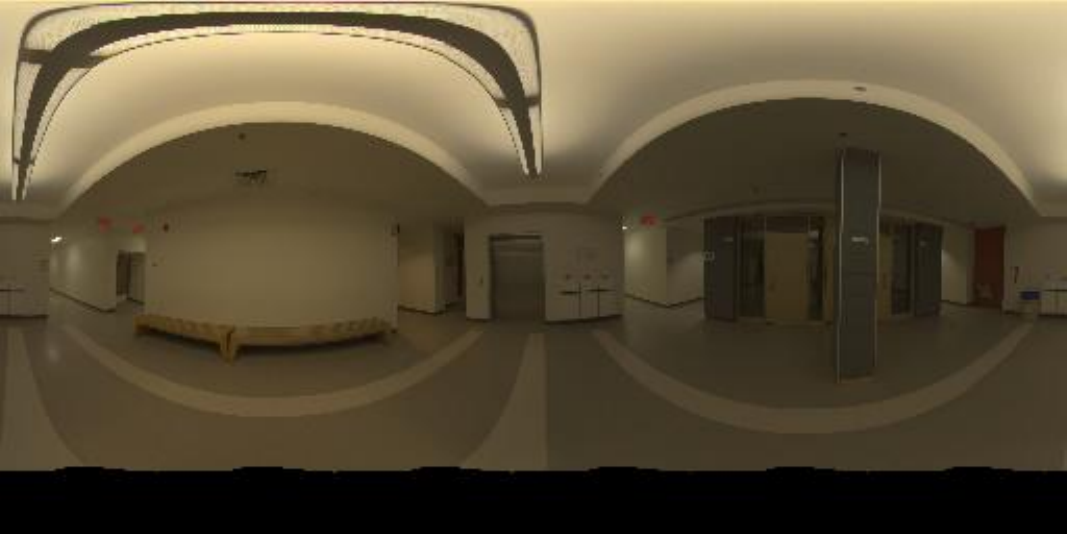}} & 
        \noindent\parbox[c]{0.14\textwidth}{\includegraphics[height=0.100\textwidth]{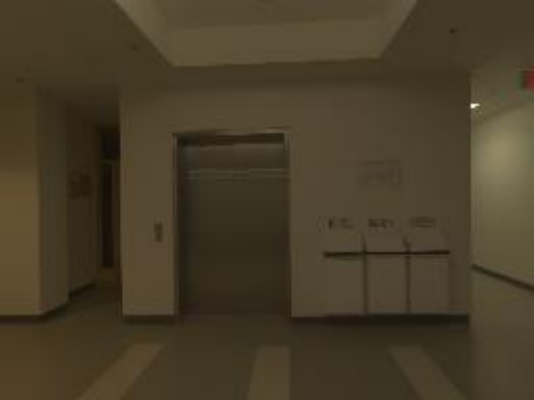}} &  
        
        \noindent\parbox[c]{0.100\textwidth}{\includegraphics[height=0.100\textwidth]{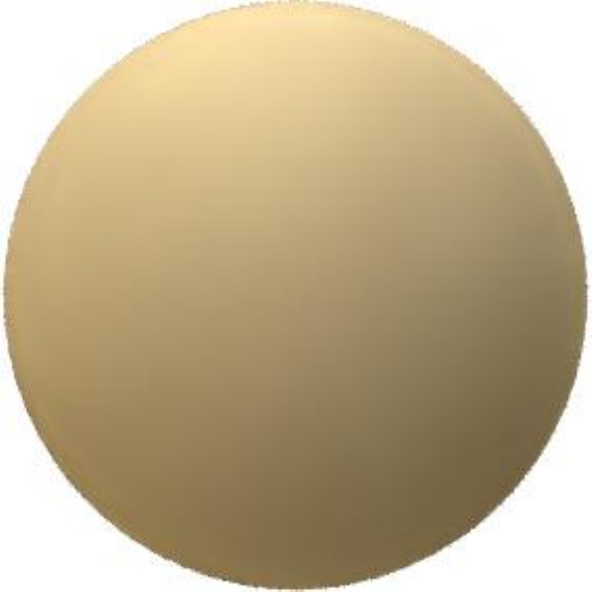}} & 
        \noindent\parbox[c]{0.100\textwidth}{\includegraphics[height=0.100\textwidth]{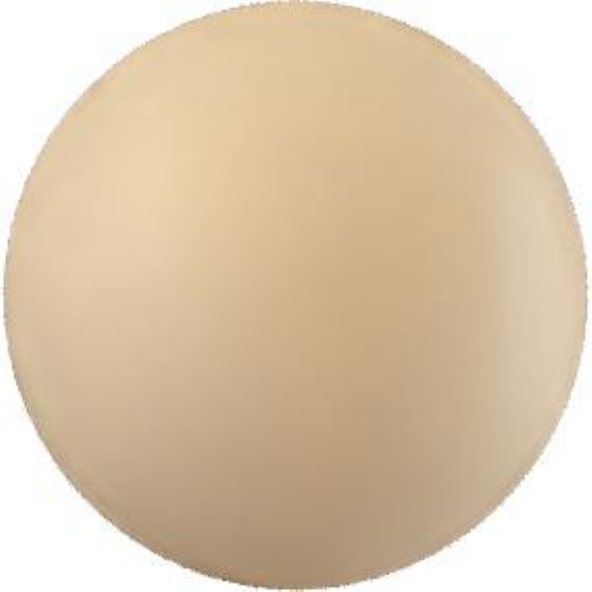}} & 
        
        \noindent\parbox[c]{0.100\textwidth}{\includegraphics[height=0.100\textwidth]{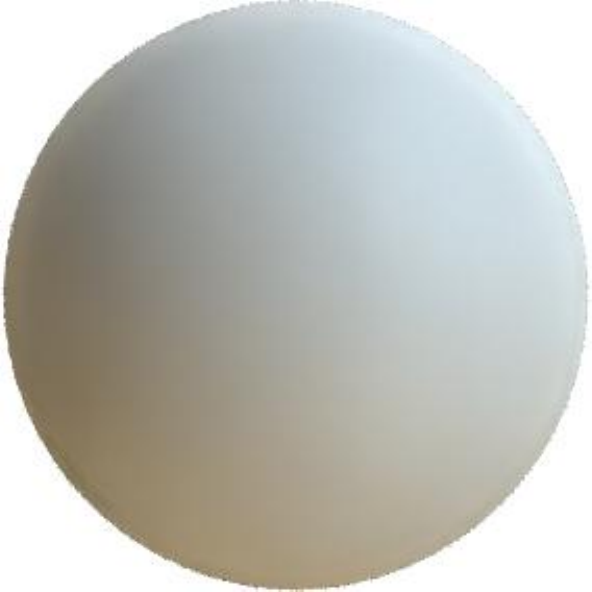}} & 
        \noindent\parbox[c]{0.100\textwidth}{\includegraphics[height=0.100\textwidth]{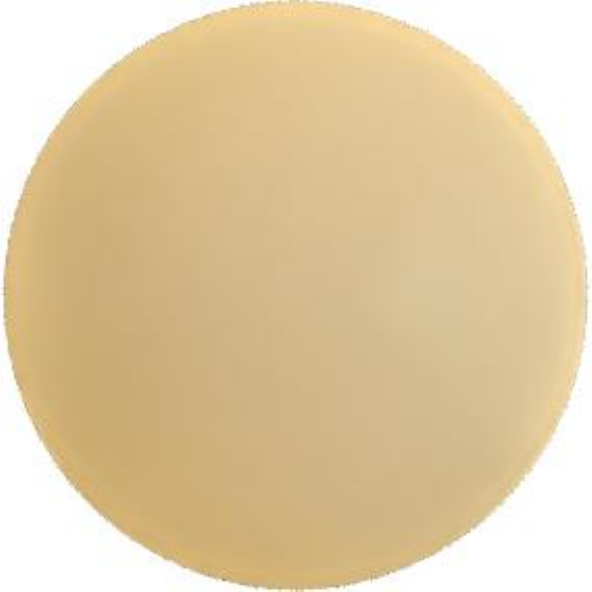}} &
        \noindent\parbox[c]{0.100\textwidth}{\includegraphics[height=0.100\textwidth]{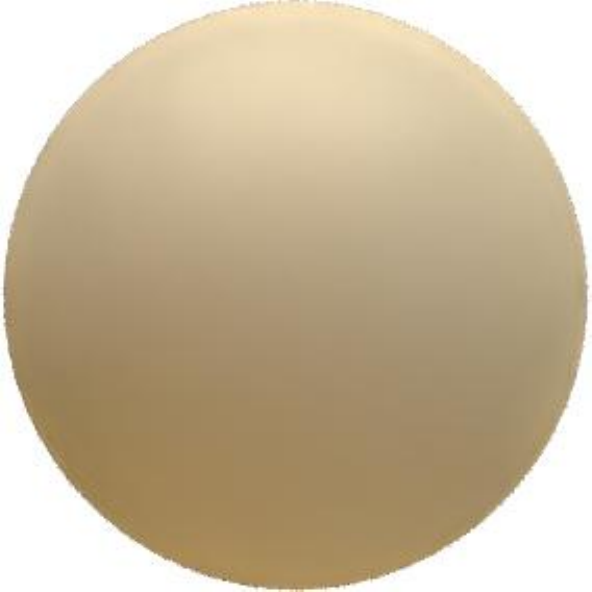}} & 
        \noindent\parbox[c]{0.100\textwidth}{\includegraphics[height=0.100\textwidth]{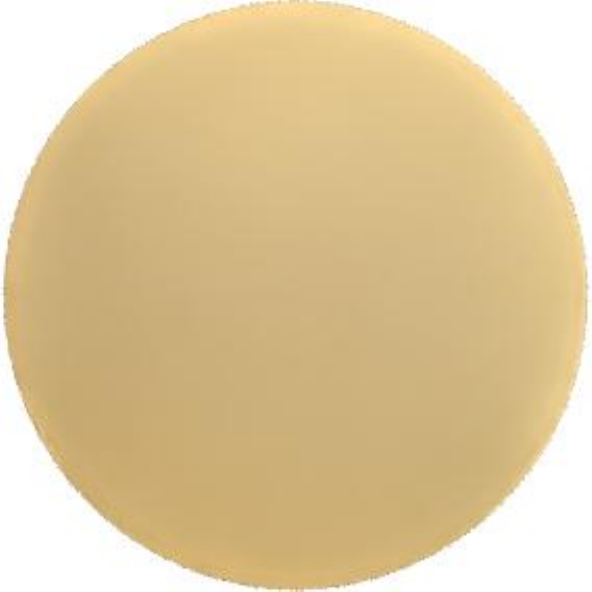}} & 
        \\

        \noindent\parbox[c]{0.205\textwidth}{\includegraphics[height=0.100\textwidth]{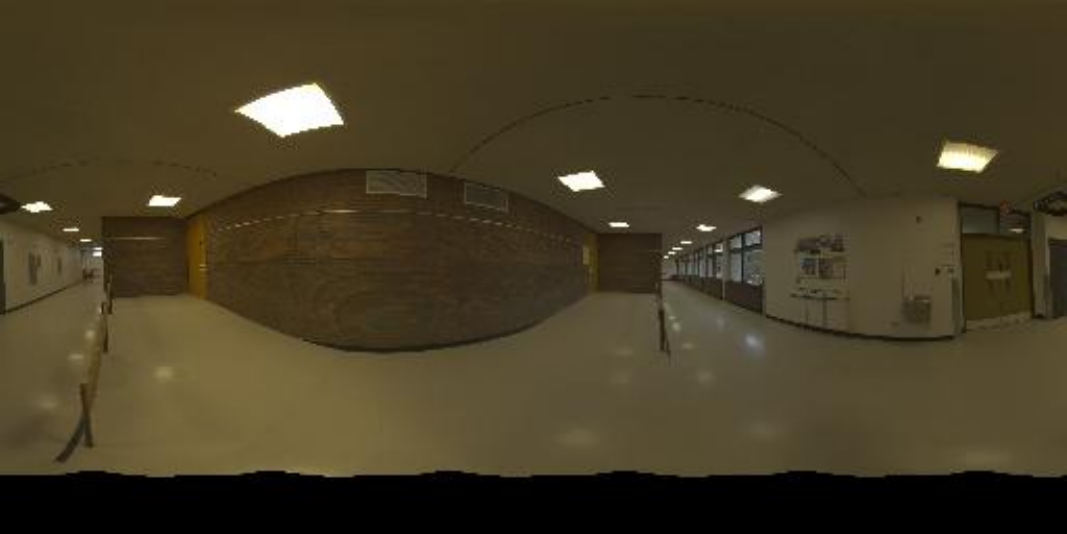}} & 
        \noindent\parbox[c]{0.14\textwidth}{\includegraphics[height=0.100\textwidth]{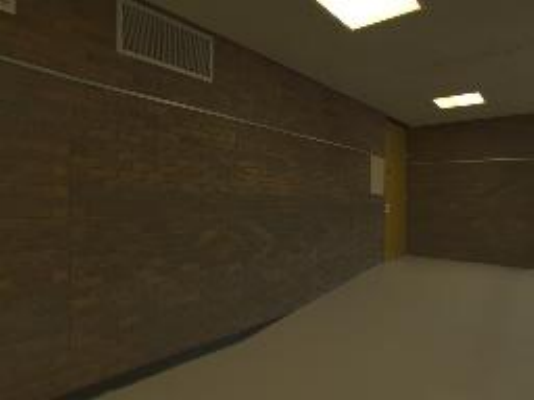}} &  
        
        \noindent\parbox[c]{0.100\textwidth}{\includegraphics[height=0.100\textwidth]{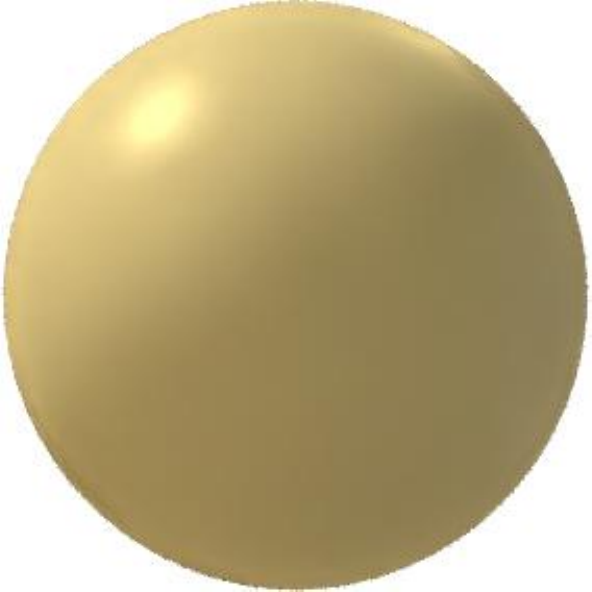}} & 
        \noindent\parbox[c]{0.100\textwidth}{\includegraphics[height=0.100\textwidth]{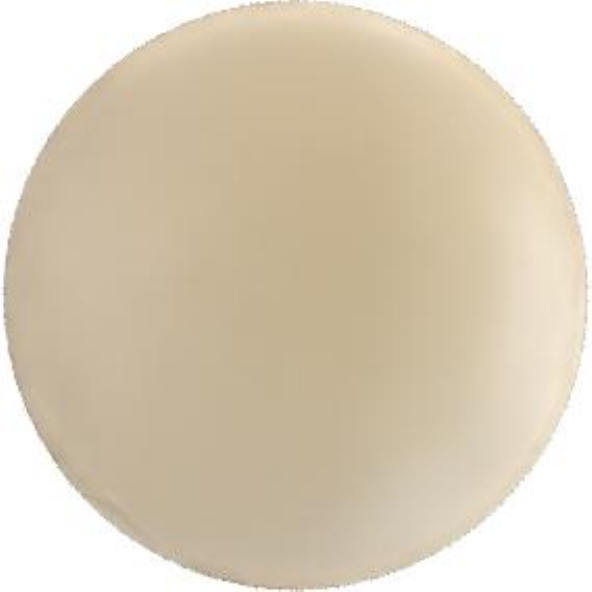}} & 
        
        \noindent\parbox[c]{0.100\textwidth}{\includegraphics[height=0.100\textwidth]{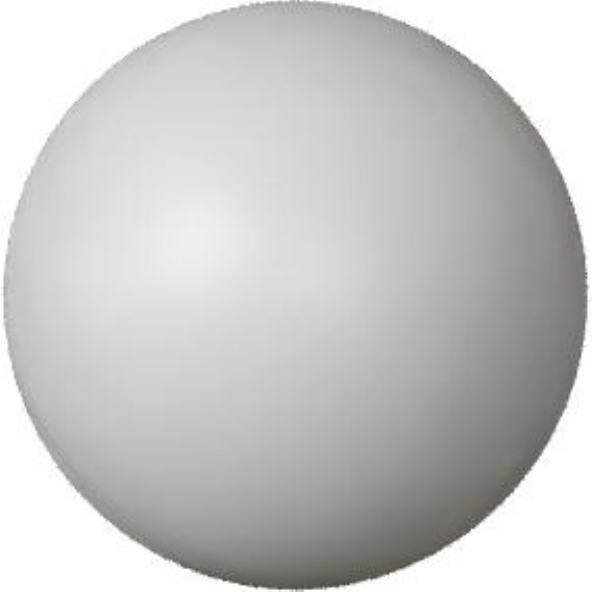}} & 
        \noindent\parbox[c]{0.100\textwidth}{\includegraphics[height=0.100\textwidth]{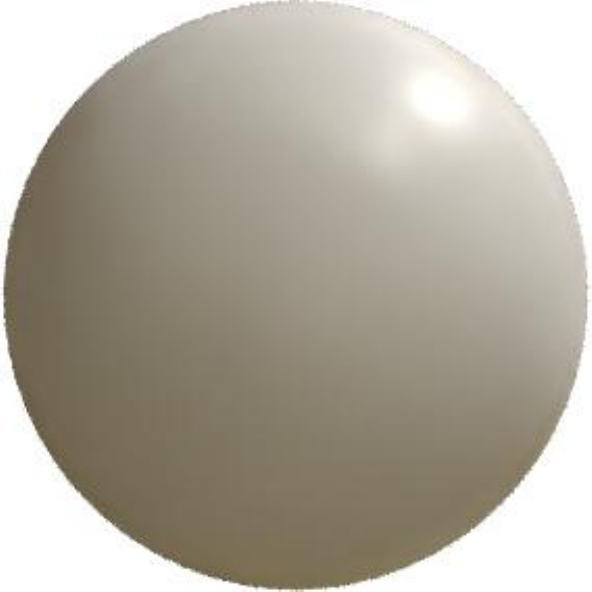}} &
        \noindent\parbox[c]{0.100\textwidth}{\includegraphics[height=0.100\textwidth]{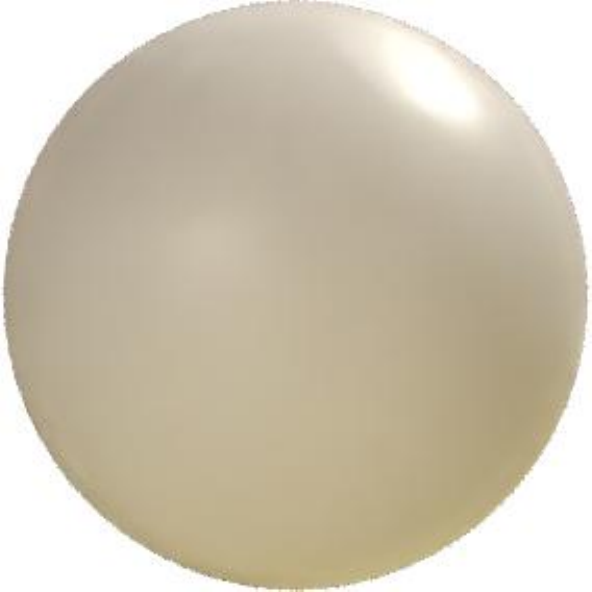}} & 
        \noindent\parbox[c]{0.100\textwidth}{\includegraphics[height=0.100\textwidth]{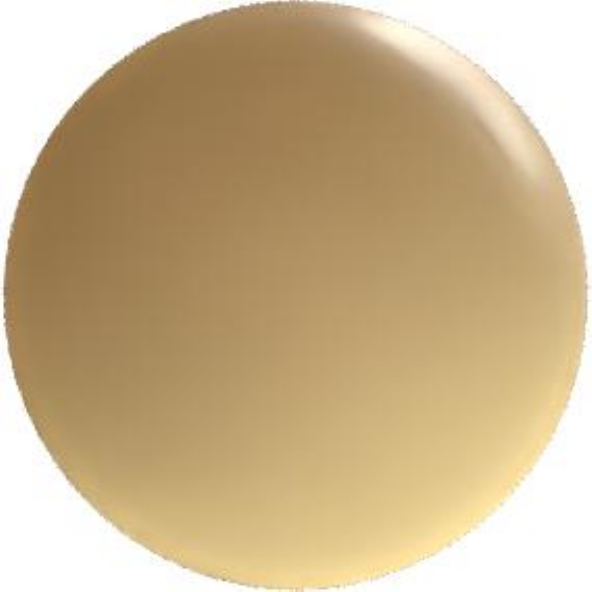}} & 
        \\

        \noindent\parbox[c]{0.205\textwidth}{\includegraphics[height=0.100\textwidth]{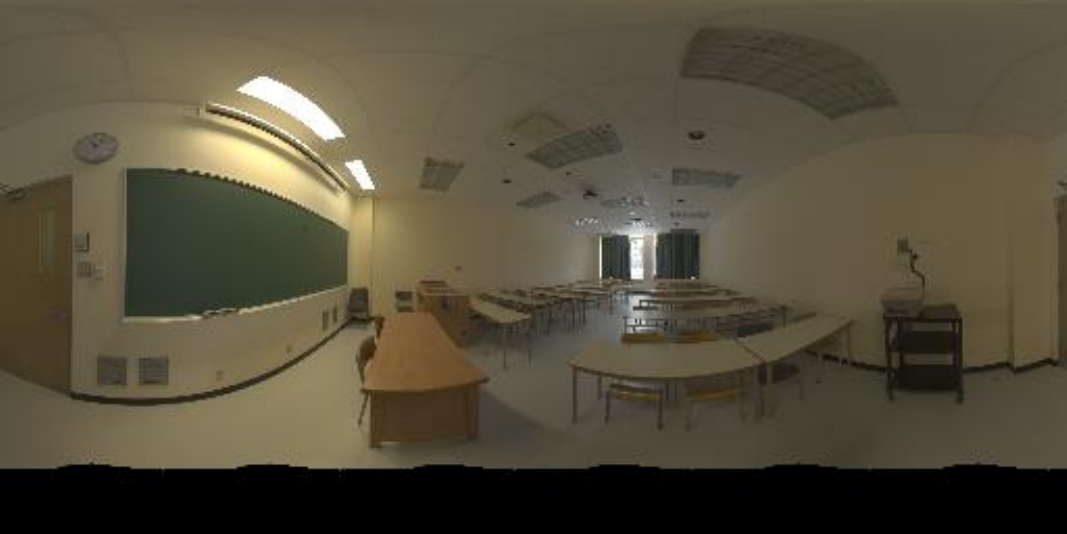}} & 
        \noindent\parbox[c]{0.14\textwidth}{\includegraphics[height=0.100\textwidth]{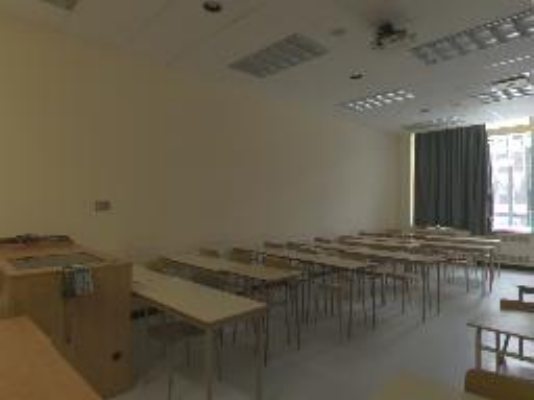}} &  
        
        \noindent\parbox[c]{0.100\textwidth}{\includegraphics[height=0.100\textwidth]{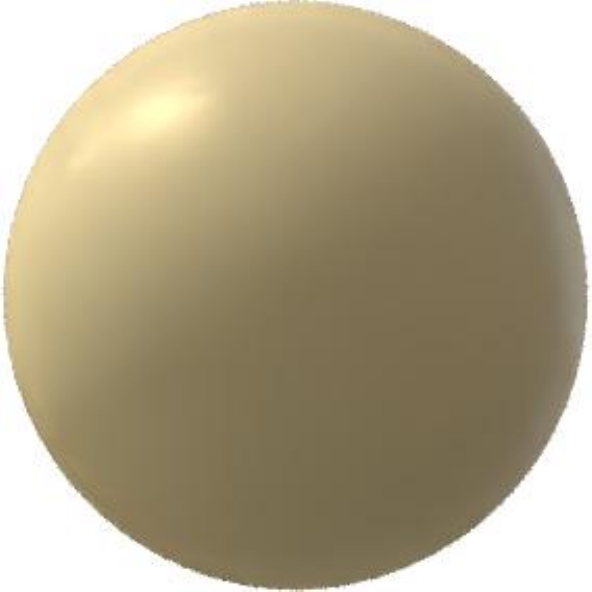}} & 
        \noindent\parbox[c]{0.100\textwidth}{\includegraphics[height=0.100\textwidth]{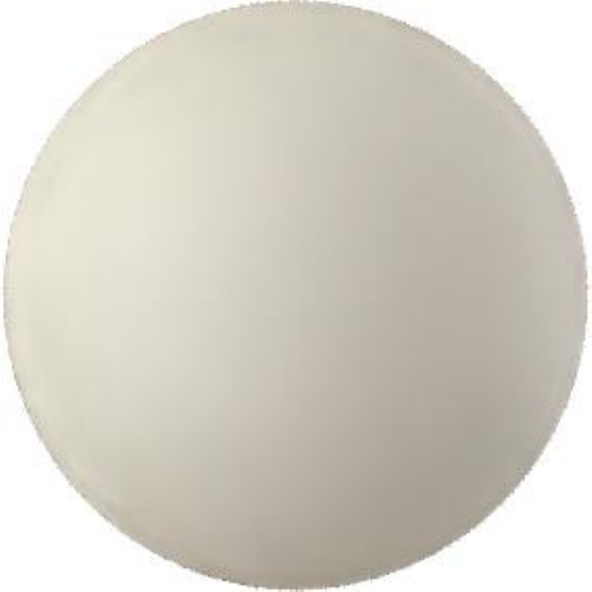}} & 
        
        \noindent\parbox[c]{0.100\textwidth}{\includegraphics[height=0.100\textwidth]{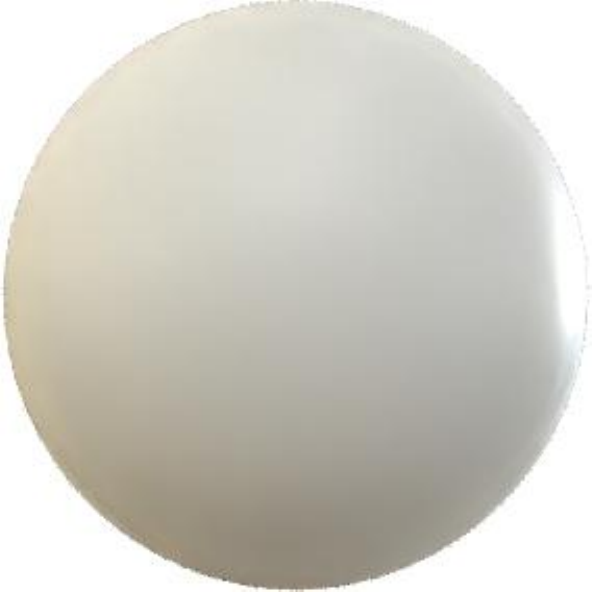}} & 
        \noindent\parbox[c]{0.100\textwidth}{\includegraphics[height=0.100\textwidth]{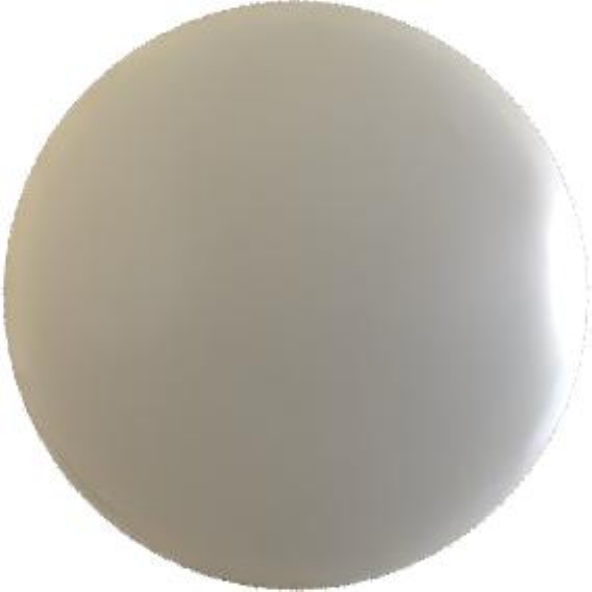}} &
        \noindent\parbox[c]{0.100\textwidth}{\includegraphics[height=0.100\textwidth]{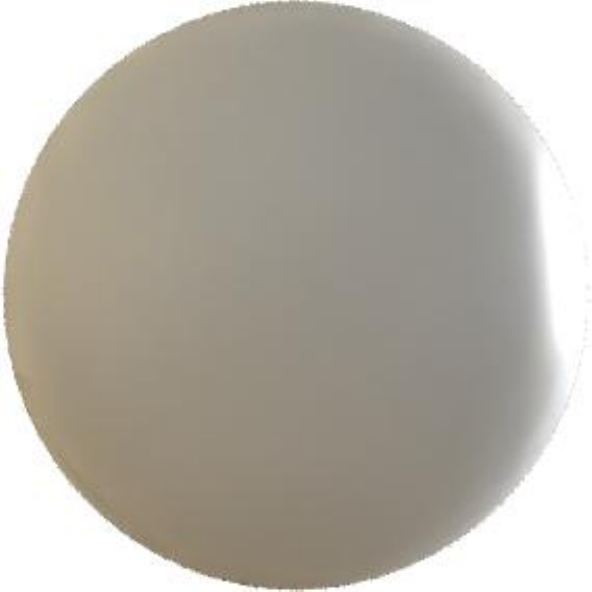}} & 
        \noindent\parbox[c]{0.100\textwidth}{\includegraphics[height=0.100\textwidth]{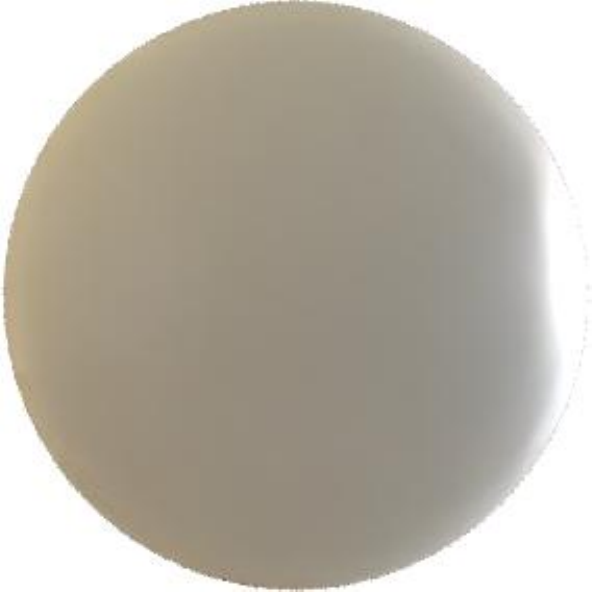}} & 
        \\

        \noindent\parbox[c]{0.205\textwidth}{\includegraphics[height=0.100\textwidth]{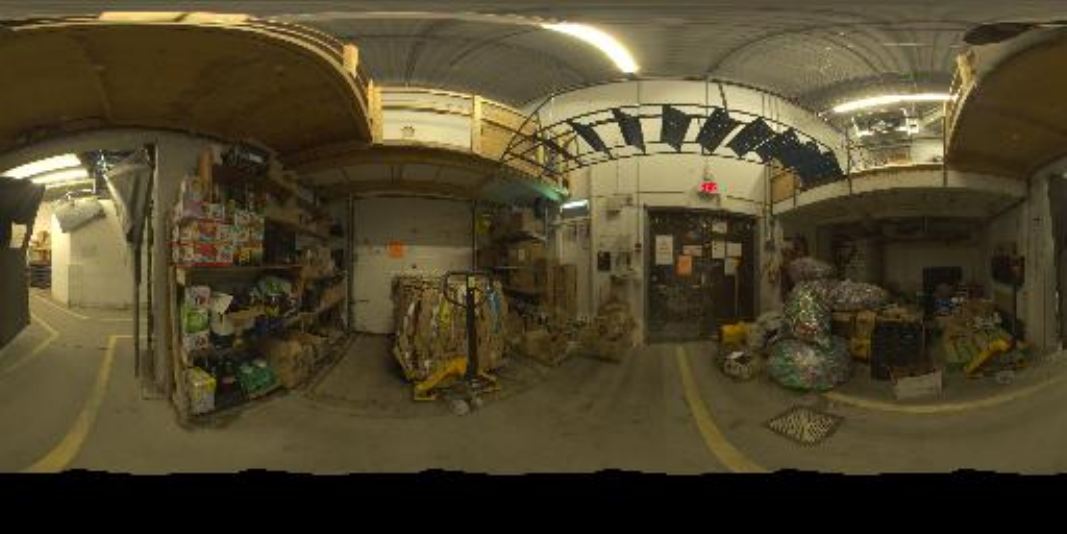}} & 
        \noindent\parbox[c]{0.14\textwidth}{\includegraphics[height=0.100\textwidth]{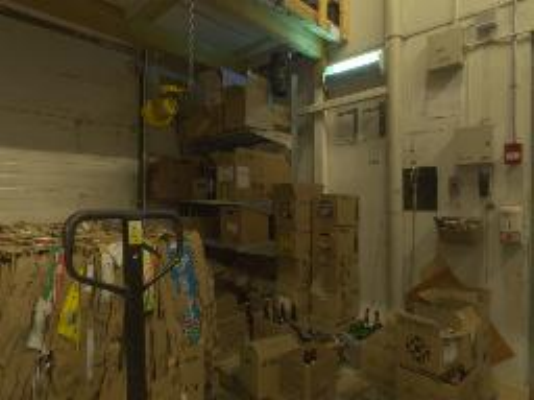}} &  
        
        \noindent\parbox[c]{0.100\textwidth}{\includegraphics[height=0.100\textwidth]{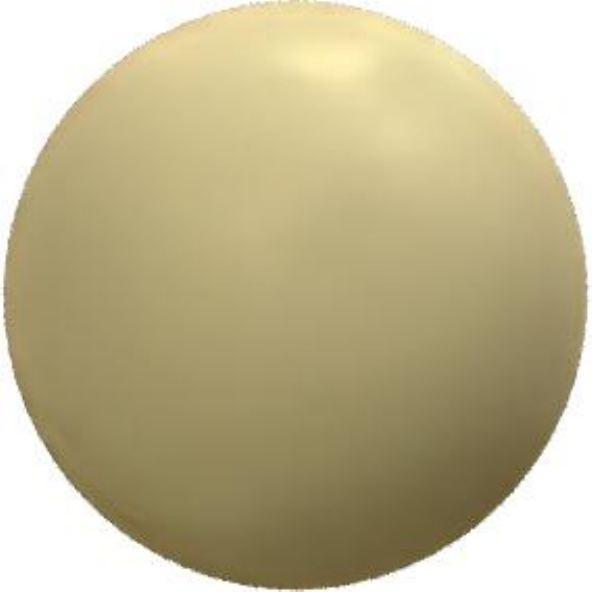}} & 
        \noindent\parbox[c]{0.100\textwidth}{\includegraphics[height=0.100\textwidth]{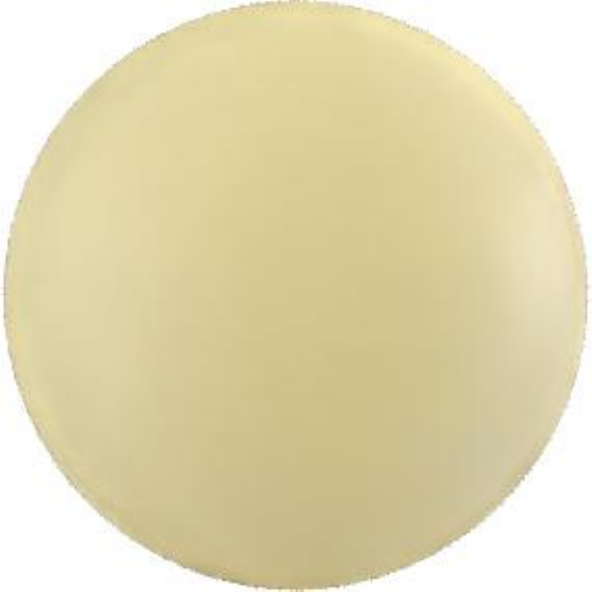}} & 
        
        \noindent\parbox[c]{0.100\textwidth}{\includegraphics[height=0.100\textwidth]{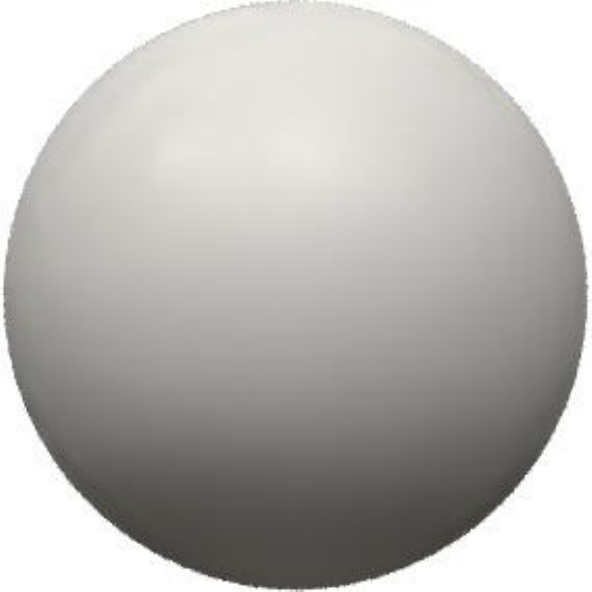}} & 
        \noindent\parbox[c]{0.100\textwidth}{\includegraphics[height=0.100\textwidth]{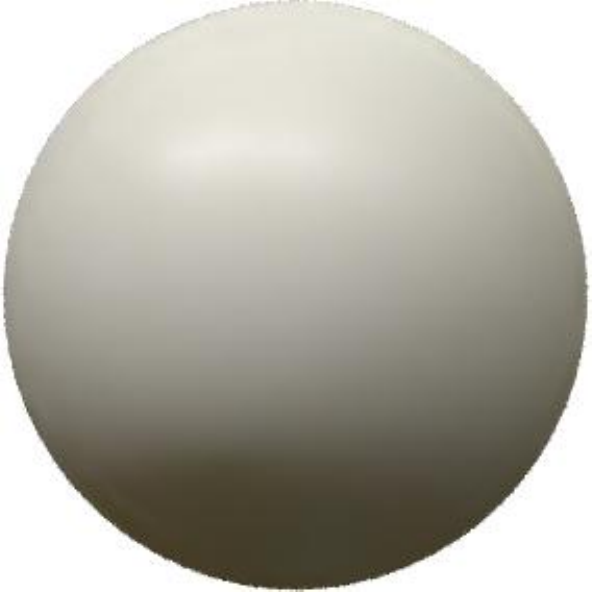}} &
        \noindent\parbox[c]{0.100\textwidth}{\includegraphics[height=0.100\textwidth]{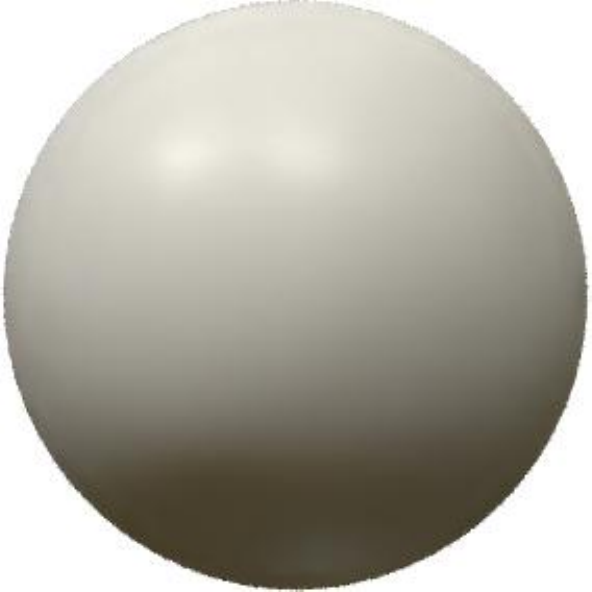}} & 
        \noindent\parbox[c]{0.100\textwidth}{\includegraphics[height=0.100\textwidth]{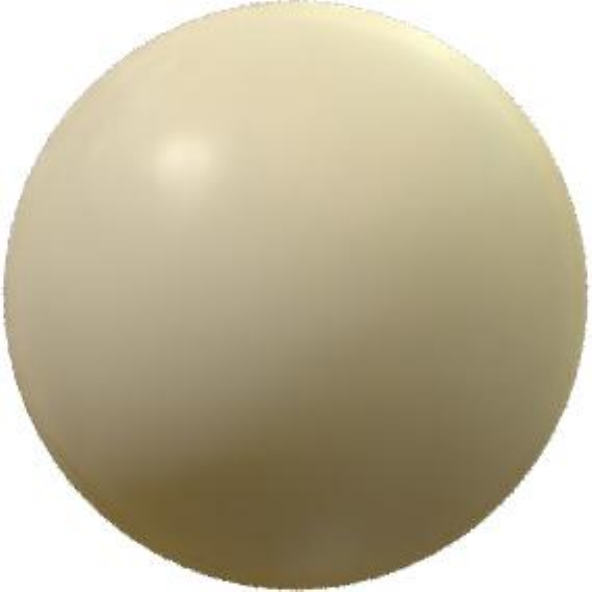}} & 
        \\

        \noindent\parbox[c]{0.205\textwidth}{\includegraphics[height=0.100\textwidth]{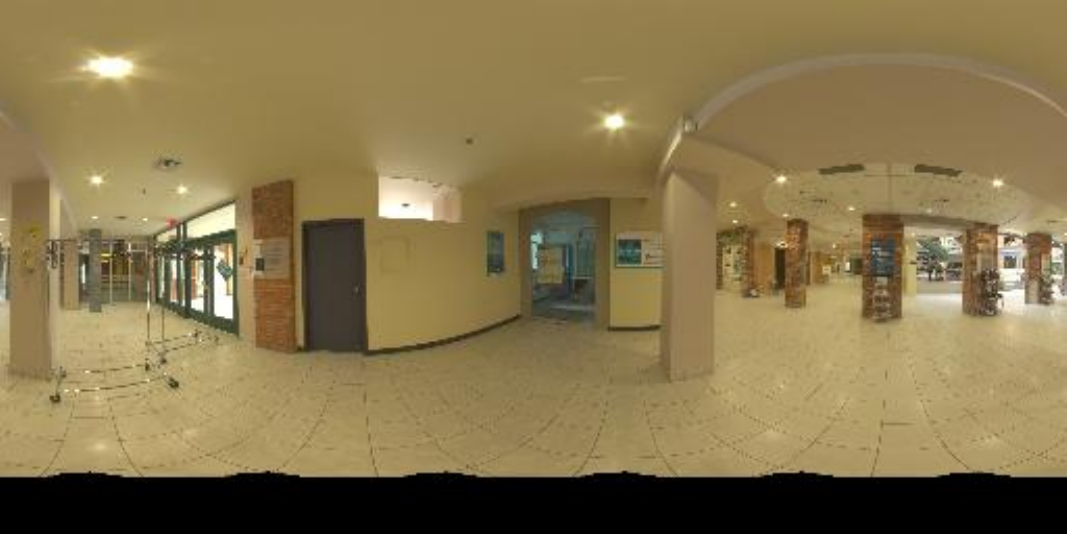}} & 
        \noindent\parbox[c]{0.14\textwidth}{\includegraphics[height=0.100\textwidth]{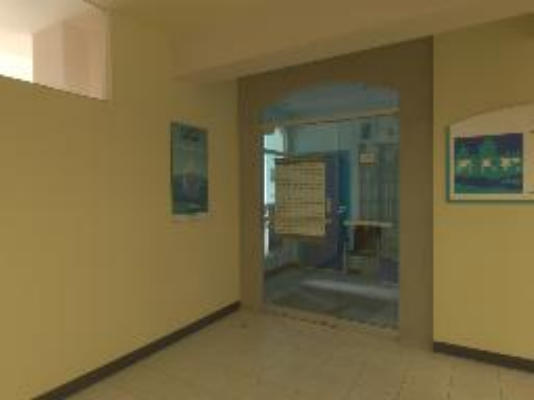}} &  
        
        \noindent\parbox[c]{0.100\textwidth}{\includegraphics[height=0.100\textwidth]{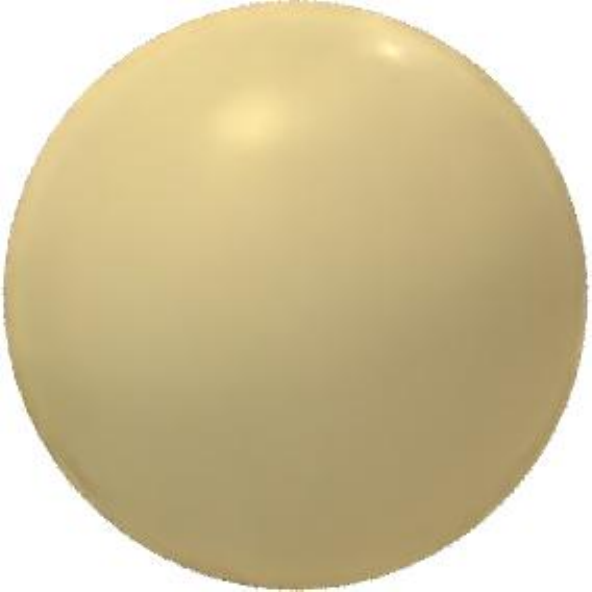}} & 
        \noindent\parbox[c]{0.100\textwidth}{\includegraphics[height=0.100\textwidth]{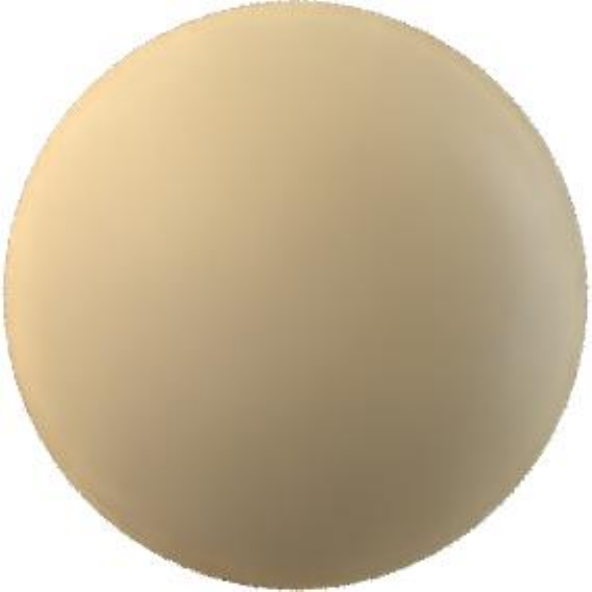}} & 
        
        \noindent\parbox[c]{0.100\textwidth}{\includegraphics[height=0.100\textwidth]{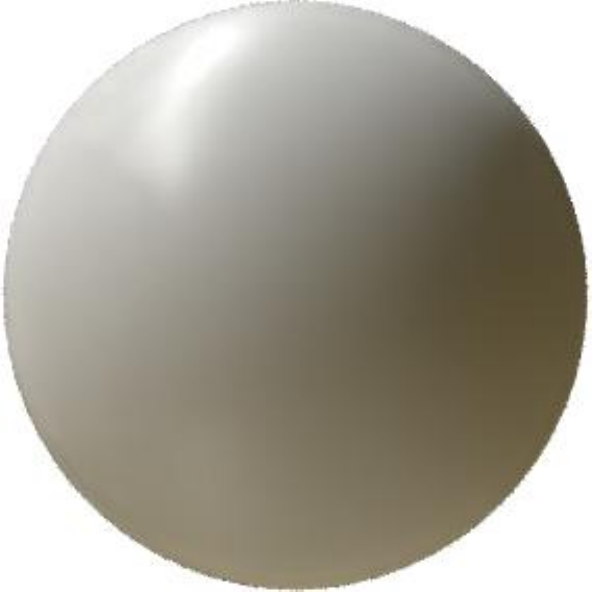}} & 
        \noindent\parbox[c]{0.100\textwidth}{\includegraphics[height=0.100\textwidth]{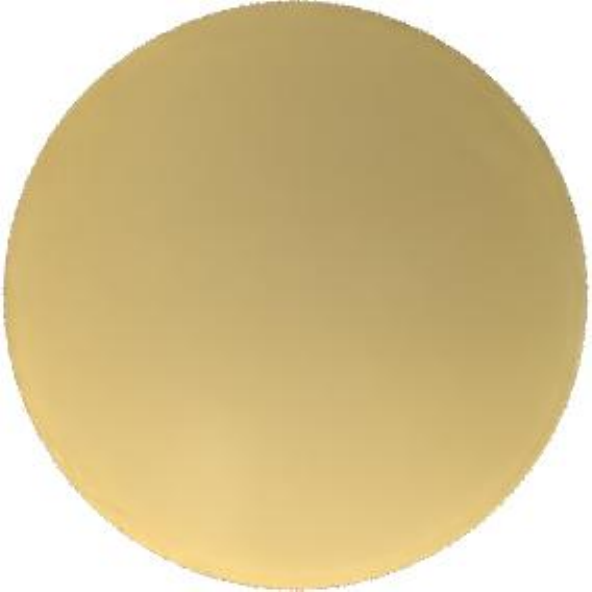}} &
        \noindent\parbox[c]{0.100\textwidth}{\includegraphics[height=0.100\textwidth]{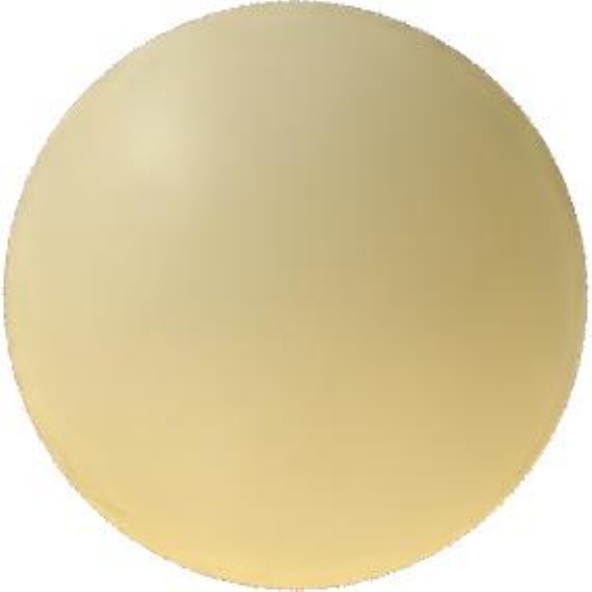}} & 
        \noindent\parbox[c]{0.100\textwidth}{\includegraphics[height=0.100\textwidth]{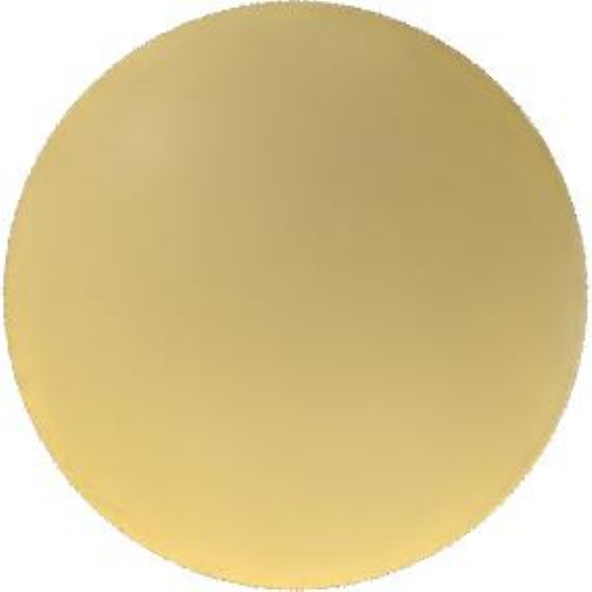}} & 
        \\

        \noindent\parbox[c]{0.205\textwidth}{\includegraphics[height=0.100\textwidth]{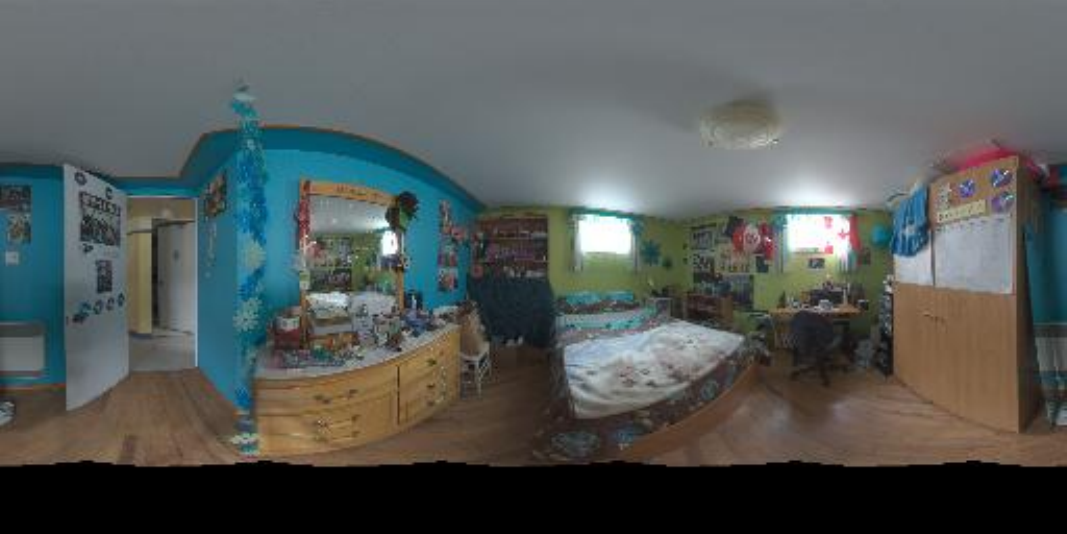}} & 
        \noindent\parbox[c]{0.14\textwidth}{\includegraphics[height=0.100\textwidth]{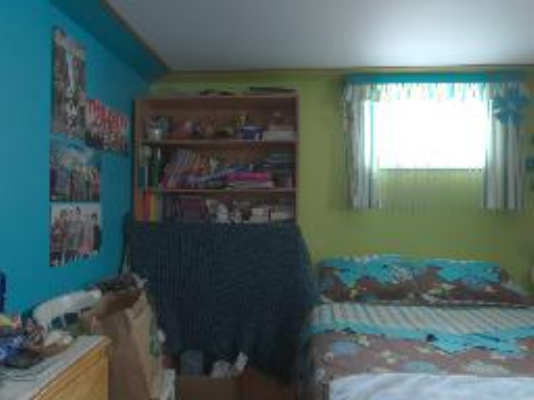}} &  
        
        \noindent\parbox[c]{0.100\textwidth}{\includegraphics[height=0.100\textwidth]{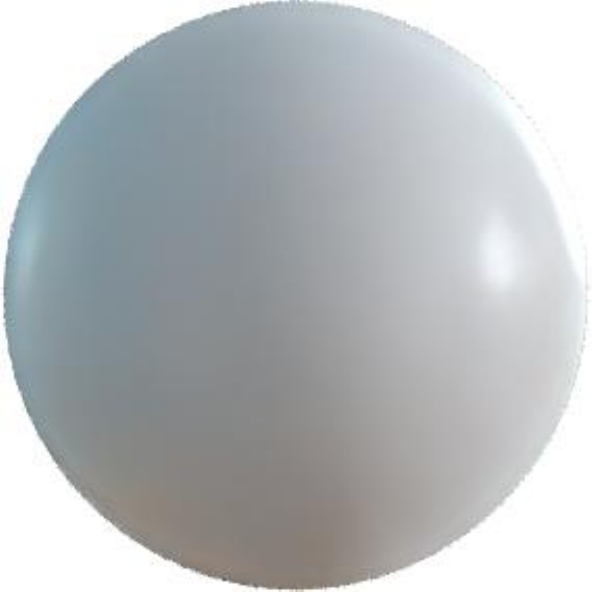}} & 
        \noindent\parbox[c]{0.100\textwidth}{\includegraphics[height=0.100\textwidth]{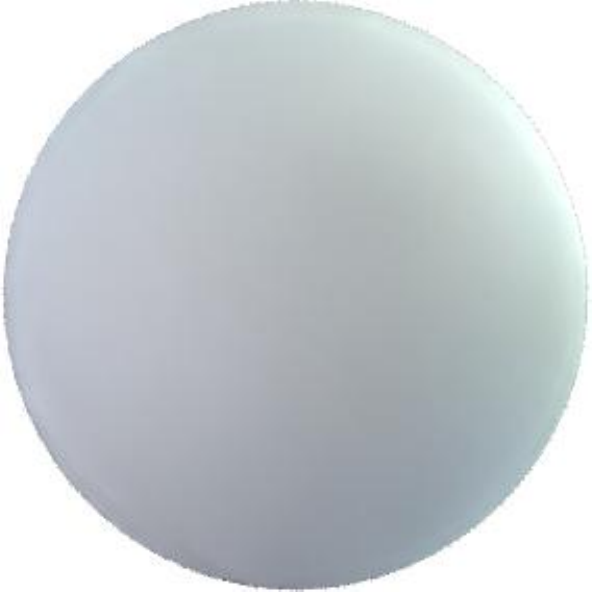}} & 
        
        \noindent\parbox[c]{0.100\textwidth}{\includegraphics[height=0.100\textwidth]{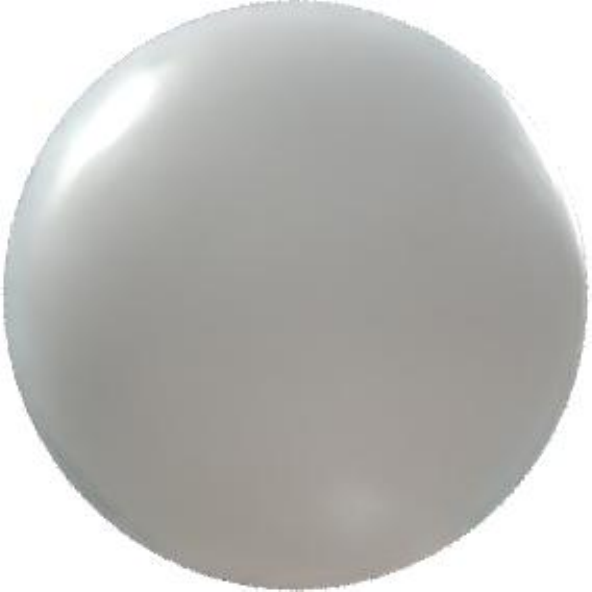}} & 
        \noindent\parbox[c]{0.100\textwidth}{\includegraphics[height=0.100\textwidth]{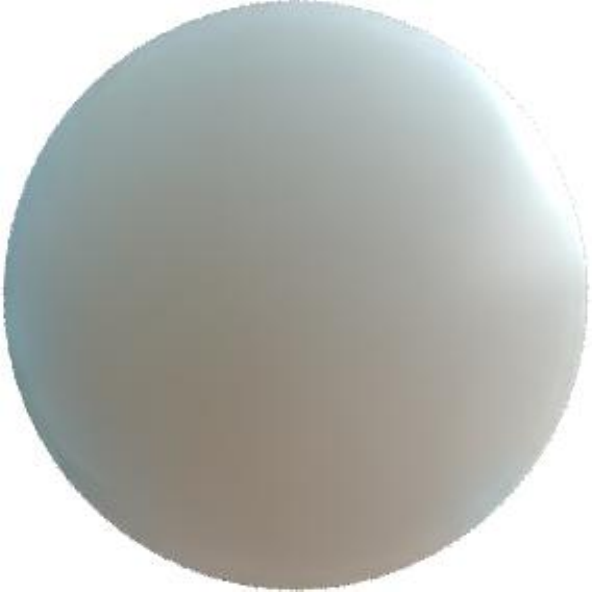}} &
        \noindent\parbox[c]{0.100\textwidth}{\includegraphics[height=0.100\textwidth]{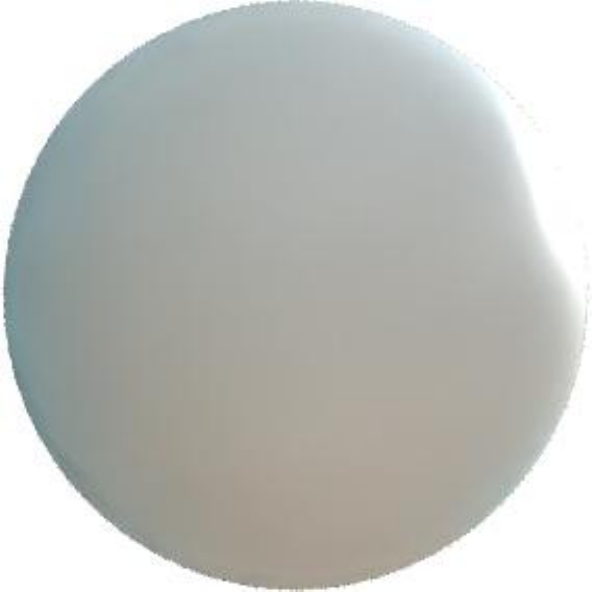}} & 
        \noindent\parbox[c]{0.100\textwidth}{\includegraphics[height=0.100\textwidth]{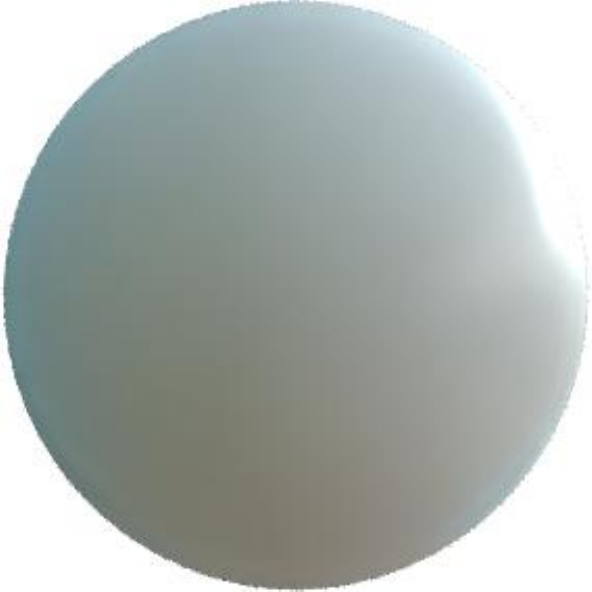}} & 
        \\

        \noindent\parbox[c]{0.205\textwidth}{\includegraphics[height=0.100\textwidth]{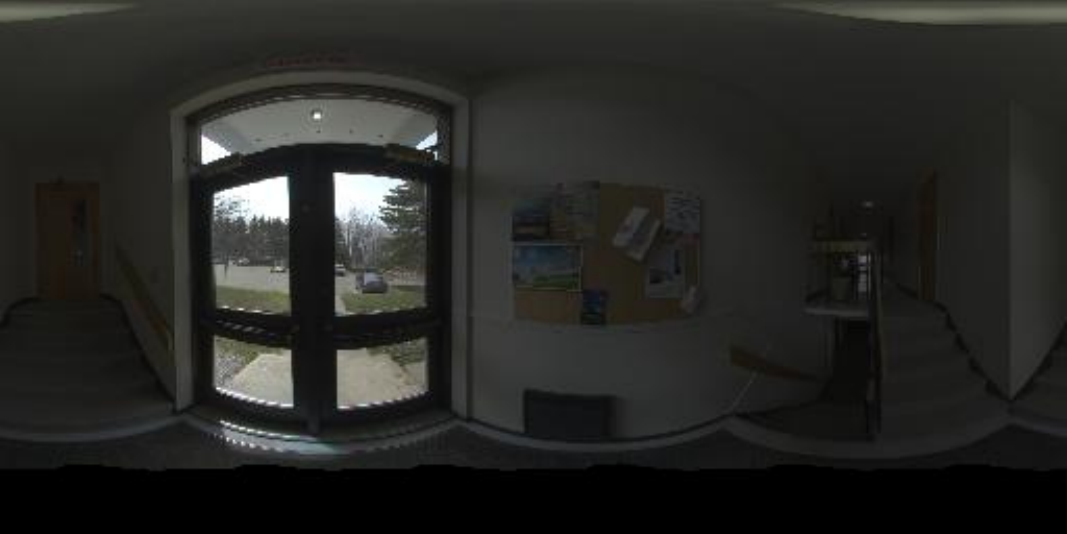}} & 
        \noindent\parbox[c]{0.14\textwidth}{\includegraphics[height=0.100\textwidth]{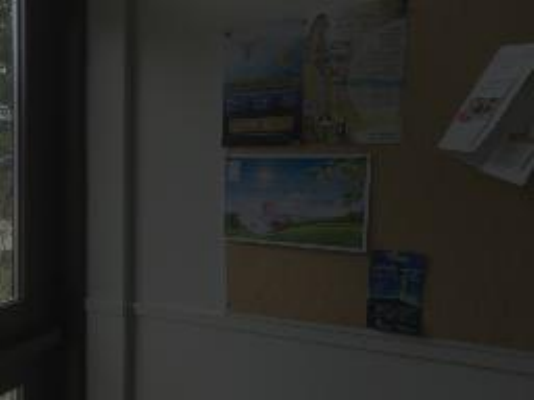}} &  
        
        \noindent\parbox[c]{0.100\textwidth}{\includegraphics[height=0.100\textwidth]{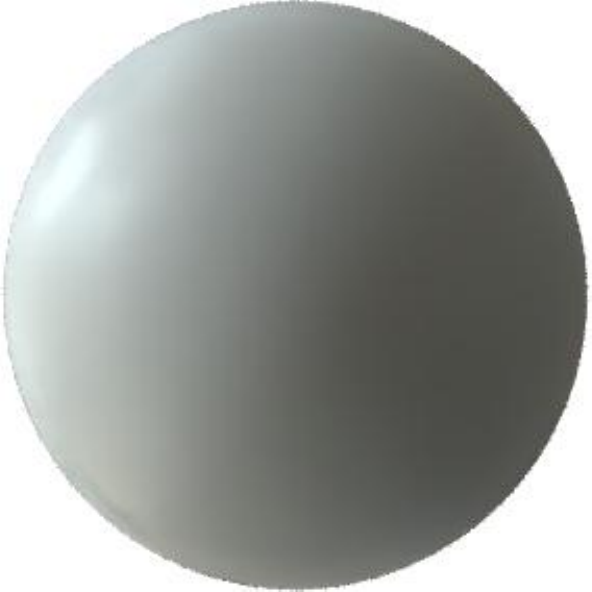}} & 
        \noindent\parbox[c]{0.100\textwidth}{\includegraphics[height=0.100\textwidth]{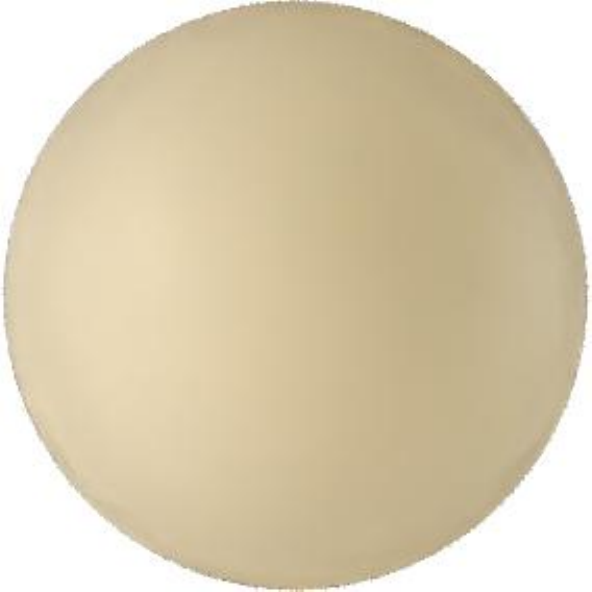}} & 
        
        \noindent\parbox[c]{0.100\textwidth}{\includegraphics[height=0.100\textwidth]{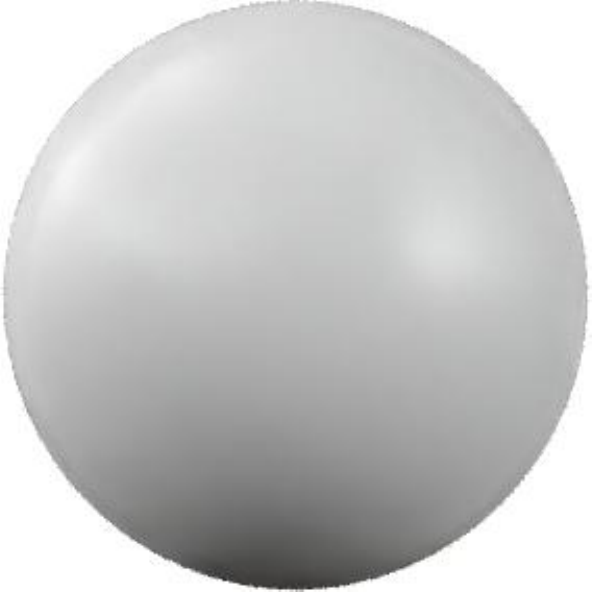}} & 
        \noindent\parbox[c]{0.100\textwidth}{\includegraphics[height=0.100\textwidth]{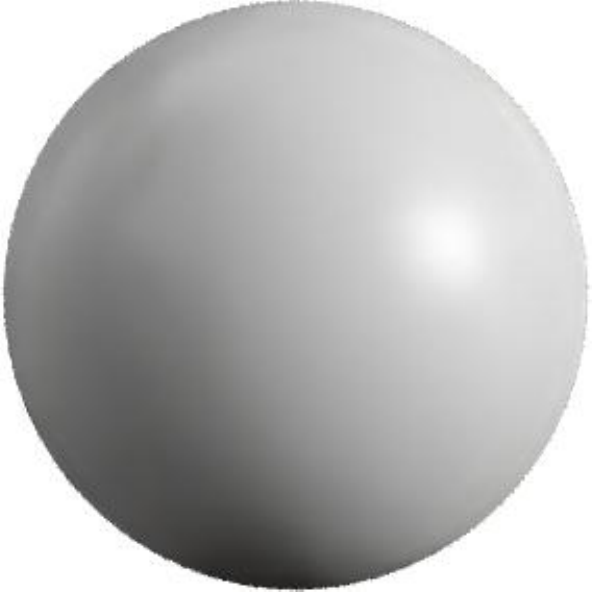}} &
        \noindent\parbox[c]{0.100\textwidth}{\includegraphics[height=0.100\textwidth]{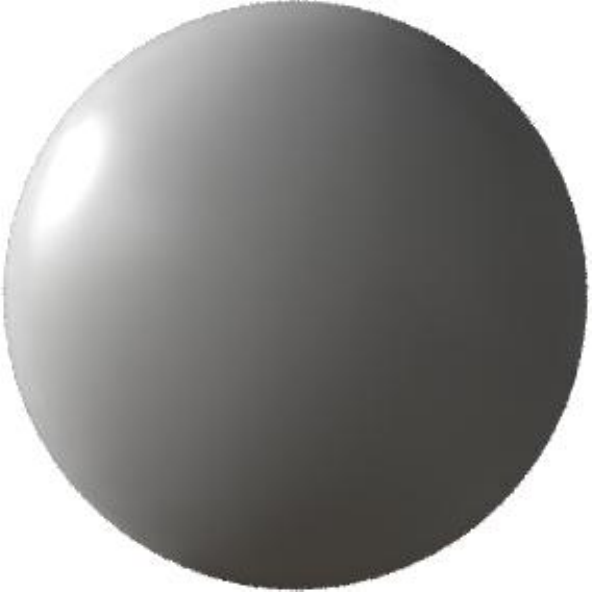}} & 
        \noindent\parbox[c]{0.100\textwidth}{\includegraphics[height=0.100\textwidth]{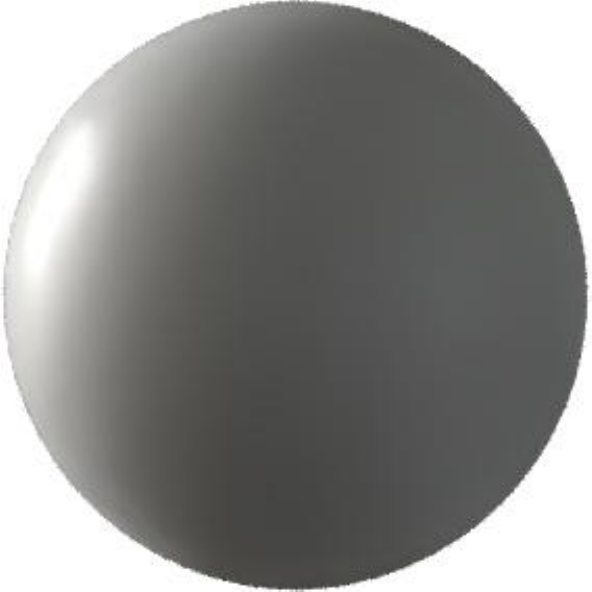}} & 
        \\
        
        \noindent\parbox[c]{0.205\textwidth}{\includegraphics[height=0.100\textwidth]{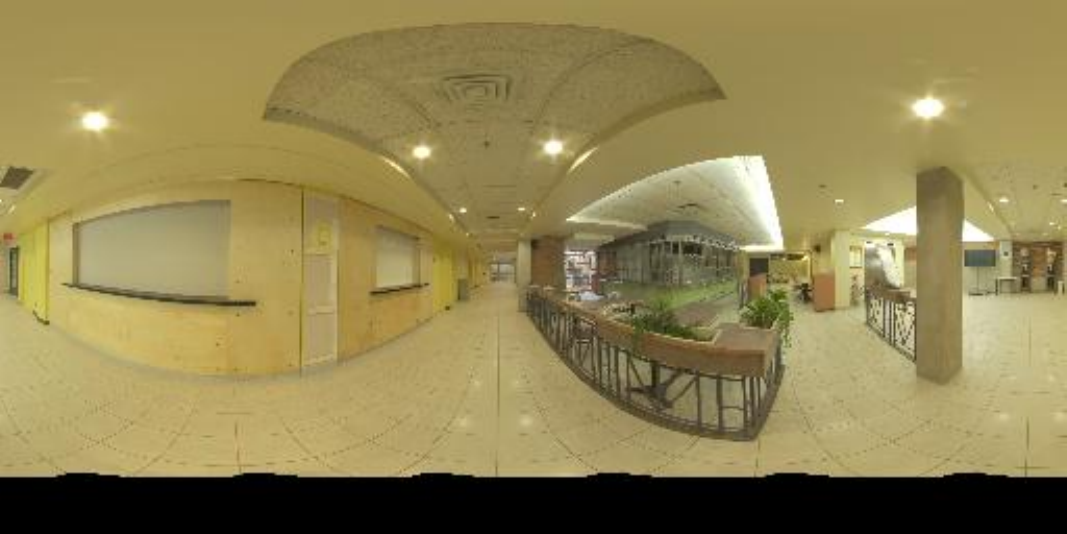}} & 
        \noindent\parbox[c]{0.14\textwidth}{\includegraphics[height=0.100\textwidth]{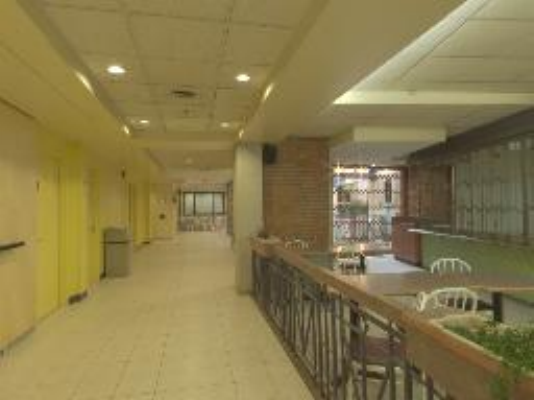}} &  
        
        \noindent\parbox[c]{0.100\textwidth}{\includegraphics[height=0.100\textwidth]{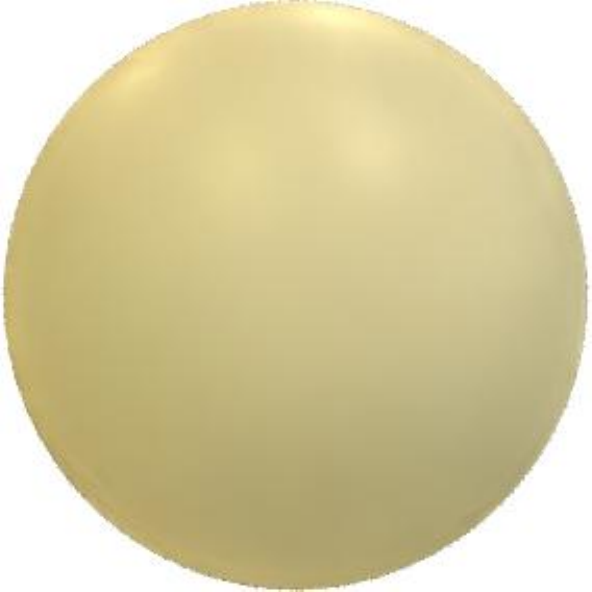}} & 
        \noindent\parbox[c]{0.100\textwidth}{\includegraphics[height=0.100\textwidth]{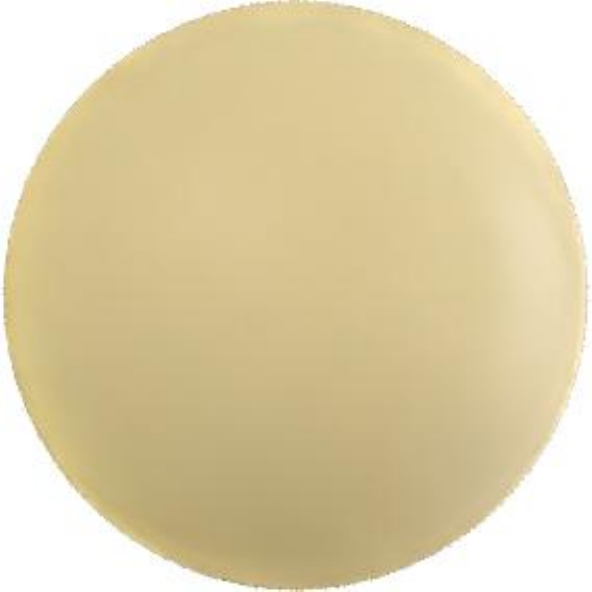}} & 
        
        \noindent\parbox[c]{0.100\textwidth}{\includegraphics[height=0.100\textwidth]{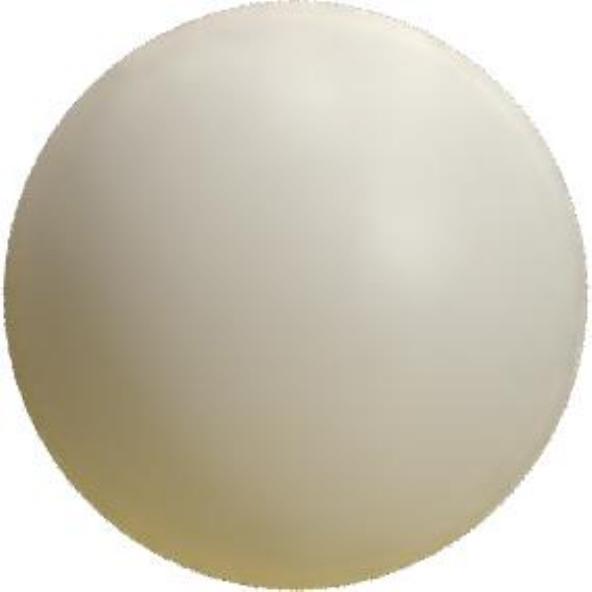}} & 
        \noindent\parbox[c]{0.100\textwidth}{\includegraphics[height=0.100\textwidth]{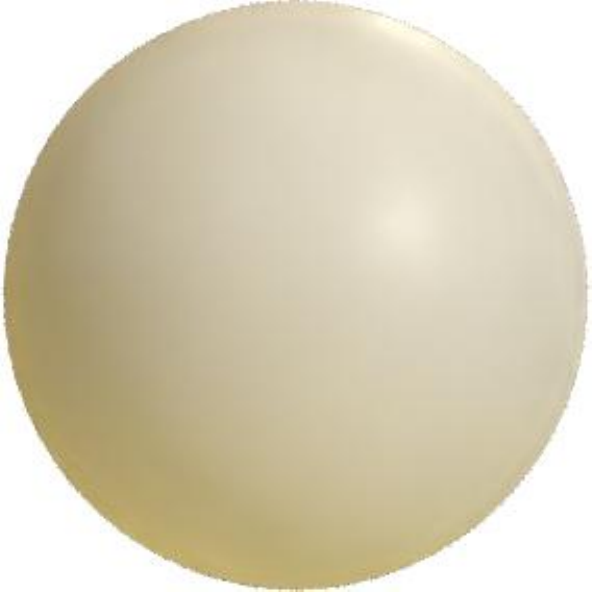}} &
        \noindent\parbox[c]{0.100\textwidth}{\includegraphics[height=0.100\textwidth]{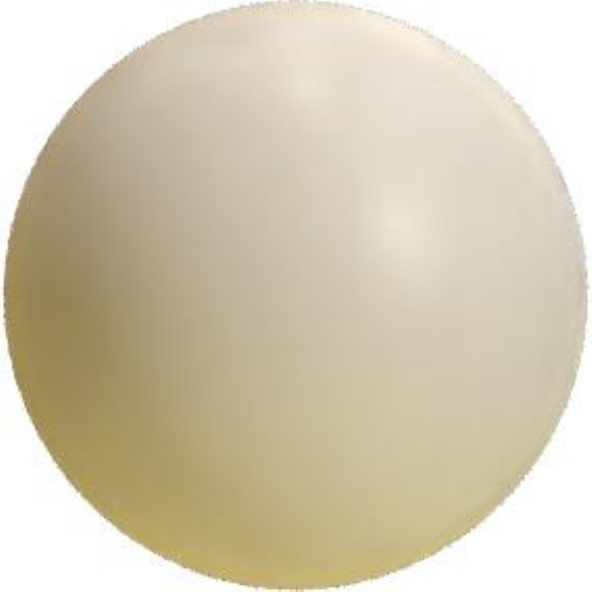}} & 
        \noindent\parbox[c]{0.100\textwidth}{\includegraphics[height=0.100\textwidth]{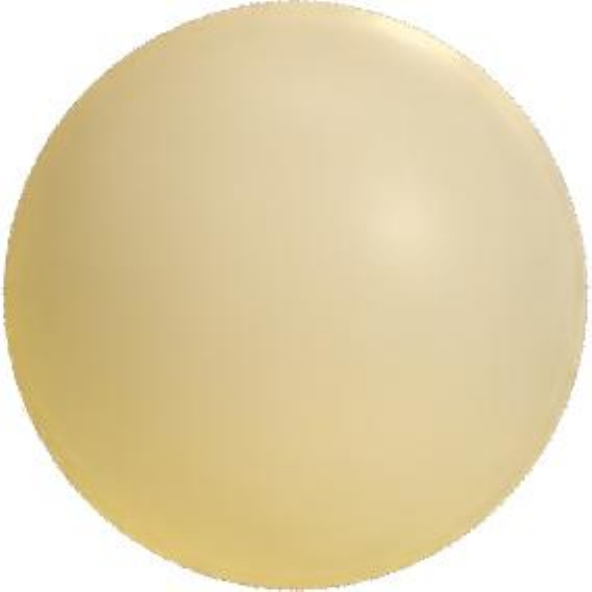}} & 
        \\

        \end{tabu}
    \caption{
    Qualitative results for the Laval indoor dataset using diffuse balls.}
    \label{fig:additional_indoor_diffuse}
\end{figure*}

\begin{figure*}[!t]
    \centering

        \begin{tabu} to \textwidth {
        @{}
        c@{}
        c@{}
        c@{}
        c@{}
        c@{}
        c@{}
        c@{}
        c@{}
        c@{}
    }

        \multicolumn{1}{c}{\shortstack{\scriptsize Ground truth map}}
        & 
        \multicolumn{1}{c}{\shortstack{\hspace{-6pt} \scriptsize Input}}
        &
        \multicolumn{1}{c}{\shortstack{\scriptsize Ground truth}}
        & 
        \multicolumn{1}{c}{\shortstack{\scriptsize StyleLight \cite{wang2022stylelight}}}
        & 
        \multicolumn{1}{c}{\shortstack{\scriptsize SDXL$^\dagger$}} &
        \multicolumn{1}{c}{\shortstack{\scriptsize \begin{tabular}[c]{@{}c@{}}SDXL$^\dagger$+LR \\ (ours, ablated)\end{tabular}}} &
        \multicolumn{1}{c}{\shortstack{\scriptsize \begin{tabular}[c]{@{}c@{}}SDXL$^\dagger$+I \\ (ours,ablated)\end{tabular}}}
        &
        \multicolumn{1}{c}{\shortstack{\scriptsize \begin{tabular}[c]{@{}c@{}}SDXL$^\dagger$+LR+I \\ (ours)\end{tabular}}} 
        \\

        \noindent\parbox[c]{0.205\textwidth}{\includegraphics[height=0.100\textwidth]{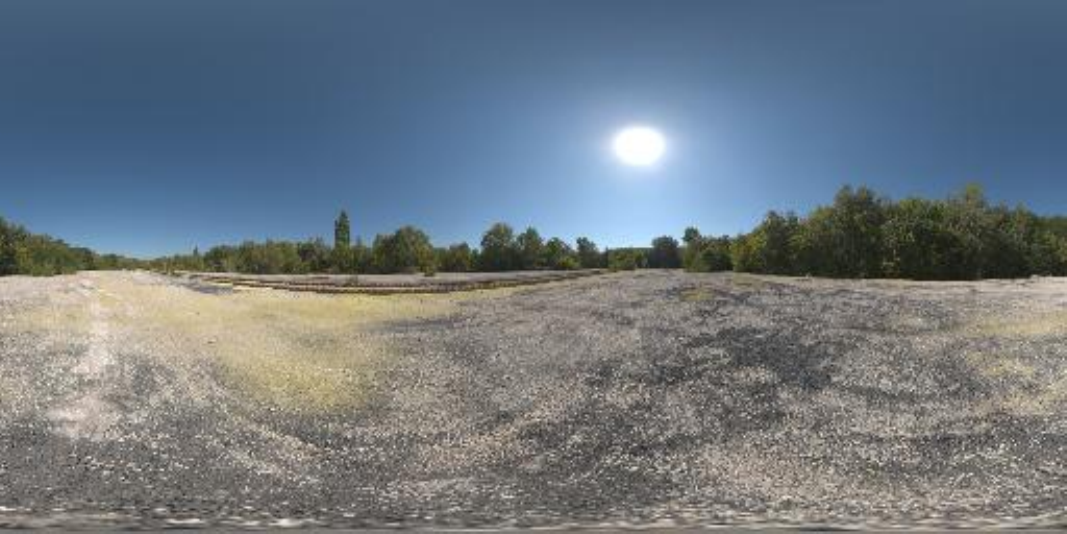}} & 
        \noindent\parbox[c]{0.14\textwidth}{\includegraphics[height=0.100\textwidth]{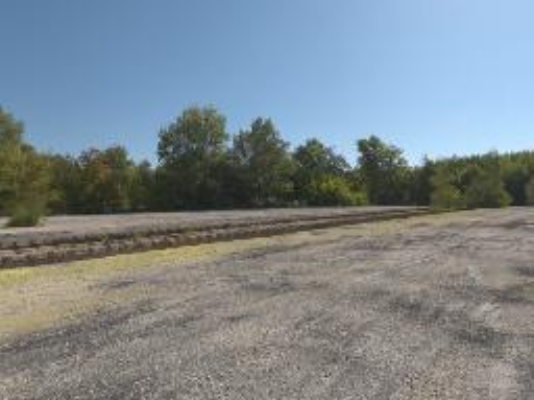}} &  
        
        \noindent\parbox[c]{0.100\textwidth}{\includegraphics[height=0.100\textwidth]{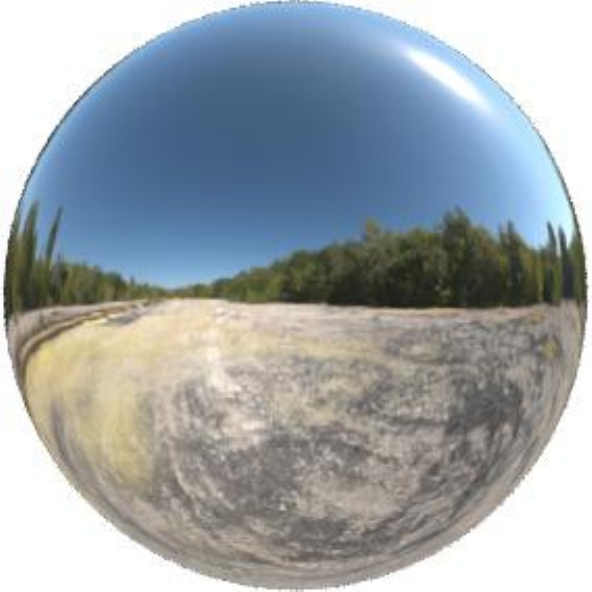}} & 
        \noindent\parbox[c]{0.100\textwidth}{\includegraphics[height=0.100\textwidth]{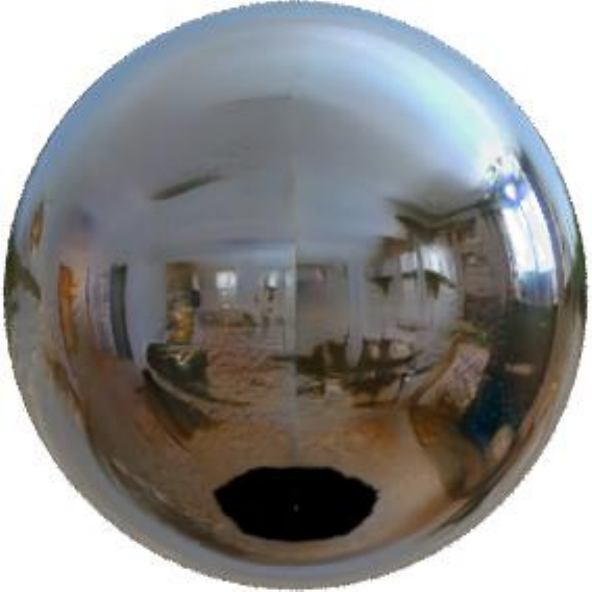}} & 
        
        \noindent\parbox[c]{0.100\textwidth}{\includegraphics[height=0.100\textwidth]{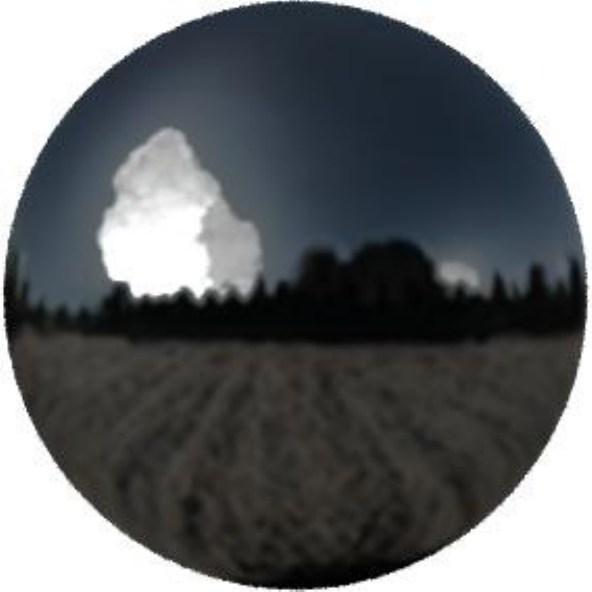}} & 
        \noindent\parbox[c]{0.100\textwidth}{\includegraphics[height=0.100\textwidth]{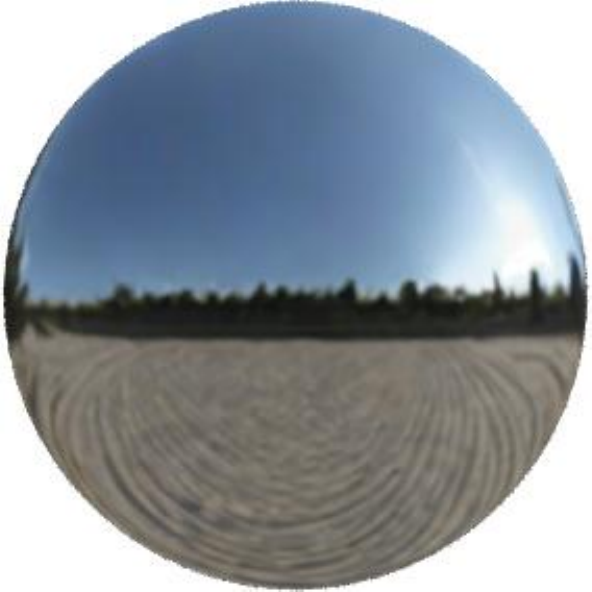}} &
        \noindent\parbox[c]{0.100\textwidth}{\includegraphics[height=0.100\textwidth]{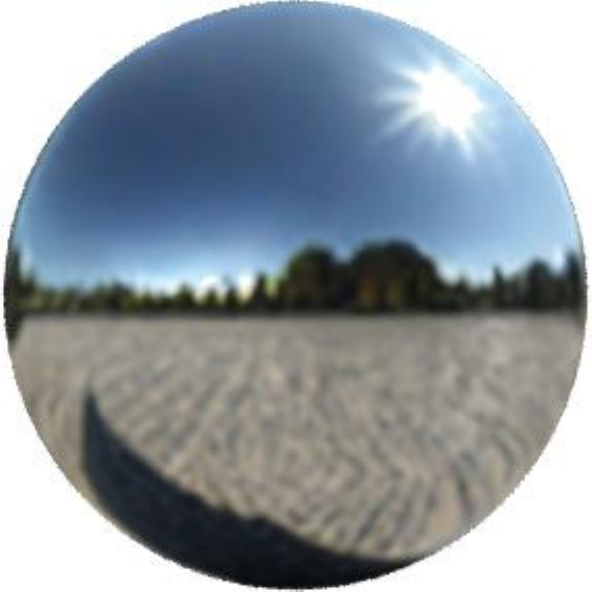}} & 
        \noindent\parbox[c]{0.100\textwidth}{\includegraphics[height=0.100\textwidth]{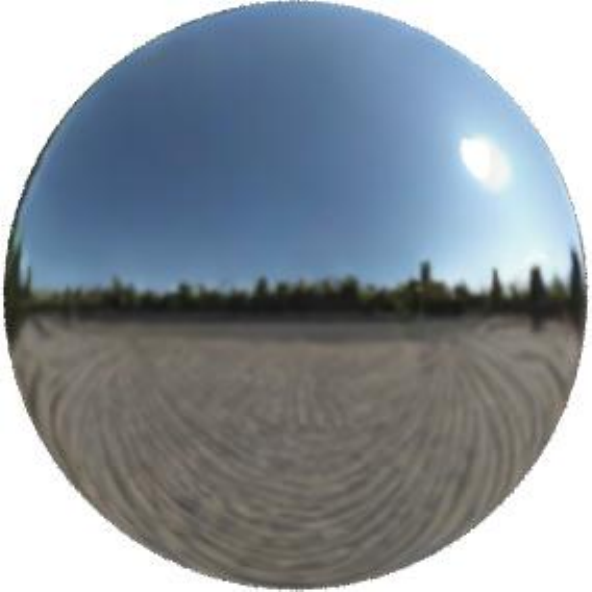}} & 
        \\

        \noindent\parbox[c]{0.205\textwidth}{\includegraphics[height=0.100\textwidth]{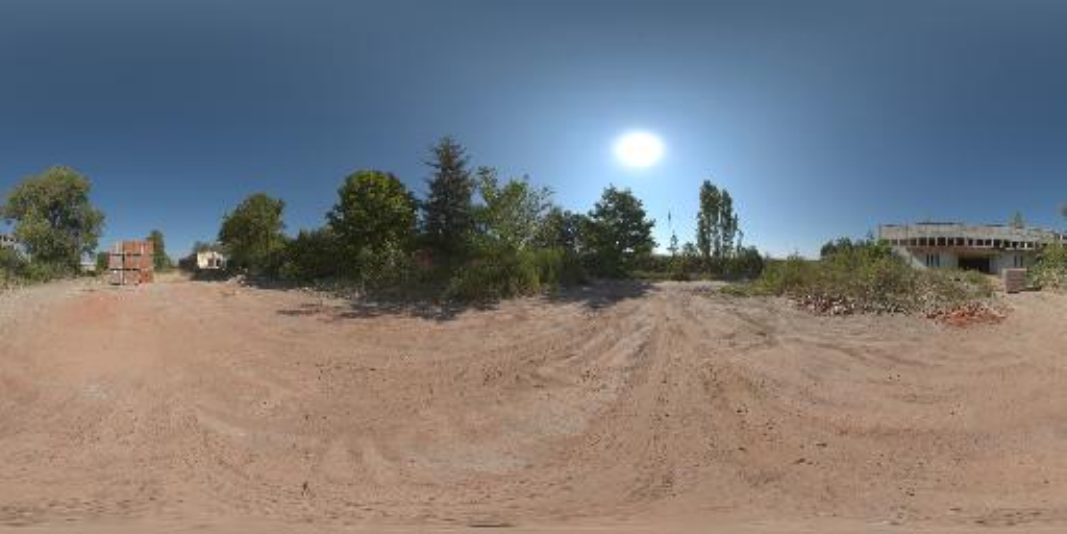}} & 
        \noindent\parbox[c]{0.14\textwidth}{\includegraphics[height=0.100\textwidth]{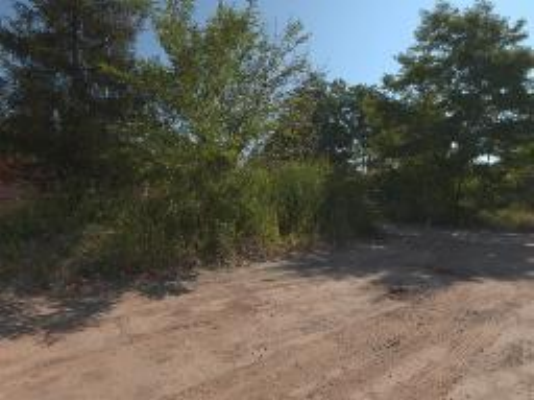}} &  
        
        \noindent\parbox[c]{0.100\textwidth}{\includegraphics[height=0.100\textwidth]{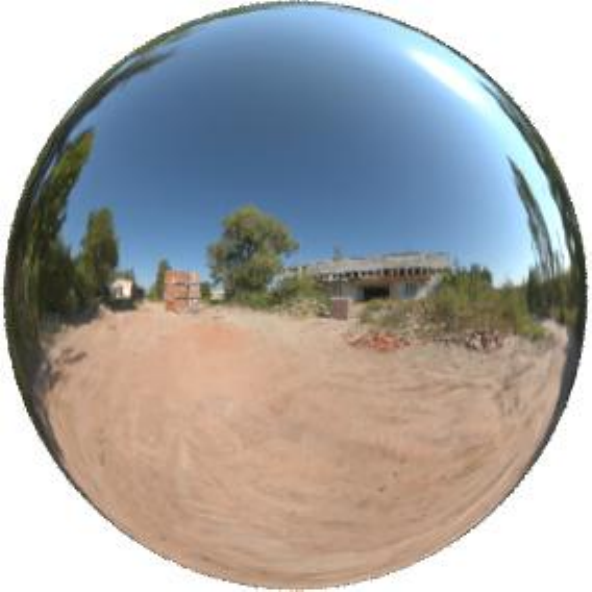}} & 
        \noindent\parbox[c]{0.100\textwidth}{\includegraphics[height=0.100\textwidth]{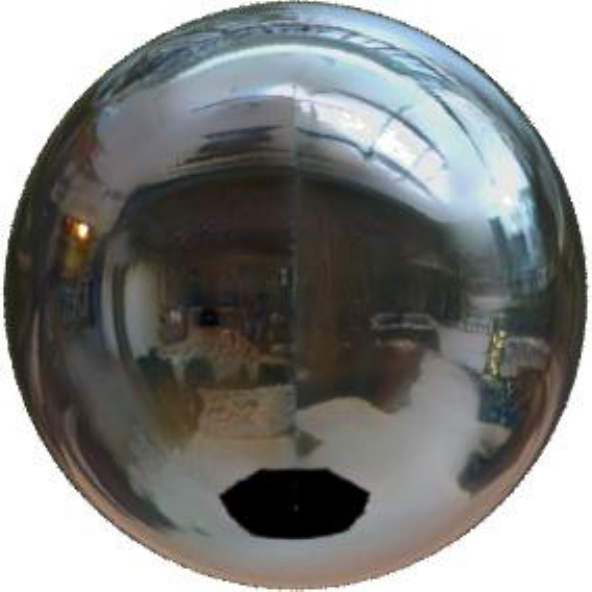}} & 
        
        \noindent\parbox[c]{0.100\textwidth}{\includegraphics[height=0.100\textwidth]{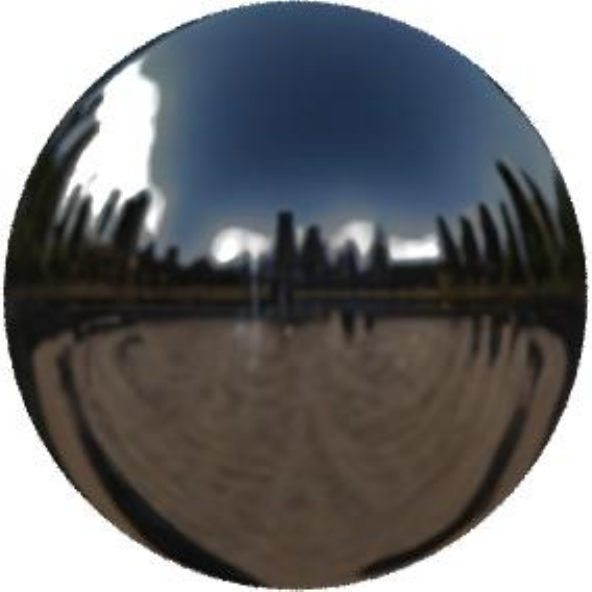}} & 
        \noindent\parbox[c]{0.100\textwidth}{\includegraphics[height=0.100\textwidth]{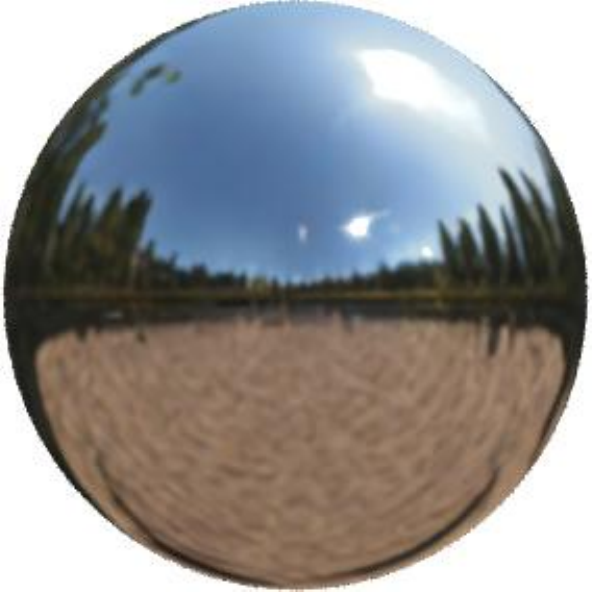}} &
        \noindent\parbox[c]{0.100\textwidth}{\includegraphics[height=0.100\textwidth]{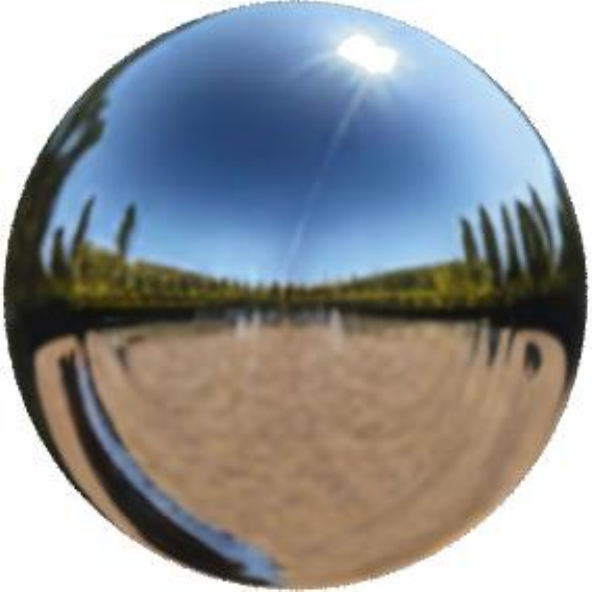}} & 
        \noindent\parbox[c]{0.100\textwidth}{\includegraphics[height=0.100\textwidth]{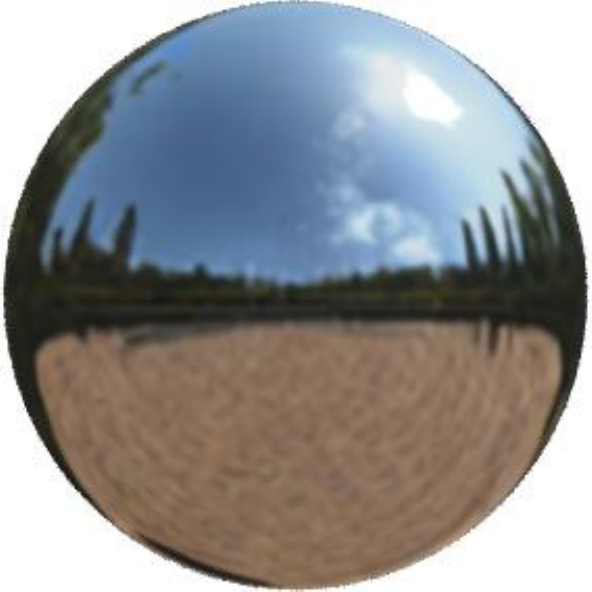}} & 
        \\

        \noindent\parbox[c]{0.205\textwidth}{\includegraphics[height=0.100\textwidth]{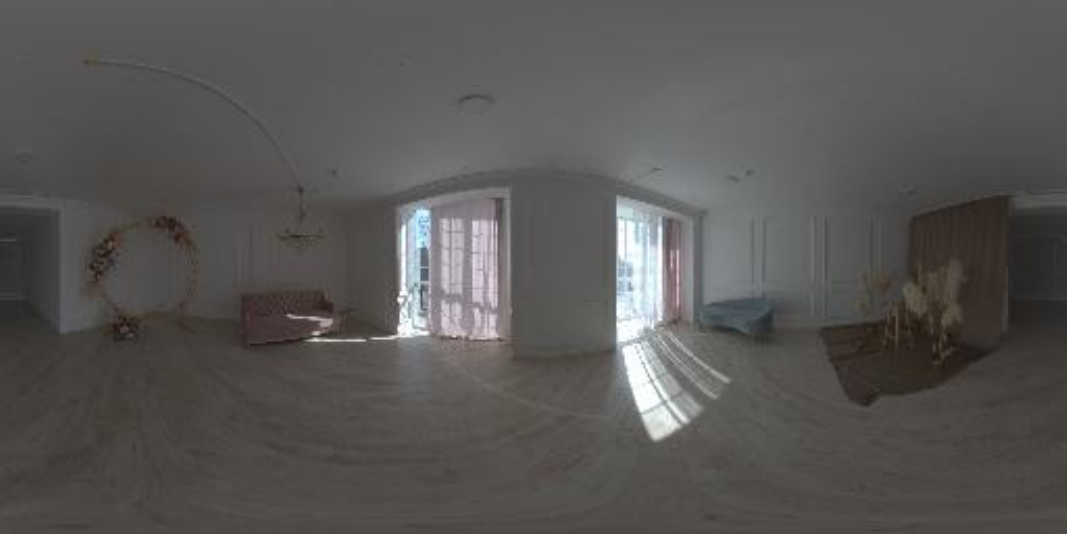}} & 
        \noindent\parbox[c]{0.14\textwidth}{\includegraphics[height=0.100\textwidth]{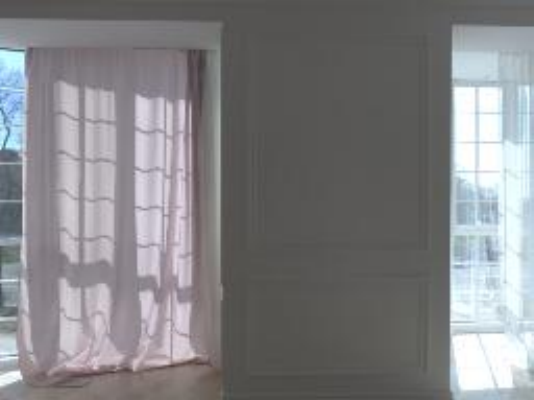}} &  
        
        \noindent\parbox[c]{0.100\textwidth}{\includegraphics[height=0.100\textwidth]{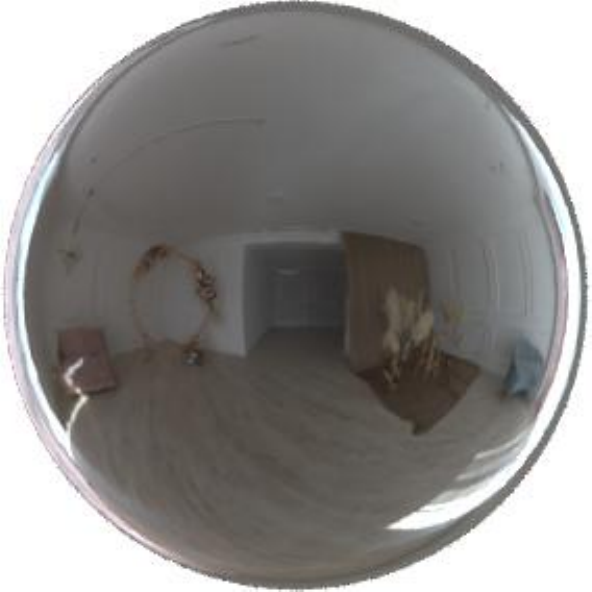}} & 
        \noindent\parbox[c]{0.100\textwidth}{\includegraphics[height=0.100\textwidth]{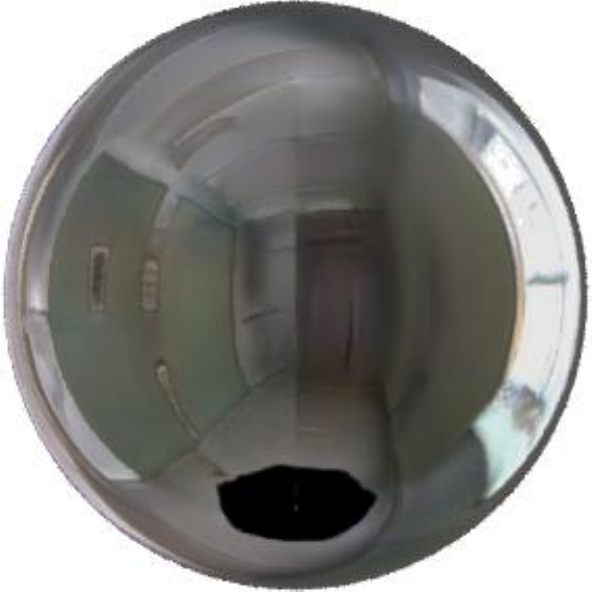}} & 
        
        \noindent\parbox[c]{0.100\textwidth}{\includegraphics[height=0.100\textwidth]{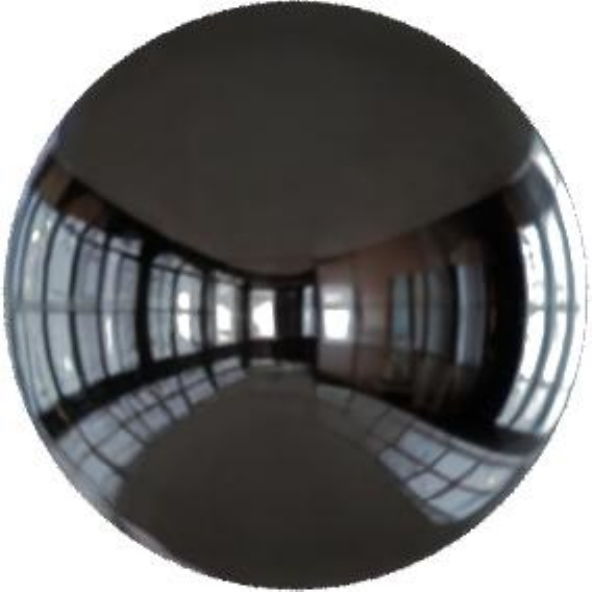}} & 
        \noindent\parbox[c]{0.100\textwidth}{\includegraphics[height=0.100\textwidth]{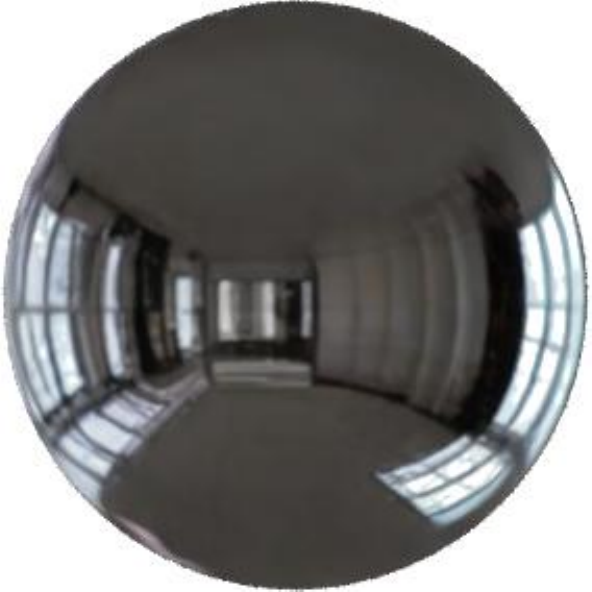}} &
        \noindent\parbox[c]{0.100\textwidth}{\includegraphics[height=0.100\textwidth]{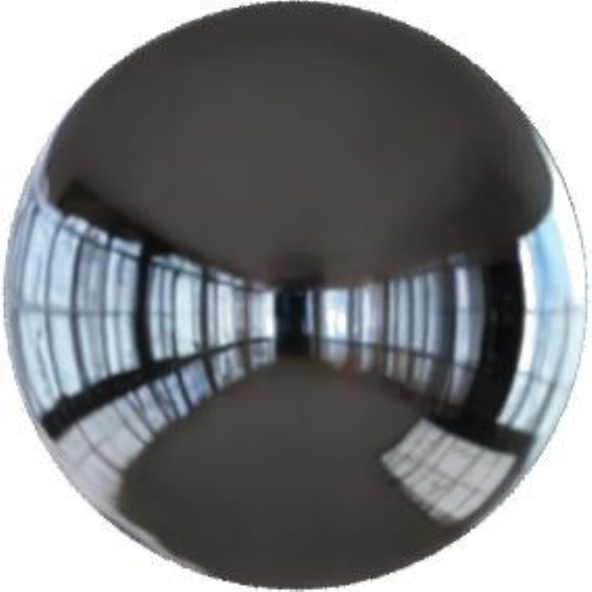}} & 
        \noindent\parbox[c]{0.100\textwidth}{\includegraphics[height=0.100\textwidth]{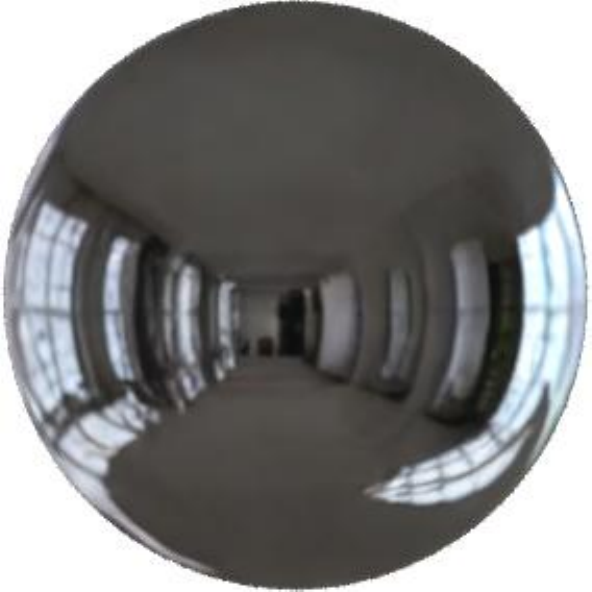}} & 
        \\

        \noindent\parbox[c]{0.205\textwidth}{\includegraphics[height=0.100\textwidth]{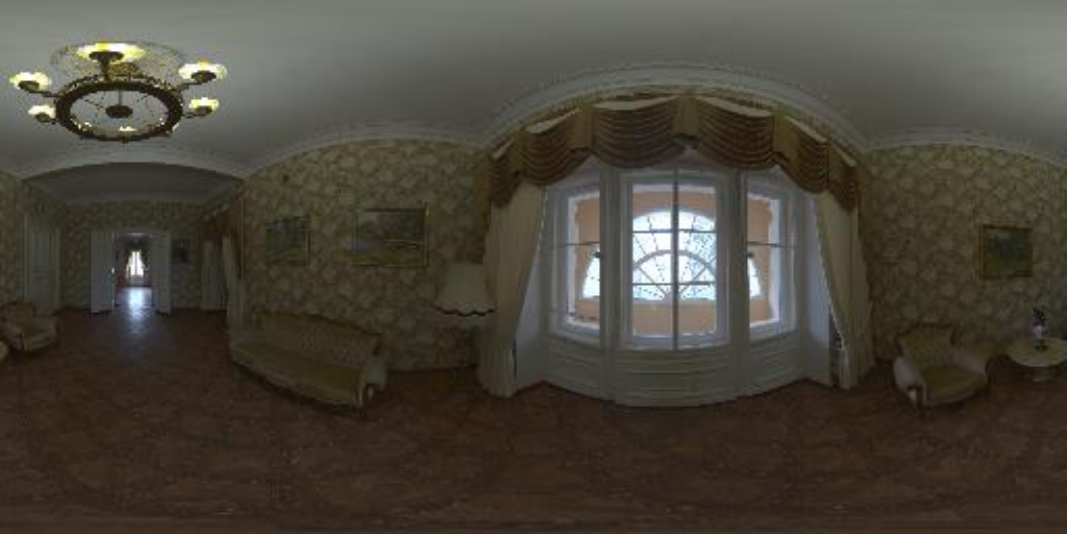}} & 
        \noindent\parbox[c]{0.14\textwidth}{\includegraphics[height=0.100\textwidth]{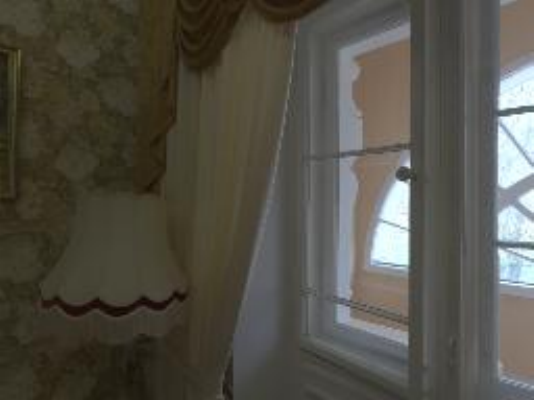}} &  
        
        \noindent\parbox[c]{0.100\textwidth}{\includegraphics[height=0.100\textwidth]{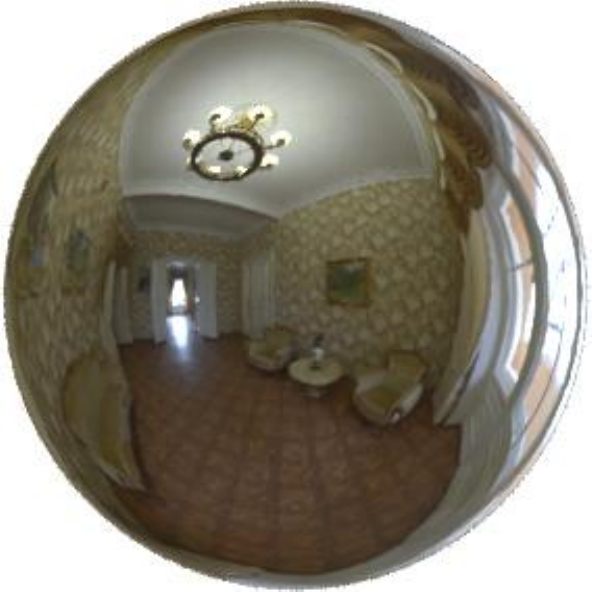}} & 
        \noindent\parbox[c]{0.100\textwidth}{\includegraphics[height=0.100\textwidth]{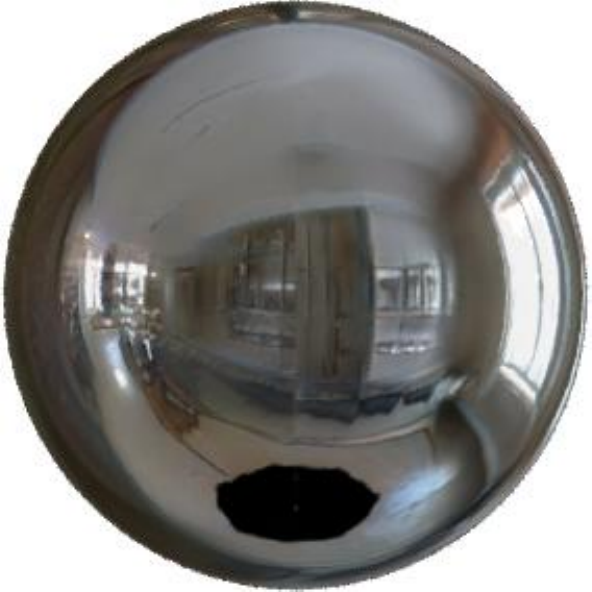}} & 
        
        \noindent\parbox[c]{0.100\textwidth}{\includegraphics[height=0.100\textwidth]{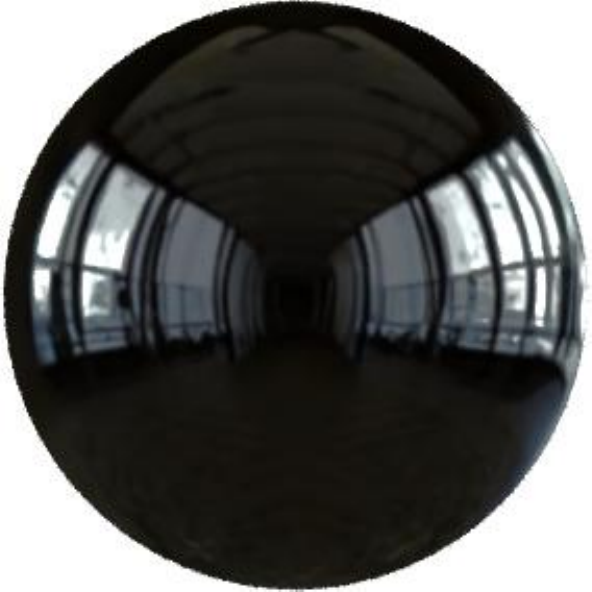}} & 
        \noindent\parbox[c]{0.100\textwidth}{\includegraphics[height=0.100\textwidth]{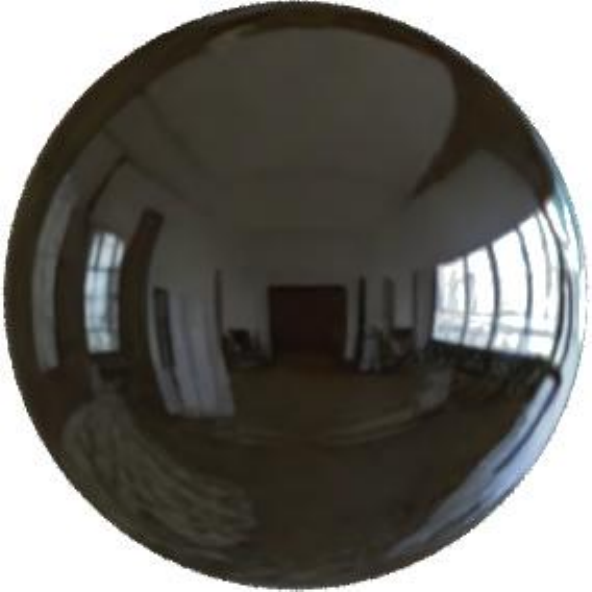}} &
        \noindent\parbox[c]{0.100\textwidth}{\includegraphics[height=0.100\textwidth]{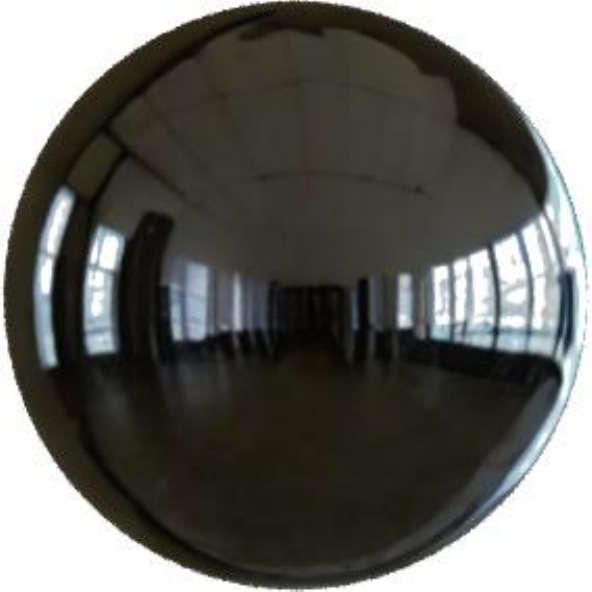}} & 
        \noindent\parbox[c]{0.100\textwidth}{\includegraphics[height=0.100\textwidth]{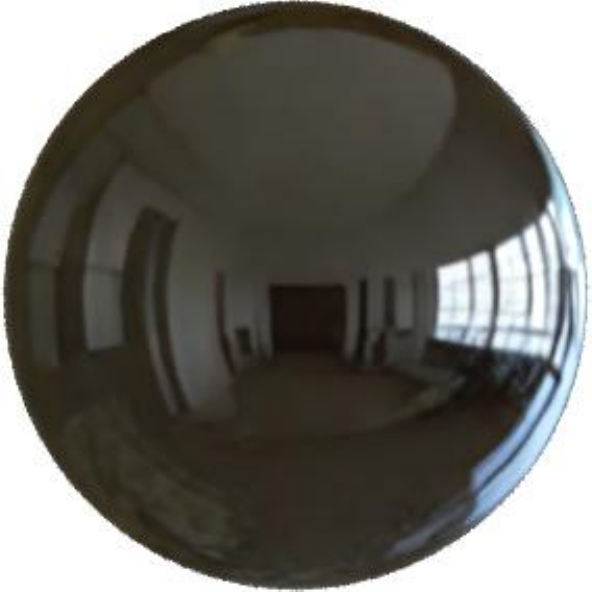}} & 
        \\

        \noindent\parbox[c]{0.205\textwidth}{\includegraphics[height=0.100\textwidth]{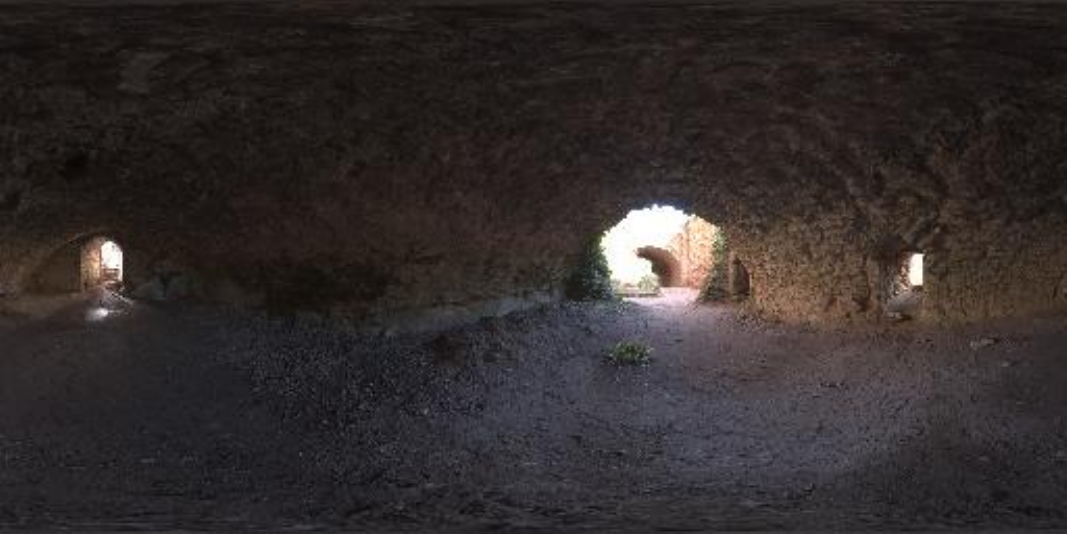}} & 
        \noindent\parbox[c]{0.14\textwidth}{\includegraphics[height=0.100\textwidth]{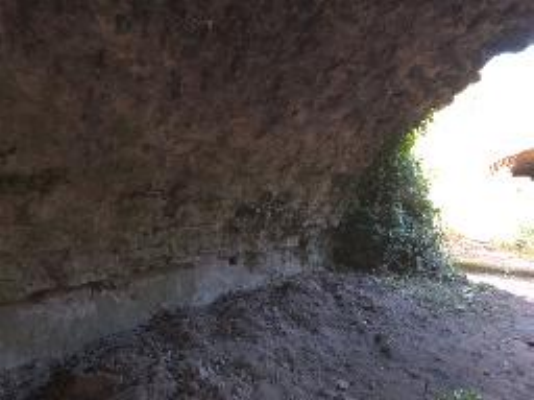}} &  
        
        \noindent\parbox[c]{0.100\textwidth}{\includegraphics[height=0.100\textwidth]{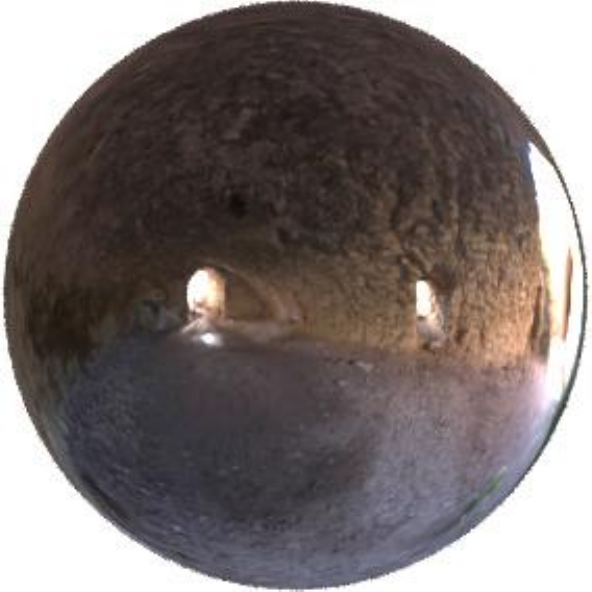}} & 
        \noindent\parbox[c]{0.100\textwidth}{\includegraphics[height=0.100\textwidth]{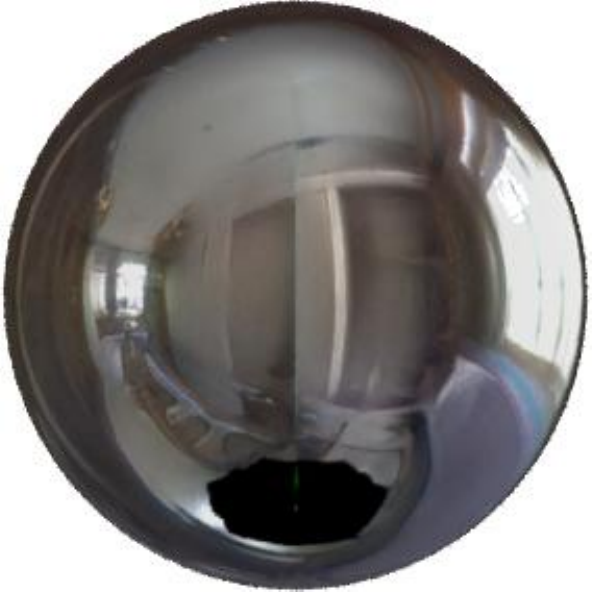}} & 
        
        \noindent\parbox[c]{0.100\textwidth}{\includegraphics[height=0.100\textwidth]{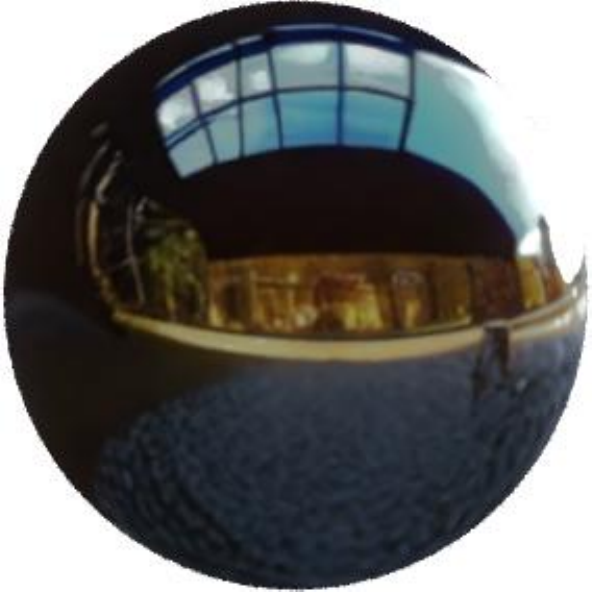}} & 
        \noindent\parbox[c]{0.100\textwidth}{\includegraphics[height=0.100\textwidth]{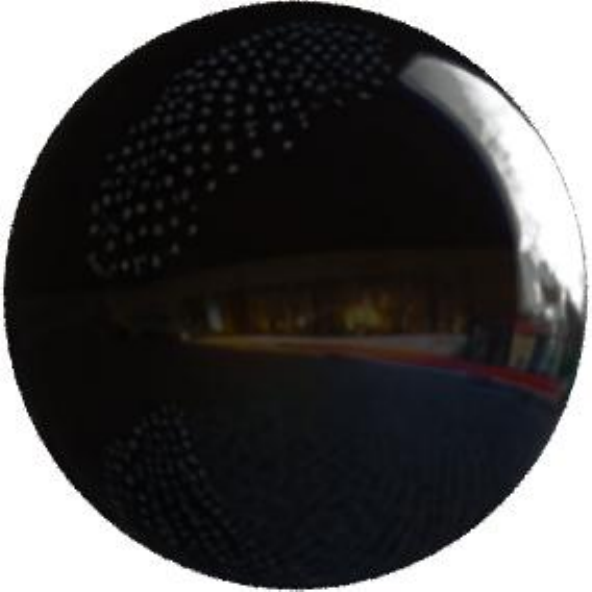}} &
        \noindent\parbox[c]{0.100\textwidth}{\includegraphics[height=0.100\textwidth]{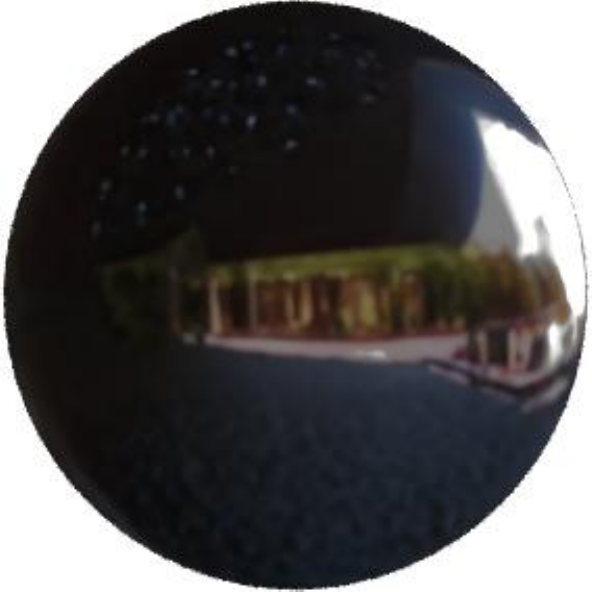}} & 
        \noindent\parbox[c]{0.100\textwidth}{\includegraphics[height=0.100\textwidth]{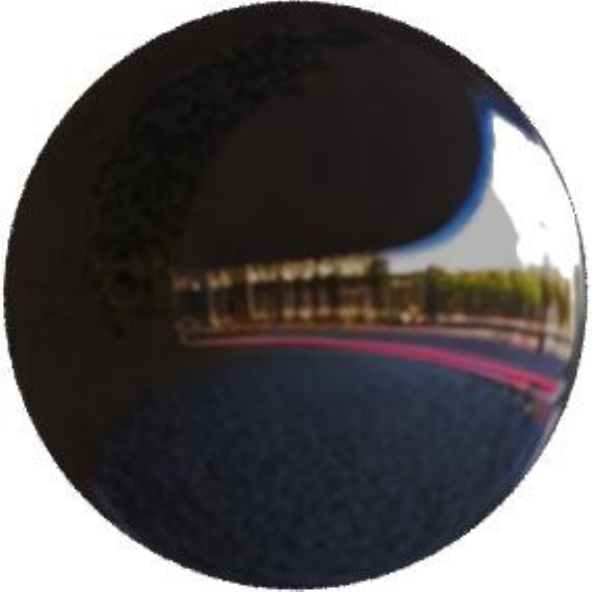}} & 
        \\

        \noindent\parbox[c]{0.205\textwidth}{\includegraphics[height=0.100\textwidth]{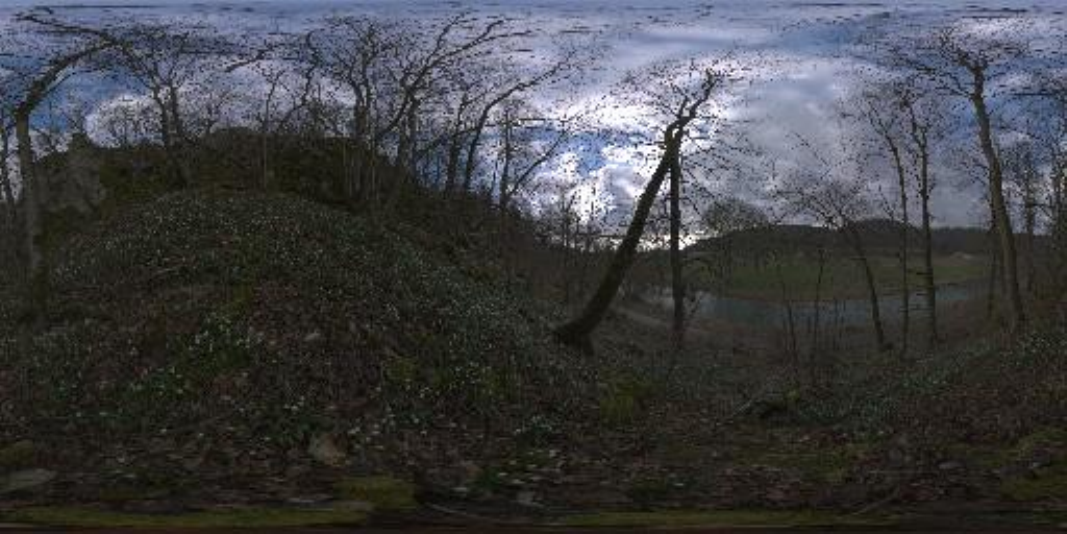}} & 
        \noindent\parbox[c]{0.14\textwidth}{\includegraphics[height=0.100\textwidth]{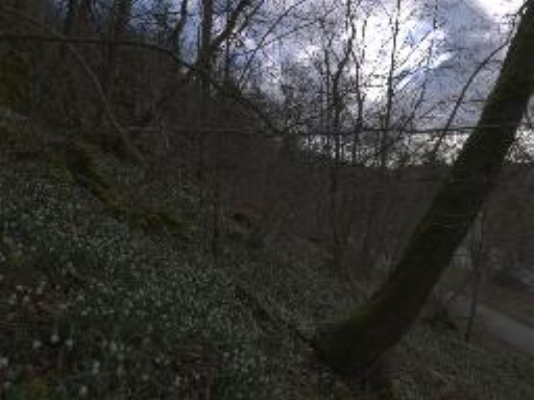}} &  
        
        \noindent\parbox[c]{0.100\textwidth}{\includegraphics[height=0.100\textwidth]{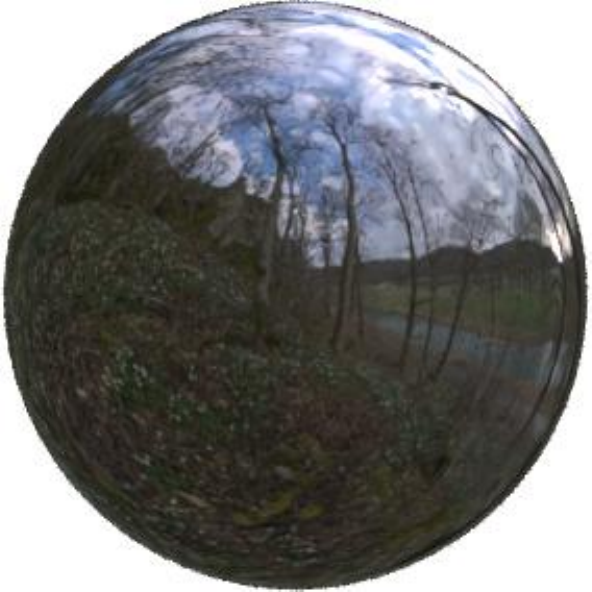}} & 
        \noindent\parbox[c]{0.100\textwidth}{\includegraphics[height=0.100\textwidth]{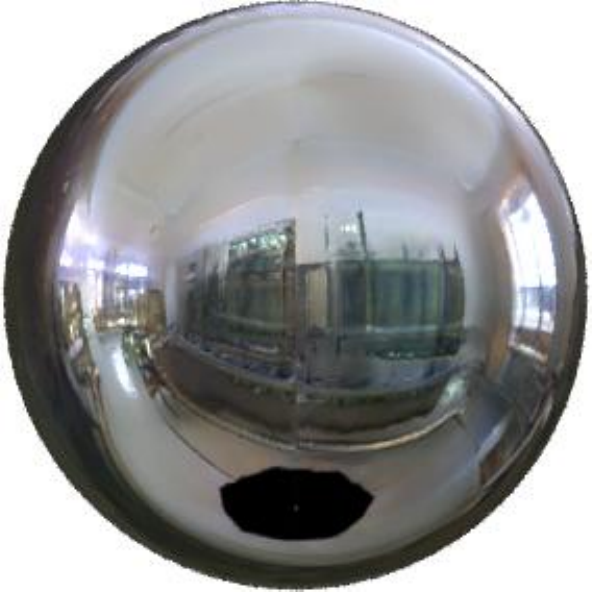}} & 
        
        \noindent\parbox[c]{0.100\textwidth}{\includegraphics[height=0.100\textwidth]{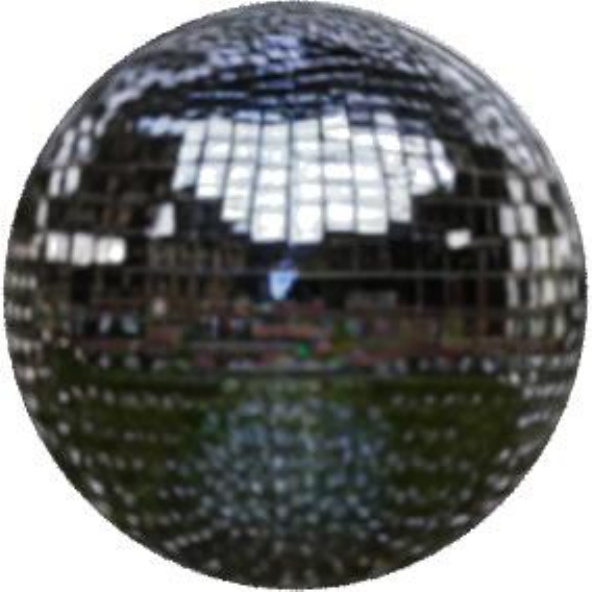}} & 
        \noindent\parbox[c]{0.100\textwidth}{\includegraphics[height=0.100\textwidth]{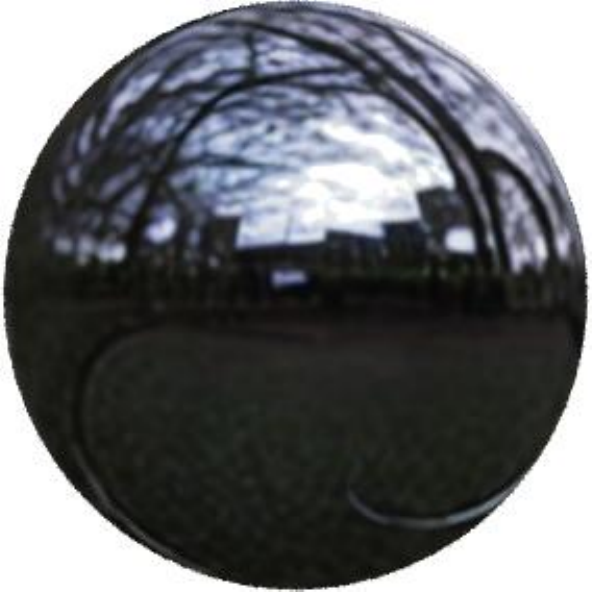}} &
        \noindent\parbox[c]{0.100\textwidth}{\includegraphics[height=0.100\textwidth]{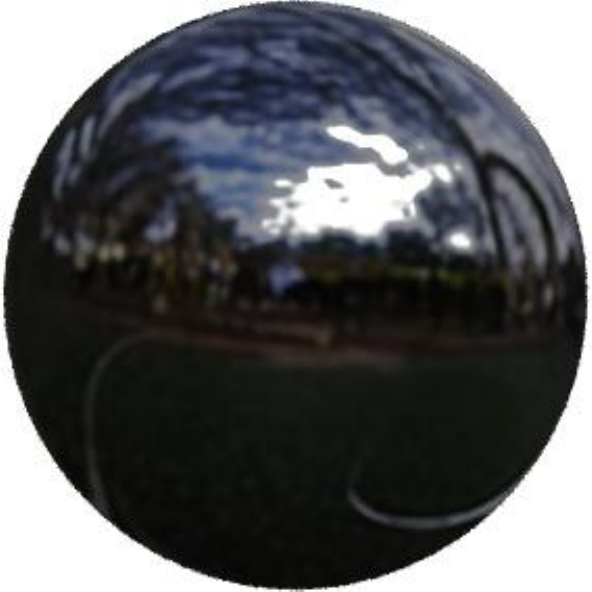}} & 
        \noindent\parbox[c]{0.100\textwidth}{\includegraphics[height=0.100\textwidth]{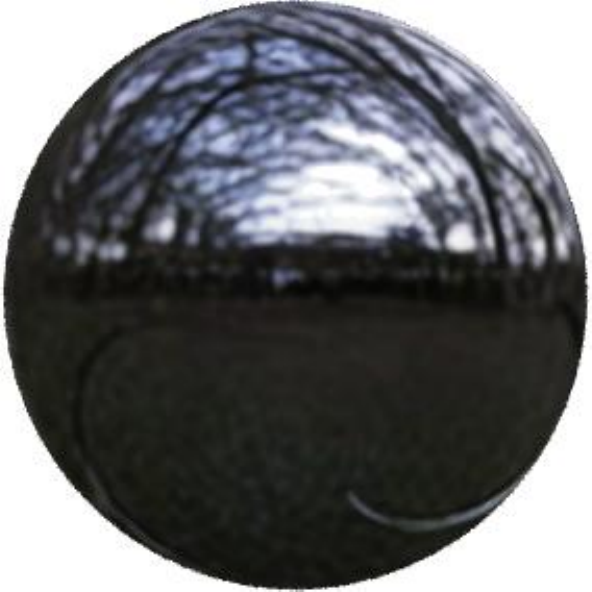}} & 
        \\

        \noindent\parbox[c]{0.205\textwidth}{\includegraphics[height=0.100\textwidth]{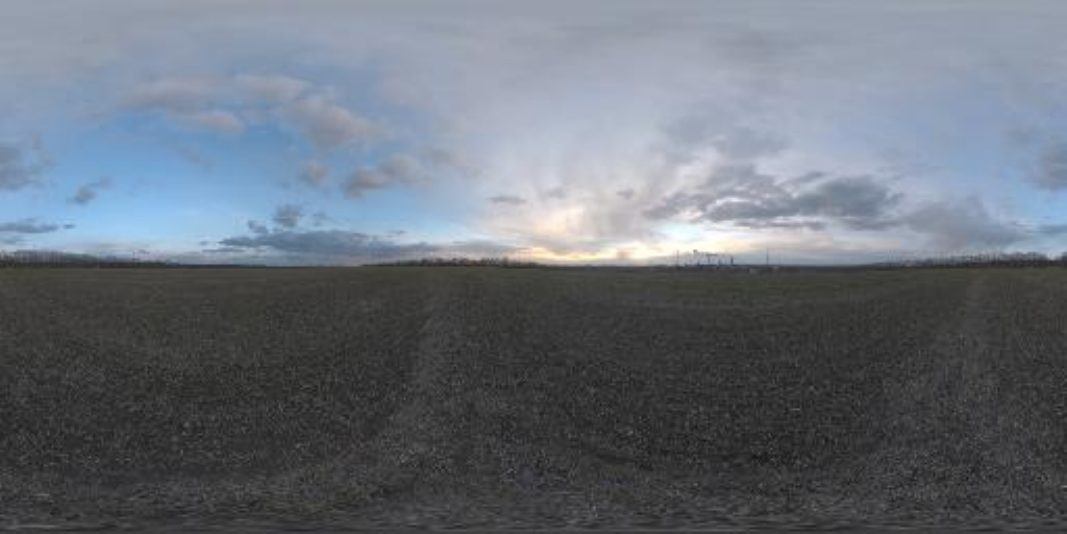}} & 
        \noindent\parbox[c]{0.14\textwidth}{\includegraphics[height=0.100\textwidth]{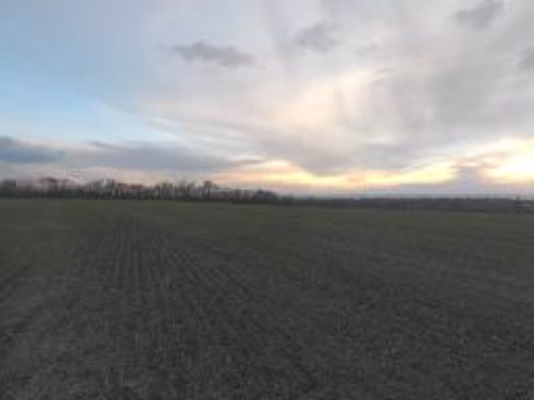}} &  
        
        \noindent\parbox[c]{0.100\textwidth}{\includegraphics[height=0.100\textwidth]{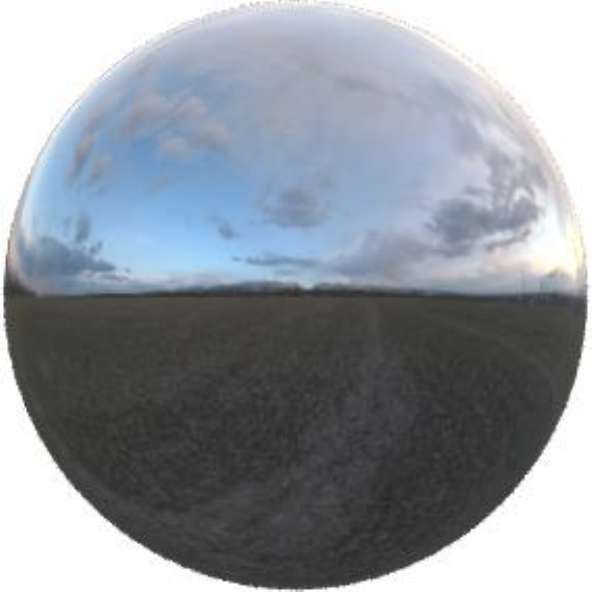}} & 
        \noindent\parbox[c]{0.100\textwidth}{\includegraphics[height=0.100\textwidth]{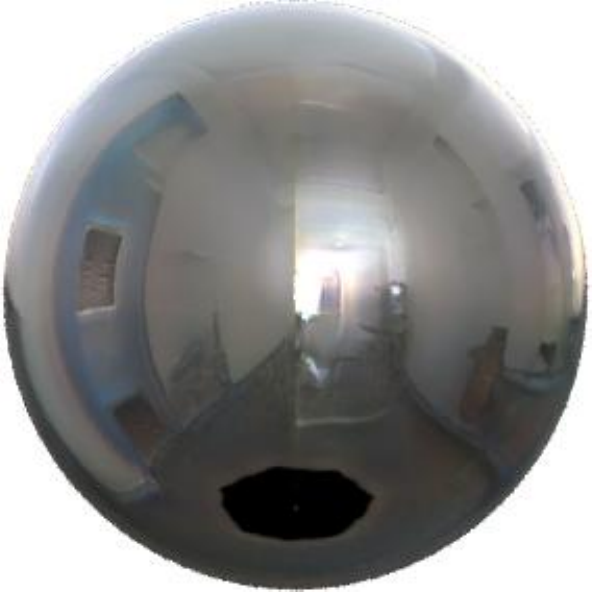}} & 
        
        \noindent\parbox[c]{0.100\textwidth}{\includegraphics[height=0.100\textwidth]{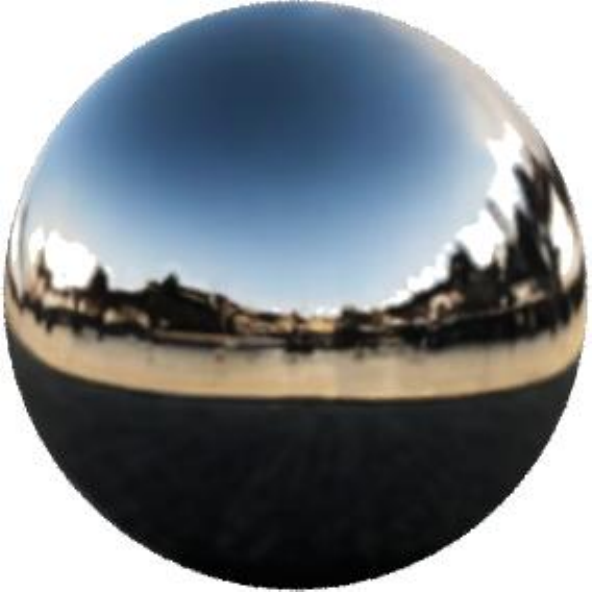}} & 
        \noindent\parbox[c]{0.100\textwidth}{\includegraphics[height=0.100\textwidth]{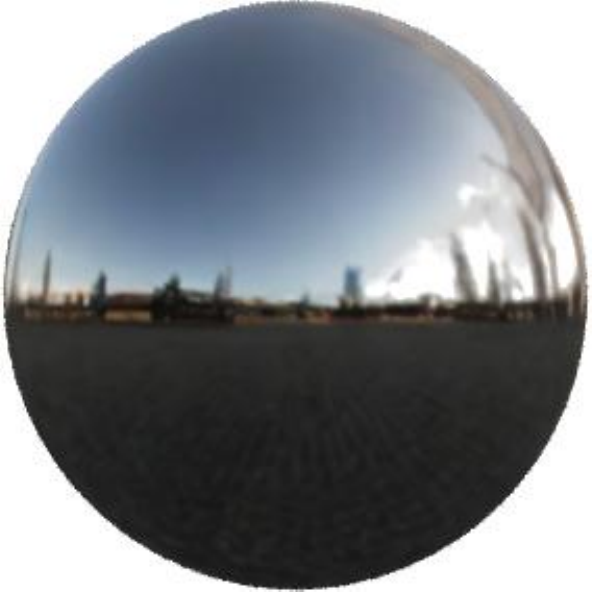}} &
        \noindent\parbox[c]{0.100\textwidth}{\includegraphics[height=0.100\textwidth]{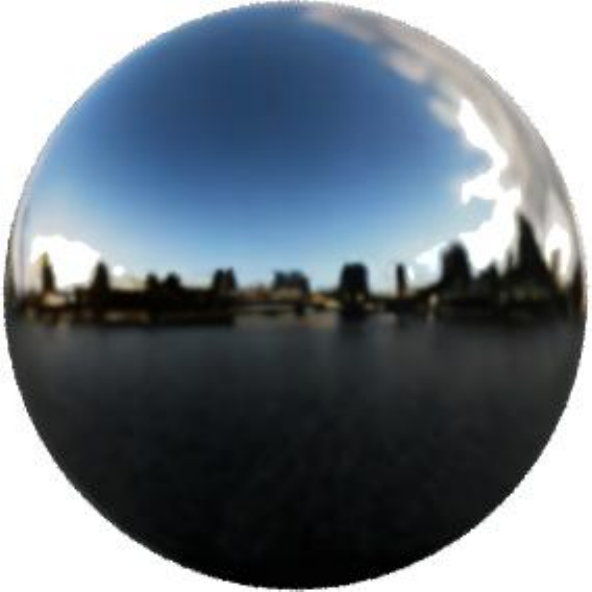}} & 
        \noindent\parbox[c]{0.100\textwidth}{\includegraphics[height=0.100\textwidth]{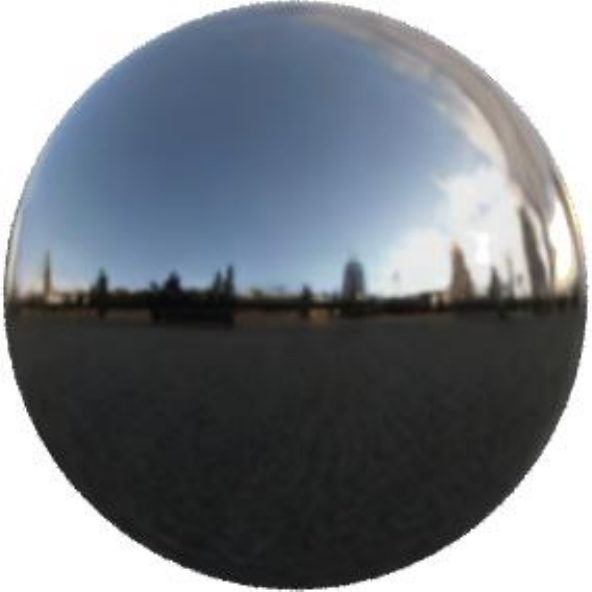}} & 
        \\

        \noindent\parbox[c]{0.205\textwidth}{\includegraphics[height=0.100\textwidth]{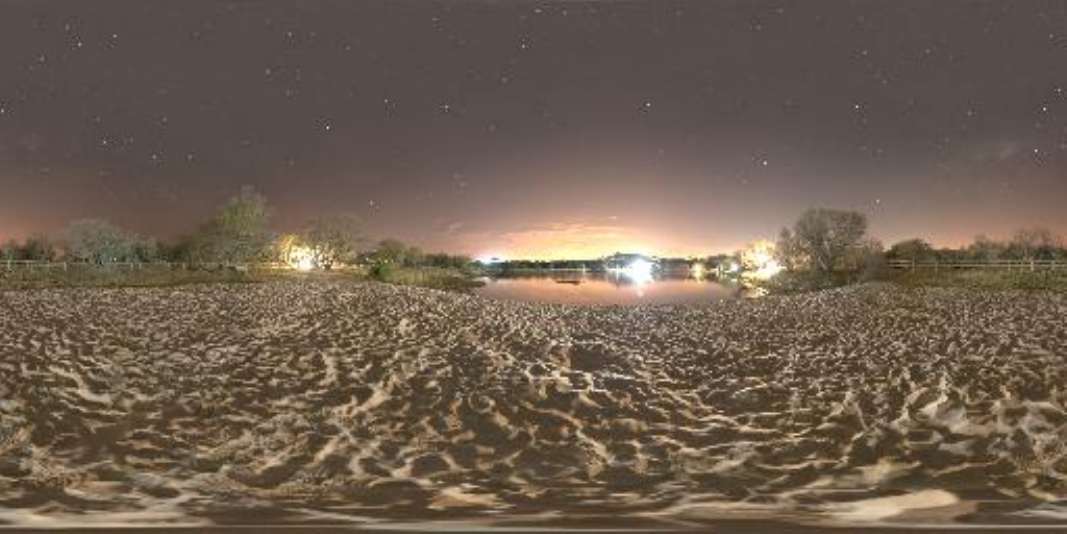}} & 
        \noindent\parbox[c]{0.14\textwidth}{\includegraphics[height=0.100\textwidth]{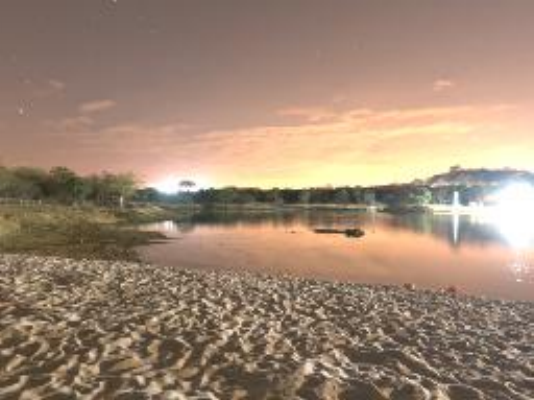}} &  
        
        \noindent\parbox[c]{0.100\textwidth}{\includegraphics[height=0.100\textwidth]{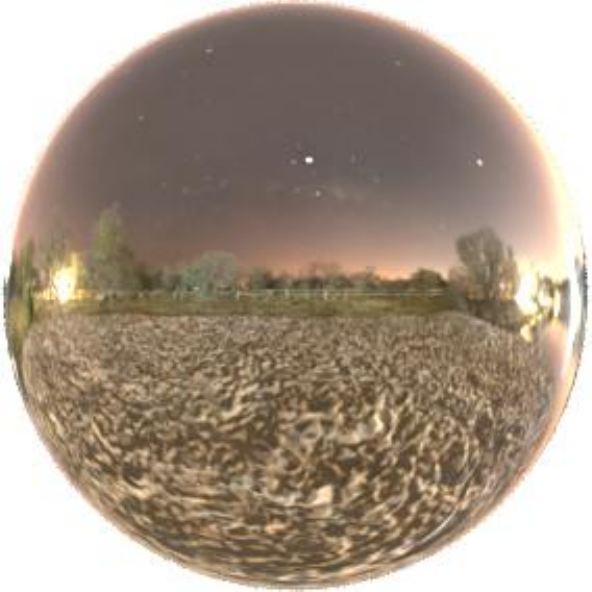}} & 
        \noindent\parbox[c]{0.100\textwidth}{\includegraphics[height=0.100\textwidth]{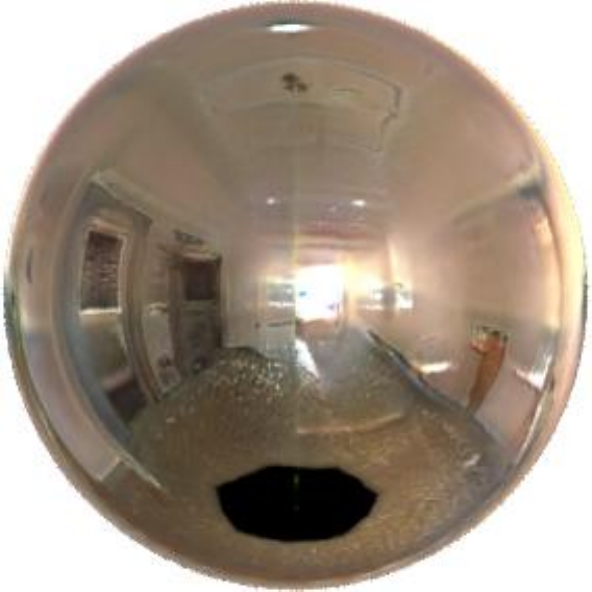}} & 
        
        \noindent\parbox[c]{0.100\textwidth}{\includegraphics[height=0.100\textwidth]{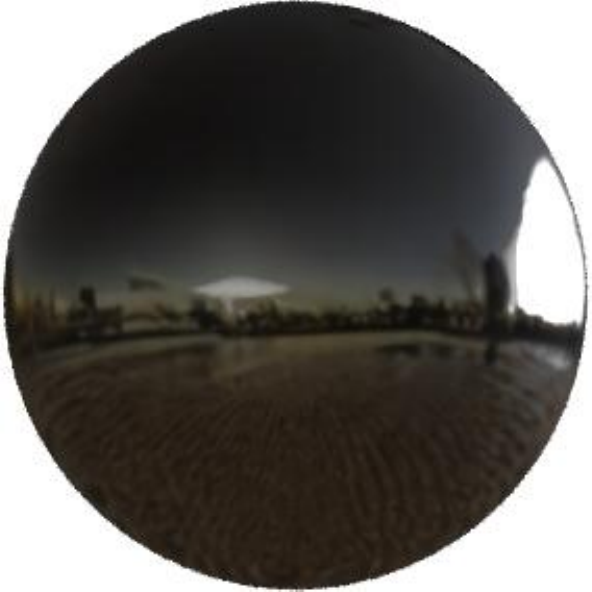}} & 
        \noindent\parbox[c]{0.100\textwidth}{\includegraphics[height=0.100\textwidth]{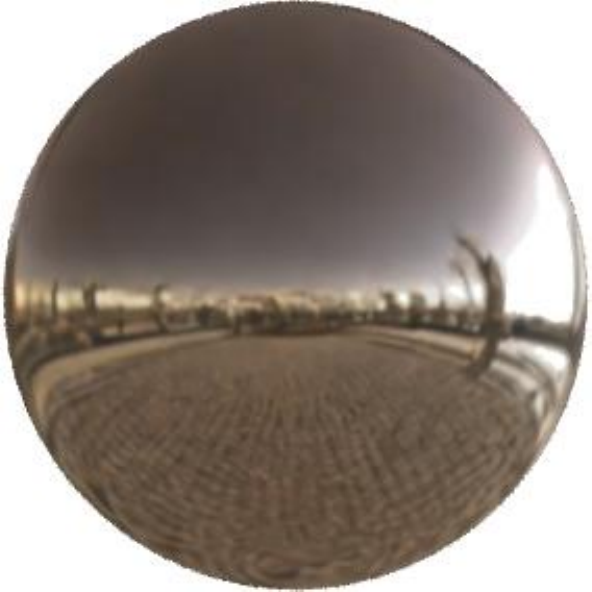}} &
        \noindent\parbox[c]{0.100\textwidth}{\includegraphics[height=0.100\textwidth]{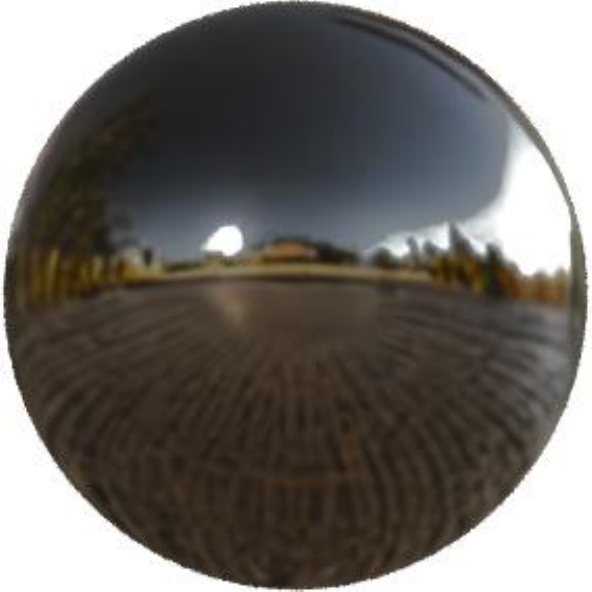}} & 
        \noindent\parbox[c]{0.100\textwidth}{\includegraphics[height=0.100\textwidth]{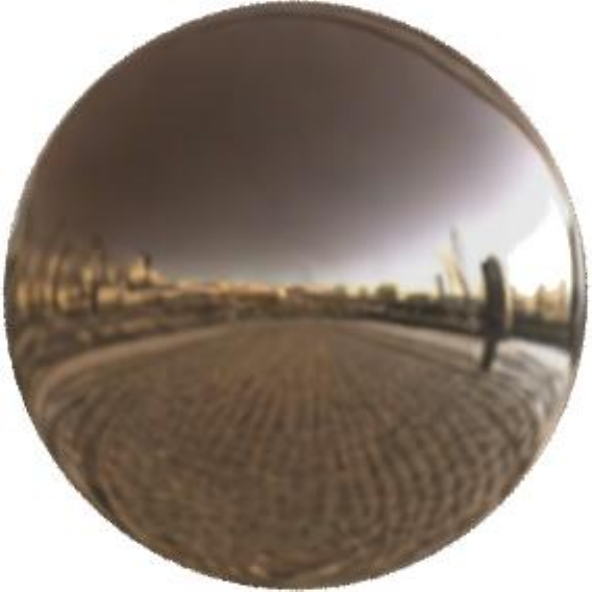}} & 
        \\

        \noindent\parbox[c]{0.205\textwidth}{\includegraphics[height=0.100\textwidth]{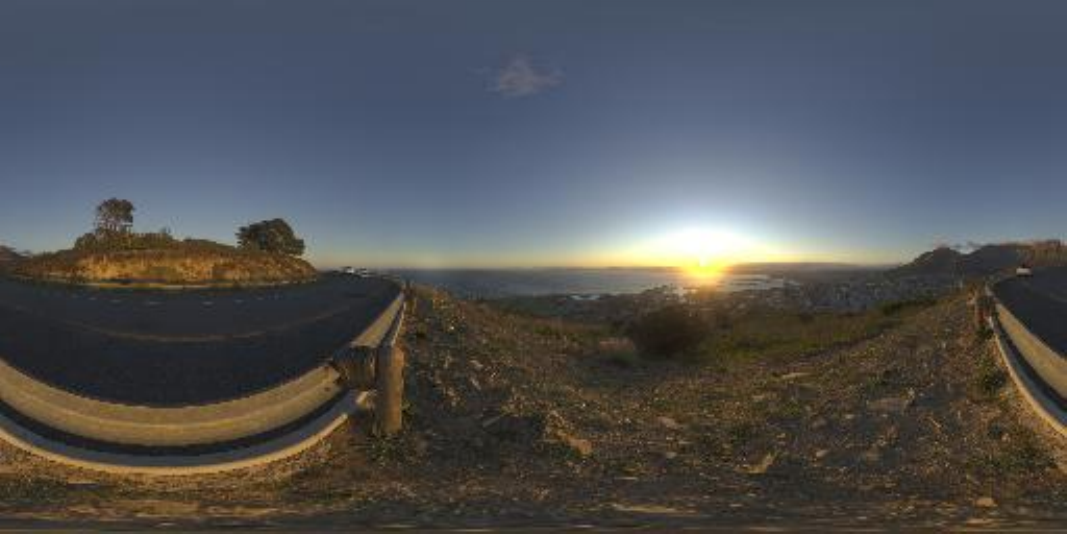}} & 
        \noindent\parbox[c]{0.14\textwidth}{\includegraphics[height=0.100\textwidth]{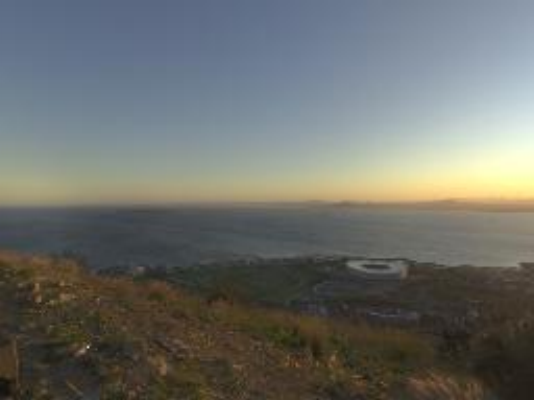}} &  
        
        \noindent\parbox[c]{0.100\textwidth}{\includegraphics[height=0.100\textwidth]{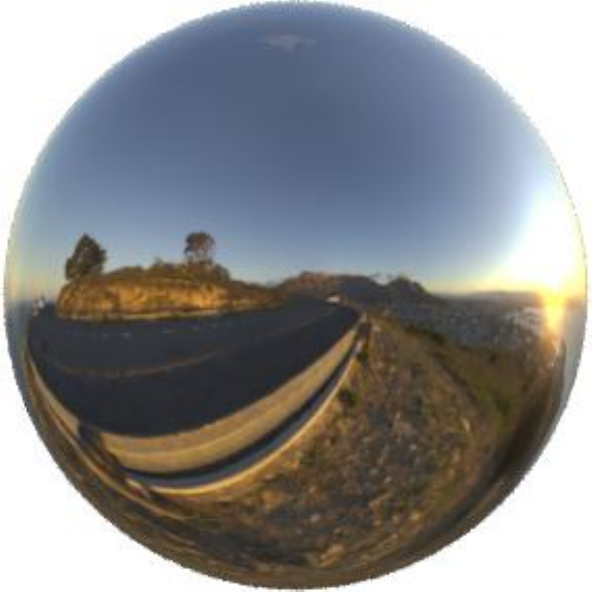}} & 
        \noindent\parbox[c]{0.100\textwidth}{\includegraphics[height=0.100\textwidth]{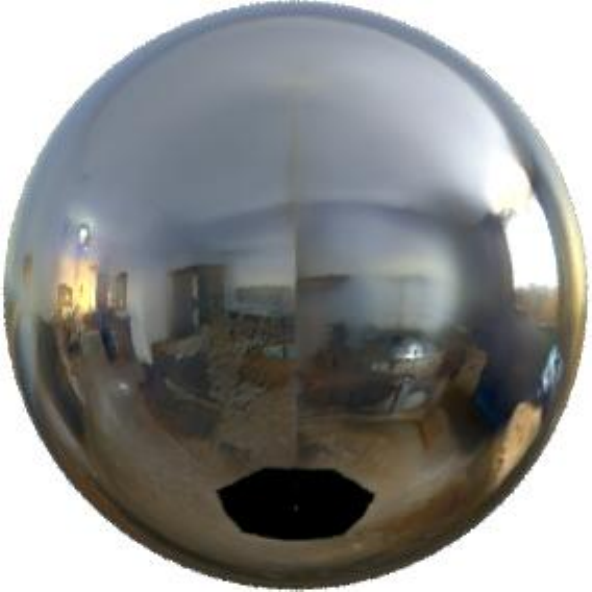}} & 
        
        \noindent\parbox[c]{0.100\textwidth}{\includegraphics[height=0.100\textwidth]{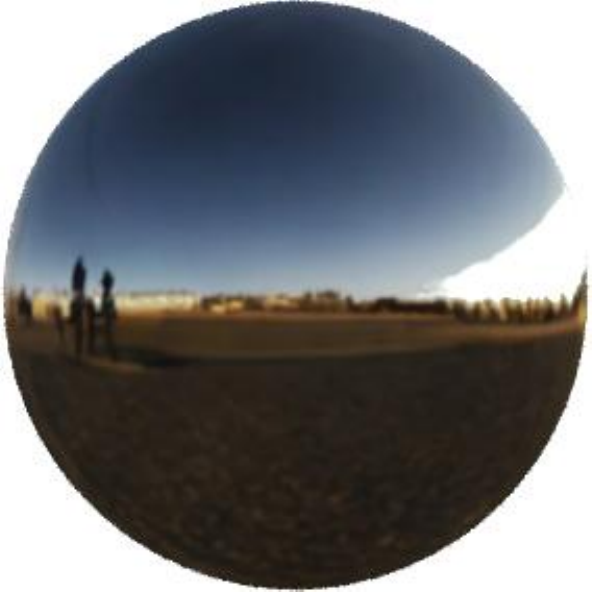}} & 
        \noindent\parbox[c]{0.100\textwidth}{\includegraphics[height=0.100\textwidth]{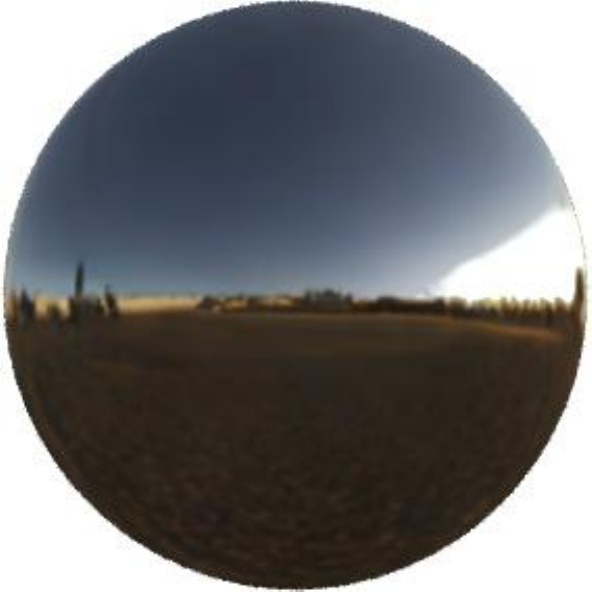}} &
        \noindent\parbox[c]{0.100\textwidth}{\includegraphics[height=0.100\textwidth]{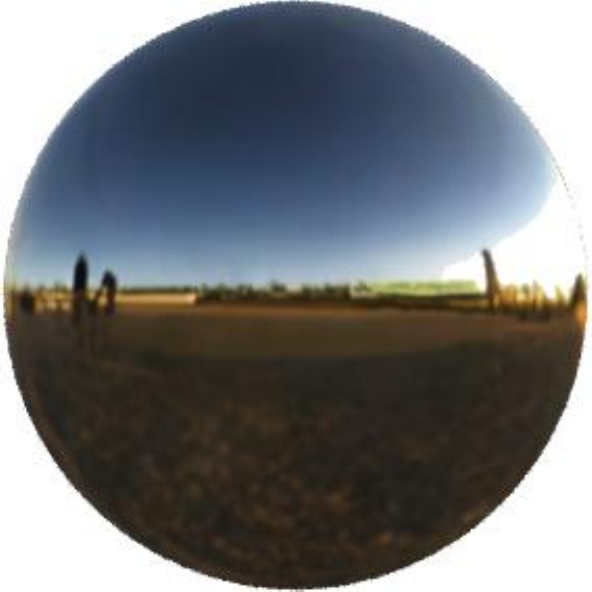}} & 
        \noindent\parbox[c]{0.100\textwidth}{\includegraphics[height=0.100\textwidth]{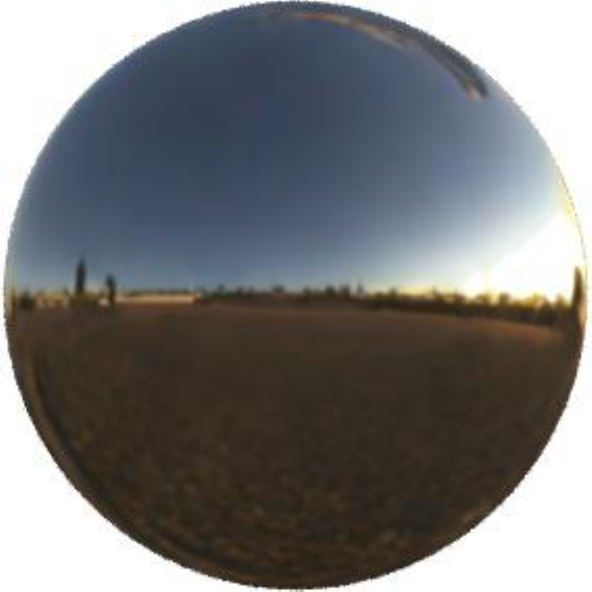}} & 
        \\

        \noindent\parbox[c]{0.205\textwidth}{\includegraphics[height=0.100\textwidth]{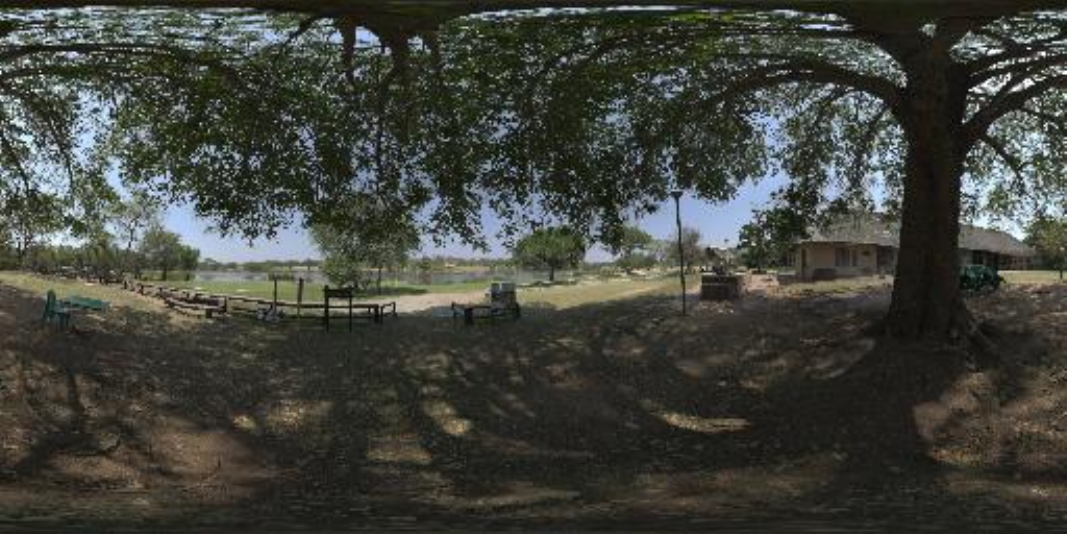}} & 
        \noindent\parbox[c]{0.14\textwidth}{\includegraphics[height=0.100\textwidth]{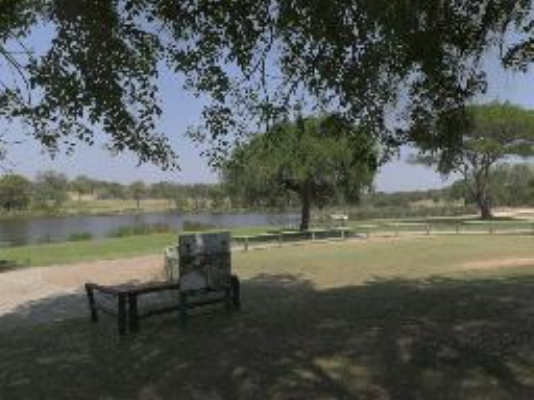}} &  
        
        \noindent\parbox[c]{0.100\textwidth}{\includegraphics[height=0.100\textwidth]{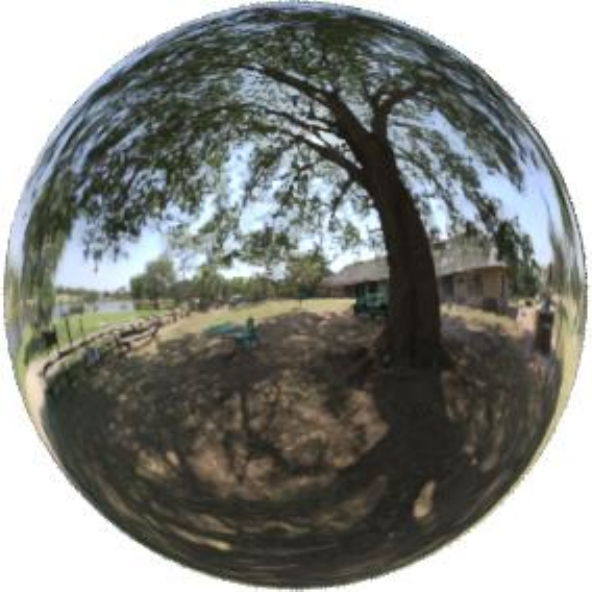}} & 
        \noindent\parbox[c]{0.100\textwidth}{\includegraphics[height=0.100\textwidth]{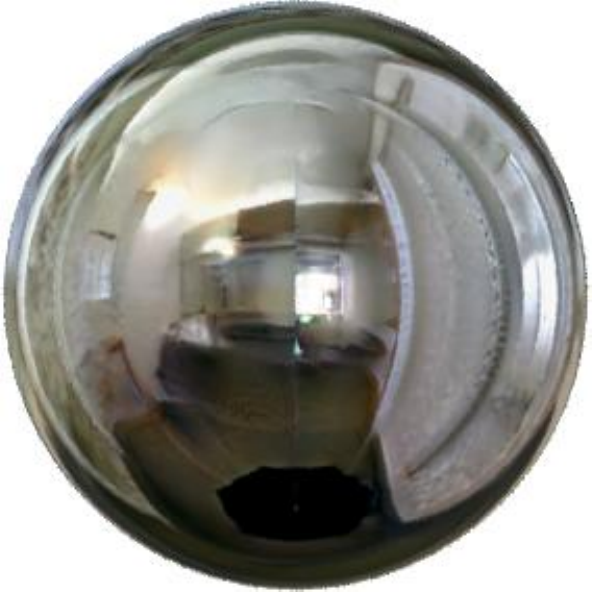}} & 
        
        \noindent\parbox[c]{0.100\textwidth}{\includegraphics[height=0.100\textwidth]{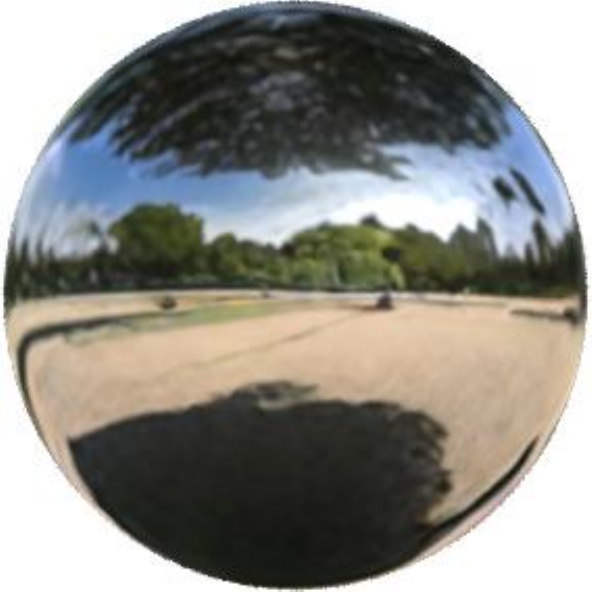}} & 
        \noindent\parbox[c]{0.100\textwidth}{\includegraphics[height=0.100\textwidth]{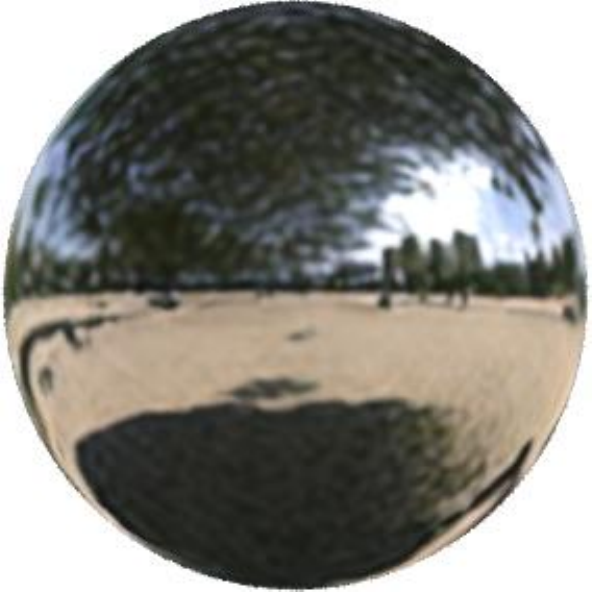}} &
        \noindent\parbox[c]{0.100\textwidth}{\includegraphics[height=0.100\textwidth]{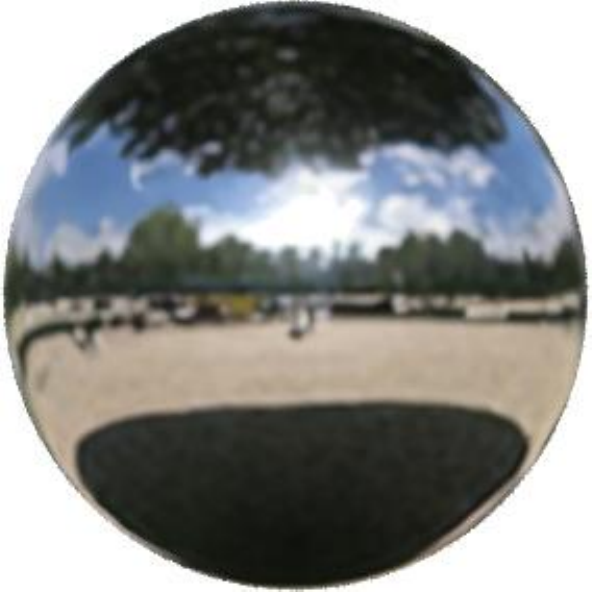}} & 
        \noindent\parbox[c]{0.100\textwidth}{\includegraphics[height=0.100\textwidth]{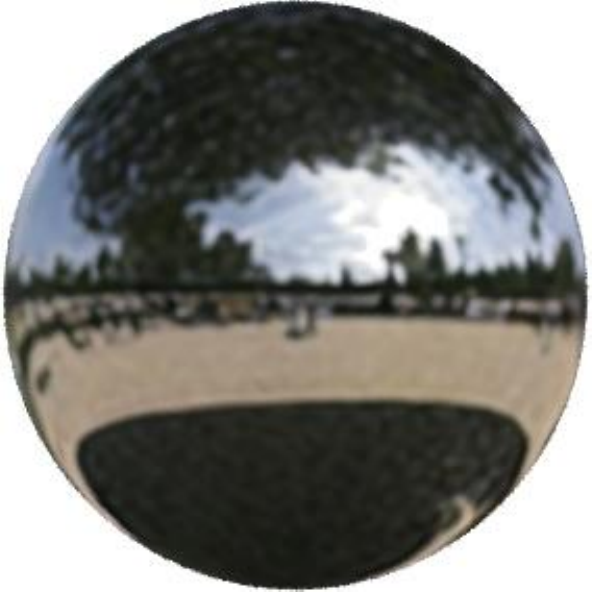}} & 
        \\

        \noindent\parbox[c]{0.205\textwidth}{\includegraphics[height=0.100\textwidth]{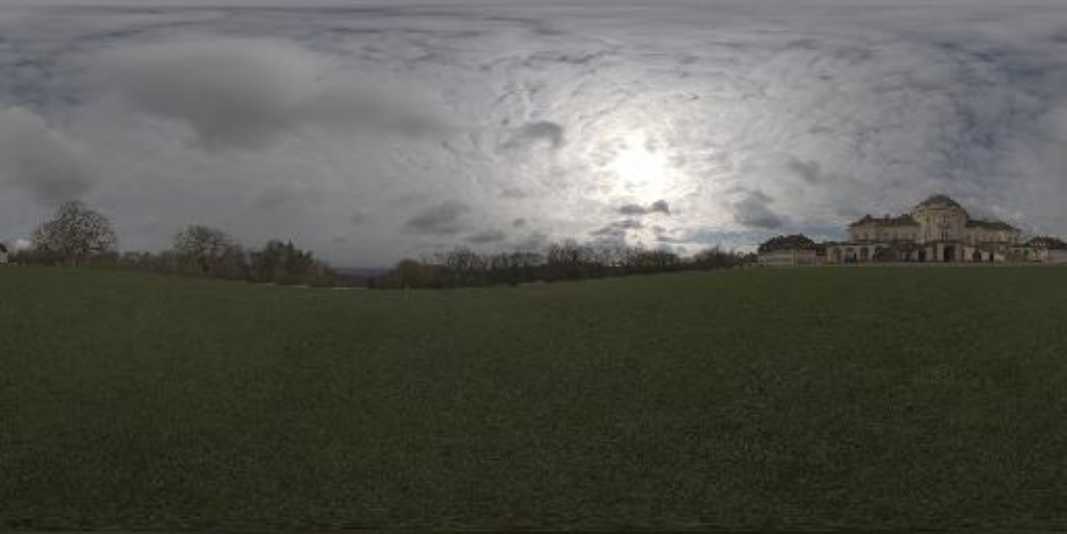}} & 
        \noindent\parbox[c]{0.14\textwidth}{\includegraphics[height=0.100\textwidth]{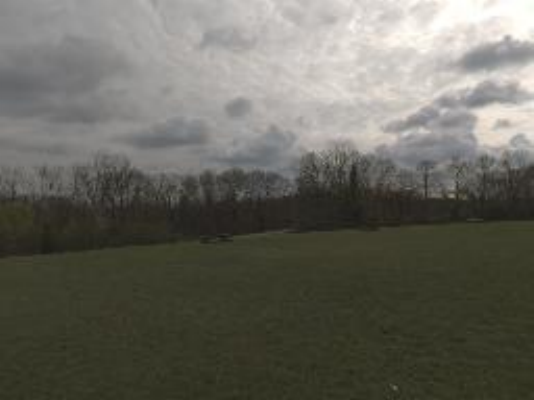}} &  
        
        \noindent\parbox[c]{0.100\textwidth}{\includegraphics[height=0.100\textwidth]{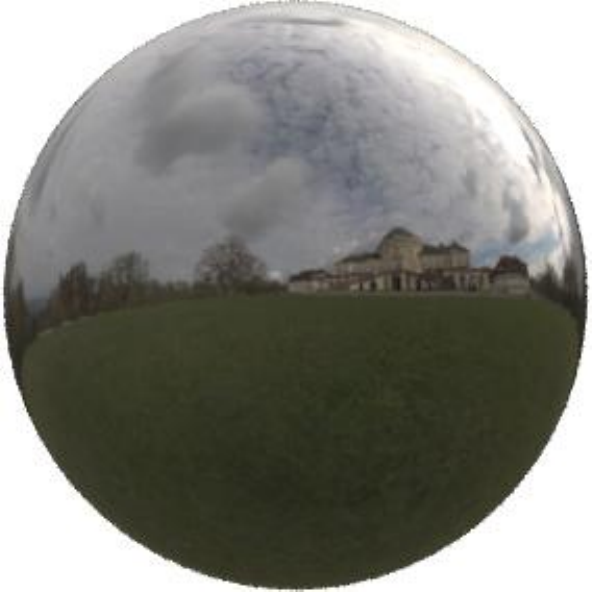}} & 
        \noindent\parbox[c]{0.100\textwidth}{\includegraphics[height=0.100\textwidth]{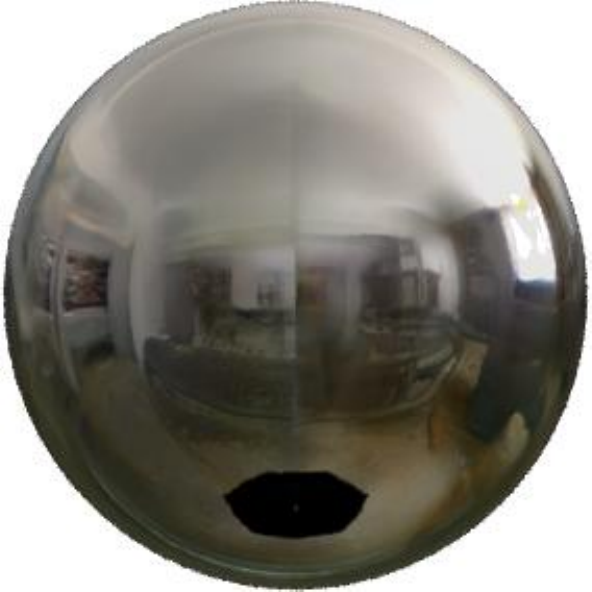}} & 
        
        \noindent\parbox[c]{0.100\textwidth}{\includegraphics[height=0.100\textwidth]{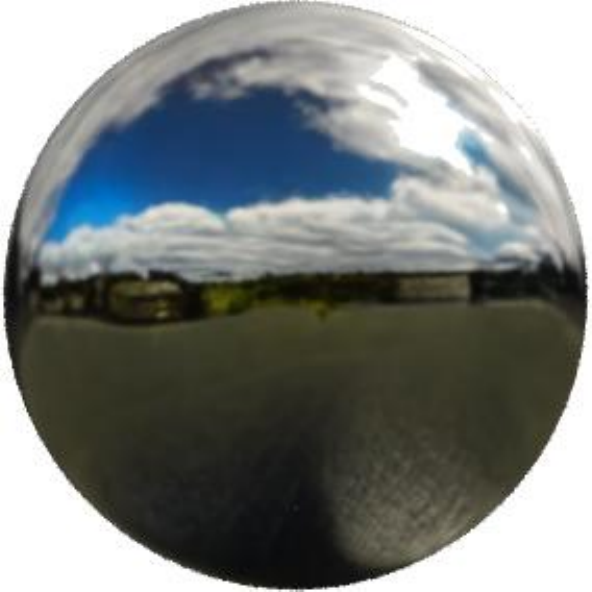}} & 
        \noindent\parbox[c]{0.100\textwidth}{\includegraphics[height=0.100\textwidth]{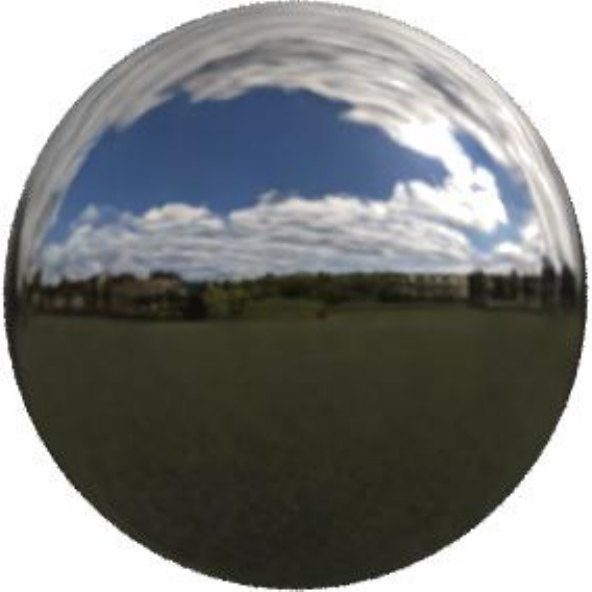}} &
        \noindent\parbox[c]{0.100\textwidth}{\includegraphics[height=0.100\textwidth]{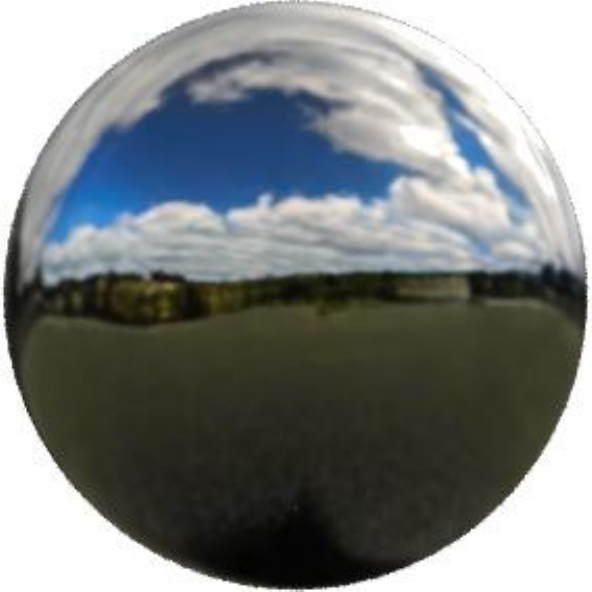}} & 
        \noindent\parbox[c]{0.100\textwidth}{\includegraphics[height=0.100\textwidth]{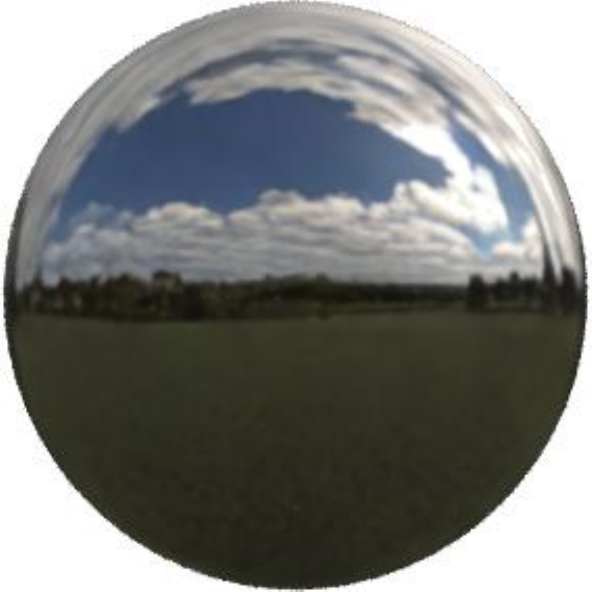}} & 
        \\

        \noindent\parbox[c]{0.205\textwidth}{\includegraphics[height=0.100\textwidth]{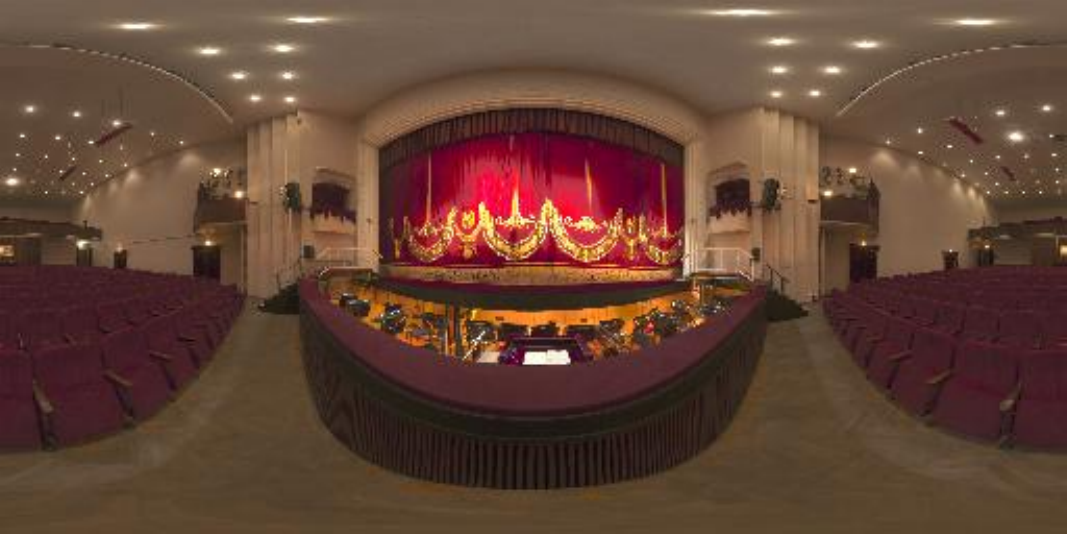}} & 
        \noindent\parbox[c]{0.14\textwidth}{\includegraphics[height=0.100\textwidth]{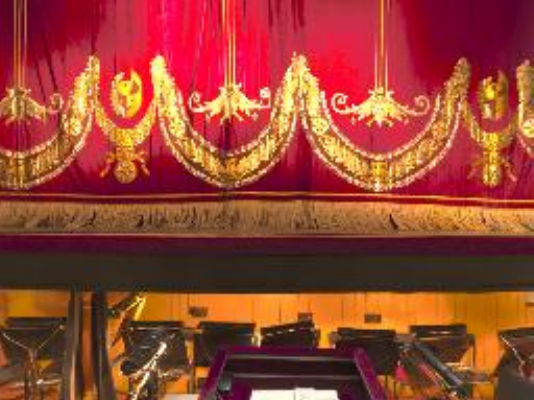}} &  
        
        \noindent\parbox[c]{0.100\textwidth}{\includegraphics[height=0.100\textwidth]{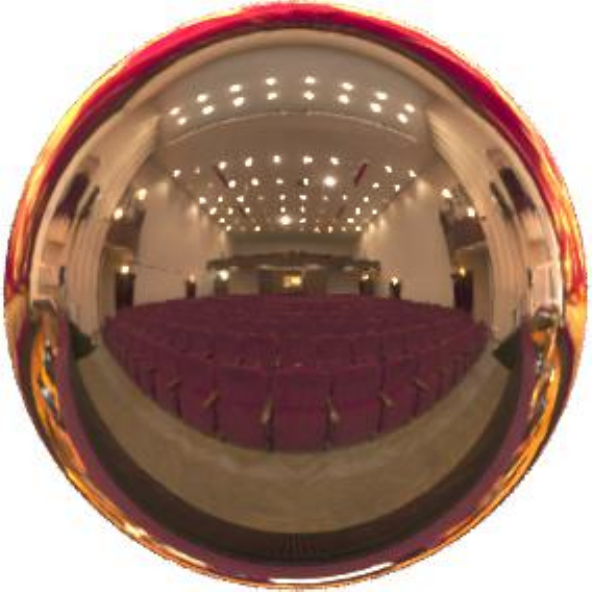}} & 
        \noindent\parbox[c]{0.100\textwidth}{\includegraphics[height=0.100\textwidth]{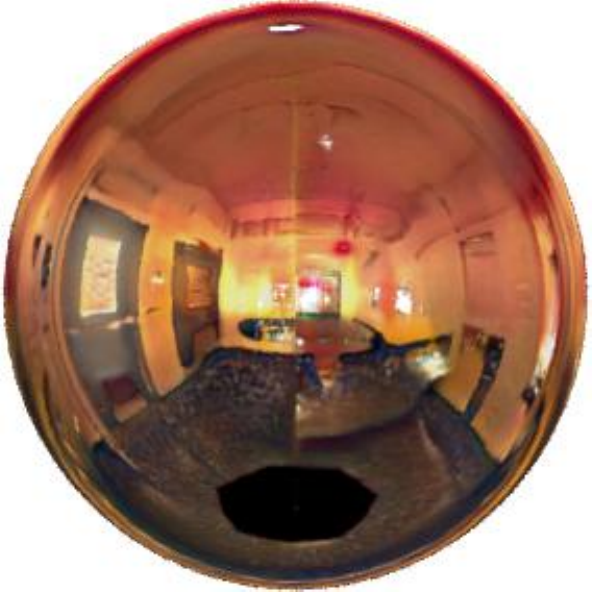}} & 
        
        \noindent\parbox[c]{0.100\textwidth}{\includegraphics[height=0.100\textwidth]{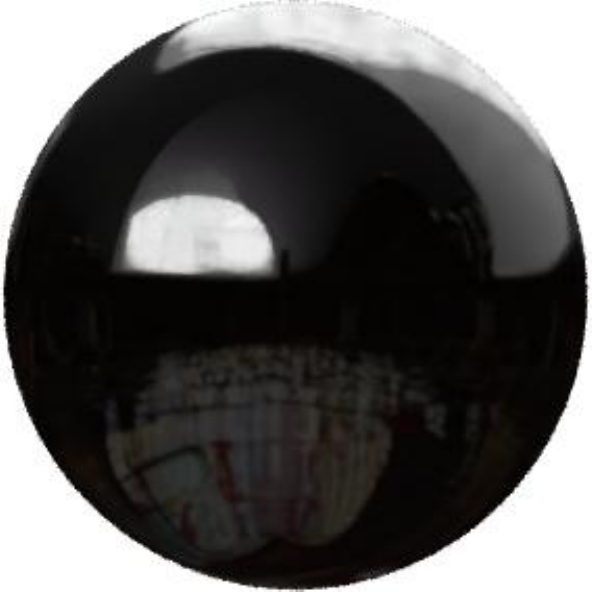}} & 
        \noindent\parbox[c]{0.100\textwidth}{\includegraphics[height=0.100\textwidth]{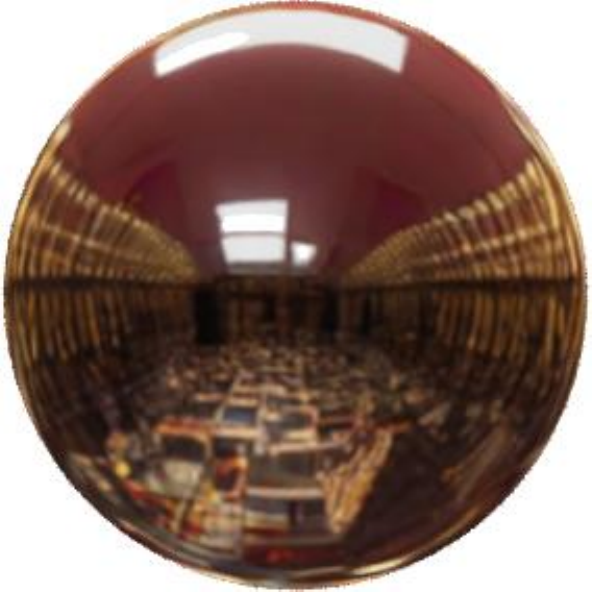}} &
        \noindent\parbox[c]{0.100\textwidth}{\includegraphics[height=0.100\textwidth]{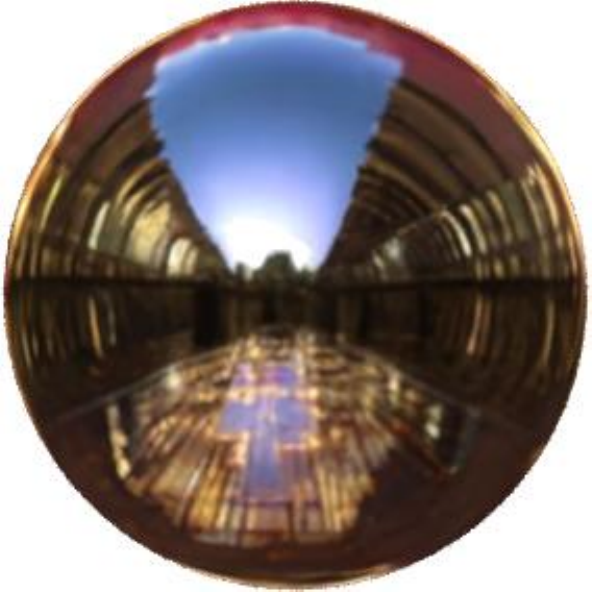}} & 
        \noindent\parbox[c]{0.100\textwidth}{\includegraphics[height=0.100\textwidth]{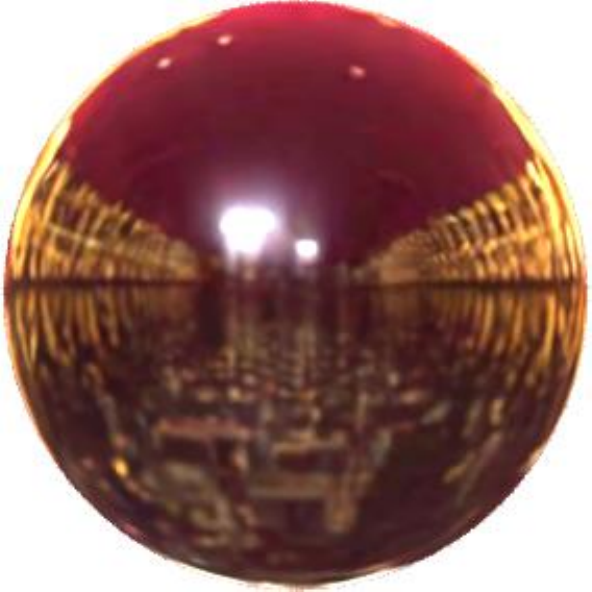}} & 
        \\
        
        \end{tabu}
    \caption{
    Qualitative results for the Poly Haven dataset using mirror balls.}
    \label{fig:additional_polyhaven_mirror}
\end{figure*}

\begin{figure*}[!t]
    \centering

        \begin{tabu} to \textwidth {
        @{}
        c@{}
        c@{}
        c@{}
        c@{}
        c@{}
        c@{}
        c@{}
        c@{}
        c@{}
    }

        \multicolumn{1}{c}{\shortstack{\scriptsize Ground truth map}}
        & 
        \multicolumn{1}{c}{\shortstack{\hspace{-6pt} \scriptsize Input}}
        &
        \multicolumn{1}{c}{\shortstack{\scriptsize Ground truth}}
        & 
        \multicolumn{1}{c}{\shortstack{\scriptsize StyleLight \cite{wang2022stylelight}}}
        & 
        \multicolumn{1}{c}{\shortstack{\scriptsize SDXL$^\dagger$}} &
        \multicolumn{1}{c}{\shortstack{\scriptsize \begin{tabular}[c]{@{}c@{}}SDXL$^\dagger$+LR \\ (ours, ablated)\end{tabular}}} &
        \multicolumn{1}{c}{\shortstack{\scriptsize \begin{tabular}[c]{@{}c@{}}SDXL$^\dagger$+I \\ (ours,ablated)\end{tabular}}}
        &
        \multicolumn{1}{c}{\shortstack{\scriptsize \begin{tabular}[c]{@{}c@{}}SDXL$^\dagger$+LR+I \\ (ours)\end{tabular}}} 
        \\

        \noindent\parbox[c]{0.205\textwidth}{\includegraphics[height=0.100\textwidth]{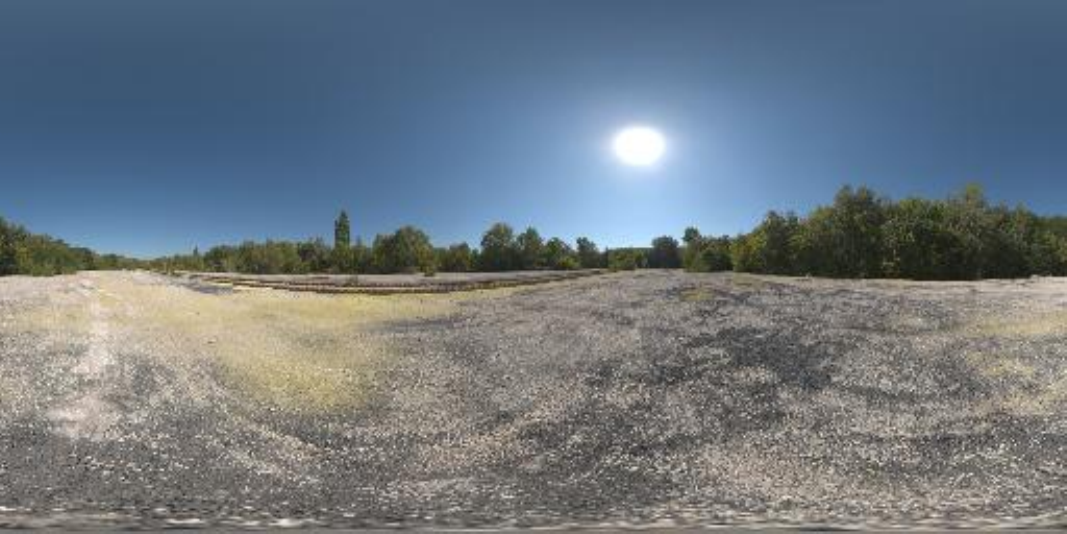}} & 
        \noindent\parbox[c]{0.14\textwidth}{\includegraphics[height=0.100\textwidth]{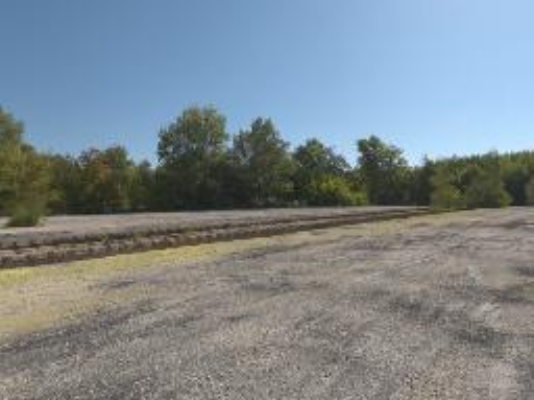}} &  
        
        \noindent\parbox[c]{0.100\textwidth}{\includegraphics[height=0.100\textwidth]{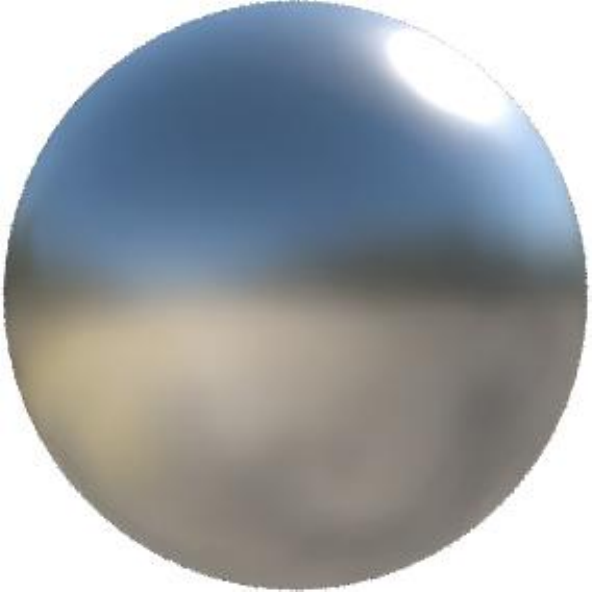}} & 
        \noindent\parbox[c]{0.100\textwidth}{\includegraphics[height=0.100\textwidth]{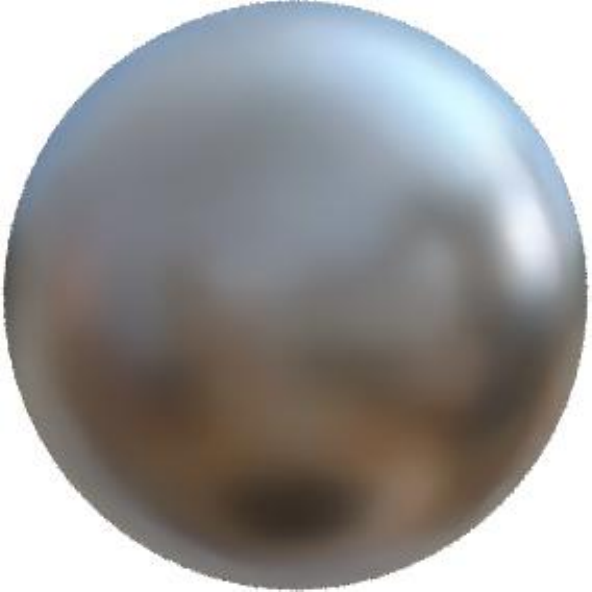}} & 
        
        \noindent\parbox[c]{0.100\textwidth}{\includegraphics[height=0.100\textwidth]{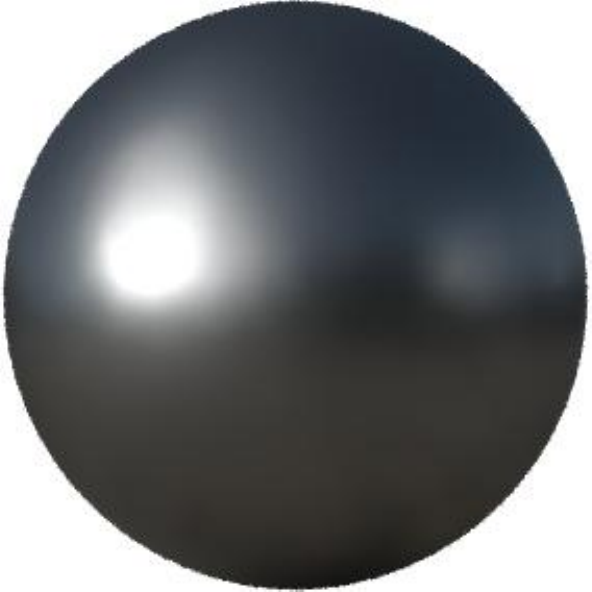}} & 
        \noindent\parbox[c]{0.100\textwidth}{\includegraphics[height=0.100\textwidth]{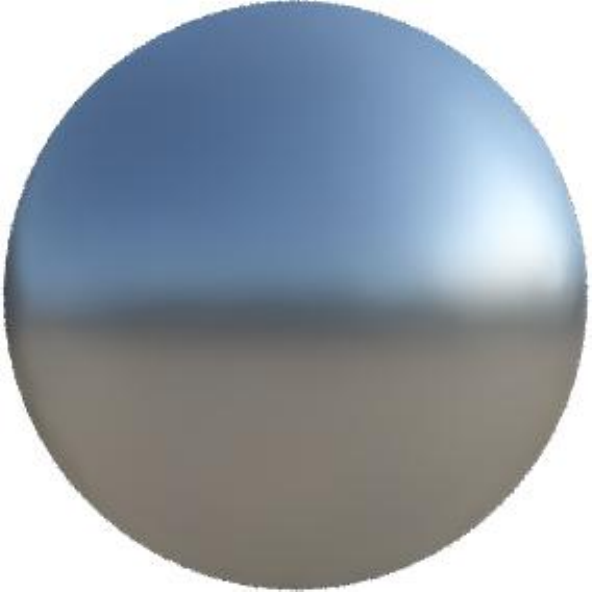}} &
        \noindent\parbox[c]{0.100\textwidth}{\includegraphics[height=0.100\textwidth]{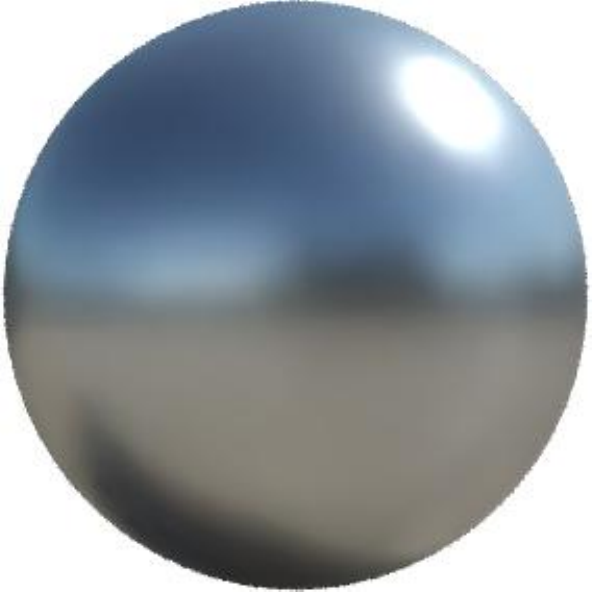}} & 
        \noindent\parbox[c]{0.100\textwidth}{\includegraphics[height=0.100\textwidth]{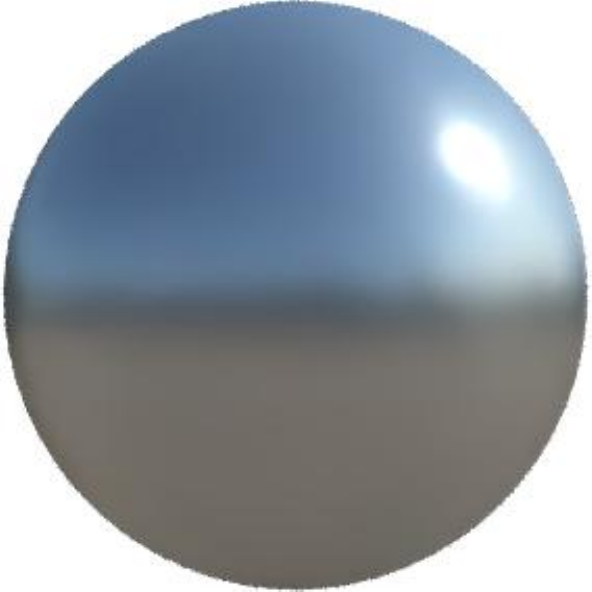}} & 
        \\

        \noindent\parbox[c]{0.205\textwidth}{\includegraphics[height=0.100\textwidth]{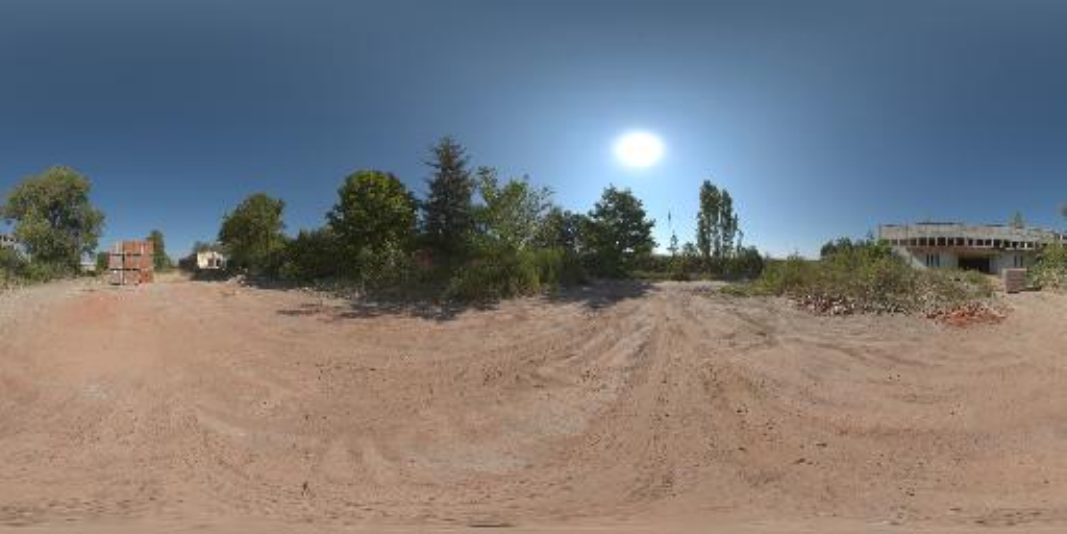}} & 
        \noindent\parbox[c]{0.14\textwidth}{\includegraphics[height=0.100\textwidth]{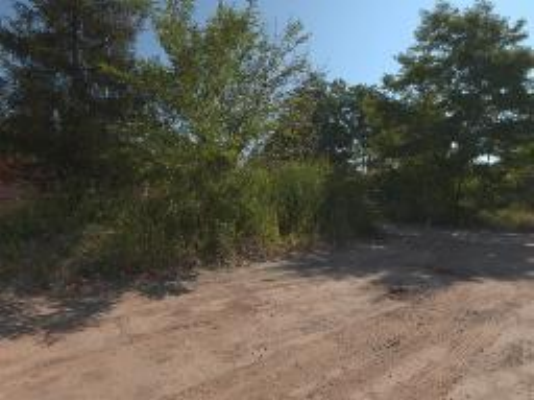}} &  
        
        \noindent\parbox[c]{0.100\textwidth}{\includegraphics[height=0.100\textwidth]{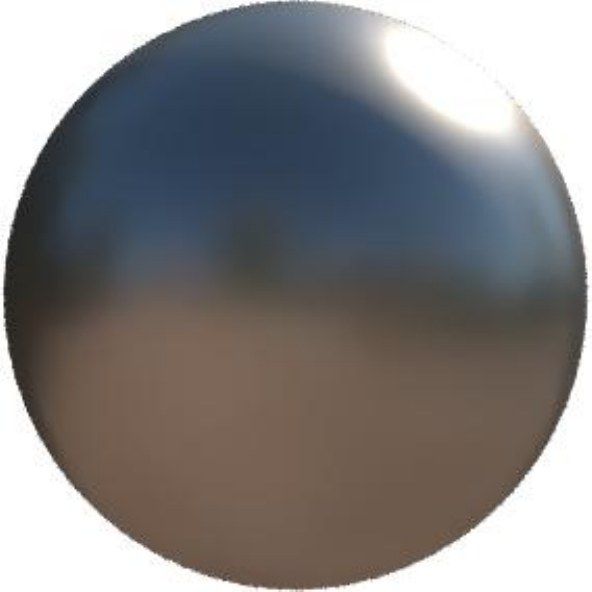}} & 
        \noindent\parbox[c]{0.100\textwidth}{\includegraphics[height=0.100\textwidth]{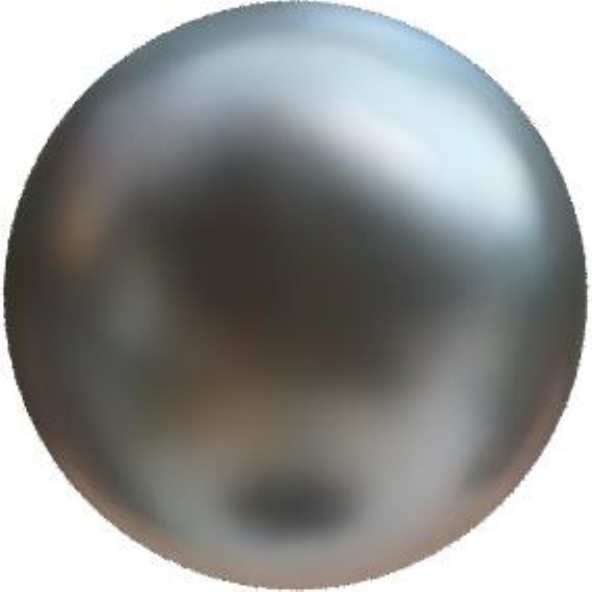}} & 
        
        \noindent\parbox[c]{0.100\textwidth}{\includegraphics[height=0.100\textwidth]{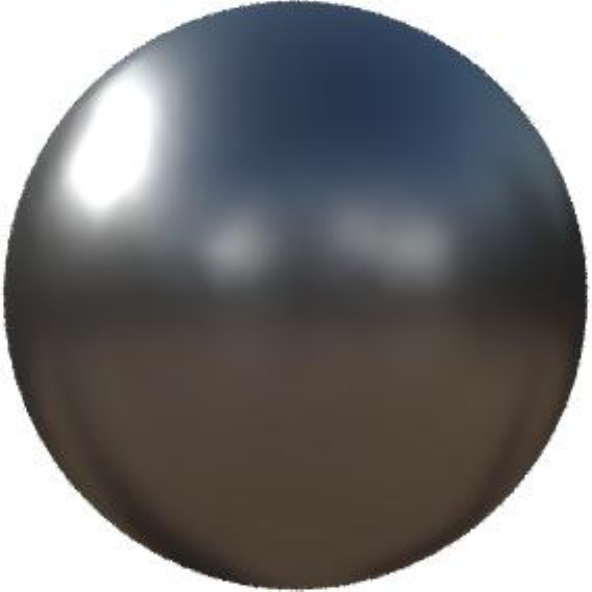}} & 
        \noindent\parbox[c]{0.100\textwidth}{\includegraphics[height=0.100\textwidth]{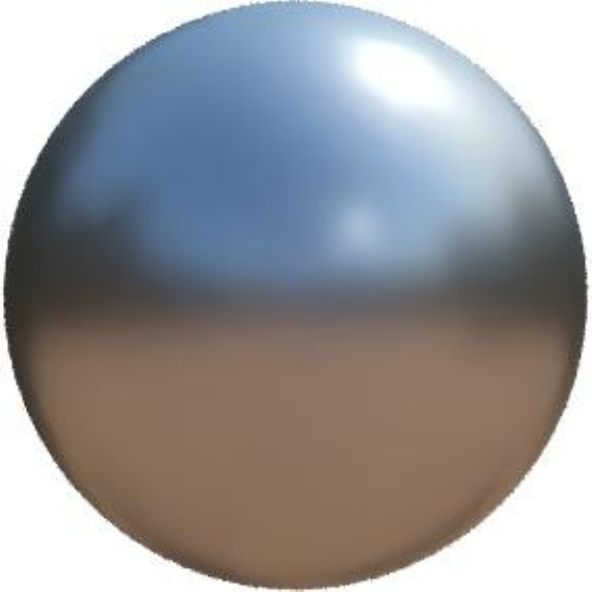}} &
        \noindent\parbox[c]{0.100\textwidth}{\includegraphics[height=0.100\textwidth]{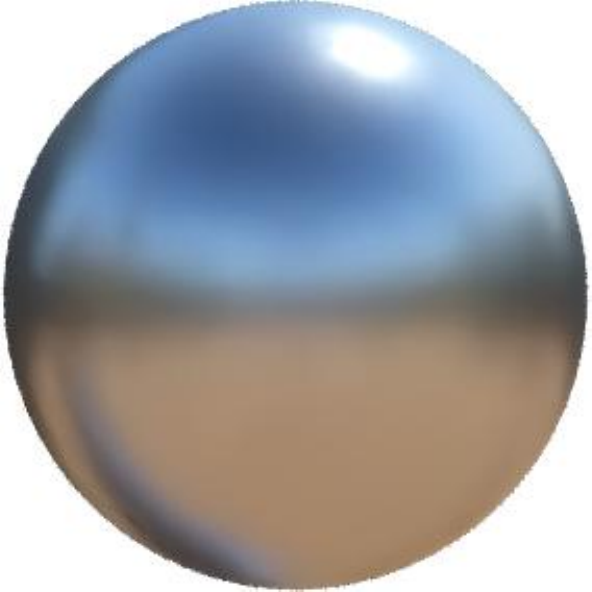}} & 
        \noindent\parbox[c]{0.100\textwidth}{\includegraphics[height=0.100\textwidth]{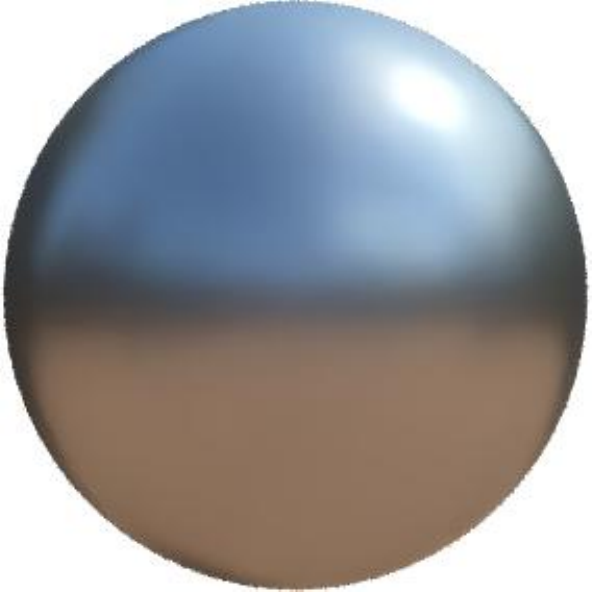}} & 
        \\

        \noindent\parbox[c]{0.205\textwidth}{\includegraphics[height=0.100\textwidth]{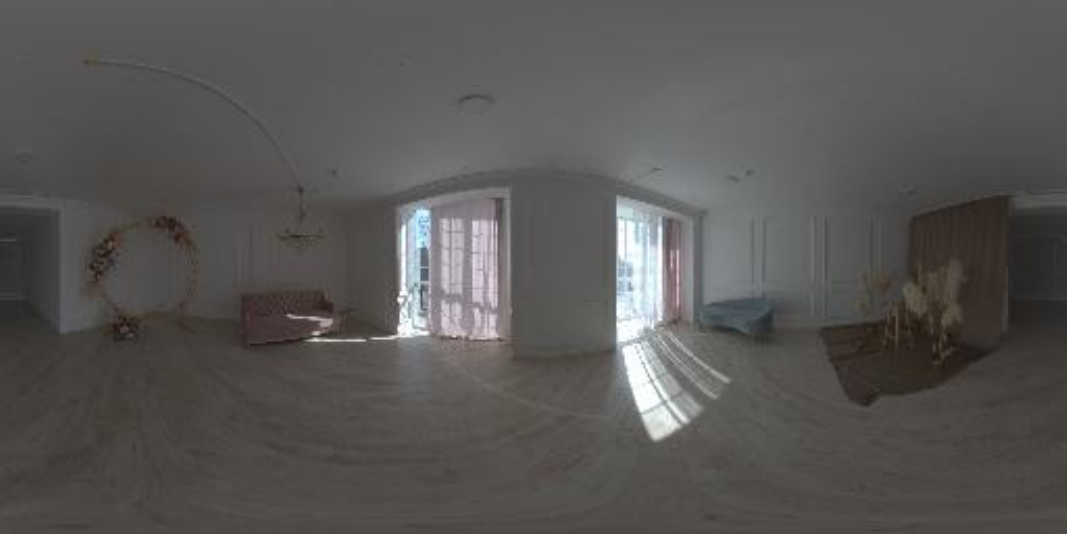}} & 
        \noindent\parbox[c]{0.14\textwidth}{\includegraphics[height=0.100\textwidth]{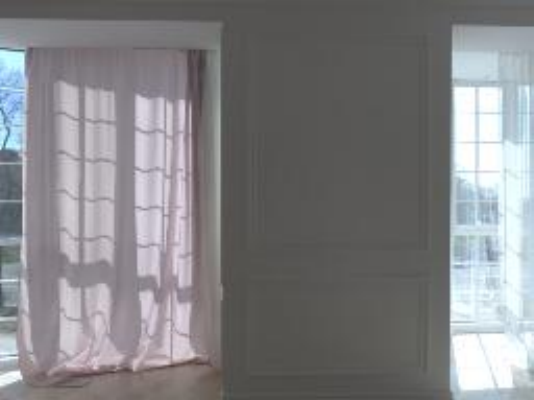}} &  
        
        \noindent\parbox[c]{0.100\textwidth}{\includegraphics[height=0.100\textwidth]{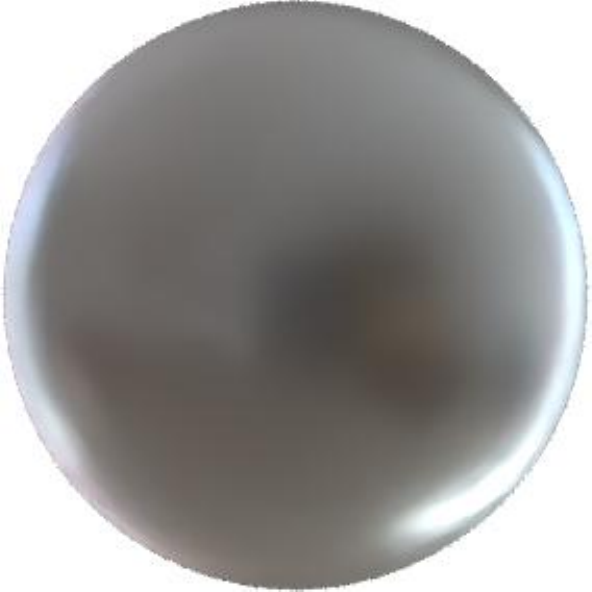}} & 
        \noindent\parbox[c]{0.100\textwidth}{\includegraphics[height=0.100\textwidth]{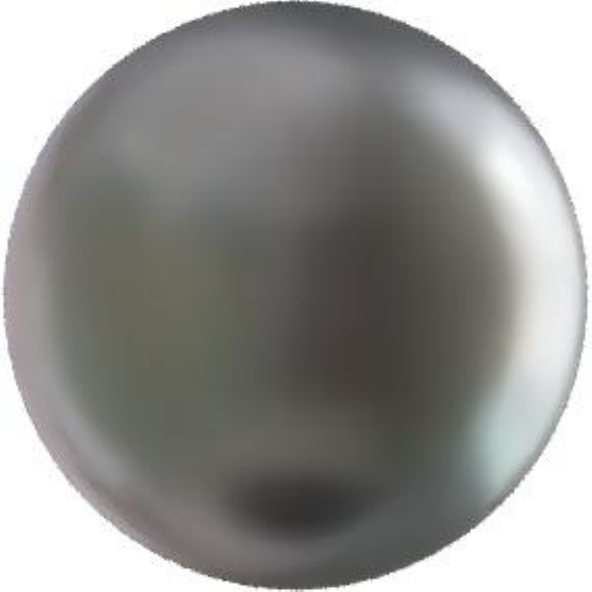}} & 
        
        \noindent\parbox[c]{0.100\textwidth}{\includegraphics[height=0.100\textwidth]{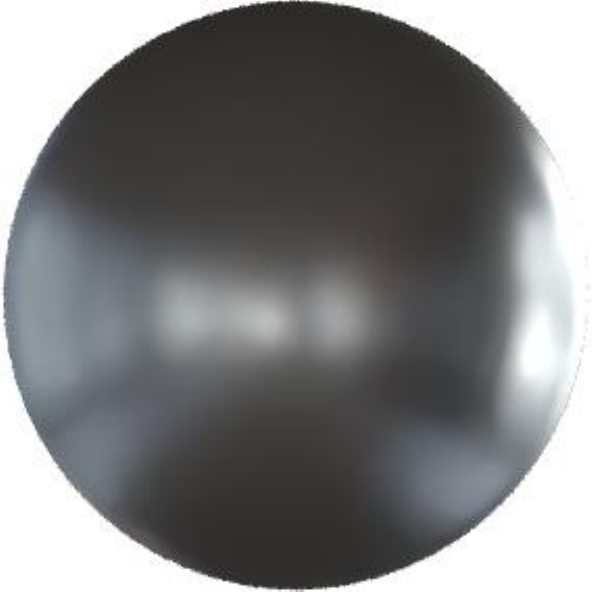}} & 
        \noindent\parbox[c]{0.100\textwidth}{\includegraphics[height=0.100\textwidth]{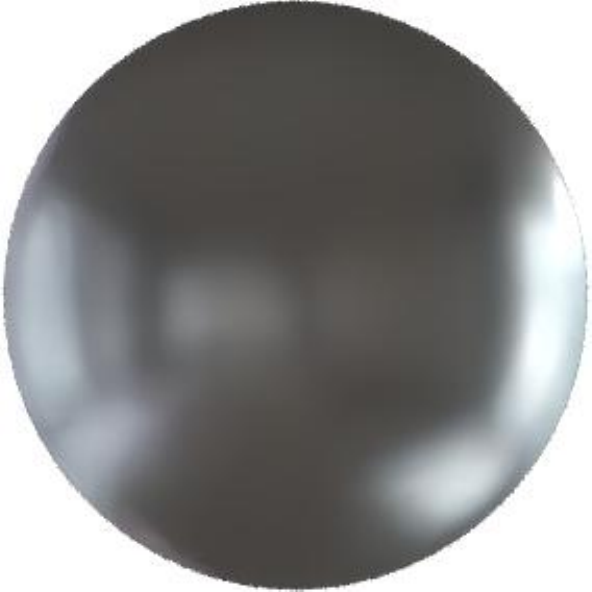}} &
        \noindent\parbox[c]{0.100\textwidth}{\includegraphics[height=0.100\textwidth]{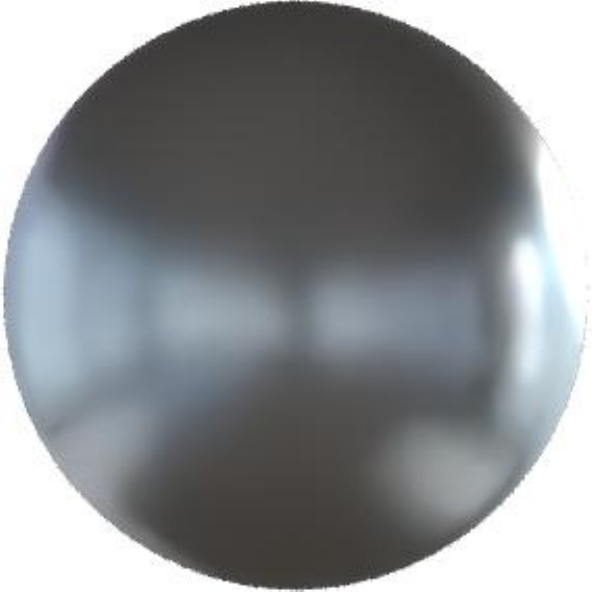}} & 
        \noindent\parbox[c]{0.100\textwidth}{\includegraphics[height=0.100\textwidth]{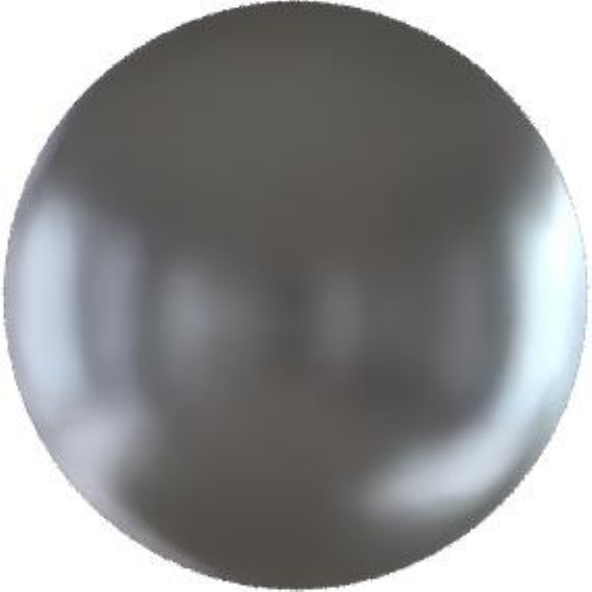}} & 
        \\

        \noindent\parbox[c]{0.205\textwidth}{\includegraphics[height=0.100\textwidth]{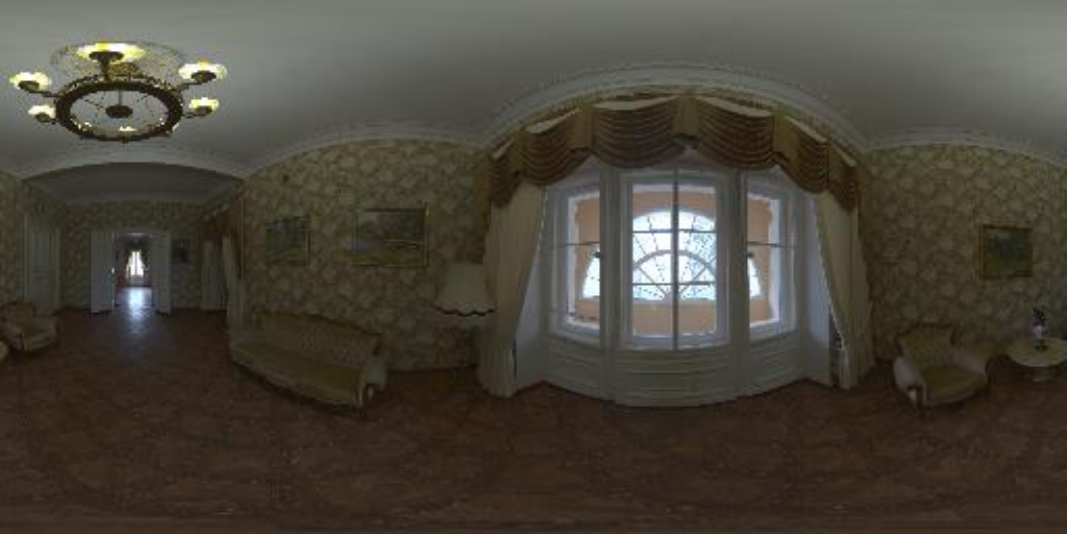}} & 
        \noindent\parbox[c]{0.14\textwidth}{\includegraphics[height=0.100\textwidth]{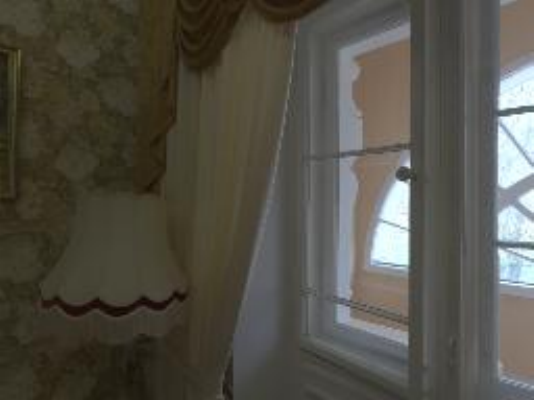}} &  
        
        \noindent\parbox[c]{0.100\textwidth}{\includegraphics[height=0.100\textwidth]{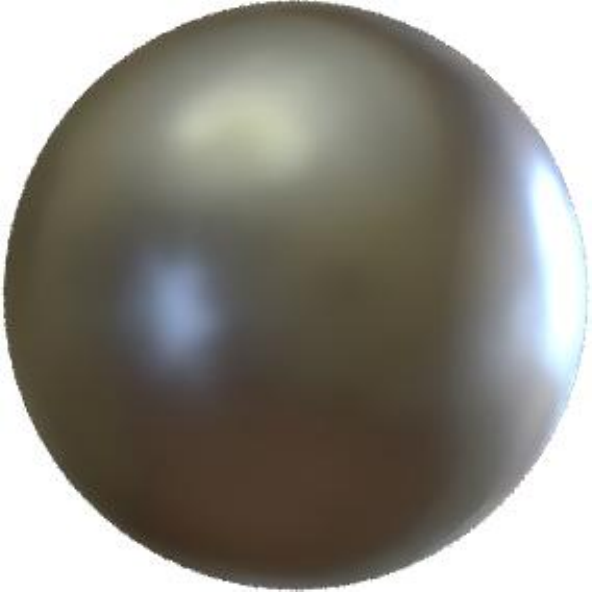}} & 
        \noindent\parbox[c]{0.100\textwidth}{\includegraphics[height=0.100\textwidth]{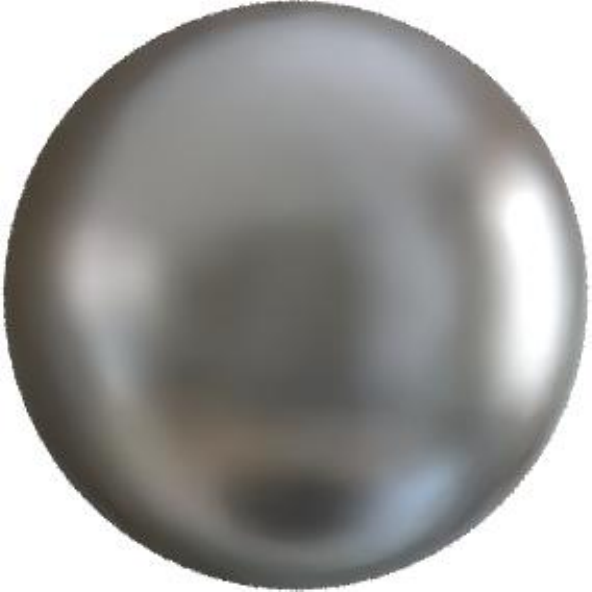}} & 
        
        \noindent\parbox[c]{0.100\textwidth}{\includegraphics[height=0.100\textwidth]{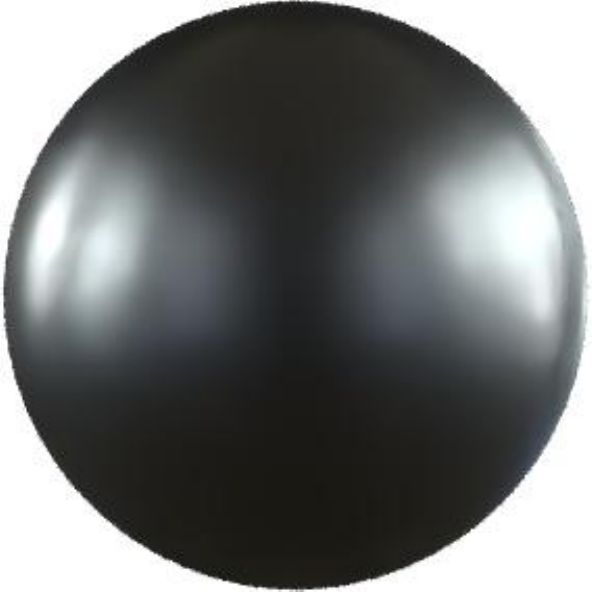}} & 
        \noindent\parbox[c]{0.100\textwidth}{\includegraphics[height=0.100\textwidth]{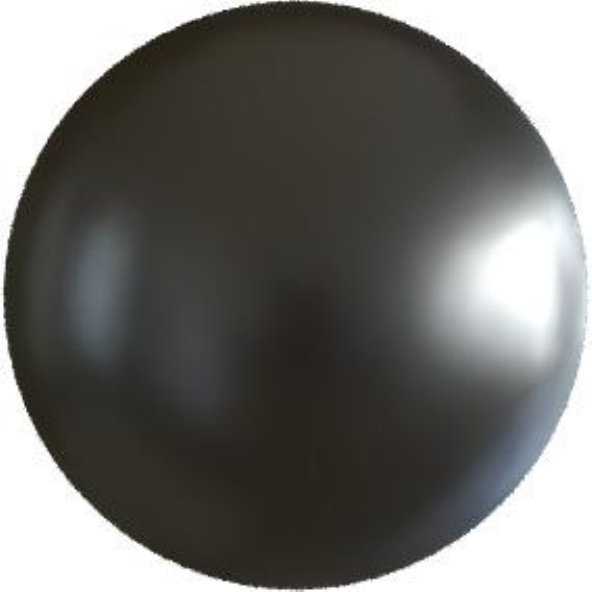}} &
        \noindent\parbox[c]{0.100\textwidth}{\includegraphics[height=0.100\textwidth]{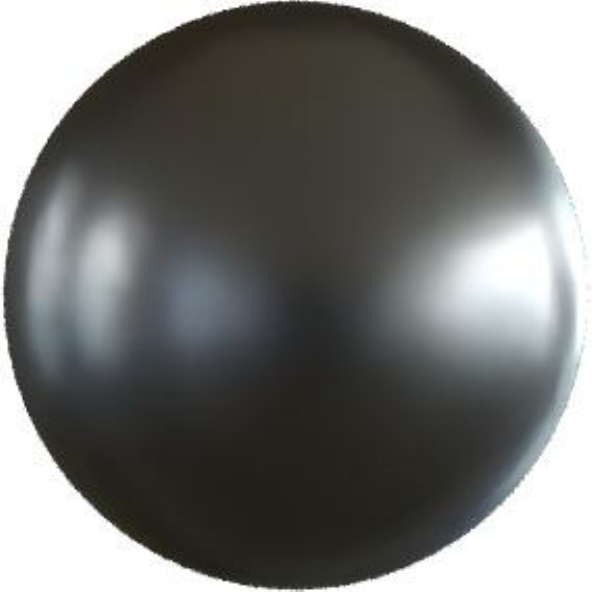}} & 
        \noindent\parbox[c]{0.100\textwidth}{\includegraphics[height=0.100\textwidth]{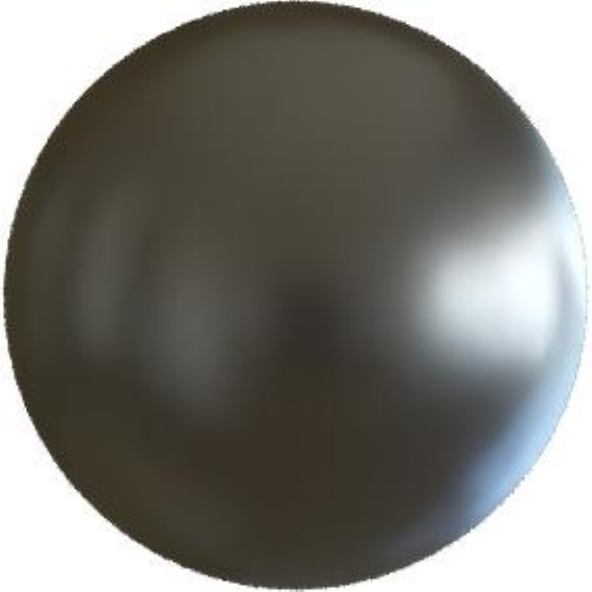}} & 
        \\

        \noindent\parbox[c]{0.205\textwidth}{\includegraphics[height=0.100\textwidth]{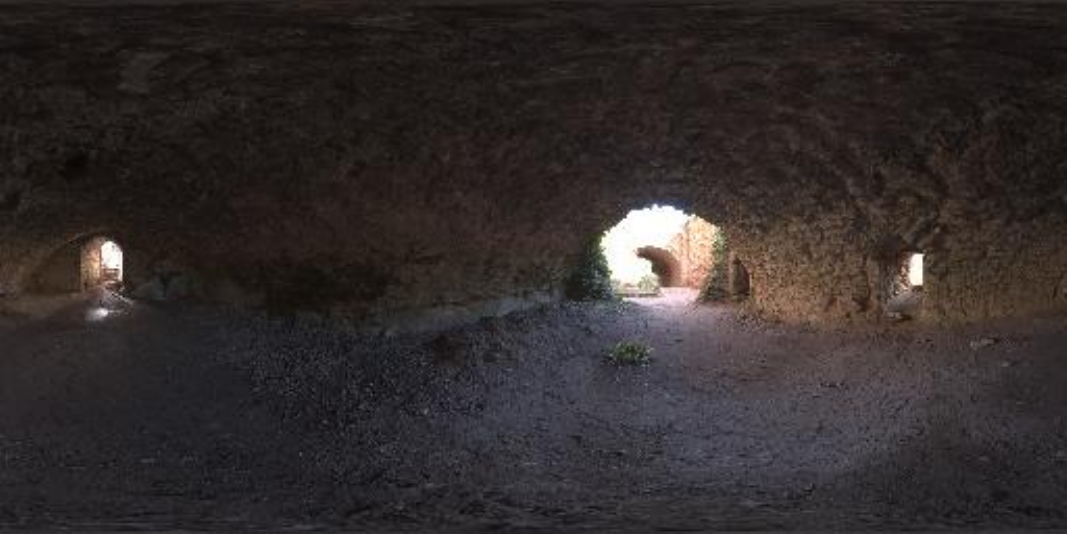}} & 
        \noindent\parbox[c]{0.14\textwidth}{\includegraphics[height=0.100\textwidth]{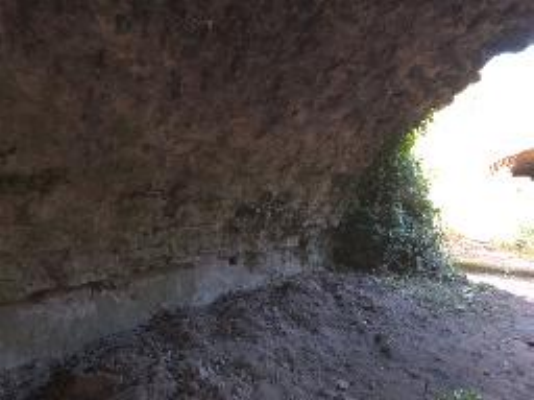}} &  
        
        \noindent\parbox[c]{0.100\textwidth}{\includegraphics[height=0.100\textwidth]{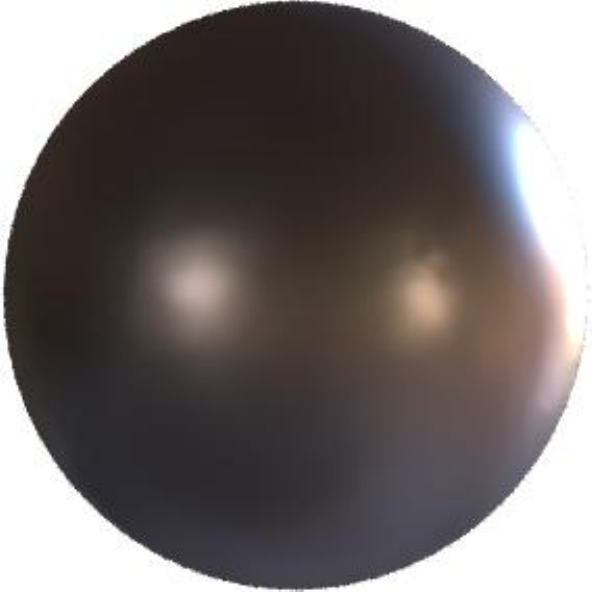}} & 
        \noindent\parbox[c]{0.100\textwidth}{\includegraphics[height=0.100\textwidth]{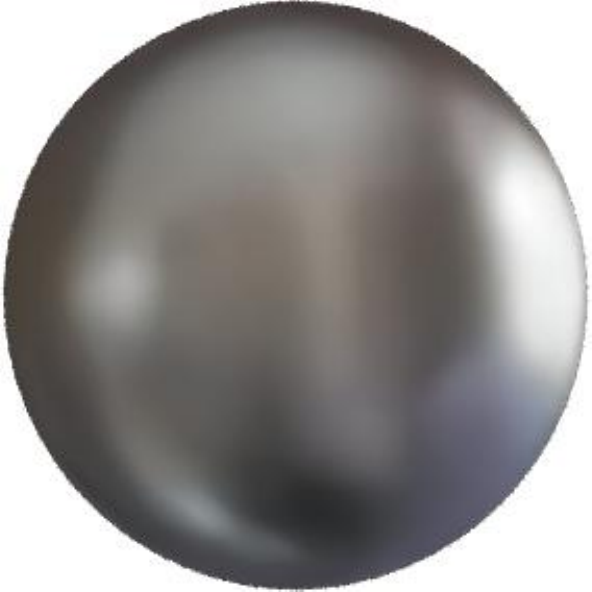}} & 
        
        \noindent\parbox[c]{0.100\textwidth}{\includegraphics[height=0.100\textwidth]{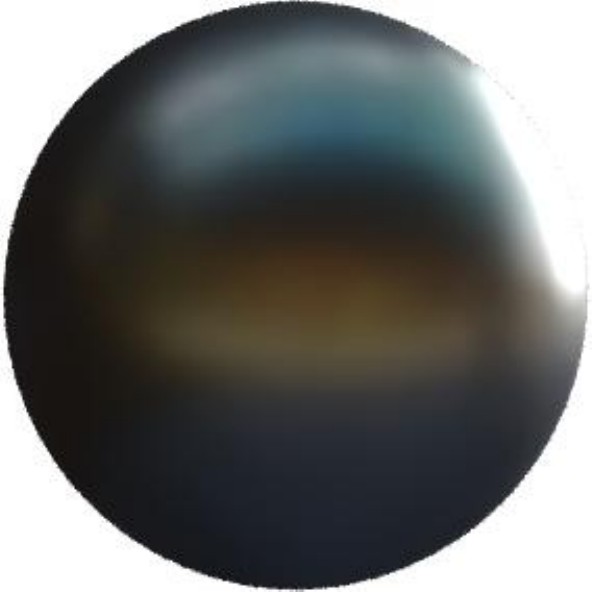}} & 
        \noindent\parbox[c]{0.100\textwidth}{\includegraphics[height=0.100\textwidth]{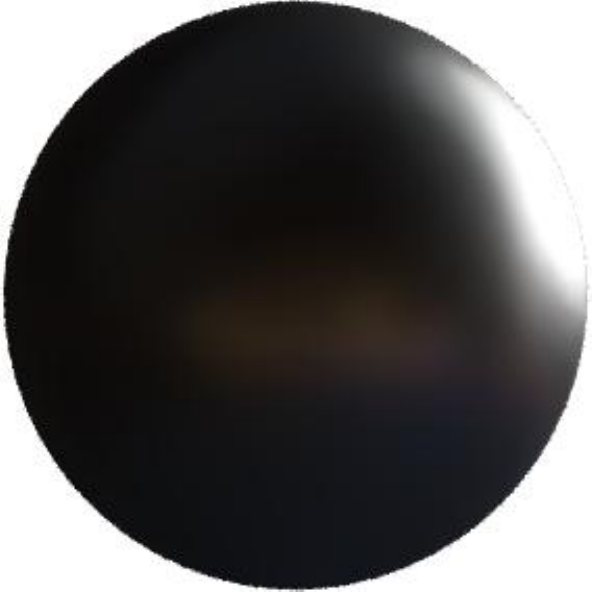}} &
        \noindent\parbox[c]{0.100\textwidth}{\includegraphics[height=0.100\textwidth]{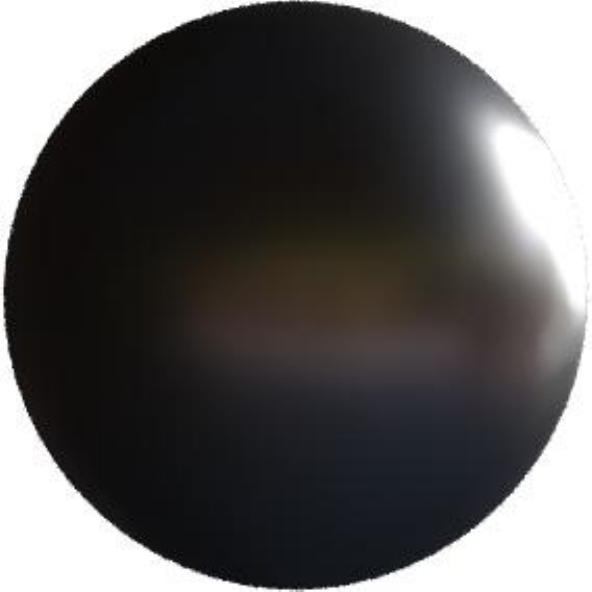}} & 
        \noindent\parbox[c]{0.100\textwidth}{\includegraphics[height=0.100\textwidth]{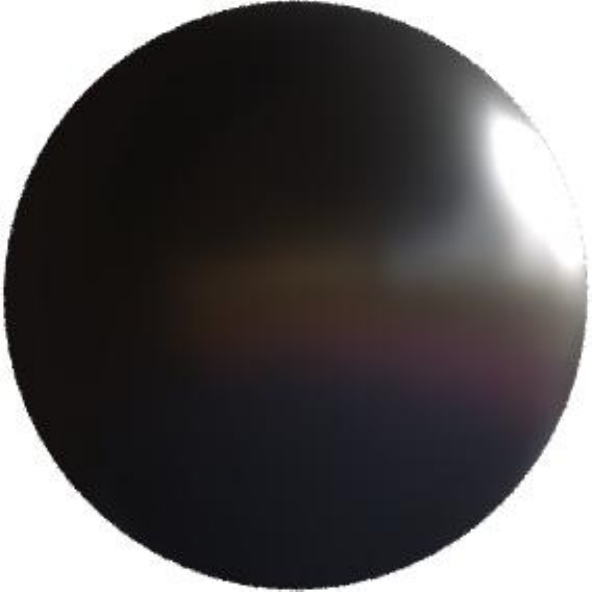}} & 
        \\

        \noindent\parbox[c]{0.205\textwidth}{\includegraphics[height=0.100\textwidth]{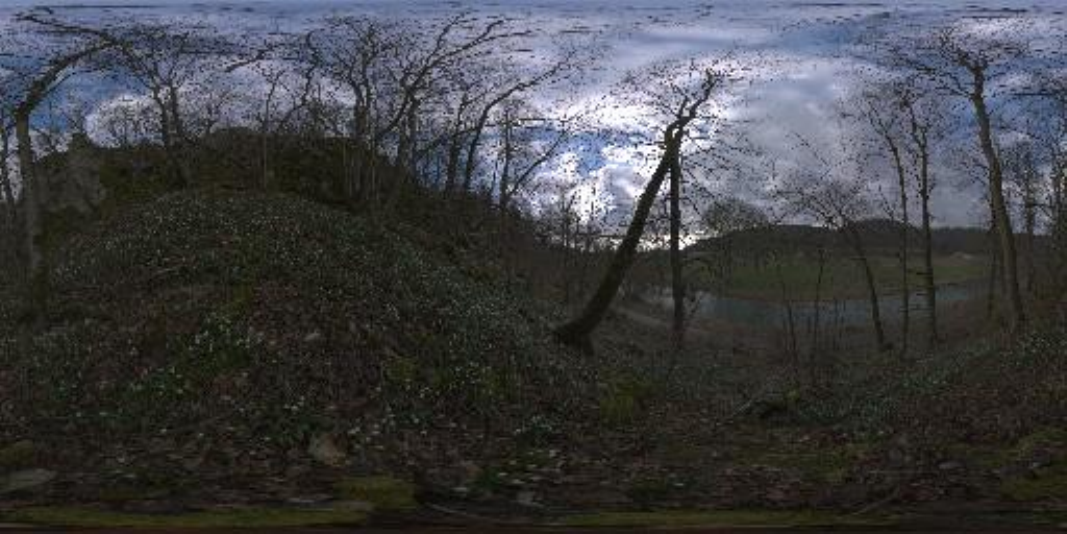}} & 
        \noindent\parbox[c]{0.14\textwidth}{\includegraphics[height=0.100\textwidth]{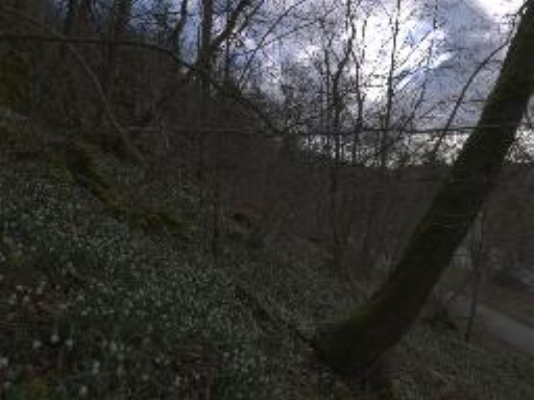}} &  
        
        \noindent\parbox[c]{0.100\textwidth}{\includegraphics[height=0.100\textwidth]{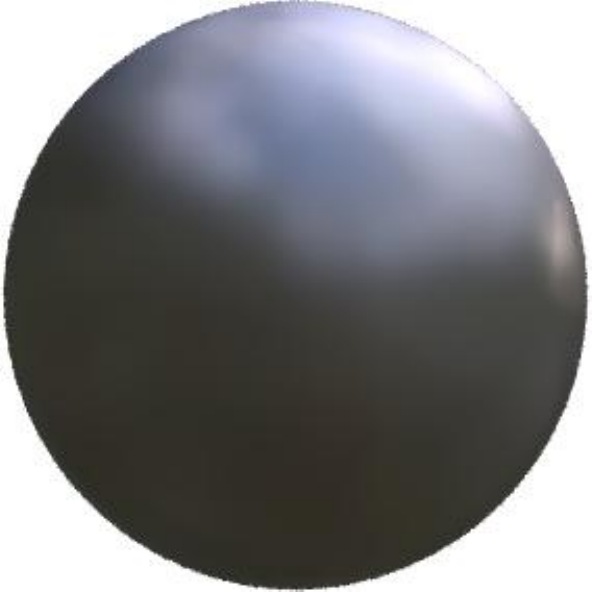}} & 
        \noindent\parbox[c]{0.100\textwidth}{\includegraphics[height=0.100\textwidth]{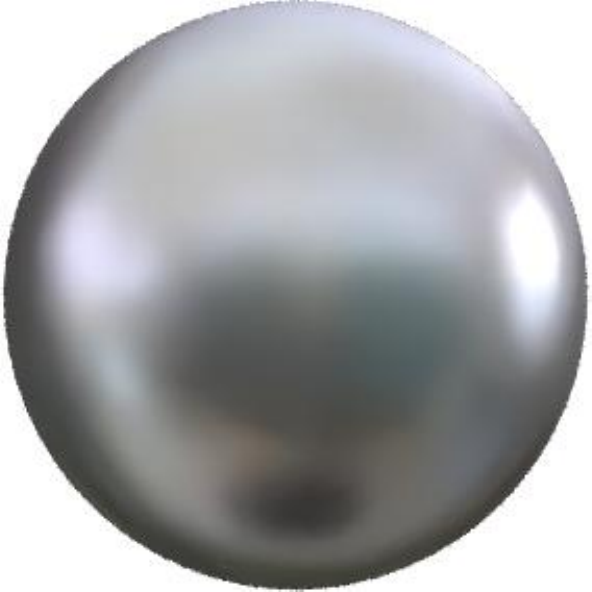}} & 
        
        \noindent\parbox[c]{0.100\textwidth}{\includegraphics[height=0.100\textwidth]{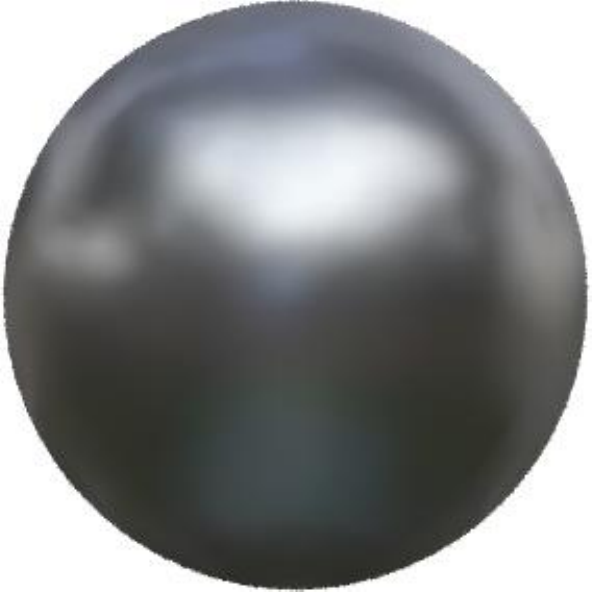}} & 
        \noindent\parbox[c]{0.100\textwidth}{\includegraphics[height=0.100\textwidth]{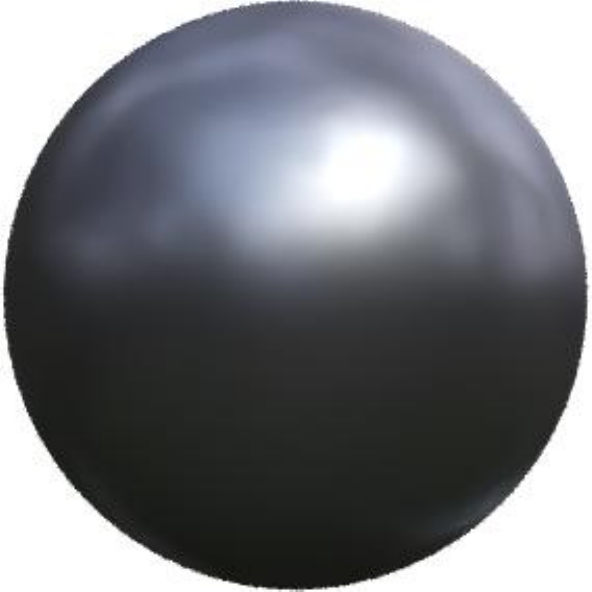}} &
        \noindent\parbox[c]{0.100\textwidth}{\includegraphics[height=0.100\textwidth]{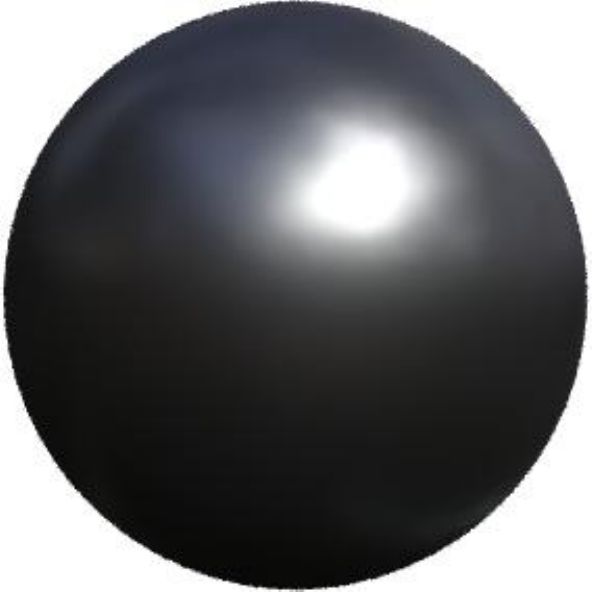}} & 
        \noindent\parbox[c]{0.100\textwidth}{\includegraphics[height=0.100\textwidth]{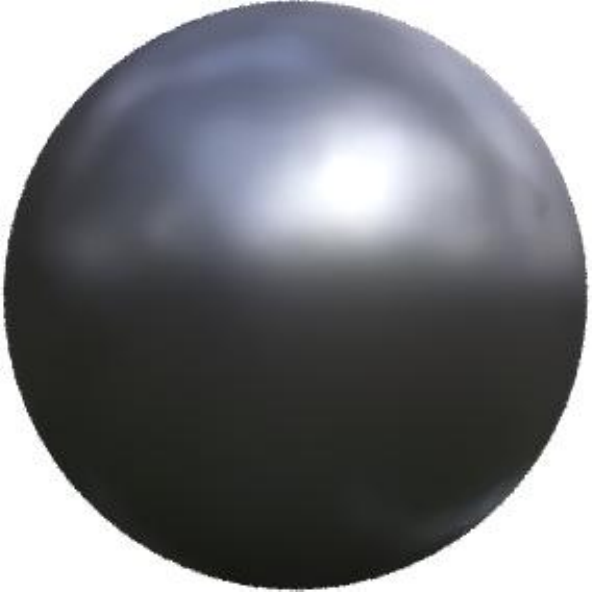}} & 
        \\

        \noindent\parbox[c]{0.205\textwidth}{\includegraphics[height=0.100\textwidth]{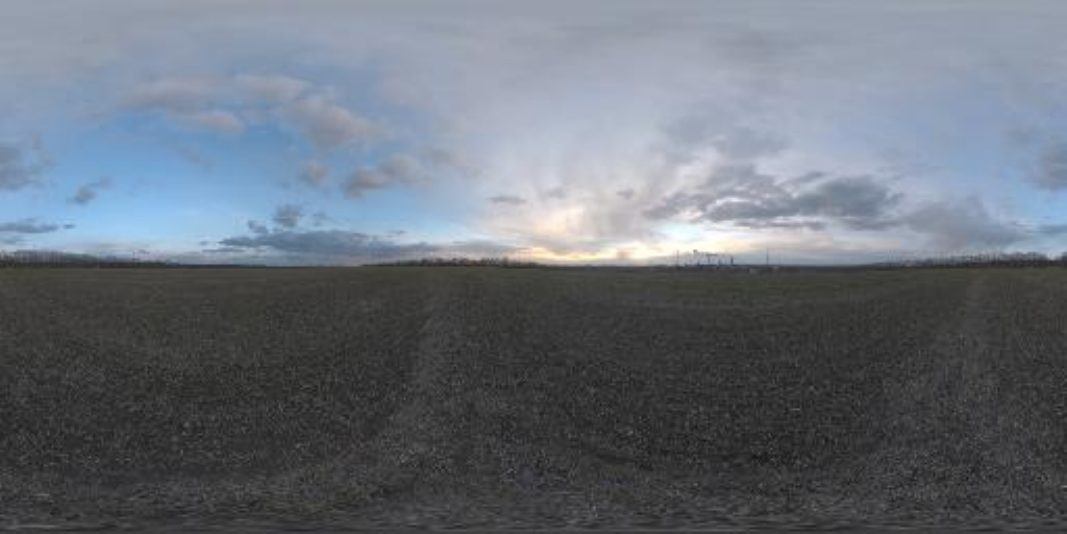}} & 
        \noindent\parbox[c]{0.14\textwidth}{\includegraphics[height=0.100\textwidth]{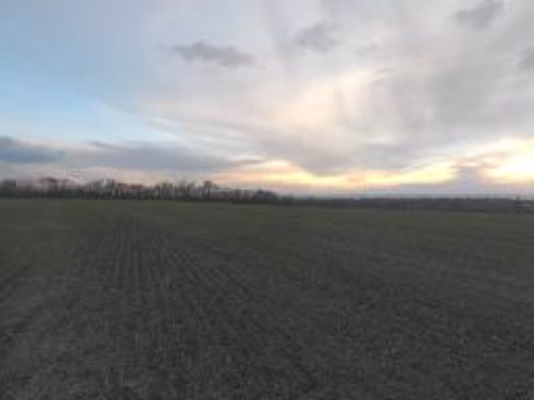}} &  
        
        \noindent\parbox[c]{0.100\textwidth}{\includegraphics[height=0.100\textwidth]{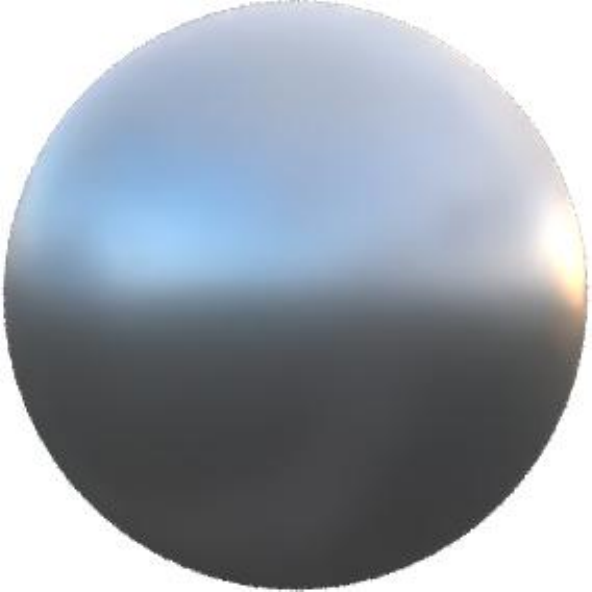}} & 
        \noindent\parbox[c]{0.100\textwidth}{\includegraphics[height=0.100\textwidth]{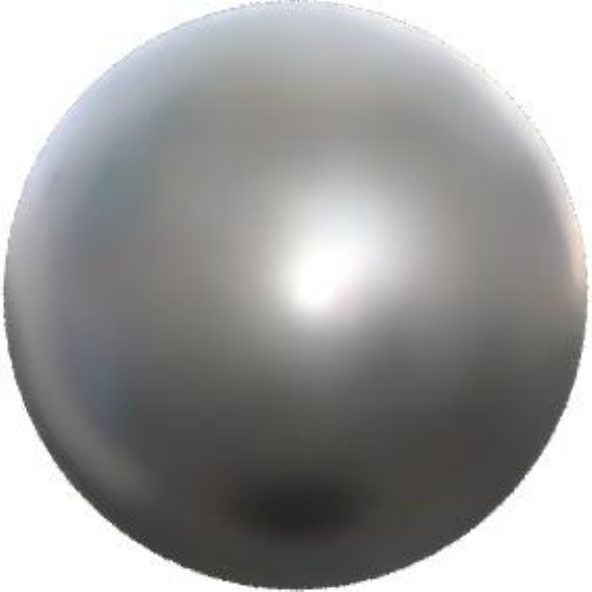}} & 
        
        \noindent\parbox[c]{0.100\textwidth}{\includegraphics[height=0.100\textwidth]{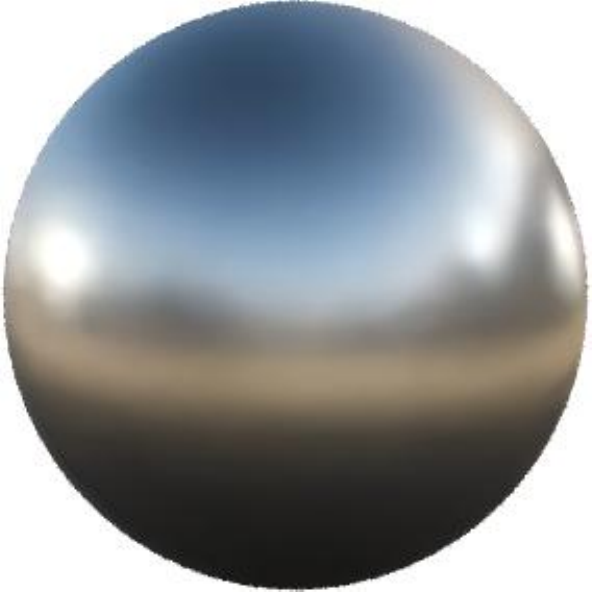}} & 
        \noindent\parbox[c]{0.100\textwidth}{\includegraphics[height=0.100\textwidth]{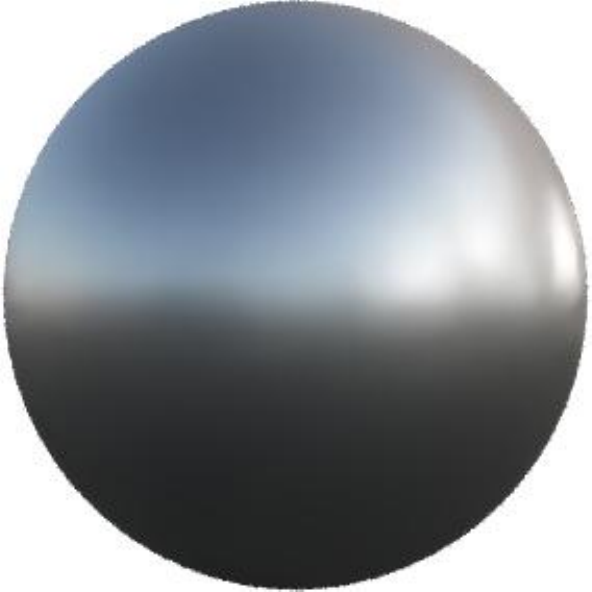}} &
        \noindent\parbox[c]{0.100\textwidth}{\includegraphics[height=0.100\textwidth]{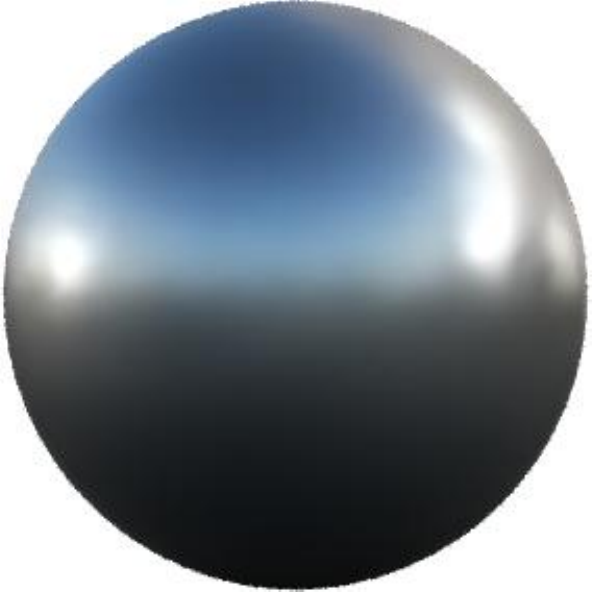}} & 
        \noindent\parbox[c]{0.100\textwidth}{\includegraphics[height=0.100\textwidth]{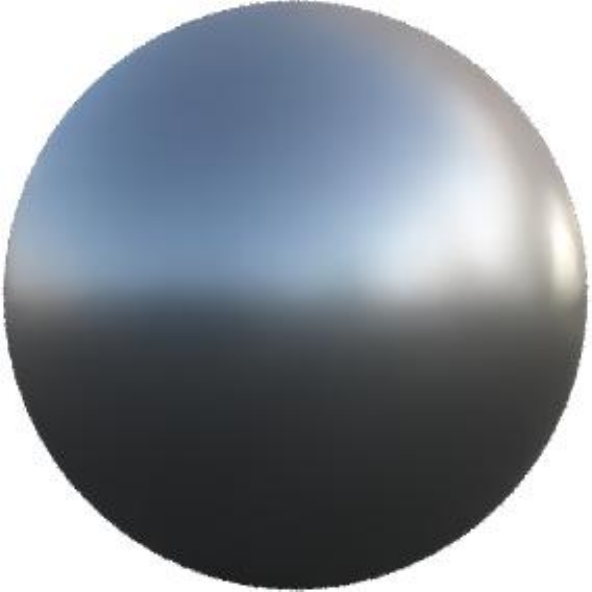}} & 
        \\

        \noindent\parbox[c]{0.205\textwidth}{\includegraphics[height=0.100\textwidth]{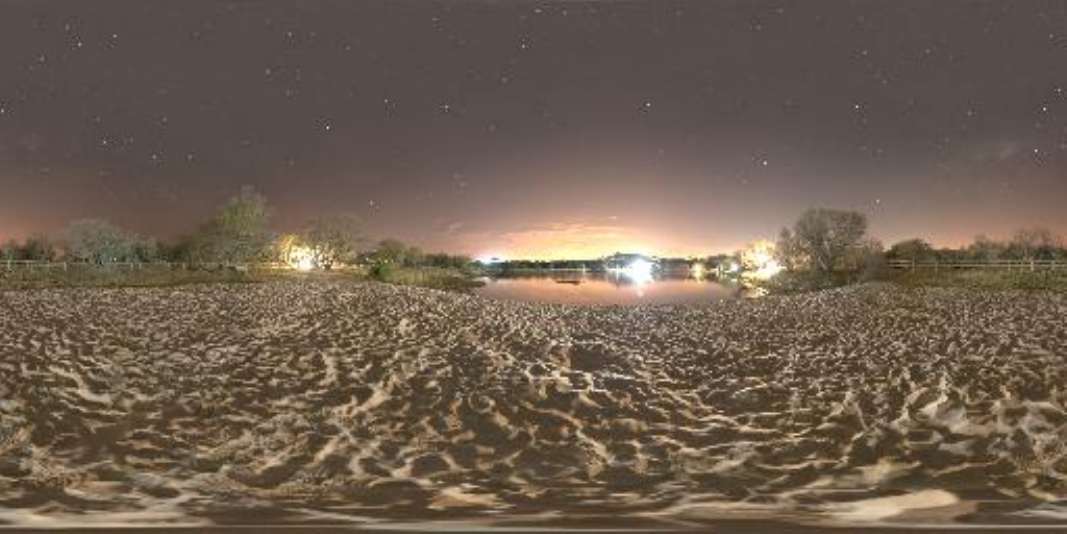}} & 
        \noindent\parbox[c]{0.14\textwidth}{\includegraphics[height=0.100\textwidth]{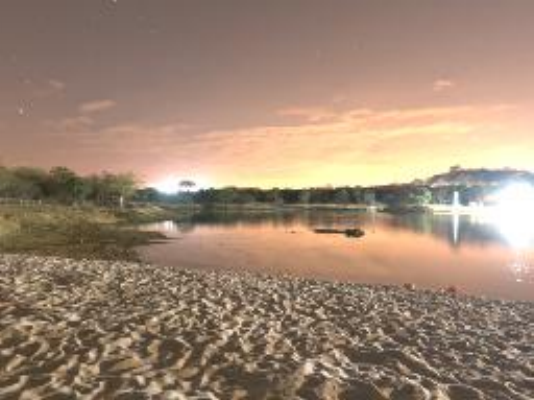}} &  
        
        \noindent\parbox[c]{0.100\textwidth}{\includegraphics[height=0.100\textwidth]{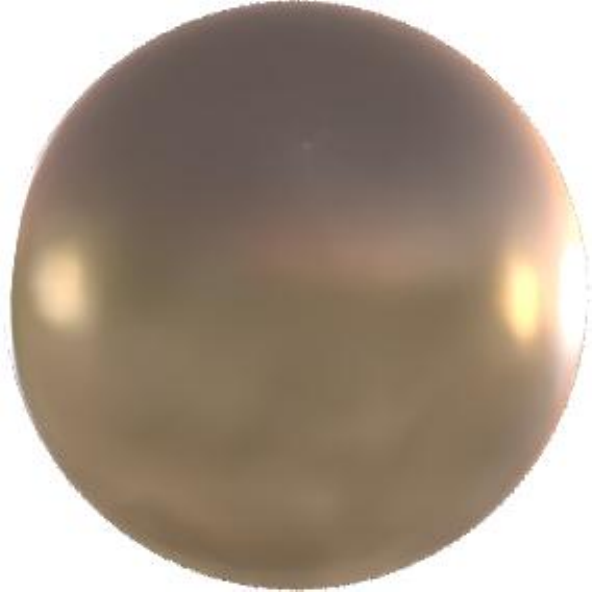}} & 
        \noindent\parbox[c]{0.100\textwidth}{\includegraphics[height=0.100\textwidth]{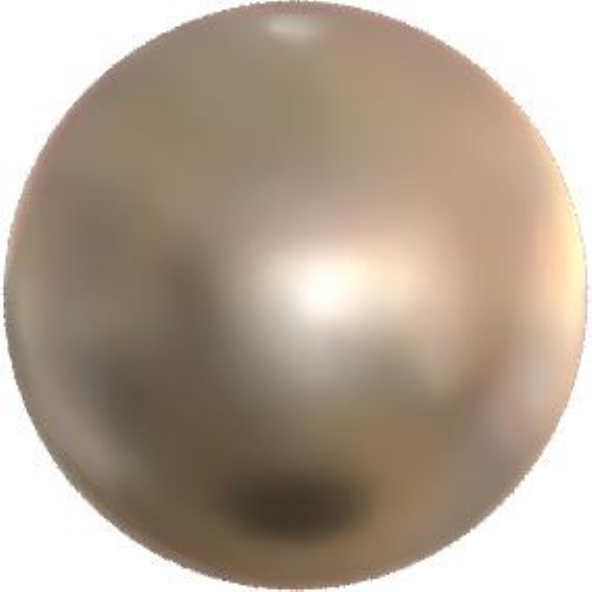}} & 
        
        \noindent\parbox[c]{0.100\textwidth}{\includegraphics[height=0.100\textwidth]{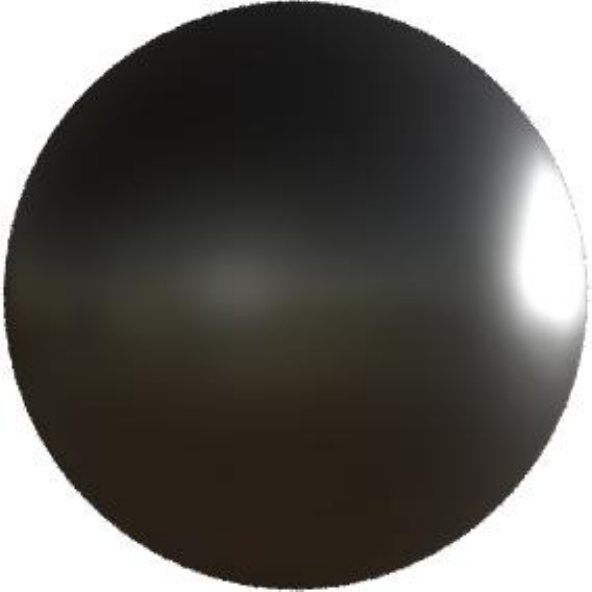}} & 
        \noindent\parbox[c]{0.100\textwidth}{\includegraphics[height=0.100\textwidth]{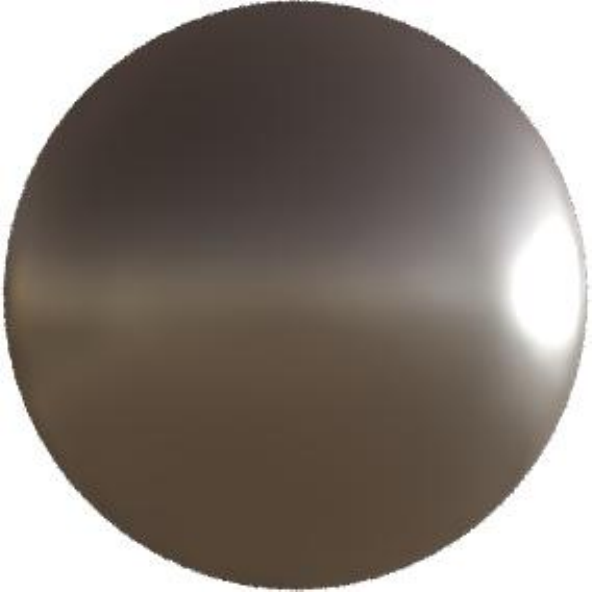}} &
        \noindent\parbox[c]{0.100\textwidth}{\includegraphics[height=0.100\textwidth]{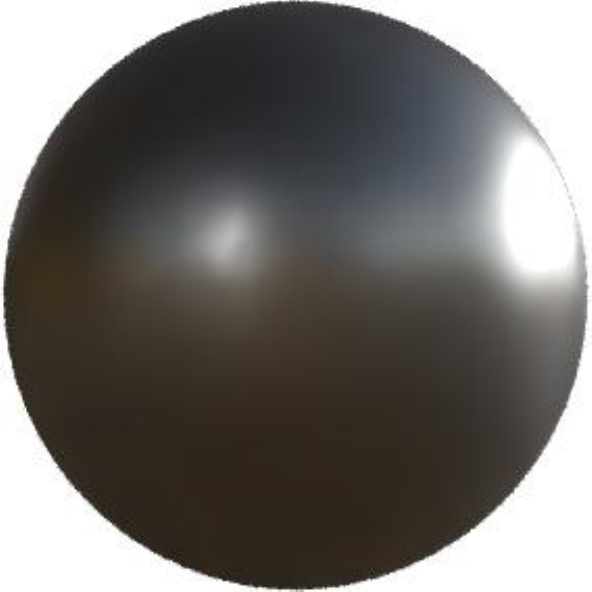}} & 
        \noindent\parbox[c]{0.100\textwidth}{\includegraphics[height=0.100\textwidth]{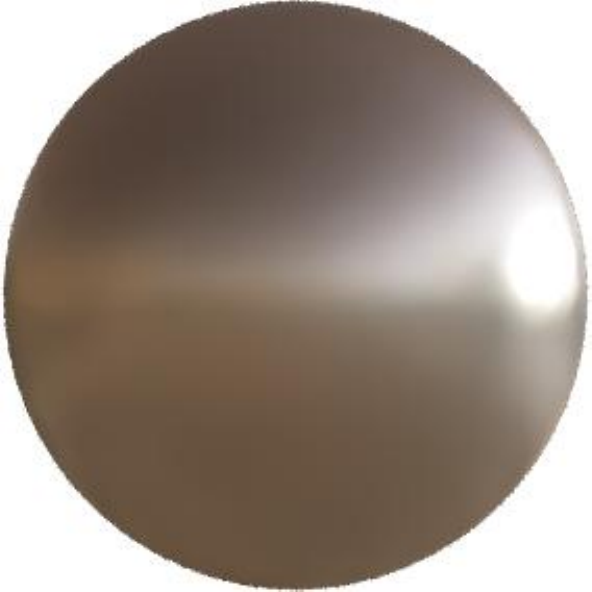}} & 
        \\

        \noindent\parbox[c]{0.205\textwidth}{\includegraphics[height=0.100\textwidth]{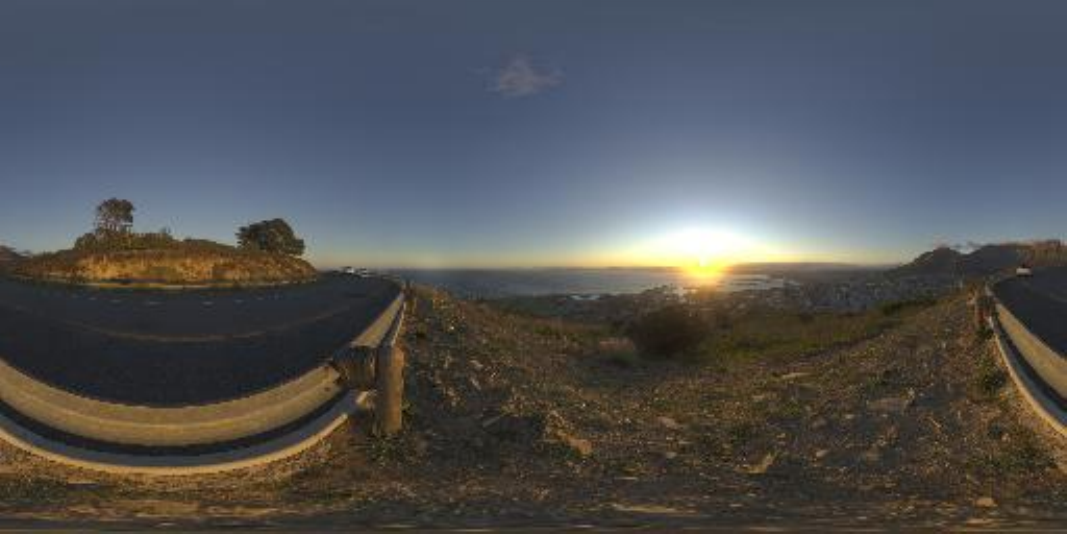}} & 
        \noindent\parbox[c]{0.14\textwidth}{\includegraphics[height=0.100\textwidth]{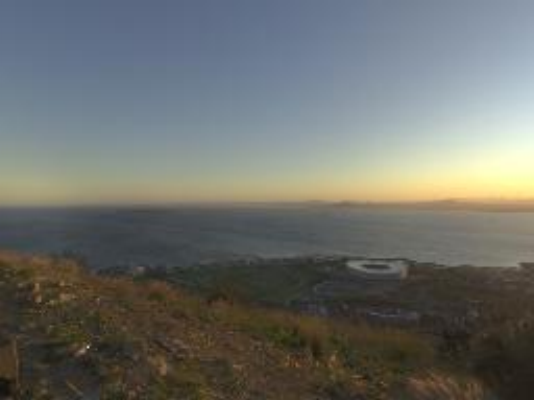}} &  
        
        \noindent\parbox[c]{0.100\textwidth}{\includegraphics[height=0.100\textwidth]{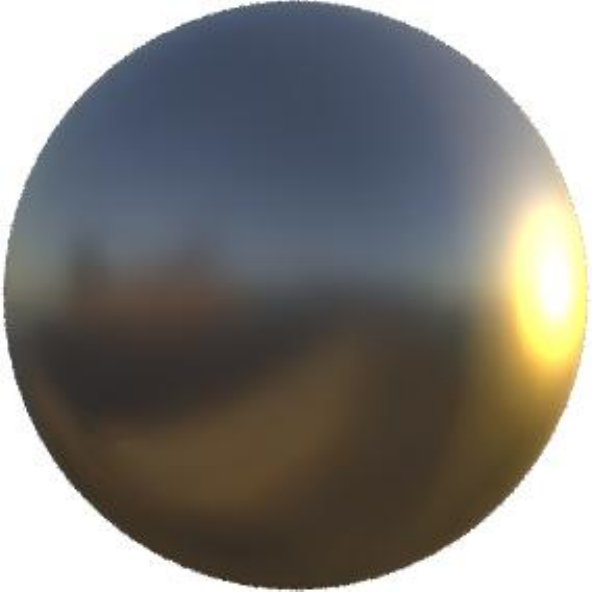}} & 
        \noindent\parbox[c]{0.100\textwidth}{\includegraphics[height=0.100\textwidth]{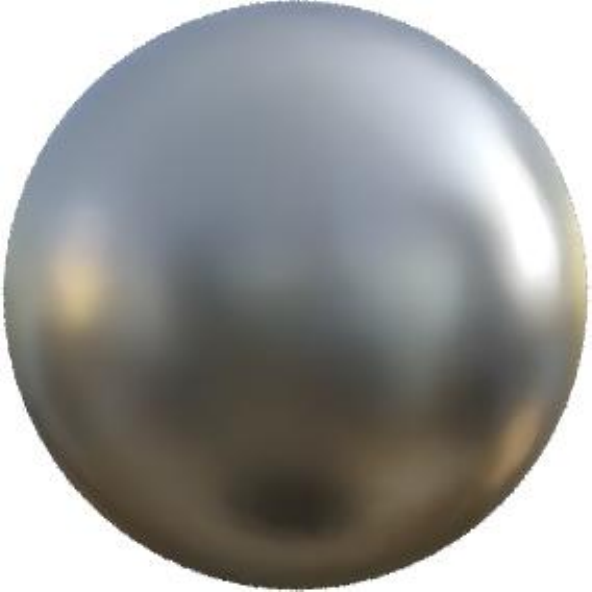}} & 
        
        \noindent\parbox[c]{0.100\textwidth}{\includegraphics[height=0.100\textwidth]{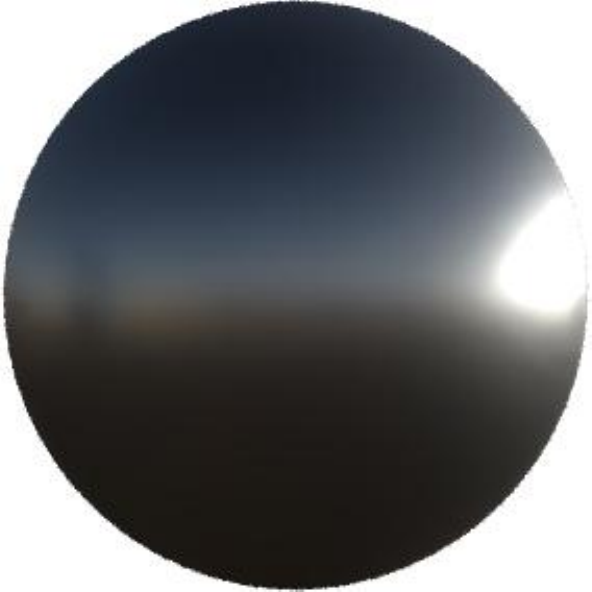}} & 
        \noindent\parbox[c]{0.100\textwidth}{\includegraphics[height=0.100\textwidth]{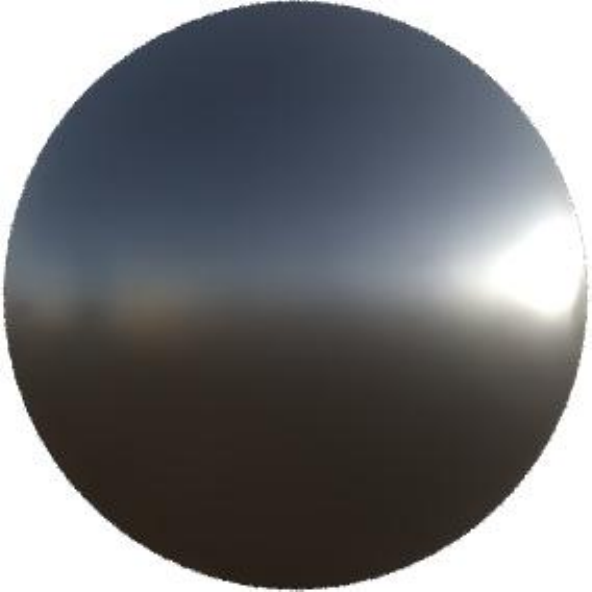}} &
        \noindent\parbox[c]{0.100\textwidth}{\includegraphics[height=0.100\textwidth]{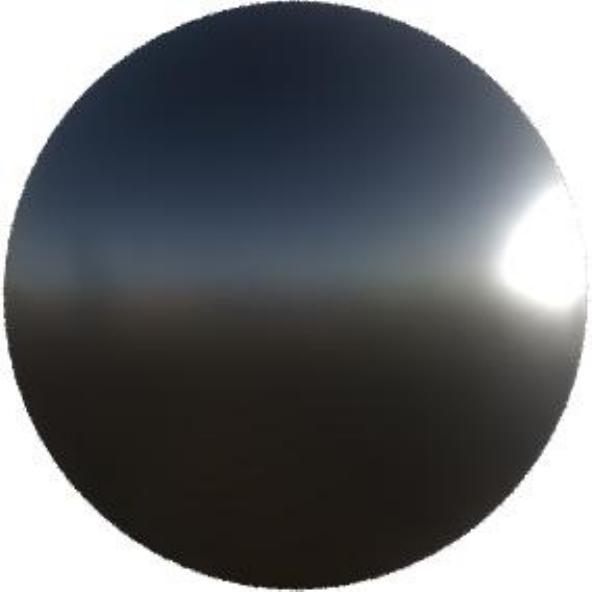}} & 
        \noindent\parbox[c]{0.100\textwidth}{\includegraphics[height=0.100\textwidth]{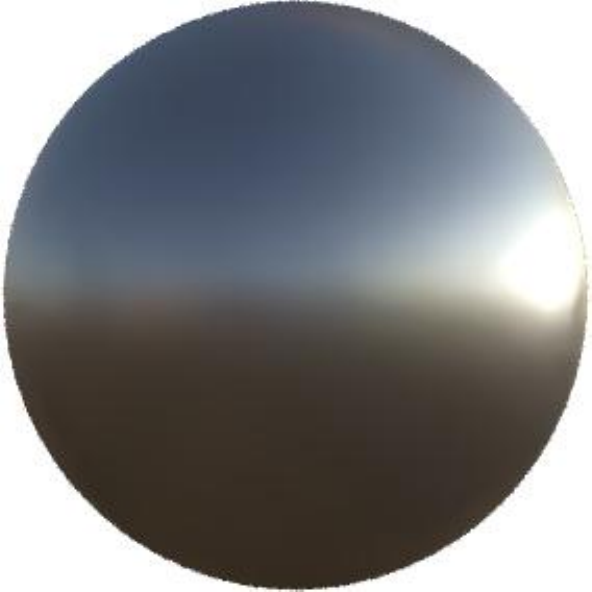}} & 
        \\

        \noindent\parbox[c]{0.205\textwidth}{\includegraphics[height=0.100\textwidth]{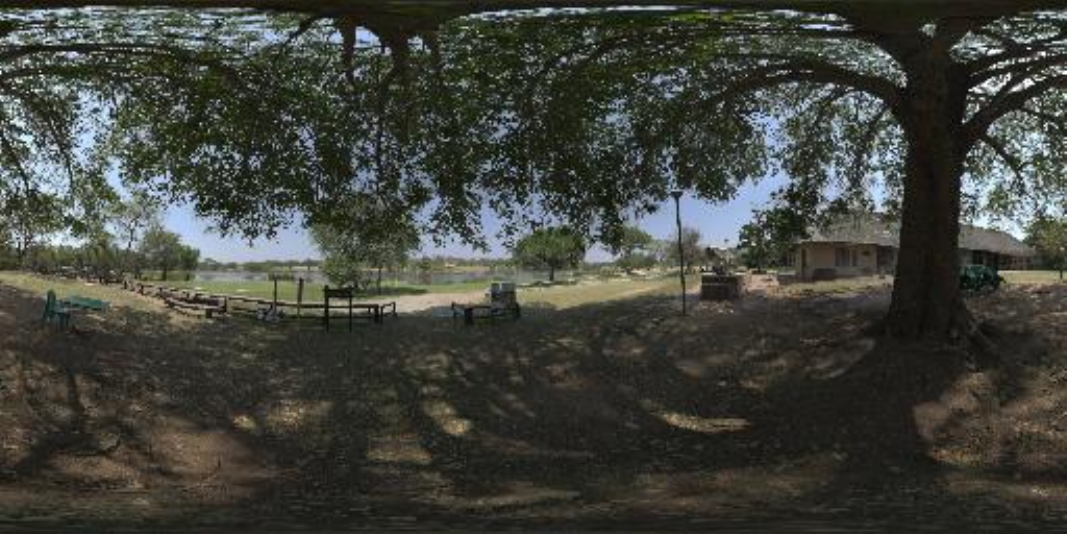}} & 
        \noindent\parbox[c]{0.14\textwidth}{\includegraphics[height=0.100\textwidth]{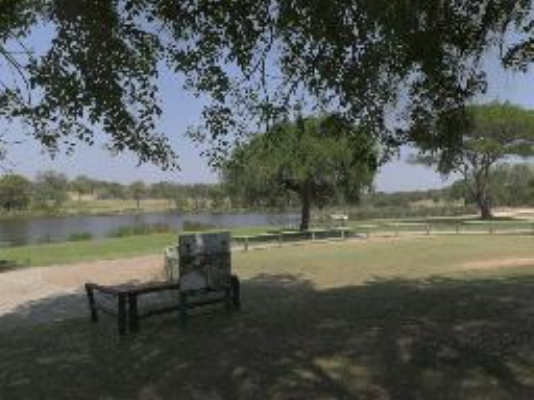}} &  
        
        \noindent\parbox[c]{0.100\textwidth}{\includegraphics[height=0.100\textwidth]{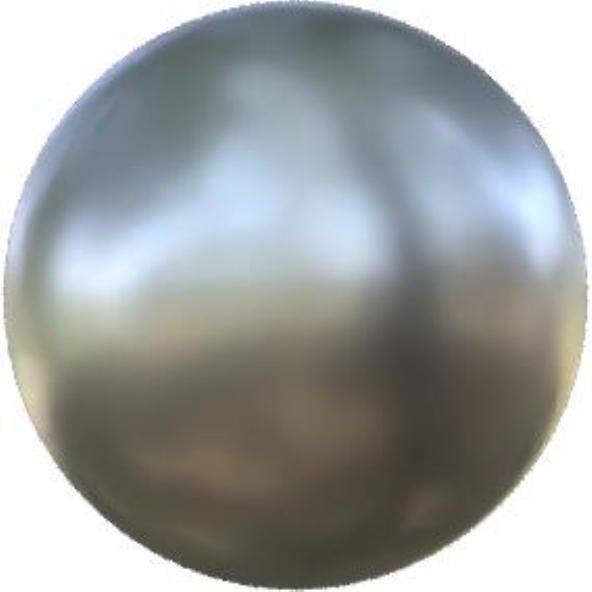}} & 
        \noindent\parbox[c]{0.100\textwidth}{\includegraphics[height=0.100\textwidth]{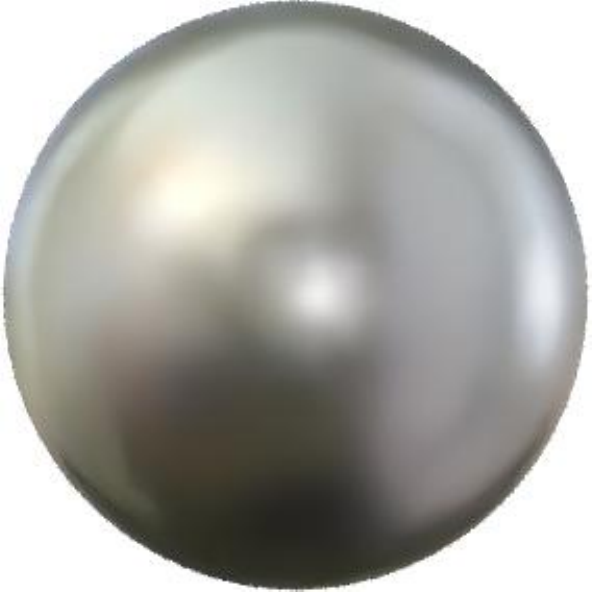}} & 
        
        \noindent\parbox[c]{0.100\textwidth}{\includegraphics[height=0.100\textwidth]{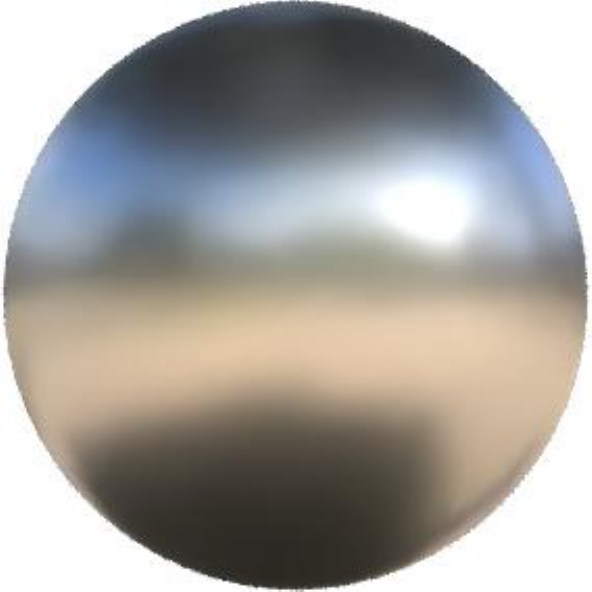}} & 
        \noindent\parbox[c]{0.100\textwidth}{\includegraphics[height=0.100\textwidth]{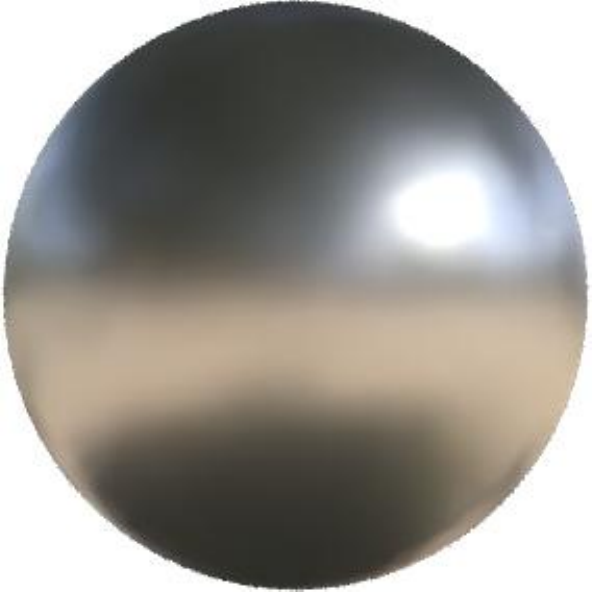}} &
        \noindent\parbox[c]{0.100\textwidth}{\includegraphics[height=0.100\textwidth]{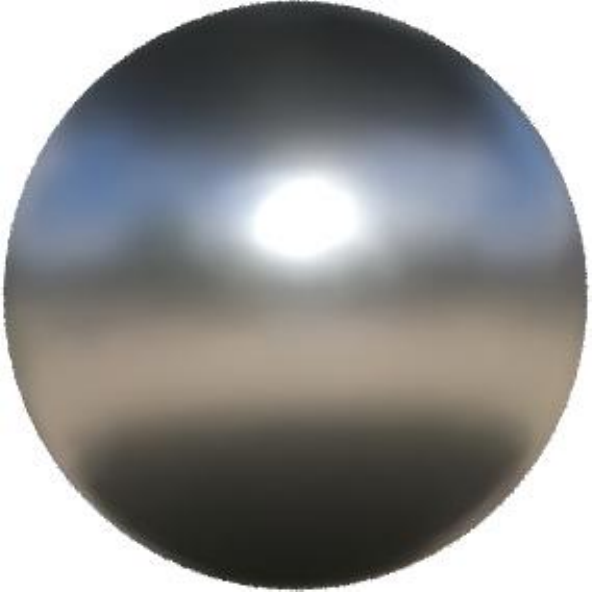}} & 
        \noindent\parbox[c]{0.100\textwidth}{\includegraphics[height=0.100\textwidth]{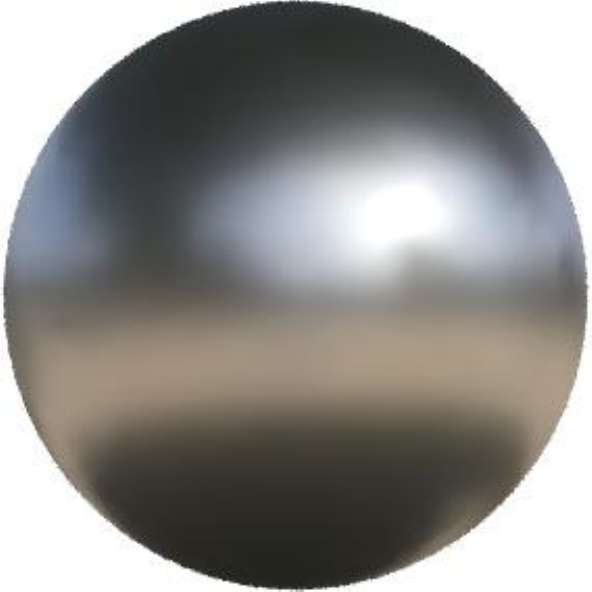}} & 
        \\

        \noindent\parbox[c]{0.205\textwidth}{\includegraphics[height=0.100\textwidth]{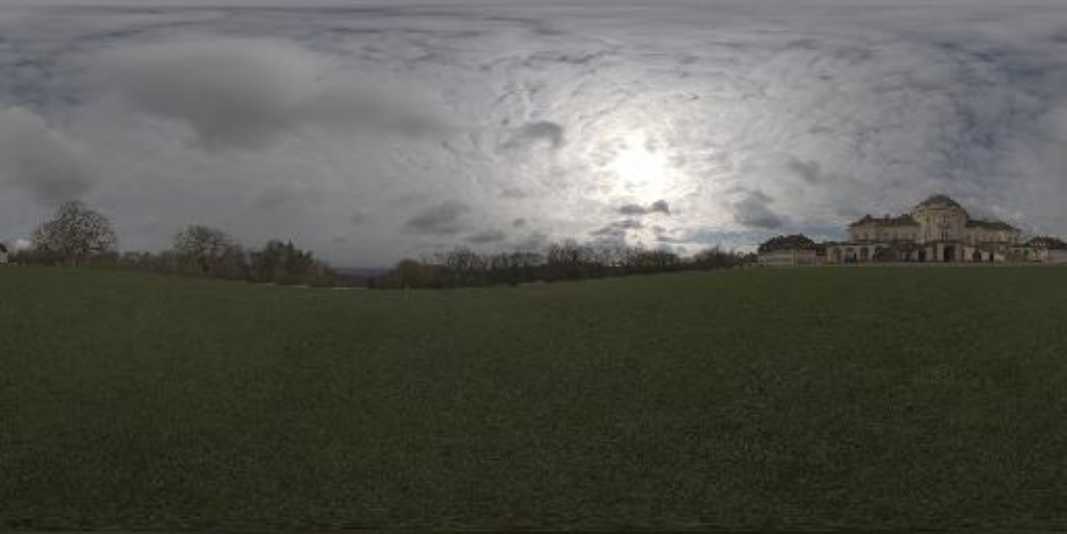}} & 
        \noindent\parbox[c]{0.14\textwidth}{\includegraphics[height=0.100\textwidth]{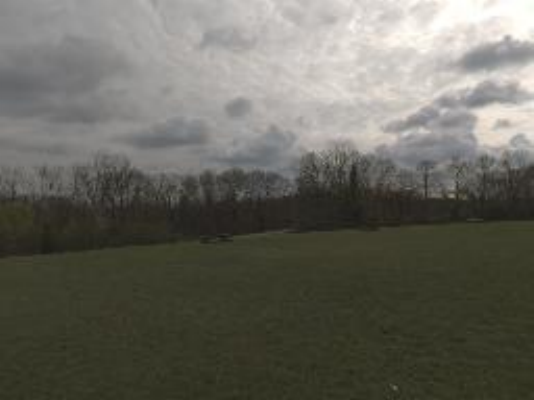}} &  
        
        \noindent\parbox[c]{0.100\textwidth}{\includegraphics[height=0.100\textwidth]{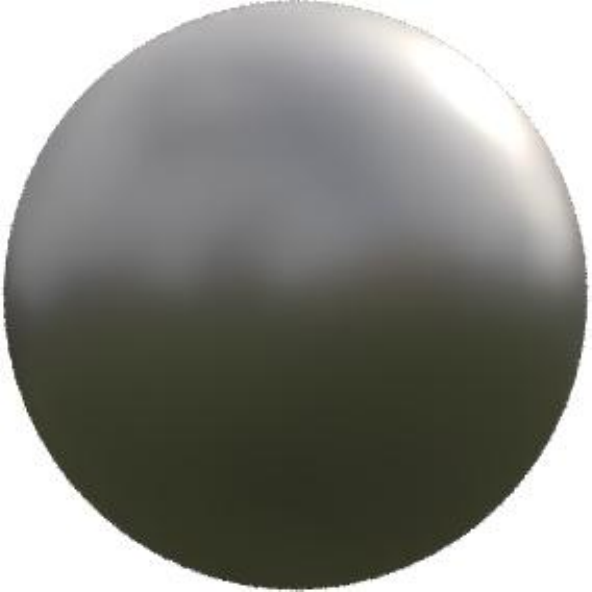}} & 
        \noindent\parbox[c]{0.100\textwidth}{\includegraphics[height=0.100\textwidth]{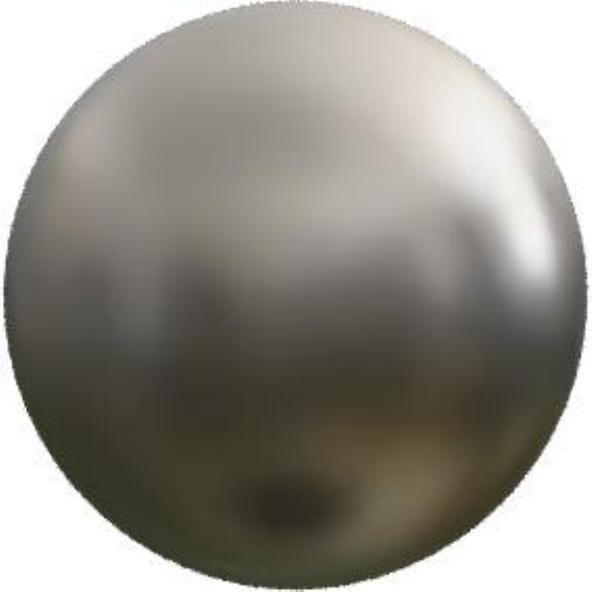}} & 
        
        \noindent\parbox[c]{0.100\textwidth}{\includegraphics[height=0.100\textwidth]{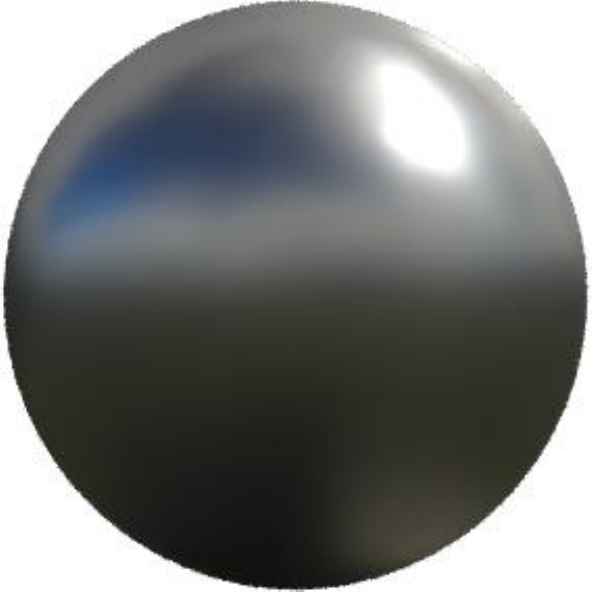}} & 
        \noindent\parbox[c]{0.100\textwidth}{\includegraphics[height=0.100\textwidth]{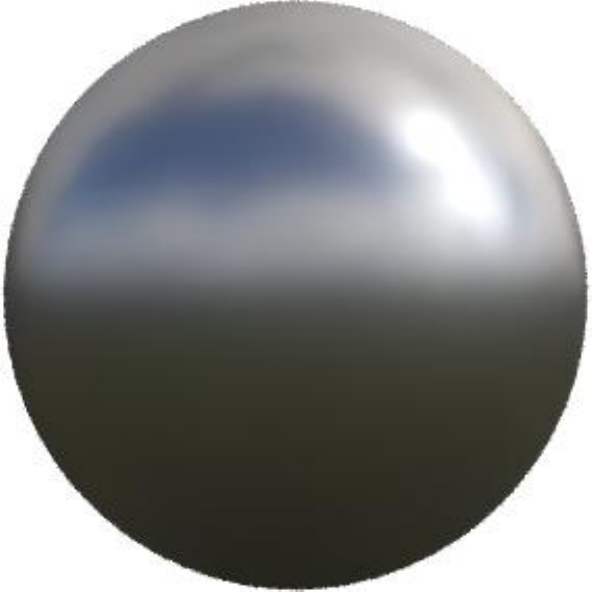}} &
        \noindent\parbox[c]{0.100\textwidth}{\includegraphics[height=0.100\textwidth]{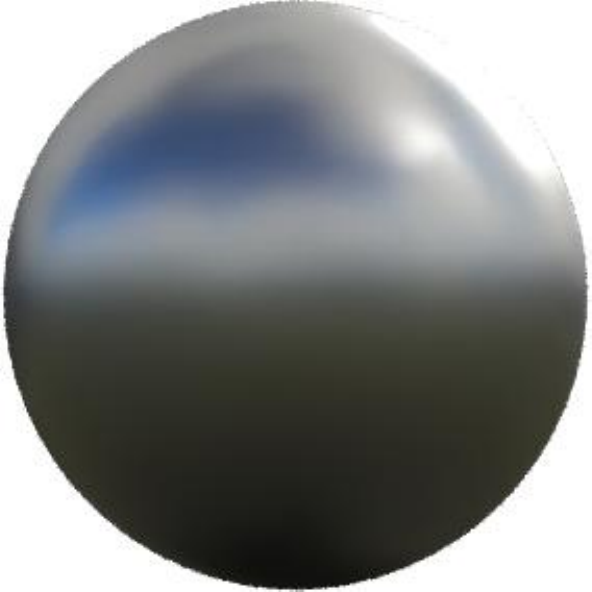}} & 
        \noindent\parbox[c]{0.100\textwidth}{\includegraphics[height=0.100\textwidth]{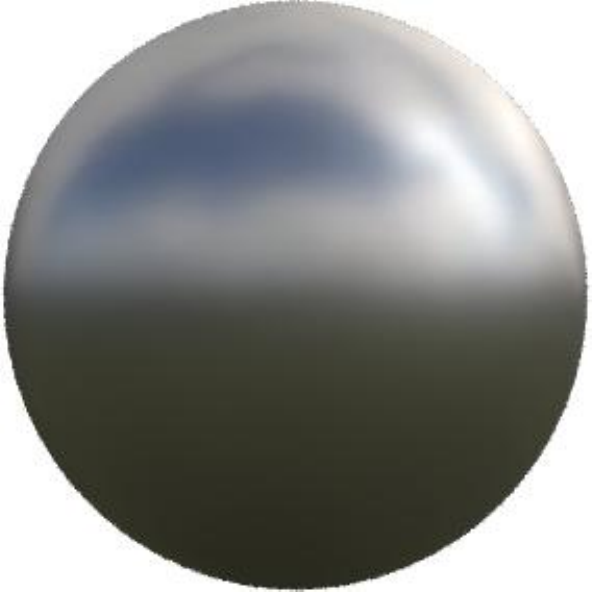}} & 
        \\

        \noindent\parbox[c]{0.205\textwidth}{\includegraphics[height=0.100\textwidth]{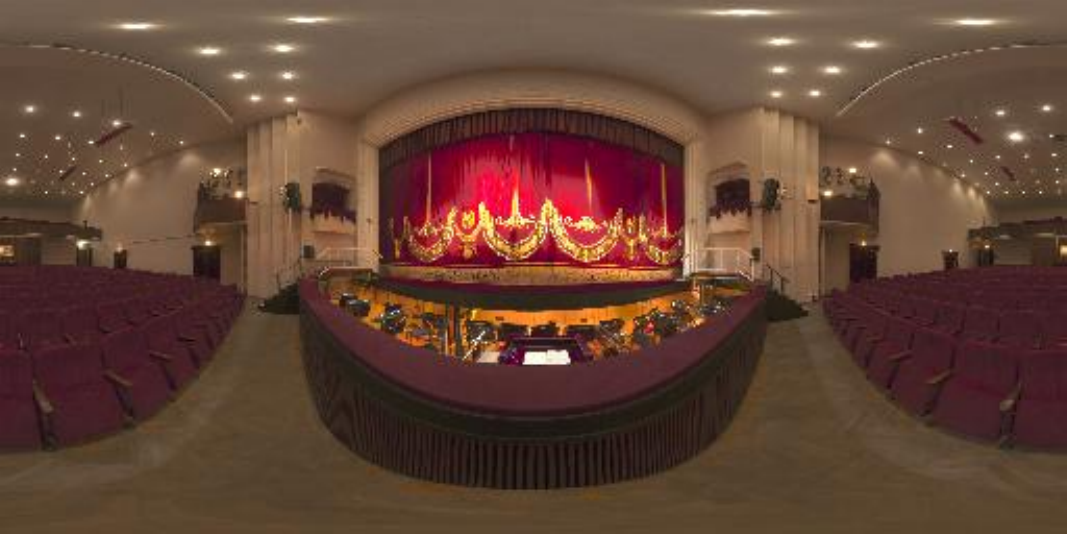}} & 
        \noindent\parbox[c]{0.14\textwidth}{\includegraphics[height=0.100\textwidth]{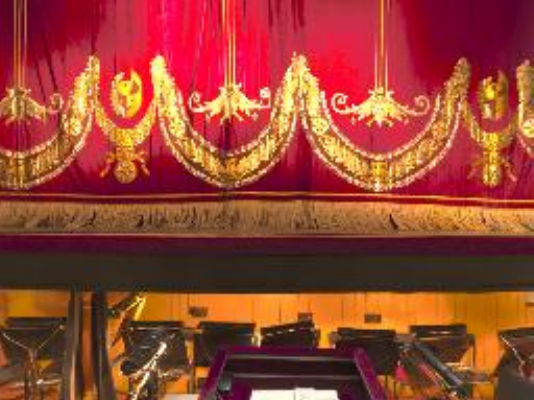}} &  
        
        \noindent\parbox[c]{0.100\textwidth}{\includegraphics[height=0.100\textwidth]{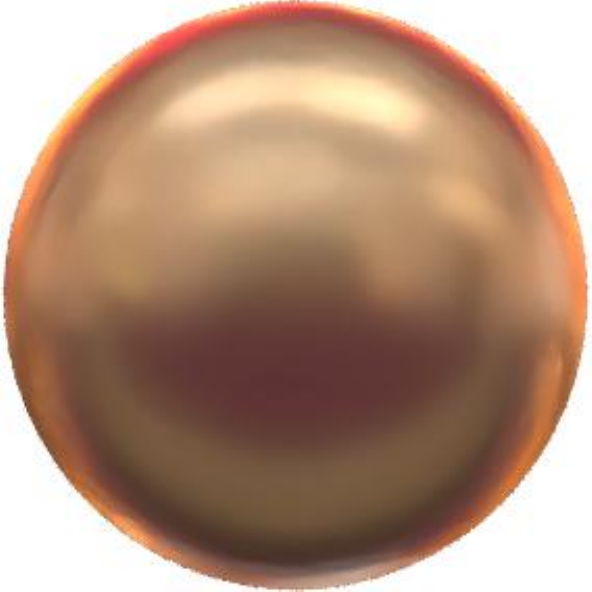}} & 
        \noindent\parbox[c]{0.100\textwidth}{\includegraphics[height=0.100\textwidth]{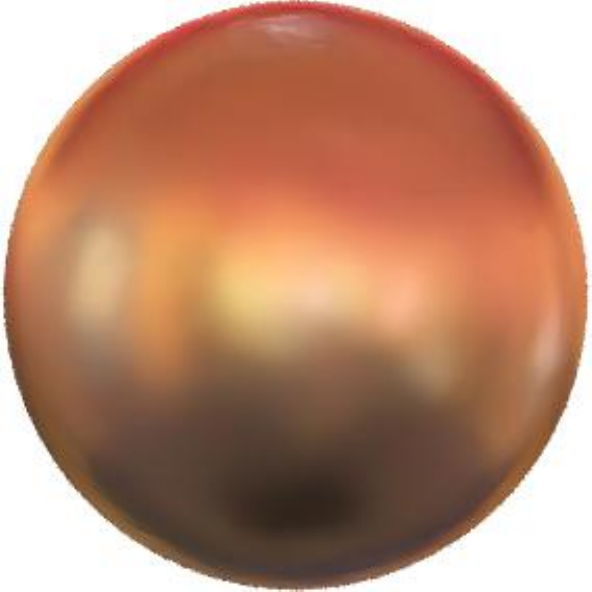}} & 
        
        \noindent\parbox[c]{0.100\textwidth}{\includegraphics[height=0.100\textwidth]{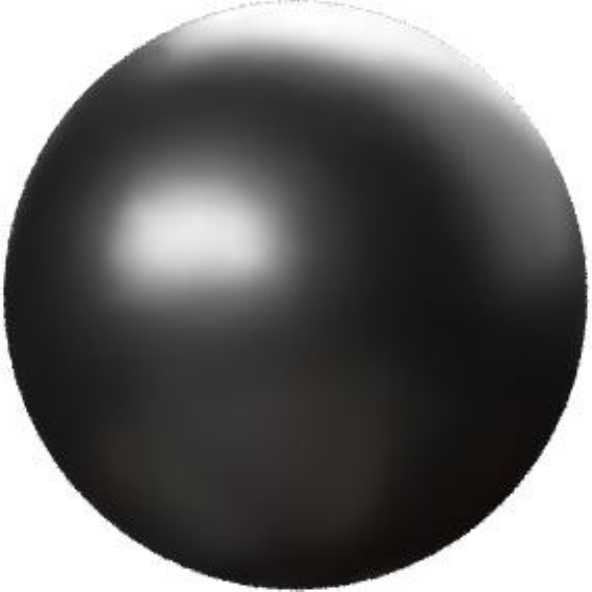}} & 
        \noindent\parbox[c]{0.100\textwidth}{\includegraphics[height=0.100\textwidth]{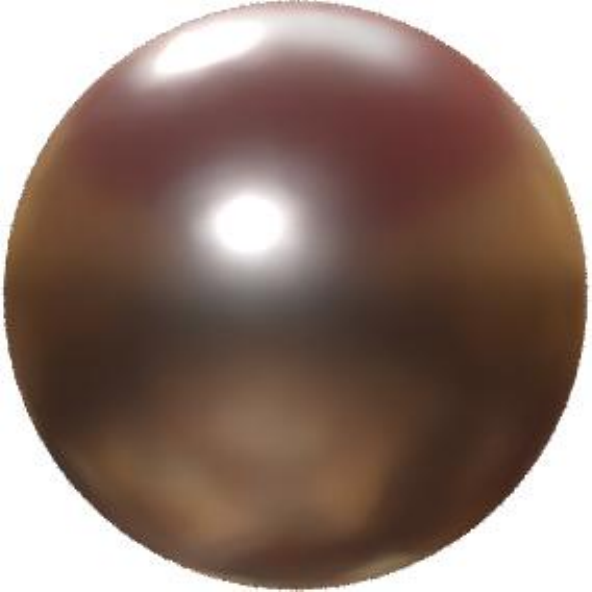}} &
        \noindent\parbox[c]{0.100\textwidth}{\includegraphics[height=0.100\textwidth]{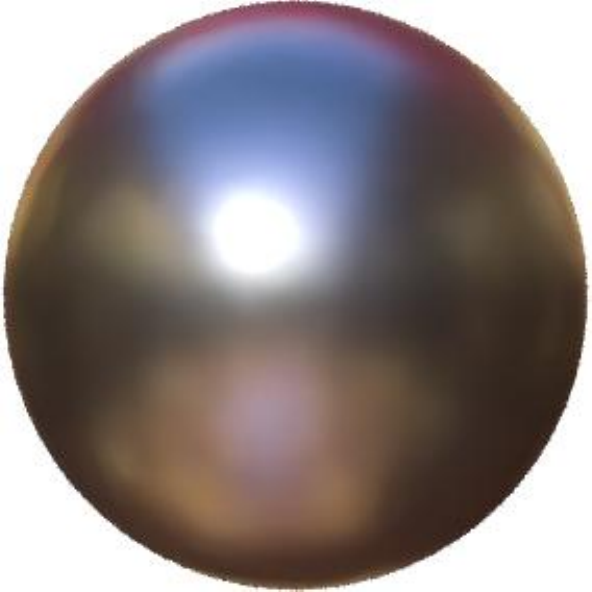}} & 
        \noindent\parbox[c]{0.100\textwidth}{\includegraphics[height=0.100\textwidth]{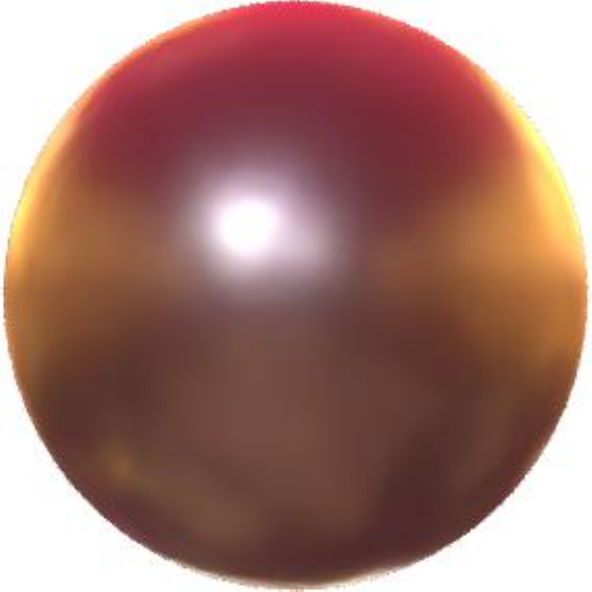}} & 
        \\
        
        \end{tabu}
    \caption{
    Qualitative results for the Poly Haven dataset using matte balls.}
    \label{fig:additional_polyhaven_matte}
\end{figure*}

\begin{figure*}[!t]
    \centering

        \begin{tabu} to \textwidth {
        @{}
        c@{}
        c@{}
        c@{}
        c@{}
        c@{}
        c@{}
        c@{}
        c@{}
        c@{}
    }

        \multicolumn{1}{c}{\shortstack{\scriptsize Ground truth map}}
        & 
        \multicolumn{1}{c}{\shortstack{\hspace{-6pt} \scriptsize Input}}
        &
        \multicolumn{1}{c}{\shortstack{\scriptsize Ground truth}}
        & 
        \multicolumn{1}{c}{\shortstack{\scriptsize StyleLight \cite{wang2022stylelight}}}
        & 
        \multicolumn{1}{c}{\shortstack{\scriptsize SDXL$^\dagger$}} &
        \multicolumn{1}{c}{\shortstack{\scriptsize \begin{tabular}[c]{@{}c@{}}SDXL$^\dagger$+LR \\ (ours, ablated)\end{tabular}}} &
        \multicolumn{1}{c}{\shortstack{\scriptsize \begin{tabular}[c]{@{}c@{}}SDXL$^\dagger$+I \\ (ours,ablated)\end{tabular}}}
        &
        \multicolumn{1}{c}{\shortstack{\scriptsize \begin{tabular}[c]{@{}c@{}}SDXL$^\dagger$+LR+I \\ (ours)\end{tabular}}} 
        \\

        \noindent\parbox[c]{0.205\textwidth}{\includegraphics[height=0.100\textwidth]{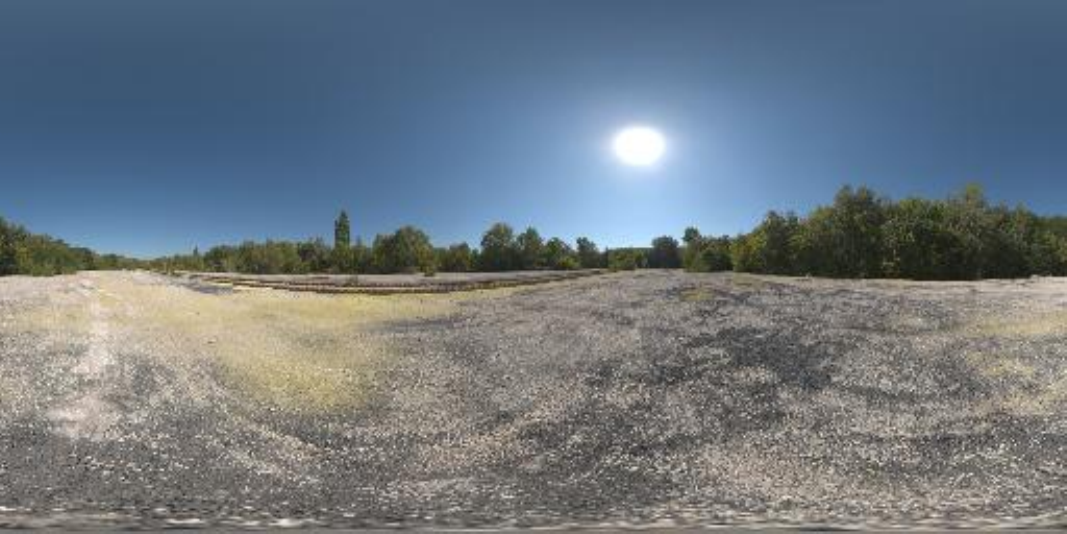}} & 
        \noindent\parbox[c]{0.14\textwidth}{\includegraphics[height=0.100\textwidth]{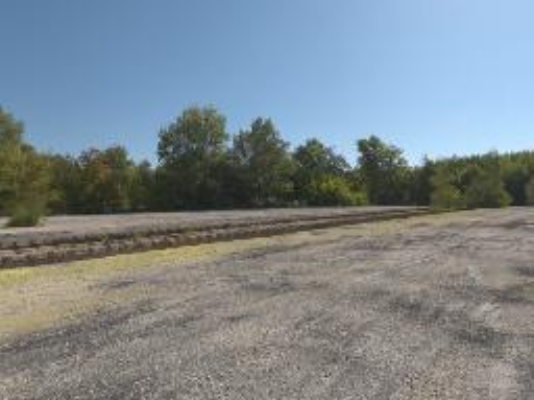}} &  
        
        \noindent\parbox[c]{0.100\textwidth}{\includegraphics[height=0.100\textwidth]{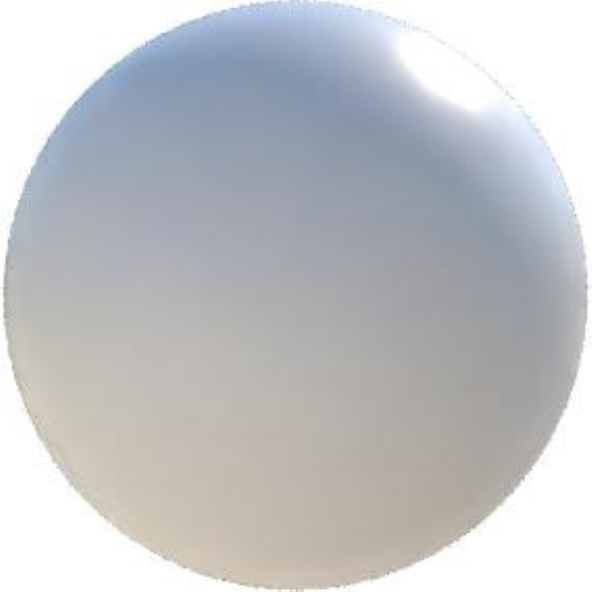}} & 
        \noindent\parbox[c]{0.100\textwidth}{\includegraphics[height=0.100\textwidth]{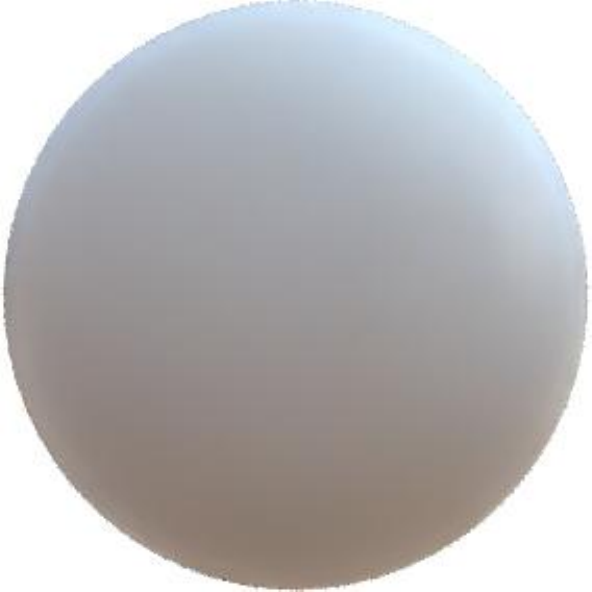}} & 
        
        \noindent\parbox[c]{0.100\textwidth}{\includegraphics[height=0.100\textwidth]{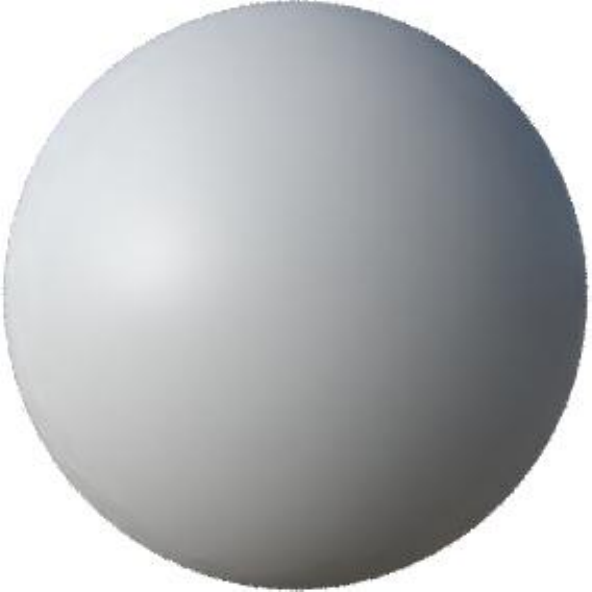}} & 
        \noindent\parbox[c]{0.100\textwidth}{\includegraphics[height=0.100\textwidth]{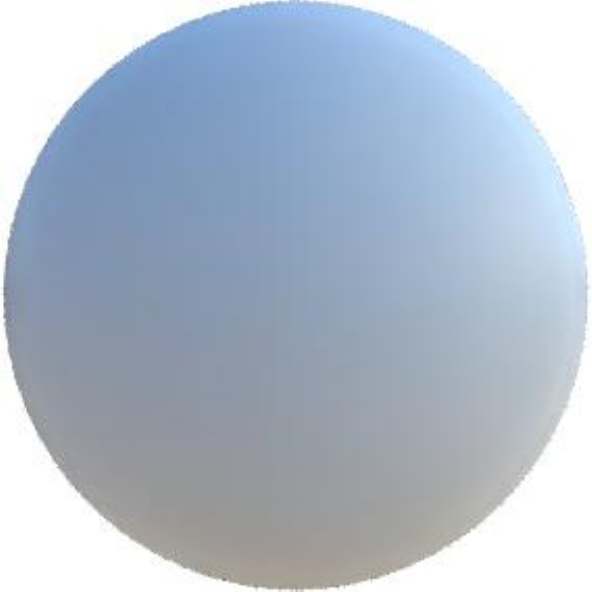}} &
        \noindent\parbox[c]{0.100\textwidth}{\includegraphics[height=0.100\textwidth]{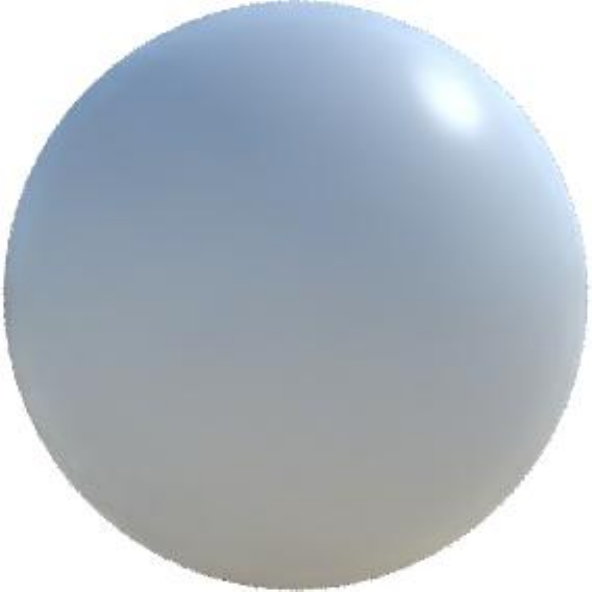}} & 
        \noindent\parbox[c]{0.100\textwidth}{\includegraphics[height=0.100\textwidth]{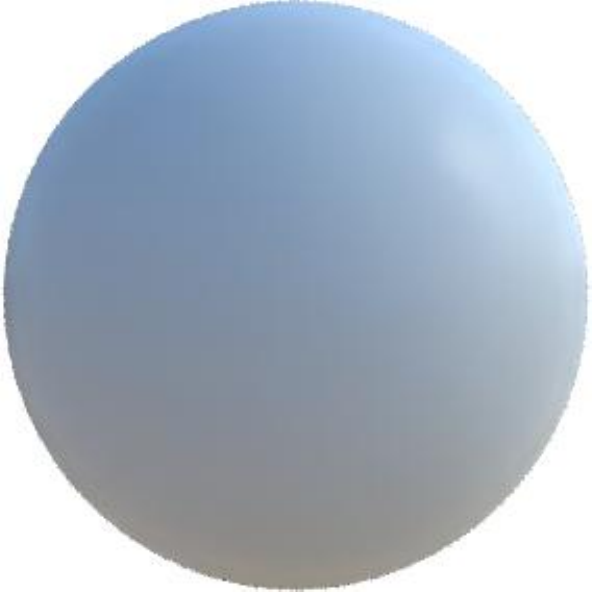}} & 
        \\

        \noindent\parbox[c]{0.205\textwidth}{\includegraphics[height=0.100\textwidth]{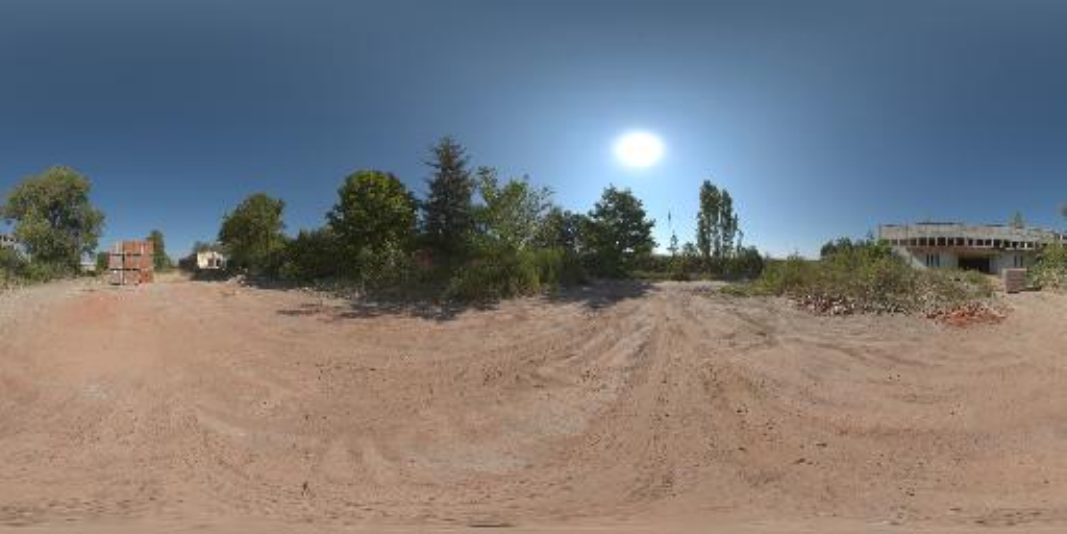}} & 
        \noindent\parbox[c]{0.14\textwidth}{\includegraphics[height=0.100\textwidth]{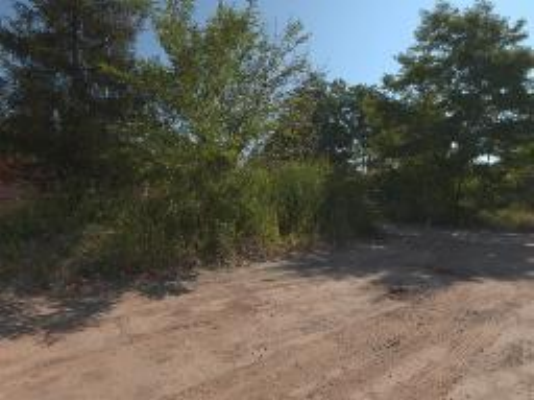}} &  
        
        \noindent\parbox[c]{0.100\textwidth}{\includegraphics[height=0.100\textwidth]{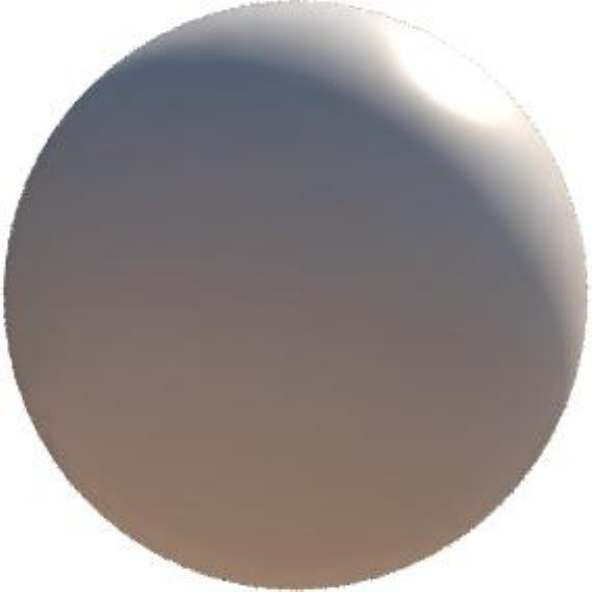}} & 
        \noindent\parbox[c]{0.100\textwidth}{\includegraphics[height=0.100\textwidth]{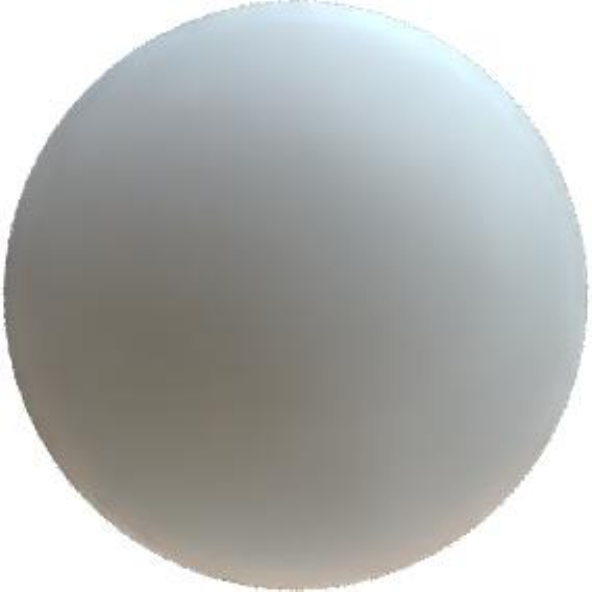}} & 
        
        \noindent\parbox[c]{0.100\textwidth}{\includegraphics[height=0.100\textwidth]{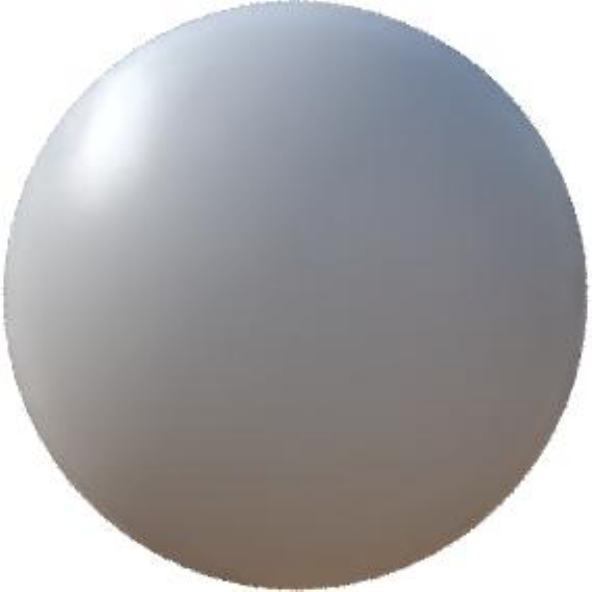}} & 
        \noindent\parbox[c]{0.100\textwidth}{\includegraphics[height=0.100\textwidth]{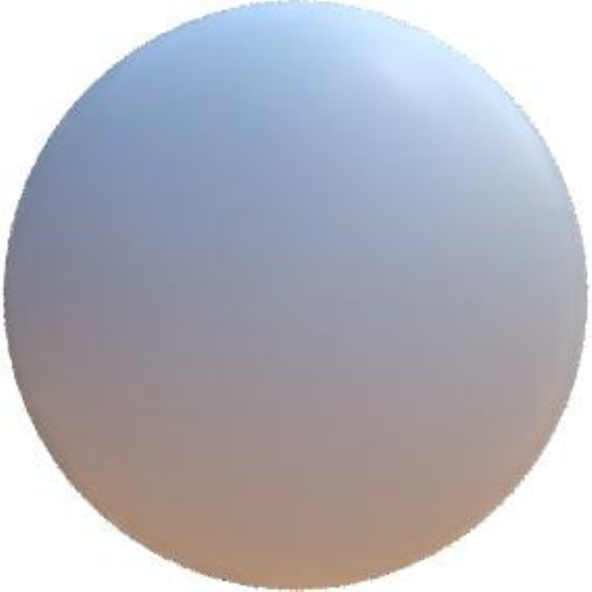}} &
        \noindent\parbox[c]{0.100\textwidth}{\includegraphics[height=0.100\textwidth]{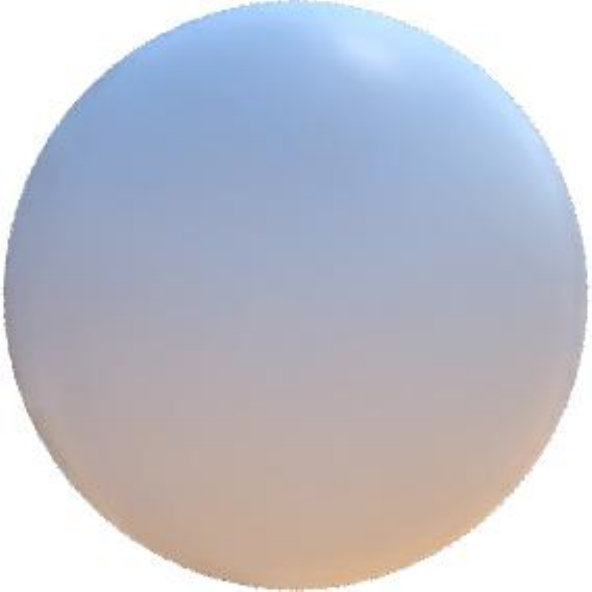}} & 
        \noindent\parbox[c]{0.100\textwidth}{\includegraphics[height=0.100\textwidth]{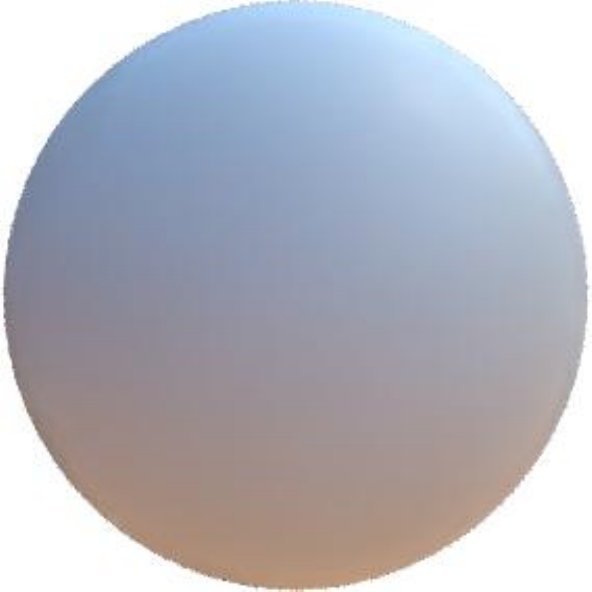}} & 
        \\

        \noindent\parbox[c]{0.205\textwidth}{\includegraphics[height=0.100\textwidth]{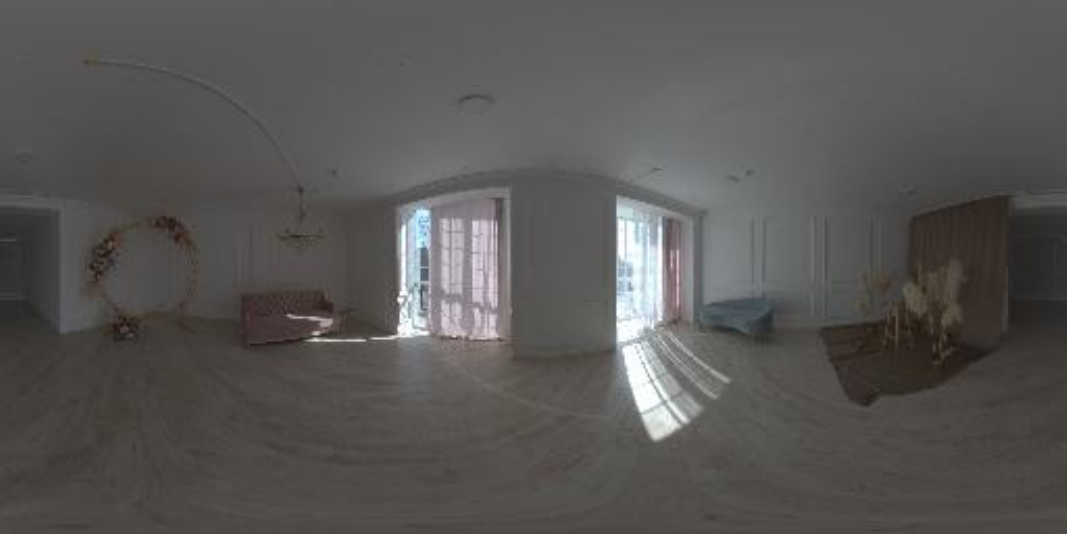}} & 
        \noindent\parbox[c]{0.14\textwidth}{\includegraphics[height=0.100\textwidth]{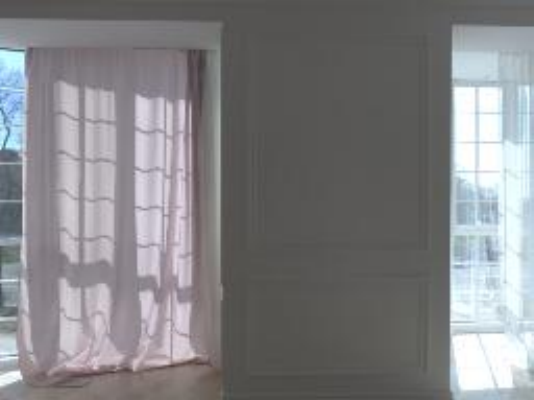}} &  
        
        \noindent\parbox[c]{0.100\textwidth}{\includegraphics[height=0.100\textwidth]{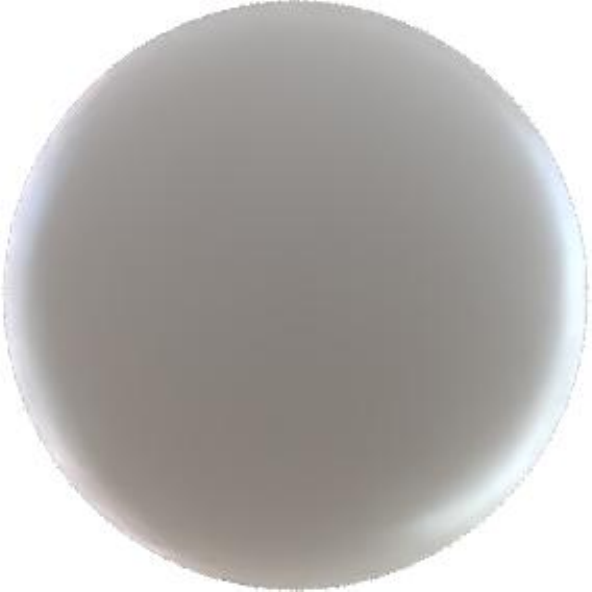}} & 
        \noindent\parbox[c]{0.100\textwidth}{\includegraphics[height=0.100\textwidth]{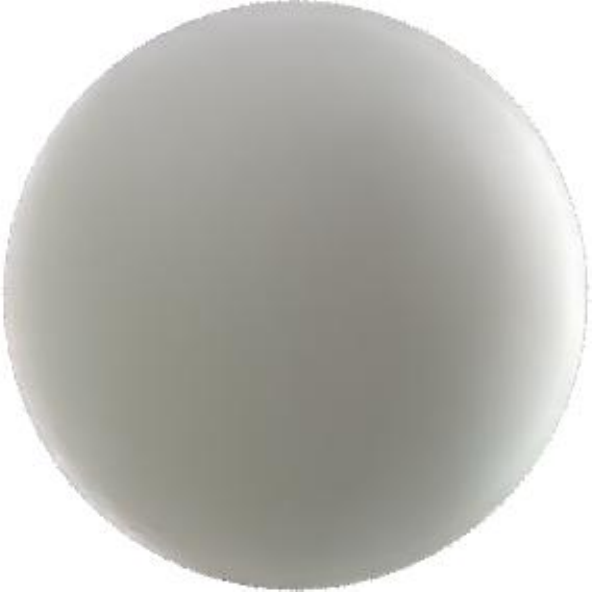}} & 
        
        \noindent\parbox[c]{0.100\textwidth}{\includegraphics[height=0.100\textwidth]{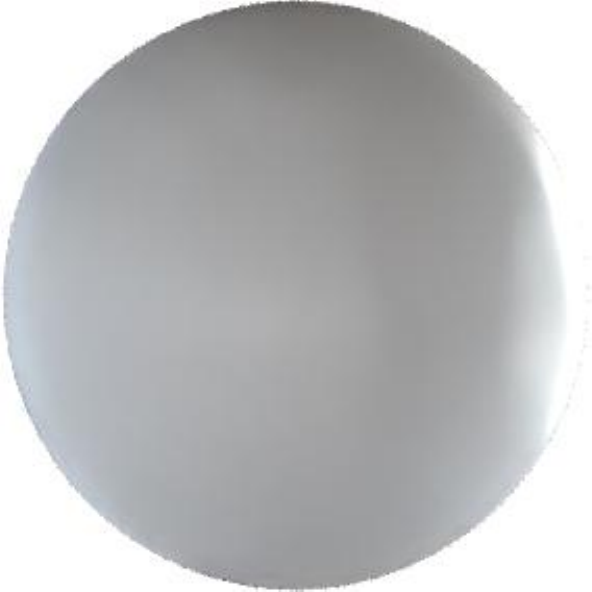}} & 
        \noindent\parbox[c]{0.100\textwidth}{\includegraphics[height=0.100\textwidth]{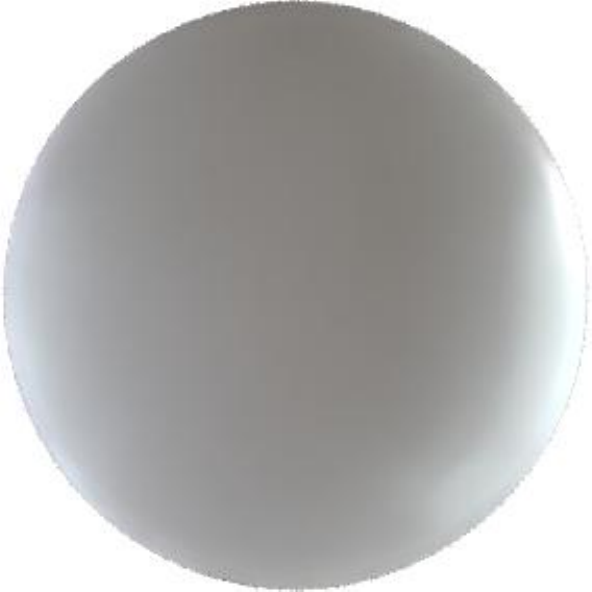}} &
        \noindent\parbox[c]{0.100\textwidth}{\includegraphics[height=0.100\textwidth]{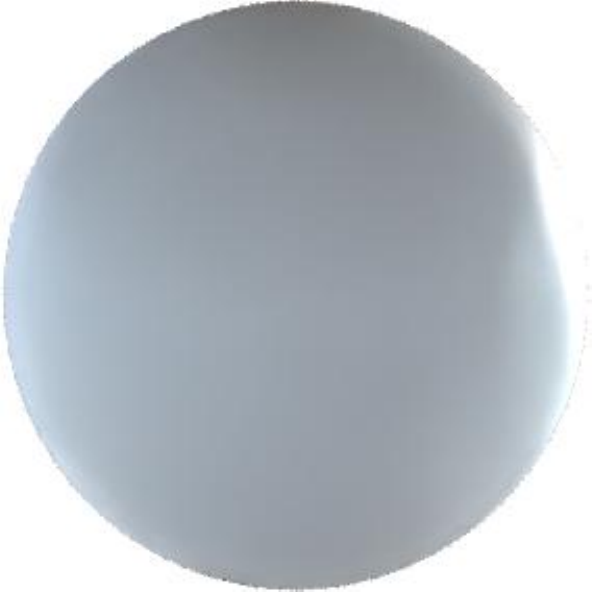}} & 
        \noindent\parbox[c]{0.100\textwidth}{\includegraphics[height=0.100\textwidth]{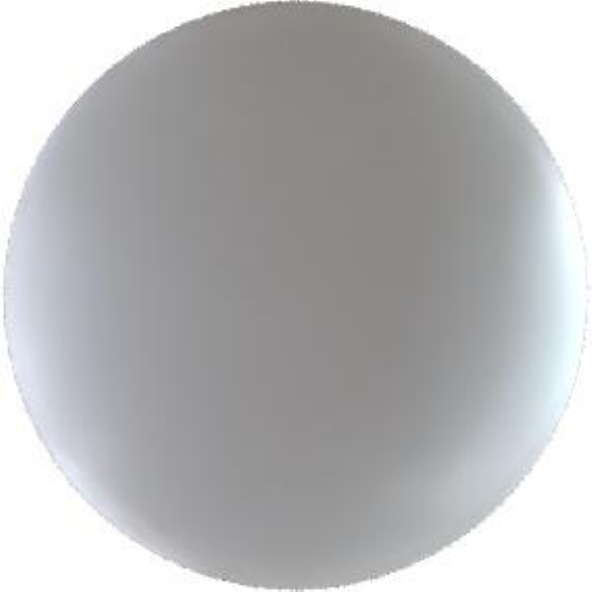}} & 
        \\

        \noindent\parbox[c]{0.205\textwidth}{\includegraphics[height=0.100\textwidth]{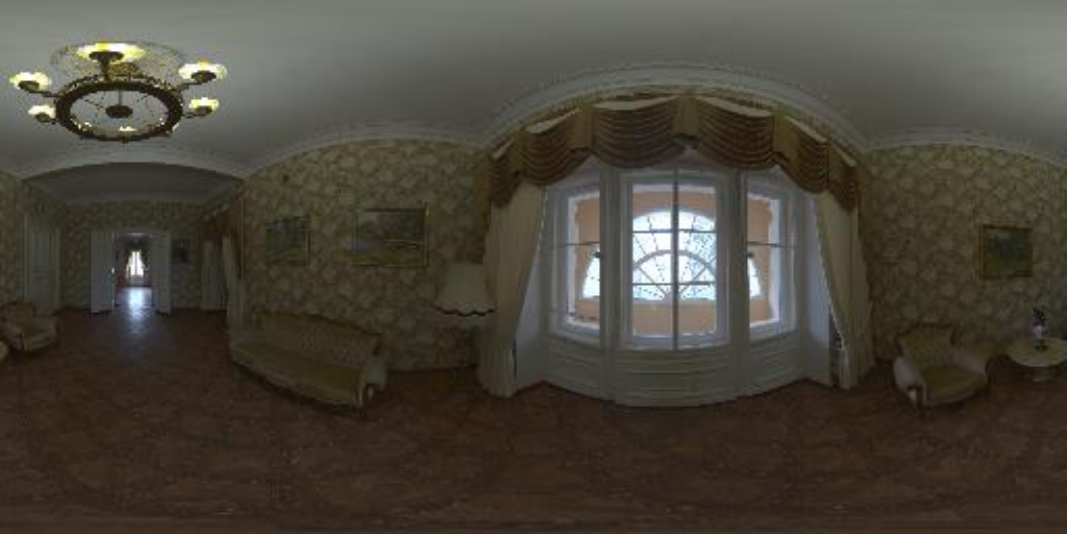}} & 
        \noindent\parbox[c]{0.14\textwidth}{\includegraphics[height=0.100\textwidth]{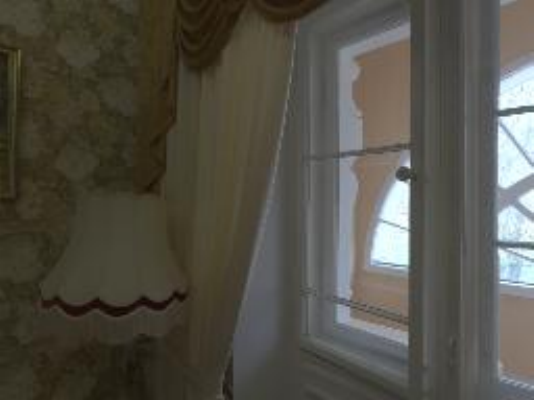}} &  
        
        \noindent\parbox[c]{0.100\textwidth}{\includegraphics[height=0.100\textwidth]{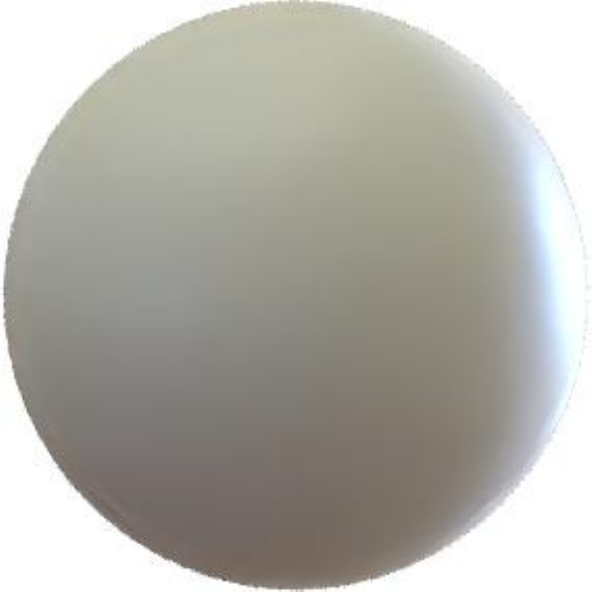}} & 
        \noindent\parbox[c]{0.100\textwidth}{\includegraphics[height=0.100\textwidth]{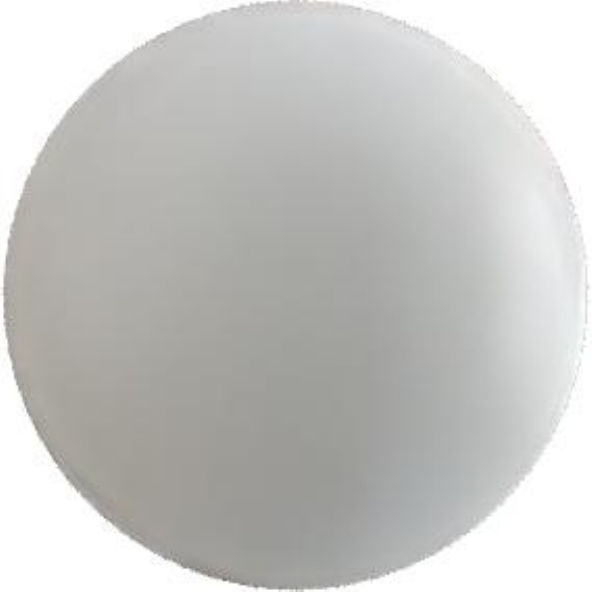}} & 
        
        \noindent\parbox[c]{0.100\textwidth}{\includegraphics[height=0.100\textwidth]{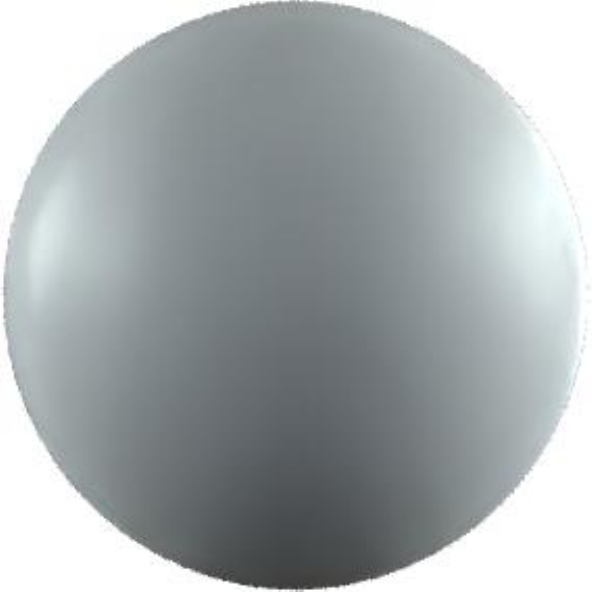}} & 
        \noindent\parbox[c]{0.100\textwidth}{\includegraphics[height=0.100\textwidth]{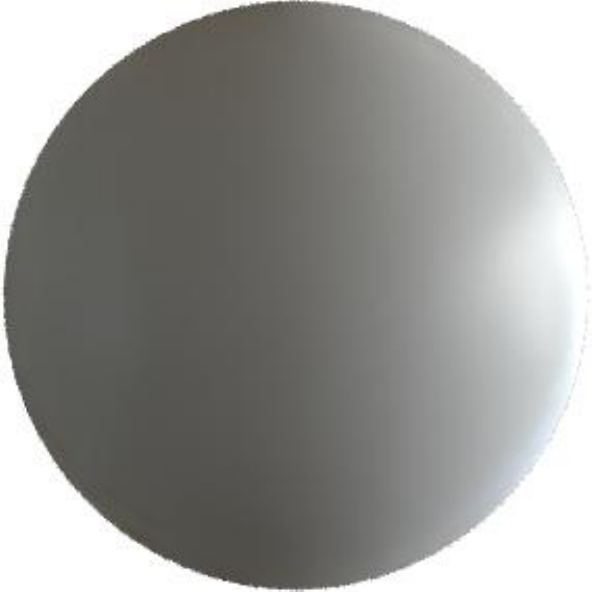}} &
        \noindent\parbox[c]{0.100\textwidth}{\includegraphics[height=0.100\textwidth]{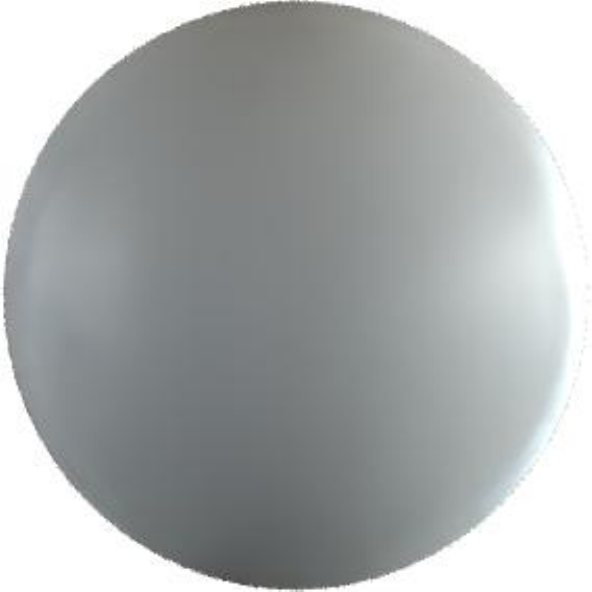}} & 
        \noindent\parbox[c]{0.100\textwidth}{\includegraphics[height=0.100\textwidth]{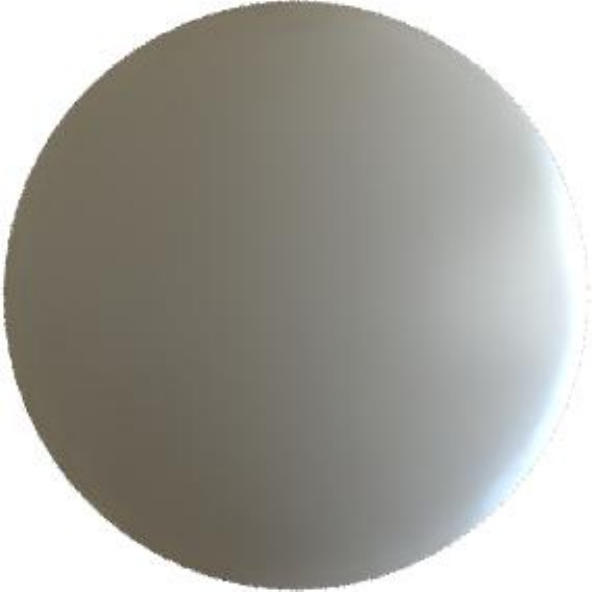}} & 
        \\

        \noindent\parbox[c]{0.205\textwidth}{\includegraphics[height=0.100\textwidth]{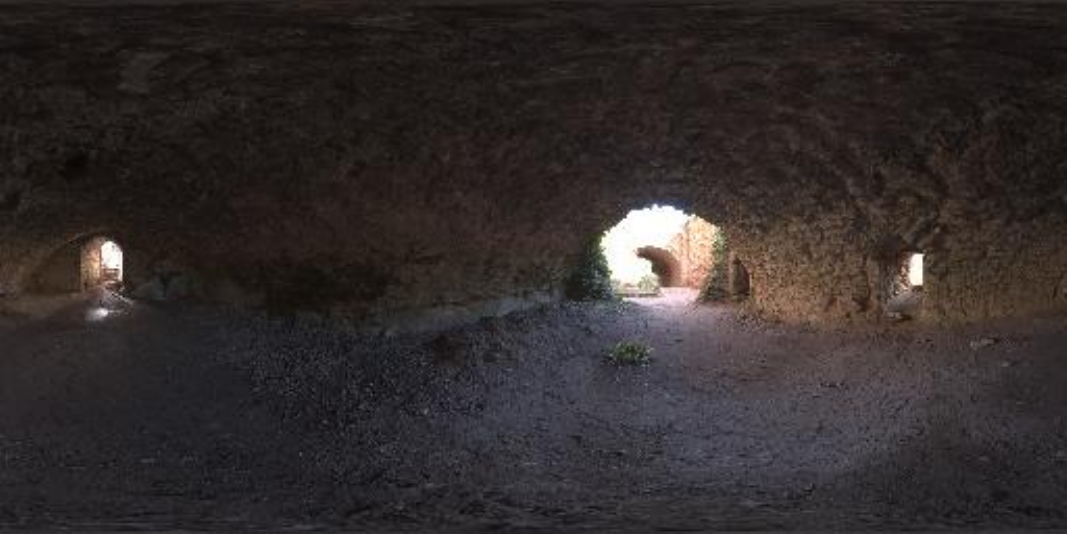}} & 
        \noindent\parbox[c]{0.14\textwidth}{\includegraphics[height=0.100\textwidth]{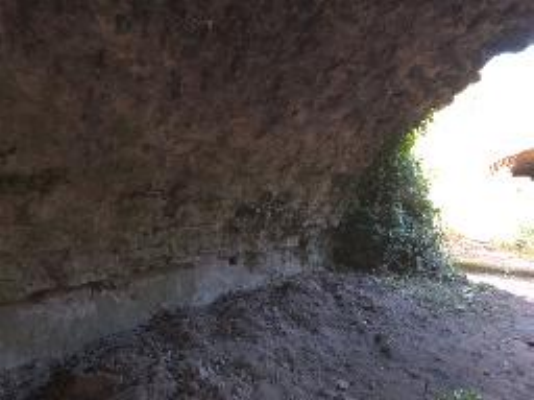}} &  
        
        \noindent\parbox[c]{0.100\textwidth}{\includegraphics[height=0.100\textwidth]{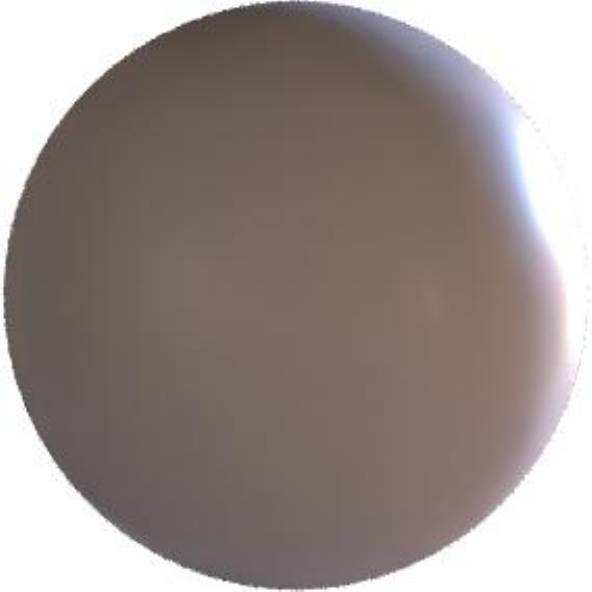}} & 
        \noindent\parbox[c]{0.100\textwidth}{\includegraphics[height=0.100\textwidth]{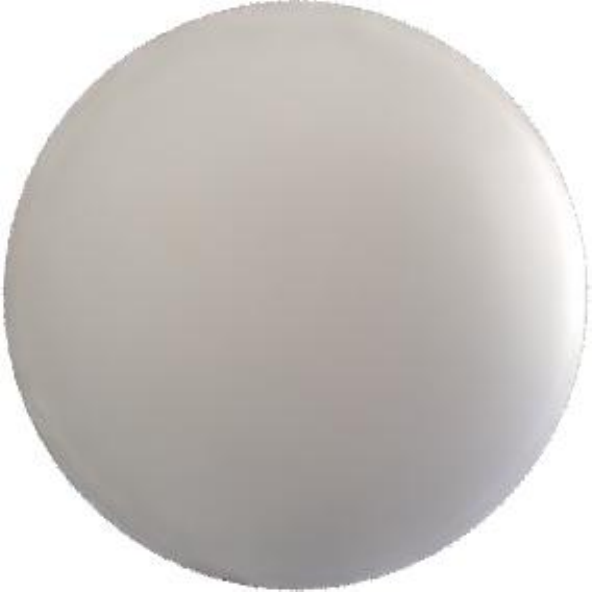}} & 
        
        \noindent\parbox[c]{0.100\textwidth}{\includegraphics[height=0.100\textwidth]{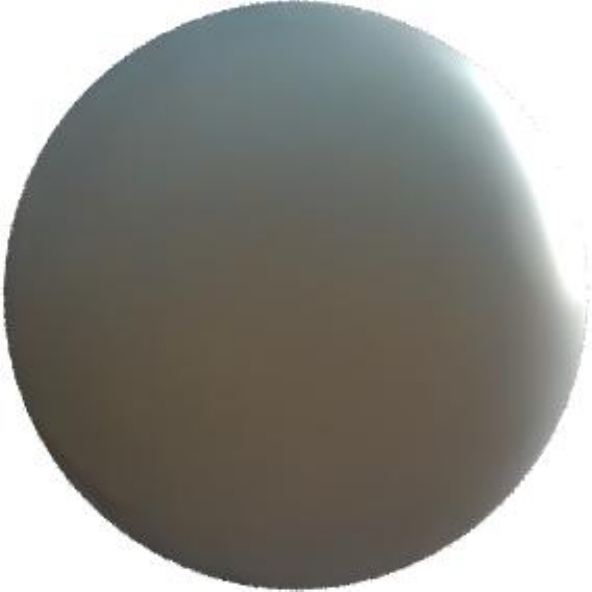}} & 
        \noindent\parbox[c]{0.100\textwidth}{\includegraphics[height=0.100\textwidth]{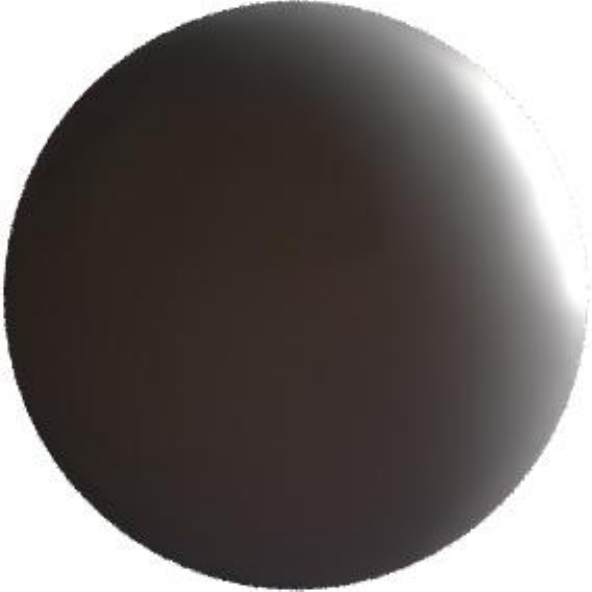}} &
        \noindent\parbox[c]{0.100\textwidth}{\includegraphics[height=0.100\textwidth]{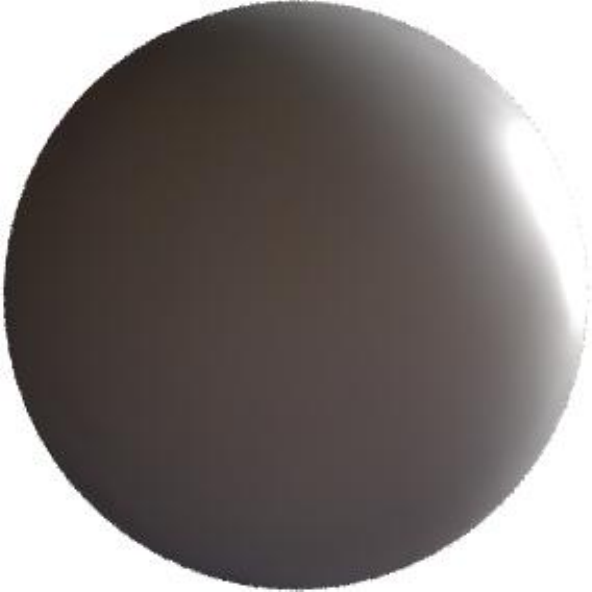}} & 
        \noindent\parbox[c]{0.100\textwidth}{\includegraphics[height=0.100\textwidth]{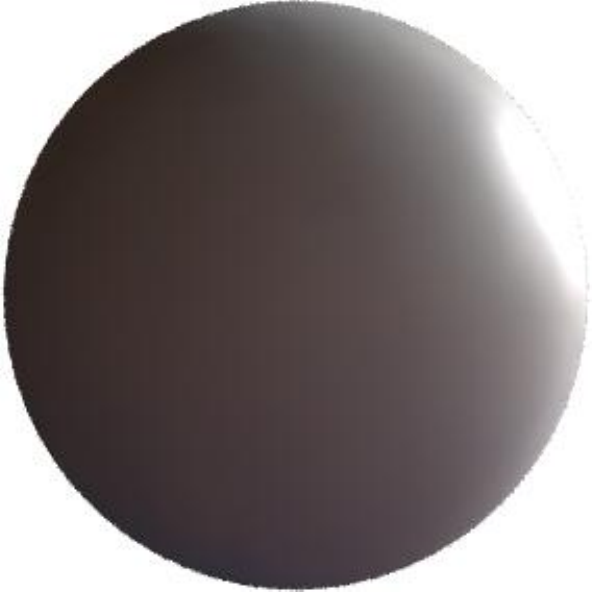}} & 
        \\

        \noindent\parbox[c]{0.205\textwidth}{\includegraphics[height=0.100\textwidth]{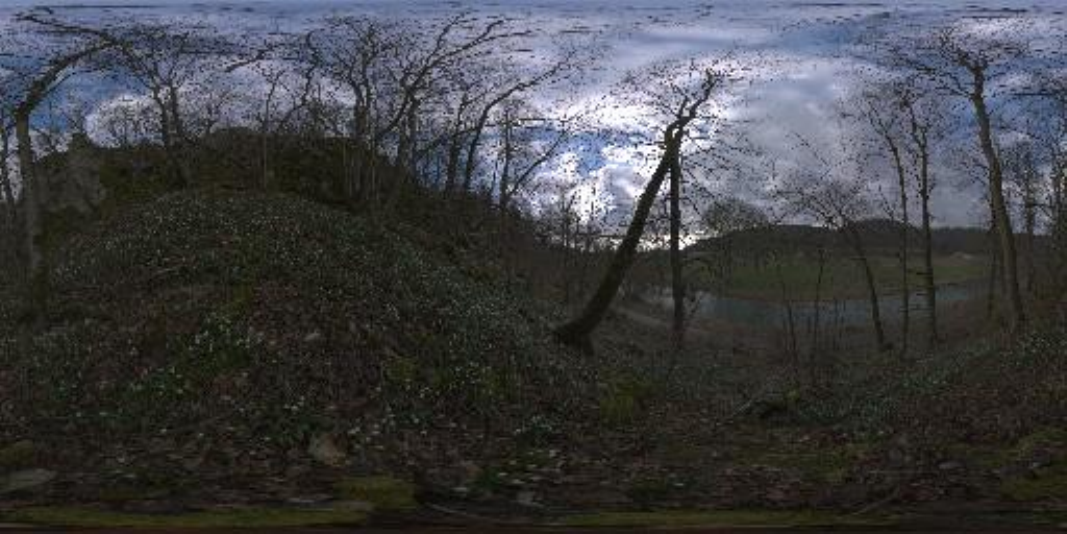}} & 
        \noindent\parbox[c]{0.14\textwidth}{\includegraphics[height=0.100\textwidth]{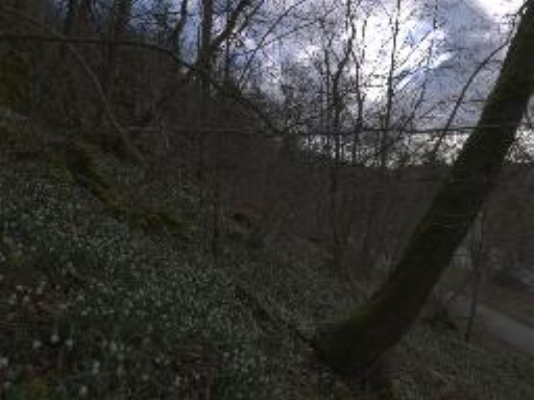}} &  
        
        \noindent\parbox[c]{0.100\textwidth}{\includegraphics[height=0.100\textwidth]{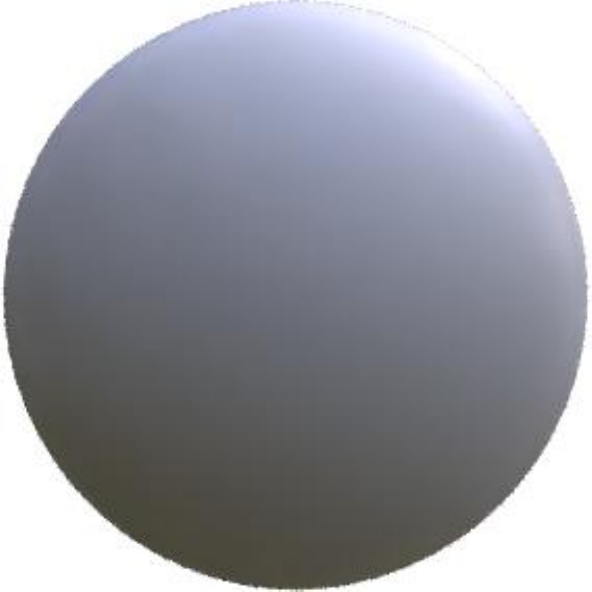}} & 
        \noindent\parbox[c]{0.100\textwidth}{\includegraphics[height=0.100\textwidth]{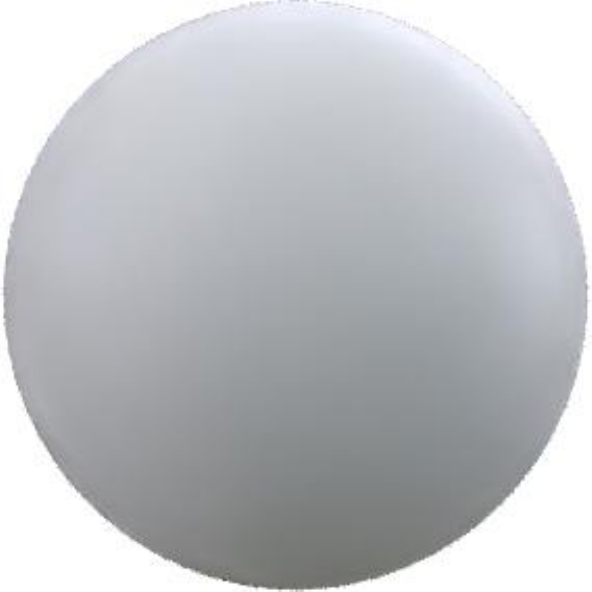}} & 
        
        \noindent\parbox[c]{0.100\textwidth}{\includegraphics[height=0.100\textwidth]{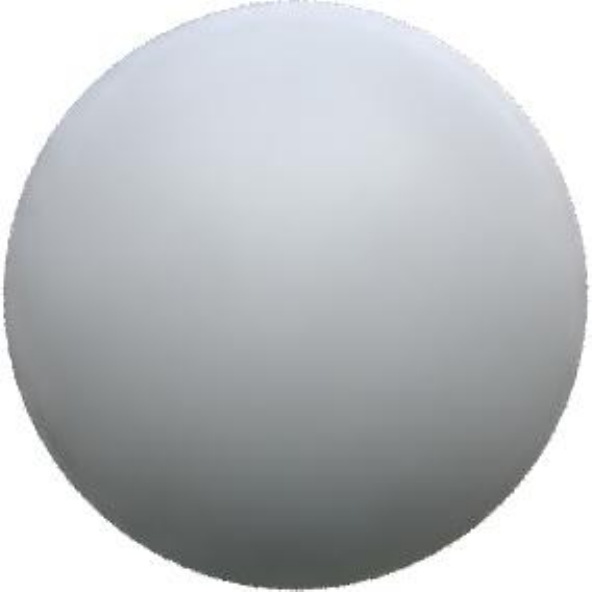}} & 
        \noindent\parbox[c]{0.100\textwidth}{\includegraphics[height=0.100\textwidth]{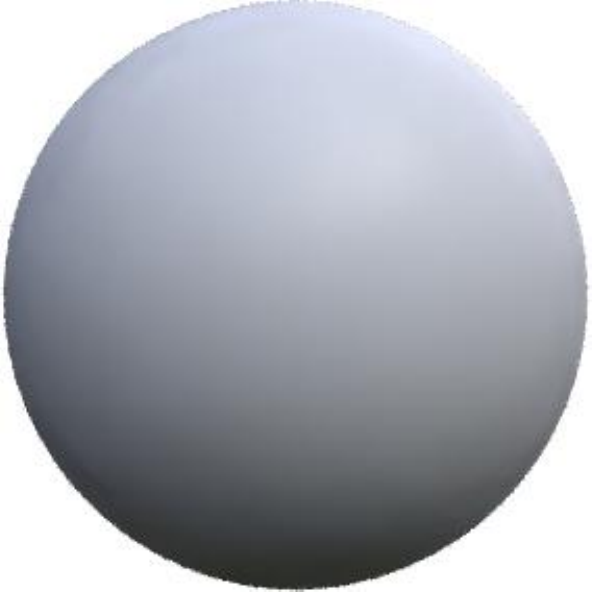}} &
        \noindent\parbox[c]{0.100\textwidth}{\includegraphics[height=0.100\textwidth]{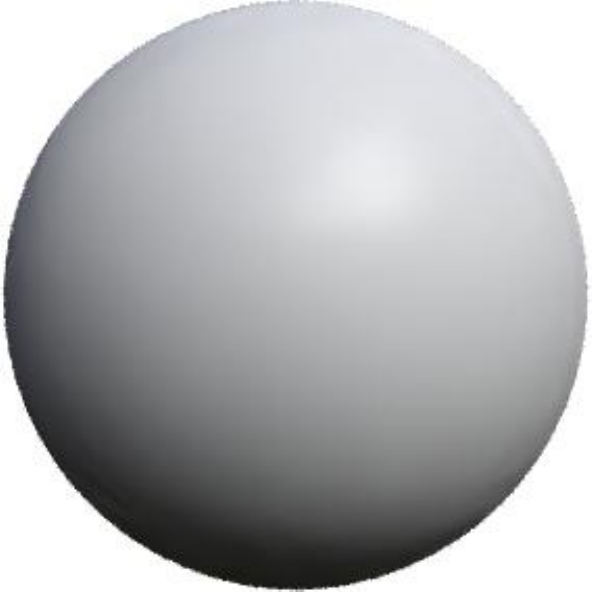}} & 
        \noindent\parbox[c]{0.100\textwidth}{\includegraphics[height=0.100\textwidth]{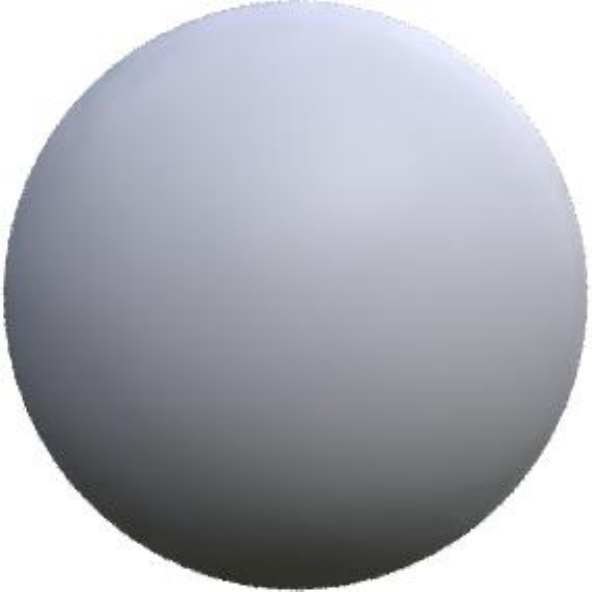}} & 
        \\

        \noindent\parbox[c]{0.205\textwidth}{\includegraphics[height=0.100\textwidth]{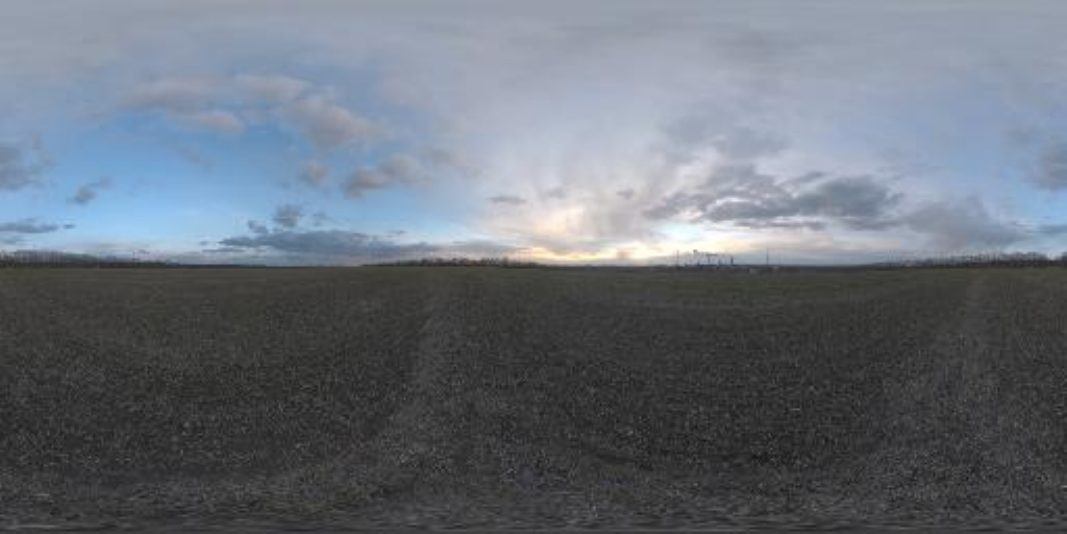}} & 
        \noindent\parbox[c]{0.14\textwidth}{\includegraphics[height=0.100\textwidth]{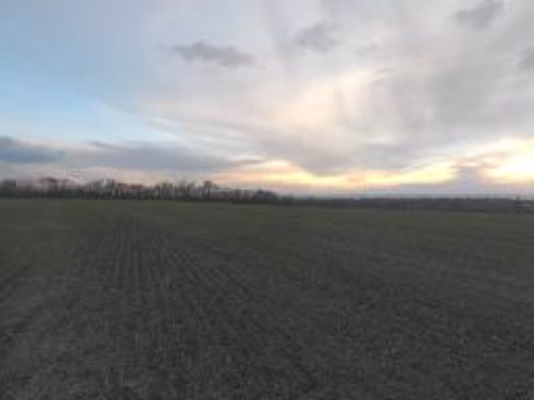}} &  
        
        \noindent\parbox[c]{0.100\textwidth}{\includegraphics[height=0.100\textwidth]{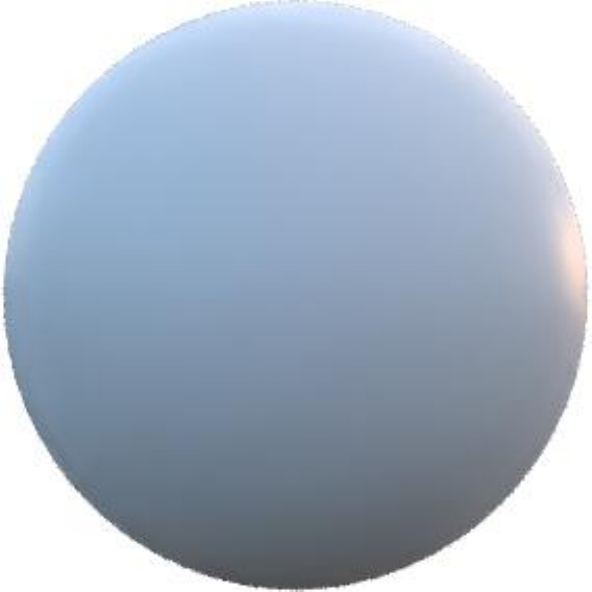}} & 
        \noindent\parbox[c]{0.100\textwidth}{\includegraphics[height=0.100\textwidth]{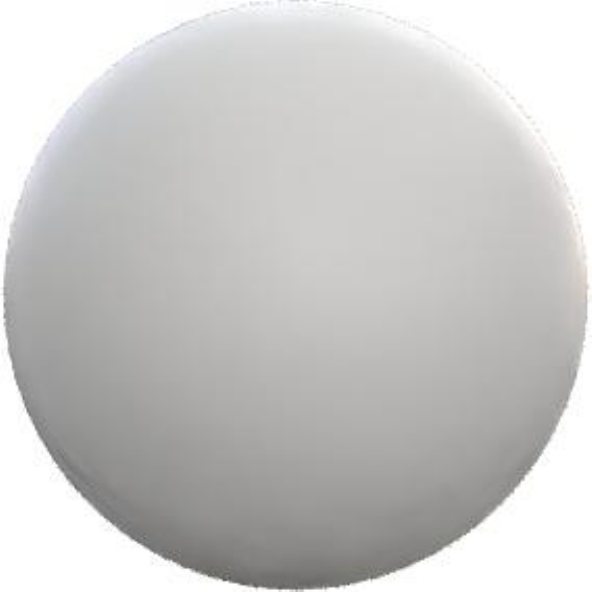}} & 
        
        \noindent\parbox[c]{0.100\textwidth}{\includegraphics[height=0.100\textwidth]{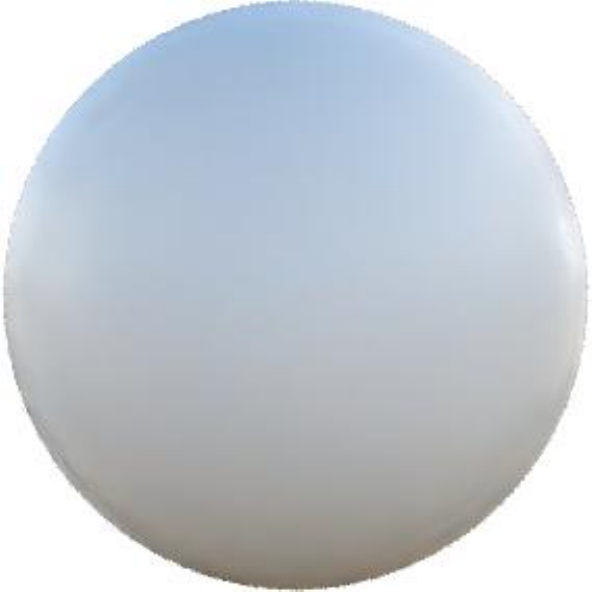}} & 
        \noindent\parbox[c]{0.100\textwidth}{\includegraphics[height=0.100\textwidth]{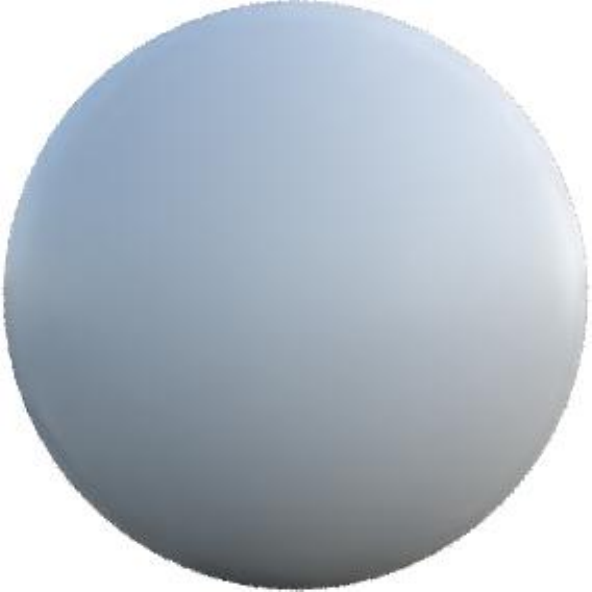}} &
        \noindent\parbox[c]{0.100\textwidth}{\includegraphics[height=0.100\textwidth]{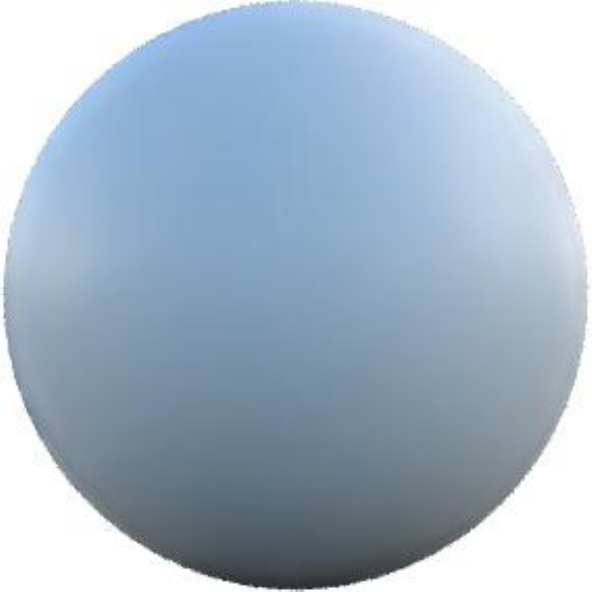}} & 
        \noindent\parbox[c]{0.100\textwidth}{\includegraphics[height=0.100\textwidth]{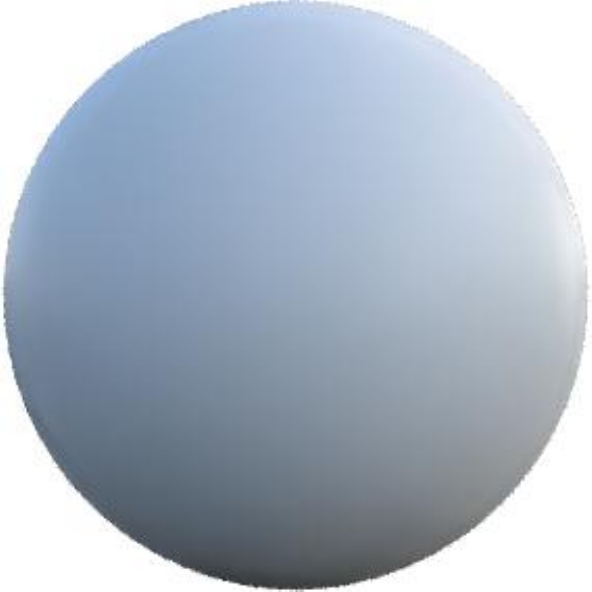}} & 
        \\

        \noindent\parbox[c]{0.205\textwidth}{\includegraphics[height=0.100\textwidth]{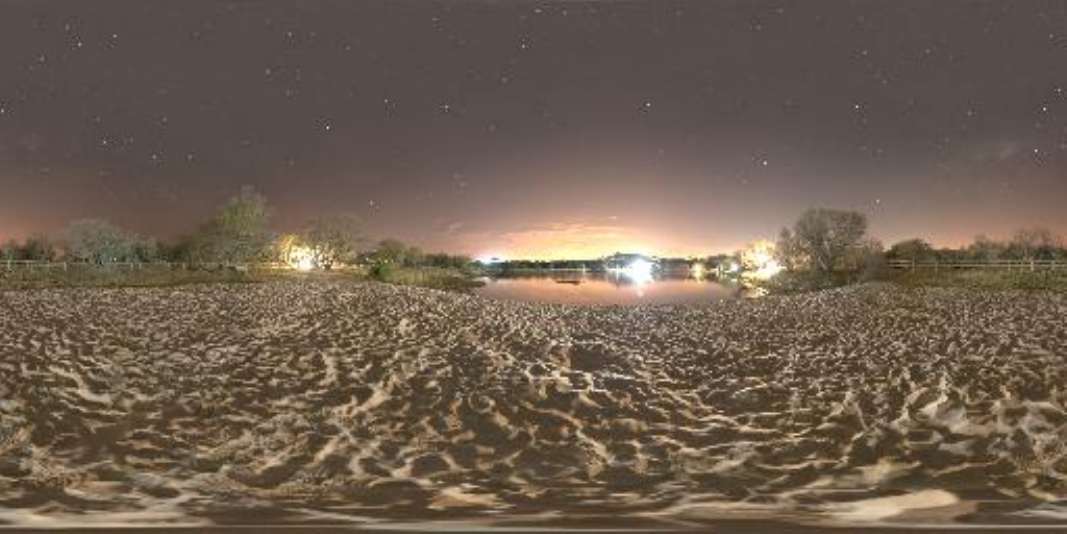}} & 
        \noindent\parbox[c]{0.14\textwidth}{\includegraphics[height=0.100\textwidth]{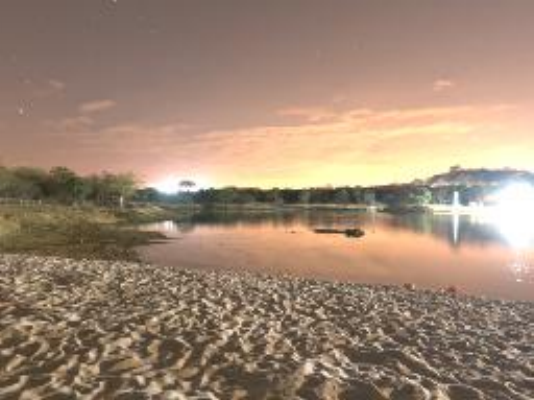}} &  
        
        \noindent\parbox[c]{0.100\textwidth}{\includegraphics[height=0.100\textwidth]{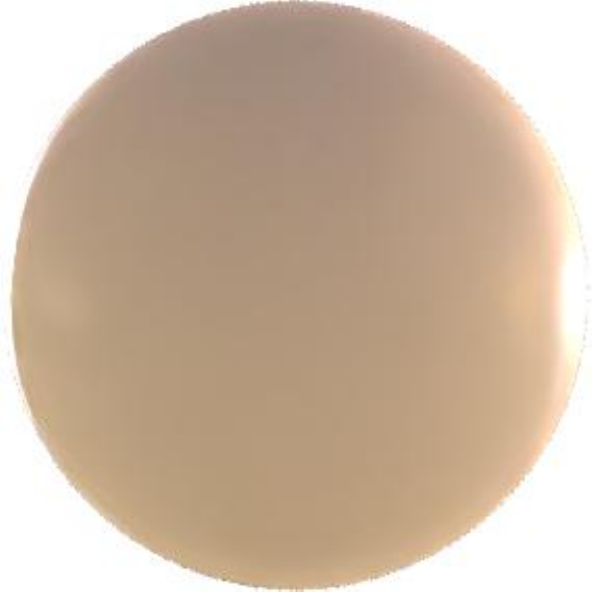}} & 
        \noindent\parbox[c]{0.100\textwidth}{\includegraphics[height=0.100\textwidth]{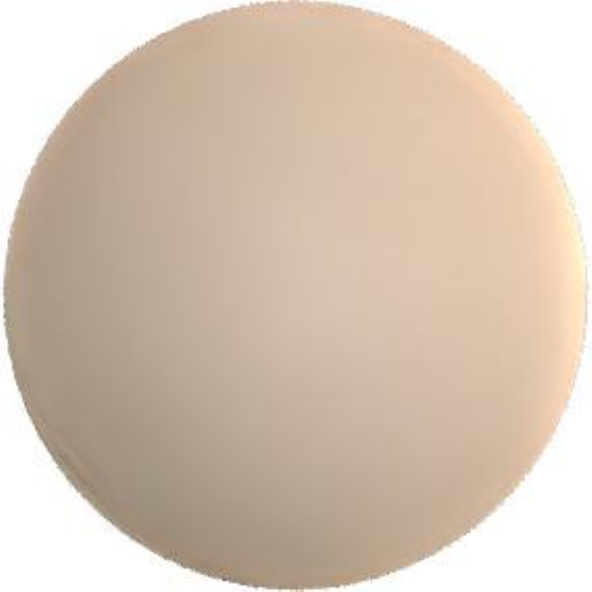}} & 
        
        \noindent\parbox[c]{0.100\textwidth}{\includegraphics[height=0.100\textwidth]{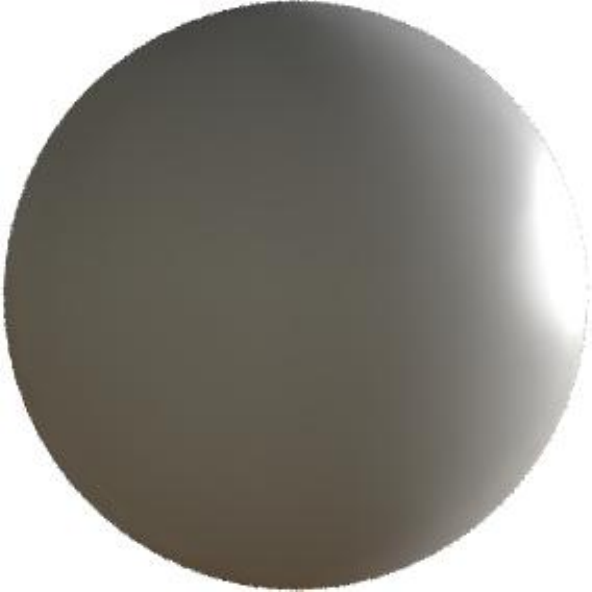}} & 
        \noindent\parbox[c]{0.100\textwidth}{\includegraphics[height=0.100\textwidth]{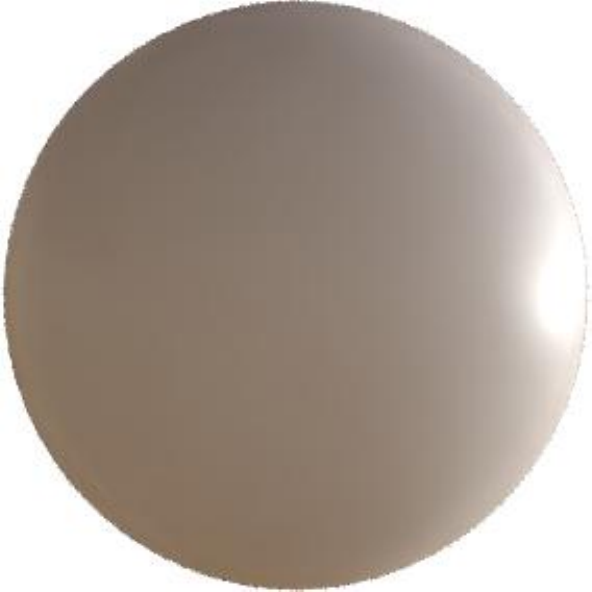}} &
        \noindent\parbox[c]{0.100\textwidth}{\includegraphics[height=0.100\textwidth]{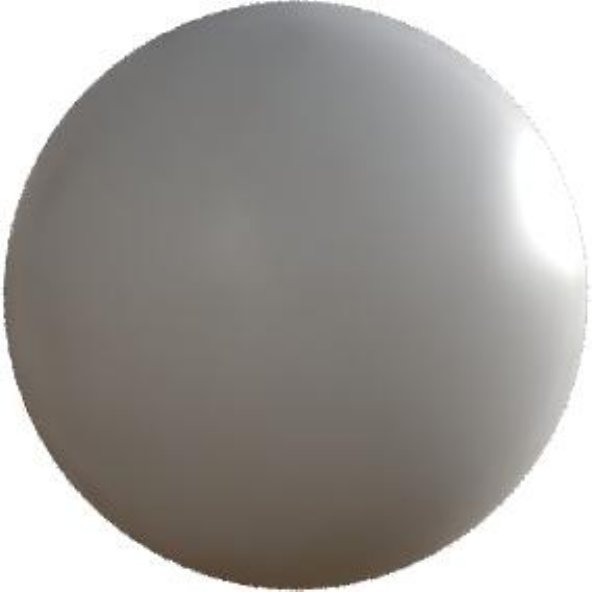}} & 
        \noindent\parbox[c]{0.100\textwidth}{\includegraphics[height=0.100\textwidth]{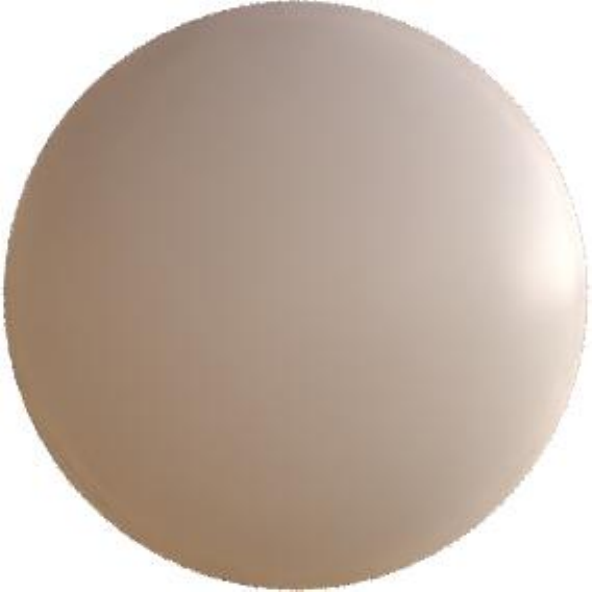}} & 
        \\

        \noindent\parbox[c]{0.205\textwidth}{\includegraphics[height=0.100\textwidth]{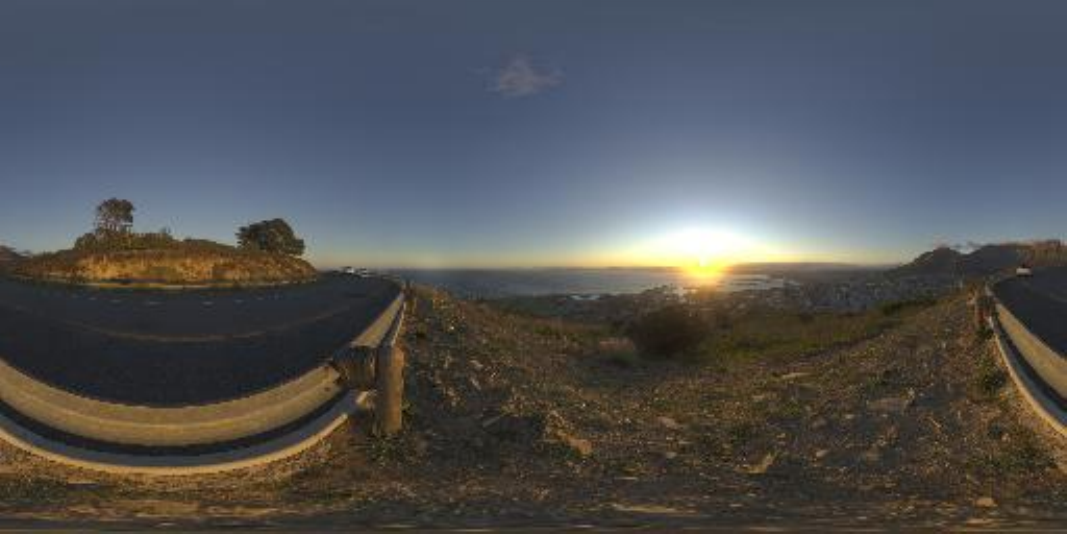}} & 
        \noindent\parbox[c]{0.14\textwidth}{\includegraphics[height=0.100\textwidth]{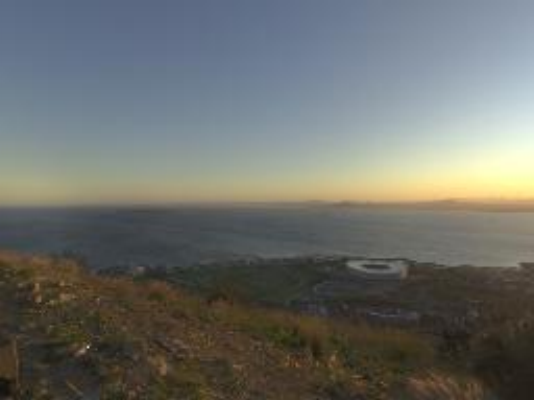}} &  
        
        \noindent\parbox[c]{0.100\textwidth}{\includegraphics[height=0.100\textwidth]{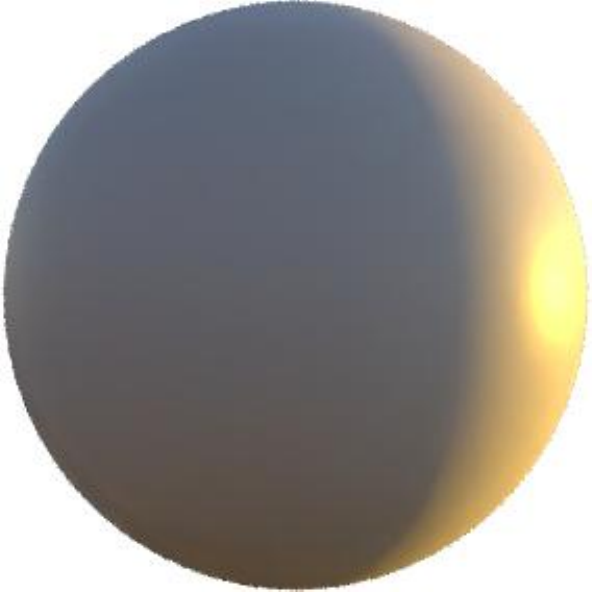}} & 
        \noindent\parbox[c]{0.100\textwidth}{\includegraphics[height=0.100\textwidth]{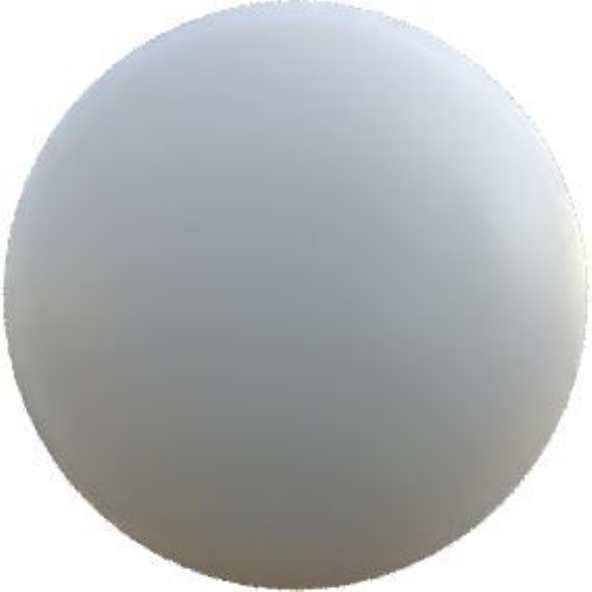}} & 
        
        \noindent\parbox[c]{0.100\textwidth}{\includegraphics[height=0.100\textwidth]{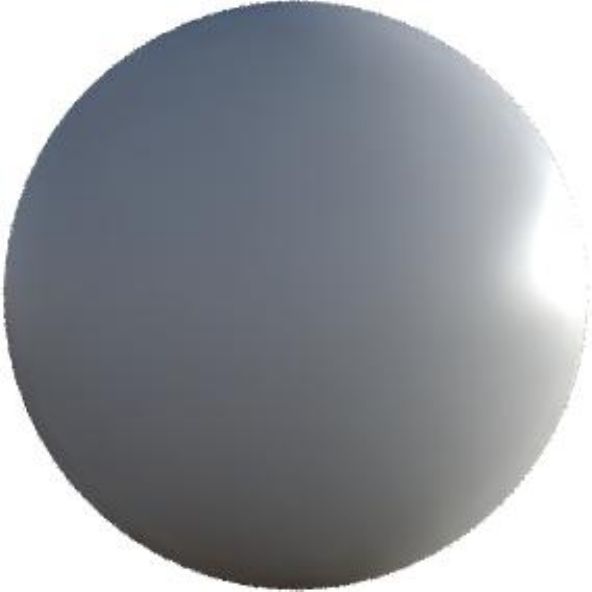}} & 
        \noindent\parbox[c]{0.100\textwidth}{\includegraphics[height=0.100\textwidth]{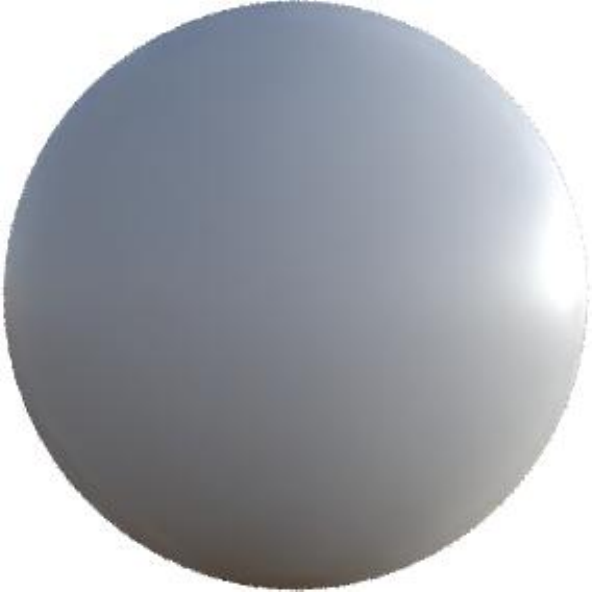}} &
        \noindent\parbox[c]{0.100\textwidth}{\includegraphics[height=0.100\textwidth]{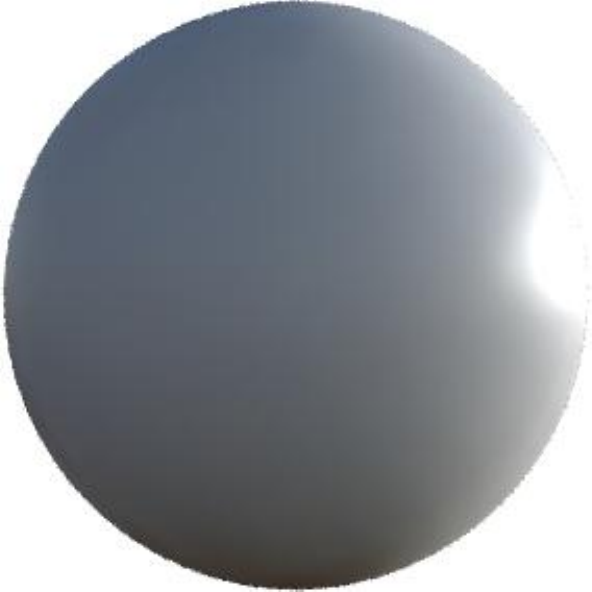}} & 
        \noindent\parbox[c]{0.100\textwidth}{\includegraphics[height=0.100\textwidth]{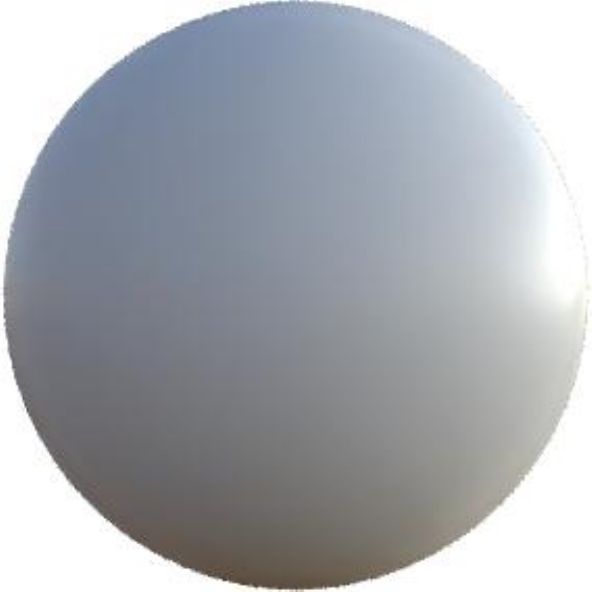}} & 
        \\

        \noindent\parbox[c]{0.205\textwidth}{\includegraphics[height=0.100\textwidth]{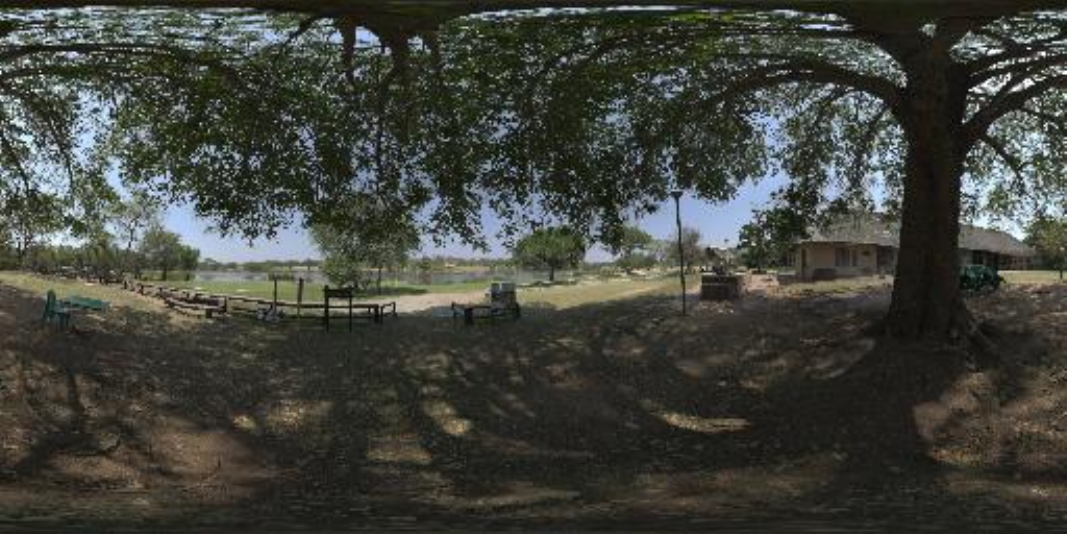}} & 
        \noindent\parbox[c]{0.14\textwidth}{\includegraphics[height=0.100\textwidth]{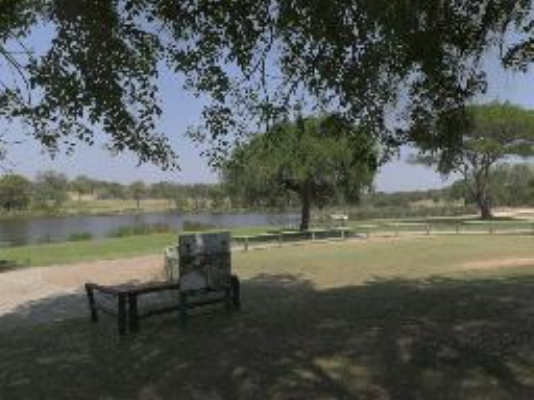}} &  
        
        \noindent\parbox[c]{0.100\textwidth}{\includegraphics[height=0.100\textwidth]{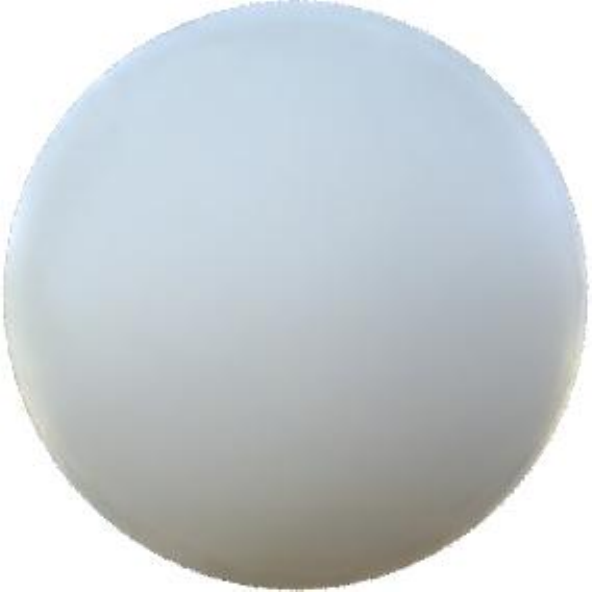}} & 
        \noindent\parbox[c]{0.100\textwidth}{\includegraphics[height=0.100\textwidth]{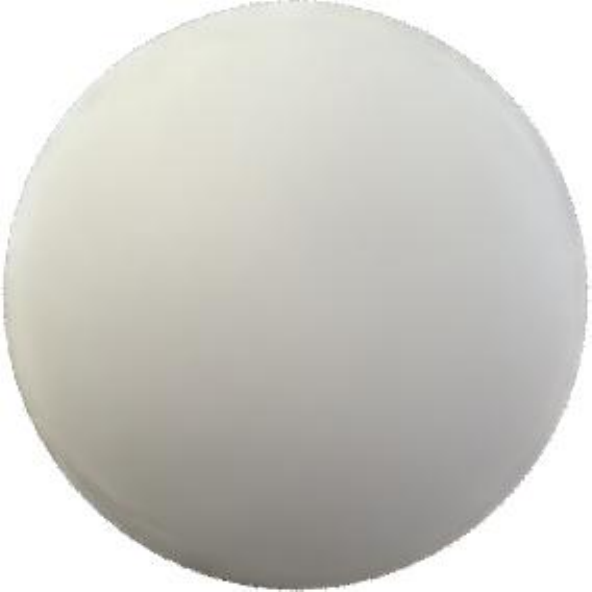}} & 
        
        \noindent\parbox[c]{0.100\textwidth}{\includegraphics[height=0.100\textwidth]{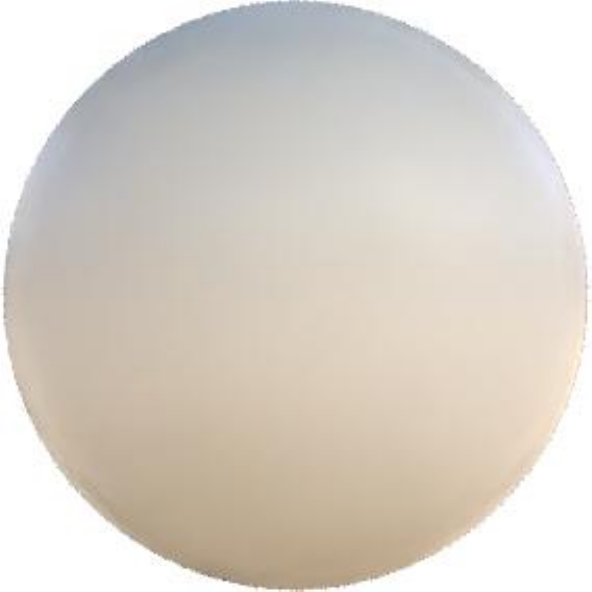}} & 
        \noindent\parbox[c]{0.100\textwidth}{\includegraphics[height=0.100\textwidth]{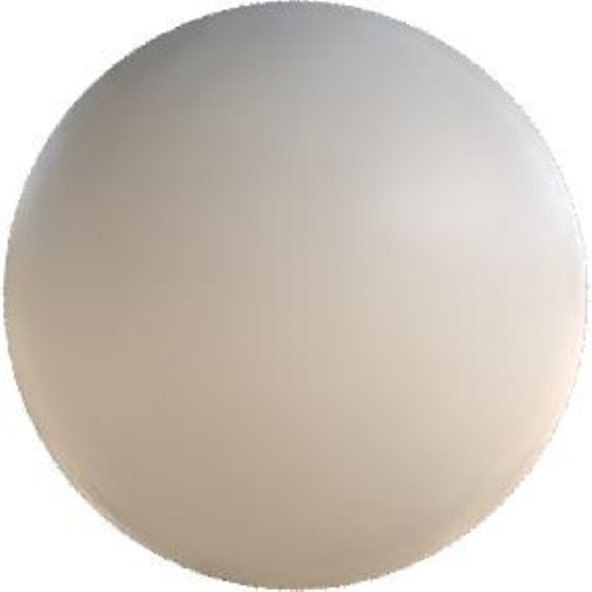}} &
        \noindent\parbox[c]{0.100\textwidth}{\includegraphics[height=0.100\textwidth]{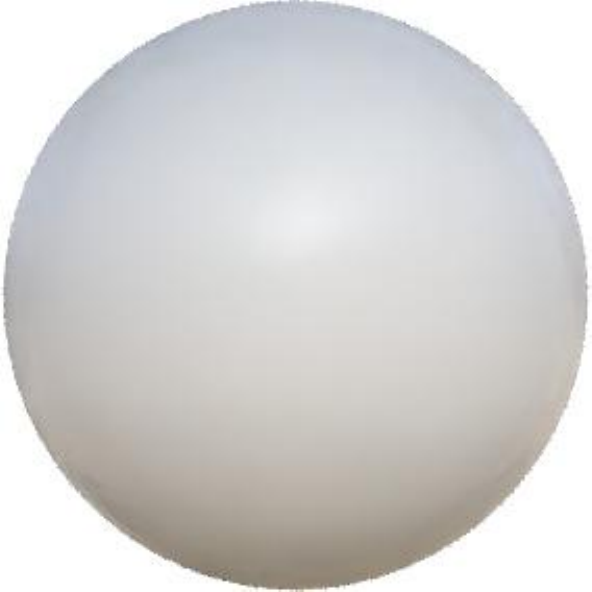}} & 
        \noindent\parbox[c]{0.100\textwidth}{\includegraphics[height=0.100\textwidth]{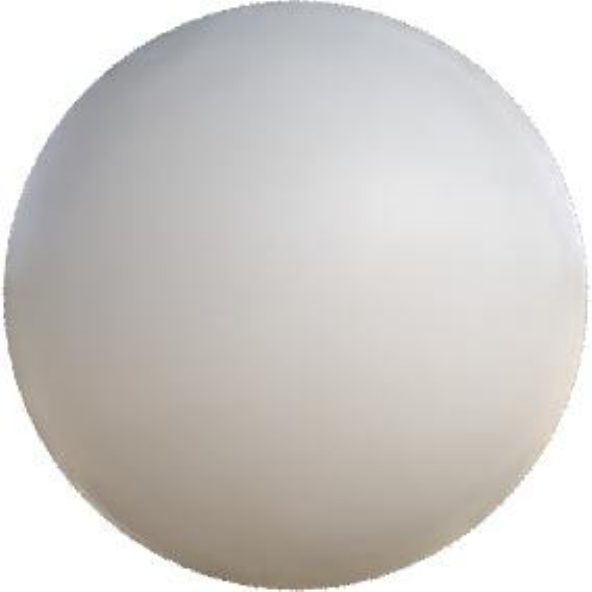}} & 
        \\

        \noindent\parbox[c]{0.205\textwidth}{\includegraphics[height=0.100\textwidth]{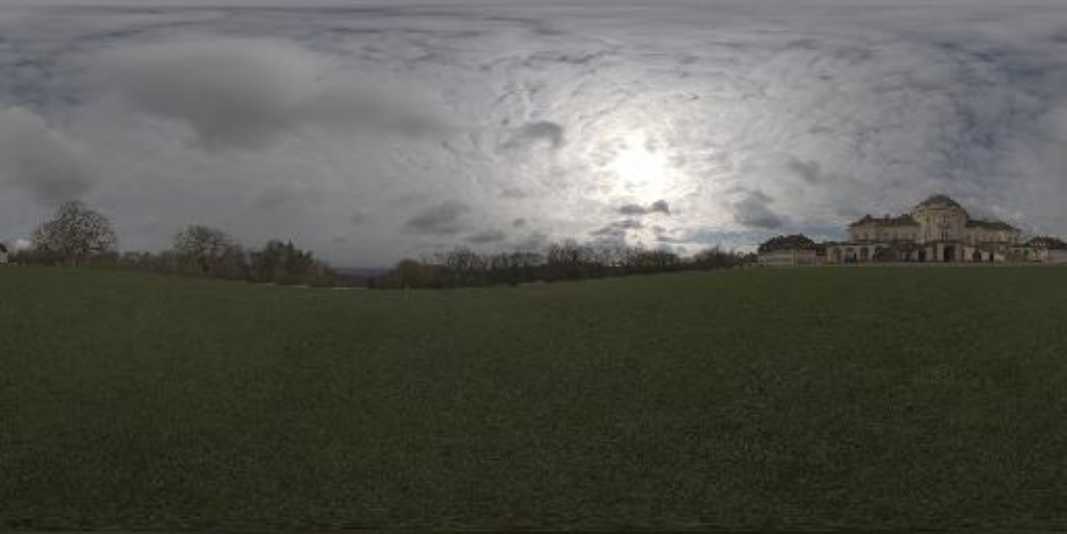}} & 
        \noindent\parbox[c]{0.14\textwidth}{\includegraphics[height=0.100\textwidth]{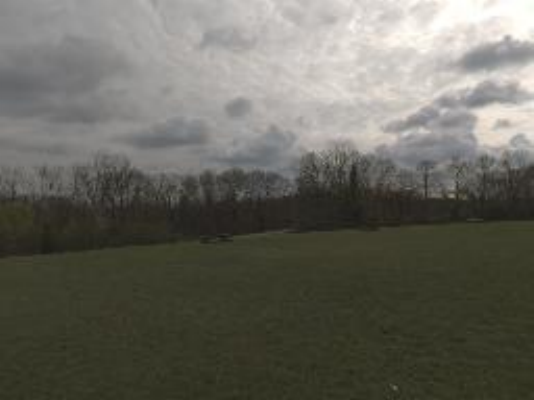}} &  
        
        \noindent\parbox[c]{0.100\textwidth}{\includegraphics[height=0.100\textwidth]{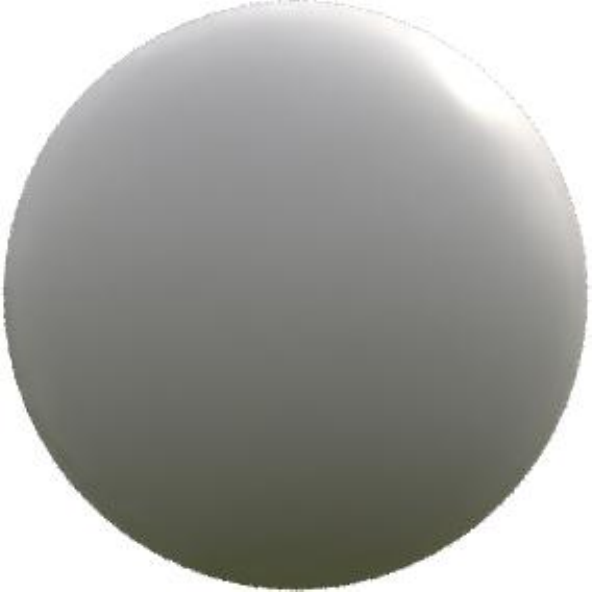}} & 
        \noindent\parbox[c]{0.100\textwidth}{\includegraphics[height=0.100\textwidth]{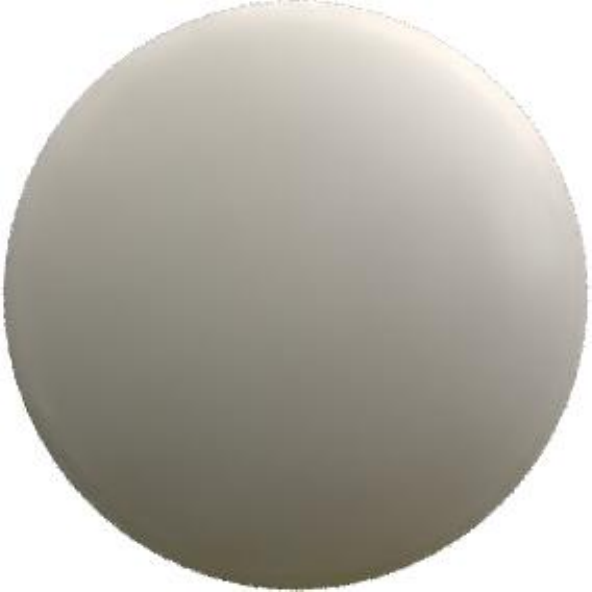}} & 
        
        \noindent\parbox[c]{0.100\textwidth}{\includegraphics[height=0.100\textwidth]{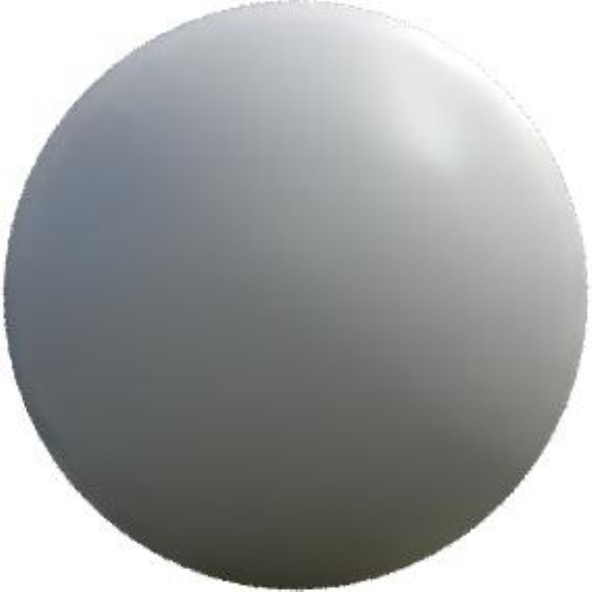}} & 
        \noindent\parbox[c]{0.100\textwidth}{\includegraphics[height=0.100\textwidth]{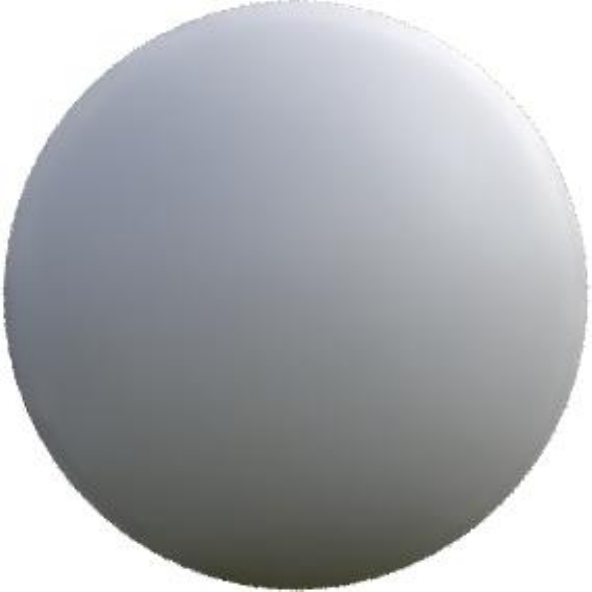}} &
        \noindent\parbox[c]{0.100\textwidth}{\includegraphics[height=0.100\textwidth]{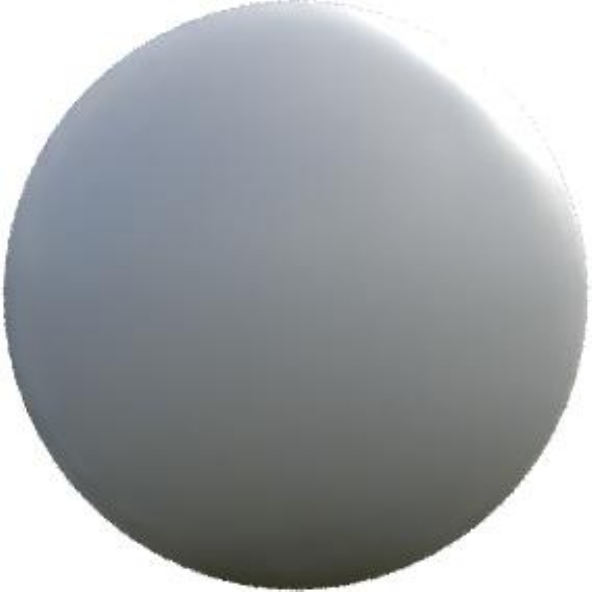}} & 
        \noindent\parbox[c]{0.100\textwidth}{\includegraphics[height=0.100\textwidth]{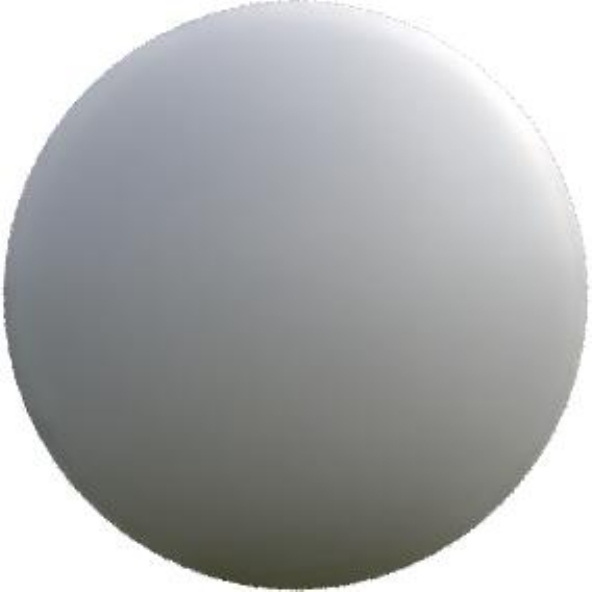}} & 
        \\

        \noindent\parbox[c]{0.205\textwidth}{\includegraphics[height=0.100\textwidth]{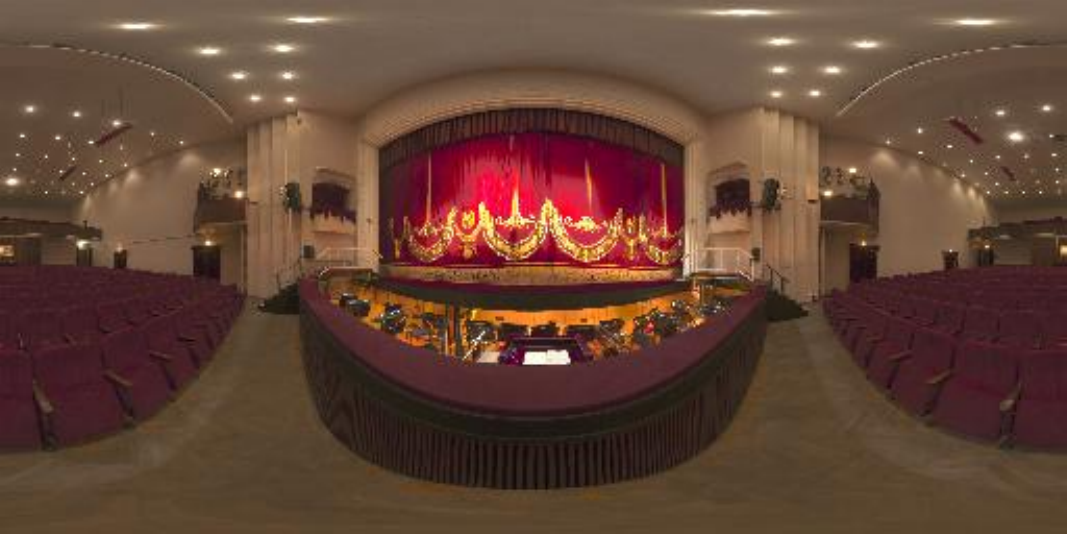}} & 
        \noindent\parbox[c]{0.14\textwidth}{\includegraphics[height=0.100\textwidth]{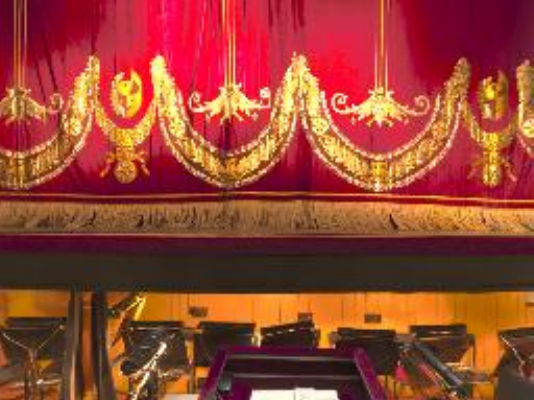}} &  
        
        \noindent\parbox[c]{0.100\textwidth}{\includegraphics[height=0.100\textwidth]{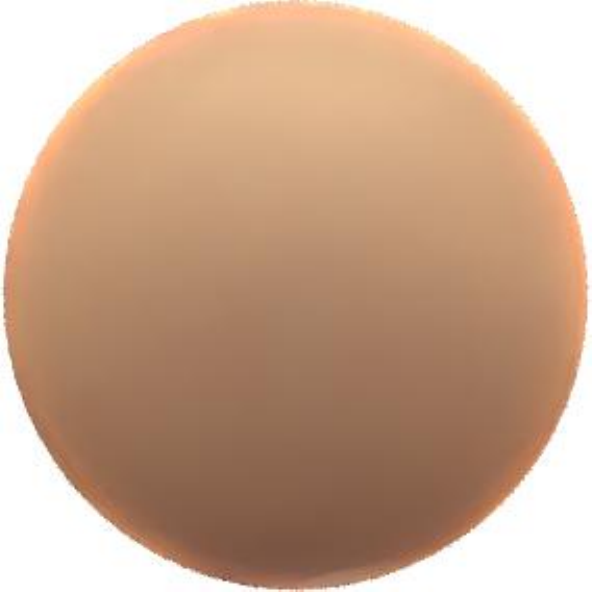}} & 
        \noindent\parbox[c]{0.100\textwidth}{\includegraphics[height=0.100\textwidth]{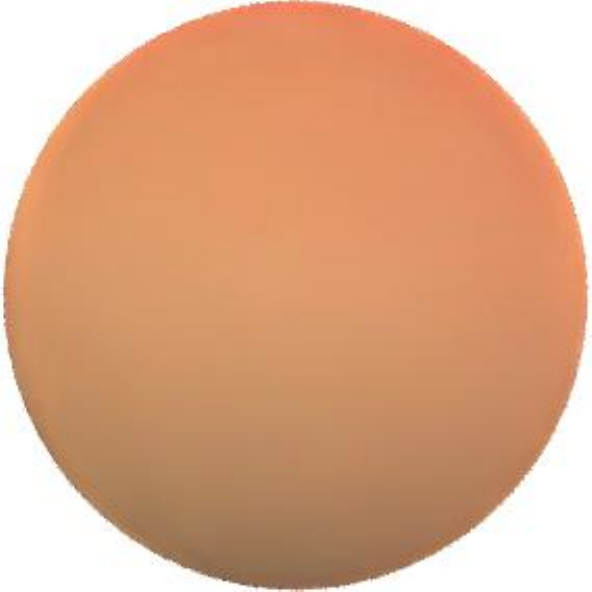}} & 
        
        \noindent\parbox[c]{0.100\textwidth}{\includegraphics[height=0.100\textwidth]{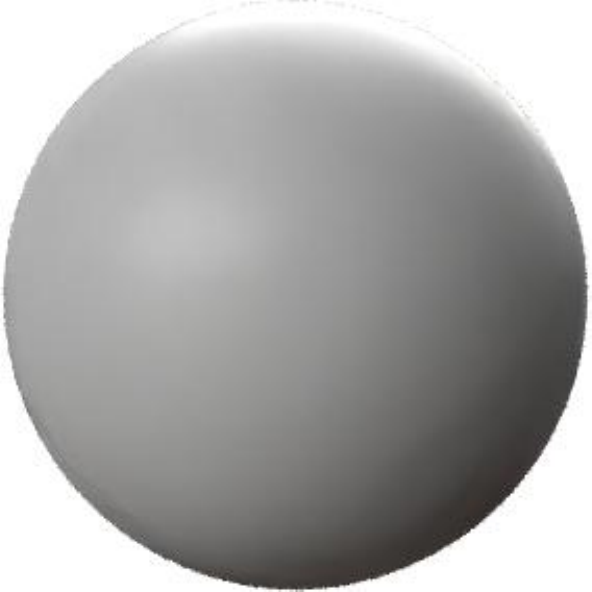}} & 
        \noindent\parbox[c]{0.100\textwidth}{\includegraphics[height=0.100\textwidth]{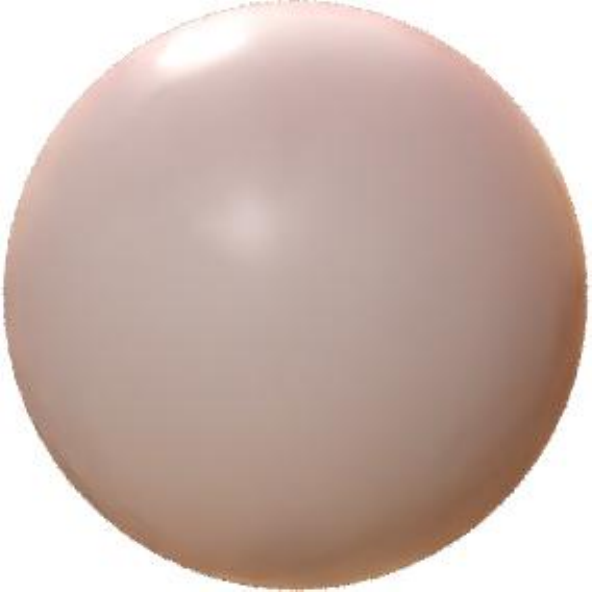}} &
        \noindent\parbox[c]{0.100\textwidth}{\includegraphics[height=0.100\textwidth]{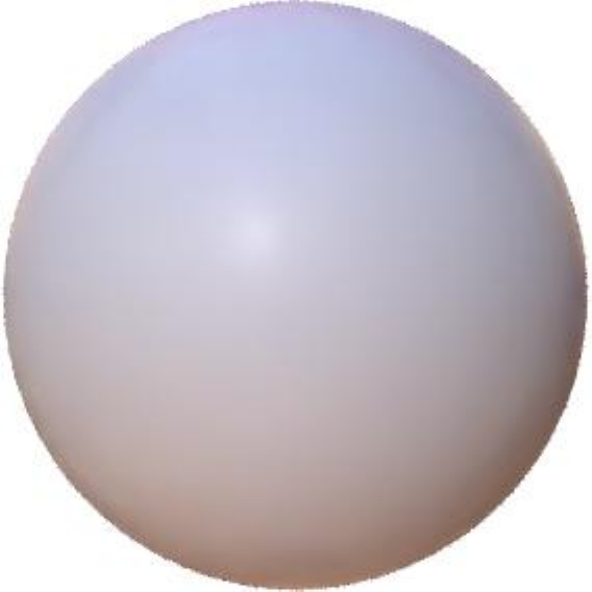}} & 
        \noindent\parbox[c]{0.100\textwidth}{\includegraphics[height=0.100\textwidth]{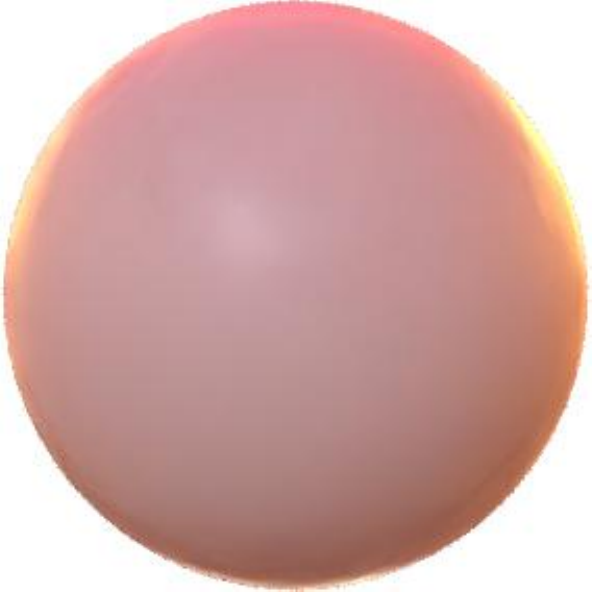}} & 
        \\
        
        \end{tabu}
    \caption{
    Qualitative results for the Poly Haven dataset using diffuse balls.}
    \label{fig:additional_polyhaven_diffuse}
\end{figure*}

Qualitative results for the Laval indoor dataset are in Figure \ref{fig:additional_indoor_mirror}-\ref{fig:additional_indoor_diffuse} and for Poly Haven in Figure \ref{fig:additional_polyhaven_mirror}-\ref{fig:additional_polyhaven_diffuse}. As discussed in the main paper, we start with SDXL$^\dagger$ to which we add our LoRA (LR) and iterative inpainting (I) to obtain our proposed algorithm, SDXL$^\dagger$+LR+I. The methods designated SDXL$^\dagger$+LR and SDXL$^\dagger$+I are ablated versions of our algorithm.

\begin{figure*}[!t]
    \centering

        \begin{tabu} to \textwidth {
        @{}
        c@{\hspace{1.5pt}}
        c@{\hspace{1.5pt}}
        c@{\hspace{1.5pt}}
        c@{\hspace{10pt}}
        c@{\hspace{1.5pt}}
        c@{\hspace{1.5pt}}
        c@{\hspace{1.5pt}}
        c@{\hspace{1.5pt}}
    }

        \multicolumn{1}{c}{\shortstack{\scriptsize Ground truth map}} & 
        \multicolumn{1}{c}{\shortstack{\scriptsize Input}} &
        \multicolumn{1}{c}{\shortstack{\scriptsize Ground Truth}} & 
        \multicolumn{1}{c}{\shortstack{\hspace{-14pt} \scriptsize Ours (scaled)}} & 
        \multicolumn{1}{c}{\shortstack{\scriptsize Ground truth map}} & 
        \multicolumn{1}{c}{\shortstack{\scriptsize Input}} &
        \multicolumn{1}{c}{\shortstack{\scriptsize Ground Truth}} &
        \multicolumn{1}{c}{\shortstack{\scriptsize Ours (scaled)}} 
        \\

        \noindent\parbox[c]{0.200\textwidth}{\includegraphics[width=0.200\textwidth]{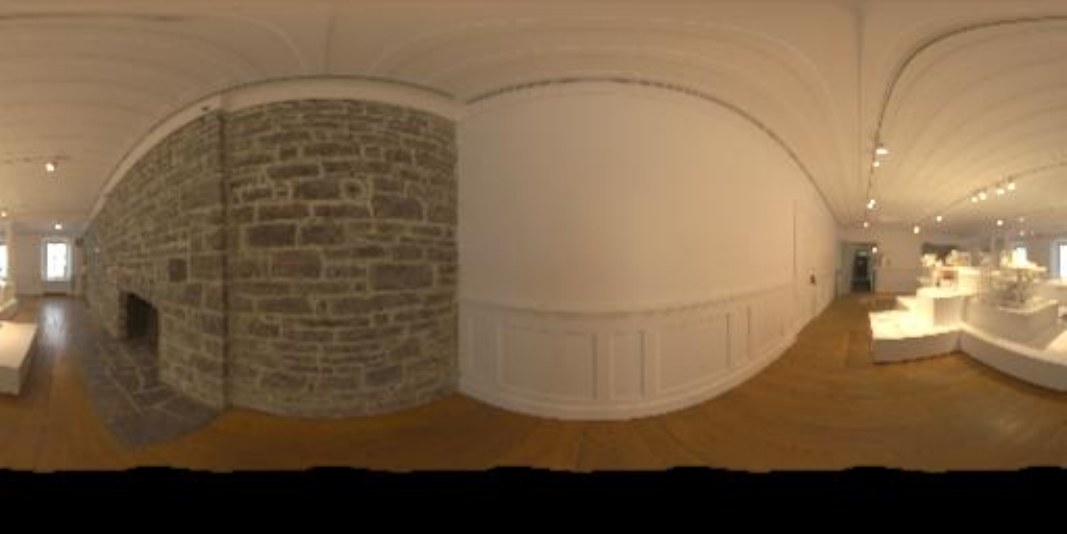}} & 
\noindent\parbox[c]{0.100\textwidth}{\includegraphics[width=0.100\textwidth]{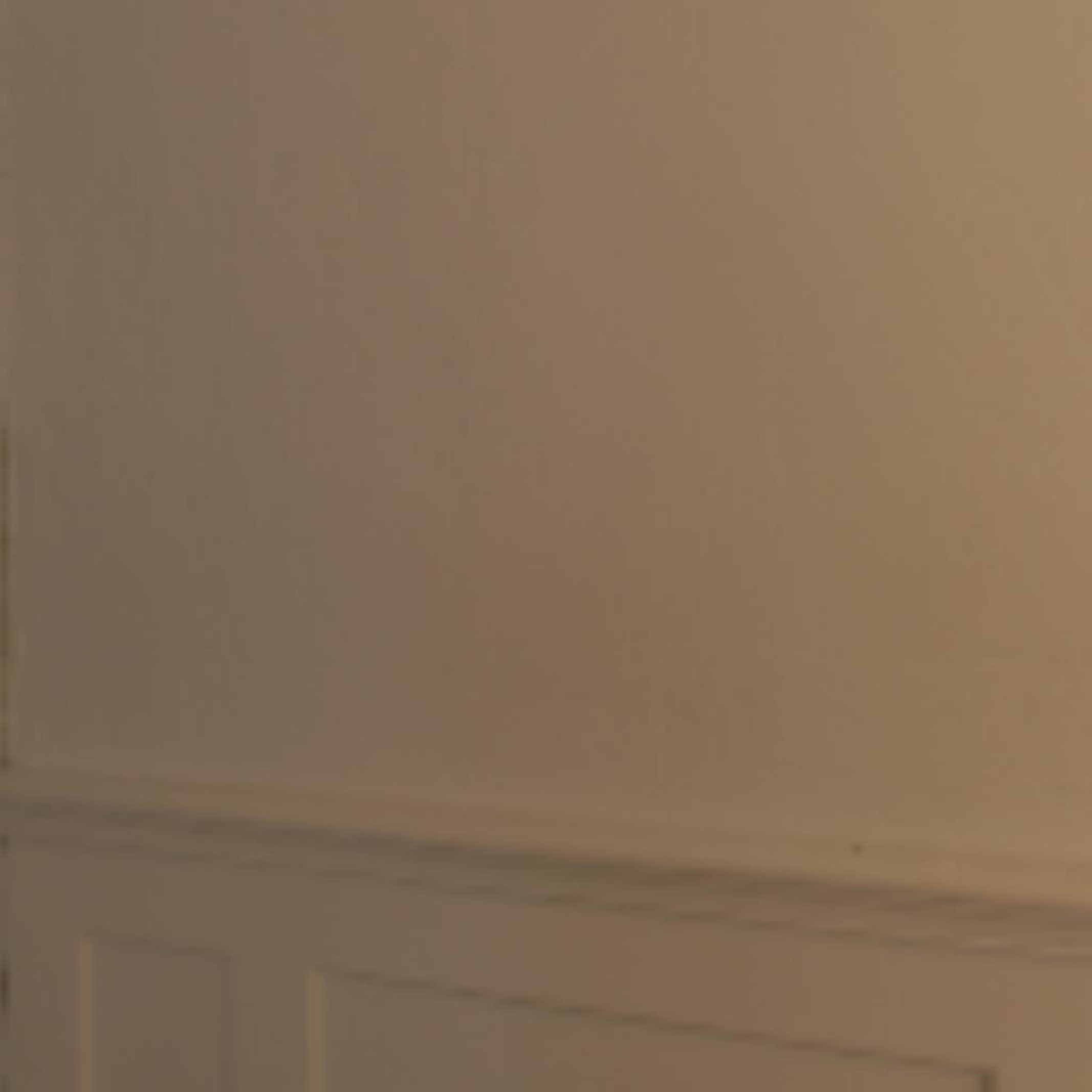}} &  
\noindent\parbox[c]{0.100\textwidth}{\includegraphics[width=0.100\textwidth]{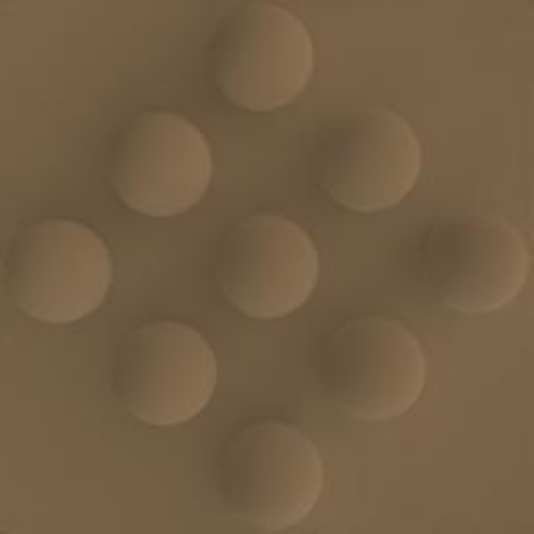}} & 
\noindent\parbox[c]{0.100\textwidth}{\includegraphics[width=0.100\textwidth]{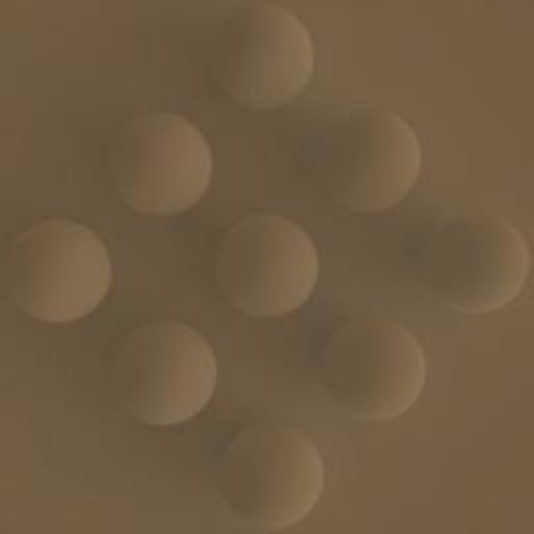}} &
\noindent\parbox[c]{0.200\textwidth}{\includegraphics[width=0.200\textwidth]{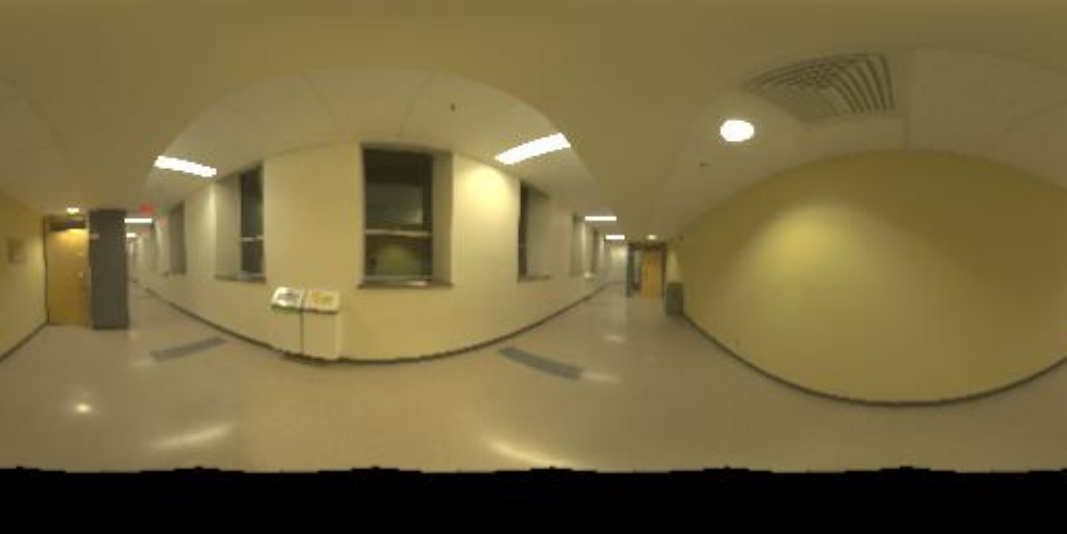}} & 
\noindent\parbox[c]{0.100\textwidth}{\includegraphics[width=0.100\textwidth]{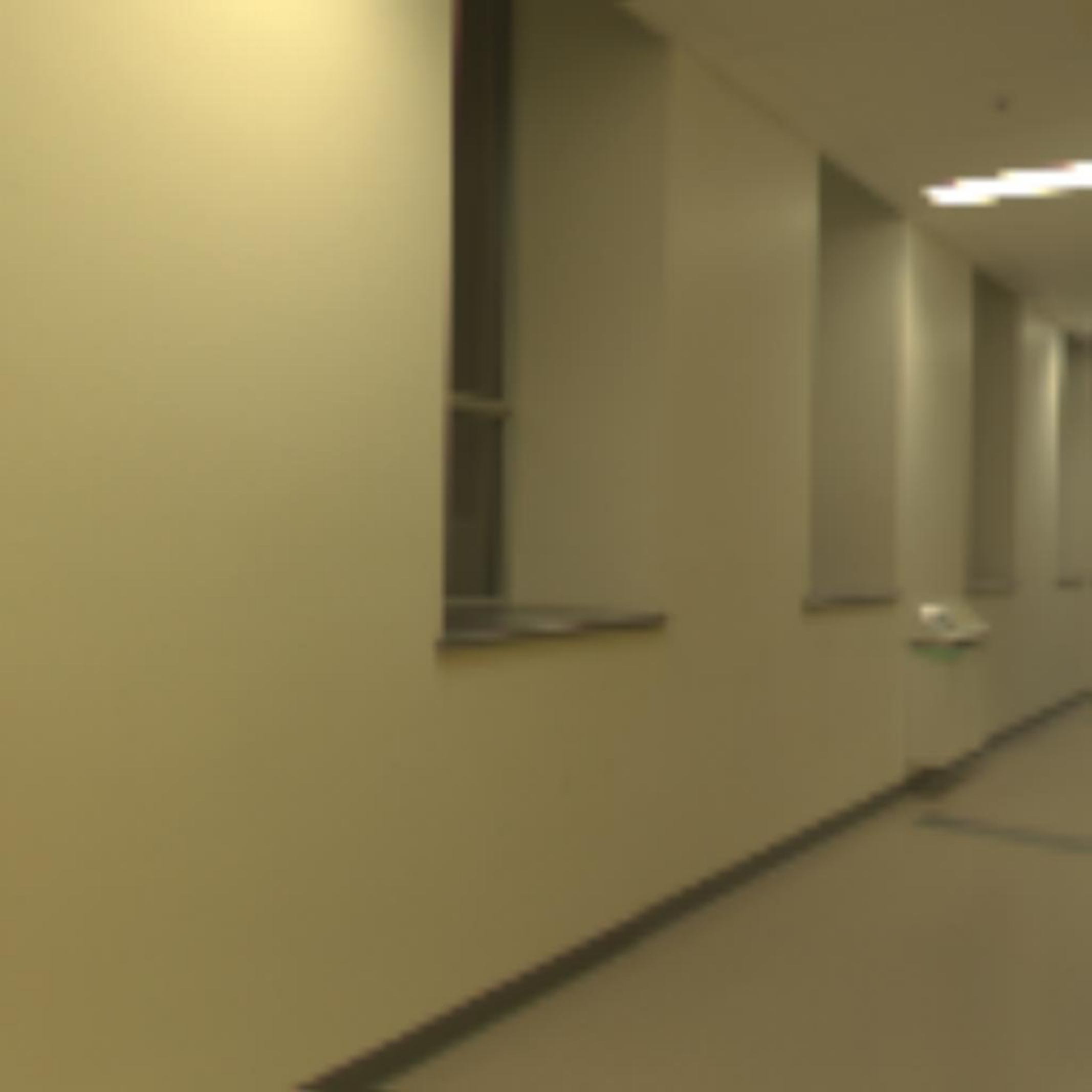}} &  
\noindent\parbox[c]{0.100\textwidth}{\includegraphics[width=0.100\textwidth]{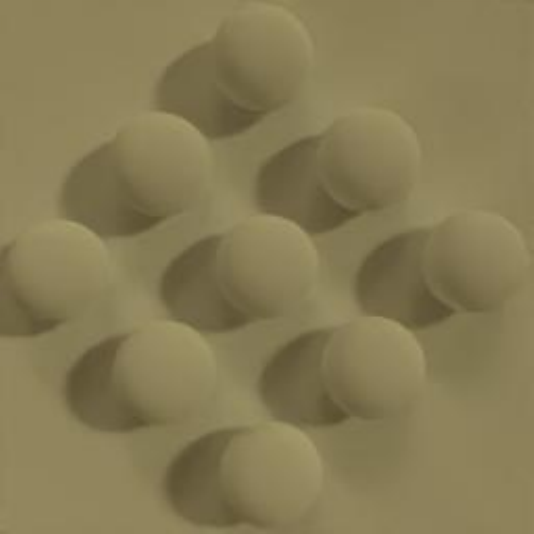}} & 
\noindent\parbox[c]{0.100\textwidth}{\includegraphics[width=0.100\textwidth]{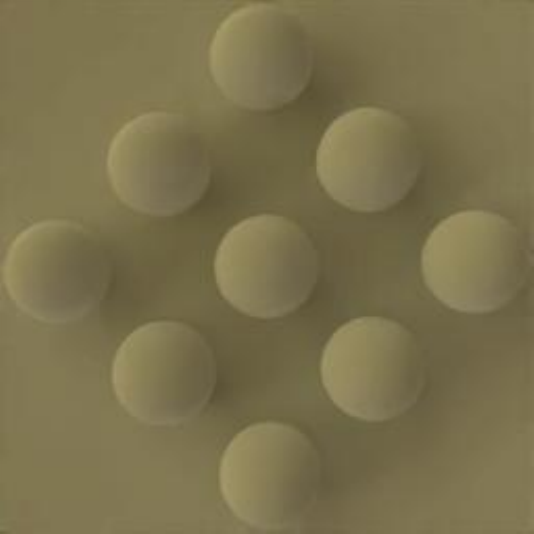}}
\\

\noindent\parbox[c]{0.200\textwidth}{\includegraphics[width=0.200\textwidth]{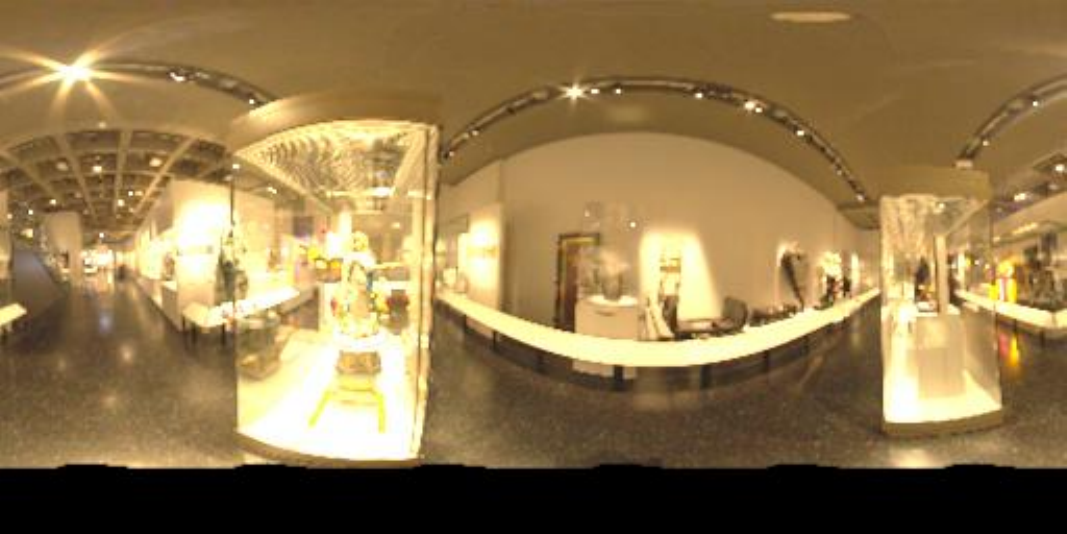}} & 
\noindent\parbox[c]{0.100\textwidth}{\includegraphics[width=0.100\textwidth]{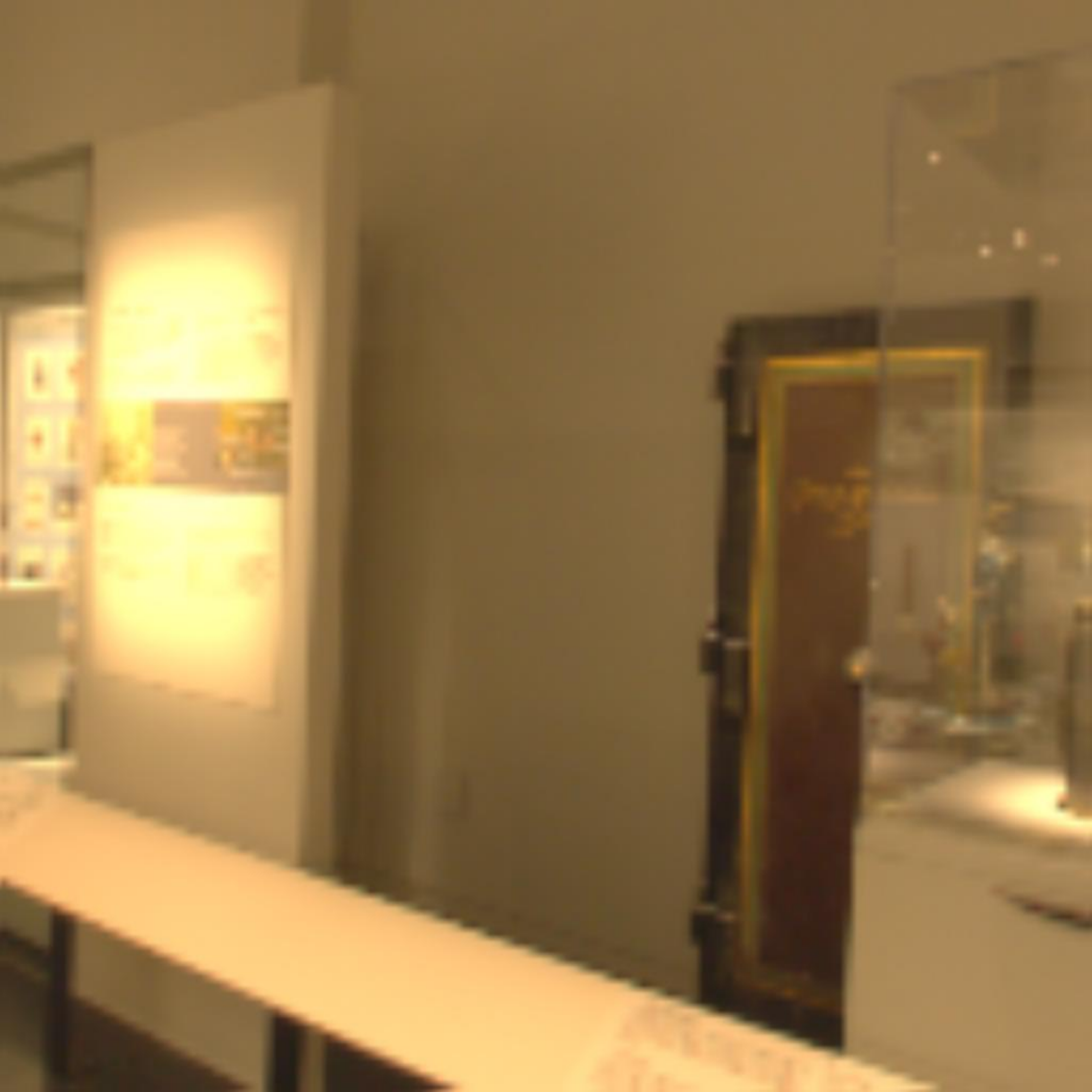}} &  
\noindent\parbox[c]{0.100\textwidth}{\includegraphics[width=0.100\textwidth]{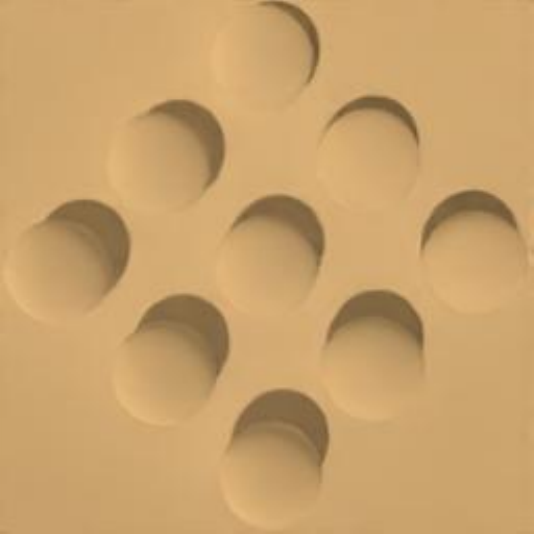}} & 
\noindent\parbox[c]{0.100\textwidth}{\includegraphics[width=0.100\textwidth]{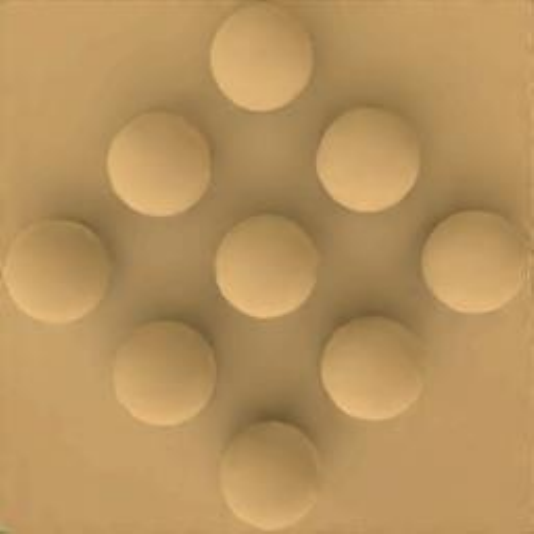}} &
\noindent\parbox[c]{0.200\textwidth}{\includegraphics[width=0.200\textwidth]{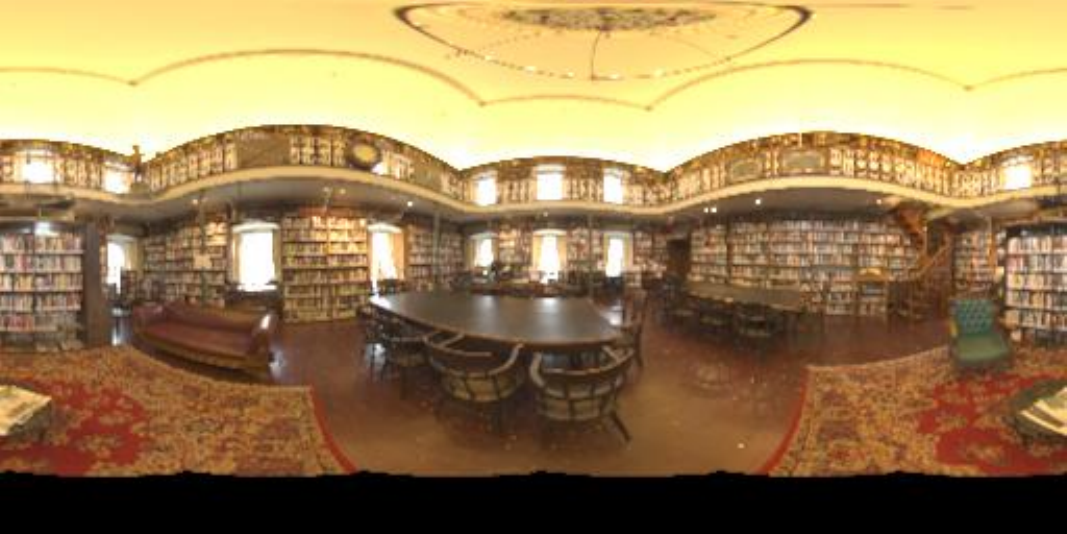}} & 
\noindent\parbox[c]{0.100\textwidth}{\includegraphics[width=0.100\textwidth]{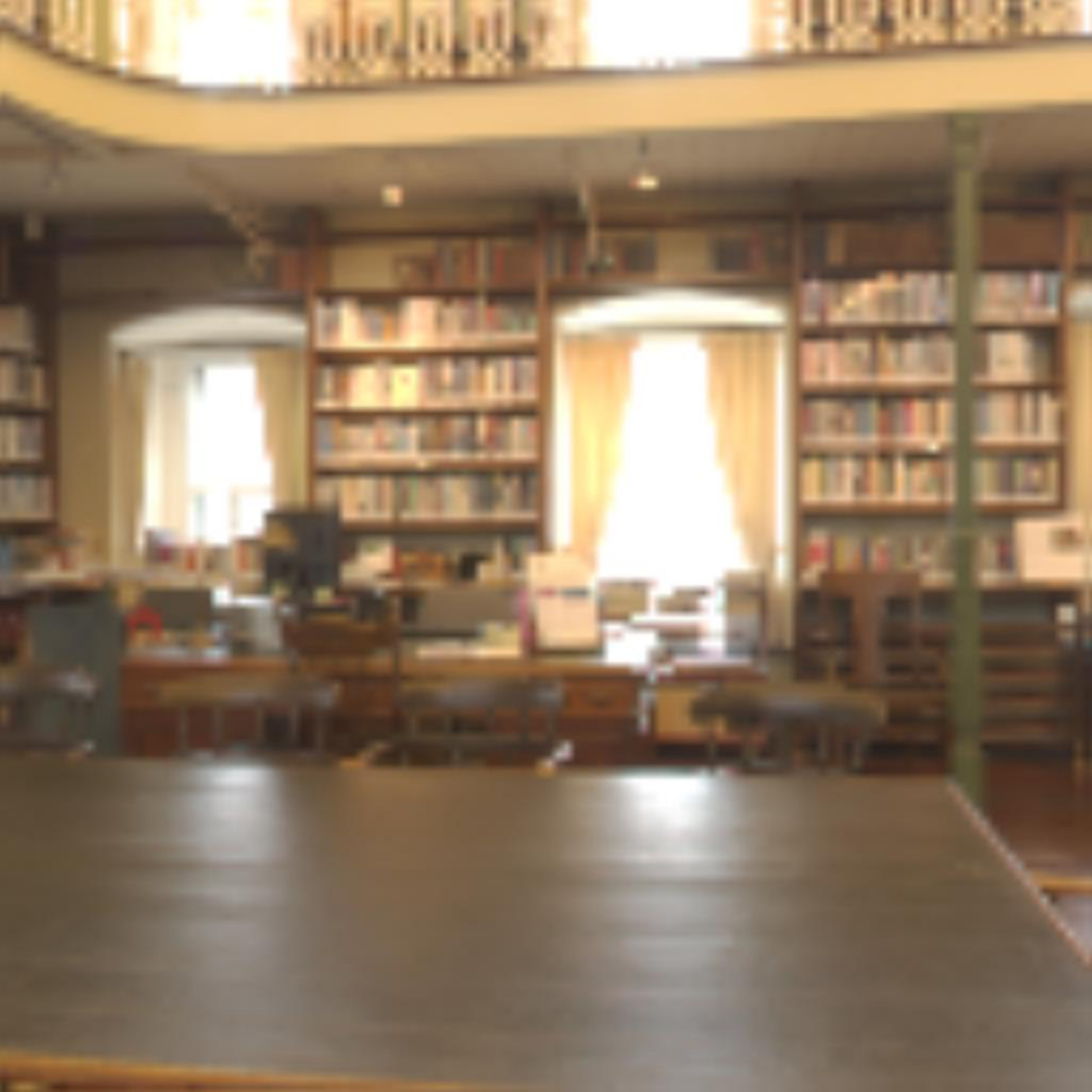}} &  
\noindent\parbox[c]{0.100\textwidth}{\includegraphics[width=0.100\textwidth]{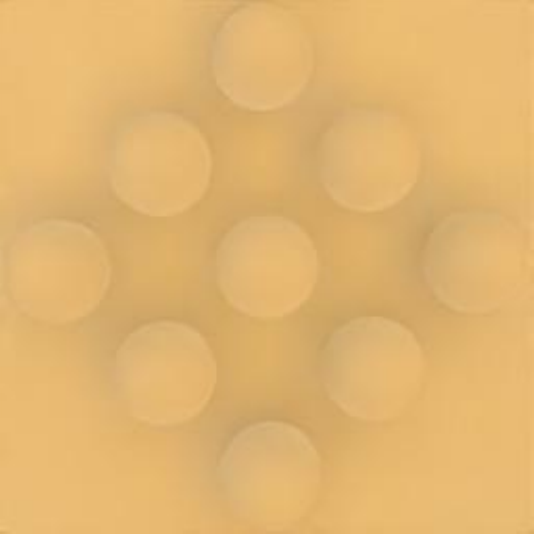}} & 
\noindent\parbox[c]{0.100\textwidth}{\includegraphics[width=0.100\textwidth]{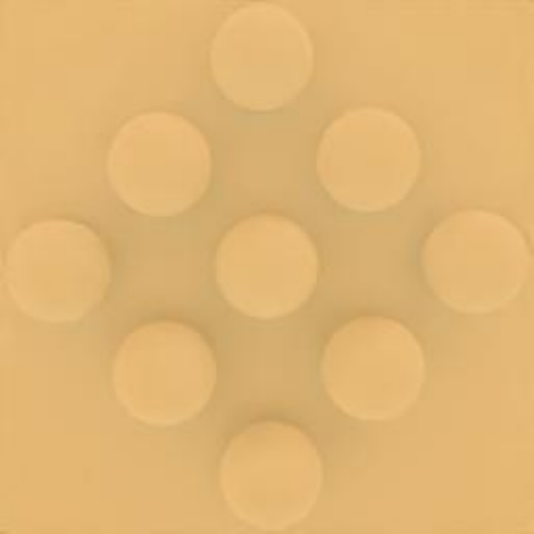}}
\\

\noindent\parbox[c]{0.200\textwidth}{\includegraphics[width=0.200\textwidth]{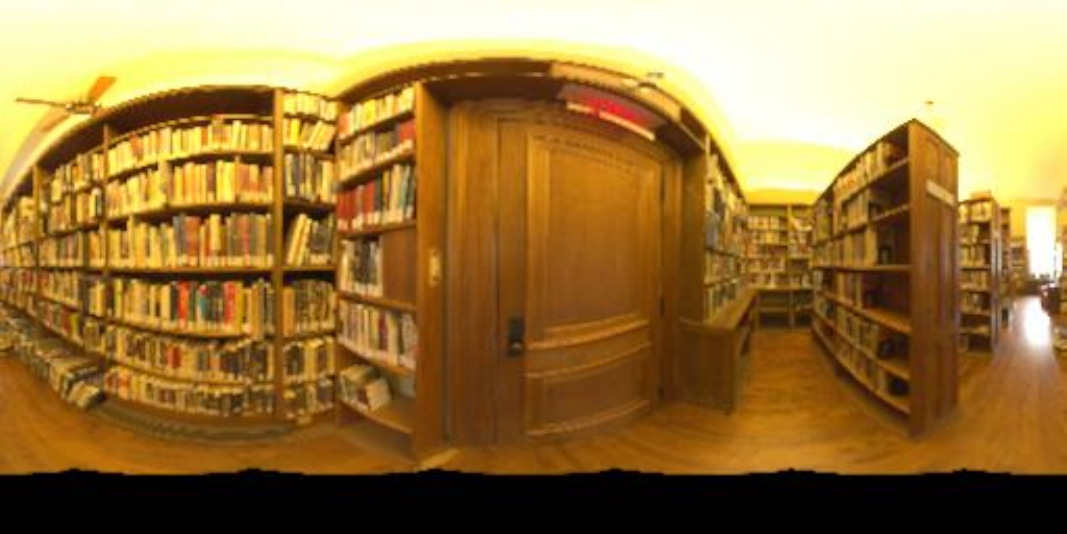}} & 
\noindent\parbox[c]{0.100\textwidth}{\includegraphics[width=0.100\textwidth]{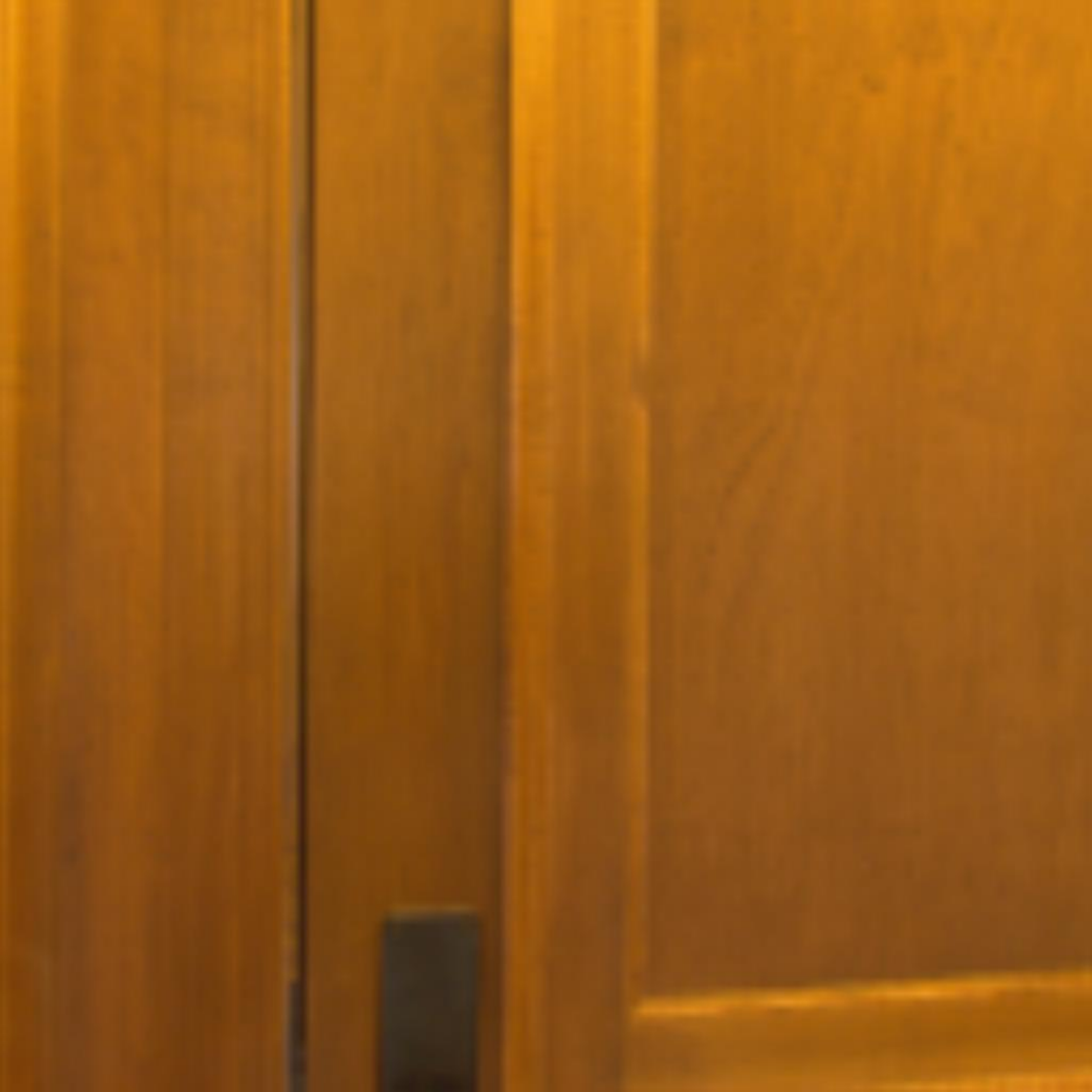}} &  
\noindent\parbox[c]{0.100\textwidth}{\includegraphics[width=0.100\textwidth]{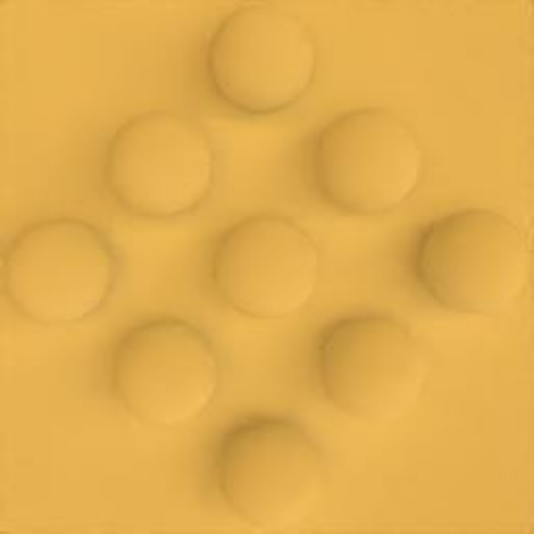}} & 
\noindent\parbox[c]{0.100\textwidth}{\includegraphics[width=0.100\textwidth]{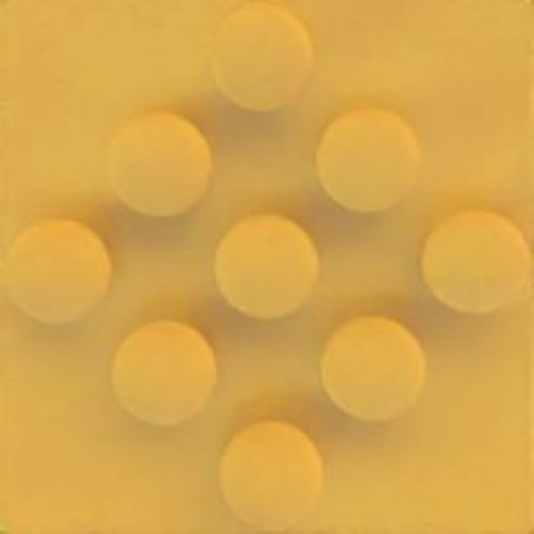}} &
\noindent\parbox[c]{0.200\textwidth}{\includegraphics[width=0.200\textwidth]{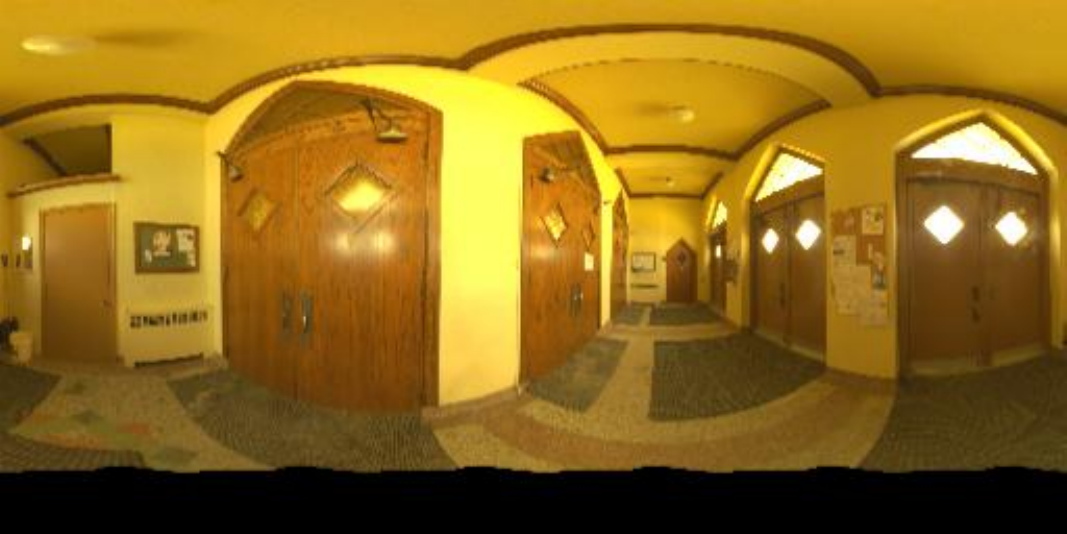}} & 
\noindent\parbox[c]{0.100\textwidth}{\includegraphics[width=0.100\textwidth]{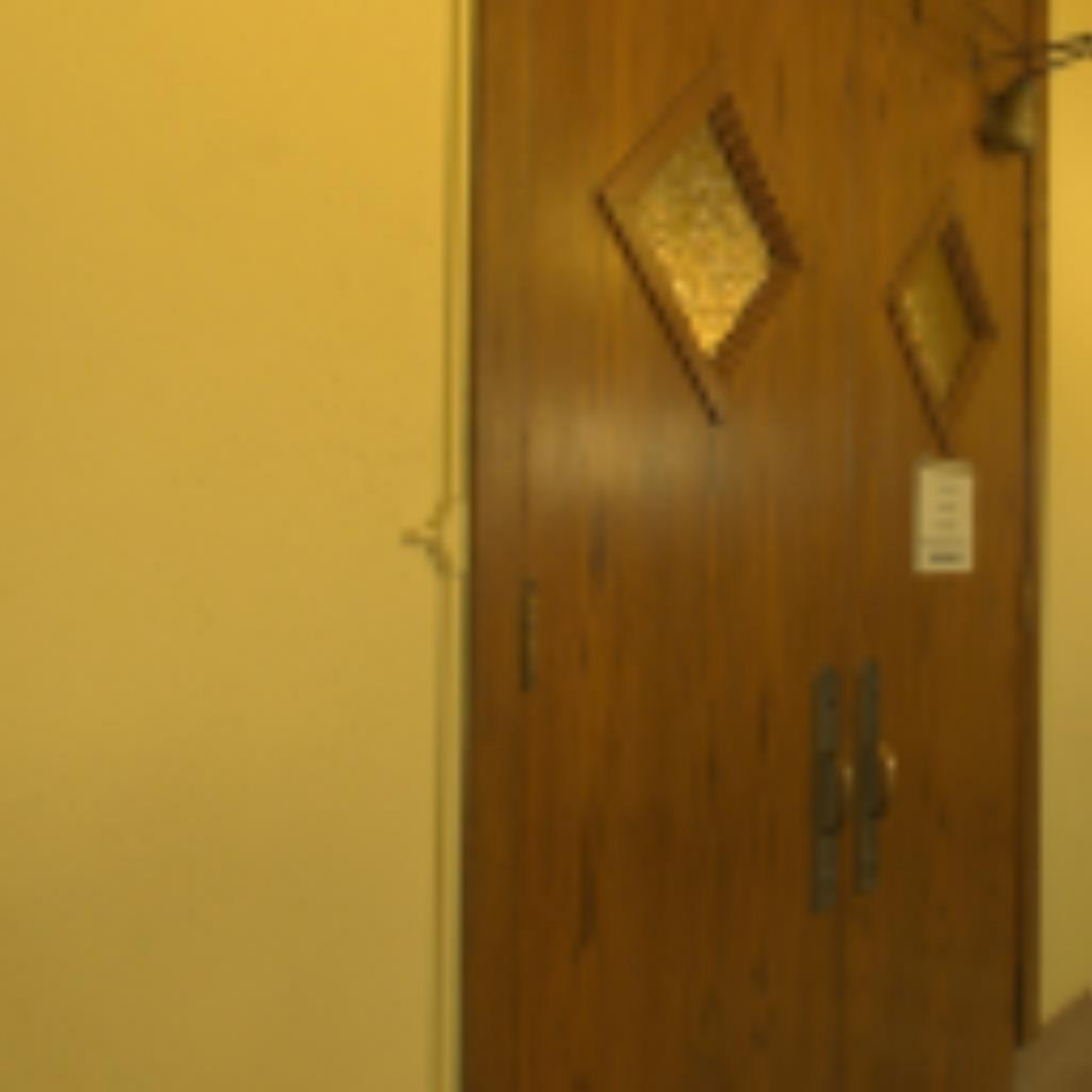}} &  
\noindent\parbox[c]{0.100\textwidth}{\includegraphics[width=0.100\textwidth]{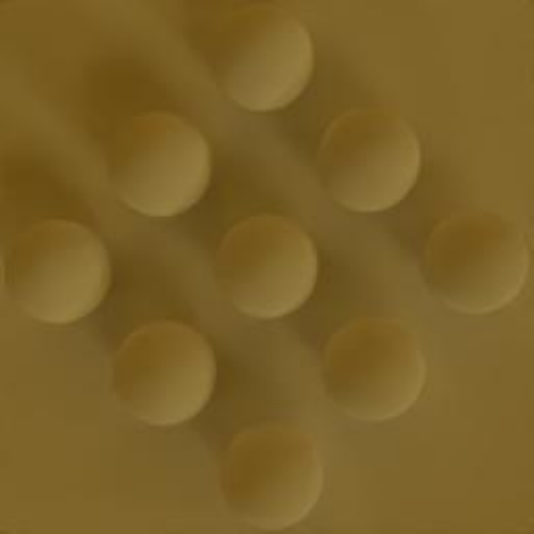}} & 
\noindent\parbox[c]{0.100\textwidth}{\includegraphics[width=0.100\textwidth]{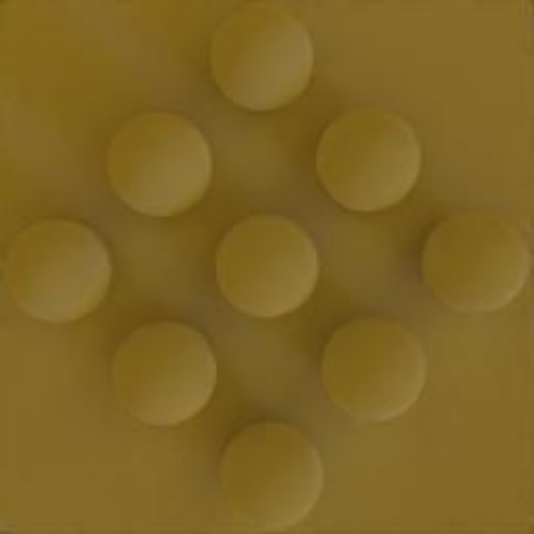}}
\\

\noindent\parbox[c]{0.200\textwidth}{\includegraphics[width=0.200\textwidth]{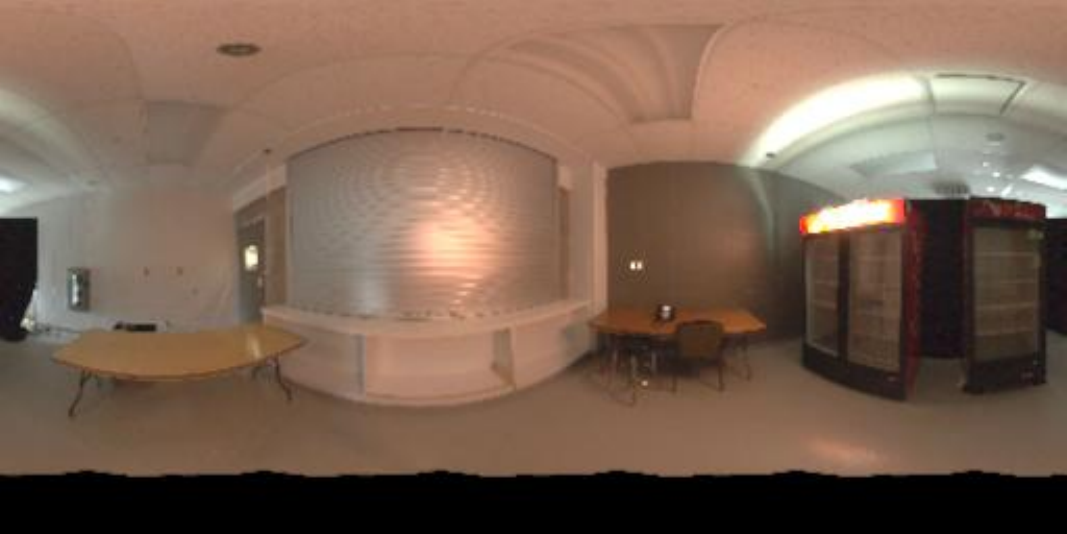}} & 
\noindent\parbox[c]{0.100\textwidth}{\includegraphics[width=0.100\textwidth]{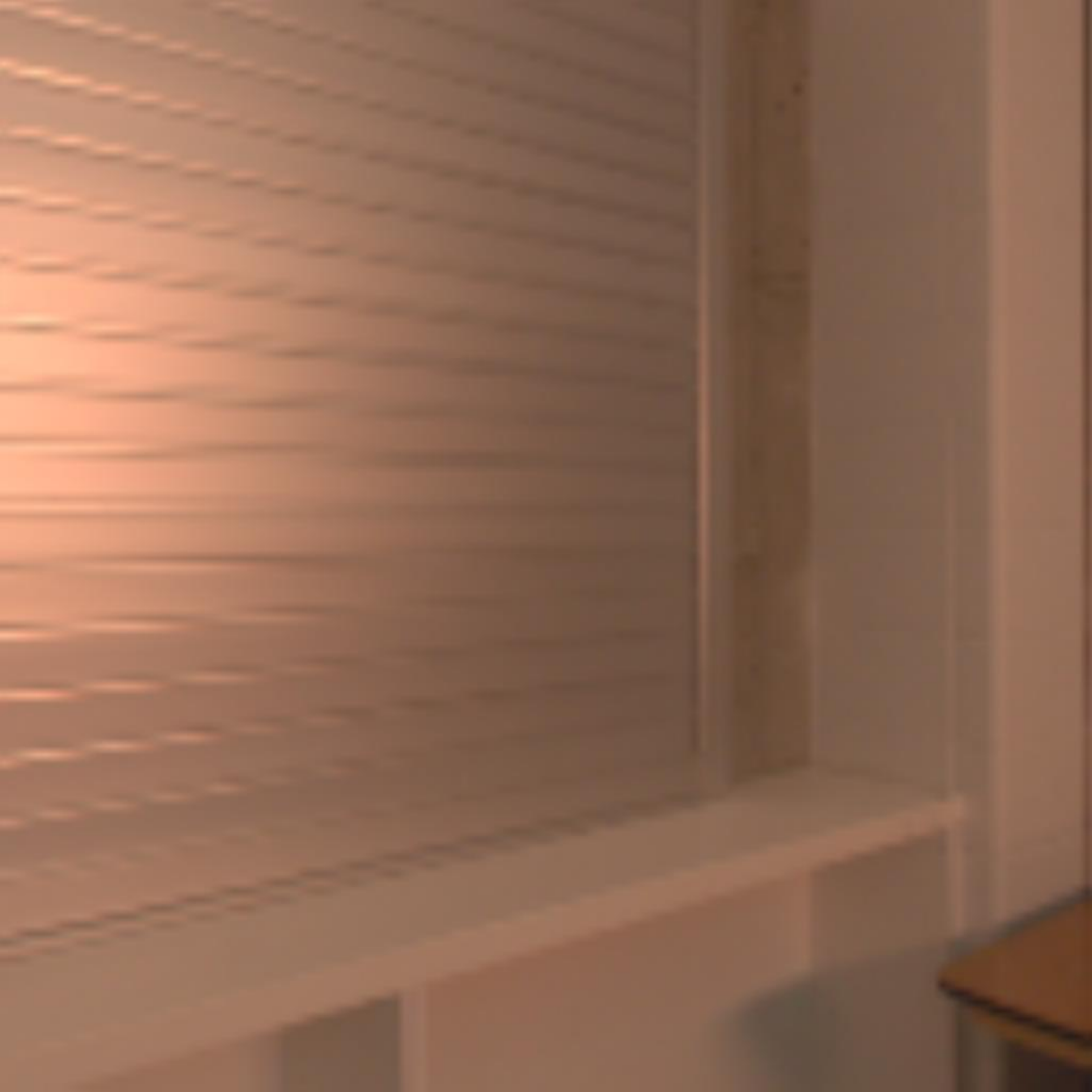}} &  
\noindent\parbox[c]{0.100\textwidth}{\includegraphics[width=0.100\textwidth]{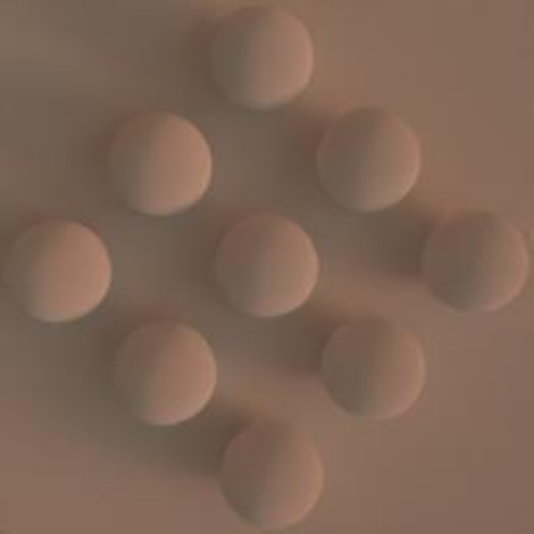}} & 
\noindent\parbox[c]{0.100\textwidth}{\includegraphics[width=0.100\textwidth]{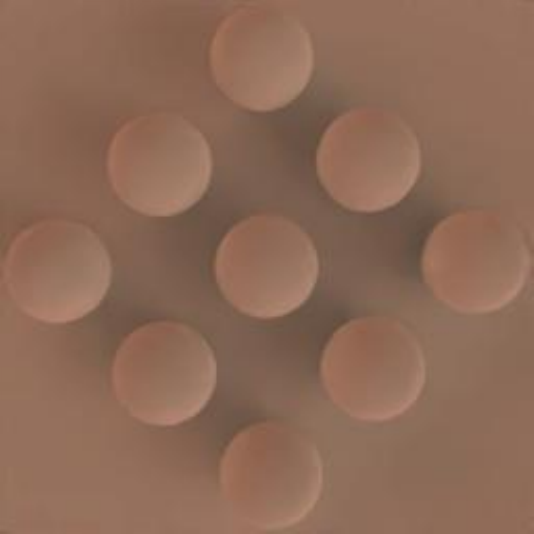}} &
\noindent\parbox[c]{0.200\textwidth}{\includegraphics[width=0.200\textwidth]{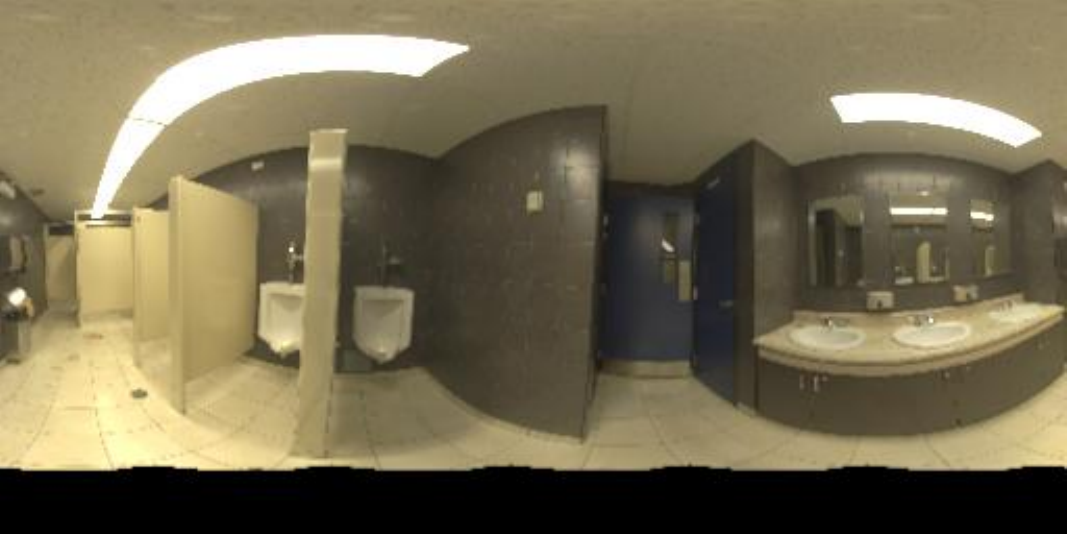}} & 
\noindent\parbox[c]{0.100\textwidth}{\includegraphics[width=0.100\textwidth]{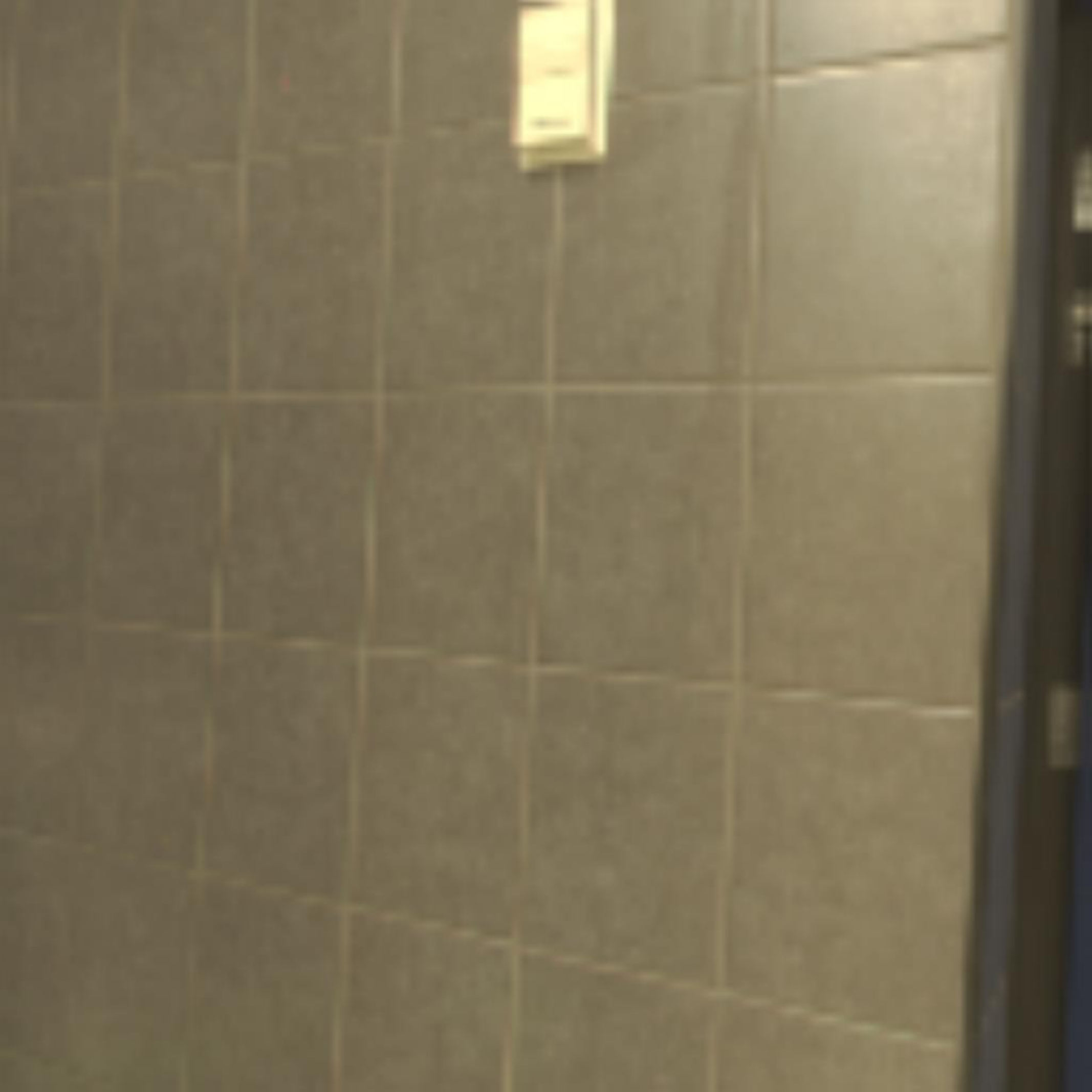}} &  
\noindent\parbox[c]{0.100\textwidth}{\includegraphics[width=0.100\textwidth]{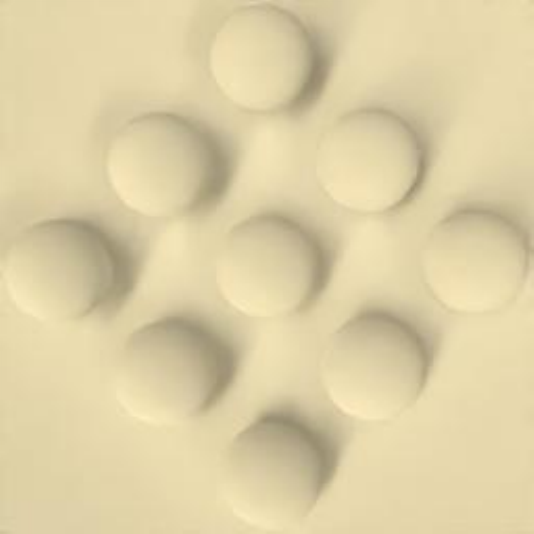}} & 
\noindent\parbox[c]{0.100\textwidth}{\includegraphics[width=0.100\textwidth]{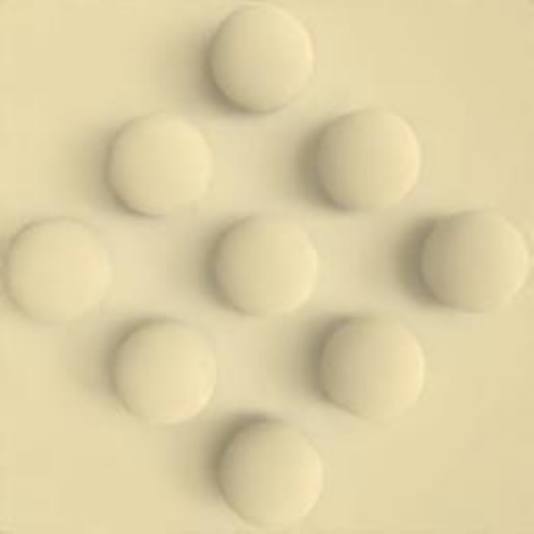}}

\\
\noindent\parbox[c]{0.200\textwidth}{\includegraphics[width=0.200\textwidth]{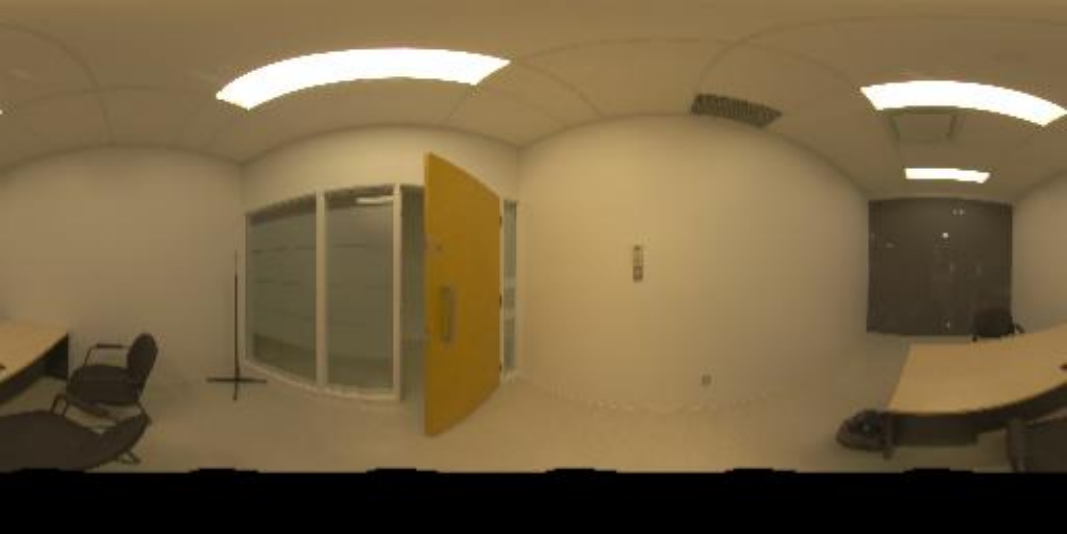}} & 
\noindent\parbox[c]{0.100\textwidth}{\includegraphics[width=0.100\textwidth]{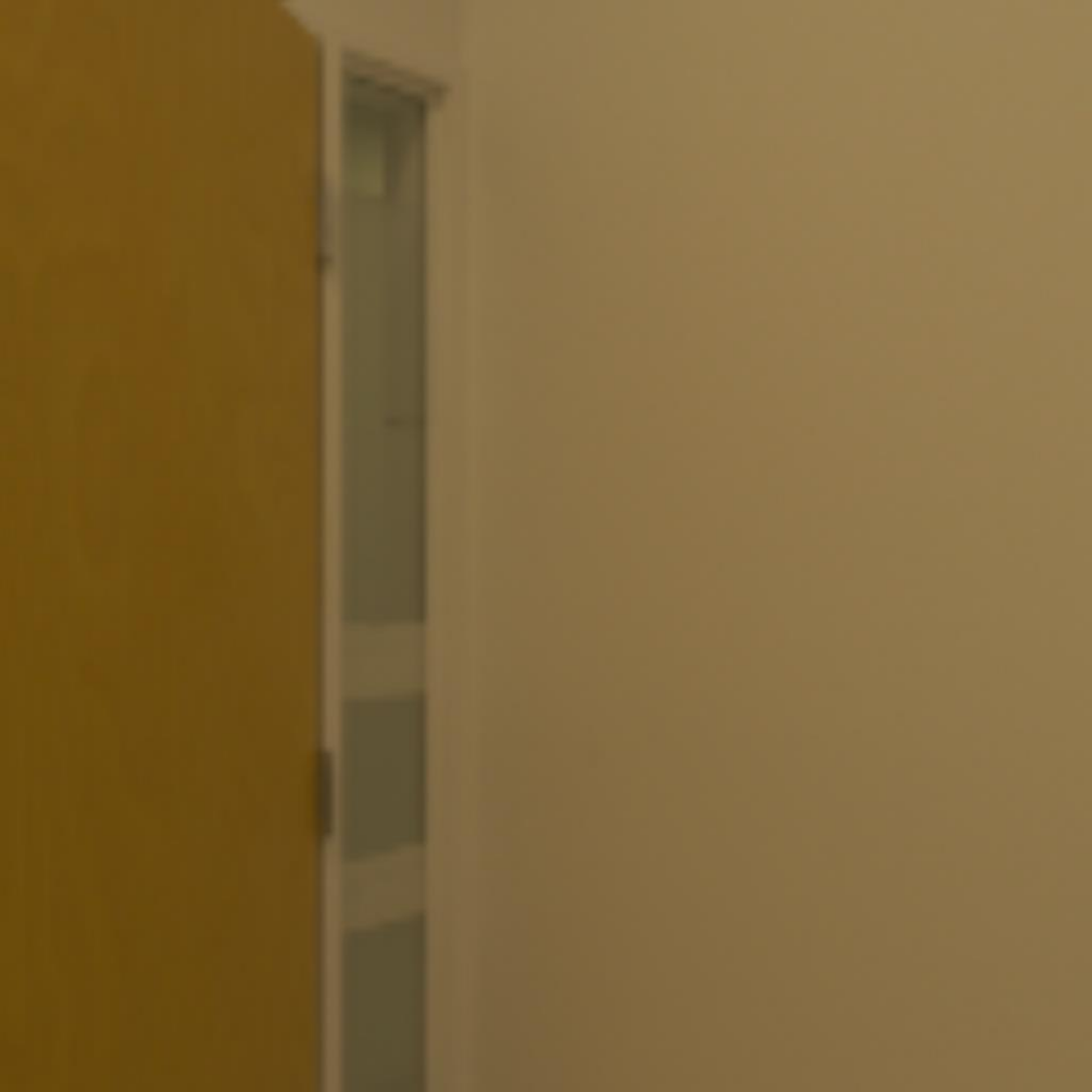}} &  
\noindent\parbox[c]{0.100\textwidth}{\includegraphics[width=0.100\textwidth]{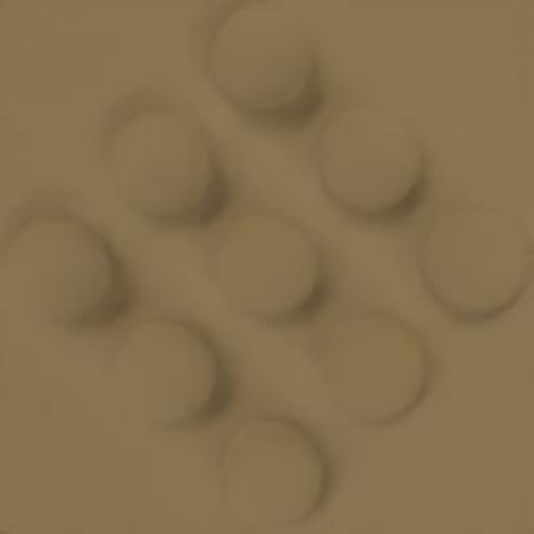}} & 
\noindent\parbox[c]{0.100\textwidth}{\includegraphics[width=0.100\textwidth]{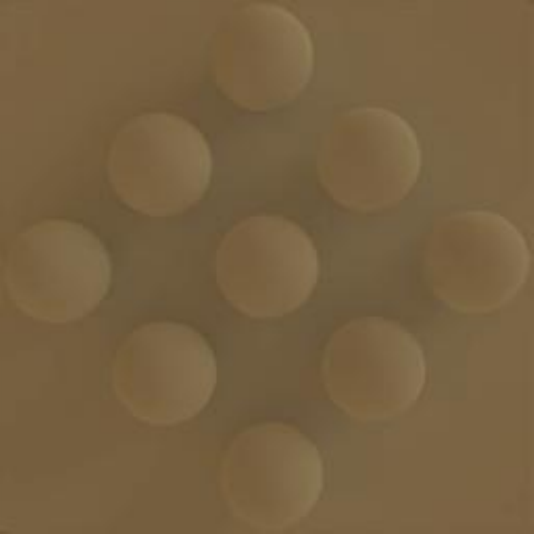}} &
\noindent\parbox[c]{0.200\textwidth}{\includegraphics[width=0.200\textwidth]{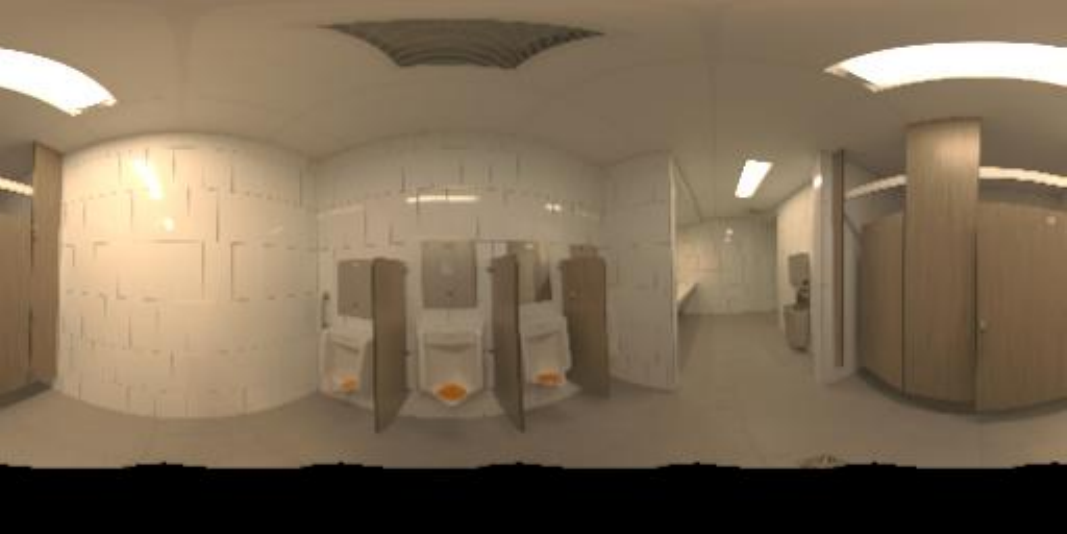}} & 
\noindent\parbox[c]{0.100\textwidth}{\includegraphics[width=0.100\textwidth]{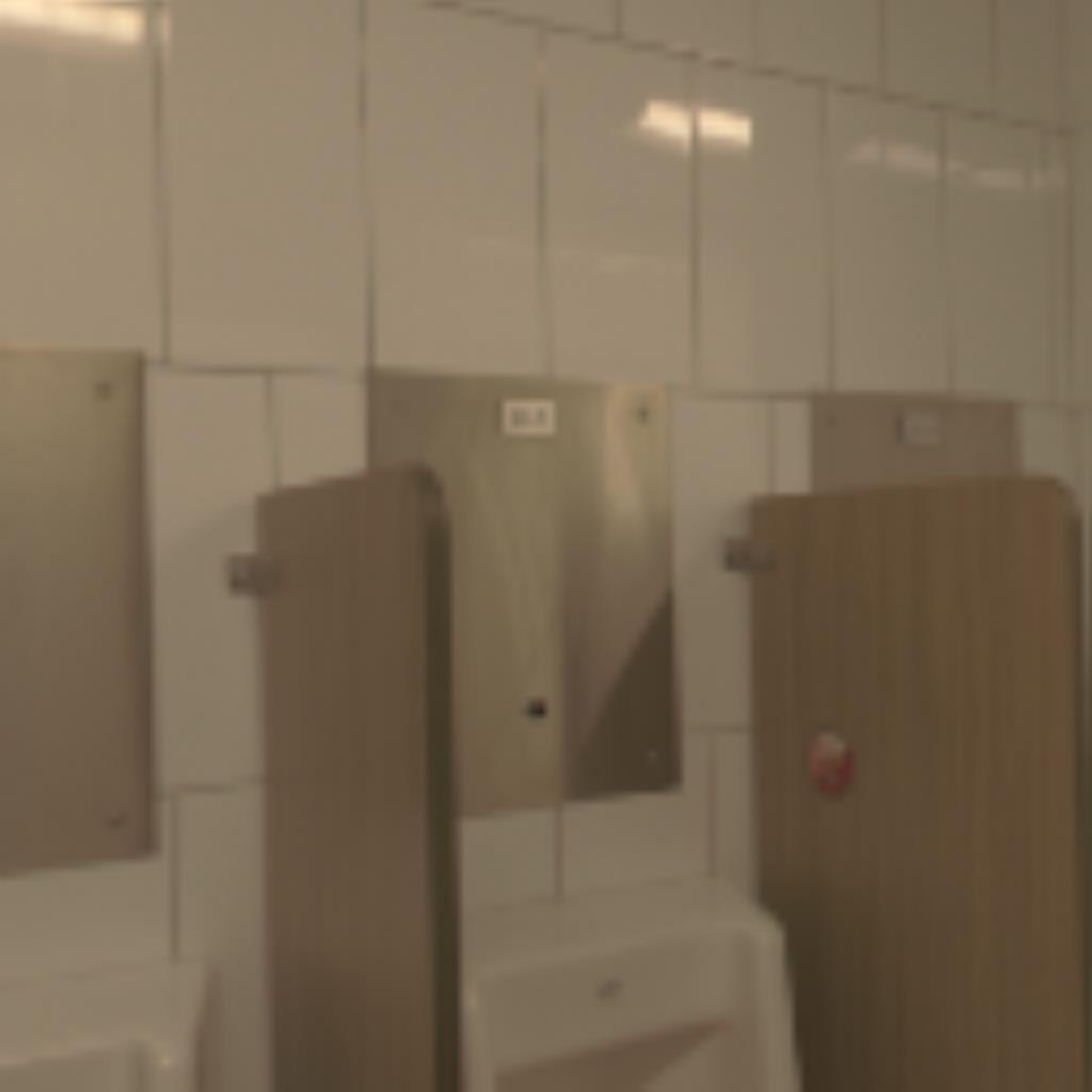}} &  
\noindent\parbox[c]{0.100\textwidth}{\includegraphics[width=0.100\textwidth]{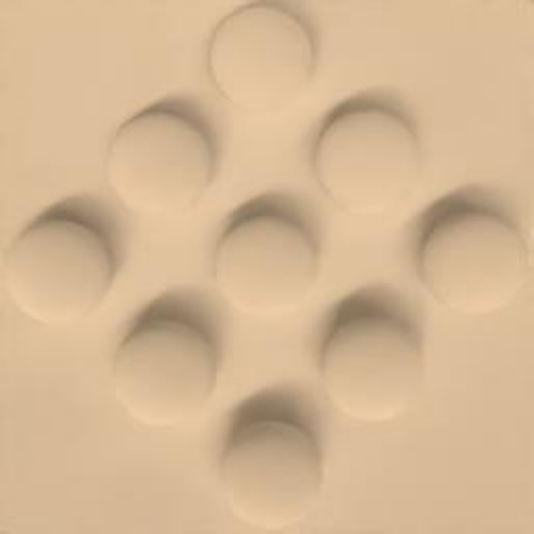}} & 
\noindent\parbox[c]{0.100\textwidth}{\includegraphics[width=0.100\textwidth]{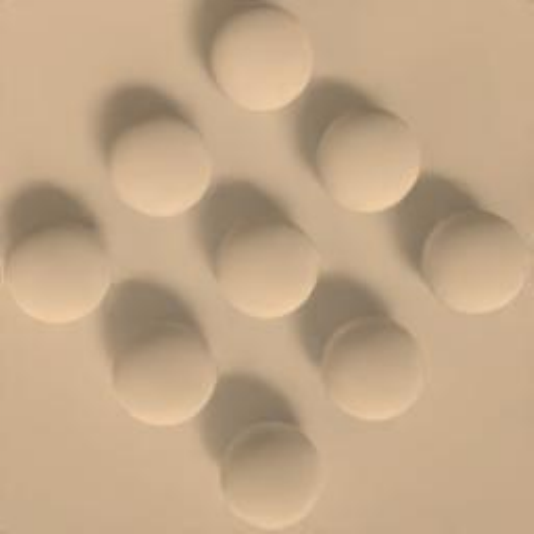}}
\\

\noindent\parbox[c]{0.200\textwidth}{\includegraphics[width=0.200\textwidth]{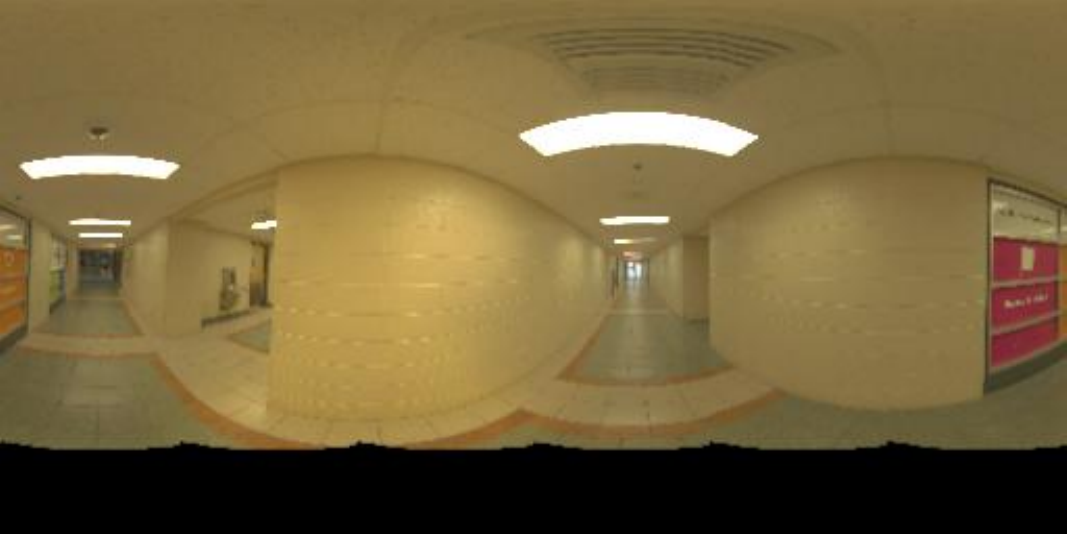}} & 
\noindent\parbox[c]{0.100\textwidth}{\includegraphics[width=0.100\textwidth]{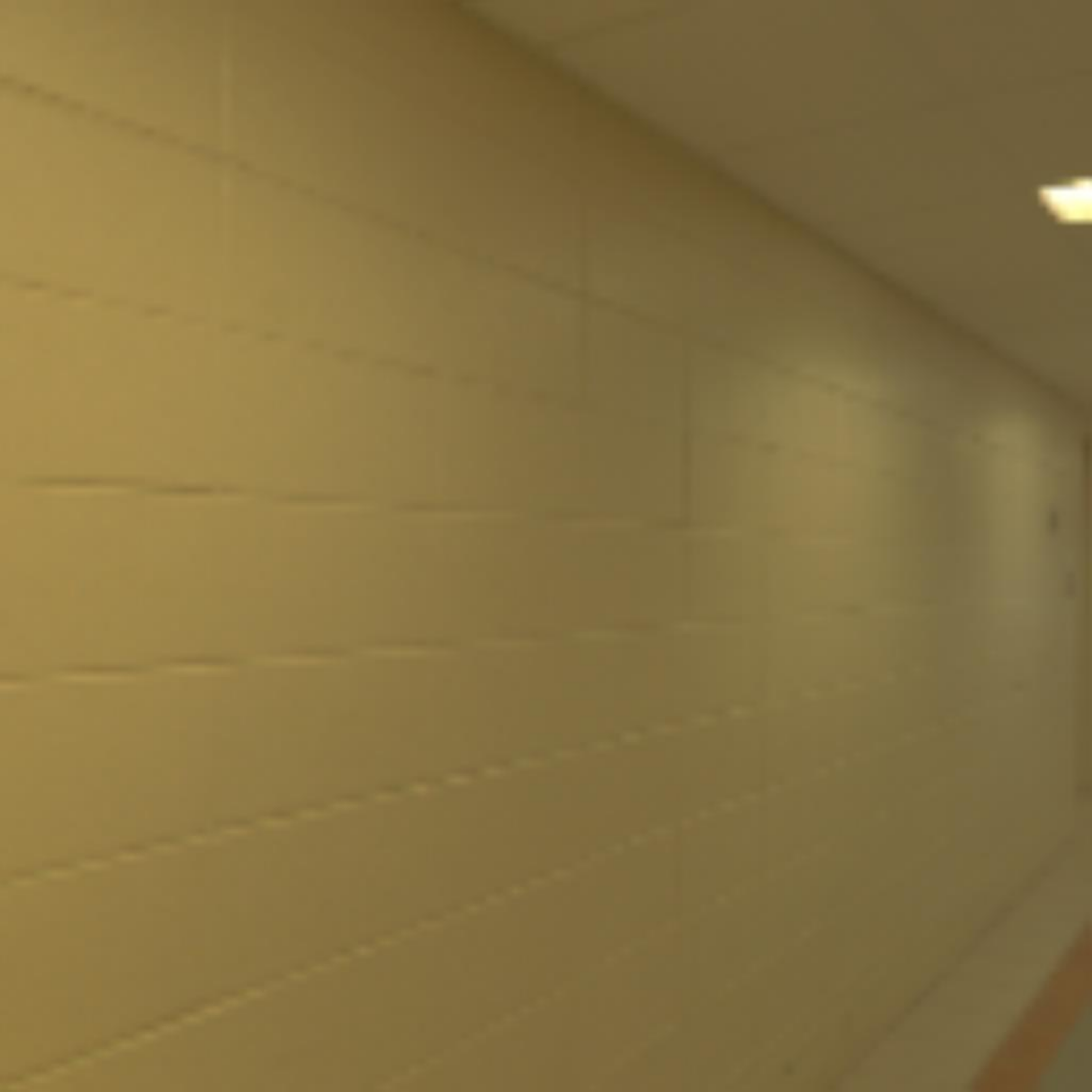}} &  
\noindent\parbox[c]{0.100\textwidth}{\includegraphics[width=0.100\textwidth]{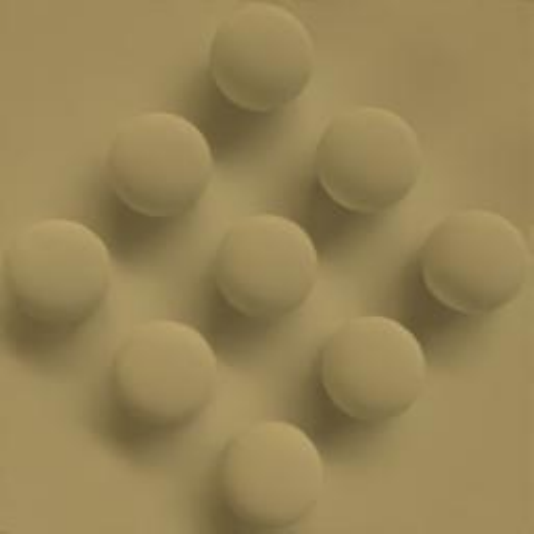}} & 
\noindent\parbox[c]{0.100\textwidth}{\includegraphics[width=0.100\textwidth]{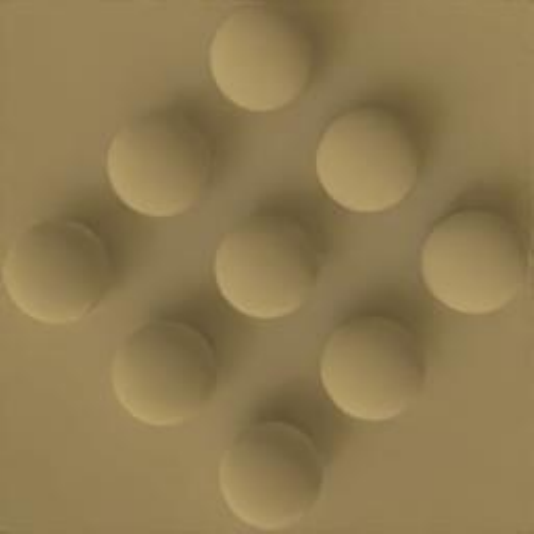}} &
\noindent\parbox[c]{0.200\textwidth}{\includegraphics[width=0.200\textwidth]{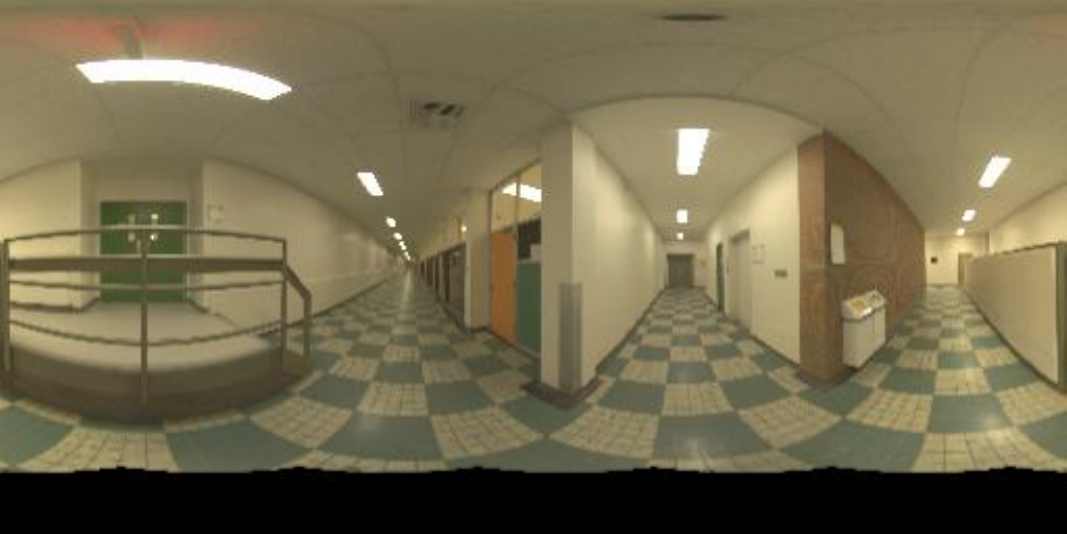}} & 
\noindent\parbox[c]{0.100\textwidth}{\includegraphics[width=0.100\textwidth]{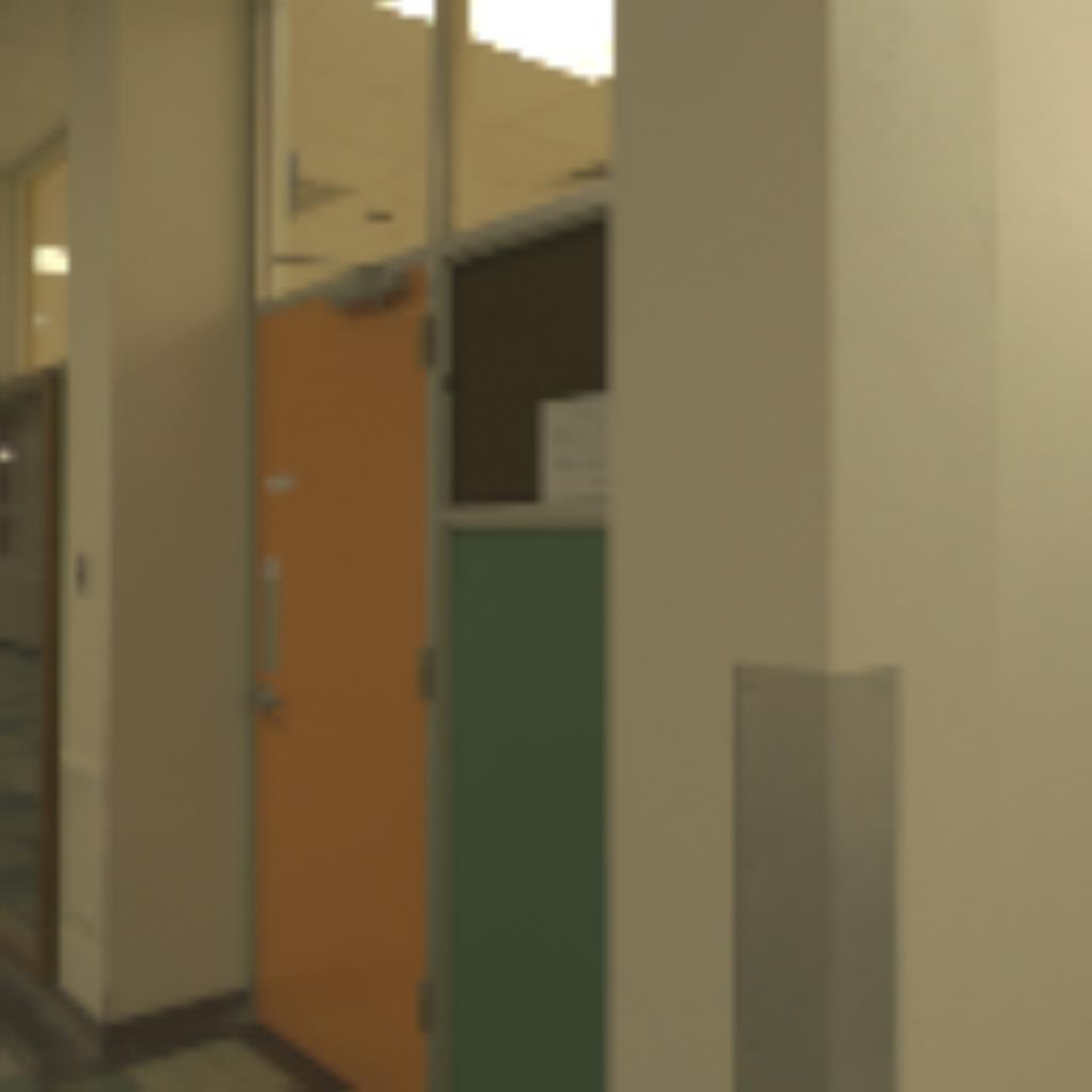}} &  
\noindent\parbox[c]{0.100\textwidth}{\includegraphics[width=0.100\textwidth]{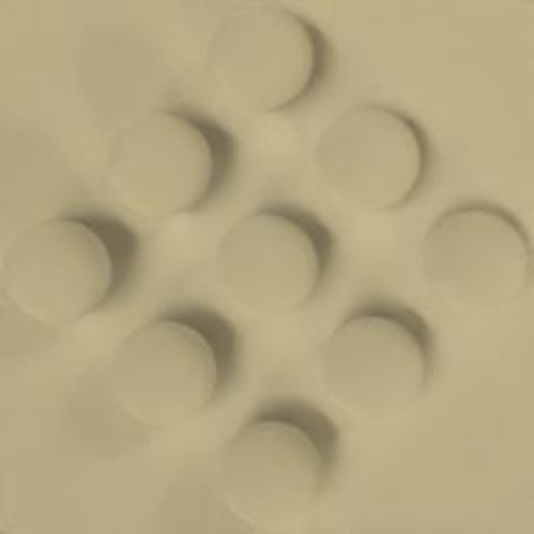}} & 
\noindent\parbox[c]{0.100\textwidth}{\includegraphics[width=0.100\textwidth]{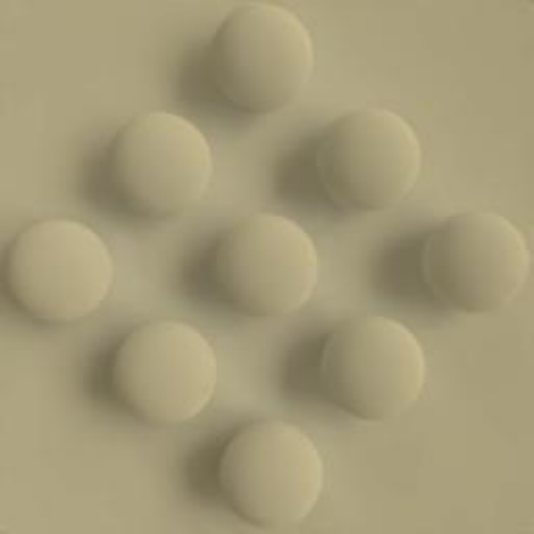}} 
\\

\noindent\parbox[c]{0.200\textwidth}{\includegraphics[width=0.200\textwidth]{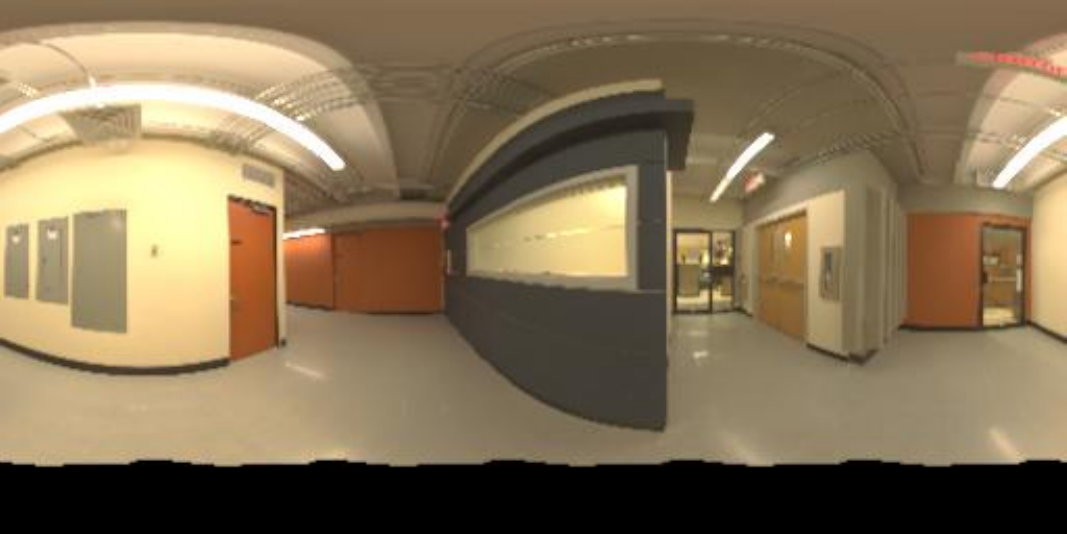}} & 
\noindent\parbox[c]{0.100\textwidth}{\includegraphics[width=0.100\textwidth]{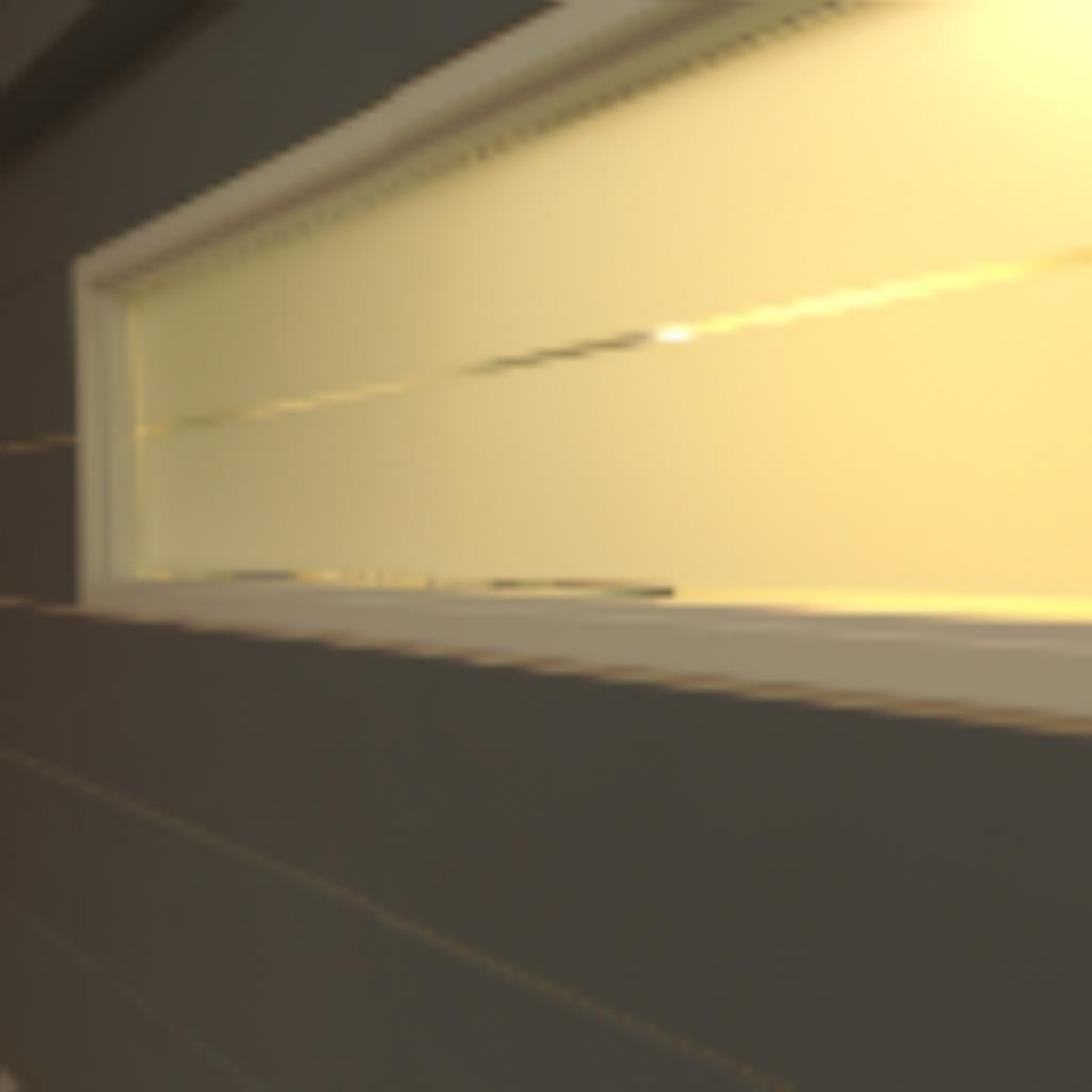}} &  
\noindent\parbox[c]{0.100\textwidth}{\includegraphics[width=0.100\textwidth]{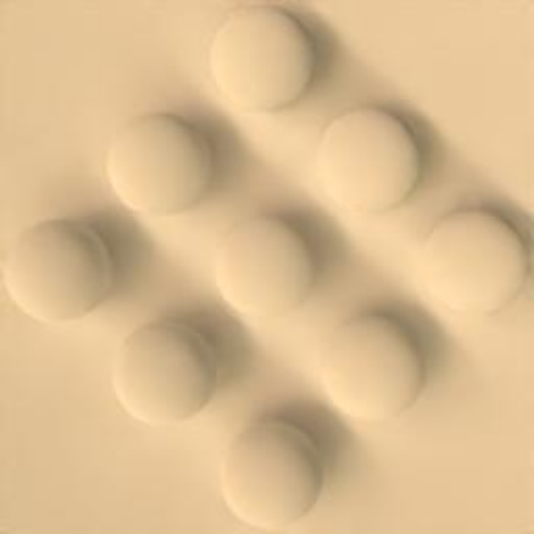}} & 
\noindent\parbox[c]{0.100\textwidth}{\includegraphics[width=0.100\textwidth]{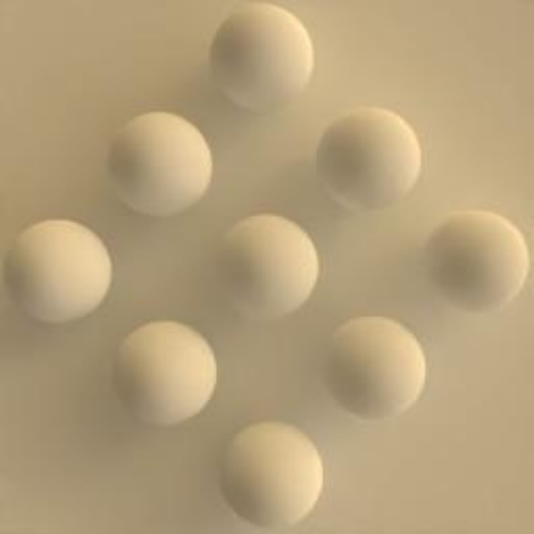}} &
\noindent\parbox[c]{0.200\textwidth}{\includegraphics[width=0.200\textwidth]{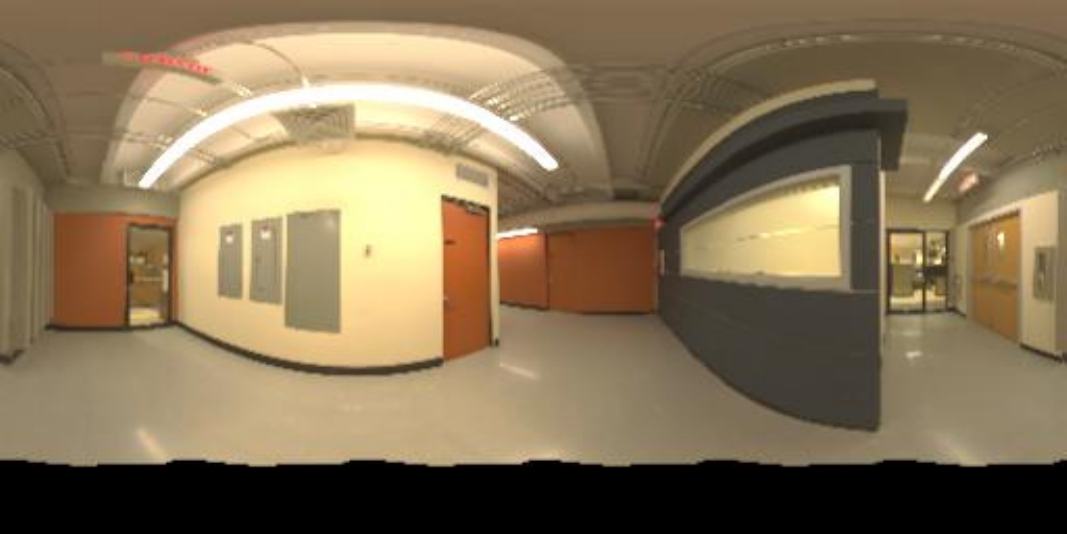}} & 
\noindent\parbox[c]{0.100\textwidth}{\includegraphics[width=0.100\textwidth]{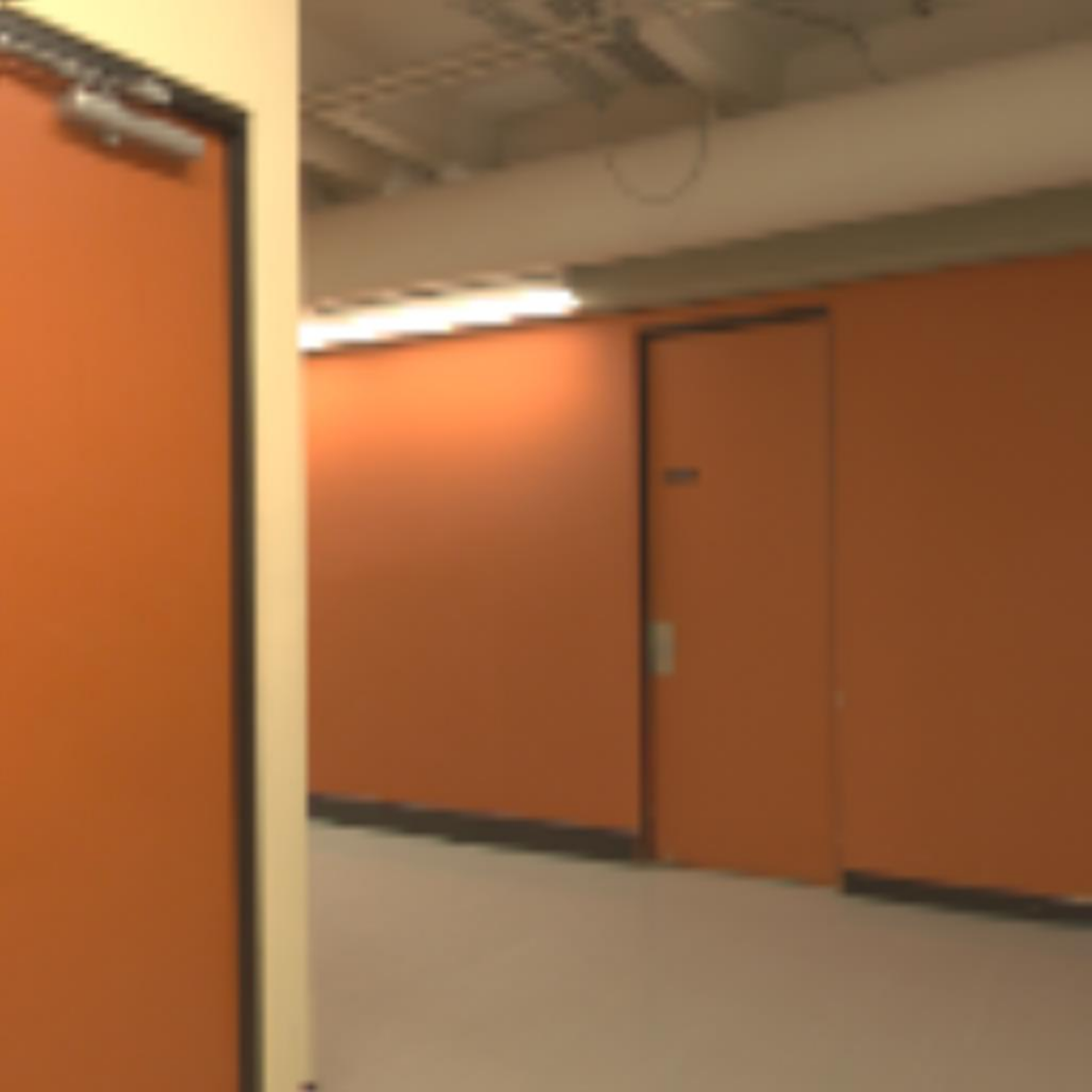}} &  
\noindent\parbox[c]{0.100\textwidth}{\includegraphics[width=0.100\textwidth]{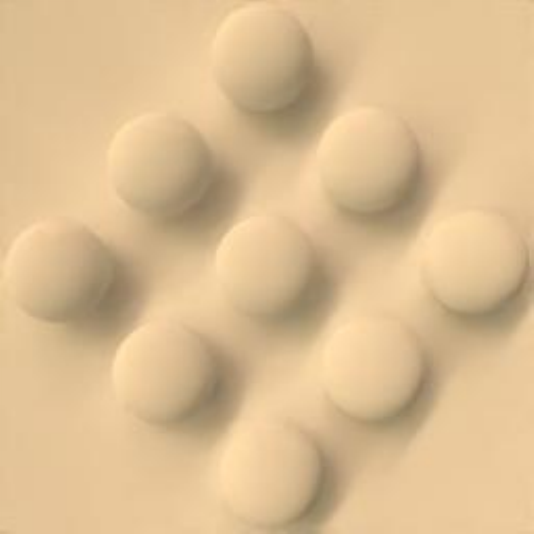}} & 
\noindent\parbox[c]{0.100\textwidth}{\includegraphics[width=0.100\textwidth]{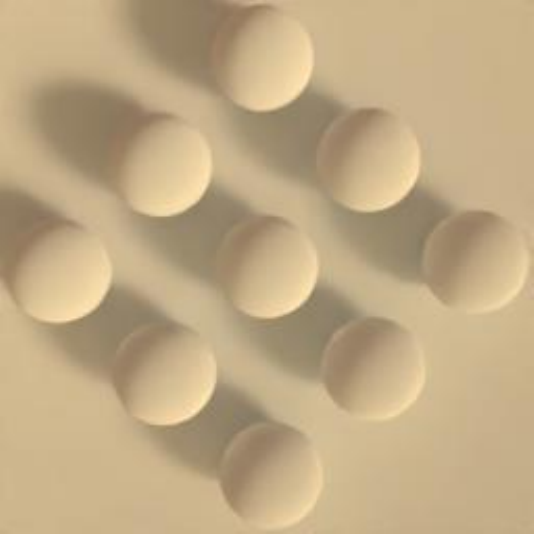}}
\\

\noindent\parbox[c]{0.200\textwidth}{\includegraphics[width=0.200\textwidth]{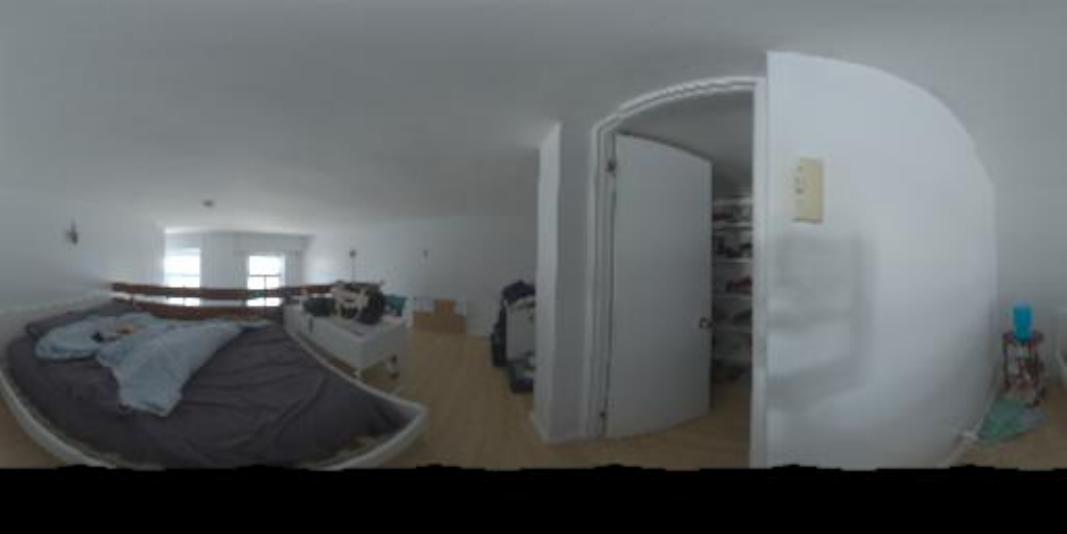}} & 
\noindent\parbox[c]{0.100\textwidth}{\includegraphics[width=0.100\textwidth]{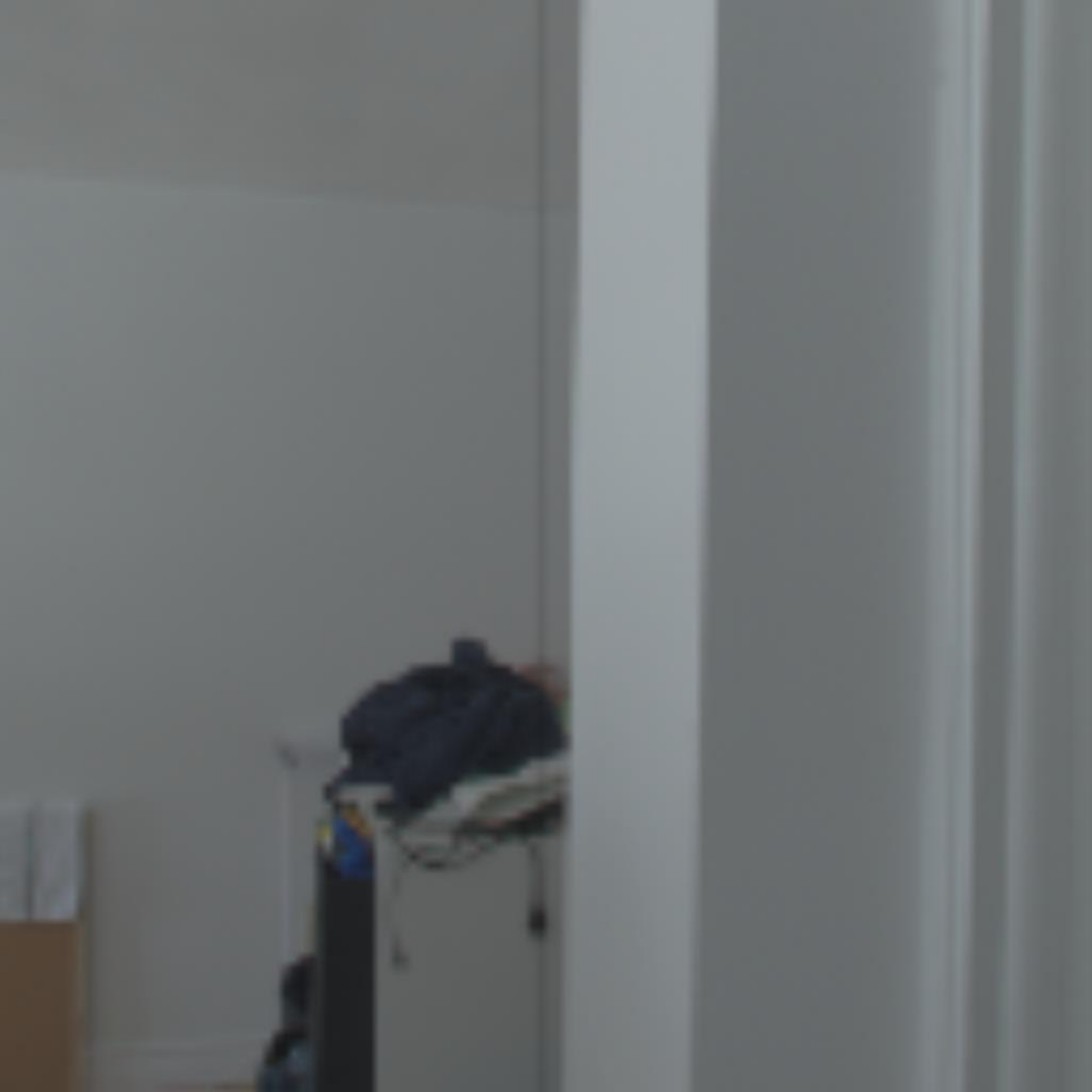}} &  
\noindent\parbox[c]{0.100\textwidth}{\includegraphics[width=0.100\textwidth]{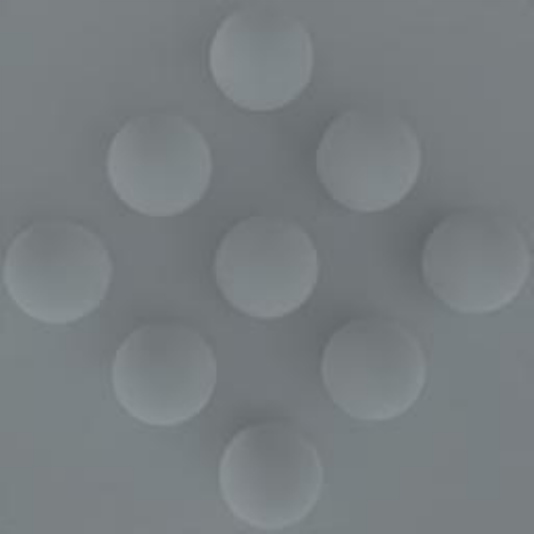}} & 
\noindent\parbox[c]{0.100\textwidth}{\includegraphics[width=0.100\textwidth]{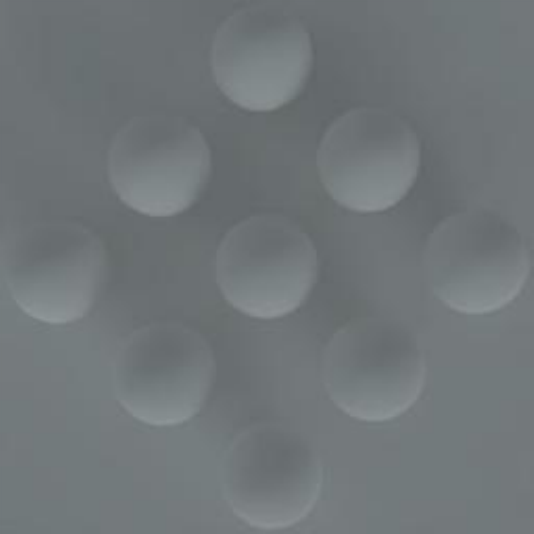}} &
\noindent\parbox[c]{0.200\textwidth}{\includegraphics[width=0.200\textwidth]{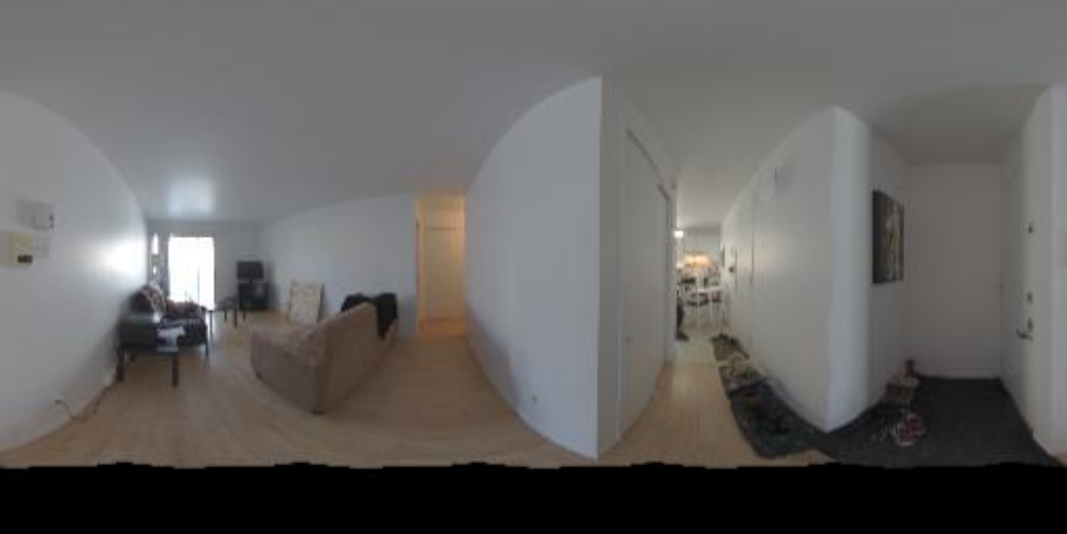}} & 
\noindent\parbox[c]{0.100\textwidth}{\includegraphics[width=0.100\textwidth]{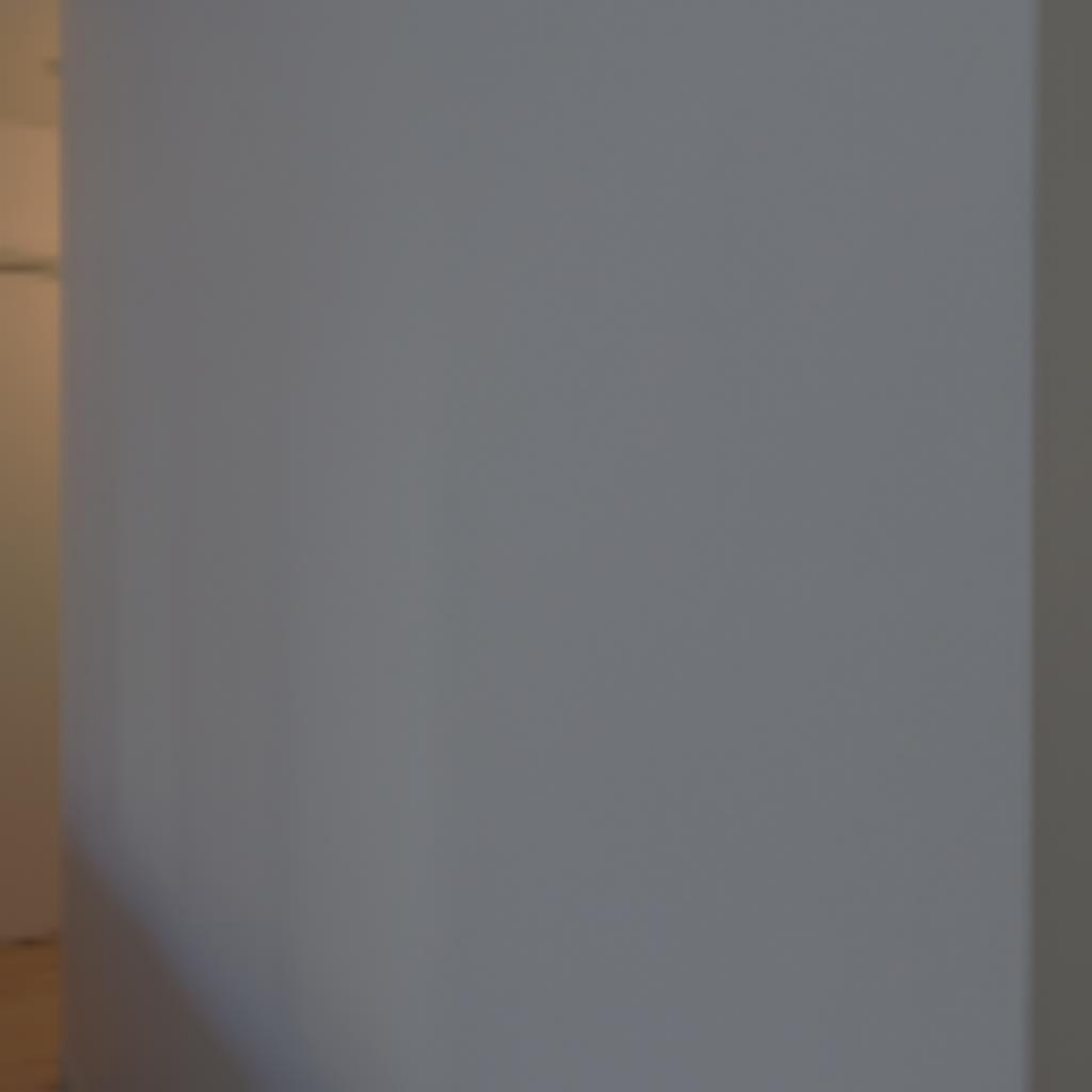}} &  
\noindent\parbox[c]{0.100\textwidth}{\includegraphics[width=0.100\textwidth]{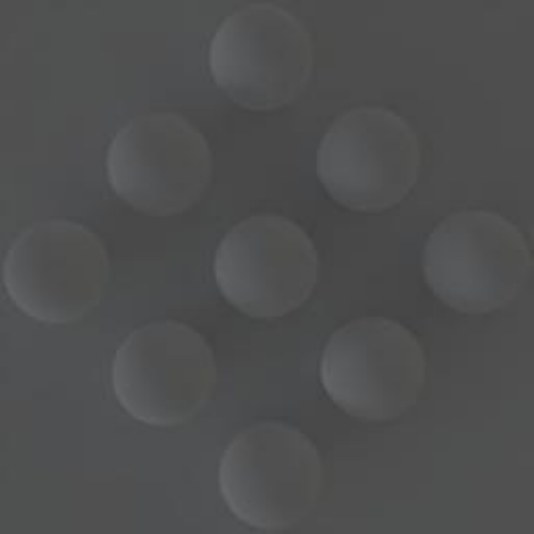}} & 
\noindent\parbox[c]{0.100\textwidth}{\includegraphics[width=0.100\textwidth]{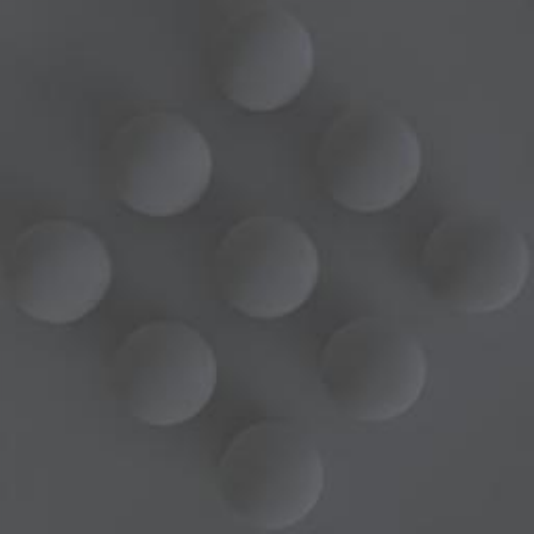}} 
\\

\noindent\parbox[c]{0.200\textwidth}{\includegraphics[width=0.200\textwidth]{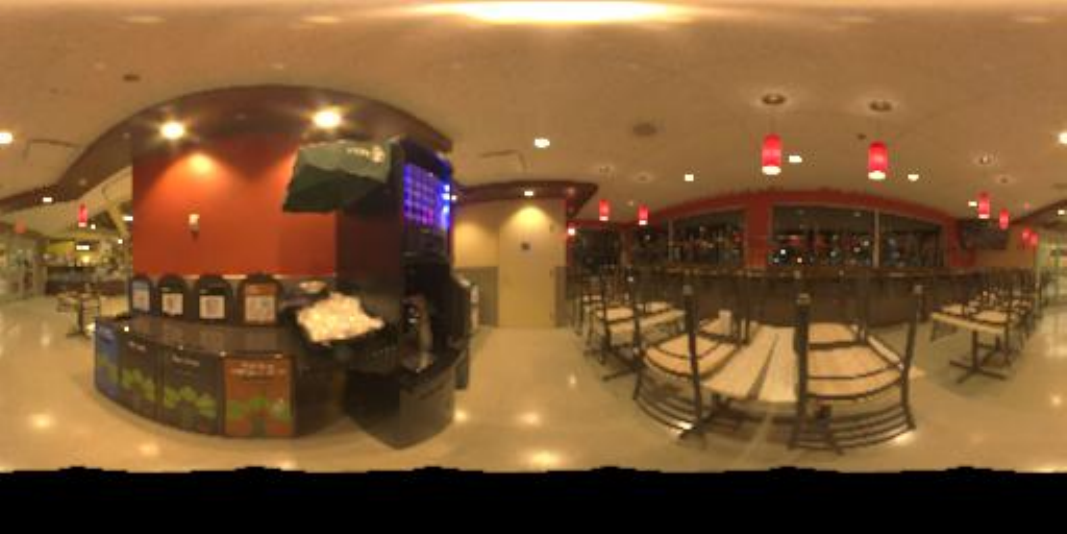}} & 
\noindent\parbox[c]{0.100\textwidth}{\includegraphics[width=0.100\textwidth]{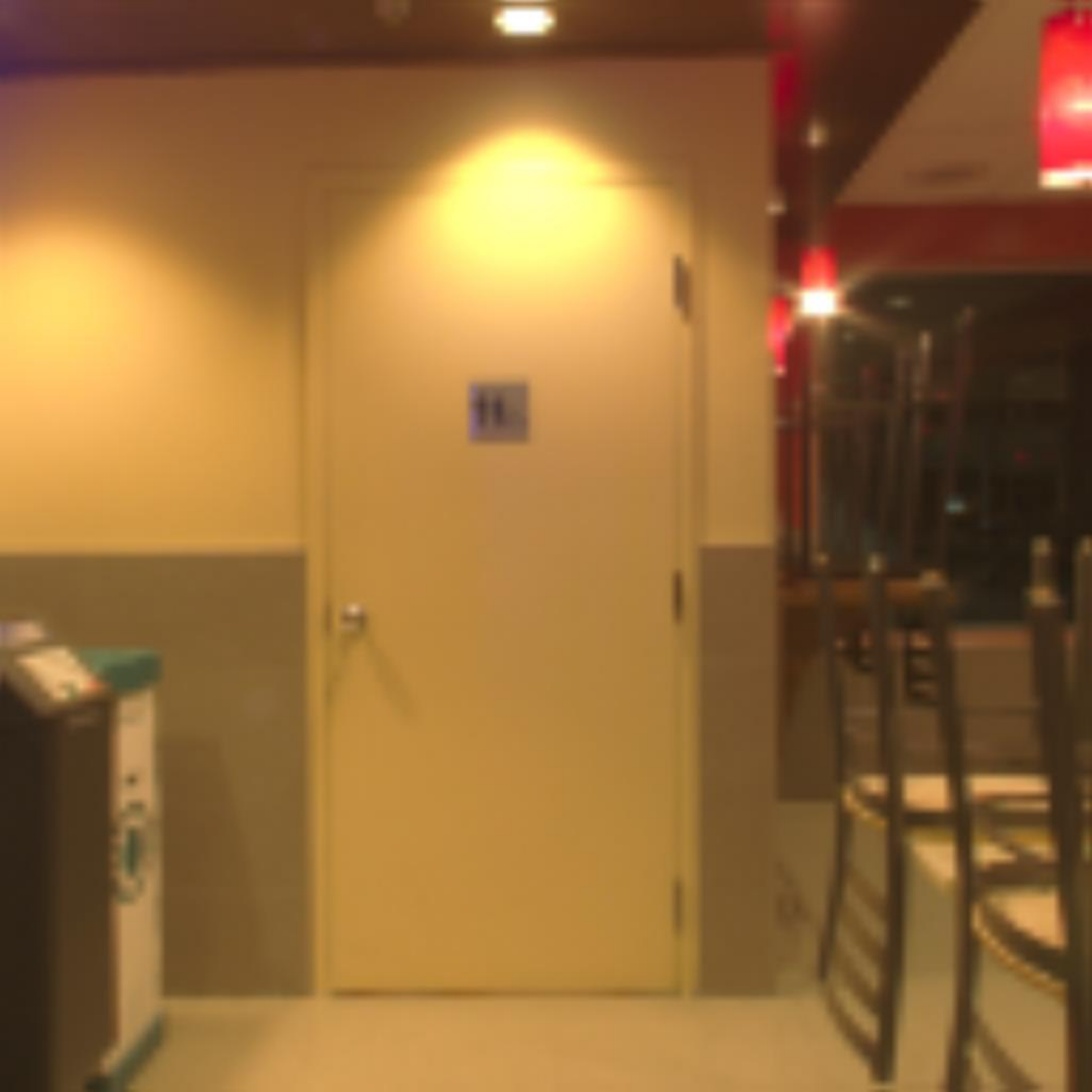}} &  
\noindent\parbox[c]{0.100\textwidth}{\includegraphics[width=0.100\textwidth]{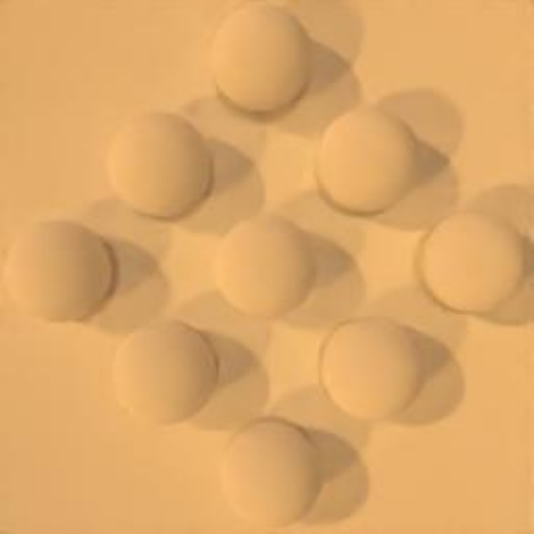}} & 
\noindent\parbox[c]{0.100\textwidth}{\includegraphics[width=0.100\textwidth]{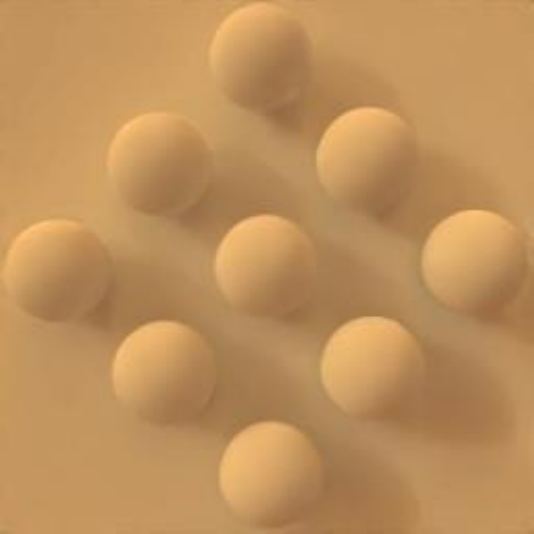}} &
\noindent\parbox[c]{0.200\textwidth}{\includegraphics[width=0.200\textwidth]{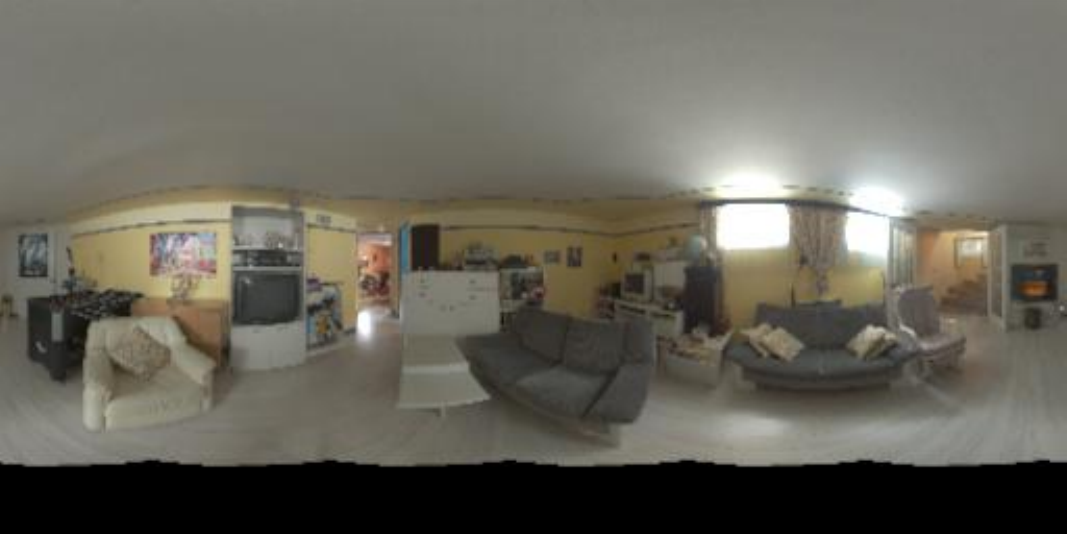}} & 
\noindent\parbox[c]{0.100\textwidth}{\includegraphics[width=0.100\textwidth]{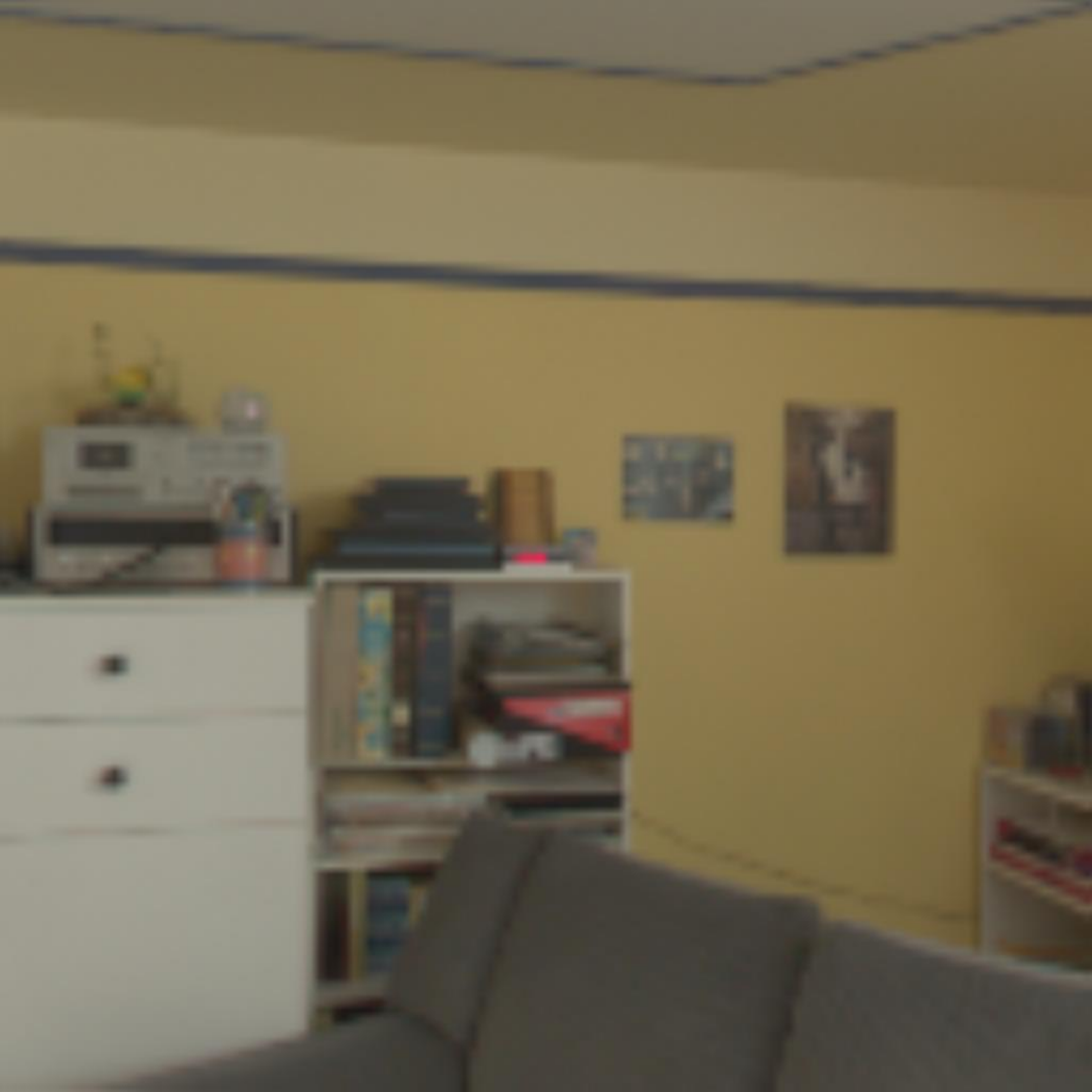}} &  
\noindent\parbox[c]{0.100\textwidth}{\includegraphics[width=0.100\textwidth]{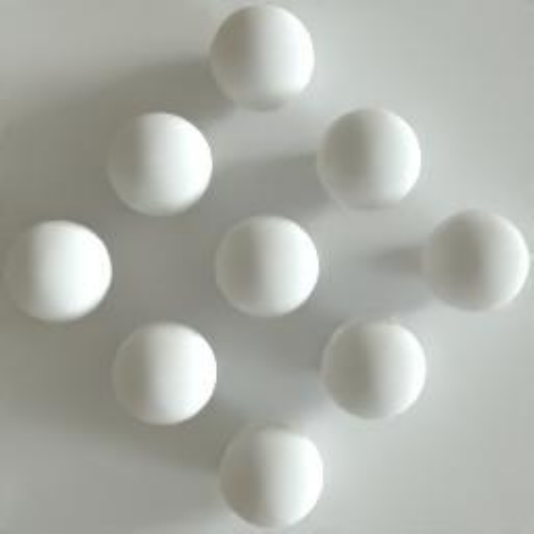}} & 
\noindent\parbox[c]{0.100\textwidth}{\includegraphics[width=0.100\textwidth]{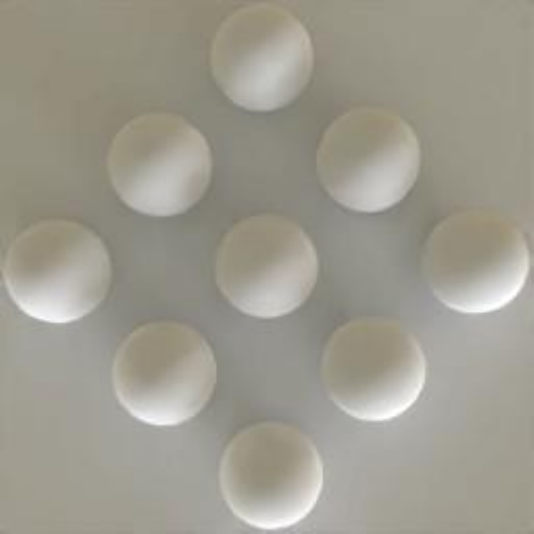}}
\\

\noindent\parbox[c]{0.200\textwidth}{\includegraphics[width=0.200\textwidth]{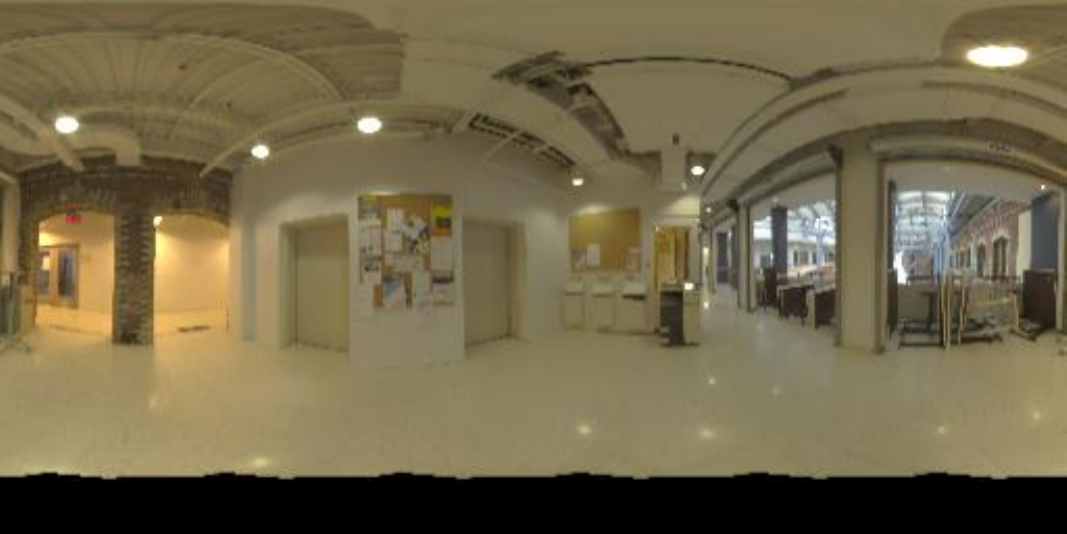}} & 
\noindent\parbox[c]{0.100\textwidth}{\includegraphics[width=0.100\textwidth]{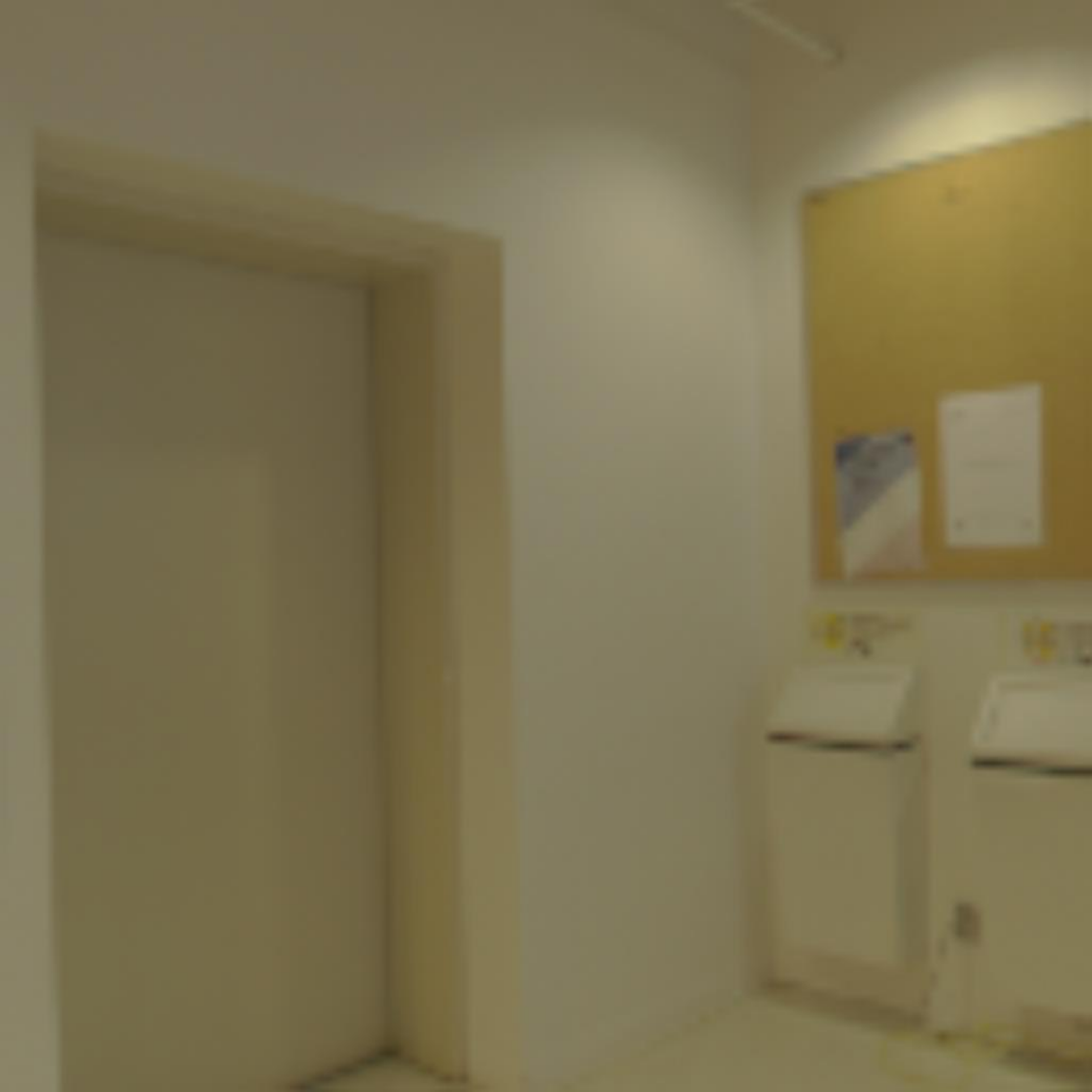}} &  
\noindent\parbox[c]{0.100\textwidth}{\includegraphics[width=0.100\textwidth]{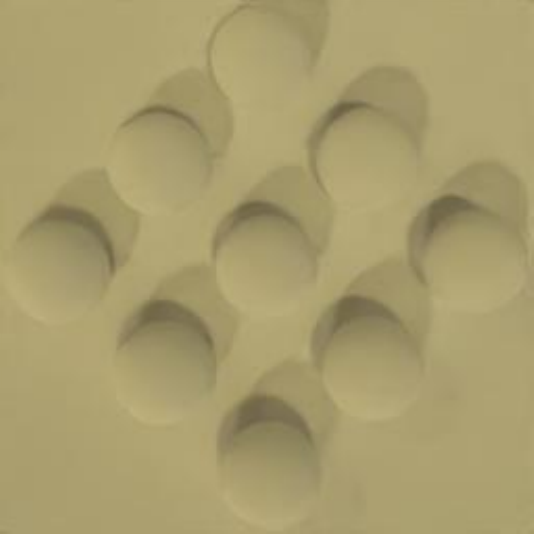}} & 
\noindent\parbox[c]{0.100\textwidth}{\includegraphics[width=0.100\textwidth]{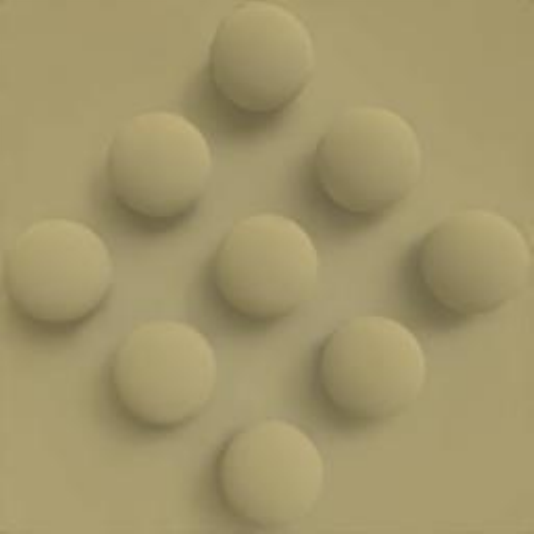}} &
\noindent\parbox[c]{0.200\textwidth}{\includegraphics[width=0.200\textwidth]{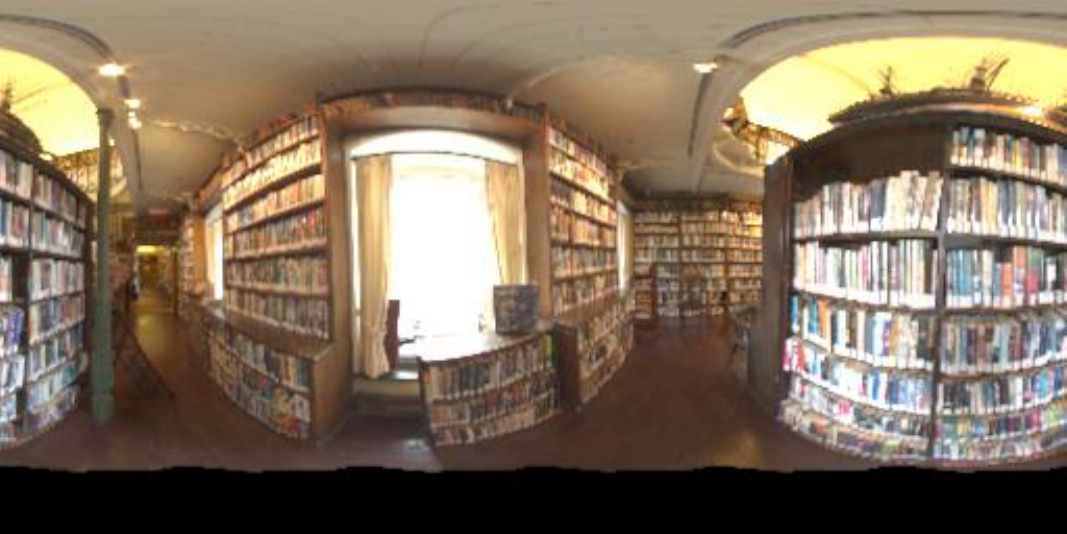}} & 
\noindent\parbox[c]{0.100\textwidth}{\includegraphics[width=0.100\textwidth]{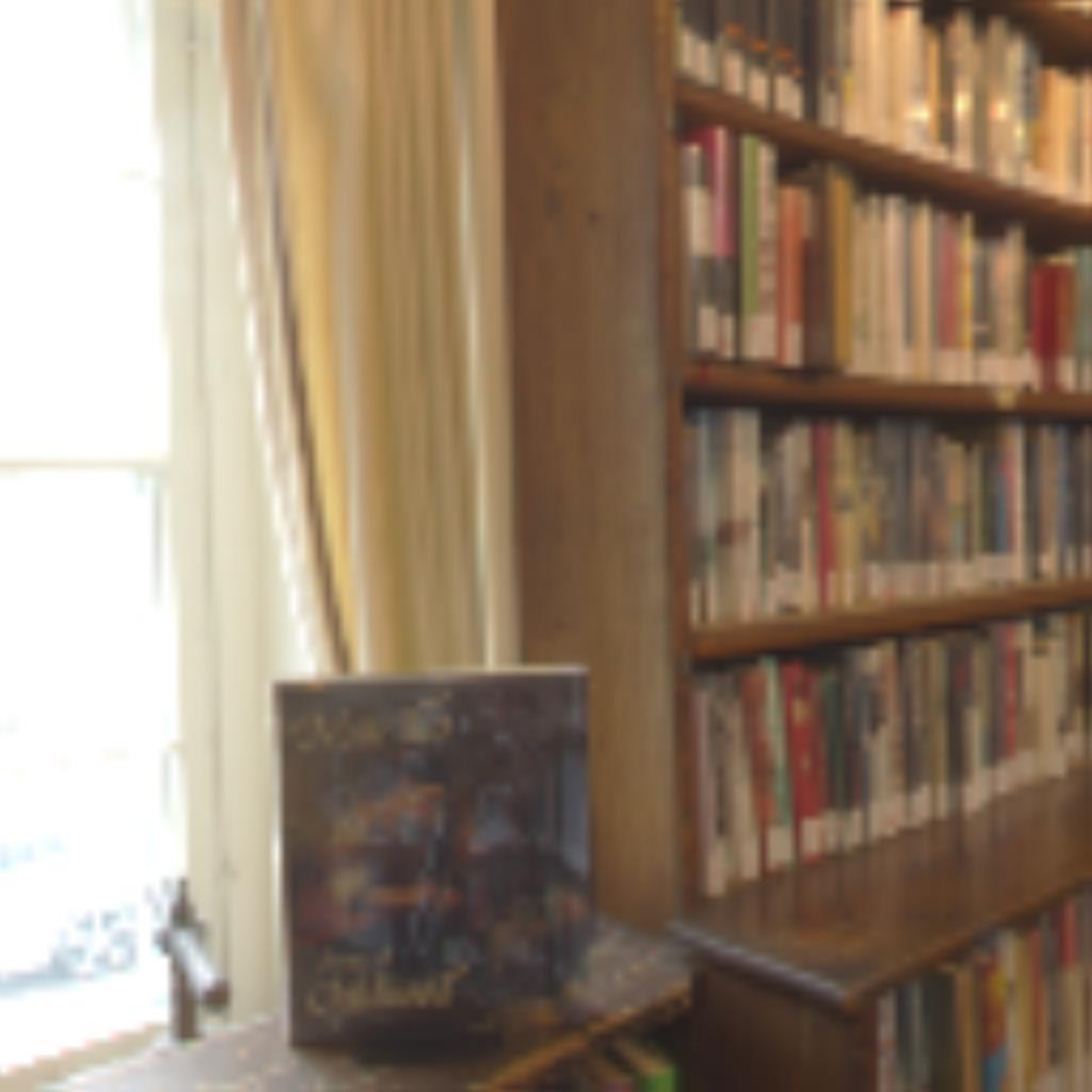}} &  
\noindent\parbox[c]{0.100\textwidth}{\includegraphics[width=0.100\textwidth]{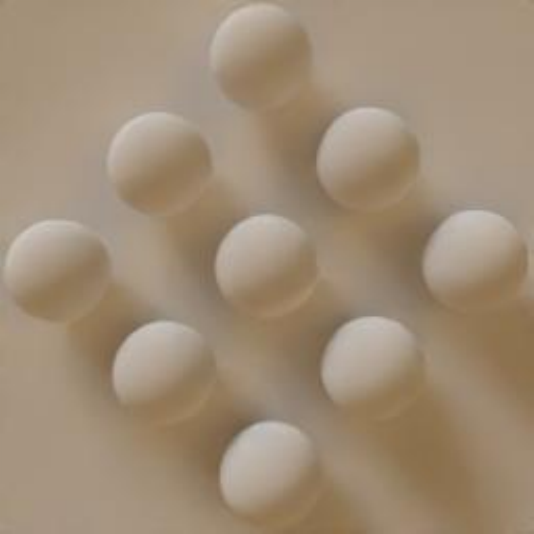}} & 
\noindent\parbox[c]{0.100\textwidth}{\includegraphics[width=0.100\textwidth]{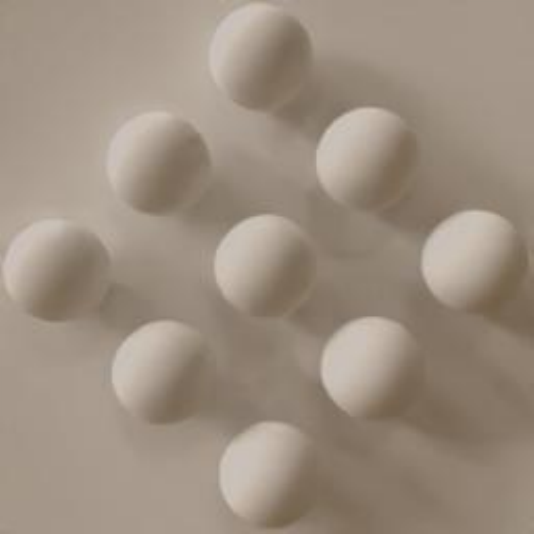}}
\\

\noindent\parbox[c]{0.200\textwidth}{\includegraphics[width=0.200\textwidth]{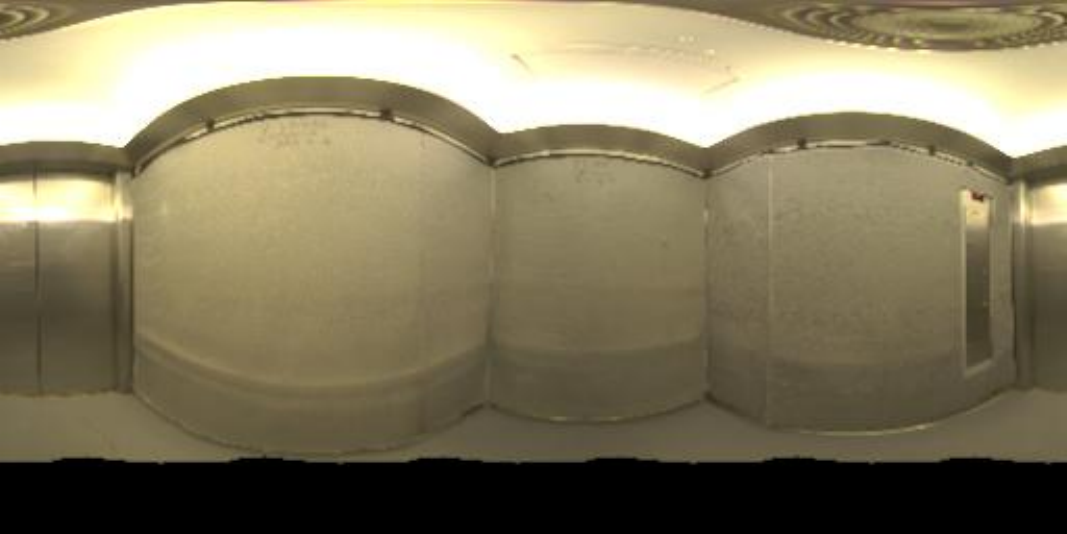}} & 
\noindent\parbox[c]{0.100\textwidth}{\includegraphics[width=0.100\textwidth]{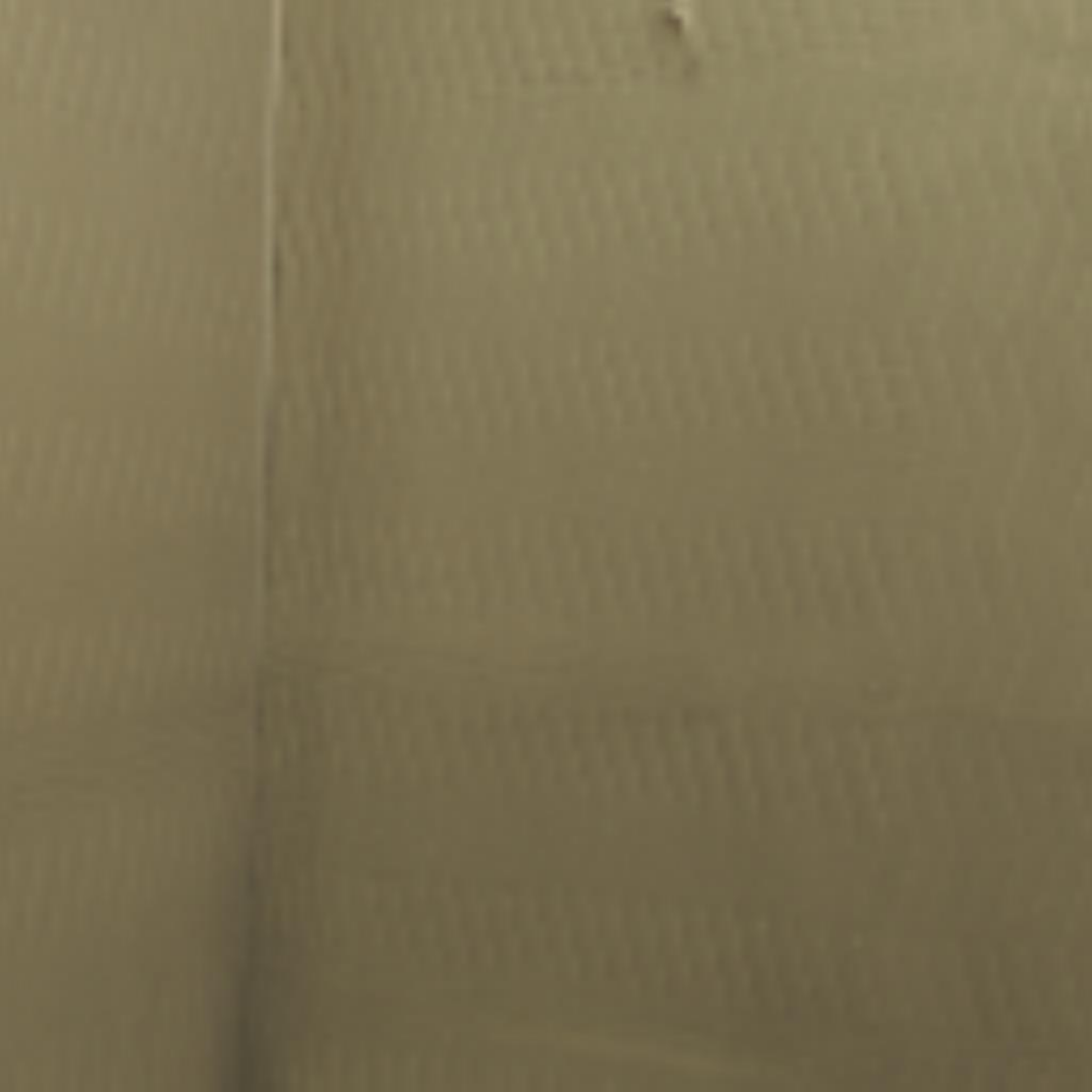}} &  
\noindent\parbox[c]{0.100\textwidth}{\includegraphics[width=0.100\textwidth]{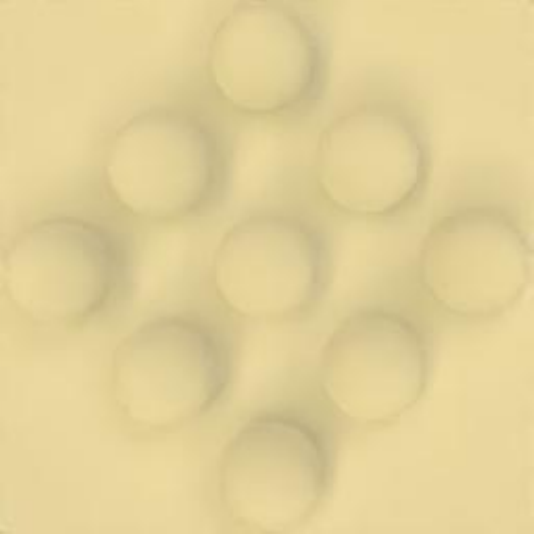}} & 
\noindent\parbox[c]{0.100\textwidth}{\includegraphics[width=0.100\textwidth]{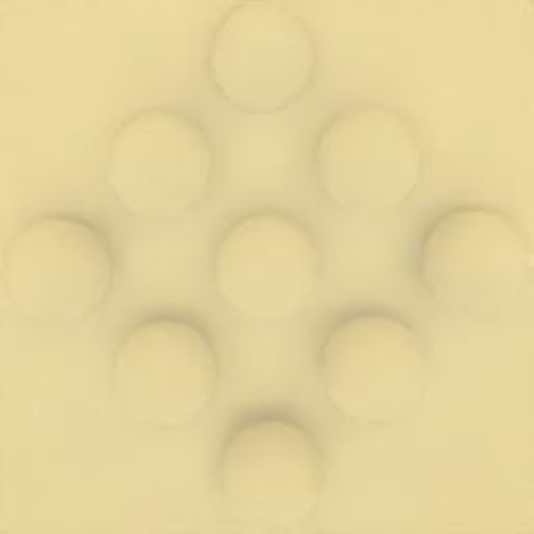}}  &
\noindent\parbox[c]{0.200\textwidth}{\includegraphics[width=0.200\textwidth]{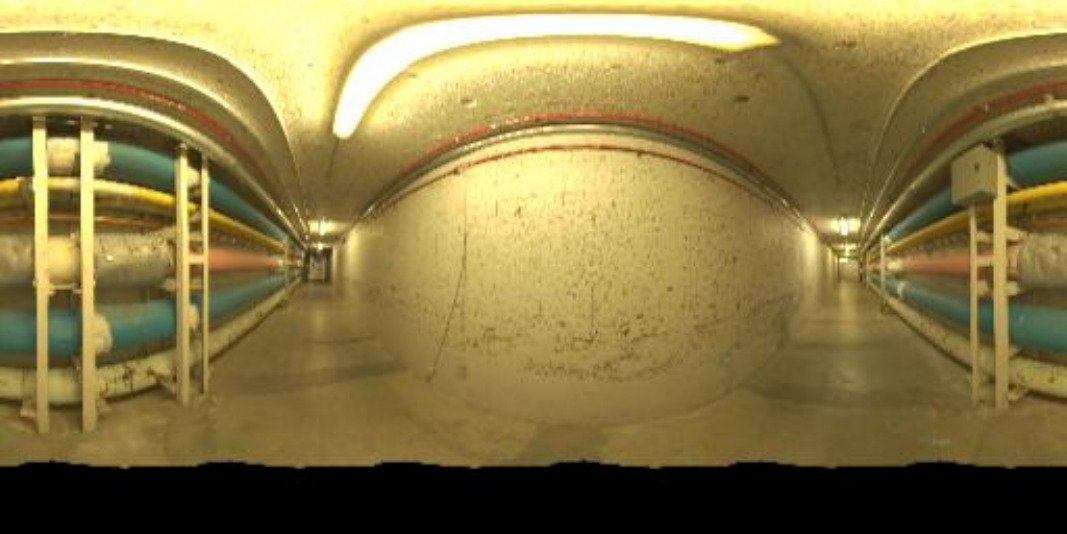}} & 
\noindent\parbox[c]{0.100\textwidth}{\includegraphics[width=0.100\textwidth]{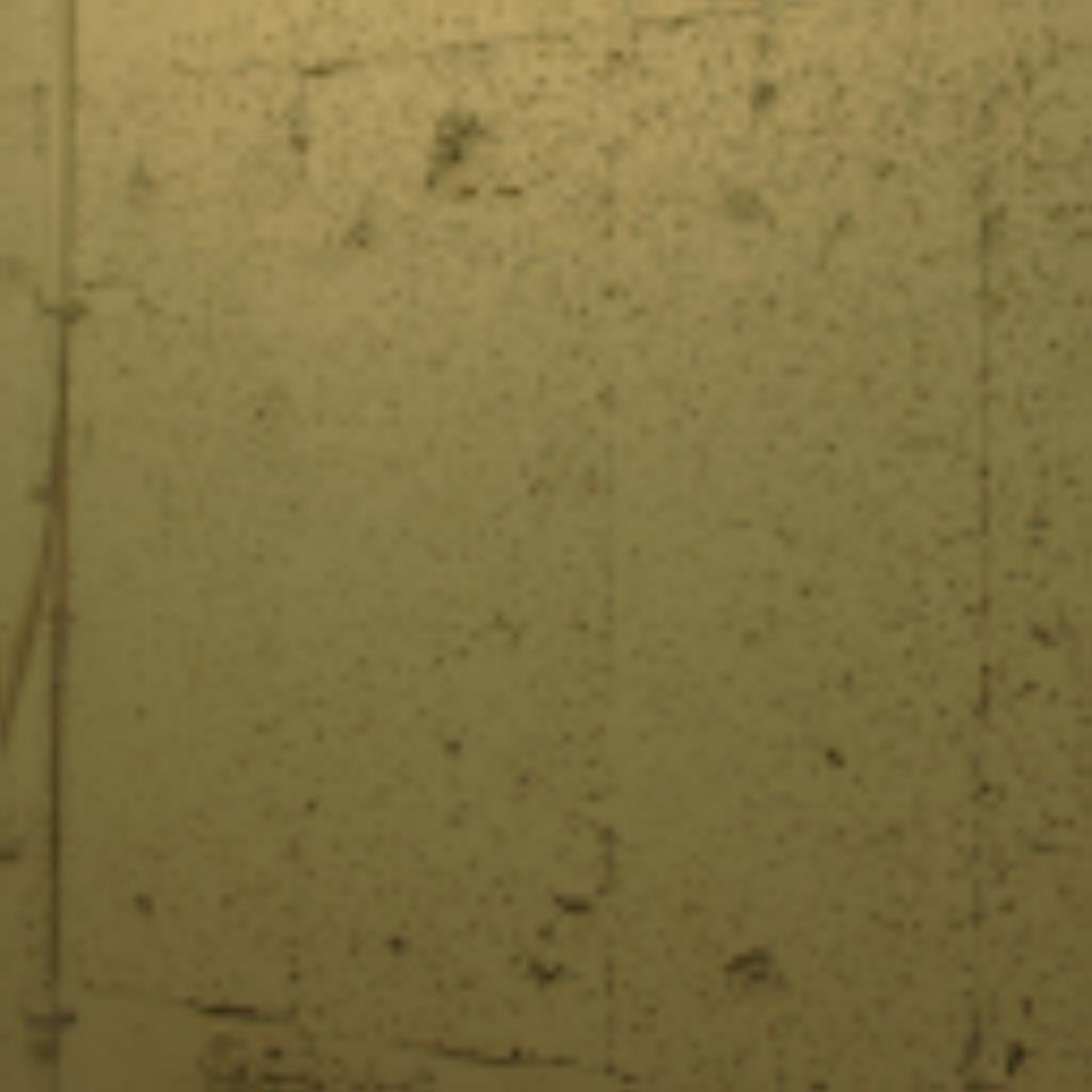}} &  
\noindent\parbox[c]{0.100\textwidth}{\includegraphics[width=0.100\textwidth]{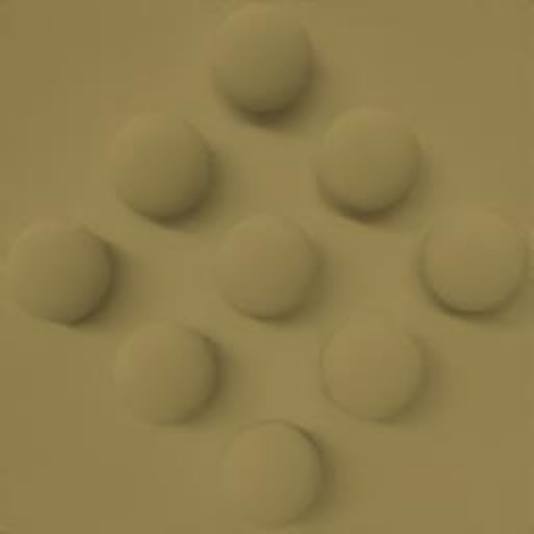}} & 
\noindent\parbox[c]{0.100\textwidth}{\includegraphics[width=0.100\textwidth]{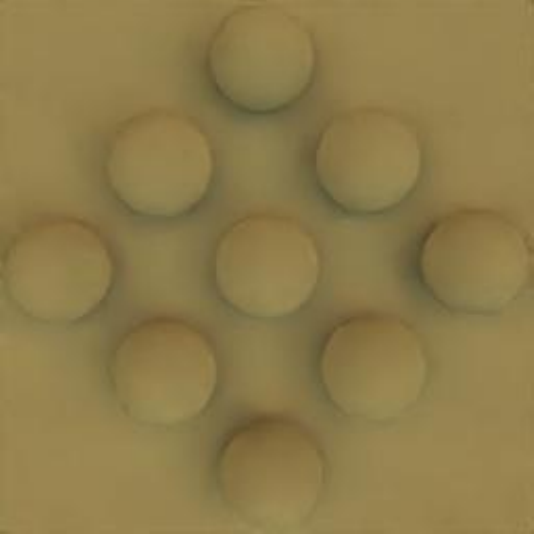}}
\\
\noindent\parbox[c]{0.200\textwidth}{\includegraphics[width=0.200\textwidth]{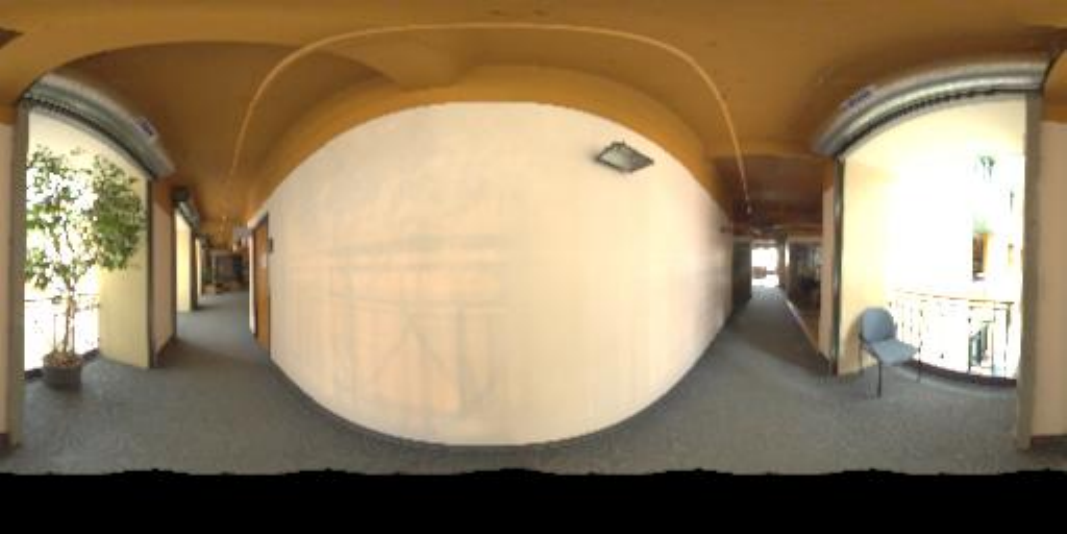}} & 
\noindent\parbox[c]{0.100\textwidth}{\includegraphics[width=0.100\textwidth]{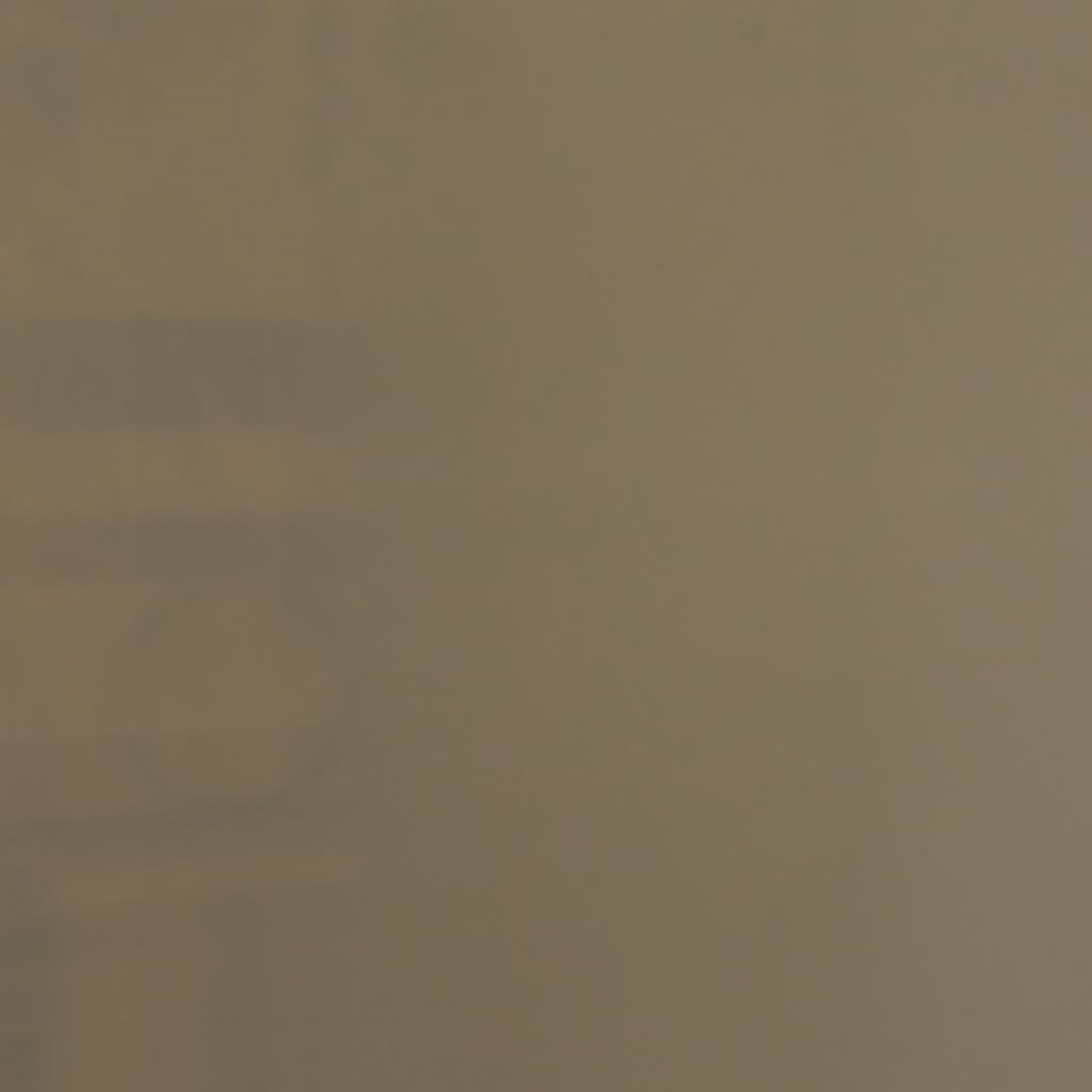}} &  
\noindent\parbox[c]{0.100\textwidth}{\includegraphics[width=0.100\textwidth]{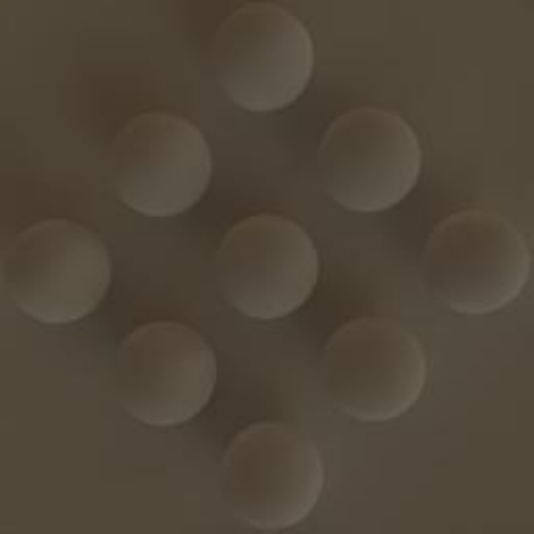}} & 
\noindent\parbox[c]{0.100\textwidth}{\includegraphics[width=0.100\textwidth]{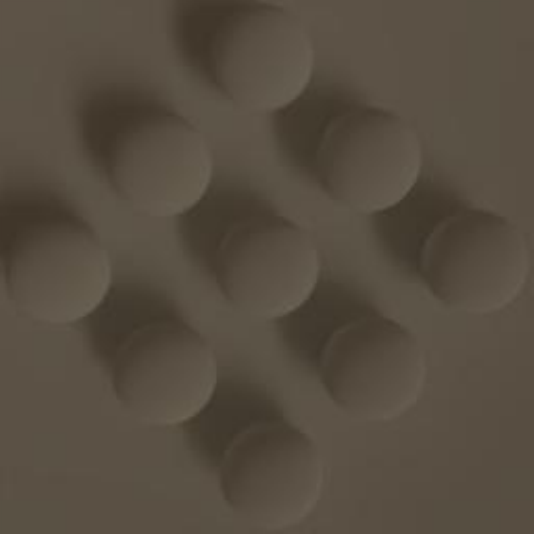}} &
\noindent\parbox[c]{0.200\textwidth}{\includegraphics[width=0.200\textwidth]{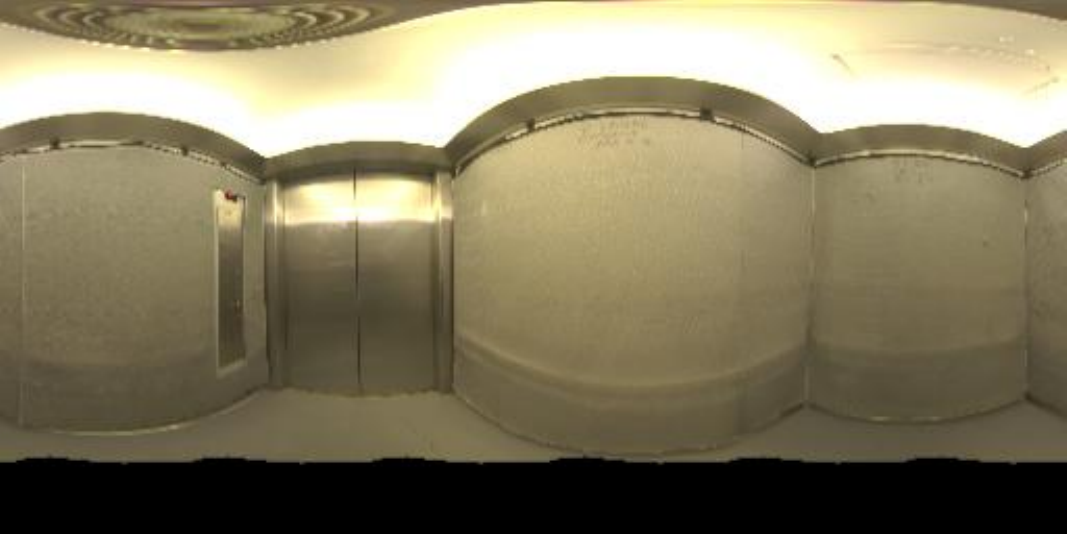}} & 
\noindent\parbox[c]{0.100\textwidth}{\includegraphics[width=0.100\textwidth]{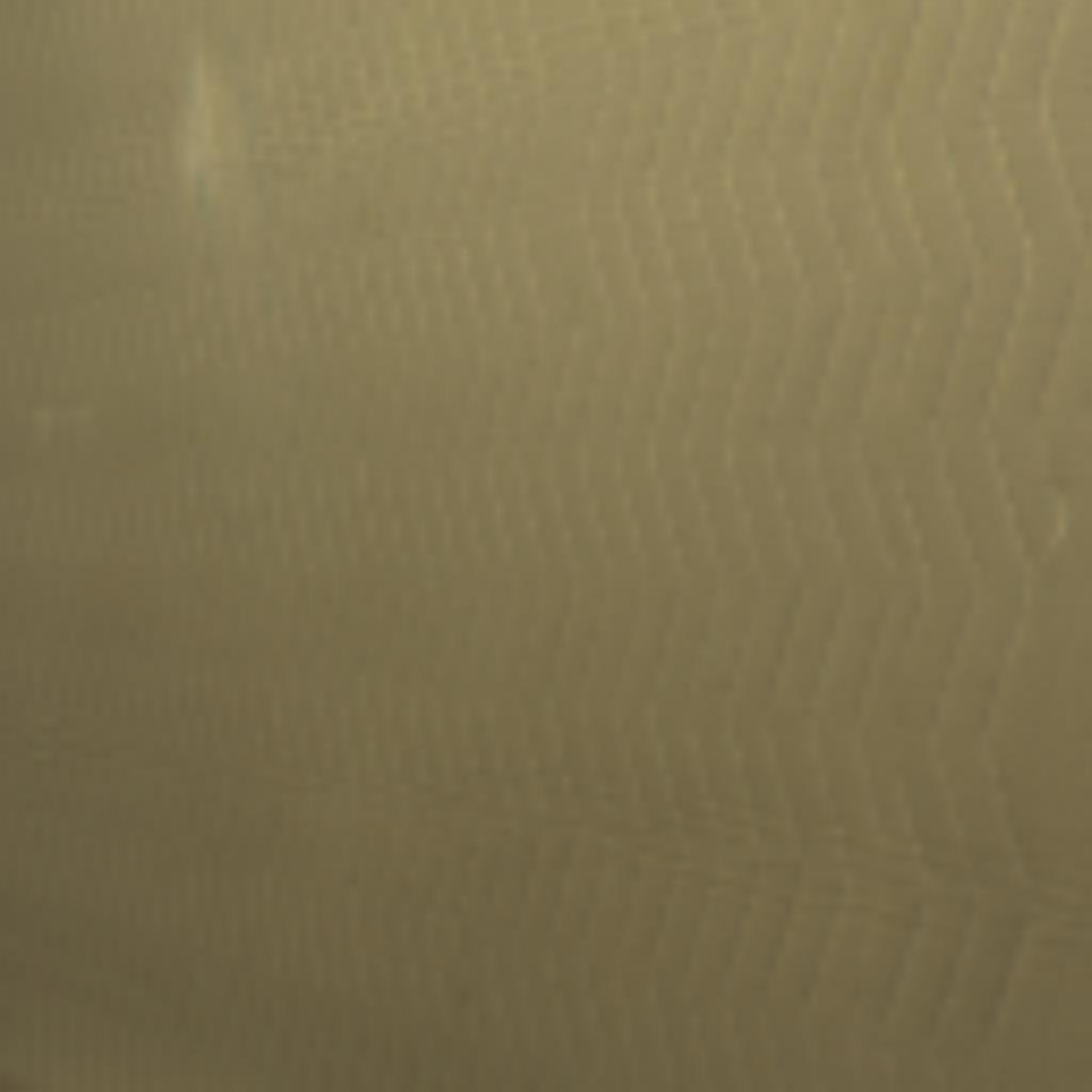}} &  
\noindent\parbox[c]{0.100\textwidth}{\includegraphics[width=0.100\textwidth]{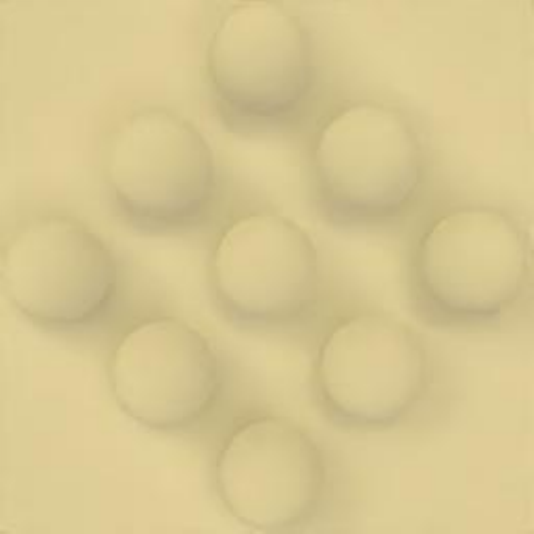}} & 
\noindent\parbox[c]{0.100\textwidth}{\includegraphics[width=0.100\textwidth]{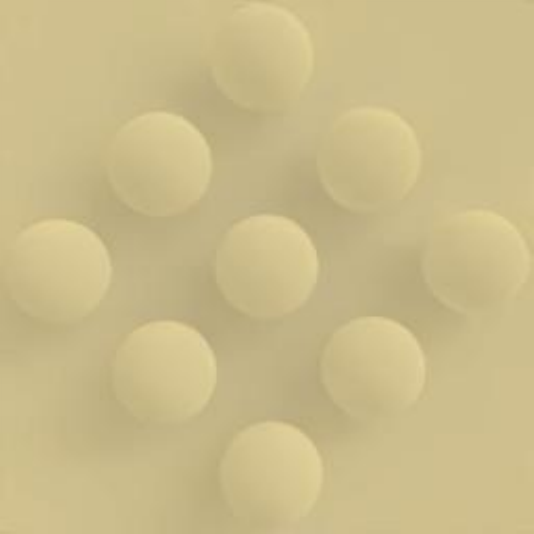}}
\\

        \end{tabu}
    \caption{
    Qualitative results for the Laval indoor dataset using an array of spheres.}
    \label{fig:additional_everlight}
\end{figure*}

\textbf{Evaluation on an array of spheres.} We show renderings of an array of spheres on a plane using our estimated lighting, following the evaluation protocol from Weber et al.  \cite{weber2022editableindoor}. We display 24 random test images from a total of 2,240 test images of the Laval Indoor dataset in Figure \ref{fig:additional_everlight}. Because the scale of the HDR images our method generates and that of the test set are different, for visualization purposes, we scale each output image so that its 0.1$^\text{st}$ and 99.9$^\text{th}$ percentiles of pixel intensity match those of the ground truth. Note that this scaling does not affect the quantitative scores reported in the main paper since the metrics are already scale-invariant.
The last row shows challenging test scenes featuring only plain, solid backgrounds without any shaded objects. Estimating lighting from such input images is highly ill-posed and multi-modal. As a result, visual assessment or evaluations using pixel-based metrics, as used in this protocol, may not be meaningful for such cases.

\subsection{In-the-wild images} \label{appendix:more_result_wild}
\begin{figure*}[!t]
    \centering

    \includegraphics[width=1.0\textwidth]{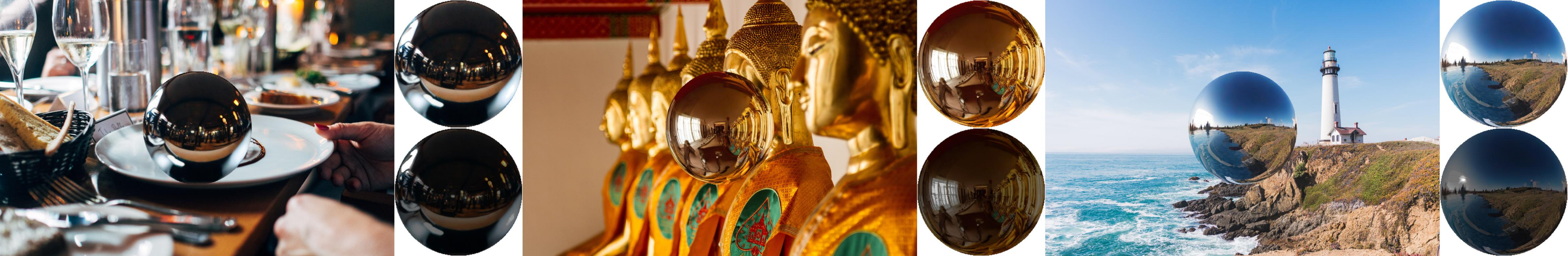}
    
    \smallskip
    \includegraphics[width=1.0\textwidth]{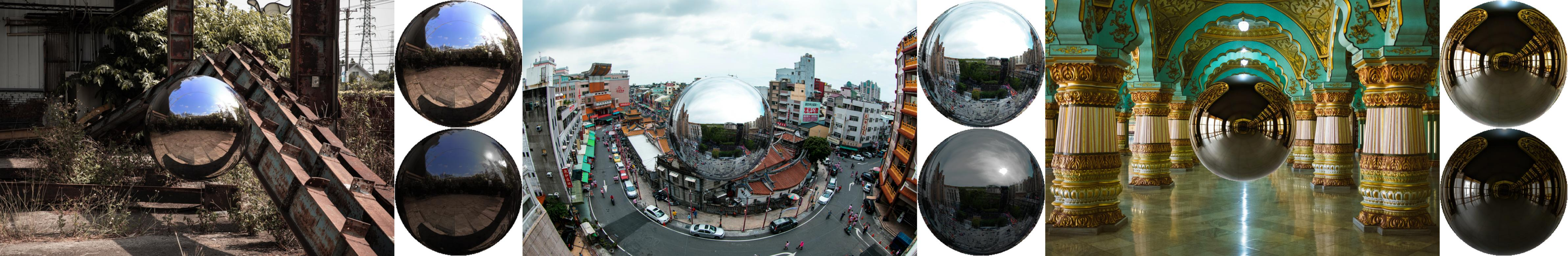}
    
    \smallskip
    \includegraphics[width=1.0\textwidth]{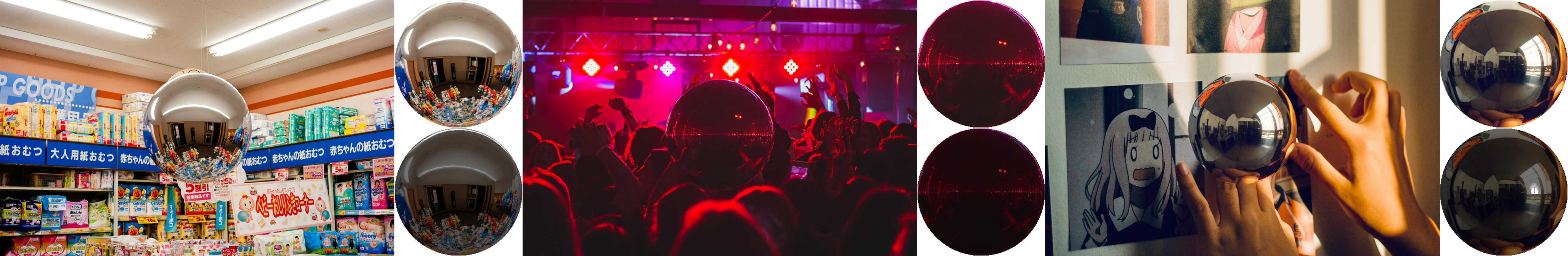}

    \smallskip
    \includegraphics[width=1.0\textwidth]{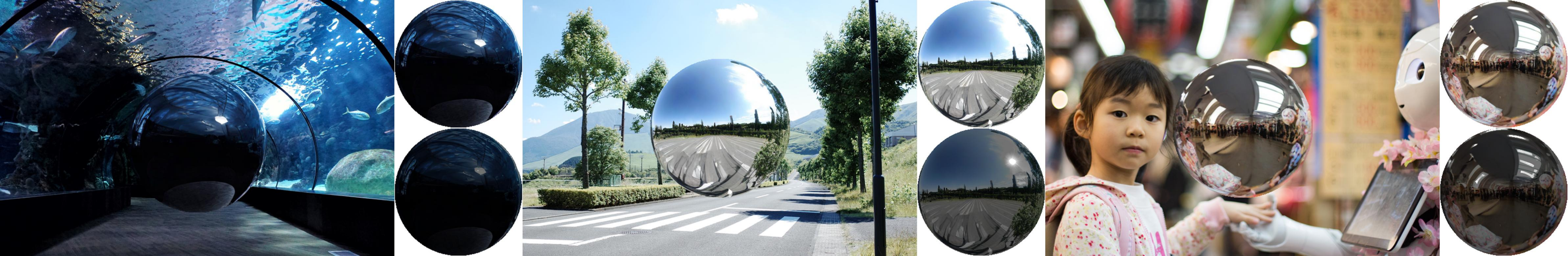}

    \smallskip
    \includegraphics[width=1.0\textwidth]{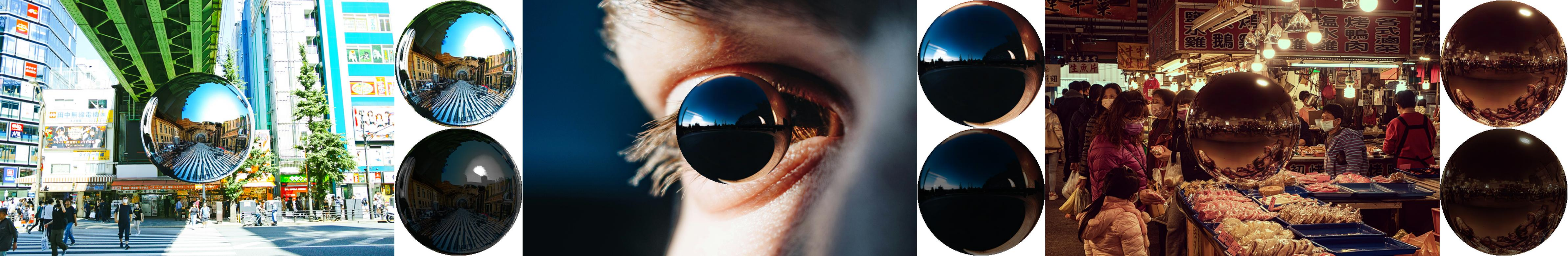}

    \smallskip
    \includegraphics[width=1.0\textwidth]{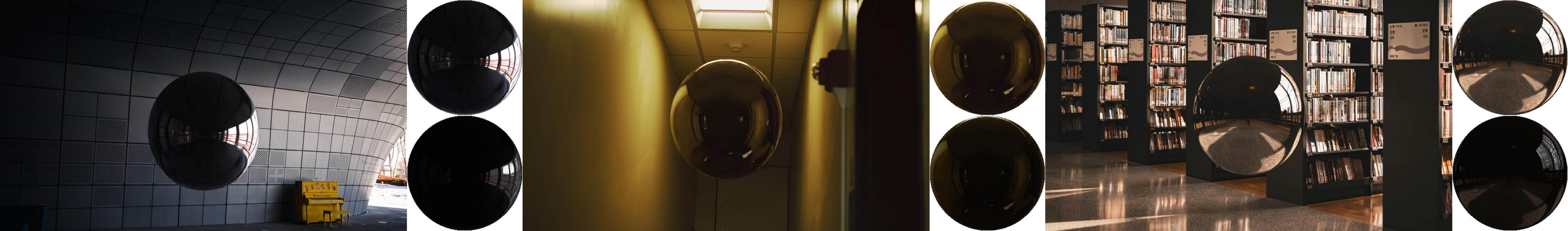}

    \smallskip
    \includegraphics[width=1.0\textwidth]{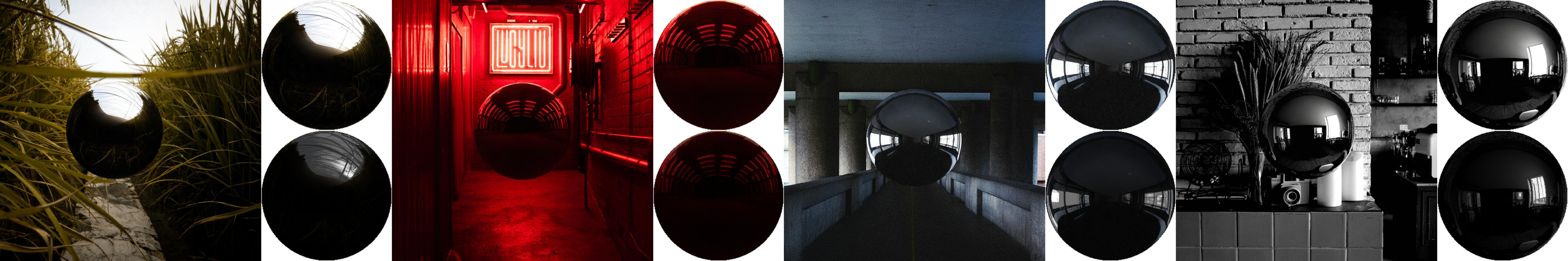}

    \caption{
    Additional qualitative results for in-the-wild scenes. For each input, we show a chrome ball generated from our pipeline and its underexposed version.}

    \label{fig:aba_wild_general}
\end{figure*}

\begin{figure*}[!t]
    \centering

    \includegraphics[width=1.0\textwidth]{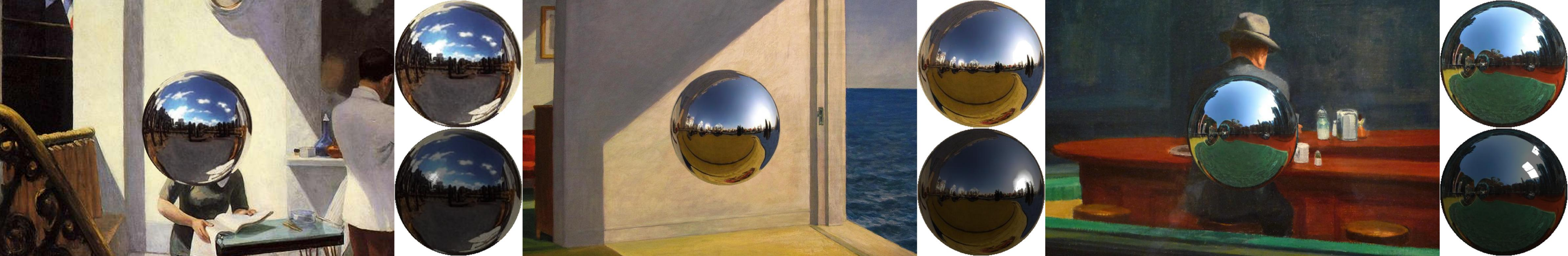}

    \smallskip
    \includegraphics[width=1.0\textwidth]{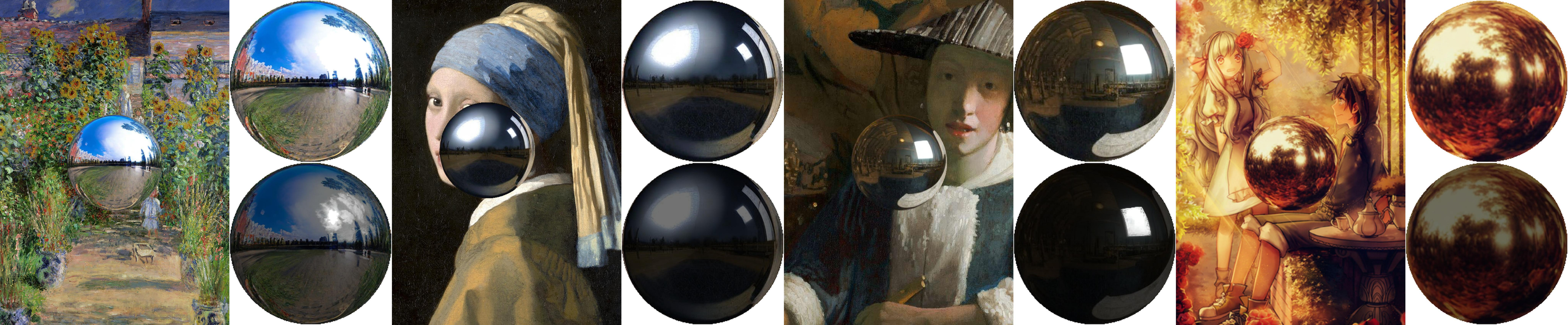}

    \smallskip
    \includegraphics[width=1.0\textwidth]{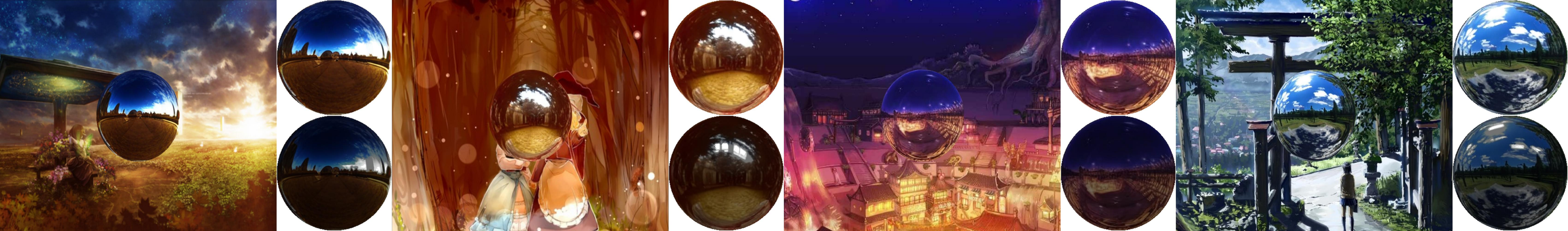}

    \caption{
    Qualitative results for artificial images such as paintings and painting-like Japanese animation-style images. For each input, we show a chrome ball generated from our pipeline and its underexposed version. Our proposed method can still perform reasonably well, albeit with some performance degradation, by leveraging the strong generative prior of SDXL \cite{podell2023sdxl}.}

    \label{fig:aba_wild_painting}
\end{figure*}

We present additional qualitative results for in-the-wild scenes in Figure \ref{fig:aba_wild_general}. Our method produces high-quality chrome balls that harmonize well with diverse scenes and lighting environments, such as a dim hallway under red neon lighting, an underwater tunnel with blue-tinted sunlight, a close-up shot of food, an outdoor view by a coastline, and a bird's-eye view from a tall building. Our method also works on non-realistic images, such as paintings or cartoons, where visual cues like shading and shadows are present (see Figure \ref{fig:aba_wild_painting}).


\section{Stochastic vs Deterministic Sampling}


In Section \ref{sec:median_algo}, we discuss the relationship between initial noise maps and semantic patterns of chrome balls. This mapping is deterministic only when using samplers derived from probability flow ODE, such as DDIM \cite{song2020denoising} and UniPC \cite{zhao2023unipc}. The ``disco'' noise map would less consistently produce ``disco'' balls if we adopted stochastic samplers such as DDPM \cite{ho2020denoising}, which introduce noise during the sampling process. This section investigates whether degeneration of chrome balls is caused by deterministic sampling and whether stochastic sampling can help mitigate this issue.   

\begin{figure}[!t]
    \centering

        \begin{tabu} to \textwidth {
        @{}
        c@{\hspace{0.5pt}}
        c@{\hspace{0.5pt}}
        c@{\hspace{0.5pt}}
        c@{\hspace{0.5pt}}
        c@{\hspace{0.5pt}}
        c@{\hspace{0.5pt}}
        c@{\hspace{0.5pt}}
    }

        \multicolumn{1}{c}{\shortstack{\scriptsize Input}} & 
        \multicolumn{1}{c}{\shortstack{\scriptsize 10 steps}} &
        \multicolumn{1}{c}{\shortstack{\scriptsize 20 steps}} & 
        \multicolumn{1}{c}{\shortstack{\scriptsize 40 steps}} & 
        \multicolumn{1}{c}{\shortstack{\scriptsize 80 steps}} &
        \multicolumn{1}{c}{\shortstack{\scriptsize 160 steps}} &
        \\

        \noindent\parbox[c]{0.11\textwidth}{\includegraphics[width=0.11\textwidth]{storage/noise_appearance_relationship_whiteBG/input/airport_seed150.pdf}} & 
        \noindent\parbox[c]{0.071\textwidth}{\includegraphics[width=0.071\textwidth]{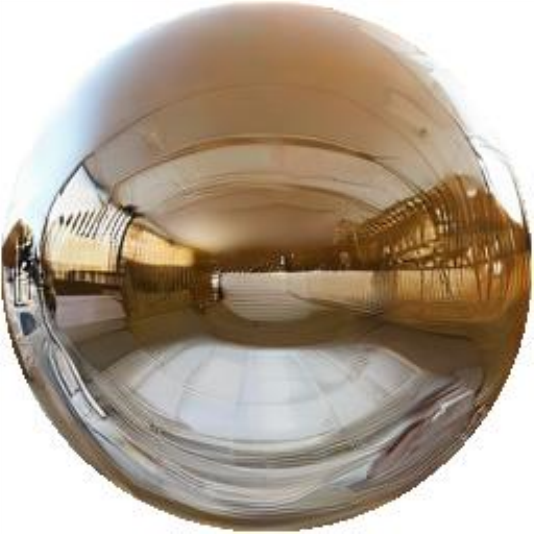}} &  
        \noindent\parbox[c]{0.071\textwidth}{\includegraphics[width=0.071\textwidth]{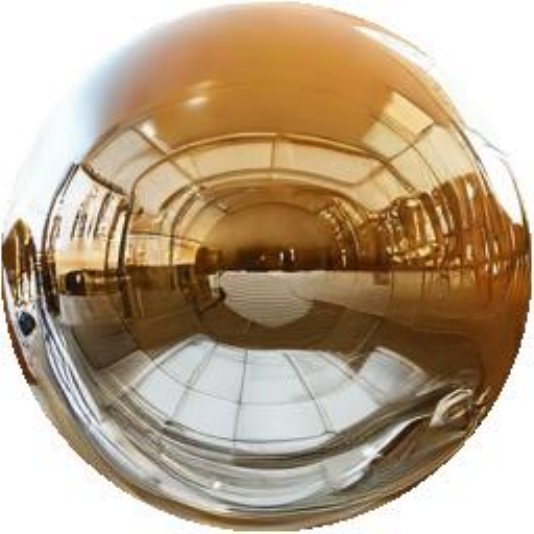}} &  
        \noindent\parbox[c]{0.071\textwidth}{\includegraphics[width=0.071\textwidth]{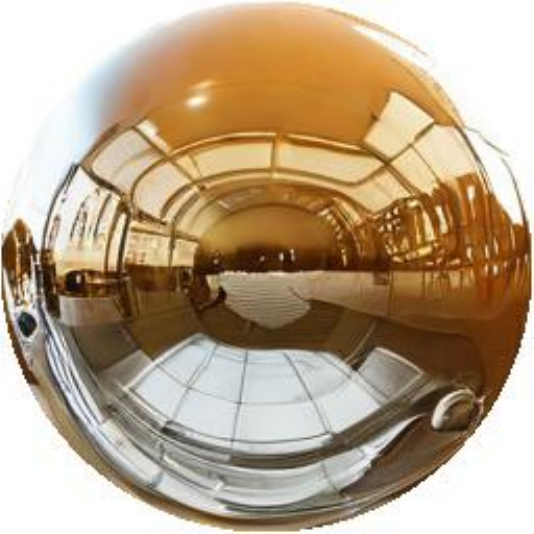}} &  
        \noindent\parbox[c]{0.071\textwidth}{\includegraphics[width=0.071\textwidth]{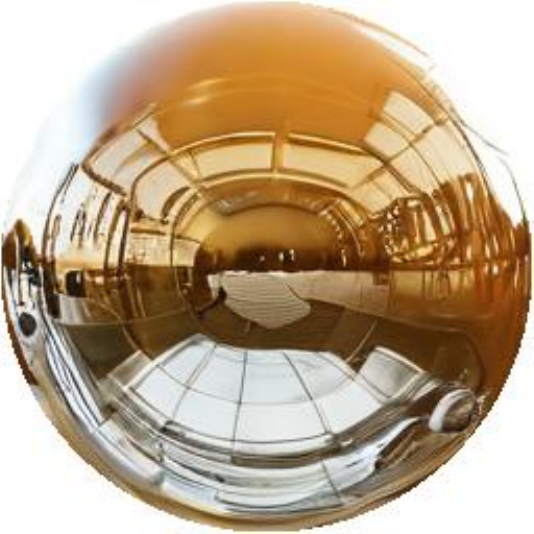}} &  
        \noindent\parbox[c]{0.071\textwidth}{\includegraphics[width=0.071\textwidth]{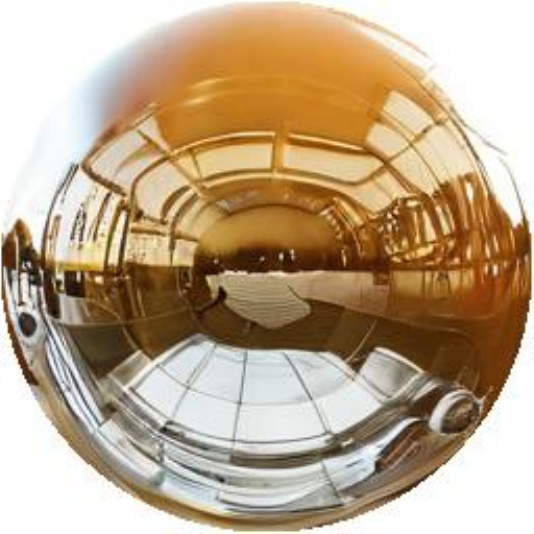}}
        \\

        & 
        \noindent\parbox[c]{0.071\textwidth}{\includegraphics[width=0.071\textwidth]{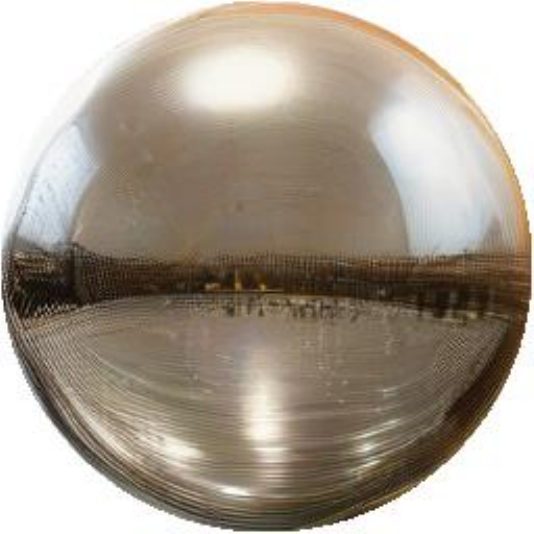}} &  
        \noindent\parbox[c]{0.071\textwidth}{\includegraphics[width=0.071\textwidth]{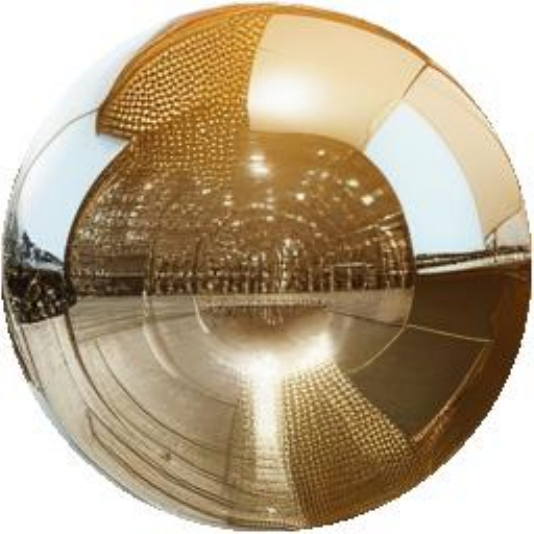}} &  
        \noindent\parbox[c]{0.071\textwidth}{\includegraphics[width=0.071\textwidth]{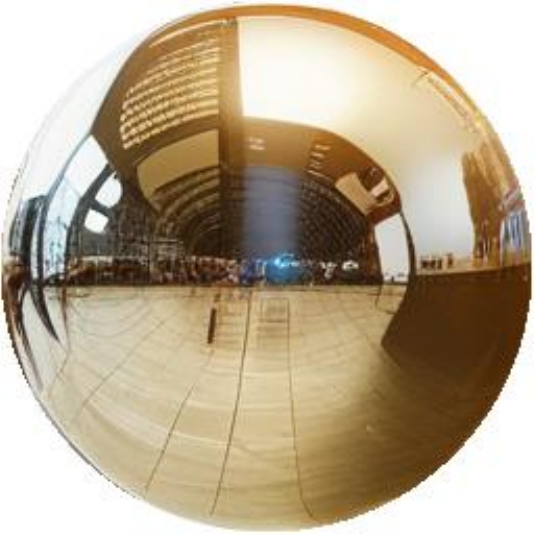}} &  
        \noindent\parbox[c]{0.071\textwidth}{\includegraphics[width=0.071\textwidth]{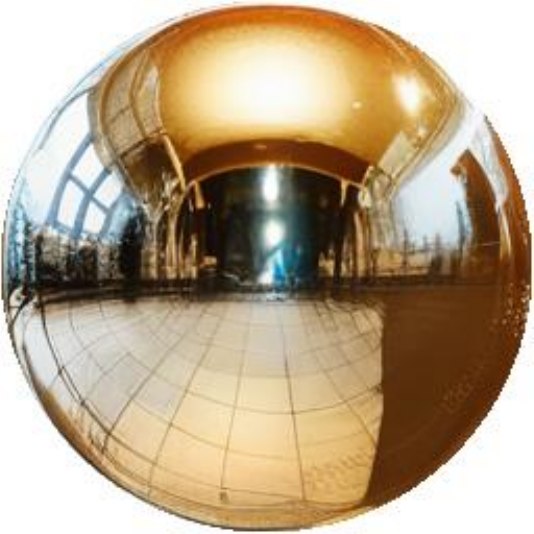}} &  
        \noindent\parbox[c]{0.071\textwidth}{\includegraphics[width=0.071\textwidth]{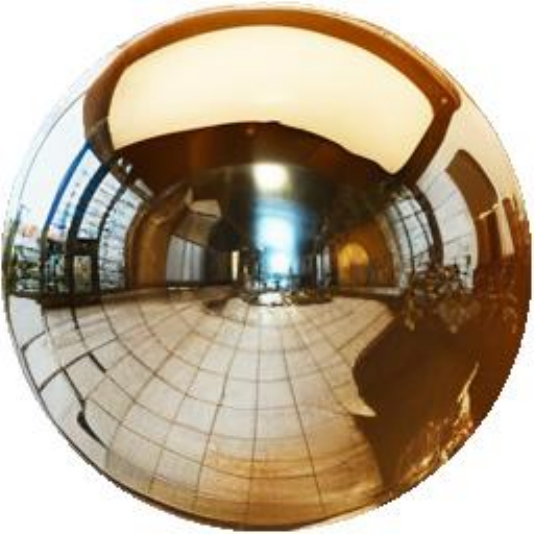}}
        \\

        \noindent\parbox[c]{0.11\textwidth}{\includegraphics[width=0.11\textwidth]{storage/noise_appearance_relationship_whiteBG/input/castle3_seed150.pdf}} & 
        \noindent\parbox[c]{0.071\textwidth}{\includegraphics[width=0.071\textwidth]{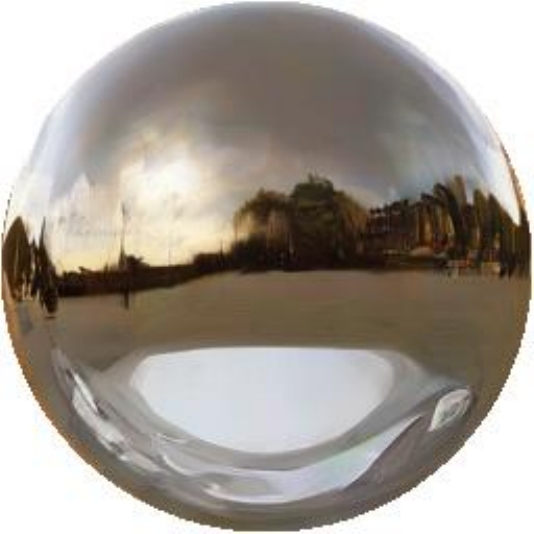}} &  
        \noindent\parbox[c]{0.071\textwidth}{\includegraphics[width=0.071\textwidth]{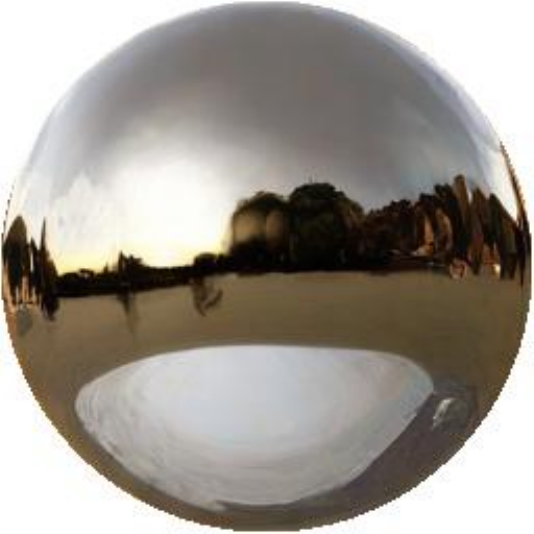}} &  
        \noindent\parbox[c]{0.071\textwidth}{\includegraphics[width=0.071\textwidth]{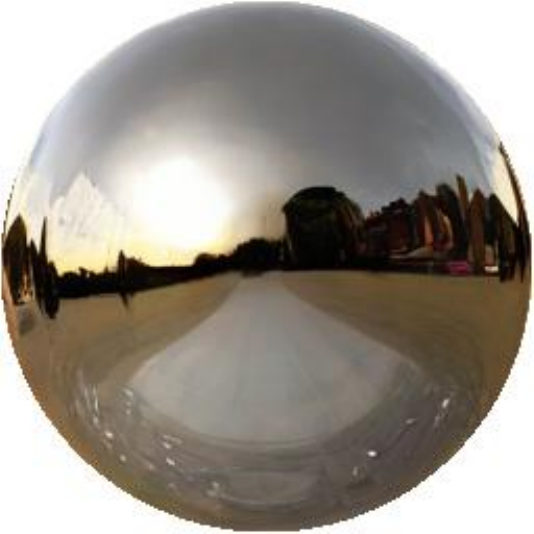}} &  
        \noindent\parbox[c]{0.071\textwidth}{\includegraphics[width=0.071\textwidth]{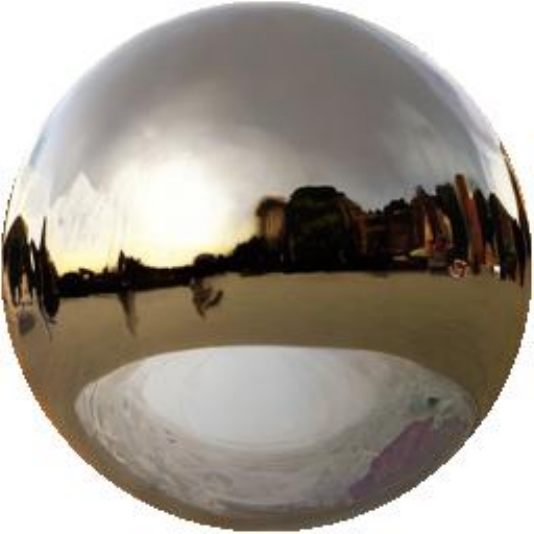}} &  
        \noindent\parbox[c]{0.071\textwidth}{\includegraphics[width=0.071\textwidth]{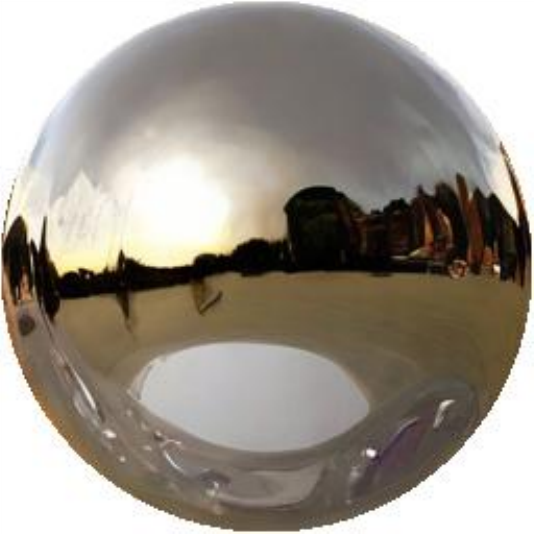}}
        \\

        & 
        \noindent\parbox[c]{0.071\textwidth}{\includegraphics[width=0.071\textwidth]{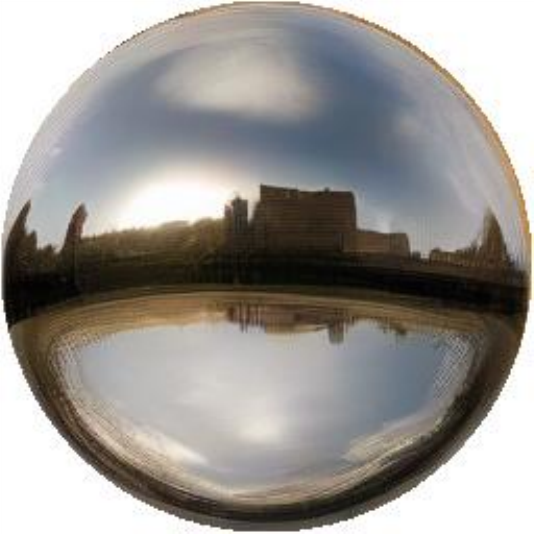}} &  
        \noindent\parbox[c]{0.071\textwidth}{\includegraphics[width=0.071\textwidth]{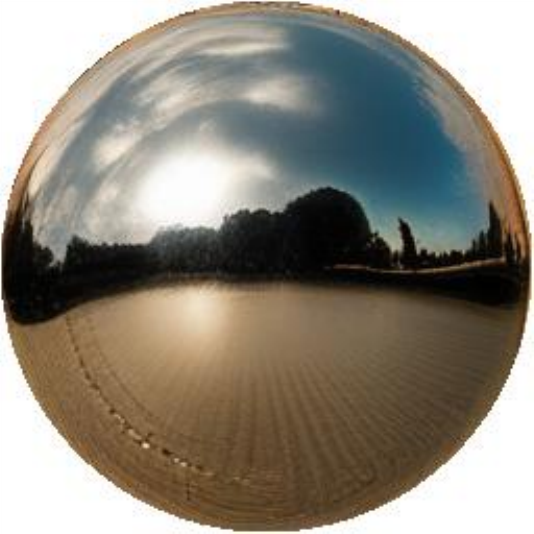}} &  
        \noindent\parbox[c]{0.071\textwidth}{\includegraphics[width=0.071\textwidth]{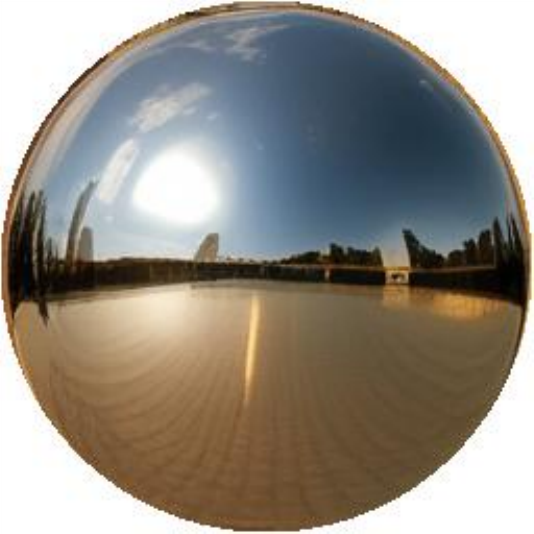}} &  
        \noindent\parbox[c]{0.071\textwidth}{\includegraphics[width=0.071\textwidth]{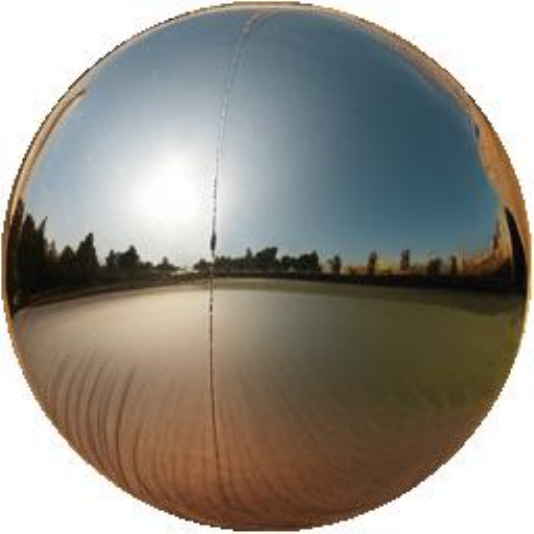}} &  
        \noindent\parbox[c]{0.071\textwidth}{\includegraphics[width=0.071\textwidth]{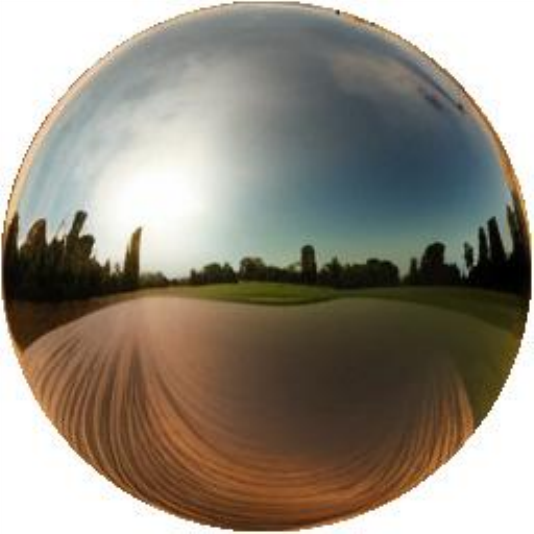}}
        \\

        \noindent\parbox[c]{0.11\textwidth}{\includegraphics[width=0.11\textwidth]{storage/noise_appearance_relationship_whiteBG/input/panda_seed150.pdf}} & 
        \noindent\parbox[c]{0.071\textwidth}{\includegraphics[width=0.071\textwidth]{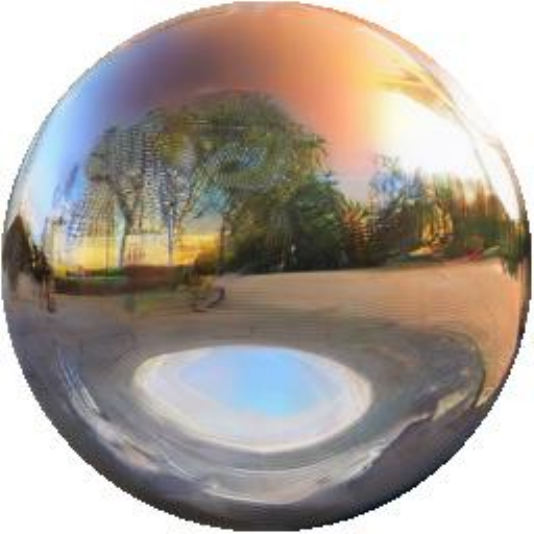}} &  
        \noindent\parbox[c]{0.071\textwidth}{\includegraphics[width=0.071\textwidth]{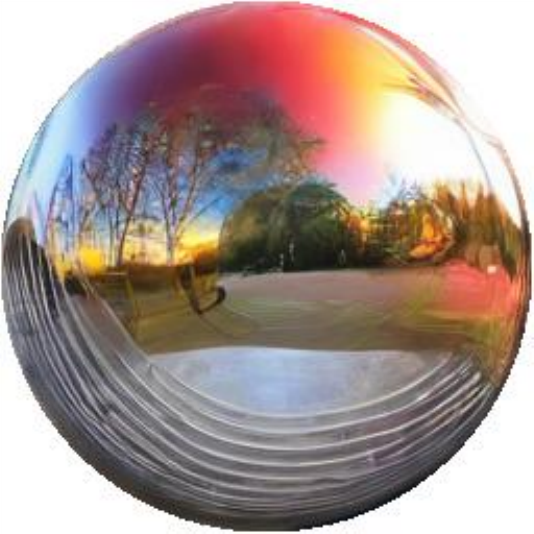}} &  
        \noindent\parbox[c]{0.071\textwidth}{\includegraphics[width=0.071\textwidth]{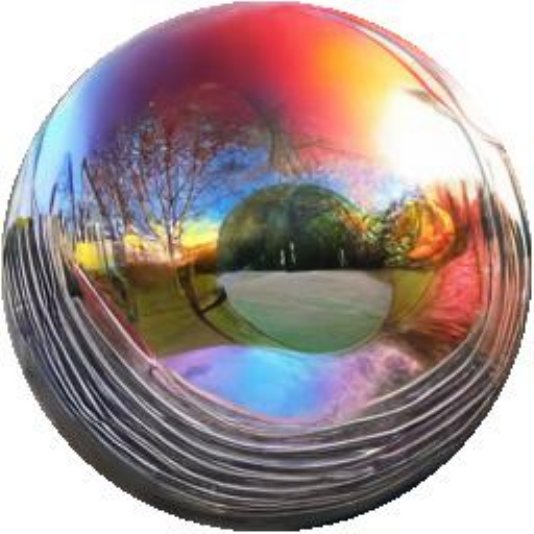}} &  
        \noindent\parbox[c]{0.071\textwidth}{\includegraphics[width=0.071\textwidth]{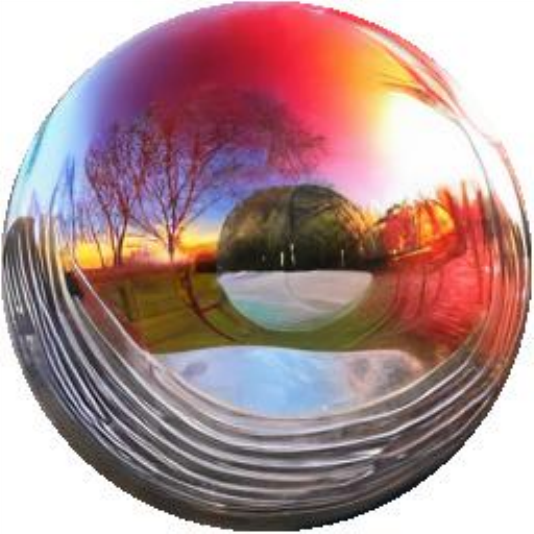}} &  
        \noindent\parbox[c]{0.071\textwidth}{\includegraphics[width=0.071\textwidth]{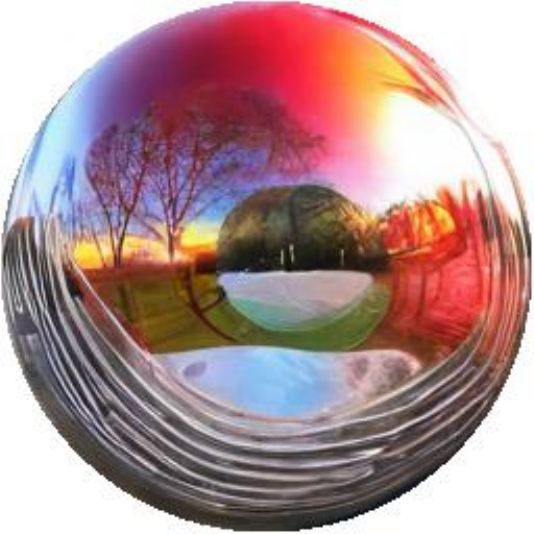}}
        \\

        & 
        \noindent\parbox[c]{0.071\textwidth}{\includegraphics[width=0.071\textwidth]{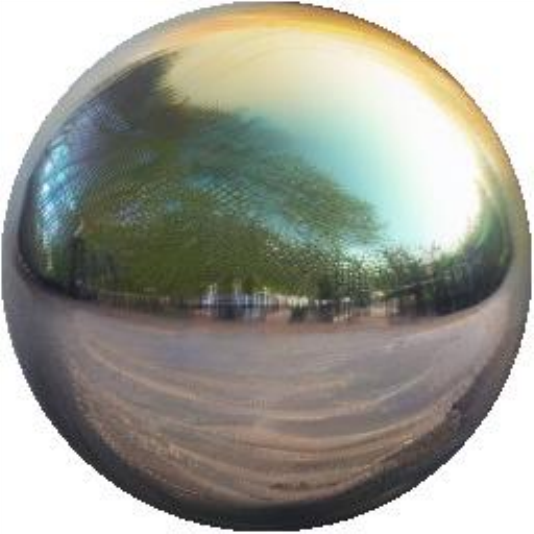}} &  
        \noindent\parbox[c]{0.071\textwidth}{\includegraphics[width=0.071\textwidth]{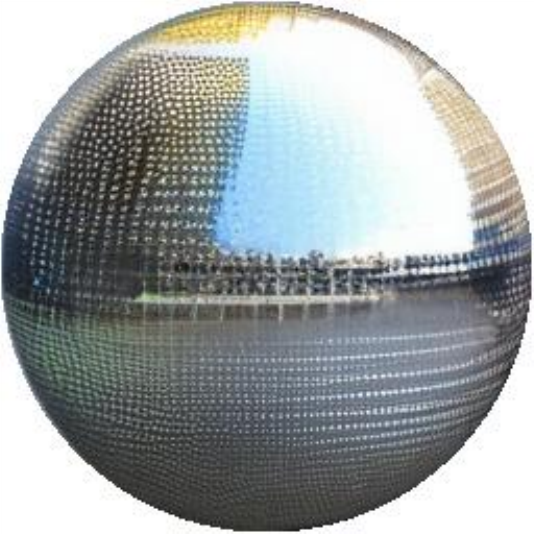}} &  
        \noindent\parbox[c]{0.071\textwidth}{\includegraphics[width=0.071\textwidth]{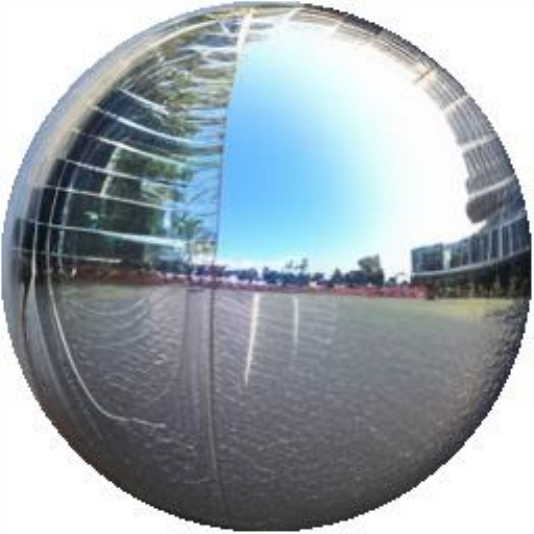}} &  
        \noindent\parbox[c]{0.071\textwidth}{\includegraphics[width=0.071\textwidth]{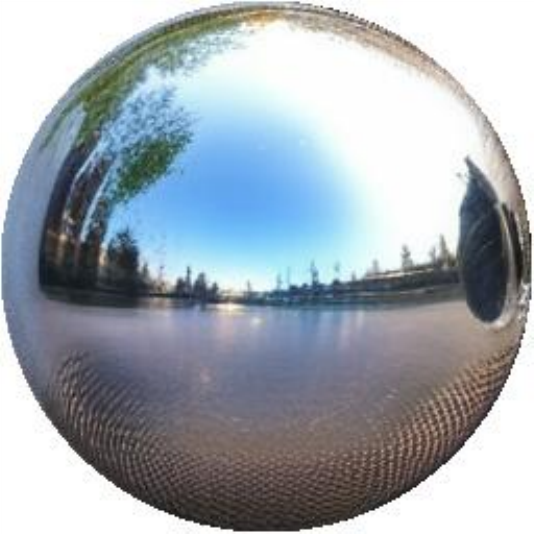}} &  
        \noindent\parbox[c]{0.071\textwidth}{\includegraphics[width=0.071\textwidth]{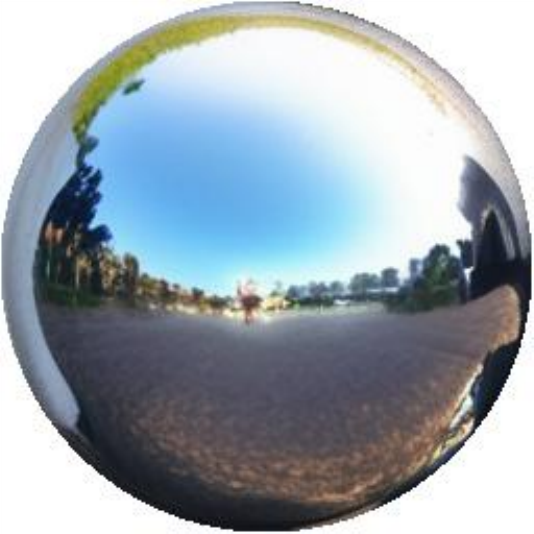}}
        \\

        \end{tabu}
    \caption{Comparison between chrome balls generated using DDIM \cite{song2020denoising} (1\textsuperscript{st} row) and DDPM \cite{ho2020denoising} (2\textsuperscript{nd} row) with different sampling steps. DDPM can sometimes mitigate the occurrence of bad patterns originating from bad initial noise maps when using high sampling steps. Prompt: ``a chrome ball''.}
    \label{fig:aba_disco_stochastic_bad}
\end{figure}

\begin{figure}[!t]
    \centering

        \begin{tabu} to \textwidth {
        @{}
        c@{\hspace{0.5pt}}
        c@{\hspace{0.5pt}}
        c@{\hspace{0.5pt}}
        c@{\hspace{0.5pt}}
        c@{\hspace{0.5pt}}
        c@{\hspace{0.5pt}}
        c@{\hspace{0.5pt}}
    }

        \multicolumn{1}{c}{\shortstack{\scriptsize Input}} & 
        \multicolumn{1}{c}{\shortstack{\scriptsize 10 steps}} &
        \multicolumn{1}{c}{\shortstack{\scriptsize 20 steps}} & 
        \multicolumn{1}{c}{\shortstack{\scriptsize 40 steps}} & 
        \multicolumn{1}{c}{\shortstack{\scriptsize 80 steps}} &
        \multicolumn{1}{c}{\shortstack{\scriptsize 160 steps}} &
        \\

        \noindent\parbox[c]{0.11\textwidth}{\includegraphics[width=0.11\textwidth]{storage/noise_appearance_relationship_whiteBG/input/airport_seed150.pdf}} & 
        \noindent\parbox[c]{0.071\textwidth}{\includegraphics[width=0.071\textwidth]{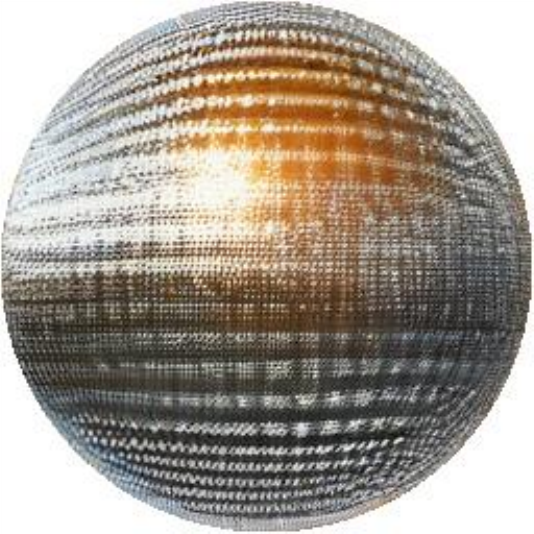}} &  
        \noindent\parbox[c]{0.071\textwidth}{\includegraphics[width=0.071\textwidth]{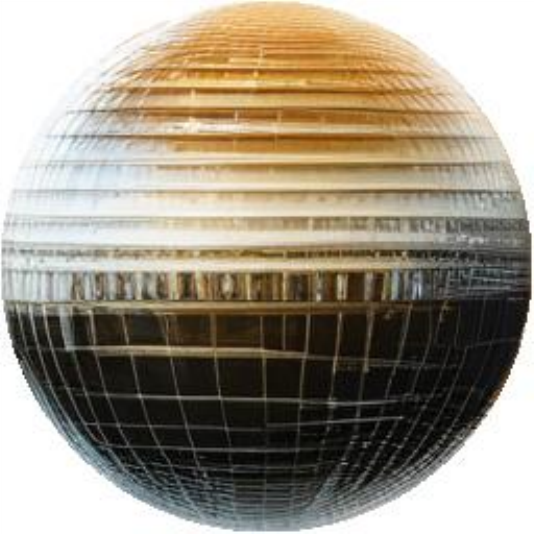}} &  
        \noindent\parbox[c]{0.071\textwidth}{\includegraphics[width=0.071\textwidth]{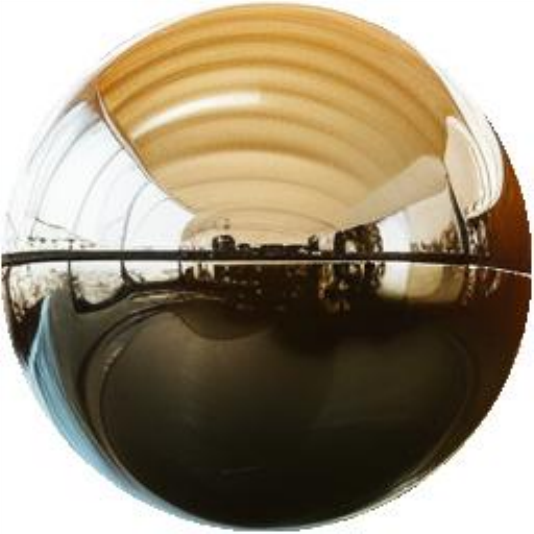}} &  
        \noindent\parbox[c]{0.071\textwidth}{\includegraphics[width=0.071\textwidth]{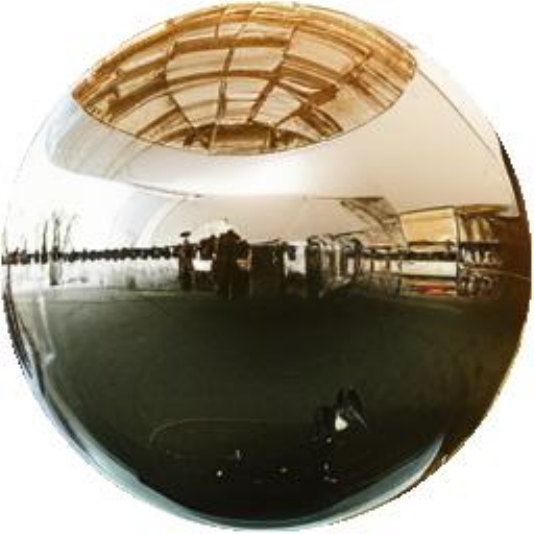}} &  
        \noindent\parbox[c]{0.071\textwidth}{\includegraphics[width=0.071\textwidth]{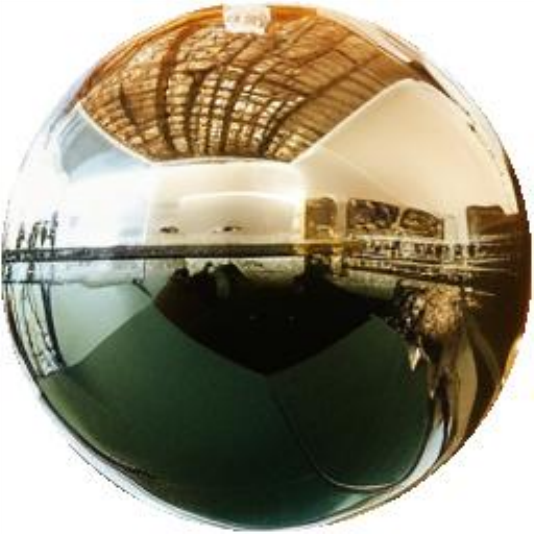}}
        \\

        \noindent\parbox[c]{0.11\textwidth}{\includegraphics[width=0.11\textwidth]{storage/noise_appearance_relationship_whiteBG/input/castle3_seed150.pdf}} & 
        \noindent\parbox[c]{0.071\textwidth}{\includegraphics[width=0.071\textwidth]{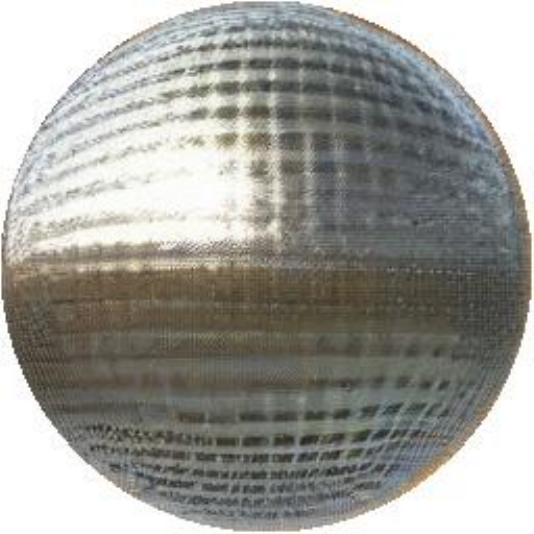}} &  
        \noindent\parbox[c]{0.071\textwidth}{\includegraphics[width=0.071\textwidth]{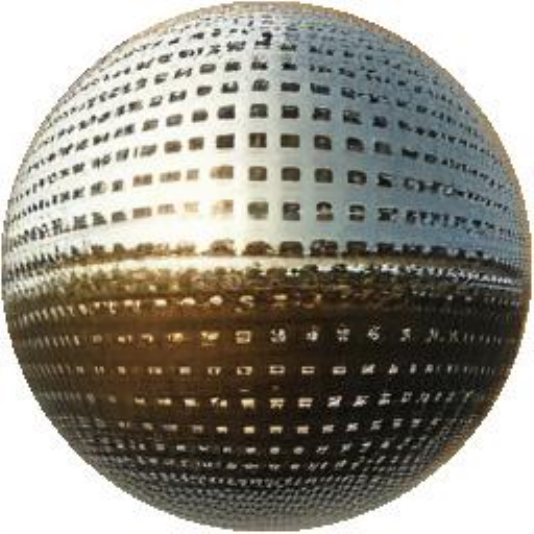}} &  
        \noindent\parbox[c]{0.071\textwidth}{\includegraphics[width=0.071\textwidth]{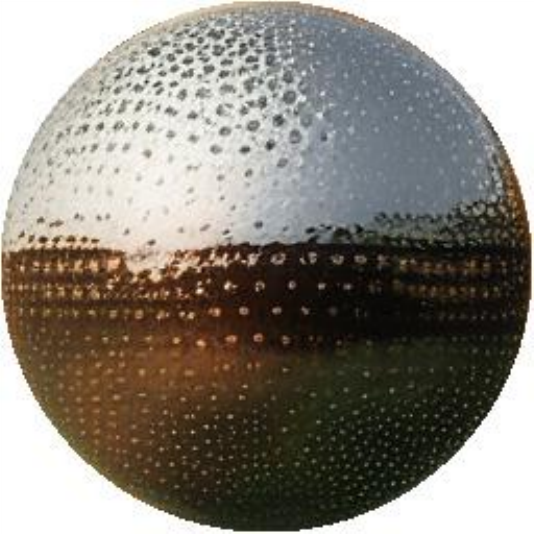}} &  
        \noindent\parbox[c]{0.071\textwidth}{\includegraphics[width=0.071\textwidth]{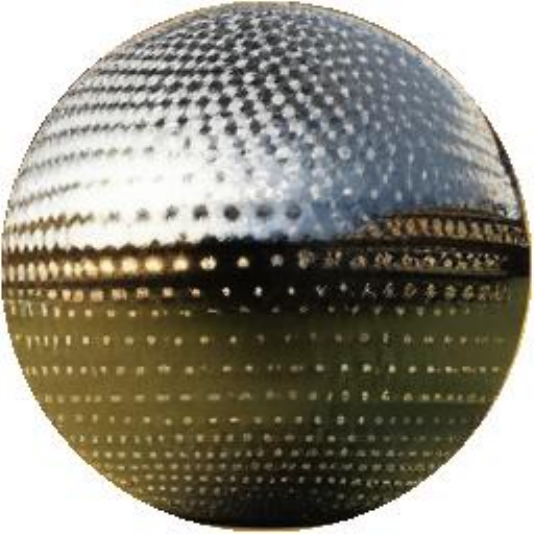}} &  
        \noindent\parbox[c]{0.071\textwidth}{\includegraphics[width=0.071\textwidth]{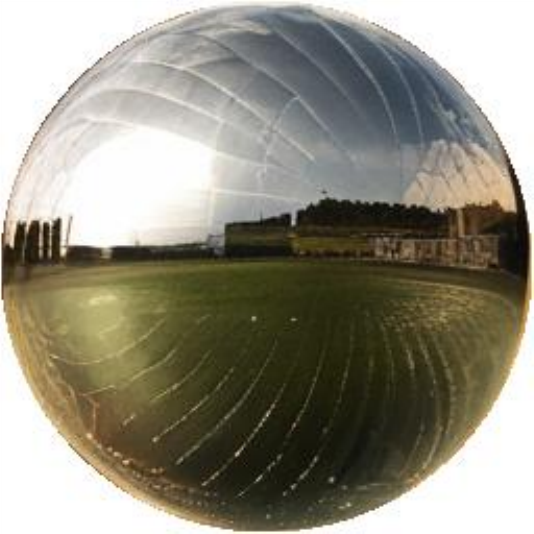}}
        \\

        \noindent\parbox[c]{0.11\textwidth}{\includegraphics[width=0.11\textwidth]{storage/noise_appearance_relationship_whiteBG/input/panda_seed150.pdf}} & 
        \noindent\parbox[c]{0.071\textwidth}{\includegraphics[width=0.071\textwidth]{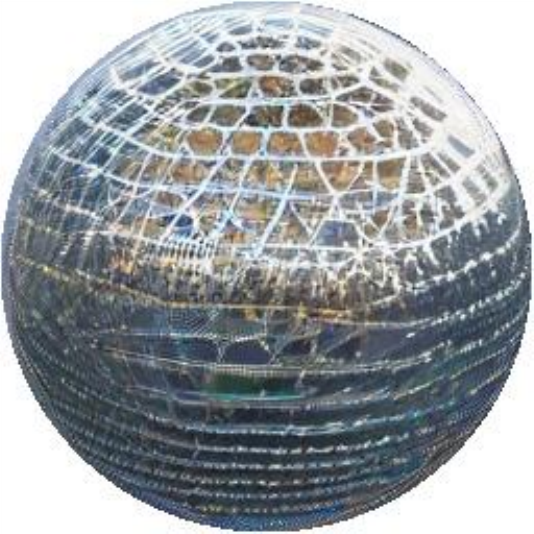}} &  
        \noindent\parbox[c]{0.071\textwidth}{\includegraphics[width=0.071\textwidth]{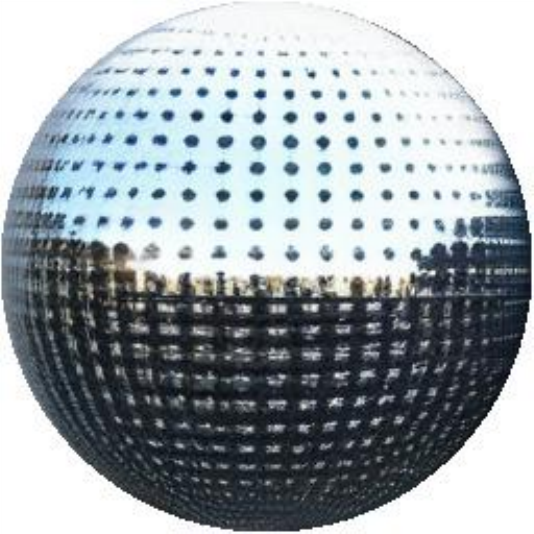}} &  
        \noindent\parbox[c]{0.071\textwidth}{\includegraphics[width=0.071\textwidth]{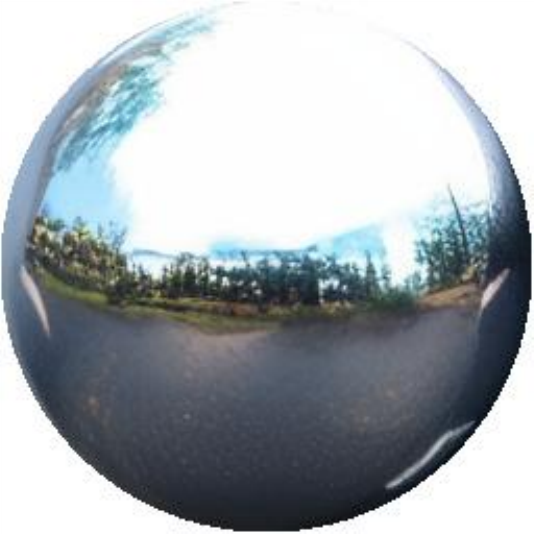}} &  
        \noindent\parbox[c]{0.071\textwidth}{\includegraphics[width=0.071\textwidth]{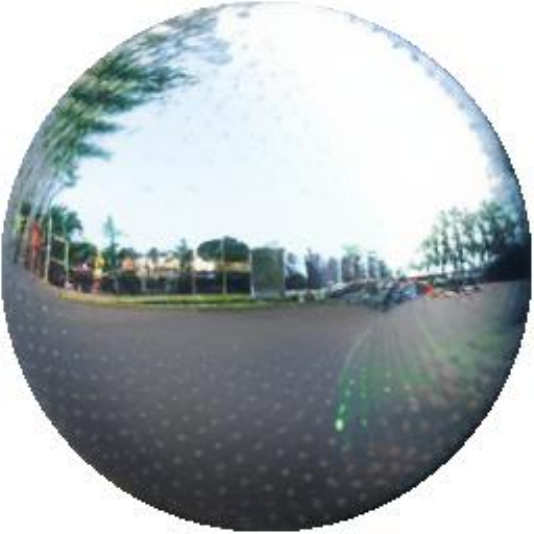}} &  
        \noindent\parbox[c]{0.071\textwidth}{\includegraphics[width=0.071\textwidth]{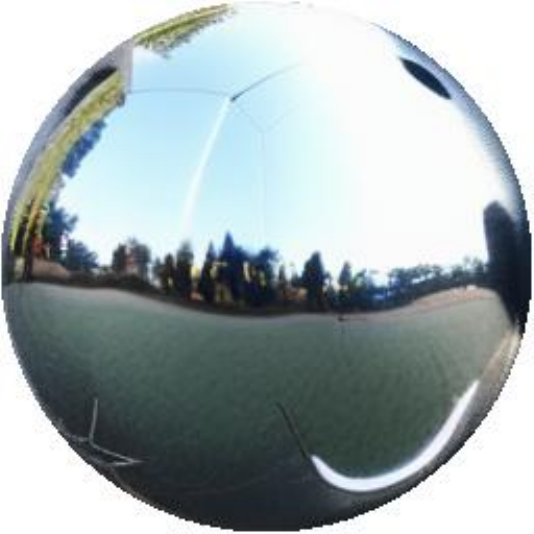}}
        \\

        \noindent\parbox[c]{0.11\textwidth}{\includegraphics[width=0.11\textwidth]{storage/noise_appearance_relationship_whiteBG/input/street2_seed150.pdf}} & 
        \noindent\parbox[c]{0.071\textwidth}{\includegraphics[width=0.071\textwidth]{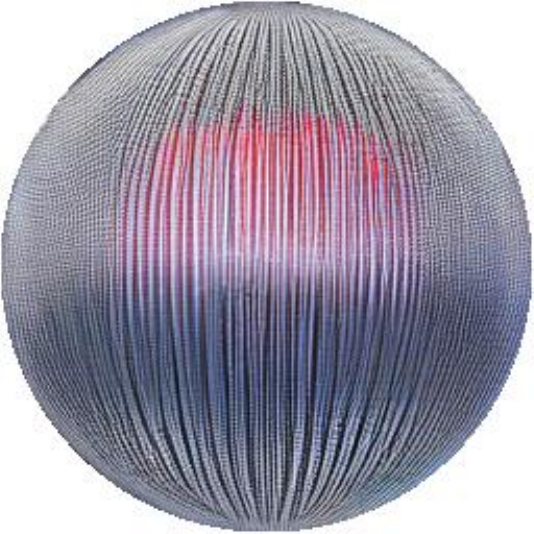}} &  
        \noindent\parbox[c]{0.071\textwidth}{\includegraphics[width=0.071\textwidth]{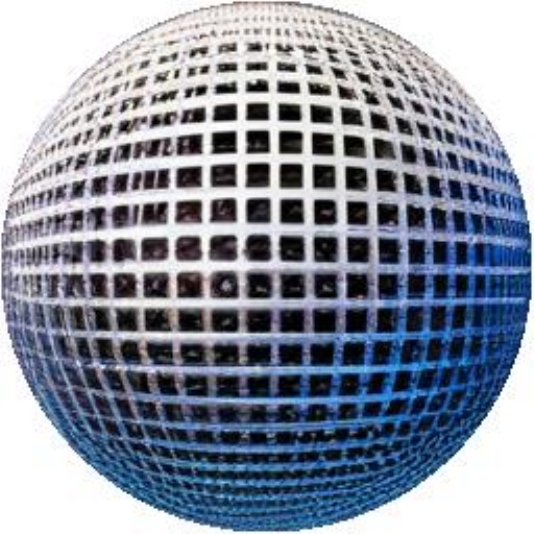}} &  
        \noindent\parbox[c]{0.071\textwidth}{\includegraphics[width=0.071\textwidth]{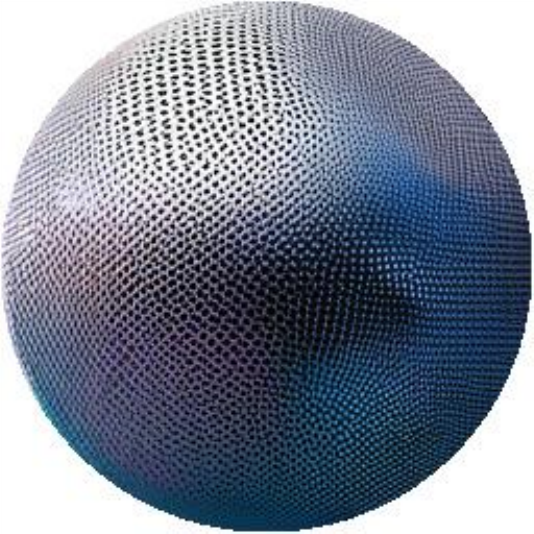}} &  
        \noindent\parbox[c]{0.071\textwidth}{\includegraphics[width=0.071\textwidth]{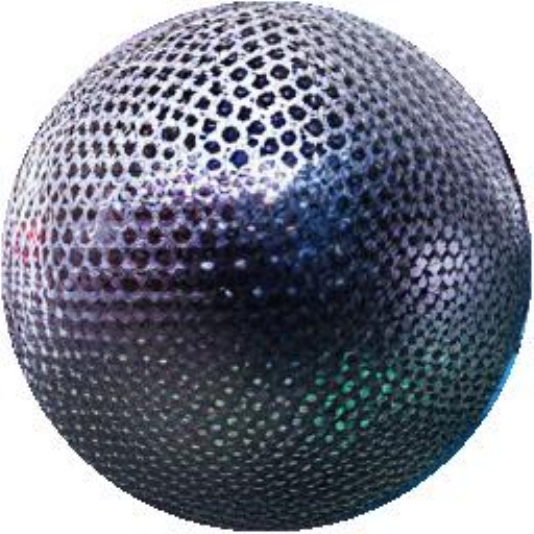}} &  
        \noindent\parbox[c]{0.071\textwidth}{\includegraphics[width=0.071\textwidth]{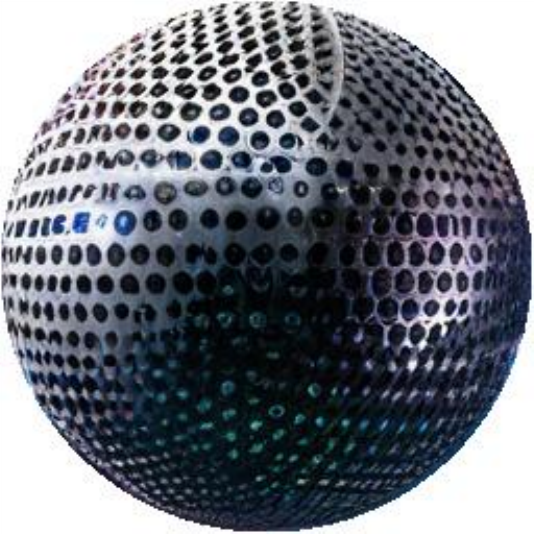}}
        \\

        \noindent\parbox[c]{0.11\textwidth}{\includegraphics[width=0.11\textwidth]{storage/noise_appearance_relationship_whiteBG/input/opera_house_seed150.pdf}} & 
        \noindent\parbox[c]{0.071\textwidth}{\includegraphics[width=0.071\textwidth]{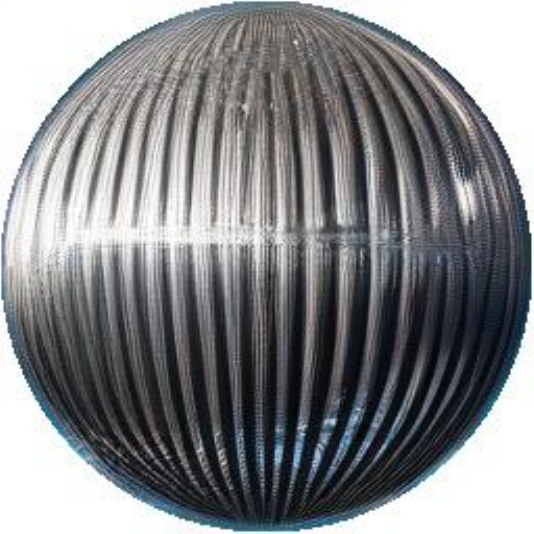}} &  
        \noindent\parbox[c]{0.071\textwidth}{\includegraphics[width=0.071\textwidth]{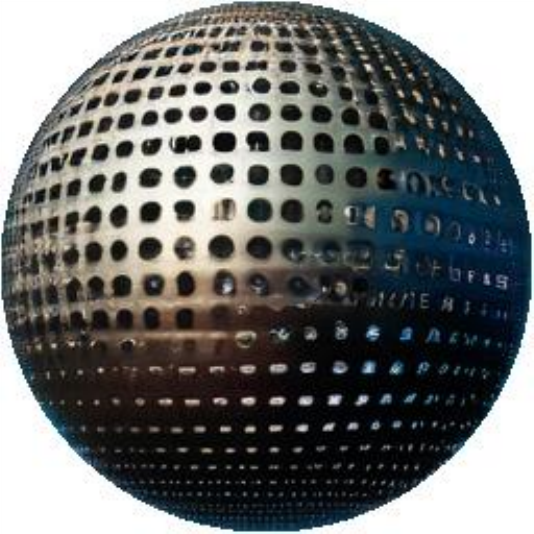}} &  
        \noindent\parbox[c]{0.071\textwidth}{\includegraphics[width=0.071\textwidth]{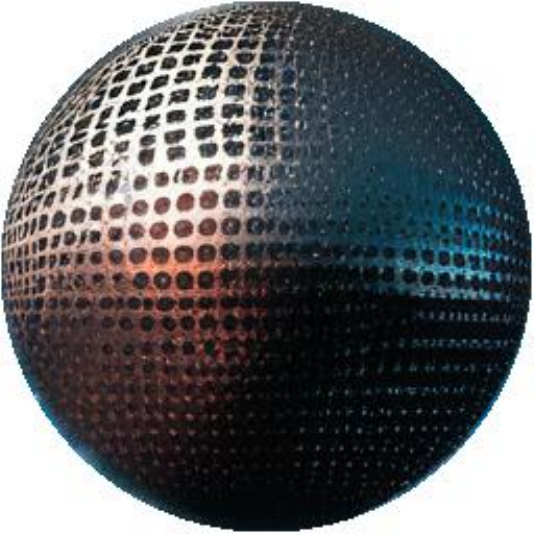}} &  
        \noindent\parbox[c]{0.071\textwidth}{\includegraphics[width=0.071\textwidth]{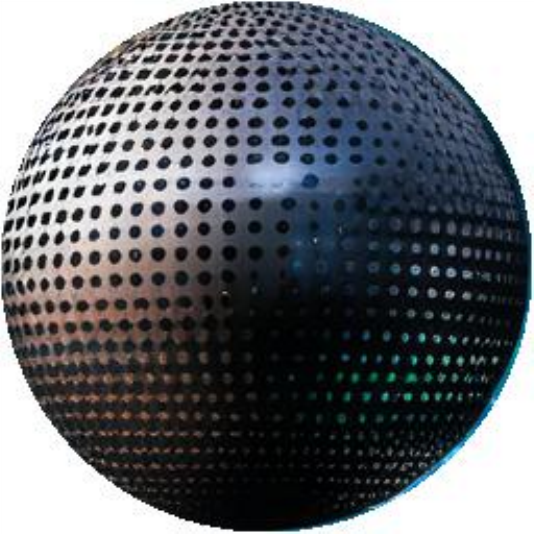}} &  
        \noindent\parbox[c]{0.071\textwidth}{\includegraphics[width=0.071\textwidth]{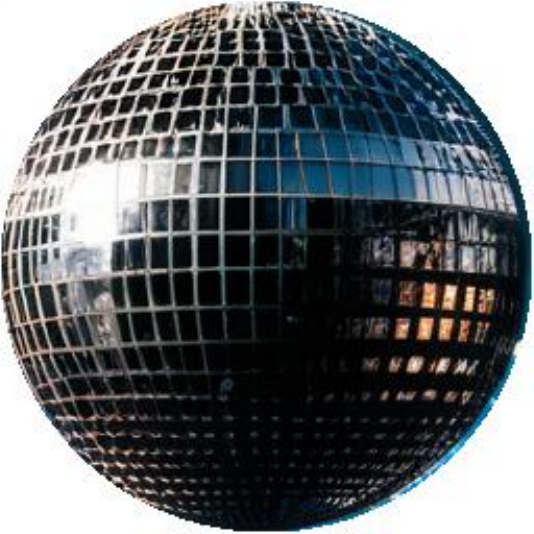}}
        \\
        
        \end{tabu}
    \caption{Chrome balls generated from the ``disco'' noise map using DDPM \cite{ho2020denoising} with different sampling steps. Switching to DDPM instead of deterministic samplers still produces ``disco'' balls, even at high sampling steps. Prompt: ``a chrome ball''.}
    \label{fig:aba_disco_stochastic_disco}
\end{figure}

We conducted an experiment comparing results from DDIM and DDPM using different numbers of sampling steps. 
Our results suggest that neither of these schemes consistently reduces occurrence of bad chrome balls. 
Specifically, using stochastic samplers may occasionally produce better chrome balls than deterministic ones at sufficiently high numbers of sampling steps (see Figure \ref{fig:aba_disco_stochastic_bad}). Unfortunately, they still yield ``disco'' balls when starting sampling with the ``disco'' noise map, as illustrated in Figure \ref{fig:aba_disco_stochastic_disco}.

\tabulinesep=0.5pt

\tabulinesep=0.5pt
\section{Spatially-Varying Light Estimation}
\begin{figure*}[ht]
    \centering

    \begin{tabu} to \textwidth {
        @{}
        c@{\hspace{2pt}}
        c@{\hspace{0.5pt}}
        c@{\hspace{0.5pt}}
        c@{\hspace{2pt}}
        c@{\hspace{0.5pt}}
        c@{\hspace{0.5pt}}
        c@{\hspace{2pt}}
        c@{\hspace{0.5pt}}
        c@{\hspace{0.5pt}}
        c@{\hspace{0.5pt}}
        c@{}
    }
        
        \multicolumn{1}{c}{\shortstack{\scriptsize Input image}} & 
        \multicolumn{3}{c}{\shortstack{\scriptsize Prediction}} &
        \multicolumn{3}{c}{\shortstack{\scriptsize Median ball (1\textsuperscript{st} iteration)}} &
        \multicolumn{3}{c}{\shortstack{\scriptsize Median ball (2\textsuperscript{nd} iteration)}} &
        \\

        \noindent\parbox[c]{0.140\textwidth}{\includegraphics[width=0.140\textwidth]{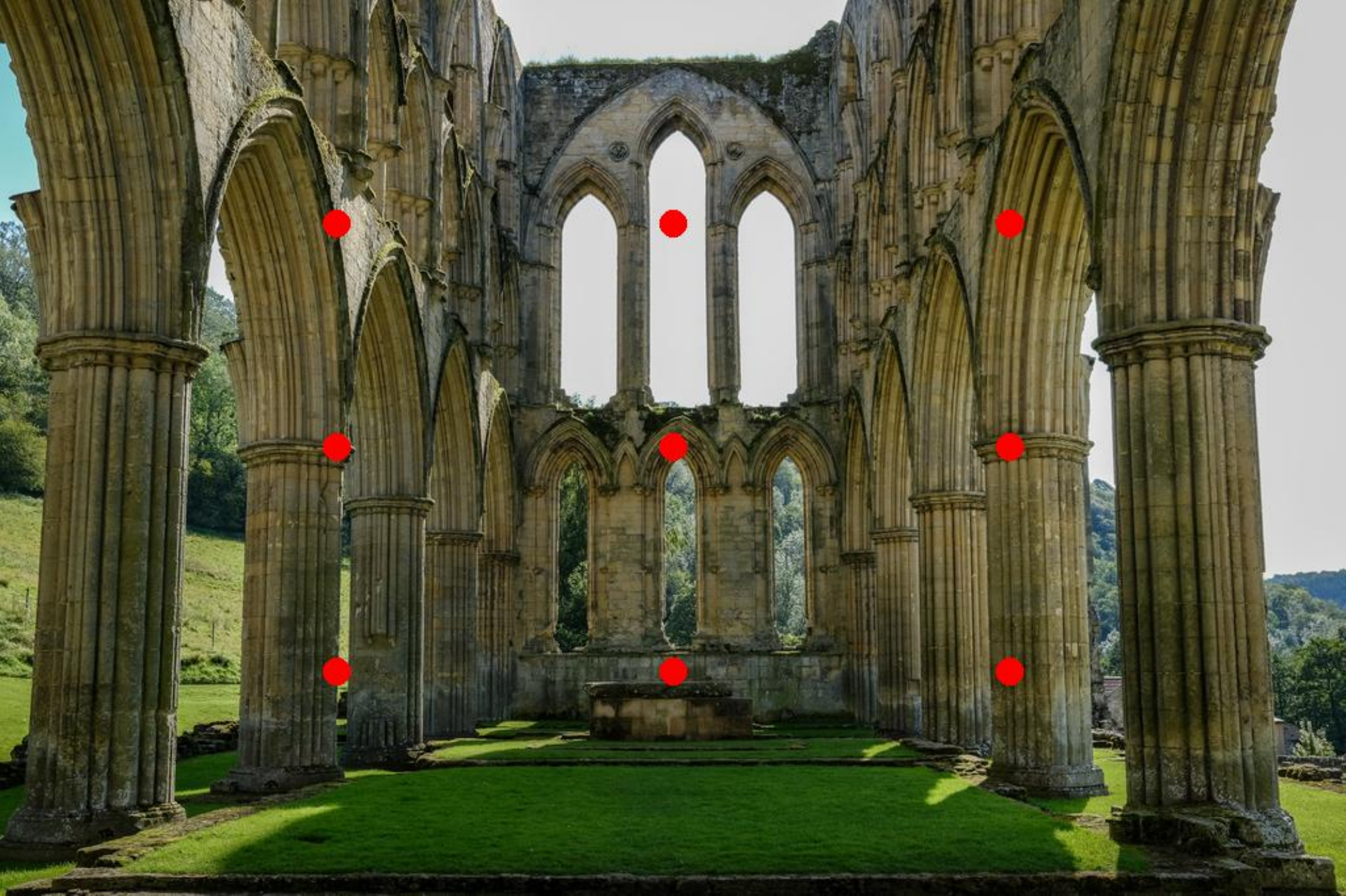}} & 
        \noindent\parbox[c]{0.092\textwidth}{\includegraphics[width=0.092\textwidth]{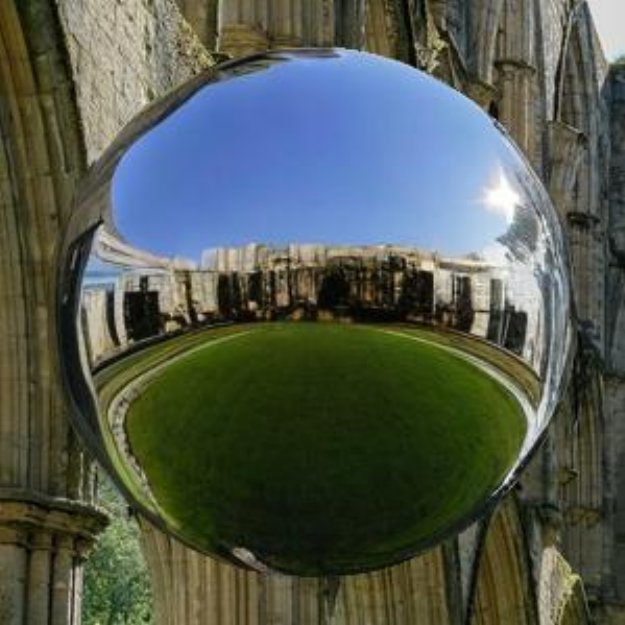}} & 
        \noindent\parbox[c]{0.092\textwidth}{\includegraphics[width=0.092\textwidth]{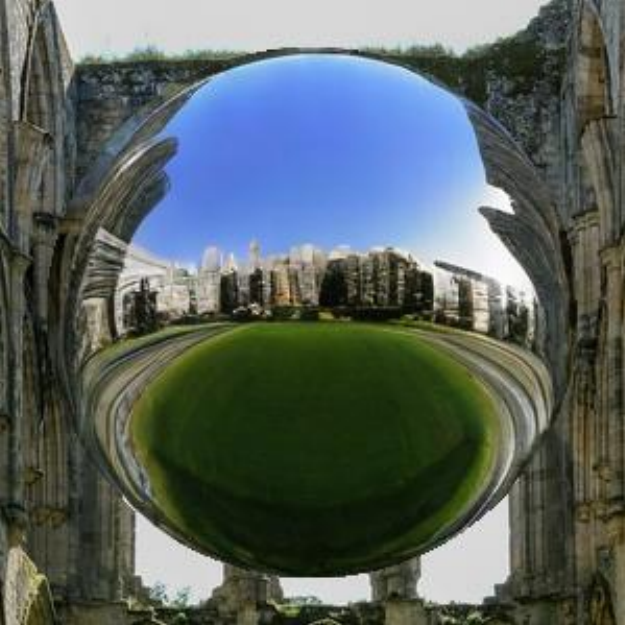}} & 
        \noindent\parbox[c]{0.092\textwidth}{\includegraphics[width=0.092\textwidth]{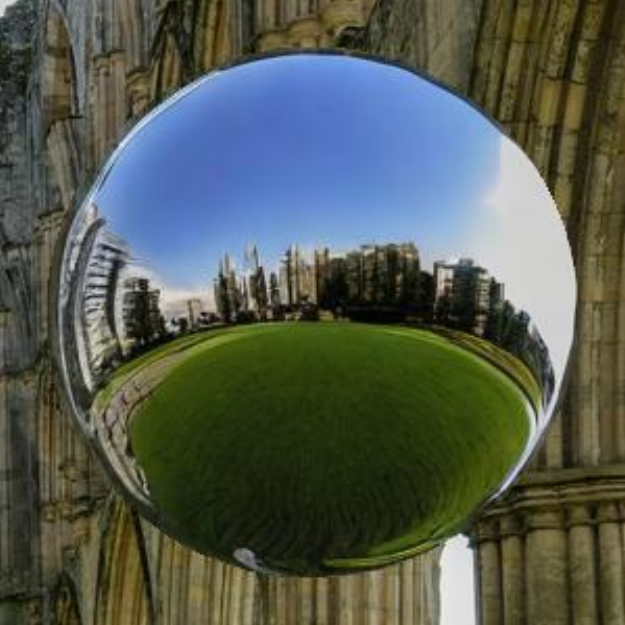}} & 
        \noindent\parbox[c]{0.092\textwidth}{\includegraphics[width=0.092\textwidth]{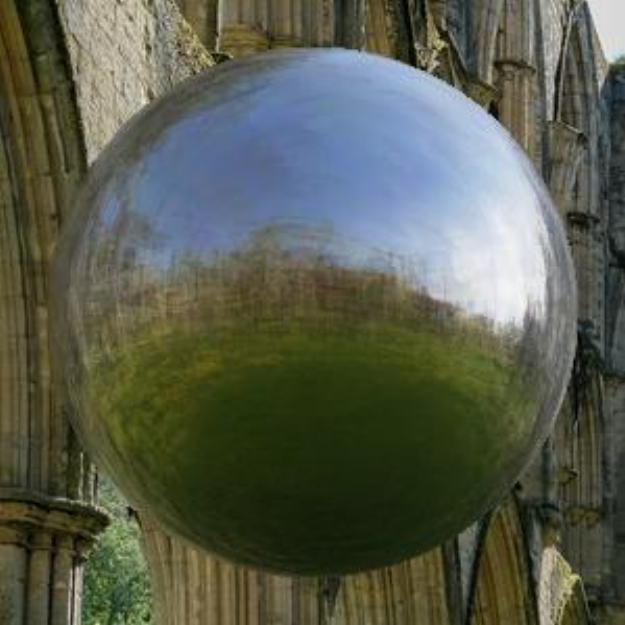}} & 
        \noindent\parbox[c]{0.092\textwidth}{\includegraphics[width=0.092\textwidth]{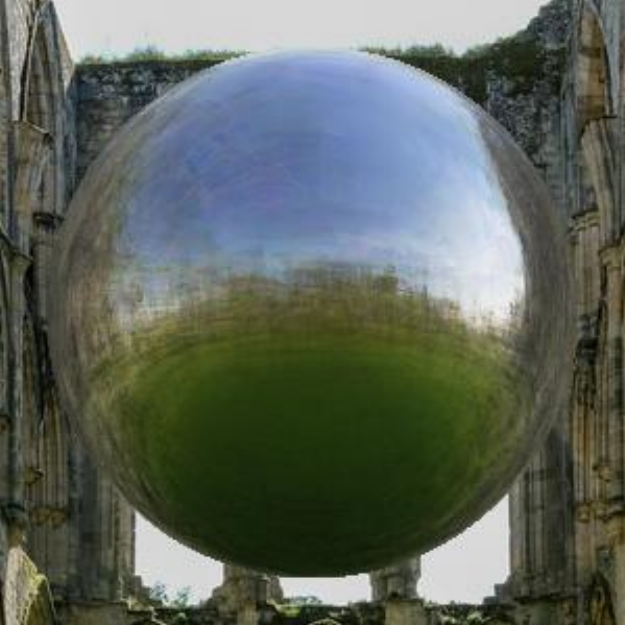}} & 
        \noindent\parbox[c]{0.092\textwidth}{\includegraphics[width=0.092\textwidth]{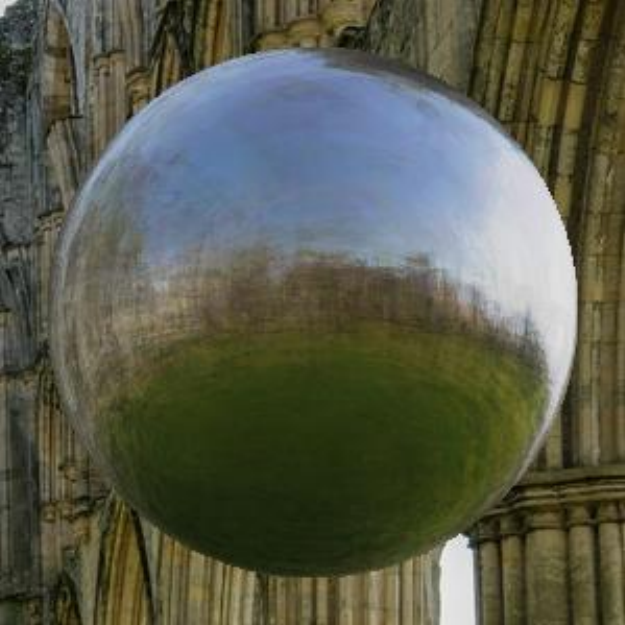}} & 
        \noindent\parbox[c]{0.092\textwidth}{\includegraphics[width=0.092\textwidth]{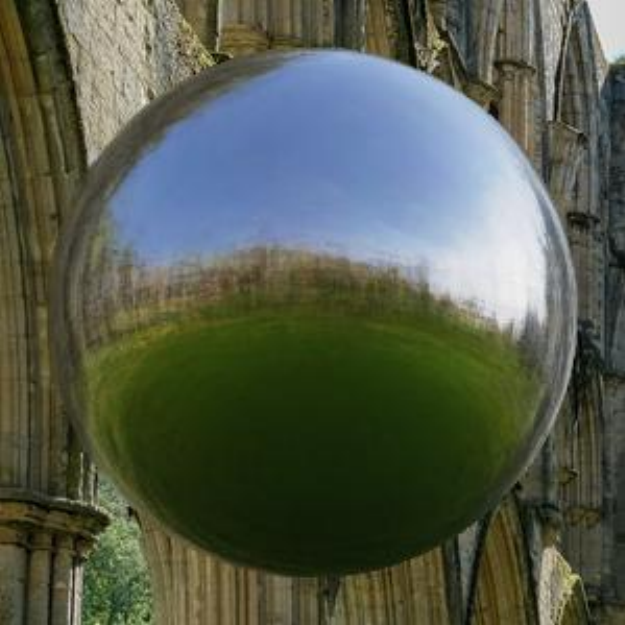}} & 
        \noindent\parbox[c]{0.092\textwidth}{\includegraphics[width=0.092\textwidth]{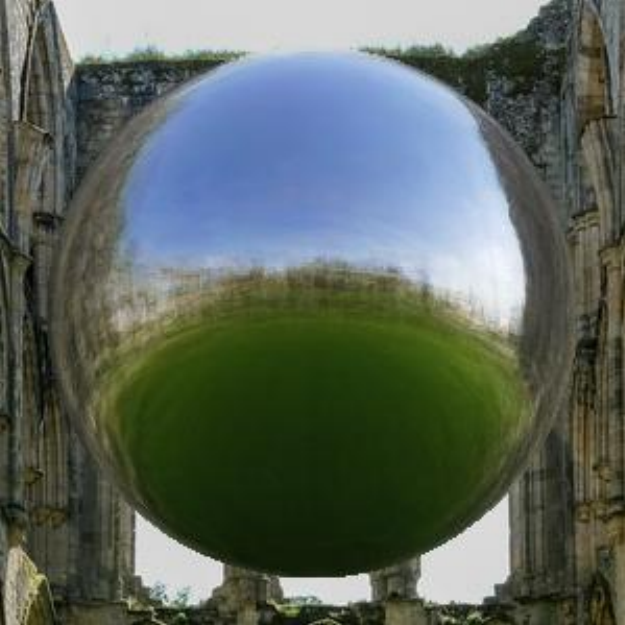}} & 
        \noindent\parbox[c]{0.092\textwidth}{\includegraphics[width=0.092\textwidth]{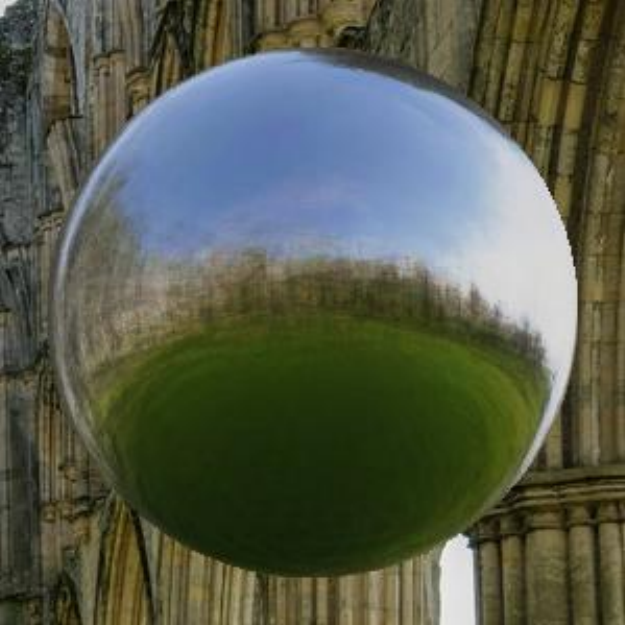}} & 
        
        \\

        & 
        \noindent\parbox[c]{0.092\textwidth}{\includegraphics[width=0.092\textwidth]{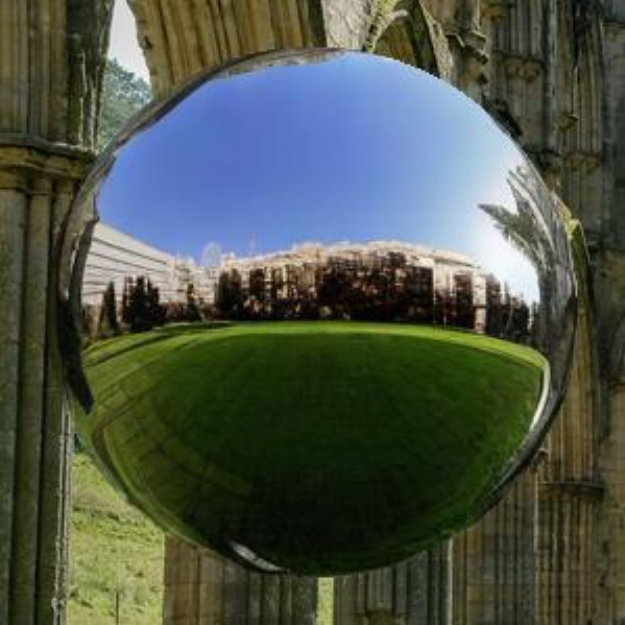}} & 
        \noindent\parbox[c]{0.092\textwidth}{\includegraphics[width=0.092\textwidth]{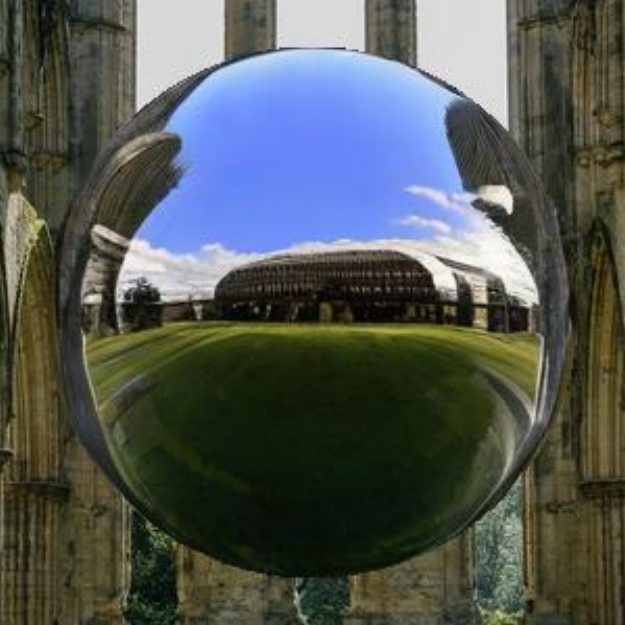}} & 
        \noindent\parbox[c]{0.092\textwidth}{\includegraphics[width=0.092\textwidth]{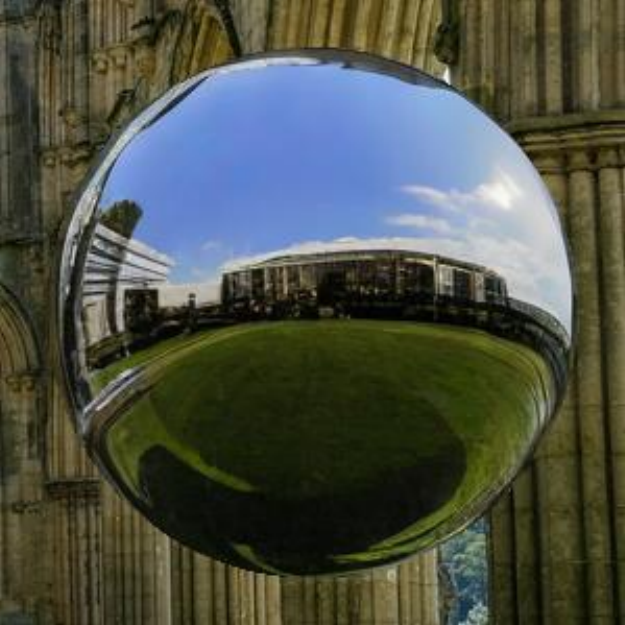}} & 
        \noindent\parbox[c]{0.092\textwidth}{\includegraphics[width=0.092\textwidth]{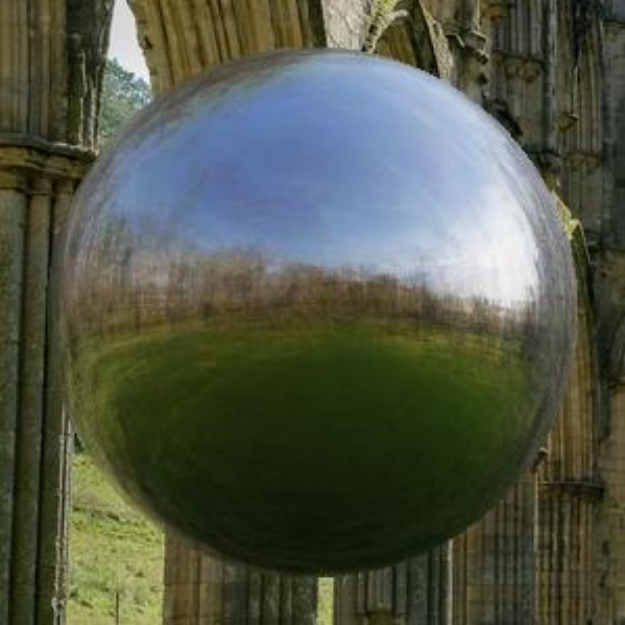}} & 
        \noindent\parbox[c]{0.092\textwidth}{\includegraphics[width=0.092\textwidth]{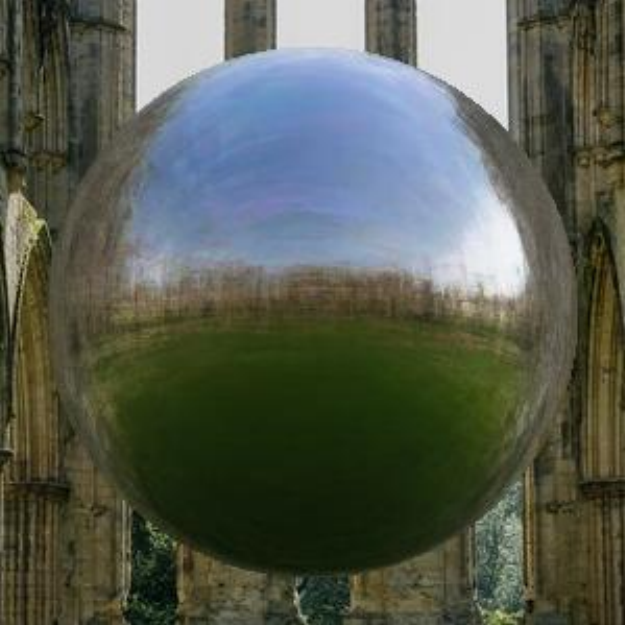}} & 
        \noindent\parbox[c]{0.092\textwidth}{\includegraphics[width=0.092\textwidth]{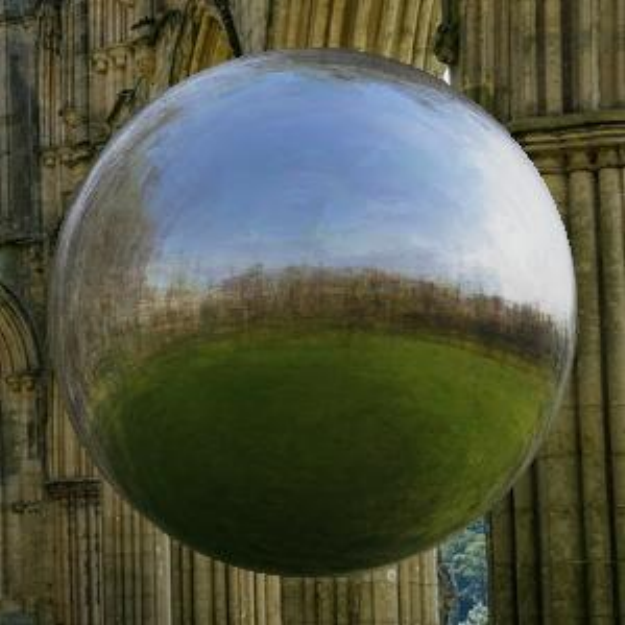}} & 
        \noindent\parbox[c]{0.092\textwidth}{\includegraphics[width=0.092\textwidth]{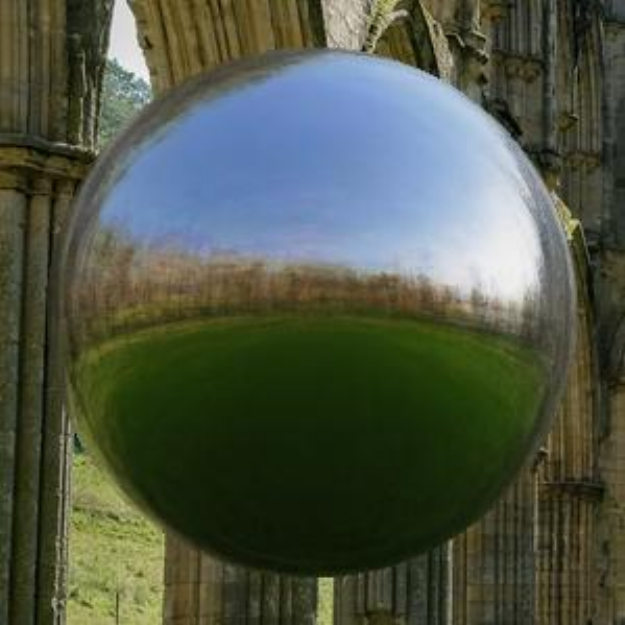}} & 
        \noindent\parbox[c]{0.092\textwidth}{\includegraphics[width=0.092\textwidth]{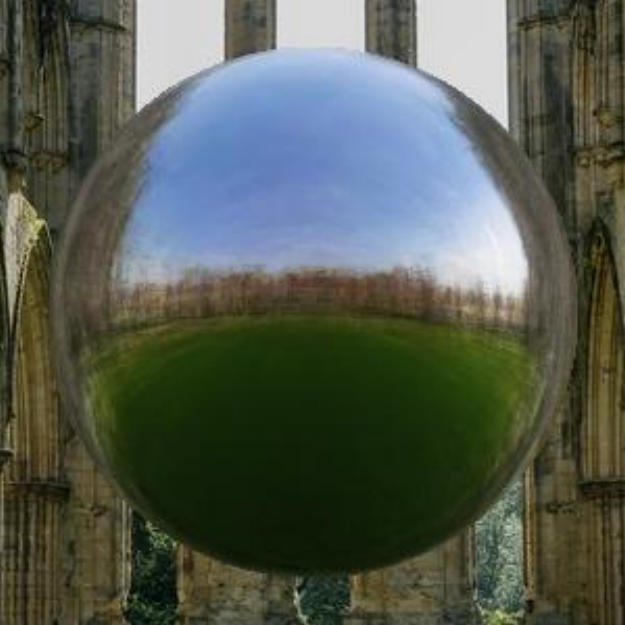}} & 
        \noindent\parbox[c]{0.092\textwidth}{\includegraphics[width=0.092\textwidth]{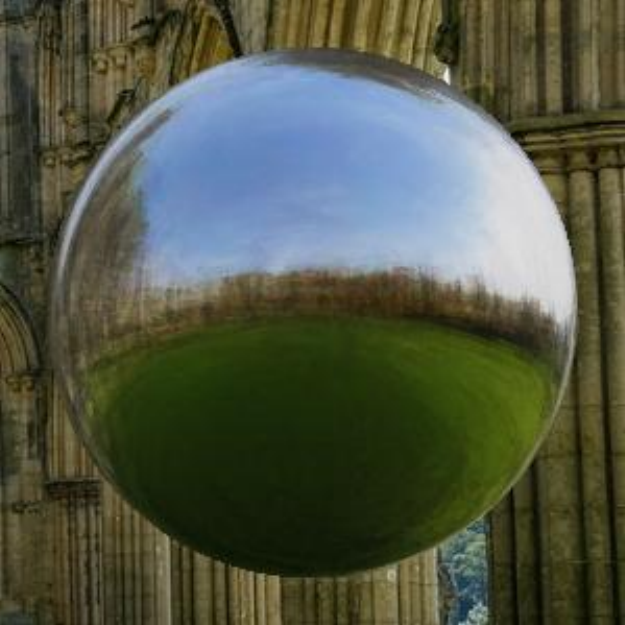}} & 
        
        \\

        & 
        \noindent\parbox[c]{0.092\textwidth}{\includegraphics[width=0.092\textwidth]{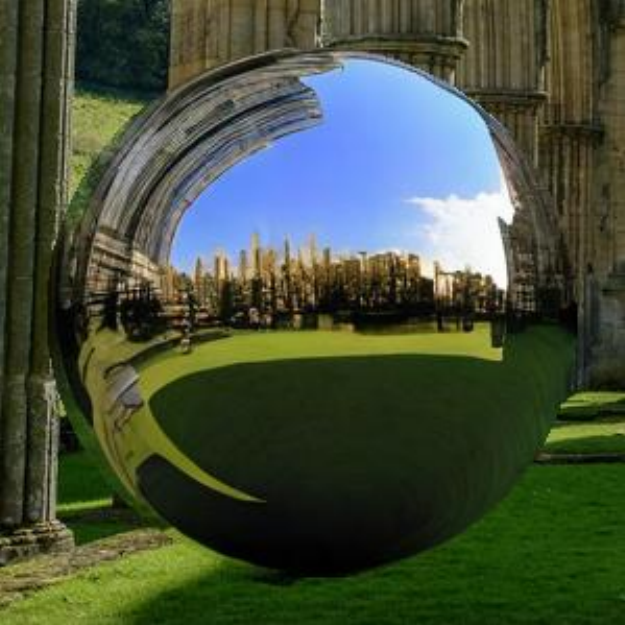}} & 
        \noindent\parbox[c]{0.092\textwidth}{\includegraphics[width=0.092\textwidth]{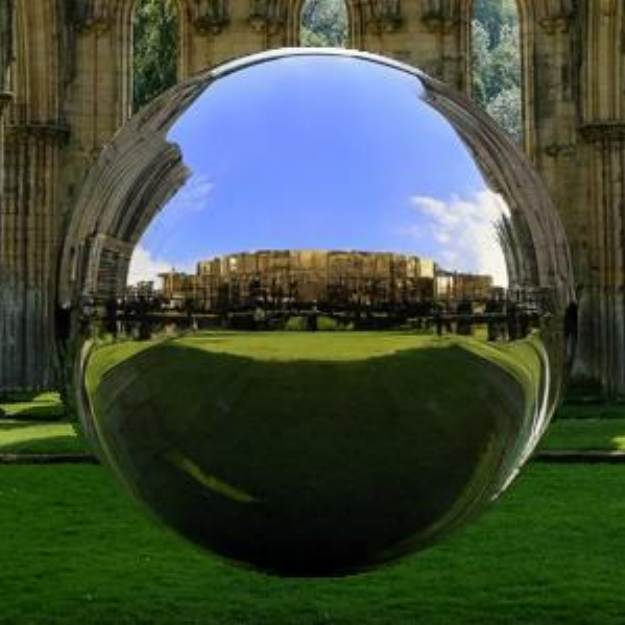}} & 
        \noindent\parbox[c]{0.092\textwidth}{\includegraphics[width=0.092\textwidth]{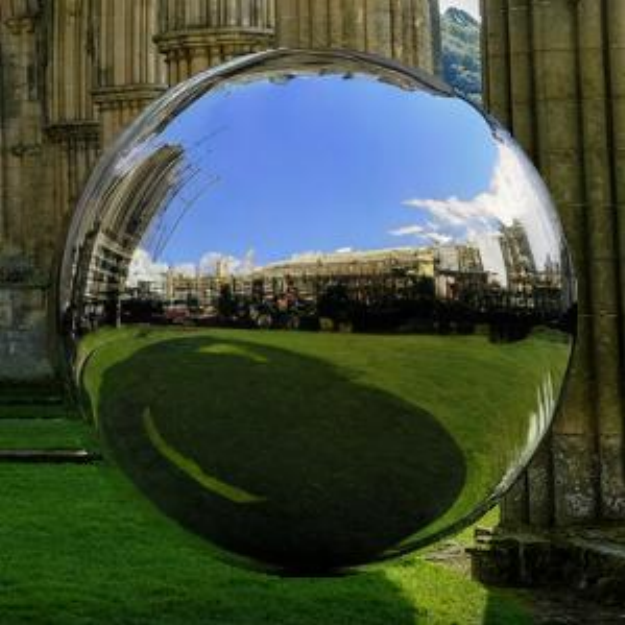}} & 
        \noindent\parbox[c]{0.092\textwidth}{\includegraphics[width=0.092\textwidth]{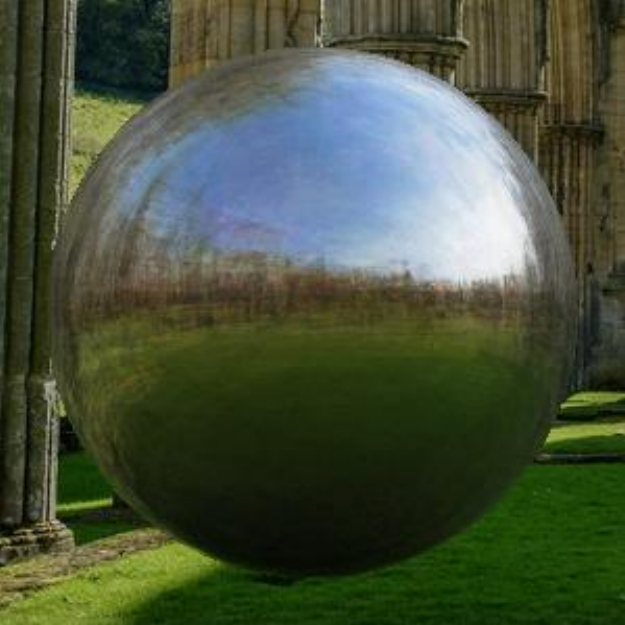}} & 
        \noindent\parbox[c]{0.092\textwidth}{\includegraphics[width=0.092\textwidth]{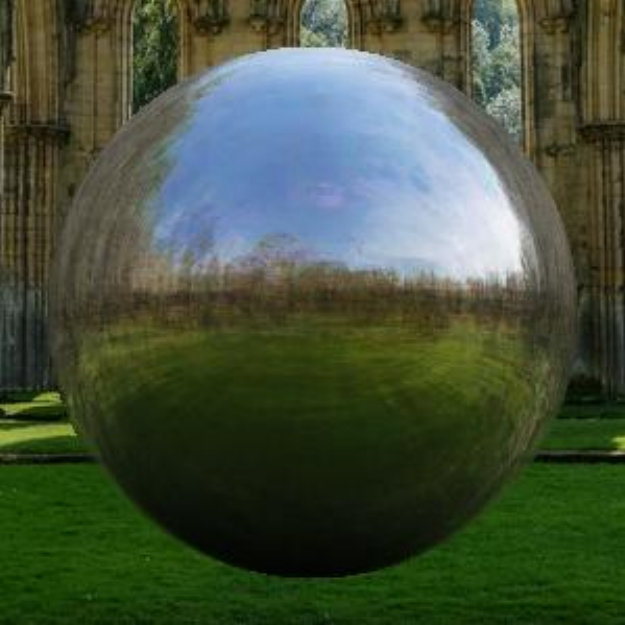}} & 
        \noindent\parbox[c]{0.092\textwidth}{\includegraphics[width=0.092\textwidth]{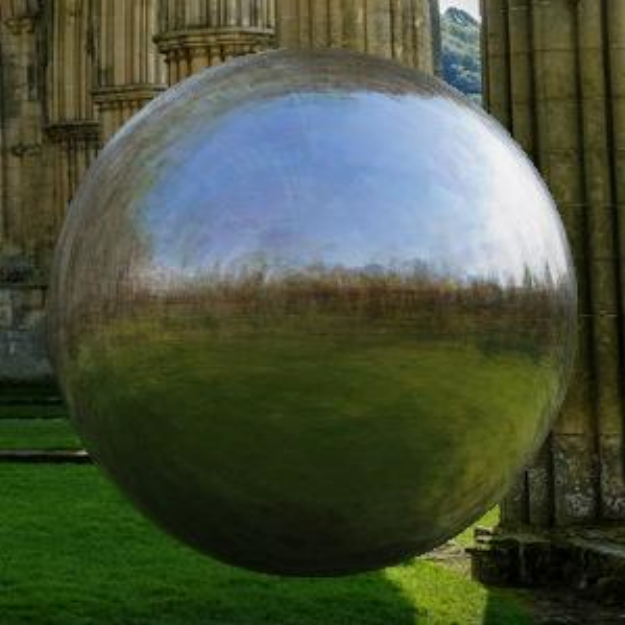}} & 
        \noindent\parbox[c]{0.092\textwidth}{\includegraphics[width=0.092\textwidth]{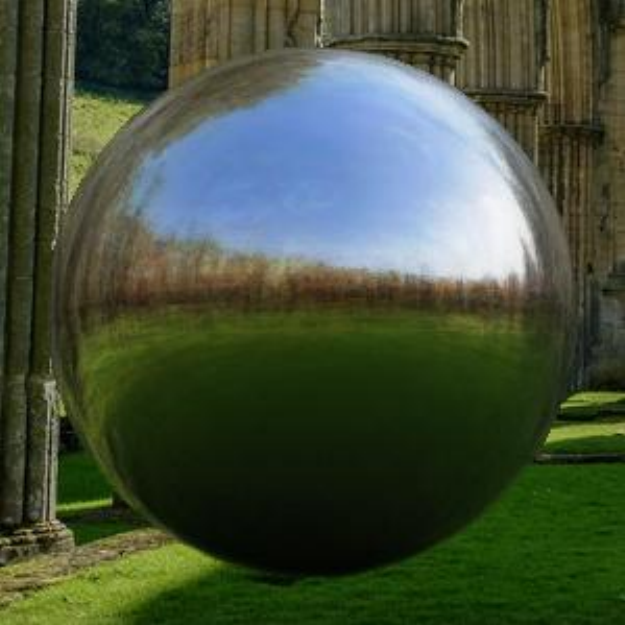}} & 
        \noindent\parbox[c]{0.092\textwidth}{\includegraphics[width=0.092\textwidth]{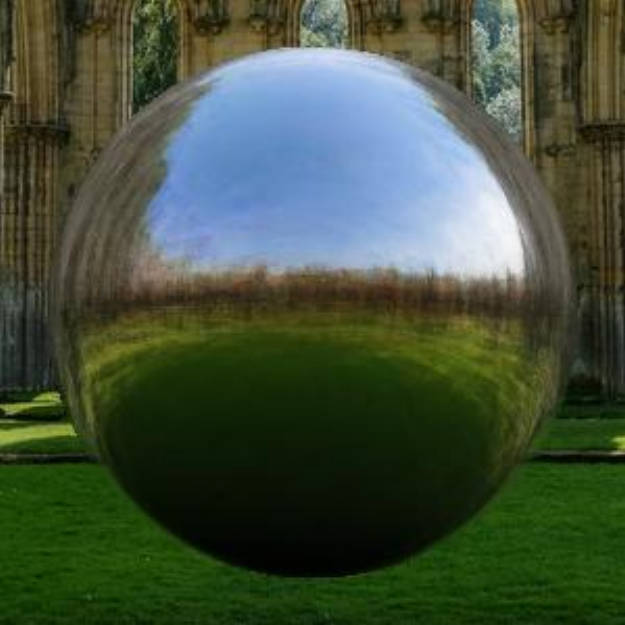}} & 
        \noindent\parbox[c]{0.092\textwidth}{\includegraphics[width=0.092\textwidth]{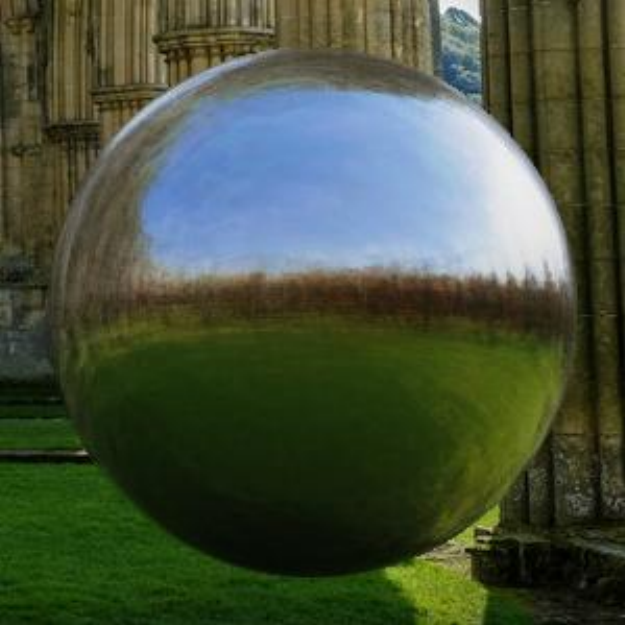}} & 
        
        \\
        
    \end{tabu}

    \smallskip
    \begin{tabu} to \textwidth {
        @{}
        c@{\hspace{2pt}}
        c@{\hspace{0.5pt}}
        c@{\hspace{0.5pt}}
        c@{\hspace{2pt}}
        c@{\hspace{0.5pt}}
        c@{\hspace{0.5pt}}
        c@{\hspace{2pt}}
        c@{\hspace{0.5pt}}
        c@{\hspace{0.5pt}}
        c@{\hspace{0.5pt}}
        c@{}
    }
        

        \noindent\parbox[c]{0.140\textwidth}{\includegraphics[width=0.140\textwidth]{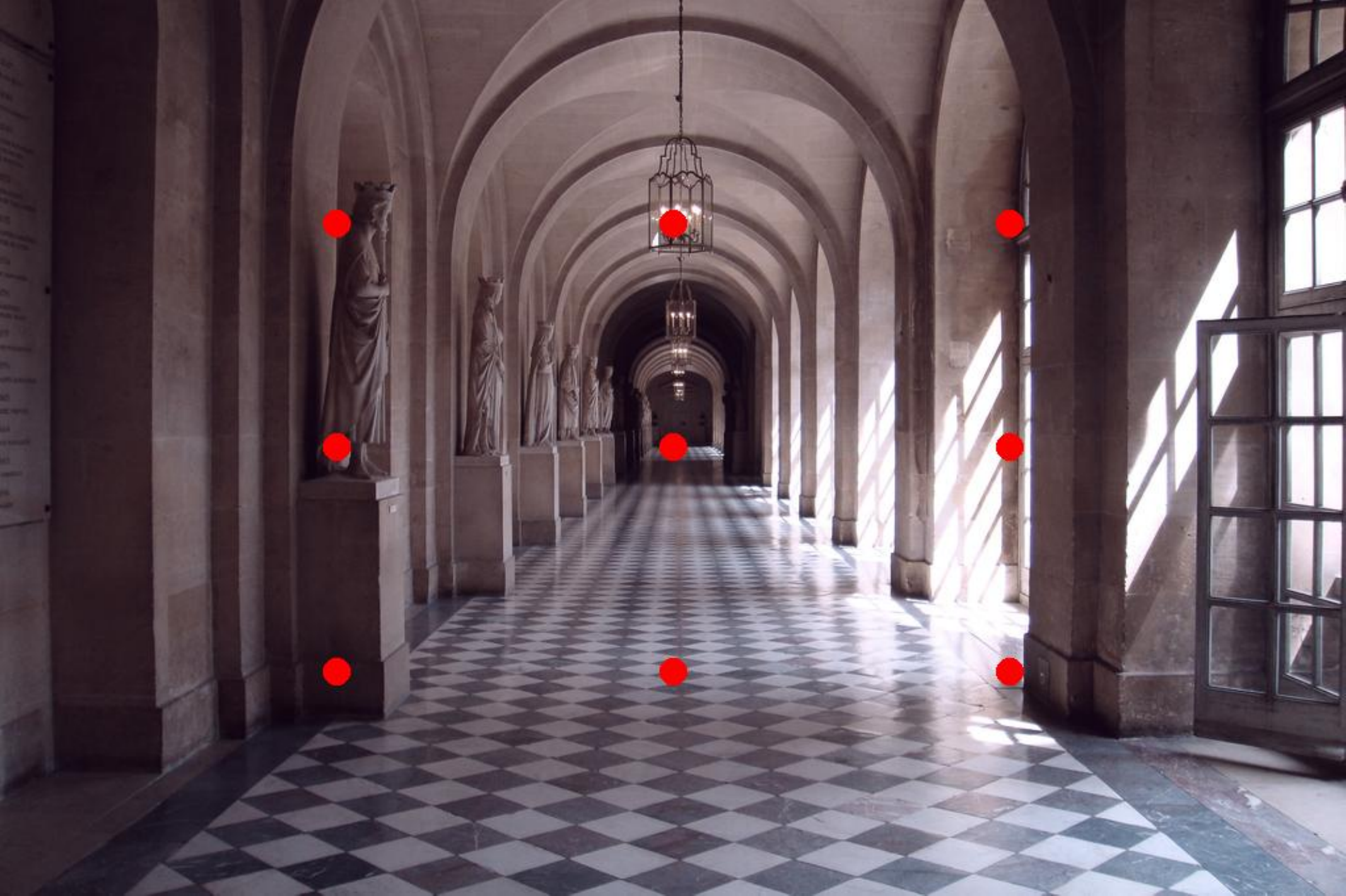}} & 
        \noindent\parbox[c]{0.092\textwidth}{\includegraphics[width=0.092\textwidth]{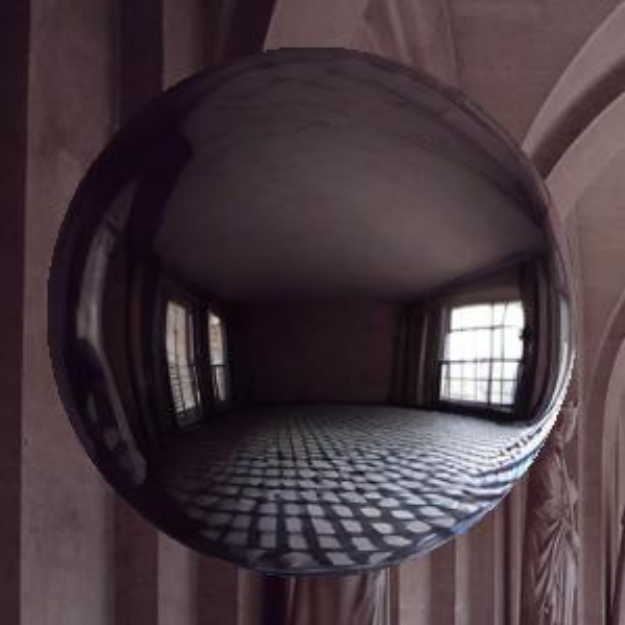}} & 
        \noindent\parbox[c]{0.092\textwidth}{\includegraphics[width=0.092\textwidth]{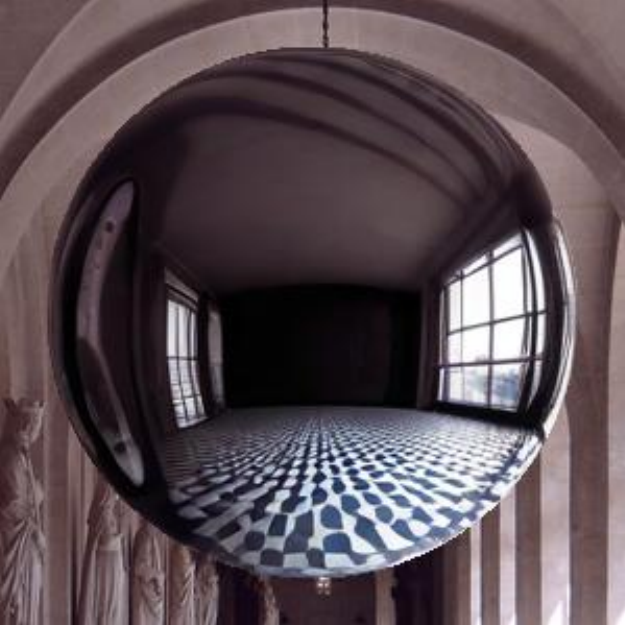}} & 
        \noindent\parbox[c]{0.092\textwidth}{\includegraphics[width=0.092\textwidth]{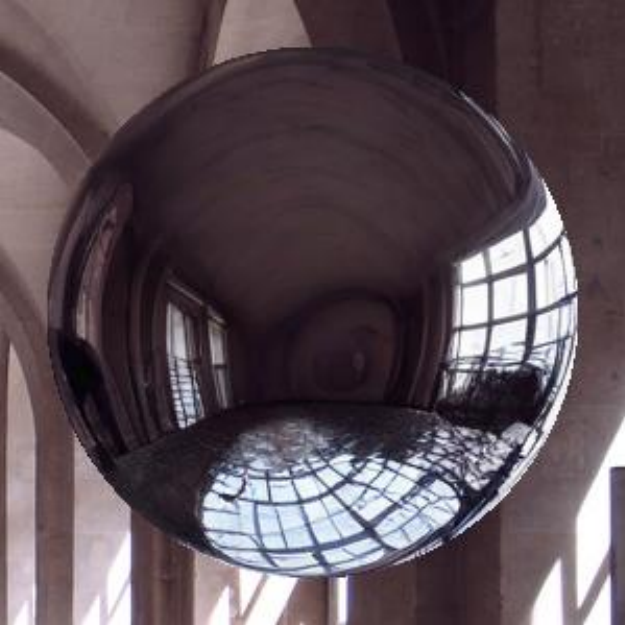}} & 
        \noindent\parbox[c]{0.092\textwidth}{\includegraphics[width=0.092\textwidth]{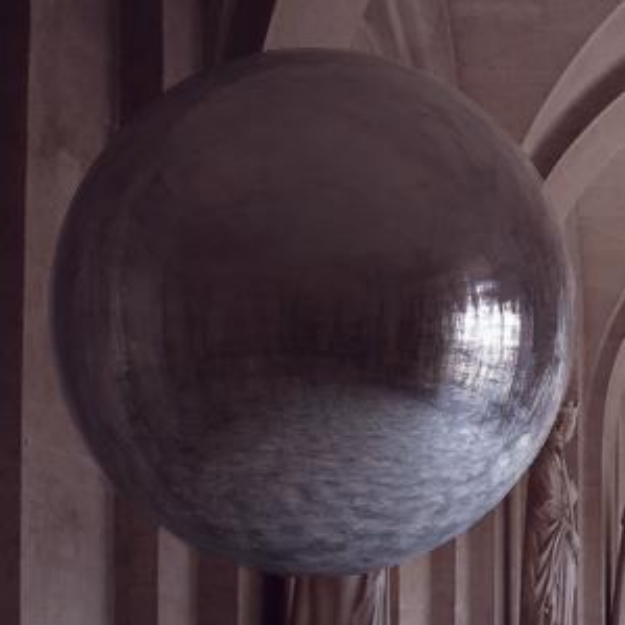}} & 
        \noindent\parbox[c]{0.092\textwidth}{\includegraphics[width=0.092\textwidth]{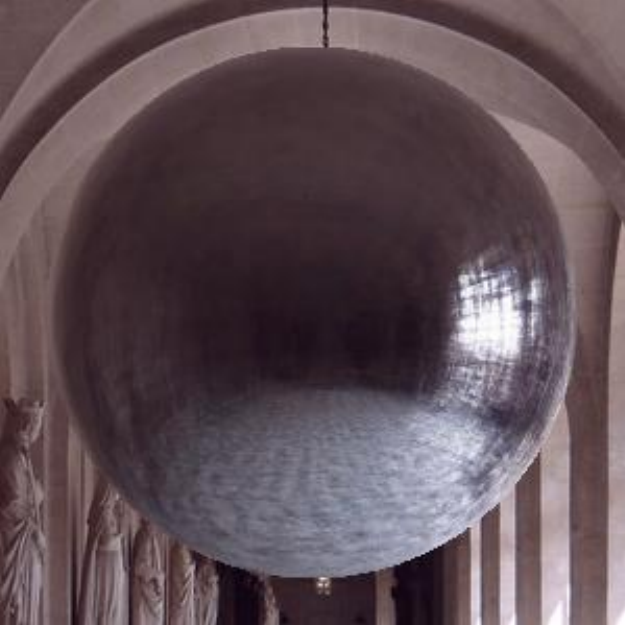}} & 
        \noindent\parbox[c]{0.092\textwidth}{\includegraphics[width=0.092\textwidth]{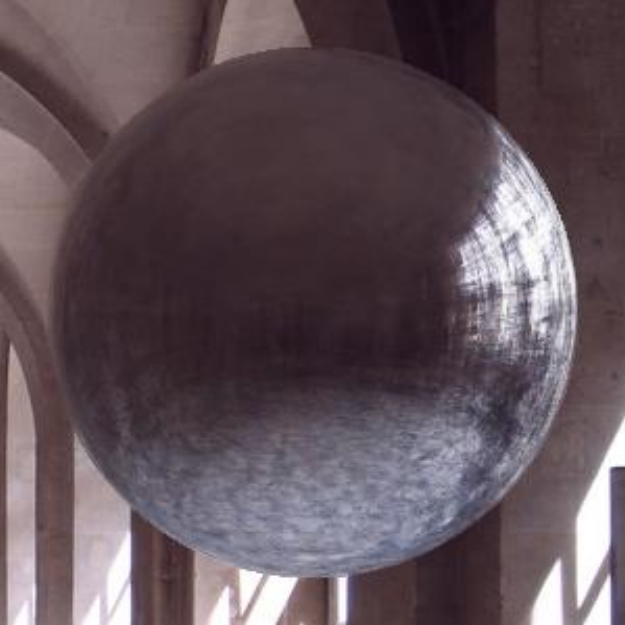}} & 
        \noindent\parbox[c]{0.092\textwidth}{\includegraphics[width=0.092\textwidth]{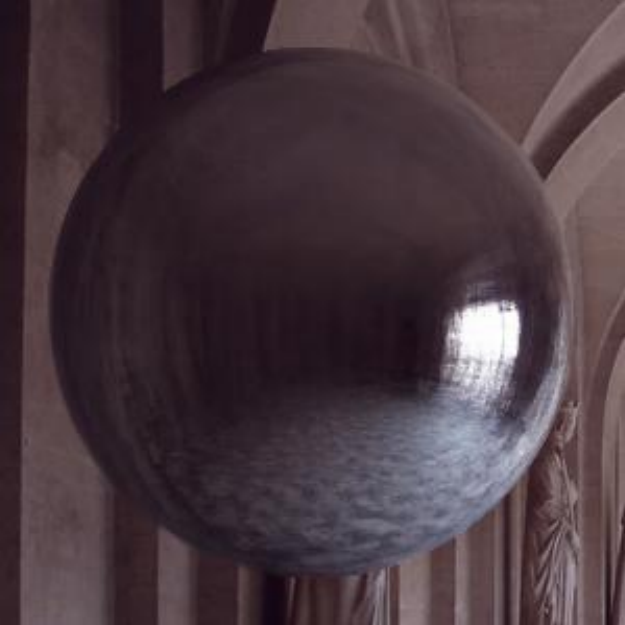}} & 
        \noindent\parbox[c]{0.092\textwidth}{\includegraphics[width=0.092\textwidth]{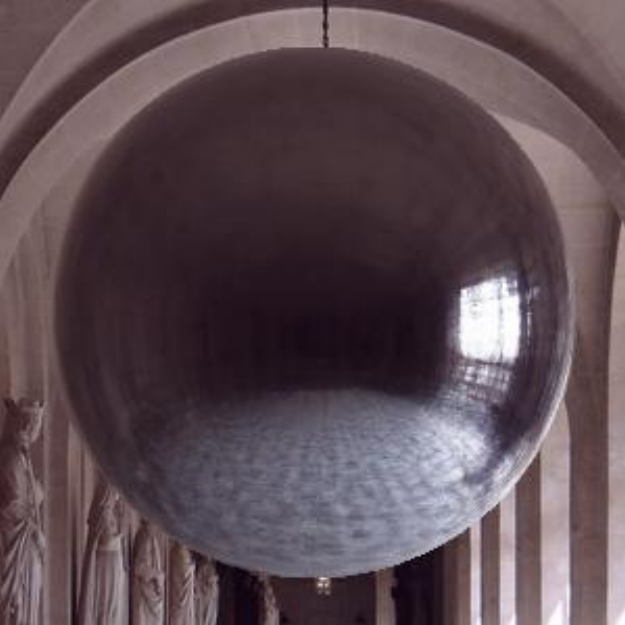}} & 
        \noindent\parbox[c]{0.092\textwidth}{\includegraphics[width=0.092\textwidth]{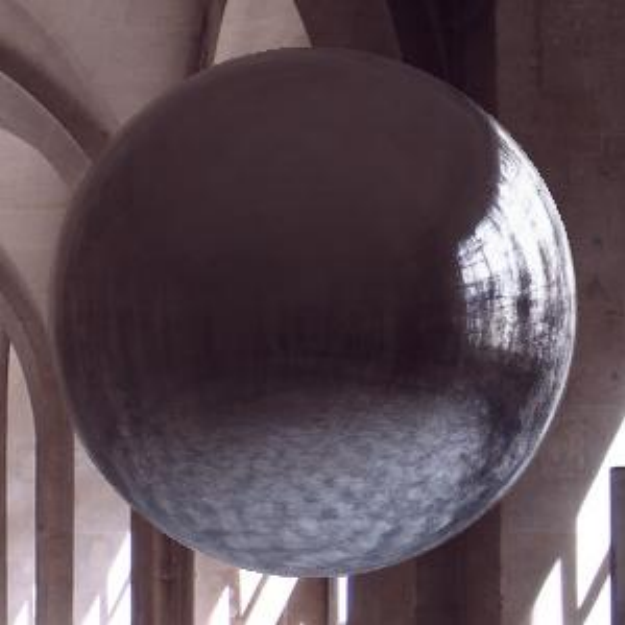}} & 
        
        \\

        & 
        \noindent\parbox[c]{0.092\textwidth}{\includegraphics[width=0.092\textwidth]{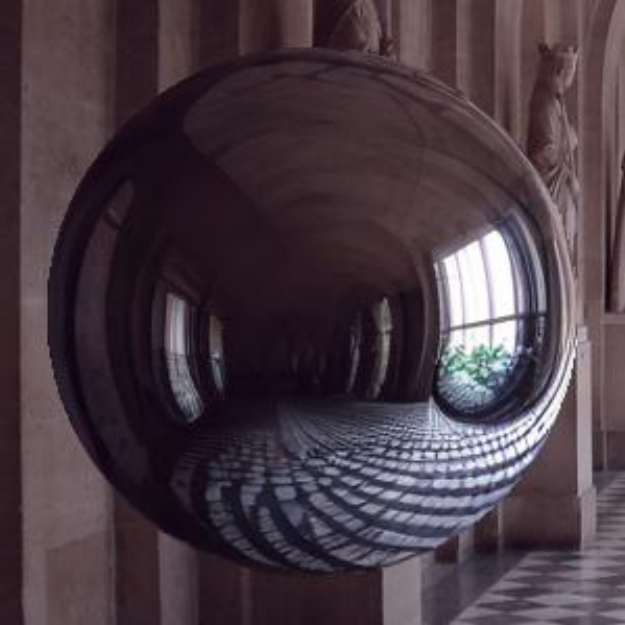}} & 
        \noindent\parbox[c]{0.092\textwidth}{\includegraphics[width=0.092\textwidth]{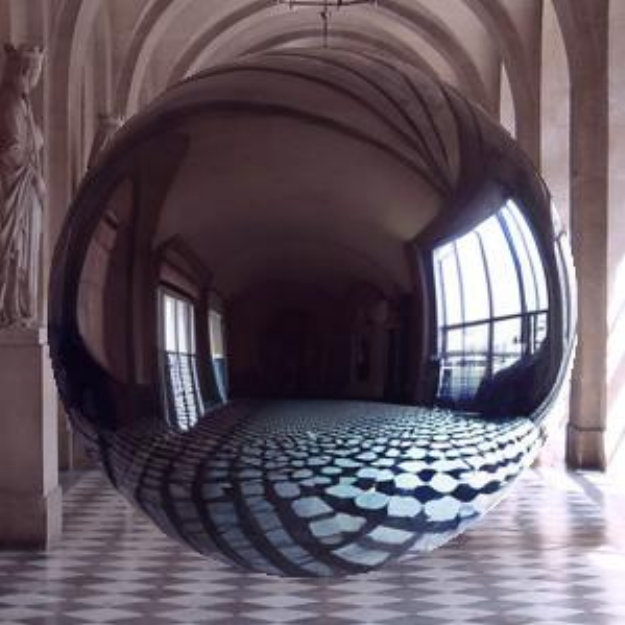}} & 
        \noindent\parbox[c]{0.092\textwidth}{\includegraphics[width=0.092\textwidth]{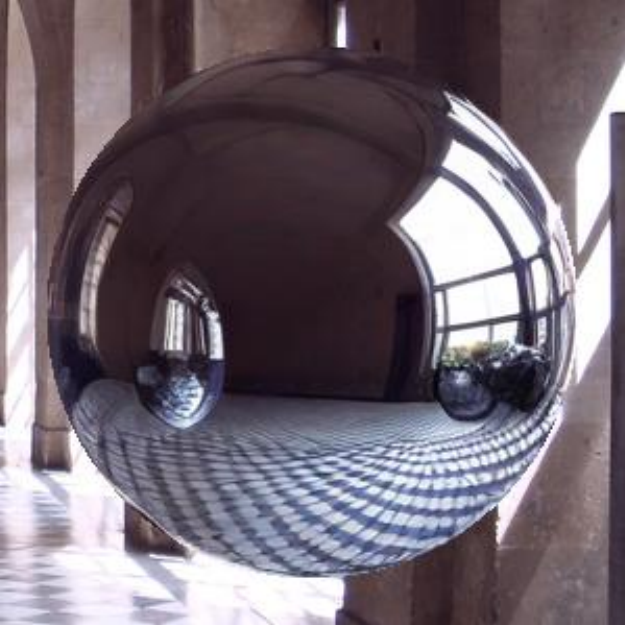}} & 
        \noindent\parbox[c]{0.092\textwidth}{\includegraphics[width=0.092\textwidth]{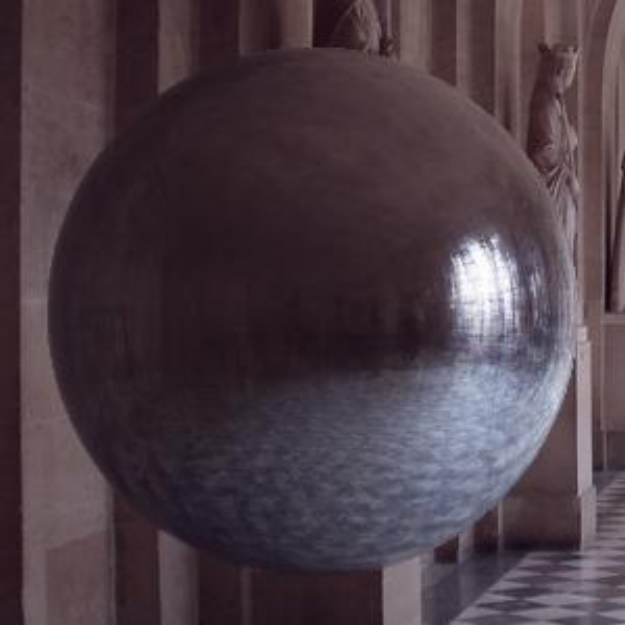}} & 
        \noindent\parbox[c]{0.092\textwidth}{\includegraphics[width=0.092\textwidth]{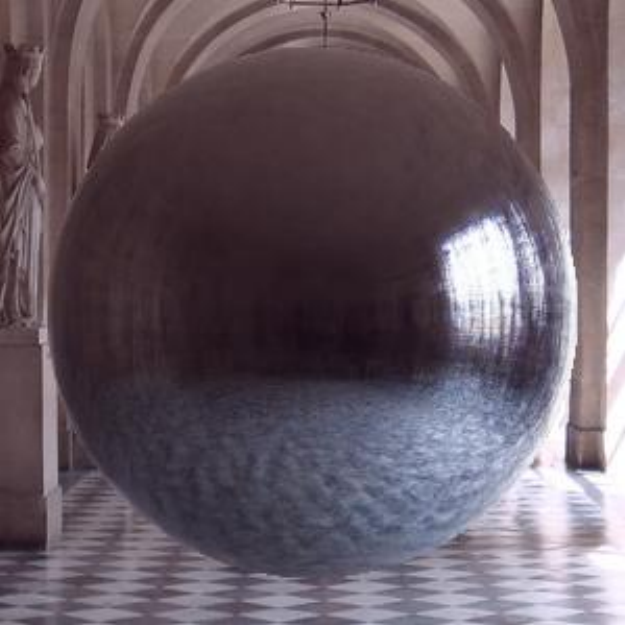}} & 
        \noindent\parbox[c]{0.092\textwidth}{\includegraphics[width=0.092\textwidth]{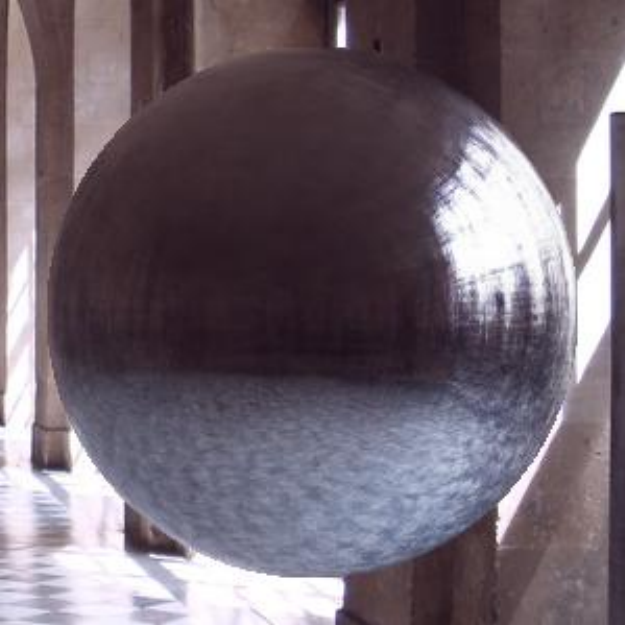}} & 
        \noindent\parbox[c]{0.092\textwidth}{\includegraphics[width=0.092\textwidth]{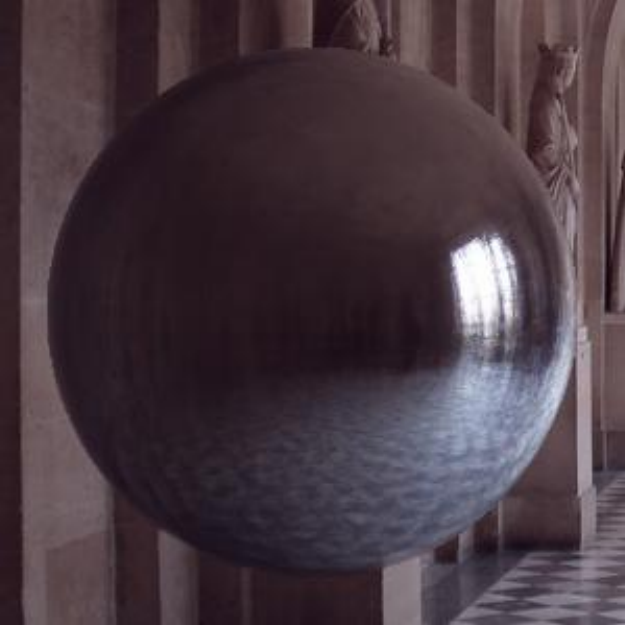}} & 
        \noindent\parbox[c]{0.092\textwidth}{\includegraphics[width=0.092\textwidth]{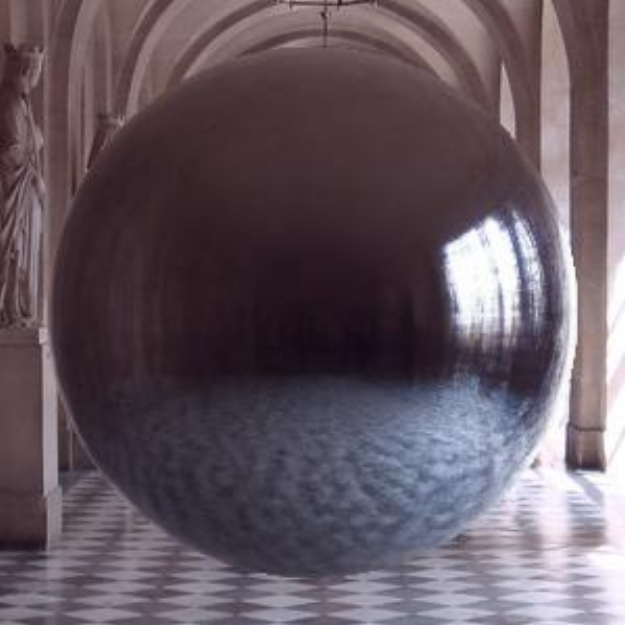}} & 
        \noindent\parbox[c]{0.092\textwidth}{\includegraphics[width=0.092\textwidth]{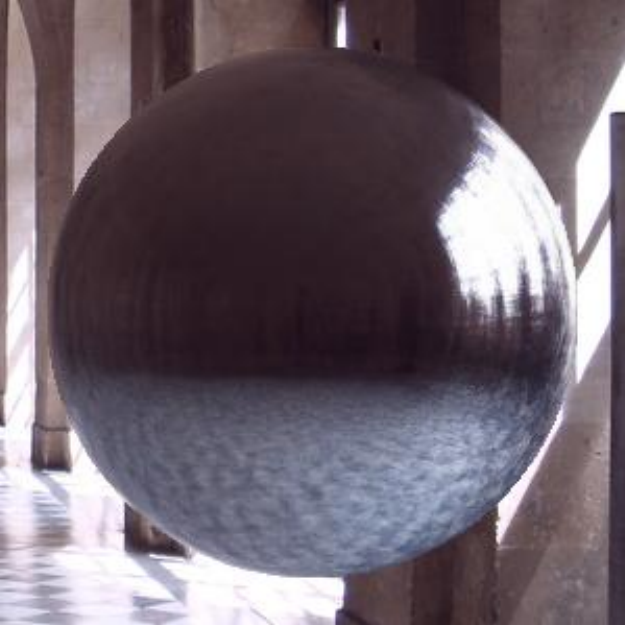}} & 
        
        \\

        & 
        \noindent\parbox[c]{0.092\textwidth}{\includegraphics[width=0.092\textwidth]{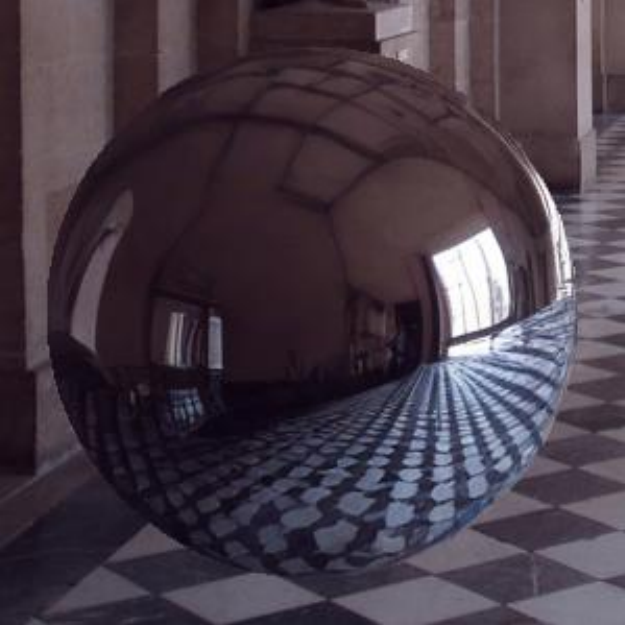}} & 
        \noindent\parbox[c]{0.092\textwidth}{\includegraphics[width=0.092\textwidth]{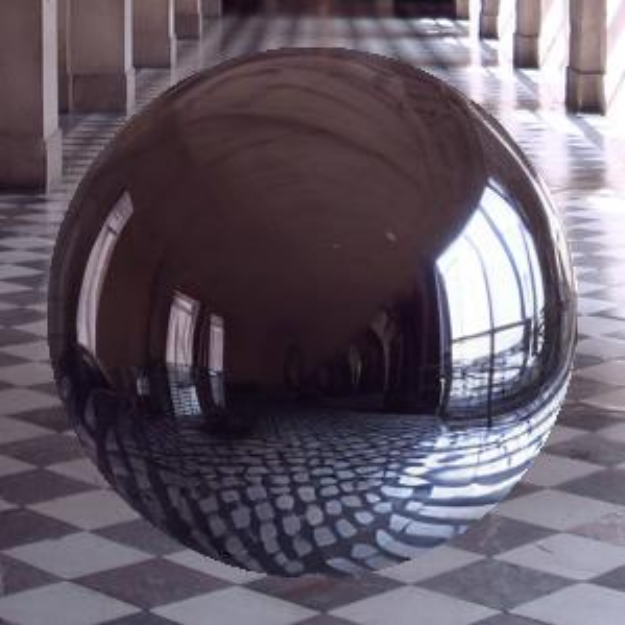}} & 
        \noindent\parbox[c]{0.092\textwidth}{\includegraphics[width=0.092\textwidth]{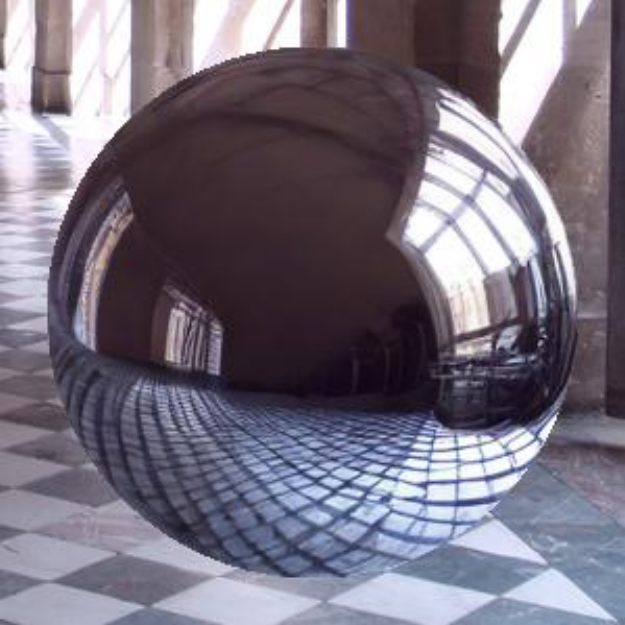}} & 
        \noindent\parbox[c]{0.092\textwidth}{\includegraphics[width=0.092\textwidth]{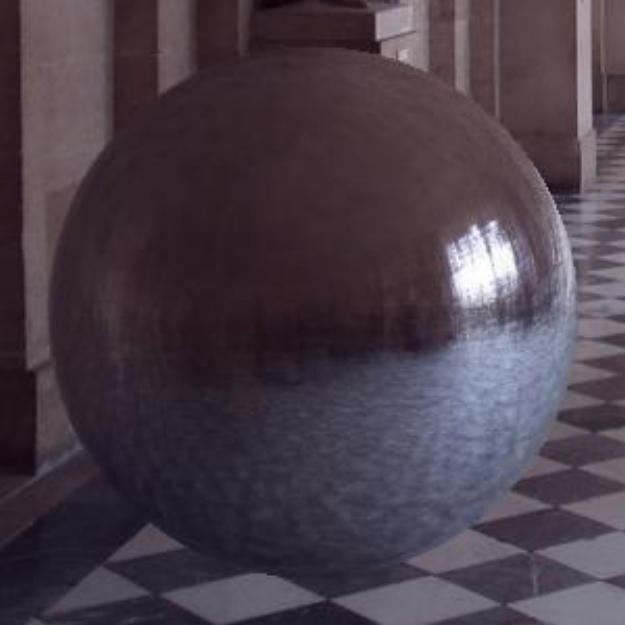}} & 
        \noindent\parbox[c]{0.092\textwidth}{\includegraphics[width=0.092\textwidth]{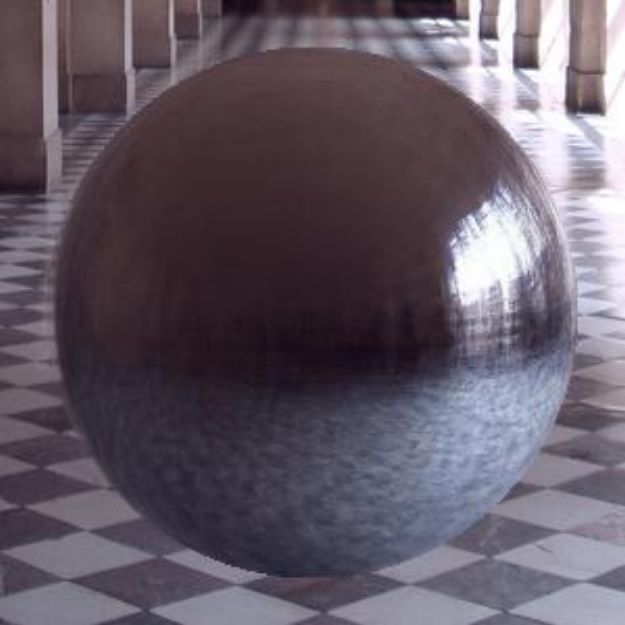}} & 
        \noindent\parbox[c]{0.092\textwidth}{\includegraphics[width=0.092\textwidth]{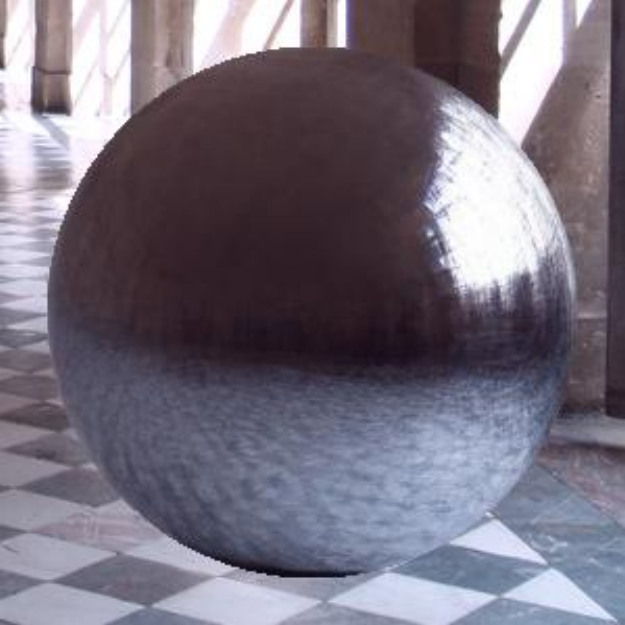}} & 
        \noindent\parbox[c]{0.092\textwidth}{\includegraphics[width=0.092\textwidth]{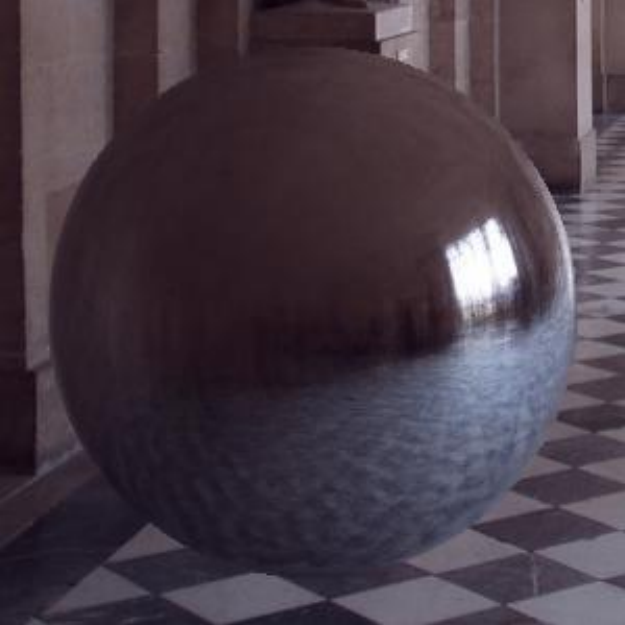}} & 
        \noindent\parbox[c]{0.092\textwidth}{\includegraphics[width=0.092\textwidth]{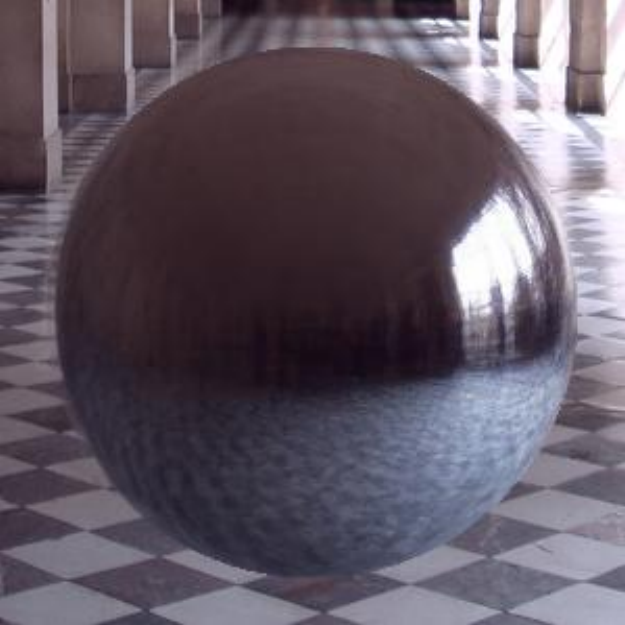}} & 
        \noindent\parbox[c]{0.092\textwidth}{\includegraphics[width=0.092\textwidth]{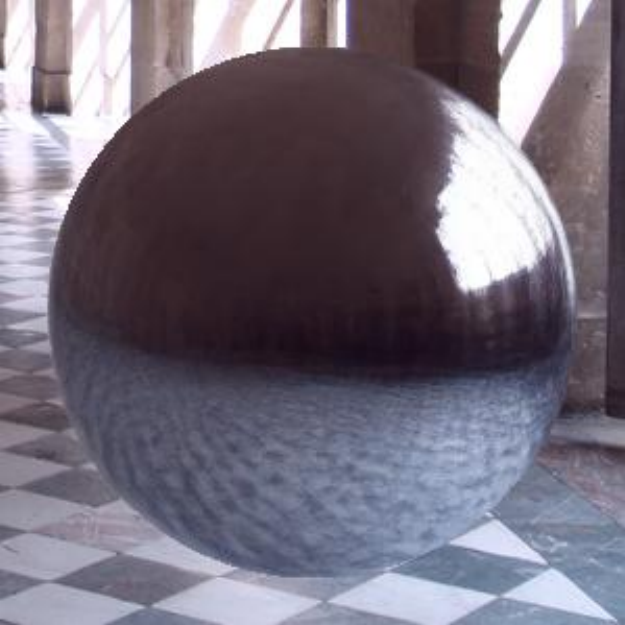}} & 
        
        \\
        
    \end{tabu}

    \smallskip
    \begin{tabu} to \textwidth {
        @{}
        c@{\hspace{2pt}}
        c@{\hspace{0.5pt}}
        c@{\hspace{0.5pt}}
        c@{\hspace{2pt}}
        c@{\hspace{0.5pt}}
        c@{\hspace{0.5pt}}
        c@{\hspace{2pt}}
        c@{\hspace{0.5pt}}
        c@{\hspace{0.5pt}}
        c@{\hspace{0.5pt}}
        c@{}
    }
        

        \noindent\parbox[c]{0.140\textwidth}{\includegraphics[width=0.140\textwidth]{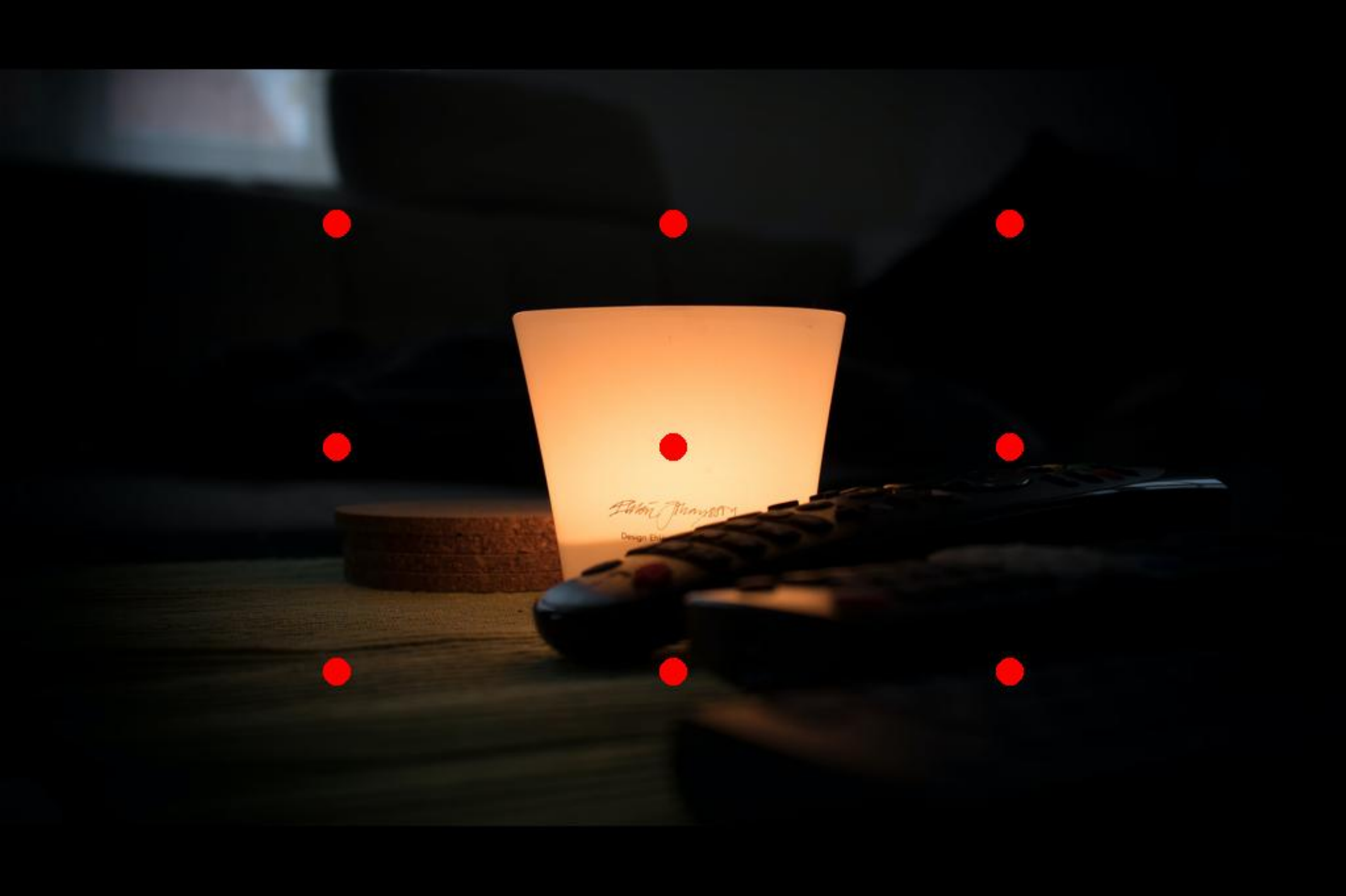}} & 
        \noindent\parbox[c]{0.092\textwidth}{\includegraphics[width=0.092\textwidth]{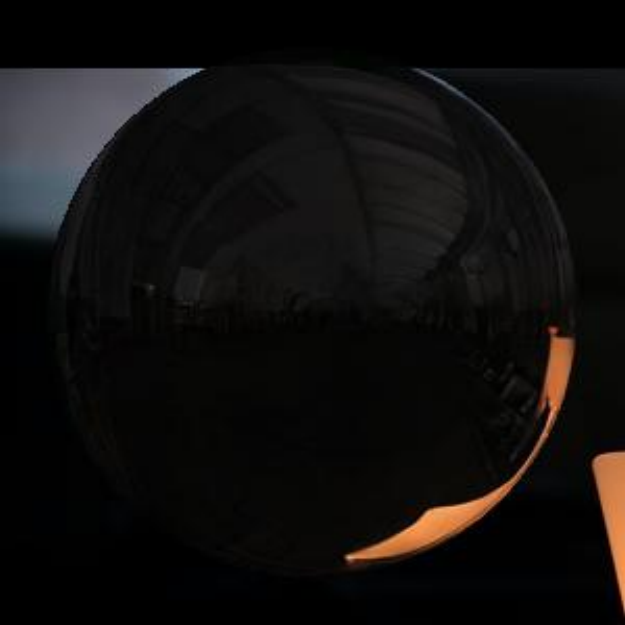}} & 
        \noindent\parbox[c]{0.092\textwidth}{\includegraphics[width=0.092\textwidth]{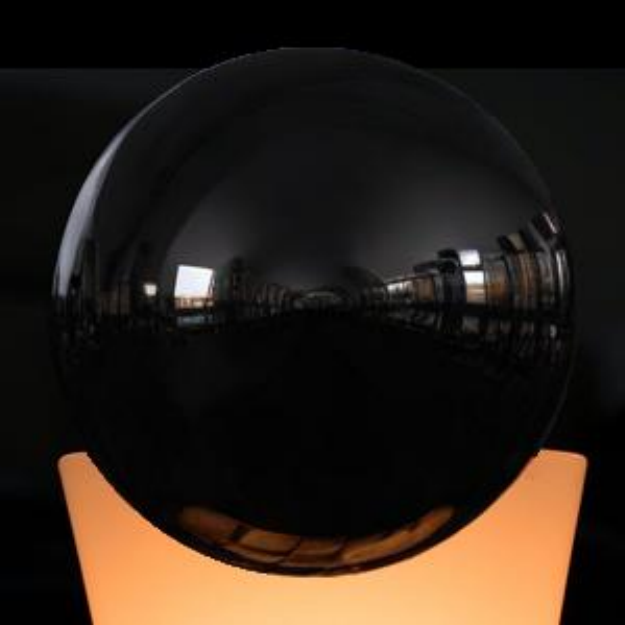}} & 
        \noindent\parbox[c]{0.092\textwidth}{\includegraphics[width=0.092\textwidth]{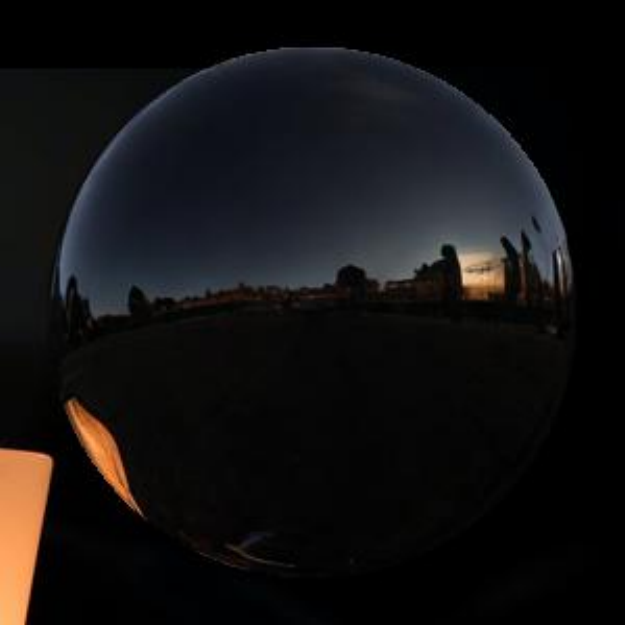}} & 
        \noindent\parbox[c]{0.092\textwidth}{\includegraphics[width=0.092\textwidth]{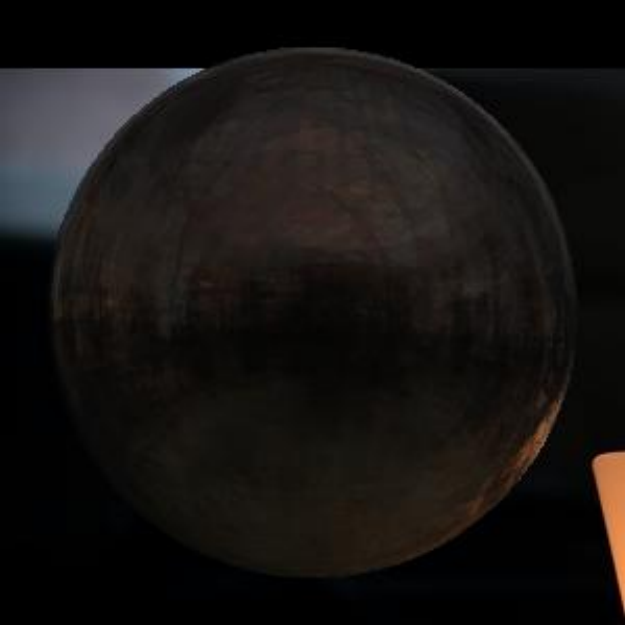}} & 
        \noindent\parbox[c]{0.092\textwidth}{\includegraphics[width=0.092\textwidth]{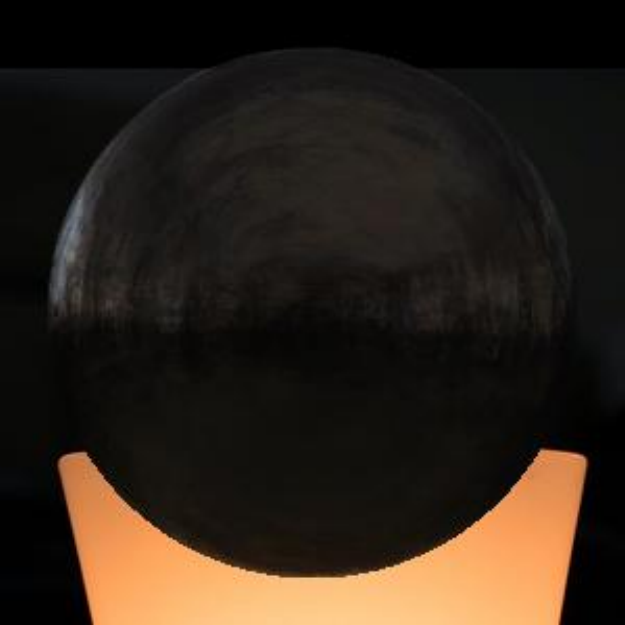}} & 
        \noindent\parbox[c]{0.092\textwidth}{\includegraphics[width=0.092\textwidth]{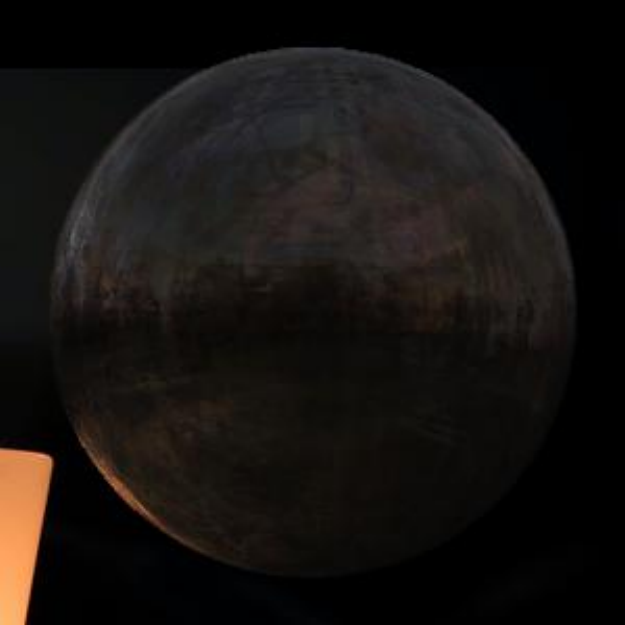}} & 
        \noindent\parbox[c]{0.092\textwidth}{\includegraphics[width=0.092\textwidth]{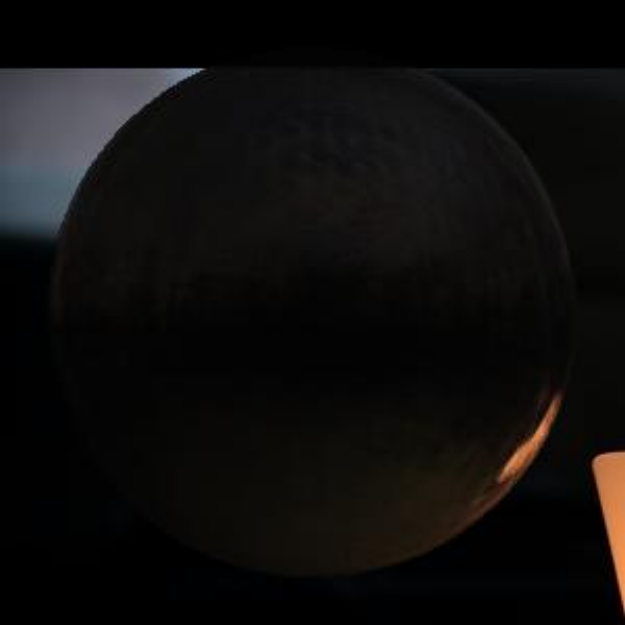}} & 
        \noindent\parbox[c]{0.092\textwidth}{\includegraphics[width=0.092\textwidth]{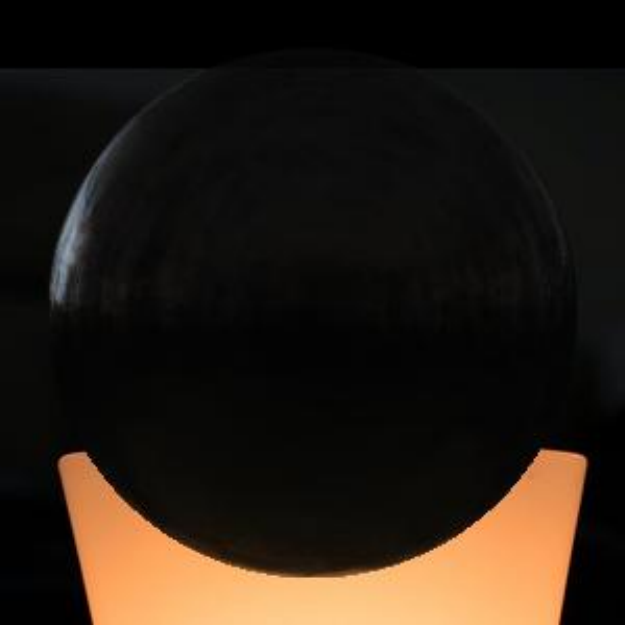}} & 
        \noindent\parbox[c]{0.092\textwidth}{\includegraphics[width=0.092\textwidth]{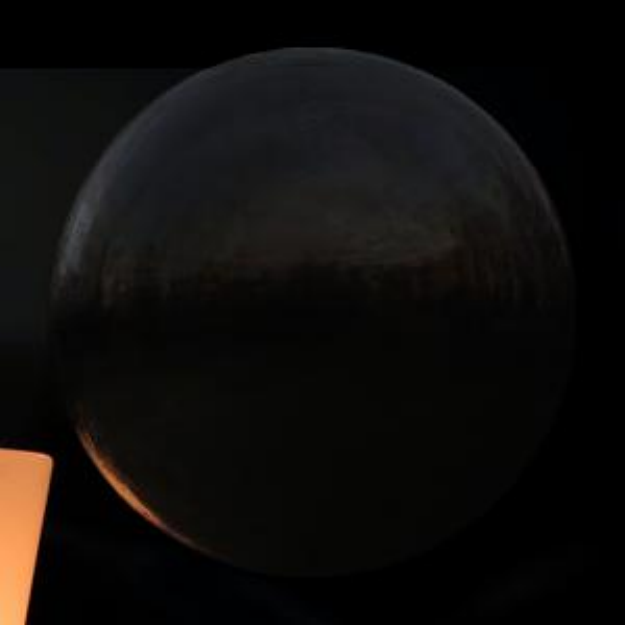}} & 
        
        \\

        & 
        \noindent\parbox[c]{0.092\textwidth}{\includegraphics[width=0.092\textwidth]{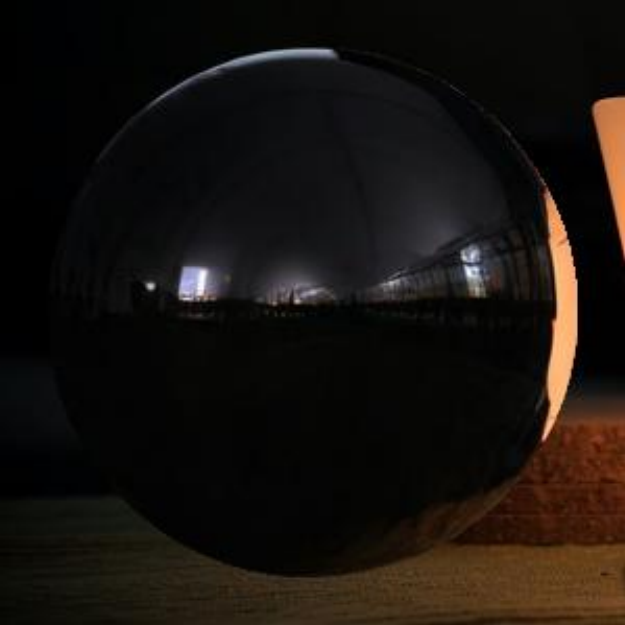}} & 
        \noindent\parbox[c]{0.092\textwidth}{\includegraphics[width=0.092\textwidth]{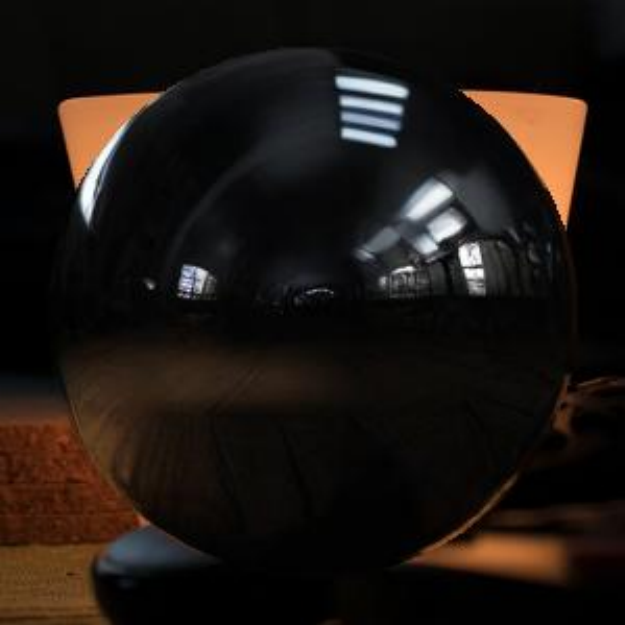}} & 
        \noindent\parbox[c]{0.092\textwidth}{\includegraphics[width=0.092\textwidth]{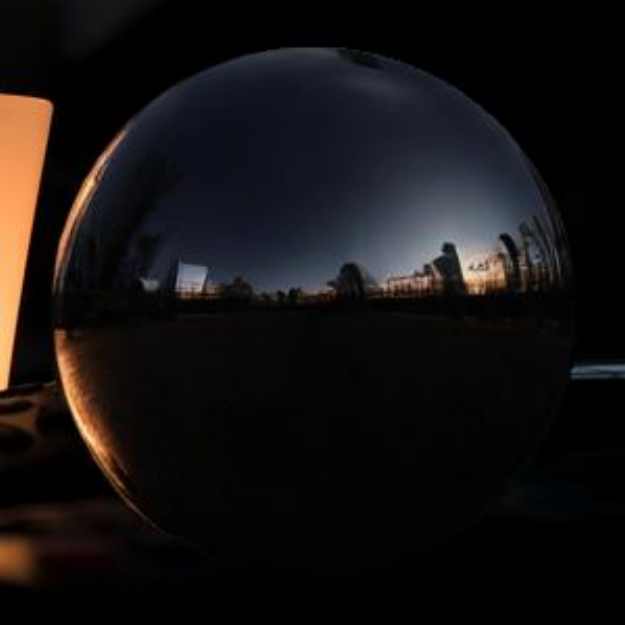}} & 
        \noindent\parbox[c]{0.092\textwidth}{\includegraphics[width=0.092\textwidth]{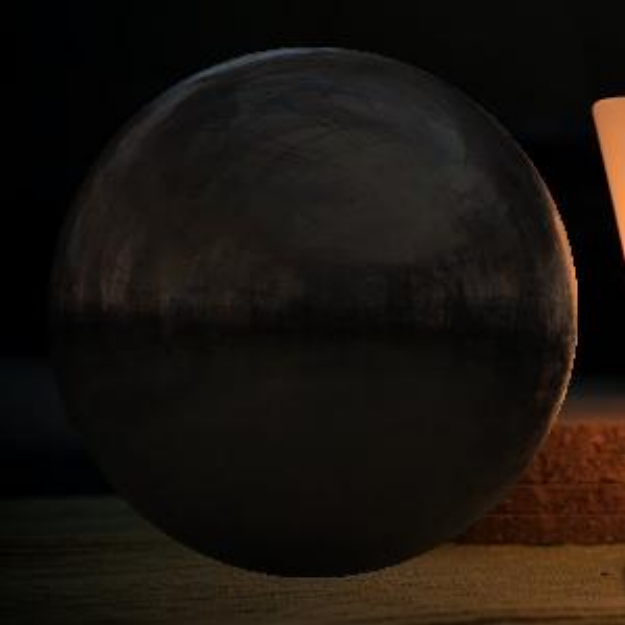}} & 
        \noindent\parbox[c]{0.092\textwidth}{\includegraphics[width=0.092\textwidth]{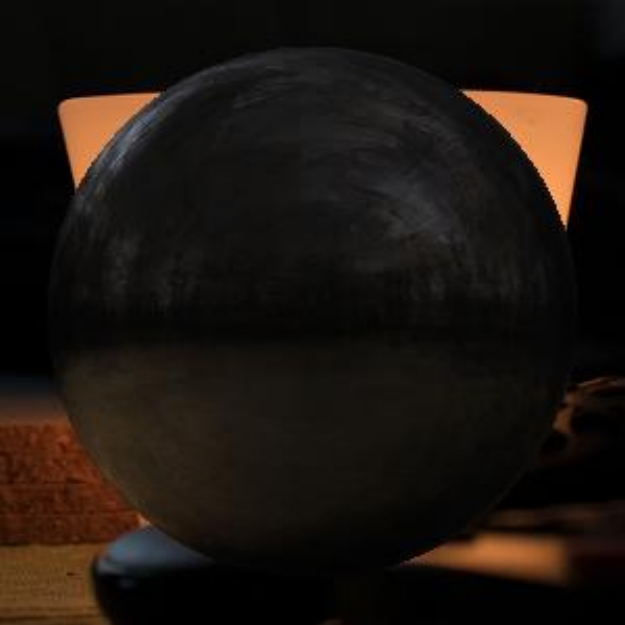}} & 
        \noindent\parbox[c]{0.092\textwidth}{\includegraphics[width=0.092\textwidth]{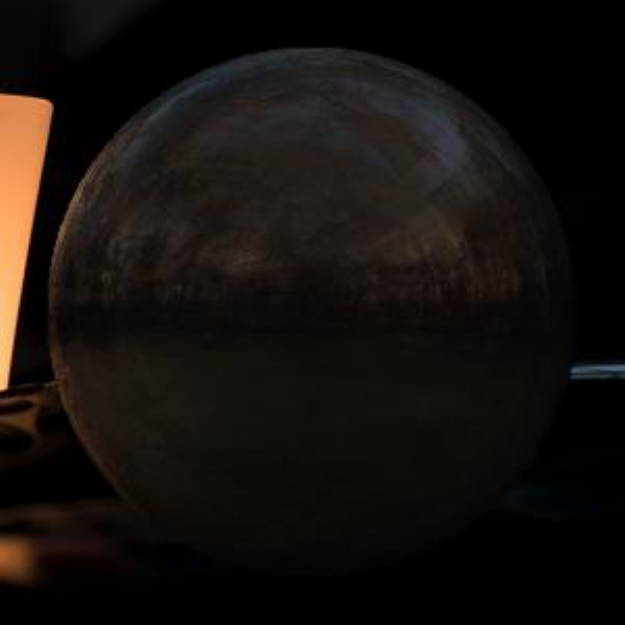}} & 
        \noindent\parbox[c]{0.092\textwidth}{\includegraphics[width=0.092\textwidth]{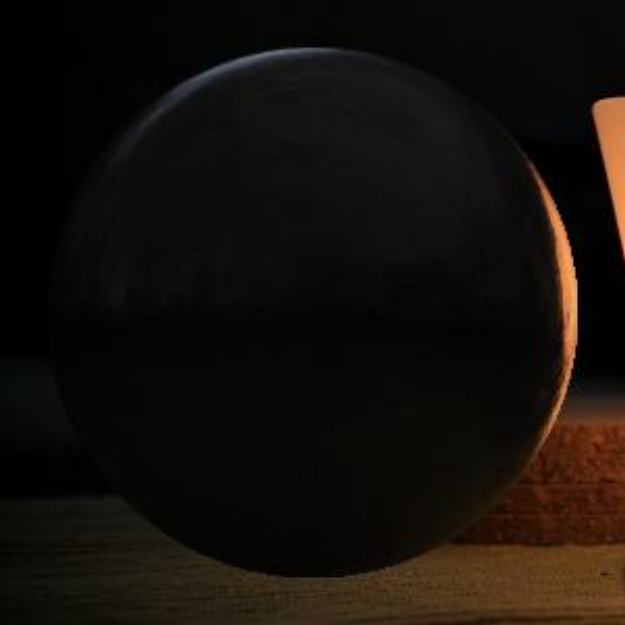}} & 
        \noindent\parbox[c]{0.092\textwidth}{\includegraphics[width=0.092\textwidth]{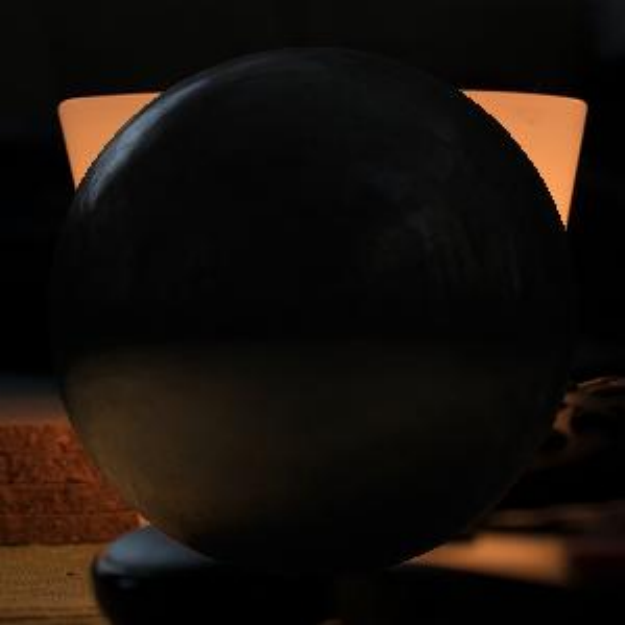}} & 
        \noindent\parbox[c]{0.092\textwidth}{\includegraphics[width=0.092\textwidth]{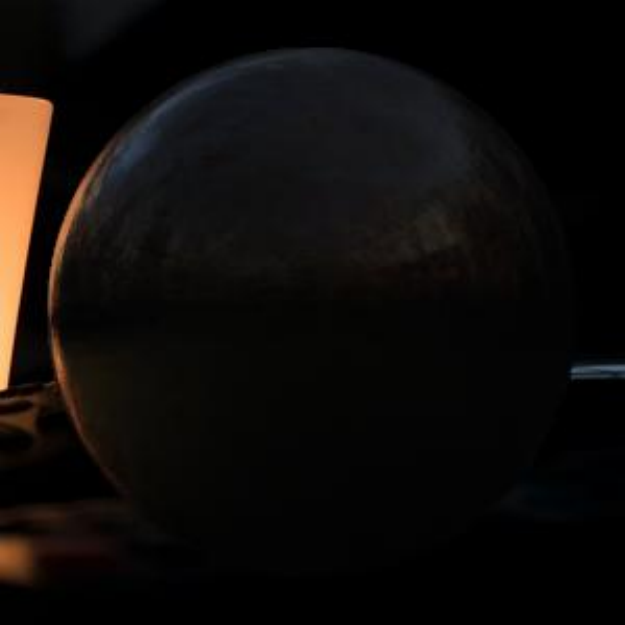}} & 
        
        \\

        & 
        \noindent\parbox[c]{0.092\textwidth}{\includegraphics[width=0.092\textwidth]{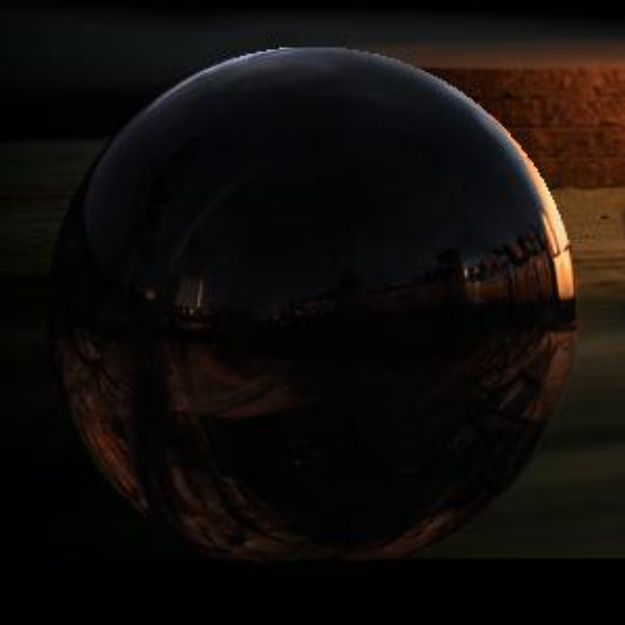}} & 
        \noindent\parbox[c]{0.092\textwidth}{\includegraphics[width=0.092\textwidth]{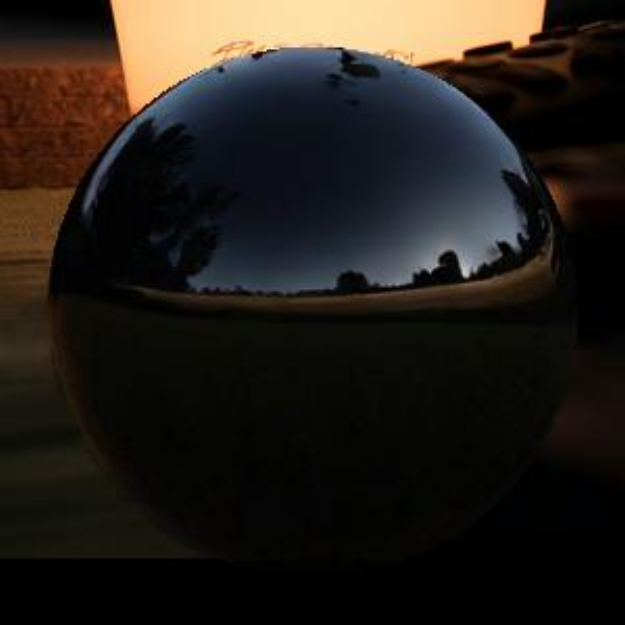}} & 
        \noindent\parbox[c]{0.092\textwidth}{\includegraphics[width=0.092\textwidth]{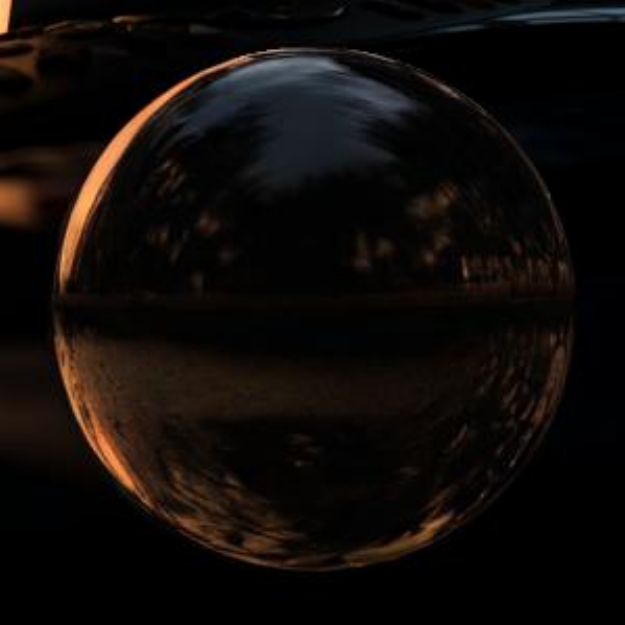}} & 
        \noindent\parbox[c]{0.092\textwidth}{\includegraphics[width=0.092\textwidth]{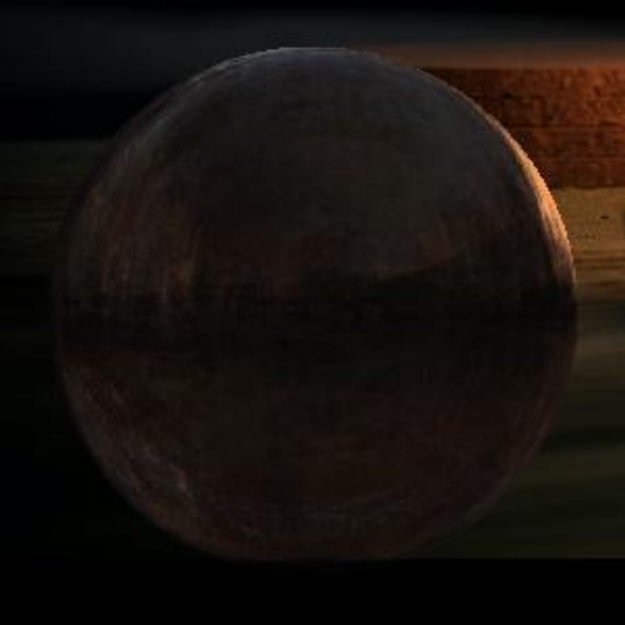}} & 
        \noindent\parbox[c]{0.092\textwidth}{\includegraphics[width=0.092\textwidth]{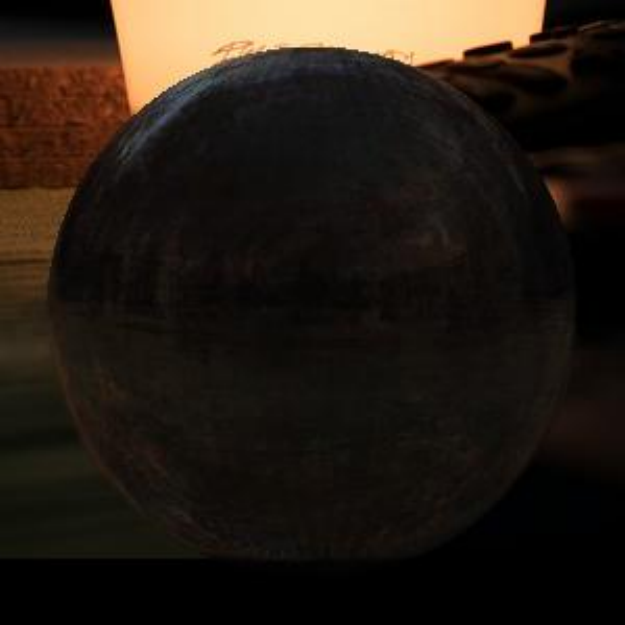}} & 
        \noindent\parbox[c]{0.092\textwidth}{\includegraphics[width=0.092\textwidth]{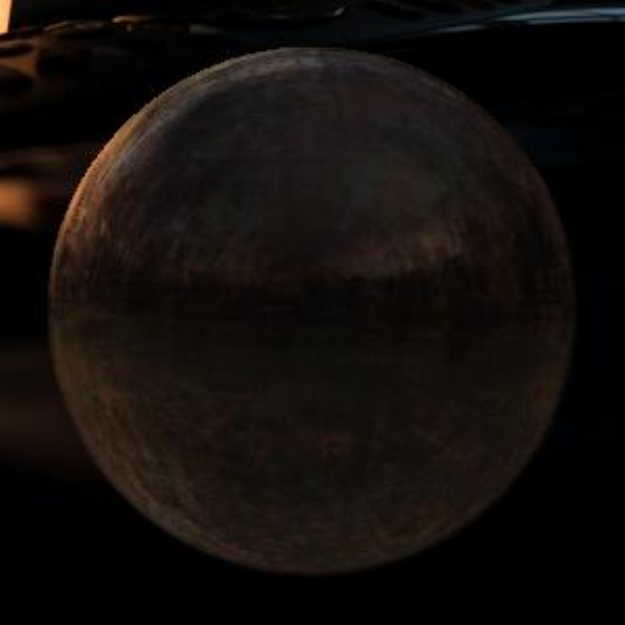}} & 
        \noindent\parbox[c]{0.092\textwidth}{\includegraphics[width=0.092\textwidth]{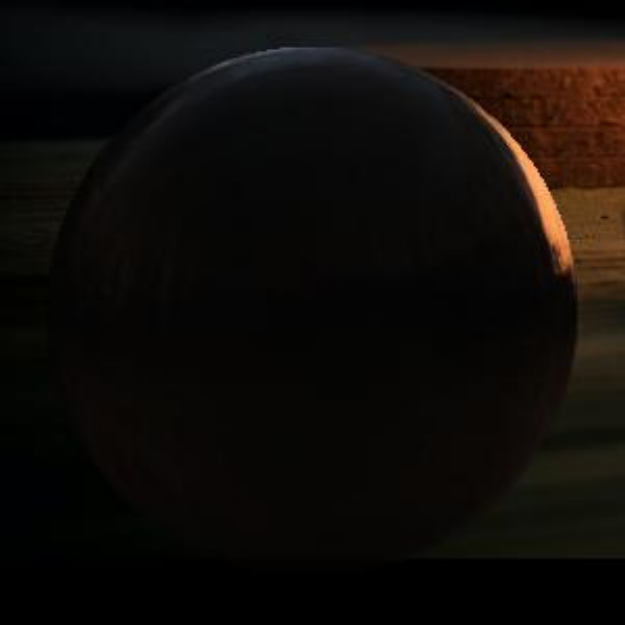}} & 
        \noindent\parbox[c]{0.092\textwidth}{\includegraphics[width=0.092\textwidth]{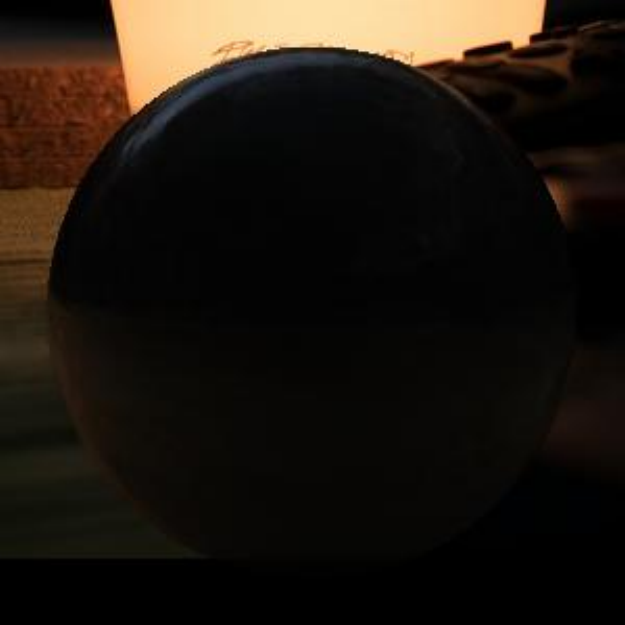}} & 
        \noindent\parbox[c]{0.092\textwidth}{\includegraphics[width=0.092\textwidth]{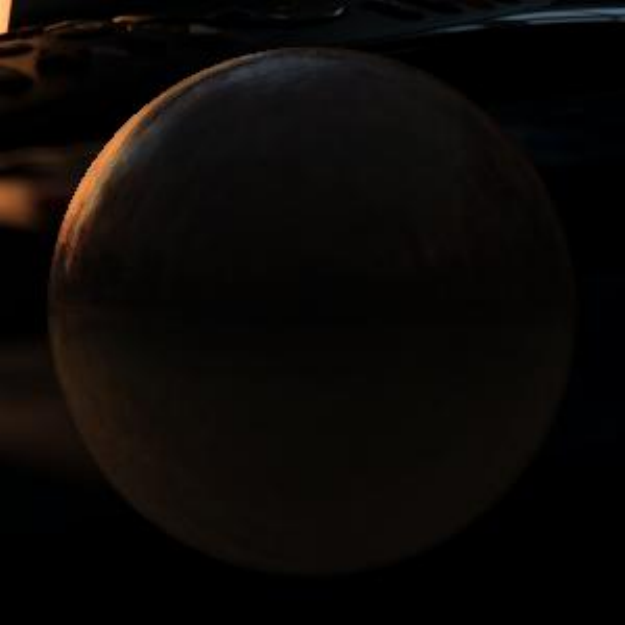}} & 
        
        \\
        
    \end{tabu}

    \caption{We show the spatially varying effects of painting a chrome ball at nine different locations, specified by red dots in the input images.
    For each input image, we present predictions from a random seed and median balls at the 1\textsuperscript{st} and 2\textsuperscript{nd} iterations. The effects can be seen in the changes of the curvature the horizon line, the size of the window, and the position of the light reflected from the lamp. These effects are more apparent in median balls as the number of iterations increases.
    } 
    
    
    
    
    \label{fig:aba_spatial_varying}
\end{figure*}

In this work, we inpaint a chrome ball in the input's center to represent global lighting in the scene and do not model any spatially varying effects by assuming orthographic projection. Nonetheless, our preliminary study suggests that the 
output from our inpainting pipeline does change according to where the chrome ball is inpatined, as illustrated in Figure \ref{fig:aba_spatial_varying}. This behavior can be leveraged for spatially-varying light estimation. 
To correctly infer spatially-varying, omnidirectional lighting, one needs to also infer the scene geometry, the depth of the inpainted chrome ball and camera parameters such as the focal length from the input image. These problems are interesting areas for future work.

\tabulinesep=0.1pt


\section{Virtual Object Insertion}
\begin{figure*}[!t]
    \centering

        \begin{tabu} to \textwidth {
        @{}
        c@{\hspace{2pt}}
        c@{\hspace{2pt}}
        c@{\hspace{2pt}}
    }

        \multicolumn{1}{c}{\shortstack{ Input object }}
        & 
        \multicolumn{1}{c}{\shortstack{ Before relighting}}
        &
        \multicolumn{1}{c}{\shortstack{ After relighting}}
        \\ 
        \multirow{ 2}{*}{\noindent\parbox[c]{0.3\textwidth}{\includegraphics[height=0.3\textwidth]{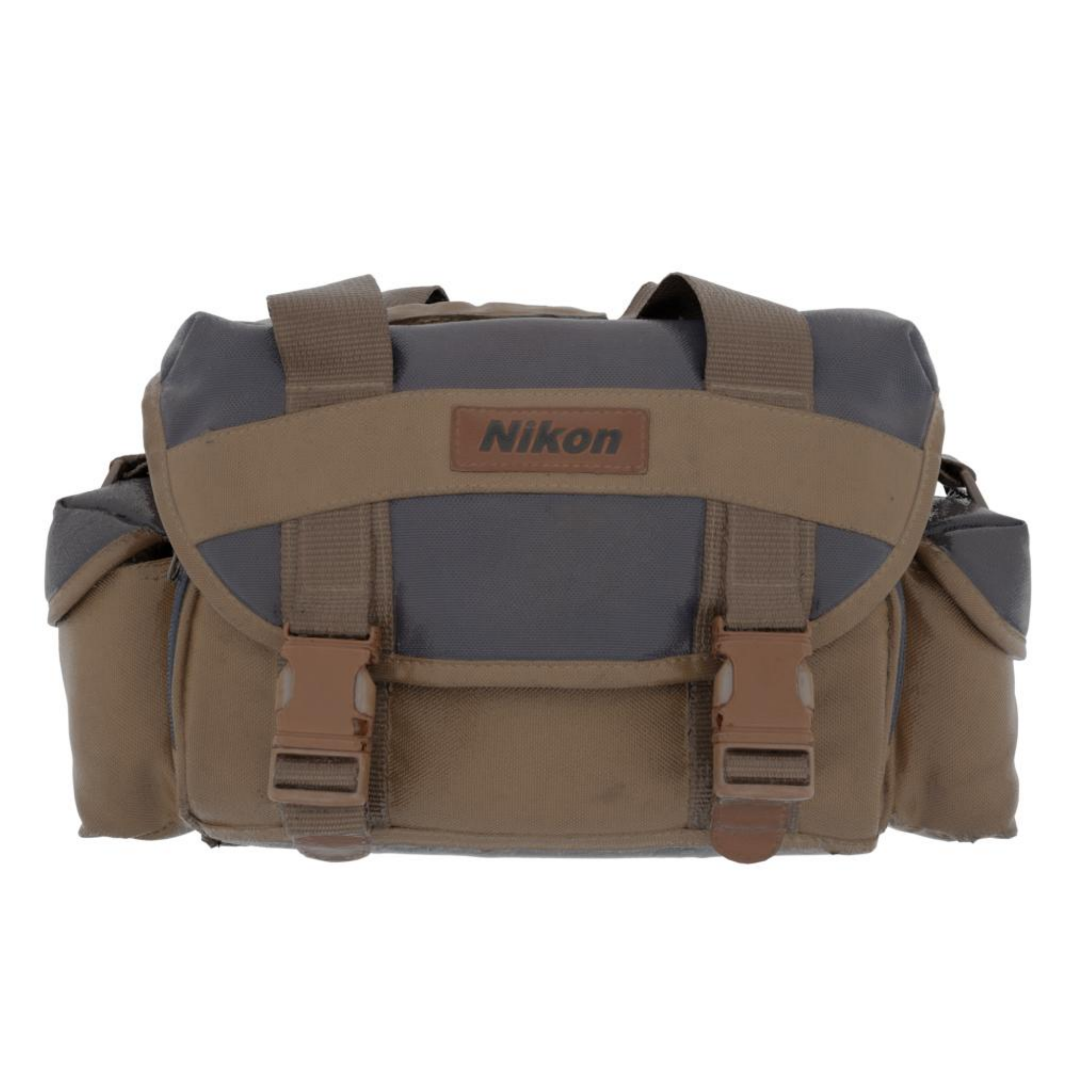}}} &
        \noindent\parbox[c]{0.3\textwidth}{\includegraphics[height=0.3\textwidth]{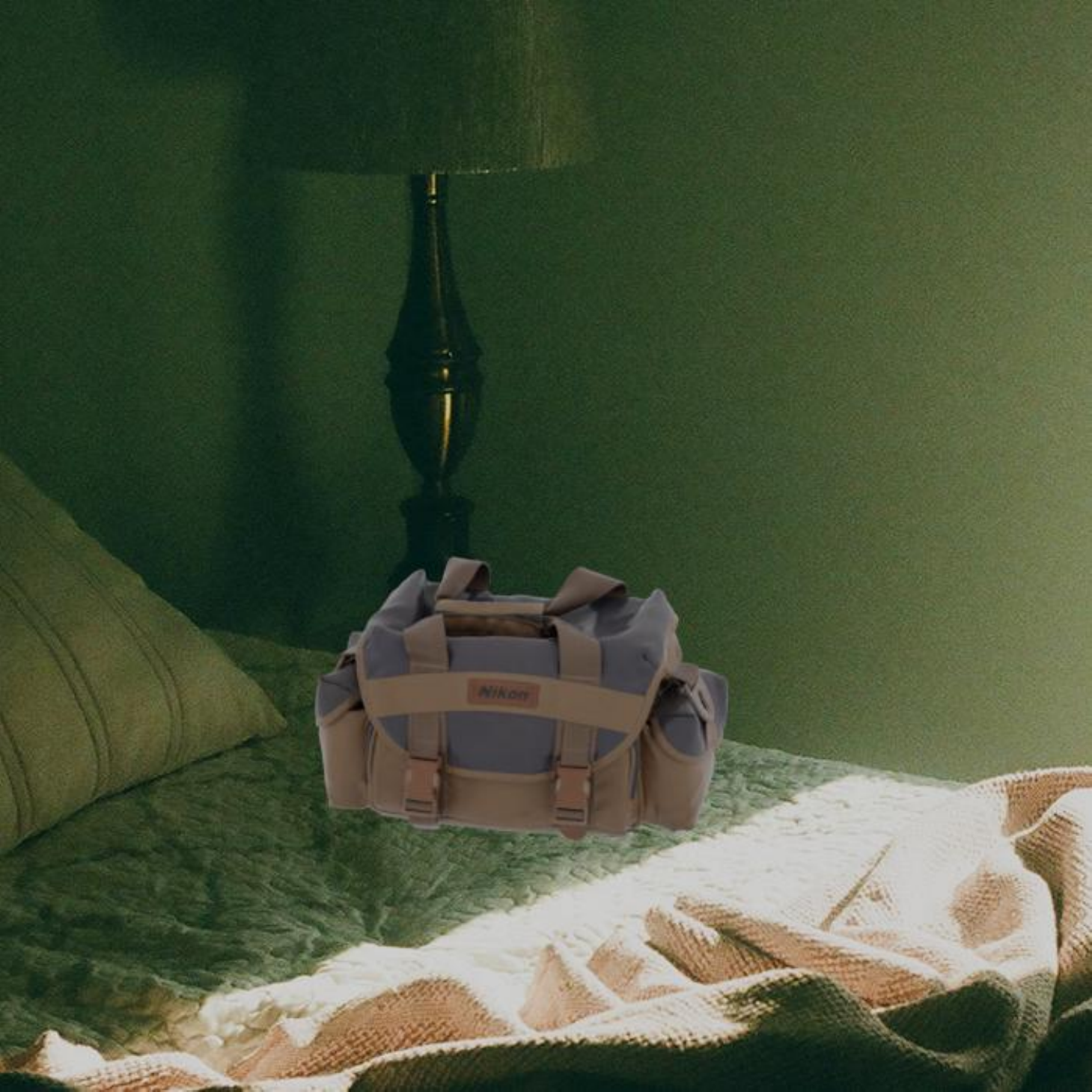}} &
        \noindent\parbox[c]{0.3\textwidth}{\includegraphics[height=0.3\textwidth]{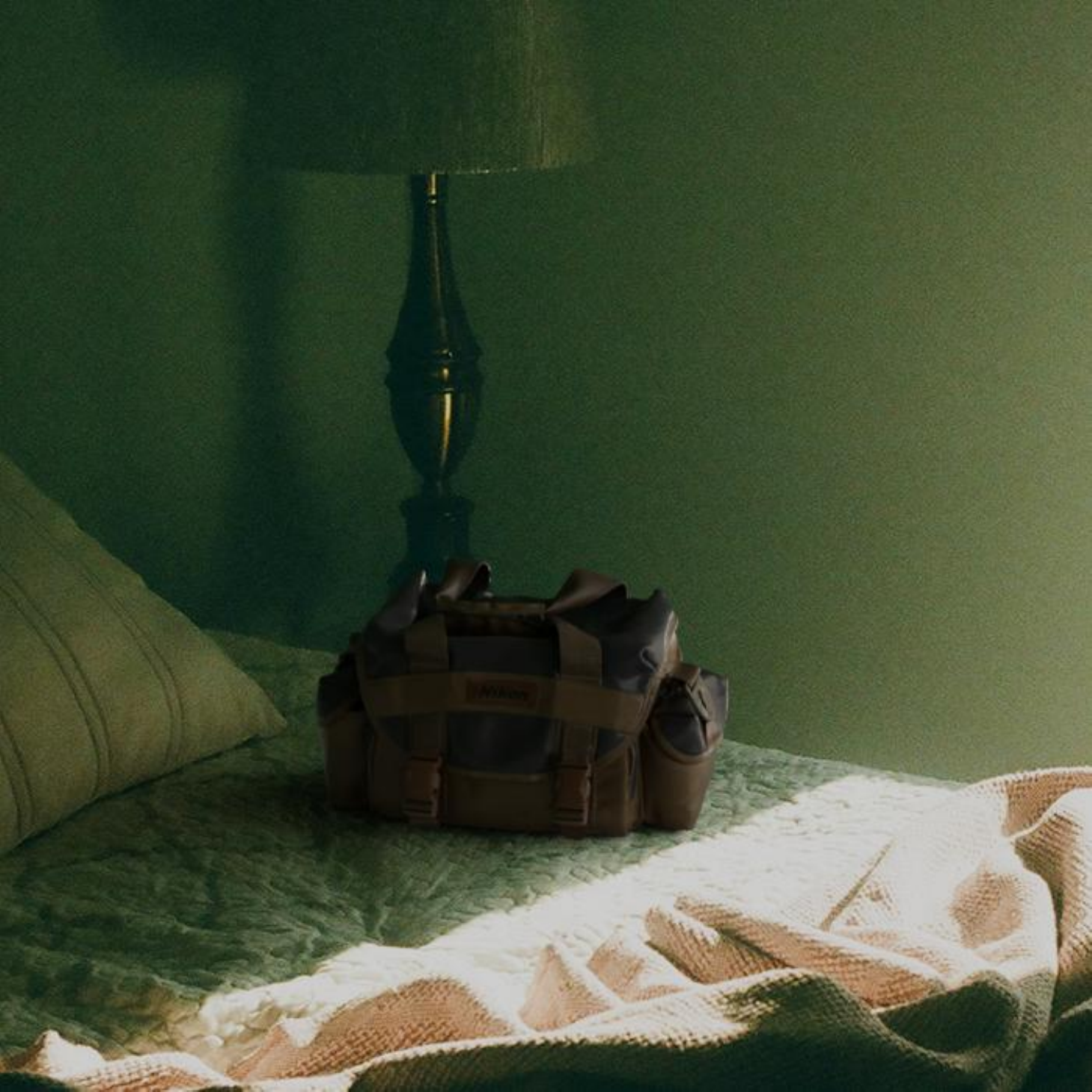}}
        \\
        &
        \noindent\parbox[c]{0.3\textwidth}{\includegraphics[height=0.3\textwidth]{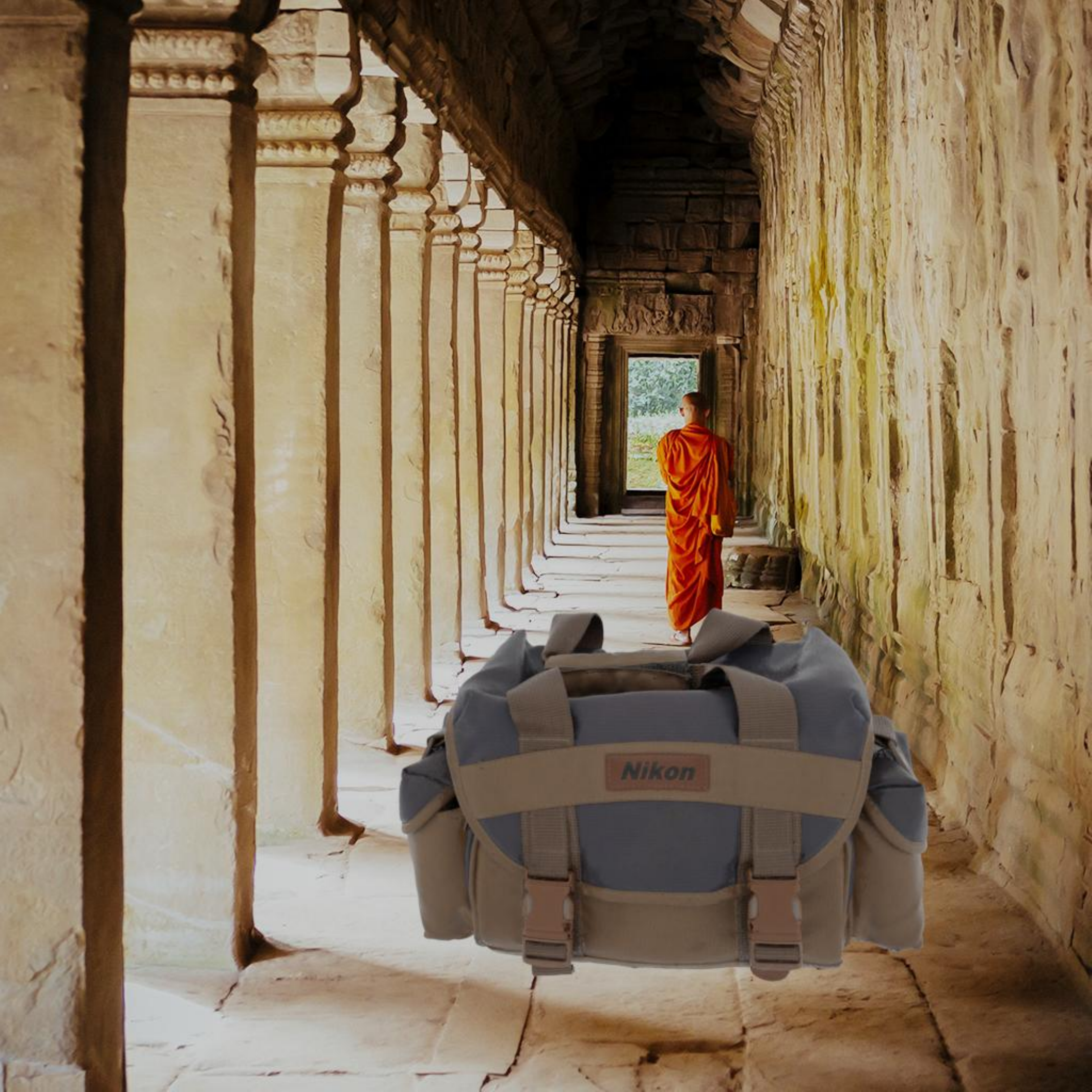}} &
        \noindent\parbox[c]{0.3\textwidth}{\includegraphics[height=0.3\textwidth]{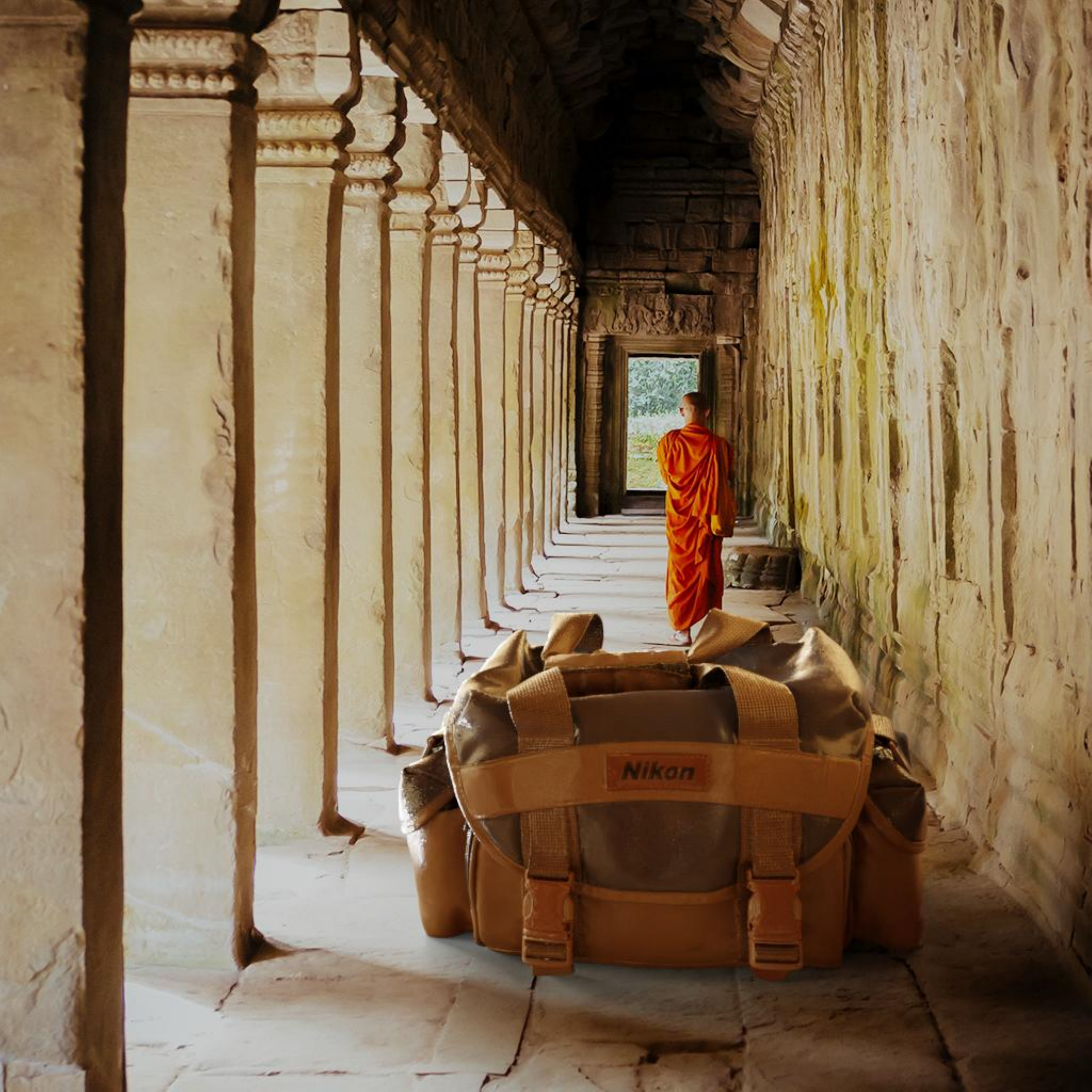}}
        \\
        \multirow{ 2}{*}{\noindent\parbox[c]{0.3\textwidth}{\includegraphics[height=0.3\textwidth]{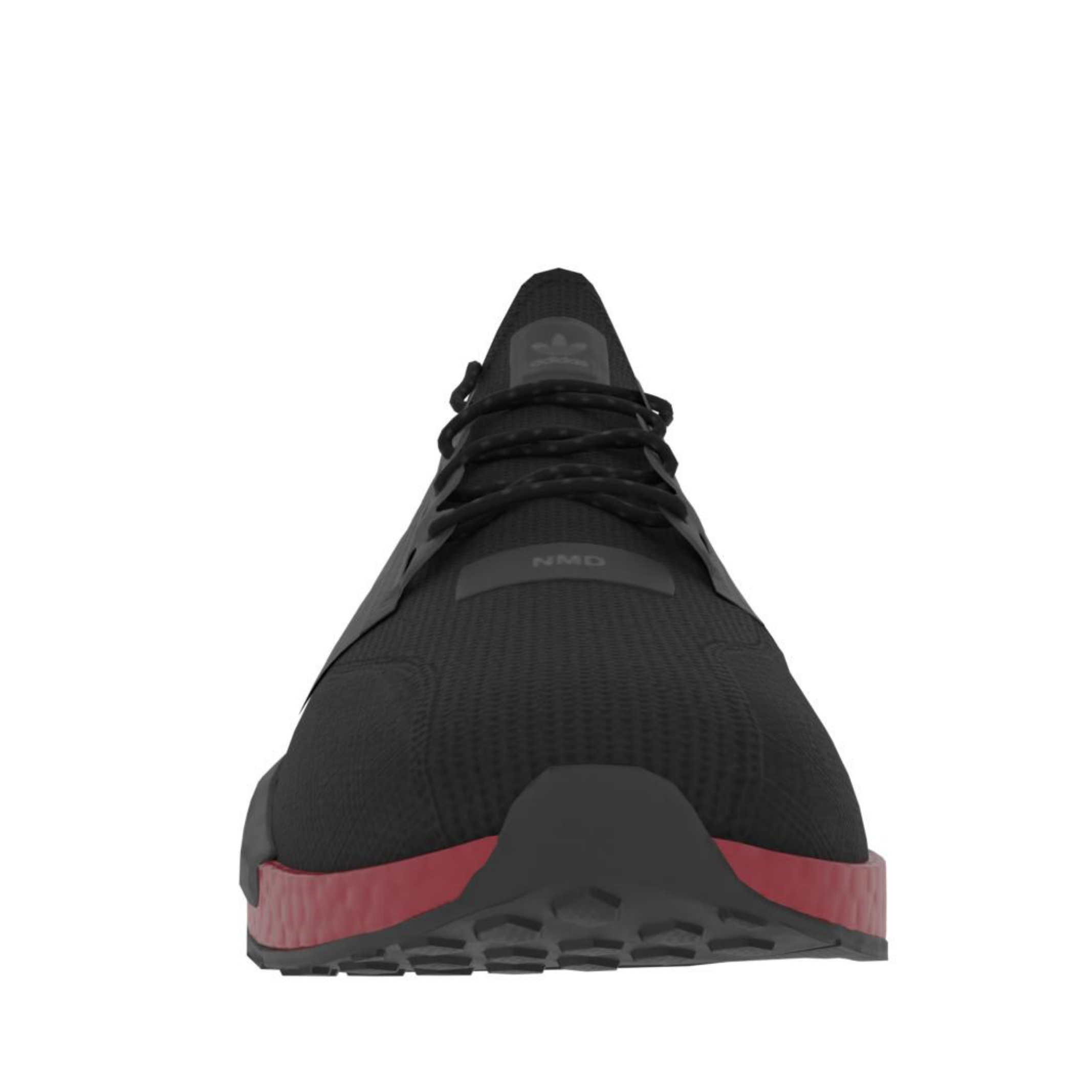}}} &
        \noindent\parbox[c]{0.3\textwidth}{\includegraphics[height=0.3\textwidth]{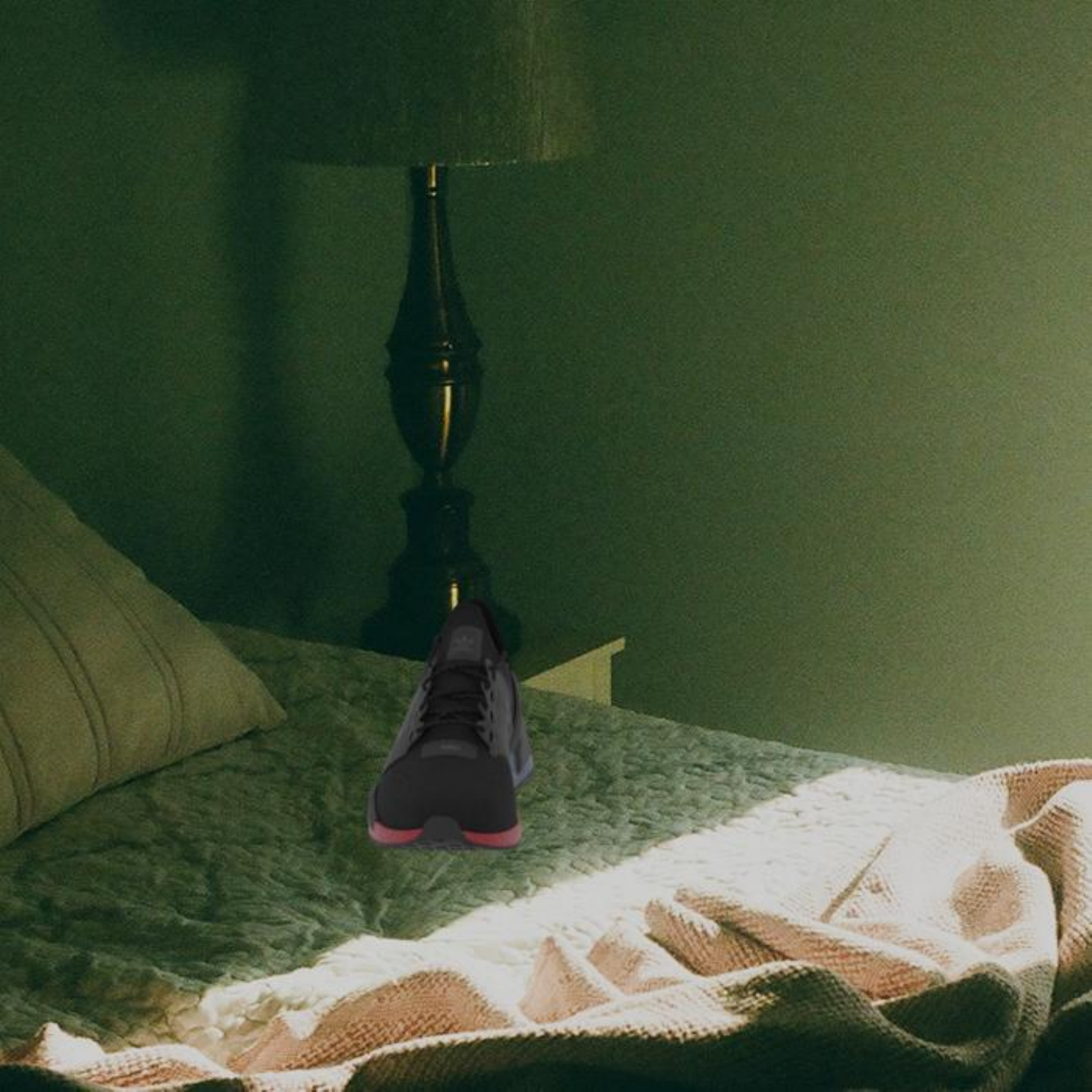}} &
        \noindent\parbox[c]{0.3\textwidth}{\includegraphics[height=0.3\textwidth]{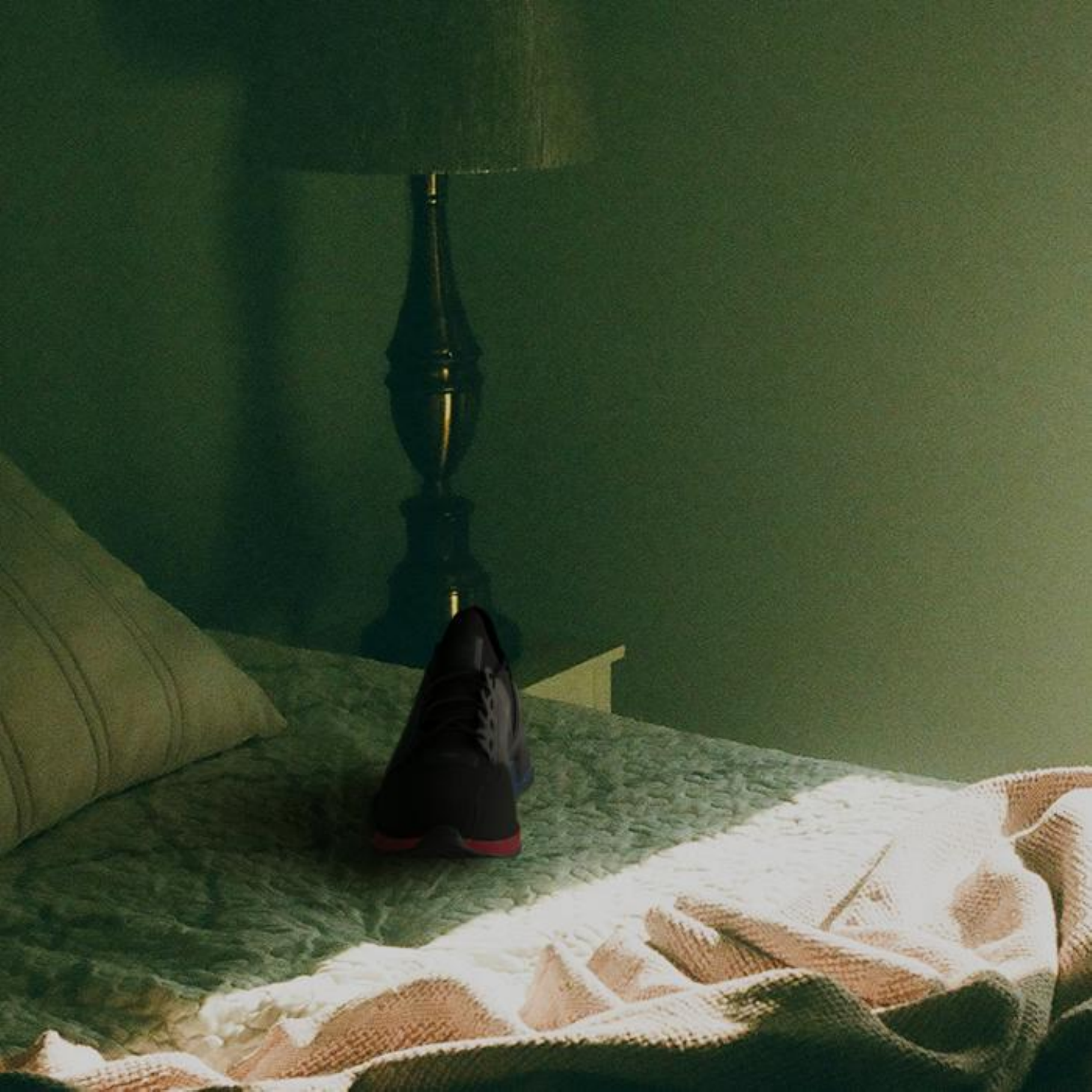}}
        \\
        &
        \noindent\parbox[c]{0.3\textwidth}{\includegraphics[height=0.3\textwidth]{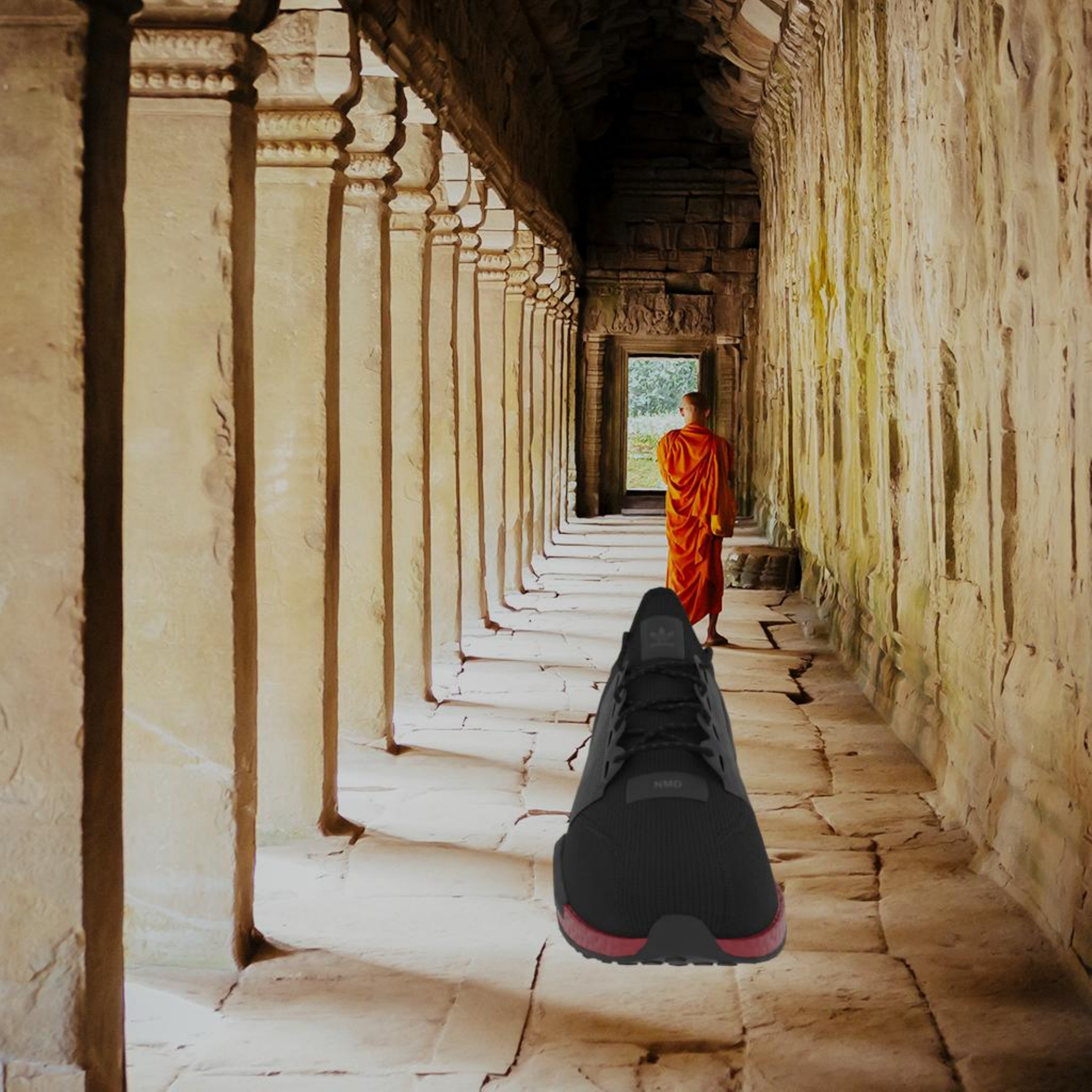}} &
        \noindent\parbox[c]{0.3\textwidth}{\includegraphics[height=0.3\textwidth]{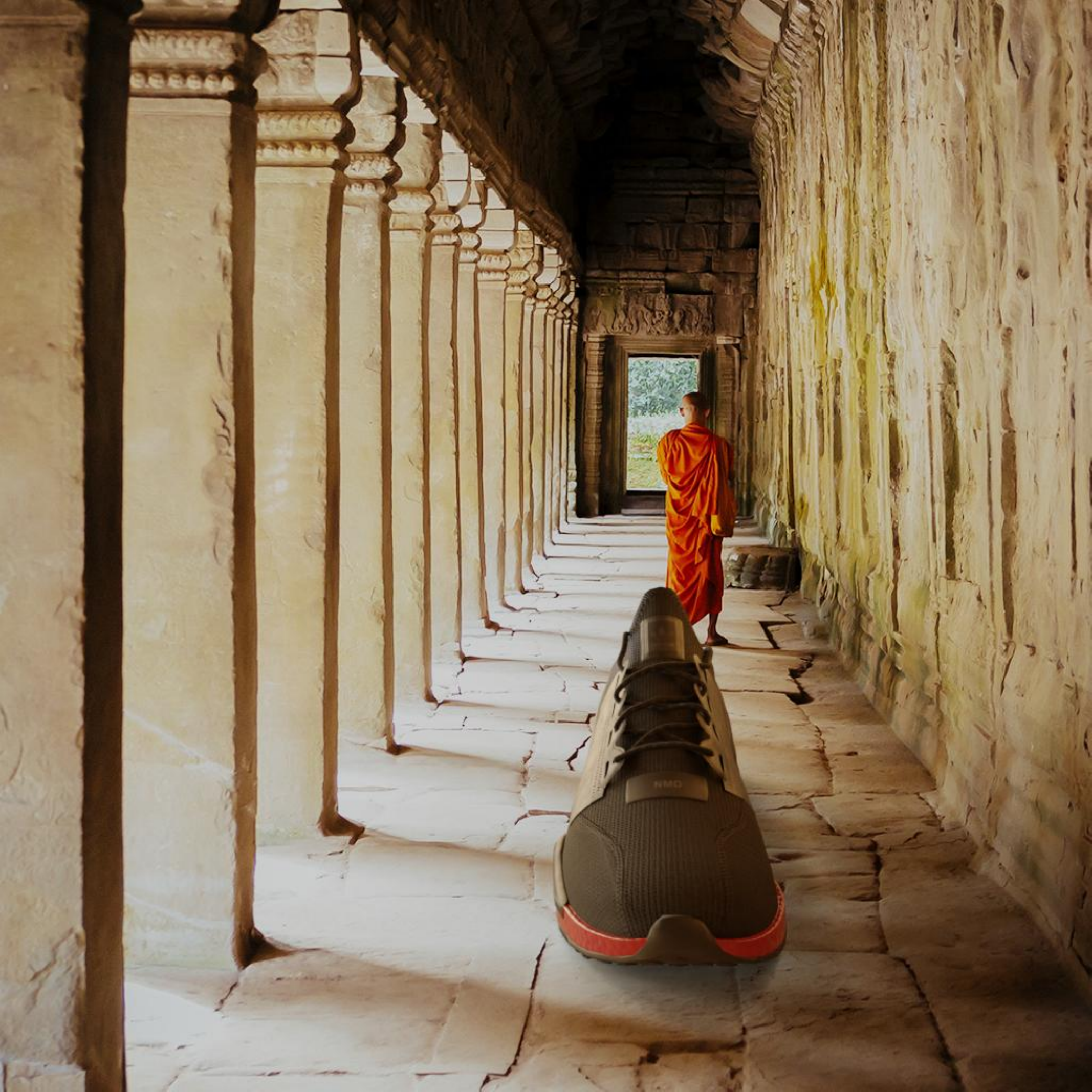}}
        \end{tabu}
    \caption{
    We synthetically render each 3D object into input images using our estimated lighting.
    }
    \label{fig:additional_object_insertion}
\end{figure*}

Virtual object insertion is a downstream application that requires light estimation. In Figure \ref{fig:additional_object_insertion}, we present qualitative results for two objects from Objaverse-XL \cite{deitke2023objaversexl}, rendered with HDR environment maps obtained through our method. 



\tabulinesep=0.5pt

\section{Additional Failure Cases}\label{app:fail}

\begin{figure*}
    \centering
 
    \begin{subfigure}{1.0\textwidth}
        \label{fig:failure-birdeye}
        \centering
        \includegraphics[width=0.32\textwidth]{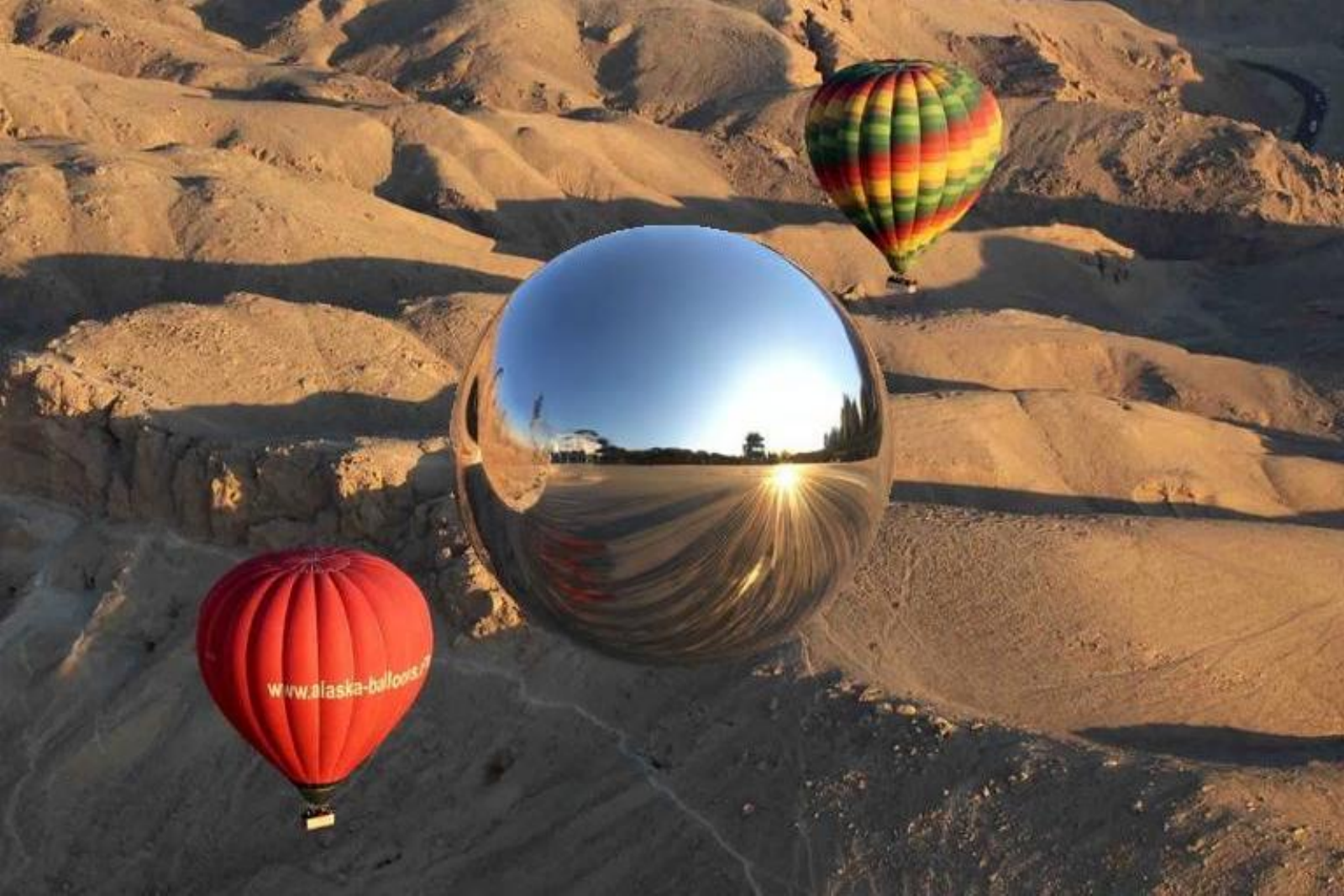}
        \includegraphics[width=0.32\textwidth]{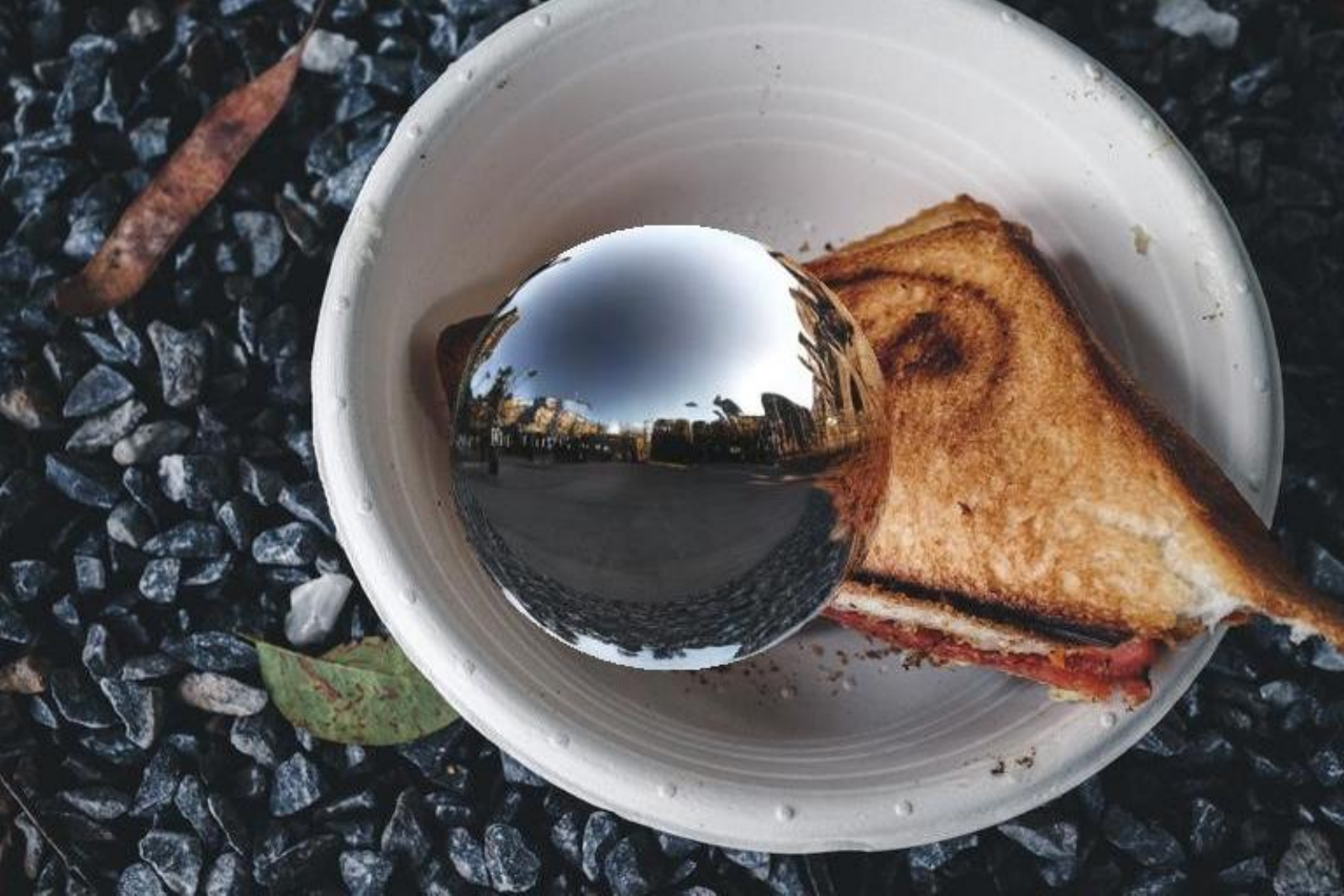}
        \includegraphics[width=0.32\textwidth]{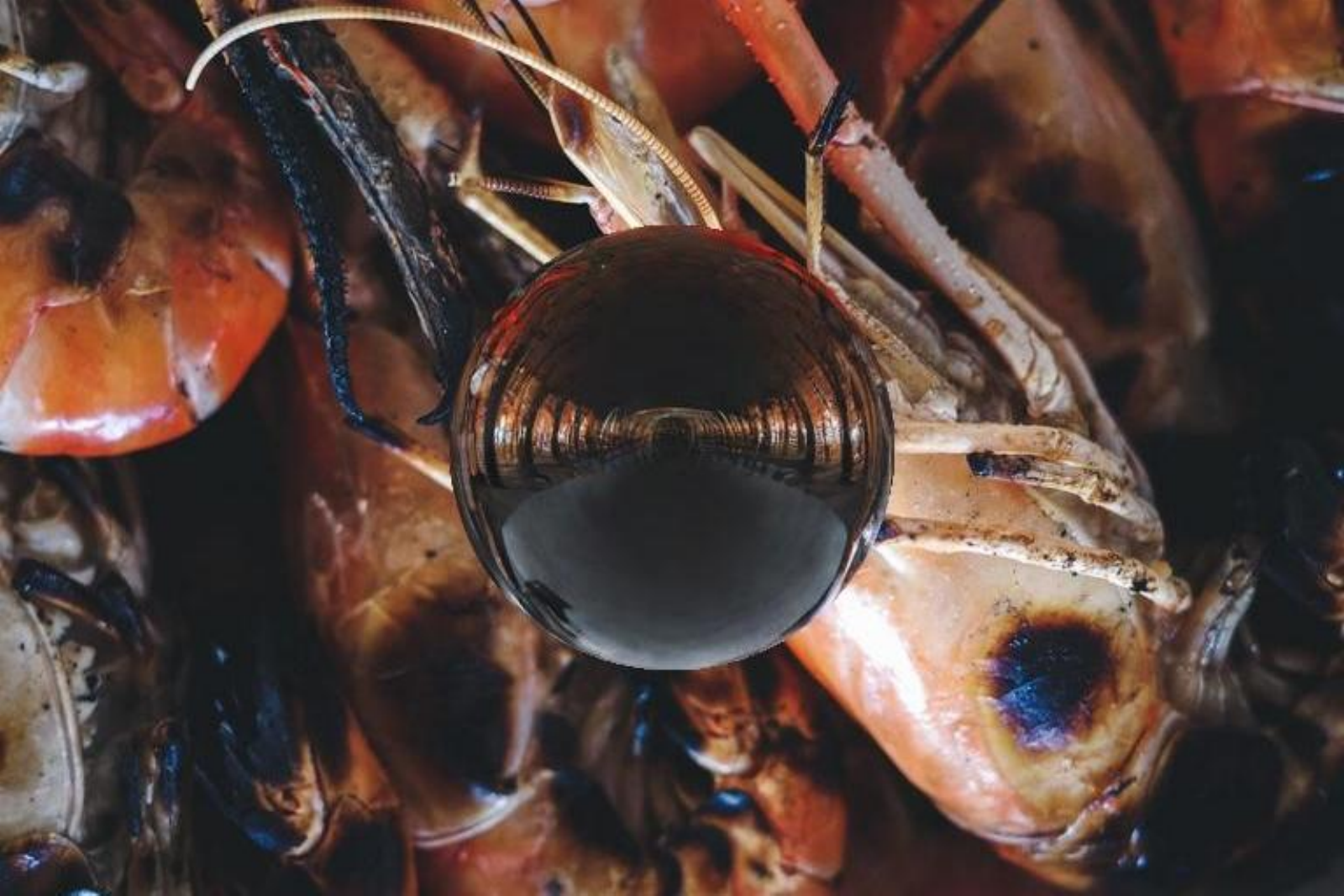}
        \caption{ Overhead and bird's-eye view images}
    \end{subfigure}
    \smallskip
    \begin{subfigure}{1.0\textwidth}
        \centering
        \includegraphics[width=0.32\textwidth]{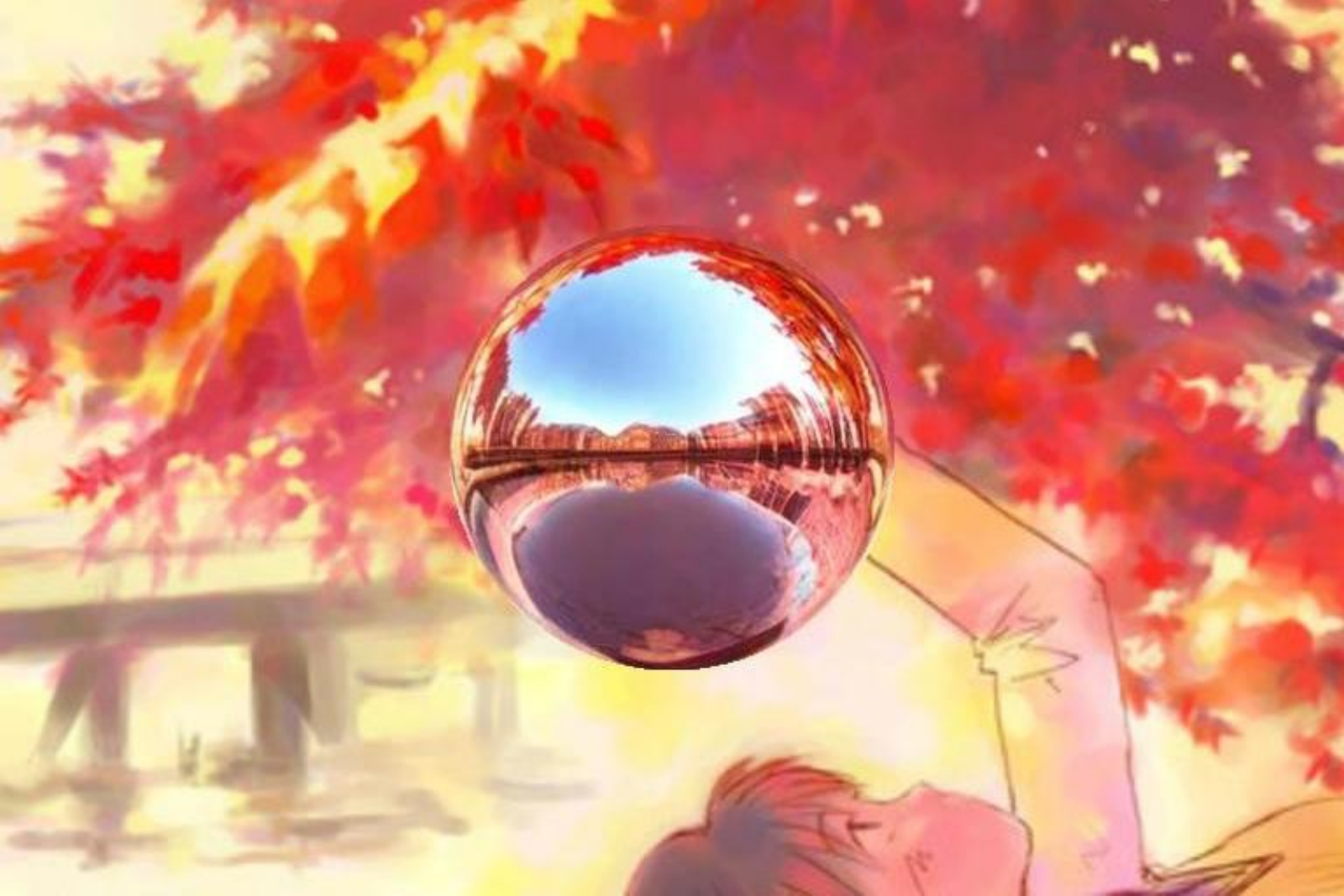}
        \includegraphics[width=0.32\textwidth]{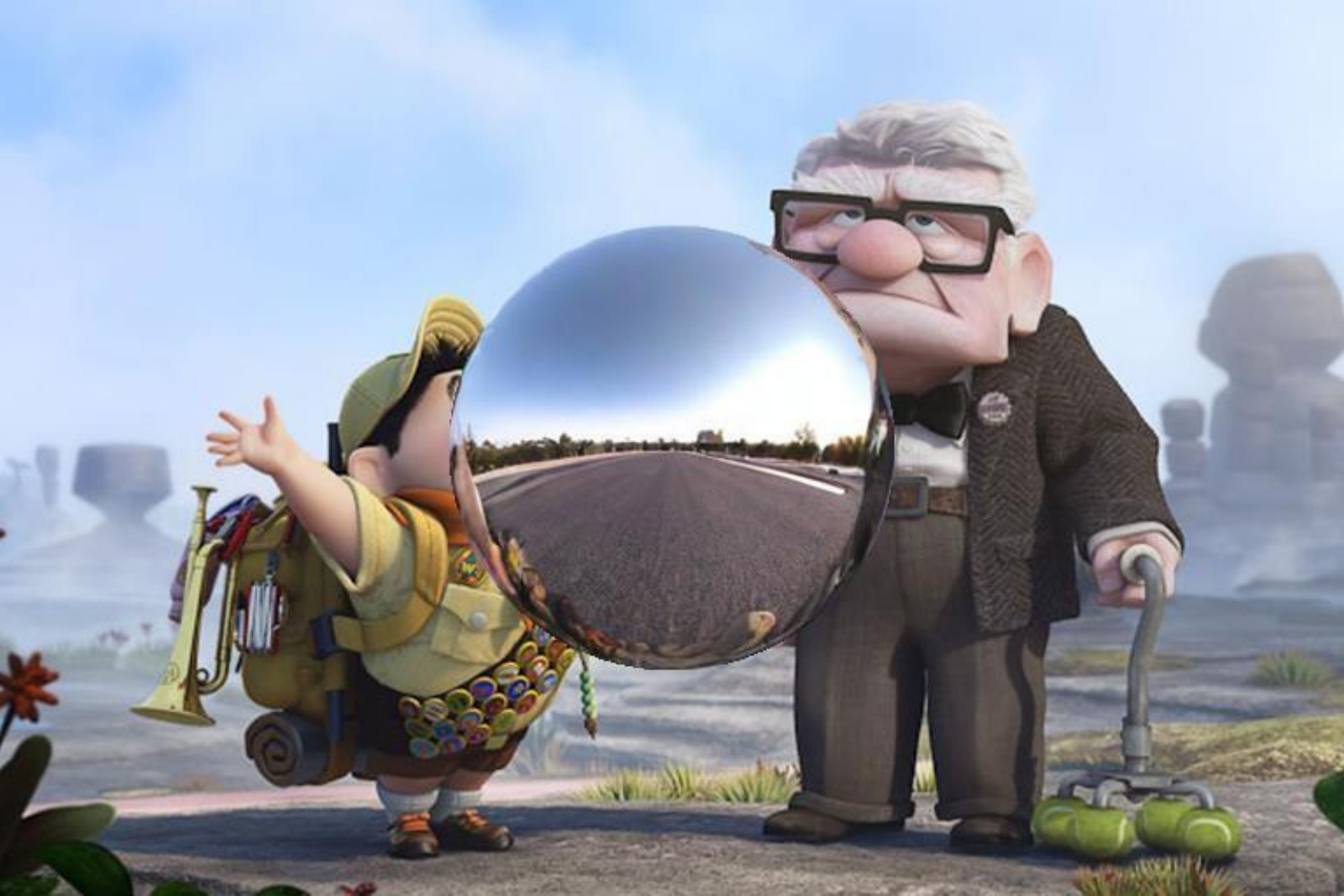}
        \includegraphics[width=0.32\textwidth]{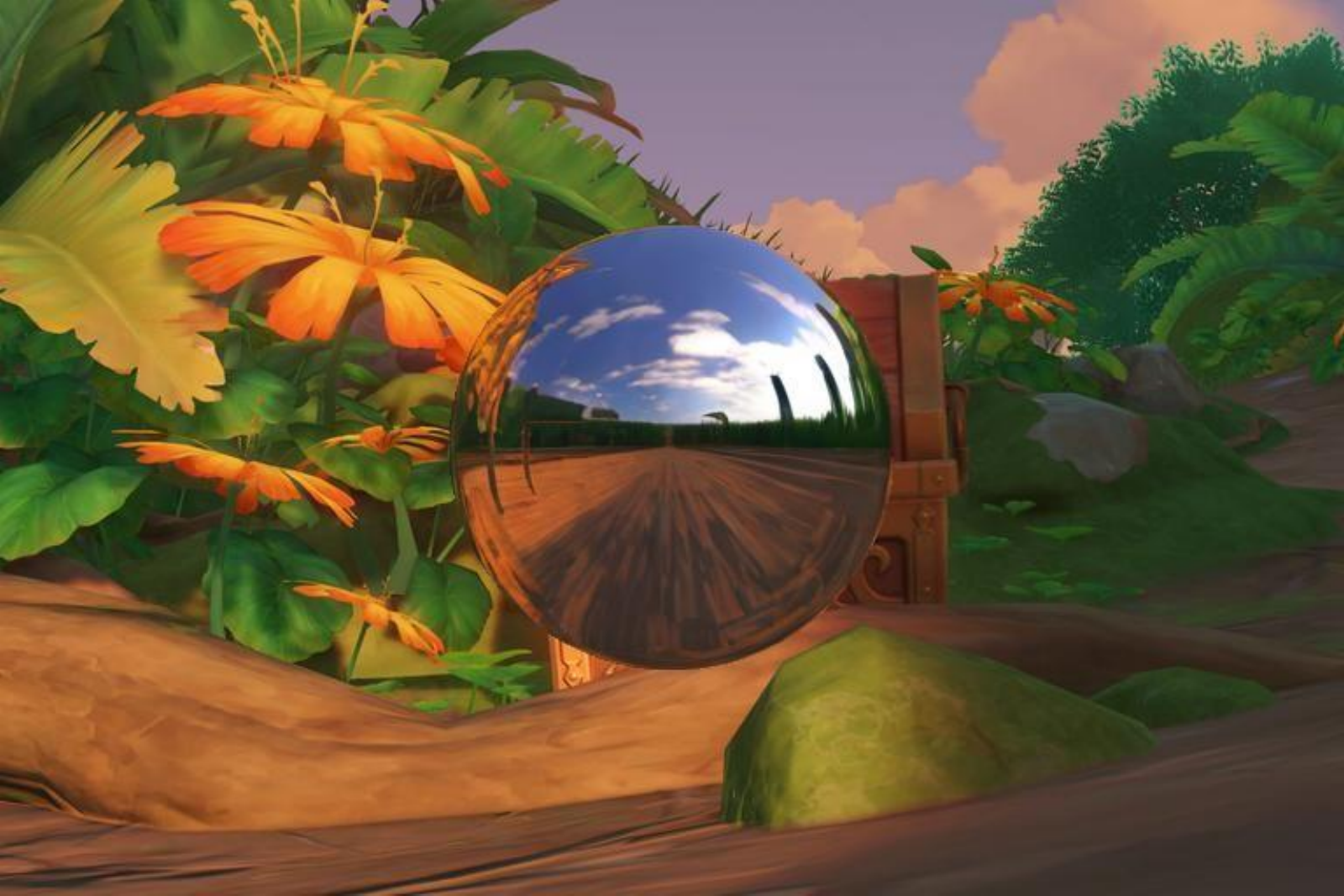}
        \caption{Images with significant style difference from natural photos}
    \end{subfigure}

    \begin{subfigure}{1.0\textwidth}
        \centering
        \includegraphics[width=0.32\textwidth]{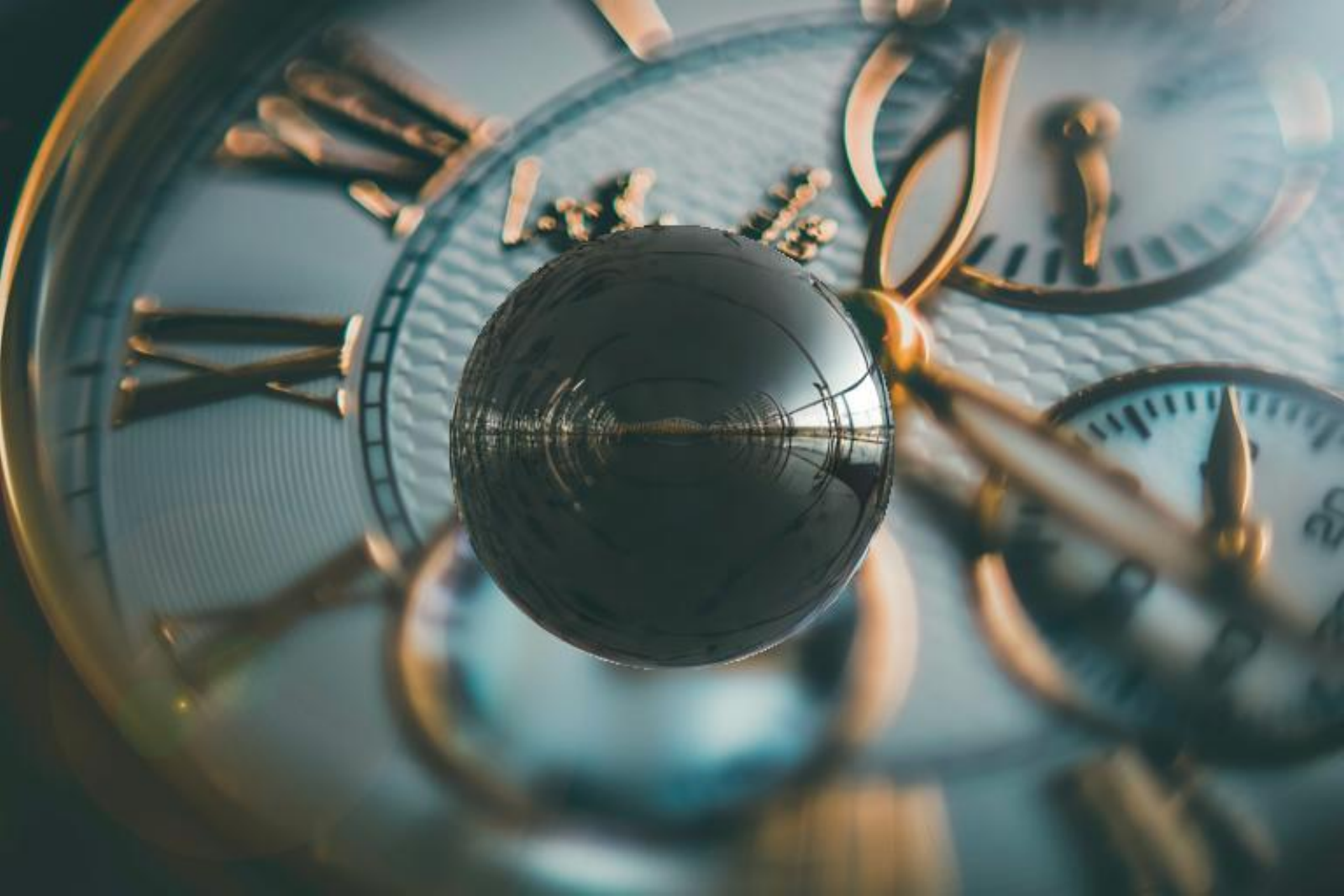}
        \includegraphics[width=0.32\textwidth]{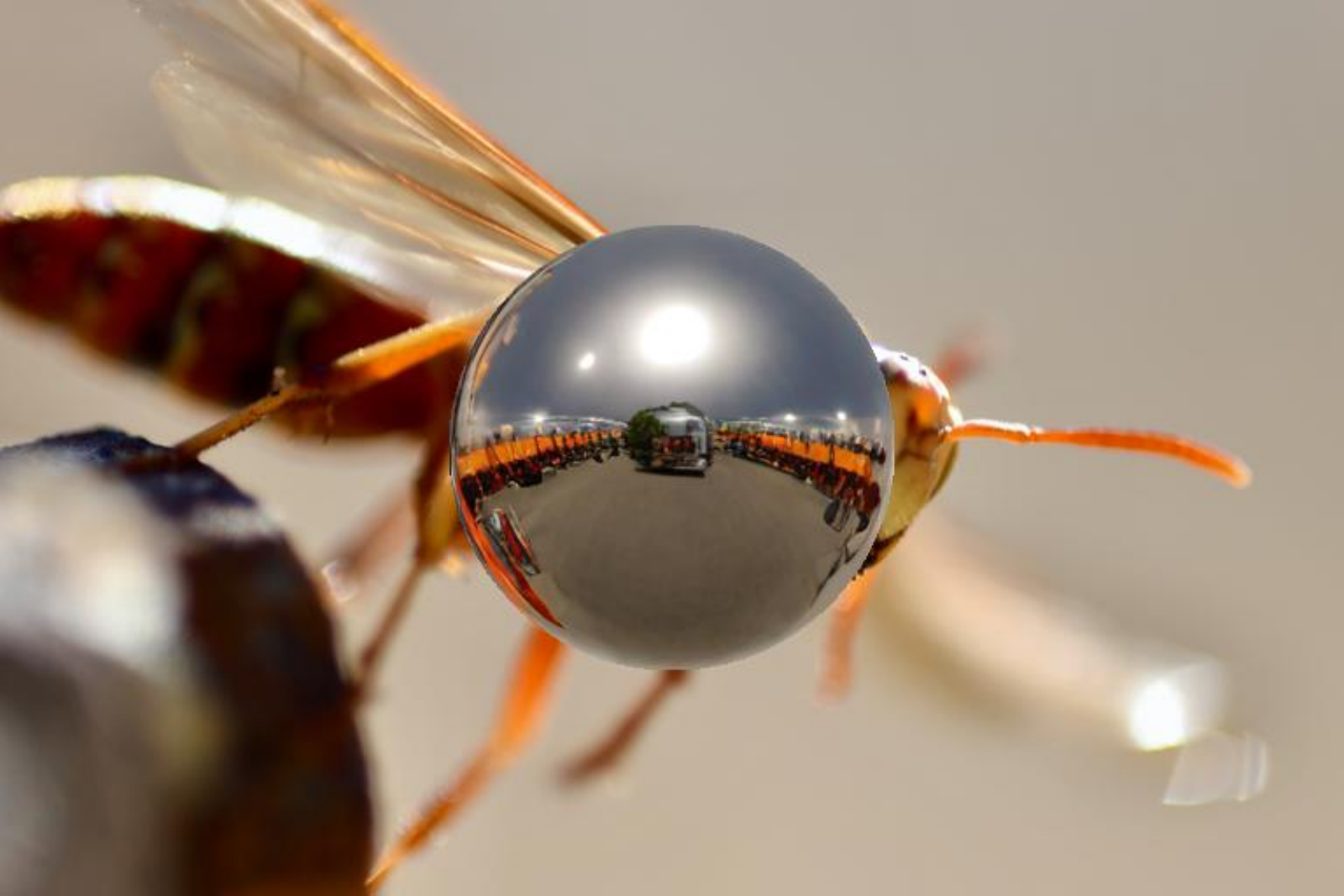}
        \includegraphics[width=0.32\textwidth]{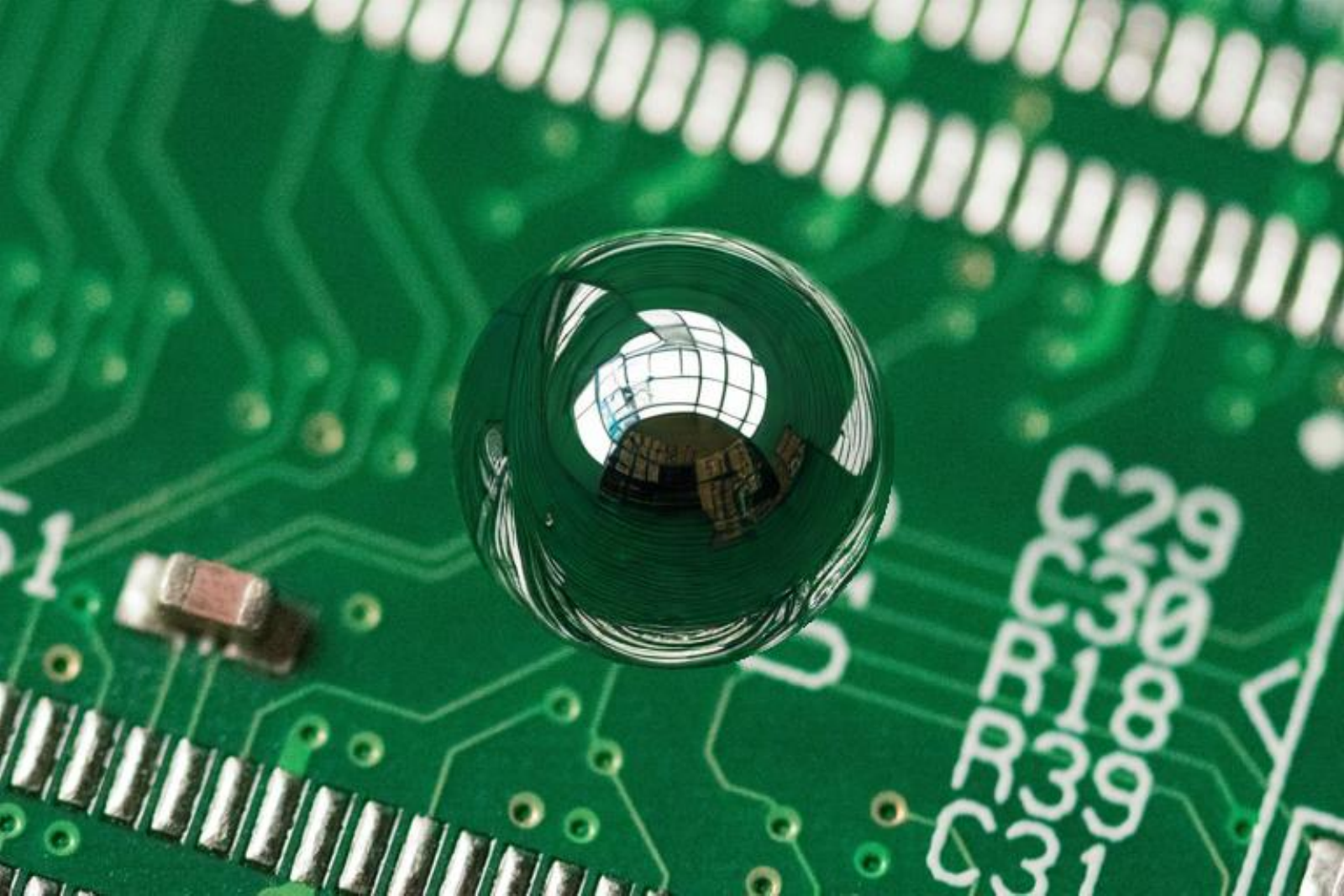}
        \caption{Marcro and close-up images}
    \end{subfigure}
    
    \begin{subfigure}{1.0\textwidth}
        \centering
        \includegraphics[width=0.32\textwidth]{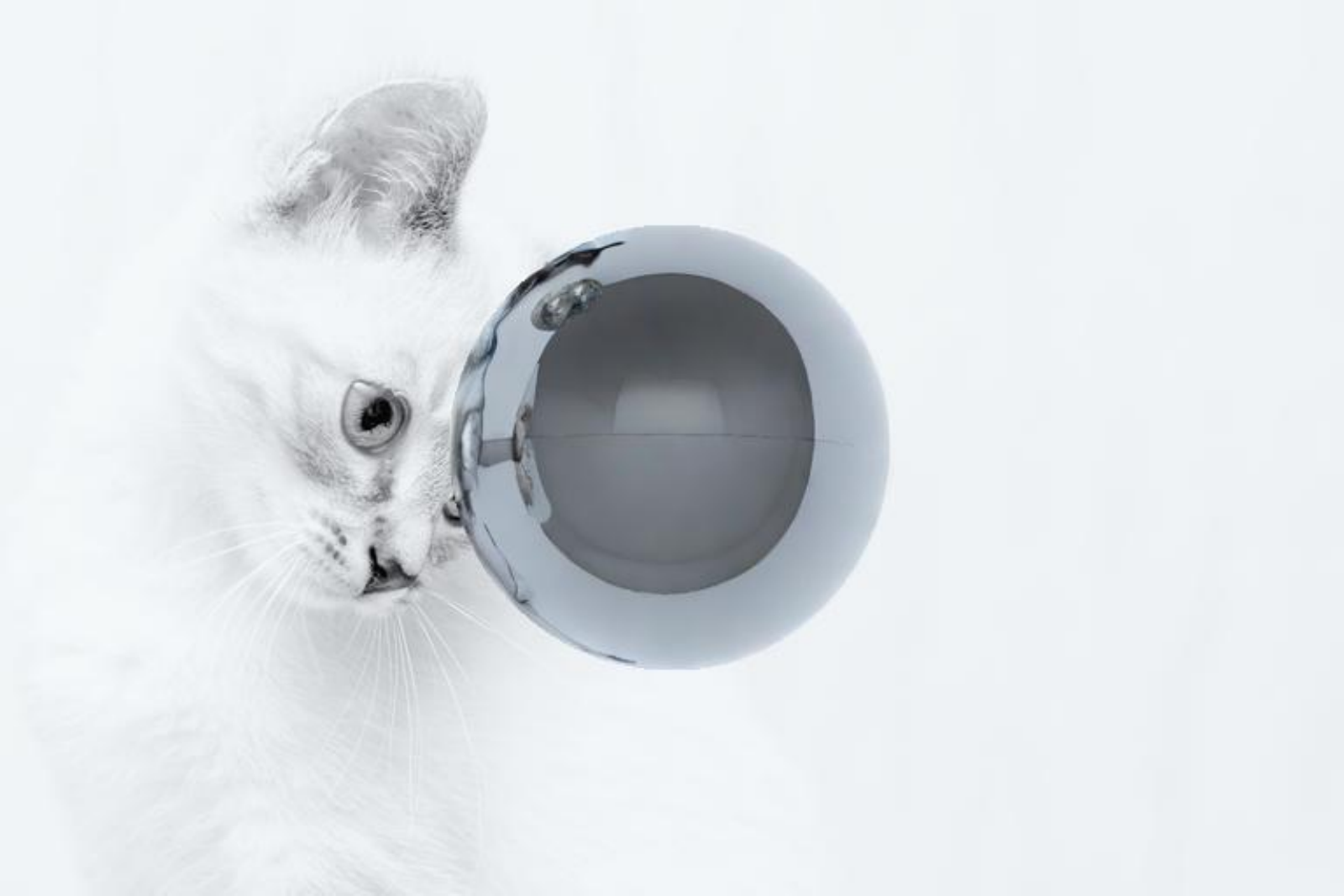}
        \includegraphics[width=0.32\textwidth]{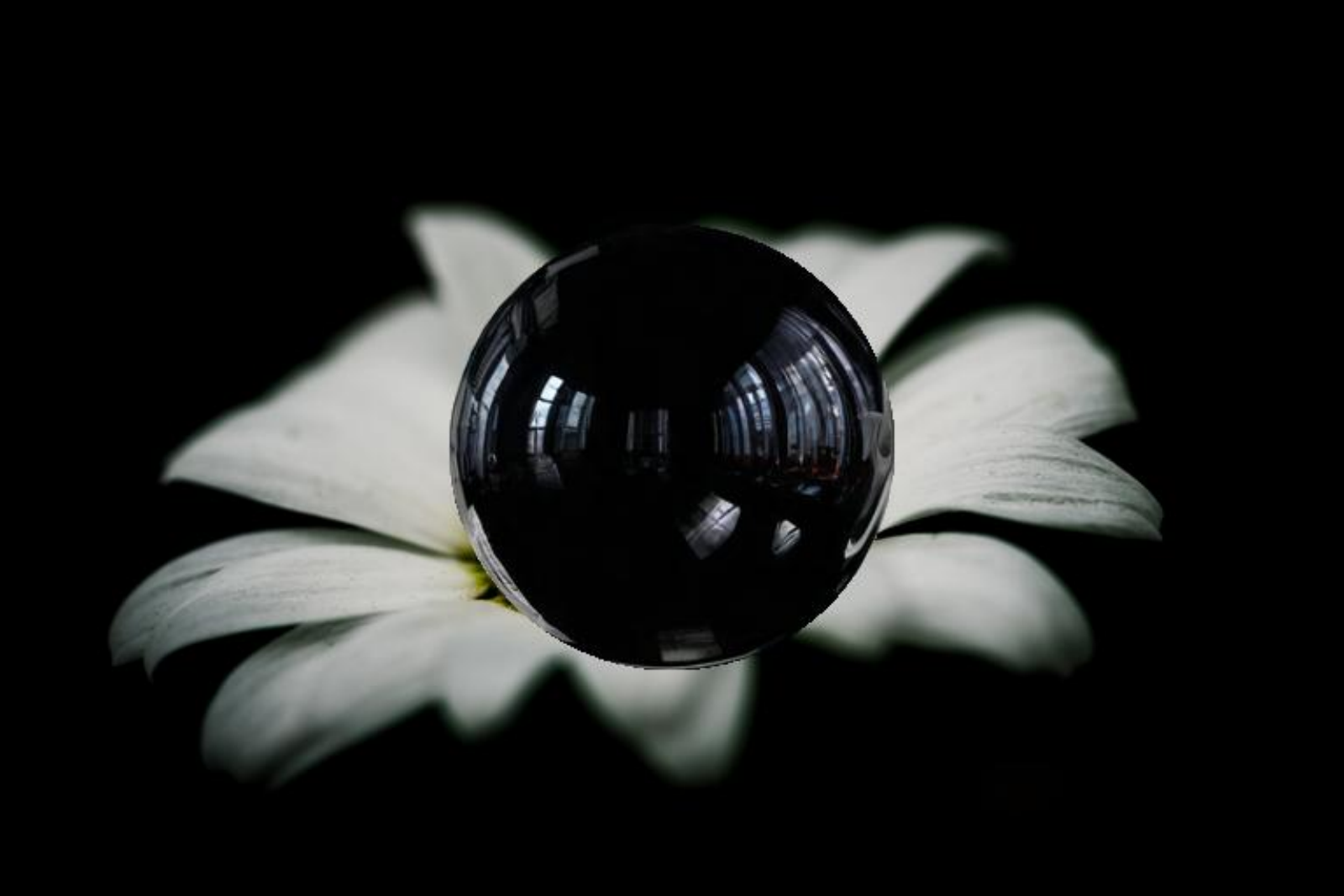}
        \includegraphics[width=0.32\textwidth]{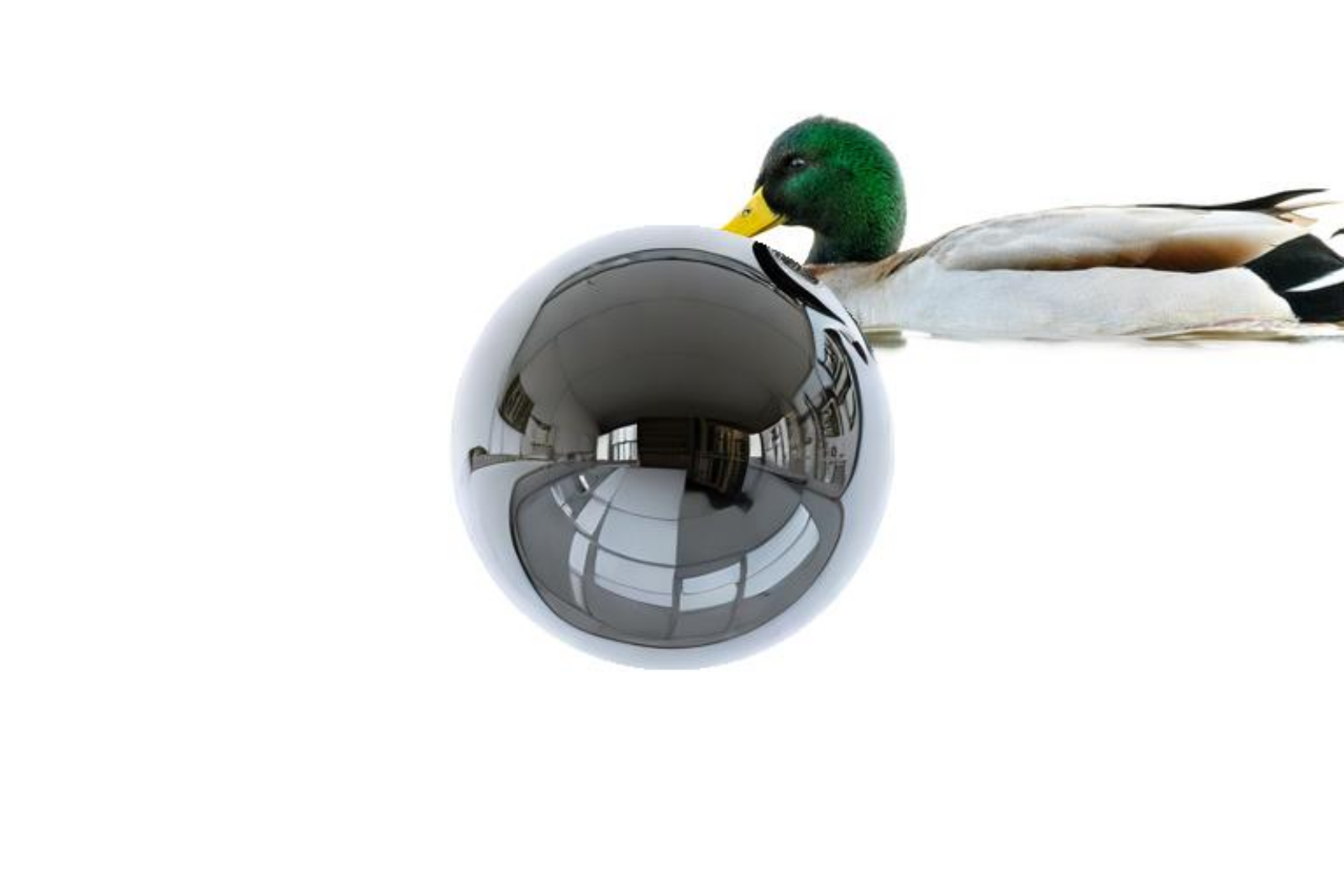}
        \caption{High-key and low-key images with solid backgrounds}
    \end{subfigure}

    \caption{\textbf{Failure modes}: (a) Our method may produce unrealistic chrome balls in overhead and bird's-eye view images, leading to incorrect curvature in the horizon line. (b) The chrome balls may not harmonize well with inputs whose styles differ significantly from natural photos.  (c) Macro and close-up images can lack sufficient shading cues, leading to less convincing estimates. (d) Images with empty or solid backgrounds often cause our method to hallucinate some details onto the balls, and the balls may appear too dark, especially on a white background. These images are from Unsplash (www.unsplash.com), Kaggle \protect\footnotemark, or other websites under the CC 4.0 license.}
    
    \label{fig:fail_all}
\end{figure*}

We present additional failure cases in Figure \ref{fig:fail_all}. Our method occasionally fails to produce chrome balls that accurately reflect surrounding environments in overhead or bird's-eye view images. For instance, the curvature of the horizon line in the scene with balloons is incorrect. 
While our method performs reasonably well for some non-realistic images like paintings, it struggles with images in drastically different styles, such as some cartoons and Japanese-style animations, which significantly differ from the training data of SDXL \cite{podell2023sdxl}. Switching to a fine-tuned model, such as AnimagineXL \footnote{\url{https://huggingface.co/Linaqruf/animagine-xl}}, to leverage a more specialized generative prior can improve its performance on specific image styles.




\footnotetext{\url{https://www.kaggle.com/datasets/mylesoneill/tagged-anime-illustrations/data}}


\section{StyleLight's Score Discrepancies} \label{appendix:stylelight-score-diff}
We used StyleLight's official implementation to produce their scores in Table \ref{tab:indoor_stylelight}. However, there are discrepancies with their reported scores due to unknown implementations of their metrics.
We discussed this with the authors on GitHub \footnote{\url{https://github.com/Wanggcong/StyleLight/issues/9}} before the submission deadline, and they clarified that additional masking of black regions and rotation of panoramas were performed before evaluation. Despite implementing these additional steps, we still could not match the scores.
The authors further suggested that we apply a consistent post-processing technique to all baselines for a fair comparison, which resulted in the scores we reported. To ensure transparency, we have made our evaluation code available at \url{https://github.com/DiffusionLight/DiffusionLight-evaluation}.












\tabulinesep=0.1pt
\tabulinesep=0.1pt


\tabulinesep=0.5pt
\begin{figure*}[!t]
    \centering

        \begin{tabu} to \textwidth {
        @{}
        c@{}
        c@{}
        c@{}
        c@{}
        c@{}
        c@{}
        c@{}
        c@{}
        c@{}
    }

        \multicolumn{1}{c}{\shortstack{\scriptsize Ground truth map}}
        & 
        \multicolumn{1}{c}{\shortstack{\hspace{-6pt} \scriptsize Input}}
        &
        \multicolumn{1}{c}{\shortstack{\scriptsize Ground truth}}
        & 
        \multicolumn{1}{c}{\shortstack{\scriptsize StyleLight \cite{wang2022stylelight}}}
        & 
        \multicolumn{1}{c}{\shortstack{\scriptsize SDXL$^\dagger$}} &
        \multicolumn{1}{c}{\shortstack{\scriptsize \begin{tabular}[c]{@{}c@{}}SDXL$^\dagger$+LR \\ (ours, ablated)\end{tabular}}} &
        \multicolumn{1}{c}{\shortstack{\scriptsize \begin{tabular}[c]{@{}c@{}}SDXL$^\dagger$+I \\ (ours,ablated)\end{tabular}}}
        &
        \multicolumn{1}{c}{\shortstack{\scriptsize \begin{tabular}[c]{@{}c@{}}SDXL$^\dagger$+LR+I \\ (ours)\end{tabular}}} 
        \\

        \noindent\parbox[c]{0.205\textwidth}{\includegraphics[height=0.100\textwidth]{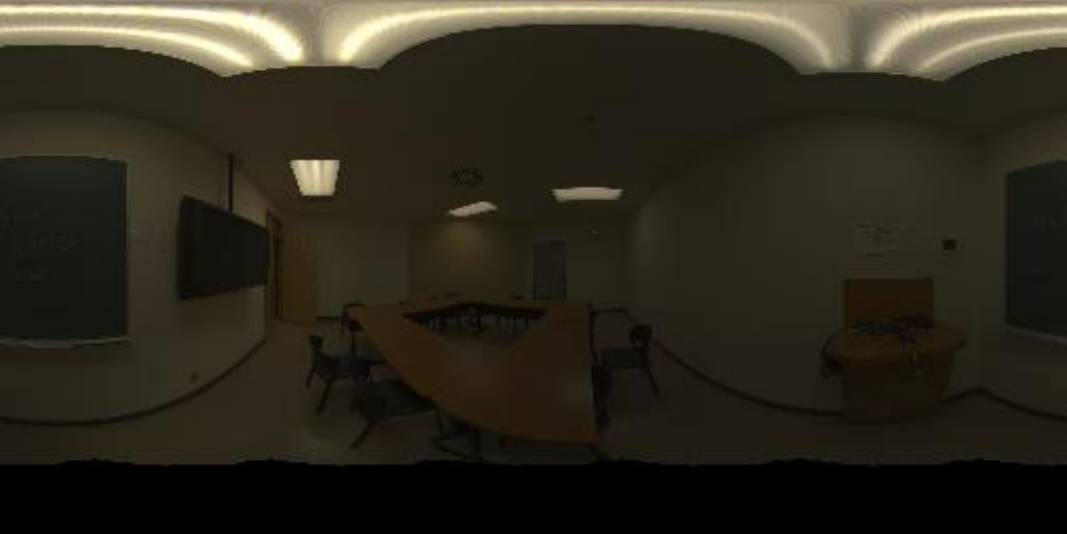}} & 
        \noindent\parbox[c]{0.14\textwidth}{\includegraphics[height=0.100\textwidth]{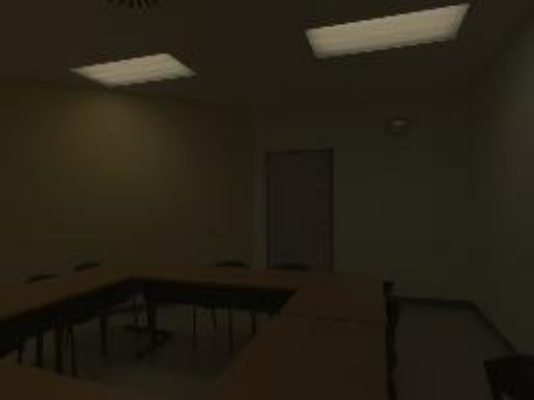}} &  
        
        \noindent\parbox[c]{0.100\textwidth}{\includegraphics[height=0.100\textwidth]{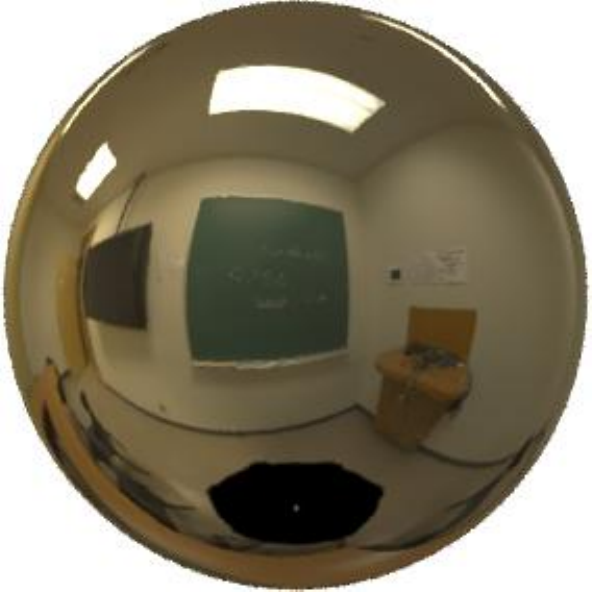}} & 
        \noindent\parbox[c]{0.100\textwidth}{\includegraphics[height=0.100\textwidth]{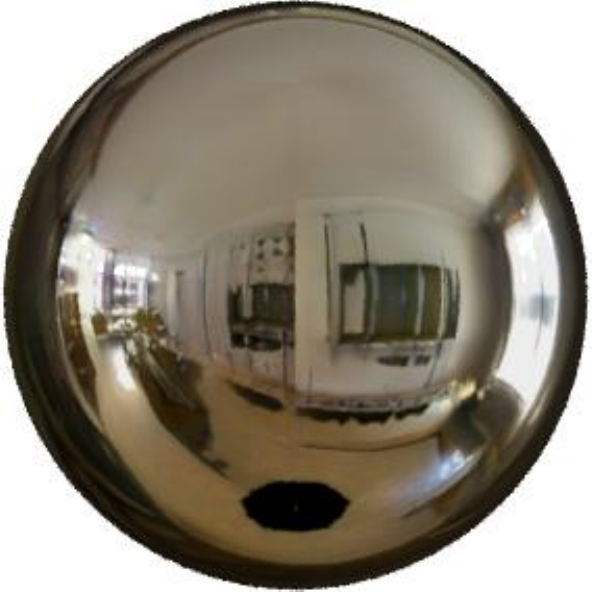}} & 
        
        \noindent\parbox[c]{0.100\textwidth}{\includegraphics[height=0.100\textwidth]{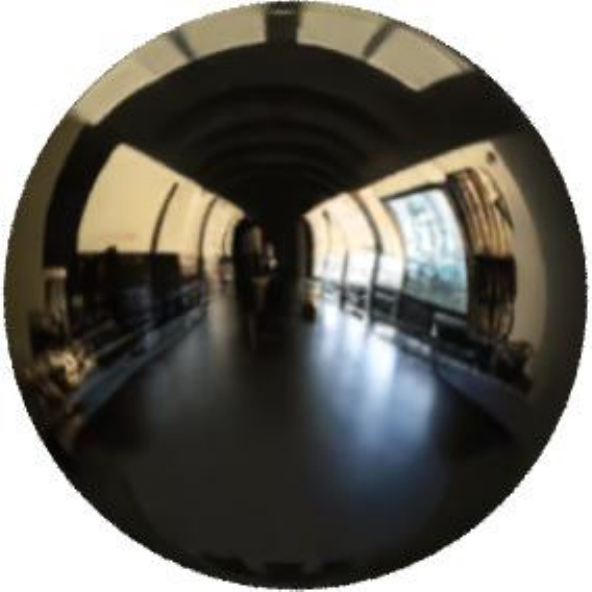}} & 
        \noindent\parbox[c]{0.100\textwidth}{\includegraphics[height=0.100\textwidth]{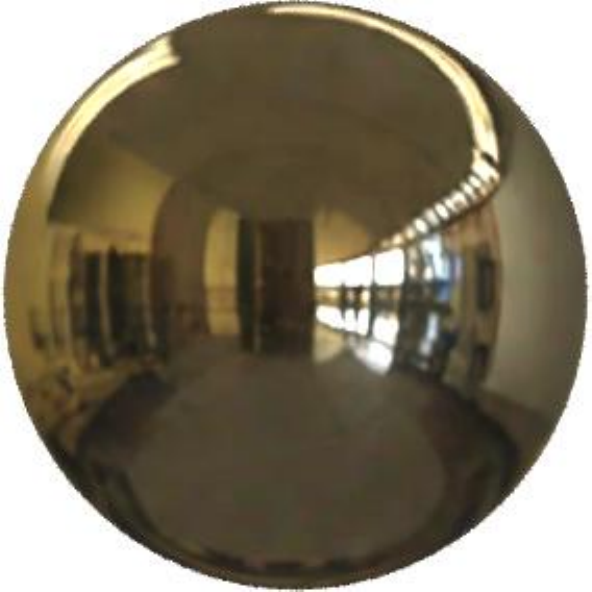}} &
        \noindent\parbox[c]{0.100\textwidth}{\includegraphics[height=0.100\textwidth]{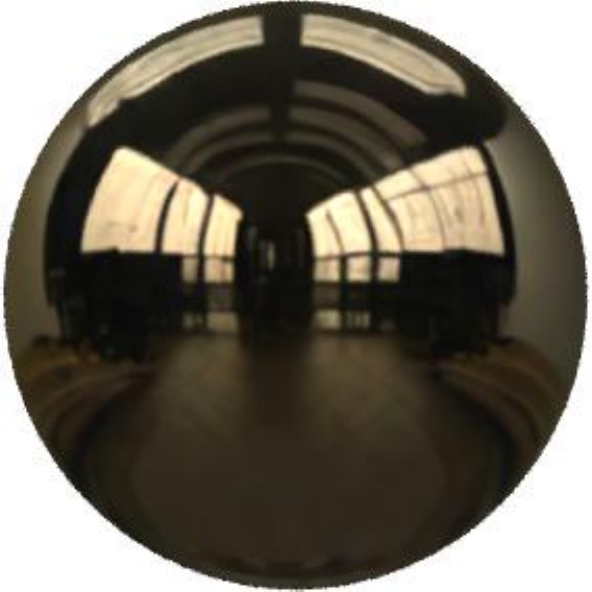}} & 
        \noindent\parbox[c]{0.100\textwidth}{\includegraphics[height=0.100\textwidth]{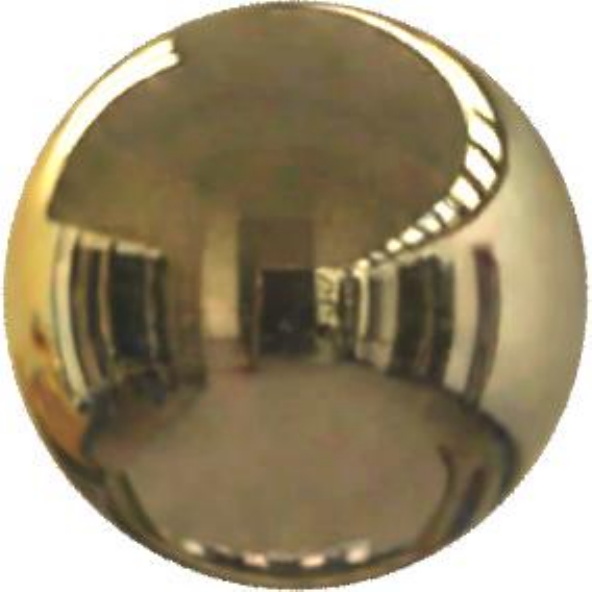}} & 
        \\

        \noindent\parbox[c]{0.205\textwidth}{\includegraphics[height=0.100\textwidth]{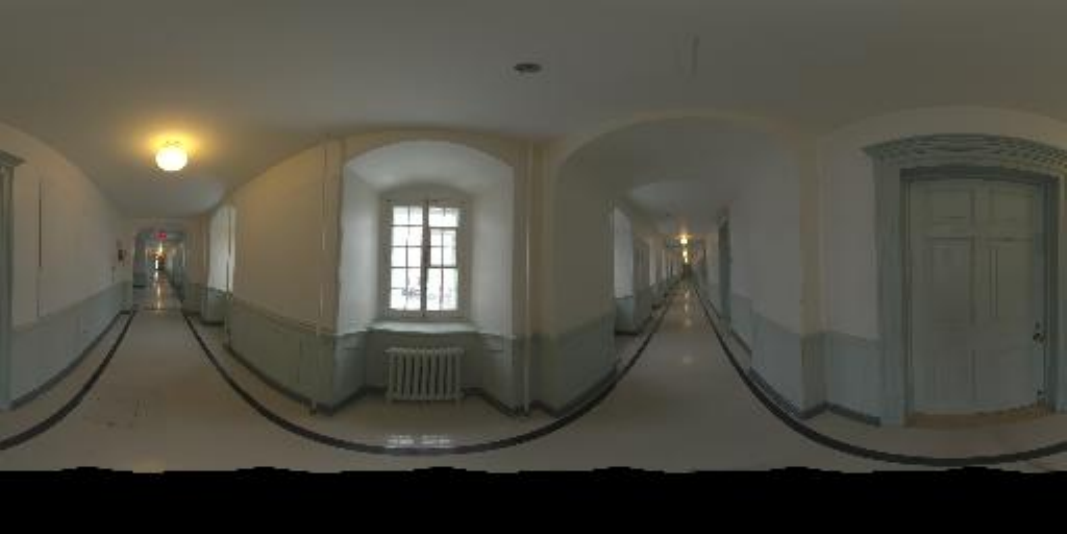}} & 
        \noindent\parbox[c]{0.14\textwidth}{\includegraphics[height=0.100\textwidth]{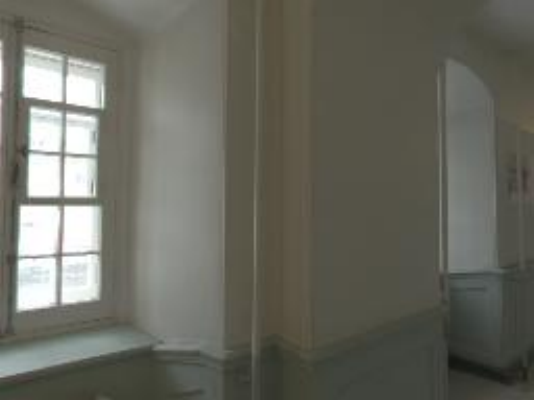}} &  
        
        \noindent\parbox[c]{0.100\textwidth}{\includegraphics[height=0.100\textwidth]{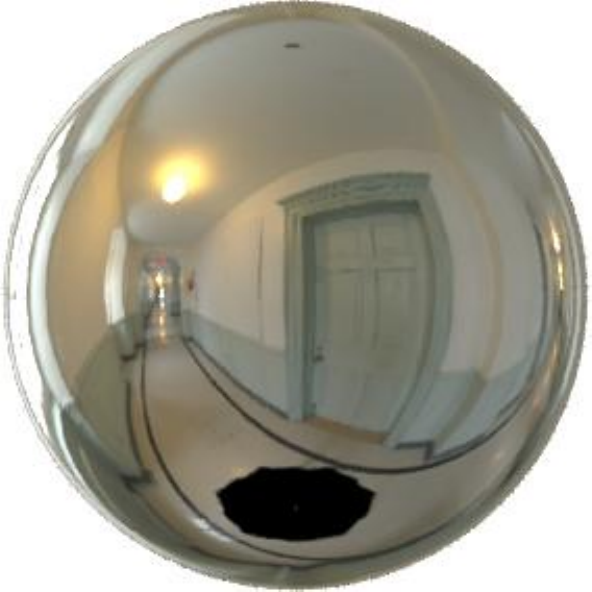}} & 
        \noindent\parbox[c]{0.100\textwidth}{\includegraphics[height=0.100\textwidth]{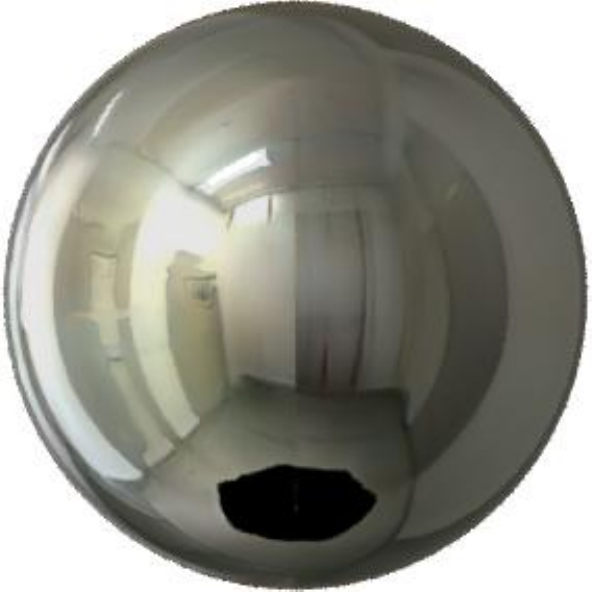}} & 
        
        \noindent\parbox[c]{0.100\textwidth}{\includegraphics[height=0.100\textwidth]{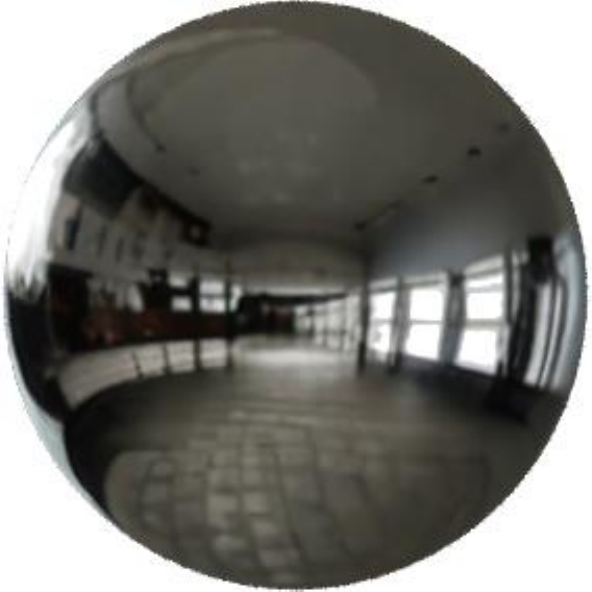}} & 
        \noindent\parbox[c]{0.100\textwidth}{\includegraphics[height=0.100\textwidth]{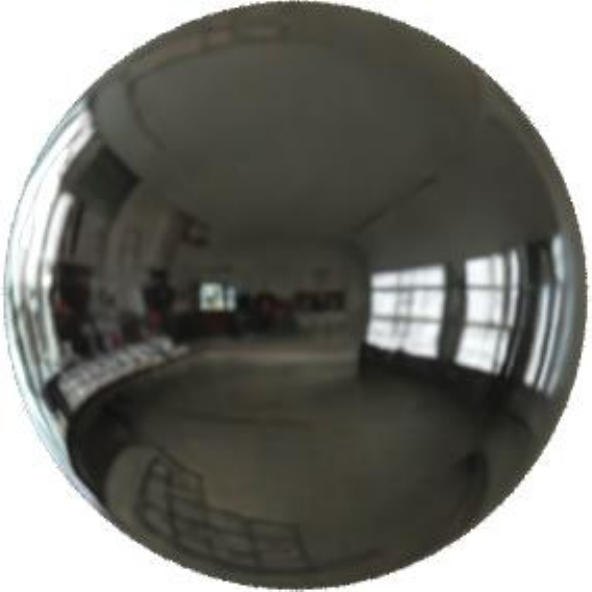}} &
        \noindent\parbox[c]{0.100\textwidth}{\includegraphics[height=0.100\textwidth]{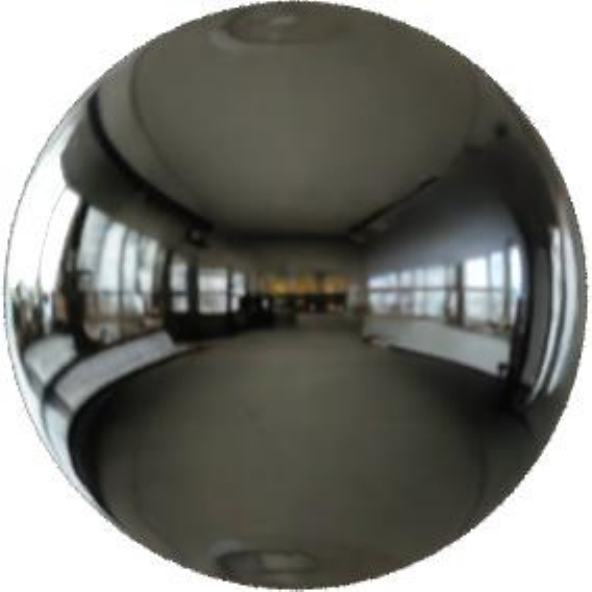}} & 
        \noindent\parbox[c]{0.100\textwidth}{\includegraphics[height=0.100\textwidth]{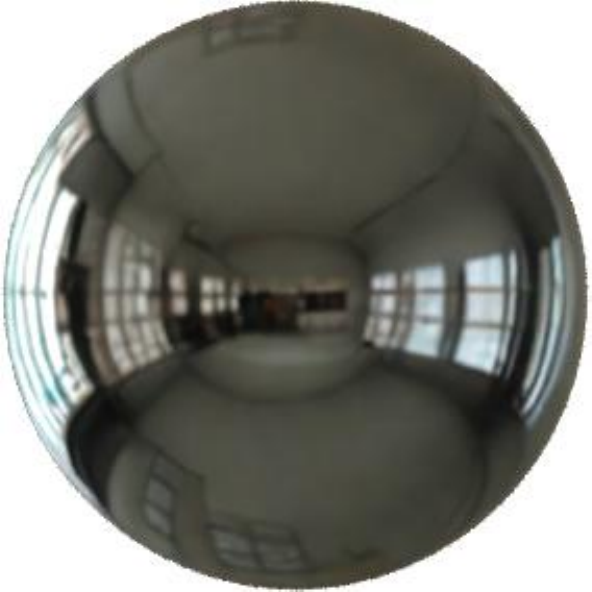}} & 
        \\

        \noindent\parbox[c]{0.205\textwidth}{\includegraphics[height=0.100\textwidth]{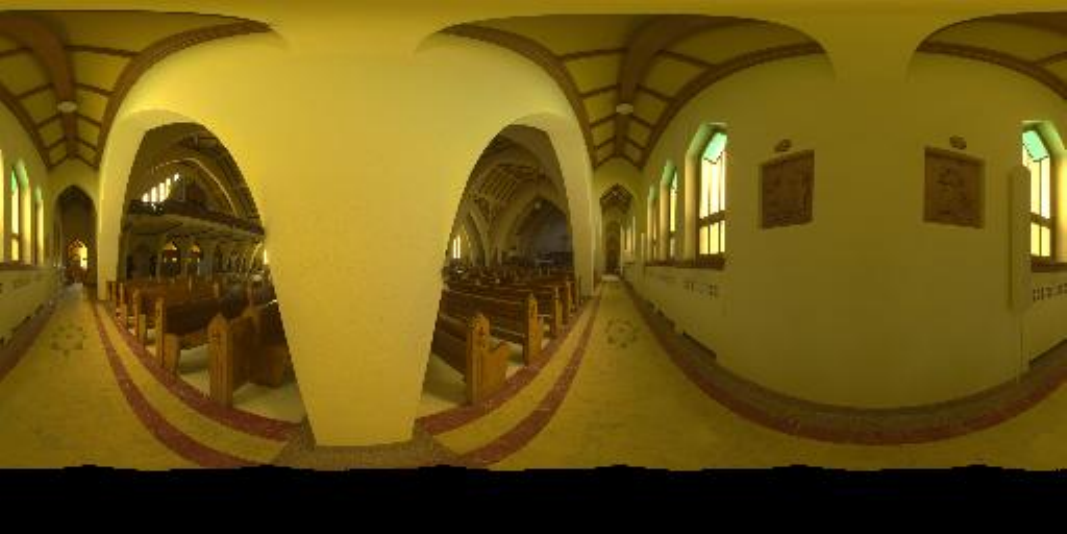}} & 
        \noindent\parbox[c]{0.14\textwidth}{\includegraphics[height=0.100\textwidth]{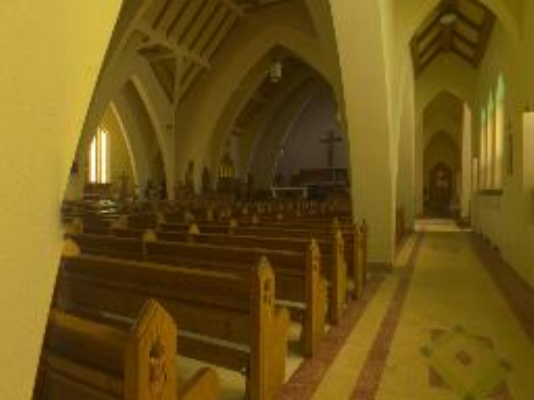}} &  
        
        \noindent\parbox[c]{0.100\textwidth}{\includegraphics[height=0.100\textwidth]{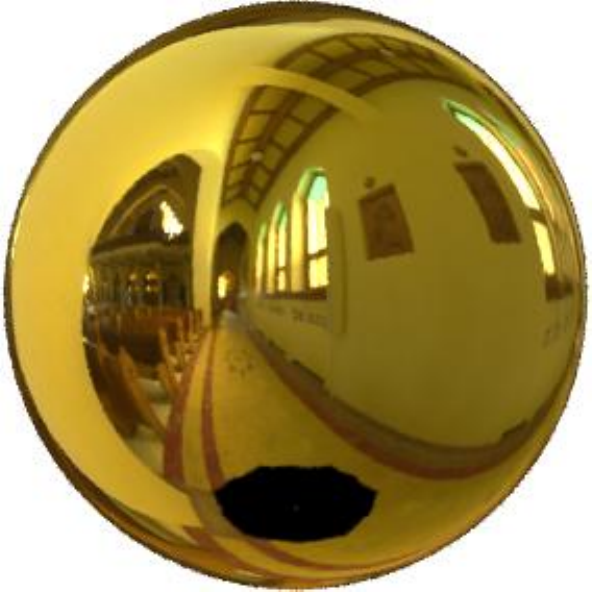}} & 
        \noindent\parbox[c]{0.100\textwidth}{\includegraphics[height=0.100\textwidth]{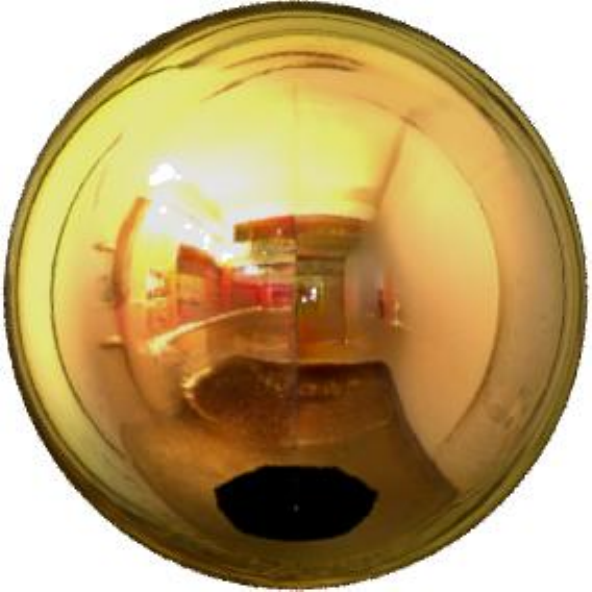}} & 
        
        \noindent\parbox[c]{0.100\textwidth}{\includegraphics[height=0.100\textwidth]{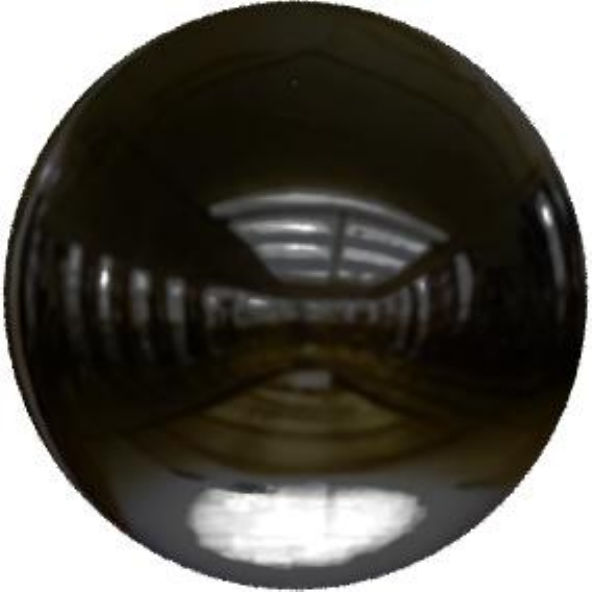}} & 
        \noindent\parbox[c]{0.100\textwidth}{\includegraphics[height=0.100\textwidth]{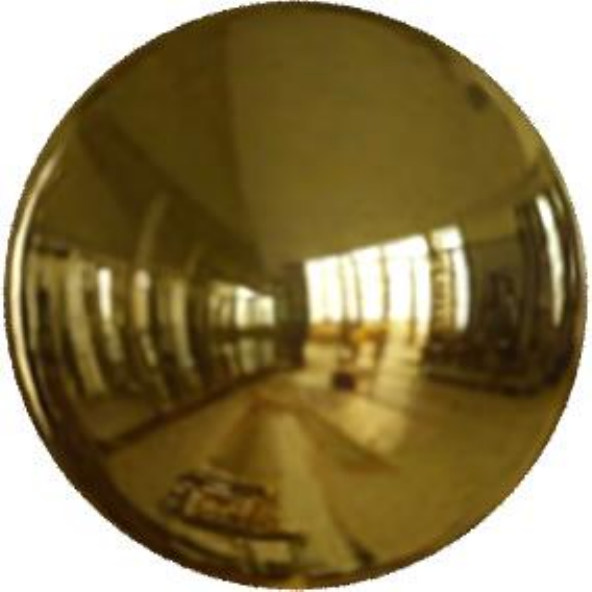}} &
        \noindent\parbox[c]{0.100\textwidth}{\includegraphics[height=0.100\textwidth]{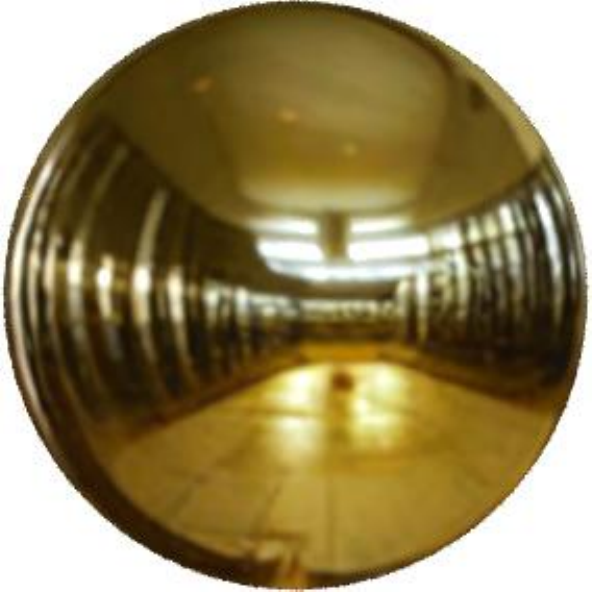}} & 
        \noindent\parbox[c]{0.100\textwidth}{\includegraphics[height=0.100\textwidth]{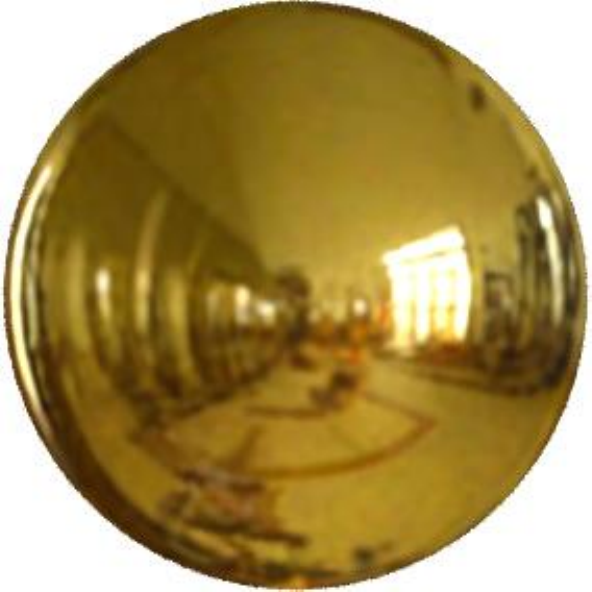}} & 
        \\

        \noindent\parbox[c]{0.205\textwidth}{\includegraphics[height=0.100\textwidth]{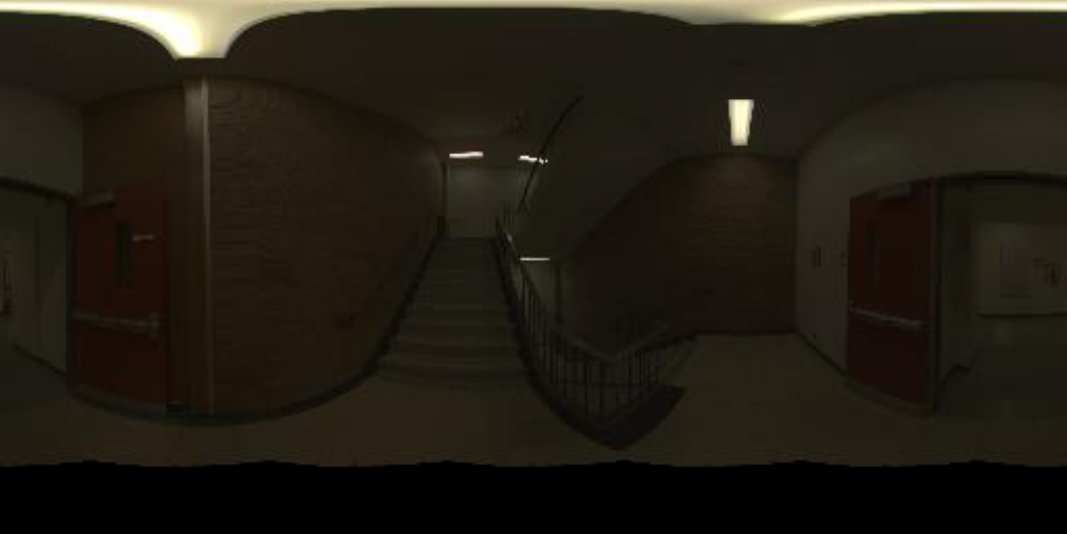}} & 
        \noindent\parbox[c]{0.14\textwidth}{\includegraphics[height=0.100\textwidth]{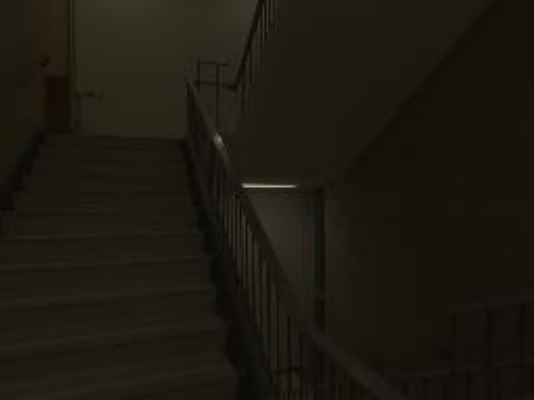}} &  
        
        \noindent\parbox[c]{0.100\textwidth}{\includegraphics[height=0.100\textwidth]{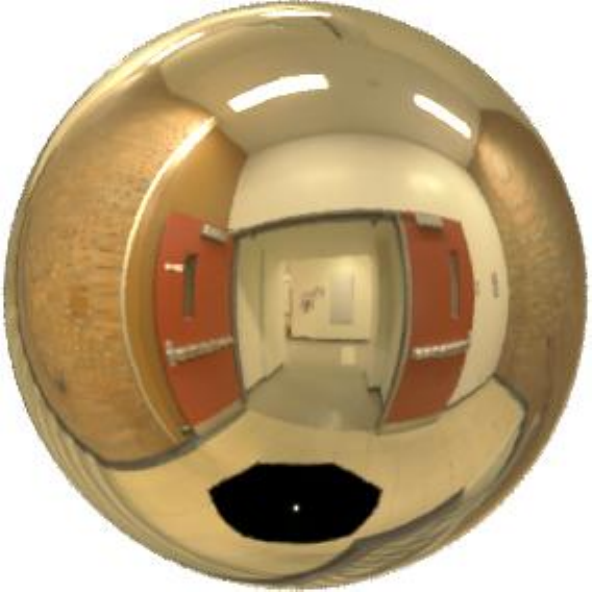}} & 
        \noindent\parbox[c]{0.100\textwidth}{\includegraphics[height=0.100\textwidth]{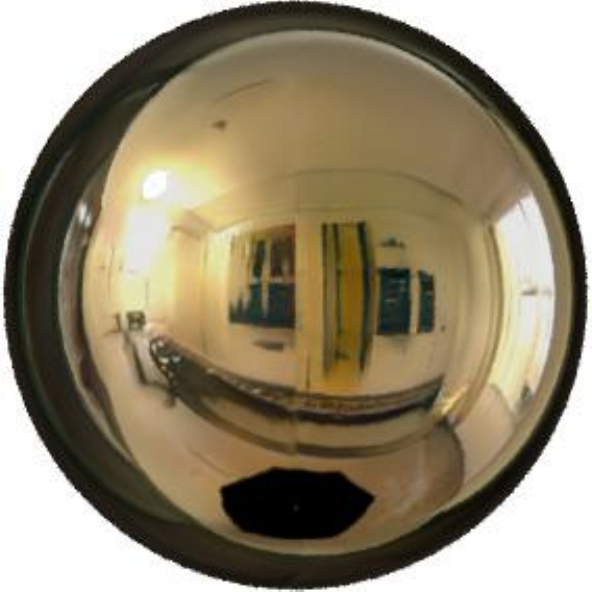}} & 
        
        \noindent\parbox[c]{0.100\textwidth}{\includegraphics[height=0.100\textwidth]{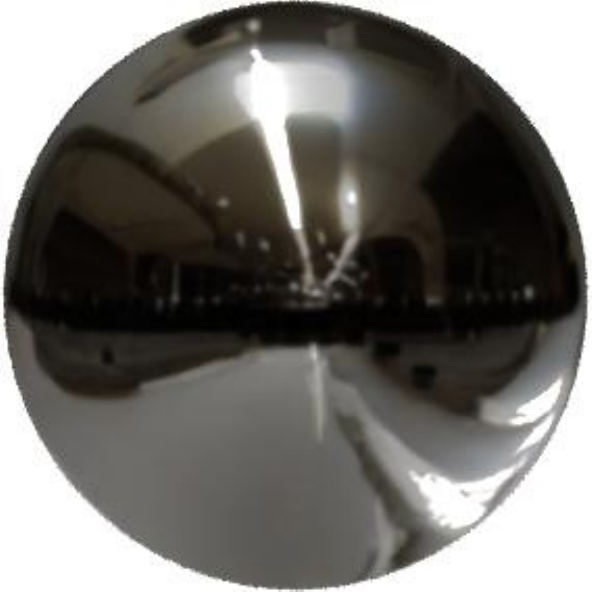}} & 
        \noindent\parbox[c]{0.100\textwidth}{\includegraphics[height=0.100\textwidth]{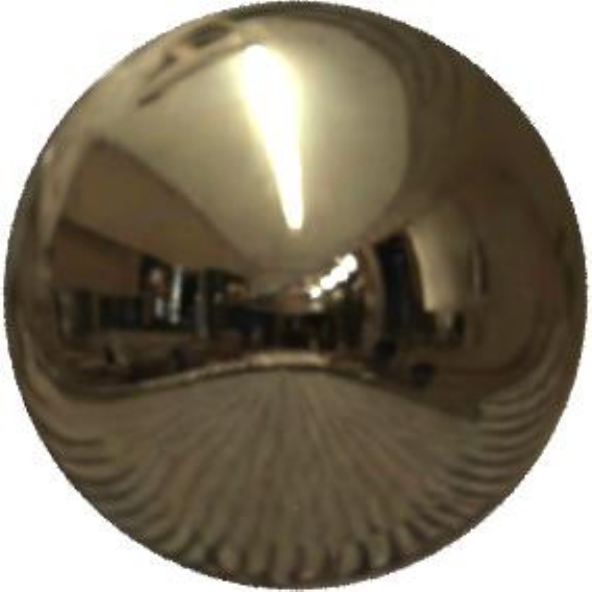}} &
        \noindent\parbox[c]{0.100\textwidth}{\includegraphics[height=0.100\textwidth]{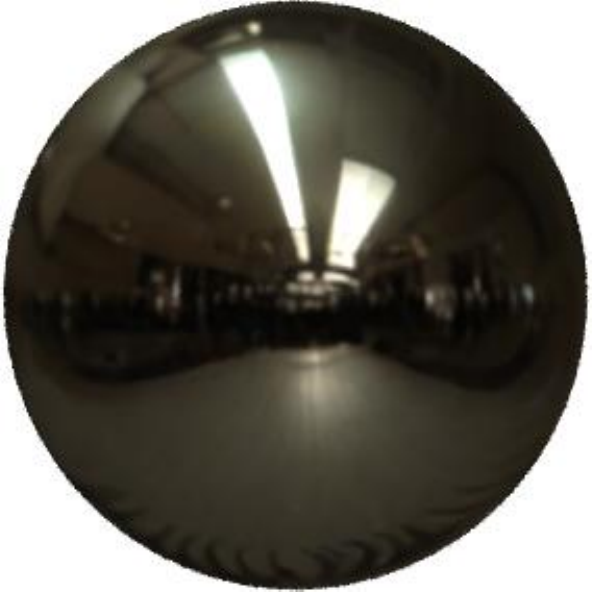}} & 
        \noindent\parbox[c]{0.100\textwidth}{\includegraphics[height=0.100\textwidth]{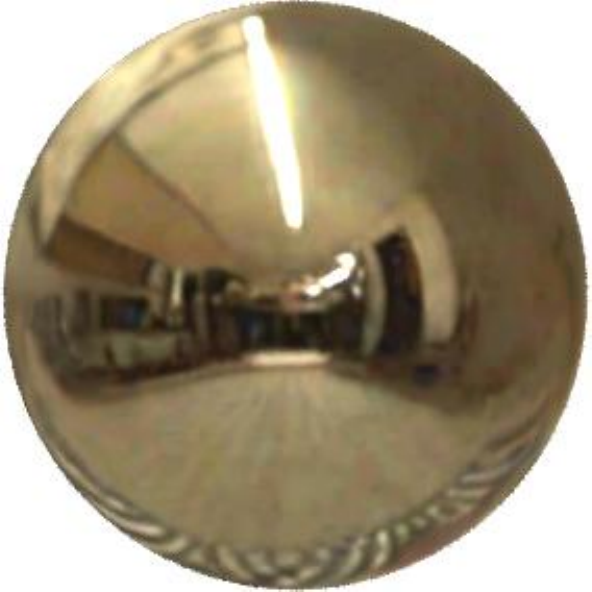}} & 
        \\

        \noindent\parbox[c]{0.205\textwidth}{\includegraphics[height=0.100\textwidth]{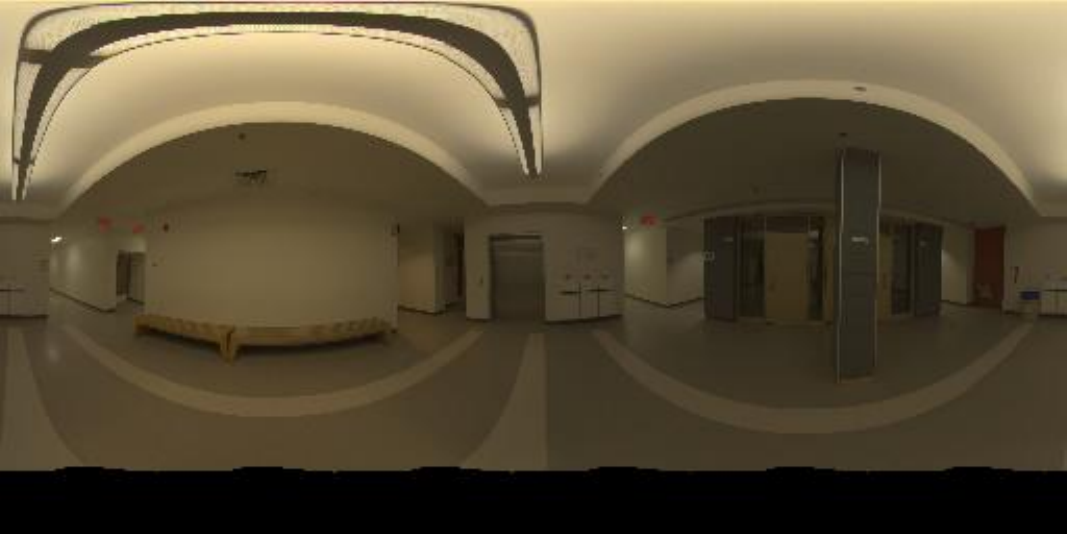}} & 
        \noindent\parbox[c]{0.14\textwidth}{\includegraphics[height=0.100\textwidth]{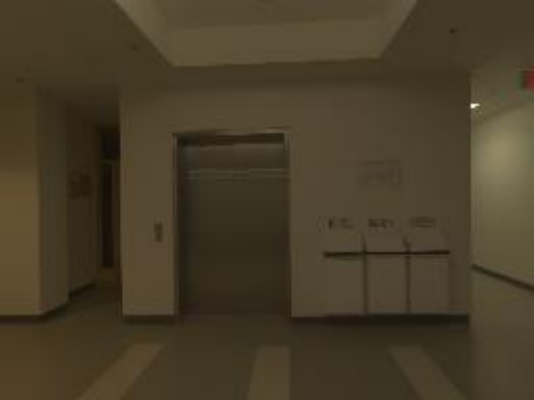}} &  
        
        \noindent\parbox[c]{0.100\textwidth}{\includegraphics[height=0.100\textwidth]{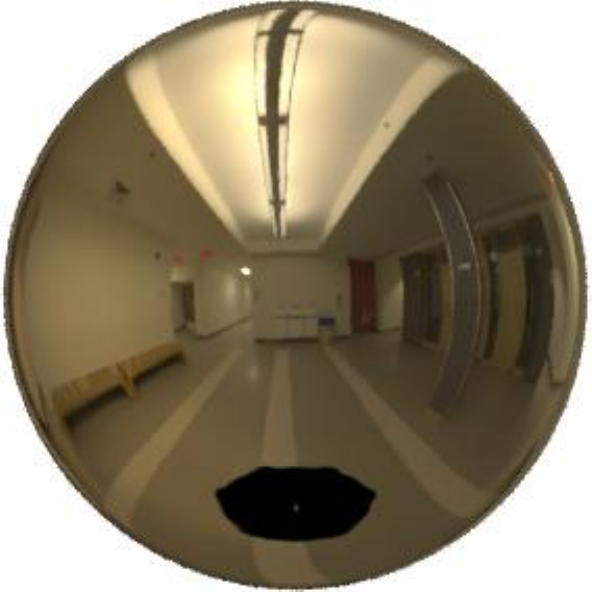}} & 
        \noindent\parbox[c]{0.100\textwidth}{\includegraphics[height=0.100\textwidth]{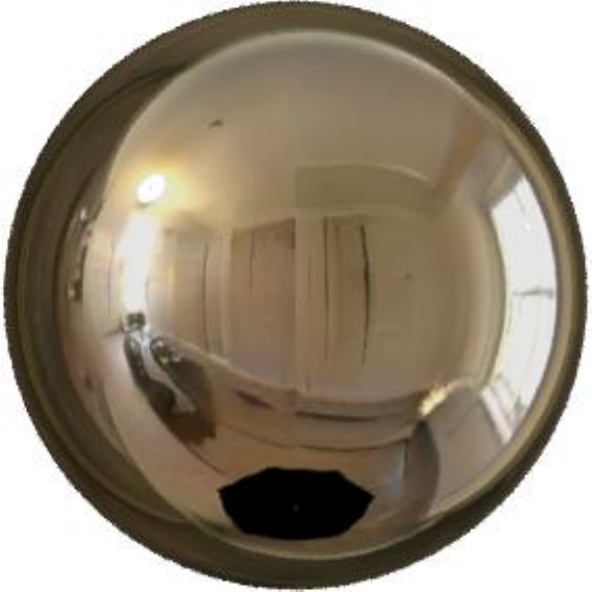}} & 
        
        \noindent\parbox[c]{0.100\textwidth}{\includegraphics[height=0.100\textwidth]{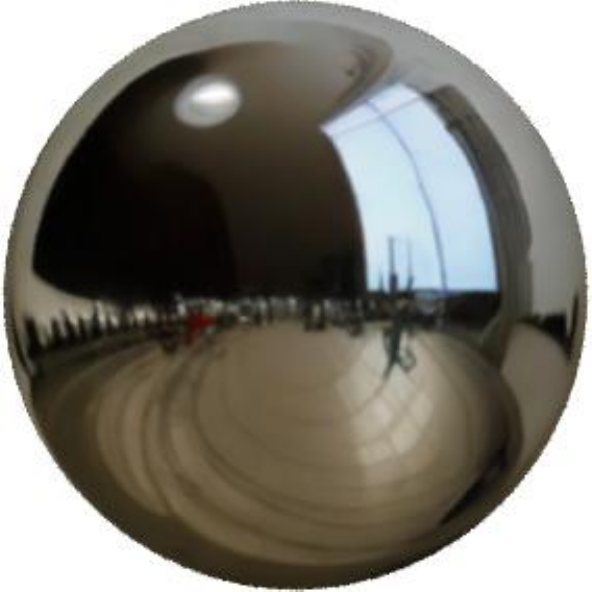}} & 
        \noindent\parbox[c]{0.100\textwidth}{\includegraphics[height=0.100\textwidth]{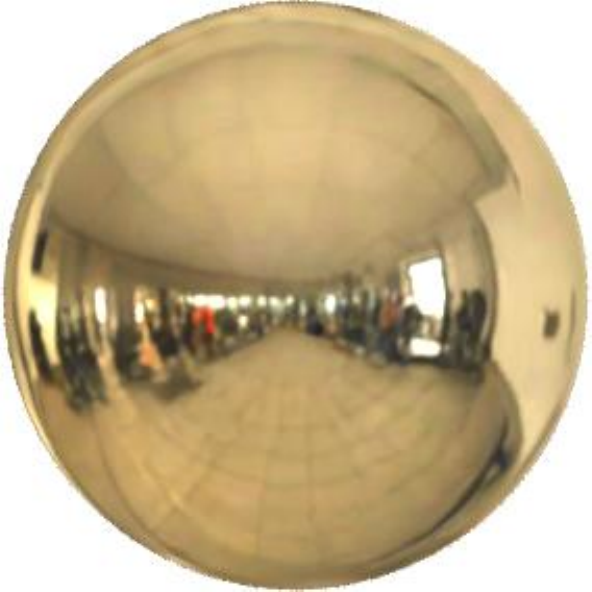}} &
        \noindent\parbox[c]{0.100\textwidth}{\includegraphics[height=0.100\textwidth]{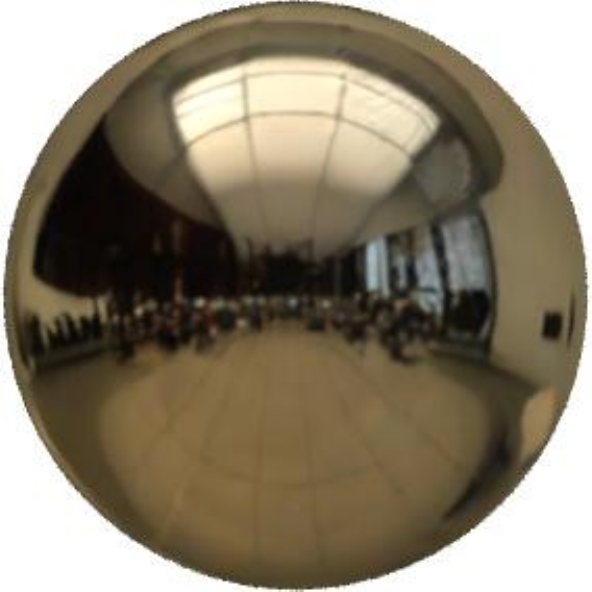}} & 
        \noindent\parbox[c]{0.100\textwidth}{\includegraphics[height=0.100\textwidth]{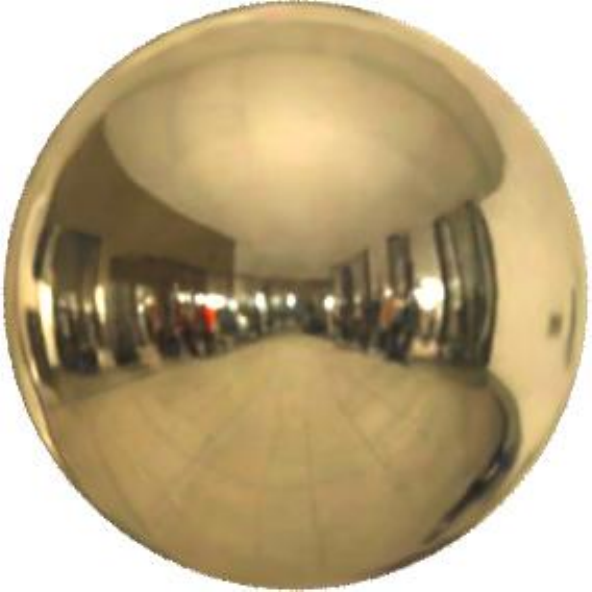}} & 
        \\

        \noindent\parbox[c]{0.205\textwidth}{\includegraphics[height=0.100\textwidth]{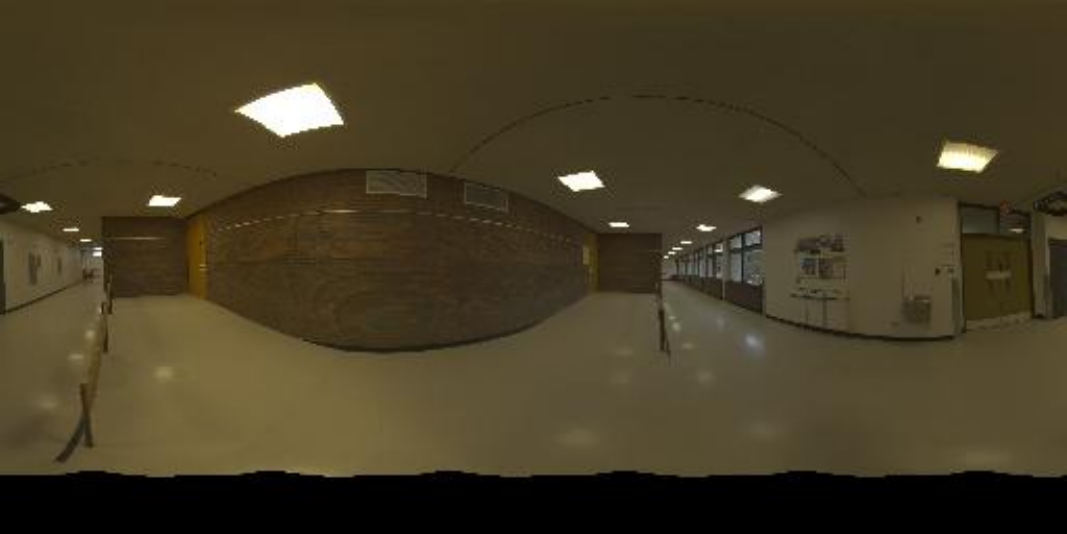}} & 
        \noindent\parbox[c]{0.14\textwidth}{\includegraphics[height=0.100\textwidth]{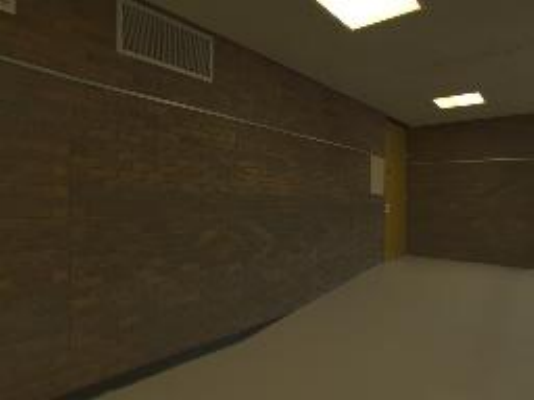}} &  
        
        \noindent\parbox[c]{0.100\textwidth}{\includegraphics[height=0.100\textwidth]{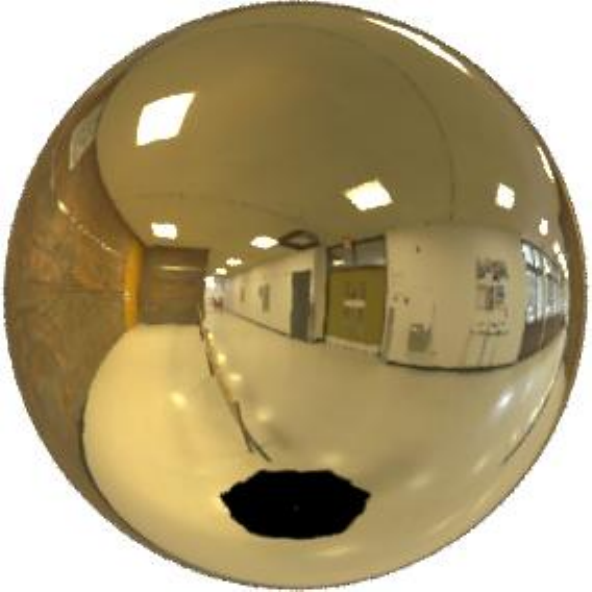}} & 
        \noindent\parbox[c]{0.100\textwidth}{\includegraphics[height=0.100\textwidth]{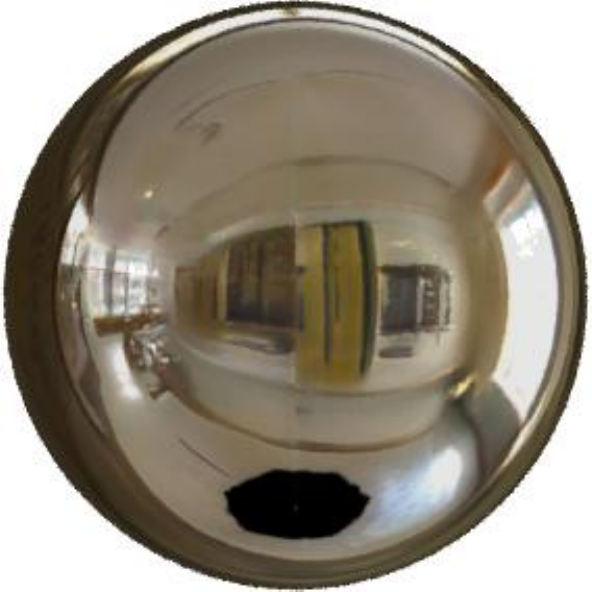}} & 
        
        \noindent\parbox[c]{0.100\textwidth}{\includegraphics[height=0.100\textwidth]{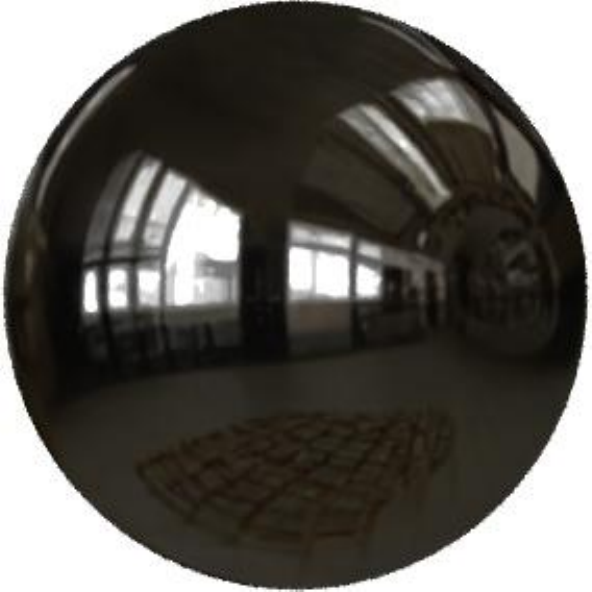}} & 
        \noindent\parbox[c]{0.100\textwidth}{\includegraphics[height=0.100\textwidth]{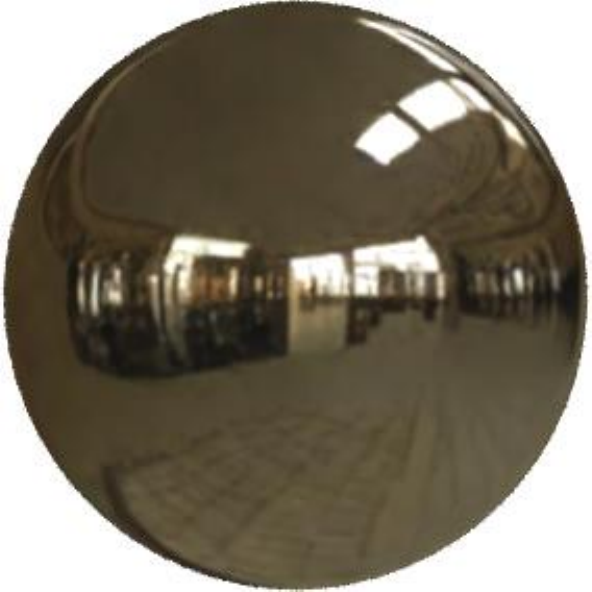}} &
        \noindent\parbox[c]{0.100\textwidth}{\includegraphics[height=0.100\textwidth]{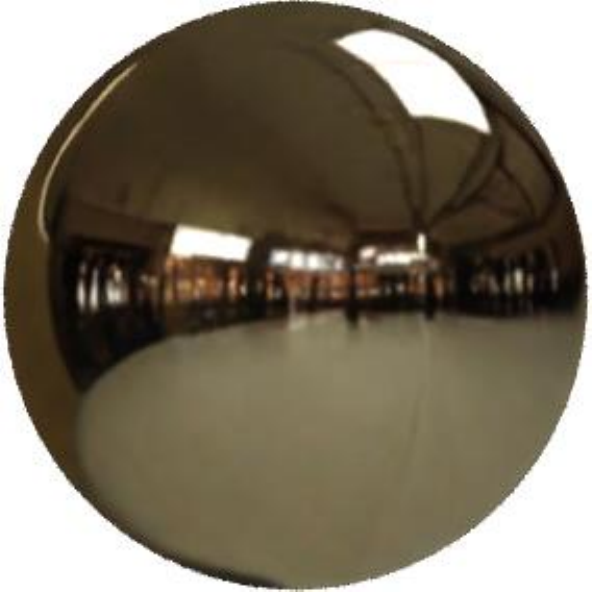}} & 
        \noindent\parbox[c]{0.100\textwidth}{\includegraphics[height=0.100\textwidth]{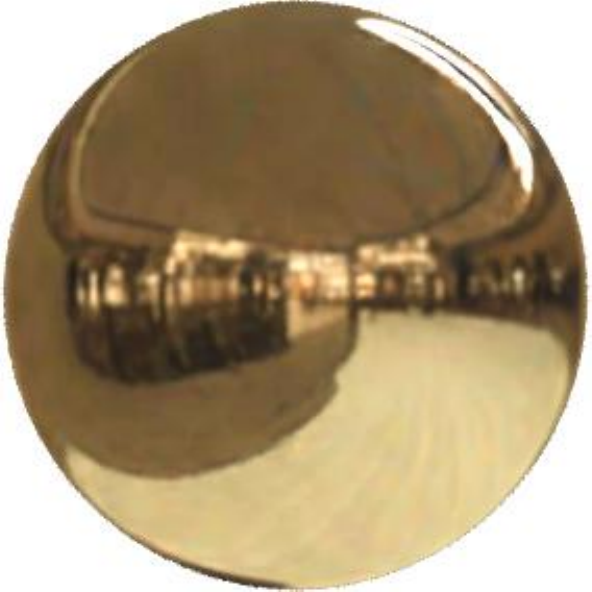}} & 
        \\

        \noindent\parbox[c]{0.205\textwidth}{\includegraphics[height=0.100\textwidth]{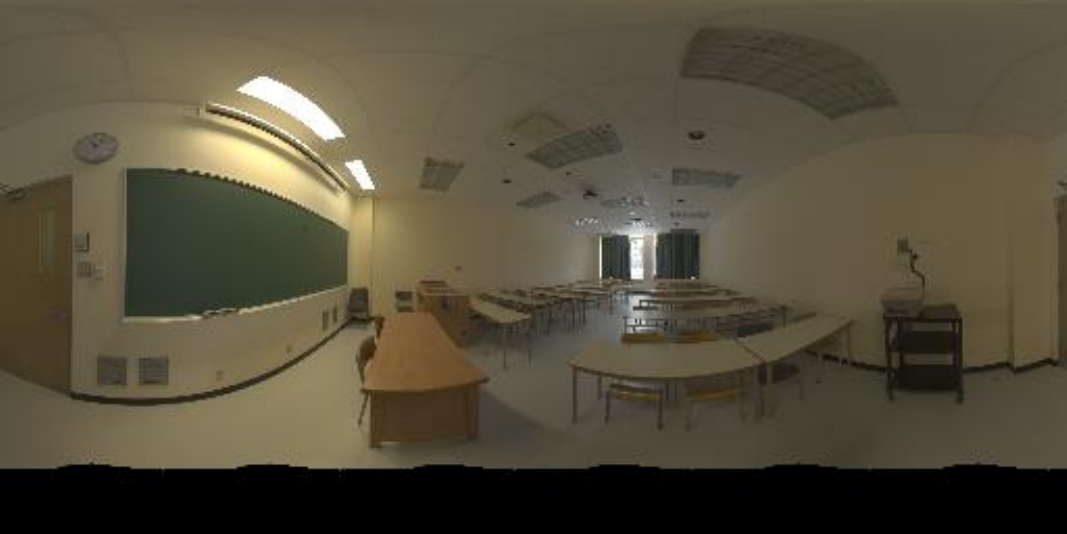}} & 
        \noindent\parbox[c]{0.14\textwidth}{\includegraphics[height=0.100\textwidth]{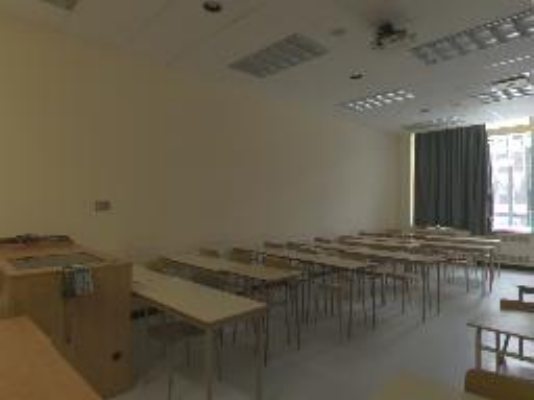}} &  
        
        \noindent\parbox[c]{0.100\textwidth}{\includegraphics[height=0.100\textwidth]{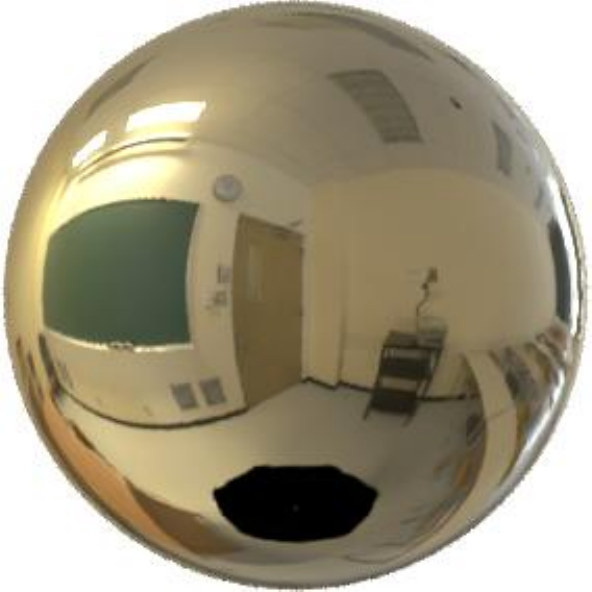}} & 
        \noindent\parbox[c]{0.100\textwidth}{\includegraphics[height=0.100\textwidth]{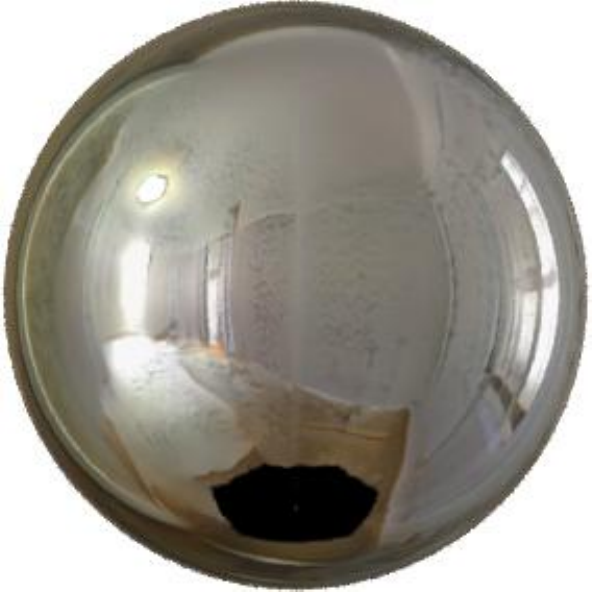}} & 
        
        \noindent\parbox[c]{0.100\textwidth}{\includegraphics[height=0.100\textwidth]{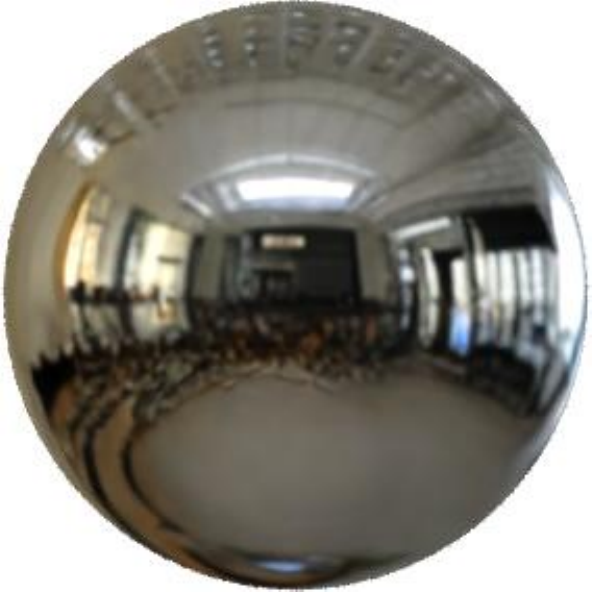}} & 
        \noindent\parbox[c]{0.100\textwidth}{\includegraphics[height=0.100\textwidth]{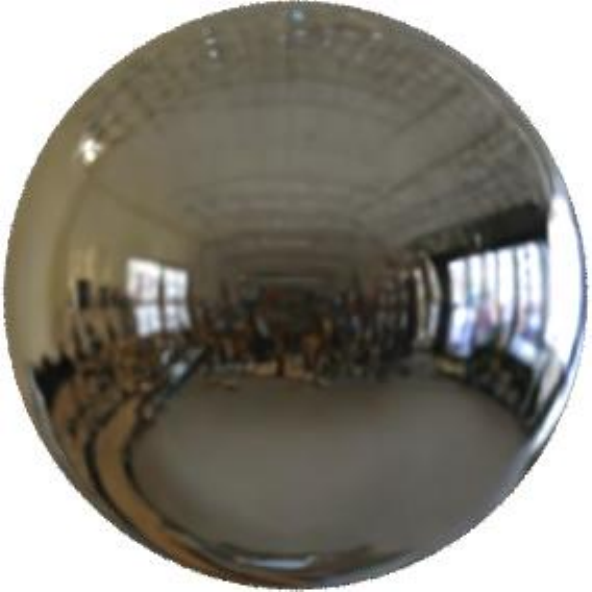}} &
        \noindent\parbox[c]{0.100\textwidth}{\includegraphics[height=0.100\textwidth]{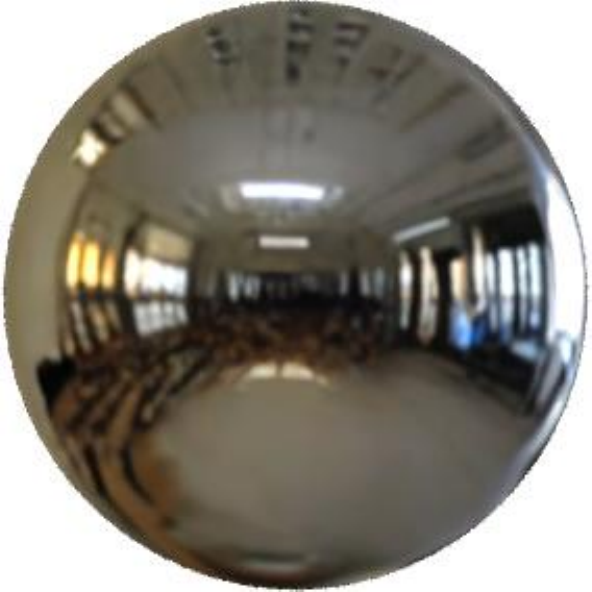}} & 
        \noindent\parbox[c]{0.100\textwidth}{\includegraphics[height=0.100\textwidth]{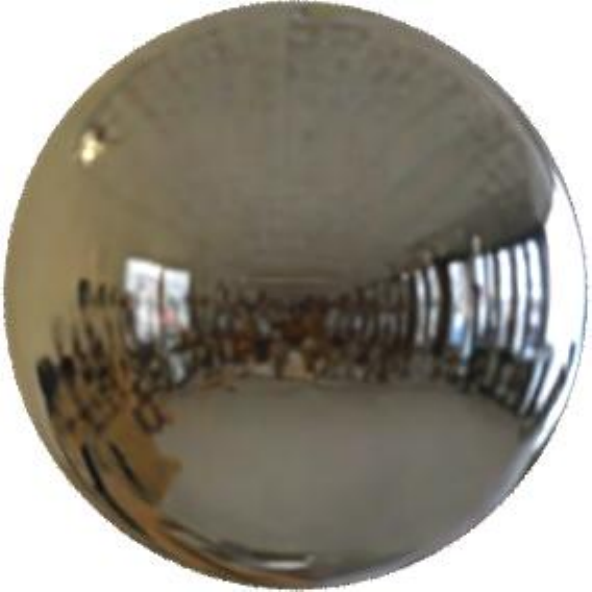}} & 
        \\

        \noindent\parbox[c]{0.205\textwidth}{\includegraphics[height=0.100\textwidth]{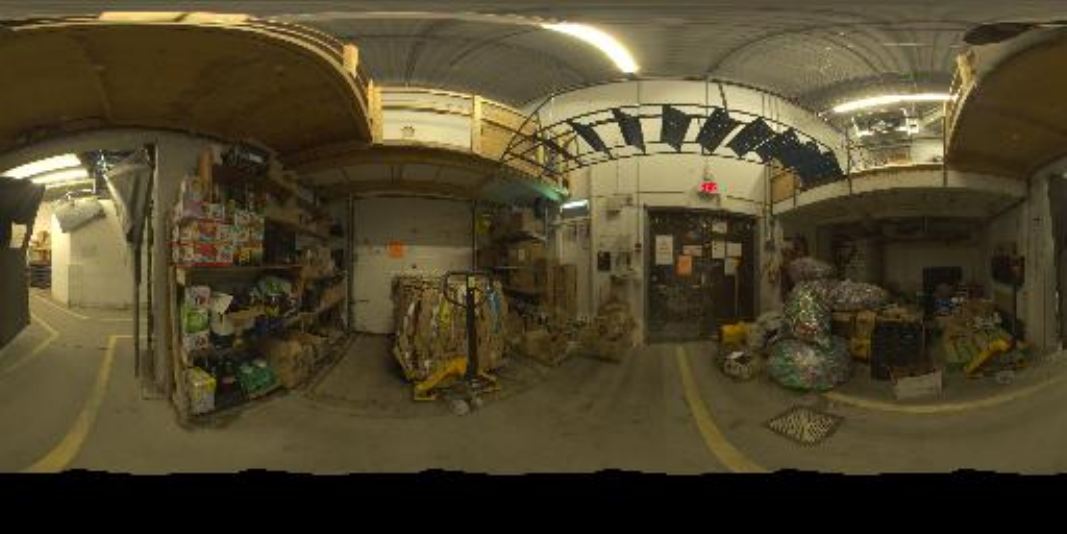}} & 
        \noindent\parbox[c]{0.14\textwidth}{\includegraphics[height=0.100\textwidth]{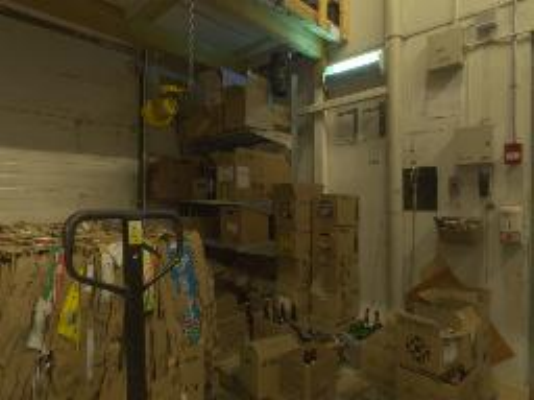}} &  
        
        \noindent\parbox[c]{0.100\textwidth}{\includegraphics[height=0.100\textwidth]{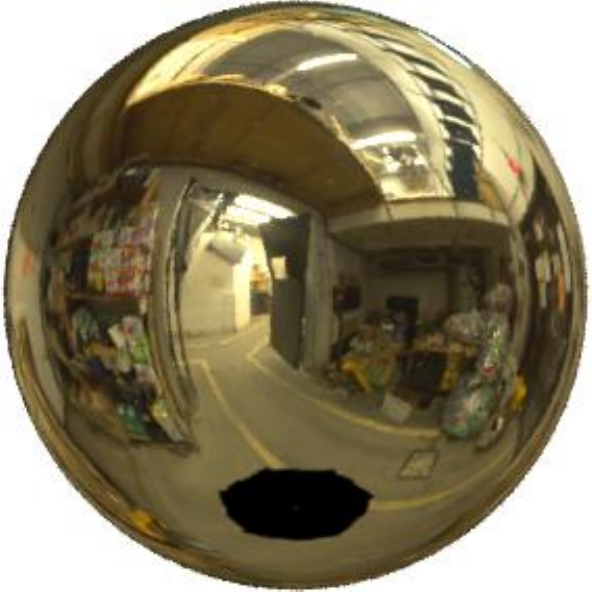}} & 
        \noindent\parbox[c]{0.100\textwidth}{\includegraphics[height=0.100\textwidth]{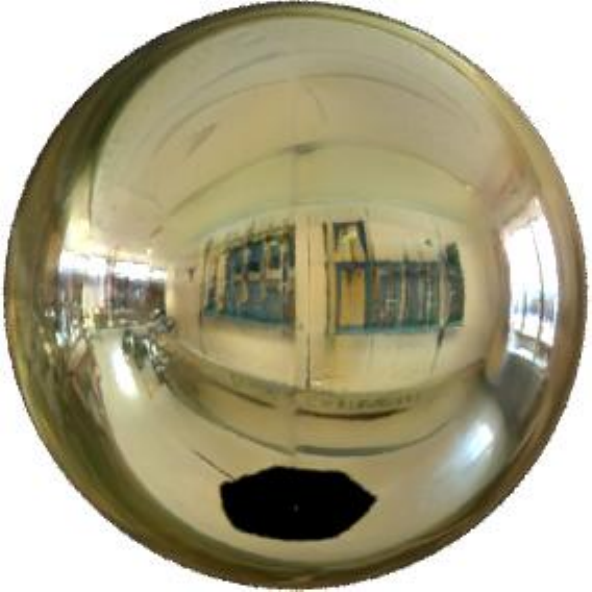}} & 
        
        \noindent\parbox[c]{0.100\textwidth}{\includegraphics[height=0.100\textwidth]{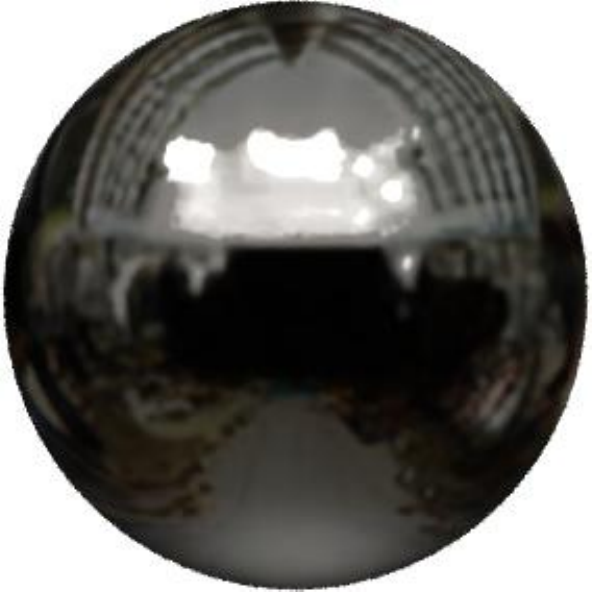}} & 
        \noindent\parbox[c]{0.100\textwidth}{\includegraphics[height=0.100\textwidth]{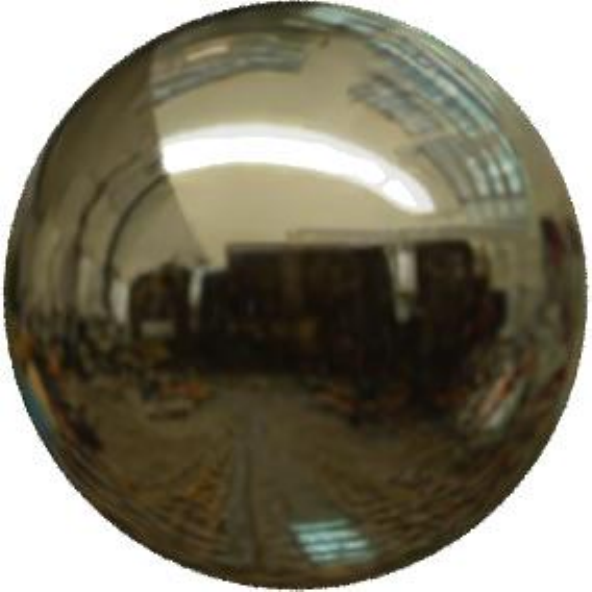}} &
        \noindent\parbox[c]{0.100\textwidth}{\includegraphics[height=0.100\textwidth]{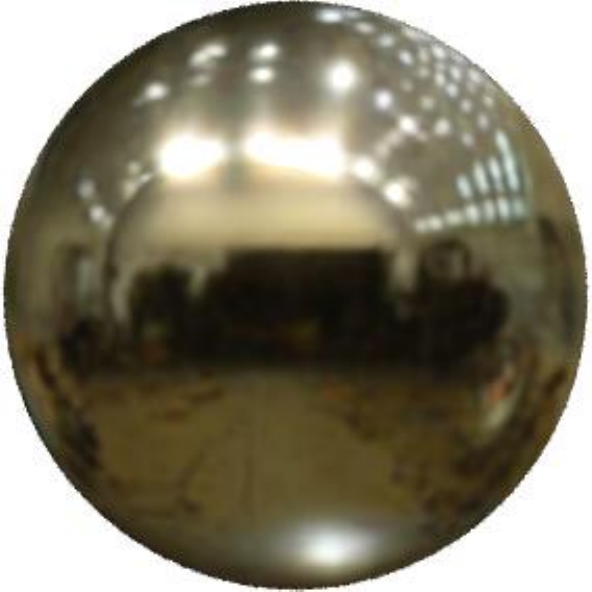}} & 
        \noindent\parbox[c]{0.100\textwidth}{\includegraphics[height=0.100\textwidth]{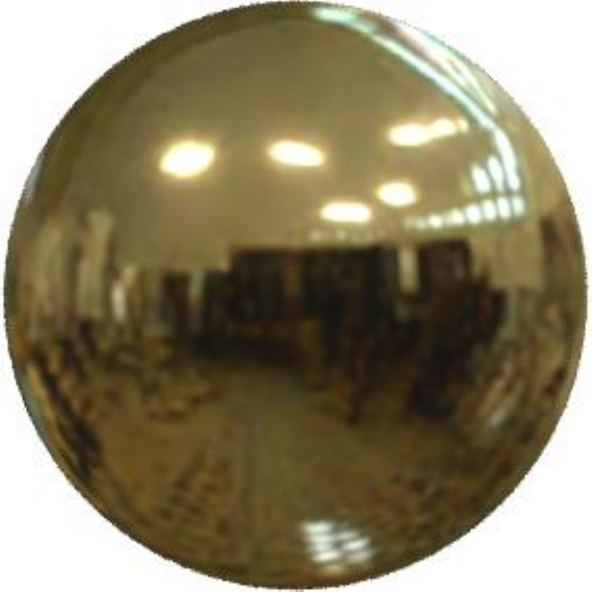}} & 
        \\

        \noindent\parbox[c]{0.205\textwidth}{\includegraphics[height=0.100\textwidth]{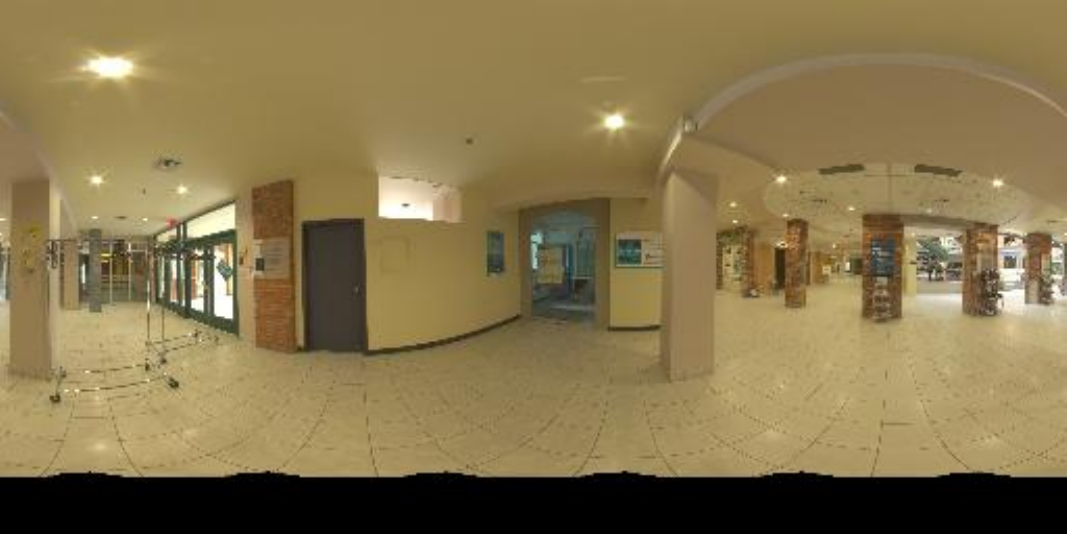}} & 
        \noindent\parbox[c]{0.14\textwidth}{\includegraphics[height=0.100\textwidth]{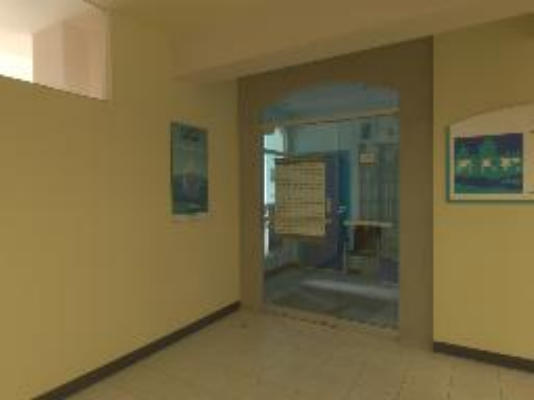}} &  
        
        \noindent\parbox[c]{0.100\textwidth}{\includegraphics[height=0.100\textwidth]{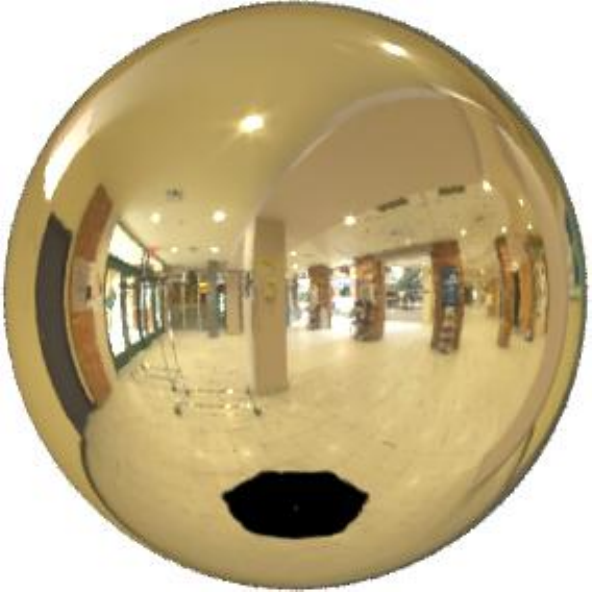}} & 
        \noindent\parbox[c]{0.100\textwidth}{\includegraphics[height=0.100\textwidth]{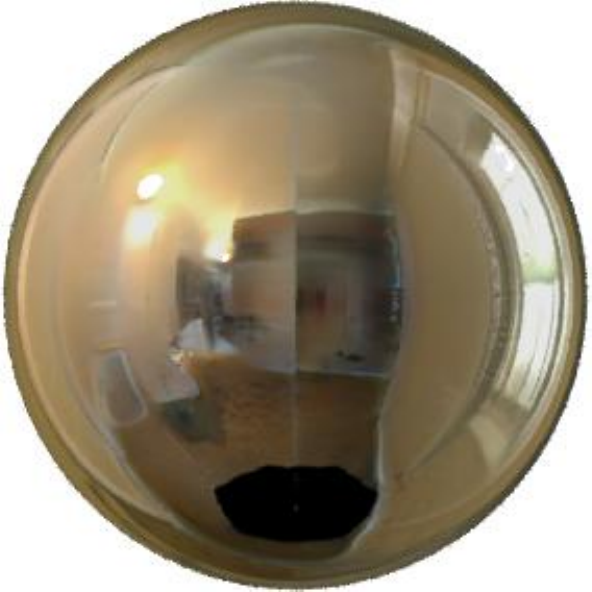}} & 
        
        \noindent\parbox[c]{0.100\textwidth}{\includegraphics[height=0.100\textwidth]{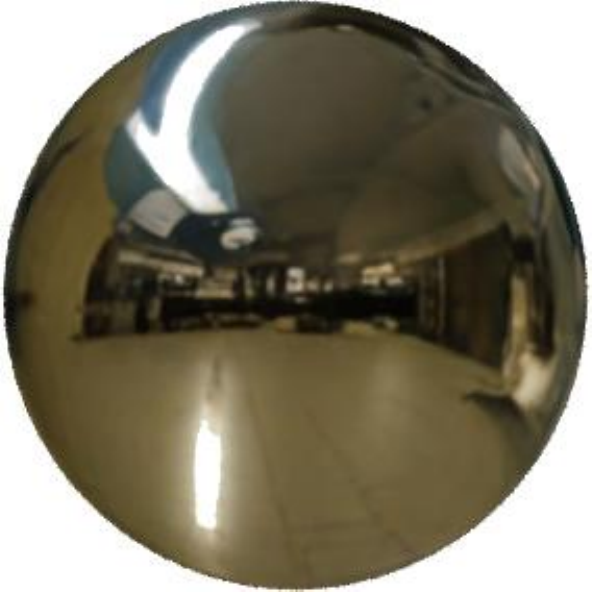}} & 
        \noindent\parbox[c]{0.100\textwidth}{\includegraphics[height=0.100\textwidth]{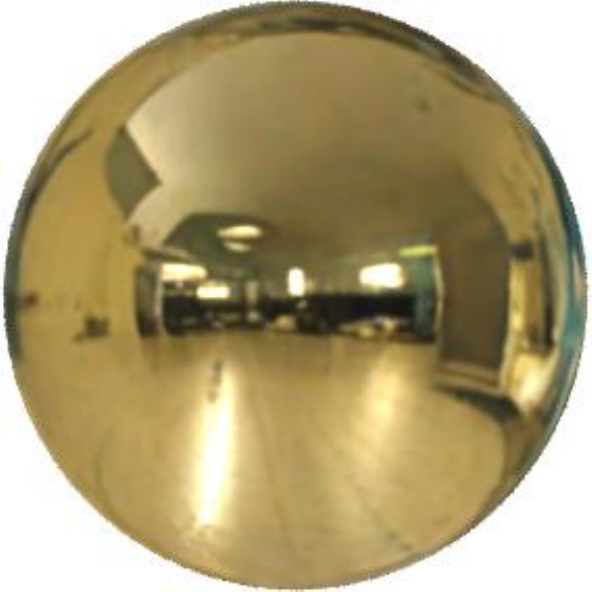}} &
        \noindent\parbox[c]{0.100\textwidth}{\includegraphics[height=0.100\textwidth]{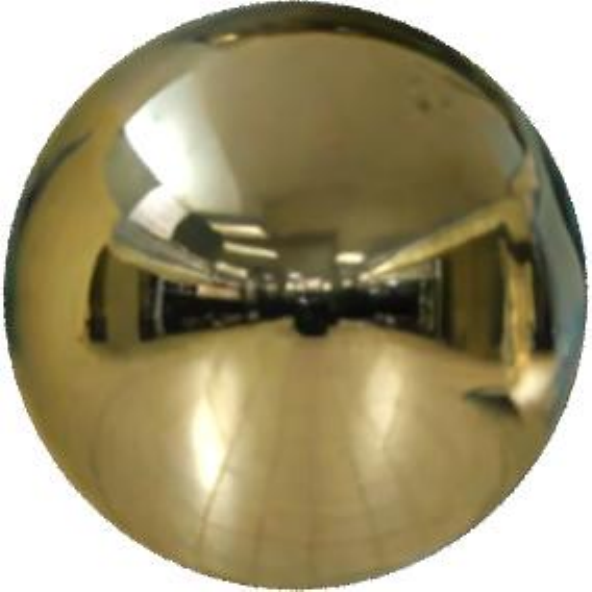}} & 
        \noindent\parbox[c]{0.100\textwidth}{\includegraphics[height=0.100\textwidth]{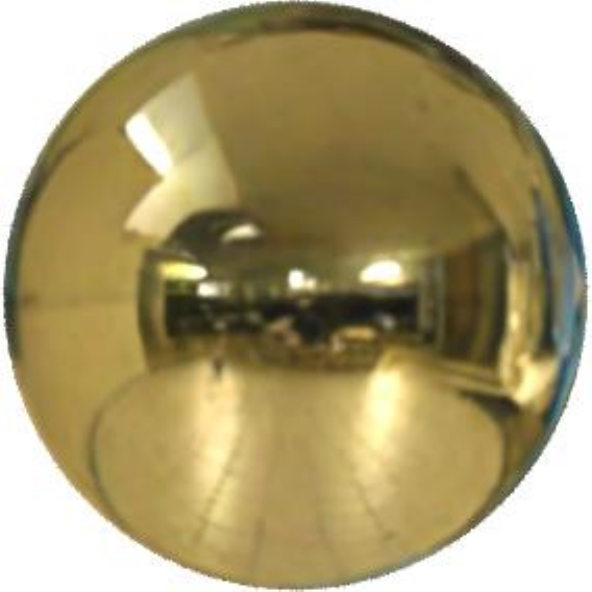}} & 
        \\

        \noindent\parbox[c]{0.205\textwidth}{\includegraphics[height=0.100\textwidth]{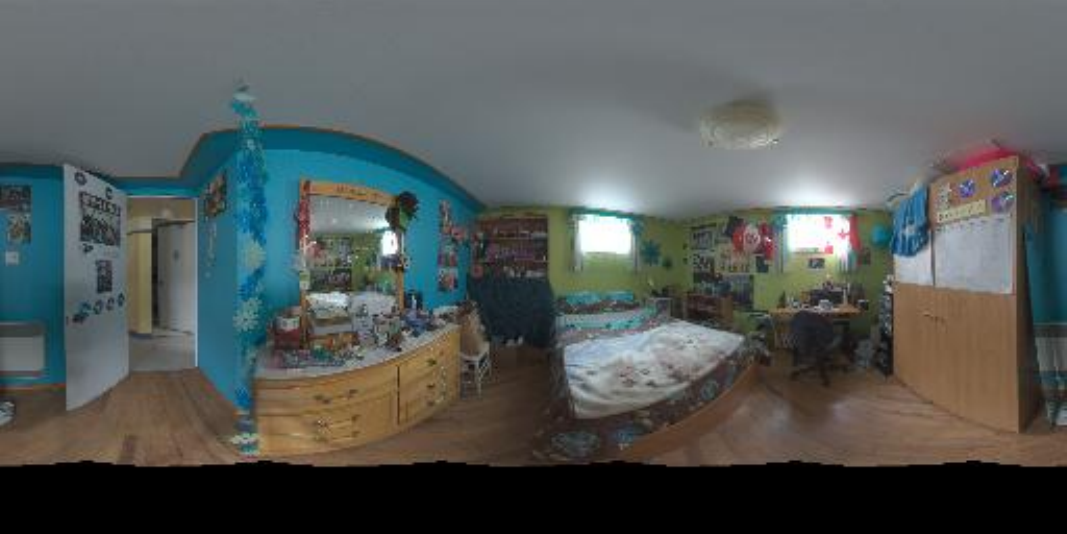}} & 
        \noindent\parbox[c]{0.14\textwidth}{\includegraphics[height=0.100\textwidth]{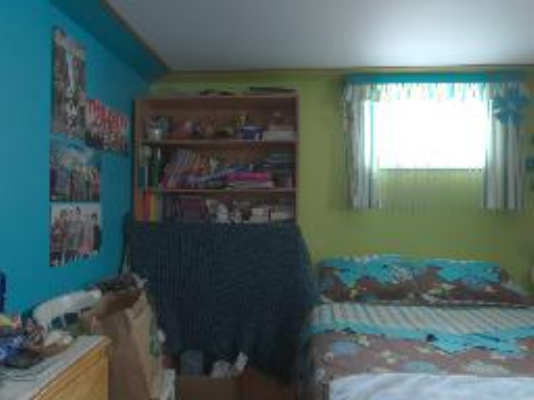}} &  
        
        \noindent\parbox[c]{0.100\textwidth}{\includegraphics[height=0.100\textwidth]{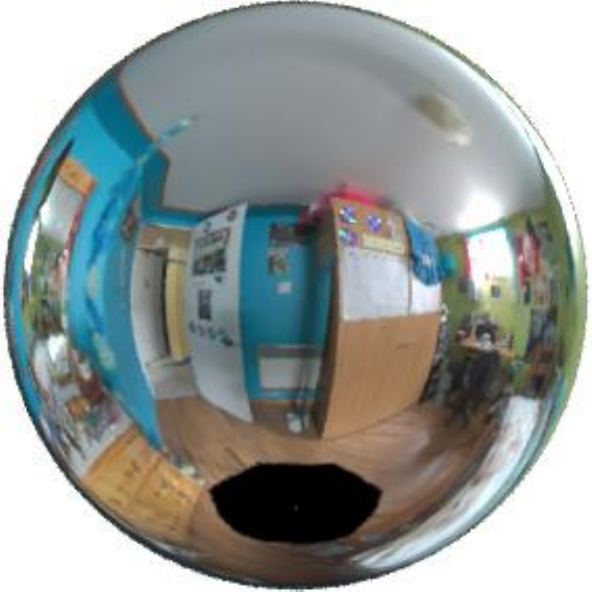}} & 
        \noindent\parbox[c]{0.100\textwidth}{\includegraphics[height=0.100\textwidth]{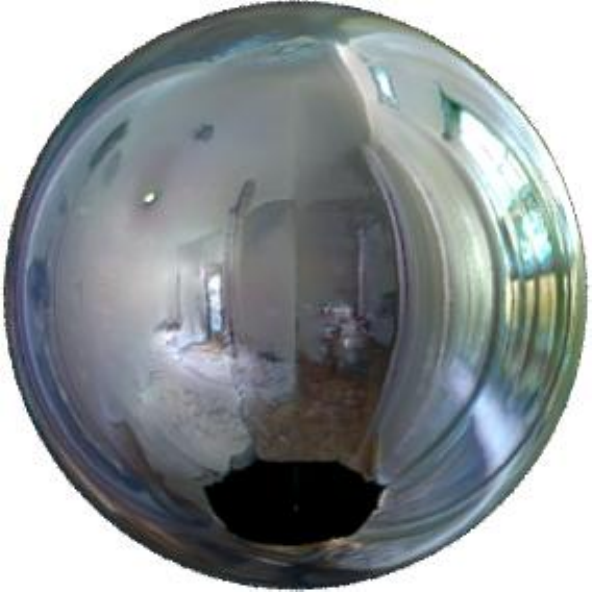}} & 
        
        \noindent\parbox[c]{0.100\textwidth}{\includegraphics[height=0.100\textwidth]{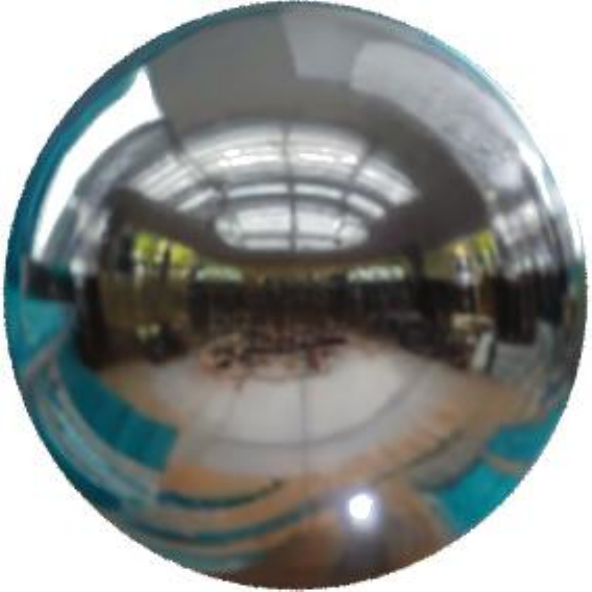}} & 
        \noindent\parbox[c]{0.100\textwidth}{\includegraphics[height=0.100\textwidth]{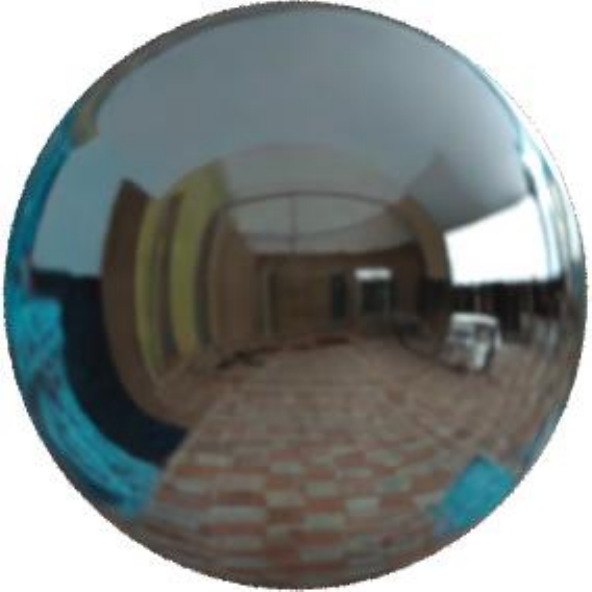}} &
        \noindent\parbox[c]{0.100\textwidth}{\includegraphics[height=0.100\textwidth]{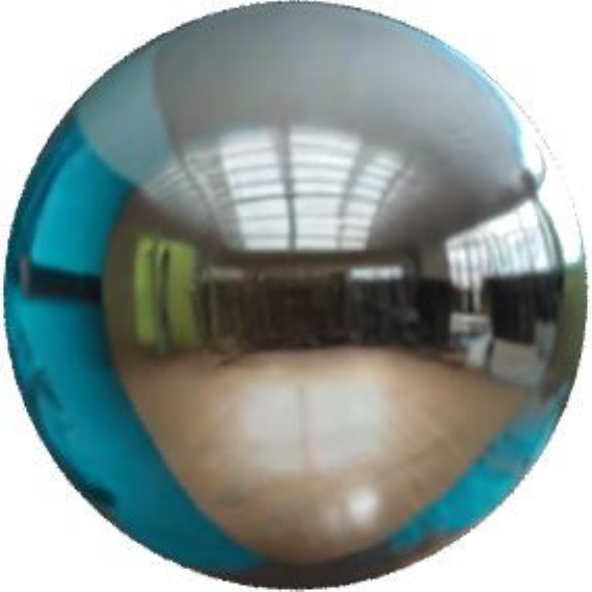}} & 
        \noindent\parbox[c]{0.100\textwidth}{\includegraphics[height=0.100\textwidth]{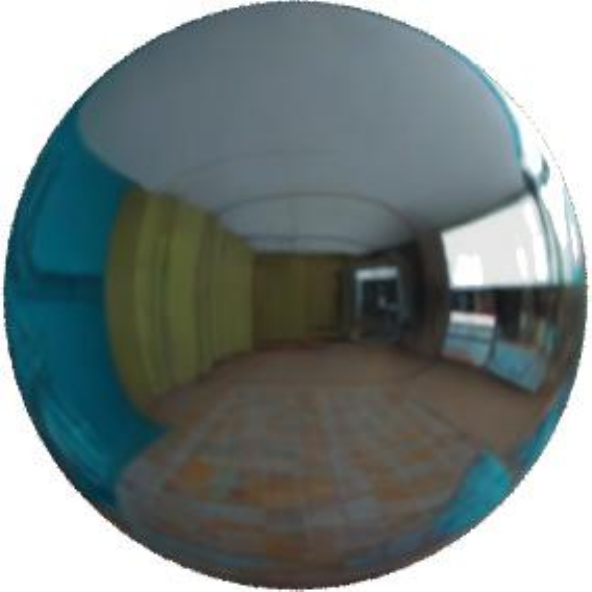}} & 
        \\

        \noindent\parbox[c]{0.205\textwidth}{\includegraphics[height=0.100\textwidth]{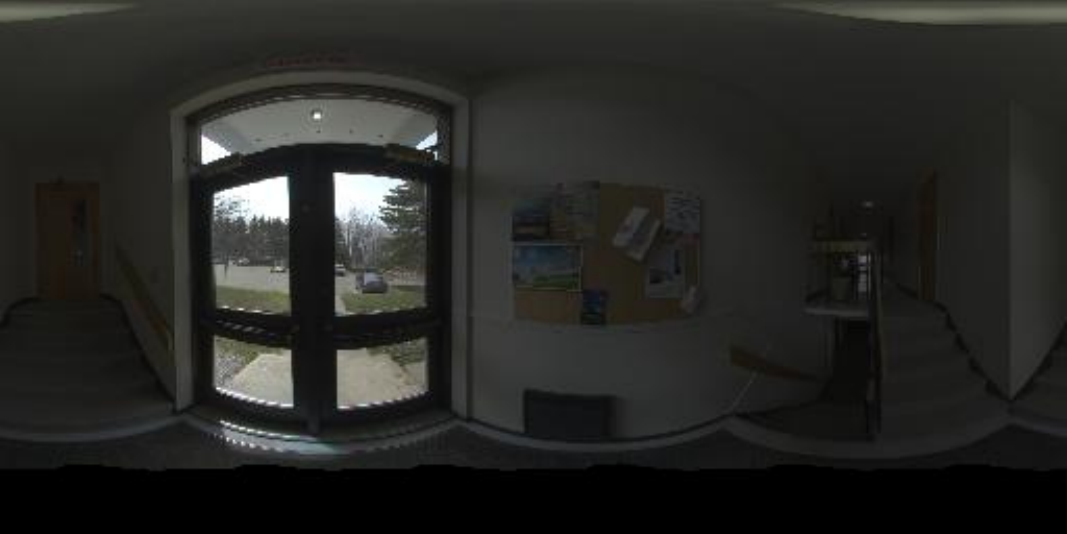}} & 
        \noindent\parbox[c]{0.14\textwidth}{\includegraphics[height=0.100\textwidth]{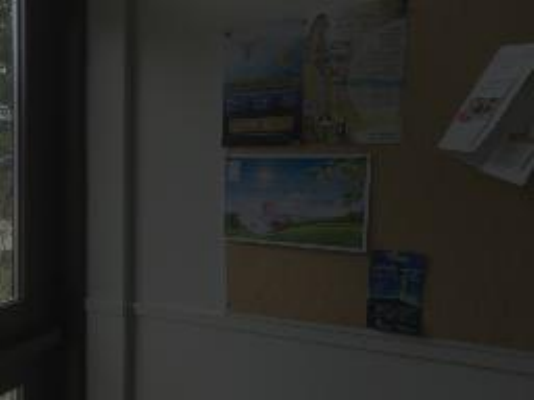}} &  
        
        \noindent\parbox[c]{0.100\textwidth}{\includegraphics[height=0.100\textwidth]{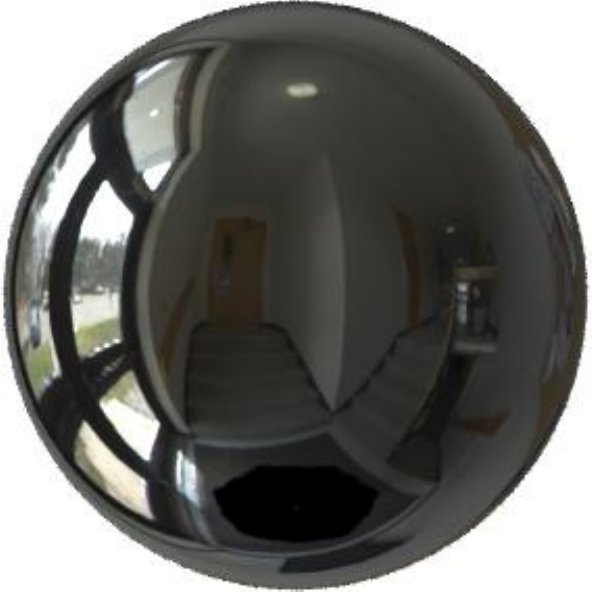}} & 
        \noindent\parbox[c]{0.100\textwidth}{\includegraphics[height=0.100\textwidth]{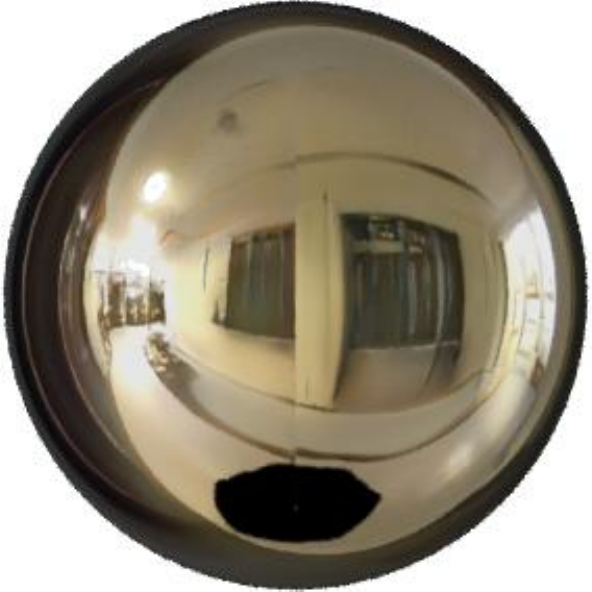}} & 
        
        \noindent\parbox[c]{0.100\textwidth}{\includegraphics[height=0.100\textwidth]{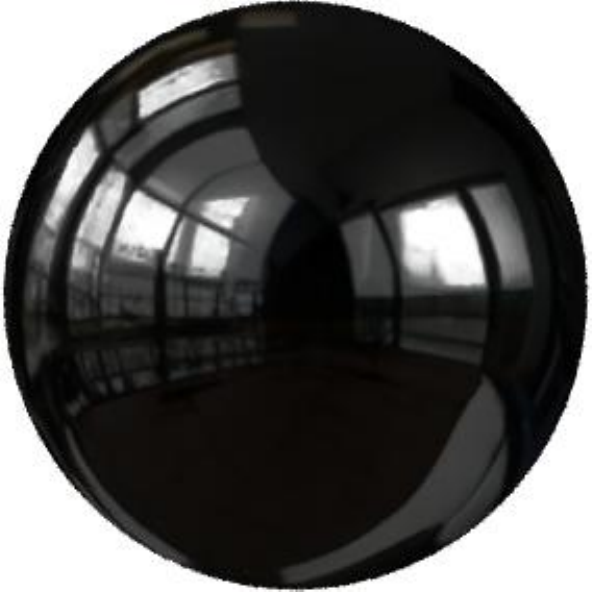}} & 
        \noindent\parbox[c]{0.100\textwidth}{\includegraphics[height=0.100\textwidth]{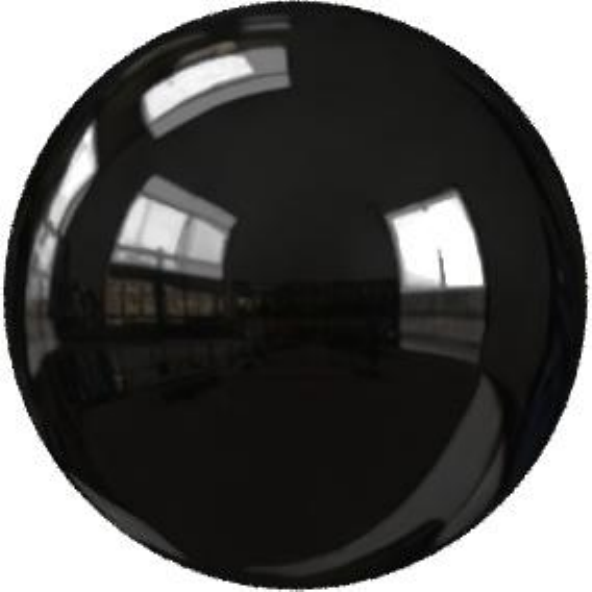}} &
        \noindent\parbox[c]{0.100\textwidth}{\includegraphics[height=0.100\textwidth]{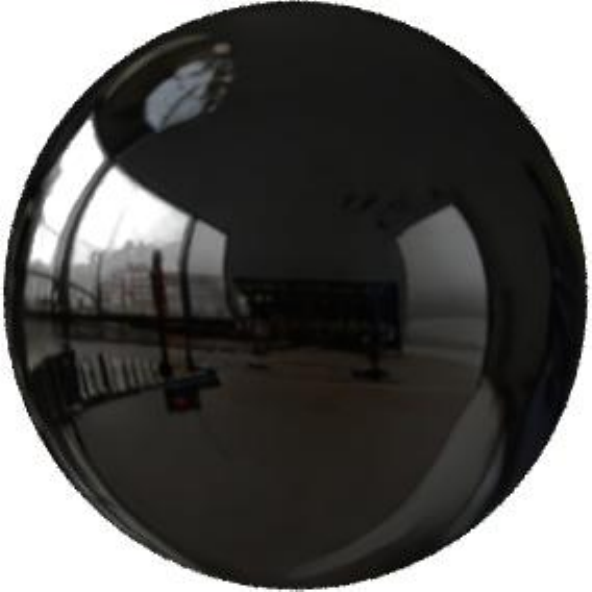}} & 
        \noindent\parbox[c]{0.100\textwidth}{\includegraphics[height=0.100\textwidth]{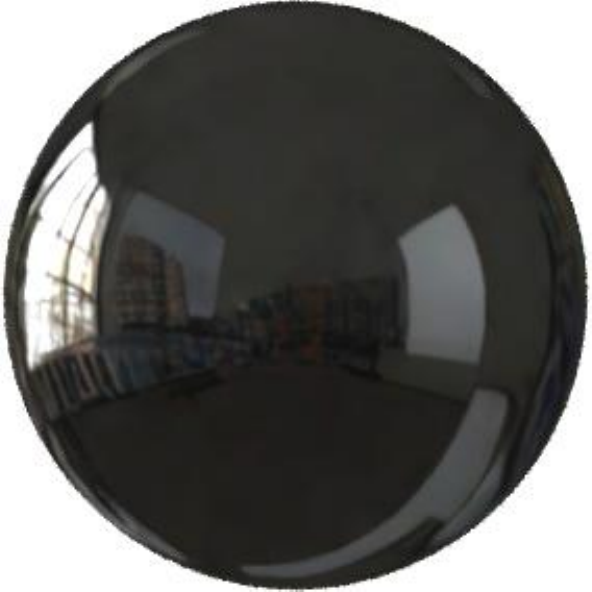}} & 
        \\
        
        \noindent\parbox[c]{0.205\textwidth}{\includegraphics[height=0.100\textwidth]{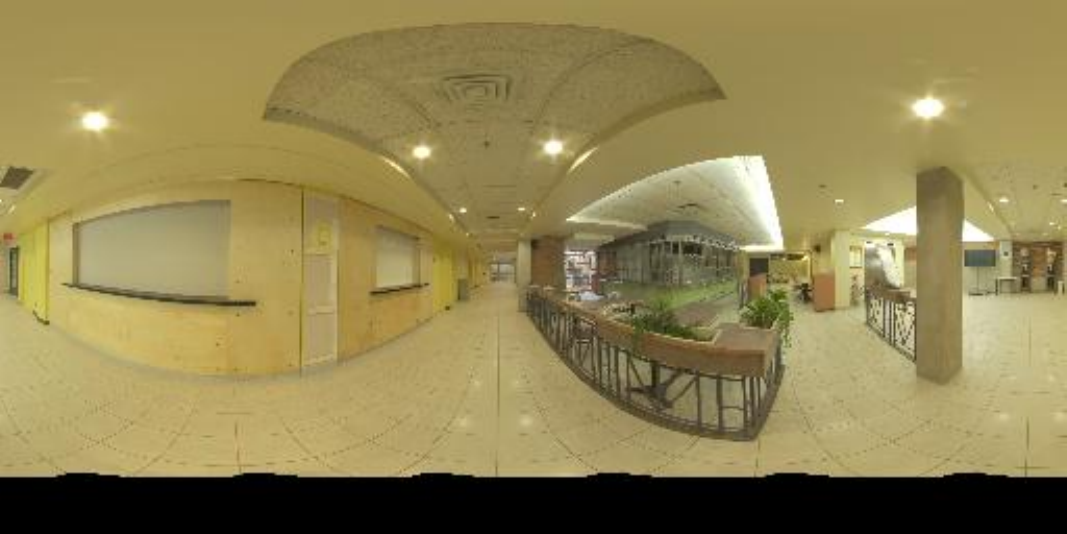}} & 
        \noindent\parbox[c]{0.14\textwidth}{\includegraphics[height=0.100\textwidth]{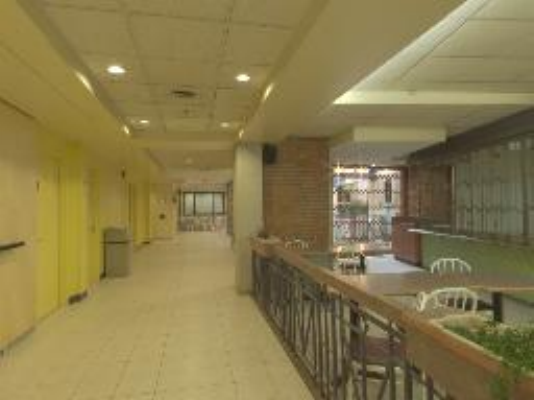}} &  
        
        \noindent\parbox[c]{0.100\textwidth}{\includegraphics[height=0.100\textwidth]{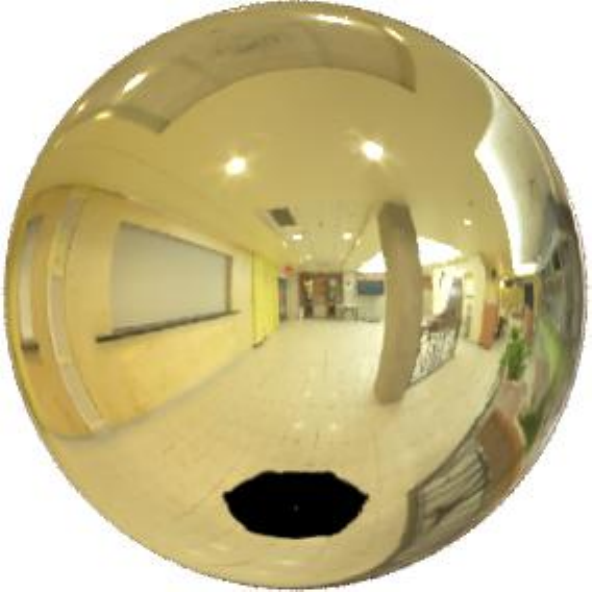}} & 
        \noindent\parbox[c]{0.100\textwidth}{\includegraphics[height=0.100\textwidth]{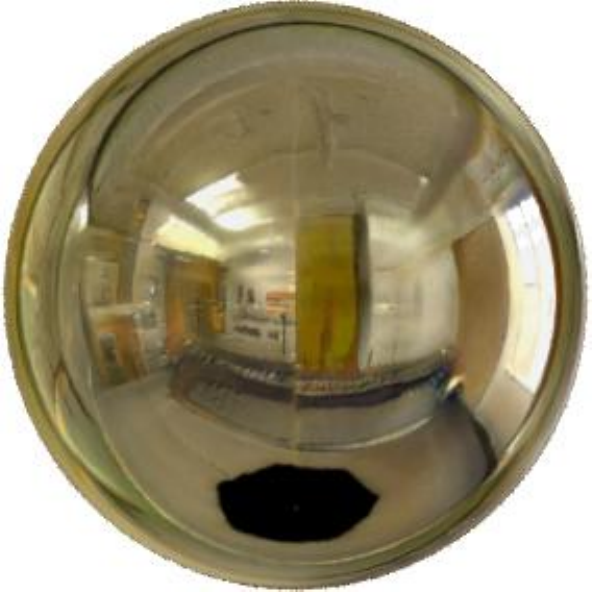}} & 
        
        \noindent\parbox[c]{0.100\textwidth}{\includegraphics[height=0.100\textwidth]{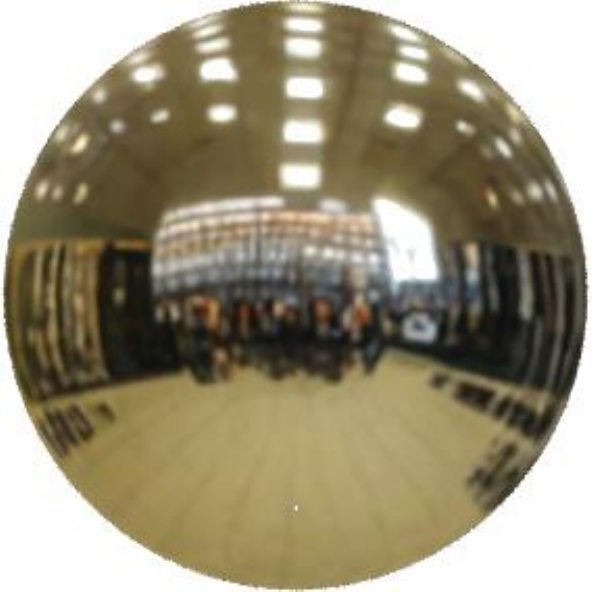}} & 
        \noindent\parbox[c]{0.100\textwidth}{\includegraphics[height=0.100\textwidth]{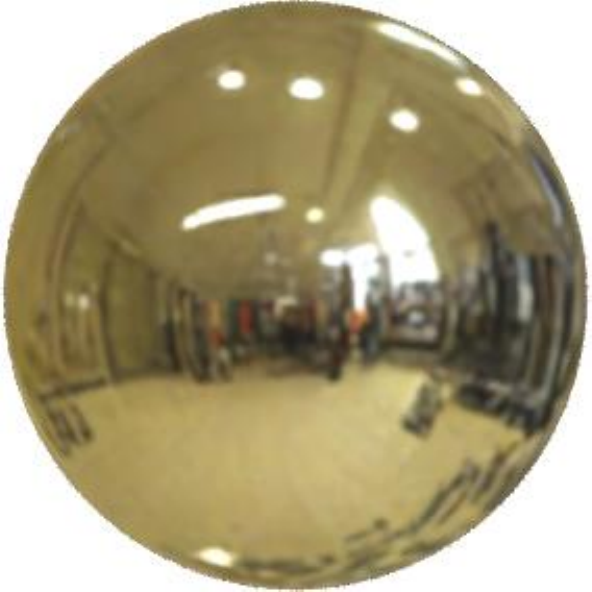}} &
        \noindent\parbox[c]{0.100\textwidth}{\includegraphics[height=0.100\textwidth]{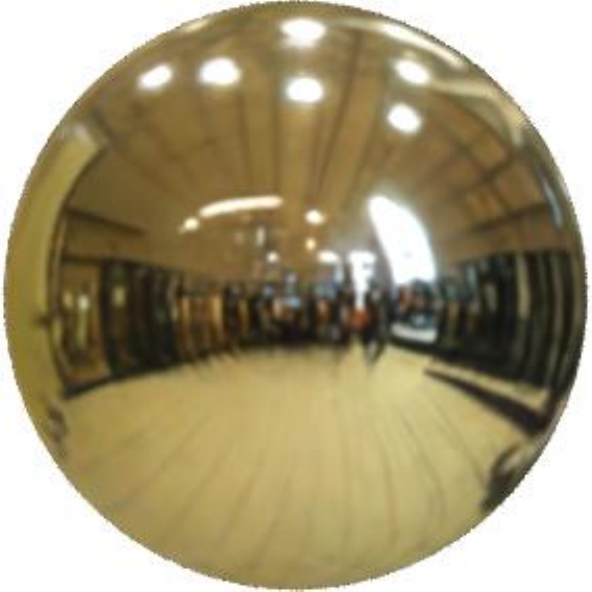}} & 
        \noindent\parbox[c]{0.100\textwidth}{\includegraphics[height=0.100\textwidth]{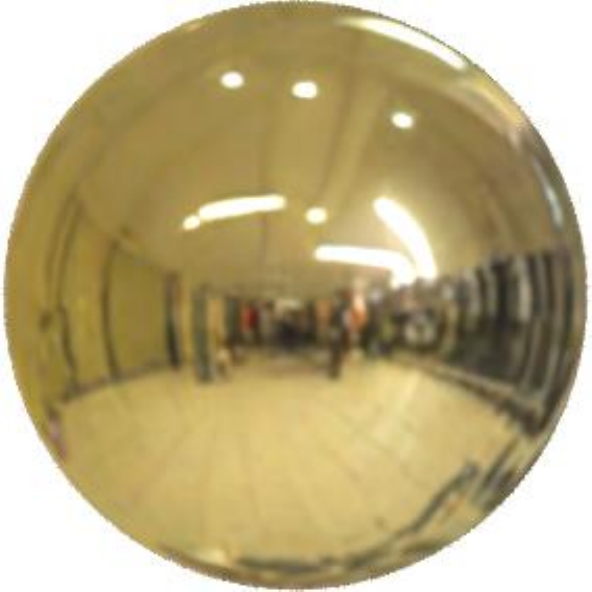}} & 
        \\

        \end{tabu}
    \caption{
    Qualitative results for the Laval indoor dataset using mirror balls.}
    \label{fig:additional_indoor_mirror}
\end{figure*}

\tabulinesep=0.5pt
\begin{figure*}[!t]
    \centering

        \begin{tabu} to \textwidth {
        @{}
        c@{}
        c@{}
        c@{}
        c@{}
        c@{}
        c@{}
        c@{}
        c@{}
        c@{}
    }

        \multicolumn{1}{c}{\shortstack{\scriptsize Ground truth map}}
        & 
        \multicolumn{1}{c}{\shortstack{\hspace{-6pt} \scriptsize Input}}
        &
        \multicolumn{1}{c}{\shortstack{\scriptsize Ground truth}}
        & 
        \multicolumn{1}{c}{\shortstack{\scriptsize StyleLight \cite{wang2022stylelight}}}
        & 
        \multicolumn{1}{c}{\shortstack{\scriptsize SDXL$^\dagger$}} &
        \multicolumn{1}{c}{\shortstack{\scriptsize \begin{tabular}[c]{@{}c@{}}SDXL$^\dagger$+LR \\ (ours, ablated)\end{tabular}}} &
        \multicolumn{1}{c}{\shortstack{\scriptsize \begin{tabular}[c]{@{}c@{}}SDXL$^\dagger$+I \\ (ours,ablated)\end{tabular}}}
        &
        \multicolumn{1}{c}{\shortstack{\scriptsize \begin{tabular}[c]{@{}c@{}}SDXL$^\dagger$+LR+I \\ (ours)\end{tabular}}} 
        \\

        \noindent\parbox[c]{0.205\textwidth}{\includegraphics[height=0.100\textwidth]{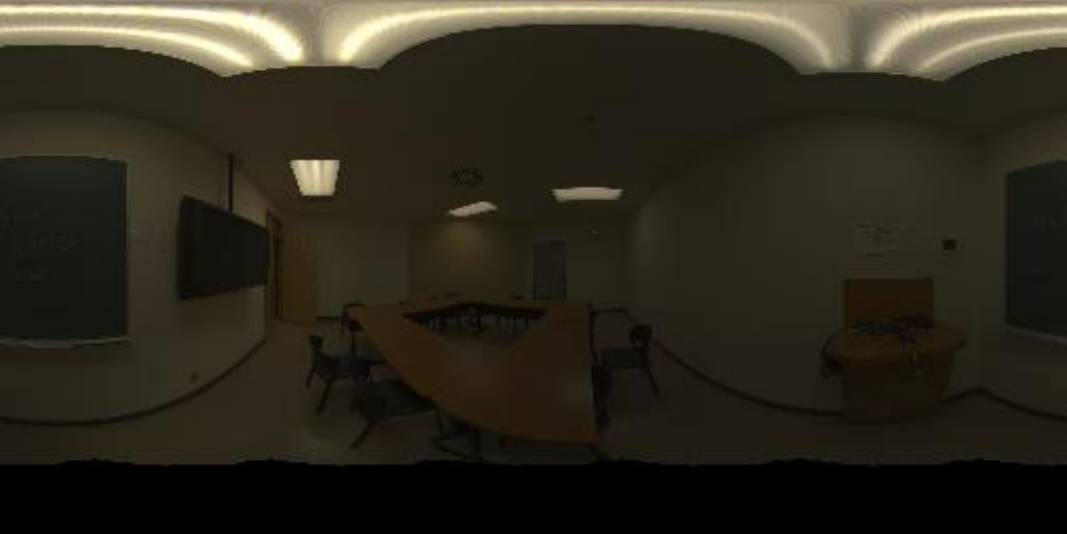}} & 
        \noindent\parbox[c]{0.14\textwidth}{\includegraphics[height=0.100\textwidth]{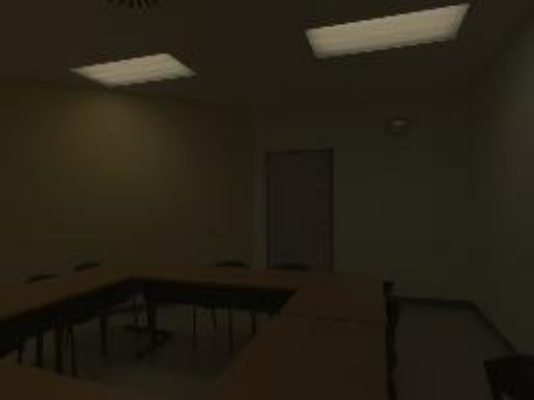}} &  
        
        \noindent\parbox[c]{0.100\textwidth}{\includegraphics[height=0.100\textwidth]{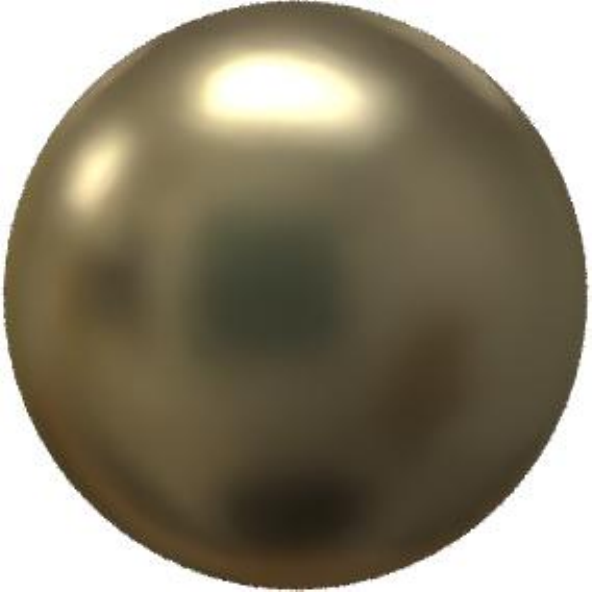}} & 
        \noindent\parbox[c]{0.100\textwidth}{\includegraphics[height=0.100\textwidth]{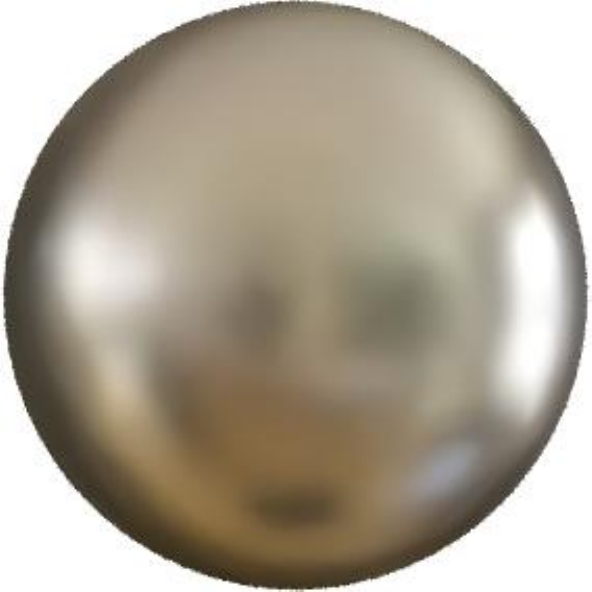}} & 
        
        \noindent\parbox[c]{0.100\textwidth}{\includegraphics[height=0.100\textwidth]{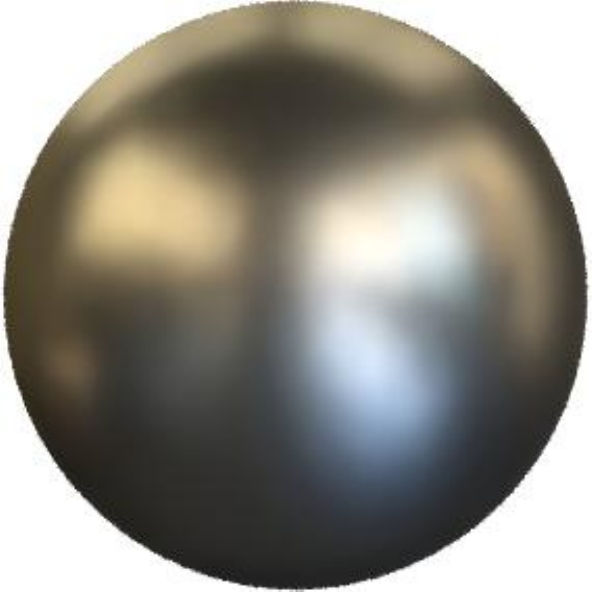}} & 
        \noindent\parbox[c]{0.100\textwidth}{\includegraphics[height=0.100\textwidth]{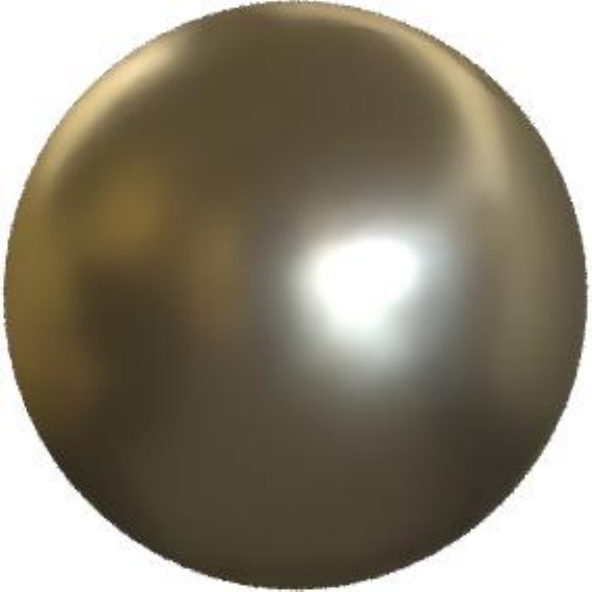}} &
        \noindent\parbox[c]{0.100\textwidth}{\includegraphics[height=0.100\textwidth]{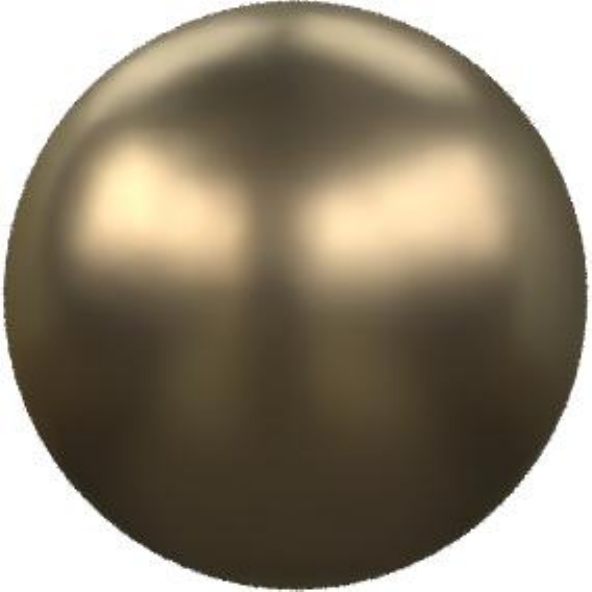}} & 
        \noindent\parbox[c]{0.100\textwidth}{\includegraphics[height=0.100\textwidth]{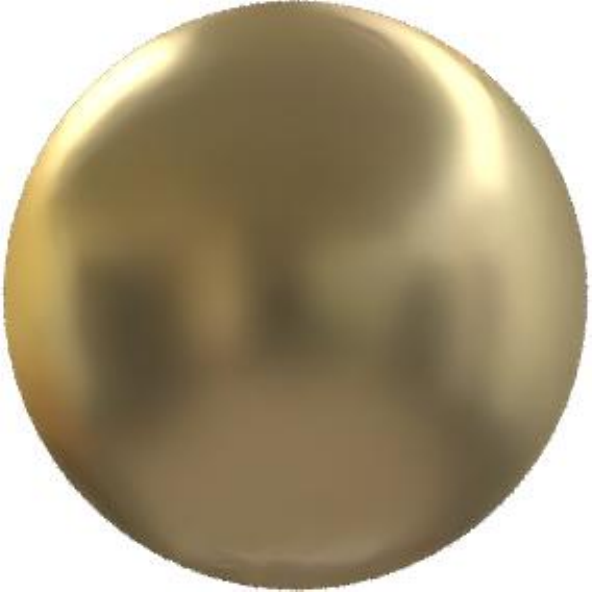}} & 
        \\

        \noindent\parbox[c]{0.205\textwidth}{\includegraphics[height=0.100\textwidth]{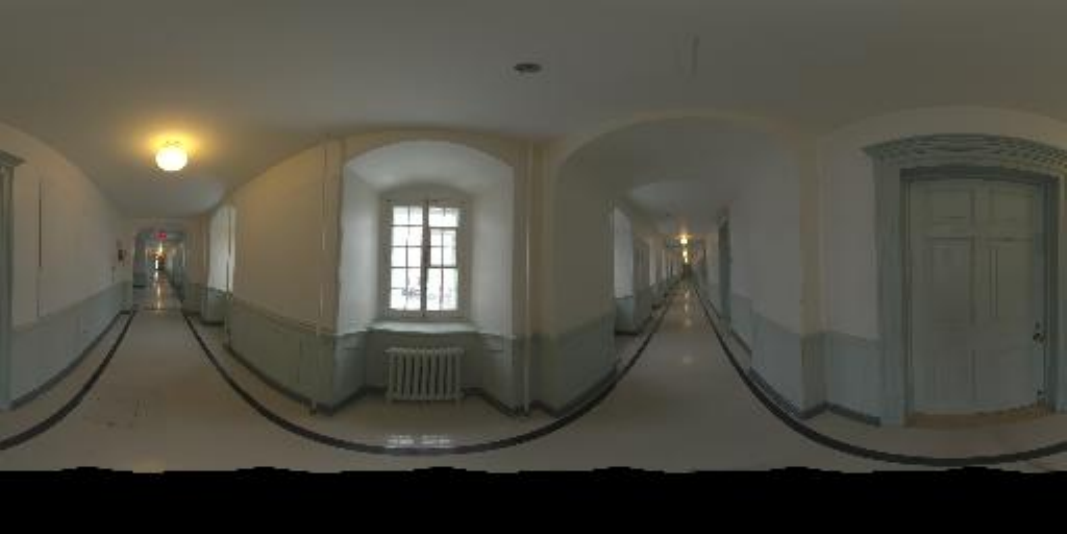}} & 
        \noindent\parbox[c]{0.14\textwidth}{\includegraphics[height=0.100\textwidth]{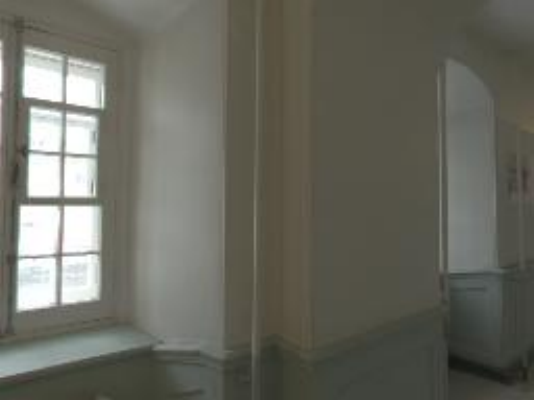}} &  
        
        \noindent\parbox[c]{0.100\textwidth}{\includegraphics[height=0.100\textwidth]{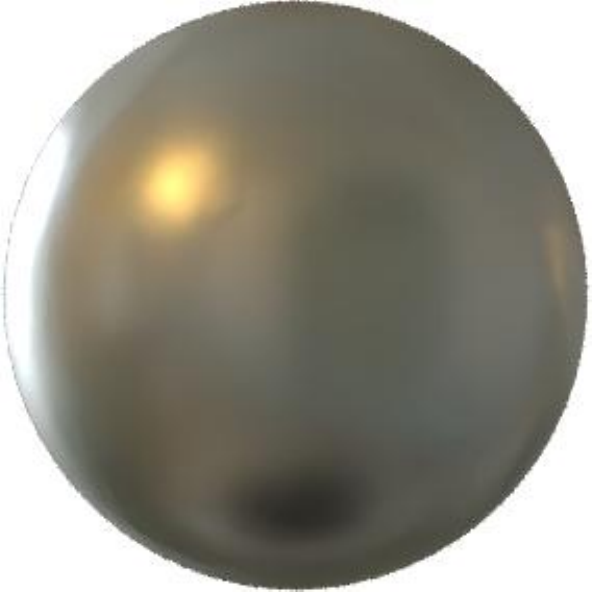}} & 
        \noindent\parbox[c]{0.100\textwidth}{\includegraphics[height=0.100\textwidth]{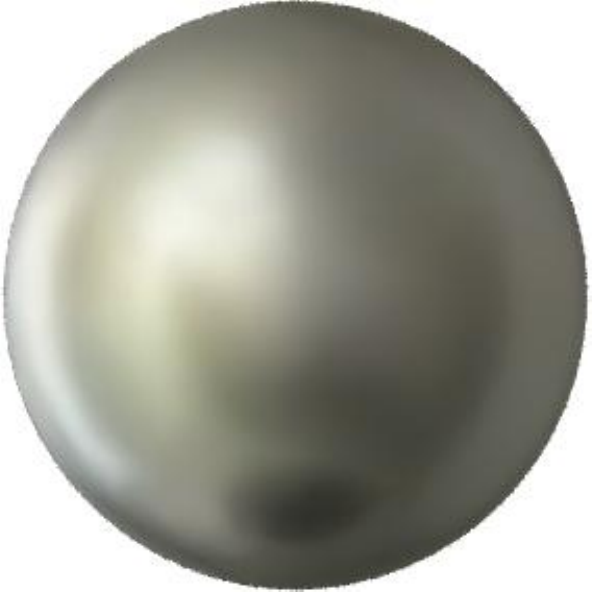}} & 
        
        \noindent\parbox[c]{0.100\textwidth}{\includegraphics[height=0.100\textwidth]{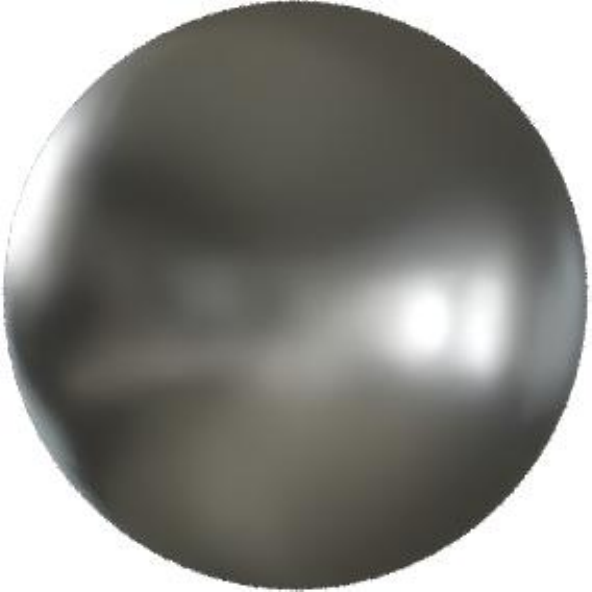}} & 
        \noindent\parbox[c]{0.100\textwidth}{\includegraphics[height=0.100\textwidth]{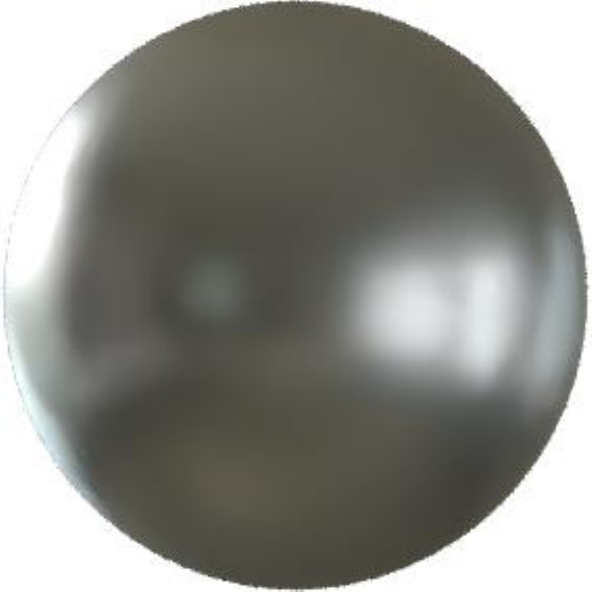}} &
        \noindent\parbox[c]{0.100\textwidth}{\includegraphics[height=0.100\textwidth]{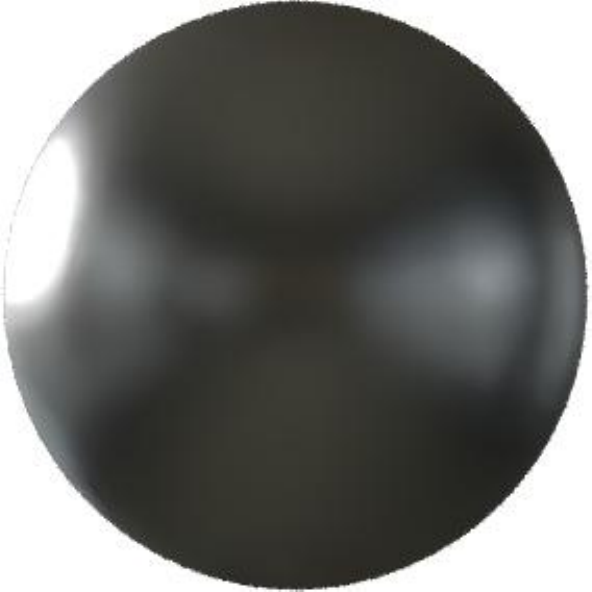}} & 
        \noindent\parbox[c]{0.100\textwidth}{\includegraphics[height=0.100\textwidth]{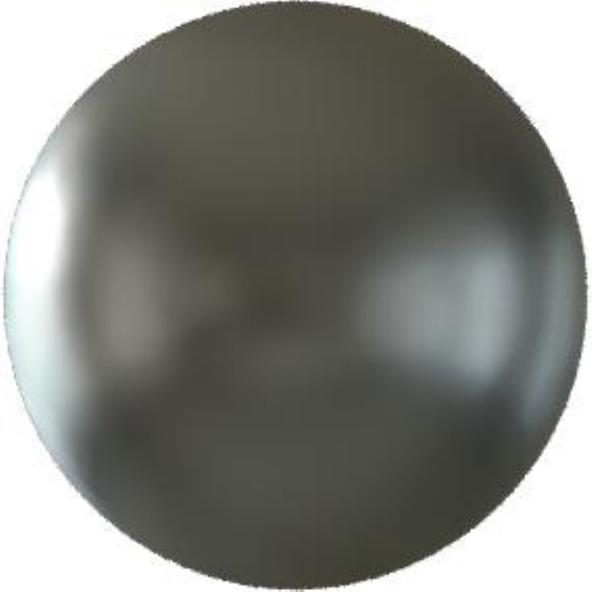}} & 
        \\

        \noindent\parbox[c]{0.205\textwidth}{\includegraphics[height=0.100\textwidth]{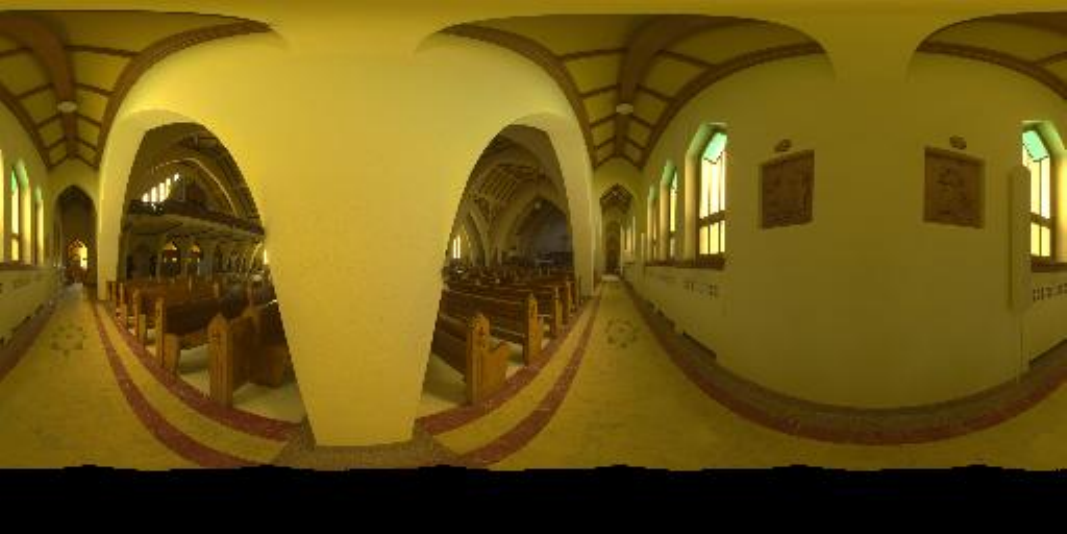}} & 
        \noindent\parbox[c]{0.14\textwidth}{\includegraphics[height=0.100\textwidth]{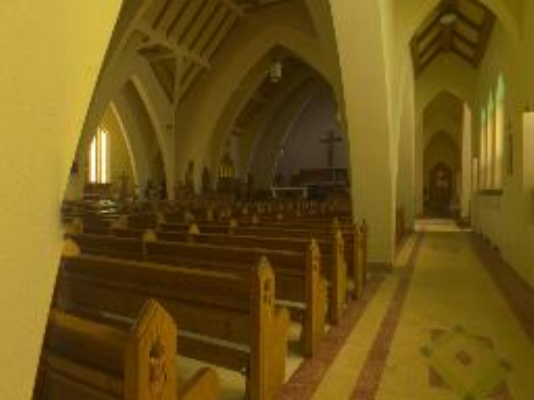}} &  
        
        \noindent\parbox[c]{0.100\textwidth}{\includegraphics[height=0.100\textwidth]{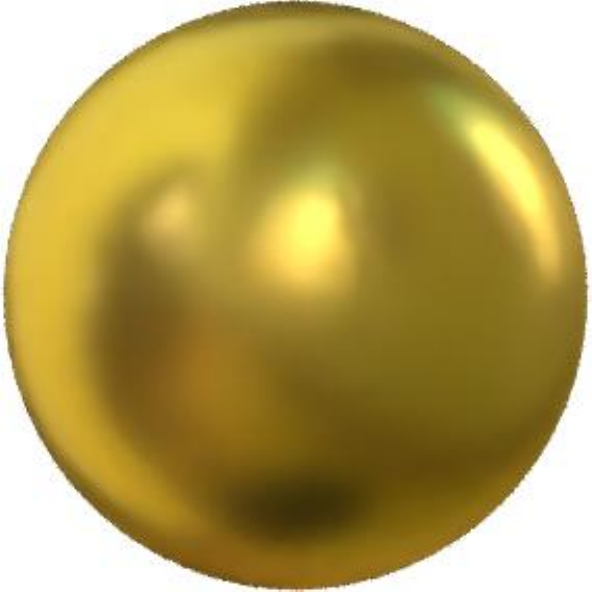}} & 
        \noindent\parbox[c]{0.100\textwidth}{\includegraphics[height=0.100\textwidth]{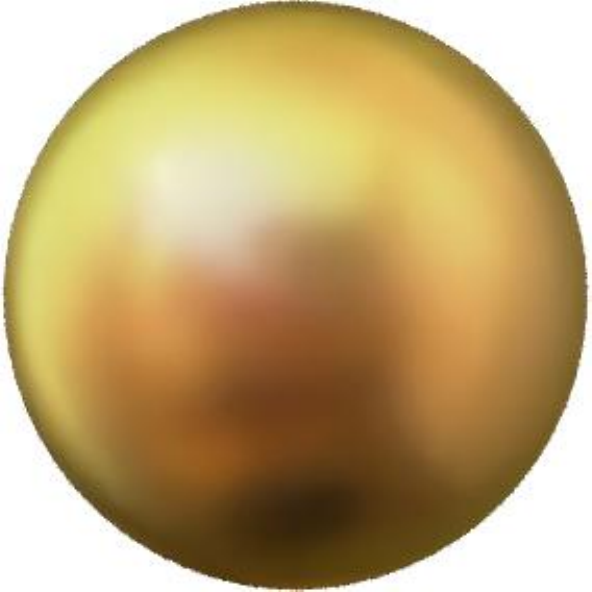}} & 
        
        \noindent\parbox[c]{0.100\textwidth}{\includegraphics[height=0.100\textwidth]{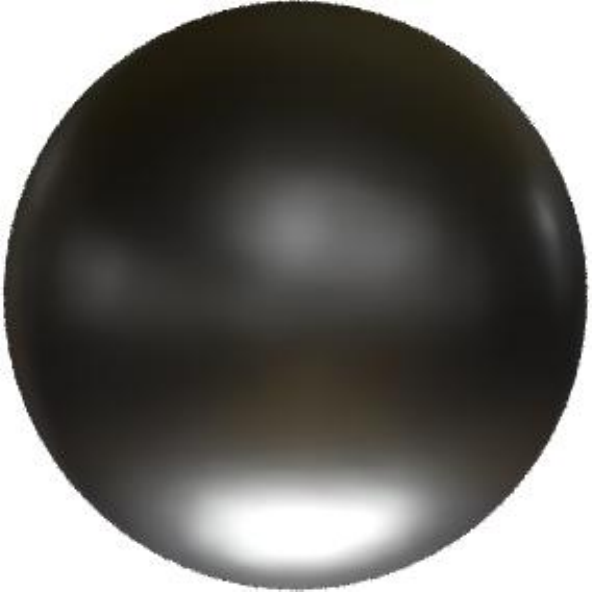}} & 
        \noindent\parbox[c]{0.100\textwidth}{\includegraphics[height=0.100\textwidth]{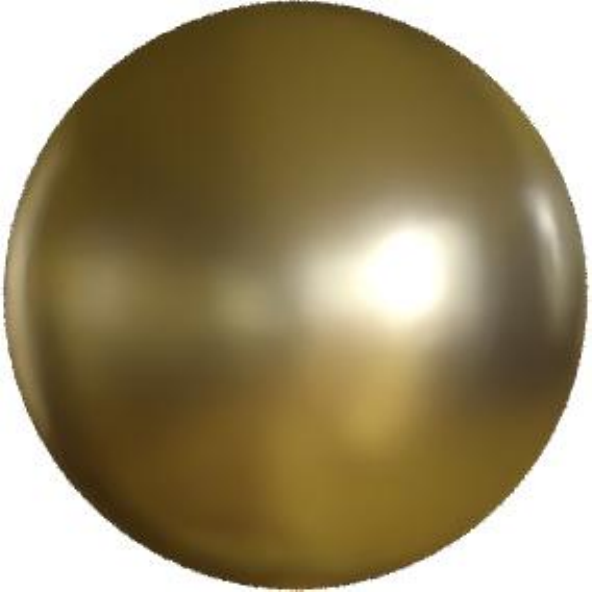}} &
        \noindent\parbox[c]{0.100\textwidth}{\includegraphics[height=0.100\textwidth]{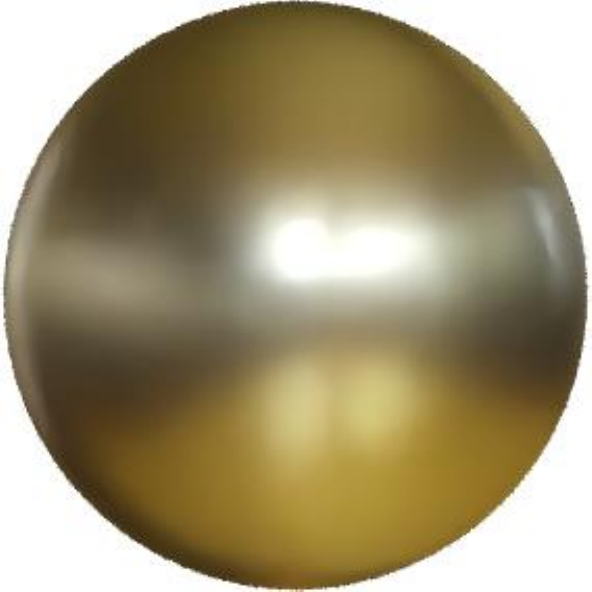}} & 
        \noindent\parbox[c]{0.100\textwidth}{\includegraphics[height=0.100\textwidth]{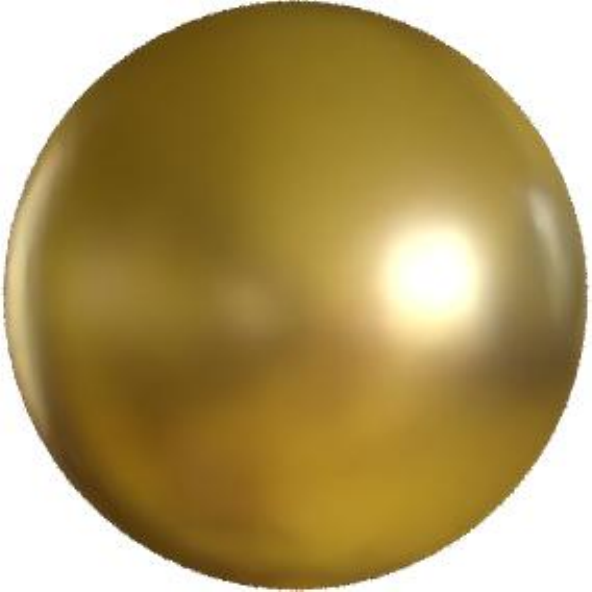}} & 
        \\

        \noindent\parbox[c]{0.205\textwidth}{\includegraphics[height=0.100\textwidth]{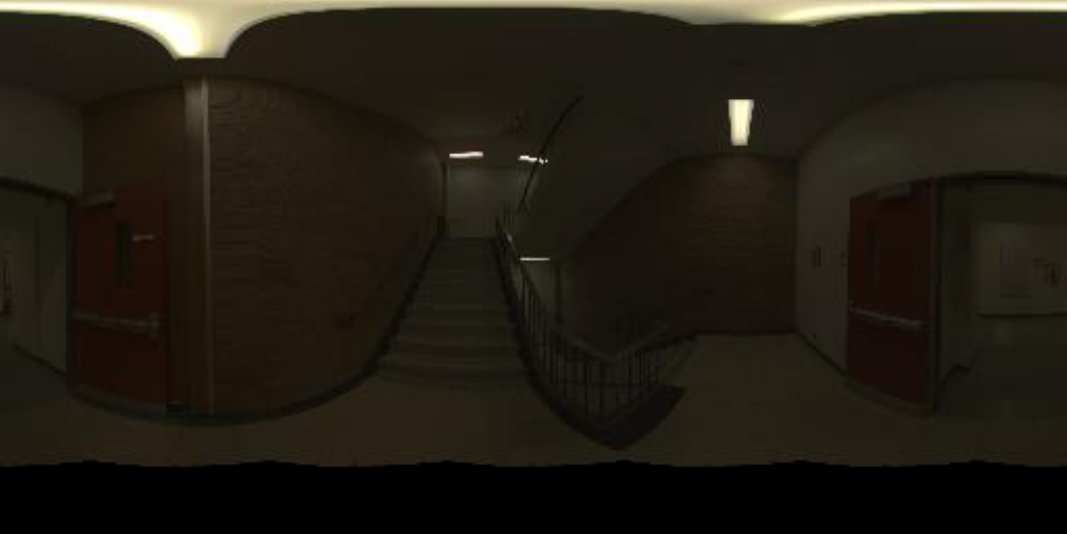}} & 
        \noindent\parbox[c]{0.14\textwidth}{\includegraphics[height=0.100\textwidth]{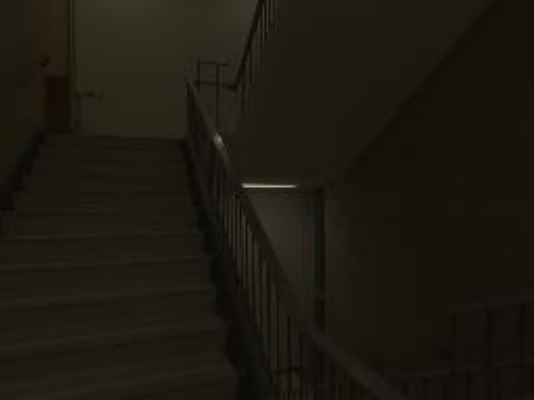}} &  
        
        \noindent\parbox[c]{0.100\textwidth}{\includegraphics[height=0.100\textwidth]{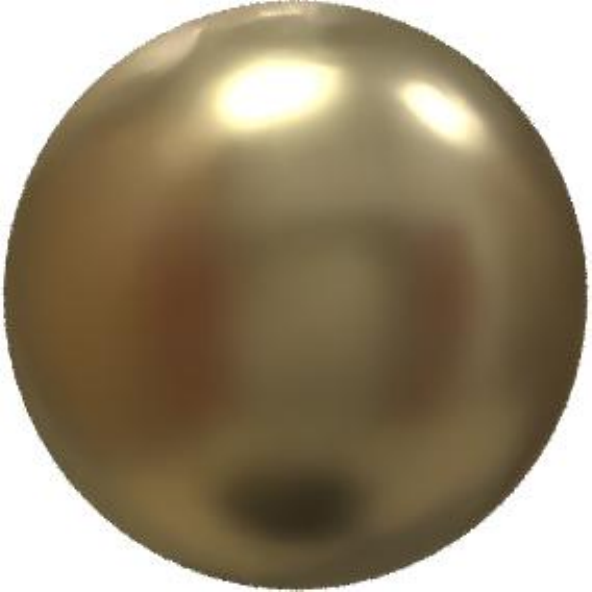}} & 
        \noindent\parbox[c]{0.100\textwidth}{\includegraphics[height=0.100\textwidth]{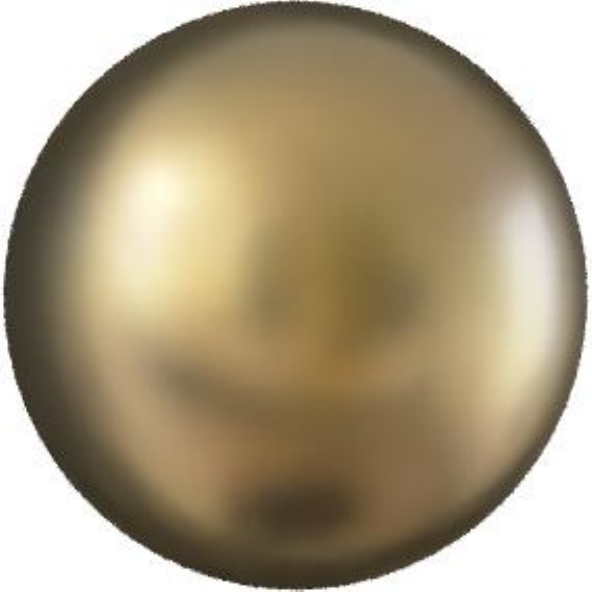}} & 
        
        \noindent\parbox[c]{0.100\textwidth}{\includegraphics[height=0.100\textwidth]{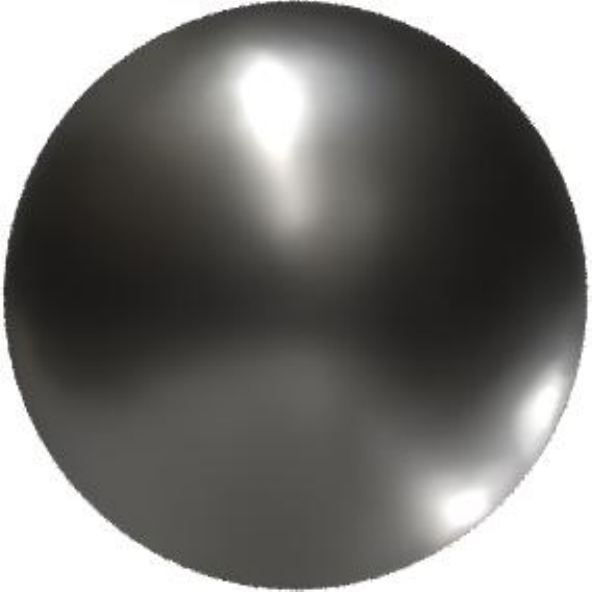}} & 
        \noindent\parbox[c]{0.100\textwidth}{\includegraphics[height=0.100\textwidth]{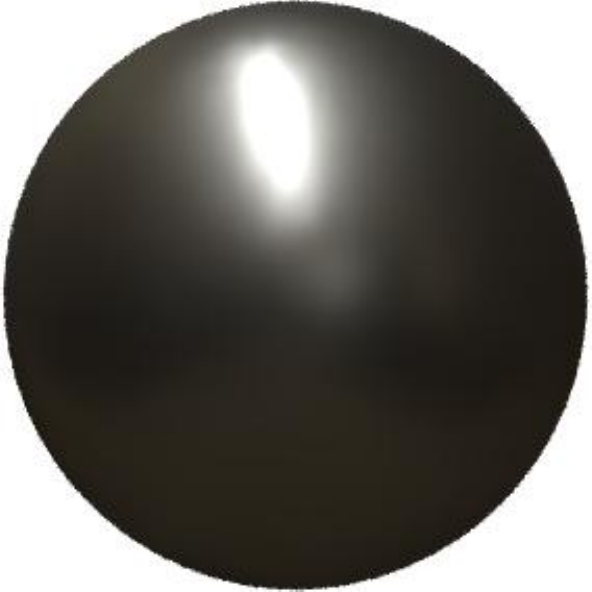}} &
        \noindent\parbox[c]{0.100\textwidth}{\includegraphics[height=0.100\textwidth]{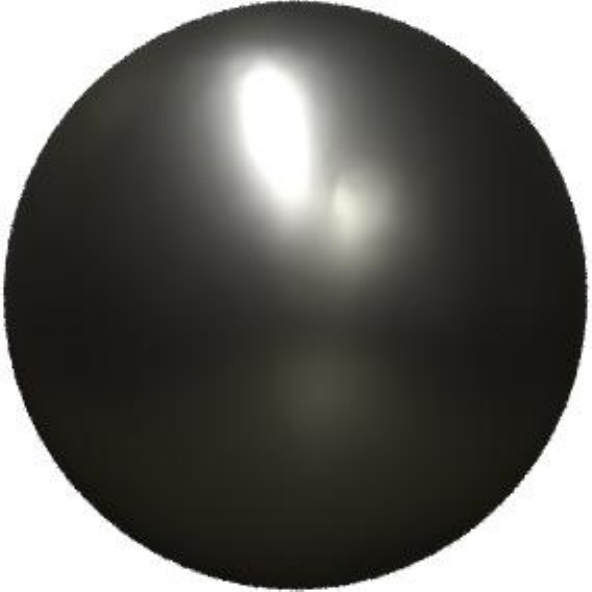}} & 
        \noindent\parbox[c]{0.100\textwidth}{\includegraphics[height=0.100\textwidth]{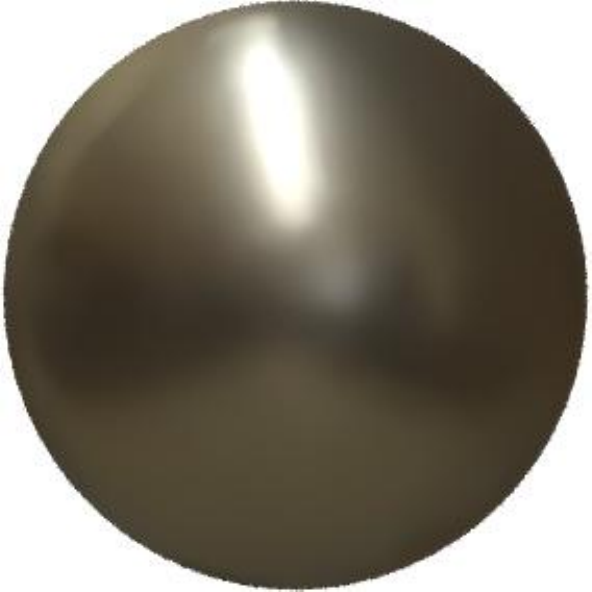}} & 
        \\

        \noindent\parbox[c]{0.205\textwidth}{\includegraphics[height=0.100\textwidth]{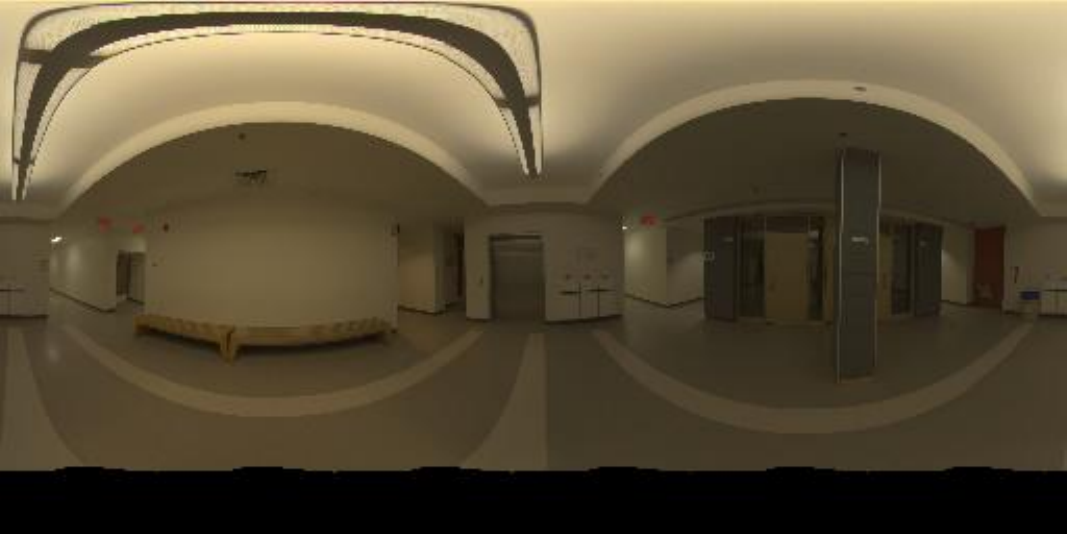}} & 
        \noindent\parbox[c]{0.14\textwidth}{\includegraphics[height=0.100\textwidth]{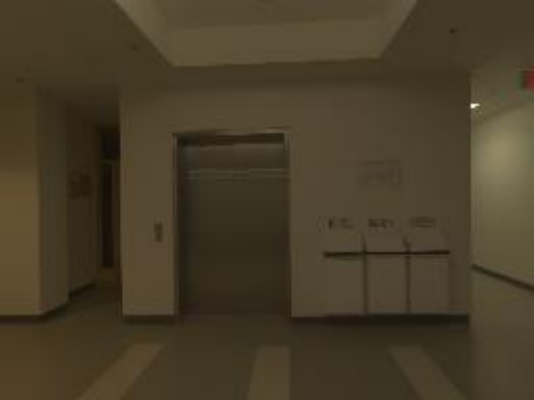}} &  
        
        \noindent\parbox[c]{0.100\textwidth}{\includegraphics[height=0.100\textwidth]{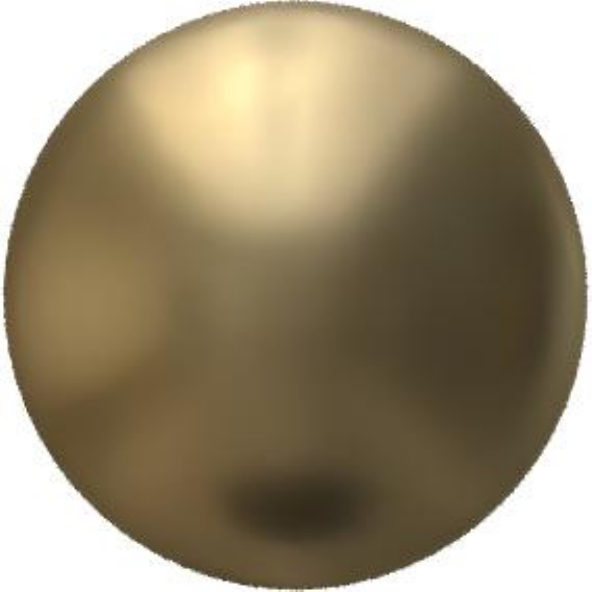}} & 
        \noindent\parbox[c]{0.100\textwidth}{\includegraphics[height=0.100\textwidth]{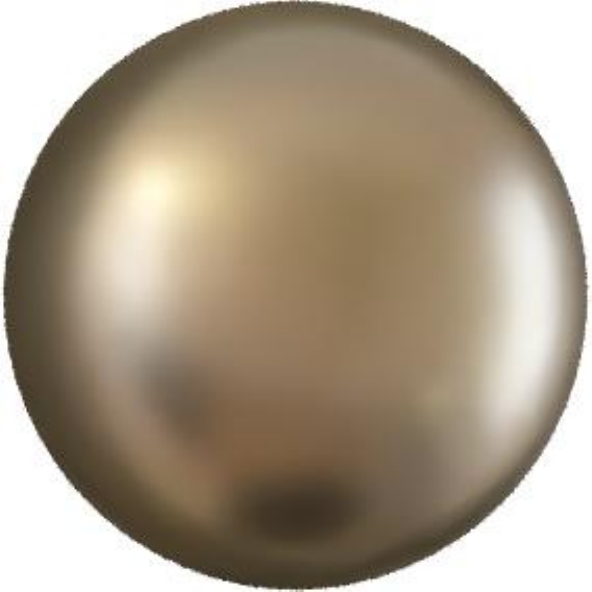}} & 
        
        \noindent\parbox[c]{0.100\textwidth}{\includegraphics[height=0.100\textwidth]{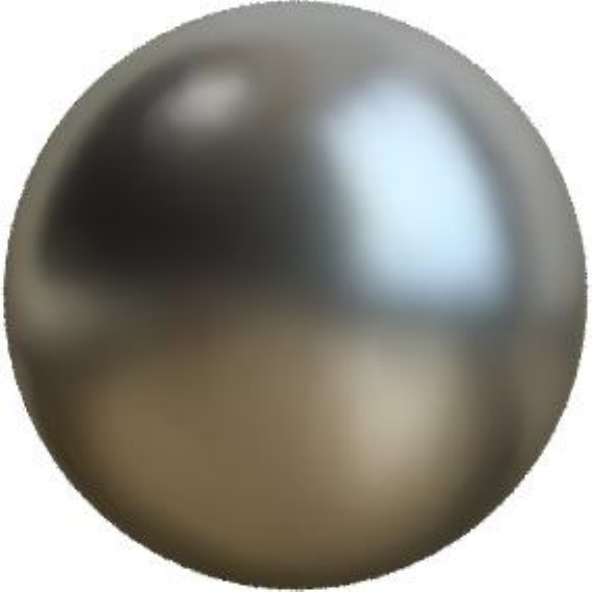}} & 
        \noindent\parbox[c]{0.100\textwidth}{\includegraphics[height=0.100\textwidth]{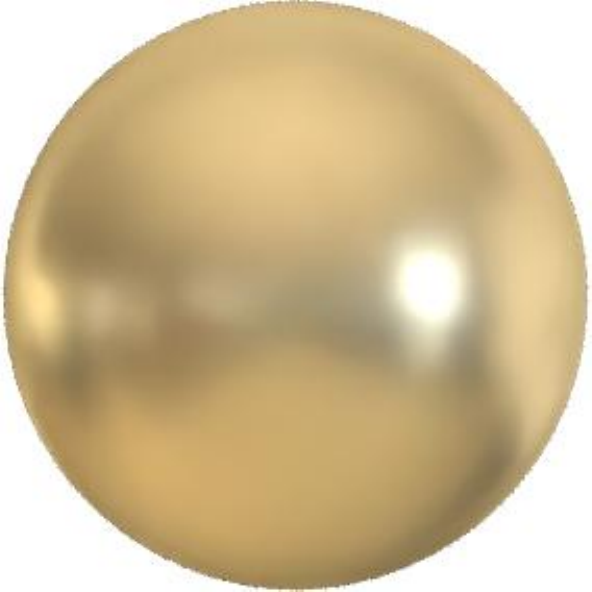}} &
        \noindent\parbox[c]{0.100\textwidth}{\includegraphics[height=0.100\textwidth]{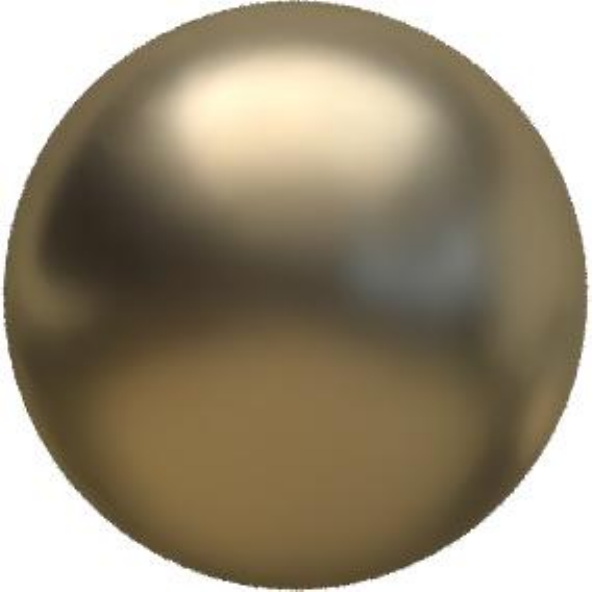}} & 
        \noindent\parbox[c]{0.100\textwidth}{\includegraphics[height=0.100\textwidth]{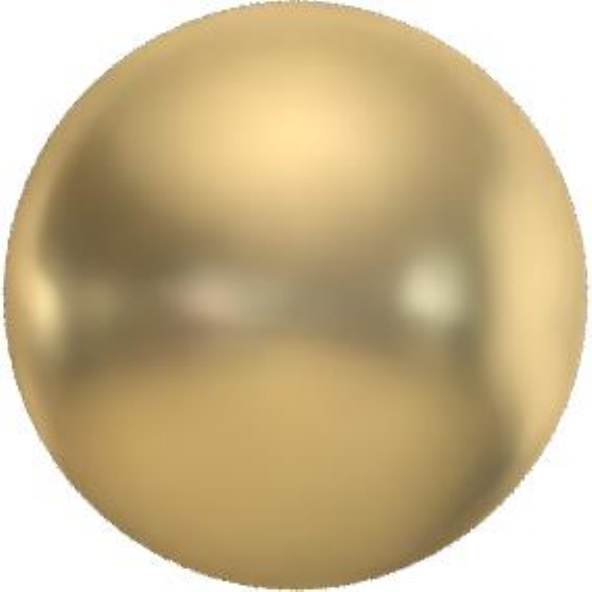}} & 
        \\

        \noindent\parbox[c]{0.205\textwidth}{\includegraphics[height=0.100\textwidth]{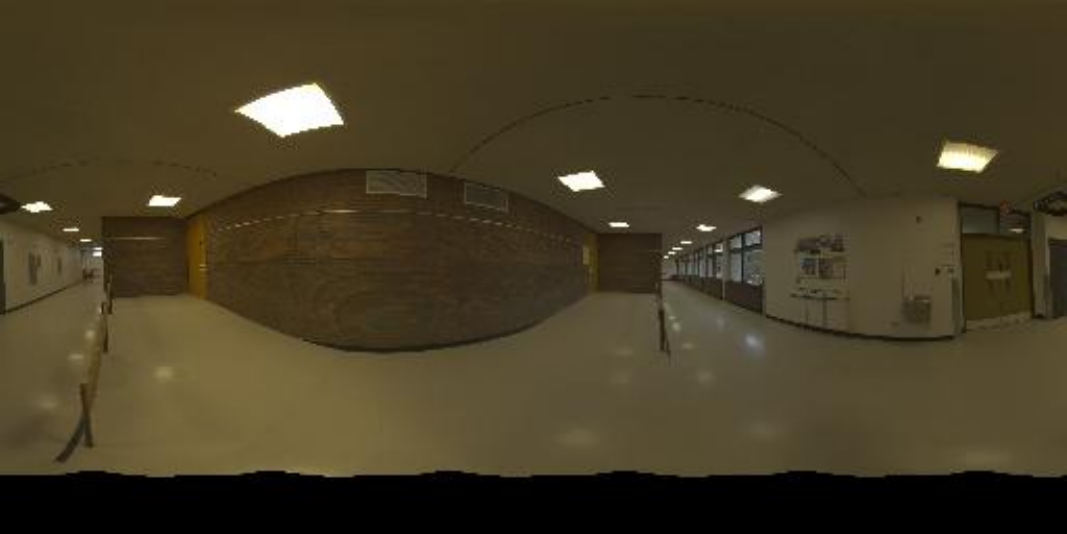}} & 
        \noindent\parbox[c]{0.14\textwidth}{\includegraphics[height=0.100\textwidth]{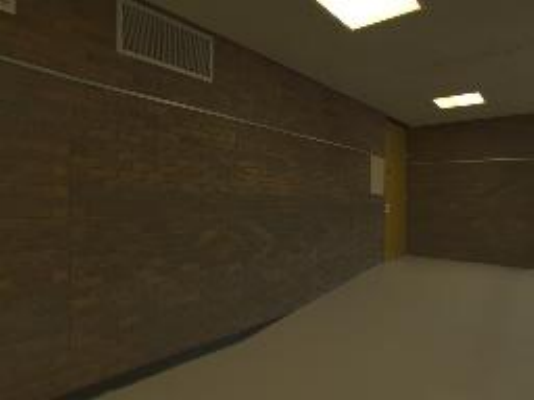}} &  
        
        \noindent\parbox[c]{0.100\textwidth}{\includegraphics[height=0.100\textwidth]{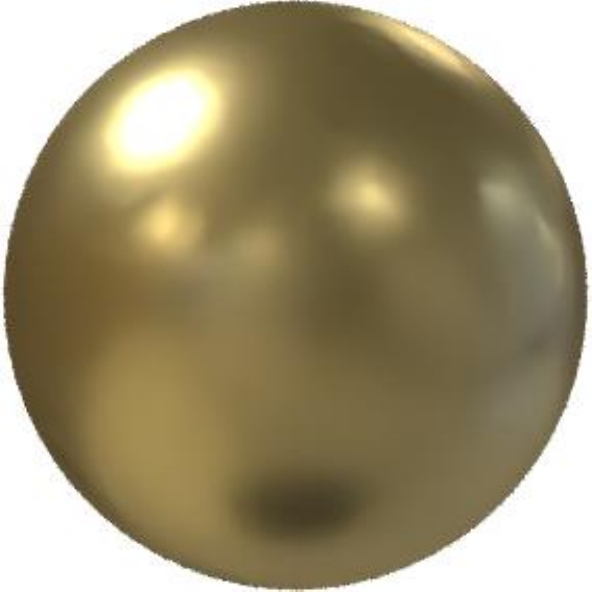}} & 
        \noindent\parbox[c]{0.100\textwidth}{\includegraphics[height=0.100\textwidth]{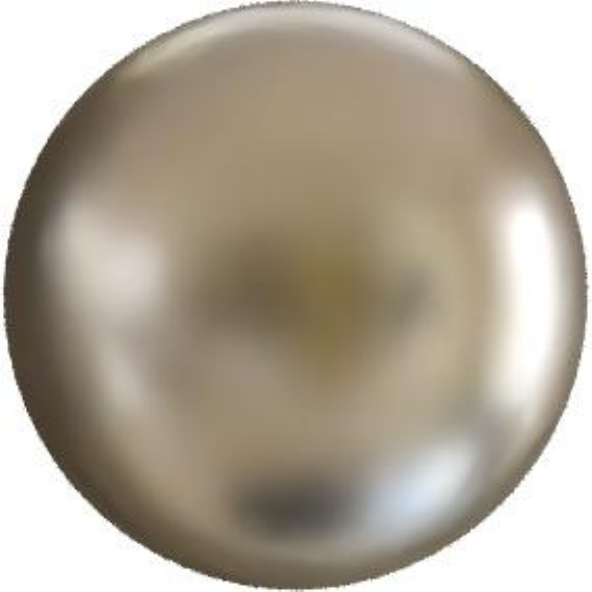}} & 
        
        \noindent\parbox[c]{0.100\textwidth}{\includegraphics[height=0.100\textwidth]{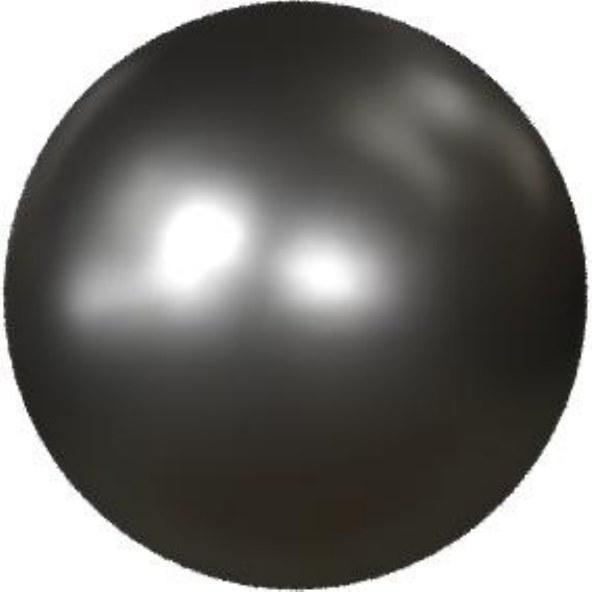}} & 
        \noindent\parbox[c]{0.100\textwidth}{\includegraphics[height=0.100\textwidth]{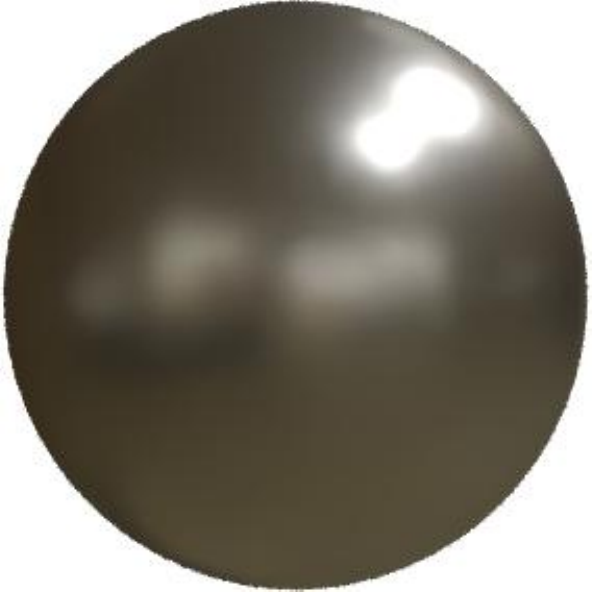}} &
        \noindent\parbox[c]{0.100\textwidth}{\includegraphics[height=0.100\textwidth]{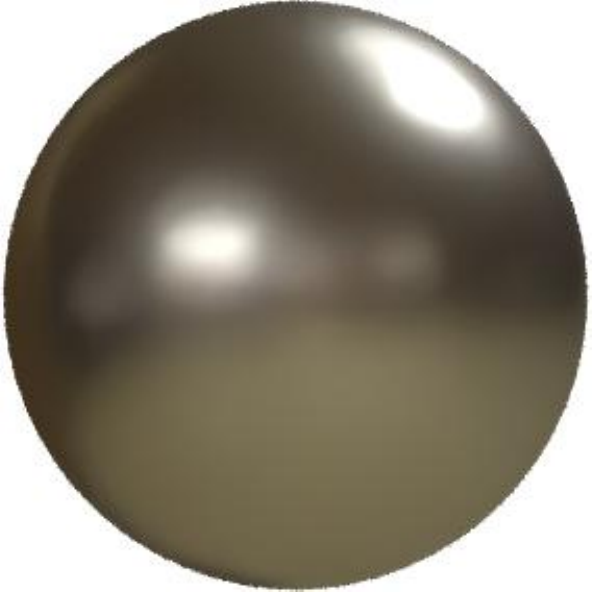}} & 
        \noindent\parbox[c]{0.100\textwidth}{\includegraphics[height=0.100\textwidth]{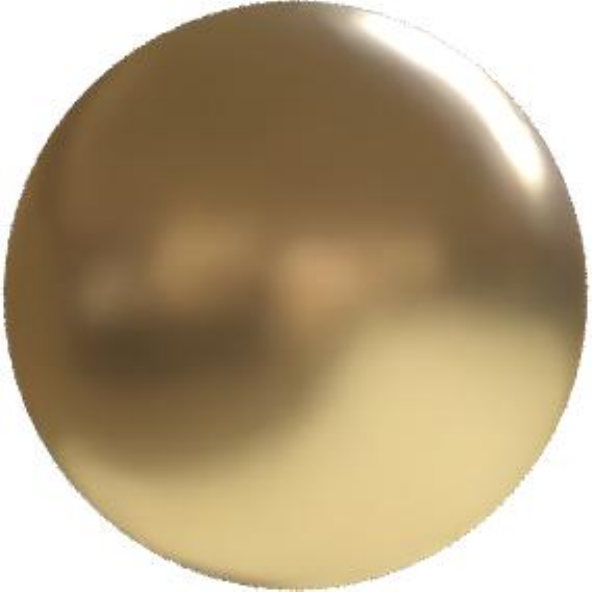}} & 
        \\

        \noindent\parbox[c]{0.205\textwidth}{\includegraphics[height=0.100\textwidth]{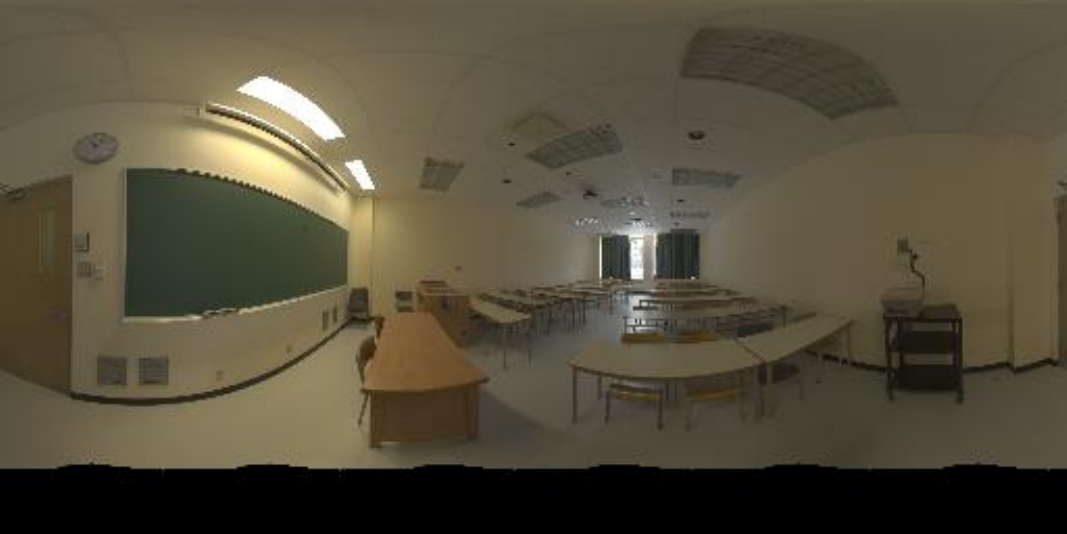}} & 
        \noindent\parbox[c]{0.14\textwidth}{\includegraphics[height=0.100\textwidth]{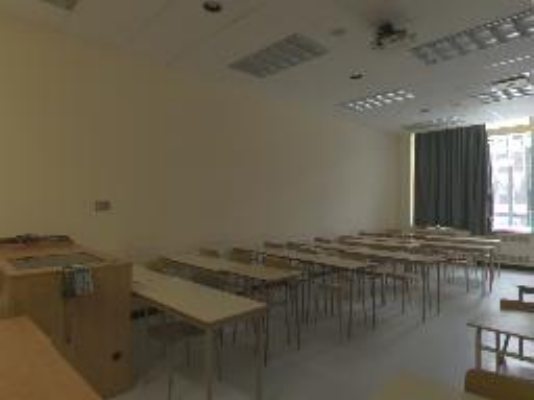}} &  
        
        \noindent\parbox[c]{0.100\textwidth}{\includegraphics[height=0.100\textwidth]{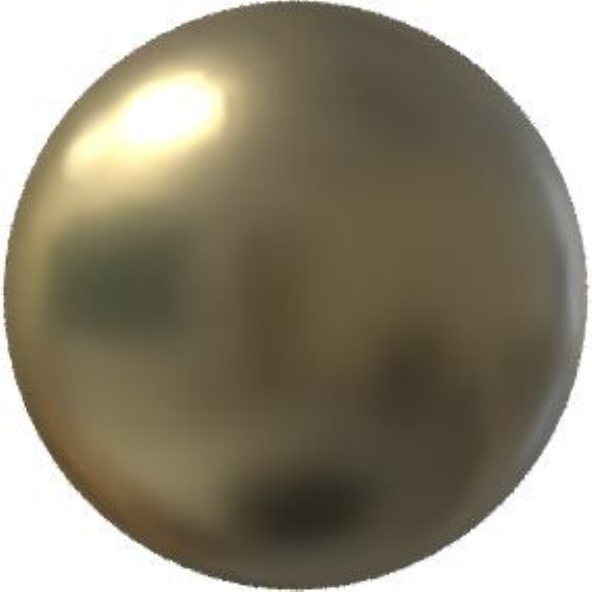}} & 
        \noindent\parbox[c]{0.100\textwidth}{\includegraphics[height=0.100\textwidth]{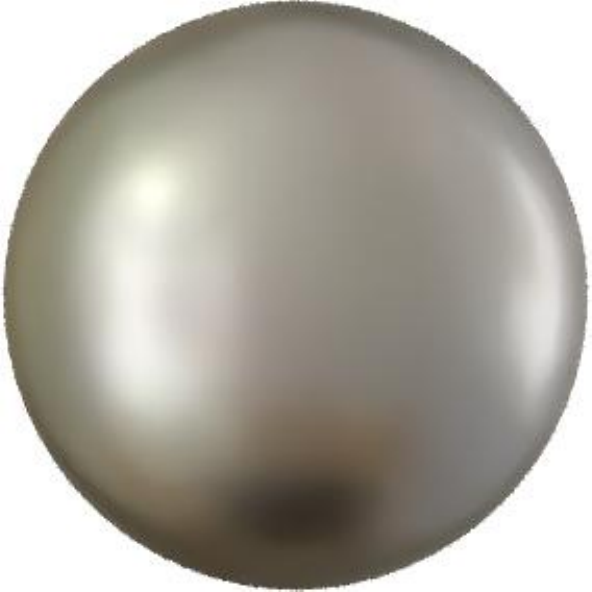}} & 
        
        \noindent\parbox[c]{0.100\textwidth}{\includegraphics[height=0.100\textwidth]{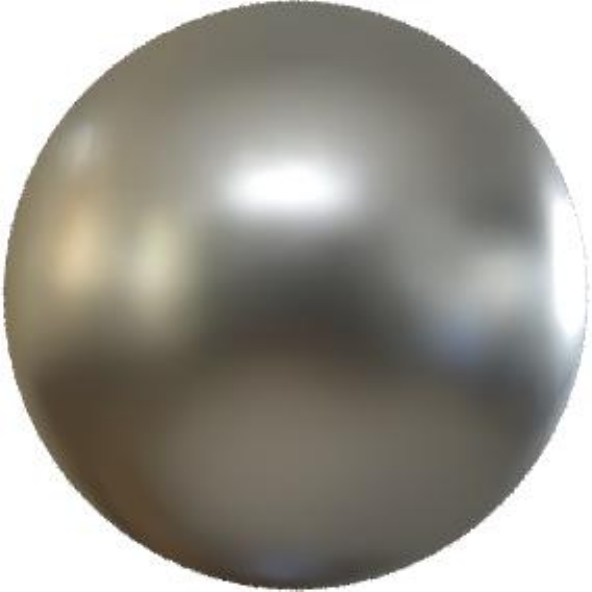}} & 
        \noindent\parbox[c]{0.100\textwidth}{\includegraphics[height=0.100\textwidth]{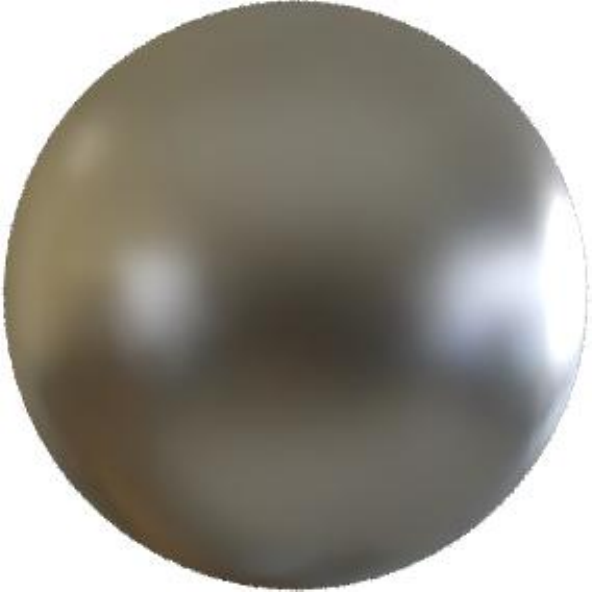}} &
        \noindent\parbox[c]{0.100\textwidth}{\includegraphics[height=0.100\textwidth]{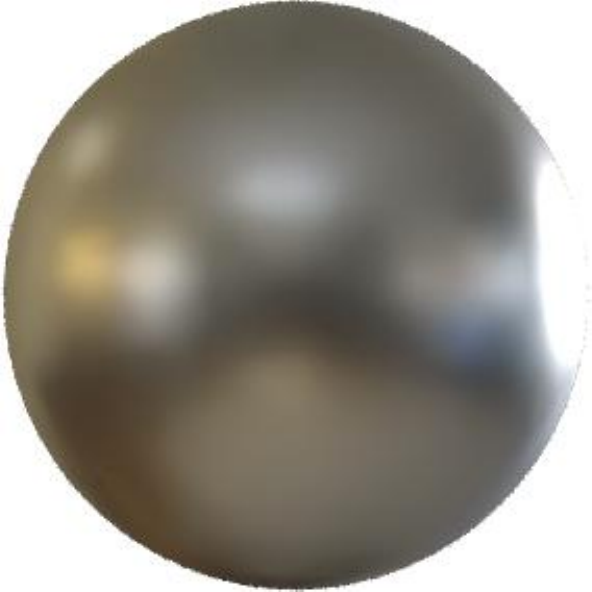}} & 
        \noindent\parbox[c]{0.100\textwidth}{\includegraphics[height=0.100\textwidth]{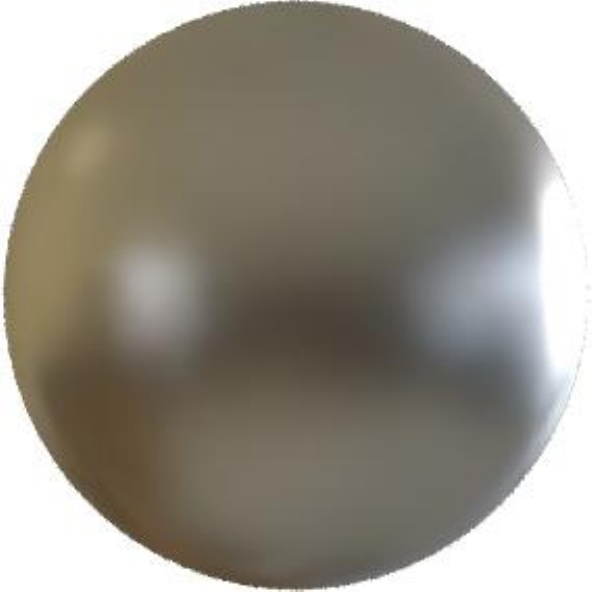}} & 
        \\

        \noindent\parbox[c]{0.205\textwidth}{\includegraphics[height=0.100\textwidth]{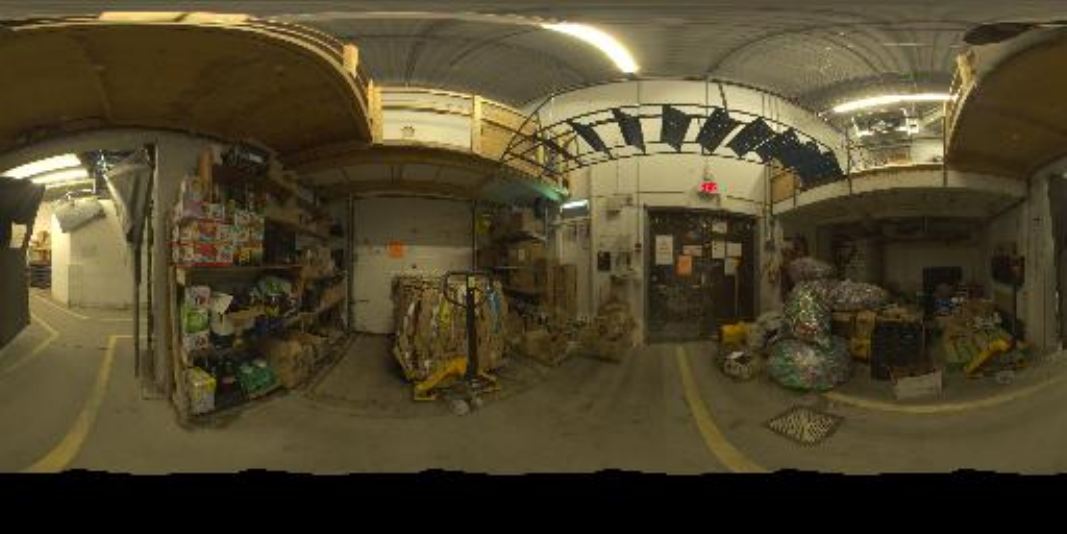}} & 
        \noindent\parbox[c]{0.14\textwidth}{\includegraphics[height=0.100\textwidth]{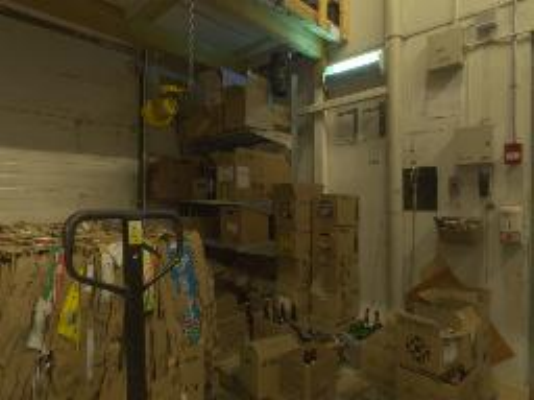}} &  
        
        \noindent\parbox[c]{0.100\textwidth}{\includegraphics[height=0.100\textwidth]{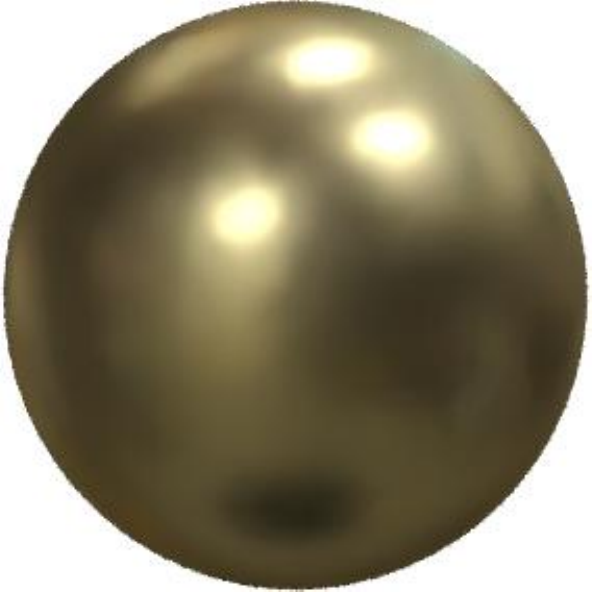}} & 
        \noindent\parbox[c]{0.100\textwidth}{\includegraphics[height=0.100\textwidth]{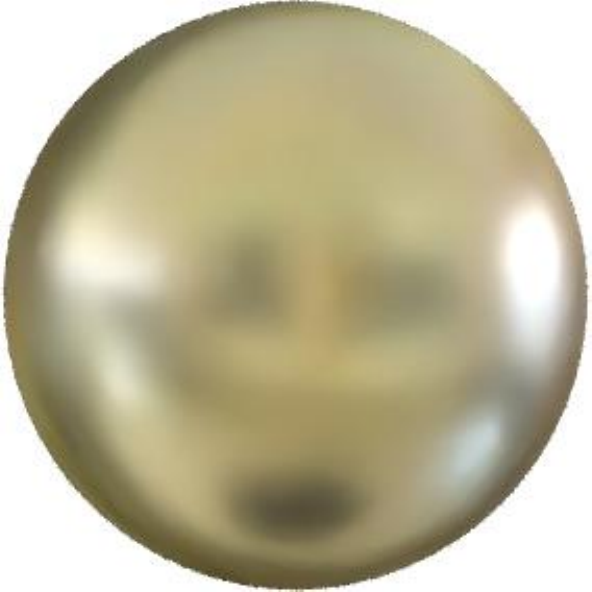}} & 
        
        \noindent\parbox[c]{0.100\textwidth}{\includegraphics[height=0.100\textwidth]{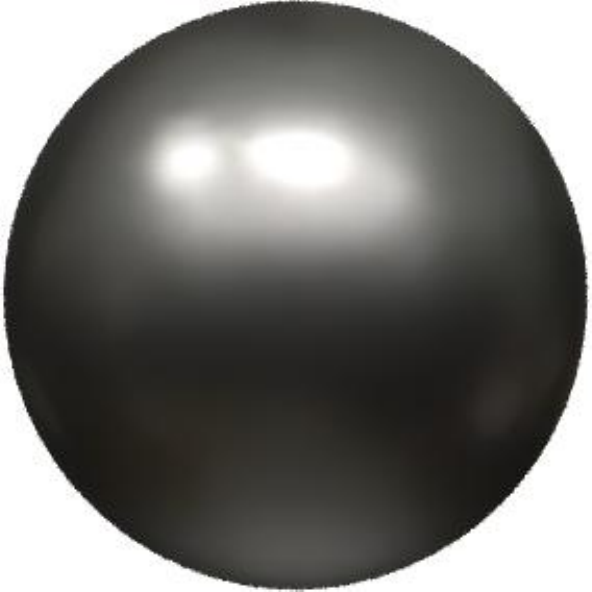}} & 
        \noindent\parbox[c]{0.100\textwidth}{\includegraphics[height=0.100\textwidth]{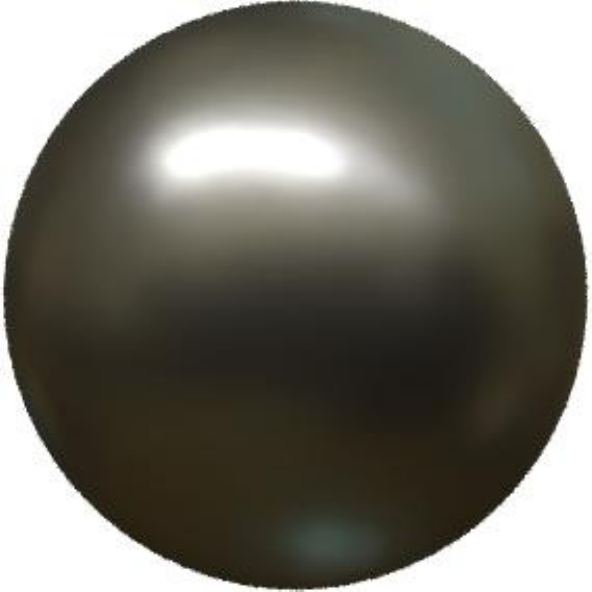}} &
        \noindent\parbox[c]{0.100\textwidth}{\includegraphics[height=0.100\textwidth]{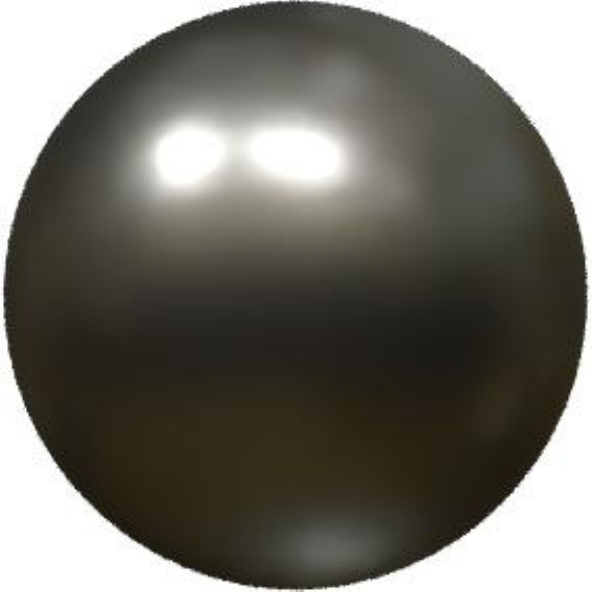}} & 
        \noindent\parbox[c]{0.100\textwidth}{\includegraphics[height=0.100\textwidth]{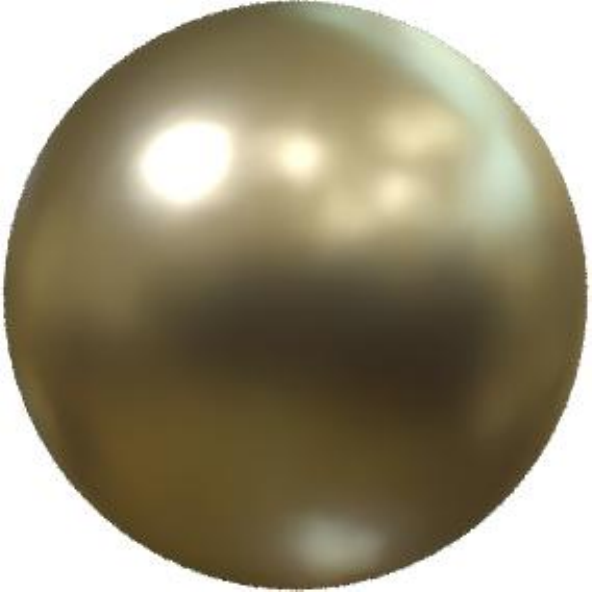}} & 
        \\

        \noindent\parbox[c]{0.205\textwidth}{\includegraphics[height=0.100\textwidth]{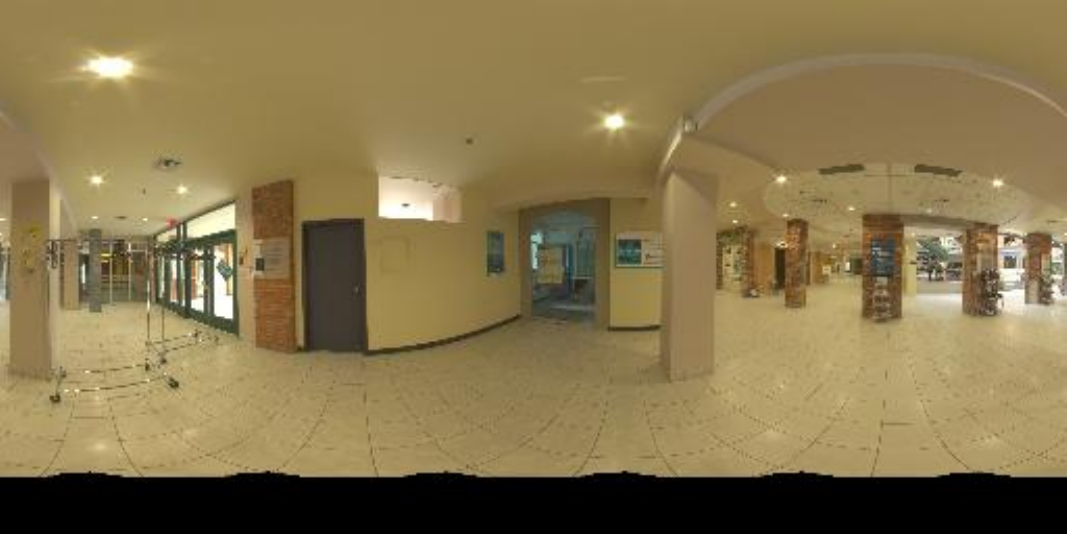}} & 
        \noindent\parbox[c]{0.14\textwidth}{\includegraphics[height=0.100\textwidth]{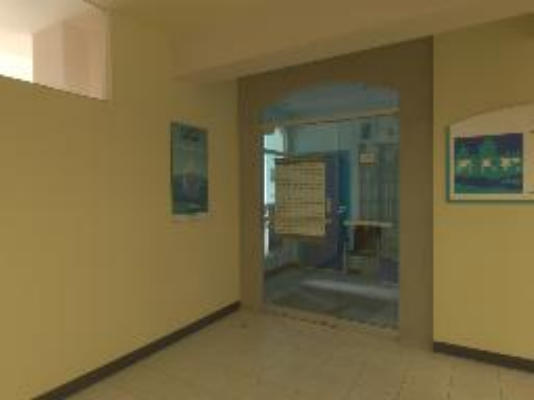}} &  
        
        \noindent\parbox[c]{0.100\textwidth}{\includegraphics[height=0.100\textwidth]{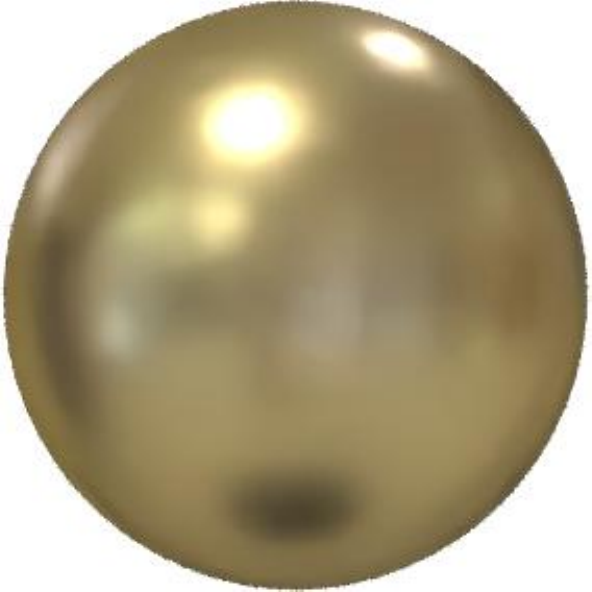}} & 
        \noindent\parbox[c]{0.100\textwidth}{\includegraphics[height=0.100\textwidth]{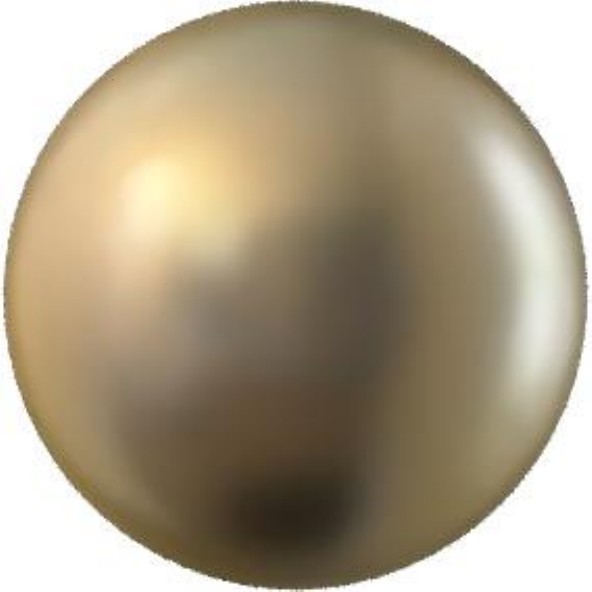}} & 
        
        \noindent\parbox[c]{0.100\textwidth}{\includegraphics[height=0.100\textwidth]{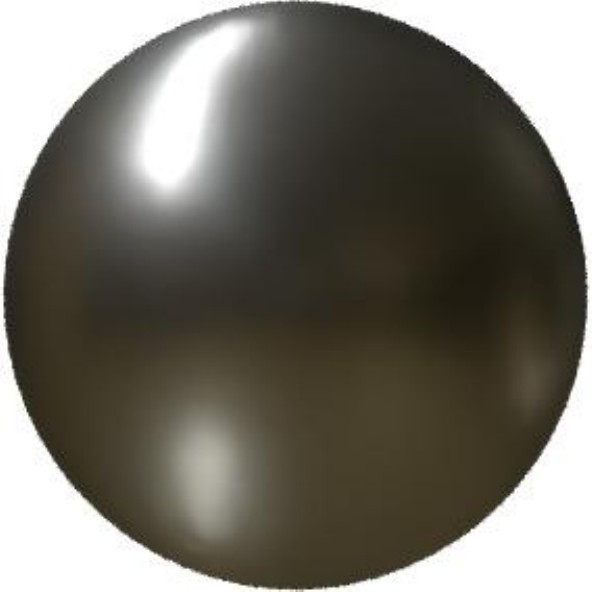}} & 
        \noindent\parbox[c]{0.100\textwidth}{\includegraphics[height=0.100\textwidth]{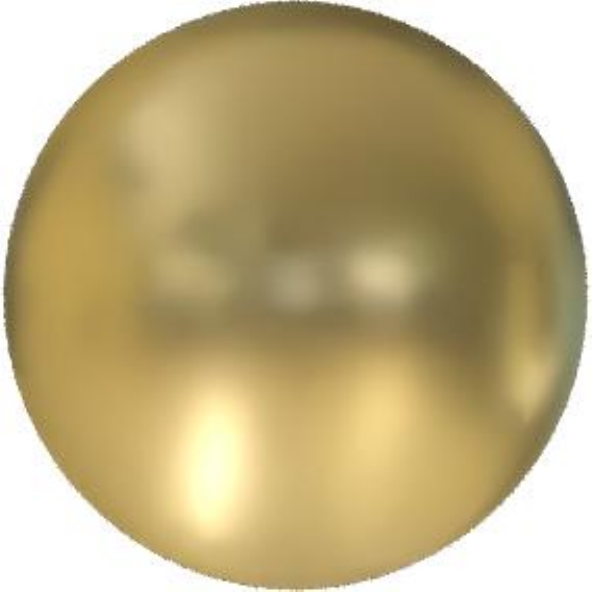}} &
        \noindent\parbox[c]{0.100\textwidth}{\includegraphics[height=0.100\textwidth]{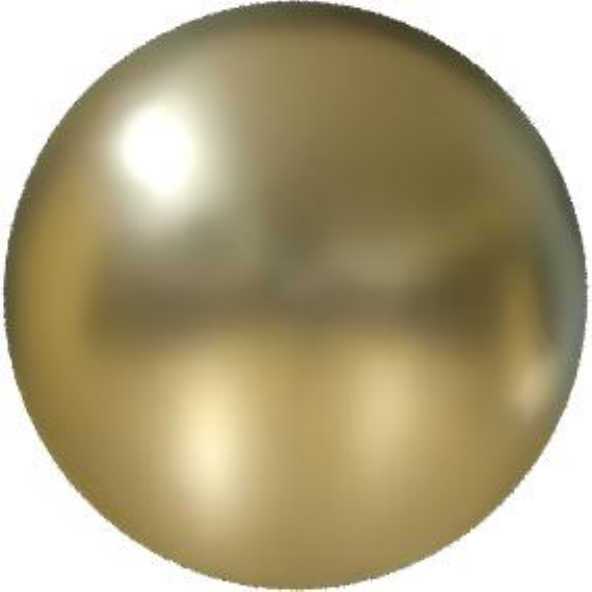}} & 
        \noindent\parbox[c]{0.100\textwidth}{\includegraphics[height=0.100\textwidth]{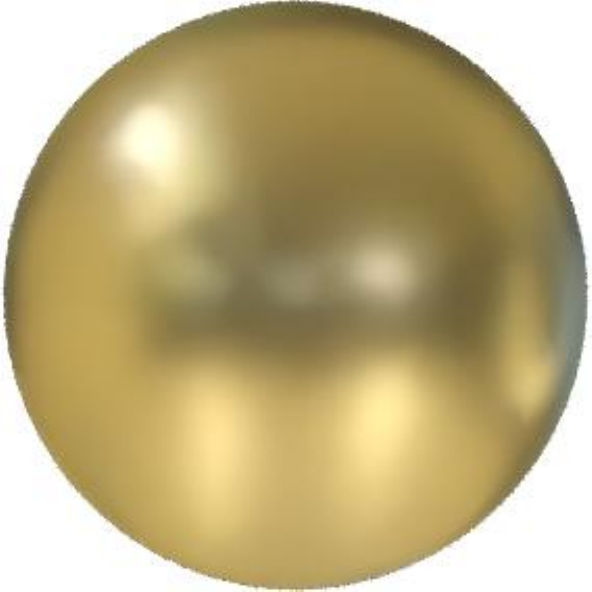}} & 
        \\

        \noindent\parbox[c]{0.205\textwidth}{\includegraphics[height=0.100\textwidth]{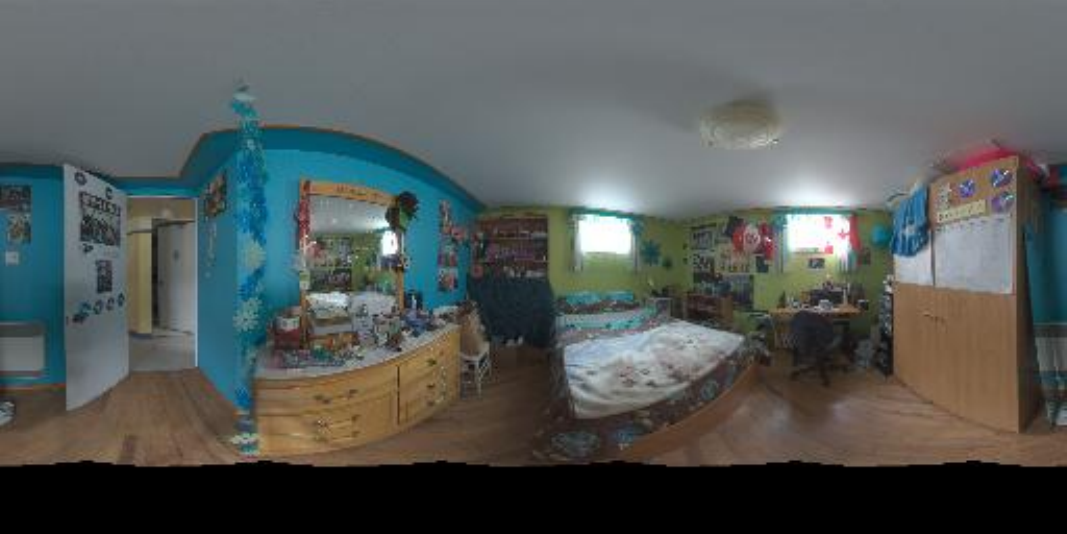}} & 
        \noindent\parbox[c]{0.14\textwidth}{\includegraphics[height=0.100\textwidth]{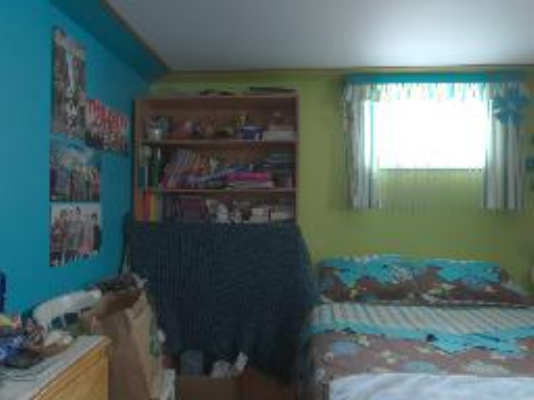}} &  
        
        \noindent\parbox[c]{0.100\textwidth}{\includegraphics[height=0.100\textwidth]{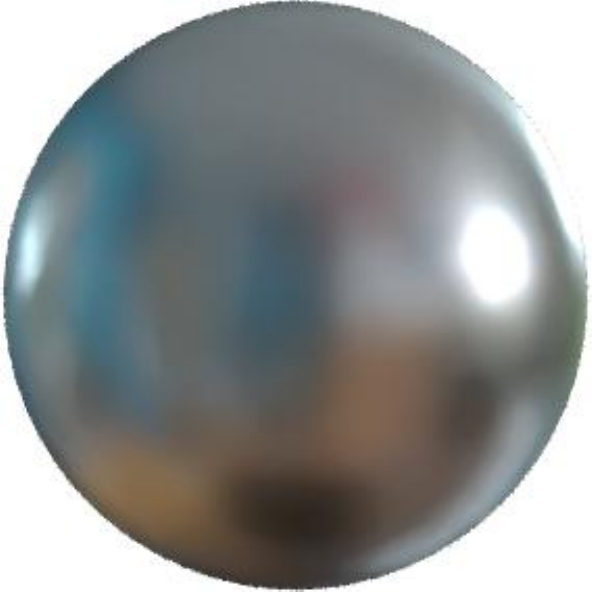}} & 
        \noindent\parbox[c]{0.100\textwidth}{\includegraphics[height=0.100\textwidth]{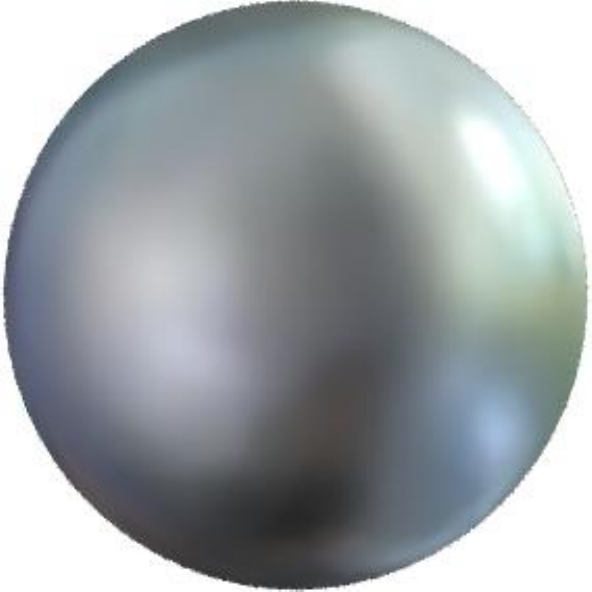}} & 
        
        \noindent\parbox[c]{0.100\textwidth}{\includegraphics[height=0.100\textwidth]{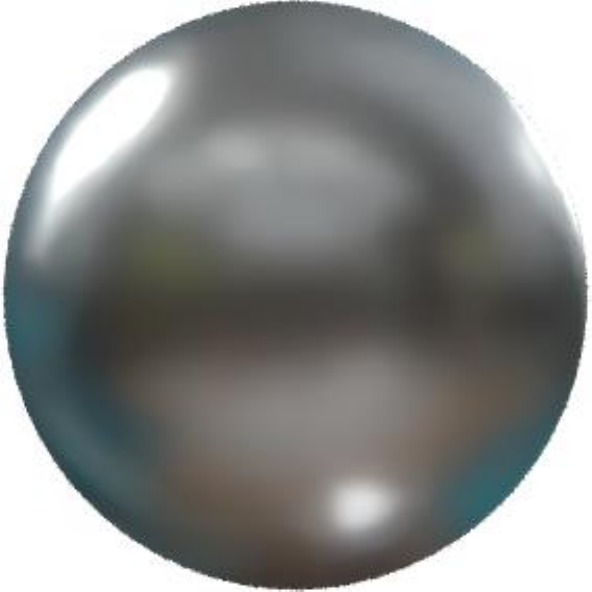}} & 
        \noindent\parbox[c]{0.100\textwidth}{\includegraphics[height=0.100\textwidth]{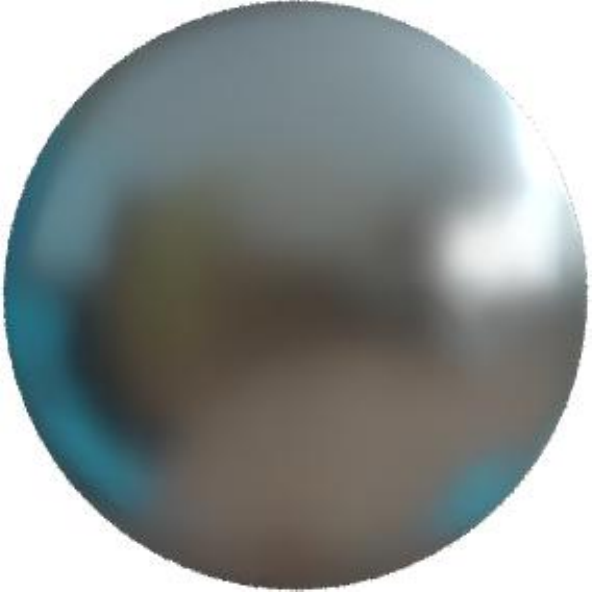}} &
        \noindent\parbox[c]{0.100\textwidth}{\includegraphics[height=0.100\textwidth]{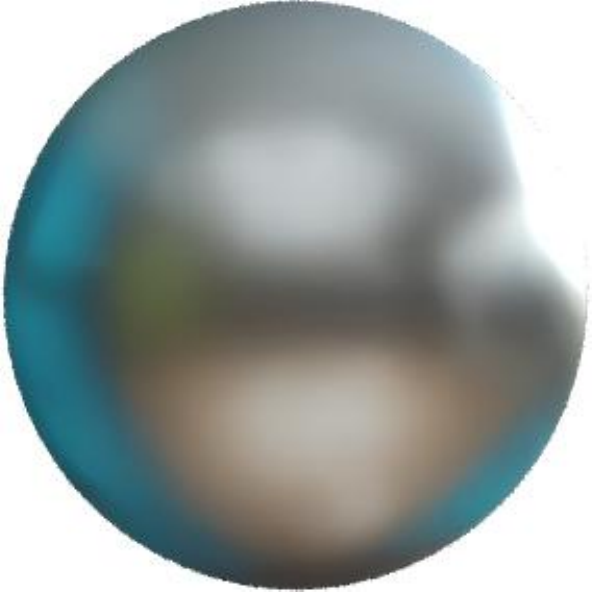}} & 
        \noindent\parbox[c]{0.100\textwidth}{\includegraphics[height=0.100\textwidth]{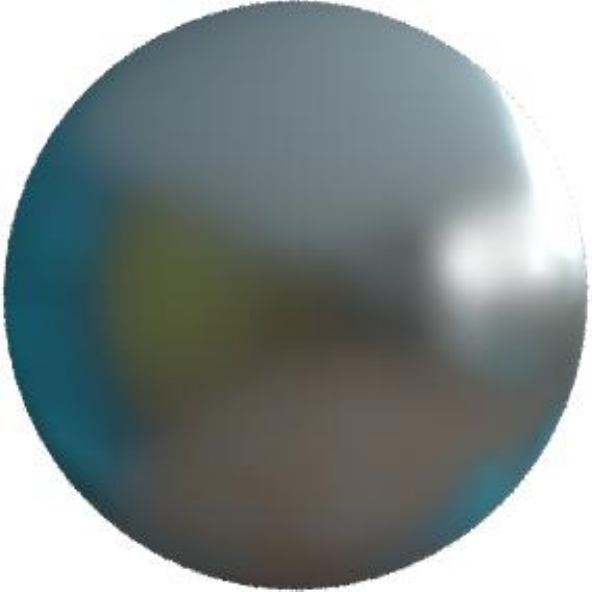}} & 
        \\

        \noindent\parbox[c]{0.205\textwidth}{\includegraphics[height=0.100\textwidth]{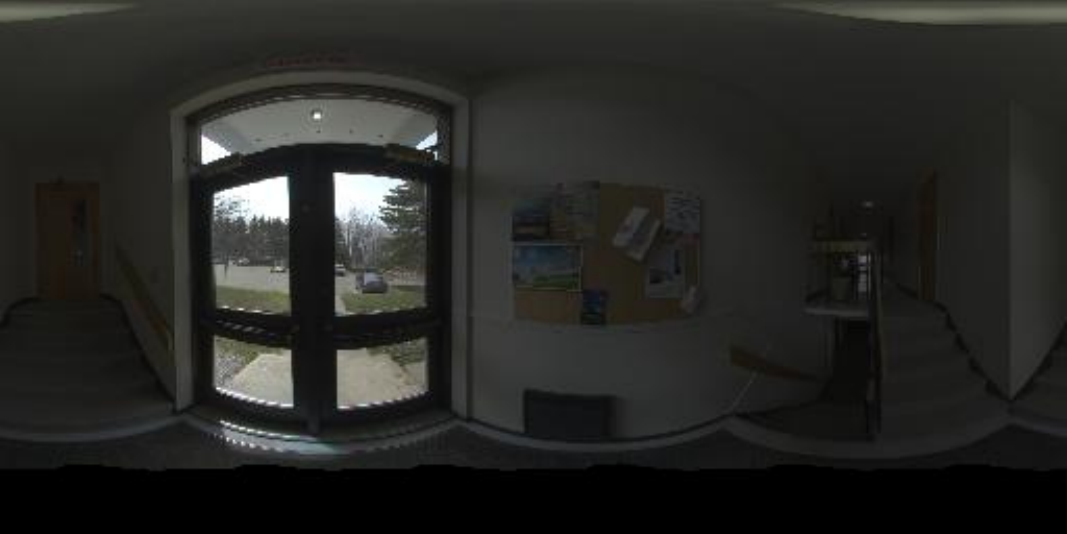}} & 
        \noindent\parbox[c]{0.14\textwidth}{\includegraphics[height=0.100\textwidth]{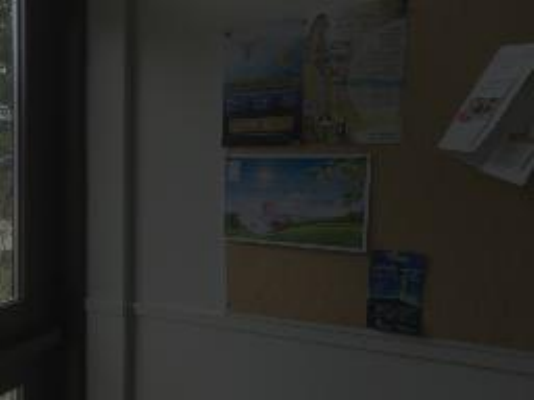}} &  
        
        \noindent\parbox[c]{0.100\textwidth}{\includegraphics[height=0.100\textwidth]{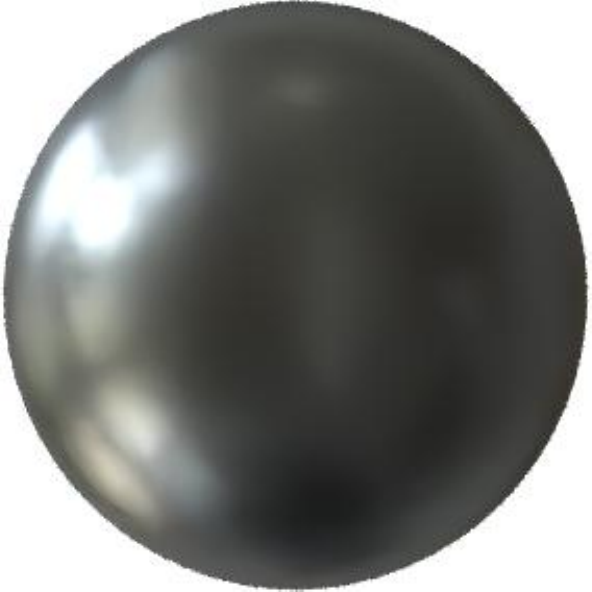}} & 
        \noindent\parbox[c]{0.100\textwidth}{\includegraphics[height=0.100\textwidth]{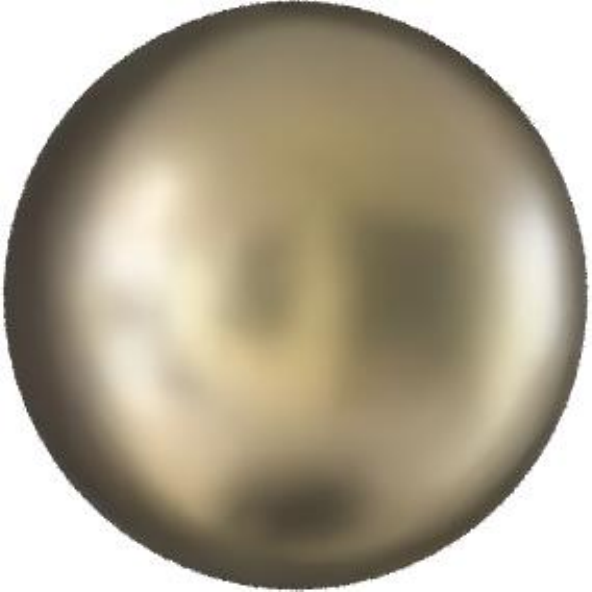}} & 
        
        \noindent\parbox[c]{0.100\textwidth}{\includegraphics[height=0.100\textwidth]{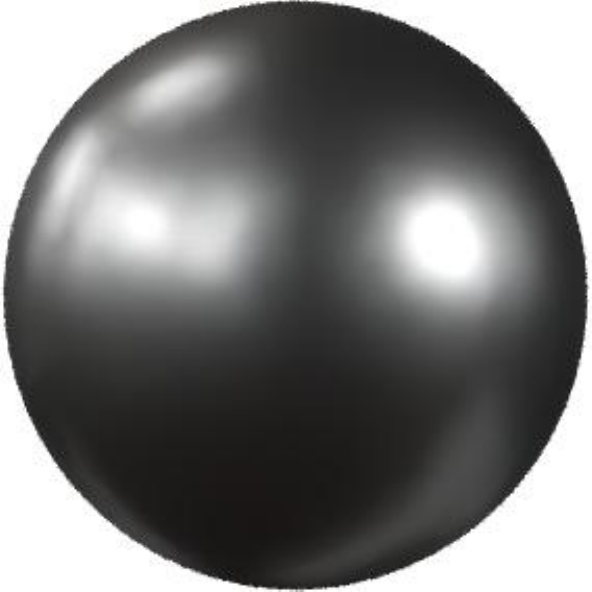}} & 
        \noindent\parbox[c]{0.100\textwidth}{\includegraphics[height=0.100\textwidth]{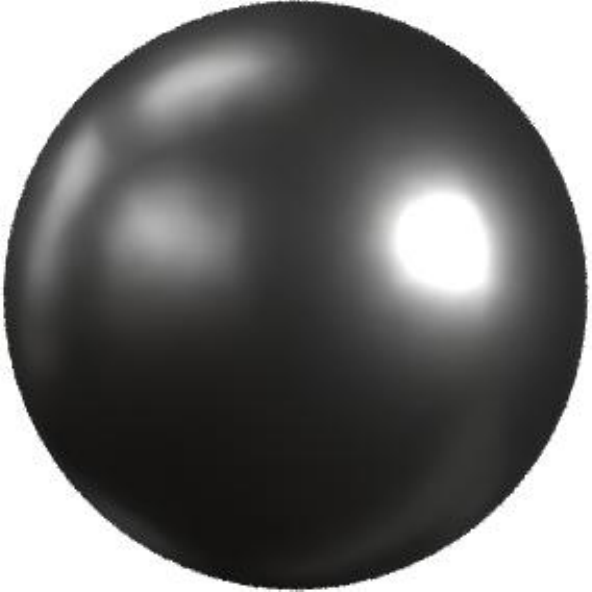}} &
        \noindent\parbox[c]{0.100\textwidth}{\includegraphics[height=0.100\textwidth]{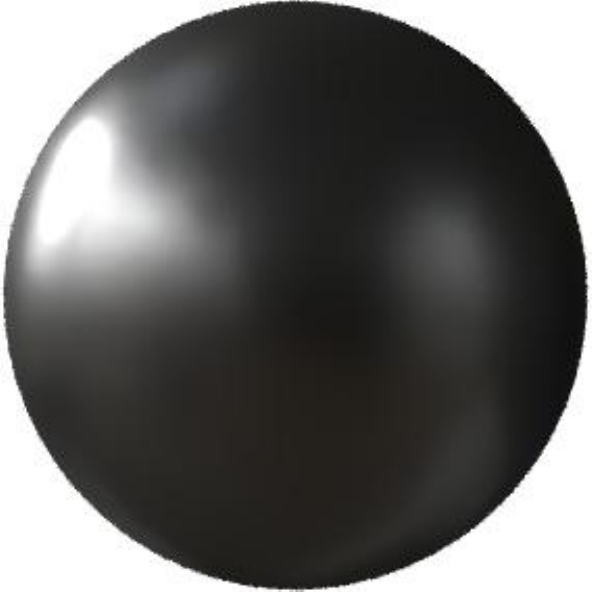}} & 
        \noindent\parbox[c]{0.100\textwidth}{\includegraphics[height=0.100\textwidth]{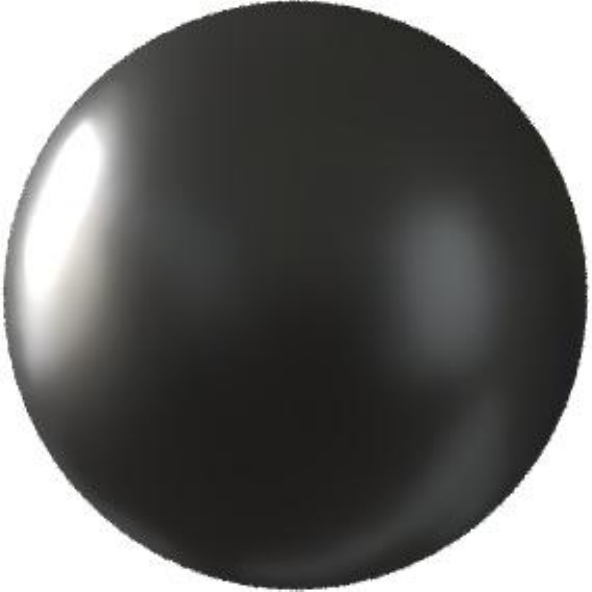}} & 
        \\
        
        \noindent\parbox[c]{0.205\textwidth}{\includegraphics[height=0.100\textwidth]{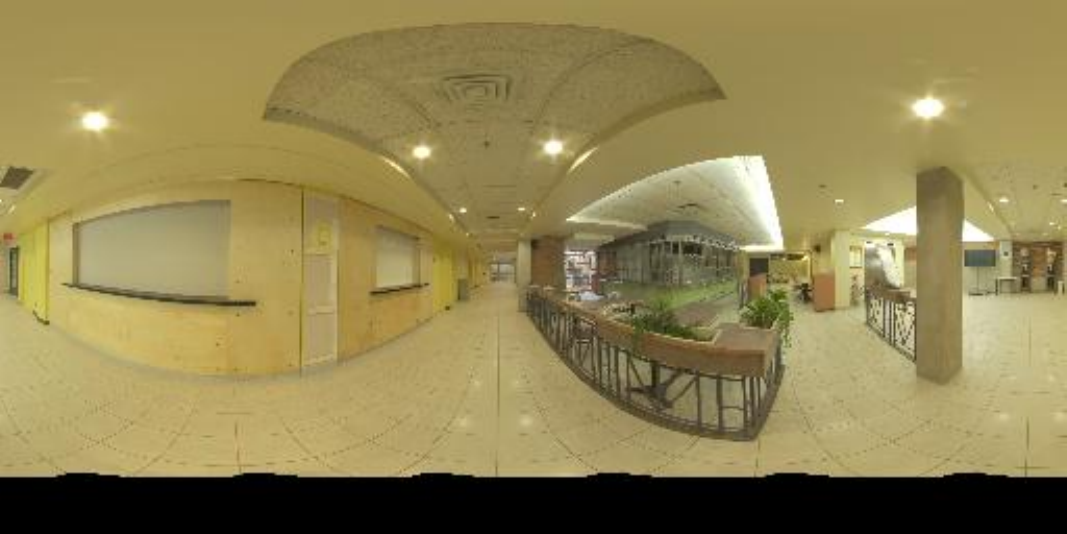}} & 
        \noindent\parbox[c]{0.14\textwidth}{\includegraphics[height=0.100\textwidth]{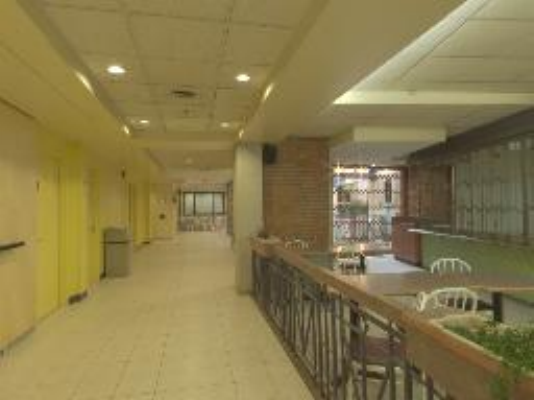}} &  
        
        \noindent\parbox[c]{0.100\textwidth}{\includegraphics[height=0.100\textwidth]{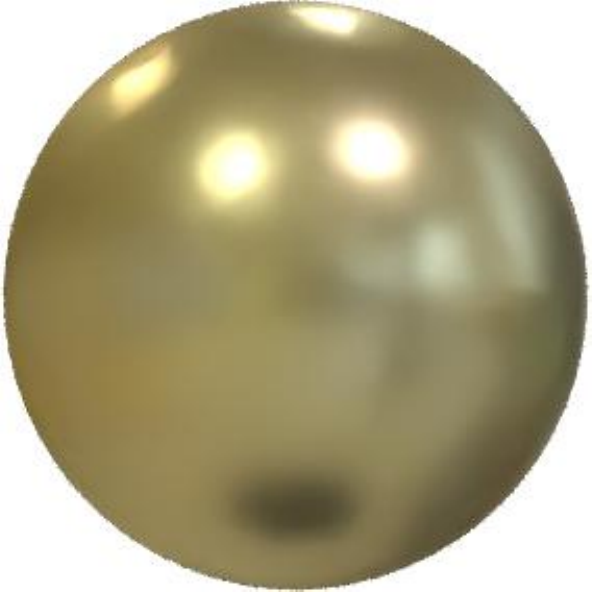}} & 
        \noindent\parbox[c]{0.100\textwidth}{\includegraphics[height=0.100\textwidth]{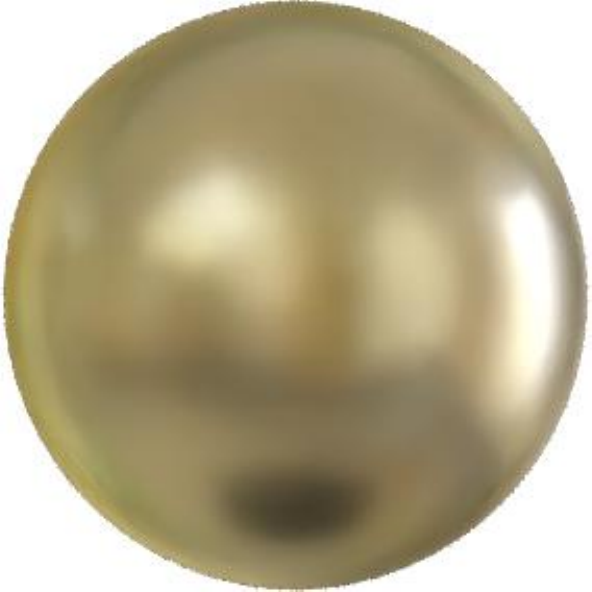}} & 
        
        \noindent\parbox[c]{0.100\textwidth}{\includegraphics[height=0.100\textwidth]{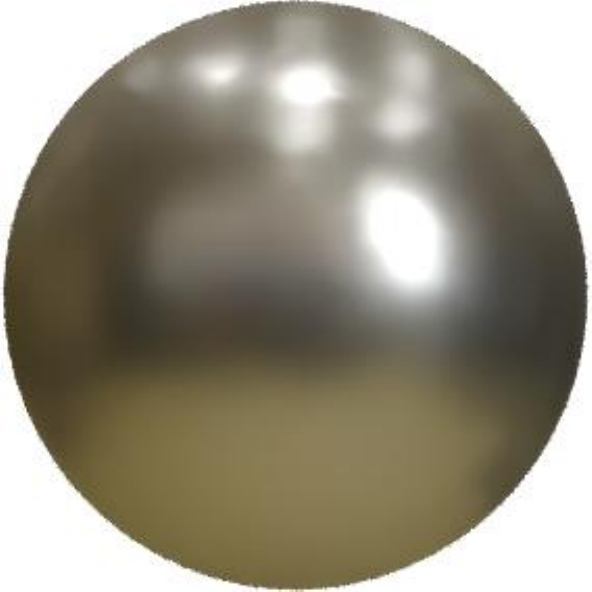}} & 
        \noindent\parbox[c]{0.100\textwidth}{\includegraphics[height=0.100\textwidth]{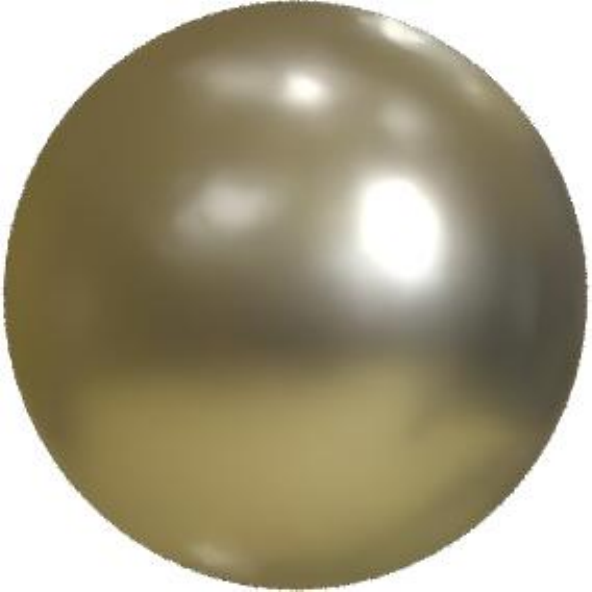}} &
        \noindent\parbox[c]{0.100\textwidth}{\includegraphics[height=0.100\textwidth]{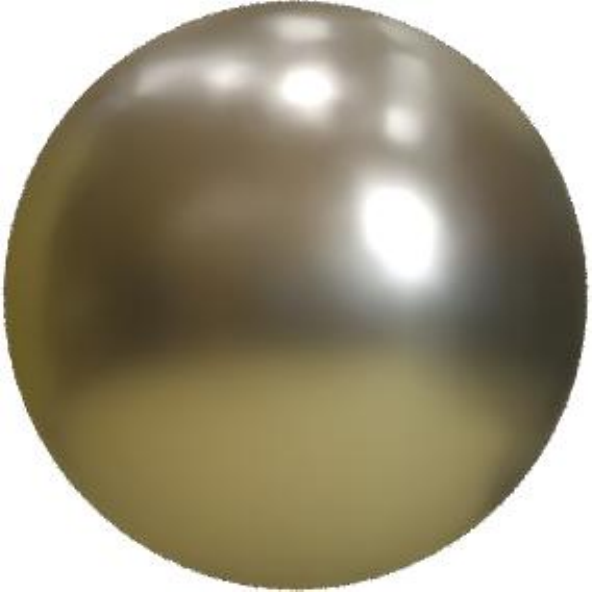}} & 
        \noindent\parbox[c]{0.100\textwidth}{\includegraphics[height=0.100\textwidth]{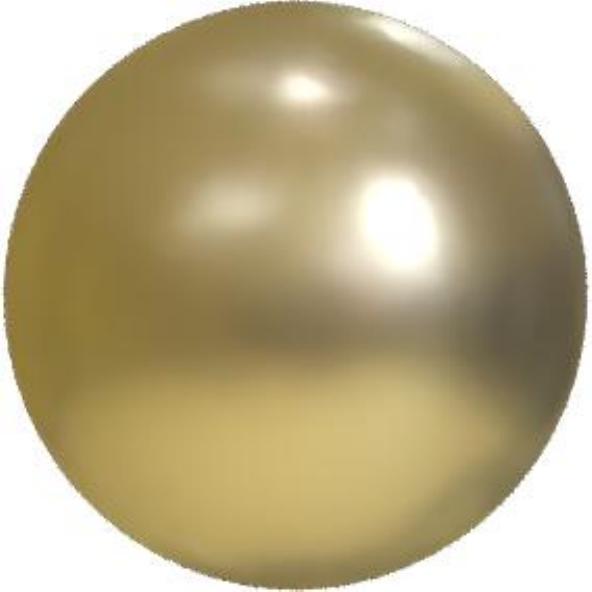}} & 
        \\

        \end{tabu}
    \caption{
    Qualitative results for the Laval indoor dataset using matte balls.}
    \label{fig:additional_indoor_matte}
\end{figure*}

\tabulinesep=0.5pt

\tabulinesep=0.5pt
\begin{figure*}[!t]
    \centering

        \begin{tabu} to \textwidth {
        @{}
        c@{}
        c@{}
        c@{}
        c@{}
        c@{}
        c@{}
    }

        \multicolumn{1}{c}{\shortstack{\hspace{-6pt} \scriptsize Input}}
        &
        \multicolumn{1}{c}{\shortstack{\scriptsize Ground truth}}
        & 
        \multicolumn{1}{c}{\shortstack{\scriptsize StyleLight \cite{wang2022stylelight}}}
        & 
        \multicolumn{1}{c}{\shortstack{\scriptsize SDXL$^\dagger$}} &
        \multicolumn{1}{c}{\shortstack{\scriptsize \begin{tabular}[c]{@{}c@{}}SDXL$^\dagger$+LR+I \\ (ours)\end{tabular}}} 
        \\

        \noindent\parbox[c]{0.14\textwidth}{\includegraphics[height=0.100\textwidth]{storage/appendix_result/indoor_mirror/9C4A0358_input.pdf}} &  
        \noindent\parbox[c]{0.205\textwidth}{\includegraphics[height=0.100\textwidth]{storage/appendix_result/indoor_mirror/9C4A0358_envmapgt.pdf}} & 

        \noindent\parbox[c]{0.205\textwidth}{\includegraphics[height=0.100\textwidth]{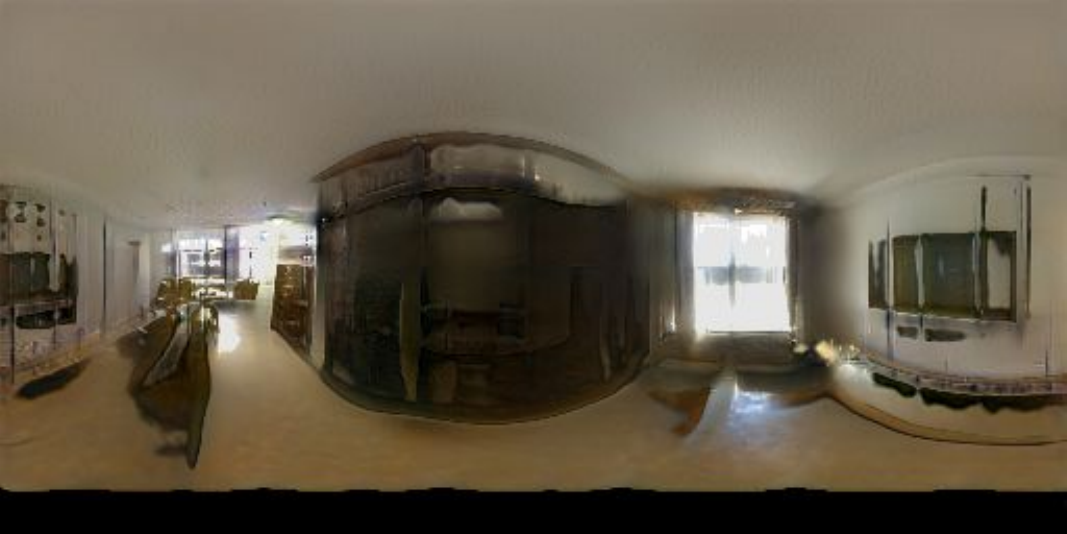}} &
        \noindent\parbox[c]{0.205\textwidth}{\includegraphics[height=0.100\textwidth]{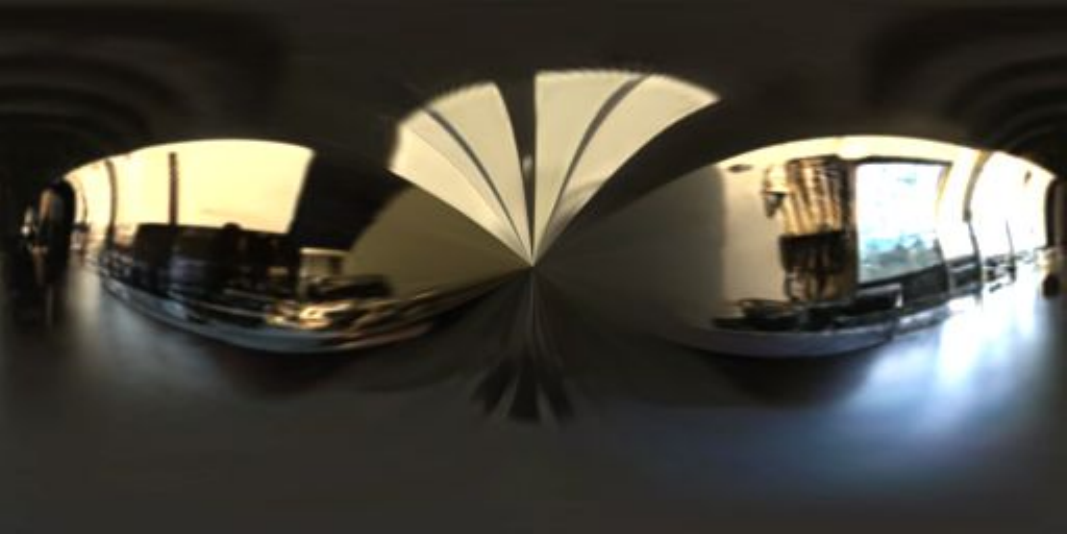}} &
        \noindent\parbox[c]{0.205\textwidth}{\includegraphics[height=0.100\textwidth]{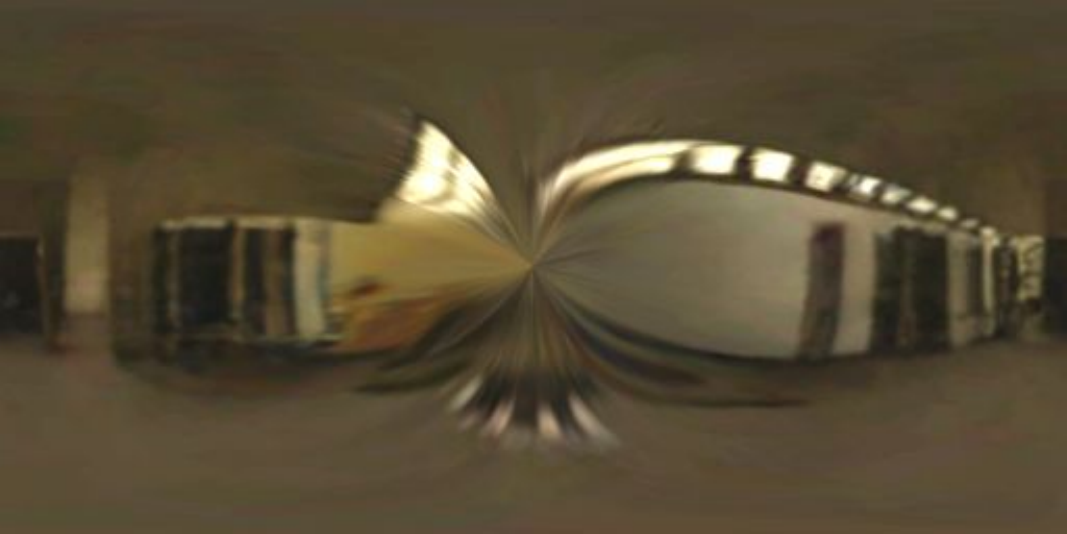}}
        
        \\

        \noindent\parbox[c]{0.14\textwidth}{\includegraphics[height=0.100\textwidth]{storage/appendix_result/indoor_mirror/9C4A2040_input.pdf}} &  
        \noindent\parbox[c]{0.205\textwidth}{\includegraphics[height=0.100\textwidth]{storage/appendix_result/indoor_mirror/9C4A2040_envmapgt.pdf}} & 

        \noindent\parbox[c]{0.205\textwidth}{\includegraphics[height=0.100\textwidth]{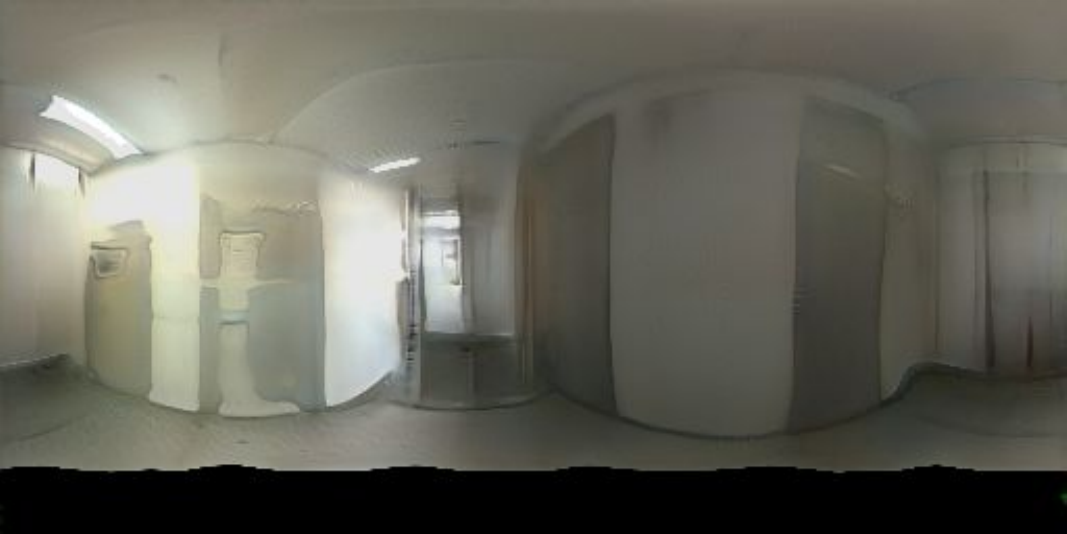}} &
        \noindent\parbox[c]{0.205\textwidth}{\includegraphics[height=0.100\textwidth]{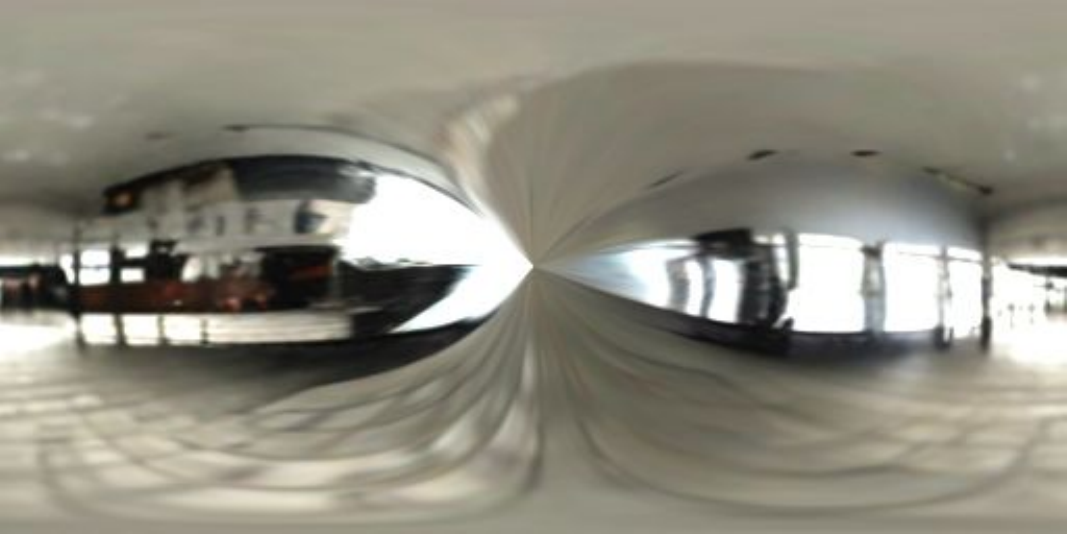}} &
        \noindent\parbox[c]{0.205\textwidth}{\includegraphics[height=0.100\textwidth]{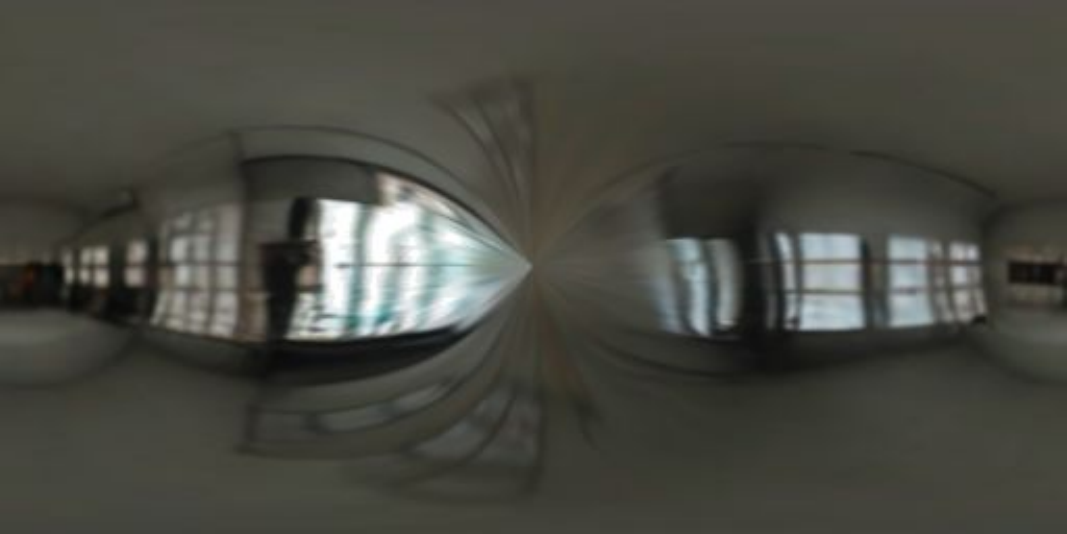}}
        
        \\

        \noindent\parbox[c]{0.14\textwidth}{\includegraphics[height=0.100\textwidth]{storage/appendix_result/indoor_mirror/9C4A2806_input.pdf}} &  
        \noindent\parbox[c]{0.205\textwidth}{\includegraphics[height=0.100\textwidth]{storage/appendix_result/indoor_mirror/9C4A2806_envmapgt.pdf}} & 
        
        \noindent\parbox[c]{0.205\textwidth}{\includegraphics[height=0.100\textwidth]{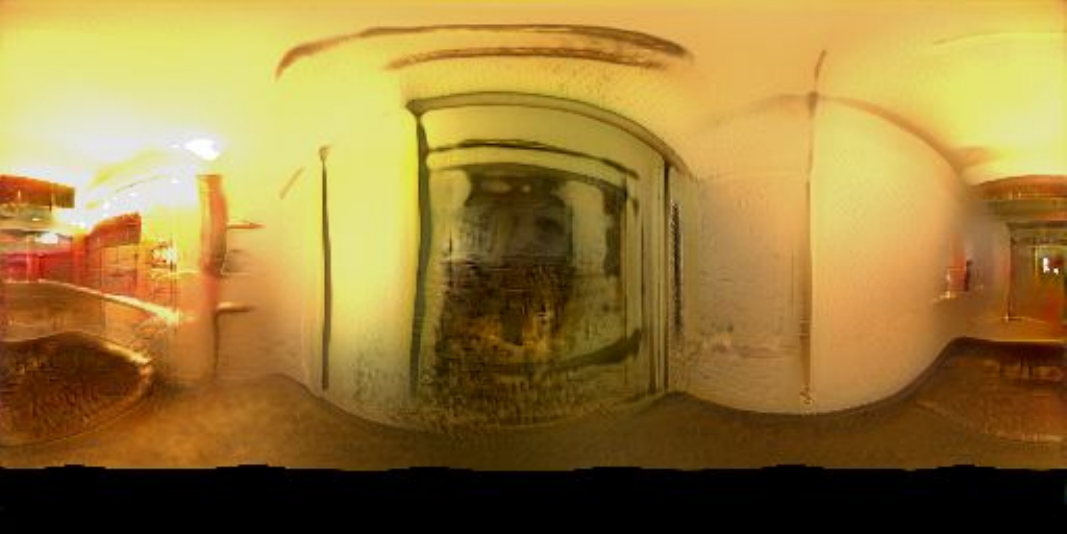}} &
        \noindent\parbox[c]{0.205\textwidth}{\includegraphics[height=0.100\textwidth]{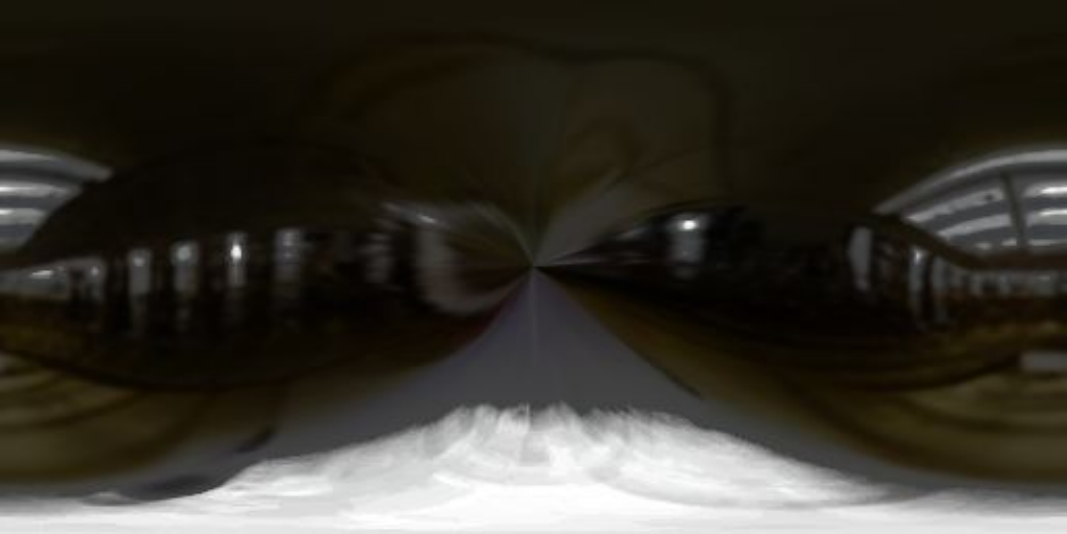}} &
        \noindent\parbox[c]{0.205\textwidth}{\includegraphics[height=0.100\textwidth]{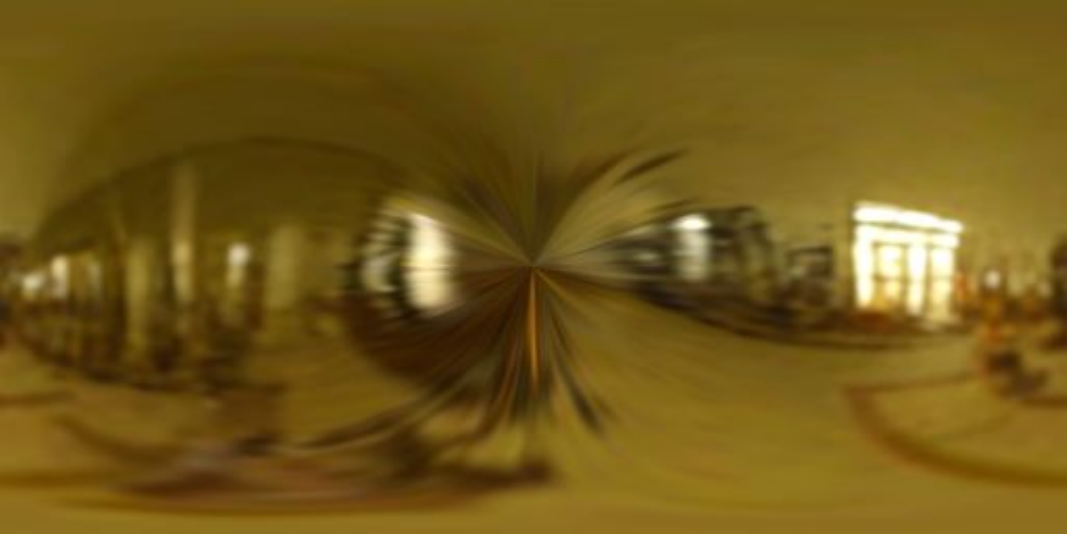}}
        
        \\

        \noindent\parbox[c]{0.14\textwidth}{\includegraphics[height=0.100\textwidth]{storage/appendix_result/indoor_mirror/9C4A3109_input.pdf}} &  
        \noindent\parbox[c]{0.205\textwidth}{\includegraphics[height=0.100\textwidth]{storage/appendix_result/indoor_mirror/9C4A3109_envmapgt.pdf}} & 

        \noindent\parbox[c]{0.205\textwidth}{\includegraphics[height=0.100\textwidth]{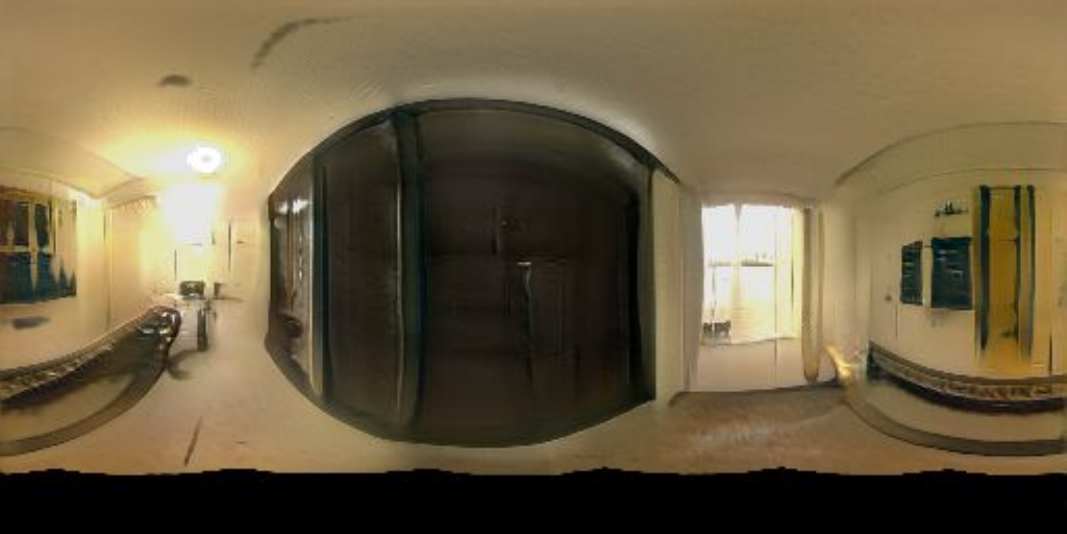}} &
        \noindent\parbox[c]{0.205\textwidth}{\includegraphics[height=0.100\textwidth]{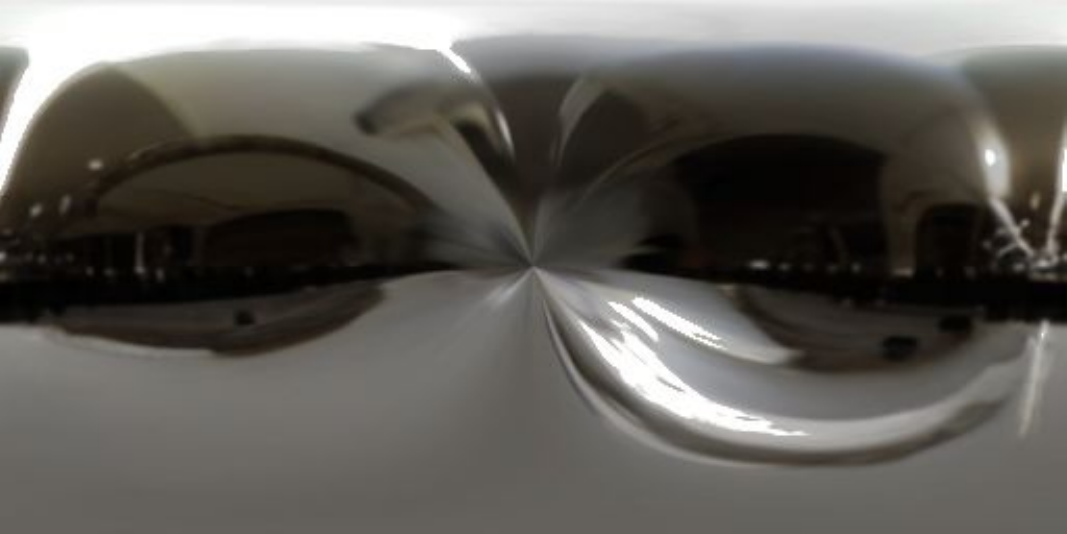}} &
        \noindent\parbox[c]{0.205\textwidth}{\includegraphics[height=0.100\textwidth]{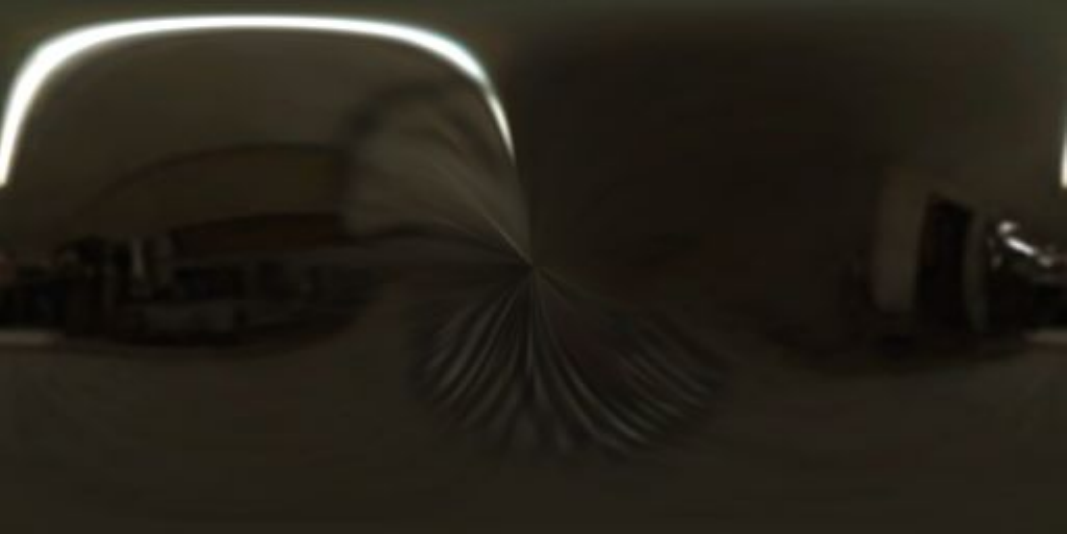}}
        
        \\

        \noindent\parbox[c]{0.14\textwidth}{\includegraphics[height=0.100\textwidth]{storage/appendix_result/indoor_mirror/9C4A3151_input.pdf}} &  
        \noindent\parbox[c]{0.205\textwidth}{\includegraphics[height=0.100\textwidth]{storage/appendix_result/indoor_mirror/9C4A3151_envmapgt.pdf}} & 

        \noindent\parbox[c]{0.205\textwidth}{\includegraphics[height=0.100\textwidth]{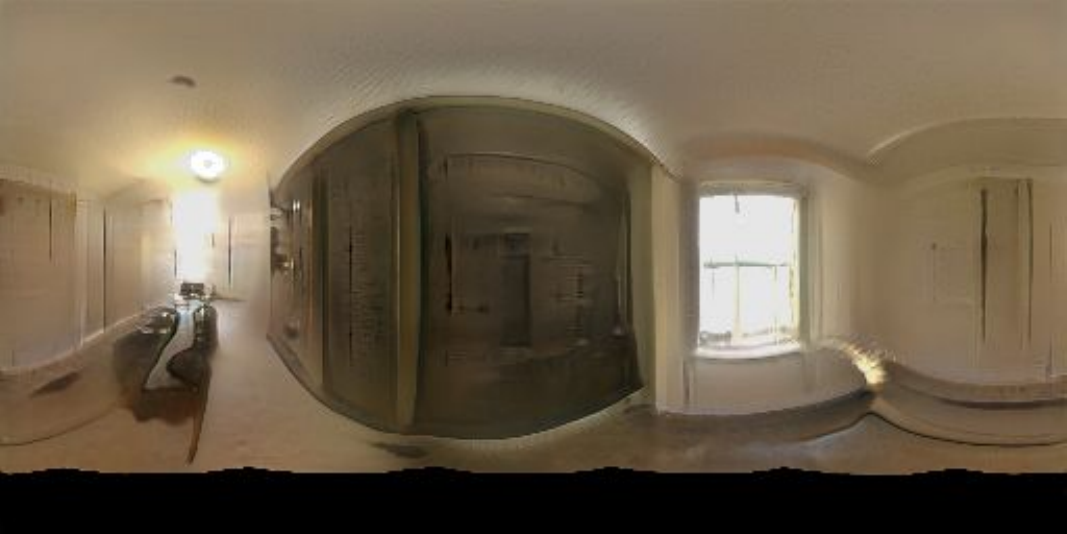}} &
        \noindent\parbox[c]{0.205\textwidth}{\includegraphics[height=0.100\textwidth]{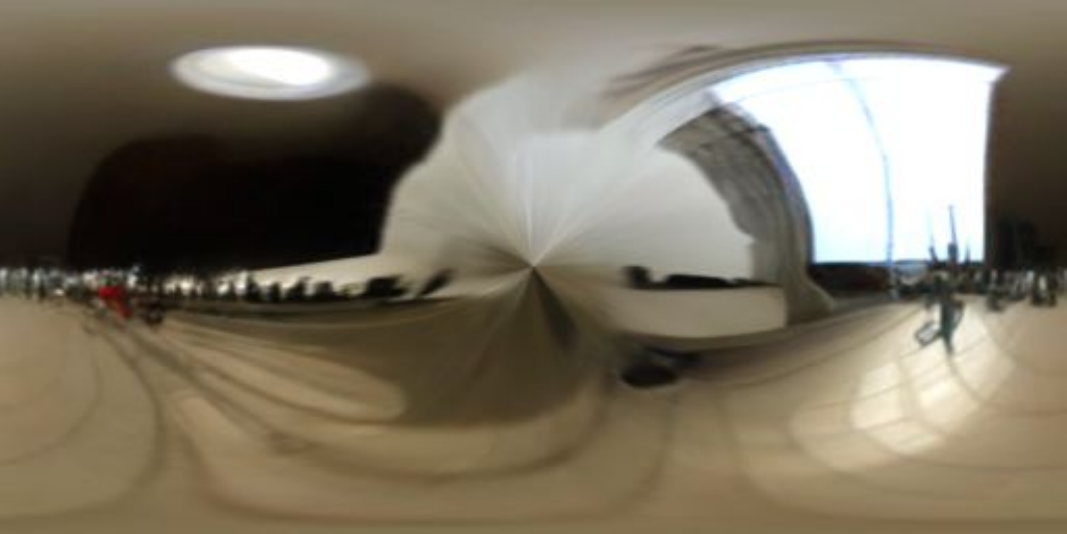}} &
        \noindent\parbox[c]{0.205\textwidth}{\includegraphics[height=0.100\textwidth]{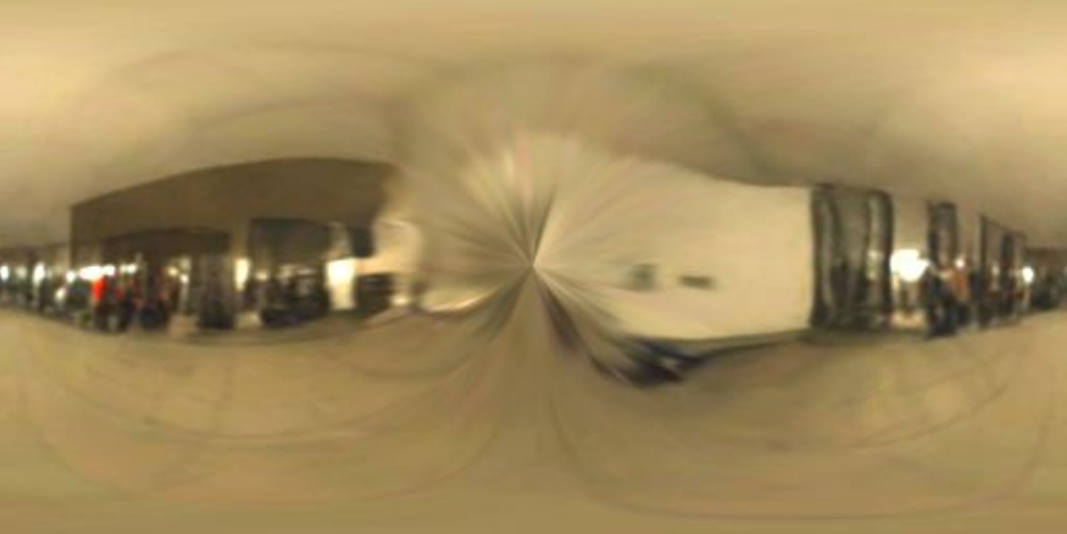}}        
        \\

        \noindent\parbox[c]{0.14\textwidth}{\includegraphics[height=0.100\textwidth]{storage/appendix_result/indoor_mirror/9C4A4878_input.pdf}} & 
        \noindent\parbox[c]{0.205\textwidth}{\includegraphics[height=0.100\textwidth]{storage/appendix_result/indoor_mirror/9C4A4878_envmapgt.pdf}} &  

        \noindent\parbox[c]{0.205\textwidth}{\includegraphics[height=0.100\textwidth]{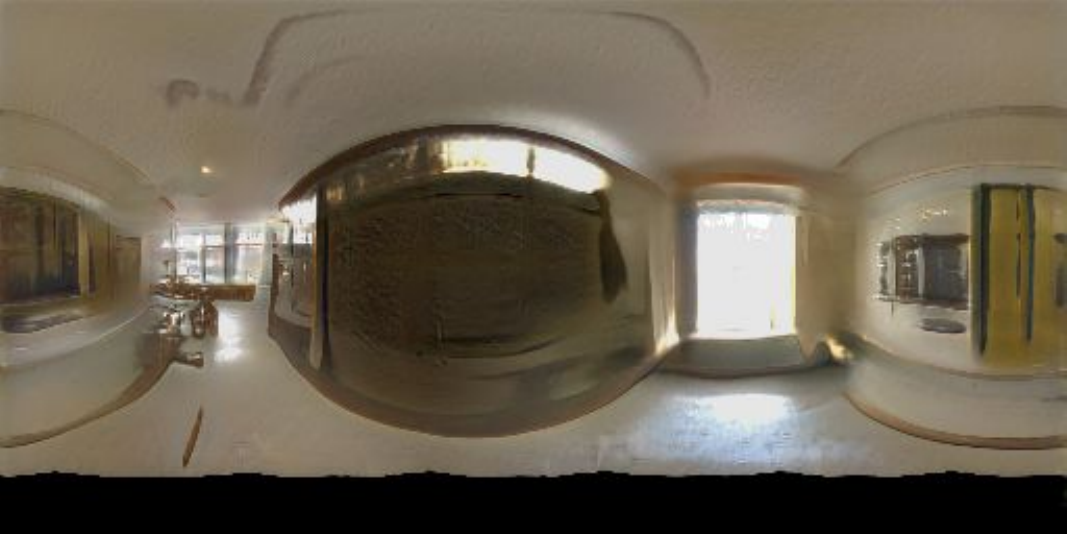}} &
        \noindent\parbox[c]{0.205\textwidth}{\includegraphics[height=0.100\textwidth]{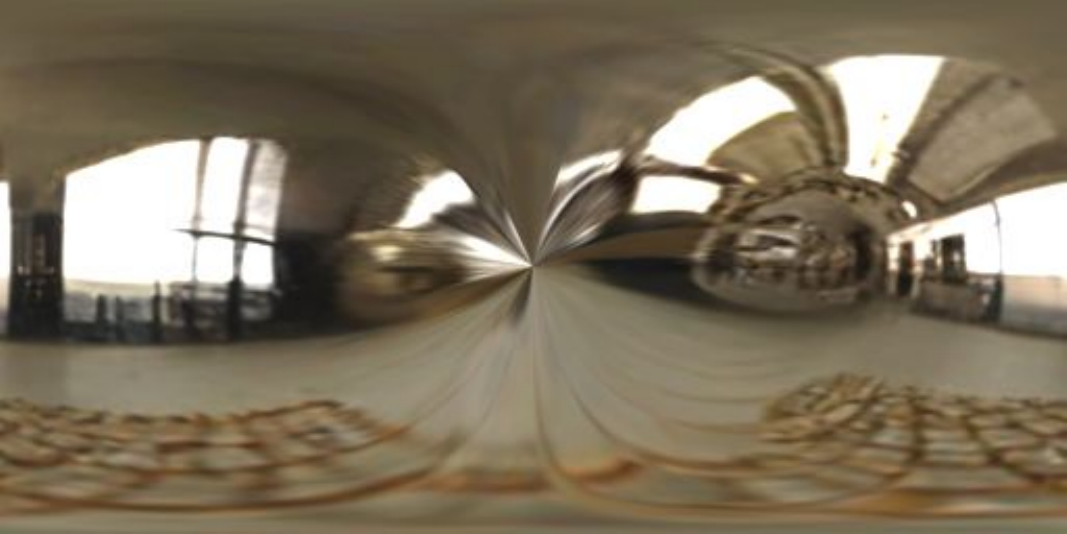}} &
        \noindent\parbox[c]{0.205\textwidth}{\includegraphics[height=0.100\textwidth]{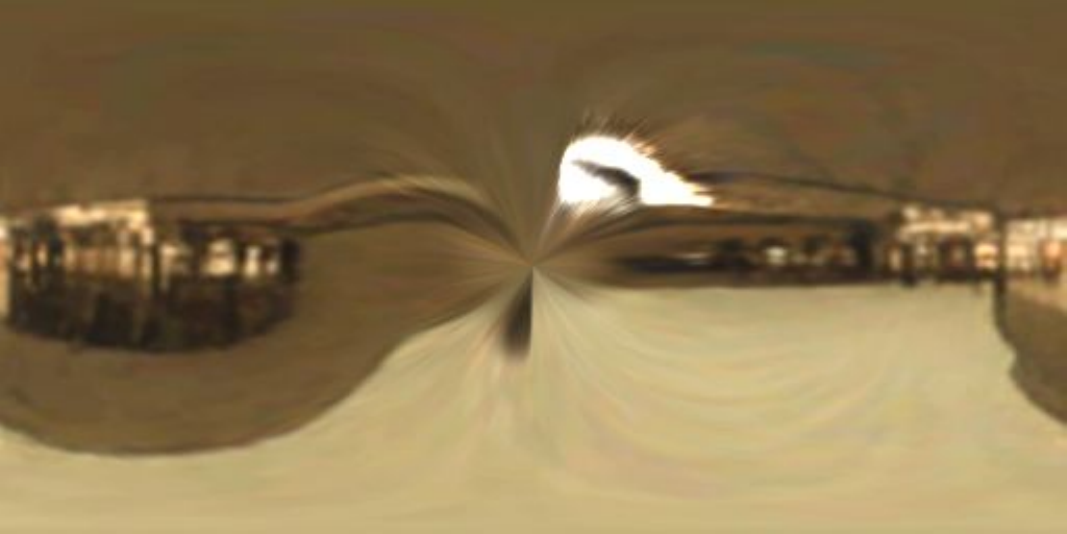}}
      
        \\

        \noindent\parbox[c]{0.14\textwidth}{\includegraphics[height=0.100\textwidth]{storage/appendix_result/indoor_mirror/9C4A9817_input.pdf}} &  
        \noindent\parbox[c]{0.205\textwidth}{\includegraphics[height=0.100\textwidth]{storage/appendix_result/indoor_mirror/9C4A9817_envmapgt.pdf}} & 

        \noindent\parbox[c]{0.205\textwidth}{\includegraphics[height=0.100\textwidth]{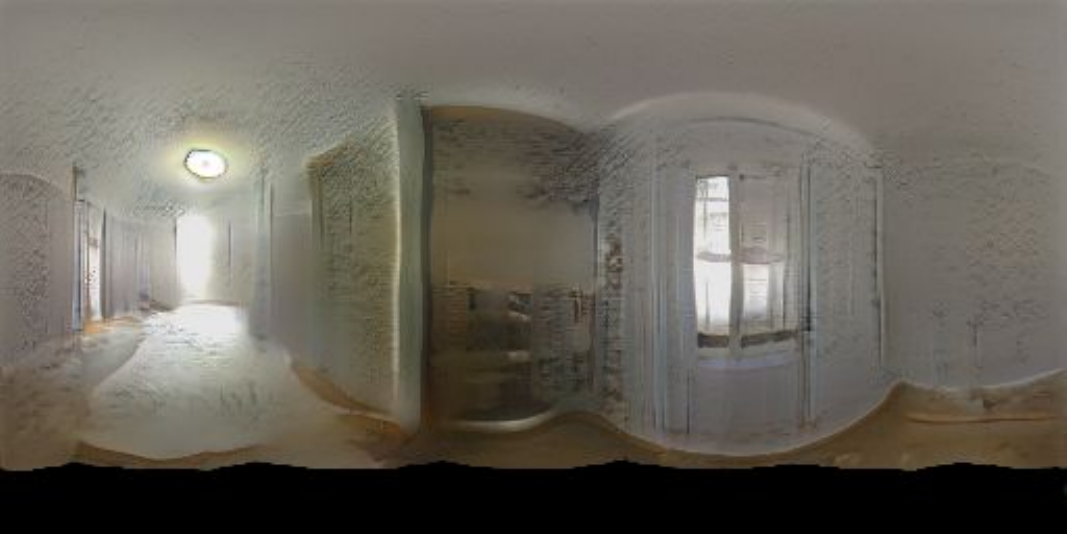}} &
        \noindent\parbox[c]{0.205\textwidth}{\includegraphics[height=0.100\textwidth]{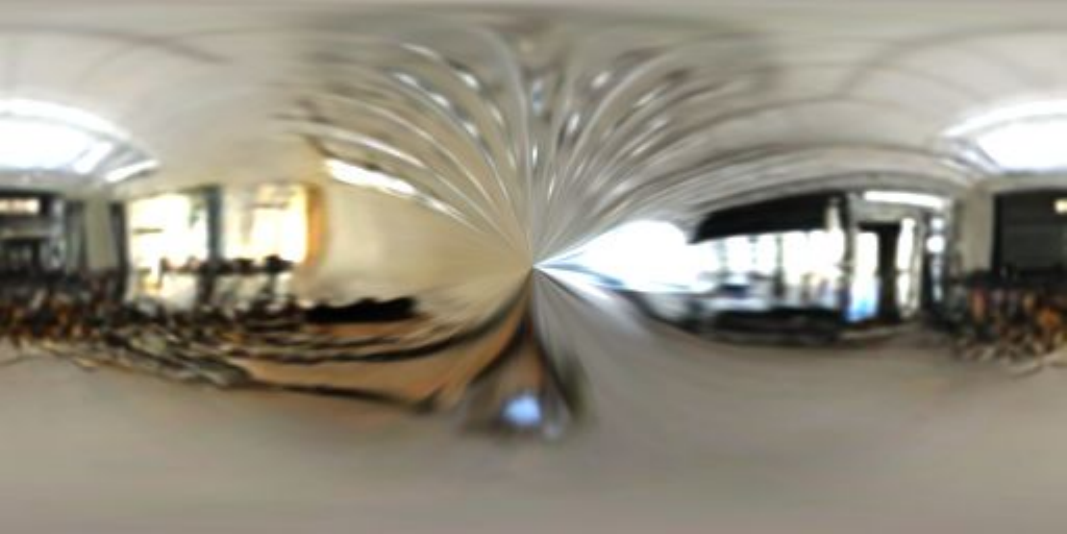}} &
        \noindent\parbox[c]{0.205\textwidth}{\includegraphics[height=0.100\textwidth]{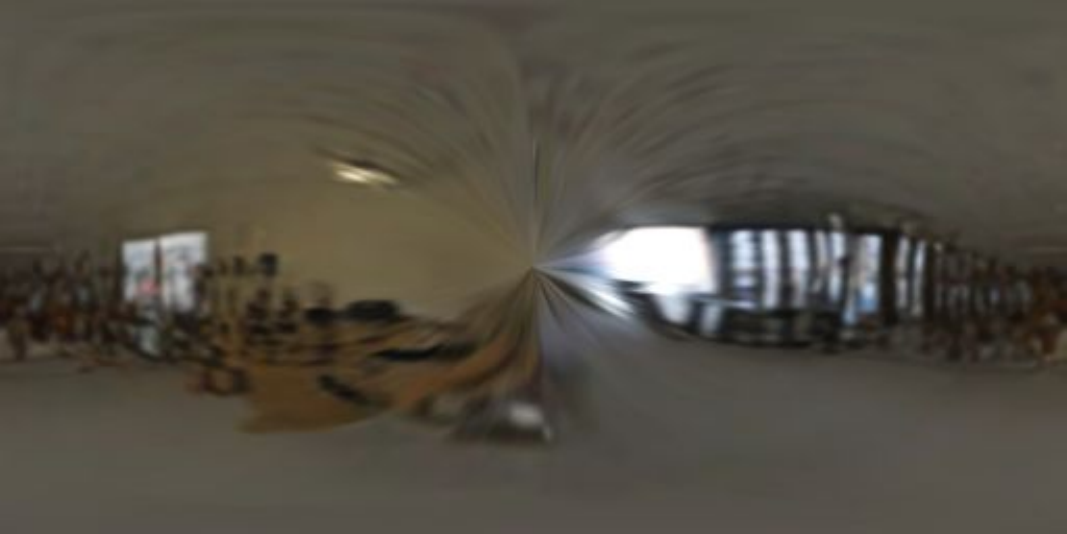}}

        \\

        \noindent\parbox[c]{0.14\textwidth}{\includegraphics[height=0.100\textwidth]{storage/appendix_result/indoor_mirror/AG8A0680_input.pdf}} & 
        \noindent\parbox[c]{0.205\textwidth}{\includegraphics[height=0.100\textwidth]{storage/appendix_result/indoor_mirror/AG8A0680_envmapgt.pdf}} & 

        \noindent\parbox[c]{0.205\textwidth}{\includegraphics[height=0.100\textwidth]{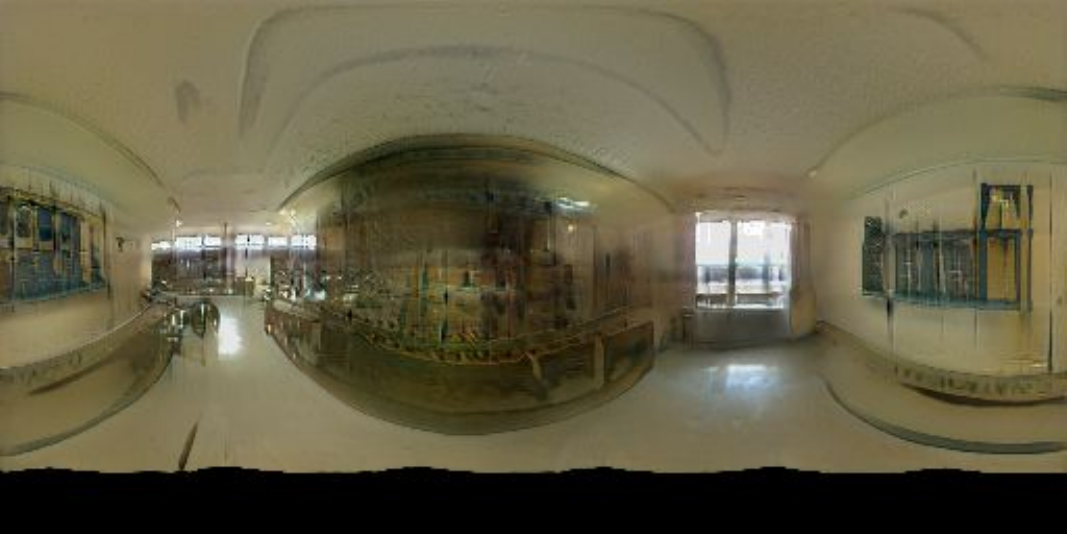}} &
        \noindent\parbox[c]{0.205\textwidth}{\includegraphics[height=0.100\textwidth]{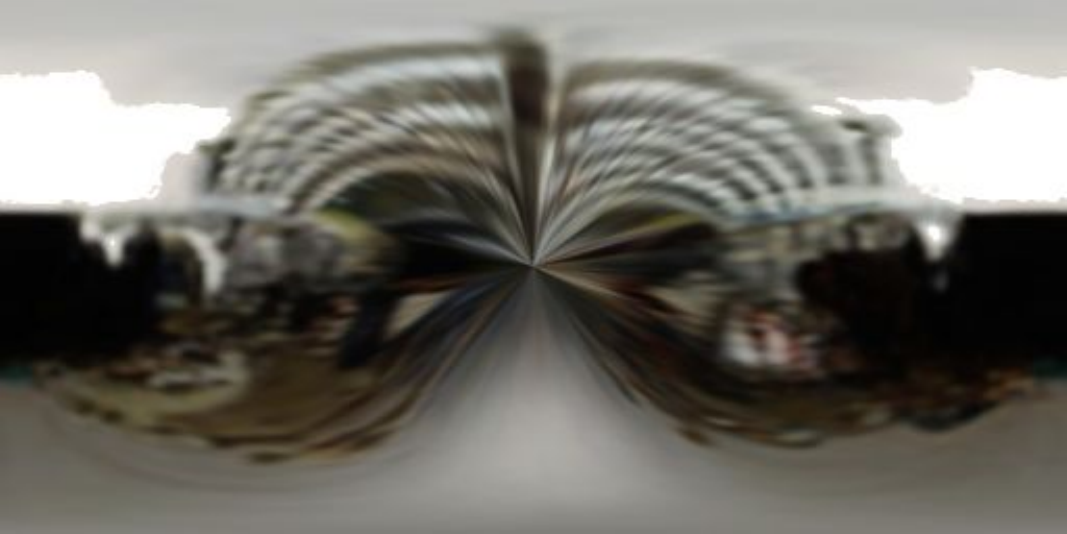}} &
        \noindent\parbox[c]{0.205\textwidth}{\includegraphics[height=0.100\textwidth]{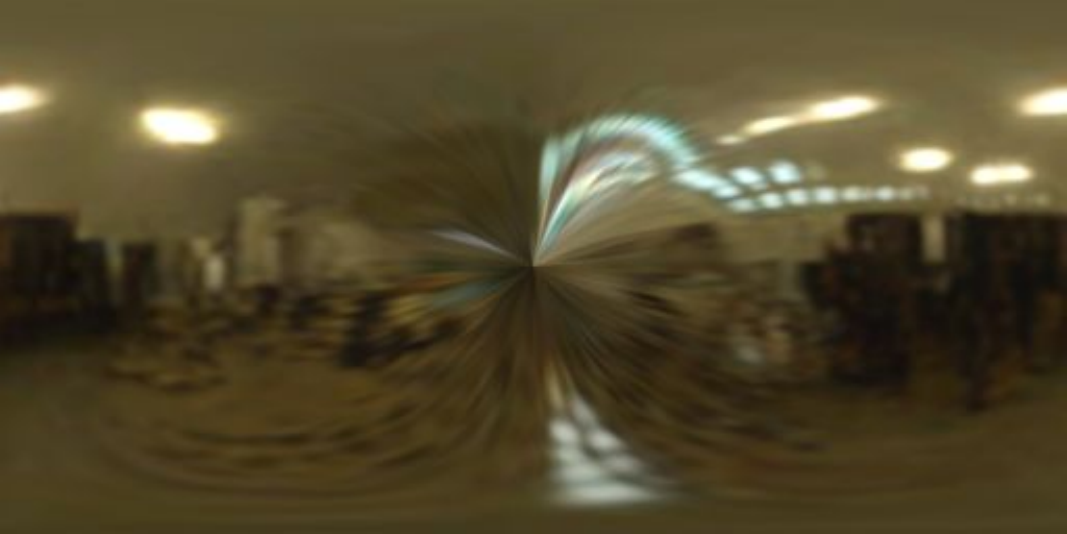}}

        \\

        \noindent\parbox[c]{0.14\textwidth}{\includegraphics[height=0.100\textwidth]{storage/appendix_result/indoor_mirror/AG8A1075_input.pdf}} & 
        \noindent\parbox[c]{0.205\textwidth}{\includegraphics[height=0.100\textwidth]{storage/appendix_result/indoor_mirror/AG8A1075_envmapgt.pdf}} & 

        \noindent\parbox[c]{0.205\textwidth}{\includegraphics[height=0.100\textwidth]{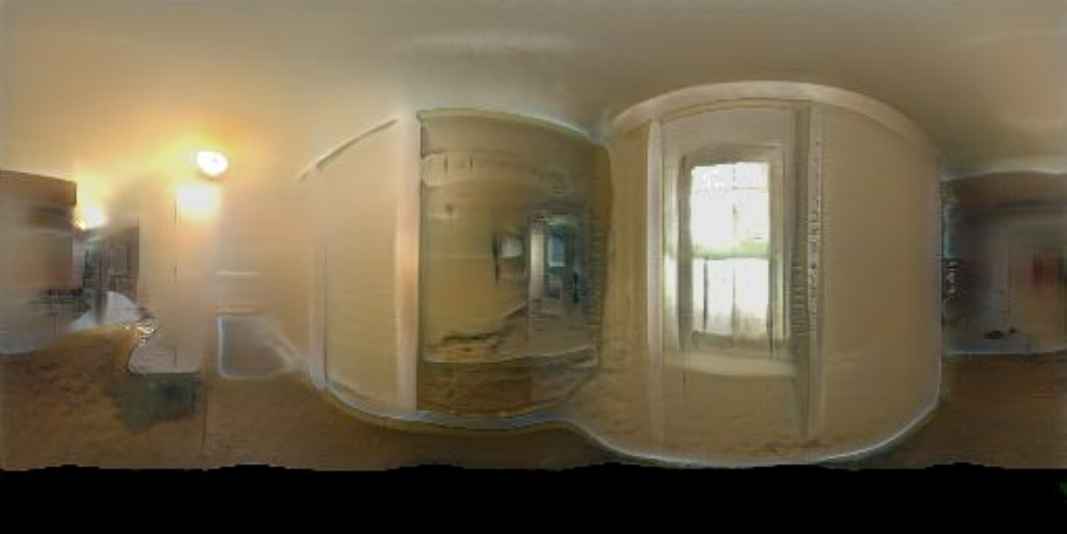}} &
        \noindent\parbox[c]{0.205\textwidth}{\includegraphics[height=0.100\textwidth]{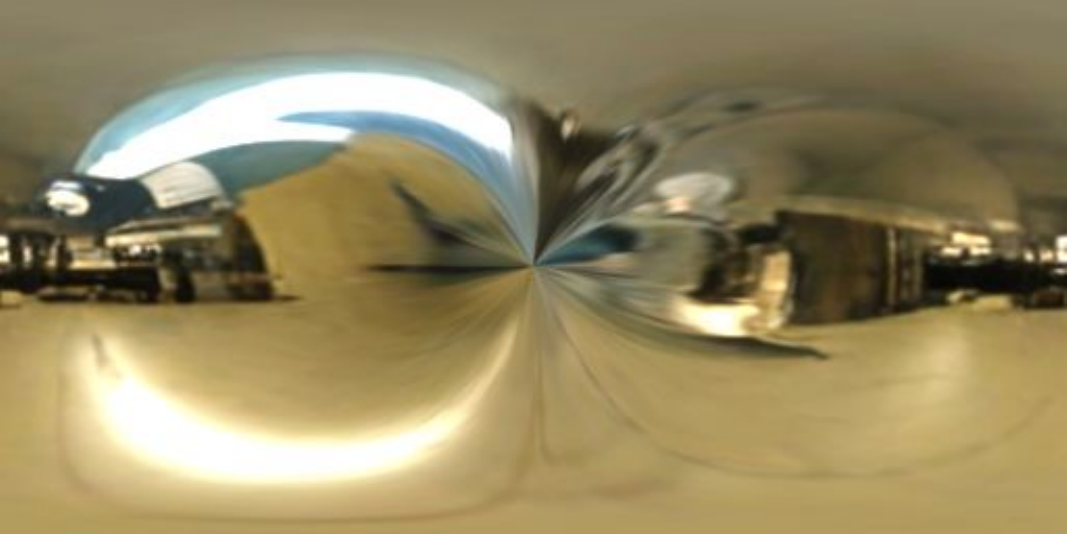}} &
        \noindent\parbox[c]{0.205\textwidth}{\includegraphics[height=0.100\textwidth]{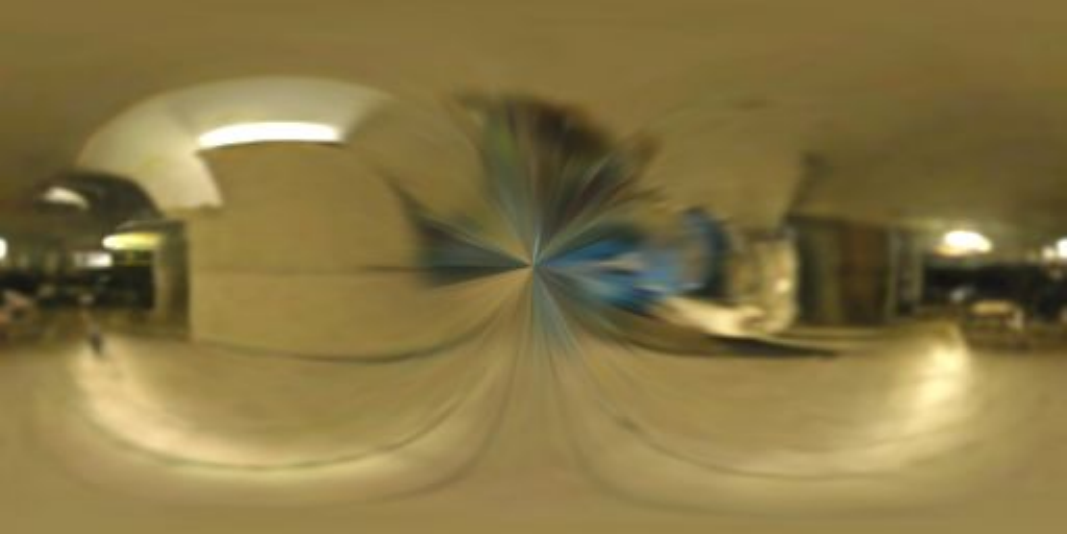}}

        \\

        \noindent\parbox[c]{0.14\textwidth}{\includegraphics[height=0.100\textwidth]{storage/appendix_result/indoor_mirror/AG8A1912_input.pdf}} & 
        \noindent\parbox[c]{0.205\textwidth}{\includegraphics[height=0.100\textwidth]{storage/appendix_result/indoor_mirror/AG8A1912_envmapgt.pdf}} & 

        \noindent\parbox[c]{0.205\textwidth}{\includegraphics[height=0.100\textwidth]{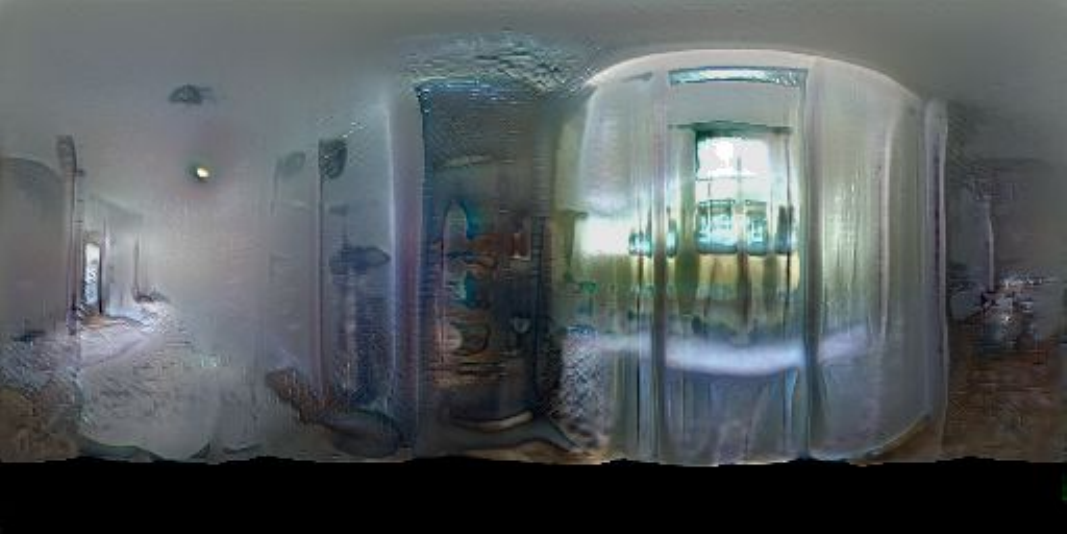}} &
        \noindent\parbox[c]{0.205\textwidth}{\includegraphics[height=0.100\textwidth]{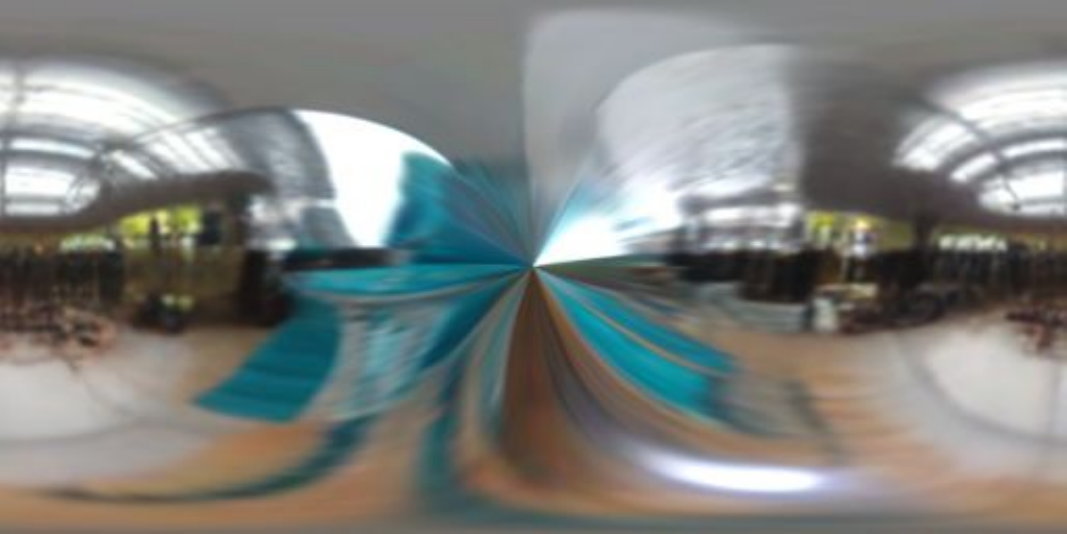}} &
        \noindent\parbox[c]{0.205\textwidth}{\includegraphics[height=0.100\textwidth]{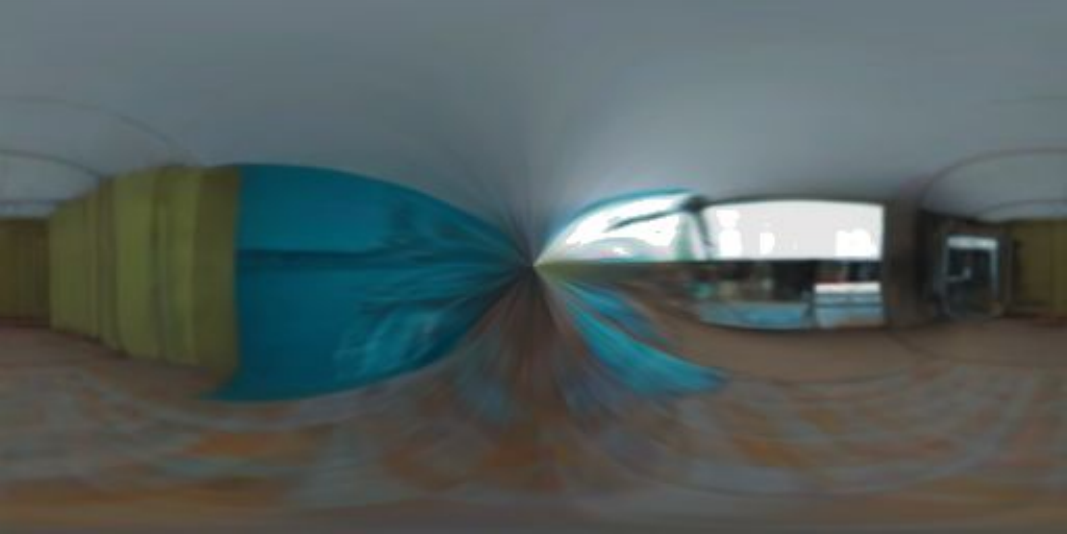}}
        
        \\

        \noindent\parbox[c]{0.14\textwidth}{\includegraphics[height=0.100\textwidth]{storage/appendix_result/indoor_mirror/AG8A4190_input.pdf}} & 
        \noindent\parbox[c]{0.205\textwidth}{\includegraphics[height=0.100\textwidth]{storage/appendix_result/indoor_mirror/AG8A4190_envmapgt.pdf}} & 

        \noindent\parbox[c]{0.205\textwidth}{\includegraphics[height=0.100\textwidth]{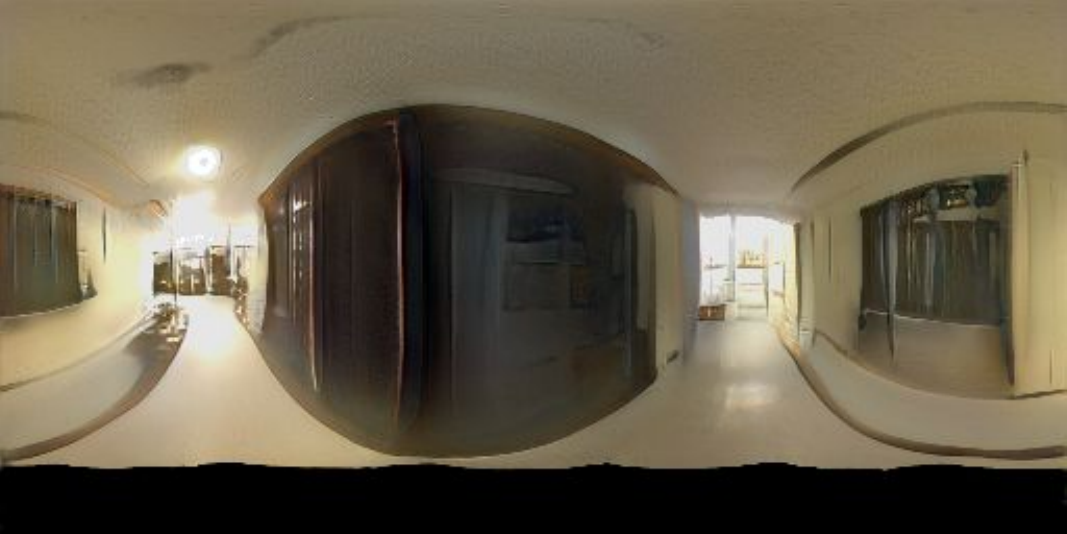}} &
        \noindent\parbox[c]{0.205\textwidth}{\includegraphics[height=0.100\textwidth]{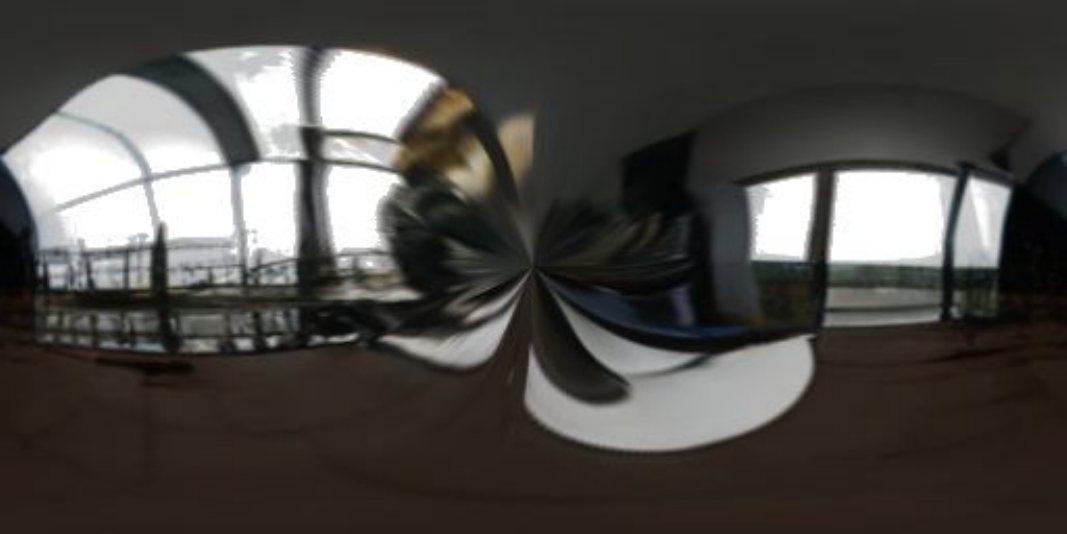}} &
        \noindent\parbox[c]{0.205\textwidth}{\includegraphics[height=0.100\textwidth]{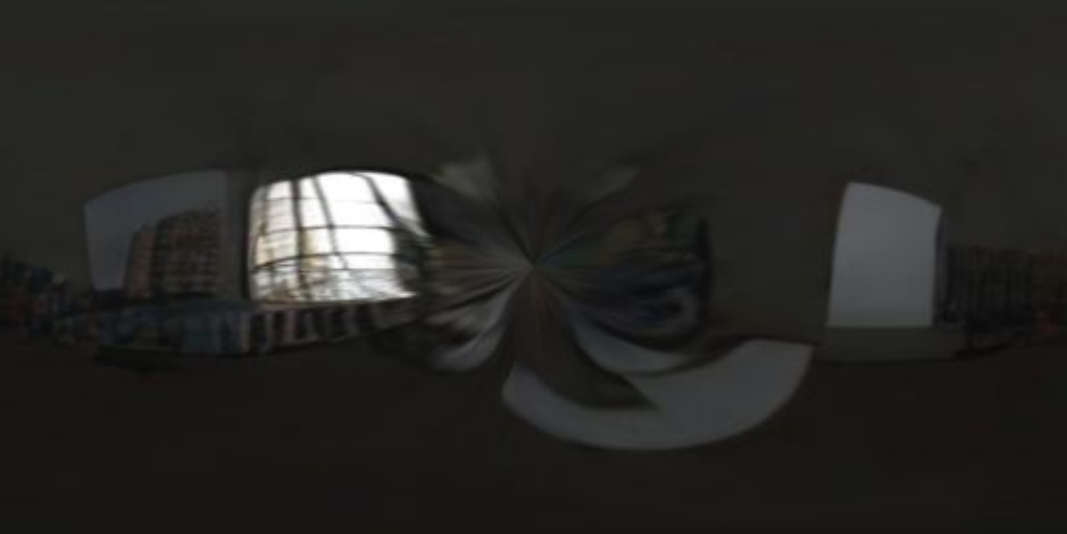}}
        \\
        
        \noindent\parbox[c]{0.14\textwidth}{\includegraphics[height=0.100\textwidth]{storage/appendix_result/indoor_mirror/AG8A8883_input.pdf}} &  
        \noindent\parbox[c]{0.205\textwidth}{\includegraphics[height=0.100\textwidth]{storage/appendix_result/indoor_mirror/AG8A8883_envmapgt.pdf}} & 

        \noindent\parbox[c]{0.205\textwidth}{\includegraphics[height=0.100\textwidth]{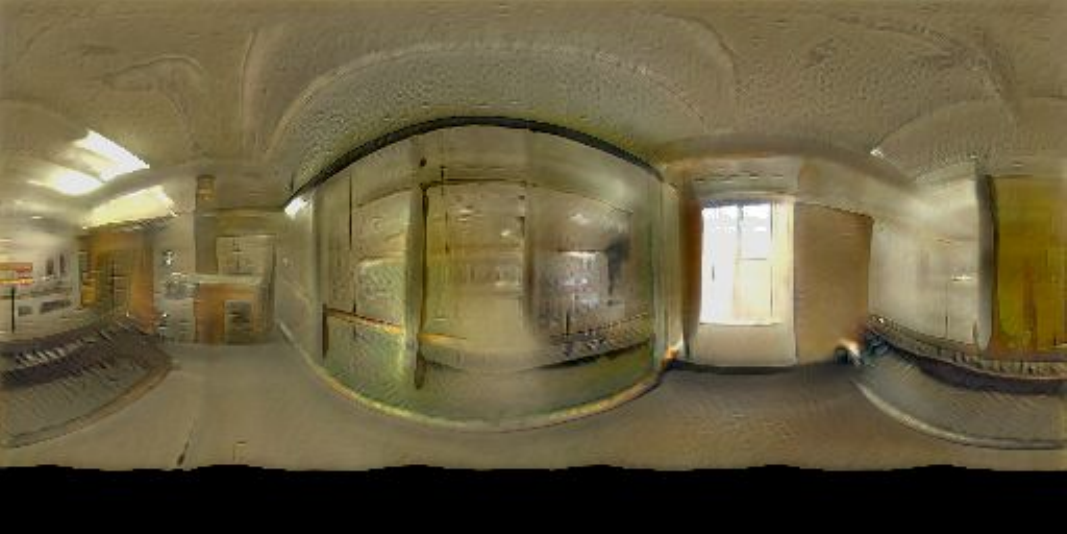}} &
        \noindent\parbox[c]{0.205\textwidth}{\includegraphics[height=0.100\textwidth]{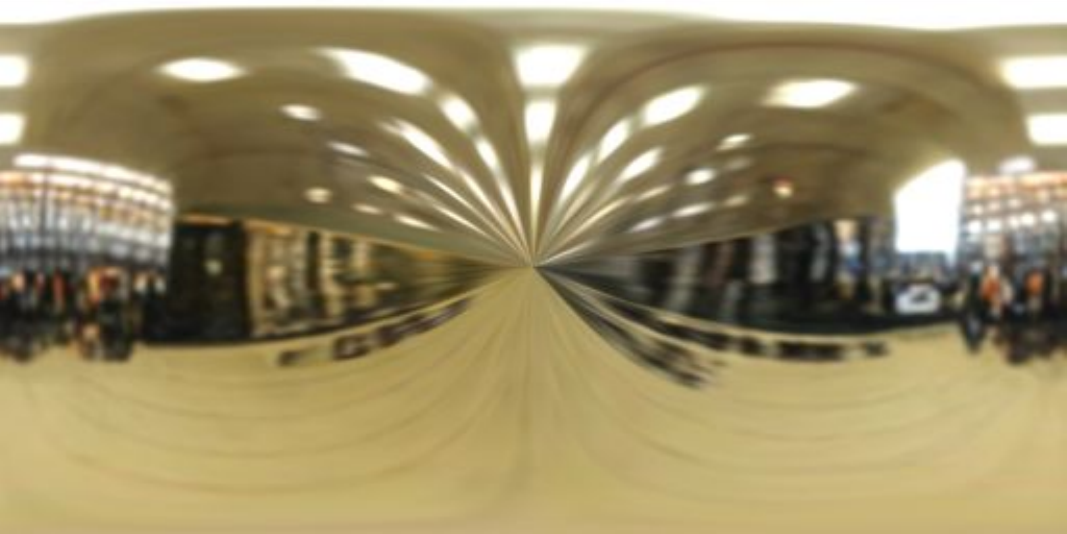}} &
        \noindent\parbox[c]{0.205\textwidth}{\includegraphics[height=0.100\textwidth]{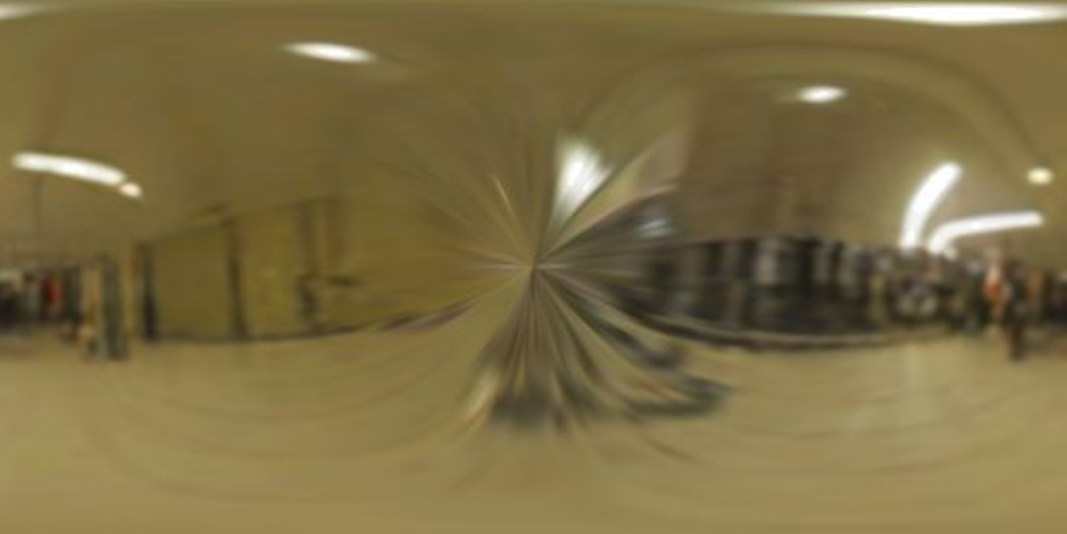}}

        \\

        \end{tabu}
    \caption{
    Unwarped equirectangular maps for the Laval indoor dataset.
    }
    \label{fig:additional_indoor_mirror}
\end{figure*}

\tabulinesep=0.5pt

\tabulinesep=0.5pt

\tabulinesep=0.5pt

\tabulinesep=0.5pt
\begin{figure*}[!t]
    \centering

        \begin{tabu} to \textwidth {
        @{}
        c@{}
        c@{}
        c@{}
        c@{}
        c@{}
        c@{}
    }

        \multicolumn{1}{c}{\shortstack{\hspace{-6pt} \scriptsize Input}}
        & 
        \multicolumn{1}{c}{\shortstack{\scriptsize Ground truth}}
        & 
        \multicolumn{1}{c}{\shortstack{\scriptsize StyleLight \cite{wang2022stylelight}}}
        &
        \multicolumn{1}{c}{\shortstack{\scriptsize SDXL$^\dagger$}} 
        &
        \multicolumn{1}{c}{\shortstack{\scriptsize \begin{tabular}[c]{@{}c@{}}SDXL$^\dagger$+LR+I \\ (ours)\end{tabular}}} 
        
        \\

        \noindent\parbox[c]{0.14\textwidth}{\includegraphics[height=0.100\textwidth]{storage/appendix_result/polyhaven_matte/abandoned_hopper_terminal_04_4k_input.pdf}} & 
        \noindent\parbox[c]{0.205\textwidth}{\includegraphics[height=0.100\textwidth]{storage/appendix_result/polyhaven_matte/abandoned_hopper_terminal_04_4k_envmapgt.pdf}} & 

        \noindent\parbox[c]{0.205\textwidth}{\includegraphics[height=0.100\textwidth]{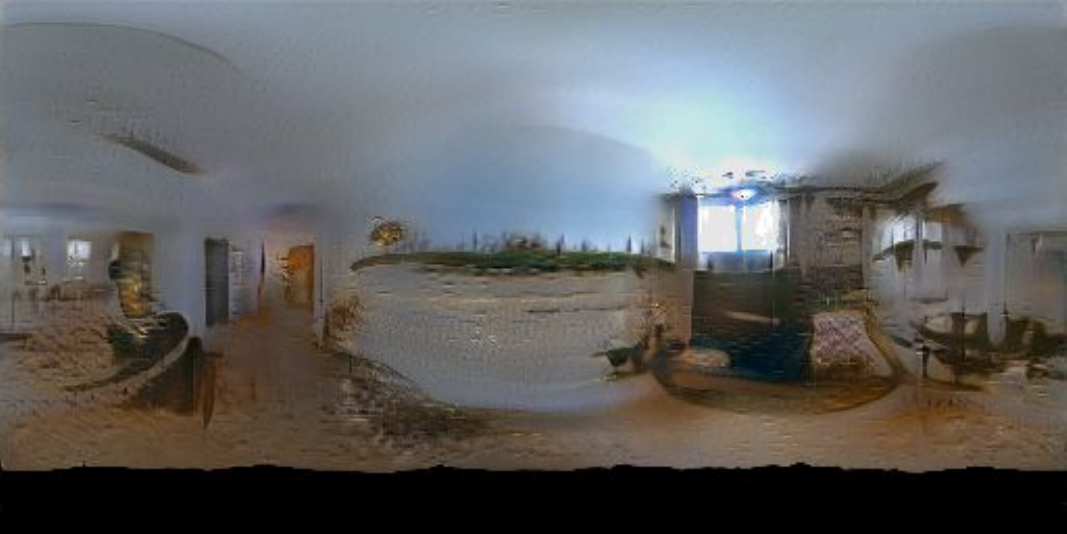}} &
        \noindent\parbox[c]{0.205\textwidth}{\includegraphics[height=0.100\textwidth]{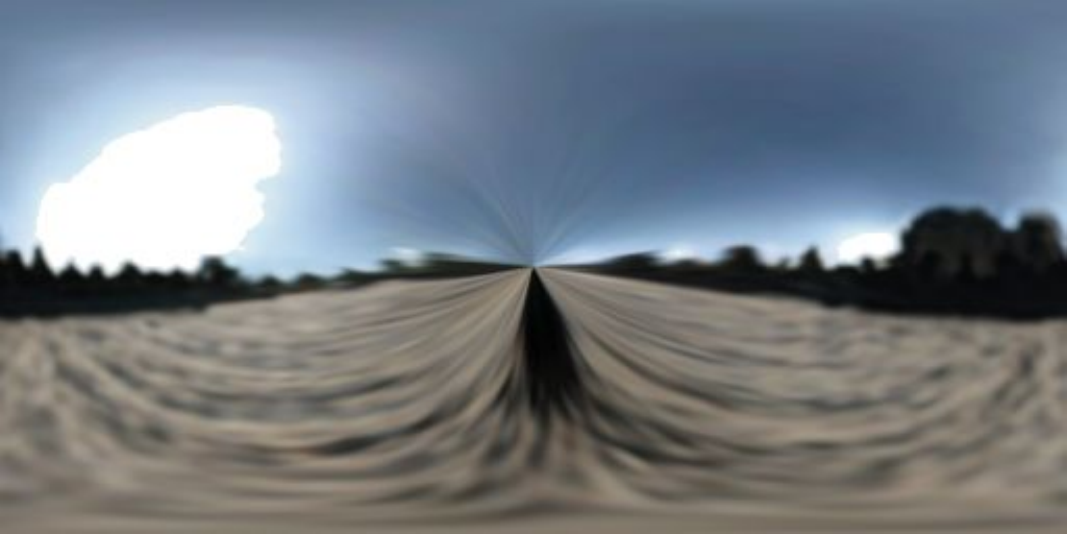}} &
        \noindent\parbox[c]{0.205\textwidth}{\includegraphics[height=0.100\textwidth]{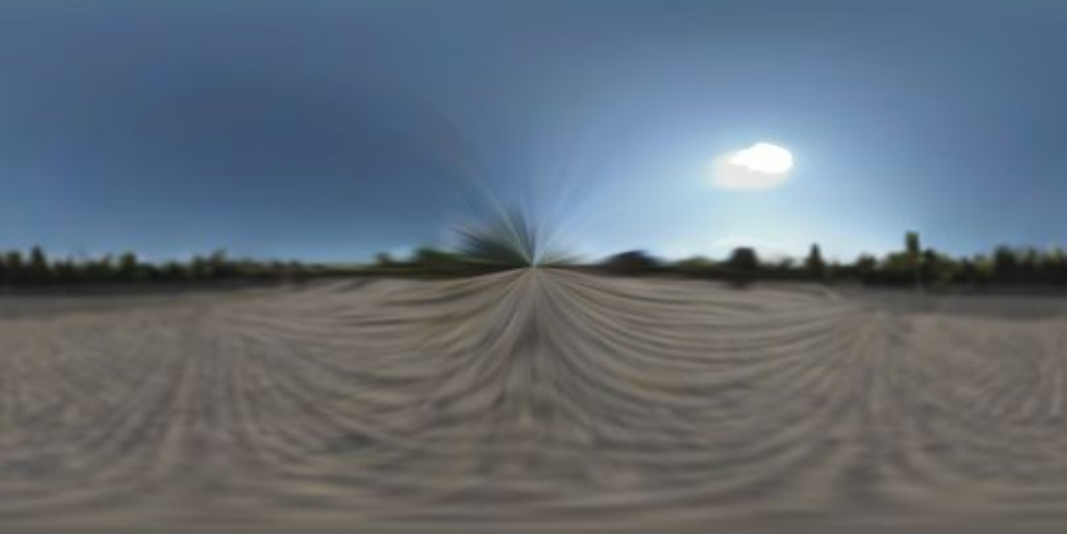}} 
        
        \\

        \noindent\parbox[c]{0.14\textwidth}{\includegraphics[height=0.100\textwidth]{storage/appendix_result/polyhaven_matte/brick_factory_01_4k_input.pdf}} & 
        \noindent\parbox[c]{0.205\textwidth}{\includegraphics[height=0.100\textwidth]{storage/appendix_result/polyhaven_matte/brick_factory_01_4k_envmapgt.pdf}} & 

        \noindent\parbox[c]{0.205\textwidth}{\includegraphics[height=0.100\textwidth]{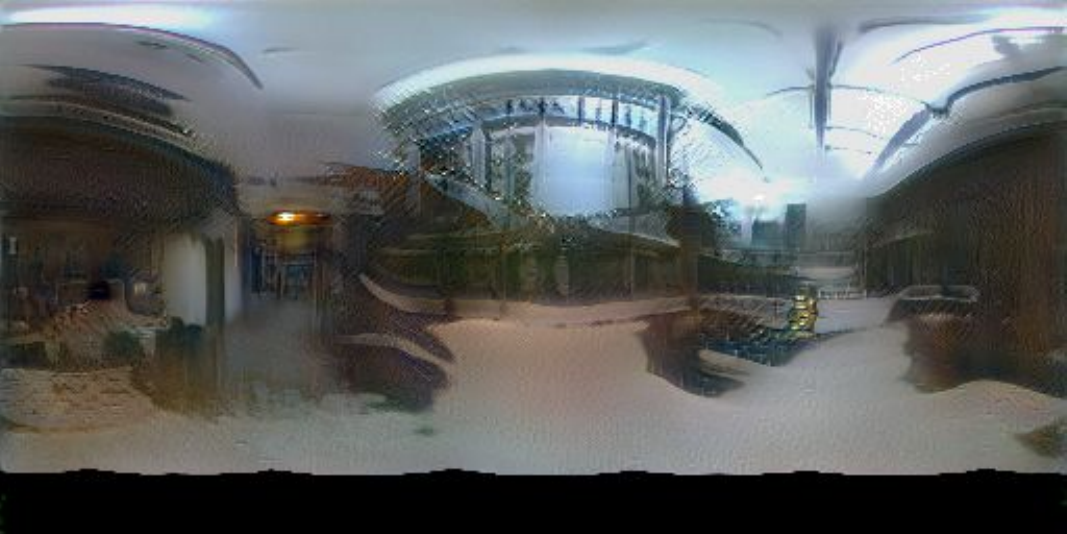}} &
        \noindent\parbox[c]{0.205\textwidth}{\includegraphics[height=0.100\textwidth]{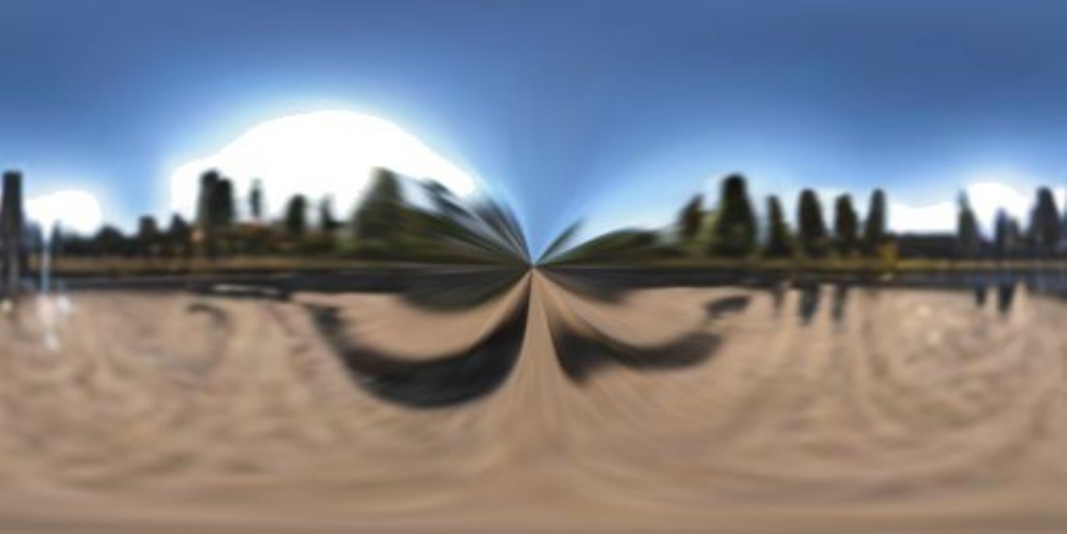}} &
        \noindent\parbox[c]{0.205\textwidth}{\includegraphics[height=0.100\textwidth]{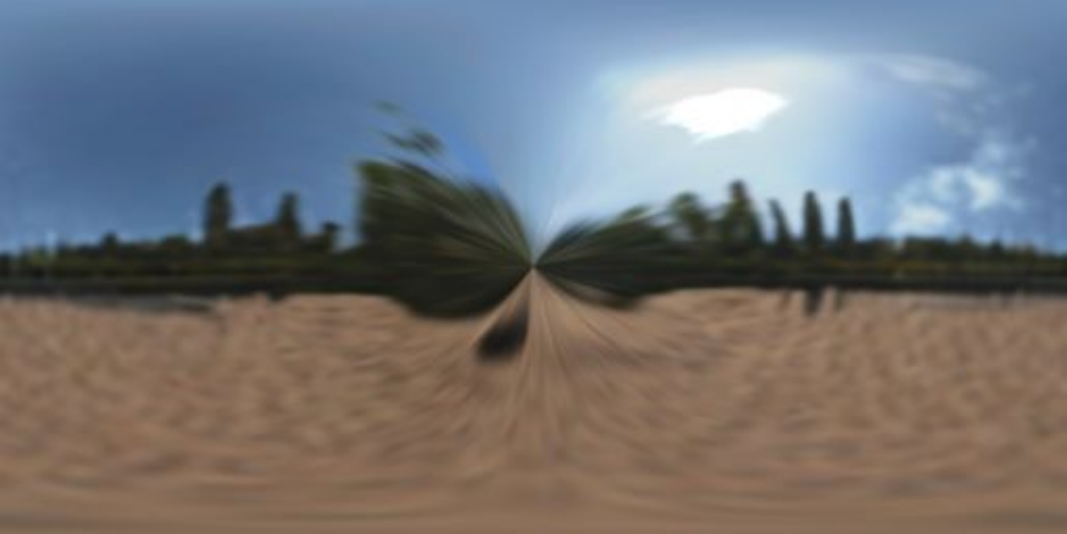}} 
        
        \\

        \noindent\parbox[c]{0.14\textwidth}{\includegraphics[height=0.100\textwidth]{storage/appendix_result/polyhaven_matte/brown_photostudio_05_4k_input.pdf}} &  
        \noindent\parbox[c]{0.205\textwidth}{\includegraphics[height=0.100\textwidth]{storage/appendix_result/polyhaven_matte/brown_photostudio_05_4k_envmapgt.pdf}} & 

        \noindent\parbox[c]{0.205\textwidth}{\includegraphics[height=0.100\textwidth]{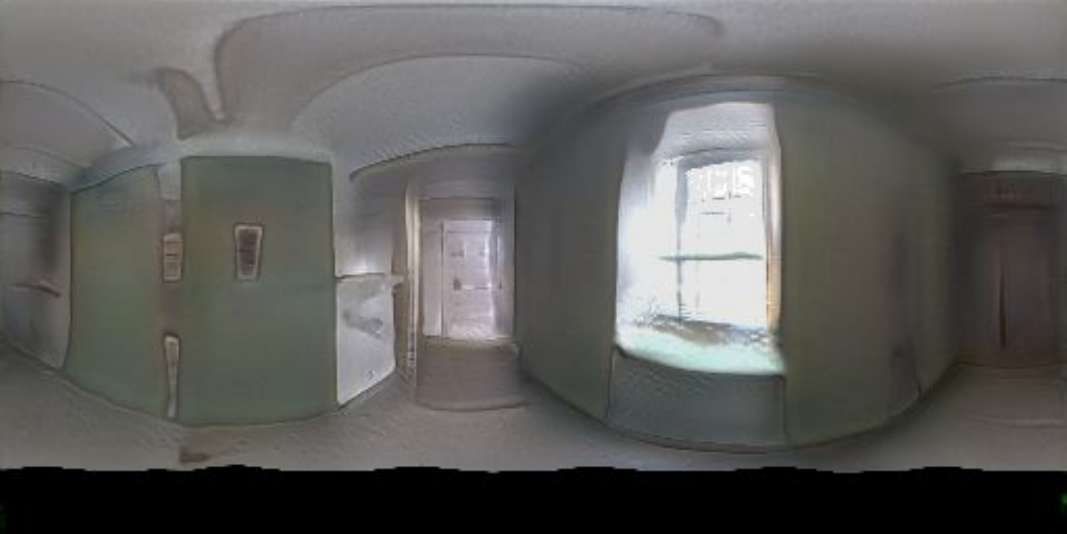}} &
        \noindent\parbox[c]{0.205\textwidth}{\includegraphics[height=0.100\textwidth]{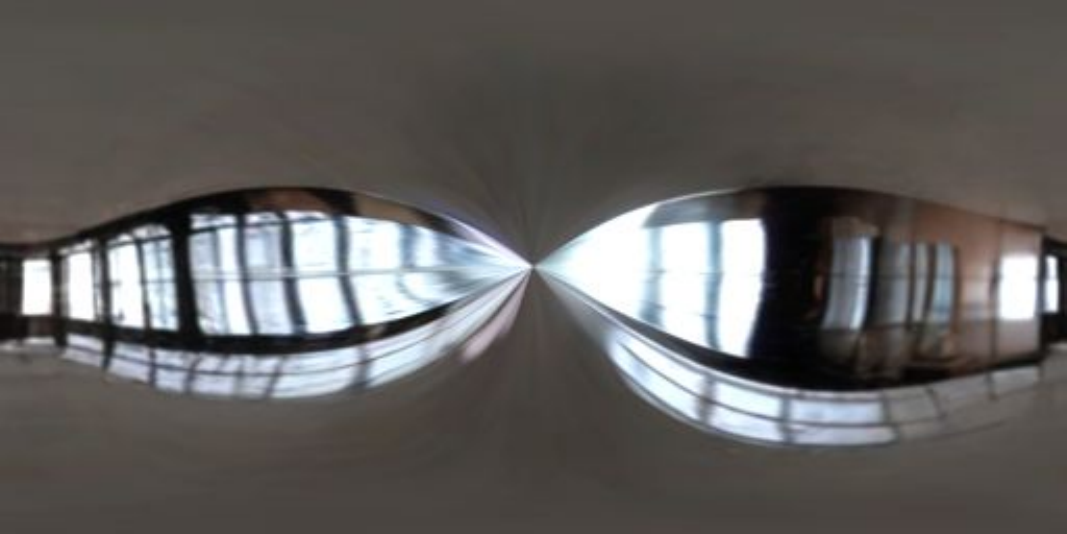}} &
        \noindent\parbox[c]{0.205\textwidth}{\includegraphics[height=0.100\textwidth]{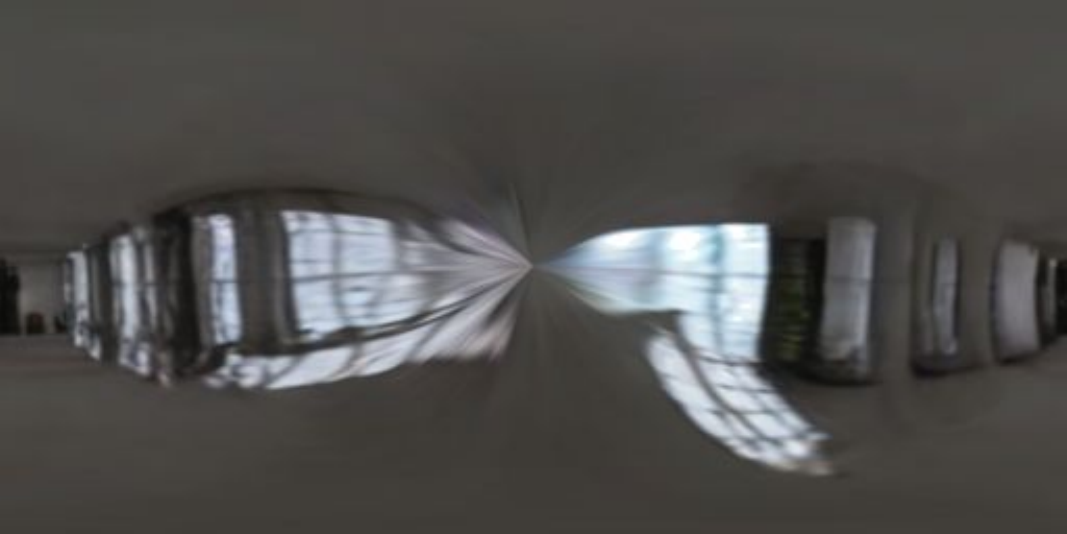}} 
        
        \\

        \noindent\parbox[c]{0.14\textwidth}{\includegraphics[height=0.100\textwidth]{storage/appendix_result/polyhaven_matte/combination_room_4k_input.pdf}} & 
        \noindent\parbox[c]{0.205\textwidth}{\includegraphics[height=0.100\textwidth]{storage/appendix_result/polyhaven_matte/combination_room_4k_envmapgt.pdf}} &  

        \noindent\parbox[c]{0.205\textwidth}{\includegraphics[height=0.100\textwidth]{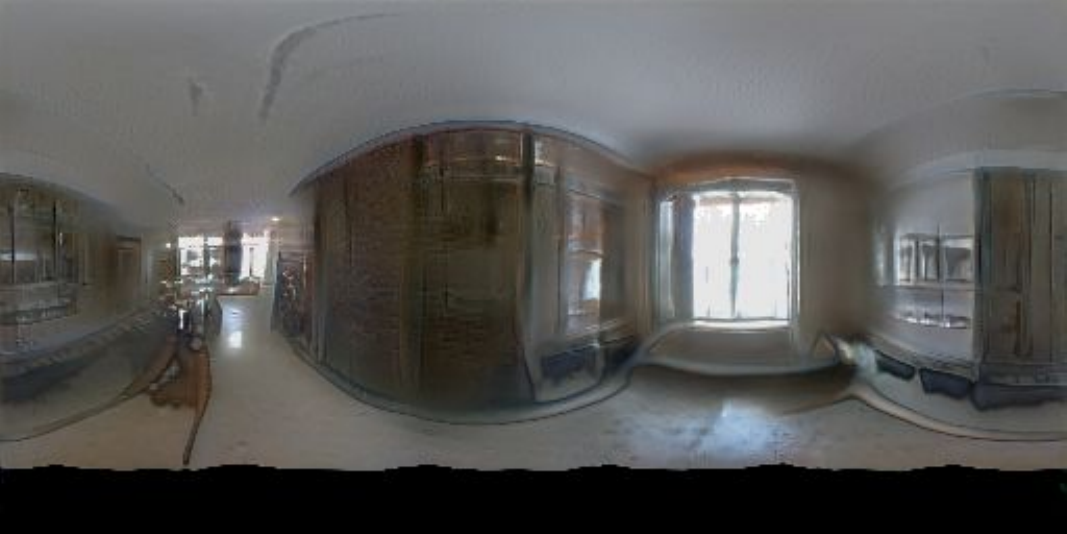}} &
        \noindent\parbox[c]{0.205\textwidth}{\includegraphics[height=0.100\textwidth]{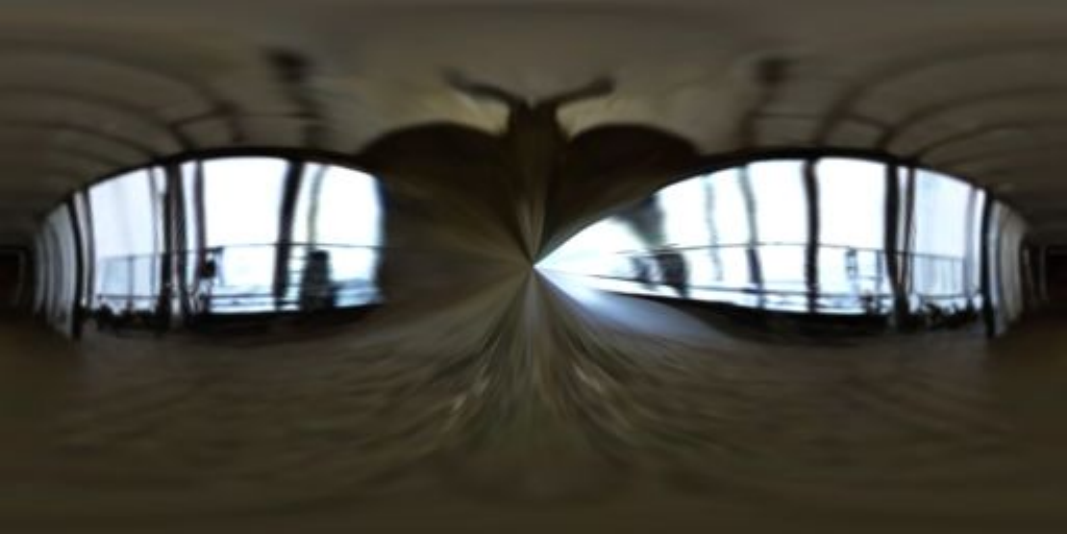}} &
        \noindent\parbox[c]{0.205\textwidth}{\includegraphics[height=0.100\textwidth]{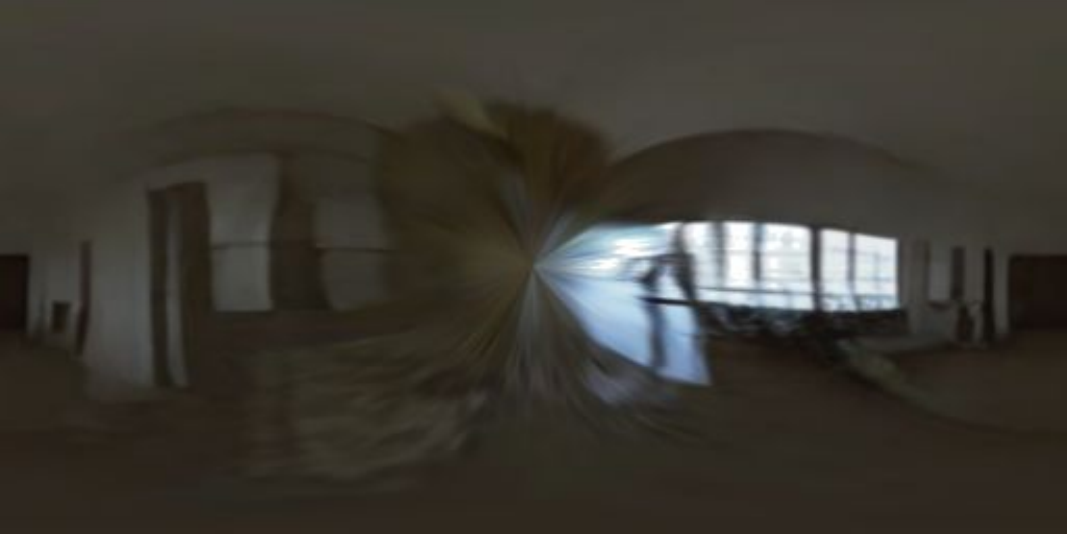}}
        
        \\

        \noindent\parbox[c]{0.14\textwidth}{\includegraphics[height=0.100\textwidth]{storage/appendix_result/polyhaven_matte/drachenfels_cellar_4k_input.pdf}} &
        \noindent\parbox[c]{0.205\textwidth}{\includegraphics[height=0.100\textwidth]{storage/appendix_result/polyhaven_matte/drachenfels_cellar_4k_envmapgt.pdf}} & 

        \noindent\parbox[c]{0.205\textwidth}{\includegraphics[height=0.100\textwidth]{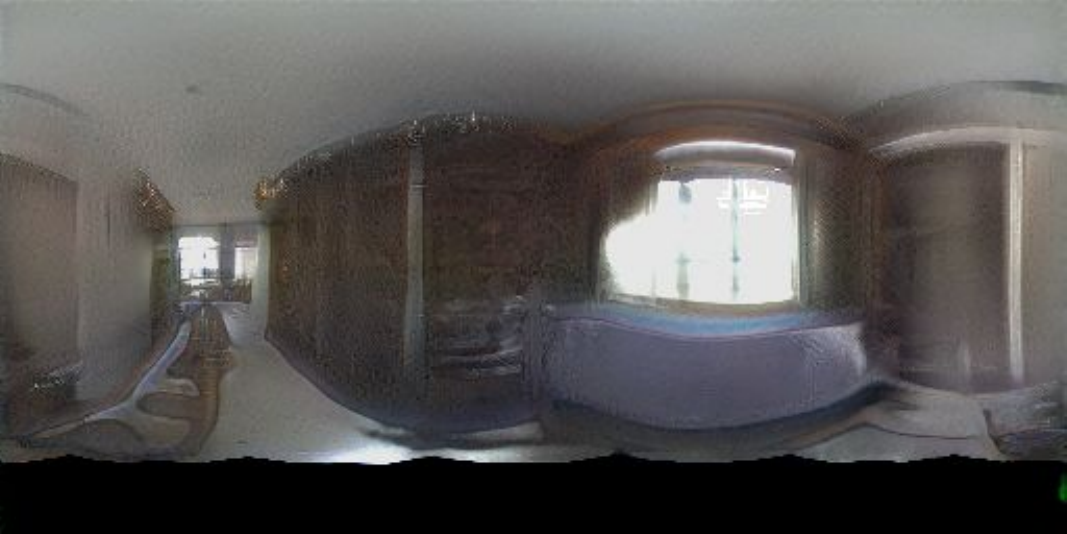}} &
        \noindent\parbox[c]{0.205\textwidth}{\includegraphics[height=0.100\textwidth]{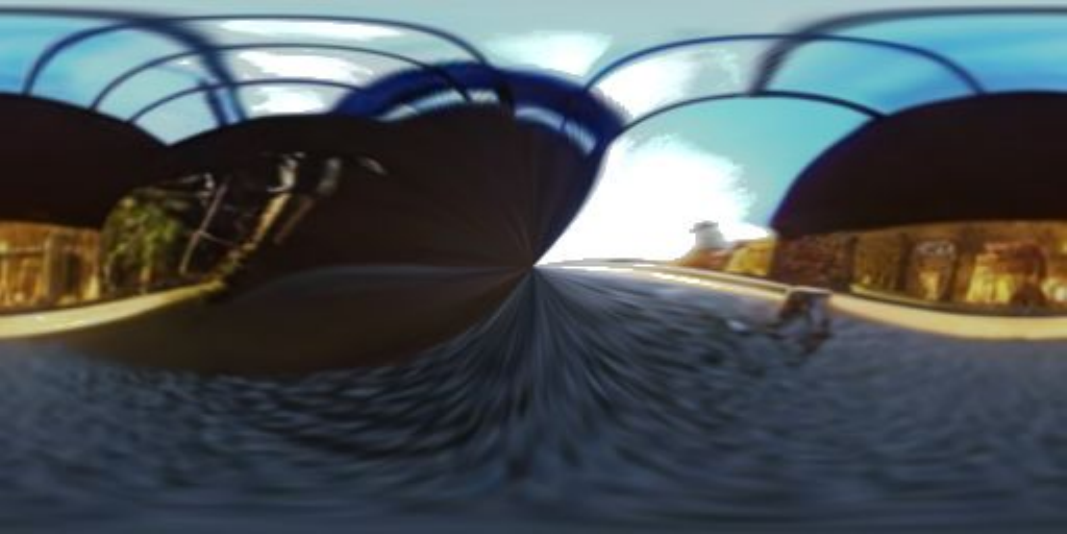}} &
        \noindent\parbox[c]{0.205\textwidth}{\includegraphics[height=0.100\textwidth]{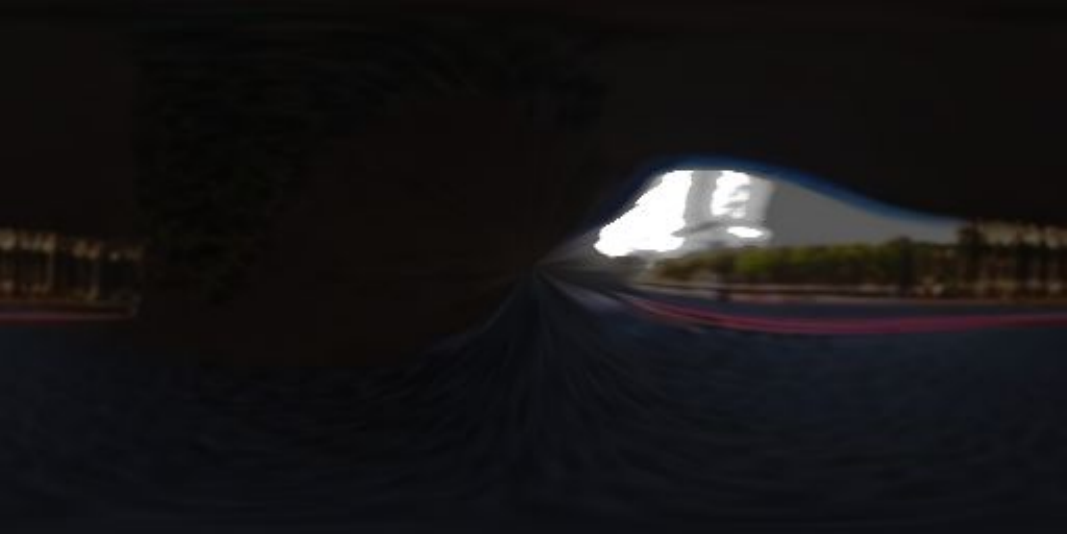}}
        
        \\

        \noindent\parbox[c]{0.14\textwidth}{\includegraphics[height=0.100\textwidth]{storage/appendix_result/polyhaven_matte/flower_hillside_4k_input.pdf}} &
        \noindent\parbox[c]{0.205\textwidth}{\includegraphics[height=0.100\textwidth]{storage/appendix_result/polyhaven_matte/flower_hillside_4k_envmapgt.pdf}} &  

        \noindent\parbox[c]{0.205\textwidth}{\includegraphics[height=0.100\textwidth]{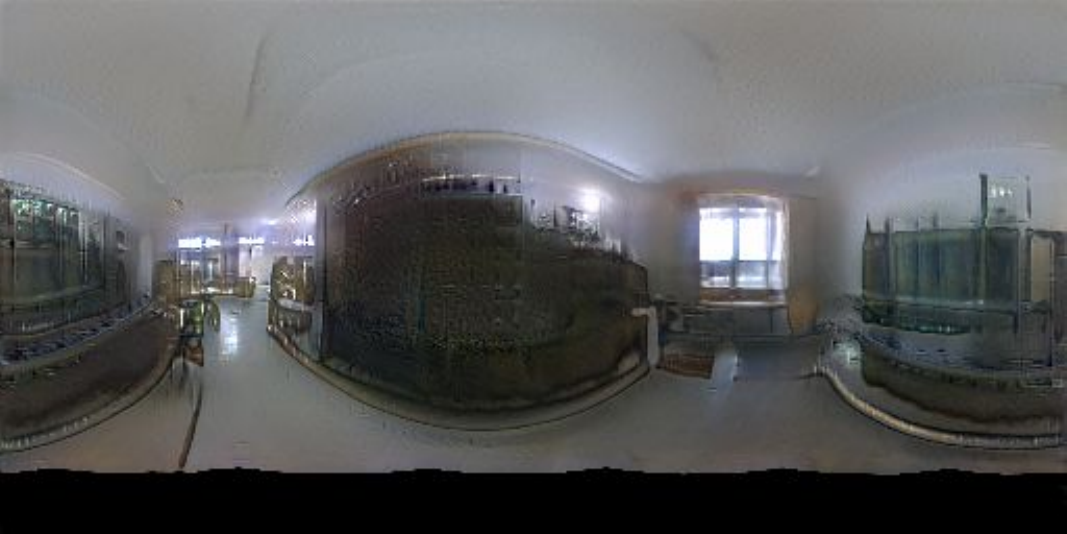}} &
        \noindent\parbox[c]{0.205\textwidth}{\includegraphics[height=0.100\textwidth]{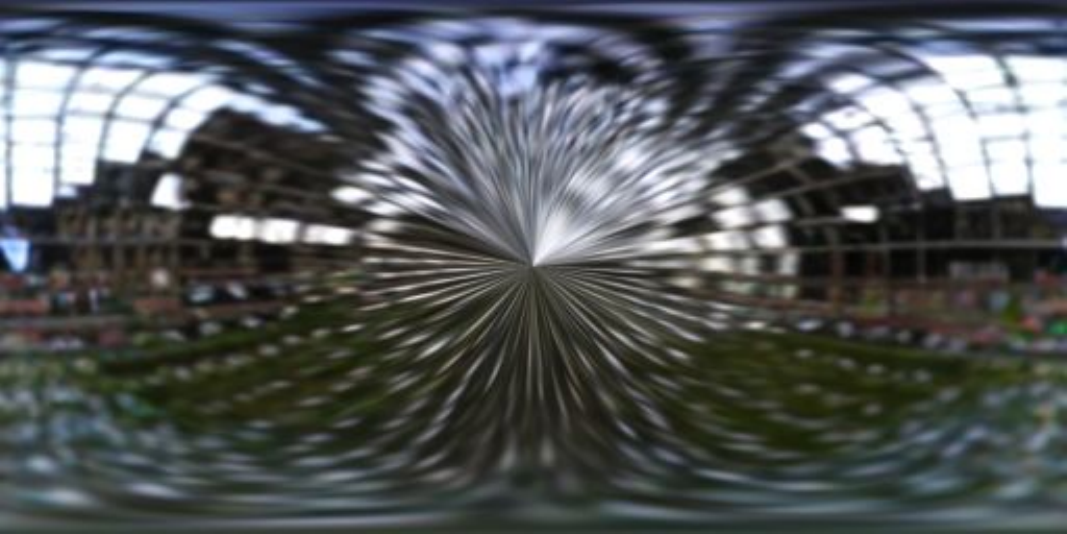}} &
        \noindent\parbox[c]{0.205\textwidth}{\includegraphics[height=0.100\textwidth]{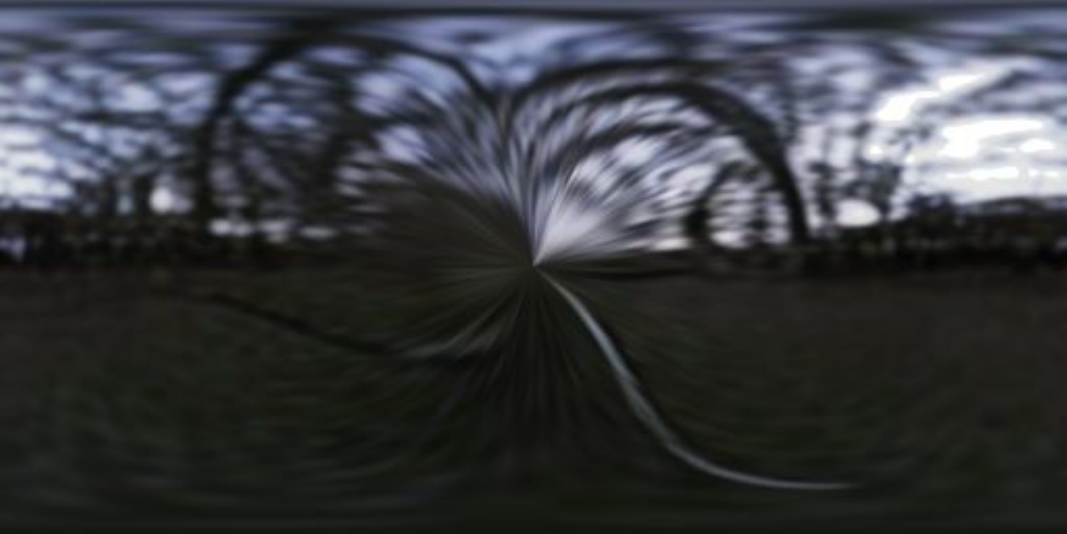}}
        
        \\

        \noindent\parbox[c]{0.14\textwidth}{\includegraphics[height=0.100\textwidth]{storage/appendix_result/polyhaven_matte/industrial_sunset_4k_input.pdf}} &
        \noindent\parbox[c]{0.205\textwidth}{\includegraphics[height=0.100\textwidth]{storage/appendix_result/polyhaven_matte/industrial_sunset_4k_envmapgt.pdf}} &   

        \noindent\parbox[c]{0.205\textwidth}{\includegraphics[height=0.100\textwidth]{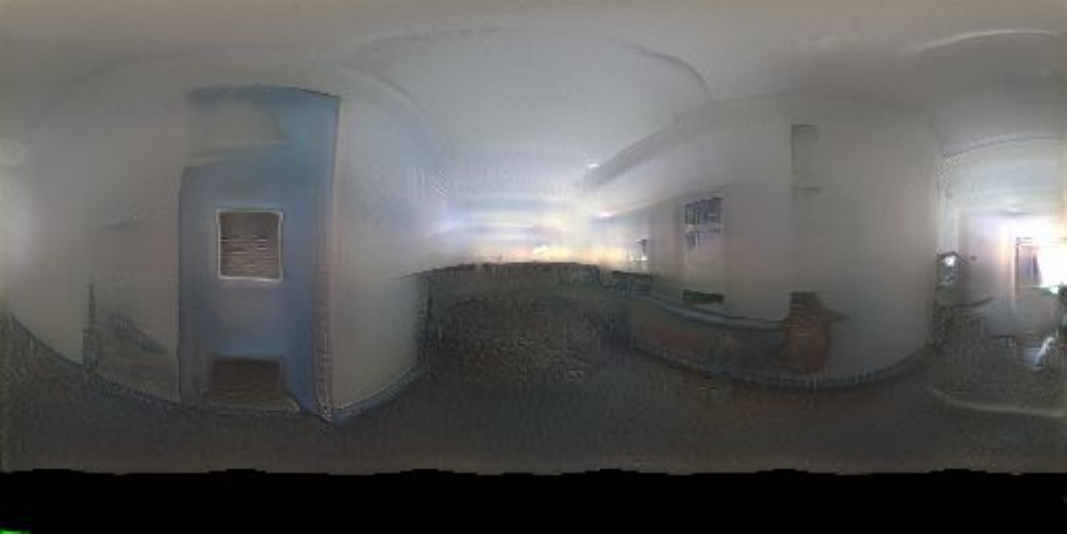}} &
        \noindent\parbox[c]{0.205\textwidth}{\includegraphics[height=0.100\textwidth]{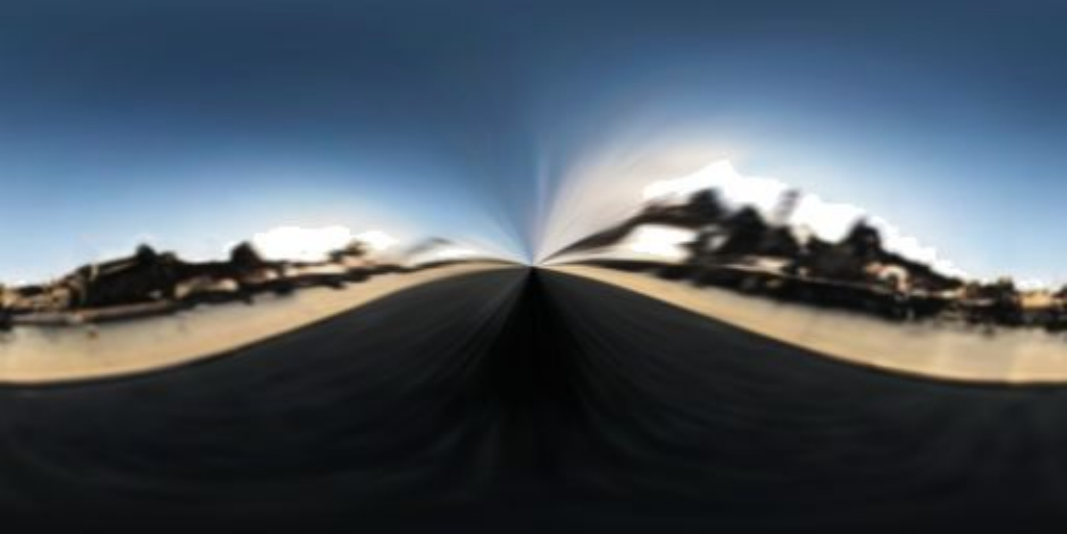}} &
        \noindent\parbox[c]{0.205\textwidth}{\includegraphics[height=0.100\textwidth]{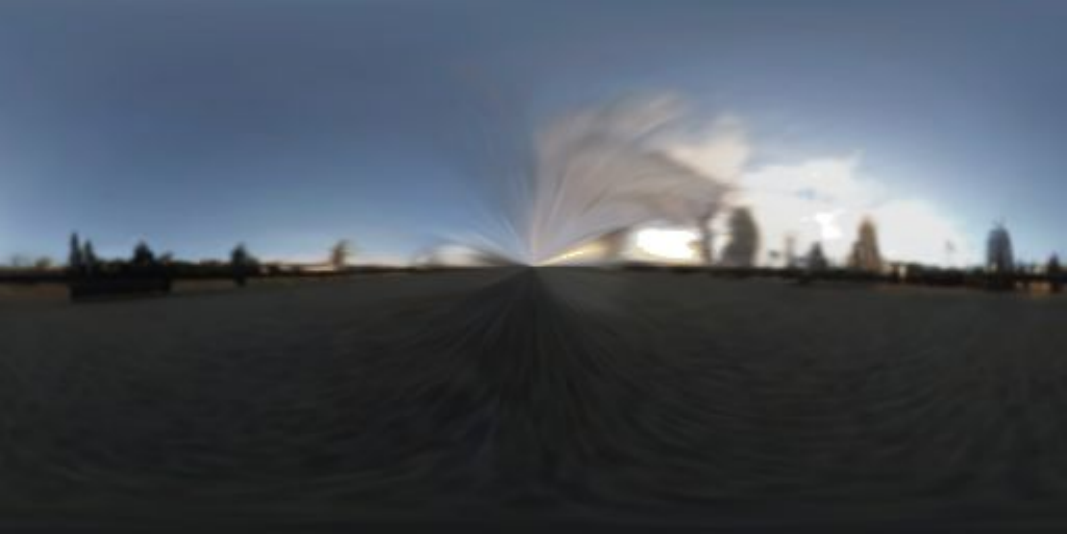}}
        \\

        \noindent\parbox[c]{0.14\textwidth}{\includegraphics[height=0.100\textwidth]{storage/appendix_result/polyhaven_matte/sandsloot_4k_input.pdf}} &  
        \noindent\parbox[c]{0.205\textwidth}{\includegraphics[height=0.100\textwidth]{storage/appendix_result/polyhaven_matte/sandsloot_4k_envmapgt.pdf}} & 

        \noindent\parbox[c]{0.205\textwidth}{\includegraphics[height=0.100\textwidth]{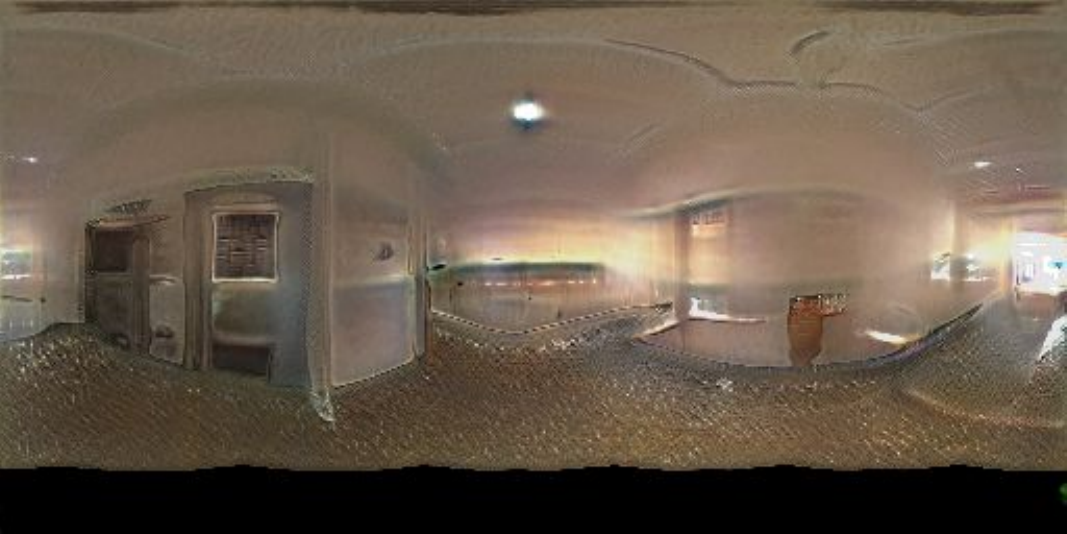}} &
        \noindent\parbox[c]{0.205\textwidth}{\includegraphics[height=0.100\textwidth]{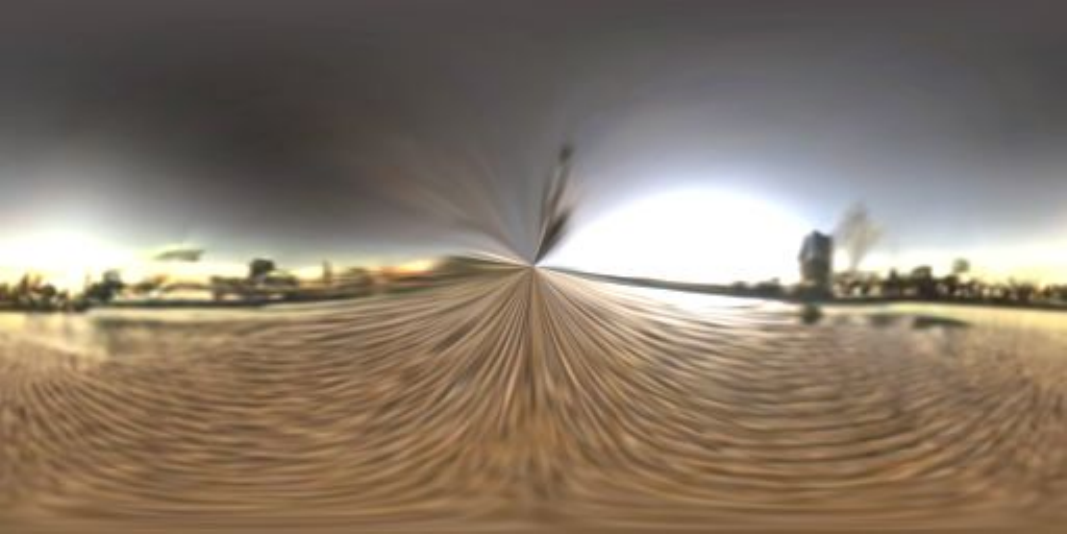}} &
        \noindent\parbox[c]{0.205\textwidth}{\includegraphics[height=0.100\textwidth]{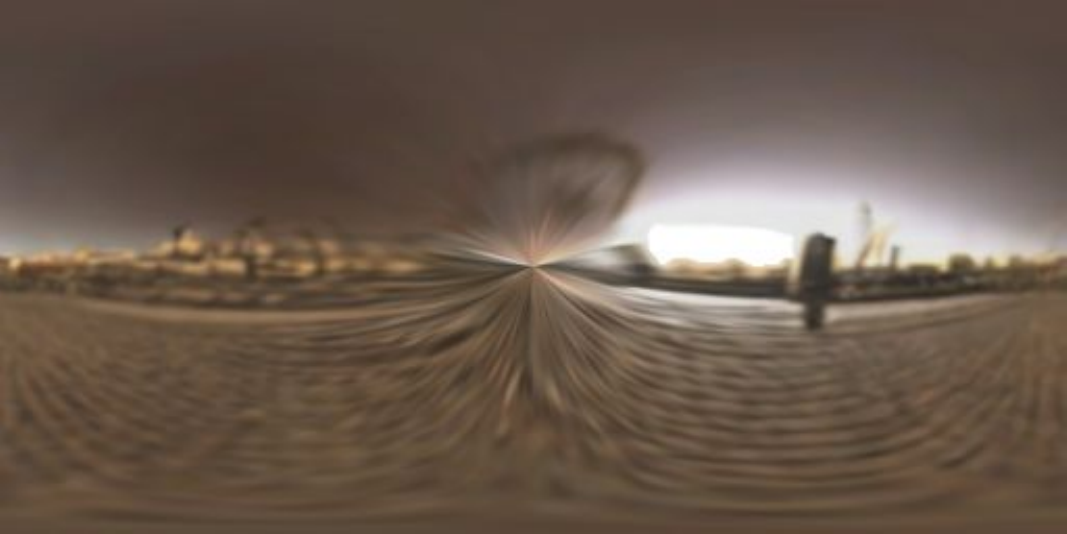}}
        \\

        \noindent\parbox[c]{0.14\textwidth}{\includegraphics[height=0.100\textwidth]{storage/appendix_result/polyhaven_matte/signal_hill_sunrise_4k_input.pdf}} & 
        \noindent\parbox[c]{0.205\textwidth}{\includegraphics[height=0.100\textwidth]{storage/appendix_result/polyhaven_matte/signal_hill_sunrise_4k_envmapgt.pdf}} &  

        \noindent\parbox[c]{0.205\textwidth}{\includegraphics[height=0.100\textwidth]{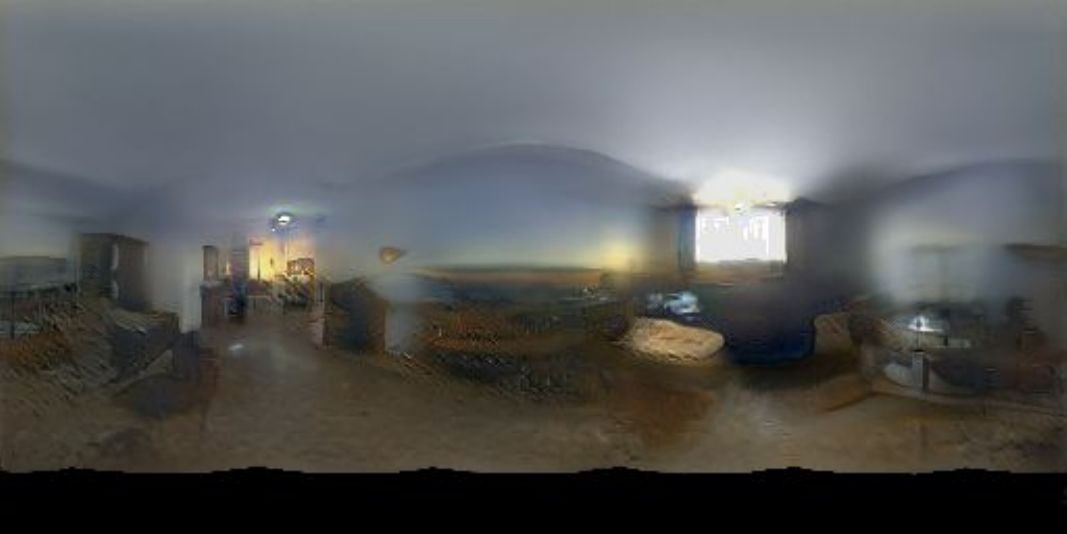}} &
        \noindent\parbox[c]{0.205\textwidth}{\includegraphics[height=0.100\textwidth]{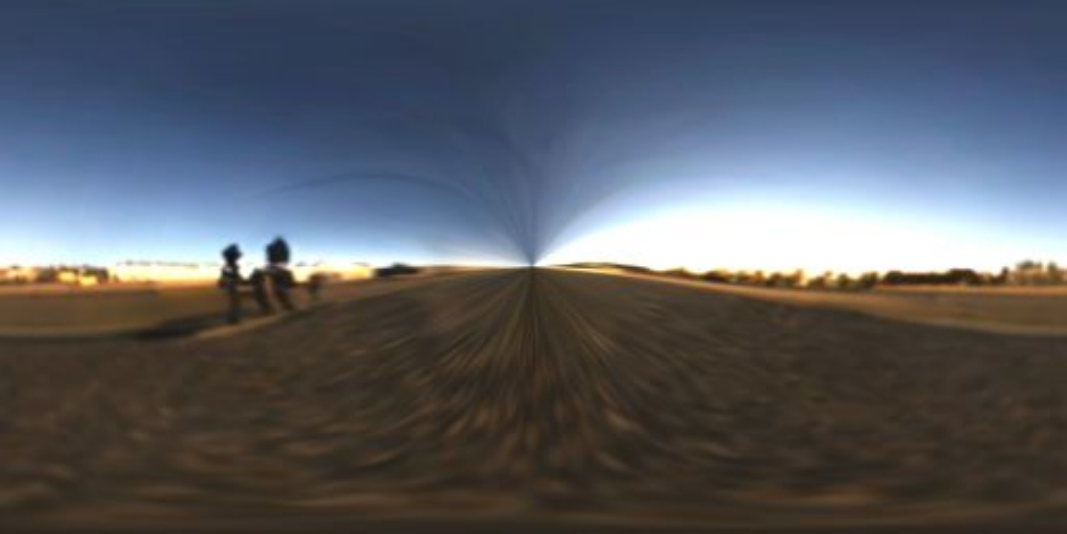}} &
        \noindent\parbox[c]{0.205\textwidth}{\includegraphics[height=0.100\textwidth]{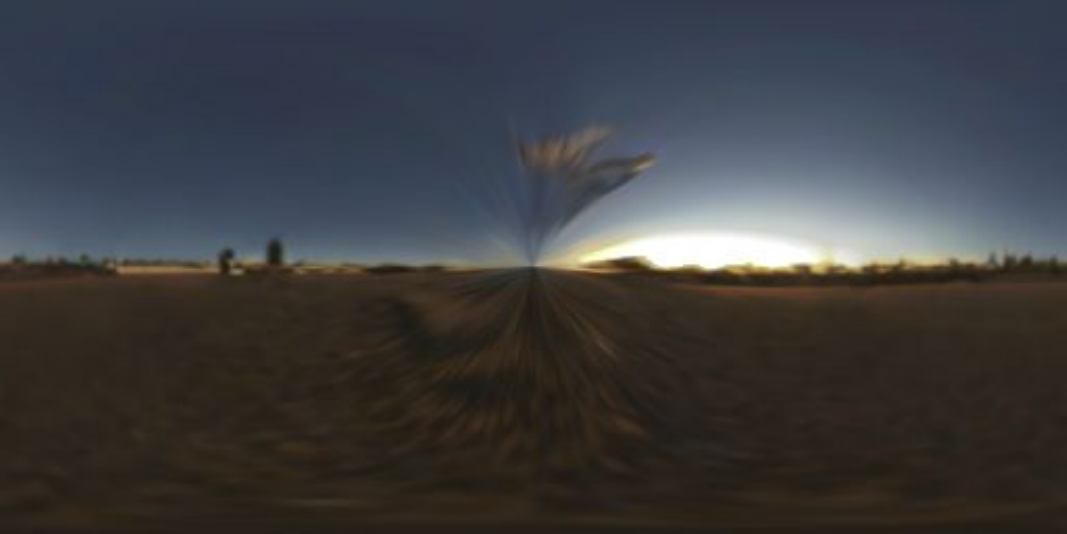}}
        \\

        \noindent\parbox[c]{0.14\textwidth}{\includegraphics[height=0.100\textwidth]{storage/appendix_result/polyhaven_matte/skukuza_golf_4k_input.pdf}} &
        \noindent\parbox[c]{0.205\textwidth}{\includegraphics[height=0.100\textwidth]{storage/appendix_result/polyhaven_matte/skukuza_golf_4k_envmapgt.pdf}} &    

        \noindent\parbox[c]{0.205\textwidth}{\includegraphics[height=0.100\textwidth]{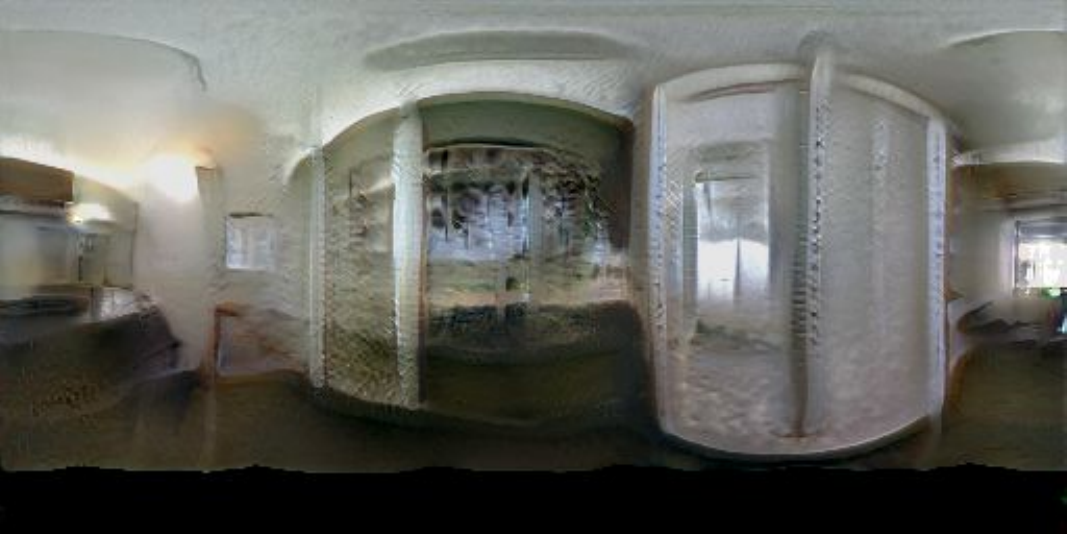}} &
        \noindent\parbox[c]{0.205\textwidth}{\includegraphics[height=0.100\textwidth]{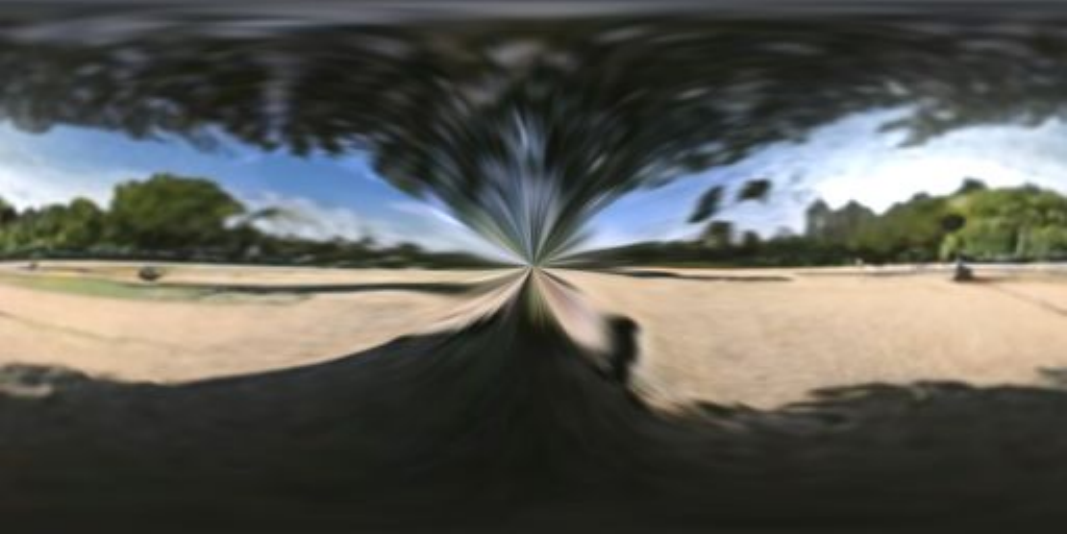}} &
        \noindent\parbox[c]{0.205\textwidth}{\includegraphics[height=0.100\textwidth]{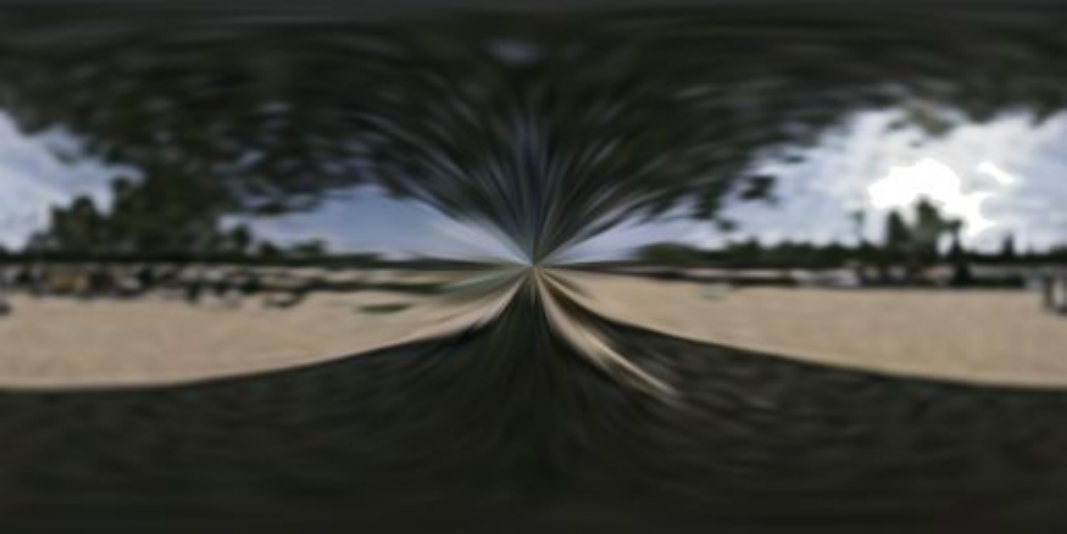}}
        \\

        \noindent\parbox[c]{0.14\textwidth}{\includegraphics[height=0.100\textwidth]{storage/appendix_result/polyhaven_matte/soliltude_4k_input.pdf}} &
        \noindent\parbox[c]{0.205\textwidth}{\includegraphics[height=0.100\textwidth]{storage/appendix_result/polyhaven_matte/soliltude_4k_envmapgt.pdf}} & 

        \noindent\parbox[c]{0.205\textwidth}{\includegraphics[height=0.100\textwidth]{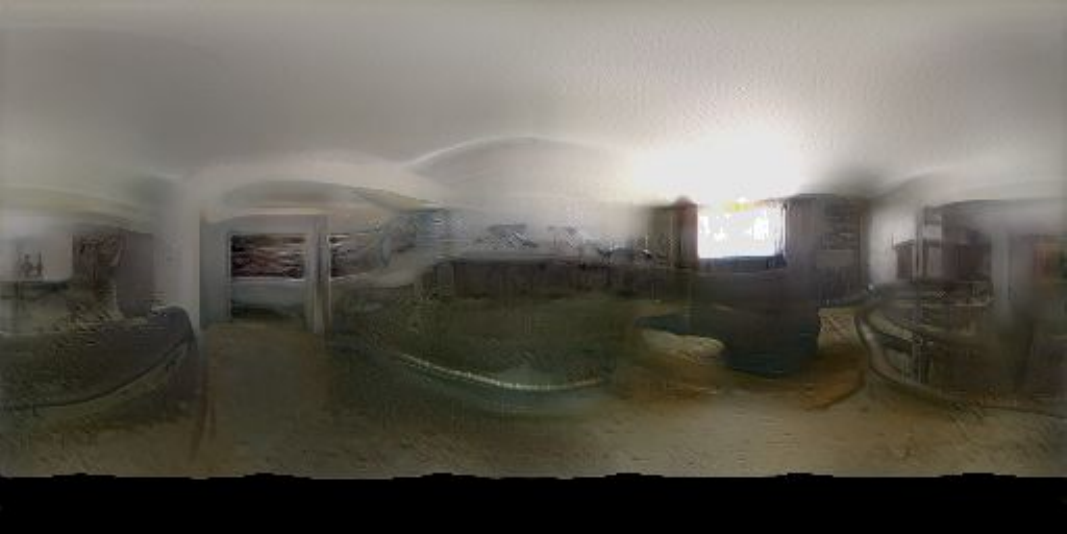}} &
        \noindent\parbox[c]{0.205\textwidth}{\includegraphics[height=0.100\textwidth]{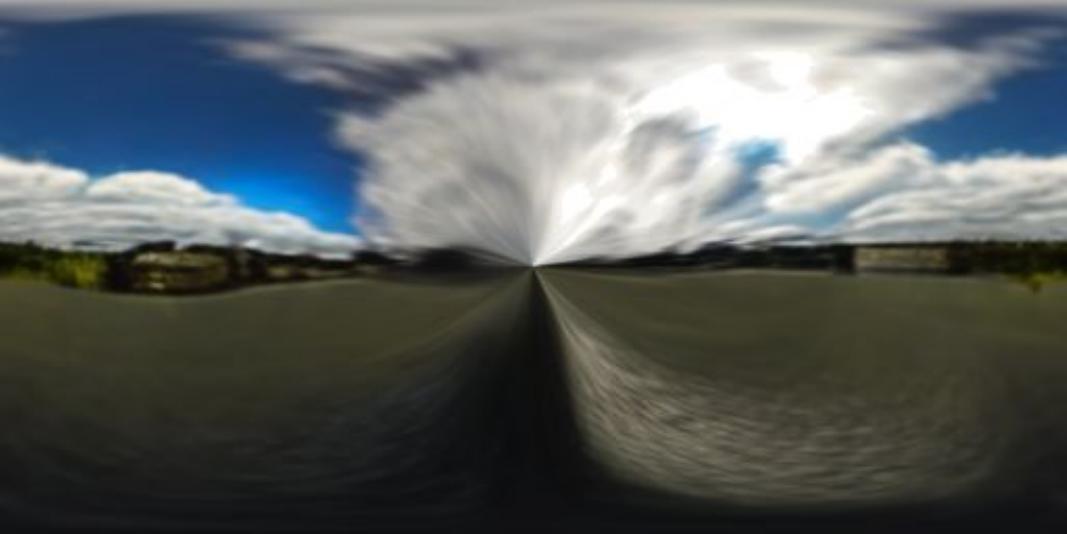}} &
        \noindent\parbox[c]{0.205\textwidth}{\includegraphics[height=0.100\textwidth]{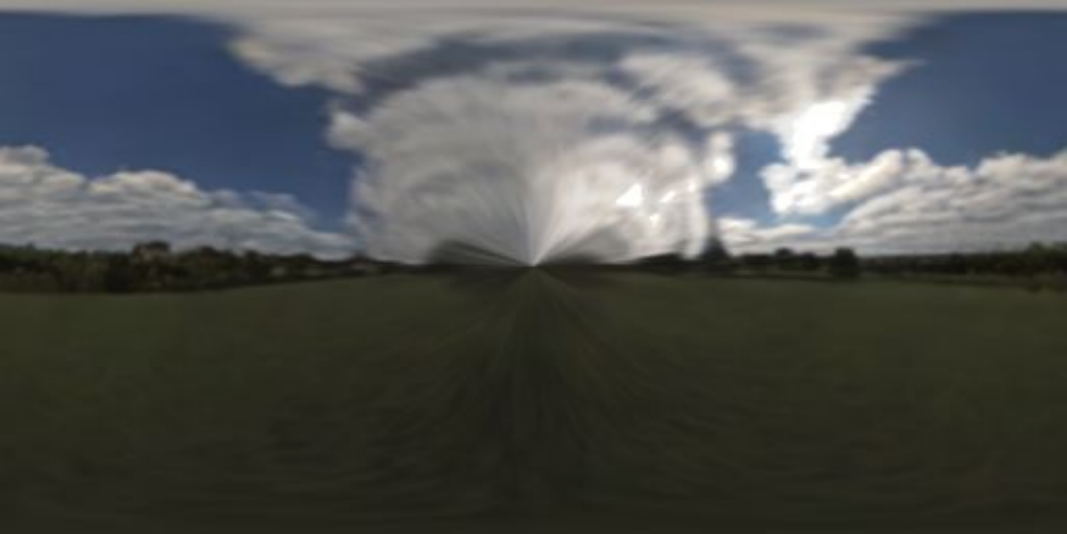}}
        
        \\

        \noindent\parbox[c]{0.14\textwidth}{\includegraphics[height=0.100\textwidth]{storage/appendix_result/polyhaven_matte/theater_01_4k_input.pdf}} &  
        \noindent\parbox[c]{0.205\textwidth}{\includegraphics[height=0.100\textwidth]{storage/appendix_result/polyhaven_matte/theater_01_4k_envmapgt.pdf}} & 

        \noindent\parbox[c]{0.205\textwidth}{\includegraphics[height=0.100\textwidth]{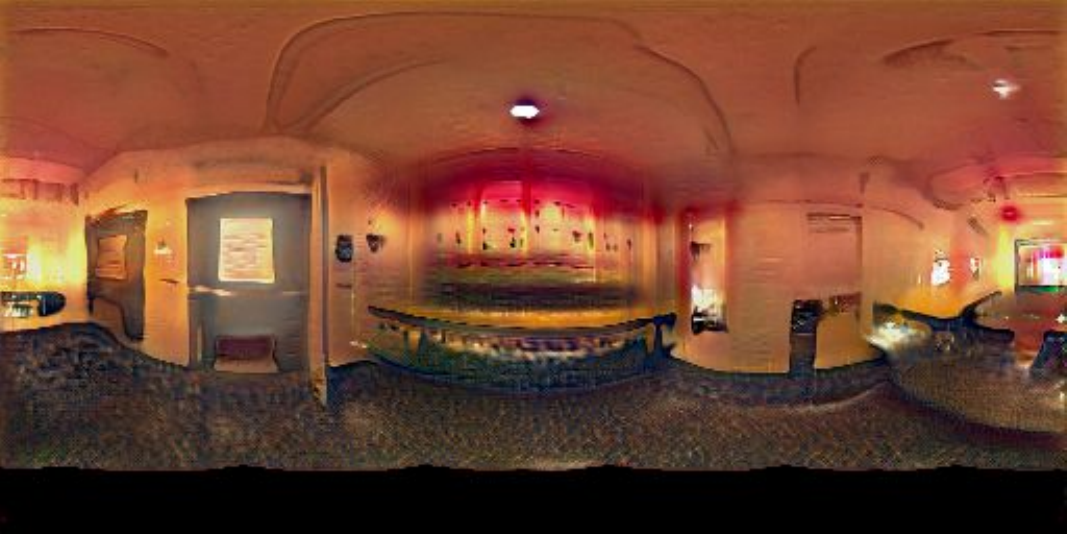}} &
        \noindent\parbox[c]{0.205\textwidth}{\includegraphics[height=0.100\textwidth]{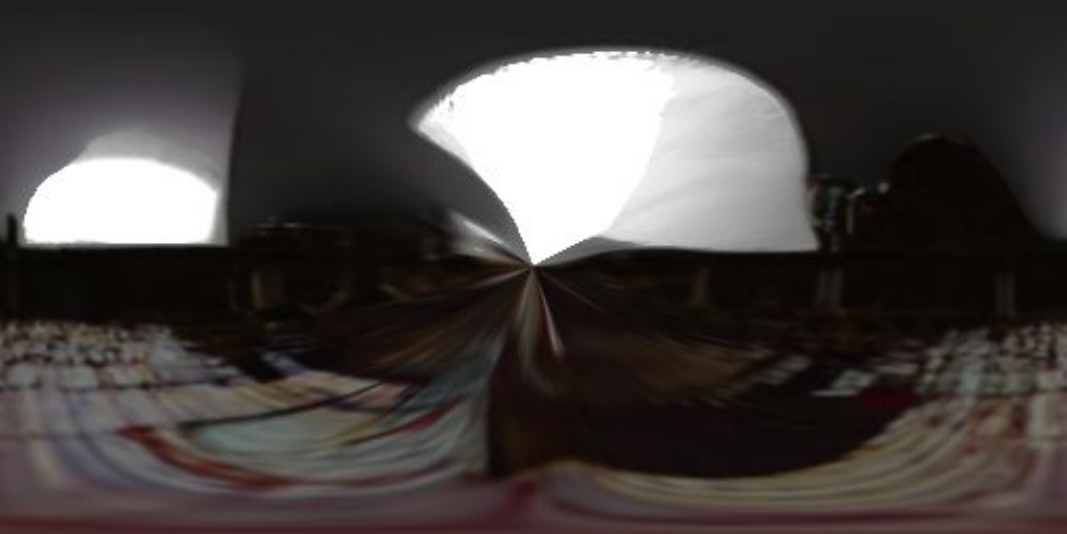}} &
        \noindent\parbox[c]{0.205\textwidth}{\includegraphics[height=0.100\textwidth]{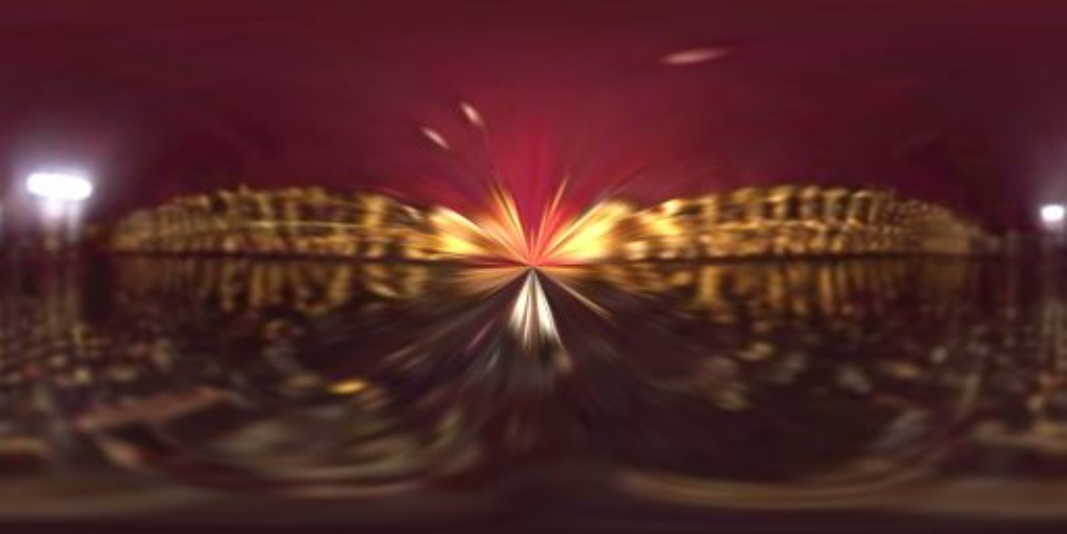}}
        
        \\
        
        \end{tabu}
    \caption{
    Unwrapped equirectangular maps for the Poly Haven dataset.}
    \label{fig:envmap_polyhaven}
\end{figure*}

\tabulinesep=0.5pt


\end{document}